\documentclass{article}

\newif\ifarxiv
\arxivtrue
\usepackage{iclr2027_conference,times}

\usepackage{amsmath,amsfonts,bm}

\def\eqref#1{equation~\ref{#1}}
\def\1{\bm{1}}

\DeclareMathAlphabet{\mathsfit}{\encodingdefault}{\sfdefault}{m}{sl}
\SetMathAlphabet{\mathsfit}{bold}{\encodingdefault}{\sfdefault}{bx}{n}

\usepackage{amssymb}       % extra math symbols
\usepackage{pifont}        % consistent check/cross glyphs in comparison tables
\usepackage[nointegrals]{wasysym} % font-based half-filled circle; keep AMS integrals
\usepackage{graphicx}
\usepackage{subcaption}    % per-panel (a)/(b)/(c) captions inside a multi-panel float
\usepackage{float}         % precise placement support for auxiliary figures
\usepackage{needspace}    % reserve the full wrapped figure before a page break
\usepackage{placeins}      % keep task examples within their source section
\usepackage{xcolor}
\usepackage{colortbl}      % subtle row/cell shading in publication tables
\usepackage{booktabs}      % \toprule/\midrule/\bottomrule for clean tables
\usepackage{multirow}
\usepackage{array}
\usepackage{makecell}
\usepackage{tabularx}
\usepackage{siunitx}       % decimal-aligned numeric columns in Table 4
\usepackage{enumitem}      % list layout options (leftmargin/itemsep) used in Contributions
\usepackage{listings}      % code/JSON listings for the annotation schema example
\usepackage[most]{tcolorbox} % breakable shaded boxes for appendix prompts/taxonomy
\usepackage{tikz}          % concept + construction schematic figures
\usetikzlibrary{arrows.meta,positioning,shapes.geometric,fit,backgrounds}
\usetikzlibrary{shapes.callouts}  % retained for editable legacy figure sources
\usepackage{amsmath}
\usepackage{hyperref}
\usepackage{url}
\usepackage{titletoc}      % partial table of contents for the standalone appendix index
\usepackage[colorinlistoftodos,textsize=footnotesize]{todonotes}
\newtcolorbox{taxonomybox}[1]{%
  enhanced,
  breakable,
  colback=gray!4,
  colframe=gray!40,
  boxrule=0.5pt,
  arc=2pt,
  left=6pt,
  right=6pt,
  top=5pt,
  bottom=5pt,
  title={#1},
  title after break={#1\ (continued)},
  fonttitle=\bfseries,
  coltitle=black,
  colbacktitle=gray!12,
  before skip=8pt,
  after skip=8pt
}

\definecolor{oduPromptBg}{RGB}{249,250,251}     % fill, barely off white
\definecolor{oduPromptRule}{RGB}{215,219,224}   % hairlines and dividers
\definecolor{oduPromptLabel}{RGB}{80,96,115}    % metadata labels, L2 headings
\definecolor{oduPromptMuted}{RGB}{132,140,150}  % omission marks, list bullets
\definecolor{oduPromptCode}{RGB}{58,64,72}      % inline variables and keys

\newcommand{\oduPromptHairline}{%
  \par\nointerlineskip
  {\color{oduPromptRule}\hrule height 0.4pt}%
  \par
}

\newcommand{\oduPromptTitle}[1]{%
  \begingroup
  \setlength{\parindent}{0pt}%
  \small\bfseries #1\par
  \endgroup
  \addvspace{2pt}%
  \oduPromptHairline
  \addvspace{4pt}%
}

\newcommand{\oduPromptLineSpread}{1.04}
\newcommand{\oduPromptParskip}{3pt}
\newtcolorbox{promptbox}[1]{%
  enhanced, breakable,
  title after break={#1\ (continued)},
  fonttitle=\small\bfseries,
  coltitle=oduPromptCode, colbacktitle=oduPromptBg,
  titlerule=0pt,
  colback=oduPromptBg,
  frame hidden,
  boxrule=0pt,
  arc=0pt, outer arc=0pt,
  borderline horizontal={0.5pt}{0pt}{oduPromptRule},
  borderline vertical={0.4pt}{0pt}{oduPromptRule!45},
  left=3.5mm, right=3.5mm, top=2.5mm, bottom=2.5mm,
  before skip=9pt, after skip=9pt,
  fontupper=\footnotesize,
  before upper={%
    \setlength{\parindent}{0pt}%
    \setlength{\parskip}{\oduPromptParskip}%
    \linespread{\oduPromptLineSpread}\selectfont
    \oduPromptTitle{#1}%
  },
}

\newcommand{\oduPromptLabelFont}[1]{%
  {\color{oduPromptLabel}\sffamily\scriptsize\bfseries\MakeUppercase{#1}}%
}
\newlist{promptmeta}{description}{1}
\setlist[promptmeta]{
  style=sameline, align=left,
  labelwidth=5.0em, leftmargin=5.4em, labelsep=0.4em,
  font=\oduPromptLabelFont,
  itemsep=2pt, topsep=2pt, parsep=0pt, partopsep=0pt,
  before=\vspace{1pt}, after=\vspace{2pt}\oduPromptHairline\vspace{5pt},
}
\newcommand{\promptrow}[2]{\item[#1] #2}

\newcommand{\PromptSection}[1]{%
  \par\addvspace{8pt}%
  {\small\bfseries #1}%
  \par\nobreak\addvspace{3pt}\nobreak%
}

\newcommand{\PromptKey}[1]{{\color{oduPromptCode}\ttfamily\footnotesize #1}}

\newlist{promptenum}{enumerate}{2}
\setlist[promptenum]{label=\arabic*., leftmargin=1.85em, labelsep=0.5em,
                     itemsep=2.5pt, topsep=3.5pt, parsep=0pt, partopsep=0pt}
\setlist[promptenum,2]{label=\alph*., leftmargin=1.5em, itemsep=2pt, topsep=2.5pt}
\newlist{promptitem}{itemize}{3}
\setlist[promptitem]{label=\textcolor{oduPromptMuted}{\raisebox{0.15ex}{\scriptsize$\bullet$}},
                     leftmargin=1.4em, labelsep=0.45em,
                     itemsep=2pt, topsep=3pt, parsep=0pt, partopsep=0pt}
\setlist[promptitem,2]{label=\textcolor{oduPromptMuted}{\textendash},
                       leftmargin=1.25em, itemsep=1.5pt, topsep=2pt}
\setlist[promptitem,3]{label=\textcolor{oduPromptMuted}{$\cdot$},
                       leftmargin=1.1em, itemsep=1.5pt, topsep=2pt}

\lstdefinestyle{oduprompt}{%
  basicstyle=\ttfamily\footnotesize\color{oduPromptCode},
  breaklines=true,
  breakatwhitespace=true,
  columns=fullflexible,
  keepspaces=true,
  showstringspaces=false,
  frame=none,
  xleftmargin=0.8em,
  aboveskip=5pt,
  belowskip=3pt,
}

\newcommand{\odunumMissing}[2]{\textcolor{red}{[\textbf{MISSING #1:} \texttt{\detokenize{#2}}]}}
\renewcommand{\num}[1]{%
  \ifcsname odunum@val@#1\endcsname\csname odunum@val@#1\endcsname
  \else\odunumMissing{NUM}{#1}\fi}
\newcommand{\numpct}[1]{%
  \ifcsname odunum@val@#1\endcsname%
    \pgfmathparse{100*(\csname odunum@val@#1\endcsname)}%
    \pgfmathprintnumber[fixed,zerofill,precision=1]{\pgfmathresult}%
  \else\odunumMissing{NUM}{#1}\fi}
\newcommand{\numpctd}[1]{%
  \ifcsname odunum@val@#1\endcsname%
    \pgfmathparse{100*(\csname odunum@val@#1\endcsname)}%
    \pgfmathprintnumber[fixed,zerofill,precision=1,print sign]{\pgfmathresult}%
  \else\odunumMissing{NUM}{#1}\fi}
\newcommand{\numn}[1]{%
  \ifcsname odunum@n@#1\endcsname\csname odunum@n@#1\endcsname
  \else\odunumMissing{N}{#1}\fi}
\newcommand{\numci}[1]{%
  \ifcsname odunum@ci@#1\endcsname\csname odunum@ci@#1\endcsname
  \else\odunumMissing{CI}{#1}\fi}
\newcommand{\numlit}[1]{#1}
\IfFileExists{numbers.tex}{% =========================================================================
\expandafter\gdef\csname odunum@val@axis_breakdown.axis_code_order_agreement.axis1.audio_visual\endcsname{0.08\ensuremath{\times}}
\expandafter\gdef\csname odunum@n@axis_breakdown.axis_code_order_agreement.axis1.audio_visual\endcsname{12}
\expandafter\gdef\csname odunum@val@axis_breakdown.axis_code_order_agreement.axis2\endcsname{0.57\ensuremath{\times}}
\expandafter\gdef\csname odunum@n@axis_breakdown.axis_code_order_agreement.axis2\endcsname{12}
\expandafter\gdef\csname odunum@val@axis_breakdown.axis_code_order_agreement.axis3\endcsname{0.70\ensuremath{\times}}
\expandafter\gdef\csname odunum@n@axis_breakdown.axis_code_order_agreement.axis3\endcsname{12}
\expandafter\gdef\csname odunum@val@axis_breakdown.axis_code_order_agreement.axis4\endcsname{0.38\ensuremath{\times}}
\expandafter\gdef\csname odunum@n@axis_breakdown.axis_code_order_agreement.axis4\endcsname{12}
\expandafter\gdef\csname odunum@val@axis_breakdown.axis_code_order_agreement.axis5.audio_only\endcsname{0.49\ensuremath{\times}}
\expandafter\gdef\csname odunum@n@axis_breakdown.axis_code_order_agreement.axis5.audio_only\endcsname{12}
\expandafter\gdef\csname odunum@val@axis_breakdown.axis_code_order_agreement.axis5.audio_visual\endcsname{0.42\ensuremath{\times}}
\expandafter\gdef\csname odunum@n@axis_breakdown.axis_code_order_agreement.axis5.audio_visual\endcsname{12}
\expandafter\gdef\csname odunum@val@axis_breakdown.axis_code_order_agreement.axis6\endcsname{0.45\ensuremath{\times}}
\expandafter\gdef\csname odunum@n@axis_breakdown.axis_code_order_agreement.axis6\endcsname{12}
\expandafter\gdef\csname odunum@val@axis_breakdown.axis_hardest.axis1.audio_only\endcsname{2}
\expandafter\gdef\csname odunum@n@axis_breakdown.axis_hardest.axis1.audio_only\endcsname{2}
\expandafter\gdef\csname odunum@val@axis_breakdown.axis_hardest.axis1.audio_visual\endcsname{4}
\expandafter\gdef\csname odunum@n@axis_breakdown.axis_hardest.axis1.audio_visual\endcsname{4}
\expandafter\gdef\csname odunum@val@axis_breakdown.axis_hardest.axis2\endcsname{4}
\expandafter\gdef\csname odunum@n@axis_breakdown.axis_hardest.axis2\endcsname{5}
\expandafter\gdef\csname odunum@val@axis_breakdown.axis_hardest.axis3\endcsname{3}
\expandafter\gdef\csname odunum@n@axis_breakdown.axis_hardest.axis3\endcsname{6}
\expandafter\gdef\csname odunum@val@axis_breakdown.axis_hardest.axis4\endcsname{3}
\expandafter\gdef\csname odunum@n@axis_breakdown.axis_hardest.axis4\endcsname{5}
\expandafter\gdef\csname odunum@val@axis_breakdown.axis_hardest.axis5.audio_only\endcsname{8}
\expandafter\gdef\csname odunum@n@axis_breakdown.axis_hardest.axis5.audio_only\endcsname{10}
\expandafter\gdef\csname odunum@val@axis_breakdown.axis_hardest.axis5.audio_visual\endcsname{2}
\expandafter\gdef\csname odunum@n@axis_breakdown.axis_hardest.axis5.audio_visual\endcsname{10}
\expandafter\gdef\csname odunum@val@axis_breakdown.axis_hardest.axis6\endcsname{8}
\expandafter\gdef\csname odunum@n@axis_breakdown.axis_hardest.axis6\endcsname{14}
\expandafter\gdef\csname odunum@val@axis_breakdown.axis_spread.axis1.audio_only\endcsname{0.051}
\expandafter\gdef\csname odunum@n@axis_breakdown.axis_spread.axis1.audio_only\endcsname{2}
\expandafter\gdef\csname odunum@val@axis_breakdown.axis_spread.axis1.audio_visual\endcsname{0.022}
\expandafter\gdef\csname odunum@n@axis_breakdown.axis_spread.axis1.audio_visual\endcsname{4}
\expandafter\gdef\csname odunum@val@axis_breakdown.axis_spread.axis2\endcsname{0.072}
\expandafter\gdef\csname odunum@n@axis_breakdown.axis_spread.axis2\endcsname{5}
\expandafter\gdef\csname odunum@val@axis_breakdown.axis_spread.axis3\endcsname{0.123}
\expandafter\gdef\csname odunum@n@axis_breakdown.axis_spread.axis3\endcsname{6}
\expandafter\gdef\csname odunum@val@axis_breakdown.axis_spread.axis4\endcsname{0.043}
\expandafter\gdef\csname odunum@n@axis_breakdown.axis_spread.axis4\endcsname{5}
\expandafter\gdef\csname odunum@val@axis_breakdown.axis_spread.axis5.audio_only\endcsname{0.115}
\expandafter\gdef\csname odunum@n@axis_breakdown.axis_spread.axis5.audio_only\endcsname{10}
\expandafter\gdef\csname odunum@val@axis_breakdown.axis_spread.axis5.audio_visual\endcsname{0.053}
\expandafter\gdef\csname odunum@n@axis_breakdown.axis_spread.axis5.audio_visual\endcsname{10}
\expandafter\gdef\csname odunum@val@axis_breakdown.axis_spread.axis6\endcsname{0.091}
\expandafter\gdef\csname odunum@n@axis_breakdown.axis_spread.axis6\endcsname{14}
\expandafter\gdef\csname odunum@val@axis_breakdown.difficulty.cut1\endcsname{0.421}
\expandafter\gdef\csname odunum@n@axis_breakdown.difficulty.cut1\endcsname{1\,429}
\expandafter\gdef\csname odunum@val@axis_breakdown.difficulty.cut2\endcsname{0.468}
\expandafter\gdef\csname odunum@n@axis_breakdown.difficulty.cut2\endcsname{1\,429}
\expandafter\gdef\csname odunum@val@axis_breakdown.difficulty.cut3\endcsname{0.522}
\expandafter\gdef\csname odunum@n@axis_breakdown.difficulty.cut3\endcsname{1\,429}
\expandafter\gdef\csname odunum@val@axis_breakdown.difficulty_corr.cascade_asr.spearman\endcsname{-0.127}
\expandafter\gdef\csname odunum@n@axis_breakdown.difficulty_corr.cascade_asr.spearman\endcsname{1\,429}
\expandafter\gdef\csname odunum@val@axis_breakdown.difficulty_corr.gemini.spearman\endcsname{-0.117}
\expandafter\gdef\csname odunum@n@axis_breakdown.difficulty_corr.gemini.spearman\endcsname{1\,429}
\expandafter\gdef\csname odunum@val@axis_breakdown.difficulty_corr.gemini35_flash_lite.spearman\endcsname{-0.085}
\expandafter\gdef\csname odunum@n@axis_breakdown.difficulty_corr.gemini35_flash_lite.spearman\endcsname{1\,429}
\expandafter\gdef\csname odunum@val@axis_breakdown.difficulty_corr.gemini37_flash.spearman\endcsname{-0.158}
\expandafter\gdef\csname odunum@n@axis_breakdown.difficulty_corr.gemini37_flash.spearman\endcsname{1\,426}
\expandafter\gdef\csname odunum@val@axis_breakdown.difficulty_corr.ming.spearman\endcsname{-0.096}
\expandafter\gdef\csname odunum@n@axis_breakdown.difficulty_corr.ming.spearman\endcsname{1\,429}
\expandafter\gdef\csname odunum@val@axis_breakdown.difficulty_corr.minicpm_o.spearman\endcsname{-0.189}
\expandafter\gdef\csname odunum@n@axis_breakdown.difficulty_corr.minicpm_o.spearman\endcsname{1\,427}
\expandafter\gdef\csname odunum@val@axis_breakdown.difficulty_corr.nemotron.spearman\endcsname{-0.175}
\expandafter\gdef\csname odunum@n@axis_breakdown.difficulty_corr.nemotron.spearman\endcsname{1\,429}
\expandafter\gdef\csname odunum@val@axis_breakdown.difficulty_corr.qwen25_omni.spearman\endcsname{-0.108}
\expandafter\gdef\csname odunum@n@axis_breakdown.difficulty_corr.qwen25_omni.spearman\endcsname{1\,429}
\expandafter\gdef\csname odunum@val@axis_breakdown.difficulty_corr.qwen3_omni_instruct.spearman\endcsname{-0.183}
\expandafter\gdef\csname odunum@n@axis_breakdown.difficulty_corr.qwen3_omni_instruct.spearman\endcsname{1\,429}
\expandafter\gdef\csname odunum@val@axis_breakdown.difficulty_corr.qwen3_omni_think.spearman\endcsname{-0.198}
\expandafter\gdef\csname odunum@n@axis_breakdown.difficulty_corr.qwen3_omni_think.spearman\endcsname{1\,429}
\expandafter\gdef\csname odunum@val@axis_breakdown.difficulty_corr.qwen_plus.spearman\endcsname{-0.102}
\expandafter\gdef\csname odunum@n@axis_breakdown.difficulty_corr.qwen_plus.spearman\endcsname{1\,429}
\expandafter\gdef\csname odunum@val@axis_breakdown.difficulty_corr.salmonn2_7b.spearman\endcsname{-0.286}
\expandafter\gdef\csname odunum@n@axis_breakdown.difficulty_corr.salmonn2_7b.spearman\endcsname{1\,415}
\expandafter\gdef\csname odunum@val@axis_breakdown.difficulty_corr.seed.spearman\endcsname{-0.181}
\expandafter\gdef\csname odunum@n@axis_breakdown.difficulty_corr.seed.spearman\endcsname{1\,429}
\expandafter\gdef\csname odunum@val@axis_breakdown.ftr.cascade_asr.axis4.4.1\endcsname{0.618}
\expandafter\gdef\csname odunum@n@axis_breakdown.ftr.cascade_asr.axis4.4.1\endcsname{110}
\expandafter\gdef\csname odunum@ci@axis_breakdown.ftr.cascade_asr.axis4.4.1\endcsname{[0.525, 0.704]}
\expandafter\gdef\csname odunum@val@axis_breakdown.ftr.cascade_asr.axis4.4.2\endcsname{0.589}
\expandafter\gdef\csname odunum@n@axis_breakdown.ftr.cascade_asr.axis4.4.2\endcsname{90}
\expandafter\gdef\csname odunum@ci@axis_breakdown.ftr.cascade_asr.axis4.4.2\endcsname{[0.486, 0.685]}
\expandafter\gdef\csname odunum@val@axis_breakdown.ftr.cascade_asr.axis4.4.3\endcsname{0.506}
\expandafter\gdef\csname odunum@n@axis_breakdown.ftr.cascade_asr.axis4.4.3\endcsname{77}
\expandafter\gdef\csname odunum@ci@axis_breakdown.ftr.cascade_asr.axis4.4.3\endcsname{[0.397, 0.615]}
\expandafter\gdef\csname odunum@val@axis_breakdown.ftr.cascade_asr.axis4.4.4\endcsname{0.634}
\expandafter\gdef\csname odunum@n@axis_breakdown.ftr.cascade_asr.axis4.4.4\endcsname{101}
\expandafter\gdef\csname odunum@ci@axis_breakdown.ftr.cascade_asr.axis4.4.4\endcsname{[0.536, 0.721]}
\expandafter\gdef\csname odunum@val@axis_breakdown.ftr.cascade_asr.axis4.4.5\endcsname{0.739}
\expandafter\gdef\csname odunum@n@axis_breakdown.ftr.cascade_asr.axis4.4.5\endcsname{69}
\expandafter\gdef\csname odunum@ci@axis_breakdown.ftr.cascade_asr.axis4.4.5\endcsname{[0.625, 0.828]}
\expandafter\gdef\csname odunum@val@axis_breakdown.ftr.cascade_asr.axis6.6.1\endcsname{0.525}
\expandafter\gdef\csname odunum@n@axis_breakdown.ftr.cascade_asr.axis6.6.1\endcsname{59}
\expandafter\gdef\csname odunum@ci@axis_breakdown.ftr.cascade_asr.axis6.6.1\endcsname{[0.400, 0.647]}
\expandafter\gdef\csname odunum@val@axis_breakdown.ftr.cascade_asr.axis6.6.12\endcsname{0.691}
\expandafter\gdef\csname odunum@n@axis_breakdown.ftr.cascade_asr.axis6.6.12\endcsname{42}
\expandafter\gdef\csname odunum@ci@axis_breakdown.ftr.cascade_asr.axis6.6.12\endcsname{[0.540, 0.809]}
\expandafter\gdef\csname odunum@val@axis_breakdown.ftr.cascade_asr.axis6.6.6\endcsname{0.667}
\expandafter\gdef\csname odunum@n@axis_breakdown.ftr.cascade_asr.axis6.6.6\endcsname{54}
\expandafter\gdef\csname odunum@ci@axis_breakdown.ftr.cascade_asr.axis6.6.6\endcsname{[0.534, 0.778]}
\expandafter\gdef\csname odunum@val@axis_breakdown.ftr.cascade_asr.axis6.6.7\endcsname{0.532}
\expandafter\gdef\csname odunum@n@axis_breakdown.ftr.cascade_asr.axis6.6.7\endcsname{47}
\expandafter\gdef\csname odunum@ci@axis_breakdown.ftr.cascade_asr.axis6.6.7\endcsname{[0.392, 0.667]}
\expandafter\gdef\csname odunum@val@axis_breakdown.ftr.gemini.axis4.4.1\endcsname{0.391}
\expandafter\gdef\csname odunum@n@axis_breakdown.ftr.gemini.axis4.4.1\endcsname{110}
\expandafter\gdef\csname odunum@ci@axis_breakdown.ftr.gemini.axis4.4.1\endcsname{[0.305, 0.484]}
\expandafter\gdef\csname odunum@val@axis_breakdown.ftr.gemini.axis4.4.2\endcsname{0.389}
\expandafter\gdef\csname odunum@n@axis_breakdown.ftr.gemini.axis4.4.2\endcsname{90}
\expandafter\gdef\csname odunum@ci@axis_breakdown.ftr.gemini.axis4.4.2\endcsname{[0.295, 0.492]}
\expandafter\gdef\csname odunum@val@axis_breakdown.ftr.gemini.axis4.4.3\endcsname{0.234}
\expandafter\gdef\csname odunum@n@axis_breakdown.ftr.gemini.axis4.4.3\endcsname{77}
\expandafter\gdef\csname odunum@ci@axis_breakdown.ftr.gemini.axis4.4.3\endcsname{[0.153, 0.340]}
\expandafter\gdef\csname odunum@val@axis_breakdown.ftr.gemini.axis4.4.4\endcsname{0.346}
\expandafter\gdef\csname odunum@n@axis_breakdown.ftr.gemini.axis4.4.4\endcsname{101}
\expandafter\gdef\csname odunum@ci@axis_breakdown.ftr.gemini.axis4.4.4\endcsname{[0.261, 0.443]}
\expandafter\gdef\csname odunum@val@axis_breakdown.ftr.gemini.axis4.4.5\endcsname{0.522}
\expandafter\gdef\csname odunum@n@axis_breakdown.ftr.gemini.axis4.4.5\endcsname{69}
\expandafter\gdef\csname odunum@ci@axis_breakdown.ftr.gemini.axis4.4.5\endcsname{[0.406, 0.635]}
\expandafter\gdef\csname odunum@val@axis_breakdown.ftr.gemini.axis6.6.1\endcsname{0.305}
\expandafter\gdef\csname odunum@n@axis_breakdown.ftr.gemini.axis6.6.1\endcsname{59}
\expandafter\gdef\csname odunum@ci@axis_breakdown.ftr.gemini.axis6.6.1\endcsname{[0.203, 0.431]}
\expandafter\gdef\csname odunum@val@axis_breakdown.ftr.gemini.axis6.6.12\endcsname{0.405}
\expandafter\gdef\csname odunum@n@axis_breakdown.ftr.gemini.axis6.6.12\endcsname{42}
\expandafter\gdef\csname odunum@ci@axis_breakdown.ftr.gemini.axis6.6.12\endcsname{[0.270, 0.555]}
\expandafter\gdef\csname odunum@val@axis_breakdown.ftr.gemini.axis6.6.6\endcsname{0.315}
\expandafter\gdef\csname odunum@n@axis_breakdown.ftr.gemini.axis6.6.6\endcsname{54}
\expandafter\gdef\csname odunum@ci@axis_breakdown.ftr.gemini.axis6.6.6\endcsname{[0.207, 0.447]}
\expandafter\gdef\csname odunum@val@axis_breakdown.ftr.gemini.axis6.6.7\endcsname{0.425}
\expandafter\gdef\csname odunum@n@axis_breakdown.ftr.gemini.axis6.6.7\endcsname{47}
\expandafter\gdef\csname odunum@ci@axis_breakdown.ftr.gemini.axis6.6.7\endcsname{[0.295, 0.567]}
\expandafter\gdef\csname odunum@val@axis_breakdown.ftr.gemini35_flash_lite.axis4.4.1\endcsname{0.464}
\expandafter\gdef\csname odunum@n@axis_breakdown.ftr.gemini35_flash_lite.axis4.4.1\endcsname{110}
\expandafter\gdef\csname odunum@ci@axis_breakdown.ftr.gemini35_flash_lite.axis4.4.1\endcsname{[0.373, 0.556]}
\expandafter\gdef\csname odunum@val@axis_breakdown.ftr.gemini35_flash_lite.axis4.4.2\endcsname{0.478}
\expandafter\gdef\csname odunum@n@axis_breakdown.ftr.gemini35_flash_lite.axis4.4.2\endcsname{90}
\expandafter\gdef\csname odunum@ci@axis_breakdown.ftr.gemini35_flash_lite.axis4.4.2\endcsname{[0.378, 0.580]}
\expandafter\gdef\csname odunum@val@axis_breakdown.ftr.gemini35_flash_lite.axis4.4.3\endcsname{0.493}
\expandafter\gdef\csname odunum@n@axis_breakdown.ftr.gemini35_flash_lite.axis4.4.3\endcsname{77}
\expandafter\gdef\csname odunum@ci@axis_breakdown.ftr.gemini35_flash_lite.axis4.4.3\endcsname{[0.385, 0.603]}
\expandafter\gdef\csname odunum@val@axis_breakdown.ftr.gemini35_flash_lite.axis4.4.4\endcsname{0.505}
\expandafter\gdef\csname odunum@n@axis_breakdown.ftr.gemini35_flash_lite.axis4.4.4\endcsname{101}
\expandafter\gdef\csname odunum@ci@axis_breakdown.ftr.gemini35_flash_lite.axis4.4.4\endcsname{[0.409, 0.600]}
\expandafter\gdef\csname odunum@val@axis_breakdown.ftr.gemini35_flash_lite.axis4.4.5\endcsname{0.551}
\expandafter\gdef\csname odunum@n@axis_breakdown.ftr.gemini35_flash_lite.axis4.4.5\endcsname{69}
\expandafter\gdef\csname odunum@ci@axis_breakdown.ftr.gemini35_flash_lite.axis4.4.5\endcsname{[0.434, 0.662]}
\expandafter\gdef\csname odunum@val@axis_breakdown.ftr.gemini35_flash_lite.axis6.6.1\endcsname{0.441}
\expandafter\gdef\csname odunum@n@axis_breakdown.ftr.gemini35_flash_lite.axis6.6.1\endcsname{59}
\expandafter\gdef\csname odunum@ci@axis_breakdown.ftr.gemini35_flash_lite.axis6.6.1\endcsname{[0.322, 0.567]}
\expandafter\gdef\csname odunum@val@axis_breakdown.ftr.gemini35_flash_lite.axis6.6.12\endcsname{0.738}
\expandafter\gdef\csname odunum@n@axis_breakdown.ftr.gemini35_flash_lite.axis6.6.12\endcsname{42}
\expandafter\gdef\csname odunum@ci@axis_breakdown.ftr.gemini35_flash_lite.axis6.6.12\endcsname{[0.589, 0.847]}
\expandafter\gdef\csname odunum@val@axis_breakdown.ftr.gemini35_flash_lite.axis6.6.6\endcsname{0.481}
\expandafter\gdef\csname odunum@n@axis_breakdown.ftr.gemini35_flash_lite.axis6.6.6\endcsname{54}
\expandafter\gdef\csname odunum@ci@axis_breakdown.ftr.gemini35_flash_lite.axis6.6.6\endcsname{[0.354, 0.611]}
\expandafter\gdef\csname odunum@val@axis_breakdown.ftr.gemini35_flash_lite.axis6.6.7\endcsname{0.511}
\expandafter\gdef\csname odunum@n@axis_breakdown.ftr.gemini35_flash_lite.axis6.6.7\endcsname{47}
\expandafter\gdef\csname odunum@ci@axis_breakdown.ftr.gemini35_flash_lite.axis6.6.7\endcsname{[0.372, 0.647]}
\expandafter\gdef\csname odunum@val@axis_breakdown.ftr.gemini37_flash.axis4.4.1\endcsname{0.193}
\expandafter\gdef\csname odunum@n@axis_breakdown.ftr.gemini37_flash.axis4.4.1\endcsname{109}
\expandafter\gdef\csname odunum@ci@axis_breakdown.ftr.gemini37_flash.axis4.4.1\endcsname{[0.130, 0.277]}
\expandafter\gdef\csname odunum@val@axis_breakdown.ftr.gemini37_flash.axis4.4.2\endcsname{0.261}
\expandafter\gdef\csname odunum@n@axis_breakdown.ftr.gemini37_flash.axis4.4.2\endcsname{88}
\expandafter\gdef\csname odunum@ci@axis_breakdown.ftr.gemini37_flash.axis4.4.2\endcsname{[0.181, 0.362]}
\expandafter\gdef\csname odunum@val@axis_breakdown.ftr.gemini37_flash.axis4.4.3\endcsname{0.132}
\expandafter\gdef\csname odunum@n@axis_breakdown.ftr.gemini37_flash.axis4.4.3\endcsname{76}
\expandafter\gdef\csname odunum@ci@axis_breakdown.ftr.gemini37_flash.axis4.4.3\endcsname{[0.073, 0.226]}
\expandafter\gdef\csname odunum@val@axis_breakdown.ftr.gemini37_flash.axis4.4.4\endcsname{0.184}
\expandafter\gdef\csname odunum@n@axis_breakdown.ftr.gemini37_flash.axis4.4.4\endcsname{98}
\expandafter\gdef\csname odunum@ci@axis_breakdown.ftr.gemini37_flash.axis4.4.4\endcsname{[0.119, 0.272]}
\expandafter\gdef\csname odunum@val@axis_breakdown.ftr.gemini37_flash.axis4.4.5\endcsname{0.449}
\expandafter\gdef\csname odunum@n@axis_breakdown.ftr.gemini37_flash.axis4.4.5\endcsname{69}
\expandafter\gdef\csname odunum@ci@axis_breakdown.ftr.gemini37_flash.axis4.4.5\endcsname{[0.338, 0.566]}
\expandafter\gdef\csname odunum@val@axis_breakdown.ftr.gemini37_flash.axis6.6.1\endcsname{0.186}
\expandafter\gdef\csname odunum@n@axis_breakdown.ftr.gemini37_flash.axis6.6.1\endcsname{59}
\expandafter\gdef\csname odunum@ci@axis_breakdown.ftr.gemini37_flash.axis6.6.1\endcsname{[0.107, 0.304]}
\expandafter\gdef\csname odunum@val@axis_breakdown.ftr.gemini37_flash.axis6.6.12\endcsname{0.238}
\expandafter\gdef\csname odunum@n@axis_breakdown.ftr.gemini37_flash.axis6.6.12\endcsname{42}
\expandafter\gdef\csname odunum@ci@axis_breakdown.ftr.gemini37_flash.axis6.6.12\endcsname{[0.135, 0.385]}
\expandafter\gdef\csname odunum@val@axis_breakdown.ftr.gemini37_flash.axis6.6.6\endcsname{0.132}
\expandafter\gdef\csname odunum@n@axis_breakdown.ftr.gemini37_flash.axis6.6.6\endcsname{53}
\expandafter\gdef\csname odunum@ci@axis_breakdown.ftr.gemini37_flash.axis6.6.6\endcsname{[0.065, 0.248]}
\expandafter\gdef\csname odunum@val@axis_breakdown.ftr.gemini37_flash.axis6.6.7\endcsname{0.326}
\expandafter\gdef\csname odunum@n@axis_breakdown.ftr.gemini37_flash.axis6.6.7\endcsname{43}
\expandafter\gdef\csname odunum@ci@axis_breakdown.ftr.gemini37_flash.axis6.6.7\endcsname{[0.205, 0.475]}
\expandafter\gdef\csname odunum@val@axis_breakdown.ftr.ming.axis4.4.1\endcsname{0.836}
\expandafter\gdef\csname odunum@n@axis_breakdown.ftr.ming.axis4.4.1\endcsname{110}
\expandafter\gdef\csname odunum@ci@axis_breakdown.ftr.ming.axis4.4.1\endcsname{[0.756, 0.894]}
\expandafter\gdef\csname odunum@val@axis_breakdown.ftr.ming.axis4.4.2\endcsname{0.900}
\expandafter\gdef\csname odunum@n@axis_breakdown.ftr.ming.axis4.4.2\endcsname{90}
\expandafter\gdef\csname odunum@ci@axis_breakdown.ftr.ming.axis4.4.2\endcsname{[0.821, 0.946]}
\expandafter\gdef\csname odunum@val@axis_breakdown.ftr.ming.axis4.4.3\endcsname{0.883}
\expandafter\gdef\csname odunum@n@axis_breakdown.ftr.ming.axis4.4.3\endcsname{77}
\expandafter\gdef\csname odunum@ci@axis_breakdown.ftr.ming.axis4.4.3\endcsname{[0.793, 0.937]}
\expandafter\gdef\csname odunum@val@axis_breakdown.ftr.ming.axis4.4.4\endcsname{0.832}
\expandafter\gdef\csname odunum@n@axis_breakdown.ftr.ming.axis4.4.4\endcsname{101}
\expandafter\gdef\csname odunum@ci@axis_breakdown.ftr.ming.axis4.4.4\endcsname{[0.747, 0.892]}
\expandafter\gdef\csname odunum@val@axis_breakdown.ftr.ming.axis4.4.5\endcsname{0.942}
\expandafter\gdef\csname odunum@n@axis_breakdown.ftr.ming.axis4.4.5\endcsname{69}
\expandafter\gdef\csname odunum@ci@axis_breakdown.ftr.ming.axis4.4.5\endcsname{[0.860, 0.977]}
\expandafter\gdef\csname odunum@val@axis_breakdown.ftr.ming.axis6.6.1\endcsname{0.814}
\expandafter\gdef\csname odunum@n@axis_breakdown.ftr.ming.axis6.6.1\endcsname{59}
\expandafter\gdef\csname odunum@ci@axis_breakdown.ftr.ming.axis6.6.1\endcsname{[0.696, 0.893]}
\expandafter\gdef\csname odunum@val@axis_breakdown.ftr.ming.axis6.6.12\endcsname{0.905}
\expandafter\gdef\csname odunum@n@axis_breakdown.ftr.ming.axis6.6.12\endcsname{42}
\expandafter\gdef\csname odunum@ci@axis_breakdown.ftr.ming.axis6.6.12\endcsname{[0.779, 0.962]}
\expandafter\gdef\csname odunum@val@axis_breakdown.ftr.ming.axis6.6.6\endcsname{0.870}
\expandafter\gdef\csname odunum@n@axis_breakdown.ftr.ming.axis6.6.6\endcsname{54}
\expandafter\gdef\csname odunum@ci@axis_breakdown.ftr.ming.axis6.6.6\endcsname{[0.756, 0.936]}
\expandafter\gdef\csname odunum@val@axis_breakdown.ftr.ming.axis6.6.7\endcsname{0.894}
\expandafter\gdef\csname odunum@n@axis_breakdown.ftr.ming.axis6.6.7\endcsname{47}
\expandafter\gdef\csname odunum@ci@axis_breakdown.ftr.ming.axis6.6.7\endcsname{[0.774, 0.954]}
\expandafter\gdef\csname odunum@val@axis_breakdown.ftr.minicpm_o.axis4.4.1\endcsname{0.809}
\expandafter\gdef\csname odunum@n@axis_breakdown.ftr.minicpm_o.axis4.4.1\endcsname{110}
\expandafter\gdef\csname odunum@ci@axis_breakdown.ftr.minicpm_o.axis4.4.1\endcsname{[0.726, 0.872]}
\expandafter\gdef\csname odunum@val@axis_breakdown.ftr.minicpm_o.axis4.4.2\endcsname{0.844}
\expandafter\gdef\csname odunum@n@axis_breakdown.ftr.minicpm_o.axis4.4.2\endcsname{90}
\expandafter\gdef\csname odunum@ci@axis_breakdown.ftr.minicpm_o.axis4.4.2\endcsname{[0.756, 0.905]}
\expandafter\gdef\csname odunum@val@axis_breakdown.ftr.minicpm_o.axis4.4.3\endcsname{0.789}
\expandafter\gdef\csname odunum@n@axis_breakdown.ftr.minicpm_o.axis4.4.3\endcsname{76}
\expandafter\gdef\csname odunum@ci@axis_breakdown.ftr.minicpm_o.axis4.4.3\endcsname{[0.685, 0.866]}
\expandafter\gdef\csname odunum@val@axis_breakdown.ftr.minicpm_o.axis4.4.4\endcsname{0.792}
\expandafter\gdef\csname odunum@n@axis_breakdown.ftr.minicpm_o.axis4.4.4\endcsname{101}
\expandafter\gdef\csname odunum@ci@axis_breakdown.ftr.minicpm_o.axis4.4.4\endcsname{[0.703, 0.860]}
\expandafter\gdef\csname odunum@val@axis_breakdown.ftr.minicpm_o.axis4.4.5\endcsname{0.841}
\expandafter\gdef\csname odunum@n@axis_breakdown.ftr.minicpm_o.axis4.4.5\endcsname{69}
\expandafter\gdef\csname odunum@ci@axis_breakdown.ftr.minicpm_o.axis4.4.5\endcsname{[0.737, 0.909]}
\expandafter\gdef\csname odunum@val@axis_breakdown.ftr.minicpm_o.axis6.6.1\endcsname{0.881}
\expandafter\gdef\csname odunum@n@axis_breakdown.ftr.minicpm_o.axis6.6.1\endcsname{59}
\expandafter\gdef\csname odunum@ci@axis_breakdown.ftr.minicpm_o.axis6.6.1\endcsname{[0.775, 0.941]}
\expandafter\gdef\csname odunum@val@axis_breakdown.ftr.minicpm_o.axis6.6.12\endcsname{0.929}
\expandafter\gdef\csname odunum@n@axis_breakdown.ftr.minicpm_o.axis6.6.12\endcsname{42}
\expandafter\gdef\csname odunum@ci@axis_breakdown.ftr.minicpm_o.axis6.6.12\endcsname{[0.810, 0.975]}
\expandafter\gdef\csname odunum@val@axis_breakdown.ftr.minicpm_o.axis6.6.6\endcsname{0.811}
\expandafter\gdef\csname odunum@n@axis_breakdown.ftr.minicpm_o.axis6.6.6\endcsname{53}
\expandafter\gdef\csname odunum@ci@axis_breakdown.ftr.minicpm_o.axis6.6.6\endcsname{[0.686, 0.894]}
\expandafter\gdef\csname odunum@val@axis_breakdown.ftr.minicpm_o.axis6.6.7\endcsname{0.851}
\expandafter\gdef\csname odunum@n@axis_breakdown.ftr.minicpm_o.axis6.6.7\endcsname{47}
\expandafter\gdef\csname odunum@ci@axis_breakdown.ftr.minicpm_o.axis6.6.7\endcsname{[0.723, 0.926]}
\expandafter\gdef\csname odunum@val@axis_breakdown.ftr.nemotron.axis4.4.1\endcsname{0.754}
\expandafter\gdef\csname odunum@n@axis_breakdown.ftr.nemotron.axis4.4.1\endcsname{110}
\expandafter\gdef\csname odunum@ci@axis_breakdown.ftr.nemotron.axis4.4.1\endcsname{[0.666, 0.825]}
\expandafter\gdef\csname odunum@val@axis_breakdown.ftr.nemotron.axis4.4.2\endcsname{0.733}
\expandafter\gdef\csname odunum@n@axis_breakdown.ftr.nemotron.axis4.4.2\endcsname{90}
\expandafter\gdef\csname odunum@ci@axis_breakdown.ftr.nemotron.axis4.4.2\endcsname{[0.634, 0.814]}
\expandafter\gdef\csname odunum@val@axis_breakdown.ftr.nemotron.axis4.4.3\endcsname{0.662}
\expandafter\gdef\csname odunum@n@axis_breakdown.ftr.nemotron.axis4.4.3\endcsname{77}
\expandafter\gdef\csname odunum@ci@axis_breakdown.ftr.nemotron.axis4.4.3\endcsname{[0.551, 0.758]}
\expandafter\gdef\csname odunum@val@axis_breakdown.ftr.nemotron.axis4.4.4\endcsname{0.792}
\expandafter\gdef\csname odunum@n@axis_breakdown.ftr.nemotron.axis4.4.4\endcsname{101}
\expandafter\gdef\csname odunum@ci@axis_breakdown.ftr.nemotron.axis4.4.4\endcsname{[0.703, 0.860]}
\expandafter\gdef\csname odunum@val@axis_breakdown.ftr.nemotron.axis4.4.5\endcsname{0.609}
\expandafter\gdef\csname odunum@n@axis_breakdown.ftr.nemotron.axis4.4.5\endcsname{69}
\expandafter\gdef\csname odunum@ci@axis_breakdown.ftr.nemotron.axis4.4.5\endcsname{[0.491, 0.715]}
\expandafter\gdef\csname odunum@val@axis_breakdown.ftr.nemotron.axis6.6.1\endcsname{0.797}
\expandafter\gdef\csname odunum@n@axis_breakdown.ftr.nemotron.axis6.6.1\endcsname{59}
\expandafter\gdef\csname odunum@ci@axis_breakdown.ftr.nemotron.axis6.6.1\endcsname{[0.677, 0.880]}
\expandafter\gdef\csname odunum@val@axis_breakdown.ftr.nemotron.axis6.6.12\endcsname{0.643}
\expandafter\gdef\csname odunum@n@axis_breakdown.ftr.nemotron.axis6.6.12\endcsname{42}
\expandafter\gdef\csname odunum@ci@axis_breakdown.ftr.nemotron.axis6.6.12\endcsname{[0.492, 0.770]}
\expandafter\gdef\csname odunum@val@axis_breakdown.ftr.nemotron.axis6.6.6\endcsname{0.704}
\expandafter\gdef\csname odunum@n@axis_breakdown.ftr.nemotron.axis6.6.6\endcsname{54}
\expandafter\gdef\csname odunum@ci@axis_breakdown.ftr.nemotron.axis6.6.6\endcsname{[0.572, 0.809]}
\expandafter\gdef\csname odunum@val@axis_breakdown.ftr.nemotron.axis6.6.7\endcsname{0.808}
\expandafter\gdef\csname odunum@n@axis_breakdown.ftr.nemotron.axis6.6.7\endcsname{47}
\expandafter\gdef\csname odunum@ci@axis_breakdown.ftr.nemotron.axis6.6.7\endcsname{[0.675, 0.896]}
\expandafter\gdef\csname odunum@val@axis_breakdown.ftr.qwen25_omni.axis4.4.1\endcsname{0.809}
\expandafter\gdef\csname odunum@n@axis_breakdown.ftr.qwen25_omni.axis4.4.1\endcsname{110}
\expandafter\gdef\csname odunum@ci@axis_breakdown.ftr.qwen25_omni.axis4.4.1\endcsname{[0.726, 0.872]}
\expandafter\gdef\csname odunum@val@axis_breakdown.ftr.qwen25_omni.axis4.4.2\endcsname{0.789}
\expandafter\gdef\csname odunum@n@axis_breakdown.ftr.qwen25_omni.axis4.4.2\endcsname{90}
\expandafter\gdef\csname odunum@ci@axis_breakdown.ftr.qwen25_omni.axis4.4.2\endcsname{[0.694, 0.860]}
\expandafter\gdef\csname odunum@val@axis_breakdown.ftr.qwen25_omni.axis4.4.3\endcsname{0.792}
\expandafter\gdef\csname odunum@n@axis_breakdown.ftr.qwen25_omni.axis4.4.3\endcsname{77}
\expandafter\gdef\csname odunum@ci@axis_breakdown.ftr.qwen25_omni.axis4.4.3\endcsname{[0.689, 0.868]}
\expandafter\gdef\csname odunum@val@axis_breakdown.ftr.qwen25_omni.axis4.4.4\endcsname{0.792}
\expandafter\gdef\csname odunum@n@axis_breakdown.ftr.qwen25_omni.axis4.4.4\endcsname{101}
\expandafter\gdef\csname odunum@ci@axis_breakdown.ftr.qwen25_omni.axis4.4.4\endcsname{[0.703, 0.860]}
\expandafter\gdef\csname odunum@val@axis_breakdown.ftr.qwen25_omni.axis4.4.5\endcsname{0.826}
\expandafter\gdef\csname odunum@n@axis_breakdown.ftr.qwen25_omni.axis4.4.5\endcsname{69}
\expandafter\gdef\csname odunum@ci@axis_breakdown.ftr.qwen25_omni.axis4.4.5\endcsname{[0.720, 0.898]}
\expandafter\gdef\csname odunum@val@axis_breakdown.ftr.qwen25_omni.axis6.6.1\endcsname{0.746}
\expandafter\gdef\csname odunum@n@axis_breakdown.ftr.qwen25_omni.axis6.6.1\endcsname{59}
\expandafter\gdef\csname odunum@ci@axis_breakdown.ftr.qwen25_omni.axis6.6.1\endcsname{[0.622, 0.839]}
\expandafter\gdef\csname odunum@val@axis_breakdown.ftr.qwen25_omni.axis6.6.12\endcsname{0.809}
\expandafter\gdef\csname odunum@n@axis_breakdown.ftr.qwen25_omni.axis6.6.12\endcsname{42}
\expandafter\gdef\csname odunum@ci@axis_breakdown.ftr.qwen25_omni.axis6.6.12\endcsname{[0.667, 0.900]}
\expandafter\gdef\csname odunum@val@axis_breakdown.ftr.qwen25_omni.axis6.6.6\endcsname{0.889}
\expandafter\gdef\csname odunum@n@axis_breakdown.ftr.qwen25_omni.axis6.6.6\endcsname{54}
\expandafter\gdef\csname odunum@ci@axis_breakdown.ftr.qwen25_omni.axis6.6.6\endcsname{[0.778, 0.948]}
\expandafter\gdef\csname odunum@val@axis_breakdown.ftr.qwen25_omni.axis6.6.7\endcsname{0.808}
\expandafter\gdef\csname odunum@n@axis_breakdown.ftr.qwen25_omni.axis6.6.7\endcsname{47}
\expandafter\gdef\csname odunum@ci@axis_breakdown.ftr.qwen25_omni.axis6.6.7\endcsname{[0.675, 0.896]}
\expandafter\gdef\csname odunum@val@axis_breakdown.ftr.qwen3_omni_instruct.axis4.4.1\endcsname{0.882}
\expandafter\gdef\csname odunum@n@axis_breakdown.ftr.qwen3_omni_instruct.axis4.4.1\endcsname{110}
\expandafter\gdef\csname odunum@ci@axis_breakdown.ftr.qwen3_omni_instruct.axis4.4.1\endcsname{[0.808, 0.930]}
\expandafter\gdef\csname odunum@val@axis_breakdown.ftr.qwen3_omni_instruct.axis4.4.2\endcsname{0.889}
\expandafter\gdef\csname odunum@n@axis_breakdown.ftr.qwen3_omni_instruct.axis4.4.2\endcsname{90}
\expandafter\gdef\csname odunum@ci@axis_breakdown.ftr.qwen3_omni_instruct.axis4.4.2\endcsname{[0.807, 0.939]}
\expandafter\gdef\csname odunum@val@axis_breakdown.ftr.qwen3_omni_instruct.axis4.4.3\endcsname{0.896}
\expandafter\gdef\csname odunum@n@axis_breakdown.ftr.qwen3_omni_instruct.axis4.4.3\endcsname{77}
\expandafter\gdef\csname odunum@ci@axis_breakdown.ftr.qwen3_omni_instruct.axis4.4.3\endcsname{[0.808, 0.946]}
\expandafter\gdef\csname odunum@val@axis_breakdown.ftr.qwen3_omni_instruct.axis4.4.4\endcsname{0.871}
\expandafter\gdef\csname odunum@n@axis_breakdown.ftr.qwen3_omni_instruct.axis4.4.4\endcsname{101}
\expandafter\gdef\csname odunum@ci@axis_breakdown.ftr.qwen3_omni_instruct.axis4.4.4\endcsname{[0.792, 0.923]}
\expandafter\gdef\csname odunum@val@axis_breakdown.ftr.qwen3_omni_instruct.axis4.4.5\endcsname{0.913}
\expandafter\gdef\csname odunum@n@axis_breakdown.ftr.qwen3_omni_instruct.axis4.4.5\endcsname{69}
\expandafter\gdef\csname odunum@ci@axis_breakdown.ftr.qwen3_omni_instruct.axis4.4.5\endcsname{[0.823, 0.960]}
\expandafter\gdef\csname odunum@val@axis_breakdown.ftr.qwen3_omni_instruct.axis6.6.1\endcsname{0.949}
\expandafter\gdef\csname odunum@n@axis_breakdown.ftr.qwen3_omni_instruct.axis6.6.1\endcsname{59}
\expandafter\gdef\csname odunum@ci@axis_breakdown.ftr.qwen3_omni_instruct.axis6.6.1\endcsname{[0.861, 0.983]}
\expandafter\gdef\csname odunum@val@axis_breakdown.ftr.qwen3_omni_instruct.axis6.6.12\endcsname{1.000}
\expandafter\gdef\csname odunum@n@axis_breakdown.ftr.qwen3_omni_instruct.axis6.6.12\endcsname{42}
\expandafter\gdef\csname odunum@ci@axis_breakdown.ftr.qwen3_omni_instruct.axis6.6.12\endcsname{[0.916, 1.000]}
\expandafter\gdef\csname odunum@val@axis_breakdown.ftr.qwen3_omni_instruct.axis6.6.6\endcsname{0.907}
\expandafter\gdef\csname odunum@n@axis_breakdown.ftr.qwen3_omni_instruct.axis6.6.6\endcsname{54}
\expandafter\gdef\csname odunum@ci@axis_breakdown.ftr.qwen3_omni_instruct.axis6.6.6\endcsname{[0.801, 0.960]}
\expandafter\gdef\csname odunum@val@axis_breakdown.ftr.qwen3_omni_instruct.axis6.6.7\endcsname{0.851}
\expandafter\gdef\csname odunum@n@axis_breakdown.ftr.qwen3_omni_instruct.axis6.6.7\endcsname{47}
\expandafter\gdef\csname odunum@ci@axis_breakdown.ftr.qwen3_omni_instruct.axis6.6.7\endcsname{[0.723, 0.926]}
\expandafter\gdef\csname odunum@val@axis_breakdown.ftr.qwen3_omni_think.axis4.4.1\endcsname{0.700}
\expandafter\gdef\csname odunum@n@axis_breakdown.ftr.qwen3_omni_think.axis4.4.1\endcsname{110}
\expandafter\gdef\csname odunum@ci@axis_breakdown.ftr.qwen3_omni_think.axis4.4.1\endcsname{[0.609, 0.778]}
\expandafter\gdef\csname odunum@val@axis_breakdown.ftr.qwen3_omni_think.axis4.4.2\endcsname{0.744}
\expandafter\gdef\csname odunum@n@axis_breakdown.ftr.qwen3_omni_think.axis4.4.2\endcsname{90}
\expandafter\gdef\csname odunum@ci@axis_breakdown.ftr.qwen3_omni_think.axis4.4.2\endcsname{[0.646, 0.823]}
\expandafter\gdef\csname odunum@val@axis_breakdown.ftr.qwen3_omni_think.axis4.4.3\endcsname{0.662}
\expandafter\gdef\csname odunum@n@axis_breakdown.ftr.qwen3_omni_think.axis4.4.3\endcsname{77}
\expandafter\gdef\csname odunum@ci@axis_breakdown.ftr.qwen3_omni_think.axis4.4.3\endcsname{[0.551, 0.758]}
\expandafter\gdef\csname odunum@val@axis_breakdown.ftr.qwen3_omni_think.axis4.4.4\endcsname{0.822}
\expandafter\gdef\csname odunum@n@axis_breakdown.ftr.qwen3_omni_think.axis4.4.4\endcsname{101}
\expandafter\gdef\csname odunum@ci@axis_breakdown.ftr.qwen3_omni_think.axis4.4.4\endcsname{[0.736, 0.884]}
\expandafter\gdef\csname odunum@val@axis_breakdown.ftr.qwen3_omni_think.axis4.4.5\endcsname{0.826}
\expandafter\gdef\csname odunum@n@axis_breakdown.ftr.qwen3_omni_think.axis4.4.5\endcsname{69}
\expandafter\gdef\csname odunum@ci@axis_breakdown.ftr.qwen3_omni_think.axis4.4.5\endcsname{[0.720, 0.898]}
\expandafter\gdef\csname odunum@val@axis_breakdown.ftr.qwen3_omni_think.axis6.6.1\endcsname{0.780}
\expandafter\gdef\csname odunum@n@axis_breakdown.ftr.qwen3_omni_think.axis6.6.1\endcsname{59}
\expandafter\gdef\csname odunum@ci@axis_breakdown.ftr.qwen3_omni_think.axis6.6.1\endcsname{[0.659, 0.866]}
\expandafter\gdef\csname odunum@val@axis_breakdown.ftr.qwen3_omni_think.axis6.6.12\endcsname{0.833}
\expandafter\gdef\csname odunum@n@axis_breakdown.ftr.qwen3_omni_think.axis6.6.12\endcsname{42}
\expandafter\gdef\csname odunum@ci@axis_breakdown.ftr.qwen3_omni_think.axis6.6.12\endcsname{[0.694, 0.917]}
\expandafter\gdef\csname odunum@val@axis_breakdown.ftr.qwen3_omni_think.axis6.6.6\endcsname{0.722}
\expandafter\gdef\csname odunum@n@axis_breakdown.ftr.qwen3_omni_think.axis6.6.6\endcsname{54}
\expandafter\gdef\csname odunum@ci@axis_breakdown.ftr.qwen3_omni_think.axis6.6.6\endcsname{[0.591, 0.824]}
\expandafter\gdef\csname odunum@val@axis_breakdown.ftr.qwen3_omni_think.axis6.6.7\endcsname{0.723}
\expandafter\gdef\csname odunum@n@axis_breakdown.ftr.qwen3_omni_think.axis6.6.7\endcsname{47}
\expandafter\gdef\csname odunum@ci@axis_breakdown.ftr.qwen3_omni_think.axis6.6.7\endcsname{[0.582, 0.831]}
\expandafter\gdef\csname odunum@val@axis_breakdown.ftr.qwen_plus.axis4.4.1\endcsname{0.664}
\expandafter\gdef\csname odunum@n@axis_breakdown.ftr.qwen_plus.axis4.4.1\endcsname{110}
\expandafter\gdef\csname odunum@ci@axis_breakdown.ftr.qwen_plus.axis4.4.1\endcsname{[0.571, 0.745]}
\expandafter\gdef\csname odunum@val@axis_breakdown.ftr.qwen_plus.axis4.4.2\endcsname{0.722}
\expandafter\gdef\csname odunum@n@axis_breakdown.ftr.qwen_plus.axis4.4.2\endcsname{90}
\expandafter\gdef\csname odunum@ci@axis_breakdown.ftr.qwen_plus.axis4.4.2\endcsname{[0.622, 0.804]}
\expandafter\gdef\csname odunum@val@axis_breakdown.ftr.qwen_plus.axis4.4.3\endcsname{0.597}
\expandafter\gdef\csname odunum@n@axis_breakdown.ftr.qwen_plus.axis4.4.3\endcsname{77}
\expandafter\gdef\csname odunum@ci@axis_breakdown.ftr.qwen_plus.axis4.4.3\endcsname{[0.486, 0.700]}
\expandafter\gdef\csname odunum@val@axis_breakdown.ftr.qwen_plus.axis4.4.4\endcsname{0.653}
\expandafter\gdef\csname odunum@n@axis_breakdown.ftr.qwen_plus.axis4.4.4\endcsname{101}
\expandafter\gdef\csname odunum@ci@axis_breakdown.ftr.qwen_plus.axis4.4.4\endcsname{[0.557, 0.739]}
\expandafter\gdef\csname odunum@val@axis_breakdown.ftr.qwen_plus.axis4.4.5\endcsname{0.725}
\expandafter\gdef\csname odunum@n@axis_breakdown.ftr.qwen_plus.axis4.4.5\endcsname{69}
\expandafter\gdef\csname odunum@ci@axis_breakdown.ftr.qwen_plus.axis4.4.5\endcsname{[0.610, 0.816]}
\expandafter\gdef\csname odunum@val@axis_breakdown.ftr.qwen_plus.axis6.6.1\endcsname{0.678}
\expandafter\gdef\csname odunum@n@axis_breakdown.ftr.qwen_plus.axis6.6.1\endcsname{59}
\expandafter\gdef\csname odunum@ci@axis_breakdown.ftr.qwen_plus.axis6.6.1\endcsname{[0.551, 0.783]}
\expandafter\gdef\csname odunum@val@axis_breakdown.ftr.qwen_plus.axis6.6.12\endcsname{0.833}
\expandafter\gdef\csname odunum@n@axis_breakdown.ftr.qwen_plus.axis6.6.12\endcsname{42}
\expandafter\gdef\csname odunum@ci@axis_breakdown.ftr.qwen_plus.axis6.6.12\endcsname{[0.694, 0.917]}
\expandafter\gdef\csname odunum@val@axis_breakdown.ftr.qwen_plus.axis6.6.6\endcsname{0.685}
\expandafter\gdef\csname odunum@n@axis_breakdown.ftr.qwen_plus.axis6.6.6\endcsname{54}
\expandafter\gdef\csname odunum@ci@axis_breakdown.ftr.qwen_plus.axis6.6.6\endcsname{[0.553, 0.793]}
\expandafter\gdef\csname odunum@val@axis_breakdown.ftr.qwen_plus.axis6.6.7\endcsname{0.723}
\expandafter\gdef\csname odunum@n@axis_breakdown.ftr.qwen_plus.axis6.6.7\endcsname{47}
\expandafter\gdef\csname odunum@ci@axis_breakdown.ftr.qwen_plus.axis6.6.7\endcsname{[0.582, 0.831]}
\expandafter\gdef\csname odunum@val@axis_breakdown.ftr.salmonn2_7b.axis4.4.1\endcsname{0.909}
\expandafter\gdef\csname odunum@n@axis_breakdown.ftr.salmonn2_7b.axis4.4.1\endcsname{110}
\expandafter\gdef\csname odunum@ci@axis_breakdown.ftr.salmonn2_7b.axis4.4.1\endcsname{[0.841, 0.950]}
\expandafter\gdef\csname odunum@val@axis_breakdown.ftr.salmonn2_7b.axis4.4.2\endcsname{0.900}
\expandafter\gdef\csname odunum@n@axis_breakdown.ftr.salmonn2_7b.axis4.4.2\endcsname{90}
\expandafter\gdef\csname odunum@ci@axis_breakdown.ftr.salmonn2_7b.axis4.4.2\endcsname{[0.821, 0.946]}
\expandafter\gdef\csname odunum@val@axis_breakdown.ftr.salmonn2_7b.axis4.4.3\endcsname{0.948}
\expandafter\gdef\csname odunum@n@axis_breakdown.ftr.salmonn2_7b.axis4.4.3\endcsname{77}
\expandafter\gdef\csname odunum@ci@axis_breakdown.ftr.salmonn2_7b.axis4.4.3\endcsname{[0.874, 0.980]}
\expandafter\gdef\csname odunum@val@axis_breakdown.ftr.salmonn2_7b.axis4.4.4\endcsname{0.919}
\expandafter\gdef\csname odunum@n@axis_breakdown.ftr.salmonn2_7b.axis4.4.4\endcsname{99}
\expandafter\gdef\csname odunum@ci@axis_breakdown.ftr.salmonn2_7b.axis4.4.4\endcsname{[0.849, 0.958]}
\expandafter\gdef\csname odunum@val@axis_breakdown.ftr.salmonn2_7b.axis4.4.5\endcsname{0.941}
\expandafter\gdef\csname odunum@n@axis_breakdown.ftr.salmonn2_7b.axis4.4.5\endcsname{68}
\expandafter\gdef\csname odunum@ci@axis_breakdown.ftr.salmonn2_7b.axis4.4.5\endcsname{[0.858, 0.977]}
\expandafter\gdef\csname odunum@val@axis_breakdown.ftr.salmonn2_7b.axis6.6.1\endcsname{0.948}
\expandafter\gdef\csname odunum@n@axis_breakdown.ftr.salmonn2_7b.axis6.6.1\endcsname{58}
\expandafter\gdef\csname odunum@ci@axis_breakdown.ftr.salmonn2_7b.axis6.6.1\endcsname{[0.859, 0.982]}
\expandafter\gdef\csname odunum@val@axis_breakdown.ftr.salmonn2_7b.axis6.6.12\endcsname{0.929}
\expandafter\gdef\csname odunum@n@axis_breakdown.ftr.salmonn2_7b.axis6.6.12\endcsname{42}
\expandafter\gdef\csname odunum@ci@axis_breakdown.ftr.salmonn2_7b.axis6.6.12\endcsname{[0.810, 0.975]}
\expandafter\gdef\csname odunum@val@axis_breakdown.ftr.salmonn2_7b.axis6.6.6\endcsname{0.962}
\expandafter\gdef\csname odunum@n@axis_breakdown.ftr.salmonn2_7b.axis6.6.6\endcsname{53}
\expandafter\gdef\csname odunum@ci@axis_breakdown.ftr.salmonn2_7b.axis6.6.6\endcsname{[0.872, 0.990]}
\expandafter\gdef\csname odunum@val@axis_breakdown.ftr.salmonn2_7b.axis6.6.7\endcsname{0.936}
\expandafter\gdef\csname odunum@n@axis_breakdown.ftr.salmonn2_7b.axis6.6.7\endcsname{47}
\expandafter\gdef\csname odunum@ci@axis_breakdown.ftr.salmonn2_7b.axis6.6.7\endcsname{[0.828, 0.978]}
\expandafter\gdef\csname odunum@val@axis_breakdown.ftr.seed.axis4.4.1\endcsname{0.836}
\expandafter\gdef\csname odunum@n@axis_breakdown.ftr.seed.axis4.4.1\endcsname{110}
\expandafter\gdef\csname odunum@ci@axis_breakdown.ftr.seed.axis4.4.1\endcsname{[0.756, 0.894]}
\expandafter\gdef\csname odunum@val@axis_breakdown.ftr.seed.axis4.4.2\endcsname{0.800}
\expandafter\gdef\csname odunum@n@axis_breakdown.ftr.seed.axis4.4.2\endcsname{90}
\expandafter\gdef\csname odunum@ci@axis_breakdown.ftr.seed.axis4.4.2\endcsname{[0.706, 0.870]}
\expandafter\gdef\csname odunum@val@axis_breakdown.ftr.seed.axis4.4.3\endcsname{0.779}
\expandafter\gdef\csname odunum@n@axis_breakdown.ftr.seed.axis4.4.3\endcsname{77}
\expandafter\gdef\csname odunum@ci@axis_breakdown.ftr.seed.axis4.4.3\endcsname{[0.675, 0.857]}
\expandafter\gdef\csname odunum@val@axis_breakdown.ftr.seed.axis4.4.4\endcsname{0.802}
\expandafter\gdef\csname odunum@n@axis_breakdown.ftr.seed.axis4.4.4\endcsname{101}
\expandafter\gdef\csname odunum@ci@axis_breakdown.ftr.seed.axis4.4.4\endcsname{[0.714, 0.868]}
\expandafter\gdef\csname odunum@val@axis_breakdown.ftr.seed.axis4.4.5\endcsname{0.870}
\expandafter\gdef\csname odunum@n@axis_breakdown.ftr.seed.axis4.4.5\endcsname{69}
\expandafter\gdef\csname odunum@ci@axis_breakdown.ftr.seed.axis4.4.5\endcsname{[0.770, 0.930]}
\expandafter\gdef\csname odunum@val@axis_breakdown.ftr.seed.axis6.6.1\endcsname{0.797}
\expandafter\gdef\csname odunum@n@axis_breakdown.ftr.seed.axis6.6.1\endcsname{59}
\expandafter\gdef\csname odunum@ci@axis_breakdown.ftr.seed.axis6.6.1\endcsname{[0.677, 0.880]}
\expandafter\gdef\csname odunum@val@axis_breakdown.ftr.seed.axis6.6.12\endcsname{0.905}
\expandafter\gdef\csname odunum@n@axis_breakdown.ftr.seed.axis6.6.12\endcsname{42}
\expandafter\gdef\csname odunum@ci@axis_breakdown.ftr.seed.axis6.6.12\endcsname{[0.779, 0.962]}
\expandafter\gdef\csname odunum@val@axis_breakdown.ftr.seed.axis6.6.6\endcsname{0.815}
\expandafter\gdef\csname odunum@n@axis_breakdown.ftr.seed.axis6.6.6\endcsname{54}
\expandafter\gdef\csname odunum@ci@axis_breakdown.ftr.seed.axis6.6.6\endcsname{[0.692, 0.896]}
\expandafter\gdef\csname odunum@val@axis_breakdown.ftr.seed.axis6.6.7\endcsname{0.808}
\expandafter\gdef\csname odunum@n@axis_breakdown.ftr.seed.axis6.6.7\endcsname{47}
\expandafter\gdef\csname odunum@ci@axis_breakdown.ftr.seed.axis6.6.7\endcsname{[0.675, 0.896]}
\expandafter\gdef\csname odunum@val@axis_breakdown.integrity.axis1.audio_only_out_of_range\endcsname{33}
\expandafter\gdef\csname odunum@n@axis_breakdown.integrity.axis1.audio_only_out_of_range\endcsname{570}
\expandafter\gdef\csname odunum@val@axis_breakdown.integrity.discrimination.n_present\endcsname{0}
\expandafter\gdef\csname odunum@n@axis_breakdown.integrity.discrimination.n_present\endcsname{2\,078}
\expandafter\gdef\csname odunum@val@axis_breakdown.integrity.empirical_difficulty.n_present\endcsname{0}
\expandafter\gdef\csname odunum@n@axis_breakdown.integrity.empirical_difficulty.n_present\endcsname{2\,078}
\expandafter\gdef\csname odunum@val@axis_breakdown.integrity.intrinsic_difficulty.n_present\endcsname{1\,801}
\expandafter\gdef\csname odunum@n@axis_breakdown.integrity.intrinsic_difficulty.n_present\endcsname{2\,078}
\expandafter\gdef\csname odunum@val@axis_breakdown.integrity.intrinsic_difficulty.negatives_at_zero\endcsname{372}
\expandafter\gdef\csname odunum@n@axis_breakdown.integrity.intrinsic_difficulty.negatives_at_zero\endcsname{372}
\expandafter\gdef\csname odunum@val@axis_breakdown.m2.cascade_asr.axis1.audio_only.1.1\endcsname{0.765}
\expandafter\gdef\csname odunum@n@axis_breakdown.m2.cascade_asr.axis1.audio_only.1.1\endcsname{158}
\expandafter\gdef\csname odunum@ci@axis_breakdown.m2.cascade_asr.axis1.audio_only.1.1\endcsname{[0.727, 0.803]}
\expandafter\gdef\csname odunum@val@axis_breakdown.m2.cascade_asr.axis1.audio_only.1.2\endcsname{0.644}
\expandafter\gdef\csname odunum@n@axis_breakdown.m2.cascade_asr.axis1.audio_only.1.2\endcsname{379}
\expandafter\gdef\csname odunum@ci@axis_breakdown.m2.cascade_asr.axis1.audio_only.1.2\endcsname{[0.613, 0.674]}
\expandafter\gdef\csname odunum@val@axis_breakdown.m2.cascade_asr.axis1.audio_visual.1.1\endcsname{0.668}
\expandafter\gdef\csname odunum@n@axis_breakdown.m2.cascade_asr.axis1.audio_visual.1.1\endcsname{94}
\expandafter\gdef\csname odunum@ci@axis_breakdown.m2.cascade_asr.axis1.audio_visual.1.1\endcsname{[0.613, 0.723]}
\expandafter\gdef\csname odunum@val@axis_breakdown.m2.cascade_asr.axis1.audio_visual.1.2\endcsname{0.515}
\expandafter\gdef\csname odunum@n@axis_breakdown.m2.cascade_asr.axis1.audio_visual.1.2\endcsname{294}
\expandafter\gdef\csname odunum@ci@axis_breakdown.m2.cascade_asr.axis1.audio_visual.1.2\endcsname{[0.482, 0.547]}
\expandafter\gdef\csname odunum@val@axis_breakdown.m2.cascade_asr.axis1.audio_visual.1.3\endcsname{0.580}
\expandafter\gdef\csname odunum@n@axis_breakdown.m2.cascade_asr.axis1.audio_visual.1.3\endcsname{291}
\expandafter\gdef\csname odunum@ci@axis_breakdown.m2.cascade_asr.axis1.audio_visual.1.3\endcsname{[0.545, 0.614]}
\expandafter\gdef\csname odunum@val@axis_breakdown.m2.cascade_asr.axis1.audio_visual.1.4\endcsname{0.540}
\expandafter\gdef\csname odunum@n@axis_breakdown.m2.cascade_asr.axis1.audio_visual.1.4\endcsname{382}
\expandafter\gdef\csname odunum@ci@axis_breakdown.m2.cascade_asr.axis1.audio_visual.1.4\endcsname{[0.513, 0.567]}
\expandafter\gdef\csname odunum@val@axis_breakdown.m2.cascade_asr.axis2.2.1\endcsname{0.434}
\expandafter\gdef\csname odunum@n@axis_breakdown.m2.cascade_asr.axis2.2.1\endcsname{143}
\expandafter\gdef\csname odunum@ci@axis_breakdown.m2.cascade_asr.axis2.2.1\endcsname{[0.383, 0.485]}
\expandafter\gdef\csname odunum@val@axis_breakdown.m2.cascade_asr.axis2.2.2\endcsname{0.653}
\expandafter\gdef\csname odunum@n@axis_breakdown.m2.cascade_asr.axis2.2.2\endcsname{440}
\expandafter\gdef\csname odunum@ci@axis_breakdown.m2.cascade_asr.axis2.2.2\endcsname{[0.630, 0.677]}
\expandafter\gdef\csname odunum@val@axis_breakdown.m2.cascade_asr.axis2.2.3\endcsname{0.688}
\expandafter\gdef\csname odunum@n@axis_breakdown.m2.cascade_asr.axis2.2.3\endcsname{531}
\expandafter\gdef\csname odunum@ci@axis_breakdown.m2.cascade_asr.axis2.2.3\endcsname{[0.667, 0.710]}
\expandafter\gdef\csname odunum@val@axis_breakdown.m2.cascade_asr.axis2.2.4\endcsname{0.504}
\expandafter\gdef\csname odunum@n@axis_breakdown.m2.cascade_asr.axis2.2.4\endcsname{261}
\expandafter\gdef\csname odunum@ci@axis_breakdown.m2.cascade_asr.axis2.2.4\endcsname{[0.469, 0.540]}
\expandafter\gdef\csname odunum@val@axis_breakdown.m2.cascade_asr.axis2.2.5\endcsname{0.517}
\expandafter\gdef\csname odunum@n@axis_breakdown.m2.cascade_asr.axis2.2.5\endcsname{256}
\expandafter\gdef\csname odunum@ci@axis_breakdown.m2.cascade_asr.axis2.2.5\endcsname{[0.477, 0.556]}
\expandafter\gdef\csname odunum@val@axis_breakdown.m2.cascade_asr.axis3.3.1\endcsname{0.591}
\expandafter\gdef\csname odunum@n@axis_breakdown.m2.cascade_asr.axis3.3.1\endcsname{109}
\expandafter\gdef\csname odunum@ci@axis_breakdown.m2.cascade_asr.axis3.3.1\endcsname{[0.538, 0.644]}
\expandafter\gdef\csname odunum@val@axis_breakdown.m2.cascade_asr.axis3.3.2\endcsname{0.576}
\expandafter\gdef\csname odunum@n@axis_breakdown.m2.cascade_asr.axis3.3.2\endcsname{116}
\expandafter\gdef\csname odunum@ci@axis_breakdown.m2.cascade_asr.axis3.3.2\endcsname{[0.528, 0.624]}
\expandafter\gdef\csname odunum@val@axis_breakdown.m2.cascade_asr.axis3.3.3\endcsname{0.471}
\expandafter\gdef\csname odunum@n@axis_breakdown.m2.cascade_asr.axis3.3.3\endcsname{207}
\expandafter\gdef\csname odunum@ci@axis_breakdown.m2.cascade_asr.axis3.3.3\endcsname{[0.424, 0.518]}
\expandafter\gdef\csname odunum@val@axis_breakdown.m2.cascade_asr.axis3.3.4\endcsname{0.643}
\expandafter\gdef\csname odunum@n@axis_breakdown.m2.cascade_asr.axis3.3.4\endcsname{404}
\expandafter\gdef\csname odunum@ci@axis_breakdown.m2.cascade_asr.axis3.3.4\endcsname{[0.615, 0.670]}
\expandafter\gdef\csname odunum@val@axis_breakdown.m2.cascade_asr.axis3.3.5\endcsname{0.507}
\expandafter\gdef\csname odunum@n@axis_breakdown.m2.cascade_asr.axis3.3.5\endcsname{294}
\expandafter\gdef\csname odunum@ci@axis_breakdown.m2.cascade_asr.axis3.3.5\endcsname{[0.475, 0.541]}
\expandafter\gdef\csname odunum@val@axis_breakdown.m2.cascade_asr.axis3.3.6\endcsname{0.681}
\expandafter\gdef\csname odunum@n@axis_breakdown.m2.cascade_asr.axis3.3.6\endcsname{501}
\expandafter\gdef\csname odunum@ci@axis_breakdown.m2.cascade_asr.axis3.3.6\endcsname{[0.659, 0.702]}
\expandafter\gdef\csname odunum@val@axis_breakdown.m2.cascade_asr.axis4.4.1\endcsname{0.615}
\expandafter\gdef\csname odunum@n@axis_breakdown.m2.cascade_asr.axis4.4.1\endcsname{202}
\expandafter\gdef\csname odunum@ci@axis_breakdown.m2.cascade_asr.axis4.4.1\endcsname{[0.577, 0.652]}
\expandafter\gdef\csname odunum@val@axis_breakdown.m2.cascade_asr.axis4.4.2\endcsname{0.611}
\expandafter\gdef\csname odunum@n@axis_breakdown.m2.cascade_asr.axis4.4.2\endcsname{457}
\expandafter\gdef\csname odunum@ci@axis_breakdown.m2.cascade_asr.axis4.4.2\endcsname{[0.585, 0.636]}
\expandafter\gdef\csname odunum@val@axis_breakdown.m2.cascade_asr.axis4.4.3\endcsname{0.547}
\expandafter\gdef\csname odunum@n@axis_breakdown.m2.cascade_asr.axis4.4.3\endcsname{164}
\expandafter\gdef\csname odunum@ci@axis_breakdown.m2.cascade_asr.axis4.4.3\endcsname{[0.497, 0.596]}
\expandafter\gdef\csname odunum@val@axis_breakdown.m2.cascade_asr.axis4.4.4\endcsname{0.582}
\expandafter\gdef\csname odunum@n@axis_breakdown.m2.cascade_asr.axis4.4.4\endcsname{505}
\expandafter\gdef\csname odunum@ci@axis_breakdown.m2.cascade_asr.axis4.4.4\endcsname{[0.556, 0.607]}
\expandafter\gdef\csname odunum@val@axis_breakdown.m2.cascade_asr.axis4.4.5\endcsname{0.633}
\expandafter\gdef\csname odunum@n@axis_breakdown.m2.cascade_asr.axis4.4.5\endcsname{303}
\expandafter\gdef\csname odunum@ci@axis_breakdown.m2.cascade_asr.axis4.4.5\endcsname{[0.601, 0.665]}
\expandafter\gdef\csname odunum@val@axis_breakdown.m2.cascade_asr.axis5.audio_only.5.1\endcsname{0.724}
\expandafter\gdef\csname odunum@n@axis_breakdown.m2.cascade_asr.axis5.audio_only.5.1\endcsname{45}
\expandafter\gdef\csname odunum@ci@axis_breakdown.m2.cascade_asr.axis5.audio_only.5.1\endcsname{[0.631, 0.810]}
\expandafter\gdef\csname odunum@val@axis_breakdown.m2.cascade_asr.axis5.audio_only.5.10\endcsname{0.707}
\expandafter\gdef\csname odunum@n@axis_breakdown.m2.cascade_asr.axis5.audio_only.5.10\endcsname{72}
\expandafter\gdef\csname odunum@ci@axis_breakdown.m2.cascade_asr.axis5.audio_only.5.10\endcsname{[0.650, 0.764]}
\expandafter\gdef\csname odunum@val@axis_breakdown.m2.cascade_asr.axis5.audio_only.5.2\endcsname{0.651}
\expandafter\gdef\csname odunum@n@axis_breakdown.m2.cascade_asr.axis5.audio_only.5.2\endcsname{75}
\expandafter\gdef\csname odunum@ci@axis_breakdown.m2.cascade_asr.axis5.audio_only.5.2\endcsname{[0.585, 0.714]}
\expandafter\gdef\csname odunum@val@axis_breakdown.m2.cascade_asr.axis5.audio_only.5.3\endcsname{0.767}
\expandafter\gdef\csname odunum@n@axis_breakdown.m2.cascade_asr.axis5.audio_only.5.3\endcsname{50}
\expandafter\gdef\csname odunum@ci@axis_breakdown.m2.cascade_asr.axis5.audio_only.5.3\endcsname{[0.695, 0.834]}
\expandafter\gdef\csname odunum@val@axis_breakdown.m2.cascade_asr.axis5.audio_only.5.4\endcsname{0.683}
\expandafter\gdef\csname odunum@n@axis_breakdown.m2.cascade_asr.axis5.audio_only.5.4\endcsname{68}
\expandafter\gdef\csname odunum@ci@axis_breakdown.m2.cascade_asr.axis5.audio_only.5.4\endcsname{[0.614, 0.749]}
\expandafter\gdef\csname odunum@val@axis_breakdown.m2.cascade_asr.axis5.audio_only.5.5\endcsname{0.720}
\expandafter\gdef\csname odunum@n@axis_breakdown.m2.cascade_asr.axis5.audio_only.5.5\endcsname{52}
\expandafter\gdef\csname odunum@ci@axis_breakdown.m2.cascade_asr.axis5.audio_only.5.5\endcsname{[0.647, 0.789]}
\expandafter\gdef\csname odunum@val@axis_breakdown.m2.cascade_asr.axis5.audio_only.5.6\endcsname{0.678}
\expandafter\gdef\csname odunum@n@axis_breakdown.m2.cascade_asr.axis5.audio_only.5.6\endcsname{52}
\expandafter\gdef\csname odunum@ci@axis_breakdown.m2.cascade_asr.axis5.audio_only.5.6\endcsname{[0.607, 0.748]}
\expandafter\gdef\csname odunum@val@axis_breakdown.m2.cascade_asr.axis5.audio_only.5.7\endcsname{0.630}
\expandafter\gdef\csname odunum@n@axis_breakdown.m2.cascade_asr.axis5.audio_only.5.7\endcsname{46}
\expandafter\gdef\csname odunum@ci@axis_breakdown.m2.cascade_asr.axis5.audio_only.5.7\endcsname{[0.543, 0.710]}
\expandafter\gdef\csname odunum@val@axis_breakdown.m2.cascade_asr.axis5.audio_only.5.8\endcsname{0.568}
\expandafter\gdef\csname odunum@n@axis_breakdown.m2.cascade_asr.axis5.audio_only.5.8\endcsname{57}
\expandafter\gdef\csname odunum@ci@axis_breakdown.m2.cascade_asr.axis5.audio_only.5.8\endcsname{[0.470, 0.666]}
\expandafter\gdef\csname odunum@val@axis_breakdown.m2.cascade_asr.axis5.audio_only.5.9\endcsname{0.725}
\expandafter\gdef\csname odunum@n@axis_breakdown.m2.cascade_asr.axis5.audio_only.5.9\endcsname{53}
\expandafter\gdef\csname odunum@ci@axis_breakdown.m2.cascade_asr.axis5.audio_only.5.9\endcsname{[0.662, 0.788]}
\expandafter\gdef\csname odunum@val@axis_breakdown.m2.cascade_asr.axis5.audio_visual.5.1\endcsname{0.577}
\expandafter\gdef\csname odunum@n@axis_breakdown.m2.cascade_asr.axis5.audio_visual.5.1\endcsname{114}
\expandafter\gdef\csname odunum@ci@axis_breakdown.m2.cascade_asr.axis5.audio_visual.5.1\endcsname{[0.522, 0.631]}
\expandafter\gdef\csname odunum@val@axis_breakdown.m2.cascade_asr.axis5.audio_visual.5.10\endcsname{0.593}
\expandafter\gdef\csname odunum@n@axis_breakdown.m2.cascade_asr.axis5.audio_visual.5.10\endcsname{123}
\expandafter\gdef\csname odunum@ci@axis_breakdown.m2.cascade_asr.axis5.audio_visual.5.10\endcsname{[0.543, 0.641]}
\expandafter\gdef\csname odunum@val@axis_breakdown.m2.cascade_asr.axis5.audio_visual.5.2\endcsname{0.454}
\expandafter\gdef\csname odunum@n@axis_breakdown.m2.cascade_asr.axis5.audio_visual.5.2\endcsname{126}
\expandafter\gdef\csname odunum@ci@axis_breakdown.m2.cascade_asr.axis5.audio_visual.5.2\endcsname{[0.405, 0.503]}
\expandafter\gdef\csname odunum@val@axis_breakdown.m2.cascade_asr.axis5.audio_visual.5.3\endcsname{0.573}
\expandafter\gdef\csname odunum@n@axis_breakdown.m2.cascade_asr.axis5.audio_visual.5.3\endcsname{100}
\expandafter\gdef\csname odunum@ci@axis_breakdown.m2.cascade_asr.axis5.audio_visual.5.3\endcsname{[0.524, 0.622]}
\expandafter\gdef\csname odunum@val@axis_breakdown.m2.cascade_asr.axis5.audio_visual.5.4\endcsname{0.534}
\expandafter\gdef\csname odunum@n@axis_breakdown.m2.cascade_asr.axis5.audio_visual.5.4\endcsname{97}
\expandafter\gdef\csname odunum@ci@axis_breakdown.m2.cascade_asr.axis5.audio_visual.5.4\endcsname{[0.479, 0.591]}
\expandafter\gdef\csname odunum@val@axis_breakdown.m2.cascade_asr.axis5.audio_visual.5.5\endcsname{0.552}
\expandafter\gdef\csname odunum@n@axis_breakdown.m2.cascade_asr.axis5.audio_visual.5.5\endcsname{99}
\expandafter\gdef\csname odunum@ci@axis_breakdown.m2.cascade_asr.axis5.audio_visual.5.5\endcsname{[0.492, 0.611]}
\expandafter\gdef\csname odunum@val@axis_breakdown.m2.cascade_asr.axis5.audio_visual.5.6\endcsname{0.594}
\expandafter\gdef\csname odunum@n@axis_breakdown.m2.cascade_asr.axis5.audio_visual.5.6\endcsname{99}
\expandafter\gdef\csname odunum@ci@axis_breakdown.m2.cascade_asr.axis5.audio_visual.5.6\endcsname{[0.547, 0.640]}
\expandafter\gdef\csname odunum@val@axis_breakdown.m2.cascade_asr.axis5.audio_visual.5.7\endcsname{0.572}
\expandafter\gdef\csname odunum@n@axis_breakdown.m2.cascade_asr.axis5.audio_visual.5.7\endcsname{104}
\expandafter\gdef\csname odunum@ci@axis_breakdown.m2.cascade_asr.axis5.audio_visual.5.7\endcsname{[0.517, 0.626]}
\expandafter\gdef\csname odunum@val@axis_breakdown.m2.cascade_asr.axis5.audio_visual.5.8\endcsname{0.537}
\expandafter\gdef\csname odunum@n@axis_breakdown.m2.cascade_asr.axis5.audio_visual.5.8\endcsname{121}
\expandafter\gdef\csname odunum@ci@axis_breakdown.m2.cascade_asr.axis5.audio_visual.5.8\endcsname{[0.485, 0.590]}
\expandafter\gdef\csname odunum@val@axis_breakdown.m2.cascade_asr.axis5.audio_visual.5.9\endcsname{0.592}
\expandafter\gdef\csname odunum@n@axis_breakdown.m2.cascade_asr.axis5.audio_visual.5.9\endcsname{78}
\expandafter\gdef\csname odunum@ci@axis_breakdown.m2.cascade_asr.axis5.audio_visual.5.9\endcsname{[0.526, 0.657]}
\expandafter\gdef\csname odunum@val@axis_breakdown.m2.cascade_asr.axis6.6.1\endcsname{0.579}
\expandafter\gdef\csname odunum@n@axis_breakdown.m2.cascade_asr.axis6.6.1\endcsname{230}
\expandafter\gdef\csname odunum@ci@axis_breakdown.m2.cascade_asr.axis6.6.1\endcsname{[0.542, 0.617]}
\expandafter\gdef\csname odunum@val@axis_breakdown.m2.cascade_asr.axis6.6.10\endcsname{0.576}
\expandafter\gdef\csname odunum@n@axis_breakdown.m2.cascade_asr.axis6.6.10\endcsname{102}
\expandafter\gdef\csname odunum@ci@axis_breakdown.m2.cascade_asr.axis6.6.10\endcsname{[0.521, 0.629]}
\expandafter\gdef\csname odunum@val@axis_breakdown.m2.cascade_asr.axis6.6.12\endcsname{0.635}
\expandafter\gdef\csname odunum@n@axis_breakdown.m2.cascade_asr.axis6.6.12\endcsname{129}
\expandafter\gdef\csname odunum@ci@axis_breakdown.m2.cascade_asr.axis6.6.12\endcsname{[0.587, 0.683]}
\expandafter\gdef\csname odunum@val@axis_breakdown.m2.cascade_asr.axis6.6.13\endcsname{0.526}
\expandafter\gdef\csname odunum@n@axis_breakdown.m2.cascade_asr.axis6.6.13\endcsname{58}
\expandafter\gdef\csname odunum@ci@axis_breakdown.m2.cascade_asr.axis6.6.13\endcsname{[0.444, 0.607]}
\expandafter\gdef\csname odunum@val@axis_breakdown.m2.cascade_asr.axis6.6.14\endcsname{0.482}
\expandafter\gdef\csname odunum@n@axis_breakdown.m2.cascade_asr.axis6.6.14\endcsname{82}
\expandafter\gdef\csname odunum@ci@axis_breakdown.m2.cascade_asr.axis6.6.14\endcsname{[0.412, 0.549]}
\expandafter\gdef\csname odunum@val@axis_breakdown.m2.cascade_asr.axis6.6.18\endcsname{0.562}
\expandafter\gdef\csname odunum@n@axis_breakdown.m2.cascade_asr.axis6.6.18\endcsname{72}
\expandafter\gdef\csname odunum@ci@axis_breakdown.m2.cascade_asr.axis6.6.18\endcsname{[0.493, 0.630]}
\expandafter\gdef\csname odunum@val@axis_breakdown.m2.cascade_asr.axis6.6.19\endcsname{0.539}
\expandafter\gdef\csname odunum@n@axis_breakdown.m2.cascade_asr.axis6.6.19\endcsname{44}
\expandafter\gdef\csname odunum@ci@axis_breakdown.m2.cascade_asr.axis6.6.19\endcsname{[0.454, 0.622]}
\expandafter\gdef\csname odunum@val@axis_breakdown.m2.cascade_asr.axis6.6.2\endcsname{0.691}
\expandafter\gdef\csname odunum@n@axis_breakdown.m2.cascade_asr.axis6.6.2\endcsname{41}
\expandafter\gdef\csname odunum@ci@axis_breakdown.m2.cascade_asr.axis6.6.2\endcsname{[0.614, 0.766]}
\expandafter\gdef\csname odunum@val@axis_breakdown.m2.cascade_asr.axis6.6.3\endcsname{0.551}
\expandafter\gdef\csname odunum@n@axis_breakdown.m2.cascade_asr.axis6.6.3\endcsname{44}
\expandafter\gdef\csname odunum@ci@axis_breakdown.m2.cascade_asr.axis6.6.3\endcsname{[0.462, 0.638]}
\expandafter\gdef\csname odunum@val@axis_breakdown.m2.cascade_asr.axis6.6.5\endcsname{0.601}
\expandafter\gdef\csname odunum@n@axis_breakdown.m2.cascade_asr.axis6.6.5\endcsname{48}
\expandafter\gdef\csname odunum@ci@axis_breakdown.m2.cascade_asr.axis6.6.5\endcsname{[0.522, 0.680]}
\expandafter\gdef\csname odunum@val@axis_breakdown.m2.cascade_asr.axis6.6.6\endcsname{0.642}
\expandafter\gdef\csname odunum@n@axis_breakdown.m2.cascade_asr.axis6.6.6\endcsname{134}
\expandafter\gdef\csname odunum@ci@axis_breakdown.m2.cascade_asr.axis6.6.6\endcsname{[0.591, 0.692]}
\expandafter\gdef\csname odunum@val@axis_breakdown.m2.cascade_asr.axis6.6.7\endcsname{0.659}
\expandafter\gdef\csname odunum@n@axis_breakdown.m2.cascade_asr.axis6.6.7\endcsname{393}
\expandafter\gdef\csname odunum@ci@axis_breakdown.m2.cascade_asr.axis6.6.7\endcsname{[0.631, 0.686]}
\expandafter\gdef\csname odunum@val@axis_breakdown.m2.cascade_asr.axis6.6.8\endcsname{0.568}
\expandafter\gdef\csname odunum@n@axis_breakdown.m2.cascade_asr.axis6.6.8\endcsname{49}
\expandafter\gdef\csname odunum@ci@axis_breakdown.m2.cascade_asr.axis6.6.8\endcsname{[0.480, 0.652]}
\expandafter\gdef\csname odunum@val@axis_breakdown.m2.cascade_asr.axis6.6.9\endcsname{0.575}
\expandafter\gdef\csname odunum@n@axis_breakdown.m2.cascade_asr.axis6.6.9\endcsname{119}
\expandafter\gdef\csname odunum@ci@axis_breakdown.m2.cascade_asr.axis6.6.9\endcsname{[0.521, 0.627]}
\expandafter\gdef\csname odunum@val@axis_breakdown.m2.cascade_asr.difficulty.q1\endcsname{0.657}
\expandafter\gdef\csname odunum@n@axis_breakdown.m2.cascade_asr.difficulty.q1\endcsname{358}
\expandafter\gdef\csname odunum@ci@axis_breakdown.m2.cascade_asr.difficulty.q1\endcsname{[0.625, 0.689]}
\expandafter\gdef\csname odunum@val@axis_breakdown.m2.cascade_asr.difficulty.q2\endcsname{0.613}
\expandafter\gdef\csname odunum@n@axis_breakdown.m2.cascade_asr.difficulty.q2\endcsname{357}
\expandafter\gdef\csname odunum@ci@axis_breakdown.m2.cascade_asr.difficulty.q2\endcsname{[0.584, 0.641]}
\expandafter\gdef\csname odunum@val@axis_breakdown.m2.cascade_asr.difficulty.q3\endcsname{0.602}
\expandafter\gdef\csname odunum@n@axis_breakdown.m2.cascade_asr.difficulty.q3\endcsname{358}
\expandafter\gdef\csname odunum@ci@axis_breakdown.m2.cascade_asr.difficulty.q3\endcsname{[0.574, 0.632]}
\expandafter\gdef\csname odunum@val@axis_breakdown.m2.cascade_asr.difficulty.q4\endcsname{0.577}
\expandafter\gdef\csname odunum@n@axis_breakdown.m2.cascade_asr.difficulty.q4\endcsname{356}
\expandafter\gdef\csname odunum@ci@axis_breakdown.m2.cascade_asr.difficulty.q4\endcsname{[0.546, 0.606]}
\expandafter\gdef\csname odunum@val@axis_breakdown.m2.gemini.axis1.audio_only.1.1\endcsname{0.710}
\expandafter\gdef\csname odunum@n@axis_breakdown.m2.gemini.axis1.audio_only.1.1\endcsname{158}
\expandafter\gdef\csname odunum@ci@axis_breakdown.m2.gemini.axis1.audio_only.1.1\endcsname{[0.670, 0.749]}
\expandafter\gdef\csname odunum@val@axis_breakdown.m2.gemini.axis1.audio_only.1.2\endcsname{0.705}
\expandafter\gdef\csname odunum@n@axis_breakdown.m2.gemini.axis1.audio_only.1.2\endcsname{379}
\expandafter\gdef\csname odunum@ci@axis_breakdown.m2.gemini.axis1.audio_only.1.2\endcsname{[0.679, 0.730]}
\expandafter\gdef\csname odunum@val@axis_breakdown.m2.gemini.axis1.audio_visual.1.1\endcsname{0.625}
\expandafter\gdef\csname odunum@n@axis_breakdown.m2.gemini.axis1.audio_visual.1.1\endcsname{94}
\expandafter\gdef\csname odunum@ci@axis_breakdown.m2.gemini.axis1.audio_visual.1.1\endcsname{[0.573, 0.679]}
\expandafter\gdef\csname odunum@val@axis_breakdown.m2.gemini.axis1.audio_visual.1.2\endcsname{0.616}
\expandafter\gdef\csname odunum@n@axis_breakdown.m2.gemini.axis1.audio_visual.1.2\endcsname{294}
\expandafter\gdef\csname odunum@ci@axis_breakdown.m2.gemini.axis1.audio_visual.1.2\endcsname{[0.587, 0.644]}
\expandafter\gdef\csname odunum@val@axis_breakdown.m2.gemini.axis1.audio_visual.1.3\endcsname{0.654}
\expandafter\gdef\csname odunum@n@axis_breakdown.m2.gemini.axis1.audio_visual.1.3\endcsname{291}
\expandafter\gdef\csname odunum@ci@axis_breakdown.m2.gemini.axis1.audio_visual.1.3\endcsname{[0.626, 0.682]}
\expandafter\gdef\csname odunum@val@axis_breakdown.m2.gemini.axis1.audio_visual.1.4\endcsname{0.633}
\expandafter\gdef\csname odunum@n@axis_breakdown.m2.gemini.axis1.audio_visual.1.4\endcsname{382}
\expandafter\gdef\csname odunum@ci@axis_breakdown.m2.gemini.axis1.audio_visual.1.4\endcsname{[0.610, 0.657]}
\expandafter\gdef\csname odunum@val@axis_breakdown.m2.gemini.axis2.2.1\endcsname{0.632}
\expandafter\gdef\csname odunum@n@axis_breakdown.m2.gemini.axis2.2.1\endcsname{143}
\expandafter\gdef\csname odunum@ci@axis_breakdown.m2.gemini.axis2.2.1\endcsname{[0.583, 0.679]}
\expandafter\gdef\csname odunum@val@axis_breakdown.m2.gemini.axis2.2.2\endcsname{0.659}
\expandafter\gdef\csname odunum@n@axis_breakdown.m2.gemini.axis2.2.2\endcsname{440}
\expandafter\gdef\csname odunum@ci@axis_breakdown.m2.gemini.axis2.2.2\endcsname{[0.637, 0.682]}
\expandafter\gdef\csname odunum@val@axis_breakdown.m2.gemini.axis2.2.3\endcsname{0.668}
\expandafter\gdef\csname odunum@n@axis_breakdown.m2.gemini.axis2.2.3\endcsname{531}
\expandafter\gdef\csname odunum@ci@axis_breakdown.m2.gemini.axis2.2.3\endcsname{[0.648, 0.688]}
\expandafter\gdef\csname odunum@val@axis_breakdown.m2.gemini.axis2.2.4\endcsname{0.646}
\expandafter\gdef\csname odunum@n@axis_breakdown.m2.gemini.axis2.2.4\endcsname{261}
\expandafter\gdef\csname odunum@ci@axis_breakdown.m2.gemini.axis2.2.4\endcsname{[0.615, 0.678]}
\expandafter\gdef\csname odunum@val@axis_breakdown.m2.gemini.axis2.2.5\endcsname{0.655}
\expandafter\gdef\csname odunum@n@axis_breakdown.m2.gemini.axis2.2.5\endcsname{256}
\expandafter\gdef\csname odunum@ci@axis_breakdown.m2.gemini.axis2.2.5\endcsname{[0.623, 0.686]}
\expandafter\gdef\csname odunum@val@axis_breakdown.m2.gemini.axis3.3.1\endcsname{0.691}
\expandafter\gdef\csname odunum@n@axis_breakdown.m2.gemini.axis3.3.1\endcsname{109}
\expandafter\gdef\csname odunum@ci@axis_breakdown.m2.gemini.axis3.3.1\endcsname{[0.648, 0.733]}
\expandafter\gdef\csname odunum@val@axis_breakdown.m2.gemini.axis3.3.2\endcsname{0.638}
\expandafter\gdef\csname odunum@n@axis_breakdown.m2.gemini.axis3.3.2\endcsname{116}
\expandafter\gdef\csname odunum@ci@axis_breakdown.m2.gemini.axis3.3.2\endcsname{[0.596, 0.679]}
\expandafter\gdef\csname odunum@val@axis_breakdown.m2.gemini.axis3.3.3\endcsname{0.603}
\expandafter\gdef\csname odunum@n@axis_breakdown.m2.gemini.axis3.3.3\endcsname{207}
\expandafter\gdef\csname odunum@ci@axis_breakdown.m2.gemini.axis3.3.3\endcsname{[0.563, 0.643]}
\expandafter\gdef\csname odunum@val@axis_breakdown.m2.gemini.axis3.3.4\endcsname{0.656}
\expandafter\gdef\csname odunum@n@axis_breakdown.m2.gemini.axis3.3.4\endcsname{404}
\expandafter\gdef\csname odunum@ci@axis_breakdown.m2.gemini.axis3.3.4\endcsname{[0.631, 0.679]}
\expandafter\gdef\csname odunum@val@axis_breakdown.m2.gemini.axis3.3.5\endcsname{0.580}
\expandafter\gdef\csname odunum@n@axis_breakdown.m2.gemini.axis3.3.5\endcsname{294}
\expandafter\gdef\csname odunum@ci@axis_breakdown.m2.gemini.axis3.3.5\endcsname{[0.551, 0.610]}
\expandafter\gdef\csname odunum@val@axis_breakdown.m2.gemini.axis3.3.6\endcsname{0.723}
\expandafter\gdef\csname odunum@n@axis_breakdown.m2.gemini.axis3.3.6\endcsname{501}
\expandafter\gdef\csname odunum@ci@axis_breakdown.m2.gemini.axis3.3.6\endcsname{[0.704, 0.743]}
\expandafter\gdef\csname odunum@val@axis_breakdown.m2.gemini.axis4.4.1\endcsname{0.644}
\expandafter\gdef\csname odunum@n@axis_breakdown.m2.gemini.axis4.4.1\endcsname{202}
\expandafter\gdef\csname odunum@ci@axis_breakdown.m2.gemini.axis4.4.1\endcsname{[0.611, 0.675]}
\expandafter\gdef\csname odunum@val@axis_breakdown.m2.gemini.axis4.4.2\endcsname{0.655}
\expandafter\gdef\csname odunum@n@axis_breakdown.m2.gemini.axis4.4.2\endcsname{457}
\expandafter\gdef\csname odunum@ci@axis_breakdown.m2.gemini.axis4.4.2\endcsname{[0.633, 0.678]}
\expandafter\gdef\csname odunum@val@axis_breakdown.m2.gemini.axis4.4.3\endcsname{0.654}
\expandafter\gdef\csname odunum@n@axis_breakdown.m2.gemini.axis4.4.3\endcsname{164}
\expandafter\gdef\csname odunum@ci@axis_breakdown.m2.gemini.axis4.4.3\endcsname{[0.613, 0.694]}
\expandafter\gdef\csname odunum@val@axis_breakdown.m2.gemini.axis4.4.4\endcsname{0.668}
\expandafter\gdef\csname odunum@n@axis_breakdown.m2.gemini.axis4.4.4\endcsname{505}
\expandafter\gdef\csname odunum@ci@axis_breakdown.m2.gemini.axis4.4.4\endcsname{[0.647, 0.690]}
\expandafter\gdef\csname odunum@val@axis_breakdown.m2.gemini.axis4.4.5\endcsname{0.652}
\expandafter\gdef\csname odunum@n@axis_breakdown.m2.gemini.axis4.4.5\endcsname{303}
\expandafter\gdef\csname odunum@ci@axis_breakdown.m2.gemini.axis4.4.5\endcsname{[0.624, 0.680]}
\expandafter\gdef\csname odunum@val@axis_breakdown.m2.gemini.axis5.audio_only.5.1\endcsname{0.706}
\expandafter\gdef\csname odunum@n@axis_breakdown.m2.gemini.axis5.audio_only.5.1\endcsname{45}
\expandafter\gdef\csname odunum@ci@axis_breakdown.m2.gemini.axis5.audio_only.5.1\endcsname{[0.623, 0.785]}
\expandafter\gdef\csname odunum@val@axis_breakdown.m2.gemini.axis5.audio_only.5.10\endcsname{0.689}
\expandafter\gdef\csname odunum@n@axis_breakdown.m2.gemini.axis5.audio_only.5.10\endcsname{72}
\expandafter\gdef\csname odunum@ci@axis_breakdown.m2.gemini.axis5.audio_only.5.10\endcsname{[0.631, 0.747]}
\expandafter\gdef\csname odunum@val@axis_breakdown.m2.gemini.axis5.audio_only.5.2\endcsname{0.695}
\expandafter\gdef\csname odunum@n@axis_breakdown.m2.gemini.axis5.audio_only.5.2\endcsname{75}
\expandafter\gdef\csname odunum@ci@axis_breakdown.m2.gemini.axis5.audio_only.5.2\endcsname{[0.636, 0.752]}
\expandafter\gdef\csname odunum@val@axis_breakdown.m2.gemini.axis5.audio_only.5.3\endcsname{0.732}
\expandafter\gdef\csname odunum@n@axis_breakdown.m2.gemini.axis5.audio_only.5.3\endcsname{50}
\expandafter\gdef\csname odunum@ci@axis_breakdown.m2.gemini.axis5.audio_only.5.3\endcsname{[0.663, 0.798]}
\expandafter\gdef\csname odunum@val@axis_breakdown.m2.gemini.axis5.audio_only.5.4\endcsname{0.646}
\expandafter\gdef\csname odunum@n@axis_breakdown.m2.gemini.axis5.audio_only.5.4\endcsname{68}
\expandafter\gdef\csname odunum@ci@axis_breakdown.m2.gemini.axis5.audio_only.5.4\endcsname{[0.585, 0.707]}
\expandafter\gdef\csname odunum@val@axis_breakdown.m2.gemini.axis5.audio_only.5.5\endcsname{0.705}
\expandafter\gdef\csname odunum@n@axis_breakdown.m2.gemini.axis5.audio_only.5.5\endcsname{52}
\expandafter\gdef\csname odunum@ci@axis_breakdown.m2.gemini.axis5.audio_only.5.5\endcsname{[0.636, 0.768]}
\expandafter\gdef\csname odunum@val@axis_breakdown.m2.gemini.axis5.audio_only.5.6\endcsname{0.748}
\expandafter\gdef\csname odunum@n@axis_breakdown.m2.gemini.axis5.audio_only.5.6\endcsname{52}
\expandafter\gdef\csname odunum@ci@axis_breakdown.m2.gemini.axis5.audio_only.5.6\endcsname{[0.692, 0.805]}
\expandafter\gdef\csname odunum@val@axis_breakdown.m2.gemini.axis5.audio_only.5.7\endcsname{0.694}
\expandafter\gdef\csname odunum@n@axis_breakdown.m2.gemini.axis5.audio_only.5.7\endcsname{46}
\expandafter\gdef\csname odunum@ci@axis_breakdown.m2.gemini.axis5.audio_only.5.7\endcsname{[0.629, 0.759]}
\expandafter\gdef\csname odunum@val@axis_breakdown.m2.gemini.axis5.audio_only.5.8\endcsname{0.729}
\expandafter\gdef\csname odunum@n@axis_breakdown.m2.gemini.axis5.audio_only.5.8\endcsname{57}
\expandafter\gdef\csname odunum@ci@axis_breakdown.m2.gemini.axis5.audio_only.5.8\endcsname{[0.661, 0.794]}
\expandafter\gdef\csname odunum@val@axis_breakdown.m2.gemini.axis5.audio_only.5.9\endcsname{0.691}
\expandafter\gdef\csname odunum@n@axis_breakdown.m2.gemini.axis5.audio_only.5.9\endcsname{53}
\expandafter\gdef\csname odunum@ci@axis_breakdown.m2.gemini.axis5.audio_only.5.9\endcsname{[0.615, 0.764]}
\expandafter\gdef\csname odunum@val@axis_breakdown.m2.gemini.axis5.audio_visual.5.1\endcsname{0.650}
\expandafter\gdef\csname odunum@n@axis_breakdown.m2.gemini.axis5.audio_visual.5.1\endcsname{114}
\expandafter\gdef\csname odunum@ci@axis_breakdown.m2.gemini.axis5.audio_visual.5.1\endcsname{[0.601, 0.698]}
\expandafter\gdef\csname odunum@val@axis_breakdown.m2.gemini.axis5.audio_visual.5.10\endcsname{0.659}
\expandafter\gdef\csname odunum@n@axis_breakdown.m2.gemini.axis5.audio_visual.5.10\endcsname{123}
\expandafter\gdef\csname odunum@ci@axis_breakdown.m2.gemini.axis5.audio_visual.5.10\endcsname{[0.620, 0.698]}
\expandafter\gdef\csname odunum@val@axis_breakdown.m2.gemini.axis5.audio_visual.5.2\endcsname{0.595}
\expandafter\gdef\csname odunum@n@axis_breakdown.m2.gemini.axis5.audio_visual.5.2\endcsname{126}
\expandafter\gdef\csname odunum@ci@axis_breakdown.m2.gemini.axis5.audio_visual.5.2\endcsname{[0.551, 0.640]}
\expandafter\gdef\csname odunum@val@axis_breakdown.m2.gemini.axis5.audio_visual.5.3\endcsname{0.624}
\expandafter\gdef\csname odunum@n@axis_breakdown.m2.gemini.axis5.audio_visual.5.3\endcsname{100}
\expandafter\gdef\csname odunum@ci@axis_breakdown.m2.gemini.axis5.audio_visual.5.3\endcsname{[0.574, 0.672]}
\expandafter\gdef\csname odunum@val@axis_breakdown.m2.gemini.axis5.audio_visual.5.4\endcsname{0.599}
\expandafter\gdef\csname odunum@n@axis_breakdown.m2.gemini.axis5.audio_visual.5.4\endcsname{97}
\expandafter\gdef\csname odunum@ci@axis_breakdown.m2.gemini.axis5.audio_visual.5.4\endcsname{[0.550, 0.647]}
\expandafter\gdef\csname odunum@val@axis_breakdown.m2.gemini.axis5.audio_visual.5.5\endcsname{0.686}
\expandafter\gdef\csname odunum@n@axis_breakdown.m2.gemini.axis5.audio_visual.5.5\endcsname{99}
\expandafter\gdef\csname odunum@ci@axis_breakdown.m2.gemini.axis5.audio_visual.5.5\endcsname{[0.638, 0.733]}
\expandafter\gdef\csname odunum@val@axis_breakdown.m2.gemini.axis5.audio_visual.5.6\endcsname{0.669}
\expandafter\gdef\csname odunum@n@axis_breakdown.m2.gemini.axis5.audio_visual.5.6\endcsname{99}
\expandafter\gdef\csname odunum@ci@axis_breakdown.m2.gemini.axis5.audio_visual.5.6\endcsname{[0.622, 0.713]}
\expandafter\gdef\csname odunum@val@axis_breakdown.m2.gemini.axis5.audio_visual.5.7\endcsname{0.619}
\expandafter\gdef\csname odunum@n@axis_breakdown.m2.gemini.axis5.audio_visual.5.7\endcsname{104}
\expandafter\gdef\csname odunum@ci@axis_breakdown.m2.gemini.axis5.audio_visual.5.7\endcsname{[0.571, 0.668]}
\expandafter\gdef\csname odunum@val@axis_breakdown.m2.gemini.axis5.audio_visual.5.8\endcsname{0.597}
\expandafter\gdef\csname odunum@n@axis_breakdown.m2.gemini.axis5.audio_visual.5.8\endcsname{121}
\expandafter\gdef\csname odunum@ci@axis_breakdown.m2.gemini.axis5.audio_visual.5.8\endcsname{[0.557, 0.635]}
\expandafter\gdef\csname odunum@val@axis_breakdown.m2.gemini.axis5.audio_visual.5.9\endcsname{0.652}
\expandafter\gdef\csname odunum@n@axis_breakdown.m2.gemini.axis5.audio_visual.5.9\endcsname{78}
\expandafter\gdef\csname odunum@ci@axis_breakdown.m2.gemini.axis5.audio_visual.5.9\endcsname{[0.598, 0.706]}
\expandafter\gdef\csname odunum@val@axis_breakdown.m2.gemini.axis6.6.1\endcsname{0.641}
\expandafter\gdef\csname odunum@n@axis_breakdown.m2.gemini.axis6.6.1\endcsname{230}
\expandafter\gdef\csname odunum@ci@axis_breakdown.m2.gemini.axis6.6.1\endcsname{[0.610, 0.672]}
\expandafter\gdef\csname odunum@val@axis_breakdown.m2.gemini.axis6.6.10\endcsname{0.646}
\expandafter\gdef\csname odunum@n@axis_breakdown.m2.gemini.axis6.6.10\endcsname{102}
\expandafter\gdef\csname odunum@ci@axis_breakdown.m2.gemini.axis6.6.10\endcsname{[0.598, 0.694]}
\expandafter\gdef\csname odunum@val@axis_breakdown.m2.gemini.axis6.6.12\endcsname{0.645}
\expandafter\gdef\csname odunum@n@axis_breakdown.m2.gemini.axis6.6.12\endcsname{129}
\expandafter\gdef\csname odunum@ci@axis_breakdown.m2.gemini.axis6.6.12\endcsname{[0.601, 0.687]}
\expandafter\gdef\csname odunum@val@axis_breakdown.m2.gemini.axis6.6.13\endcsname{0.628}
\expandafter\gdef\csname odunum@n@axis_breakdown.m2.gemini.axis6.6.13\endcsname{58}
\expandafter\gdef\csname odunum@ci@axis_breakdown.m2.gemini.axis6.6.13\endcsname{[0.555, 0.698]}
\expandafter\gdef\csname odunum@val@axis_breakdown.m2.gemini.axis6.6.14\endcsname{0.620}
\expandafter\gdef\csname odunum@n@axis_breakdown.m2.gemini.axis6.6.14\endcsname{82}
\expandafter\gdef\csname odunum@ci@axis_breakdown.m2.gemini.axis6.6.14\endcsname{[0.565, 0.674]}
\expandafter\gdef\csname odunum@val@axis_breakdown.m2.gemini.axis6.6.18\endcsname{0.640}
\expandafter\gdef\csname odunum@n@axis_breakdown.m2.gemini.axis6.6.18\endcsname{72}
\expandafter\gdef\csname odunum@ci@axis_breakdown.m2.gemini.axis6.6.18\endcsname{[0.581, 0.698]}
\expandafter\gdef\csname odunum@val@axis_breakdown.m2.gemini.axis6.6.19\endcsname{0.591}
\expandafter\gdef\csname odunum@n@axis_breakdown.m2.gemini.axis6.6.19\endcsname{44}
\expandafter\gdef\csname odunum@ci@axis_breakdown.m2.gemini.axis6.6.19\endcsname{[0.506, 0.673]}
\expandafter\gdef\csname odunum@val@axis_breakdown.m2.gemini.axis6.6.2\endcsname{0.692}
\expandafter\gdef\csname odunum@n@axis_breakdown.m2.gemini.axis6.6.2\endcsname{41}
\expandafter\gdef\csname odunum@ci@axis_breakdown.m2.gemini.axis6.6.2\endcsname{[0.611, 0.770]}
\expandafter\gdef\csname odunum@val@axis_breakdown.m2.gemini.axis6.6.3\endcsname{0.718}
\expandafter\gdef\csname odunum@n@axis_breakdown.m2.gemini.axis6.6.3\endcsname{44}
\expandafter\gdef\csname odunum@ci@axis_breakdown.m2.gemini.axis6.6.3\endcsname{[0.642, 0.790]}
\expandafter\gdef\csname odunum@val@axis_breakdown.m2.gemini.axis6.6.5\endcsname{0.606}
\expandafter\gdef\csname odunum@n@axis_breakdown.m2.gemini.axis6.6.5\endcsname{48}
\expandafter\gdef\csname odunum@ci@axis_breakdown.m2.gemini.axis6.6.5\endcsname{[0.531, 0.677]}
\expandafter\gdef\csname odunum@val@axis_breakdown.m2.gemini.axis6.6.6\endcsname{0.680}
\expandafter\gdef\csname odunum@n@axis_breakdown.m2.gemini.axis6.6.6\endcsname{134}
\expandafter\gdef\csname odunum@ci@axis_breakdown.m2.gemini.axis6.6.6\endcsname{[0.639, 0.722]}
\expandafter\gdef\csname odunum@val@axis_breakdown.m2.gemini.axis6.6.7\endcsname{0.698}
\expandafter\gdef\csname odunum@n@axis_breakdown.m2.gemini.axis6.6.7\endcsname{393}
\expandafter\gdef\csname odunum@ci@axis_breakdown.m2.gemini.axis6.6.7\endcsname{[0.674, 0.721]}
\expandafter\gdef\csname odunum@val@axis_breakdown.m2.gemini.axis6.6.8\endcsname{0.603}
\expandafter\gdef\csname odunum@n@axis_breakdown.m2.gemini.axis6.6.8\endcsname{49}
\expandafter\gdef\csname odunum@ci@axis_breakdown.m2.gemini.axis6.6.8\endcsname{[0.537, 0.668]}
\expandafter\gdef\csname odunum@val@axis_breakdown.m2.gemini.axis6.6.9\endcsname{0.642}
\expandafter\gdef\csname odunum@n@axis_breakdown.m2.gemini.axis6.6.9\endcsname{119}
\expandafter\gdef\csname odunum@ci@axis_breakdown.m2.gemini.axis6.6.9\endcsname{[0.597, 0.686]}
\expandafter\gdef\csname odunum@val@axis_breakdown.m2.gemini.difficulty.q1\endcsname{0.692}
\expandafter\gdef\csname odunum@n@axis_breakdown.m2.gemini.difficulty.q1\endcsname{358}
\expandafter\gdef\csname odunum@ci@axis_breakdown.m2.gemini.difficulty.q1\endcsname{[0.664, 0.719]}
\expandafter\gdef\csname odunum@val@axis_breakdown.m2.gemini.difficulty.q2\endcsname{0.650}
\expandafter\gdef\csname odunum@n@axis_breakdown.m2.gemini.difficulty.q2\endcsname{357}
\expandafter\gdef\csname odunum@ci@axis_breakdown.m2.gemini.difficulty.q2\endcsname{[0.625, 0.675]}
\expandafter\gdef\csname odunum@val@axis_breakdown.m2.gemini.difficulty.q3\endcsname{0.656}
\expandafter\gdef\csname odunum@n@axis_breakdown.m2.gemini.difficulty.q3\endcsname{358}
\expandafter\gdef\csname odunum@ci@axis_breakdown.m2.gemini.difficulty.q3\endcsname{[0.631, 0.681]}
\expandafter\gdef\csname odunum@val@axis_breakdown.m2.gemini.difficulty.q4\endcsname{0.633}
\expandafter\gdef\csname odunum@n@axis_breakdown.m2.gemini.difficulty.q4\endcsname{356}
\expandafter\gdef\csname odunum@ci@axis_breakdown.m2.gemini.difficulty.q4\endcsname{[0.607, 0.658]}
\expandafter\gdef\csname odunum@val@axis_breakdown.m2.gemini35_flash_lite.axis1.audio_only.1.1\endcsname{0.650}
\expandafter\gdef\csname odunum@n@axis_breakdown.m2.gemini35_flash_lite.axis1.audio_only.1.1\endcsname{158}
\expandafter\gdef\csname odunum@ci@axis_breakdown.m2.gemini35_flash_lite.axis1.audio_only.1.1\endcsname{[0.600, 0.699]}
\expandafter\gdef\csname odunum@val@axis_breakdown.m2.gemini35_flash_lite.axis1.audio_only.1.2\endcsname{0.524}
\expandafter\gdef\csname odunum@n@axis_breakdown.m2.gemini35_flash_lite.axis1.audio_only.1.2\endcsname{379}
\expandafter\gdef\csname odunum@ci@axis_breakdown.m2.gemini35_flash_lite.axis1.audio_only.1.2\endcsname{[0.491, 0.556]}
\expandafter\gdef\csname odunum@val@axis_breakdown.m2.gemini35_flash_lite.axis1.audio_visual.1.1\endcsname{0.498}
\expandafter\gdef\csname odunum@n@axis_breakdown.m2.gemini35_flash_lite.axis1.audio_visual.1.1\endcsname{94}
\expandafter\gdef\csname odunum@ci@axis_breakdown.m2.gemini35_flash_lite.axis1.audio_visual.1.1\endcsname{[0.434, 0.562]}
\expandafter\gdef\csname odunum@val@axis_breakdown.m2.gemini35_flash_lite.axis1.audio_visual.1.2\endcsname{0.490}
\expandafter\gdef\csname odunum@n@axis_breakdown.m2.gemini35_flash_lite.axis1.audio_visual.1.2\endcsname{294}
\expandafter\gdef\csname odunum@ci@axis_breakdown.m2.gemini35_flash_lite.axis1.audio_visual.1.2\endcsname{[0.452, 0.526]}
\expandafter\gdef\csname odunum@val@axis_breakdown.m2.gemini35_flash_lite.axis1.audio_visual.1.3\endcsname{0.507}
\expandafter\gdef\csname odunum@n@axis_breakdown.m2.gemini35_flash_lite.axis1.audio_visual.1.3\endcsname{291}
\expandafter\gdef\csname odunum@ci@axis_breakdown.m2.gemini35_flash_lite.axis1.audio_visual.1.3\endcsname{[0.468, 0.544]}
\expandafter\gdef\csname odunum@val@axis_breakdown.m2.gemini35_flash_lite.axis1.audio_visual.1.4\endcsname{0.500}
\expandafter\gdef\csname odunum@n@axis_breakdown.m2.gemini35_flash_lite.axis1.audio_visual.1.4\endcsname{382}
\expandafter\gdef\csname odunum@ci@axis_breakdown.m2.gemini35_flash_lite.axis1.audio_visual.1.4\endcsname{[0.470, 0.532]}
\expandafter\gdef\csname odunum@val@axis_breakdown.m2.gemini35_flash_lite.axis2.2.1\endcsname{0.419}
\expandafter\gdef\csname odunum@n@axis_breakdown.m2.gemini35_flash_lite.axis2.2.1\endcsname{143}
\expandafter\gdef\csname odunum@ci@axis_breakdown.m2.gemini35_flash_lite.axis2.2.1\endcsname{[0.357, 0.479]}
\expandafter\gdef\csname odunum@val@axis_breakdown.m2.gemini35_flash_lite.axis2.2.2\endcsname{0.572}
\expandafter\gdef\csname odunum@n@axis_breakdown.m2.gemini35_flash_lite.axis2.2.2\endcsname{440}
\expandafter\gdef\csname odunum@ci@axis_breakdown.m2.gemini35_flash_lite.axis2.2.2\endcsname{[0.543, 0.600]}
\expandafter\gdef\csname odunum@val@axis_breakdown.m2.gemini35_flash_lite.axis2.2.3\endcsname{0.618}
\expandafter\gdef\csname odunum@n@axis_breakdown.m2.gemini35_flash_lite.axis2.2.3\endcsname{531}
\expandafter\gdef\csname odunum@ci@axis_breakdown.m2.gemini35_flash_lite.axis2.2.3\endcsname{[0.594, 0.641]}
\expandafter\gdef\csname odunum@val@axis_breakdown.m2.gemini35_flash_lite.axis2.2.4\endcsname{0.400}
\expandafter\gdef\csname odunum@n@axis_breakdown.m2.gemini35_flash_lite.axis2.2.4\endcsname{261}
\expandafter\gdef\csname odunum@ci@axis_breakdown.m2.gemini35_flash_lite.axis2.2.4\endcsname{[0.360, 0.441]}
\expandafter\gdef\csname odunum@val@axis_breakdown.m2.gemini35_flash_lite.axis2.2.5\endcsname{0.419}
\expandafter\gdef\csname odunum@n@axis_breakdown.m2.gemini35_flash_lite.axis2.2.5\endcsname{256}
\expandafter\gdef\csname odunum@ci@axis_breakdown.m2.gemini35_flash_lite.axis2.2.5\endcsname{[0.376, 0.459]}
\expandafter\gdef\csname odunum@val@axis_breakdown.m2.gemini35_flash_lite.axis3.3.1\endcsname{0.556}
\expandafter\gdef\csname odunum@n@axis_breakdown.m2.gemini35_flash_lite.axis3.3.1\endcsname{109}
\expandafter\gdef\csname odunum@ci@axis_breakdown.m2.gemini35_flash_lite.axis3.3.1\endcsname{[0.501, 0.609]}
\expandafter\gdef\csname odunum@val@axis_breakdown.m2.gemini35_flash_lite.axis3.3.2\endcsname{0.544}
\expandafter\gdef\csname odunum@n@axis_breakdown.m2.gemini35_flash_lite.axis3.3.2\endcsname{116}
\expandafter\gdef\csname odunum@ci@axis_breakdown.m2.gemini35_flash_lite.axis3.3.2\endcsname{[0.492, 0.595]}
\expandafter\gdef\csname odunum@val@axis_breakdown.m2.gemini35_flash_lite.axis3.3.3\endcsname{0.343}
\expandafter\gdef\csname odunum@n@axis_breakdown.m2.gemini35_flash_lite.axis3.3.3\endcsname{207}
\expandafter\gdef\csname odunum@ci@axis_breakdown.m2.gemini35_flash_lite.axis3.3.3\endcsname{[0.294, 0.394]}
\expandafter\gdef\csname odunum@val@axis_breakdown.m2.gemini35_flash_lite.axis3.3.4\endcsname{0.536}
\expandafter\gdef\csname odunum@n@axis_breakdown.m2.gemini35_flash_lite.axis3.3.4\endcsname{404}
\expandafter\gdef\csname odunum@ci@axis_breakdown.m2.gemini35_flash_lite.axis3.3.4\endcsname{[0.506, 0.566]}
\expandafter\gdef\csname odunum@val@axis_breakdown.m2.gemini35_flash_lite.axis3.3.5\endcsname{0.492}
\expandafter\gdef\csname odunum@n@axis_breakdown.m2.gemini35_flash_lite.axis3.3.5\endcsname{294}
\expandafter\gdef\csname odunum@ci@axis_breakdown.m2.gemini35_flash_lite.axis3.3.5\endcsname{[0.454, 0.531]}
\expandafter\gdef\csname odunum@val@axis_breakdown.m2.gemini35_flash_lite.axis3.3.6\endcsname{0.588}
\expandafter\gdef\csname odunum@n@axis_breakdown.m2.gemini35_flash_lite.axis3.3.6\endcsname{501}
\expandafter\gdef\csname odunum@ci@axis_breakdown.m2.gemini35_flash_lite.axis3.3.6\endcsname{[0.561, 0.615]}
\expandafter\gdef\csname odunum@val@axis_breakdown.m2.gemini35_flash_lite.axis4.4.1\endcsname{0.533}
\expandafter\gdef\csname odunum@n@axis_breakdown.m2.gemini35_flash_lite.axis4.4.1\endcsname{202}
\expandafter\gdef\csname odunum@ci@axis_breakdown.m2.gemini35_flash_lite.axis4.4.1\endcsname{[0.492, 0.575]}
\expandafter\gdef\csname odunum@val@axis_breakdown.m2.gemini35_flash_lite.axis4.4.2\endcsname{0.554}
\expandafter\gdef\csname odunum@n@axis_breakdown.m2.gemini35_flash_lite.axis4.4.2\endcsname{457}
\expandafter\gdef\csname odunum@ci@axis_breakdown.m2.gemini35_flash_lite.axis4.4.2\endcsname{[0.525, 0.582]}
\expandafter\gdef\csname odunum@val@axis_breakdown.m2.gemini35_flash_lite.axis4.4.3\endcsname{0.454}
\expandafter\gdef\csname odunum@n@axis_breakdown.m2.gemini35_flash_lite.axis4.4.3\endcsname{164}
\expandafter\gdef\csname odunum@ci@axis_breakdown.m2.gemini35_flash_lite.axis4.4.3\endcsname{[0.401, 0.506]}
\expandafter\gdef\csname odunum@val@axis_breakdown.m2.gemini35_flash_lite.axis4.4.4\endcsname{0.488}
\expandafter\gdef\csname odunum@n@axis_breakdown.m2.gemini35_flash_lite.axis4.4.4\endcsname{505}
\expandafter\gdef\csname odunum@ci@axis_breakdown.m2.gemini35_flash_lite.axis4.4.4\endcsname{[0.459, 0.518]}
\expandafter\gdef\csname odunum@val@axis_breakdown.m2.gemini35_flash_lite.axis4.4.5\endcsname{0.557}
\expandafter\gdef\csname odunum@n@axis_breakdown.m2.gemini35_flash_lite.axis4.4.5\endcsname{303}
\expandafter\gdef\csname odunum@ci@axis_breakdown.m2.gemini35_flash_lite.axis4.4.5\endcsname{[0.519, 0.594]}
\expandafter\gdef\csname odunum@val@axis_breakdown.m2.gemini35_flash_lite.axis5.audio_only.5.1\endcsname{0.597}
\expandafter\gdef\csname odunum@n@axis_breakdown.m2.gemini35_flash_lite.axis5.audio_only.5.1\endcsname{45}
\expandafter\gdef\csname odunum@ci@axis_breakdown.m2.gemini35_flash_lite.axis5.audio_only.5.1\endcsname{[0.501, 0.690]}
\expandafter\gdef\csname odunum@val@axis_breakdown.m2.gemini35_flash_lite.axis5.audio_only.5.10\endcsname{0.640}
\expandafter\gdef\csname odunum@n@axis_breakdown.m2.gemini35_flash_lite.axis5.audio_only.5.10\endcsname{72}
\expandafter\gdef\csname odunum@ci@axis_breakdown.m2.gemini35_flash_lite.axis5.audio_only.5.10\endcsname{[0.573, 0.706]}
\expandafter\gdef\csname odunum@val@axis_breakdown.m2.gemini35_flash_lite.axis5.audio_only.5.2\endcsname{0.564}
\expandafter\gdef\csname odunum@n@axis_breakdown.m2.gemini35_flash_lite.axis5.audio_only.5.2\endcsname{75}
\expandafter\gdef\csname odunum@ci@axis_breakdown.m2.gemini35_flash_lite.axis5.audio_only.5.2\endcsname{[0.488, 0.638]}
\expandafter\gdef\csname odunum@val@axis_breakdown.m2.gemini35_flash_lite.axis5.audio_only.5.3\endcsname{0.658}
\expandafter\gdef\csname odunum@n@axis_breakdown.m2.gemini35_flash_lite.axis5.audio_only.5.3\endcsname{50}
\expandafter\gdef\csname odunum@ci@axis_breakdown.m2.gemini35_flash_lite.axis5.audio_only.5.3\endcsname{[0.575, 0.739]}
\expandafter\gdef\csname odunum@val@axis_breakdown.m2.gemini35_flash_lite.axis5.audio_only.5.4\endcsname{0.501}
\expandafter\gdef\csname odunum@n@axis_breakdown.m2.gemini35_flash_lite.axis5.audio_only.5.4\endcsname{68}
\expandafter\gdef\csname odunum@ci@axis_breakdown.m2.gemini35_flash_lite.axis5.audio_only.5.4\endcsname{[0.420, 0.580]}
\expandafter\gdef\csname odunum@val@axis_breakdown.m2.gemini35_flash_lite.axis5.audio_only.5.5\endcsname{0.545}
\expandafter\gdef\csname odunum@n@axis_breakdown.m2.gemini35_flash_lite.axis5.audio_only.5.5\endcsname{52}
\expandafter\gdef\csname odunum@ci@axis_breakdown.m2.gemini35_flash_lite.axis5.audio_only.5.5\endcsname{[0.454, 0.634]}
\expandafter\gdef\csname odunum@val@axis_breakdown.m2.gemini35_flash_lite.axis5.audio_only.5.6\endcsname{0.623}
\expandafter\gdef\csname odunum@n@axis_breakdown.m2.gemini35_flash_lite.axis5.audio_only.5.6\endcsname{52}
\expandafter\gdef\csname odunum@ci@axis_breakdown.m2.gemini35_flash_lite.axis5.audio_only.5.6\endcsname{[0.551, 0.693]}
\expandafter\gdef\csname odunum@val@axis_breakdown.m2.gemini35_flash_lite.axis5.audio_only.5.7\endcsname{0.574}
\expandafter\gdef\csname odunum@n@axis_breakdown.m2.gemini35_flash_lite.axis5.audio_only.5.7\endcsname{46}
\expandafter\gdef\csname odunum@ci@axis_breakdown.m2.gemini35_flash_lite.axis5.audio_only.5.7\endcsname{[0.474, 0.670]}
\expandafter\gdef\csname odunum@val@axis_breakdown.m2.gemini35_flash_lite.axis5.audio_only.5.8\endcsname{0.341}
\expandafter\gdef\csname odunum@n@axis_breakdown.m2.gemini35_flash_lite.axis5.audio_only.5.8\endcsname{57}
\expandafter\gdef\csname odunum@ci@axis_breakdown.m2.gemini35_flash_lite.axis5.audio_only.5.8\endcsname{[0.252, 0.432]}
\expandafter\gdef\csname odunum@val@axis_breakdown.m2.gemini35_flash_lite.axis5.audio_only.5.9\endcsname{0.616}
\expandafter\gdef\csname odunum@n@axis_breakdown.m2.gemini35_flash_lite.axis5.audio_only.5.9\endcsname{53}
\expandafter\gdef\csname odunum@ci@axis_breakdown.m2.gemini35_flash_lite.axis5.audio_only.5.9\endcsname{[0.531, 0.700]}
\expandafter\gdef\csname odunum@val@axis_breakdown.m2.gemini35_flash_lite.axis5.audio_visual.5.1\endcsname{0.518}
\expandafter\gdef\csname odunum@n@axis_breakdown.m2.gemini35_flash_lite.axis5.audio_visual.5.1\endcsname{114}
\expandafter\gdef\csname odunum@ci@axis_breakdown.m2.gemini35_flash_lite.axis5.audio_visual.5.1\endcsname{[0.456, 0.580]}
\expandafter\gdef\csname odunum@val@axis_breakdown.m2.gemini35_flash_lite.axis5.audio_visual.5.10\endcsname{0.519}
\expandafter\gdef\csname odunum@n@axis_breakdown.m2.gemini35_flash_lite.axis5.audio_visual.5.10\endcsname{123}
\expandafter\gdef\csname odunum@ci@axis_breakdown.m2.gemini35_flash_lite.axis5.audio_visual.5.10\endcsname{[0.466, 0.571]}
\expandafter\gdef\csname odunum@val@axis_breakdown.m2.gemini35_flash_lite.axis5.audio_visual.5.2\endcsname{0.422}
\expandafter\gdef\csname odunum@n@axis_breakdown.m2.gemini35_flash_lite.axis5.audio_visual.5.2\endcsname{126}
\expandafter\gdef\csname odunum@ci@axis_breakdown.m2.gemini35_flash_lite.axis5.audio_visual.5.2\endcsname{[0.364, 0.480]}
\expandafter\gdef\csname odunum@val@axis_breakdown.m2.gemini35_flash_lite.axis5.audio_visual.5.3\endcsname{0.483}
\expandafter\gdef\csname odunum@n@axis_breakdown.m2.gemini35_flash_lite.axis5.audio_visual.5.3\endcsname{100}
\expandafter\gdef\csname odunum@ci@axis_breakdown.m2.gemini35_flash_lite.axis5.audio_visual.5.3\endcsname{[0.424, 0.541]}
\expandafter\gdef\csname odunum@val@axis_breakdown.m2.gemini35_flash_lite.axis5.audio_visual.5.4\endcsname{0.455}
\expandafter\gdef\csname odunum@n@axis_breakdown.m2.gemini35_flash_lite.axis5.audio_visual.5.4\endcsname{97}
\expandafter\gdef\csname odunum@ci@axis_breakdown.m2.gemini35_flash_lite.axis5.audio_visual.5.4\endcsname{[0.388, 0.518]}
\expandafter\gdef\csname odunum@val@axis_breakdown.m2.gemini35_flash_lite.axis5.audio_visual.5.5\endcsname{0.557}
\expandafter\gdef\csname odunum@n@axis_breakdown.m2.gemini35_flash_lite.axis5.audio_visual.5.5\endcsname{99}
\expandafter\gdef\csname odunum@ci@axis_breakdown.m2.gemini35_flash_lite.axis5.audio_visual.5.5\endcsname{[0.491, 0.623]}
\expandafter\gdef\csname odunum@val@axis_breakdown.m2.gemini35_flash_lite.axis5.audio_visual.5.6\endcsname{0.525}
\expandafter\gdef\csname odunum@n@axis_breakdown.m2.gemini35_flash_lite.axis5.audio_visual.5.6\endcsname{99}
\expandafter\gdef\csname odunum@ci@axis_breakdown.m2.gemini35_flash_lite.axis5.audio_visual.5.6\endcsname{[0.458, 0.590]}
\expandafter\gdef\csname odunum@val@axis_breakdown.m2.gemini35_flash_lite.axis5.audio_visual.5.7\endcsname{0.530}
\expandafter\gdef\csname odunum@n@axis_breakdown.m2.gemini35_flash_lite.axis5.audio_visual.5.7\endcsname{104}
\expandafter\gdef\csname odunum@ci@axis_breakdown.m2.gemini35_flash_lite.axis5.audio_visual.5.7\endcsname{[0.477, 0.585]}
\expandafter\gdef\csname odunum@val@axis_breakdown.m2.gemini35_flash_lite.axis5.audio_visual.5.8\endcsname{0.470}
\expandafter\gdef\csname odunum@n@axis_breakdown.m2.gemini35_flash_lite.axis5.audio_visual.5.8\endcsname{121}
\expandafter\gdef\csname odunum@ci@axis_breakdown.m2.gemini35_flash_lite.axis5.audio_visual.5.8\endcsname{[0.411, 0.529]}
\expandafter\gdef\csname odunum@val@axis_breakdown.m2.gemini35_flash_lite.axis5.audio_visual.5.9\endcsname{0.533}
\expandafter\gdef\csname odunum@n@axis_breakdown.m2.gemini35_flash_lite.axis5.audio_visual.5.9\endcsname{78}
\expandafter\gdef\csname odunum@ci@axis_breakdown.m2.gemini35_flash_lite.axis5.audio_visual.5.9\endcsname{[0.463, 0.601]}
\expandafter\gdef\csname odunum@val@axis_breakdown.m2.gemini35_flash_lite.axis6.6.1\endcsname{0.515}
\expandafter\gdef\csname odunum@n@axis_breakdown.m2.gemini35_flash_lite.axis6.6.1\endcsname{230}
\expandafter\gdef\csname odunum@ci@axis_breakdown.m2.gemini35_flash_lite.axis6.6.1\endcsname{[0.473, 0.557]}
\expandafter\gdef\csname odunum@val@axis_breakdown.m2.gemini35_flash_lite.axis6.6.10\endcsname{0.527}
\expandafter\gdef\csname odunum@n@axis_breakdown.m2.gemini35_flash_lite.axis6.6.10\endcsname{102}
\expandafter\gdef\csname odunum@ci@axis_breakdown.m2.gemini35_flash_lite.axis6.6.10\endcsname{[0.468, 0.586]}
\expandafter\gdef\csname odunum@val@axis_breakdown.m2.gemini35_flash_lite.axis6.6.12\endcsname{0.557}
\expandafter\gdef\csname odunum@n@axis_breakdown.m2.gemini35_flash_lite.axis6.6.12\endcsname{129}
\expandafter\gdef\csname odunum@ci@axis_breakdown.m2.gemini35_flash_lite.axis6.6.12\endcsname{[0.502, 0.611]}
\expandafter\gdef\csname odunum@val@axis_breakdown.m2.gemini35_flash_lite.axis6.6.13\endcsname{0.466}
\expandafter\gdef\csname odunum@n@axis_breakdown.m2.gemini35_flash_lite.axis6.6.13\endcsname{58}
\expandafter\gdef\csname odunum@ci@axis_breakdown.m2.gemini35_flash_lite.axis6.6.13\endcsname{[0.378, 0.554]}
\expandafter\gdef\csname odunum@val@axis_breakdown.m2.gemini35_flash_lite.axis6.6.14\endcsname{0.463}
\expandafter\gdef\csname odunum@n@axis_breakdown.m2.gemini35_flash_lite.axis6.6.14\endcsname{82}
\expandafter\gdef\csname odunum@ci@axis_breakdown.m2.gemini35_flash_lite.axis6.6.14\endcsname{[0.391, 0.534]}
\expandafter\gdef\csname odunum@val@axis_breakdown.m2.gemini35_flash_lite.axis6.6.18\endcsname{0.492}
\expandafter\gdef\csname odunum@n@axis_breakdown.m2.gemini35_flash_lite.axis6.6.18\endcsname{72}
\expandafter\gdef\csname odunum@ci@axis_breakdown.m2.gemini35_flash_lite.axis6.6.18\endcsname{[0.416, 0.568]}
\expandafter\gdef\csname odunum@val@axis_breakdown.m2.gemini35_flash_lite.axis6.6.19\endcsname{0.453}
\expandafter\gdef\csname odunum@n@axis_breakdown.m2.gemini35_flash_lite.axis6.6.19\endcsname{44}
\expandafter\gdef\csname odunum@ci@axis_breakdown.m2.gemini35_flash_lite.axis6.6.19\endcsname{[0.352, 0.552]}
\expandafter\gdef\csname odunum@val@axis_breakdown.m2.gemini35_flash_lite.axis6.6.2\endcsname{0.574}
\expandafter\gdef\csname odunum@n@axis_breakdown.m2.gemini35_flash_lite.axis6.6.2\endcsname{41}
\expandafter\gdef\csname odunum@ci@axis_breakdown.m2.gemini35_flash_lite.axis6.6.2\endcsname{[0.485, 0.665]}
\expandafter\gdef\csname odunum@val@axis_breakdown.m2.gemini35_flash_lite.axis6.6.3\endcsname{0.560}
\expandafter\gdef\csname odunum@n@axis_breakdown.m2.gemini35_flash_lite.axis6.6.3\endcsname{44}
\expandafter\gdef\csname odunum@ci@axis_breakdown.m2.gemini35_flash_lite.axis6.6.3\endcsname{[0.460, 0.658]}
\expandafter\gdef\csname odunum@val@axis_breakdown.m2.gemini35_flash_lite.axis6.6.5\endcsname{0.537}
\expandafter\gdef\csname odunum@n@axis_breakdown.m2.gemini35_flash_lite.axis6.6.5\endcsname{48}
\expandafter\gdef\csname odunum@ci@axis_breakdown.m2.gemini35_flash_lite.axis6.6.5\endcsname{[0.438, 0.637]}
\expandafter\gdef\csname odunum@val@axis_breakdown.m2.gemini35_flash_lite.axis6.6.6\endcsname{0.514}
\expandafter\gdef\csname odunum@n@axis_breakdown.m2.gemini35_flash_lite.axis6.6.6\endcsname{134}
\expandafter\gdef\csname odunum@ci@axis_breakdown.m2.gemini35_flash_lite.axis6.6.6\endcsname{[0.455, 0.573]}
\expandafter\gdef\csname odunum@val@axis_breakdown.m2.gemini35_flash_lite.axis6.6.7\endcsname{0.570}
\expandafter\gdef\csname odunum@n@axis_breakdown.m2.gemini35_flash_lite.axis6.6.7\endcsname{393}
\expandafter\gdef\csname odunum@ci@axis_breakdown.m2.gemini35_flash_lite.axis6.6.7\endcsname{[0.539, 0.602]}
\expandafter\gdef\csname odunum@val@axis_breakdown.m2.gemini35_flash_lite.axis6.6.8\endcsname{0.442}
\expandafter\gdef\csname odunum@n@axis_breakdown.m2.gemini35_flash_lite.axis6.6.8\endcsname{49}
\expandafter\gdef\csname odunum@ci@axis_breakdown.m2.gemini35_flash_lite.axis6.6.8\endcsname{[0.354, 0.527]}
\expandafter\gdef\csname odunum@val@axis_breakdown.m2.gemini35_flash_lite.axis6.6.9\endcsname{0.507}
\expandafter\gdef\csname odunum@n@axis_breakdown.m2.gemini35_flash_lite.axis6.6.9\endcsname{119}
\expandafter\gdef\csname odunum@ci@axis_breakdown.m2.gemini35_flash_lite.axis6.6.9\endcsname{[0.450, 0.563]}
\expandafter\gdef\csname odunum@val@axis_breakdown.m2.gemini35_flash_lite.difficulty.q1\endcsname{0.576}
\expandafter\gdef\csname odunum@n@axis_breakdown.m2.gemini35_flash_lite.difficulty.q1\endcsname{358}
\expandafter\gdef\csname odunum@ci@axis_breakdown.m2.gemini35_flash_lite.difficulty.q1\endcsname{[0.541, 0.611]}
\expandafter\gdef\csname odunum@val@axis_breakdown.m2.gemini35_flash_lite.difficulty.q2\endcsname{0.531}
\expandafter\gdef\csname odunum@n@axis_breakdown.m2.gemini35_flash_lite.difficulty.q2\endcsname{357}
\expandafter\gdef\csname odunum@ci@axis_breakdown.m2.gemini35_flash_lite.difficulty.q2\endcsname{[0.498, 0.565]}
\expandafter\gdef\csname odunum@val@axis_breakdown.m2.gemini35_flash_lite.difficulty.q3\endcsname{0.532}
\expandafter\gdef\csname odunum@n@axis_breakdown.m2.gemini35_flash_lite.difficulty.q3\endcsname{358}
\expandafter\gdef\csname odunum@ci@axis_breakdown.m2.gemini35_flash_lite.difficulty.q3\endcsname{[0.498, 0.566]}
\expandafter\gdef\csname odunum@val@axis_breakdown.m2.gemini35_flash_lite.difficulty.q4\endcsname{0.531}
\expandafter\gdef\csname odunum@n@axis_breakdown.m2.gemini35_flash_lite.difficulty.q4\endcsname{356}
\expandafter\gdef\csname odunum@ci@axis_breakdown.m2.gemini35_flash_lite.difficulty.q4\endcsname{[0.498, 0.562]}
\expandafter\gdef\csname odunum@val@axis_breakdown.m2.gemini37_flash.axis1.audio_only.1.1\endcsname{0.718}
\expandafter\gdef\csname odunum@n@axis_breakdown.m2.gemini37_flash.axis1.audio_only.1.1\endcsname{156}
\expandafter\gdef\csname odunum@ci@axis_breakdown.m2.gemini37_flash.axis1.audio_only.1.1\endcsname{[0.675, 0.759]}
\expandafter\gdef\csname odunum@val@axis_breakdown.m2.gemini37_flash.axis1.audio_only.1.2\endcsname{0.636}
\expandafter\gdef\csname odunum@n@axis_breakdown.m2.gemini37_flash.axis1.audio_only.1.2\endcsname{378}
\expandafter\gdef\csname odunum@ci@axis_breakdown.m2.gemini37_flash.axis1.audio_only.1.2\endcsname{[0.608, 0.665]}
\expandafter\gdef\csname odunum@val@axis_breakdown.m2.gemini37_flash.axis1.audio_visual.1.1\endcsname{0.621}
\expandafter\gdef\csname odunum@n@axis_breakdown.m2.gemini37_flash.axis1.audio_visual.1.1\endcsname{93}
\expandafter\gdef\csname odunum@ci@axis_breakdown.m2.gemini37_flash.axis1.audio_visual.1.1\endcsname{[0.564, 0.676]}
\expandafter\gdef\csname odunum@val@axis_breakdown.m2.gemini37_flash.axis1.audio_visual.1.2\endcsname{0.600}
\expandafter\gdef\csname odunum@n@axis_breakdown.m2.gemini37_flash.axis1.audio_visual.1.2\endcsname{294}
\expandafter\gdef\csname odunum@ci@axis_breakdown.m2.gemini37_flash.axis1.audio_visual.1.2\endcsname{[0.567, 0.633]}
\expandafter\gdef\csname odunum@val@axis_breakdown.m2.gemini37_flash.axis1.audio_visual.1.3\endcsname{0.612}
\expandafter\gdef\csname odunum@n@axis_breakdown.m2.gemini37_flash.axis1.audio_visual.1.3\endcsname{291}
\expandafter\gdef\csname odunum@ci@axis_breakdown.m2.gemini37_flash.axis1.audio_visual.1.3\endcsname{[0.579, 0.645]}
\expandafter\gdef\csname odunum@val@axis_breakdown.m2.gemini37_flash.axis1.audio_visual.1.4\endcsname{0.590}
\expandafter\gdef\csname odunum@n@axis_breakdown.m2.gemini37_flash.axis1.audio_visual.1.4\endcsname{382}
\expandafter\gdef\csname odunum@ci@axis_breakdown.m2.gemini37_flash.axis1.audio_visual.1.4\endcsname{[0.562, 0.618]}
\expandafter\gdef\csname odunum@val@axis_breakdown.m2.gemini37_flash.axis2.2.1\endcsname{0.576}
\expandafter\gdef\csname odunum@n@axis_breakdown.m2.gemini37_flash.axis2.2.1\endcsname{143}
\expandafter\gdef\csname odunum@ci@axis_breakdown.m2.gemini37_flash.axis2.2.1\endcsname{[0.522, 0.629]}
\expandafter\gdef\csname odunum@val@axis_breakdown.m2.gemini37_flash.axis2.2.2\endcsname{0.639}
\expandafter\gdef\csname odunum@n@axis_breakdown.m2.gemini37_flash.axis2.2.2\endcsname{439}
\expandafter\gdef\csname odunum@ci@axis_breakdown.m2.gemini37_flash.axis2.2.2\endcsname{[0.615, 0.662]}
\expandafter\gdef\csname odunum@val@axis_breakdown.m2.gemini37_flash.axis2.2.3\endcsname{0.666}
\expandafter\gdef\csname odunum@n@axis_breakdown.m2.gemini37_flash.axis2.2.3\endcsname{528}
\expandafter\gdef\csname odunum@ci@axis_breakdown.m2.gemini37_flash.axis2.2.3\endcsname{[0.644, 0.688]}
\expandafter\gdef\csname odunum@val@axis_breakdown.m2.gemini37_flash.axis2.2.4\endcsname{0.588}
\expandafter\gdef\csname odunum@n@axis_breakdown.m2.gemini37_flash.axis2.2.4\endcsname{261}
\expandafter\gdef\csname odunum@ci@axis_breakdown.m2.gemini37_flash.axis2.2.4\endcsname{[0.551, 0.625]}
\expandafter\gdef\csname odunum@val@axis_breakdown.m2.gemini37_flash.axis2.2.5\endcsname{0.565}
\expandafter\gdef\csname odunum@n@axis_breakdown.m2.gemini37_flash.axis2.2.5\endcsname{256}
\expandafter\gdef\csname odunum@ci@axis_breakdown.m2.gemini37_flash.axis2.2.5\endcsname{[0.528, 0.603]}
\expandafter\gdef\csname odunum@val@axis_breakdown.m2.gemini37_flash.axis3.3.1\endcsname{0.649}
\expandafter\gdef\csname odunum@n@axis_breakdown.m2.gemini37_flash.axis3.3.1\endcsname{109}
\expandafter\gdef\csname odunum@ci@axis_breakdown.m2.gemini37_flash.axis3.3.1\endcsname{[0.597, 0.700]}
\expandafter\gdef\csname odunum@val@axis_breakdown.m2.gemini37_flash.axis3.3.2\endcsname{0.621}
\expandafter\gdef\csname odunum@n@axis_breakdown.m2.gemini37_flash.axis3.3.2\endcsname{116}
\expandafter\gdef\csname odunum@ci@axis_breakdown.m2.gemini37_flash.axis3.3.2\endcsname{[0.572, 0.668]}
\expandafter\gdef\csname odunum@val@axis_breakdown.m2.gemini37_flash.axis3.3.3\endcsname{0.526}
\expandafter\gdef\csname odunum@n@axis_breakdown.m2.gemini37_flash.axis3.3.3\endcsname{206}
\expandafter\gdef\csname odunum@ci@axis_breakdown.m2.gemini37_flash.axis3.3.3\endcsname{[0.482, 0.568]}
\expandafter\gdef\csname odunum@val@axis_breakdown.m2.gemini37_flash.axis3.3.4\endcsname{0.640}
\expandafter\gdef\csname odunum@n@axis_breakdown.m2.gemini37_flash.axis3.3.4\endcsname{403}
\expandafter\gdef\csname odunum@ci@axis_breakdown.m2.gemini37_flash.axis3.3.4\endcsname{[0.613, 0.666]}
\expandafter\gdef\csname odunum@val@axis_breakdown.m2.gemini37_flash.axis3.3.5\endcsname{0.530}
\expandafter\gdef\csname odunum@n@axis_breakdown.m2.gemini37_flash.axis3.3.5\endcsname{293}
\expandafter\gdef\csname odunum@ci@axis_breakdown.m2.gemini37_flash.axis3.3.5\endcsname{[0.499, 0.562]}
\expandafter\gdef\csname odunum@val@axis_breakdown.m2.gemini37_flash.axis3.3.6\endcsname{0.697}
\expandafter\gdef\csname odunum@n@axis_breakdown.m2.gemini37_flash.axis3.3.6\endcsname{500}
\expandafter\gdef\csname odunum@ci@axis_breakdown.m2.gemini37_flash.axis3.3.6\endcsname{[0.675, 0.718]}
\expandafter\gdef\csname odunum@val@axis_breakdown.m2.gemini37_flash.axis4.4.1\endcsname{0.638}
\expandafter\gdef\csname odunum@n@axis_breakdown.m2.gemini37_flash.axis4.4.1\endcsname{202}
\expandafter\gdef\csname odunum@ci@axis_breakdown.m2.gemini37_flash.axis4.4.1\endcsname{[0.604, 0.672]}
\expandafter\gdef\csname odunum@val@axis_breakdown.m2.gemini37_flash.axis4.4.2\endcsname{0.614}
\expandafter\gdef\csname odunum@n@axis_breakdown.m2.gemini37_flash.axis4.4.2\endcsname{457}
\expandafter\gdef\csname odunum@ci@axis_breakdown.m2.gemini37_flash.axis4.4.2\endcsname{[0.589, 0.638]}
\expandafter\gdef\csname odunum@val@axis_breakdown.m2.gemini37_flash.axis4.4.3\endcsname{0.612}
\expandafter\gdef\csname odunum@n@axis_breakdown.m2.gemini37_flash.axis4.4.3\endcsname{163}
\expandafter\gdef\csname odunum@ci@axis_breakdown.m2.gemini37_flash.axis4.4.3\endcsname{[0.567, 0.657]}
\expandafter\gdef\csname odunum@val@axis_breakdown.m2.gemini37_flash.axis4.4.4\endcsname{0.611}
\expandafter\gdef\csname odunum@n@axis_breakdown.m2.gemini37_flash.axis4.4.4\endcsname{503}
\expandafter\gdef\csname odunum@ci@axis_breakdown.m2.gemini37_flash.axis4.4.4\endcsname{[0.587, 0.637]}
\expandafter\gdef\csname odunum@val@axis_breakdown.m2.gemini37_flash.axis4.4.5\endcsname{0.649}
\expandafter\gdef\csname odunum@n@axis_breakdown.m2.gemini37_flash.axis4.4.5\endcsname{302}
\expandafter\gdef\csname odunum@ci@axis_breakdown.m2.gemini37_flash.axis4.4.5\endcsname{[0.616, 0.681]}
\expandafter\gdef\csname odunum@val@axis_breakdown.m2.gemini37_flash.axis5.audio_only.5.1\endcsname{0.724}
\expandafter\gdef\csname odunum@n@axis_breakdown.m2.gemini37_flash.axis5.audio_only.5.1\endcsname{45}
\expandafter\gdef\csname odunum@ci@axis_breakdown.m2.gemini37_flash.axis5.audio_only.5.1\endcsname{[0.633, 0.811]}
\expandafter\gdef\csname odunum@val@axis_breakdown.m2.gemini37_flash.axis5.audio_only.5.10\endcsname{0.680}
\expandafter\gdef\csname odunum@n@axis_breakdown.m2.gemini37_flash.axis5.audio_only.5.10\endcsname{72}
\expandafter\gdef\csname odunum@ci@axis_breakdown.m2.gemini37_flash.axis5.audio_only.5.10\endcsname{[0.617, 0.741]}
\expandafter\gdef\csname odunum@val@axis_breakdown.m2.gemini37_flash.axis5.audio_only.5.2\endcsname{0.626}
\expandafter\gdef\csname odunum@n@axis_breakdown.m2.gemini37_flash.axis5.audio_only.5.2\endcsname{75}
\expandafter\gdef\csname odunum@ci@axis_breakdown.m2.gemini37_flash.axis5.audio_only.5.2\endcsname{[0.559, 0.691]}
\expandafter\gdef\csname odunum@val@axis_breakdown.m2.gemini37_flash.axis5.audio_only.5.3\endcsname{0.652}
\expandafter\gdef\csname odunum@n@axis_breakdown.m2.gemini37_flash.axis5.audio_only.5.3\endcsname{50}
\expandafter\gdef\csname odunum@ci@axis_breakdown.m2.gemini37_flash.axis5.audio_only.5.3\endcsname{[0.569, 0.734]}
\expandafter\gdef\csname odunum@val@axis_breakdown.m2.gemini37_flash.axis5.audio_only.5.4\endcsname{0.647}
\expandafter\gdef\csname odunum@n@axis_breakdown.m2.gemini37_flash.axis5.audio_only.5.4\endcsname{67}
\expandafter\gdef\csname odunum@ci@axis_breakdown.m2.gemini37_flash.axis5.audio_only.5.4\endcsname{[0.584, 0.708]}
\expandafter\gdef\csname odunum@val@axis_breakdown.m2.gemini37_flash.axis5.audio_only.5.5\endcsname{0.683}
\expandafter\gdef\csname odunum@n@axis_breakdown.m2.gemini37_flash.axis5.audio_only.5.5\endcsname{52}
\expandafter\gdef\csname odunum@ci@axis_breakdown.m2.gemini37_flash.axis5.audio_only.5.5\endcsname{[0.614, 0.749]}
\expandafter\gdef\csname odunum@val@axis_breakdown.m2.gemini37_flash.axis5.audio_only.5.6\endcsname{0.683}
\expandafter\gdef\csname odunum@n@axis_breakdown.m2.gemini37_flash.axis5.audio_only.5.6\endcsname{52}
\expandafter\gdef\csname odunum@ci@axis_breakdown.m2.gemini37_flash.axis5.audio_only.5.6\endcsname{[0.618, 0.746]}
\expandafter\gdef\csname odunum@val@axis_breakdown.m2.gemini37_flash.axis5.audio_only.5.7\endcsname{0.644}
\expandafter\gdef\csname odunum@n@axis_breakdown.m2.gemini37_flash.axis5.audio_only.5.7\endcsname{45}
\expandafter\gdef\csname odunum@ci@axis_breakdown.m2.gemini37_flash.axis5.audio_only.5.7\endcsname{[0.578, 0.713]}
\expandafter\gdef\csname odunum@val@axis_breakdown.m2.gemini37_flash.axis5.audio_only.5.8\endcsname{0.583}
\expandafter\gdef\csname odunum@n@axis_breakdown.m2.gemini37_flash.axis5.audio_only.5.8\endcsname{57}
\expandafter\gdef\csname odunum@ci@axis_breakdown.m2.gemini37_flash.axis5.audio_only.5.8\endcsname{[0.492, 0.671]}
\expandafter\gdef\csname odunum@val@axis_breakdown.m2.gemini37_flash.axis5.audio_only.5.9\endcsname{0.715}
\expandafter\gdef\csname odunum@n@axis_breakdown.m2.gemini37_flash.axis5.audio_only.5.9\endcsname{52}
\expandafter\gdef\csname odunum@ci@axis_breakdown.m2.gemini37_flash.axis5.audio_only.5.9\endcsname{[0.648, 0.780]}
\expandafter\gdef\csname odunum@val@axis_breakdown.m2.gemini37_flash.axis5.audio_visual.5.1\endcsname{0.598}
\expandafter\gdef\csname odunum@n@axis_breakdown.m2.gemini37_flash.axis5.audio_visual.5.1\endcsname{114}
\expandafter\gdef\csname odunum@ci@axis_breakdown.m2.gemini37_flash.axis5.audio_visual.5.1\endcsname{[0.545, 0.649]}
\expandafter\gdef\csname odunum@val@axis_breakdown.m2.gemini37_flash.axis5.audio_visual.5.10\endcsname{0.624}
\expandafter\gdef\csname odunum@n@axis_breakdown.m2.gemini37_flash.axis5.audio_visual.5.10\endcsname{123}
\expandafter\gdef\csname odunum@ci@axis_breakdown.m2.gemini37_flash.axis5.audio_visual.5.10\endcsname{[0.577, 0.670]}
\expandafter\gdef\csname odunum@val@axis_breakdown.m2.gemini37_flash.axis5.audio_visual.5.2\endcsname{0.551}
\expandafter\gdef\csname odunum@n@axis_breakdown.m2.gemini37_flash.axis5.audio_visual.5.2\endcsname{126}
\expandafter\gdef\csname odunum@ci@axis_breakdown.m2.gemini37_flash.axis5.audio_visual.5.2\endcsname{[0.503, 0.600]}
\expandafter\gdef\csname odunum@val@axis_breakdown.m2.gemini37_flash.axis5.audio_visual.5.3\endcsname{0.600}
\expandafter\gdef\csname odunum@n@axis_breakdown.m2.gemini37_flash.axis5.audio_visual.5.3\endcsname{100}
\expandafter\gdef\csname odunum@ci@axis_breakdown.m2.gemini37_flash.axis5.audio_visual.5.3\endcsname{[0.543, 0.654]}
\expandafter\gdef\csname odunum@val@axis_breakdown.m2.gemini37_flash.axis5.audio_visual.5.4\endcsname{0.572}
\expandafter\gdef\csname odunum@n@axis_breakdown.m2.gemini37_flash.axis5.audio_visual.5.4\endcsname{97}
\expandafter\gdef\csname odunum@ci@axis_breakdown.m2.gemini37_flash.axis5.audio_visual.5.4\endcsname{[0.511, 0.633]}
\expandafter\gdef\csname odunum@val@axis_breakdown.m2.gemini37_flash.axis5.audio_visual.5.5\endcsname{0.632}
\expandafter\gdef\csname odunum@n@axis_breakdown.m2.gemini37_flash.axis5.audio_visual.5.5\endcsname{99}
\expandafter\gdef\csname odunum@ci@axis_breakdown.m2.gemini37_flash.axis5.audio_visual.5.5\endcsname{[0.580, 0.684]}
\expandafter\gdef\csname odunum@val@axis_breakdown.m2.gemini37_flash.axis5.audio_visual.5.6\endcsname{0.640}
\expandafter\gdef\csname odunum@n@axis_breakdown.m2.gemini37_flash.axis5.audio_visual.5.6\endcsname{99}
\expandafter\gdef\csname odunum@ci@axis_breakdown.m2.gemini37_flash.axis5.audio_visual.5.6\endcsname{[0.587, 0.691]}
\expandafter\gdef\csname odunum@val@axis_breakdown.m2.gemini37_flash.axis5.audio_visual.5.7\endcsname{0.614}
\expandafter\gdef\csname odunum@n@axis_breakdown.m2.gemini37_flash.axis5.audio_visual.5.7\endcsname{104}
\expandafter\gdef\csname odunum@ci@axis_breakdown.m2.gemini37_flash.axis5.audio_visual.5.7\endcsname{[0.555, 0.671]}
\expandafter\gdef\csname odunum@val@axis_breakdown.m2.gemini37_flash.axis5.audio_visual.5.8\endcsname{0.542}
\expandafter\gdef\csname odunum@n@axis_breakdown.m2.gemini37_flash.axis5.audio_visual.5.8\endcsname{120}
\expandafter\gdef\csname odunum@ci@axis_breakdown.m2.gemini37_flash.axis5.audio_visual.5.8\endcsname{[0.493, 0.592]}
\expandafter\gdef\csname odunum@val@axis_breakdown.m2.gemini37_flash.axis5.audio_visual.5.9\endcsname{0.680}
\expandafter\gdef\csname odunum@n@axis_breakdown.m2.gemini37_flash.axis5.audio_visual.5.9\endcsname{78}
\expandafter\gdef\csname odunum@ci@axis_breakdown.m2.gemini37_flash.axis5.audio_visual.5.9\endcsname{[0.625, 0.734]}
\expandafter\gdef\csname odunum@val@axis_breakdown.m2.gemini37_flash.axis6.6.1\endcsname{0.596}
\expandafter\gdef\csname odunum@n@axis_breakdown.m2.gemini37_flash.axis6.6.1\endcsname{230}
\expandafter\gdef\csname odunum@ci@axis_breakdown.m2.gemini37_flash.axis6.6.1\endcsname{[0.557, 0.633]}
\expandafter\gdef\csname odunum@val@axis_breakdown.m2.gemini37_flash.axis6.6.10\endcsname{0.621}
\expandafter\gdef\csname odunum@n@axis_breakdown.m2.gemini37_flash.axis6.6.10\endcsname{101}
\expandafter\gdef\csname odunum@ci@axis_breakdown.m2.gemini37_flash.axis6.6.10\endcsname{[0.565, 0.675]}
\expandafter\gdef\csname odunum@val@axis_breakdown.m2.gemini37_flash.axis6.6.12\endcsname{0.623}
\expandafter\gdef\csname odunum@n@axis_breakdown.m2.gemini37_flash.axis6.6.12\endcsname{128}
\expandafter\gdef\csname odunum@ci@axis_breakdown.m2.gemini37_flash.axis6.6.12\endcsname{[0.572, 0.673]}
\expandafter\gdef\csname odunum@val@axis_breakdown.m2.gemini37_flash.axis6.6.13\endcsname{0.607}
\expandafter\gdef\csname odunum@n@axis_breakdown.m2.gemini37_flash.axis6.6.13\endcsname{58}
\expandafter\gdef\csname odunum@ci@axis_breakdown.m2.gemini37_flash.axis6.6.13\endcsname{[0.534, 0.678]}
\expandafter\gdef\csname odunum@val@axis_breakdown.m2.gemini37_flash.axis6.6.14\endcsname{0.614}
\expandafter\gdef\csname odunum@n@axis_breakdown.m2.gemini37_flash.axis6.6.14\endcsname{82}
\expandafter\gdef\csname odunum@ci@axis_breakdown.m2.gemini37_flash.axis6.6.14\endcsname{[0.556, 0.670]}
\expandafter\gdef\csname odunum@val@axis_breakdown.m2.gemini37_flash.axis6.6.18\endcsname{0.584}
\expandafter\gdef\csname odunum@n@axis_breakdown.m2.gemini37_flash.axis6.6.18\endcsname{72}
\expandafter\gdef\csname odunum@ci@axis_breakdown.m2.gemini37_flash.axis6.6.18\endcsname{[0.515, 0.652]}
\expandafter\gdef\csname odunum@val@axis_breakdown.m2.gemini37_flash.axis6.6.19\endcsname{0.559}
\expandafter\gdef\csname odunum@n@axis_breakdown.m2.gemini37_flash.axis6.6.19\endcsname{44}
\expandafter\gdef\csname odunum@ci@axis_breakdown.m2.gemini37_flash.axis6.6.19\endcsname{[0.475, 0.642]}
\expandafter\gdef\csname odunum@val@axis_breakdown.m2.gemini37_flash.axis6.6.2\endcsname{0.657}
\expandafter\gdef\csname odunum@n@axis_breakdown.m2.gemini37_flash.axis6.6.2\endcsname{41}
\expandafter\gdef\csname odunum@ci@axis_breakdown.m2.gemini37_flash.axis6.6.2\endcsname{[0.584, 0.731]}
\expandafter\gdef\csname odunum@val@axis_breakdown.m2.gemini37_flash.axis6.6.3\endcsname{0.631}
\expandafter\gdef\csname odunum@n@axis_breakdown.m2.gemini37_flash.axis6.6.3\endcsname{44}
\expandafter\gdef\csname odunum@ci@axis_breakdown.m2.gemini37_flash.axis6.6.3\endcsname{[0.536, 0.724]}
\expandafter\gdef\csname odunum@val@axis_breakdown.m2.gemini37_flash.axis6.6.5\endcsname{0.611}
\expandafter\gdef\csname odunum@n@axis_breakdown.m2.gemini37_flash.axis6.6.5\endcsname{48}
\expandafter\gdef\csname odunum@ci@axis_breakdown.m2.gemini37_flash.axis6.6.5\endcsname{[0.537, 0.686]}
\expandafter\gdef\csname odunum@val@axis_breakdown.m2.gemini37_flash.axis6.6.6\endcsname{0.657}
\expandafter\gdef\csname odunum@n@axis_breakdown.m2.gemini37_flash.axis6.6.6\endcsname{134}
\expandafter\gdef\csname odunum@ci@axis_breakdown.m2.gemini37_flash.axis6.6.6\endcsname{[0.609, 0.704]}
\expandafter\gdef\csname odunum@val@axis_breakdown.m2.gemini37_flash.axis6.6.7\endcsname{0.646}
\expandafter\gdef\csname odunum@n@axis_breakdown.m2.gemini37_flash.axis6.6.7\endcsname{392}
\expandafter\gdef\csname odunum@ci@axis_breakdown.m2.gemini37_flash.axis6.6.7\endcsname{[0.618, 0.674]}
\expandafter\gdef\csname odunum@val@axis_breakdown.m2.gemini37_flash.axis6.6.8\endcsname{0.610}
\expandafter\gdef\csname odunum@n@axis_breakdown.m2.gemini37_flash.axis6.6.8\endcsname{49}
\expandafter\gdef\csname odunum@ci@axis_breakdown.m2.gemini37_flash.axis6.6.8\endcsname{[0.548, 0.670]}
\expandafter\gdef\csname odunum@val@axis_breakdown.m2.gemini37_flash.axis6.6.9\endcsname{0.635}
\expandafter\gdef\csname odunum@n@axis_breakdown.m2.gemini37_flash.axis6.6.9\endcsname{119}
\expandafter\gdef\csname odunum@ci@axis_breakdown.m2.gemini37_flash.axis6.6.9\endcsname{[0.587, 0.681]}
\expandafter\gdef\csname odunum@val@axis_breakdown.m2.gemini37_flash.difficulty.q1\endcsname{0.670}
\expandafter\gdef\csname odunum@n@axis_breakdown.m2.gemini37_flash.difficulty.q1\endcsname{357}
\expandafter\gdef\csname odunum@ci@axis_breakdown.m2.gemini37_flash.difficulty.q1\endcsname{[0.642, 0.699]}
\expandafter\gdef\csname odunum@val@axis_breakdown.m2.gemini37_flash.difficulty.q2\endcsname{0.638}
\expandafter\gdef\csname odunum@n@axis_breakdown.m2.gemini37_flash.difficulty.q2\endcsname{356}
\expandafter\gdef\csname odunum@ci@axis_breakdown.m2.gemini37_flash.difficulty.q2\endcsname{[0.609, 0.666]}
\expandafter\gdef\csname odunum@val@axis_breakdown.m2.gemini37_flash.difficulty.q3\endcsname{0.619}
\expandafter\gdef\csname odunum@n@axis_breakdown.m2.gemini37_flash.difficulty.q3\endcsname{357}
\expandafter\gdef\csname odunum@ci@axis_breakdown.m2.gemini37_flash.difficulty.q3\endcsname{[0.590, 0.647]}
\expandafter\gdef\csname odunum@val@axis_breakdown.m2.gemini37_flash.difficulty.q4\endcsname{0.579}
\expandafter\gdef\csname odunum@n@axis_breakdown.m2.gemini37_flash.difficulty.q4\endcsname{356}
\expandafter\gdef\csname odunum@ci@axis_breakdown.m2.gemini37_flash.difficulty.q4\endcsname{[0.549, 0.607]}
\expandafter\gdef\csname odunum@val@axis_breakdown.m2.ming.axis1.audio_only.1.1\endcsname{0.543}
\expandafter\gdef\csname odunum@n@axis_breakdown.m2.ming.axis1.audio_only.1.1\endcsname{158}
\expandafter\gdef\csname odunum@ci@axis_breakdown.m2.ming.axis1.audio_only.1.1\endcsname{[0.493, 0.592]}
\expandafter\gdef\csname odunum@val@axis_breakdown.m2.ming.axis1.audio_only.1.2\endcsname{0.487}
\expandafter\gdef\csname odunum@n@axis_breakdown.m2.ming.axis1.audio_only.1.2\endcsname{379}
\expandafter\gdef\csname odunum@ci@axis_breakdown.m2.ming.axis1.audio_only.1.2\endcsname{[0.457, 0.517]}
\expandafter\gdef\csname odunum@val@axis_breakdown.m2.ming.axis1.audio_visual.1.1\endcsname{0.559}
\expandafter\gdef\csname odunum@n@axis_breakdown.m2.ming.axis1.audio_visual.1.1\endcsname{94}
\expandafter\gdef\csname odunum@ci@axis_breakdown.m2.ming.axis1.audio_visual.1.1\endcsname{[0.505, 0.613]}
\expandafter\gdef\csname odunum@val@axis_breakdown.m2.ming.axis1.audio_visual.1.2\endcsname{0.496}
\expandafter\gdef\csname odunum@n@axis_breakdown.m2.ming.axis1.audio_visual.1.2\endcsname{294}
\expandafter\gdef\csname odunum@ci@axis_breakdown.m2.ming.axis1.audio_visual.1.2\endcsname{[0.466, 0.526]}
\expandafter\gdef\csname odunum@val@axis_breakdown.m2.ming.axis1.audio_visual.1.3\endcsname{0.490}
\expandafter\gdef\csname odunum@n@axis_breakdown.m2.ming.axis1.audio_visual.1.3\endcsname{291}
\expandafter\gdef\csname odunum@ci@axis_breakdown.m2.ming.axis1.audio_visual.1.3\endcsname{[0.459, 0.522]}
\expandafter\gdef\csname odunum@val@axis_breakdown.m2.ming.axis1.audio_visual.1.4\endcsname{0.483}
\expandafter\gdef\csname odunum@n@axis_breakdown.m2.ming.axis1.audio_visual.1.4\endcsname{382}
\expandafter\gdef\csname odunum@ci@axis_breakdown.m2.ming.axis1.audio_visual.1.4\endcsname{[0.457, 0.510]}
\expandafter\gdef\csname odunum@val@axis_breakdown.m2.ming.axis2.2.1\endcsname{0.485}
\expandafter\gdef\csname odunum@n@axis_breakdown.m2.ming.axis2.2.1\endcsname{143}
\expandafter\gdef\csname odunum@ci@axis_breakdown.m2.ming.axis2.2.1\endcsname{[0.436, 0.534]}
\expandafter\gdef\csname odunum@val@axis_breakdown.m2.ming.axis2.2.2\endcsname{0.526}
\expandafter\gdef\csname odunum@n@axis_breakdown.m2.ming.axis2.2.2\endcsname{440}
\expandafter\gdef\csname odunum@ci@axis_breakdown.m2.ming.axis2.2.2\endcsname{[0.502, 0.550]}
\expandafter\gdef\csname odunum@val@axis_breakdown.m2.ming.axis2.2.3\endcsname{0.522}
\expandafter\gdef\csname odunum@n@axis_breakdown.m2.ming.axis2.2.3\endcsname{531}
\expandafter\gdef\csname odunum@ci@axis_breakdown.m2.ming.axis2.2.3\endcsname{[0.498, 0.545]}
\expandafter\gdef\csname odunum@val@axis_breakdown.m2.ming.axis2.2.4\endcsname{0.461}
\expandafter\gdef\csname odunum@n@axis_breakdown.m2.ming.axis2.2.4\endcsname{261}
\expandafter\gdef\csname odunum@ci@axis_breakdown.m2.ming.axis2.2.4\endcsname{[0.428, 0.494]}
\expandafter\gdef\csname odunum@val@axis_breakdown.m2.ming.axis2.2.5\endcsname{0.451}
\expandafter\gdef\csname odunum@n@axis_breakdown.m2.ming.axis2.2.5\endcsname{256}
\expandafter\gdef\csname odunum@ci@axis_breakdown.m2.ming.axis2.2.5\endcsname{[0.416, 0.487]}
\expandafter\gdef\csname odunum@val@axis_breakdown.m2.ming.axis3.3.1\endcsname{0.514}
\expandafter\gdef\csname odunum@n@axis_breakdown.m2.ming.axis3.3.1\endcsname{109}
\expandafter\gdef\csname odunum@ci@axis_breakdown.m2.ming.axis3.3.1\endcsname{[0.460, 0.568]}
\expandafter\gdef\csname odunum@val@axis_breakdown.m2.ming.axis3.3.2\endcsname{0.470}
\expandafter\gdef\csname odunum@n@axis_breakdown.m2.ming.axis3.3.2\endcsname{116}
\expandafter\gdef\csname odunum@ci@axis_breakdown.m2.ming.axis3.3.2\endcsname{[0.423, 0.515]}
\expandafter\gdef\csname odunum@val@axis_breakdown.m2.ming.axis3.3.3\endcsname{0.431}
\expandafter\gdef\csname odunum@n@axis_breakdown.m2.ming.axis3.3.3\endcsname{207}
\expandafter\gdef\csname odunum@ci@axis_breakdown.m2.ming.axis3.3.3\endcsname{[0.392, 0.471]}
\expandafter\gdef\csname odunum@val@axis_breakdown.m2.ming.axis3.3.4\endcsname{0.491}
\expandafter\gdef\csname odunum@n@axis_breakdown.m2.ming.axis3.3.4\endcsname{404}
\expandafter\gdef\csname odunum@ci@axis_breakdown.m2.ming.axis3.3.4\endcsname{[0.463, 0.519]}
\expandafter\gdef\csname odunum@val@axis_breakdown.m2.ming.axis3.3.5\endcsname{0.458}
\expandafter\gdef\csname odunum@n@axis_breakdown.m2.ming.axis3.3.5\endcsname{294}
\expandafter\gdef\csname odunum@ci@axis_breakdown.m2.ming.axis3.3.5\endcsname{[0.429, 0.488]}
\expandafter\gdef\csname odunum@val@axis_breakdown.m2.ming.axis3.3.6\endcsname{0.561}
\expandafter\gdef\csname odunum@n@axis_breakdown.m2.ming.axis3.3.6\endcsname{501}
\expandafter\gdef\csname odunum@ci@axis_breakdown.m2.ming.axis3.3.6\endcsname{[0.536, 0.585]}
\expandafter\gdef\csname odunum@val@axis_breakdown.m2.ming.axis4.4.1\endcsname{0.518}
\expandafter\gdef\csname odunum@n@axis_breakdown.m2.ming.axis4.4.1\endcsname{202}
\expandafter\gdef\csname odunum@ci@axis_breakdown.m2.ming.axis4.4.1\endcsname{[0.482, 0.554]}
\expandafter\gdef\csname odunum@val@axis_breakdown.m2.ming.axis4.4.2\endcsname{0.495}
\expandafter\gdef\csname odunum@n@axis_breakdown.m2.ming.axis4.4.2\endcsname{457}
\expandafter\gdef\csname odunum@ci@axis_breakdown.m2.ming.axis4.4.2\endcsname{[0.470, 0.519]}
\expandafter\gdef\csname odunum@val@axis_breakdown.m2.ming.axis4.4.3\endcsname{0.498}
\expandafter\gdef\csname odunum@n@axis_breakdown.m2.ming.axis4.4.3\endcsname{164}
\expandafter\gdef\csname odunum@ci@axis_breakdown.m2.ming.axis4.4.3\endcsname{[0.453, 0.542]}
\expandafter\gdef\csname odunum@val@axis_breakdown.m2.ming.axis4.4.4\endcsname{0.488}
\expandafter\gdef\csname odunum@n@axis_breakdown.m2.ming.axis4.4.4\endcsname{505}
\expandafter\gdef\csname odunum@ci@axis_breakdown.m2.ming.axis4.4.4\endcsname{[0.463, 0.511]}
\expandafter\gdef\csname odunum@val@axis_breakdown.m2.ming.axis4.4.5\endcsname{0.512}
\expandafter\gdef\csname odunum@n@axis_breakdown.m2.ming.axis4.4.5\endcsname{303}
\expandafter\gdef\csname odunum@ci@axis_breakdown.m2.ming.axis4.4.5\endcsname{[0.478, 0.546]}
\expandafter\gdef\csname odunum@val@axis_breakdown.m2.ming.axis5.audio_only.5.1\endcsname{0.532}
\expandafter\gdef\csname odunum@n@axis_breakdown.m2.ming.axis5.audio_only.5.1\endcsname{45}
\expandafter\gdef\csname odunum@ci@axis_breakdown.m2.ming.axis5.audio_only.5.1\endcsname{[0.448, 0.616]}
\expandafter\gdef\csname odunum@val@axis_breakdown.m2.ming.axis5.audio_only.5.10\endcsname{0.534}
\expandafter\gdef\csname odunum@n@axis_breakdown.m2.ming.axis5.audio_only.5.10\endcsname{72}
\expandafter\gdef\csname odunum@ci@axis_breakdown.m2.ming.axis5.audio_only.5.10\endcsname{[0.465, 0.601]}
\expandafter\gdef\csname odunum@val@axis_breakdown.m2.ming.axis5.audio_only.5.2\endcsname{0.452}
\expandafter\gdef\csname odunum@n@axis_breakdown.m2.ming.axis5.audio_only.5.2\endcsname{75}
\expandafter\gdef\csname odunum@ci@axis_breakdown.m2.ming.axis5.audio_only.5.2\endcsname{[0.383, 0.519]}
\expandafter\gdef\csname odunum@val@axis_breakdown.m2.ming.axis5.audio_only.5.3\endcsname{0.604}
\expandafter\gdef\csname odunum@n@axis_breakdown.m2.ming.axis5.audio_only.5.3\endcsname{50}
\expandafter\gdef\csname odunum@ci@axis_breakdown.m2.ming.axis5.audio_only.5.3\endcsname{[0.511, 0.695]}
\expandafter\gdef\csname odunum@val@axis_breakdown.m2.ming.axis5.audio_only.5.4\endcsname{0.488}
\expandafter\gdef\csname odunum@n@axis_breakdown.m2.ming.axis5.audio_only.5.4\endcsname{68}
\expandafter\gdef\csname odunum@ci@axis_breakdown.m2.ming.axis5.audio_only.5.4\endcsname{[0.413, 0.563]}
\expandafter\gdef\csname odunum@val@axis_breakdown.m2.ming.axis5.audio_only.5.5\endcsname{0.501}
\expandafter\gdef\csname odunum@n@axis_breakdown.m2.ming.axis5.audio_only.5.5\endcsname{52}
\expandafter\gdef\csname odunum@ci@axis_breakdown.m2.ming.axis5.audio_only.5.5\endcsname{[0.427, 0.579]}
\expandafter\gdef\csname odunum@val@axis_breakdown.m2.ming.axis5.audio_only.5.6\endcsname{0.540}
\expandafter\gdef\csname odunum@n@axis_breakdown.m2.ming.axis5.audio_only.5.6\endcsname{52}
\expandafter\gdef\csname odunum@ci@axis_breakdown.m2.ming.axis5.audio_only.5.6\endcsname{[0.473, 0.608]}
\expandafter\gdef\csname odunum@val@axis_breakdown.m2.ming.axis5.audio_only.5.7\endcsname{0.471}
\expandafter\gdef\csname odunum@n@axis_breakdown.m2.ming.axis5.audio_only.5.7\endcsname{46}
\expandafter\gdef\csname odunum@ci@axis_breakdown.m2.ming.axis5.audio_only.5.7\endcsname{[0.396, 0.548]}
\expandafter\gdef\csname odunum@val@axis_breakdown.m2.ming.axis5.audio_only.5.8\endcsname{0.432}
\expandafter\gdef\csname odunum@n@axis_breakdown.m2.ming.axis5.audio_only.5.8\endcsname{57}
\expandafter\gdef\csname odunum@ci@axis_breakdown.m2.ming.axis5.audio_only.5.8\endcsname{[0.346, 0.518]}
\expandafter\gdef\csname odunum@val@axis_breakdown.m2.ming.axis5.audio_only.5.9\endcsname{0.528}
\expandafter\gdef\csname odunum@n@axis_breakdown.m2.ming.axis5.audio_only.5.9\endcsname{53}
\expandafter\gdef\csname odunum@ci@axis_breakdown.m2.ming.axis5.audio_only.5.9\endcsname{[0.443, 0.613]}
\expandafter\gdef\csname odunum@val@axis_breakdown.m2.ming.axis5.audio_visual.5.1\endcsname{0.514}
\expandafter\gdef\csname odunum@n@axis_breakdown.m2.ming.axis5.audio_visual.5.1\endcsname{114}
\expandafter\gdef\csname odunum@ci@axis_breakdown.m2.ming.axis5.audio_visual.5.1\endcsname{[0.468, 0.560]}
\expandafter\gdef\csname odunum@val@axis_breakdown.m2.ming.axis5.audio_visual.5.10\endcsname{0.556}
\expandafter\gdef\csname odunum@n@axis_breakdown.m2.ming.axis5.audio_visual.5.10\endcsname{123}
\expandafter\gdef\csname odunum@ci@axis_breakdown.m2.ming.axis5.audio_visual.5.10\endcsname{[0.508, 0.603]}
\expandafter\gdef\csname odunum@val@axis_breakdown.m2.ming.axis5.audio_visual.5.2\endcsname{0.448}
\expandafter\gdef\csname odunum@n@axis_breakdown.m2.ming.axis5.audio_visual.5.2\endcsname{126}
\expandafter\gdef\csname odunum@ci@axis_breakdown.m2.ming.axis5.audio_visual.5.2\endcsname{[0.405, 0.491]}
\expandafter\gdef\csname odunum@val@axis_breakdown.m2.ming.axis5.audio_visual.5.3\endcsname{0.486}
\expandafter\gdef\csname odunum@n@axis_breakdown.m2.ming.axis5.audio_visual.5.3\endcsname{100}
\expandafter\gdef\csname odunum@ci@axis_breakdown.m2.ming.axis5.audio_visual.5.3\endcsname{[0.436, 0.538]}
\expandafter\gdef\csname odunum@val@axis_breakdown.m2.ming.axis5.audio_visual.5.4\endcsname{0.452}
\expandafter\gdef\csname odunum@n@axis_breakdown.m2.ming.axis5.audio_visual.5.4\endcsname{97}
\expandafter\gdef\csname odunum@ci@axis_breakdown.m2.ming.axis5.audio_visual.5.4\endcsname{[0.396, 0.510]}
\expandafter\gdef\csname odunum@val@axis_breakdown.m2.ming.axis5.audio_visual.5.5\endcsname{0.528}
\expandafter\gdef\csname odunum@n@axis_breakdown.m2.ming.axis5.audio_visual.5.5\endcsname{99}
\expandafter\gdef\csname odunum@ci@axis_breakdown.m2.ming.axis5.audio_visual.5.5\endcsname{[0.479, 0.576]}
\expandafter\gdef\csname odunum@val@axis_breakdown.m2.ming.axis5.audio_visual.5.6\endcsname{0.485}
\expandafter\gdef\csname odunum@n@axis_breakdown.m2.ming.axis5.audio_visual.5.6\endcsname{99}
\expandafter\gdef\csname odunum@ci@axis_breakdown.m2.ming.axis5.audio_visual.5.6\endcsname{[0.428, 0.541]}
\expandafter\gdef\csname odunum@val@axis_breakdown.m2.ming.axis5.audio_visual.5.7\endcsname{0.531}
\expandafter\gdef\csname odunum@n@axis_breakdown.m2.ming.axis5.audio_visual.5.7\endcsname{104}
\expandafter\gdef\csname odunum@ci@axis_breakdown.m2.ming.axis5.audio_visual.5.7\endcsname{[0.484, 0.580]}
\expandafter\gdef\csname odunum@val@axis_breakdown.m2.ming.axis5.audio_visual.5.8\endcsname{0.450}
\expandafter\gdef\csname odunum@n@axis_breakdown.m2.ming.axis5.audio_visual.5.8\endcsname{121}
\expandafter\gdef\csname odunum@ci@axis_breakdown.m2.ming.axis5.audio_visual.5.8\endcsname{[0.403, 0.498]}
\expandafter\gdef\csname odunum@val@axis_breakdown.m2.ming.axis5.audio_visual.5.9\endcsname{0.508}
\expandafter\gdef\csname odunum@n@axis_breakdown.m2.ming.axis5.audio_visual.5.9\endcsname{78}
\expandafter\gdef\csname odunum@ci@axis_breakdown.m2.ming.axis5.audio_visual.5.9\endcsname{[0.449, 0.572]}
\expandafter\gdef\csname odunum@val@axis_breakdown.m2.ming.axis6.6.1\endcsname{0.493}
\expandafter\gdef\csname odunum@n@axis_breakdown.m2.ming.axis6.6.1\endcsname{230}
\expandafter\gdef\csname odunum@ci@axis_breakdown.m2.ming.axis6.6.1\endcsname{[0.457, 0.530]}
\expandafter\gdef\csname odunum@val@axis_breakdown.m2.ming.axis6.6.10\endcsname{0.480}
\expandafter\gdef\csname odunum@n@axis_breakdown.m2.ming.axis6.6.10\endcsname{102}
\expandafter\gdef\csname odunum@ci@axis_breakdown.m2.ming.axis6.6.10\endcsname{[0.428, 0.532]}
\expandafter\gdef\csname odunum@val@axis_breakdown.m2.ming.axis6.6.12\endcsname{0.478}
\expandafter\gdef\csname odunum@n@axis_breakdown.m2.ming.axis6.6.12\endcsname{129}
\expandafter\gdef\csname odunum@ci@axis_breakdown.m2.ming.axis6.6.12\endcsname{[0.427, 0.531]}
\expandafter\gdef\csname odunum@val@axis_breakdown.m2.ming.axis6.6.13\endcsname{0.515}
\expandafter\gdef\csname odunum@n@axis_breakdown.m2.ming.axis6.6.13\endcsname{58}
\expandafter\gdef\csname odunum@ci@axis_breakdown.m2.ming.axis6.6.13\endcsname{[0.453, 0.579]}
\expandafter\gdef\csname odunum@val@axis_breakdown.m2.ming.axis6.6.14\endcsname{0.440}
\expandafter\gdef\csname odunum@n@axis_breakdown.m2.ming.axis6.6.14\endcsname{82}
\expandafter\gdef\csname odunum@ci@axis_breakdown.m2.ming.axis6.6.14\endcsname{[0.386, 0.494]}
\expandafter\gdef\csname odunum@val@axis_breakdown.m2.ming.axis6.6.18\endcsname{0.487}
\expandafter\gdef\csname odunum@n@axis_breakdown.m2.ming.axis6.6.18\endcsname{72}
\expandafter\gdef\csname odunum@ci@axis_breakdown.m2.ming.axis6.6.18\endcsname{[0.423, 0.549]}
\expandafter\gdef\csname odunum@val@axis_breakdown.m2.ming.axis6.6.19\endcsname{0.482}
\expandafter\gdef\csname odunum@n@axis_breakdown.m2.ming.axis6.6.19\endcsname{44}
\expandafter\gdef\csname odunum@ci@axis_breakdown.m2.ming.axis6.6.19\endcsname{[0.415, 0.552]}
\expandafter\gdef\csname odunum@val@axis_breakdown.m2.ming.axis6.6.2\endcsname{0.495}
\expandafter\gdef\csname odunum@n@axis_breakdown.m2.ming.axis6.6.2\endcsname{41}
\expandafter\gdef\csname odunum@ci@axis_breakdown.m2.ming.axis6.6.2\endcsname{[0.417, 0.575]}
\expandafter\gdef\csname odunum@val@axis_breakdown.m2.ming.axis6.6.3\endcsname{0.518}
\expandafter\gdef\csname odunum@n@axis_breakdown.m2.ming.axis6.6.3\endcsname{44}
\expandafter\gdef\csname odunum@ci@axis_breakdown.m2.ming.axis6.6.3\endcsname{[0.431, 0.605]}
\expandafter\gdef\csname odunum@val@axis_breakdown.m2.ming.axis6.6.5\endcsname{0.488}
\expandafter\gdef\csname odunum@n@axis_breakdown.m2.ming.axis6.6.5\endcsname{48}
\expandafter\gdef\csname odunum@ci@axis_breakdown.m2.ming.axis6.6.5\endcsname{[0.417, 0.560]}
\expandafter\gdef\csname odunum@val@axis_breakdown.m2.ming.axis6.6.6\endcsname{0.499}
\expandafter\gdef\csname odunum@n@axis_breakdown.m2.ming.axis6.6.6\endcsname{134}
\expandafter\gdef\csname odunum@ci@axis_breakdown.m2.ming.axis6.6.6\endcsname{[0.447, 0.550]}
\expandafter\gdef\csname odunum@val@axis_breakdown.m2.ming.axis6.6.7\endcsname{0.544}
\expandafter\gdef\csname odunum@n@axis_breakdown.m2.ming.axis6.6.7\endcsname{393}
\expandafter\gdef\csname odunum@ci@axis_breakdown.m2.ming.axis6.6.7\endcsname{[0.517, 0.571]}
\expandafter\gdef\csname odunum@val@axis_breakdown.m2.ming.axis6.6.8\endcsname{0.469}
\expandafter\gdef\csname odunum@n@axis_breakdown.m2.ming.axis6.6.8\endcsname{49}
\expandafter\gdef\csname odunum@ci@axis_breakdown.m2.ming.axis6.6.8\endcsname{[0.390, 0.545]}
\expandafter\gdef\csname odunum@val@axis_breakdown.m2.ming.axis6.6.9\endcsname{0.480}
\expandafter\gdef\csname odunum@n@axis_breakdown.m2.ming.axis6.6.9\endcsname{119}
\expandafter\gdef\csname odunum@ci@axis_breakdown.m2.ming.axis6.6.9\endcsname{[0.429, 0.532]}
\expandafter\gdef\csname odunum@val@axis_breakdown.m2.ming.difficulty.q1\endcsname{0.518}
\expandafter\gdef\csname odunum@n@axis_breakdown.m2.ming.difficulty.q1\endcsname{358}
\expandafter\gdef\csname odunum@ci@axis_breakdown.m2.ming.difficulty.q1\endcsname{[0.487, 0.551]}
\expandafter\gdef\csname odunum@val@axis_breakdown.m2.ming.difficulty.q2\endcsname{0.514}
\expandafter\gdef\csname odunum@n@axis_breakdown.m2.ming.difficulty.q2\endcsname{357}
\expandafter\gdef\csname odunum@ci@axis_breakdown.m2.ming.difficulty.q2\endcsname{[0.486, 0.542]}
\expandafter\gdef\csname odunum@val@axis_breakdown.m2.ming.difficulty.q3\endcsname{0.506}
\expandafter\gdef\csname odunum@n@axis_breakdown.m2.ming.difficulty.q3\endcsname{358}
\expandafter\gdef\csname odunum@ci@axis_breakdown.m2.ming.difficulty.q3\endcsname{[0.477, 0.534]}
\expandafter\gdef\csname odunum@val@axis_breakdown.m2.ming.difficulty.q4\endcsname{0.470}
\expandafter\gdef\csname odunum@n@axis_breakdown.m2.ming.difficulty.q4\endcsname{356}
\expandafter\gdef\csname odunum@ci@axis_breakdown.m2.ming.difficulty.q4\endcsname{[0.443, 0.498]}
\expandafter\gdef\csname odunum@val@axis_breakdown.m2.minicpm_o.axis1.audio_only.1.1\endcsname{0.503}
\expandafter\gdef\csname odunum@n@axis_breakdown.m2.minicpm_o.axis1.audio_only.1.1\endcsname{158}
\expandafter\gdef\csname odunum@ci@axis_breakdown.m2.minicpm_o.axis1.audio_only.1.1\endcsname{[0.442, 0.563]}
\expandafter\gdef\csname odunum@val@axis_breakdown.m2.minicpm_o.axis1.audio_only.1.2\endcsname{0.466}
\expandafter\gdef\csname odunum@n@axis_breakdown.m2.minicpm_o.axis1.audio_only.1.2\endcsname{378}
\expandafter\gdef\csname odunum@ci@axis_breakdown.m2.minicpm_o.axis1.audio_only.1.2\endcsname{[0.432, 0.499]}
\expandafter\gdef\csname odunum@val@axis_breakdown.m2.minicpm_o.axis1.audio_visual.1.1\endcsname{0.345}
\expandafter\gdef\csname odunum@n@axis_breakdown.m2.minicpm_o.axis1.audio_visual.1.1\endcsname{94}
\expandafter\gdef\csname odunum@ci@axis_breakdown.m2.minicpm_o.axis1.audio_visual.1.1\endcsname{[0.279, 0.410]}
\expandafter\gdef\csname odunum@val@axis_breakdown.m2.minicpm_o.axis1.audio_visual.1.2\endcsname{0.385}
\expandafter\gdef\csname odunum@n@axis_breakdown.m2.minicpm_o.axis1.audio_visual.1.2\endcsname{293}
\expandafter\gdef\csname odunum@ci@axis_breakdown.m2.minicpm_o.axis1.audio_visual.1.2\endcsname{[0.349, 0.422]}
\expandafter\gdef\csname odunum@val@axis_breakdown.m2.minicpm_o.axis1.audio_visual.1.3\endcsname{0.343}
\expandafter\gdef\csname odunum@n@axis_breakdown.m2.minicpm_o.axis1.audio_visual.1.3\endcsname{291}
\expandafter\gdef\csname odunum@ci@axis_breakdown.m2.minicpm_o.axis1.audio_visual.1.3\endcsname{[0.309, 0.377]}
\expandafter\gdef\csname odunum@val@axis_breakdown.m2.minicpm_o.axis1.audio_visual.1.4\endcsname{0.343}
\expandafter\gdef\csname odunum@n@axis_breakdown.m2.minicpm_o.axis1.audio_visual.1.4\endcsname{382}
\expandafter\gdef\csname odunum@ci@axis_breakdown.m2.minicpm_o.axis1.audio_visual.1.4\endcsname{[0.315, 0.372]}
\expandafter\gdef\csname odunum@val@axis_breakdown.m2.minicpm_o.axis2.2.1\endcsname{0.394}
\expandafter\gdef\csname odunum@n@axis_breakdown.m2.minicpm_o.axis2.2.1\endcsname{143}
\expandafter\gdef\csname odunum@ci@axis_breakdown.m2.minicpm_o.axis2.2.1\endcsname{[0.342, 0.447]}
\expandafter\gdef\csname odunum@val@axis_breakdown.m2.minicpm_o.axis2.2.2\endcsname{0.399}
\expandafter\gdef\csname odunum@n@axis_breakdown.m2.minicpm_o.axis2.2.2\endcsname{440}
\expandafter\gdef\csname odunum@ci@axis_breakdown.m2.minicpm_o.axis2.2.2\endcsname{[0.369, 0.429]}
\expandafter\gdef\csname odunum@val@axis_breakdown.m2.minicpm_o.axis2.2.3\endcsname{0.392}
\expandafter\gdef\csname odunum@n@axis_breakdown.m2.minicpm_o.axis2.2.3\endcsname{529}
\expandafter\gdef\csname odunum@ci@axis_breakdown.m2.minicpm_o.axis2.2.3\endcsname{[0.362, 0.422]}
\expandafter\gdef\csname odunum@val@axis_breakdown.m2.minicpm_o.axis2.2.4\endcsname{0.387}
\expandafter\gdef\csname odunum@n@axis_breakdown.m2.minicpm_o.axis2.2.4\endcsname{261}
\expandafter\gdef\csname odunum@ci@axis_breakdown.m2.minicpm_o.axis2.2.4\endcsname{[0.351, 0.422]}
\expandafter\gdef\csname odunum@val@axis_breakdown.m2.minicpm_o.axis2.2.5\endcsname{0.425}
\expandafter\gdef\csname odunum@n@axis_breakdown.m2.minicpm_o.axis2.2.5\endcsname{256}
\expandafter\gdef\csname odunum@ci@axis_breakdown.m2.minicpm_o.axis2.2.5\endcsname{[0.389, 0.462]}
\expandafter\gdef\csname odunum@val@axis_breakdown.m2.minicpm_o.axis3.3.1\endcsname{0.424}
\expandafter\gdef\csname odunum@n@axis_breakdown.m2.minicpm_o.axis3.3.1\endcsname{109}
\expandafter\gdef\csname odunum@ci@axis_breakdown.m2.minicpm_o.axis3.3.1\endcsname{[0.363, 0.486]}
\expandafter\gdef\csname odunum@val@axis_breakdown.m2.minicpm_o.axis3.3.2\endcsname{0.385}
\expandafter\gdef\csname odunum@n@axis_breakdown.m2.minicpm_o.axis3.3.2\endcsname{116}
\expandafter\gdef\csname odunum@ci@axis_breakdown.m2.minicpm_o.axis3.3.2\endcsname{[0.325, 0.443]}
\expandafter\gdef\csname odunum@val@axis_breakdown.m2.minicpm_o.axis3.3.3\endcsname{0.314}
\expandafter\gdef\csname odunum@n@axis_breakdown.m2.minicpm_o.axis3.3.3\endcsname{207}
\expandafter\gdef\csname odunum@ci@axis_breakdown.m2.minicpm_o.axis3.3.3\endcsname{[0.269, 0.361]}
\expandafter\gdef\csname odunum@val@axis_breakdown.m2.minicpm_o.axis3.3.4\endcsname{0.441}
\expandafter\gdef\csname odunum@n@axis_breakdown.m2.minicpm_o.axis3.3.4\endcsname{404}
\expandafter\gdef\csname odunum@ci@axis_breakdown.m2.minicpm_o.axis3.3.4\endcsname{[0.407, 0.475]}
\expandafter\gdef\csname odunum@val@axis_breakdown.m2.minicpm_o.axis3.3.5\endcsname{0.382}
\expandafter\gdef\csname odunum@n@axis_breakdown.m2.minicpm_o.axis3.3.5\endcsname{294}
\expandafter\gdef\csname odunum@ci@axis_breakdown.m2.minicpm_o.axis3.3.5\endcsname{[0.348, 0.415]}
\expandafter\gdef\csname odunum@val@axis_breakdown.m2.minicpm_o.axis3.3.6\endcsname{0.405}
\expandafter\gdef\csname odunum@n@axis_breakdown.m2.minicpm_o.axis3.3.6\endcsname{499}
\expandafter\gdef\csname odunum@ci@axis_breakdown.m2.minicpm_o.axis3.3.6\endcsname{[0.377, 0.433]}
\expandafter\gdef\csname odunum@val@axis_breakdown.m2.minicpm_o.axis4.4.1\endcsname{0.354}
\expandafter\gdef\csname odunum@n@axis_breakdown.m2.minicpm_o.axis4.4.1\endcsname{201}
\expandafter\gdef\csname odunum@ci@axis_breakdown.m2.minicpm_o.axis4.4.1\endcsname{[0.312, 0.396]}
\expandafter\gdef\csname odunum@val@axis_breakdown.m2.minicpm_o.axis4.4.2\endcsname{0.419}
\expandafter\gdef\csname odunum@n@axis_breakdown.m2.minicpm_o.axis4.4.2\endcsname{457}
\expandafter\gdef\csname odunum@ci@axis_breakdown.m2.minicpm_o.axis4.4.2\endcsname{[0.390, 0.449]}
\expandafter\gdef\csname odunum@val@axis_breakdown.m2.minicpm_o.axis4.4.3\endcsname{0.327}
\expandafter\gdef\csname odunum@n@axis_breakdown.m2.minicpm_o.axis4.4.3\endcsname{164}
\expandafter\gdef\csname odunum@ci@axis_breakdown.m2.minicpm_o.axis4.4.3\endcsname{[0.281, 0.373]}
\expandafter\gdef\csname odunum@val@axis_breakdown.m2.minicpm_o.axis4.4.4\endcsname{0.387}
\expandafter\gdef\csname odunum@n@axis_breakdown.m2.minicpm_o.axis4.4.4\endcsname{505}
\expandafter\gdef\csname odunum@ci@axis_breakdown.m2.minicpm_o.axis4.4.4\endcsname{[0.359, 0.415]}
\expandafter\gdef\csname odunum@val@axis_breakdown.m2.minicpm_o.axis4.4.5\endcsname{0.453}
\expandafter\gdef\csname odunum@n@axis_breakdown.m2.minicpm_o.axis4.4.5\endcsname{302}
\expandafter\gdef\csname odunum@ci@axis_breakdown.m2.minicpm_o.axis4.4.5\endcsname{[0.415, 0.492]}
\expandafter\gdef\csname odunum@val@axis_breakdown.m2.minicpm_o.axis5.audio_only.5.1\endcsname{0.587}
\expandafter\gdef\csname odunum@n@axis_breakdown.m2.minicpm_o.axis5.audio_only.5.1\endcsname{45}
\expandafter\gdef\csname odunum@ci@axis_breakdown.m2.minicpm_o.axis5.audio_only.5.1\endcsname{[0.479, 0.689]}
\expandafter\gdef\csname odunum@val@axis_breakdown.m2.minicpm_o.axis5.audio_only.5.10\endcsname{0.468}
\expandafter\gdef\csname odunum@n@axis_breakdown.m2.minicpm_o.axis5.audio_only.5.10\endcsname{72}
\expandafter\gdef\csname odunum@ci@axis_breakdown.m2.minicpm_o.axis5.audio_only.5.10\endcsname{[0.385, 0.549]}
\expandafter\gdef\csname odunum@val@axis_breakdown.m2.minicpm_o.axis5.audio_only.5.2\endcsname{0.470}
\expandafter\gdef\csname odunum@n@axis_breakdown.m2.minicpm_o.axis5.audio_only.5.2\endcsname{75}
\expandafter\gdef\csname odunum@ci@axis_breakdown.m2.minicpm_o.axis5.audio_only.5.2\endcsname{[0.394, 0.547]}
\expandafter\gdef\csname odunum@val@axis_breakdown.m2.minicpm_o.axis5.audio_only.5.3\endcsname{0.534}
\expandafter\gdef\csname odunum@n@axis_breakdown.m2.minicpm_o.axis5.audio_only.5.3\endcsname{50}
\expandafter\gdef\csname odunum@ci@axis_breakdown.m2.minicpm_o.axis5.audio_only.5.3\endcsname{[0.433, 0.633]}
\expandafter\gdef\csname odunum@val@axis_breakdown.m2.minicpm_o.axis5.audio_only.5.4\endcsname{0.441}
\expandafter\gdef\csname odunum@n@axis_breakdown.m2.minicpm_o.axis5.audio_only.5.4\endcsname{68}
\expandafter\gdef\csname odunum@ci@axis_breakdown.m2.minicpm_o.axis5.audio_only.5.4\endcsname{[0.365, 0.517]}
\expandafter\gdef\csname odunum@val@axis_breakdown.m2.minicpm_o.axis5.audio_only.5.5\endcsname{0.415}
\expandafter\gdef\csname odunum@n@axis_breakdown.m2.minicpm_o.axis5.audio_only.5.5\endcsname{51}
\expandafter\gdef\csname odunum@ci@axis_breakdown.m2.minicpm_o.axis5.audio_only.5.5\endcsname{[0.327, 0.502]}
\expandafter\gdef\csname odunum@val@axis_breakdown.m2.minicpm_o.axis5.audio_only.5.6\endcsname{0.464}
\expandafter\gdef\csname odunum@n@axis_breakdown.m2.minicpm_o.axis5.audio_only.5.6\endcsname{52}
\expandafter\gdef\csname odunum@ci@axis_breakdown.m2.minicpm_o.axis5.audio_only.5.6\endcsname{[0.368, 0.558]}
\expandafter\gdef\csname odunum@val@axis_breakdown.m2.minicpm_o.axis5.audio_only.5.7\endcsname{0.467}
\expandafter\gdef\csname odunum@n@axis_breakdown.m2.minicpm_o.axis5.audio_only.5.7\endcsname{46}
\expandafter\gdef\csname odunum@ci@axis_breakdown.m2.minicpm_o.axis5.audio_only.5.7\endcsname{[0.365, 0.566]}
\expandafter\gdef\csname odunum@val@axis_breakdown.m2.minicpm_o.axis5.audio_only.5.8\endcsname{0.453}
\expandafter\gdef\csname odunum@n@axis_breakdown.m2.minicpm_o.axis5.audio_only.5.8\endcsname{57}
\expandafter\gdef\csname odunum@ci@axis_breakdown.m2.minicpm_o.axis5.audio_only.5.8\endcsname{[0.366, 0.540]}
\expandafter\gdef\csname odunum@val@axis_breakdown.m2.minicpm_o.axis5.audio_only.5.9\endcsname{0.525}
\expandafter\gdef\csname odunum@n@axis_breakdown.m2.minicpm_o.axis5.audio_only.5.9\endcsname{53}
\expandafter\gdef\csname odunum@ci@axis_breakdown.m2.minicpm_o.axis5.audio_only.5.9\endcsname{[0.426, 0.622]}
\expandafter\gdef\csname odunum@val@axis_breakdown.m2.minicpm_o.axis5.audio_visual.5.1\endcsname{0.391}
\expandafter\gdef\csname odunum@n@axis_breakdown.m2.minicpm_o.axis5.audio_visual.5.1\endcsname{114}
\expandafter\gdef\csname odunum@ci@axis_breakdown.m2.minicpm_o.axis5.audio_visual.5.1\endcsname{[0.328, 0.455]}
\expandafter\gdef\csname odunum@val@axis_breakdown.m2.minicpm_o.axis5.audio_visual.5.10\endcsname{0.338}
\expandafter\gdef\csname odunum@n@axis_breakdown.m2.minicpm_o.axis5.audio_visual.5.10\endcsname{123}
\expandafter\gdef\csname odunum@ci@axis_breakdown.m2.minicpm_o.axis5.audio_visual.5.10\endcsname{[0.284, 0.392]}
\expandafter\gdef\csname odunum@val@axis_breakdown.m2.minicpm_o.axis5.audio_visual.5.2\endcsname{0.391}
\expandafter\gdef\csname odunum@n@axis_breakdown.m2.minicpm_o.axis5.audio_visual.5.2\endcsname{126}
\expandafter\gdef\csname odunum@ci@axis_breakdown.m2.minicpm_o.axis5.audio_visual.5.2\endcsname{[0.340, 0.443]}
\expandafter\gdef\csname odunum@val@axis_breakdown.m2.minicpm_o.axis5.audio_visual.5.3\endcsname{0.337}
\expandafter\gdef\csname odunum@n@axis_breakdown.m2.minicpm_o.axis5.audio_visual.5.3\endcsname{99}
\expandafter\gdef\csname odunum@ci@axis_breakdown.m2.minicpm_o.axis5.audio_visual.5.3\endcsname{[0.283, 0.394]}
\expandafter\gdef\csname odunum@val@axis_breakdown.m2.minicpm_o.axis5.audio_visual.5.4\endcsname{0.353}
\expandafter\gdef\csname odunum@n@axis_breakdown.m2.minicpm_o.axis5.audio_visual.5.4\endcsname{97}
\expandafter\gdef\csname odunum@ci@axis_breakdown.m2.minicpm_o.axis5.audio_visual.5.4\endcsname{[0.293, 0.414]}
\expandafter\gdef\csname odunum@val@axis_breakdown.m2.minicpm_o.axis5.audio_visual.5.5\endcsname{0.309}
\expandafter\gdef\csname odunum@n@axis_breakdown.m2.minicpm_o.axis5.audio_visual.5.5\endcsname{99}
\expandafter\gdef\csname odunum@ci@axis_breakdown.m2.minicpm_o.axis5.audio_visual.5.5\endcsname{[0.252, 0.369]}
\expandafter\gdef\csname odunum@val@axis_breakdown.m2.minicpm_o.axis5.audio_visual.5.6\endcsname{0.392}
\expandafter\gdef\csname odunum@n@axis_breakdown.m2.minicpm_o.axis5.audio_visual.5.6\endcsname{99}
\expandafter\gdef\csname odunum@ci@axis_breakdown.m2.minicpm_o.axis5.audio_visual.5.6\endcsname{[0.335, 0.452]}
\expandafter\gdef\csname odunum@val@axis_breakdown.m2.minicpm_o.axis5.audio_visual.5.7\endcsname{0.293}
\expandafter\gdef\csname odunum@n@axis_breakdown.m2.minicpm_o.axis5.audio_visual.5.7\endcsname{104}
\expandafter\gdef\csname odunum@ci@axis_breakdown.m2.minicpm_o.axis5.audio_visual.5.7\endcsname{[0.234, 0.356]}
\expandafter\gdef\csname odunum@val@axis_breakdown.m2.minicpm_o.axis5.audio_visual.5.8\endcsname{0.343}
\expandafter\gdef\csname odunum@n@axis_breakdown.m2.minicpm_o.axis5.audio_visual.5.8\endcsname{121}
\expandafter\gdef\csname odunum@ci@axis_breakdown.m2.minicpm_o.axis5.audio_visual.5.8\endcsname{[0.288, 0.398]}
\expandafter\gdef\csname odunum@val@axis_breakdown.m2.minicpm_o.axis5.audio_visual.5.9\endcsname{0.407}
\expandafter\gdef\csname odunum@n@axis_breakdown.m2.minicpm_o.axis5.audio_visual.5.9\endcsname{78}
\expandafter\gdef\csname odunum@ci@axis_breakdown.m2.minicpm_o.axis5.audio_visual.5.9\endcsname{[0.349, 0.465]}
\expandafter\gdef\csname odunum@val@axis_breakdown.m2.minicpm_o.axis6.6.1\endcsname{0.364}
\expandafter\gdef\csname odunum@n@axis_breakdown.m2.minicpm_o.axis6.6.1\endcsname{230}
\expandafter\gdef\csname odunum@ci@axis_breakdown.m2.minicpm_o.axis6.6.1\endcsname{[0.322, 0.407]}
\expandafter\gdef\csname odunum@val@axis_breakdown.m2.minicpm_o.axis6.6.10\endcsname{0.418}
\expandafter\gdef\csname odunum@n@axis_breakdown.m2.minicpm_o.axis6.6.10\endcsname{102}
\expandafter\gdef\csname odunum@ci@axis_breakdown.m2.minicpm_o.axis6.6.10\endcsname{[0.355, 0.484]}
\expandafter\gdef\csname odunum@val@axis_breakdown.m2.minicpm_o.axis6.6.12\endcsname{0.407}
\expandafter\gdef\csname odunum@n@axis_breakdown.m2.minicpm_o.axis6.6.12\endcsname{128}
\expandafter\gdef\csname odunum@ci@axis_breakdown.m2.minicpm_o.axis6.6.12\endcsname{[0.348, 0.466]}
\expandafter\gdef\csname odunum@val@axis_breakdown.m2.minicpm_o.axis6.6.13\endcsname{0.363}
\expandafter\gdef\csname odunum@n@axis_breakdown.m2.minicpm_o.axis6.6.13\endcsname{58}
\expandafter\gdef\csname odunum@ci@axis_breakdown.m2.minicpm_o.axis6.6.13\endcsname{[0.281, 0.445]}
\expandafter\gdef\csname odunum@val@axis_breakdown.m2.minicpm_o.axis6.6.14\endcsname{0.375}
\expandafter\gdef\csname odunum@n@axis_breakdown.m2.minicpm_o.axis6.6.14\endcsname{82}
\expandafter\gdef\csname odunum@ci@axis_breakdown.m2.minicpm_o.axis6.6.14\endcsname{[0.316, 0.435]}
\expandafter\gdef\csname odunum@val@axis_breakdown.m2.minicpm_o.axis6.6.18\endcsname{0.423}
\expandafter\gdef\csname odunum@n@axis_breakdown.m2.minicpm_o.axis6.6.18\endcsname{72}
\expandafter\gdef\csname odunum@ci@axis_breakdown.m2.minicpm_o.axis6.6.18\endcsname{[0.362, 0.484]}
\expandafter\gdef\csname odunum@val@axis_breakdown.m2.minicpm_o.axis6.6.19\endcsname{0.430}
\expandafter\gdef\csname odunum@n@axis_breakdown.m2.minicpm_o.axis6.6.19\endcsname{44}
\expandafter\gdef\csname odunum@ci@axis_breakdown.m2.minicpm_o.axis6.6.19\endcsname{[0.339, 0.520]}
\expandafter\gdef\csname odunum@val@axis_breakdown.m2.minicpm_o.axis6.6.2\endcsname{0.480}
\expandafter\gdef\csname odunum@n@axis_breakdown.m2.minicpm_o.axis6.6.2\endcsname{41}
\expandafter\gdef\csname odunum@ci@axis_breakdown.m2.minicpm_o.axis6.6.2\endcsname{[0.386, 0.572]}
\expandafter\gdef\csname odunum@val@axis_breakdown.m2.minicpm_o.axis6.6.3\endcsname{0.389}
\expandafter\gdef\csname odunum@n@axis_breakdown.m2.minicpm_o.axis6.6.3\endcsname{44}
\expandafter\gdef\csname odunum@ci@axis_breakdown.m2.minicpm_o.axis6.6.3\endcsname{[0.298, 0.485]}
\expandafter\gdef\csname odunum@val@axis_breakdown.m2.minicpm_o.axis6.6.5\endcsname{0.366}
\expandafter\gdef\csname odunum@n@axis_breakdown.m2.minicpm_o.axis6.6.5\endcsname{48}
\expandafter\gdef\csname odunum@ci@axis_breakdown.m2.minicpm_o.axis6.6.5\endcsname{[0.270, 0.463]}
\expandafter\gdef\csname odunum@val@axis_breakdown.m2.minicpm_o.axis6.6.6\endcsname{0.440}
\expandafter\gdef\csname odunum@n@axis_breakdown.m2.minicpm_o.axis6.6.6\endcsname{134}
\expandafter\gdef\csname odunum@ci@axis_breakdown.m2.minicpm_o.axis6.6.6\endcsname{[0.384, 0.496]}
\expandafter\gdef\csname odunum@val@axis_breakdown.m2.minicpm_o.axis6.6.7\endcsname{0.431}
\expandafter\gdef\csname odunum@n@axis_breakdown.m2.minicpm_o.axis6.6.7\endcsname{392}
\expandafter\gdef\csname odunum@ci@axis_breakdown.m2.minicpm_o.axis6.6.7\endcsname{[0.398, 0.465]}
\expandafter\gdef\csname odunum@val@axis_breakdown.m2.minicpm_o.axis6.6.8\endcsname{0.338}
\expandafter\gdef\csname odunum@n@axis_breakdown.m2.minicpm_o.axis6.6.8\endcsname{49}
\expandafter\gdef\csname odunum@ci@axis_breakdown.m2.minicpm_o.axis6.6.8\endcsname{[0.239, 0.438]}
\expandafter\gdef\csname odunum@val@axis_breakdown.m2.minicpm_o.axis6.6.9\endcsname{0.374}
\expandafter\gdef\csname odunum@n@axis_breakdown.m2.minicpm_o.axis6.6.9\endcsname{119}
\expandafter\gdef\csname odunum@ci@axis_breakdown.m2.minicpm_o.axis6.6.9\endcsname{[0.314, 0.433]}
\expandafter\gdef\csname odunum@val@axis_breakdown.m2.minicpm_o.difficulty.q1\endcsname{0.491}
\expandafter\gdef\csname odunum@n@axis_breakdown.m2.minicpm_o.difficulty.q1\endcsname{357}
\expandafter\gdef\csname odunum@ci@axis_breakdown.m2.minicpm_o.difficulty.q1\endcsname{[0.455, 0.527]}
\expandafter\gdef\csname odunum@val@axis_breakdown.m2.minicpm_o.difficulty.q2\endcsname{0.434}
\expandafter\gdef\csname odunum@n@axis_breakdown.m2.minicpm_o.difficulty.q2\endcsname{357}
\expandafter\gdef\csname odunum@ci@axis_breakdown.m2.minicpm_o.difficulty.q2\endcsname{[0.401, 0.467]}
\expandafter\gdef\csname odunum@val@axis_breakdown.m2.minicpm_o.difficulty.q3\endcsname{0.398}
\expandafter\gdef\csname odunum@n@axis_breakdown.m2.minicpm_o.difficulty.q3\endcsname{358}
\expandafter\gdef\csname odunum@ci@axis_breakdown.m2.minicpm_o.difficulty.q3\endcsname{[0.365, 0.431]}
\expandafter\gdef\csname odunum@val@axis_breakdown.m2.minicpm_o.difficulty.q4\endcsname{0.356}
\expandafter\gdef\csname odunum@n@axis_breakdown.m2.minicpm_o.difficulty.q4\endcsname{355}
\expandafter\gdef\csname odunum@ci@axis_breakdown.m2.minicpm_o.difficulty.q4\endcsname{[0.327, 0.385]}
\expandafter\gdef\csname odunum@val@axis_breakdown.m2.nemotron.axis1.audio_only.1.1\endcsname{0.340}
\expandafter\gdef\csname odunum@n@axis_breakdown.m2.nemotron.axis1.audio_only.1.1\endcsname{158}
\expandafter\gdef\csname odunum@ci@axis_breakdown.m2.nemotron.axis1.audio_only.1.1\endcsname{[0.281, 0.400]}
\expandafter\gdef\csname odunum@val@axis_breakdown.m2.nemotron.axis1.audio_only.1.2\endcsname{0.336}
\expandafter\gdef\csname odunum@n@axis_breakdown.m2.nemotron.axis1.audio_only.1.2\endcsname{379}
\expandafter\gdef\csname odunum@ci@axis_breakdown.m2.nemotron.axis1.audio_only.1.2\endcsname{[0.300, 0.372]}
\expandafter\gdef\csname odunum@val@axis_breakdown.m2.nemotron.axis1.audio_visual.1.1\endcsname{0.209}
\expandafter\gdef\csname odunum@n@axis_breakdown.m2.nemotron.axis1.audio_visual.1.1\endcsname{94}
\expandafter\gdef\csname odunum@ci@axis_breakdown.m2.nemotron.axis1.audio_visual.1.1\endcsname{[0.150, 0.272]}
\expandafter\gdef\csname odunum@val@axis_breakdown.m2.nemotron.axis1.audio_visual.1.2\endcsname{0.284}
\expandafter\gdef\csname odunum@n@axis_breakdown.m2.nemotron.axis1.audio_visual.1.2\endcsname{294}
\expandafter\gdef\csname odunum@ci@axis_breakdown.m2.nemotron.axis1.audio_visual.1.2\endcsname{[0.248, 0.321]}
\expandafter\gdef\csname odunum@val@axis_breakdown.m2.nemotron.axis1.audio_visual.1.3\endcsname{0.233}
\expandafter\gdef\csname odunum@n@axis_breakdown.m2.nemotron.axis1.audio_visual.1.3\endcsname{291}
\expandafter\gdef\csname odunum@ci@axis_breakdown.m2.nemotron.axis1.audio_visual.1.3\endcsname{[0.198, 0.269]}
\expandafter\gdef\csname odunum@val@axis_breakdown.m2.nemotron.axis1.audio_visual.1.4\endcsname{0.295}
\expandafter\gdef\csname odunum@n@axis_breakdown.m2.nemotron.axis1.audio_visual.1.4\endcsname{382}
\expandafter\gdef\csname odunum@ci@axis_breakdown.m2.nemotron.axis1.audio_visual.1.4\endcsname{[0.262, 0.328]}
\expandafter\gdef\csname odunum@val@axis_breakdown.m2.nemotron.axis2.2.1\endcsname{0.289}
\expandafter\gdef\csname odunum@n@axis_breakdown.m2.nemotron.axis2.2.1\endcsname{143}
\expandafter\gdef\csname odunum@ci@axis_breakdown.m2.nemotron.axis2.2.1\endcsname{[0.236, 0.343]}
\expandafter\gdef\csname odunum@val@axis_breakdown.m2.nemotron.axis2.2.2\endcsname{0.313}
\expandafter\gdef\csname odunum@n@axis_breakdown.m2.nemotron.axis2.2.2\endcsname{440}
\expandafter\gdef\csname odunum@ci@axis_breakdown.m2.nemotron.axis2.2.2\endcsname{[0.282, 0.345]}
\expandafter\gdef\csname odunum@val@axis_breakdown.m2.nemotron.axis2.2.3\endcsname{0.299}
\expandafter\gdef\csname odunum@n@axis_breakdown.m2.nemotron.axis2.2.3\endcsname{531}
\expandafter\gdef\csname odunum@ci@axis_breakdown.m2.nemotron.axis2.2.3\endcsname{[0.271, 0.329]}
\expandafter\gdef\csname odunum@val@axis_breakdown.m2.nemotron.axis2.2.4\endcsname{0.275}
\expandafter\gdef\csname odunum@n@axis_breakdown.m2.nemotron.axis2.2.4\endcsname{261}
\expandafter\gdef\csname odunum@ci@axis_breakdown.m2.nemotron.axis2.2.4\endcsname{[0.234, 0.315]}
\expandafter\gdef\csname odunum@val@axis_breakdown.m2.nemotron.axis2.2.5\endcsname{0.261}
\expandafter\gdef\csname odunum@n@axis_breakdown.m2.nemotron.axis2.2.5\endcsname{256}
\expandafter\gdef\csname odunum@ci@axis_breakdown.m2.nemotron.axis2.2.5\endcsname{[0.224, 0.301]}
\expandafter\gdef\csname odunum@val@axis_breakdown.m2.nemotron.axis3.3.1\endcsname{0.303}
\expandafter\gdef\csname odunum@n@axis_breakdown.m2.nemotron.axis3.3.1\endcsname{109}
\expandafter\gdef\csname odunum@ci@axis_breakdown.m2.nemotron.axis3.3.1\endcsname{[0.240, 0.369]}
\expandafter\gdef\csname odunum@val@axis_breakdown.m2.nemotron.axis3.3.2\endcsname{0.300}
\expandafter\gdef\csname odunum@n@axis_breakdown.m2.nemotron.axis3.3.2\endcsname{116}
\expandafter\gdef\csname odunum@ci@axis_breakdown.m2.nemotron.axis3.3.2\endcsname{[0.243, 0.361]}
\expandafter\gdef\csname odunum@val@axis_breakdown.m2.nemotron.axis3.3.3\endcsname{0.180}
\expandafter\gdef\csname odunum@n@axis_breakdown.m2.nemotron.axis3.3.3\endcsname{207}
\expandafter\gdef\csname odunum@ci@axis_breakdown.m2.nemotron.axis3.3.3\endcsname{[0.141, 0.221]}
\expandafter\gdef\csname odunum@val@axis_breakdown.m2.nemotron.axis3.3.4\endcsname{0.319}
\expandafter\gdef\csname odunum@n@axis_breakdown.m2.nemotron.axis3.3.4\endcsname{404}
\expandafter\gdef\csname odunum@ci@axis_breakdown.m2.nemotron.axis3.3.4\endcsname{[0.285, 0.354]}
\expandafter\gdef\csname odunum@val@axis_breakdown.m2.nemotron.axis3.3.5\endcsname{0.329}
\expandafter\gdef\csname odunum@n@axis_breakdown.m2.nemotron.axis3.3.5\endcsname{294}
\expandafter\gdef\csname odunum@ci@axis_breakdown.m2.nemotron.axis3.3.5\endcsname{[0.291, 0.367]}
\expandafter\gdef\csname odunum@val@axis_breakdown.m2.nemotron.axis3.3.6\endcsname{0.291}
\expandafter\gdef\csname odunum@n@axis_breakdown.m2.nemotron.axis3.3.6\endcsname{501}
\expandafter\gdef\csname odunum@ci@axis_breakdown.m2.nemotron.axis3.3.6\endcsname{[0.262, 0.321]}
\expandafter\gdef\csname odunum@val@axis_breakdown.m2.nemotron.axis4.4.1\endcsname{0.305}
\expandafter\gdef\csname odunum@n@axis_breakdown.m2.nemotron.axis4.4.1\endcsname{202}
\expandafter\gdef\csname odunum@ci@axis_breakdown.m2.nemotron.axis4.4.1\endcsname{[0.259, 0.351]}
\expandafter\gdef\csname odunum@val@axis_breakdown.m2.nemotron.axis4.4.2\endcsname{0.316}
\expandafter\gdef\csname odunum@n@axis_breakdown.m2.nemotron.axis4.4.2\endcsname{457}
\expandafter\gdef\csname odunum@ci@axis_breakdown.m2.nemotron.axis4.4.2\endcsname{[0.286, 0.347]}
\expandafter\gdef\csname odunum@val@axis_breakdown.m2.nemotron.axis4.4.3\endcsname{0.254}
\expandafter\gdef\csname odunum@n@axis_breakdown.m2.nemotron.axis4.4.3\endcsname{164}
\expandafter\gdef\csname odunum@ci@axis_breakdown.m2.nemotron.axis4.4.3\endcsname{[0.204, 0.307]}
\expandafter\gdef\csname odunum@val@axis_breakdown.m2.nemotron.axis4.4.4\endcsname{0.281}
\expandafter\gdef\csname odunum@n@axis_breakdown.m2.nemotron.axis4.4.4\endcsname{505}
\expandafter\gdef\csname odunum@ci@axis_breakdown.m2.nemotron.axis4.4.4\endcsname{[0.252, 0.310]}
\expandafter\gdef\csname odunum@val@axis_breakdown.m2.nemotron.axis4.4.5\endcsname{0.288}
\expandafter\gdef\csname odunum@n@axis_breakdown.m2.nemotron.axis4.4.5\endcsname{303}
\expandafter\gdef\csname odunum@ci@axis_breakdown.m2.nemotron.axis4.4.5\endcsname{[0.251, 0.326]}
\expandafter\gdef\csname odunum@val@axis_breakdown.m2.nemotron.axis5.audio_only.5.1\endcsname{0.358}
\expandafter\gdef\csname odunum@n@axis_breakdown.m2.nemotron.axis5.audio_only.5.1\endcsname{45}
\expandafter\gdef\csname odunum@ci@axis_breakdown.m2.nemotron.axis5.audio_only.5.1\endcsname{[0.249, 0.472]}
\expandafter\gdef\csname odunum@val@axis_breakdown.m2.nemotron.axis5.audio_only.5.10\endcsname{0.351}
\expandafter\gdef\csname odunum@n@axis_breakdown.m2.nemotron.axis5.audio_only.5.10\endcsname{72}
\expandafter\gdef\csname odunum@ci@axis_breakdown.m2.nemotron.axis5.audio_only.5.10\endcsname{[0.272, 0.436]}
\expandafter\gdef\csname odunum@val@axis_breakdown.m2.nemotron.axis5.audio_only.5.2\endcsname{0.359}
\expandafter\gdef\csname odunum@n@axis_breakdown.m2.nemotron.axis5.audio_only.5.2\endcsname{75}
\expandafter\gdef\csname odunum@ci@axis_breakdown.m2.nemotron.axis5.audio_only.5.2\endcsname{[0.277, 0.445]}
\expandafter\gdef\csname odunum@val@axis_breakdown.m2.nemotron.axis5.audio_only.5.3\endcsname{0.322}
\expandafter\gdef\csname odunum@n@axis_breakdown.m2.nemotron.axis5.audio_only.5.3\endcsname{50}
\expandafter\gdef\csname odunum@ci@axis_breakdown.m2.nemotron.axis5.audio_only.5.3\endcsname{[0.226, 0.424]}
\expandafter\gdef\csname odunum@val@axis_breakdown.m2.nemotron.axis5.audio_only.5.4\endcsname{0.355}
\expandafter\gdef\csname odunum@n@axis_breakdown.m2.nemotron.axis5.audio_only.5.4\endcsname{68}
\expandafter\gdef\csname odunum@ci@axis_breakdown.m2.nemotron.axis5.audio_only.5.4\endcsname{[0.271, 0.441]}
\expandafter\gdef\csname odunum@val@axis_breakdown.m2.nemotron.axis5.audio_only.5.5\endcsname{0.280}
\expandafter\gdef\csname odunum@n@axis_breakdown.m2.nemotron.axis5.audio_only.5.5\endcsname{52}
\expandafter\gdef\csname odunum@ci@axis_breakdown.m2.nemotron.axis5.audio_only.5.5\endcsname{[0.189, 0.378]}
\expandafter\gdef\csname odunum@val@axis_breakdown.m2.nemotron.axis5.audio_only.5.6\endcsname{0.419}
\expandafter\gdef\csname odunum@n@axis_breakdown.m2.nemotron.axis5.audio_only.5.6\endcsname{52}
\expandafter\gdef\csname odunum@ci@axis_breakdown.m2.nemotron.axis5.audio_only.5.6\endcsname{[0.321, 0.515]}
\expandafter\gdef\csname odunum@val@axis_breakdown.m2.nemotron.axis5.audio_only.5.7\endcsname{0.358}
\expandafter\gdef\csname odunum@n@axis_breakdown.m2.nemotron.axis5.audio_only.5.7\endcsname{46}
\expandafter\gdef\csname odunum@ci@axis_breakdown.m2.nemotron.axis5.audio_only.5.7\endcsname{[0.255, 0.466]}
\expandafter\gdef\csname odunum@val@axis_breakdown.m2.nemotron.axis5.audio_only.5.8\endcsname{0.242}
\expandafter\gdef\csname odunum@n@axis_breakdown.m2.nemotron.axis5.audio_only.5.8\endcsname{57}
\expandafter\gdef\csname odunum@ci@axis_breakdown.m2.nemotron.axis5.audio_only.5.8\endcsname{[0.167, 0.322]}
\expandafter\gdef\csname odunum@val@axis_breakdown.m2.nemotron.axis5.audio_only.5.9\endcsname{0.329}
\expandafter\gdef\csname odunum@n@axis_breakdown.m2.nemotron.axis5.audio_only.5.9\endcsname{53}
\expandafter\gdef\csname odunum@ci@axis_breakdown.m2.nemotron.axis5.audio_only.5.9\endcsname{[0.229, 0.437]}
\expandafter\gdef\csname odunum@val@axis_breakdown.m2.nemotron.axis5.audio_visual.5.1\endcsname{0.245}
\expandafter\gdef\csname odunum@n@axis_breakdown.m2.nemotron.axis5.audio_visual.5.1\endcsname{114}
\expandafter\gdef\csname odunum@ci@axis_breakdown.m2.nemotron.axis5.audio_visual.5.1\endcsname{[0.186, 0.306]}
\expandafter\gdef\csname odunum@val@axis_breakdown.m2.nemotron.axis5.audio_visual.5.10\endcsname{0.249}
\expandafter\gdef\csname odunum@n@axis_breakdown.m2.nemotron.axis5.audio_visual.5.10\endcsname{123}
\expandafter\gdef\csname odunum@ci@axis_breakdown.m2.nemotron.axis5.audio_visual.5.10\endcsname{[0.192, 0.309]}
\expandafter\gdef\csname odunum@val@axis_breakdown.m2.nemotron.axis5.audio_visual.5.2\endcsname{0.290}
\expandafter\gdef\csname odunum@n@axis_breakdown.m2.nemotron.axis5.audio_visual.5.2\endcsname{126}
\expandafter\gdef\csname odunum@ci@axis_breakdown.m2.nemotron.axis5.audio_visual.5.2\endcsname{[0.236, 0.349]}
\expandafter\gdef\csname odunum@val@axis_breakdown.m2.nemotron.axis5.audio_visual.5.3\endcsname{0.272}
\expandafter\gdef\csname odunum@n@axis_breakdown.m2.nemotron.axis5.audio_visual.5.3\endcsname{100}
\expandafter\gdef\csname odunum@ci@axis_breakdown.m2.nemotron.axis5.audio_visual.5.3\endcsname{[0.215, 0.331]}
\expandafter\gdef\csname odunum@val@axis_breakdown.m2.nemotron.axis5.audio_visual.5.4\endcsname{0.230}
\expandafter\gdef\csname odunum@n@axis_breakdown.m2.nemotron.axis5.audio_visual.5.4\endcsname{97}
\expandafter\gdef\csname odunum@ci@axis_breakdown.m2.nemotron.axis5.audio_visual.5.4\endcsname{[0.176, 0.288]}
\expandafter\gdef\csname odunum@val@axis_breakdown.m2.nemotron.axis5.audio_visual.5.5\endcsname{0.269}
\expandafter\gdef\csname odunum@n@axis_breakdown.m2.nemotron.axis5.audio_visual.5.5\endcsname{99}
\expandafter\gdef\csname odunum@ci@axis_breakdown.m2.nemotron.axis5.audio_visual.5.5\endcsname{[0.207, 0.332]}
\expandafter\gdef\csname odunum@val@axis_breakdown.m2.nemotron.axis5.audio_visual.5.6\endcsname{0.313}
\expandafter\gdef\csname odunum@n@axis_breakdown.m2.nemotron.axis5.audio_visual.5.6\endcsname{99}
\expandafter\gdef\csname odunum@ci@axis_breakdown.m2.nemotron.axis5.audio_visual.5.6\endcsname{[0.248, 0.381]}
\expandafter\gdef\csname odunum@val@axis_breakdown.m2.nemotron.axis5.audio_visual.5.7\endcsname{0.254}
\expandafter\gdef\csname odunum@n@axis_breakdown.m2.nemotron.axis5.audio_visual.5.7\endcsname{104}
\expandafter\gdef\csname odunum@ci@axis_breakdown.m2.nemotron.axis5.audio_visual.5.7\endcsname{[0.198, 0.313]}
\expandafter\gdef\csname odunum@val@axis_breakdown.m2.nemotron.axis5.audio_visual.5.8\endcsname{0.282}
\expandafter\gdef\csname odunum@n@axis_breakdown.m2.nemotron.axis5.audio_visual.5.8\endcsname{121}
\expandafter\gdef\csname odunum@ci@axis_breakdown.m2.nemotron.axis5.audio_visual.5.8\endcsname{[0.227, 0.338]}
\expandafter\gdef\csname odunum@val@axis_breakdown.m2.nemotron.axis5.audio_visual.5.9\endcsname{0.272}
\expandafter\gdef\csname odunum@n@axis_breakdown.m2.nemotron.axis5.audio_visual.5.9\endcsname{78}
\expandafter\gdef\csname odunum@ci@axis_breakdown.m2.nemotron.axis5.audio_visual.5.9\endcsname{[0.202, 0.347]}
\expandafter\gdef\csname odunum@val@axis_breakdown.m2.nemotron.axis6.6.1\endcsname{0.245}
\expandafter\gdef\csname odunum@n@axis_breakdown.m2.nemotron.axis6.6.1\endcsname{230}
\expandafter\gdef\csname odunum@ci@axis_breakdown.m2.nemotron.axis6.6.1\endcsname{[0.205, 0.287]}
\expandafter\gdef\csname odunum@val@axis_breakdown.m2.nemotron.axis6.6.10\endcsname{0.317}
\expandafter\gdef\csname odunum@n@axis_breakdown.m2.nemotron.axis6.6.10\endcsname{102}
\expandafter\gdef\csname odunum@ci@axis_breakdown.m2.nemotron.axis6.6.10\endcsname{[0.252, 0.384]}
\expandafter\gdef\csname odunum@val@axis_breakdown.m2.nemotron.axis6.6.12\endcsname{0.264}
\expandafter\gdef\csname odunum@n@axis_breakdown.m2.nemotron.axis6.6.12\endcsname{129}
\expandafter\gdef\csname odunum@ci@axis_breakdown.m2.nemotron.axis6.6.12\endcsname{[0.209, 0.321]}
\expandafter\gdef\csname odunum@val@axis_breakdown.m2.nemotron.axis6.6.13\endcsname{0.256}
\expandafter\gdef\csname odunum@n@axis_breakdown.m2.nemotron.axis6.6.13\endcsname{58}
\expandafter\gdef\csname odunum@ci@axis_breakdown.m2.nemotron.axis6.6.13\endcsname{[0.175, 0.340]}
\expandafter\gdef\csname odunum@val@axis_breakdown.m2.nemotron.axis6.6.14\endcsname{0.335}
\expandafter\gdef\csname odunum@n@axis_breakdown.m2.nemotron.axis6.6.14\endcsname{82}
\expandafter\gdef\csname odunum@ci@axis_breakdown.m2.nemotron.axis6.6.14\endcsname{[0.271, 0.401]}
\expandafter\gdef\csname odunum@val@axis_breakdown.m2.nemotron.axis6.6.18\endcsname{0.310}
\expandafter\gdef\csname odunum@n@axis_breakdown.m2.nemotron.axis6.6.18\endcsname{72}
\expandafter\gdef\csname odunum@ci@axis_breakdown.m2.nemotron.axis6.6.18\endcsname{[0.237, 0.386]}
\expandafter\gdef\csname odunum@val@axis_breakdown.m2.nemotron.axis6.6.19\endcsname{0.305}
\expandafter\gdef\csname odunum@n@axis_breakdown.m2.nemotron.axis6.6.19\endcsname{44}
\expandafter\gdef\csname odunum@ci@axis_breakdown.m2.nemotron.axis6.6.19\endcsname{[0.205, 0.409]}
\expandafter\gdef\csname odunum@val@axis_breakdown.m2.nemotron.axis6.6.2\endcsname{0.318}
\expandafter\gdef\csname odunum@n@axis_breakdown.m2.nemotron.axis6.6.2\endcsname{41}
\expandafter\gdef\csname odunum@ci@axis_breakdown.m2.nemotron.axis6.6.2\endcsname{[0.220, 0.419]}
\expandafter\gdef\csname odunum@val@axis_breakdown.m2.nemotron.axis6.6.3\endcsname{0.276}
\expandafter\gdef\csname odunum@n@axis_breakdown.m2.nemotron.axis6.6.3\endcsname{44}
\expandafter\gdef\csname odunum@ci@axis_breakdown.m2.nemotron.axis6.6.3\endcsname{[0.176, 0.385]}
\expandafter\gdef\csname odunum@val@axis_breakdown.m2.nemotron.axis6.6.5\endcsname{0.257}
\expandafter\gdef\csname odunum@n@axis_breakdown.m2.nemotron.axis6.6.5\endcsname{48}
\expandafter\gdef\csname odunum@ci@axis_breakdown.m2.nemotron.axis6.6.5\endcsname{[0.168, 0.351]}
\expandafter\gdef\csname odunum@val@axis_breakdown.m2.nemotron.axis6.6.6\endcsname{0.299}
\expandafter\gdef\csname odunum@n@axis_breakdown.m2.nemotron.axis6.6.6\endcsname{134}
\expandafter\gdef\csname odunum@ci@axis_breakdown.m2.nemotron.axis6.6.6\endcsname{[0.238, 0.362]}
\expandafter\gdef\csname odunum@val@axis_breakdown.m2.nemotron.axis6.6.7\endcsname{0.337}
\expandafter\gdef\csname odunum@n@axis_breakdown.m2.nemotron.axis6.6.7\endcsname{393}
\expandafter\gdef\csname odunum@ci@axis_breakdown.m2.nemotron.axis6.6.7\endcsname{[0.302, 0.372]}
\expandafter\gdef\csname odunum@val@axis_breakdown.m2.nemotron.axis6.6.8\endcsname{0.251}
\expandafter\gdef\csname odunum@n@axis_breakdown.m2.nemotron.axis6.6.8\endcsname{49}
\expandafter\gdef\csname odunum@ci@axis_breakdown.m2.nemotron.axis6.6.8\endcsname{[0.163, 0.344]}
\expandafter\gdef\csname odunum@val@axis_breakdown.m2.nemotron.axis6.6.9\endcsname{0.282}
\expandafter\gdef\csname odunum@n@axis_breakdown.m2.nemotron.axis6.6.9\endcsname{119}
\expandafter\gdef\csname odunum@ci@axis_breakdown.m2.nemotron.axis6.6.9\endcsname{[0.223, 0.340]}
\expandafter\gdef\csname odunum@val@axis_breakdown.m2.nemotron.difficulty.q1\endcsname{0.418}
\expandafter\gdef\csname odunum@n@axis_breakdown.m2.nemotron.difficulty.q1\endcsname{358}
\expandafter\gdef\csname odunum@ci@axis_breakdown.m2.nemotron.difficulty.q1\endcsname{[0.379, 0.457]}
\expandafter\gdef\csname odunum@val@axis_breakdown.m2.nemotron.difficulty.q2\endcsname{0.321}
\expandafter\gdef\csname odunum@n@axis_breakdown.m2.nemotron.difficulty.q2\endcsname{357}
\expandafter\gdef\csname odunum@ci@axis_breakdown.m2.nemotron.difficulty.q2\endcsname{[0.286, 0.356]}
\expandafter\gdef\csname odunum@val@axis_breakdown.m2.nemotron.difficulty.q3\endcsname{0.304}
\expandafter\gdef\csname odunum@n@axis_breakdown.m2.nemotron.difficulty.q3\endcsname{358}
\expandafter\gdef\csname odunum@ci@axis_breakdown.m2.nemotron.difficulty.q3\endcsname{[0.270, 0.339]}
\expandafter\gdef\csname odunum@val@axis_breakdown.m2.nemotron.difficulty.q4\endcsname{0.271}
\expandafter\gdef\csname odunum@n@axis_breakdown.m2.nemotron.difficulty.q4\endcsname{356}
\expandafter\gdef\csname odunum@ci@axis_breakdown.m2.nemotron.difficulty.q4\endcsname{[0.241, 0.302]}
\expandafter\gdef\csname odunum@val@axis_breakdown.m2.qwen25_omni.axis1.audio_only.1.1\endcsname{0.510}
\expandafter\gdef\csname odunum@n@axis_breakdown.m2.qwen25_omni.axis1.audio_only.1.1\endcsname{158}
\expandafter\gdef\csname odunum@ci@axis_breakdown.m2.qwen25_omni.axis1.audio_only.1.1\endcsname{[0.458, 0.564]}
\expandafter\gdef\csname odunum@val@axis_breakdown.m2.qwen25_omni.axis1.audio_only.1.2\endcsname{0.436}
\expandafter\gdef\csname odunum@n@axis_breakdown.m2.qwen25_omni.axis1.audio_only.1.2\endcsname{379}
\expandafter\gdef\csname odunum@ci@axis_breakdown.m2.qwen25_omni.axis1.audio_only.1.2\endcsname{[0.403, 0.469]}
\expandafter\gdef\csname odunum@val@axis_breakdown.m2.qwen25_omni.axis1.audio_visual.1.1\endcsname{0.447}
\expandafter\gdef\csname odunum@n@axis_breakdown.m2.qwen25_omni.axis1.audio_visual.1.1\endcsname{94}
\expandafter\gdef\csname odunum@ci@axis_breakdown.m2.qwen25_omni.axis1.audio_visual.1.1\endcsname{[0.383, 0.510]}
\expandafter\gdef\csname odunum@val@axis_breakdown.m2.qwen25_omni.axis1.audio_visual.1.2\endcsname{0.421}
\expandafter\gdef\csname odunum@n@axis_breakdown.m2.qwen25_omni.axis1.audio_visual.1.2\endcsname{294}
\expandafter\gdef\csname odunum@ci@axis_breakdown.m2.qwen25_omni.axis1.audio_visual.1.2\endcsname{[0.389, 0.454]}
\expandafter\gdef\csname odunum@val@axis_breakdown.m2.qwen25_omni.axis1.audio_visual.1.3\endcsname{0.437}
\expandafter\gdef\csname odunum@n@axis_breakdown.m2.qwen25_omni.axis1.audio_visual.1.3\endcsname{291}
\expandafter\gdef\csname odunum@ci@axis_breakdown.m2.qwen25_omni.axis1.audio_visual.1.3\endcsname{[0.402, 0.474]}
\expandafter\gdef\csname odunum@val@axis_breakdown.m2.qwen25_omni.axis1.audio_visual.1.4\endcsname{0.401}
\expandafter\gdef\csname odunum@n@axis_breakdown.m2.qwen25_omni.axis1.audio_visual.1.4\endcsname{382}
\expandafter\gdef\csname odunum@ci@axis_breakdown.m2.qwen25_omni.axis1.audio_visual.1.4\endcsname{[0.374, 0.428]}
\expandafter\gdef\csname odunum@val@axis_breakdown.m2.qwen25_omni.axis2.2.1\endcsname{0.350}
\expandafter\gdef\csname odunum@n@axis_breakdown.m2.qwen25_omni.axis2.2.1\endcsname{143}
\expandafter\gdef\csname odunum@ci@axis_breakdown.m2.qwen25_omni.axis2.2.1\endcsname{[0.298, 0.401]}
\expandafter\gdef\csname odunum@val@axis_breakdown.m2.qwen25_omni.axis2.2.2\endcsname{0.478}
\expandafter\gdef\csname odunum@n@axis_breakdown.m2.qwen25_omni.axis2.2.2\endcsname{440}
\expandafter\gdef\csname odunum@ci@axis_breakdown.m2.qwen25_omni.axis2.2.2\endcsname{[0.449, 0.506]}
\expandafter\gdef\csname odunum@val@axis_breakdown.m2.qwen25_omni.axis2.2.3\endcsname{0.472}
\expandafter\gdef\csname odunum@n@axis_breakdown.m2.qwen25_omni.axis2.2.3\endcsname{531}
\expandafter\gdef\csname odunum@ci@axis_breakdown.m2.qwen25_omni.axis2.2.3\endcsname{[0.445, 0.499]}
\expandafter\gdef\csname odunum@val@axis_breakdown.m2.qwen25_omni.axis2.2.4\endcsname{0.396}
\expandafter\gdef\csname odunum@n@axis_breakdown.m2.qwen25_omni.axis2.2.4\endcsname{261}
\expandafter\gdef\csname odunum@ci@axis_breakdown.m2.qwen25_omni.axis2.2.4\endcsname{[0.361, 0.430]}
\expandafter\gdef\csname odunum@val@axis_breakdown.m2.qwen25_omni.axis2.2.5\endcsname{0.362}
\expandafter\gdef\csname odunum@n@axis_breakdown.m2.qwen25_omni.axis2.2.5\endcsname{256}
\expandafter\gdef\csname odunum@ci@axis_breakdown.m2.qwen25_omni.axis2.2.5\endcsname{[0.326, 0.398]}
\expandafter\gdef\csname odunum@val@axis_breakdown.m2.qwen25_omni.axis3.3.1\endcsname{0.456}
\expandafter\gdef\csname odunum@n@axis_breakdown.m2.qwen25_omni.axis3.3.1\endcsname{109}
\expandafter\gdef\csname odunum@ci@axis_breakdown.m2.qwen25_omni.axis3.3.1\endcsname{[0.401, 0.509]}
\expandafter\gdef\csname odunum@val@axis_breakdown.m2.qwen25_omni.axis3.3.2\endcsname{0.469}
\expandafter\gdef\csname odunum@n@axis_breakdown.m2.qwen25_omni.axis3.3.2\endcsname{116}
\expandafter\gdef\csname odunum@ci@axis_breakdown.m2.qwen25_omni.axis3.3.2\endcsname{[0.416, 0.523]}
\expandafter\gdef\csname odunum@val@axis_breakdown.m2.qwen25_omni.axis3.3.3\endcsname{0.322}
\expandafter\gdef\csname odunum@n@axis_breakdown.m2.qwen25_omni.axis3.3.3\endcsname{207}
\expandafter\gdef\csname odunum@ci@axis_breakdown.m2.qwen25_omni.axis3.3.3\endcsname{[0.280, 0.363]}
\expandafter\gdef\csname odunum@val@axis_breakdown.m2.qwen25_omni.axis3.3.4\endcsname{0.438}
\expandafter\gdef\csname odunum@n@axis_breakdown.m2.qwen25_omni.axis3.3.4\endcsname{404}
\expandafter\gdef\csname odunum@ci@axis_breakdown.m2.qwen25_omni.axis3.3.4\endcsname{[0.407, 0.469]}
\expandafter\gdef\csname odunum@val@axis_breakdown.m2.qwen25_omni.axis3.3.5\endcsname{0.450}
\expandafter\gdef\csname odunum@n@axis_breakdown.m2.qwen25_omni.axis3.3.5\endcsname{294}
\expandafter\gdef\csname odunum@ci@axis_breakdown.m2.qwen25_omni.axis3.3.5\endcsname{[0.416, 0.483]}
\expandafter\gdef\csname odunum@val@axis_breakdown.m2.qwen25_omni.axis3.3.6\endcsname{0.452}
\expandafter\gdef\csname odunum@n@axis_breakdown.m2.qwen25_omni.axis3.3.6\endcsname{501}
\expandafter\gdef\csname odunum@ci@axis_breakdown.m2.qwen25_omni.axis3.3.6\endcsname{[0.426, 0.479]}
\expandafter\gdef\csname odunum@val@axis_breakdown.m2.qwen25_omni.axis4.4.1\endcsname{0.462}
\expandafter\gdef\csname odunum@n@axis_breakdown.m2.qwen25_omni.axis4.4.1\endcsname{202}
\expandafter\gdef\csname odunum@ci@axis_breakdown.m2.qwen25_omni.axis4.4.1\endcsname{[0.422, 0.501]}
\expandafter\gdef\csname odunum@val@axis_breakdown.m2.qwen25_omni.axis4.4.2\endcsname{0.446}
\expandafter\gdef\csname odunum@n@axis_breakdown.m2.qwen25_omni.axis4.4.2\endcsname{457}
\expandafter\gdef\csname odunum@ci@axis_breakdown.m2.qwen25_omni.axis4.4.2\endcsname{[0.419, 0.473]}
\expandafter\gdef\csname odunum@val@axis_breakdown.m2.qwen25_omni.axis4.4.3\endcsname{0.375}
\expandafter\gdef\csname odunum@n@axis_breakdown.m2.qwen25_omni.axis4.4.3\endcsname{164}
\expandafter\gdef\csname odunum@ci@axis_breakdown.m2.qwen25_omni.axis4.4.3\endcsname{[0.323, 0.427]}
\expandafter\gdef\csname odunum@val@axis_breakdown.m2.qwen25_omni.axis4.4.4\endcsname{0.435}
\expandafter\gdef\csname odunum@n@axis_breakdown.m2.qwen25_omni.axis4.4.4\endcsname{505}
\expandafter\gdef\csname odunum@ci@axis_breakdown.m2.qwen25_omni.axis4.4.4\endcsname{[0.409, 0.462]}
\expandafter\gdef\csname odunum@val@axis_breakdown.m2.qwen25_omni.axis4.4.5\endcsname{0.422}
\expandafter\gdef\csname odunum@n@axis_breakdown.m2.qwen25_omni.axis4.4.5\endcsname{303}
\expandafter\gdef\csname odunum@ci@axis_breakdown.m2.qwen25_omni.axis4.4.5\endcsname{[0.387, 0.457]}
\expandafter\gdef\csname odunum@val@axis_breakdown.m2.qwen25_omni.axis5.audio_only.5.1\endcsname{0.533}
\expandafter\gdef\csname odunum@n@axis_breakdown.m2.qwen25_omni.axis5.audio_only.5.1\endcsname{45}
\expandafter\gdef\csname odunum@ci@axis_breakdown.m2.qwen25_omni.axis5.audio_only.5.1\endcsname{[0.426, 0.643]}
\expandafter\gdef\csname odunum@val@axis_breakdown.m2.qwen25_omni.axis5.audio_only.5.10\endcsname{0.444}
\expandafter\gdef\csname odunum@n@axis_breakdown.m2.qwen25_omni.axis5.audio_only.5.10\endcsname{72}
\expandafter\gdef\csname odunum@ci@axis_breakdown.m2.qwen25_omni.axis5.audio_only.5.10\endcsname{[0.367, 0.521]}
\expandafter\gdef\csname odunum@val@axis_breakdown.m2.qwen25_omni.axis5.audio_only.5.2\endcsname{0.482}
\expandafter\gdef\csname odunum@n@axis_breakdown.m2.qwen25_omni.axis5.audio_only.5.2\endcsname{75}
\expandafter\gdef\csname odunum@ci@axis_breakdown.m2.qwen25_omni.axis5.audio_only.5.2\endcsname{[0.405, 0.560]}
\expandafter\gdef\csname odunum@val@axis_breakdown.m2.qwen25_omni.axis5.audio_only.5.3\endcsname{0.512}
\expandafter\gdef\csname odunum@n@axis_breakdown.m2.qwen25_omni.axis5.audio_only.5.3\endcsname{50}
\expandafter\gdef\csname odunum@ci@axis_breakdown.m2.qwen25_omni.axis5.audio_only.5.3\endcsname{[0.423, 0.601]}
\expandafter\gdef\csname odunum@val@axis_breakdown.m2.qwen25_omni.axis5.audio_only.5.4\endcsname{0.386}
\expandafter\gdef\csname odunum@n@axis_breakdown.m2.qwen25_omni.axis5.audio_only.5.4\endcsname{68}
\expandafter\gdef\csname odunum@ci@axis_breakdown.m2.qwen25_omni.axis5.audio_only.5.4\endcsname{[0.317, 0.457]}
\expandafter\gdef\csname odunum@val@axis_breakdown.m2.qwen25_omni.axis5.audio_only.5.5\endcsname{0.397}
\expandafter\gdef\csname odunum@n@axis_breakdown.m2.qwen25_omni.axis5.audio_only.5.5\endcsname{52}
\expandafter\gdef\csname odunum@ci@axis_breakdown.m2.qwen25_omni.axis5.audio_only.5.5\endcsname{[0.317, 0.478]}
\expandafter\gdef\csname odunum@val@axis_breakdown.m2.qwen25_omni.axis5.audio_only.5.6\endcsname{0.470}
\expandafter\gdef\csname odunum@n@axis_breakdown.m2.qwen25_omni.axis5.audio_only.5.6\endcsname{52}
\expandafter\gdef\csname odunum@ci@axis_breakdown.m2.qwen25_omni.axis5.audio_only.5.6\endcsname{[0.380, 0.560]}
\expandafter\gdef\csname odunum@val@axis_breakdown.m2.qwen25_omni.axis5.audio_only.5.7\endcsname{0.503}
\expandafter\gdef\csname odunum@n@axis_breakdown.m2.qwen25_omni.axis5.audio_only.5.7\endcsname{46}
\expandafter\gdef\csname odunum@ci@axis_breakdown.m2.qwen25_omni.axis5.audio_only.5.7\endcsname{[0.410, 0.592]}
\expandafter\gdef\csname odunum@val@axis_breakdown.m2.qwen25_omni.axis5.audio_only.5.8\endcsname{0.358}
\expandafter\gdef\csname odunum@n@axis_breakdown.m2.qwen25_omni.axis5.audio_only.5.8\endcsname{57}
\expandafter\gdef\csname odunum@ci@axis_breakdown.m2.qwen25_omni.axis5.audio_only.5.8\endcsname{[0.279, 0.440]}
\expandafter\gdef\csname odunum@val@axis_breakdown.m2.qwen25_omni.axis5.audio_only.5.9\endcsname{0.523}
\expandafter\gdef\csname odunum@n@axis_breakdown.m2.qwen25_omni.axis5.audio_only.5.9\endcsname{53}
\expandafter\gdef\csname odunum@ci@axis_breakdown.m2.qwen25_omni.axis5.audio_only.5.9\endcsname{[0.424, 0.621]}
\expandafter\gdef\csname odunum@val@axis_breakdown.m2.qwen25_omni.axis5.audio_visual.5.1\endcsname{0.449}
\expandafter\gdef\csname odunum@n@axis_breakdown.m2.qwen25_omni.axis5.audio_visual.5.1\endcsname{114}
\expandafter\gdef\csname odunum@ci@axis_breakdown.m2.qwen25_omni.axis5.audio_visual.5.1\endcsname{[0.395, 0.505]}
\expandafter\gdef\csname odunum@val@axis_breakdown.m2.qwen25_omni.axis5.audio_visual.5.10\endcsname{0.458}
\expandafter\gdef\csname odunum@n@axis_breakdown.m2.qwen25_omni.axis5.audio_visual.5.10\endcsname{123}
\expandafter\gdef\csname odunum@ci@axis_breakdown.m2.qwen25_omni.axis5.audio_visual.5.10\endcsname{[0.407, 0.507]}
\expandafter\gdef\csname odunum@val@axis_breakdown.m2.qwen25_omni.axis5.audio_visual.5.2\endcsname{0.431}
\expandafter\gdef\csname odunum@n@axis_breakdown.m2.qwen25_omni.axis5.audio_visual.5.2\endcsname{126}
\expandafter\gdef\csname odunum@ci@axis_breakdown.m2.qwen25_omni.axis5.audio_visual.5.2\endcsname{[0.385, 0.479]}
\expandafter\gdef\csname odunum@val@axis_breakdown.m2.qwen25_omni.axis5.audio_visual.5.3\endcsname{0.422}
\expandafter\gdef\csname odunum@n@axis_breakdown.m2.qwen25_omni.axis5.audio_visual.5.3\endcsname{100}
\expandafter\gdef\csname odunum@ci@axis_breakdown.m2.qwen25_omni.axis5.audio_visual.5.3\endcsname{[0.370, 0.474]}
\expandafter\gdef\csname odunum@val@axis_breakdown.m2.qwen25_omni.axis5.audio_visual.5.4\endcsname{0.386}
\expandafter\gdef\csname odunum@n@axis_breakdown.m2.qwen25_omni.axis5.audio_visual.5.4\endcsname{97}
\expandafter\gdef\csname odunum@ci@axis_breakdown.m2.qwen25_omni.axis5.audio_visual.5.4\endcsname{[0.328, 0.443]}
\expandafter\gdef\csname odunum@val@axis_breakdown.m2.qwen25_omni.axis5.audio_visual.5.5\endcsname{0.424}
\expandafter\gdef\csname odunum@n@axis_breakdown.m2.qwen25_omni.axis5.audio_visual.5.5\endcsname{99}
\expandafter\gdef\csname odunum@ci@axis_breakdown.m2.qwen25_omni.axis5.audio_visual.5.5\endcsname{[0.364, 0.483]}
\expandafter\gdef\csname odunum@val@axis_breakdown.m2.qwen25_omni.axis5.audio_visual.5.6\endcsname{0.389}
\expandafter\gdef\csname odunum@n@axis_breakdown.m2.qwen25_omni.axis5.audio_visual.5.6\endcsname{99}
\expandafter\gdef\csname odunum@ci@axis_breakdown.m2.qwen25_omni.axis5.audio_visual.5.6\endcsname{[0.331, 0.448]}
\expandafter\gdef\csname odunum@val@axis_breakdown.m2.qwen25_omni.axis5.audio_visual.5.7\endcsname{0.407}
\expandafter\gdef\csname odunum@n@axis_breakdown.m2.qwen25_omni.axis5.audio_visual.5.7\endcsname{104}
\expandafter\gdef\csname odunum@ci@axis_breakdown.m2.qwen25_omni.axis5.audio_visual.5.7\endcsname{[0.354, 0.462]}
\expandafter\gdef\csname odunum@val@axis_breakdown.m2.qwen25_omni.axis5.audio_visual.5.8\endcsname{0.403}
\expandafter\gdef\csname odunum@n@axis_breakdown.m2.qwen25_omni.axis5.audio_visual.5.8\endcsname{121}
\expandafter\gdef\csname odunum@ci@axis_breakdown.m2.qwen25_omni.axis5.audio_visual.5.8\endcsname{[0.352, 0.454]}
\expandafter\gdef\csname odunum@val@axis_breakdown.m2.qwen25_omni.axis5.audio_visual.5.9\endcsname{0.425}
\expandafter\gdef\csname odunum@n@axis_breakdown.m2.qwen25_omni.axis5.audio_visual.5.9\endcsname{78}
\expandafter\gdef\csname odunum@ci@axis_breakdown.m2.qwen25_omni.axis5.audio_visual.5.9\endcsname{[0.359, 0.490]}
\expandafter\gdef\csname odunum@val@axis_breakdown.m2.qwen25_omni.axis6.6.1\endcsname{0.403}
\expandafter\gdef\csname odunum@n@axis_breakdown.m2.qwen25_omni.axis6.6.1\endcsname{230}
\expandafter\gdef\csname odunum@ci@axis_breakdown.m2.qwen25_omni.axis6.6.1\endcsname{[0.364, 0.444]}
\expandafter\gdef\csname odunum@val@axis_breakdown.m2.qwen25_omni.axis6.6.10\endcsname{0.442}
\expandafter\gdef\csname odunum@n@axis_breakdown.m2.qwen25_omni.axis6.6.10\endcsname{102}
\expandafter\gdef\csname odunum@ci@axis_breakdown.m2.qwen25_omni.axis6.6.10\endcsname{[0.389, 0.496]}
\expandafter\gdef\csname odunum@val@axis_breakdown.m2.qwen25_omni.axis6.6.12\endcsname{0.446}
\expandafter\gdef\csname odunum@n@axis_breakdown.m2.qwen25_omni.axis6.6.12\endcsname{129}
\expandafter\gdef\csname odunum@ci@axis_breakdown.m2.qwen25_omni.axis6.6.12\endcsname{[0.395, 0.498]}
\expandafter\gdef\csname odunum@val@axis_breakdown.m2.qwen25_omni.axis6.6.13\endcsname{0.376}
\expandafter\gdef\csname odunum@n@axis_breakdown.m2.qwen25_omni.axis6.6.13\endcsname{58}
\expandafter\gdef\csname odunum@ci@axis_breakdown.m2.qwen25_omni.axis6.6.13\endcsname{[0.291, 0.465]}
\expandafter\gdef\csname odunum@val@axis_breakdown.m2.qwen25_omni.axis6.6.14\endcsname{0.375}
\expandafter\gdef\csname odunum@n@axis_breakdown.m2.qwen25_omni.axis6.6.14\endcsname{82}
\expandafter\gdef\csname odunum@ci@axis_breakdown.m2.qwen25_omni.axis6.6.14\endcsname{[0.318, 0.432]}
\expandafter\gdef\csname odunum@val@axis_breakdown.m2.qwen25_omni.axis6.6.18\endcsname{0.424}
\expandafter\gdef\csname odunum@n@axis_breakdown.m2.qwen25_omni.axis6.6.18\endcsname{72}
\expandafter\gdef\csname odunum@ci@axis_breakdown.m2.qwen25_omni.axis6.6.18\endcsname{[0.356, 0.494]}
\expandafter\gdef\csname odunum@val@axis_breakdown.m2.qwen25_omni.axis6.6.19\endcsname{0.465}
\expandafter\gdef\csname odunum@n@axis_breakdown.m2.qwen25_omni.axis6.6.19\endcsname{44}
\expandafter\gdef\csname odunum@ci@axis_breakdown.m2.qwen25_omni.axis6.6.19\endcsname{[0.370, 0.562]}
\expandafter\gdef\csname odunum@val@axis_breakdown.m2.qwen25_omni.axis6.6.2\endcsname{0.500}
\expandafter\gdef\csname odunum@n@axis_breakdown.m2.qwen25_omni.axis6.6.2\endcsname{41}
\expandafter\gdef\csname odunum@ci@axis_breakdown.m2.qwen25_omni.axis6.6.2\endcsname{[0.397, 0.604]}
\expandafter\gdef\csname odunum@val@axis_breakdown.m2.qwen25_omni.axis6.6.3\endcsname{0.448}
\expandafter\gdef\csname odunum@n@axis_breakdown.m2.qwen25_omni.axis6.6.3\endcsname{44}
\expandafter\gdef\csname odunum@ci@axis_breakdown.m2.qwen25_omni.axis6.6.3\endcsname{[0.360, 0.534]}
\expandafter\gdef\csname odunum@val@axis_breakdown.m2.qwen25_omni.axis6.6.5\endcsname{0.466}
\expandafter\gdef\csname odunum@n@axis_breakdown.m2.qwen25_omni.axis6.6.5\endcsname{48}
\expandafter\gdef\csname odunum@ci@axis_breakdown.m2.qwen25_omni.axis6.6.5\endcsname{[0.375, 0.562]}
\expandafter\gdef\csname odunum@val@axis_breakdown.m2.qwen25_omni.axis6.6.6\endcsname{0.449}
\expandafter\gdef\csname odunum@n@axis_breakdown.m2.qwen25_omni.axis6.6.6\endcsname{134}
\expandafter\gdef\csname odunum@ci@axis_breakdown.m2.qwen25_omni.axis6.6.6\endcsname{[0.393, 0.505]}
\expandafter\gdef\csname odunum@val@axis_breakdown.m2.qwen25_omni.axis6.6.7\endcsname{0.455}
\expandafter\gdef\csname odunum@n@axis_breakdown.m2.qwen25_omni.axis6.6.7\endcsname{393}
\expandafter\gdef\csname odunum@ci@axis_breakdown.m2.qwen25_omni.axis6.6.7\endcsname{[0.425, 0.485]}
\expandafter\gdef\csname odunum@val@axis_breakdown.m2.qwen25_omni.axis6.6.8\endcsname{0.370}
\expandafter\gdef\csname odunum@n@axis_breakdown.m2.qwen25_omni.axis6.6.8\endcsname{49}
\expandafter\gdef\csname odunum@ci@axis_breakdown.m2.qwen25_omni.axis6.6.8\endcsname{[0.281, 0.459]}
\expandafter\gdef\csname odunum@val@axis_breakdown.m2.qwen25_omni.axis6.6.9\endcsname{0.456}
\expandafter\gdef\csname odunum@n@axis_breakdown.m2.qwen25_omni.axis6.6.9\endcsname{119}
\expandafter\gdef\csname odunum@ci@axis_breakdown.m2.qwen25_omni.axis6.6.9\endcsname{[0.406, 0.507]}
\expandafter\gdef\csname odunum@val@axis_breakdown.m2.qwen25_omni.difficulty.q1\endcsname{0.492}
\expandafter\gdef\csname odunum@n@axis_breakdown.m2.qwen25_omni.difficulty.q1\endcsname{358}
\expandafter\gdef\csname odunum@ci@axis_breakdown.m2.qwen25_omni.difficulty.q1\endcsname{[0.458, 0.526]}
\expandafter\gdef\csname odunum@val@axis_breakdown.m2.qwen25_omni.difficulty.q2\endcsname{0.456}
\expandafter\gdef\csname odunum@n@axis_breakdown.m2.qwen25_omni.difficulty.q2\endcsname{357}
\expandafter\gdef\csname odunum@ci@axis_breakdown.m2.qwen25_omni.difficulty.q2\endcsname{[0.425, 0.487]}
\expandafter\gdef\csname odunum@val@axis_breakdown.m2.qwen25_omni.difficulty.q3\endcsname{0.453}
\expandafter\gdef\csname odunum@n@axis_breakdown.m2.qwen25_omni.difficulty.q3\endcsname{358}
\expandafter\gdef\csname odunum@ci@axis_breakdown.m2.qwen25_omni.difficulty.q3\endcsname{[0.422, 0.484]}
\expandafter\gdef\csname odunum@val@axis_breakdown.m2.qwen25_omni.difficulty.q4\endcsname{0.421}
\expandafter\gdef\csname odunum@n@axis_breakdown.m2.qwen25_omni.difficulty.q4\endcsname{356}
\expandafter\gdef\csname odunum@ci@axis_breakdown.m2.qwen25_omni.difficulty.q4\endcsname{[0.393, 0.450]}
\expandafter\gdef\csname odunum@val@axis_breakdown.m2.qwen3_omni_instruct.axis1.audio_only.1.1\endcsname{0.633}
\expandafter\gdef\csname odunum@n@axis_breakdown.m2.qwen3_omni_instruct.axis1.audio_only.1.1\endcsname{158}
\expandafter\gdef\csname odunum@ci@axis_breakdown.m2.qwen3_omni_instruct.axis1.audio_only.1.1\endcsname{[0.587, 0.680]}
\expandafter\gdef\csname odunum@val@axis_breakdown.m2.qwen3_omni_instruct.axis1.audio_only.1.2\endcsname{0.600}
\expandafter\gdef\csname odunum@n@axis_breakdown.m2.qwen3_omni_instruct.axis1.audio_only.1.2\endcsname{379}
\expandafter\gdef\csname odunum@ci@axis_breakdown.m2.qwen3_omni_instruct.axis1.audio_only.1.2\endcsname{[0.574, 0.626]}
\expandafter\gdef\csname odunum@val@axis_breakdown.m2.qwen3_omni_instruct.axis1.audio_visual.1.1\endcsname{0.518}
\expandafter\gdef\csname odunum@n@axis_breakdown.m2.qwen3_omni_instruct.axis1.audio_visual.1.1\endcsname{94}
\expandafter\gdef\csname odunum@ci@axis_breakdown.m2.qwen3_omni_instruct.axis1.audio_visual.1.1\endcsname{[0.461, 0.577]}
\expandafter\gdef\csname odunum@val@axis_breakdown.m2.qwen3_omni_instruct.axis1.audio_visual.1.2\endcsname{0.552}
\expandafter\gdef\csname odunum@n@axis_breakdown.m2.qwen3_omni_instruct.axis1.audio_visual.1.2\endcsname{294}
\expandafter\gdef\csname odunum@ci@axis_breakdown.m2.qwen3_omni_instruct.axis1.audio_visual.1.2\endcsname{[0.523, 0.583]}
\expandafter\gdef\csname odunum@val@axis_breakdown.m2.qwen3_omni_instruct.axis1.audio_visual.1.3\endcsname{0.534}
\expandafter\gdef\csname odunum@n@axis_breakdown.m2.qwen3_omni_instruct.axis1.audio_visual.1.3\endcsname{291}
\expandafter\gdef\csname odunum@ci@axis_breakdown.m2.qwen3_omni_instruct.axis1.audio_visual.1.3\endcsname{[0.503, 0.564]}
\expandafter\gdef\csname odunum@val@axis_breakdown.m2.qwen3_omni_instruct.axis1.audio_visual.1.4\endcsname{0.514}
\expandafter\gdef\csname odunum@n@axis_breakdown.m2.qwen3_omni_instruct.axis1.audio_visual.1.4\endcsname{382}
\expandafter\gdef\csname odunum@ci@axis_breakdown.m2.qwen3_omni_instruct.axis1.audio_visual.1.4\endcsname{[0.488, 0.540]}
\expandafter\gdef\csname odunum@val@axis_breakdown.m2.qwen3_omni_instruct.axis2.2.1\endcsname{0.562}
\expandafter\gdef\csname odunum@n@axis_breakdown.m2.qwen3_omni_instruct.axis2.2.1\endcsname{143}
\expandafter\gdef\csname odunum@ci@axis_breakdown.m2.qwen3_omni_instruct.axis2.2.1\endcsname{[0.518, 0.607]}
\expandafter\gdef\csname odunum@val@axis_breakdown.m2.qwen3_omni_instruct.axis2.2.2\endcsname{0.552}
\expandafter\gdef\csname odunum@n@axis_breakdown.m2.qwen3_omni_instruct.axis2.2.2\endcsname{440}
\expandafter\gdef\csname odunum@ci@axis_breakdown.m2.qwen3_omni_instruct.axis2.2.2\endcsname{[0.527, 0.577]}
\expandafter\gdef\csname odunum@val@axis_breakdown.m2.qwen3_omni_instruct.axis2.2.3\endcsname{0.564}
\expandafter\gdef\csname odunum@n@axis_breakdown.m2.qwen3_omni_instruct.axis2.2.3\endcsname{531}
\expandafter\gdef\csname odunum@ci@axis_breakdown.m2.qwen3_omni_instruct.axis2.2.3\endcsname{[0.542, 0.587]}
\expandafter\gdef\csname odunum@val@axis_breakdown.m2.qwen3_omni_instruct.axis2.2.4\endcsname{0.535}
\expandafter\gdef\csname odunum@n@axis_breakdown.m2.qwen3_omni_instruct.axis2.2.4\endcsname{261}
\expandafter\gdef\csname odunum@ci@axis_breakdown.m2.qwen3_omni_instruct.axis2.2.4\endcsname{[0.501, 0.569]}
\expandafter\gdef\csname odunum@val@axis_breakdown.m2.qwen3_omni_instruct.axis2.2.5\endcsname{0.577}
\expandafter\gdef\csname odunum@n@axis_breakdown.m2.qwen3_omni_instruct.axis2.2.5\endcsname{256}
\expandafter\gdef\csname odunum@ci@axis_breakdown.m2.qwen3_omni_instruct.axis2.2.5\endcsname{[0.546, 0.608]}
\expandafter\gdef\csname odunum@val@axis_breakdown.m2.qwen3_omni_instruct.axis3.3.1\endcsname{0.554}
\expandafter\gdef\csname odunum@n@axis_breakdown.m2.qwen3_omni_instruct.axis3.3.1\endcsname{109}
\expandafter\gdef\csname odunum@ci@axis_breakdown.m2.qwen3_omni_instruct.axis3.3.1\endcsname{[0.502, 0.604]}
\expandafter\gdef\csname odunum@val@axis_breakdown.m2.qwen3_omni_instruct.axis3.3.2\endcsname{0.553}
\expandafter\gdef\csname odunum@n@axis_breakdown.m2.qwen3_omni_instruct.axis3.3.2\endcsname{116}
\expandafter\gdef\csname odunum@ci@axis_breakdown.m2.qwen3_omni_instruct.axis3.3.2\endcsname{[0.506, 0.601]}
\expandafter\gdef\csname odunum@val@axis_breakdown.m2.qwen3_omni_instruct.axis3.3.3\endcsname{0.510}
\expandafter\gdef\csname odunum@n@axis_breakdown.m2.qwen3_omni_instruct.axis3.3.3\endcsname{207}
\expandafter\gdef\csname odunum@ci@axis_breakdown.m2.qwen3_omni_instruct.axis3.3.3\endcsname{[0.474, 0.548]}
\expandafter\gdef\csname odunum@val@axis_breakdown.m2.qwen3_omni_instruct.axis3.3.4\endcsname{0.553}
\expandafter\gdef\csname odunum@n@axis_breakdown.m2.qwen3_omni_instruct.axis3.3.4\endcsname{404}
\expandafter\gdef\csname odunum@ci@axis_breakdown.m2.qwen3_omni_instruct.axis3.3.4\endcsname{[0.525, 0.579]}
\expandafter\gdef\csname odunum@val@axis_breakdown.m2.qwen3_omni_instruct.axis3.3.5\endcsname{0.525}
\expandafter\gdef\csname odunum@n@axis_breakdown.m2.qwen3_omni_instruct.axis3.3.5\endcsname{294}
\expandafter\gdef\csname odunum@ci@axis_breakdown.m2.qwen3_omni_instruct.axis3.3.5\endcsname{[0.493, 0.556]}
\expandafter\gdef\csname odunum@val@axis_breakdown.m2.qwen3_omni_instruct.axis3.3.6\endcsname{0.604}
\expandafter\gdef\csname odunum@n@axis_breakdown.m2.qwen3_omni_instruct.axis3.3.6\endcsname{501}
\expandafter\gdef\csname odunum@ci@axis_breakdown.m2.qwen3_omni_instruct.axis3.3.6\endcsname{[0.581, 0.626]}
\expandafter\gdef\csname odunum@val@axis_breakdown.m2.qwen3_omni_instruct.axis4.4.1\endcsname{0.529}
\expandafter\gdef\csname odunum@n@axis_breakdown.m2.qwen3_omni_instruct.axis4.4.1\endcsname{202}
\expandafter\gdef\csname odunum@ci@axis_breakdown.m2.qwen3_omni_instruct.axis4.4.1\endcsname{[0.495, 0.565]}
\expandafter\gdef\csname odunum@val@axis_breakdown.m2.qwen3_omni_instruct.axis4.4.2\endcsname{0.570}
\expandafter\gdef\csname odunum@n@axis_breakdown.m2.qwen3_omni_instruct.axis4.4.2\endcsname{457}
\expandafter\gdef\csname odunum@ci@axis_breakdown.m2.qwen3_omni_instruct.axis4.4.2\endcsname{[0.546, 0.593]}
\expandafter\gdef\csname odunum@val@axis_breakdown.m2.qwen3_omni_instruct.axis4.4.3\endcsname{0.521}
\expandafter\gdef\csname odunum@n@axis_breakdown.m2.qwen3_omni_instruct.axis4.4.3\endcsname{164}
\expandafter\gdef\csname odunum@ci@axis_breakdown.m2.qwen3_omni_instruct.axis4.4.3\endcsname{[0.474, 0.568]}
\expandafter\gdef\csname odunum@val@axis_breakdown.m2.qwen3_omni_instruct.axis4.4.4\endcsname{0.561}
\expandafter\gdef\csname odunum@n@axis_breakdown.m2.qwen3_omni_instruct.axis4.4.4\endcsname{505}
\expandafter\gdef\csname odunum@ci@axis_breakdown.m2.qwen3_omni_instruct.axis4.4.4\endcsname{[0.537, 0.585]}
\expandafter\gdef\csname odunum@val@axis_breakdown.m2.qwen3_omni_instruct.axis4.4.5\endcsname{0.574}
\expandafter\gdef\csname odunum@n@axis_breakdown.m2.qwen3_omni_instruct.axis4.4.5\endcsname{303}
\expandafter\gdef\csname odunum@ci@axis_breakdown.m2.qwen3_omni_instruct.axis4.4.5\endcsname{[0.547, 0.602]}
\expandafter\gdef\csname odunum@val@axis_breakdown.m2.qwen3_omni_instruct.axis5.audio_only.5.1\endcsname{0.630}
\expandafter\gdef\csname odunum@n@axis_breakdown.m2.qwen3_omni_instruct.axis5.audio_only.5.1\endcsname{45}
\expandafter\gdef\csname odunum@ci@axis_breakdown.m2.qwen3_omni_instruct.axis5.audio_only.5.1\endcsname{[0.563, 0.700]}
\expandafter\gdef\csname odunum@val@axis_breakdown.m2.qwen3_omni_instruct.axis5.audio_only.5.10\endcsname{0.567}
\expandafter\gdef\csname odunum@n@axis_breakdown.m2.qwen3_omni_instruct.axis5.audio_only.5.10\endcsname{72}
\expandafter\gdef\csname odunum@ci@axis_breakdown.m2.qwen3_omni_instruct.axis5.audio_only.5.10\endcsname{[0.504, 0.629]}
\expandafter\gdef\csname odunum@val@axis_breakdown.m2.qwen3_omni_instruct.axis5.audio_only.5.2\endcsname{0.634}
\expandafter\gdef\csname odunum@n@axis_breakdown.m2.qwen3_omni_instruct.axis5.audio_only.5.2\endcsname{75}
\expandafter\gdef\csname odunum@ci@axis_breakdown.m2.qwen3_omni_instruct.axis5.audio_only.5.2\endcsname{[0.571, 0.696]}
\expandafter\gdef\csname odunum@val@axis_breakdown.m2.qwen3_omni_instruct.axis5.audio_only.5.3\endcsname{0.699}
\expandafter\gdef\csname odunum@n@axis_breakdown.m2.qwen3_omni_instruct.axis5.audio_only.5.3\endcsname{50}
\expandafter\gdef\csname odunum@ci@axis_breakdown.m2.qwen3_omni_instruct.axis5.audio_only.5.3\endcsname{[0.624, 0.771]}
\expandafter\gdef\csname odunum@val@axis_breakdown.m2.qwen3_omni_instruct.axis5.audio_only.5.4\endcsname{0.568}
\expandafter\gdef\csname odunum@n@axis_breakdown.m2.qwen3_omni_instruct.axis5.audio_only.5.4\endcsname{68}
\expandafter\gdef\csname odunum@ci@axis_breakdown.m2.qwen3_omni_instruct.axis5.audio_only.5.4\endcsname{[0.508, 0.631]}
\expandafter\gdef\csname odunum@val@axis_breakdown.m2.qwen3_omni_instruct.axis5.audio_only.5.5\endcsname{0.609}
\expandafter\gdef\csname odunum@n@axis_breakdown.m2.qwen3_omni_instruct.axis5.audio_only.5.5\endcsname{52}
\expandafter\gdef\csname odunum@ci@axis_breakdown.m2.qwen3_omni_instruct.axis5.audio_only.5.5\endcsname{[0.542, 0.676]}
\expandafter\gdef\csname odunum@val@axis_breakdown.m2.qwen3_omni_instruct.axis5.audio_only.5.6\endcsname{0.636}
\expandafter\gdef\csname odunum@n@axis_breakdown.m2.qwen3_omni_instruct.axis5.audio_only.5.6\endcsname{52}
\expandafter\gdef\csname odunum@ci@axis_breakdown.m2.qwen3_omni_instruct.axis5.audio_only.5.6\endcsname{[0.563, 0.708]}
\expandafter\gdef\csname odunum@val@axis_breakdown.m2.qwen3_omni_instruct.axis5.audio_only.5.7\endcsname{0.565}
\expandafter\gdef\csname odunum@n@axis_breakdown.m2.qwen3_omni_instruct.axis5.audio_only.5.7\endcsname{46}
\expandafter\gdef\csname odunum@ci@axis_breakdown.m2.qwen3_omni_instruct.axis5.audio_only.5.7\endcsname{[0.494, 0.639]}
\expandafter\gdef\csname odunum@val@axis_breakdown.m2.qwen3_omni_instruct.axis5.audio_only.5.8\endcsname{0.557}
\expandafter\gdef\csname odunum@n@axis_breakdown.m2.qwen3_omni_instruct.axis5.audio_only.5.8\endcsname{57}
\expandafter\gdef\csname odunum@ci@axis_breakdown.m2.qwen3_omni_instruct.axis5.audio_only.5.8\endcsname{[0.492, 0.622]}
\expandafter\gdef\csname odunum@val@axis_breakdown.m2.qwen3_omni_instruct.axis5.audio_only.5.9\endcsname{0.649}
\expandafter\gdef\csname odunum@n@axis_breakdown.m2.qwen3_omni_instruct.axis5.audio_only.5.9\endcsname{53}
\expandafter\gdef\csname odunum@ci@axis_breakdown.m2.qwen3_omni_instruct.axis5.audio_only.5.9\endcsname{[0.560, 0.735]}
\expandafter\gdef\csname odunum@val@axis_breakdown.m2.qwen3_omni_instruct.axis5.audio_visual.5.1\endcsname{0.538}
\expandafter\gdef\csname odunum@n@axis_breakdown.m2.qwen3_omni_instruct.axis5.audio_visual.5.1\endcsname{114}
\expandafter\gdef\csname odunum@ci@axis_breakdown.m2.qwen3_omni_instruct.axis5.audio_visual.5.1\endcsname{[0.487, 0.588]}
\expandafter\gdef\csname odunum@val@axis_breakdown.m2.qwen3_omni_instruct.axis5.audio_visual.5.10\endcsname{0.536}
\expandafter\gdef\csname odunum@n@axis_breakdown.m2.qwen3_omni_instruct.axis5.audio_visual.5.10\endcsname{123}
\expandafter\gdef\csname odunum@ci@axis_breakdown.m2.qwen3_omni_instruct.axis5.audio_visual.5.10\endcsname{[0.491, 0.580]}
\expandafter\gdef\csname odunum@val@axis_breakdown.m2.qwen3_omni_instruct.axis5.audio_visual.5.2\endcsname{0.524}
\expandafter\gdef\csname odunum@n@axis_breakdown.m2.qwen3_omni_instruct.axis5.audio_visual.5.2\endcsname{126}
\expandafter\gdef\csname odunum@ci@axis_breakdown.m2.qwen3_omni_instruct.axis5.audio_visual.5.2\endcsname{[0.478, 0.572]}
\expandafter\gdef\csname odunum@val@axis_breakdown.m2.qwen3_omni_instruct.axis5.audio_visual.5.3\endcsname{0.527}
\expandafter\gdef\csname odunum@n@axis_breakdown.m2.qwen3_omni_instruct.axis5.audio_visual.5.3\endcsname{100}
\expandafter\gdef\csname odunum@ci@axis_breakdown.m2.qwen3_omni_instruct.axis5.audio_visual.5.3\endcsname{[0.477, 0.577]}
\expandafter\gdef\csname odunum@val@axis_breakdown.m2.qwen3_omni_instruct.axis5.audio_visual.5.4\endcsname{0.511}
\expandafter\gdef\csname odunum@n@axis_breakdown.m2.qwen3_omni_instruct.axis5.audio_visual.5.4\endcsname{97}
\expandafter\gdef\csname odunum@ci@axis_breakdown.m2.qwen3_omni_instruct.axis5.audio_visual.5.4\endcsname{[0.458, 0.564]}
\expandafter\gdef\csname odunum@val@axis_breakdown.m2.qwen3_omni_instruct.axis5.audio_visual.5.5\endcsname{0.565}
\expandafter\gdef\csname odunum@n@axis_breakdown.m2.qwen3_omni_instruct.axis5.audio_visual.5.5\endcsname{99}
\expandafter\gdef\csname odunum@ci@axis_breakdown.m2.qwen3_omni_instruct.axis5.audio_visual.5.5\endcsname{[0.512, 0.619]}
\expandafter\gdef\csname odunum@val@axis_breakdown.m2.qwen3_omni_instruct.axis5.audio_visual.5.6\endcsname{0.497}
\expandafter\gdef\csname odunum@n@axis_breakdown.m2.qwen3_omni_instruct.axis5.audio_visual.5.6\endcsname{99}
\expandafter\gdef\csname odunum@ci@axis_breakdown.m2.qwen3_omni_instruct.axis5.audio_visual.5.6\endcsname{[0.449, 0.546]}
\expandafter\gdef\csname odunum@val@axis_breakdown.m2.qwen3_omni_instruct.axis5.audio_visual.5.7\endcsname{0.542}
\expandafter\gdef\csname odunum@n@axis_breakdown.m2.qwen3_omni_instruct.axis5.audio_visual.5.7\endcsname{104}
\expandafter\gdef\csname odunum@ci@axis_breakdown.m2.qwen3_omni_instruct.axis5.audio_visual.5.7\endcsname{[0.494, 0.592]}
\expandafter\gdef\csname odunum@val@axis_breakdown.m2.qwen3_omni_instruct.axis5.audio_visual.5.8\endcsname{0.512}
\expandafter\gdef\csname odunum@n@axis_breakdown.m2.qwen3_omni_instruct.axis5.audio_visual.5.8\endcsname{121}
\expandafter\gdef\csname odunum@ci@axis_breakdown.m2.qwen3_omni_instruct.axis5.audio_visual.5.8\endcsname{[0.464, 0.560]}
\expandafter\gdef\csname odunum@val@axis_breakdown.m2.qwen3_omni_instruct.axis5.audio_visual.5.9\endcsname{0.564}
\expandafter\gdef\csname odunum@n@axis_breakdown.m2.qwen3_omni_instruct.axis5.audio_visual.5.9\endcsname{78}
\expandafter\gdef\csname odunum@ci@axis_breakdown.m2.qwen3_omni_instruct.axis5.audio_visual.5.9\endcsname{[0.505, 0.624]}
\expandafter\gdef\csname odunum@val@axis_breakdown.m2.qwen3_omni_instruct.axis6.6.1\endcsname{0.555}
\expandafter\gdef\csname odunum@n@axis_breakdown.m2.qwen3_omni_instruct.axis6.6.1\endcsname{230}
\expandafter\gdef\csname odunum@ci@axis_breakdown.m2.qwen3_omni_instruct.axis6.6.1\endcsname{[0.521, 0.591]}
\expandafter\gdef\csname odunum@val@axis_breakdown.m2.qwen3_omni_instruct.axis6.6.10\endcsname{0.536}
\expandafter\gdef\csname odunum@n@axis_breakdown.m2.qwen3_omni_instruct.axis6.6.10\endcsname{102}
\expandafter\gdef\csname odunum@ci@axis_breakdown.m2.qwen3_omni_instruct.axis6.6.10\endcsname{[0.482, 0.589]}
\expandafter\gdef\csname odunum@val@axis_breakdown.m2.qwen3_omni_instruct.axis6.6.12\endcsname{0.560}
\expandafter\gdef\csname odunum@n@axis_breakdown.m2.qwen3_omni_instruct.axis6.6.12\endcsname{129}
\expandafter\gdef\csname odunum@ci@axis_breakdown.m2.qwen3_omni_instruct.axis6.6.12\endcsname{[0.517, 0.604]}
\expandafter\gdef\csname odunum@val@axis_breakdown.m2.qwen3_omni_instruct.axis6.6.13\endcsname{0.511}
\expandafter\gdef\csname odunum@n@axis_breakdown.m2.qwen3_omni_instruct.axis6.6.13\endcsname{58}
\expandafter\gdef\csname odunum@ci@axis_breakdown.m2.qwen3_omni_instruct.axis6.6.13\endcsname{[0.439, 0.582]}
\expandafter\gdef\csname odunum@val@axis_breakdown.m2.qwen3_omni_instruct.axis6.6.14\endcsname{0.531}
\expandafter\gdef\csname odunum@n@axis_breakdown.m2.qwen3_omni_instruct.axis6.6.14\endcsname{82}
\expandafter\gdef\csname odunum@ci@axis_breakdown.m2.qwen3_omni_instruct.axis6.6.14\endcsname{[0.477, 0.586]}
\expandafter\gdef\csname odunum@val@axis_breakdown.m2.qwen3_omni_instruct.axis6.6.18\endcsname{0.572}
\expandafter\gdef\csname odunum@n@axis_breakdown.m2.qwen3_omni_instruct.axis6.6.18\endcsname{72}
\expandafter\gdef\csname odunum@ci@axis_breakdown.m2.qwen3_omni_instruct.axis6.6.18\endcsname{[0.514, 0.631]}
\expandafter\gdef\csname odunum@val@axis_breakdown.m2.qwen3_omni_instruct.axis6.6.19\endcsname{0.548}
\expandafter\gdef\csname odunum@n@axis_breakdown.m2.qwen3_omni_instruct.axis6.6.19\endcsname{44}
\expandafter\gdef\csname odunum@ci@axis_breakdown.m2.qwen3_omni_instruct.axis6.6.19\endcsname{[0.473, 0.625]}
\expandafter\gdef\csname odunum@val@axis_breakdown.m2.qwen3_omni_instruct.axis6.6.2\endcsname{0.619}
\expandafter\gdef\csname odunum@n@axis_breakdown.m2.qwen3_omni_instruct.axis6.6.2\endcsname{41}
\expandafter\gdef\csname odunum@ci@axis_breakdown.m2.qwen3_omni_instruct.axis6.6.2\endcsname{[0.541, 0.699]}
\expandafter\gdef\csname odunum@val@axis_breakdown.m2.qwen3_omni_instruct.axis6.6.3\endcsname{0.544}
\expandafter\gdef\csname odunum@n@axis_breakdown.m2.qwen3_omni_instruct.axis6.6.3\endcsname{44}
\expandafter\gdef\csname odunum@ci@axis_breakdown.m2.qwen3_omni_instruct.axis6.6.3\endcsname{[0.458, 0.632]}
\expandafter\gdef\csname odunum@val@axis_breakdown.m2.qwen3_omni_instruct.axis6.6.5\endcsname{0.514}
\expandafter\gdef\csname odunum@n@axis_breakdown.m2.qwen3_omni_instruct.axis6.6.5\endcsname{48}
\expandafter\gdef\csname odunum@ci@axis_breakdown.m2.qwen3_omni_instruct.axis6.6.5\endcsname{[0.431, 0.600]}
\expandafter\gdef\csname odunum@val@axis_breakdown.m2.qwen3_omni_instruct.axis6.6.6\endcsname{0.574}
\expandafter\gdef\csname odunum@n@axis_breakdown.m2.qwen3_omni_instruct.axis6.6.6\endcsname{134}
\expandafter\gdef\csname odunum@ci@axis_breakdown.m2.qwen3_omni_instruct.axis6.6.6\endcsname{[0.529, 0.621]}
\expandafter\gdef\csname odunum@val@axis_breakdown.m2.qwen3_omni_instruct.axis6.6.7\endcsname{0.587}
\expandafter\gdef\csname odunum@n@axis_breakdown.m2.qwen3_omni_instruct.axis6.6.7\endcsname{393}
\expandafter\gdef\csname odunum@ci@axis_breakdown.m2.qwen3_omni_instruct.axis6.6.7\endcsname{[0.559, 0.614]}
\expandafter\gdef\csname odunum@val@axis_breakdown.m2.qwen3_omni_instruct.axis6.6.8\endcsname{0.502}
\expandafter\gdef\csname odunum@n@axis_breakdown.m2.qwen3_omni_instruct.axis6.6.8\endcsname{49}
\expandafter\gdef\csname odunum@ci@axis_breakdown.m2.qwen3_omni_instruct.axis6.6.8\endcsname{[0.438, 0.564]}
\expandafter\gdef\csname odunum@val@axis_breakdown.m2.qwen3_omni_instruct.axis6.6.9\endcsname{0.549}
\expandafter\gdef\csname odunum@n@axis_breakdown.m2.qwen3_omni_instruct.axis6.6.9\endcsname{119}
\expandafter\gdef\csname odunum@ci@axis_breakdown.m2.qwen3_omni_instruct.axis6.6.9\endcsname{[0.501, 0.597]}
\expandafter\gdef\csname odunum@val@axis_breakdown.m2.qwen3_omni_instruct.difficulty.q1\endcsname{0.626}
\expandafter\gdef\csname odunum@n@axis_breakdown.m2.qwen3_omni_instruct.difficulty.q1\endcsname{358}
\expandafter\gdef\csname odunum@ci@axis_breakdown.m2.qwen3_omni_instruct.difficulty.q1\endcsname{[0.599, 0.654]}
\expandafter\gdef\csname odunum@val@axis_breakdown.m2.qwen3_omni_instruct.difficulty.q2\endcsname{0.569}
\expandafter\gdef\csname odunum@n@axis_breakdown.m2.qwen3_omni_instruct.difficulty.q2\endcsname{357}
\expandafter\gdef\csname odunum@ci@axis_breakdown.m2.qwen3_omni_instruct.difficulty.q2\endcsname{[0.543, 0.595]}
\expandafter\gdef\csname odunum@val@axis_breakdown.m2.qwen3_omni_instruct.difficulty.q3\endcsname{0.557}
\expandafter\gdef\csname odunum@n@axis_breakdown.m2.qwen3_omni_instruct.difficulty.q3\endcsname{358}
\expandafter\gdef\csname odunum@ci@axis_breakdown.m2.qwen3_omni_instruct.difficulty.q3\endcsname{[0.531, 0.584]}
\expandafter\gdef\csname odunum@val@axis_breakdown.m2.qwen3_omni_instruct.difficulty.q4\endcsname{0.518}
\expandafter\gdef\csname odunum@n@axis_breakdown.m2.qwen3_omni_instruct.difficulty.q4\endcsname{356}
\expandafter\gdef\csname odunum@ci@axis_breakdown.m2.qwen3_omni_instruct.difficulty.q4\endcsname{[0.491, 0.546]}
\expandafter\gdef\csname odunum@val@axis_breakdown.m2.qwen3_omni_think.axis1.audio_only.1.1\endcsname{0.642}
\expandafter\gdef\csname odunum@n@axis_breakdown.m2.qwen3_omni_think.axis1.audio_only.1.1\endcsname{158}
\expandafter\gdef\csname odunum@ci@axis_breakdown.m2.qwen3_omni_think.axis1.audio_only.1.1\endcsname{[0.597, 0.685]}
\expandafter\gdef\csname odunum@val@axis_breakdown.m2.qwen3_omni_think.axis1.audio_only.1.2\endcsname{0.627}
\expandafter\gdef\csname odunum@n@axis_breakdown.m2.qwen3_omni_think.axis1.audio_only.1.2\endcsname{379}
\expandafter\gdef\csname odunum@ci@axis_breakdown.m2.qwen3_omni_think.axis1.audio_only.1.2\endcsname{[0.599, 0.654]}
\expandafter\gdef\csname odunum@val@axis_breakdown.m2.qwen3_omni_think.axis1.audio_visual.1.1\endcsname{0.594}
\expandafter\gdef\csname odunum@n@axis_breakdown.m2.qwen3_omni_think.axis1.audio_visual.1.1\endcsname{94}
\expandafter\gdef\csname odunum@ci@axis_breakdown.m2.qwen3_omni_think.axis1.audio_visual.1.1\endcsname{[0.537, 0.649]}
\expandafter\gdef\csname odunum@val@axis_breakdown.m2.qwen3_omni_think.axis1.audio_visual.1.2\endcsname{0.570}
\expandafter\gdef\csname odunum@n@axis_breakdown.m2.qwen3_omni_think.axis1.audio_visual.1.2\endcsname{294}
\expandafter\gdef\csname odunum@ci@axis_breakdown.m2.qwen3_omni_think.axis1.audio_visual.1.2\endcsname{[0.539, 0.600]}
\expandafter\gdef\csname odunum@val@axis_breakdown.m2.qwen3_omni_think.axis1.audio_visual.1.3\endcsname{0.536}
\expandafter\gdef\csname odunum@n@axis_breakdown.m2.qwen3_omni_think.axis1.audio_visual.1.3\endcsname{291}
\expandafter\gdef\csname odunum@ci@axis_breakdown.m2.qwen3_omni_think.axis1.audio_visual.1.3\endcsname{[0.506, 0.566]}
\expandafter\gdef\csname odunum@val@axis_breakdown.m2.qwen3_omni_think.axis1.audio_visual.1.4\endcsname{0.552}
\expandafter\gdef\csname odunum@n@axis_breakdown.m2.qwen3_omni_think.axis1.audio_visual.1.4\endcsname{382}
\expandafter\gdef\csname odunum@ci@axis_breakdown.m2.qwen3_omni_think.axis1.audio_visual.1.4\endcsname{[0.527, 0.578]}
\expandafter\gdef\csname odunum@val@axis_breakdown.m2.qwen3_omni_think.axis2.2.1\endcsname{0.570}
\expandafter\gdef\csname odunum@n@axis_breakdown.m2.qwen3_omni_think.axis2.2.1\endcsname{143}
\expandafter\gdef\csname odunum@ci@axis_breakdown.m2.qwen3_omni_think.axis2.2.1\endcsname{[0.521, 0.618]}
\expandafter\gdef\csname odunum@val@axis_breakdown.m2.qwen3_omni_think.axis2.2.2\endcsname{0.581}
\expandafter\gdef\csname odunum@n@axis_breakdown.m2.qwen3_omni_think.axis2.2.2\endcsname{440}
\expandafter\gdef\csname odunum@ci@axis_breakdown.m2.qwen3_omni_think.axis2.2.2\endcsname{[0.557, 0.604]}
\expandafter\gdef\csname odunum@val@axis_breakdown.m2.qwen3_omni_think.axis2.2.3\endcsname{0.611}
\expandafter\gdef\csname odunum@n@axis_breakdown.m2.qwen3_omni_think.axis2.2.3\endcsname{531}
\expandafter\gdef\csname odunum@ci@axis_breakdown.m2.qwen3_omni_think.axis2.2.3\endcsname{[0.589, 0.634]}
\expandafter\gdef\csname odunum@val@axis_breakdown.m2.qwen3_omni_think.axis2.2.4\endcsname{0.542}
\expandafter\gdef\csname odunum@n@axis_breakdown.m2.qwen3_omni_think.axis2.2.4\endcsname{261}
\expandafter\gdef\csname odunum@ci@axis_breakdown.m2.qwen3_omni_think.axis2.2.4\endcsname{[0.508, 0.577]}
\expandafter\gdef\csname odunum@val@axis_breakdown.m2.qwen3_omni_think.axis2.2.5\endcsname{0.572}
\expandafter\gdef\csname odunum@n@axis_breakdown.m2.qwen3_omni_think.axis2.2.5\endcsname{256}
\expandafter\gdef\csname odunum@ci@axis_breakdown.m2.qwen3_omni_think.axis2.2.5\endcsname{[0.541, 0.604]}
\expandafter\gdef\csname odunum@val@axis_breakdown.m2.qwen3_omni_think.axis3.3.1\endcsname{0.582}
\expandafter\gdef\csname odunum@n@axis_breakdown.m2.qwen3_omni_think.axis3.3.1\endcsname{109}
\expandafter\gdef\csname odunum@ci@axis_breakdown.m2.qwen3_omni_think.axis3.3.1\endcsname{[0.536, 0.630]}
\expandafter\gdef\csname odunum@val@axis_breakdown.m2.qwen3_omni_think.axis3.3.2\endcsname{0.562}
\expandafter\gdef\csname odunum@n@axis_breakdown.m2.qwen3_omni_think.axis3.3.2\endcsname{116}
\expandafter\gdef\csname odunum@ci@axis_breakdown.m2.qwen3_omni_think.axis3.3.2\endcsname{[0.515, 0.609]}
\expandafter\gdef\csname odunum@val@axis_breakdown.m2.qwen3_omni_think.axis3.3.3\endcsname{0.542}
\expandafter\gdef\csname odunum@n@axis_breakdown.m2.qwen3_omni_think.axis3.3.3\endcsname{207}
\expandafter\gdef\csname odunum@ci@axis_breakdown.m2.qwen3_omni_think.axis3.3.3\endcsname{[0.504, 0.580]}
\expandafter\gdef\csname odunum@val@axis_breakdown.m2.qwen3_omni_think.axis3.3.4\endcsname{0.591}
\expandafter\gdef\csname odunum@n@axis_breakdown.m2.qwen3_omni_think.axis3.3.4\endcsname{404}
\expandafter\gdef\csname odunum@ci@axis_breakdown.m2.qwen3_omni_think.axis3.3.4\endcsname{[0.564, 0.618]}
\expandafter\gdef\csname odunum@val@axis_breakdown.m2.qwen3_omni_think.axis3.3.5\endcsname{0.521}
\expandafter\gdef\csname odunum@n@axis_breakdown.m2.qwen3_omni_think.axis3.3.5\endcsname{294}
\expandafter\gdef\csname odunum@ci@axis_breakdown.m2.qwen3_omni_think.axis3.3.5\endcsname{[0.490, 0.552]}
\expandafter\gdef\csname odunum@val@axis_breakdown.m2.qwen3_omni_think.axis3.3.6\endcsname{0.632}
\expandafter\gdef\csname odunum@n@axis_breakdown.m2.qwen3_omni_think.axis3.3.6\endcsname{501}
\expandafter\gdef\csname odunum@ci@axis_breakdown.m2.qwen3_omni_think.axis3.3.6\endcsname{[0.609, 0.654]}
\expandafter\gdef\csname odunum@val@axis_breakdown.m2.qwen3_omni_think.axis4.4.1\endcsname{0.569}
\expandafter\gdef\csname odunum@n@axis_breakdown.m2.qwen3_omni_think.axis4.4.1\endcsname{202}
\expandafter\gdef\csname odunum@ci@axis_breakdown.m2.qwen3_omni_think.axis4.4.1\endcsname{[0.534, 0.605]}
\expandafter\gdef\csname odunum@val@axis_breakdown.m2.qwen3_omni_think.axis4.4.2\endcsname{0.600}
\expandafter\gdef\csname odunum@n@axis_breakdown.m2.qwen3_omni_think.axis4.4.2\endcsname{457}
\expandafter\gdef\csname odunum@ci@axis_breakdown.m2.qwen3_omni_think.axis4.4.2\endcsname{[0.575, 0.624]}
\expandafter\gdef\csname odunum@val@axis_breakdown.m2.qwen3_omni_think.axis4.4.3\endcsname{0.582}
\expandafter\gdef\csname odunum@n@axis_breakdown.m2.qwen3_omni_think.axis4.4.3\endcsname{164}
\expandafter\gdef\csname odunum@ci@axis_breakdown.m2.qwen3_omni_think.axis4.4.3\endcsname{[0.536, 0.627]}
\expandafter\gdef\csname odunum@val@axis_breakdown.m2.qwen3_omni_think.axis4.4.4\endcsname{0.574}
\expandafter\gdef\csname odunum@n@axis_breakdown.m2.qwen3_omni_think.axis4.4.4\endcsname{505}
\expandafter\gdef\csname odunum@ci@axis_breakdown.m2.qwen3_omni_think.axis4.4.4\endcsname{[0.550, 0.598]}
\expandafter\gdef\csname odunum@val@axis_breakdown.m2.qwen3_omni_think.axis4.4.5\endcsname{0.577}
\expandafter\gdef\csname odunum@n@axis_breakdown.m2.qwen3_omni_think.axis4.4.5\endcsname{303}
\expandafter\gdef\csname odunum@ci@axis_breakdown.m2.qwen3_omni_think.axis4.4.5\endcsname{[0.548, 0.605]}
\expandafter\gdef\csname odunum@val@axis_breakdown.m2.qwen3_omni_think.axis5.audio_only.5.1\endcsname{0.653}
\expandafter\gdef\csname odunum@n@axis_breakdown.m2.qwen3_omni_think.axis5.audio_only.5.1\endcsname{45}
\expandafter\gdef\csname odunum@ci@axis_breakdown.m2.qwen3_omni_think.axis5.audio_only.5.1\endcsname{[0.559, 0.746]}
\expandafter\gdef\csname odunum@val@axis_breakdown.m2.qwen3_omni_think.axis5.audio_only.5.10\endcsname{0.598}
\expandafter\gdef\csname odunum@n@axis_breakdown.m2.qwen3_omni_think.axis5.audio_only.5.10\endcsname{72}
\expandafter\gdef\csname odunum@ci@axis_breakdown.m2.qwen3_omni_think.axis5.audio_only.5.10\endcsname{[0.526, 0.669]}
\expandafter\gdef\csname odunum@val@axis_breakdown.m2.qwen3_omni_think.axis5.audio_only.5.2\endcsname{0.628}
\expandafter\gdef\csname odunum@n@axis_breakdown.m2.qwen3_omni_think.axis5.audio_only.5.2\endcsname{75}
\expandafter\gdef\csname odunum@ci@axis_breakdown.m2.qwen3_omni_think.axis5.audio_only.5.2\endcsname{[0.564, 0.690]}
\expandafter\gdef\csname odunum@val@axis_breakdown.m2.qwen3_omni_think.axis5.audio_only.5.3\endcsname{0.682}
\expandafter\gdef\csname odunum@n@axis_breakdown.m2.qwen3_omni_think.axis5.audio_only.5.3\endcsname{50}
\expandafter\gdef\csname odunum@ci@axis_breakdown.m2.qwen3_omni_think.axis5.audio_only.5.3\endcsname{[0.609, 0.753]}
\expandafter\gdef\csname odunum@val@axis_breakdown.m2.qwen3_omni_think.axis5.audio_only.5.4\endcsname{0.608}
\expandafter\gdef\csname odunum@n@axis_breakdown.m2.qwen3_omni_think.axis5.audio_only.5.4\endcsname{68}
\expandafter\gdef\csname odunum@ci@axis_breakdown.m2.qwen3_omni_think.axis5.audio_only.5.4\endcsname{[0.550, 0.666]}
\expandafter\gdef\csname odunum@val@axis_breakdown.m2.qwen3_omni_think.axis5.audio_only.5.5\endcsname{0.629}
\expandafter\gdef\csname odunum@n@axis_breakdown.m2.qwen3_omni_think.axis5.audio_only.5.5\endcsname{52}
\expandafter\gdef\csname odunum@ci@axis_breakdown.m2.qwen3_omni_think.axis5.audio_only.5.5\endcsname{[0.559, 0.698]}
\expandafter\gdef\csname odunum@val@axis_breakdown.m2.qwen3_omni_think.axis5.audio_only.5.6\endcsname{0.656}
\expandafter\gdef\csname odunum@n@axis_breakdown.m2.qwen3_omni_think.axis5.audio_only.5.6\endcsname{52}
\expandafter\gdef\csname odunum@ci@axis_breakdown.m2.qwen3_omni_think.axis5.audio_only.5.6\endcsname{[0.600, 0.711]}
\expandafter\gdef\csname odunum@val@axis_breakdown.m2.qwen3_omni_think.axis5.audio_only.5.7\endcsname{0.554}
\expandafter\gdef\csname odunum@n@axis_breakdown.m2.qwen3_omni_think.axis5.audio_only.5.7\endcsname{46}
\expandafter\gdef\csname odunum@ci@axis_breakdown.m2.qwen3_omni_think.axis5.audio_only.5.7\endcsname{[0.479, 0.633]}
\expandafter\gdef\csname odunum@val@axis_breakdown.m2.qwen3_omni_think.axis5.audio_only.5.8\endcsname{0.594}
\expandafter\gdef\csname odunum@n@axis_breakdown.m2.qwen3_omni_think.axis5.audio_only.5.8\endcsname{57}
\expandafter\gdef\csname odunum@ci@axis_breakdown.m2.qwen3_omni_think.axis5.audio_only.5.8\endcsname{[0.513, 0.675]}
\expandafter\gdef\csname odunum@val@axis_breakdown.m2.qwen3_omni_think.axis5.audio_only.5.9\endcsname{0.717}
\expandafter\gdef\csname odunum@n@axis_breakdown.m2.qwen3_omni_think.axis5.audio_only.5.9\endcsname{53}
\expandafter\gdef\csname odunum@ci@axis_breakdown.m2.qwen3_omni_think.axis5.audio_only.5.9\endcsname{[0.649, 0.784]}
\expandafter\gdef\csname odunum@val@axis_breakdown.m2.qwen3_omni_think.axis5.audio_visual.5.1\endcsname{0.544}
\expandafter\gdef\csname odunum@n@axis_breakdown.m2.qwen3_omni_think.axis5.audio_visual.5.1\endcsname{114}
\expandafter\gdef\csname odunum@ci@axis_breakdown.m2.qwen3_omni_think.axis5.audio_visual.5.1\endcsname{[0.495, 0.593]}
\expandafter\gdef\csname odunum@val@axis_breakdown.m2.qwen3_omni_think.axis5.audio_visual.5.10\endcsname{0.617}
\expandafter\gdef\csname odunum@n@axis_breakdown.m2.qwen3_omni_think.axis5.audio_visual.5.10\endcsname{123}
\expandafter\gdef\csname odunum@ci@axis_breakdown.m2.qwen3_omni_think.axis5.audio_visual.5.10\endcsname{[0.569, 0.664]}
\expandafter\gdef\csname odunum@val@axis_breakdown.m2.qwen3_omni_think.axis5.audio_visual.5.2\endcsname{0.514}
\expandafter\gdef\csname odunum@n@axis_breakdown.m2.qwen3_omni_think.axis5.audio_visual.5.2\endcsname{126}
\expandafter\gdef\csname odunum@ci@axis_breakdown.m2.qwen3_omni_think.axis5.audio_visual.5.2\endcsname{[0.469, 0.560]}
\expandafter\gdef\csname odunum@val@axis_breakdown.m2.qwen3_omni_think.axis5.audio_visual.5.3\endcsname{0.533}
\expandafter\gdef\csname odunum@n@axis_breakdown.m2.qwen3_omni_think.axis5.audio_visual.5.3\endcsname{100}
\expandafter\gdef\csname odunum@ci@axis_breakdown.m2.qwen3_omni_think.axis5.audio_visual.5.3\endcsname{[0.482, 0.585]}
\expandafter\gdef\csname odunum@val@axis_breakdown.m2.qwen3_omni_think.axis5.audio_visual.5.4\endcsname{0.503}
\expandafter\gdef\csname odunum@n@axis_breakdown.m2.qwen3_omni_think.axis5.audio_visual.5.4\endcsname{97}
\expandafter\gdef\csname odunum@ci@axis_breakdown.m2.qwen3_omni_think.axis5.audio_visual.5.4\endcsname{[0.449, 0.558]}
\expandafter\gdef\csname odunum@val@axis_breakdown.m2.qwen3_omni_think.axis5.audio_visual.5.5\endcsname{0.590}
\expandafter\gdef\csname odunum@n@axis_breakdown.m2.qwen3_omni_think.axis5.audio_visual.5.5\endcsname{99}
\expandafter\gdef\csname odunum@ci@axis_breakdown.m2.qwen3_omni_think.axis5.audio_visual.5.5\endcsname{[0.544, 0.638]}
\expandafter\gdef\csname odunum@val@axis_breakdown.m2.qwen3_omni_think.axis5.audio_visual.5.6\endcsname{0.590}
\expandafter\gdef\csname odunum@n@axis_breakdown.m2.qwen3_omni_think.axis5.audio_visual.5.6\endcsname{99}
\expandafter\gdef\csname odunum@ci@axis_breakdown.m2.qwen3_omni_think.axis5.audio_visual.5.6\endcsname{[0.544, 0.638]}
\expandafter\gdef\csname odunum@val@axis_breakdown.m2.qwen3_omni_think.axis5.audio_visual.5.7\endcsname{0.557}
\expandafter\gdef\csname odunum@n@axis_breakdown.m2.qwen3_omni_think.axis5.audio_visual.5.7\endcsname{104}
\expandafter\gdef\csname odunum@ci@axis_breakdown.m2.qwen3_omni_think.axis5.audio_visual.5.7\endcsname{[0.504, 0.608]}
\expandafter\gdef\csname odunum@val@axis_breakdown.m2.qwen3_omni_think.axis5.audio_visual.5.8\endcsname{0.509}
\expandafter\gdef\csname odunum@n@axis_breakdown.m2.qwen3_omni_think.axis5.audio_visual.5.8\endcsname{121}
\expandafter\gdef\csname odunum@ci@axis_breakdown.m2.qwen3_omni_think.axis5.audio_visual.5.8\endcsname{[0.464, 0.555]}
\expandafter\gdef\csname odunum@val@axis_breakdown.m2.qwen3_omni_think.axis5.audio_visual.5.9\endcsname{0.628}
\expandafter\gdef\csname odunum@n@axis_breakdown.m2.qwen3_omni_think.axis5.audio_visual.5.9\endcsname{78}
\expandafter\gdef\csname odunum@ci@axis_breakdown.m2.qwen3_omni_think.axis5.audio_visual.5.9\endcsname{[0.574, 0.683]}
\expandafter\gdef\csname odunum@val@axis_breakdown.m2.qwen3_omni_think.axis6.6.1\endcsname{0.564}
\expandafter\gdef\csname odunum@n@axis_breakdown.m2.qwen3_omni_think.axis6.6.1\endcsname{230}
\expandafter\gdef\csname odunum@ci@axis_breakdown.m2.qwen3_omni_think.axis6.6.1\endcsname{[0.529, 0.599]}
\expandafter\gdef\csname odunum@val@axis_breakdown.m2.qwen3_omni_think.axis6.6.10\endcsname{0.601}
\expandafter\gdef\csname odunum@n@axis_breakdown.m2.qwen3_omni_think.axis6.6.10\endcsname{102}
\expandafter\gdef\csname odunum@ci@axis_breakdown.m2.qwen3_omni_think.axis6.6.10\endcsname{[0.553, 0.650]}
\expandafter\gdef\csname odunum@val@axis_breakdown.m2.qwen3_omni_think.axis6.6.12\endcsname{0.588}
\expandafter\gdef\csname odunum@n@axis_breakdown.m2.qwen3_omni_think.axis6.6.12\endcsname{129}
\expandafter\gdef\csname odunum@ci@axis_breakdown.m2.qwen3_omni_think.axis6.6.12\endcsname{[0.542, 0.634]}
\expandafter\gdef\csname odunum@val@axis_breakdown.m2.qwen3_omni_think.axis6.6.13\endcsname{0.548}
\expandafter\gdef\csname odunum@n@axis_breakdown.m2.qwen3_omni_think.axis6.6.13\endcsname{58}
\expandafter\gdef\csname odunum@ci@axis_breakdown.m2.qwen3_omni_think.axis6.6.13\endcsname{[0.470, 0.623]}
\expandafter\gdef\csname odunum@val@axis_breakdown.m2.qwen3_omni_think.axis6.6.14\endcsname{0.541}
\expandafter\gdef\csname odunum@n@axis_breakdown.m2.qwen3_omni_think.axis6.6.14\endcsname{82}
\expandafter\gdef\csname odunum@ci@axis_breakdown.m2.qwen3_omni_think.axis6.6.14\endcsname{[0.491, 0.591]}
\expandafter\gdef\csname odunum@val@axis_breakdown.m2.qwen3_omni_think.axis6.6.18\endcsname{0.607}
\expandafter\gdef\csname odunum@n@axis_breakdown.m2.qwen3_omni_think.axis6.6.18\endcsname{72}
\expandafter\gdef\csname odunum@ci@axis_breakdown.m2.qwen3_omni_think.axis6.6.18\endcsname{[0.546, 0.669]}
\expandafter\gdef\csname odunum@val@axis_breakdown.m2.qwen3_omni_think.axis6.6.19\endcsname{0.547}
\expandafter\gdef\csname odunum@n@axis_breakdown.m2.qwen3_omni_think.axis6.6.19\endcsname{44}
\expandafter\gdef\csname odunum@ci@axis_breakdown.m2.qwen3_omni_think.axis6.6.19\endcsname{[0.472, 0.620]}
\expandafter\gdef\csname odunum@val@axis_breakdown.m2.qwen3_omni_think.axis6.6.2\endcsname{0.623}
\expandafter\gdef\csname odunum@n@axis_breakdown.m2.qwen3_omni_think.axis6.6.2\endcsname{41}
\expandafter\gdef\csname odunum@ci@axis_breakdown.m2.qwen3_omni_think.axis6.6.2\endcsname{[0.554, 0.693]}
\expandafter\gdef\csname odunum@val@axis_breakdown.m2.qwen3_omni_think.axis6.6.3\endcsname{0.592}
\expandafter\gdef\csname odunum@n@axis_breakdown.m2.qwen3_omni_think.axis6.6.3\endcsname{44}
\expandafter\gdef\csname odunum@ci@axis_breakdown.m2.qwen3_omni_think.axis6.6.3\endcsname{[0.505, 0.677]}
\expandafter\gdef\csname odunum@val@axis_breakdown.m2.qwen3_omni_think.axis6.6.5\endcsname{0.569}
\expandafter\gdef\csname odunum@n@axis_breakdown.m2.qwen3_omni_think.axis6.6.5\endcsname{48}
\expandafter\gdef\csname odunum@ci@axis_breakdown.m2.qwen3_omni_think.axis6.6.5\endcsname{[0.487, 0.651]}
\expandafter\gdef\csname odunum@val@axis_breakdown.m2.qwen3_omni_think.axis6.6.6\endcsname{0.566}
\expandafter\gdef\csname odunum@n@axis_breakdown.m2.qwen3_omni_think.axis6.6.6\endcsname{134}
\expandafter\gdef\csname odunum@ci@axis_breakdown.m2.qwen3_omni_think.axis6.6.6\endcsname{[0.516, 0.616]}
\expandafter\gdef\csname odunum@val@axis_breakdown.m2.qwen3_omni_think.axis6.6.7\endcsname{0.626}
\expandafter\gdef\csname odunum@n@axis_breakdown.m2.qwen3_omni_think.axis6.6.7\endcsname{393}
\expandafter\gdef\csname odunum@ci@axis_breakdown.m2.qwen3_omni_think.axis6.6.7\endcsname{[0.598, 0.653]}
\expandafter\gdef\csname odunum@val@axis_breakdown.m2.qwen3_omni_think.axis6.6.8\endcsname{0.510}
\expandafter\gdef\csname odunum@n@axis_breakdown.m2.qwen3_omni_think.axis6.6.8\endcsname{49}
\expandafter\gdef\csname odunum@ci@axis_breakdown.m2.qwen3_omni_think.axis6.6.8\endcsname{[0.445, 0.578]}
\expandafter\gdef\csname odunum@val@axis_breakdown.m2.qwen3_omni_think.axis6.6.9\endcsname{0.561}
\expandafter\gdef\csname odunum@n@axis_breakdown.m2.qwen3_omni_think.axis6.6.9\endcsname{119}
\expandafter\gdef\csname odunum@ci@axis_breakdown.m2.qwen3_omni_think.axis6.6.9\endcsname{[0.517, 0.606]}
\expandafter\gdef\csname odunum@val@axis_breakdown.m2.qwen3_omni_think.difficulty.q1\endcsname{0.652}
\expandafter\gdef\csname odunum@n@axis_breakdown.m2.qwen3_omni_think.difficulty.q1\endcsname{358}
\expandafter\gdef\csname odunum@ci@axis_breakdown.m2.qwen3_omni_think.difficulty.q1\endcsname{[0.622, 0.682]}
\expandafter\gdef\csname odunum@val@axis_breakdown.m2.qwen3_omni_think.difficulty.q2\endcsname{0.600}
\expandafter\gdef\csname odunum@n@axis_breakdown.m2.qwen3_omni_think.difficulty.q2\endcsname{357}
\expandafter\gdef\csname odunum@ci@axis_breakdown.m2.qwen3_omni_think.difficulty.q2\endcsname{[0.575, 0.625]}
\expandafter\gdef\csname odunum@val@axis_breakdown.m2.qwen3_omni_think.difficulty.q3\endcsname{0.589}
\expandafter\gdef\csname odunum@n@axis_breakdown.m2.qwen3_omni_think.difficulty.q3\endcsname{358}
\expandafter\gdef\csname odunum@ci@axis_breakdown.m2.qwen3_omni_think.difficulty.q3\endcsname{[0.561, 0.616]}
\expandafter\gdef\csname odunum@val@axis_breakdown.m2.qwen3_omni_think.difficulty.q4\endcsname{0.536}
\expandafter\gdef\csname odunum@n@axis_breakdown.m2.qwen3_omni_think.difficulty.q4\endcsname{356}
\expandafter\gdef\csname odunum@ci@axis_breakdown.m2.qwen3_omni_think.difficulty.q4\endcsname{[0.510, 0.563]}
\expandafter\gdef\csname odunum@val@axis_breakdown.m2.qwen_plus.axis1.audio_only.1.1\endcsname{0.750}
\expandafter\gdef\csname odunum@n@axis_breakdown.m2.qwen_plus.axis1.audio_only.1.1\endcsname{158}
\expandafter\gdef\csname odunum@ci@axis_breakdown.m2.qwen_plus.axis1.audio_only.1.1\endcsname{[0.710, 0.787]}
\expandafter\gdef\csname odunum@val@axis_breakdown.m2.qwen_plus.axis1.audio_only.1.2\endcsname{0.698}
\expandafter\gdef\csname odunum@n@axis_breakdown.m2.qwen_plus.axis1.audio_only.1.2\endcsname{379}
\expandafter\gdef\csname odunum@ci@axis_breakdown.m2.qwen_plus.axis1.audio_only.1.2\endcsname{[0.671, 0.725]}
\expandafter\gdef\csname odunum@val@axis_breakdown.m2.qwen_plus.axis1.audio_visual.1.1\endcsname{0.727}
\expandafter\gdef\csname odunum@n@axis_breakdown.m2.qwen_plus.axis1.audio_visual.1.1\endcsname{94}
\expandafter\gdef\csname odunum@ci@axis_breakdown.m2.qwen_plus.axis1.audio_visual.1.1\endcsname{[0.676, 0.776]}
\expandafter\gdef\csname odunum@val@axis_breakdown.m2.qwen_plus.axis1.audio_visual.1.2\endcsname{0.635}
\expandafter\gdef\csname odunum@n@axis_breakdown.m2.qwen_plus.axis1.audio_visual.1.2\endcsname{294}
\expandafter\gdef\csname odunum@ci@axis_breakdown.m2.qwen_plus.axis1.audio_visual.1.2\endcsname{[0.605, 0.666]}
\expandafter\gdef\csname odunum@val@axis_breakdown.m2.qwen_plus.axis1.audio_visual.1.3\endcsname{0.673}
\expandafter\gdef\csname odunum@n@axis_breakdown.m2.qwen_plus.axis1.audio_visual.1.3\endcsname{290}
\expandafter\gdef\csname odunum@ci@axis_breakdown.m2.qwen_plus.axis1.audio_visual.1.3\endcsname{[0.643, 0.703]}
\expandafter\gdef\csname odunum@val@axis_breakdown.m2.qwen_plus.axis1.audio_visual.1.4\endcsname{0.646}
\expandafter\gdef\csname odunum@n@axis_breakdown.m2.qwen_plus.axis1.audio_visual.1.4\endcsname{382}
\expandafter\gdef\csname odunum@ci@axis_breakdown.m2.qwen_plus.axis1.audio_visual.1.4\endcsname{[0.621, 0.671]}
\expandafter\gdef\csname odunum@val@axis_breakdown.m2.qwen_plus.axis2.2.1\endcsname{0.623}
\expandafter\gdef\csname odunum@n@axis_breakdown.m2.qwen_plus.axis2.2.1\endcsname{143}
\expandafter\gdef\csname odunum@ci@axis_breakdown.m2.qwen_plus.axis2.2.1\endcsname{[0.573, 0.672]}
\expandafter\gdef\csname odunum@val@axis_breakdown.m2.qwen_plus.axis2.2.2\endcsname{0.701}
\expandafter\gdef\csname odunum@n@axis_breakdown.m2.qwen_plus.axis2.2.2\endcsname{439}
\expandafter\gdef\csname odunum@ci@axis_breakdown.m2.qwen_plus.axis2.2.2\endcsname{[0.679, 0.723]}
\expandafter\gdef\csname odunum@val@axis_breakdown.m2.qwen_plus.axis2.2.3\endcsname{0.724}
\expandafter\gdef\csname odunum@n@axis_breakdown.m2.qwen_plus.axis2.2.3\endcsname{531}
\expandafter\gdef\csname odunum@ci@axis_breakdown.m2.qwen_plus.axis2.2.3\endcsname{[0.704, 0.743]}
\expandafter\gdef\csname odunum@val@axis_breakdown.m2.qwen_plus.axis2.2.4\endcsname{0.589}
\expandafter\gdef\csname odunum@n@axis_breakdown.m2.qwen_plus.axis2.2.4\endcsname{261}
\expandafter\gdef\csname odunum@ci@axis_breakdown.m2.qwen_plus.axis2.2.4\endcsname{[0.553, 0.624]}
\expandafter\gdef\csname odunum@val@axis_breakdown.m2.qwen_plus.axis2.2.5\endcsname{0.665}
\expandafter\gdef\csname odunum@n@axis_breakdown.m2.qwen_plus.axis2.2.5\endcsname{256}
\expandafter\gdef\csname odunum@ci@axis_breakdown.m2.qwen_plus.axis2.2.5\endcsname{[0.632, 0.697]}
\expandafter\gdef\csname odunum@val@axis_breakdown.m2.qwen_plus.axis3.3.1\endcsname{0.645}
\expandafter\gdef\csname odunum@n@axis_breakdown.m2.qwen_plus.axis3.3.1\endcsname{109}
\expandafter\gdef\csname odunum@ci@axis_breakdown.m2.qwen_plus.axis3.3.1\endcsname{[0.591, 0.698]}
\expandafter\gdef\csname odunum@val@axis_breakdown.m2.qwen_plus.axis3.3.2\endcsname{0.696}
\expandafter\gdef\csname odunum@n@axis_breakdown.m2.qwen_plus.axis3.3.2\endcsname{115}
\expandafter\gdef\csname odunum@ci@axis_breakdown.m2.qwen_plus.axis3.3.2\endcsname{[0.652, 0.741]}
\expandafter\gdef\csname odunum@val@axis_breakdown.m2.qwen_plus.axis3.3.3\endcsname{0.642}
\expandafter\gdef\csname odunum@n@axis_breakdown.m2.qwen_plus.axis3.3.3\endcsname{207}
\expandafter\gdef\csname odunum@ci@axis_breakdown.m2.qwen_plus.axis3.3.3\endcsname{[0.602, 0.681]}
\expandafter\gdef\csname odunum@val@axis_breakdown.m2.qwen_plus.axis3.3.4\endcsname{0.672}
\expandafter\gdef\csname odunum@n@axis_breakdown.m2.qwen_plus.axis3.3.4\endcsname{404}
\expandafter\gdef\csname odunum@ci@axis_breakdown.m2.qwen_plus.axis3.3.4\endcsname{[0.647, 0.697]}
\expandafter\gdef\csname odunum@val@axis_breakdown.m2.qwen_plus.axis3.3.5\endcsname{0.631}
\expandafter\gdef\csname odunum@n@axis_breakdown.m2.qwen_plus.axis3.3.5\endcsname{294}
\expandafter\gdef\csname odunum@ci@axis_breakdown.m2.qwen_plus.axis3.3.5\endcsname{[0.599, 0.663]}
\expandafter\gdef\csname odunum@val@axis_breakdown.m2.qwen_plus.axis3.3.6\endcsname{0.728}
\expandafter\gdef\csname odunum@n@axis_breakdown.m2.qwen_plus.axis3.3.6\endcsname{501}
\expandafter\gdef\csname odunum@ci@axis_breakdown.m2.qwen_plus.axis3.3.6\endcsname{[0.708, 0.749]}
\expandafter\gdef\csname odunum@val@axis_breakdown.m2.qwen_plus.axis4.4.1\endcsname{0.686}
\expandafter\gdef\csname odunum@n@axis_breakdown.m2.qwen_plus.axis4.4.1\endcsname{202}
\expandafter\gdef\csname odunum@ci@axis_breakdown.m2.qwen_plus.axis4.4.1\endcsname{[0.653, 0.719]}
\expandafter\gdef\csname odunum@val@axis_breakdown.m2.qwen_plus.axis4.4.2\endcsname{0.672}
\expandafter\gdef\csname odunum@n@axis_breakdown.m2.qwen_plus.axis4.4.2\endcsname{457}
\expandafter\gdef\csname odunum@ci@axis_breakdown.m2.qwen_plus.axis4.4.2\endcsname{[0.649, 0.695]}
\expandafter\gdef\csname odunum@val@axis_breakdown.m2.qwen_plus.axis4.4.3\endcsname{0.650}
\expandafter\gdef\csname odunum@n@axis_breakdown.m2.qwen_plus.axis4.4.3\endcsname{164}
\expandafter\gdef\csname odunum@ci@axis_breakdown.m2.qwen_plus.axis4.4.3\endcsname{[0.605, 0.694]}
\expandafter\gdef\csname odunum@val@axis_breakdown.m2.qwen_plus.axis4.4.4\endcsname{0.680}
\expandafter\gdef\csname odunum@n@axis_breakdown.m2.qwen_plus.axis4.4.4\endcsname{505}
\expandafter\gdef\csname odunum@ci@axis_breakdown.m2.qwen_plus.axis4.4.4\endcsname{[0.657, 0.703]}
\expandafter\gdef\csname odunum@val@axis_breakdown.m2.qwen_plus.axis4.4.5\endcsname{0.693}
\expandafter\gdef\csname odunum@n@axis_breakdown.m2.qwen_plus.axis4.4.5\endcsname{302}
\expandafter\gdef\csname odunum@ci@axis_breakdown.m2.qwen_plus.axis4.4.5\endcsname{[0.664, 0.723]}
\expandafter\gdef\csname odunum@val@axis_breakdown.m2.qwen_plus.axis5.audio_only.5.1\endcsname{0.752}
\expandafter\gdef\csname odunum@n@axis_breakdown.m2.qwen_plus.axis5.audio_only.5.1\endcsname{45}
\expandafter\gdef\csname odunum@ci@axis_breakdown.m2.qwen_plus.axis5.audio_only.5.1\endcsname{[0.674, 0.822]}
\expandafter\gdef\csname odunum@val@axis_breakdown.m2.qwen_plus.axis5.audio_only.5.10\endcsname{0.684}
\expandafter\gdef\csname odunum@n@axis_breakdown.m2.qwen_plus.axis5.audio_only.5.10\endcsname{72}
\expandafter\gdef\csname odunum@ci@axis_breakdown.m2.qwen_plus.axis5.audio_only.5.10\endcsname{[0.624, 0.746]}
\expandafter\gdef\csname odunum@val@axis_breakdown.m2.qwen_plus.axis5.audio_only.5.2\endcsname{0.728}
\expandafter\gdef\csname odunum@n@axis_breakdown.m2.qwen_plus.axis5.audio_only.5.2\endcsname{75}
\expandafter\gdef\csname odunum@ci@axis_breakdown.m2.qwen_plus.axis5.audio_only.5.2\endcsname{[0.669, 0.785]}
\expandafter\gdef\csname odunum@val@axis_breakdown.m2.qwen_plus.axis5.audio_only.5.3\endcsname{0.748}
\expandafter\gdef\csname odunum@n@axis_breakdown.m2.qwen_plus.axis5.audio_only.5.3\endcsname{50}
\expandafter\gdef\csname odunum@ci@axis_breakdown.m2.qwen_plus.axis5.audio_only.5.3\endcsname{[0.680, 0.813]}
\expandafter\gdef\csname odunum@val@axis_breakdown.m2.qwen_plus.axis5.audio_only.5.4\endcsname{0.688}
\expandafter\gdef\csname odunum@n@axis_breakdown.m2.qwen_plus.axis5.audio_only.5.4\endcsname{68}
\expandafter\gdef\csname odunum@ci@axis_breakdown.m2.qwen_plus.axis5.audio_only.5.4\endcsname{[0.627, 0.748]}
\expandafter\gdef\csname odunum@val@axis_breakdown.m2.qwen_plus.axis5.audio_only.5.5\endcsname{0.738}
\expandafter\gdef\csname odunum@n@axis_breakdown.m2.qwen_plus.axis5.audio_only.5.5\endcsname{52}
\expandafter\gdef\csname odunum@ci@axis_breakdown.m2.qwen_plus.axis5.audio_only.5.5\endcsname{[0.670, 0.802]}
\expandafter\gdef\csname odunum@val@axis_breakdown.m2.qwen_plus.axis5.audio_only.5.6\endcsname{0.709}
\expandafter\gdef\csname odunum@n@axis_breakdown.m2.qwen_plus.axis5.audio_only.5.6\endcsname{52}
\expandafter\gdef\csname odunum@ci@axis_breakdown.m2.qwen_plus.axis5.audio_only.5.6\endcsname{[0.643, 0.773]}
\expandafter\gdef\csname odunum@val@axis_breakdown.m2.qwen_plus.axis5.audio_only.5.7\endcsname{0.699}
\expandafter\gdef\csname odunum@n@axis_breakdown.m2.qwen_plus.axis5.audio_only.5.7\endcsname{46}
\expandafter\gdef\csname odunum@ci@axis_breakdown.m2.qwen_plus.axis5.audio_only.5.7\endcsname{[0.630, 0.770]}
\expandafter\gdef\csname odunum@val@axis_breakdown.m2.qwen_plus.axis5.audio_only.5.8\endcsname{0.691}
\expandafter\gdef\csname odunum@n@axis_breakdown.m2.qwen_plus.axis5.audio_only.5.8\endcsname{57}
\expandafter\gdef\csname odunum@ci@axis_breakdown.m2.qwen_plus.axis5.audio_only.5.8\endcsname{[0.607, 0.770]}
\expandafter\gdef\csname odunum@val@axis_breakdown.m2.qwen_plus.axis5.audio_only.5.9\endcsname{0.737}
\expandafter\gdef\csname odunum@n@axis_breakdown.m2.qwen_plus.axis5.audio_only.5.9\endcsname{53}
\expandafter\gdef\csname odunum@ci@axis_breakdown.m2.qwen_plus.axis5.audio_only.5.9\endcsname{[0.662, 0.808]}
\expandafter\gdef\csname odunum@val@axis_breakdown.m2.qwen_plus.axis5.audio_visual.5.1\endcsname{0.654}
\expandafter\gdef\csname odunum@n@axis_breakdown.m2.qwen_plus.axis5.audio_visual.5.1\endcsname{114}
\expandafter\gdef\csname odunum@ci@axis_breakdown.m2.qwen_plus.axis5.audio_visual.5.1\endcsname{[0.606, 0.702]}
\expandafter\gdef\csname odunum@val@axis_breakdown.m2.qwen_plus.axis5.audio_visual.5.10\endcsname{0.715}
\expandafter\gdef\csname odunum@n@axis_breakdown.m2.qwen_plus.axis5.audio_visual.5.10\endcsname{123}
\expandafter\gdef\csname odunum@ci@axis_breakdown.m2.qwen_plus.axis5.audio_visual.5.10\endcsname{[0.676, 0.752]}
\expandafter\gdef\csname odunum@val@axis_breakdown.m2.qwen_plus.axis5.audio_visual.5.2\endcsname{0.593}
\expandafter\gdef\csname odunum@n@axis_breakdown.m2.qwen_plus.axis5.audio_visual.5.2\endcsname{126}
\expandafter\gdef\csname odunum@ci@axis_breakdown.m2.qwen_plus.axis5.audio_visual.5.2\endcsname{[0.547, 0.640]}
\expandafter\gdef\csname odunum@val@axis_breakdown.m2.qwen_plus.axis5.audio_visual.5.3\endcsname{0.667}
\expandafter\gdef\csname odunum@n@axis_breakdown.m2.qwen_plus.axis5.audio_visual.5.3\endcsname{100}
\expandafter\gdef\csname odunum@ci@axis_breakdown.m2.qwen_plus.axis5.audio_visual.5.3\endcsname{[0.620, 0.712]}
\expandafter\gdef\csname odunum@val@axis_breakdown.m2.qwen_plus.axis5.audio_visual.5.4\endcsname{0.628}
\expandafter\gdef\csname odunum@n@axis_breakdown.m2.qwen_plus.axis5.audio_visual.5.4\endcsname{97}
\expandafter\gdef\csname odunum@ci@axis_breakdown.m2.qwen_plus.axis5.audio_visual.5.4\endcsname{[0.575, 0.681]}
\expandafter\gdef\csname odunum@val@axis_breakdown.m2.qwen_plus.axis5.audio_visual.5.5\endcsname{0.714}
\expandafter\gdef\csname odunum@n@axis_breakdown.m2.qwen_plus.axis5.audio_visual.5.5\endcsname{98}
\expandafter\gdef\csname odunum@ci@axis_breakdown.m2.qwen_plus.axis5.audio_visual.5.5\endcsname{[0.661, 0.767]}
\expandafter\gdef\csname odunum@val@axis_breakdown.m2.qwen_plus.axis5.audio_visual.5.6\endcsname{0.675}
\expandafter\gdef\csname odunum@n@axis_breakdown.m2.qwen_plus.axis5.audio_visual.5.6\endcsname{99}
\expandafter\gdef\csname odunum@ci@axis_breakdown.m2.qwen_plus.axis5.audio_visual.5.6\endcsname{[0.626, 0.722]}
\expandafter\gdef\csname odunum@val@axis_breakdown.m2.qwen_plus.axis5.audio_visual.5.7\endcsname{0.644}
\expandafter\gdef\csname odunum@n@axis_breakdown.m2.qwen_plus.axis5.audio_visual.5.7\endcsname{104}
\expandafter\gdef\csname odunum@ci@axis_breakdown.m2.qwen_plus.axis5.audio_visual.5.7\endcsname{[0.593, 0.693]}
\expandafter\gdef\csname odunum@val@axis_breakdown.m2.qwen_plus.axis5.audio_visual.5.8\endcsname{0.632}
\expandafter\gdef\csname odunum@n@axis_breakdown.m2.qwen_plus.axis5.audio_visual.5.8\endcsname{121}
\expandafter\gdef\csname odunum@ci@axis_breakdown.m2.qwen_plus.axis5.audio_visual.5.8\endcsname{[0.585, 0.677]}
\expandafter\gdef\csname odunum@val@axis_breakdown.m2.qwen_plus.axis5.audio_visual.5.9\endcsname{0.670}
\expandafter\gdef\csname odunum@n@axis_breakdown.m2.qwen_plus.axis5.audio_visual.5.9\endcsname{78}
\expandafter\gdef\csname odunum@ci@axis_breakdown.m2.qwen_plus.axis5.audio_visual.5.9\endcsname{[0.609, 0.729]}
\expandafter\gdef\csname odunum@val@axis_breakdown.m2.qwen_plus.axis6.6.1\endcsname{0.684}
\expandafter\gdef\csname odunum@n@axis_breakdown.m2.qwen_plus.axis6.6.1\endcsname{230}
\expandafter\gdef\csname odunum@ci@axis_breakdown.m2.qwen_plus.axis6.6.1\endcsname{[0.649, 0.719]}
\expandafter\gdef\csname odunum@val@axis_breakdown.m2.qwen_plus.axis6.6.10\endcsname{0.698}
\expandafter\gdef\csname odunum@n@axis_breakdown.m2.qwen_plus.axis6.6.10\endcsname{102}
\expandafter\gdef\csname odunum@ci@axis_breakdown.m2.qwen_plus.axis6.6.10\endcsname{[0.656, 0.738]}
\expandafter\gdef\csname odunum@val@axis_breakdown.m2.qwen_plus.axis6.6.12\endcsname{0.672}
\expandafter\gdef\csname odunum@n@axis_breakdown.m2.qwen_plus.axis6.6.12\endcsname{129}
\expandafter\gdef\csname odunum@ci@axis_breakdown.m2.qwen_plus.axis6.6.12\endcsname{[0.627, 0.717]}
\expandafter\gdef\csname odunum@val@axis_breakdown.m2.qwen_plus.axis6.6.13\endcsname{0.653}
\expandafter\gdef\csname odunum@n@axis_breakdown.m2.qwen_plus.axis6.6.13\endcsname{58}
\expandafter\gdef\csname odunum@ci@axis_breakdown.m2.qwen_plus.axis6.6.13\endcsname{[0.585, 0.719]}
\expandafter\gdef\csname odunum@val@axis_breakdown.m2.qwen_plus.axis6.6.14\endcsname{0.616}
\expandafter\gdef\csname odunum@n@axis_breakdown.m2.qwen_plus.axis6.6.14\endcsname{82}
\expandafter\gdef\csname odunum@ci@axis_breakdown.m2.qwen_plus.axis6.6.14\endcsname{[0.559, 0.671]}
\expandafter\gdef\csname odunum@val@axis_breakdown.m2.qwen_plus.axis6.6.18\endcsname{0.700}
\expandafter\gdef\csname odunum@n@axis_breakdown.m2.qwen_plus.axis6.6.18\endcsname{72}
\expandafter\gdef\csname odunum@ci@axis_breakdown.m2.qwen_plus.axis6.6.18\endcsname{[0.644, 0.756]}
\expandafter\gdef\csname odunum@val@axis_breakdown.m2.qwen_plus.axis6.6.19\endcsname{0.622}
\expandafter\gdef\csname odunum@n@axis_breakdown.m2.qwen_plus.axis6.6.19\endcsname{44}
\expandafter\gdef\csname odunum@ci@axis_breakdown.m2.qwen_plus.axis6.6.19\endcsname{[0.519, 0.719]}
\expandafter\gdef\csname odunum@val@axis_breakdown.m2.qwen_plus.axis6.6.2\endcsname{0.710}
\expandafter\gdef\csname odunum@n@axis_breakdown.m2.qwen_plus.axis6.6.2\endcsname{41}
\expandafter\gdef\csname odunum@ci@axis_breakdown.m2.qwen_plus.axis6.6.2\endcsname{[0.637, 0.780]}
\expandafter\gdef\csname odunum@val@axis_breakdown.m2.qwen_plus.axis6.6.3\endcsname{0.685}
\expandafter\gdef\csname odunum@n@axis_breakdown.m2.qwen_plus.axis6.6.3\endcsname{44}
\expandafter\gdef\csname odunum@ci@axis_breakdown.m2.qwen_plus.axis6.6.3\endcsname{[0.595, 0.772]}
\expandafter\gdef\csname odunum@val@axis_breakdown.m2.qwen_plus.axis6.6.5\endcsname{0.698}
\expandafter\gdef\csname odunum@n@axis_breakdown.m2.qwen_plus.axis6.6.5\endcsname{48}
\expandafter\gdef\csname odunum@ci@axis_breakdown.m2.qwen_plus.axis6.6.5\endcsname{[0.628, 0.767]}
\expandafter\gdef\csname odunum@val@axis_breakdown.m2.qwen_plus.axis6.6.6\endcsname{0.704}
\expandafter\gdef\csname odunum@n@axis_breakdown.m2.qwen_plus.axis6.6.6\endcsname{133}
\expandafter\gdef\csname odunum@ci@axis_breakdown.m2.qwen_plus.axis6.6.6\endcsname{[0.659, 0.749]}
\expandafter\gdef\csname odunum@val@axis_breakdown.m2.qwen_plus.axis6.6.7\endcsname{0.699}
\expandafter\gdef\csname odunum@n@axis_breakdown.m2.qwen_plus.axis6.6.7\endcsname{393}
\expandafter\gdef\csname odunum@ci@axis_breakdown.m2.qwen_plus.axis6.6.7\endcsname{[0.673, 0.724]}
\expandafter\gdef\csname odunum@val@axis_breakdown.m2.qwen_plus.axis6.6.8\endcsname{0.592}
\expandafter\gdef\csname odunum@n@axis_breakdown.m2.qwen_plus.axis6.6.8\endcsname{49}
\expandafter\gdef\csname odunum@ci@axis_breakdown.m2.qwen_plus.axis6.6.8\endcsname{[0.516, 0.667]}
\expandafter\gdef\csname odunum@val@axis_breakdown.m2.qwen_plus.axis6.6.9\endcsname{0.678}
\expandafter\gdef\csname odunum@n@axis_breakdown.m2.qwen_plus.axis6.6.9\endcsname{119}
\expandafter\gdef\csname odunum@ci@axis_breakdown.m2.qwen_plus.axis6.6.9\endcsname{[0.632, 0.723]}
\expandafter\gdef\csname odunum@val@axis_breakdown.m2.qwen_plus.difficulty.q1\endcsname{0.706}
\expandafter\gdef\csname odunum@n@axis_breakdown.m2.qwen_plus.difficulty.q1\endcsname{358}
\expandafter\gdef\csname odunum@ci@axis_breakdown.m2.qwen_plus.difficulty.q1\endcsname{[0.677, 0.735]}
\expandafter\gdef\csname odunum@val@axis_breakdown.m2.qwen_plus.difficulty.q2\endcsname{0.685}
\expandafter\gdef\csname odunum@n@axis_breakdown.m2.qwen_plus.difficulty.q2\endcsname{357}
\expandafter\gdef\csname odunum@ci@axis_breakdown.m2.qwen_plus.difficulty.q2\endcsname{[0.659, 0.711]}
\expandafter\gdef\csname odunum@val@axis_breakdown.m2.qwen_plus.difficulty.q3\endcsname{0.670}
\expandafter\gdef\csname odunum@n@axis_breakdown.m2.qwen_plus.difficulty.q3\endcsname{358}
\expandafter\gdef\csname odunum@ci@axis_breakdown.m2.qwen_plus.difficulty.q3\endcsname{[0.643, 0.696]}
\expandafter\gdef\csname odunum@val@axis_breakdown.m2.qwen_plus.difficulty.q4\endcsname{0.671}
\expandafter\gdef\csname odunum@n@axis_breakdown.m2.qwen_plus.difficulty.q4\endcsname{356}
\expandafter\gdef\csname odunum@ci@axis_breakdown.m2.qwen_plus.difficulty.q4\endcsname{[0.644, 0.697]}
\expandafter\gdef\csname odunum@val@axis_breakdown.m2.salmonn2_7b.axis1.audio_only.1.1\endcsname{0.209}
\expandafter\gdef\csname odunum@n@axis_breakdown.m2.salmonn2_7b.axis1.audio_only.1.1\endcsname{158}
\expandafter\gdef\csname odunum@ci@axis_breakdown.m2.salmonn2_7b.axis1.audio_only.1.1\endcsname{[0.162, 0.260]}
\expandafter\gdef\csname odunum@val@axis_breakdown.m2.salmonn2_7b.axis1.audio_only.1.2\endcsname{0.180}
\expandafter\gdef\csname odunum@n@axis_breakdown.m2.salmonn2_7b.axis1.audio_only.1.2\endcsname{377}
\expandafter\gdef\csname odunum@ci@axis_breakdown.m2.salmonn2_7b.axis1.audio_only.1.2\endcsname{[0.154, 0.206]}
\expandafter\gdef\csname odunum@val@axis_breakdown.m2.salmonn2_7b.axis1.audio_visual.1.1\endcsname{0.050}
\expandafter\gdef\csname odunum@n@axis_breakdown.m2.salmonn2_7b.axis1.audio_visual.1.1\endcsname{93}
\expandafter\gdef\csname odunum@ci@axis_breakdown.m2.salmonn2_7b.axis1.audio_visual.1.1\endcsname{[0.023, 0.081]}
\expandafter\gdef\csname odunum@val@axis_breakdown.m2.salmonn2_7b.axis1.audio_visual.1.2\endcsname{0.080}
\expandafter\gdef\csname odunum@n@axis_breakdown.m2.salmonn2_7b.axis1.audio_visual.1.2\endcsname{289}
\expandafter\gdef\csname odunum@ci@axis_breakdown.m2.salmonn2_7b.axis1.audio_visual.1.2\endcsname{[0.060, 0.100]}
\expandafter\gdef\csname odunum@val@axis_breakdown.m2.salmonn2_7b.axis1.audio_visual.1.3\endcsname{0.056}
\expandafter\gdef\csname odunum@n@axis_breakdown.m2.salmonn2_7b.axis1.audio_visual.1.3\endcsname{287}
\expandafter\gdef\csname odunum@ci@axis_breakdown.m2.salmonn2_7b.axis1.audio_visual.1.3\endcsname{[0.040, 0.073]}
\expandafter\gdef\csname odunum@val@axis_breakdown.m2.salmonn2_7b.axis1.audio_visual.1.4\endcsname{0.069}
\expandafter\gdef\csname odunum@n@axis_breakdown.m2.salmonn2_7b.axis1.audio_visual.1.4\endcsname{377}
\expandafter\gdef\csname odunum@ci@axis_breakdown.m2.salmonn2_7b.axis1.audio_visual.1.4\endcsname{[0.052, 0.087]}
\expandafter\gdef\csname odunum@val@axis_breakdown.m2.salmonn2_7b.axis2.2.1\endcsname{0.099}
\expandafter\gdef\csname odunum@n@axis_breakdown.m2.salmonn2_7b.axis2.2.1\endcsname{139}
\expandafter\gdef\csname odunum@ci@axis_breakdown.m2.salmonn2_7b.axis2.2.1\endcsname{[0.069, 0.132]}
\expandafter\gdef\csname odunum@val@axis_breakdown.m2.salmonn2_7b.axis2.2.2\endcsname{0.105}
\expandafter\gdef\csname odunum@n@axis_breakdown.m2.salmonn2_7b.axis2.2.2\endcsname{437}
\expandafter\gdef\csname odunum@ci@axis_breakdown.m2.salmonn2_7b.axis2.2.2\endcsname{[0.085, 0.126]}
\expandafter\gdef\csname odunum@val@axis_breakdown.m2.salmonn2_7b.axis2.2.3\endcsname{0.077}
\expandafter\gdef\csname odunum@n@axis_breakdown.m2.salmonn2_7b.axis2.2.3\endcsname{527}
\expandafter\gdef\csname odunum@ci@axis_breakdown.m2.salmonn2_7b.axis2.2.3\endcsname{[0.061, 0.094]}
\expandafter\gdef\csname odunum@val@axis_breakdown.m2.salmonn2_7b.axis2.2.4\endcsname{0.145}
\expandafter\gdef\csname odunum@n@axis_breakdown.m2.salmonn2_7b.axis2.2.4\endcsname{260}
\expandafter\gdef\csname odunum@ci@axis_breakdown.m2.salmonn2_7b.axis2.2.4\endcsname{[0.116, 0.176]}
\expandafter\gdef\csname odunum@val@axis_breakdown.m2.salmonn2_7b.axis2.2.5\endcsname{0.143}
\expandafter\gdef\csname odunum@n@axis_breakdown.m2.salmonn2_7b.axis2.2.5\endcsname{251}
\expandafter\gdef\csname odunum@ci@axis_breakdown.m2.salmonn2_7b.axis2.2.5\endcsname{[0.116, 0.171]}
\expandafter\gdef\csname odunum@val@axis_breakdown.m2.salmonn2_7b.axis3.3.1\endcsname{0.119}
\expandafter\gdef\csname odunum@n@axis_breakdown.m2.salmonn2_7b.axis3.3.1\endcsname{109}
\expandafter\gdef\csname odunum@ci@axis_breakdown.m2.salmonn2_7b.axis3.3.1\endcsname{[0.076, 0.167]}
\expandafter\gdef\csname odunum@val@axis_breakdown.m2.salmonn2_7b.axis3.3.2\endcsname{0.091}
\expandafter\gdef\csname odunum@n@axis_breakdown.m2.salmonn2_7b.axis3.3.2\endcsname{116}
\expandafter\gdef\csname odunum@ci@axis_breakdown.m2.salmonn2_7b.axis3.3.2\endcsname{[0.060, 0.125]}
\expandafter\gdef\csname odunum@val@axis_breakdown.m2.salmonn2_7b.axis3.3.3\endcsname{0.071}
\expandafter\gdef\csname odunum@n@axis_breakdown.m2.salmonn2_7b.axis3.3.3\endcsname{205}
\expandafter\gdef\csname odunum@ci@axis_breakdown.m2.salmonn2_7b.axis3.3.3\endcsname{[0.046, 0.097]}
\expandafter\gdef\csname odunum@val@axis_breakdown.m2.salmonn2_7b.axis3.3.4\endcsname{0.092}
\expandafter\gdef\csname odunum@n@axis_breakdown.m2.salmonn2_7b.axis3.3.4\endcsname{398}
\expandafter\gdef\csname odunum@ci@axis_breakdown.m2.salmonn2_7b.axis3.3.4\endcsname{[0.072, 0.113]}
\expandafter\gdef\csname odunum@val@axis_breakdown.m2.salmonn2_7b.axis3.3.5\endcsname{0.158}
\expandafter\gdef\csname odunum@n@axis_breakdown.m2.salmonn2_7b.axis3.3.5\endcsname{292}
\expandafter\gdef\csname odunum@ci@axis_breakdown.m2.salmonn2_7b.axis3.3.5\endcsname{[0.131, 0.185]}
\expandafter\gdef\csname odunum@val@axis_breakdown.m2.salmonn2_7b.axis3.3.6\endcsname{0.108}
\expandafter\gdef\csname odunum@n@axis_breakdown.m2.salmonn2_7b.axis3.3.6\endcsname{494}
\expandafter\gdef\csname odunum@ci@axis_breakdown.m2.salmonn2_7b.axis3.3.6\endcsname{[0.090, 0.128]}
\expandafter\gdef\csname odunum@val@axis_breakdown.m2.salmonn2_7b.axis4.4.1\endcsname{0.080}
\expandafter\gdef\csname odunum@n@axis_breakdown.m2.salmonn2_7b.axis4.4.1\endcsname{201}
\expandafter\gdef\csname odunum@ci@axis_breakdown.m2.salmonn2_7b.axis4.4.1\endcsname{[0.055, 0.106]}
\expandafter\gdef\csname odunum@val@axis_breakdown.m2.salmonn2_7b.axis4.4.2\endcsname{0.126}
\expandafter\gdef\csname odunum@n@axis_breakdown.m2.salmonn2_7b.axis4.4.2\endcsname{450}
\expandafter\gdef\csname odunum@ci@axis_breakdown.m2.salmonn2_7b.axis4.4.2\endcsname{[0.106, 0.147]}
\expandafter\gdef\csname odunum@val@axis_breakdown.m2.salmonn2_7b.axis4.4.3\endcsname{0.100}
\expandafter\gdef\csname odunum@n@axis_breakdown.m2.salmonn2_7b.axis4.4.3\endcsname{164}
\expandafter\gdef\csname odunum@ci@axis_breakdown.m2.salmonn2_7b.axis4.4.3\endcsname{[0.070, 0.134]}
\expandafter\gdef\csname odunum@val@axis_breakdown.m2.salmonn2_7b.axis4.4.4\endcsname{0.095}
\expandafter\gdef\csname odunum@n@axis_breakdown.m2.salmonn2_7b.axis4.4.4\endcsname{498}
\expandafter\gdef\csname odunum@ci@axis_breakdown.m2.salmonn2_7b.axis4.4.4\endcsname{[0.078, 0.113]}
\expandafter\gdef\csname odunum@val@axis_breakdown.m2.salmonn2_7b.axis4.4.5\endcsname{0.124}
\expandafter\gdef\csname odunum@n@axis_breakdown.m2.salmonn2_7b.axis4.4.5\endcsname{301}
\expandafter\gdef\csname odunum@ci@axis_breakdown.m2.salmonn2_7b.axis4.4.5\endcsname{[0.097, 0.152]}
\expandafter\gdef\csname odunum@val@axis_breakdown.m2.salmonn2_7b.axis5.audio_only.5.1\endcsname{0.127}
\expandafter\gdef\csname odunum@n@axis_breakdown.m2.salmonn2_7b.axis5.audio_only.5.1\endcsname{45}
\expandafter\gdef\csname odunum@ci@axis_breakdown.m2.salmonn2_7b.axis5.audio_only.5.1\endcsname{[0.063, 0.203]}
\expandafter\gdef\csname odunum@val@axis_breakdown.m2.salmonn2_7b.axis5.audio_only.5.10\endcsname{0.215}
\expandafter\gdef\csname odunum@n@axis_breakdown.m2.salmonn2_7b.axis5.audio_only.5.10\endcsname{72}
\expandafter\gdef\csname odunum@ci@axis_breakdown.m2.salmonn2_7b.axis5.audio_only.5.10\endcsname{[0.149, 0.287]}
\expandafter\gdef\csname odunum@val@axis_breakdown.m2.salmonn2_7b.axis5.audio_only.5.2\endcsname{0.190}
\expandafter\gdef\csname odunum@n@axis_breakdown.m2.salmonn2_7b.axis5.audio_only.5.2\endcsname{75}
\expandafter\gdef\csname odunum@ci@axis_breakdown.m2.salmonn2_7b.axis5.audio_only.5.2\endcsname{[0.130, 0.254]}
\expandafter\gdef\csname odunum@val@axis_breakdown.m2.salmonn2_7b.axis5.audio_only.5.3\endcsname{0.213}
\expandafter\gdef\csname odunum@n@axis_breakdown.m2.salmonn2_7b.axis5.audio_only.5.3\endcsname{50}
\expandafter\gdef\csname odunum@ci@axis_breakdown.m2.salmonn2_7b.axis5.audio_only.5.3\endcsname{[0.127, 0.307]}
\expandafter\gdef\csname odunum@val@axis_breakdown.m2.salmonn2_7b.axis5.audio_only.5.4\endcsname{0.186}
\expandafter\gdef\csname odunum@n@axis_breakdown.m2.salmonn2_7b.axis5.audio_only.5.4\endcsname{67}
\expandafter\gdef\csname odunum@ci@axis_breakdown.m2.salmonn2_7b.axis5.audio_only.5.4\endcsname{[0.126, 0.250]}
\expandafter\gdef\csname odunum@val@axis_breakdown.m2.salmonn2_7b.axis5.audio_only.5.5\endcsname{0.142}
\expandafter\gdef\csname odunum@n@axis_breakdown.m2.salmonn2_7b.axis5.audio_only.5.5\endcsname{52}
\expandafter\gdef\csname odunum@ci@axis_breakdown.m2.salmonn2_7b.axis5.audio_only.5.5\endcsname{[0.085, 0.206]}
\expandafter\gdef\csname odunum@val@axis_breakdown.m2.salmonn2_7b.axis5.audio_only.5.6\endcsname{0.285}
\expandafter\gdef\csname odunum@n@axis_breakdown.m2.salmonn2_7b.axis5.audio_only.5.6\endcsname{52}
\expandafter\gdef\csname odunum@ci@axis_breakdown.m2.salmonn2_7b.axis5.audio_only.5.6\endcsname{[0.198, 0.378]}
\expandafter\gdef\csname odunum@val@axis_breakdown.m2.salmonn2_7b.axis5.audio_only.5.7\endcsname{0.194}
\expandafter\gdef\csname odunum@n@axis_breakdown.m2.salmonn2_7b.axis5.audio_only.5.7\endcsname{46}
\expandafter\gdef\csname odunum@ci@axis_breakdown.m2.salmonn2_7b.axis5.audio_only.5.7\endcsname{[0.127, 0.267]}
\expandafter\gdef\csname odunum@val@axis_breakdown.m2.salmonn2_7b.axis5.audio_only.5.8\endcsname{0.122}
\expandafter\gdef\csname odunum@n@axis_breakdown.m2.salmonn2_7b.axis5.audio_only.5.8\endcsname{57}
\expandafter\gdef\csname odunum@ci@axis_breakdown.m2.salmonn2_7b.axis5.audio_only.5.8\endcsname{[0.069, 0.184]}
\expandafter\gdef\csname odunum@val@axis_breakdown.m2.salmonn2_7b.axis5.audio_only.5.9\endcsname{0.150}
\expandafter\gdef\csname odunum@n@axis_breakdown.m2.salmonn2_7b.axis5.audio_only.5.9\endcsname{52}
\expandafter\gdef\csname odunum@ci@axis_breakdown.m2.salmonn2_7b.axis5.audio_only.5.9\endcsname{[0.082, 0.229]}
\expandafter\gdef\csname odunum@val@axis_breakdown.m2.salmonn2_7b.axis5.audio_visual.5.1\endcsname{0.038}
\expandafter\gdef\csname odunum@n@axis_breakdown.m2.salmonn2_7b.axis5.audio_visual.5.1\endcsname{113}
\expandafter\gdef\csname odunum@ci@axis_breakdown.m2.salmonn2_7b.axis5.audio_visual.5.1\endcsname{[0.018, 0.061]}
\expandafter\gdef\csname odunum@val@axis_breakdown.m2.salmonn2_7b.axis5.audio_visual.5.10\endcsname{0.075}
\expandafter\gdef\csname odunum@n@axis_breakdown.m2.salmonn2_7b.axis5.audio_visual.5.10\endcsname{121}
\expandafter\gdef\csname odunum@ci@axis_breakdown.m2.salmonn2_7b.axis5.audio_visual.5.10\endcsname{[0.046, 0.107]}
\expandafter\gdef\csname odunum@val@axis_breakdown.m2.salmonn2_7b.axis5.audio_visual.5.2\endcsname{0.077}
\expandafter\gdef\csname odunum@n@axis_breakdown.m2.salmonn2_7b.axis5.audio_visual.5.2\endcsname{123}
\expandafter\gdef\csname odunum@ci@axis_breakdown.m2.salmonn2_7b.axis5.audio_visual.5.2\endcsname{[0.046, 0.113]}
\expandafter\gdef\csname odunum@val@axis_breakdown.m2.salmonn2_7b.axis5.audio_visual.5.3\endcsname{0.073}
\expandafter\gdef\csname odunum@n@axis_breakdown.m2.salmonn2_7b.axis5.audio_visual.5.3\endcsname{99}
\expandafter\gdef\csname odunum@ci@axis_breakdown.m2.salmonn2_7b.axis5.audio_visual.5.3\endcsname{[0.043, 0.108]}
\expandafter\gdef\csname odunum@val@axis_breakdown.m2.salmonn2_7b.axis5.audio_visual.5.4\endcsname{0.066}
\expandafter\gdef\csname odunum@n@axis_breakdown.m2.salmonn2_7b.axis5.audio_visual.5.4\endcsname{96}
\expandafter\gdef\csname odunum@ci@axis_breakdown.m2.salmonn2_7b.axis5.audio_visual.5.4\endcsname{[0.037, 0.099]}
\expandafter\gdef\csname odunum@val@axis_breakdown.m2.salmonn2_7b.axis5.audio_visual.5.5\endcsname{0.030}
\expandafter\gdef\csname odunum@n@axis_breakdown.m2.salmonn2_7b.axis5.audio_visual.5.5\endcsname{99}
\expandafter\gdef\csname odunum@ci@axis_breakdown.m2.salmonn2_7b.axis5.audio_visual.5.5\endcsname{[0.009, 0.055]}
\expandafter\gdef\csname odunum@val@axis_breakdown.m2.salmonn2_7b.axis5.audio_visual.5.6\endcsname{0.090}
\expandafter\gdef\csname odunum@n@axis_breakdown.m2.salmonn2_7b.axis5.audio_visual.5.6\endcsname{95}
\expandafter\gdef\csname odunum@ci@axis_breakdown.m2.salmonn2_7b.axis5.audio_visual.5.6\endcsname{[0.053, 0.132]}
\expandafter\gdef\csname odunum@val@axis_breakdown.m2.salmonn2_7b.axis5.audio_visual.5.7\endcsname{0.082}
\expandafter\gdef\csname odunum@n@axis_breakdown.m2.salmonn2_7b.axis5.audio_visual.5.7\endcsname{102}
\expandafter\gdef\csname odunum@ci@axis_breakdown.m2.salmonn2_7b.axis5.audio_visual.5.7\endcsname{[0.050, 0.116]}
\expandafter\gdef\csname odunum@val@axis_breakdown.m2.salmonn2_7b.axis5.audio_visual.5.8\endcsname{0.091}
\expandafter\gdef\csname odunum@n@axis_breakdown.m2.salmonn2_7b.axis5.audio_visual.5.8\endcsname{120}
\expandafter\gdef\csname odunum@ci@axis_breakdown.m2.salmonn2_7b.axis5.audio_visual.5.8\endcsname{[0.064, 0.122]}
\expandafter\gdef\csname odunum@val@axis_breakdown.m2.salmonn2_7b.axis5.audio_visual.5.9\endcsname{0.030}
\expandafter\gdef\csname odunum@n@axis_breakdown.m2.salmonn2_7b.axis5.audio_visual.5.9\endcsname{78}
\expandafter\gdef\csname odunum@ci@axis_breakdown.m2.salmonn2_7b.axis5.audio_visual.5.9\endcsname{[0.010, 0.054]}
\expandafter\gdef\csname odunum@val@axis_breakdown.m2.salmonn2_7b.axis6.6.1\endcsname{0.087}
\expandafter\gdef\csname odunum@n@axis_breakdown.m2.salmonn2_7b.axis6.6.1\endcsname{228}
\expandafter\gdef\csname odunum@ci@axis_breakdown.m2.salmonn2_7b.axis6.6.1\endcsname{[0.063, 0.112]}
\expandafter\gdef\csname odunum@val@axis_breakdown.m2.salmonn2_7b.axis6.6.10\endcsname{0.098}
\expandafter\gdef\csname odunum@n@axis_breakdown.m2.salmonn2_7b.axis6.6.10\endcsname{101}
\expandafter\gdef\csname odunum@ci@axis_breakdown.m2.salmonn2_7b.axis6.6.10\endcsname{[0.062, 0.138]}
\expandafter\gdef\csname odunum@val@axis_breakdown.m2.salmonn2_7b.axis6.6.12\endcsname{0.098}
\expandafter\gdef\csname odunum@n@axis_breakdown.m2.salmonn2_7b.axis6.6.12\endcsname{128}
\expandafter\gdef\csname odunum@ci@axis_breakdown.m2.salmonn2_7b.axis6.6.12\endcsname{[0.063, 0.137]}
\expandafter\gdef\csname odunum@val@axis_breakdown.m2.salmonn2_7b.axis6.6.13\endcsname{0.099}
\expandafter\gdef\csname odunum@n@axis_breakdown.m2.salmonn2_7b.axis6.6.13\endcsname{57}
\expandafter\gdef\csname odunum@ci@axis_breakdown.m2.salmonn2_7b.axis6.6.13\endcsname{[0.053, 0.153]}
\expandafter\gdef\csname odunum@val@axis_breakdown.m2.salmonn2_7b.axis6.6.14\endcsname{0.177}
\expandafter\gdef\csname odunum@n@axis_breakdown.m2.salmonn2_7b.axis6.6.14\endcsname{82}
\expandafter\gdef\csname odunum@ci@axis_breakdown.m2.salmonn2_7b.axis6.6.14\endcsname{[0.123, 0.235]}
\expandafter\gdef\csname odunum@val@axis_breakdown.m2.salmonn2_7b.axis6.6.18\endcsname{0.084}
\expandafter\gdef\csname odunum@n@axis_breakdown.m2.salmonn2_7b.axis6.6.18\endcsname{72}
\expandafter\gdef\csname odunum@ci@axis_breakdown.m2.salmonn2_7b.axis6.6.18\endcsname{[0.045, 0.128]}
\expandafter\gdef\csname odunum@val@axis_breakdown.m2.salmonn2_7b.axis6.6.19\endcsname{0.154}
\expandafter\gdef\csname odunum@n@axis_breakdown.m2.salmonn2_7b.axis6.6.19\endcsname{44}
\expandafter\gdef\csname odunum@ci@axis_breakdown.m2.salmonn2_7b.axis6.6.19\endcsname{[0.085, 0.233]}
\expandafter\gdef\csname odunum@val@axis_breakdown.m2.salmonn2_7b.axis6.6.2\endcsname{0.084}
\expandafter\gdef\csname odunum@n@axis_breakdown.m2.salmonn2_7b.axis6.6.2\endcsname{41}
\expandafter\gdef\csname odunum@ci@axis_breakdown.m2.salmonn2_7b.axis6.6.2\endcsname{[0.029, 0.153]}
\expandafter\gdef\csname odunum@val@axis_breakdown.m2.salmonn2_7b.axis6.6.3\endcsname{0.093}
\expandafter\gdef\csname odunum@n@axis_breakdown.m2.salmonn2_7b.axis6.6.3\endcsname{43}
\expandafter\gdef\csname odunum@ci@axis_breakdown.m2.salmonn2_7b.axis6.6.3\endcsname{[0.041, 0.157]}
\expandafter\gdef\csname odunum@val@axis_breakdown.m2.salmonn2_7b.axis6.6.5\endcsname{0.053}
\expandafter\gdef\csname odunum@n@axis_breakdown.m2.salmonn2_7b.axis6.6.5\endcsname{48}
\expandafter\gdef\csname odunum@ci@axis_breakdown.m2.salmonn2_7b.axis6.6.5\endcsname{[0.022, 0.092]}
\expandafter\gdef\csname odunum@val@axis_breakdown.m2.salmonn2_7b.axis6.6.6\endcsname{0.095}
\expandafter\gdef\csname odunum@n@axis_breakdown.m2.salmonn2_7b.axis6.6.6\endcsname{133}
\expandafter\gdef\csname odunum@ci@axis_breakdown.m2.salmonn2_7b.axis6.6.6\endcsname{[0.059, 0.133]}
\expandafter\gdef\csname odunum@val@axis_breakdown.m2.salmonn2_7b.axis6.6.7\endcsname{0.132}
\expandafter\gdef\csname odunum@n@axis_breakdown.m2.salmonn2_7b.axis6.6.7\endcsname{390}
\expandafter\gdef\csname odunum@ci@axis_breakdown.m2.salmonn2_7b.axis6.6.7\endcsname{[0.109, 0.157]}
\expandafter\gdef\csname odunum@val@axis_breakdown.m2.salmonn2_7b.axis6.6.8\endcsname{0.092}
\expandafter\gdef\csname odunum@n@axis_breakdown.m2.salmonn2_7b.axis6.6.8\endcsname{47}
\expandafter\gdef\csname odunum@ci@axis_breakdown.m2.salmonn2_7b.axis6.6.8\endcsname{[0.041, 0.152]}
\expandafter\gdef\csname odunum@val@axis_breakdown.m2.salmonn2_7b.axis6.6.9\endcsname{0.107}
\expandafter\gdef\csname odunum@n@axis_breakdown.m2.salmonn2_7b.axis6.6.9\endcsname{116}
\expandafter\gdef\csname odunum@ci@axis_breakdown.m2.salmonn2_7b.axis6.6.9\endcsname{[0.072, 0.146]}
\expandafter\gdef\csname odunum@val@axis_breakdown.m2.salmonn2_7b.difficulty.q1\endcsname{0.225}
\expandafter\gdef\csname odunum@n@axis_breakdown.m2.salmonn2_7b.difficulty.q1\endcsname{353}
\expandafter\gdef\csname odunum@ci@axis_breakdown.m2.salmonn2_7b.difficulty.q1\endcsname{[0.196, 0.255]}
\expandafter\gdef\csname odunum@val@axis_breakdown.m2.salmonn2_7b.difficulty.q2\endcsname{0.096}
\expandafter\gdef\csname odunum@n@axis_breakdown.m2.salmonn2_7b.difficulty.q2\endcsname{353}
\expandafter\gdef\csname odunum@ci@axis_breakdown.m2.salmonn2_7b.difficulty.q2\endcsname{[0.076, 0.117]}
\expandafter\gdef\csname odunum@val@axis_breakdown.m2.salmonn2_7b.difficulty.q3\endcsname{0.100}
\expandafter\gdef\csname odunum@n@axis_breakdown.m2.salmonn2_7b.difficulty.q3\endcsname{354}
\expandafter\gdef\csname odunum@ci@axis_breakdown.m2.salmonn2_7b.difficulty.q3\endcsname{[0.079, 0.123]}
\expandafter\gdef\csname odunum@val@axis_breakdown.m2.salmonn2_7b.difficulty.q4\endcsname{0.066}
\expandafter\gdef\csname odunum@n@axis_breakdown.m2.salmonn2_7b.difficulty.q4\endcsname{355}
\expandafter\gdef\csname odunum@ci@axis_breakdown.m2.salmonn2_7b.difficulty.q4\endcsname{[0.050, 0.083]}
\expandafter\gdef\csname odunum@val@axis_breakdown.m2.seed.axis1.audio_only.1.1\endcsname{0.800}
\expandafter\gdef\csname odunum@n@axis_breakdown.m2.seed.axis1.audio_only.1.1\endcsname{158}
\expandafter\gdef\csname odunum@ci@axis_breakdown.m2.seed.axis1.audio_only.1.1\endcsname{[0.759, 0.839]}
\expandafter\gdef\csname odunum@val@axis_breakdown.m2.seed.axis1.audio_only.1.2\endcsname{0.769}
\expandafter\gdef\csname odunum@n@axis_breakdown.m2.seed.axis1.audio_only.1.2\endcsname{379}
\expandafter\gdef\csname odunum@ci@axis_breakdown.m2.seed.axis1.audio_only.1.2\endcsname{[0.746, 0.792]}
\expandafter\gdef\csname odunum@val@axis_breakdown.m2.seed.axis1.audio_visual.1.1\endcsname{0.637}
\expandafter\gdef\csname odunum@n@axis_breakdown.m2.seed.axis1.audio_visual.1.1\endcsname{94}
\expandafter\gdef\csname odunum@ci@axis_breakdown.m2.seed.axis1.audio_visual.1.1\endcsname{[0.570, 0.703]}
\expandafter\gdef\csname odunum@val@axis_breakdown.m2.seed.axis1.audio_visual.1.2\endcsname{0.638}
\expandafter\gdef\csname odunum@n@axis_breakdown.m2.seed.axis1.audio_visual.1.2\endcsname{294}
\expandafter\gdef\csname odunum@ci@axis_breakdown.m2.seed.axis1.audio_visual.1.2\endcsname{[0.605, 0.672]}
\expandafter\gdef\csname odunum@val@axis_breakdown.m2.seed.axis1.audio_visual.1.3\endcsname{0.650}
\expandafter\gdef\csname odunum@n@axis_breakdown.m2.seed.axis1.audio_visual.1.3\endcsname{291}
\expandafter\gdef\csname odunum@ci@axis_breakdown.m2.seed.axis1.audio_visual.1.3\endcsname{[0.611, 0.687]}
\expandafter\gdef\csname odunum@val@axis_breakdown.m2.seed.axis1.audio_visual.1.4\endcsname{0.638}
\expandafter\gdef\csname odunum@n@axis_breakdown.m2.seed.axis1.audio_visual.1.4\endcsname{382}
\expandafter\gdef\csname odunum@ci@axis_breakdown.m2.seed.axis1.audio_visual.1.4\endcsname{[0.608, 0.668]}
\expandafter\gdef\csname odunum@val@axis_breakdown.m2.seed.axis2.2.1\endcsname{0.665}
\expandafter\gdef\csname odunum@n@axis_breakdown.m2.seed.axis2.2.1\endcsname{143}
\expandafter\gdef\csname odunum@ci@axis_breakdown.m2.seed.axis2.2.1\endcsname{[0.612, 0.718]}
\expandafter\gdef\csname odunum@val@axis_breakdown.m2.seed.axis2.2.2\endcsname{0.690}
\expandafter\gdef\csname odunum@n@axis_breakdown.m2.seed.axis2.2.2\endcsname{440}
\expandafter\gdef\csname odunum@ci@axis_breakdown.m2.seed.axis2.2.2\endcsname{[0.663, 0.716]}
\expandafter\gdef\csname odunum@val@axis_breakdown.m2.seed.axis2.2.3\endcsname{0.726}
\expandafter\gdef\csname odunum@n@axis_breakdown.m2.seed.axis2.2.3\endcsname{531}
\expandafter\gdef\csname odunum@ci@axis_breakdown.m2.seed.axis2.2.3\endcsname{[0.703, 0.749]}
\expandafter\gdef\csname odunum@val@axis_breakdown.m2.seed.axis2.2.4\endcsname{0.625}
\expandafter\gdef\csname odunum@n@axis_breakdown.m2.seed.axis2.2.4\endcsname{261}
\expandafter\gdef\csname odunum@ci@axis_breakdown.m2.seed.axis2.2.4\endcsname{[0.584, 0.667]}
\expandafter\gdef\csname odunum@val@axis_breakdown.m2.seed.axis2.2.5\endcsname{0.695}
\expandafter\gdef\csname odunum@n@axis_breakdown.m2.seed.axis2.2.5\endcsname{256}
\expandafter\gdef\csname odunum@ci@axis_breakdown.m2.seed.axis2.2.5\endcsname{[0.660, 0.729]}
\expandafter\gdef\csname odunum@val@axis_breakdown.m2.seed.axis3.3.1\endcsname{0.672}
\expandafter\gdef\csname odunum@n@axis_breakdown.m2.seed.axis3.3.1\endcsname{109}
\expandafter\gdef\csname odunum@ci@axis_breakdown.m2.seed.axis3.3.1\endcsname{[0.611, 0.730]}
\expandafter\gdef\csname odunum@val@axis_breakdown.m2.seed.axis3.3.2\endcsname{0.661}
\expandafter\gdef\csname odunum@n@axis_breakdown.m2.seed.axis3.3.2\endcsname{116}
\expandafter\gdef\csname odunum@ci@axis_breakdown.m2.seed.axis3.3.2\endcsname{[0.608, 0.713]}
\expandafter\gdef\csname odunum@val@axis_breakdown.m2.seed.axis3.3.3\endcsname{0.669}
\expandafter\gdef\csname odunum@n@axis_breakdown.m2.seed.axis3.3.3\endcsname{207}
\expandafter\gdef\csname odunum@ci@axis_breakdown.m2.seed.axis3.3.3\endcsname{[0.630, 0.709]}
\expandafter\gdef\csname odunum@val@axis_breakdown.m2.seed.axis3.3.4\endcsname{0.683}
\expandafter\gdef\csname odunum@n@axis_breakdown.m2.seed.axis3.3.4\endcsname{404}
\expandafter\gdef\csname odunum@ci@axis_breakdown.m2.seed.axis3.3.4\endcsname{[0.652, 0.713]}
\expandafter\gdef\csname odunum@val@axis_breakdown.m2.seed.axis3.3.5\endcsname{0.634}
\expandafter\gdef\csname odunum@n@axis_breakdown.m2.seed.axis3.3.5\endcsname{294}
\expandafter\gdef\csname odunum@ci@axis_breakdown.m2.seed.axis3.3.5\endcsname{[0.600, 0.668]}
\expandafter\gdef\csname odunum@val@axis_breakdown.m2.seed.axis3.3.6\endcsname{0.747}
\expandafter\gdef\csname odunum@n@axis_breakdown.m2.seed.axis3.3.6\endcsname{501}
\expandafter\gdef\csname odunum@ci@axis_breakdown.m2.seed.axis3.3.6\endcsname{[0.722, 0.771]}
\expandafter\gdef\csname odunum@val@axis_breakdown.m2.seed.axis4.4.1\endcsname{0.653}
\expandafter\gdef\csname odunum@n@axis_breakdown.m2.seed.axis4.4.1\endcsname{202}
\expandafter\gdef\csname odunum@ci@axis_breakdown.m2.seed.axis4.4.1\endcsname{[0.611, 0.695]}
\expandafter\gdef\csname odunum@val@axis_breakdown.m2.seed.axis4.4.2\endcsname{0.698}
\expandafter\gdef\csname odunum@n@axis_breakdown.m2.seed.axis4.4.2\endcsname{457}
\expandafter\gdef\csname odunum@ci@axis_breakdown.m2.seed.axis4.4.2\endcsname{[0.673, 0.723]}
\expandafter\gdef\csname odunum@val@axis_breakdown.m2.seed.axis4.4.3\endcsname{0.720}
\expandafter\gdef\csname odunum@n@axis_breakdown.m2.seed.axis4.4.3\endcsname{164}
\expandafter\gdef\csname odunum@ci@axis_breakdown.m2.seed.axis4.4.3\endcsname{[0.672, 0.766]}
\expandafter\gdef\csname odunum@val@axis_breakdown.m2.seed.axis4.4.4\endcsname{0.673}
\expandafter\gdef\csname odunum@n@axis_breakdown.m2.seed.axis4.4.4\endcsname{505}
\expandafter\gdef\csname odunum@ci@axis_breakdown.m2.seed.axis4.4.4\endcsname{[0.646, 0.700]}
\expandafter\gdef\csname odunum@val@axis_breakdown.m2.seed.axis4.4.5\endcsname{0.715}
\expandafter\gdef\csname odunum@n@axis_breakdown.m2.seed.axis4.4.5\endcsname{303}
\expandafter\gdef\csname odunum@ci@axis_breakdown.m2.seed.axis4.4.5\endcsname{[0.681, 0.748]}
\expandafter\gdef\csname odunum@val@axis_breakdown.m2.seed.axis5.audio_only.5.1\endcsname{0.854}
\expandafter\gdef\csname odunum@n@axis_breakdown.m2.seed.axis5.audio_only.5.1\endcsname{45}
\expandafter\gdef\csname odunum@ci@axis_breakdown.m2.seed.axis5.audio_only.5.1\endcsname{[0.787, 0.913]}
\expandafter\gdef\csname odunum@val@axis_breakdown.m2.seed.axis5.audio_only.5.10\endcsname{0.793}
\expandafter\gdef\csname odunum@n@axis_breakdown.m2.seed.axis5.audio_only.5.10\endcsname{72}
\expandafter\gdef\csname odunum@ci@axis_breakdown.m2.seed.axis5.audio_only.5.10\endcsname{[0.745, 0.839]}
\expandafter\gdef\csname odunum@val@axis_breakdown.m2.seed.axis5.audio_only.5.2\endcsname{0.769}
\expandafter\gdef\csname odunum@n@axis_breakdown.m2.seed.axis5.audio_only.5.2\endcsname{75}
\expandafter\gdef\csname odunum@ci@axis_breakdown.m2.seed.axis5.audio_only.5.2\endcsname{[0.714, 0.823]}
\expandafter\gdef\csname odunum@val@axis_breakdown.m2.seed.axis5.audio_only.5.3\endcsname{0.793}
\expandafter\gdef\csname odunum@n@axis_breakdown.m2.seed.axis5.audio_only.5.3\endcsname{50}
\expandafter\gdef\csname odunum@ci@axis_breakdown.m2.seed.axis5.audio_only.5.3\endcsname{[0.717, 0.860]}
\expandafter\gdef\csname odunum@val@axis_breakdown.m2.seed.axis5.audio_only.5.4\endcsname{0.782}
\expandafter\gdef\csname odunum@n@axis_breakdown.m2.seed.axis5.audio_only.5.4\endcsname{68}
\expandafter\gdef\csname odunum@ci@axis_breakdown.m2.seed.axis5.audio_only.5.4\endcsname{[0.721, 0.840]}
\expandafter\gdef\csname odunum@val@axis_breakdown.m2.seed.axis5.audio_only.5.5\endcsname{0.770}
\expandafter\gdef\csname odunum@n@axis_breakdown.m2.seed.axis5.audio_only.5.5\endcsname{52}
\expandafter\gdef\csname odunum@ci@axis_breakdown.m2.seed.axis5.audio_only.5.5\endcsname{[0.708, 0.830]}
\expandafter\gdef\csname odunum@val@axis_breakdown.m2.seed.axis5.audio_only.5.6\endcsname{0.749}
\expandafter\gdef\csname odunum@n@axis_breakdown.m2.seed.axis5.audio_only.5.6\endcsname{52}
\expandafter\gdef\csname odunum@ci@axis_breakdown.m2.seed.axis5.audio_only.5.6\endcsname{[0.675, 0.819]}
\expandafter\gdef\csname odunum@val@axis_breakdown.m2.seed.axis5.audio_only.5.7\endcsname{0.764}
\expandafter\gdef\csname odunum@n@axis_breakdown.m2.seed.axis5.audio_only.5.7\endcsname{46}
\expandafter\gdef\csname odunum@ci@axis_breakdown.m2.seed.axis5.audio_only.5.7\endcsname{[0.698, 0.829]}
\expandafter\gdef\csname odunum@val@axis_breakdown.m2.seed.axis5.audio_only.5.8\endcsname{0.749}
\expandafter\gdef\csname odunum@n@axis_breakdown.m2.seed.axis5.audio_only.5.8\endcsname{57}
\expandafter\gdef\csname odunum@ci@axis_breakdown.m2.seed.axis5.audio_only.5.8\endcsname{[0.682, 0.813]}
\expandafter\gdef\csname odunum@val@axis_breakdown.m2.seed.axis5.audio_only.5.9\endcsname{0.793}
\expandafter\gdef\csname odunum@n@axis_breakdown.m2.seed.axis5.audio_only.5.9\endcsname{53}
\expandafter\gdef\csname odunum@ci@axis_breakdown.m2.seed.axis5.audio_only.5.9\endcsname{[0.728, 0.854]}
\expandafter\gdef\csname odunum@val@axis_breakdown.m2.seed.axis5.audio_visual.5.1\endcsname{0.660}
\expandafter\gdef\csname odunum@n@axis_breakdown.m2.seed.axis5.audio_visual.5.1\endcsname{114}
\expandafter\gdef\csname odunum@ci@axis_breakdown.m2.seed.axis5.audio_visual.5.1\endcsname{[0.603, 0.715]}
\expandafter\gdef\csname odunum@val@axis_breakdown.m2.seed.axis5.audio_visual.5.10\endcsname{0.640}
\expandafter\gdef\csname odunum@n@axis_breakdown.m2.seed.axis5.audio_visual.5.10\endcsname{123}
\expandafter\gdef\csname odunum@ci@axis_breakdown.m2.seed.axis5.audio_visual.5.10\endcsname{[0.580, 0.699]}
\expandafter\gdef\csname odunum@val@axis_breakdown.m2.seed.axis5.audio_visual.5.2\endcsname{0.626}
\expandafter\gdef\csname odunum@n@axis_breakdown.m2.seed.axis5.audio_visual.5.2\endcsname{126}
\expandafter\gdef\csname odunum@ci@axis_breakdown.m2.seed.axis5.audio_visual.5.2\endcsname{[0.579, 0.671]}
\expandafter\gdef\csname odunum@val@axis_breakdown.m2.seed.axis5.audio_visual.5.3\endcsname{0.647}
\expandafter\gdef\csname odunum@n@axis_breakdown.m2.seed.axis5.audio_visual.5.3\endcsname{100}
\expandafter\gdef\csname odunum@ci@axis_breakdown.m2.seed.axis5.audio_visual.5.3\endcsname{[0.582, 0.711]}
\expandafter\gdef\csname odunum@val@axis_breakdown.m2.seed.axis5.audio_visual.5.4\endcsname{0.643}
\expandafter\gdef\csname odunum@n@axis_breakdown.m2.seed.axis5.audio_visual.5.4\endcsname{97}
\expandafter\gdef\csname odunum@ci@axis_breakdown.m2.seed.axis5.audio_visual.5.4\endcsname{[0.586, 0.698]}
\expandafter\gdef\csname odunum@val@axis_breakdown.m2.seed.axis5.audio_visual.5.5\endcsname{0.683}
\expandafter\gdef\csname odunum@n@axis_breakdown.m2.seed.axis5.audio_visual.5.5\endcsname{99}
\expandafter\gdef\csname odunum@ci@axis_breakdown.m2.seed.axis5.audio_visual.5.5\endcsname{[0.622, 0.744]}
\expandafter\gdef\csname odunum@val@axis_breakdown.m2.seed.axis5.audio_visual.5.6\endcsname{0.636}
\expandafter\gdef\csname odunum@n@axis_breakdown.m2.seed.axis5.audio_visual.5.6\endcsname{99}
\expandafter\gdef\csname odunum@ci@axis_breakdown.m2.seed.axis5.audio_visual.5.6\endcsname{[0.579, 0.691]}
\expandafter\gdef\csname odunum@val@axis_breakdown.m2.seed.axis5.audio_visual.5.7\endcsname{0.623}
\expandafter\gdef\csname odunum@n@axis_breakdown.m2.seed.axis5.audio_visual.5.7\endcsname{104}
\expandafter\gdef\csname odunum@ci@axis_breakdown.m2.seed.axis5.audio_visual.5.7\endcsname{[0.558, 0.686]}
\expandafter\gdef\csname odunum@val@axis_breakdown.m2.seed.axis5.audio_visual.5.8\endcsname{0.619}
\expandafter\gdef\csname odunum@n@axis_breakdown.m2.seed.axis5.audio_visual.5.8\endcsname{121}
\expandafter\gdef\csname odunum@ci@axis_breakdown.m2.seed.axis5.audio_visual.5.8\endcsname{[0.562, 0.674]}
\expandafter\gdef\csname odunum@val@axis_breakdown.m2.seed.axis5.audio_visual.5.9\endcsname{0.646}
\expandafter\gdef\csname odunum@n@axis_breakdown.m2.seed.axis5.audio_visual.5.9\endcsname{78}
\expandafter\gdef\csname odunum@ci@axis_breakdown.m2.seed.axis5.audio_visual.5.9\endcsname{[0.572, 0.719]}
\expandafter\gdef\csname odunum@val@axis_breakdown.m2.seed.axis6.6.1\endcsname{0.650}
\expandafter\gdef\csname odunum@n@axis_breakdown.m2.seed.axis6.6.1\endcsname{230}
\expandafter\gdef\csname odunum@ci@axis_breakdown.m2.seed.axis6.6.1\endcsname{[0.605, 0.692]}
\expandafter\gdef\csname odunum@val@axis_breakdown.m2.seed.axis6.6.10\endcsname{0.681}
\expandafter\gdef\csname odunum@n@axis_breakdown.m2.seed.axis6.6.10\endcsname{102}
\expandafter\gdef\csname odunum@ci@axis_breakdown.m2.seed.axis6.6.10\endcsname{[0.627, 0.734]}
\expandafter\gdef\csname odunum@val@axis_breakdown.m2.seed.axis6.6.12\endcsname{0.688}
\expandafter\gdef\csname odunum@n@axis_breakdown.m2.seed.axis6.6.12\endcsname{129}
\expandafter\gdef\csname odunum@ci@axis_breakdown.m2.seed.axis6.6.12\endcsname{[0.633, 0.740]}
\expandafter\gdef\csname odunum@val@axis_breakdown.m2.seed.axis6.6.13\endcsname{0.740}
\expandafter\gdef\csname odunum@n@axis_breakdown.m2.seed.axis6.6.13\endcsname{58}
\expandafter\gdef\csname odunum@ci@axis_breakdown.m2.seed.axis6.6.13\endcsname{[0.677, 0.800]}
\expandafter\gdef\csname odunum@val@axis_breakdown.m2.seed.axis6.6.14\endcsname{0.637}
\expandafter\gdef\csname odunum@n@axis_breakdown.m2.seed.axis6.6.14\endcsname{82}
\expandafter\gdef\csname odunum@ci@axis_breakdown.m2.seed.axis6.6.14\endcsname{[0.580, 0.691]}
\expandafter\gdef\csname odunum@val@axis_breakdown.m2.seed.axis6.6.18\endcsname{0.649}
\expandafter\gdef\csname odunum@n@axis_breakdown.m2.seed.axis6.6.18\endcsname{72}
\expandafter\gdef\csname odunum@ci@axis_breakdown.m2.seed.axis6.6.18\endcsname{[0.577, 0.717]}
\expandafter\gdef\csname odunum@val@axis_breakdown.m2.seed.axis6.6.19\endcsname{0.661}
\expandafter\gdef\csname odunum@n@axis_breakdown.m2.seed.axis6.6.19\endcsname{44}
\expandafter\gdef\csname odunum@ci@axis_breakdown.m2.seed.axis6.6.19\endcsname{[0.577, 0.743]}
\expandafter\gdef\csname odunum@val@axis_breakdown.m2.seed.axis6.6.2\endcsname{0.785}
\expandafter\gdef\csname odunum@n@axis_breakdown.m2.seed.axis6.6.2\endcsname{41}
\expandafter\gdef\csname odunum@ci@axis_breakdown.m2.seed.axis6.6.2\endcsname{[0.720, 0.848]}
\expandafter\gdef\csname odunum@val@axis_breakdown.m2.seed.axis6.6.3\endcsname{0.726}
\expandafter\gdef\csname odunum@n@axis_breakdown.m2.seed.axis6.6.3\endcsname{44}
\expandafter\gdef\csname odunum@ci@axis_breakdown.m2.seed.axis6.6.3\endcsname{[0.636, 0.810]}
\expandafter\gdef\csname odunum@val@axis_breakdown.m2.seed.axis6.6.5\endcsname{0.751}
\expandafter\gdef\csname odunum@n@axis_breakdown.m2.seed.axis6.6.5\endcsname{48}
\expandafter\gdef\csname odunum@ci@axis_breakdown.m2.seed.axis6.6.5\endcsname{[0.676, 0.819]}
\expandafter\gdef\csname odunum@val@axis_breakdown.m2.seed.axis6.6.6\endcsname{0.685}
\expandafter\gdef\csname odunum@n@axis_breakdown.m2.seed.axis6.6.6\endcsname{134}
\expandafter\gdef\csname odunum@ci@axis_breakdown.m2.seed.axis6.6.6\endcsname{[0.632, 0.737]}
\expandafter\gdef\csname odunum@val@axis_breakdown.m2.seed.axis6.6.7\endcsname{0.714}
\expandafter\gdef\csname odunum@n@axis_breakdown.m2.seed.axis6.6.7\endcsname{393}
\expandafter\gdef\csname odunum@ci@axis_breakdown.m2.seed.axis6.6.7\endcsname{[0.684, 0.744]}
\expandafter\gdef\csname odunum@val@axis_breakdown.m2.seed.axis6.6.8\endcsname{0.696}
\expandafter\gdef\csname odunum@n@axis_breakdown.m2.seed.axis6.6.8\endcsname{49}
\expandafter\gdef\csname odunum@ci@axis_breakdown.m2.seed.axis6.6.8\endcsname{[0.617, 0.772]}
\expandafter\gdef\csname odunum@val@axis_breakdown.m2.seed.axis6.6.9\endcsname{0.698}
\expandafter\gdef\csname odunum@n@axis_breakdown.m2.seed.axis6.6.9\endcsname{119}
\expandafter\gdef\csname odunum@ci@axis_breakdown.m2.seed.axis6.6.9\endcsname{[0.647, 0.746]}
\expandafter\gdef\csname odunum@val@axis_breakdown.m2.seed.difficulty.q1\endcsname{0.760}
\expandafter\gdef\csname odunum@n@axis_breakdown.m2.seed.difficulty.q1\endcsname{358}
\expandafter\gdef\csname odunum@ci@axis_breakdown.m2.seed.difficulty.q1\endcsname{[0.732, 0.789]}
\expandafter\gdef\csname odunum@val@axis_breakdown.m2.seed.difficulty.q2\endcsname{0.710}
\expandafter\gdef\csname odunum@n@axis_breakdown.m2.seed.difficulty.q2\endcsname{357}
\expandafter\gdef\csname odunum@ci@axis_breakdown.m2.seed.difficulty.q2\endcsname{[0.682, 0.738]}
\expandafter\gdef\csname odunum@val@axis_breakdown.m2.seed.difficulty.q3\endcsname{0.654}
\expandafter\gdef\csname odunum@n@axis_breakdown.m2.seed.difficulty.q3\endcsname{358}
\expandafter\gdef\csname odunum@ci@axis_breakdown.m2.seed.difficulty.q3\endcsname{[0.622, 0.687]}
\expandafter\gdef\csname odunum@val@axis_breakdown.m2.seed.difficulty.q4\endcsname{0.665}
\expandafter\gdef\csname odunum@n@axis_breakdown.m2.seed.difficulty.q4\endcsname{356}
\expandafter\gdef\csname odunum@ci@axis_breakdown.m2.seed.difficulty.q4\endcsname{[0.635, 0.694]}
\expandafter\gdef\csname odunum@val@axis_breakdown.panel_mean_m2.axis1.audio_only.1.1\endcsname{0.598}
\expandafter\gdef\csname odunum@n@axis_breakdown.panel_mean_m2.axis1.audio_only.1.1\endcsname{13}
\expandafter\gdef\csname odunum@ci@axis_breakdown.panel_mean_m2.axis1.audio_only.1.1\endcsname{[0.503, 0.683]}
\expandafter\gdef\csname odunum@val@axis_breakdown.panel_mean_m2.axis1.audio_only.1.2\endcsname{0.547}
\expandafter\gdef\csname odunum@n@axis_breakdown.panel_mean_m2.axis1.audio_only.1.2\endcsname{13}
\expandafter\gdef\csname odunum@ci@axis_breakdown.panel_mean_m2.axis1.audio_only.1.2\endcsname{[0.458, 0.630]}
\expandafter\gdef\csname odunum@val@axis_breakdown.panel_mean_m2.axis1.audio_visual.1.1\endcsname{0.500}
\expandafter\gdef\csname odunum@n@axis_breakdown.panel_mean_m2.axis1.audio_visual.1.1\endcsname{13}
\expandafter\gdef\csname odunum@ci@axis_breakdown.panel_mean_m2.axis1.audio_visual.1.1\endcsname{[0.390, 0.595]}
\expandafter\gdef\csname odunum@val@axis_breakdown.panel_mean_m2.axis1.audio_visual.1.2\endcsname{0.483}
\expandafter\gdef\csname odunum@n@axis_breakdown.panel_mean_m2.axis1.audio_visual.1.2\endcsname{13}
\expandafter\gdef\csname odunum@ci@axis_breakdown.panel_mean_m2.axis1.audio_visual.1.2\endcsname{[0.393, 0.560]}
\expandafter\gdef\csname odunum@val@axis_breakdown.panel_mean_m2.axis1.audio_visual.1.3\endcsname{0.485}
\expandafter\gdef\csname odunum@n@axis_breakdown.panel_mean_m2.axis1.audio_visual.1.3\endcsname{13}
\expandafter\gdef\csname odunum@ci@axis_breakdown.panel_mean_m2.axis1.audio_visual.1.3\endcsname{[0.383, 0.573]}
\expandafter\gdef\csname odunum@val@axis_breakdown.panel_mean_m2.axis1.audio_visual.1.4\endcsname{0.477}
\expandafter\gdef\csname odunum@n@axis_breakdown.panel_mean_m2.axis1.audio_visual.1.4\endcsname{13}
\expandafter\gdef\csname odunum@ci@axis_breakdown.panel_mean_m2.axis1.audio_visual.1.4\endcsname{[0.385, 0.558]}
\expandafter\gdef\csname odunum@val@axis_breakdown.panel_mean_m2.axis2.2.1\endcsname{0.469}
\expandafter\gdef\csname odunum@n@axis_breakdown.panel_mean_m2.axis2.2.1\endcsname{13}
\expandafter\gdef\csname odunum@ci@axis_breakdown.panel_mean_m2.axis2.2.1\endcsname{[0.381, 0.550]}
\expandafter\gdef\csname odunum@val@axis_breakdown.panel_mean_m2.axis2.2.2\endcsname{0.528}
\expandafter\gdef\csname odunum@n@axis_breakdown.panel_mean_m2.axis2.2.2\endcsname{13}
\expandafter\gdef\csname odunum@ci@axis_breakdown.panel_mean_m2.axis2.2.2\endcsname{[0.431, 0.610]}
\expandafter\gdef\csname odunum@val@axis_breakdown.panel_mean_m2.axis2.2.3\endcsname{0.541}
\expandafter\gdef\csname odunum@n@axis_breakdown.panel_mean_m2.axis2.2.3\endcsname{13}
\expandafter\gdef\csname odunum@ci@axis_breakdown.panel_mean_m2.axis2.2.3\endcsname{[0.434, 0.632]}
\expandafter\gdef\csname odunum@val@axis_breakdown.panel_mean_m2.axis2.2.4\endcsname{0.469}
\expandafter\gdef\csname odunum@n@axis_breakdown.panel_mean_m2.axis2.2.4\endcsname{13}
\expandafter\gdef\csname odunum@ci@axis_breakdown.panel_mean_m2.axis2.2.4\endcsname{[0.389, 0.541]}
\expandafter\gdef\csname odunum@val@axis_breakdown.panel_mean_m2.axis2.2.5\endcsname{0.485}
\expandafter\gdef\csname odunum@n@axis_breakdown.panel_mean_m2.axis2.2.5\endcsname{13}
\expandafter\gdef\csname odunum@ci@axis_breakdown.panel_mean_m2.axis2.2.5\endcsname{[0.397, 0.568]}
\expandafter\gdef\csname odunum@val@axis_breakdown.panel_mean_m2.axis3.3.1\endcsname{0.520}
\expandafter\gdef\csname odunum@n@axis_breakdown.panel_mean_m2.axis3.3.1\endcsname{13}
\expandafter\gdef\csname odunum@ci@axis_breakdown.panel_mean_m2.axis3.3.1\endcsname{[0.428, 0.598]}
\expandafter\gdef\csname odunum@val@axis_breakdown.panel_mean_m2.axis3.3.2\endcsname{0.505}
\expandafter\gdef\csname odunum@n@axis_breakdown.panel_mean_m2.axis3.3.2\endcsname{13}
\expandafter\gdef\csname odunum@ci@axis_breakdown.panel_mean_m2.axis3.3.2\endcsname{[0.411, 0.586]}
\expandafter\gdef\csname odunum@val@axis_breakdown.panel_mean_m2.axis3.3.3\endcsname{0.433}
\expandafter\gdef\csname odunum@n@axis_breakdown.panel_mean_m2.axis3.3.3\endcsname{13}
\expandafter\gdef\csname odunum@ci@axis_breakdown.panel_mean_m2.axis3.3.3\endcsname{[0.336, 0.524]}
\expandafter\gdef\csname odunum@val@axis_breakdown.panel_mean_m2.axis3.3.4\endcsname{0.520}
\expandafter\gdef\csname odunum@n@axis_breakdown.panel_mean_m2.axis3.3.4\endcsname{13}
\expandafter\gdef\csname odunum@ci@axis_breakdown.panel_mean_m2.axis3.3.4\endcsname{[0.425, 0.600]}
\expandafter\gdef\csname odunum@val@axis_breakdown.panel_mean_m2.axis3.3.5\endcsname{0.477}
\expandafter\gdef\csname odunum@n@axis_breakdown.panel_mean_m2.axis3.3.5\endcsname{13}
\expandafter\gdef\csname odunum@ci@axis_breakdown.panel_mean_m2.axis3.3.5\endcsname{[0.404, 0.541]}
\expandafter\gdef\csname odunum@val@axis_breakdown.panel_mean_m2.axis3.3.6\endcsname{0.555}
\expandafter\gdef\csname odunum@n@axis_breakdown.panel_mean_m2.axis3.3.6\endcsname{13}
\expandafter\gdef\csname odunum@ci@axis_breakdown.panel_mean_m2.axis3.3.6\endcsname{[0.449, 0.649]}
\expandafter\gdef\csname odunum@val@axis_breakdown.panel_mean_m2.axis4.4.1\endcsname{0.506}
\expandafter\gdef\csname odunum@n@axis_breakdown.panel_mean_m2.axis4.4.1\endcsname{13}
\expandafter\gdef\csname odunum@ci@axis_breakdown.panel_mean_m2.axis4.4.1\endcsname{[0.409, 0.590]}
\expandafter\gdef\csname odunum@val@axis_breakdown.panel_mean_m2.axis4.4.2\endcsname{0.521}
\expandafter\gdef\csname odunum@n@axis_breakdown.panel_mean_m2.axis4.4.2\endcsname{13}
\expandafter\gdef\csname odunum@ci@axis_breakdown.panel_mean_m2.axis4.4.2\endcsname{[0.431, 0.599]}
\expandafter\gdef\csname odunum@val@axis_breakdown.panel_mean_m2.axis4.4.3\endcsname{0.484}
\expandafter\gdef\csname odunum@n@axis_breakdown.panel_mean_m2.axis4.4.3\endcsname{13}
\expandafter\gdef\csname odunum@ci@axis_breakdown.panel_mean_m2.axis4.4.3\endcsname{[0.387, 0.574]}
\expandafter\gdef\csname odunum@val@axis_breakdown.panel_mean_m2.axis4.4.4\endcsname{0.502}
\expandafter\gdef\csname odunum@n@axis_breakdown.panel_mean_m2.axis4.4.4\endcsname{13}
\expandafter\gdef\csname odunum@ci@axis_breakdown.panel_mean_m2.axis4.4.4\endcsname{[0.406, 0.585]}
\expandafter\gdef\csname odunum@val@axis_breakdown.panel_mean_m2.axis4.4.5\endcsname{0.527}
\expandafter\gdef\csname odunum@n@axis_breakdown.panel_mean_m2.axis4.4.5\endcsname{13}
\expandafter\gdef\csname odunum@ci@axis_breakdown.panel_mean_m2.axis4.4.5\endcsname{[0.432, 0.609]}
\expandafter\gdef\csname odunum@val@axis_breakdown.panel_mean_m2.axis5.audio_only.5.1\endcsname{0.598}
\expandafter\gdef\csname odunum@n@axis_breakdown.panel_mean_m2.axis5.audio_only.5.1\endcsname{13}
\expandafter\gdef\csname odunum@ci@axis_breakdown.panel_mean_m2.axis5.audio_only.5.1\endcsname{[0.494, 0.689]}
\expandafter\gdef\csname odunum@val@axis_breakdown.panel_mean_m2.axis5.audio_only.5.10\endcsname{0.567}
\expandafter\gdef\csname odunum@n@axis_breakdown.panel_mean_m2.axis5.audio_only.5.10\endcsname{13}
\expandafter\gdef\csname odunum@ci@axis_breakdown.panel_mean_m2.axis5.audio_only.5.10\endcsname{[0.479, 0.648]}
\expandafter\gdef\csname odunum@val@axis_breakdown.panel_mean_m2.axis5.audio_only.5.2\endcsname{0.558}
\expandafter\gdef\csname odunum@n@axis_breakdown.panel_mean_m2.axis5.audio_only.5.2\endcsname{13}
\expandafter\gdef\csname odunum@ci@axis_breakdown.panel_mean_m2.axis5.audio_only.5.2\endcsname{[0.470, 0.638]}
\expandafter\gdef\csname odunum@val@axis_breakdown.panel_mean_m2.axis5.audio_only.5.3\endcsname{0.609}
\expandafter\gdef\csname odunum@n@axis_breakdown.panel_mean_m2.axis5.audio_only.5.3\endcsname{13}
\expandafter\gdef\csname odunum@ci@axis_breakdown.panel_mean_m2.axis5.audio_only.5.3\endcsname{[0.511, 0.693]}
\expandafter\gdef\csname odunum@val@axis_breakdown.panel_mean_m2.axis5.audio_only.5.4\endcsname{0.537}
\expandafter\gdef\csname odunum@n@axis_breakdown.panel_mean_m2.axis5.audio_only.5.4\endcsname{13}
\expandafter\gdef\csname odunum@ci@axis_breakdown.panel_mean_m2.axis5.audio_only.5.4\endcsname{[0.448, 0.621]}
\expandafter\gdef\csname odunum@val@axis_breakdown.panel_mean_m2.axis5.audio_only.5.5\endcsname{0.549}
\expandafter\gdef\csname odunum@n@axis_breakdown.panel_mean_m2.axis5.audio_only.5.5\endcsname{13}
\expandafter\gdef\csname odunum@ci@axis_breakdown.panel_mean_m2.axis5.audio_only.5.5\endcsname{[0.444, 0.645]}
\expandafter\gdef\csname odunum@val@axis_breakdown.panel_mean_m2.axis5.audio_only.5.6\endcsname{0.589}
\expandafter\gdef\csname odunum@n@axis_breakdown.panel_mean_m2.axis5.audio_only.5.6\endcsname{13}
\expandafter\gdef\csname odunum@ci@axis_breakdown.panel_mean_m2.axis5.audio_only.5.6\endcsname{[0.512, 0.661]}
\expandafter\gdef\csname odunum@val@axis_breakdown.panel_mean_m2.axis5.audio_only.5.7\endcsname{0.547}
\expandafter\gdef\csname odunum@n@axis_breakdown.panel_mean_m2.axis5.audio_only.5.7\endcsname{13}
\expandafter\gdef\csname odunum@ci@axis_breakdown.panel_mean_m2.axis5.audio_only.5.7\endcsname{[0.464, 0.623]}
\expandafter\gdef\csname odunum@val@axis_breakdown.panel_mean_m2.axis5.audio_only.5.8\endcsname{0.494}
\expandafter\gdef\csname odunum@n@axis_breakdown.panel_mean_m2.axis5.audio_only.5.8\endcsname{13}
\expandafter\gdef\csname odunum@ci@axis_breakdown.panel_mean_m2.axis5.audio_only.5.8\endcsname{[0.392, 0.592]}
\expandafter\gdef\csname odunum@val@axis_breakdown.panel_mean_m2.axis5.audio_only.5.9\endcsname{0.592}
\expandafter\gdef\csname odunum@n@axis_breakdown.panel_mean_m2.axis5.audio_only.5.9\endcsname{13}
\expandafter\gdef\csname odunum@ci@axis_breakdown.panel_mean_m2.axis5.audio_only.5.9\endcsname{[0.490, 0.680]}
\expandafter\gdef\csname odunum@val@axis_breakdown.panel_mean_m2.axis5.audio_visual.5.1\endcsname{0.490}
\expandafter\gdef\csname odunum@n@axis_breakdown.panel_mean_m2.axis5.audio_visual.5.1\endcsname{13}
\expandafter\gdef\csname odunum@ci@axis_breakdown.panel_mean_m2.axis5.audio_visual.5.1\endcsname{[0.389, 0.575]}
\expandafter\gdef\csname odunum@val@axis_breakdown.panel_mean_m2.axis5.audio_visual.5.10\endcsname{0.506}
\expandafter\gdef\csname odunum@n@axis_breakdown.panel_mean_m2.axis5.audio_visual.5.10\endcsname{13}
\expandafter\gdef\csname odunum@ci@axis_breakdown.panel_mean_m2.axis5.audio_visual.5.10\endcsname{[0.403, 0.595]}
\expandafter\gdef\csname odunum@val@axis_breakdown.panel_mean_m2.axis5.audio_visual.5.2\endcsname{0.455}
\expandafter\gdef\csname odunum@n@axis_breakdown.panel_mean_m2.axis5.audio_visual.5.2\endcsname{13}
\expandafter\gdef\csname odunum@ci@axis_breakdown.panel_mean_m2.axis5.audio_visual.5.2\endcsname{[0.372, 0.526]}
\expandafter\gdef\csname odunum@val@axis_breakdown.panel_mean_m2.axis5.audio_visual.5.3\endcsname{0.480}
\expandafter\gdef\csname odunum@n@axis_breakdown.panel_mean_m2.axis5.audio_visual.5.3\endcsname{13}
\expandafter\gdef\csname odunum@ci@axis_breakdown.panel_mean_m2.axis5.audio_visual.5.3\endcsname{[0.385, 0.564]}
\expandafter\gdef\csname odunum@val@axis_breakdown.panel_mean_m2.axis5.audio_visual.5.4\endcsname{0.456}
\expandafter\gdef\csname odunum@n@axis_breakdown.panel_mean_m2.axis5.audio_visual.5.4\endcsname{13}
\expandafter\gdef\csname odunum@ci@axis_breakdown.panel_mean_m2.axis5.audio_visual.5.4\endcsname{[0.364, 0.537]}
\expandafter\gdef\csname odunum@val@axis_breakdown.panel_mean_m2.axis5.audio_visual.5.5\endcsname{0.503}
\expandafter\gdef\csname odunum@n@axis_breakdown.panel_mean_m2.axis5.audio_visual.5.5\endcsname{13}
\expandafter\gdef\csname odunum@ci@axis_breakdown.panel_mean_m2.axis5.audio_visual.5.5\endcsname{[0.392, 0.599]}
\expandafter\gdef\csname odunum@val@axis_breakdown.panel_mean_m2.axis5.audio_visual.5.6\endcsname{0.500}
\expandafter\gdef\csname odunum@n@axis_breakdown.panel_mean_m2.axis5.audio_visual.5.6\endcsname{13}
\expandafter\gdef\csname odunum@ci@axis_breakdown.panel_mean_m2.axis5.audio_visual.5.6\endcsname{[0.407, 0.583]}
\expandafter\gdef\csname odunum@val@axis_breakdown.panel_mean_m2.axis5.audio_visual.5.7\endcsname{0.482}
\expandafter\gdef\csname odunum@n@axis_breakdown.panel_mean_m2.axis5.audio_visual.5.7\endcsname{13}
\expandafter\gdef\csname odunum@ci@axis_breakdown.panel_mean_m2.axis5.audio_visual.5.7\endcsname{[0.385, 0.565]}
\expandafter\gdef\csname odunum@val@axis_breakdown.panel_mean_m2.axis5.audio_visual.5.8\endcsname{0.461}
\expandafter\gdef\csname odunum@n@axis_breakdown.panel_mean_m2.axis5.audio_visual.5.8\endcsname{13}
\expandafter\gdef\csname odunum@ci@axis_breakdown.panel_mean_m2.axis5.audio_visual.5.8\endcsname{[0.376, 0.535]}
\expandafter\gdef\csname odunum@val@axis_breakdown.panel_mean_m2.axis5.audio_visual.5.9\endcsname{0.508}
\expandafter\gdef\csname odunum@n@axis_breakdown.panel_mean_m2.axis5.audio_visual.5.9\endcsname{13}
\expandafter\gdef\csname odunum@ci@axis_breakdown.panel_mean_m2.axis5.audio_visual.5.9\endcsname{[0.403, 0.598]}
\expandafter\gdef\csname odunum@val@axis_breakdown.panel_mean_m2.axis6.6.1\endcsname{0.490}
\expandafter\gdef\csname odunum@n@axis_breakdown.panel_mean_m2.axis6.6.1\endcsname{13}
\expandafter\gdef\csname odunum@ci@axis_breakdown.panel_mean_m2.axis6.6.1\endcsname{[0.393, 0.576]}
\expandafter\gdef\csname odunum@val@axis_breakdown.panel_mean_m2.axis6.6.10\endcsname{0.511}
\expandafter\gdef\csname odunum@n@axis_breakdown.panel_mean_m2.axis6.6.10\endcsname{13}
\expandafter\gdef\csname odunum@ci@axis_breakdown.panel_mean_m2.axis6.6.10\endcsname{[0.419, 0.592]}
\expandafter\gdef\csname odunum@val@axis_breakdown.panel_mean_m2.axis6.6.12\endcsname{0.512}
\expandafter\gdef\csname odunum@n@axis_breakdown.panel_mean_m2.axis6.6.12\endcsname{13}
\expandafter\gdef\csname odunum@ci@axis_breakdown.panel_mean_m2.axis6.6.12\endcsname{[0.415, 0.596]}
\expandafter\gdef\csname odunum@val@axis_breakdown.panel_mean_m2.axis6.6.13\endcsname{0.484}
\expandafter\gdef\csname odunum@n@axis_breakdown.panel_mean_m2.axis6.6.13\endcsname{13}
\expandafter\gdef\csname odunum@ci@axis_breakdown.panel_mean_m2.axis6.6.13\endcsname{[0.388, 0.572]}
\expandafter\gdef\csname odunum@val@axis_breakdown.panel_mean_m2.axis6.6.14\endcsname{0.477}
\expandafter\gdef\csname odunum@n@axis_breakdown.panel_mean_m2.axis6.6.14\endcsname{13}
\expandafter\gdef\csname odunum@ci@axis_breakdown.panel_mean_m2.axis6.6.14\endcsname{[0.403, 0.546]}
\expandafter\gdef\csname odunum@val@axis_breakdown.panel_mean_m2.axis6.6.18\endcsname{0.503}
\expandafter\gdef\csname odunum@n@axis_breakdown.panel_mean_m2.axis6.6.18\endcsname{13}
\expandafter\gdef\csname odunum@ci@axis_breakdown.panel_mean_m2.axis6.6.18\endcsname{[0.408, 0.583]}
\expandafter\gdef\csname odunum@val@axis_breakdown.panel_mean_m2.axis6.6.19\endcsname{0.489}
\expandafter\gdef\csname odunum@n@axis_breakdown.panel_mean_m2.axis6.6.19\endcsname{13}
\expandafter\gdef\csname odunum@ci@axis_breakdown.panel_mean_m2.axis6.6.19\endcsname{[0.413, 0.555]}
\expandafter\gdef\csname odunum@val@axis_breakdown.panel_mean_m2.axis6.6.2\endcsname{0.556}
\expandafter\gdef\csname odunum@n@axis_breakdown.panel_mean_m2.axis6.6.2\endcsname{13}
\expandafter\gdef\csname odunum@ci@axis_breakdown.panel_mean_m2.axis6.6.2\endcsname{[0.452, 0.646]}
\expandafter\gdef\csname odunum@val@axis_breakdown.panel_mean_m2.axis6.6.3\endcsname{0.518}
\expandafter\gdef\csname odunum@n@axis_breakdown.panel_mean_m2.axis6.6.3\endcsname{13}
\expandafter\gdef\csname odunum@ci@axis_breakdown.panel_mean_m2.axis6.6.3\endcsname{[0.417, 0.606]}
\expandafter\gdef\csname odunum@val@axis_breakdown.panel_mean_m2.axis6.6.5\endcsname{0.501}
\expandafter\gdef\csname odunum@n@axis_breakdown.panel_mean_m2.axis6.6.5\endcsname{13}
\expandafter\gdef\csname odunum@ci@axis_breakdown.panel_mean_m2.axis6.6.5\endcsname{[0.398, 0.594]}
\expandafter\gdef\csname odunum@val@axis_breakdown.panel_mean_m2.axis6.6.6\endcsname{0.523}
\expandafter\gdef\csname odunum@n@axis_breakdown.panel_mean_m2.axis6.6.6\endcsname{13}
\expandafter\gdef\csname odunum@ci@axis_breakdown.panel_mean_m2.axis6.6.6\endcsname{[0.426, 0.608]}
\expandafter\gdef\csname odunum@val@axis_breakdown.panel_mean_m2.axis6.6.7\endcsname{0.546}
\expandafter\gdef\csname odunum@n@axis_breakdown.panel_mean_m2.axis6.6.7\endcsname{13}
\expandafter\gdef\csname odunum@ci@axis_breakdown.panel_mean_m2.axis6.6.7\endcsname{[0.451, 0.627]}
\expandafter\gdef\csname odunum@val@axis_breakdown.panel_mean_m2.axis6.6.8\endcsname{0.465}
\expandafter\gdef\csname odunum@n@axis_breakdown.panel_mean_m2.axis6.6.8\endcsname{13}
\expandafter\gdef\csname odunum@ci@axis_breakdown.panel_mean_m2.axis6.6.8\endcsname{[0.374, 0.548]}
\expandafter\gdef\csname odunum@val@axis_breakdown.panel_mean_m2.axis6.6.9\endcsname{0.503}
\expandafter\gdef\csname odunum@n@axis_breakdown.panel_mean_m2.axis6.6.9\endcsname{13}
\expandafter\gdef\csname odunum@ci@axis_breakdown.panel_mean_m2.axis6.6.9\endcsname{[0.410, 0.586]}
\expandafter\gdef\csname odunum@val@axis_breakdown.panel_mean_m2.difficulty.q1\endcsname{0.576}
\expandafter\gdef\csname odunum@n@axis_breakdown.panel_mean_m2.difficulty.q1\endcsname{13}
\expandafter\gdef\csname odunum@ci@axis_breakdown.panel_mean_m2.difficulty.q1\endcsname{[0.495, 0.647]}
\expandafter\gdef\csname odunum@val@axis_breakdown.panel_mean_m2.difficulty.q2\endcsname{0.524}
\expandafter\gdef\csname odunum@n@axis_breakdown.panel_mean_m2.difficulty.q2\endcsname{13}
\expandafter\gdef\csname odunum@ci@axis_breakdown.panel_mean_m2.difficulty.q2\endcsname{[0.429, 0.605]}
\expandafter\gdef\csname odunum@val@axis_breakdown.panel_mean_m2.difficulty.q3\endcsname{0.511}
\expandafter\gdef\csname odunum@n@axis_breakdown.panel_mean_m2.difficulty.q3\endcsname{13}
\expandafter\gdef\csname odunum@ci@axis_breakdown.panel_mean_m2.difficulty.q3\endcsname{[0.418, 0.589]}
\expandafter\gdef\csname odunum@val@axis_breakdown.panel_mean_m2.difficulty.q4\endcsname{0.484}
\expandafter\gdef\csname odunum@n@axis_breakdown.panel_mean_m2.difficulty.q4\endcsname{13}
\expandafter\gdef\csname odunum@ci@axis_breakdown.panel_mean_m2.difficulty.q4\endcsname{[0.388, 0.567]}
\expandafter\gdef\csname odunum@val@axis_breakdown.pop.axis1.audio_only.1.1.n\endcsname{158}
\expandafter\gdef\csname odunum@n@axis_breakdown.pop.axis1.audio_only.1.1.n\endcsname{570}
\expandafter\gdef\csname odunum@val@axis_breakdown.pop.axis1.audio_only.1.2.n\endcsname{379}
\expandafter\gdef\csname odunum@n@axis_breakdown.pop.axis1.audio_only.1.2.n\endcsname{570}
\expandafter\gdef\csname odunum@val@axis_breakdown.pop.axis1.audio_only.1.3.n\endcsname{32}
\expandafter\gdef\csname odunum@n@axis_breakdown.pop.axis1.audio_only.1.3.n\endcsname{570}
\expandafter\gdef\csname odunum@val@axis_breakdown.pop.axis1.audio_only.1.4.n\endcsname{1}
\expandafter\gdef\csname odunum@n@axis_breakdown.pop.axis1.audio_only.1.4.n\endcsname{570}
\expandafter\gdef\csname odunum@val@axis_breakdown.pop.axis1.audio_visual.1.1.n\endcsname{94}
\expandafter\gdef\csname odunum@n@axis_breakdown.pop.axis1.audio_visual.1.1.n\endcsname{1\,061}
\expandafter\gdef\csname odunum@val@axis_breakdown.pop.axis1.audio_visual.1.2.n\endcsname{294}
\expandafter\gdef\csname odunum@n@axis_breakdown.pop.axis1.audio_visual.1.2.n\endcsname{1\,061}
\expandafter\gdef\csname odunum@val@axis_breakdown.pop.axis1.audio_visual.1.3.n\endcsname{291}
\expandafter\gdef\csname odunum@n@axis_breakdown.pop.axis1.audio_visual.1.3.n\endcsname{1\,061}
\expandafter\gdef\csname odunum@val@axis_breakdown.pop.axis1.audio_visual.1.4.n\endcsname{382}
\expandafter\gdef\csname odunum@n@axis_breakdown.pop.axis1.audio_visual.1.4.n\endcsname{1\,061}
\expandafter\gdef\csname odunum@val@axis_breakdown.pop.axis1.n_labelled_neg\endcsname{0}
\expandafter\gdef\csname odunum@n@axis_breakdown.pop.axis1.n_labelled_neg\endcsname{447}
\expandafter\gdef\csname odunum@val@axis_breakdown.pop.axis1.n_labelled_pos\endcsname{1\,631}
\expandafter\gdef\csname odunum@n@axis_breakdown.pop.axis1.n_labelled_pos\endcsname{1\,631}
\expandafter\gdef\csname odunum@val@axis_breakdown.pop.axis2.2.1.n\endcsname{143}
\expandafter\gdef\csname odunum@n@axis_breakdown.pop.axis2.2.1.n\endcsname{1\,631}
\expandafter\gdef\csname odunum@val@axis_breakdown.pop.axis2.2.2.n\endcsname{440}
\expandafter\gdef\csname odunum@n@axis_breakdown.pop.axis2.2.2.n\endcsname{1\,631}
\expandafter\gdef\csname odunum@val@axis_breakdown.pop.axis2.2.3.n\endcsname{531}
\expandafter\gdef\csname odunum@n@axis_breakdown.pop.axis2.2.3.n\endcsname{1\,631}
\expandafter\gdef\csname odunum@val@axis_breakdown.pop.axis2.2.4.n\endcsname{261}
\expandafter\gdef\csname odunum@n@axis_breakdown.pop.axis2.2.4.n\endcsname{1\,631}
\expandafter\gdef\csname odunum@val@axis_breakdown.pop.axis2.2.5.n\endcsname{256}
\expandafter\gdef\csname odunum@n@axis_breakdown.pop.axis2.2.5.n\endcsname{1\,631}
\expandafter\gdef\csname odunum@val@axis_breakdown.pop.axis2.n_labelled_neg\endcsname{0}
\expandafter\gdef\csname odunum@n@axis_breakdown.pop.axis2.n_labelled_neg\endcsname{447}
\expandafter\gdef\csname odunum@val@axis_breakdown.pop.axis2.n_labelled_pos\endcsname{1\,631}
\expandafter\gdef\csname odunum@n@axis_breakdown.pop.axis2.n_labelled_pos\endcsname{1\,631}
\expandafter\gdef\csname odunum@val@axis_breakdown.pop.axis3.3.1.n\endcsname{109}
\expandafter\gdef\csname odunum@n@axis_breakdown.pop.axis3.3.1.n\endcsname{1\,631}
\expandafter\gdef\csname odunum@val@axis_breakdown.pop.axis3.3.2.n\endcsname{116}
\expandafter\gdef\csname odunum@n@axis_breakdown.pop.axis3.3.2.n\endcsname{1\,631}
\expandafter\gdef\csname odunum@val@axis_breakdown.pop.axis3.3.3.n\endcsname{207}
\expandafter\gdef\csname odunum@n@axis_breakdown.pop.axis3.3.3.n\endcsname{1\,631}
\expandafter\gdef\csname odunum@val@axis_breakdown.pop.axis3.3.4.n\endcsname{404}
\expandafter\gdef\csname odunum@n@axis_breakdown.pop.axis3.3.4.n\endcsname{1\,631}
\expandafter\gdef\csname odunum@val@axis_breakdown.pop.axis3.3.5.n\endcsname{294}
\expandafter\gdef\csname odunum@n@axis_breakdown.pop.axis3.3.5.n\endcsname{1\,631}
\expandafter\gdef\csname odunum@val@axis_breakdown.pop.axis3.3.6.n\endcsname{501}
\expandafter\gdef\csname odunum@n@axis_breakdown.pop.axis3.3.6.n\endcsname{1\,631}
\expandafter\gdef\csname odunum@val@axis_breakdown.pop.axis3.n_labelled_neg\endcsname{0}
\expandafter\gdef\csname odunum@n@axis_breakdown.pop.axis3.n_labelled_neg\endcsname{447}
\expandafter\gdef\csname odunum@val@axis_breakdown.pop.axis3.n_labelled_pos\endcsname{1\,631}
\expandafter\gdef\csname odunum@n@axis_breakdown.pop.axis3.n_labelled_pos\endcsname{1\,631}
\expandafter\gdef\csname odunum@val@axis_breakdown.pop.axis4.4.1.n\endcsname{202}
\expandafter\gdef\csname odunum@n@axis_breakdown.pop.axis4.4.1.n\endcsname{1\,631}
\expandafter\gdef\csname odunum@val@axis_breakdown.pop.axis4.4.2.n\endcsname{457}
\expandafter\gdef\csname odunum@n@axis_breakdown.pop.axis4.4.2.n\endcsname{1\,631}
\expandafter\gdef\csname odunum@val@axis_breakdown.pop.axis4.4.3.n\endcsname{164}
\expandafter\gdef\csname odunum@n@axis_breakdown.pop.axis4.4.3.n\endcsname{1\,631}
\expandafter\gdef\csname odunum@val@axis_breakdown.pop.axis4.4.4.n\endcsname{505}
\expandafter\gdef\csname odunum@n@axis_breakdown.pop.axis4.4.4.n\endcsname{1\,631}
\expandafter\gdef\csname odunum@val@axis_breakdown.pop.axis4.4.5.n\endcsname{303}
\expandafter\gdef\csname odunum@n@axis_breakdown.pop.axis4.4.5.n\endcsname{1\,631}
\expandafter\gdef\csname odunum@val@axis_breakdown.pop.axis4.n_labelled_neg\endcsname{447}
\expandafter\gdef\csname odunum@n@axis_breakdown.pop.axis4.n_labelled_neg\endcsname{447}
\expandafter\gdef\csname odunum@val@axis_breakdown.pop.axis4.n_labelled_pos\endcsname{1\,631}
\expandafter\gdef\csname odunum@n@axis_breakdown.pop.axis4.n_labelled_pos\endcsname{1\,631}
\expandafter\gdef\csname odunum@val@axis_breakdown.pop.axis5.audio_only.5.1.n\endcsname{45}
\expandafter\gdef\csname odunum@n@axis_breakdown.pop.axis5.audio_only.5.1.n\endcsname{570}
\expandafter\gdef\csname odunum@val@axis_breakdown.pop.axis5.audio_only.5.10.n\endcsname{72}
\expandafter\gdef\csname odunum@n@axis_breakdown.pop.axis5.audio_only.5.10.n\endcsname{570}
\expandafter\gdef\csname odunum@val@axis_breakdown.pop.axis5.audio_only.5.2.n\endcsname{75}
\expandafter\gdef\csname odunum@n@axis_breakdown.pop.axis5.audio_only.5.2.n\endcsname{570}
\expandafter\gdef\csname odunum@val@axis_breakdown.pop.axis5.audio_only.5.3.n\endcsname{50}
\expandafter\gdef\csname odunum@n@axis_breakdown.pop.axis5.audio_only.5.3.n\endcsname{570}
\expandafter\gdef\csname odunum@val@axis_breakdown.pop.axis5.audio_only.5.4.n\endcsname{68}
\expandafter\gdef\csname odunum@n@axis_breakdown.pop.axis5.audio_only.5.4.n\endcsname{570}
\expandafter\gdef\csname odunum@val@axis_breakdown.pop.axis5.audio_only.5.5.n\endcsname{52}
\expandafter\gdef\csname odunum@n@axis_breakdown.pop.axis5.audio_only.5.5.n\endcsname{570}
\expandafter\gdef\csname odunum@val@axis_breakdown.pop.axis5.audio_only.5.6.n\endcsname{52}
\expandafter\gdef\csname odunum@n@axis_breakdown.pop.axis5.audio_only.5.6.n\endcsname{570}
\expandafter\gdef\csname odunum@val@axis_breakdown.pop.axis5.audio_only.5.7.n\endcsname{46}
\expandafter\gdef\csname odunum@n@axis_breakdown.pop.axis5.audio_only.5.7.n\endcsname{570}
\expandafter\gdef\csname odunum@val@axis_breakdown.pop.axis5.audio_only.5.8.n\endcsname{57}
\expandafter\gdef\csname odunum@n@axis_breakdown.pop.axis5.audio_only.5.8.n\endcsname{570}
\expandafter\gdef\csname odunum@val@axis_breakdown.pop.axis5.audio_only.5.9.n\endcsname{53}
\expandafter\gdef\csname odunum@n@axis_breakdown.pop.axis5.audio_only.5.9.n\endcsname{570}
\expandafter\gdef\csname odunum@val@axis_breakdown.pop.axis5.audio_visual.5.1.n\endcsname{114}
\expandafter\gdef\csname odunum@n@axis_breakdown.pop.axis5.audio_visual.5.1.n\endcsname{1\,061}
\expandafter\gdef\csname odunum@val@axis_breakdown.pop.axis5.audio_visual.5.10.n\endcsname{123}
\expandafter\gdef\csname odunum@n@axis_breakdown.pop.axis5.audio_visual.5.10.n\endcsname{1\,061}
\expandafter\gdef\csname odunum@val@axis_breakdown.pop.axis5.audio_visual.5.2.n\endcsname{126}
\expandafter\gdef\csname odunum@n@axis_breakdown.pop.axis5.audio_visual.5.2.n\endcsname{1\,061}
\expandafter\gdef\csname odunum@val@axis_breakdown.pop.axis5.audio_visual.5.3.n\endcsname{100}
\expandafter\gdef\csname odunum@n@axis_breakdown.pop.axis5.audio_visual.5.3.n\endcsname{1\,061}
\expandafter\gdef\csname odunum@val@axis_breakdown.pop.axis5.audio_visual.5.4.n\endcsname{97}
\expandafter\gdef\csname odunum@n@axis_breakdown.pop.axis5.audio_visual.5.4.n\endcsname{1\,061}
\expandafter\gdef\csname odunum@val@axis_breakdown.pop.axis5.audio_visual.5.5.n\endcsname{99}
\expandafter\gdef\csname odunum@n@axis_breakdown.pop.axis5.audio_visual.5.5.n\endcsname{1\,061}
\expandafter\gdef\csname odunum@val@axis_breakdown.pop.axis5.audio_visual.5.6.n\endcsname{99}
\expandafter\gdef\csname odunum@n@axis_breakdown.pop.axis5.audio_visual.5.6.n\endcsname{1\,061}
\expandafter\gdef\csname odunum@val@axis_breakdown.pop.axis5.audio_visual.5.7.n\endcsname{104}
\expandafter\gdef\csname odunum@n@axis_breakdown.pop.axis5.audio_visual.5.7.n\endcsname{1\,061}
\expandafter\gdef\csname odunum@val@axis_breakdown.pop.axis5.audio_visual.5.8.n\endcsname{121}
\expandafter\gdef\csname odunum@n@axis_breakdown.pop.axis5.audio_visual.5.8.n\endcsname{1\,061}
\expandafter\gdef\csname odunum@val@axis_breakdown.pop.axis5.audio_visual.5.9.n\endcsname{78}
\expandafter\gdef\csname odunum@n@axis_breakdown.pop.axis5.audio_visual.5.9.n\endcsname{1\,061}
\expandafter\gdef\csname odunum@val@axis_breakdown.pop.axis5.n_labelled_neg\endcsname{0}
\expandafter\gdef\csname odunum@n@axis_breakdown.pop.axis5.n_labelled_neg\endcsname{447}
\expandafter\gdef\csname odunum@val@axis_breakdown.pop.axis5.n_labelled_pos\endcsname{1\,631}
\expandafter\gdef\csname odunum@n@axis_breakdown.pop.axis5.n_labelled_pos\endcsname{1\,631}
\expandafter\gdef\csname odunum@val@axis_breakdown.pop.axis6.6.1.n\endcsname{230}
\expandafter\gdef\csname odunum@n@axis_breakdown.pop.axis6.6.1.n\endcsname{1\,631}
\expandafter\gdef\csname odunum@val@axis_breakdown.pop.axis6.6.10.n\endcsname{102}
\expandafter\gdef\csname odunum@n@axis_breakdown.pop.axis6.6.10.n\endcsname{1\,631}
\expandafter\gdef\csname odunum@val@axis_breakdown.pop.axis6.6.11.n\endcsname{8}
\expandafter\gdef\csname odunum@n@axis_breakdown.pop.axis6.6.11.n\endcsname{1\,631}
\expandafter\gdef\csname odunum@val@axis_breakdown.pop.axis6.6.12.n\endcsname{129}
\expandafter\gdef\csname odunum@n@axis_breakdown.pop.axis6.6.12.n\endcsname{1\,631}
\expandafter\gdef\csname odunum@val@axis_breakdown.pop.axis6.6.13.n\endcsname{58}
\expandafter\gdef\csname odunum@n@axis_breakdown.pop.axis6.6.13.n\endcsname{1\,631}
\expandafter\gdef\csname odunum@val@axis_breakdown.pop.axis6.6.14.n\endcsname{82}
\expandafter\gdef\csname odunum@n@axis_breakdown.pop.axis6.6.14.n\endcsname{1\,631}
\expandafter\gdef\csname odunum@val@axis_breakdown.pop.axis6.6.15.n\endcsname{24}
\expandafter\gdef\csname odunum@n@axis_breakdown.pop.axis6.6.15.n\endcsname{1\,631}
\expandafter\gdef\csname odunum@val@axis_breakdown.pop.axis6.6.16.n\endcsname{18}
\expandafter\gdef\csname odunum@n@axis_breakdown.pop.axis6.6.16.n\endcsname{1\,631}
\expandafter\gdef\csname odunum@val@axis_breakdown.pop.axis6.6.17.n\endcsname{12}
\expandafter\gdef\csname odunum@n@axis_breakdown.pop.axis6.6.17.n\endcsname{1\,631}
\expandafter\gdef\csname odunum@val@axis_breakdown.pop.axis6.6.18.n\endcsname{72}
\expandafter\gdef\csname odunum@n@axis_breakdown.pop.axis6.6.18.n\endcsname{1\,631}
\expandafter\gdef\csname odunum@val@axis_breakdown.pop.axis6.6.19.n\endcsname{44}
\expandafter\gdef\csname odunum@n@axis_breakdown.pop.axis6.6.19.n\endcsname{1\,631}
\expandafter\gdef\csname odunum@val@axis_breakdown.pop.axis6.6.2.n\endcsname{41}
\expandafter\gdef\csname odunum@n@axis_breakdown.pop.axis6.6.2.n\endcsname{1\,631}
\expandafter\gdef\csname odunum@val@axis_breakdown.pop.axis6.6.3.n\endcsname{44}
\expandafter\gdef\csname odunum@n@axis_breakdown.pop.axis6.6.3.n\endcsname{1\,631}
\expandafter\gdef\csname odunum@val@axis_breakdown.pop.axis6.6.4.n\endcsname{24}
\expandafter\gdef\csname odunum@n@axis_breakdown.pop.axis6.6.4.n\endcsname{1\,631}
\expandafter\gdef\csname odunum@val@axis_breakdown.pop.axis6.6.5.n\endcsname{48}
\expandafter\gdef\csname odunum@n@axis_breakdown.pop.axis6.6.5.n\endcsname{1\,631}
\expandafter\gdef\csname odunum@val@axis_breakdown.pop.axis6.6.6.n\endcsname{134}
\expandafter\gdef\csname odunum@n@axis_breakdown.pop.axis6.6.6.n\endcsname{1\,631}
\expandafter\gdef\csname odunum@val@axis_breakdown.pop.axis6.6.7.n\endcsname{393}
\expandafter\gdef\csname odunum@n@axis_breakdown.pop.axis6.6.7.n\endcsname{1\,631}
\expandafter\gdef\csname odunum@val@axis_breakdown.pop.axis6.6.8.n\endcsname{49}
\expandafter\gdef\csname odunum@n@axis_breakdown.pop.axis6.6.8.n\endcsname{1\,631}
\expandafter\gdef\csname odunum@val@axis_breakdown.pop.axis6.6.9.n\endcsname{119}
\expandafter\gdef\csname odunum@n@axis_breakdown.pop.axis6.6.9.n\endcsname{1\,631}
\expandafter\gdef\csname odunum@val@axis_breakdown.pop.axis6.n_labelled_neg\endcsname{447}
\expandafter\gdef\csname odunum@n@axis_breakdown.pop.axis6.n_labelled_neg\endcsname{447}
\expandafter\gdef\csname odunum@val@axis_breakdown.pop.axis6.n_labelled_pos\endcsname{1\,631}
\expandafter\gdef\csname odunum@n@axis_breakdown.pop.axis6.n_labelled_pos\endcsname{1\,631}
\expandafter\gdef\csname odunum@val@axis_breakdown.pop.difficulty.q1.n\endcsname{358}
\expandafter\gdef\csname odunum@n@axis_breakdown.pop.difficulty.q1.n\endcsname{1\,429}
\expandafter\gdef\csname odunum@val@axis_breakdown.pop.difficulty.q2.n\endcsname{357}
\expandafter\gdef\csname odunum@n@axis_breakdown.pop.difficulty.q2.n\endcsname{1\,429}
\expandafter\gdef\csname odunum@val@axis_breakdown.pop.difficulty.q3.n\endcsname{358}
\expandafter\gdef\csname odunum@n@axis_breakdown.pop.difficulty.q3.n\endcsname{1\,429}
\expandafter\gdef\csname odunum@val@axis_breakdown.pop.difficulty.q4.n\endcsname{356}
\expandafter\gdef\csname odunum@n@axis_breakdown.pop.difficulty.q4.n\endcsname{1\,429}
\expandafter\gdef\csname odunum@val@benchmark_main.cascade.n_complete\endcsname{1}
\expandafter\gdef\csname odunum@n@benchmark_main.cascade.n_complete\endcsname{1}
\expandafter\gdef\csname odunum@val@benchmark_main.cascade.reader\endcsname{GPT-5.4 (openai.gpt-5.4)}
\expandafter\gdef\csname odunum@n@benchmark_main.cascade.reader\endcsname{1}
\expandafter\gdef\csname odunum@val@benchmark_main.complete.models\endcsname{Gemini 3.1 Pro, Gemini 3.7 Flash, Gemini 3.5 Flash Lite, Qwen3.5-Omni-Plus, Seed 2.0 Lite, GPT-Realtime-2, Qwen3-Omni-30B-A3B, Qwen3-Omni-30B-A3B-Instruct, Qwen2.5-Omni-7B, Ming-Flash-Omni 2.0, MiniCPM-o 4.5, Nemotron 3 Nano Omni 30B-A3B, video-SALMONN2+ 7B, Kimi-Audio-7B-Instruct}
\expandafter\gdef\csname odunum@n@benchmark_main.complete.models\endcsname{14}
\expandafter\gdef\csname odunum@val@benchmark_main.complete.n_models\endcsname{14}
\expandafter\gdef\csname odunum@n@benchmark_main.complete.n_models\endcsname{14}
\expandafter\gdef\csname odunum@val@benchmark_main.complete.threshold\endcsname{99\%}
\expandafter\gdef\csname odunum@n@benchmark_main.complete.threshold\endcsname{15}
\expandafter\gdef\csname odunum@val@benchmark_main.coverage_of_subset.cascade_asr\endcsname{1.000}
\expandafter\gdef\csname odunum@n@benchmark_main.coverage_of_subset.cascade_asr\endcsname{2\,078}
\expandafter\gdef\csname odunum@val@benchmark_main.coverage_of_subset.gemini\endcsname{1.000}
\expandafter\gdef\csname odunum@n@benchmark_main.coverage_of_subset.gemini\endcsname{2\,078}
\expandafter\gdef\csname odunum@val@benchmark_main.coverage_of_subset.gemini35_flash_lite\endcsname{1.000}
\expandafter\gdef\csname odunum@n@benchmark_main.coverage_of_subset.gemini35_flash_lite\endcsname{2\,078}
\expandafter\gdef\csname odunum@val@benchmark_main.coverage_of_subset.gemini37_flash\endcsname{0.995}
\expandafter\gdef\csname odunum@n@benchmark_main.coverage_of_subset.gemini37_flash\endcsname{2\,078}
\expandafter\gdef\csname odunum@val@benchmark_main.coverage_of_subset.gpt_realtime\endcsname{1.000}
\expandafter\gdef\csname odunum@n@benchmark_main.coverage_of_subset.gpt_realtime\endcsname{731}
\expandafter\gdef\csname odunum@val@benchmark_main.coverage_of_subset.kimi_audio_7b_instruct\endcsname{0.999}
\expandafter\gdef\csname odunum@n@benchmark_main.coverage_of_subset.kimi_audio_7b_instruct\endcsname{731}
\expandafter\gdef\csname odunum@val@benchmark_main.coverage_of_subset.ming\endcsname{1.000}
\expandafter\gdef\csname odunum@n@benchmark_main.coverage_of_subset.ming\endcsname{2\,078}
\expandafter\gdef\csname odunum@val@benchmark_main.coverage_of_subset.minicpm_o\endcsname{0.999}
\expandafter\gdef\csname odunum@n@benchmark_main.coverage_of_subset.minicpm_o\endcsname{2\,078}
\expandafter\gdef\csname odunum@val@benchmark_main.coverage_of_subset.nemotron\endcsname{1.000}
\expandafter\gdef\csname odunum@n@benchmark_main.coverage_of_subset.nemotron\endcsname{2\,078}
\expandafter\gdef\csname odunum@val@benchmark_main.coverage_of_subset.qwen25_omni\endcsname{1.000}
\expandafter\gdef\csname odunum@n@benchmark_main.coverage_of_subset.qwen25_omni\endcsname{2\,078}
\expandafter\gdef\csname odunum@val@benchmark_main.coverage_of_subset.qwen3_omni_instruct\endcsname{1.000}
\expandafter\gdef\csname odunum@n@benchmark_main.coverage_of_subset.qwen3_omni_instruct\endcsname{2\,078}
\expandafter\gdef\csname odunum@val@benchmark_main.coverage_of_subset.qwen3_omni_think\endcsname{1.000}
\expandafter\gdef\csname odunum@n@benchmark_main.coverage_of_subset.qwen3_omni_think\endcsname{2\,078}
\expandafter\gdef\csname odunum@val@benchmark_main.coverage_of_subset.qwen_plus\endcsname{1.000}
\expandafter\gdef\csname odunum@n@benchmark_main.coverage_of_subset.qwen_plus\endcsname{2\,078}
\expandafter\gdef\csname odunum@val@benchmark_main.coverage_of_subset.salmonn2_7b\endcsname{0.990}
\expandafter\gdef\csname odunum@n@benchmark_main.coverage_of_subset.salmonn2_7b\endcsname{2\,078}
\expandafter\gdef\csname odunum@val@benchmark_main.coverage_of_subset.seed\endcsname{1.000}
\expandafter\gdef\csname odunum@n@benchmark_main.coverage_of_subset.seed\endcsname{2\,078}
\expandafter\gdef\csname odunum@val@benchmark_main.family.cascade_asr\endcsname{text-only cascade}
\expandafter\gdef\csname odunum@n@benchmark_main.family.cascade_asr\endcsname{1}
\expandafter\gdef\csname odunum@val@benchmark_main.family.gemini\endcsname{hosted API}
\expandafter\gdef\csname odunum@n@benchmark_main.family.gemini\endcsname{1}
\expandafter\gdef\csname odunum@val@benchmark_main.family.gemini35_flash_lite\endcsname{hosted API}
\expandafter\gdef\csname odunum@n@benchmark_main.family.gemini35_flash_lite\endcsname{1}
\expandafter\gdef\csname odunum@val@benchmark_main.family.gemini37_flash\endcsname{hosted API}
\expandafter\gdef\csname odunum@n@benchmark_main.family.gemini37_flash\endcsname{1}
\expandafter\gdef\csname odunum@val@benchmark_main.family.gpt_realtime\endcsname{hosted API}
\expandafter\gdef\csname odunum@n@benchmark_main.family.gpt_realtime\endcsname{1}
\expandafter\gdef\csname odunum@val@benchmark_main.family.kimi_audio_7b_instruct\endcsname{open-source}
\expandafter\gdef\csname odunum@n@benchmark_main.family.kimi_audio_7b_instruct\endcsname{1}
\expandafter\gdef\csname odunum@val@benchmark_main.family.ming\endcsname{open-source}
\expandafter\gdef\csname odunum@n@benchmark_main.family.ming\endcsname{1}
\expandafter\gdef\csname odunum@val@benchmark_main.family.minicpm_o\endcsname{open-source}
\expandafter\gdef\csname odunum@n@benchmark_main.family.minicpm_o\endcsname{1}
\expandafter\gdef\csname odunum@val@benchmark_main.family.nemotron\endcsname{open-source}
\expandafter\gdef\csname odunum@n@benchmark_main.family.nemotron\endcsname{1}
\expandafter\gdef\csname odunum@val@benchmark_main.family.qwen25_omni\endcsname{open-source}
\expandafter\gdef\csname odunum@n@benchmark_main.family.qwen25_omni\endcsname{1}
\expandafter\gdef\csname odunum@val@benchmark_main.family.qwen3_omni_instruct\endcsname{open-source}
\expandafter\gdef\csname odunum@n@benchmark_main.family.qwen3_omni_instruct\endcsname{1}
\expandafter\gdef\csname odunum@val@benchmark_main.family.qwen3_omni_think\endcsname{open-source}
\expandafter\gdef\csname odunum@n@benchmark_main.family.qwen3_omni_think\endcsname{1}
\expandafter\gdef\csname odunum@val@benchmark_main.family.qwen_plus\endcsname{hosted API}
\expandafter\gdef\csname odunum@n@benchmark_main.family.qwen_plus\endcsname{1}
\expandafter\gdef\csname odunum@val@benchmark_main.family.salmonn2_7b\endcsname{open-source}
\expandafter\gdef\csname odunum@n@benchmark_main.family.salmonn2_7b\endcsname{1}
\expandafter\gdef\csname odunum@val@benchmark_main.family.seed\endcsname{hosted API}
\expandafter\gdef\csname odunum@n@benchmark_main.family.seed\endcsname{1}
\expandafter\gdef\csname odunum@val@benchmark_main.frontend.cascade_asr\endcsname{FunASR (fun-asr, via api\_usage/batch\_transcribe\_funasr.py)}
\expandafter\gdef\csname odunum@n@benchmark_main.frontend.cascade_asr\endcsname{1}
\expandafter\gdef\csname odunum@val@benchmark_main.ftr.cascade_asr\endcsname{0.615}
\expandafter\gdef\csname odunum@n@benchmark_main.ftr.cascade_asr\endcsname{447}
\expandafter\gdef\csname odunum@val@benchmark_main.ftr.cascade_asr.ao\endcsname{0.404}
\expandafter\gdef\csname odunum@n@benchmark_main.ftr.cascade_asr.ao\endcsname{161}
\expandafter\gdef\csname odunum@val@benchmark_main.ftr.cascade_asr.av\endcsname{0.734}
\expandafter\gdef\csname odunum@n@benchmark_main.ftr.cascade_asr.av\endcsname{286}
\expandafter\gdef\csname odunum@val@benchmark_main.ftr.gemini\endcsname{0.374}
\expandafter\gdef\csname odunum@n@benchmark_main.ftr.gemini\endcsname{447}
\expandafter\gdef\csname odunum@val@benchmark_main.ftr.gemini.ao\endcsname{0.317}
\expandafter\gdef\csname odunum@n@benchmark_main.ftr.gemini.ao\endcsname{161}
\expandafter\gdef\csname odunum@val@benchmark_main.ftr.gemini.av\endcsname{0.406}
\expandafter\gdef\csname odunum@n@benchmark_main.ftr.gemini.av\endcsname{286}
\expandafter\gdef\csname odunum@val@benchmark_main.ftr.gemini35_flash_lite\endcsname{0.494}
\expandafter\gdef\csname odunum@n@benchmark_main.ftr.gemini35_flash_lite\endcsname{447}
\expandafter\gdef\csname odunum@val@benchmark_main.ftr.gemini35_flash_lite.ao\endcsname{0.385}
\expandafter\gdef\csname odunum@n@benchmark_main.ftr.gemini35_flash_lite.ao\endcsname{161}
\expandafter\gdef\csname odunum@val@benchmark_main.ftr.gemini35_flash_lite.av\endcsname{0.556}
\expandafter\gdef\csname odunum@n@benchmark_main.ftr.gemini35_flash_lite.av\endcsname{286}
\expandafter\gdef\csname odunum@val@benchmark_main.ftr.gemini37_flash\endcsname{0.234}
\expandafter\gdef\csname odunum@n@benchmark_main.ftr.gemini37_flash\endcsname{440}
\expandafter\gdef\csname odunum@val@benchmark_main.ftr.gemini37_flash.ao\endcsname{0.160}
\expandafter\gdef\csname odunum@n@benchmark_main.ftr.gemini37_flash.ao\endcsname{156}
\expandafter\gdef\csname odunum@val@benchmark_main.ftr.gemini37_flash.av\endcsname{0.275}
\expandafter\gdef\csname odunum@n@benchmark_main.ftr.gemini37_flash.av\endcsname{284}
\expandafter\gdef\csname odunum@val@benchmark_main.ftr.gpt_realtime\endcsname{0.634}
\expandafter\gdef\csname odunum@n@benchmark_main.ftr.gpt_realtime\endcsname{161}
\expandafter\gdef\csname odunum@val@benchmark_main.ftr.gpt_realtime.ao\endcsname{0.634}
\expandafter\gdef\csname odunum@n@benchmark_main.ftr.gpt_realtime.ao\endcsname{161}
\expandafter\gdef\csname odunum@val@benchmark_main.ftr.kimi_audio_7b_instruct\endcsname{0.820}
\expandafter\gdef\csname odunum@n@benchmark_main.ftr.kimi_audio_7b_instruct\endcsname{161}
\expandafter\gdef\csname odunum@val@benchmark_main.ftr.kimi_audio_7b_instruct.ao\endcsname{0.820}
\expandafter\gdef\csname odunum@n@benchmark_main.ftr.kimi_audio_7b_instruct.ao\endcsname{161}
\expandafter\gdef\csname odunum@val@benchmark_main.ftr.ming\endcsname{0.872}
\expandafter\gdef\csname odunum@n@benchmark_main.ftr.ming\endcsname{447}
\expandafter\gdef\csname odunum@val@benchmark_main.ftr.ming.ao\endcsname{0.758}
\expandafter\gdef\csname odunum@n@benchmark_main.ftr.ming.ao\endcsname{161}
\expandafter\gdef\csname odunum@val@benchmark_main.ftr.ming.av\endcsname{0.937}
\expandafter\gdef\csname odunum@n@benchmark_main.ftr.ming.av\endcsname{286}
\expandafter\gdef\csname odunum@val@benchmark_main.ftr.minicpm_o\endcsname{0.814}
\expandafter\gdef\csname odunum@n@benchmark_main.ftr.minicpm_o\endcsname{446}
\expandafter\gdef\csname odunum@val@benchmark_main.ftr.minicpm_o.ao\endcsname{0.745}
\expandafter\gdef\csname odunum@n@benchmark_main.ftr.minicpm_o.ao\endcsname{161}
\expandafter\gdef\csname odunum@val@benchmark_main.ftr.minicpm_o.av\endcsname{0.853}
\expandafter\gdef\csname odunum@n@benchmark_main.ftr.minicpm_o.av\endcsname{285}
\expandafter\gdef\csname odunum@val@benchmark_main.ftr.nemotron\endcsname{0.720}
\expandafter\gdef\csname odunum@n@benchmark_main.ftr.nemotron\endcsname{447}
\expandafter\gdef\csname odunum@val@benchmark_main.ftr.nemotron.ao\endcsname{0.720}
\expandafter\gdef\csname odunum@n@benchmark_main.ftr.nemotron.ao\endcsname{161}
\expandafter\gdef\csname odunum@val@benchmark_main.ftr.nemotron.av\endcsname{0.720}
\expandafter\gdef\csname odunum@n@benchmark_main.ftr.nemotron.av\endcsname{286}
\expandafter\gdef\csname odunum@val@benchmark_main.ftr.qwen25_omni\endcsname{0.801}
\expandafter\gdef\csname odunum@n@benchmark_main.ftr.qwen25_omni\endcsname{447}
\expandafter\gdef\csname odunum@val@benchmark_main.ftr.qwen25_omni.ao\endcsname{0.665}
\expandafter\gdef\csname odunum@n@benchmark_main.ftr.qwen25_omni.ao\endcsname{161}
\expandafter\gdef\csname odunum@val@benchmark_main.ftr.qwen25_omni.av\endcsname{0.878}
\expandafter\gdef\csname odunum@n@benchmark_main.ftr.qwen25_omni.av\endcsname{286}
\expandafter\gdef\csname odunum@val@benchmark_main.ftr.qwen3_omni_instruct\endcsname{0.888}
\expandafter\gdef\csname odunum@n@benchmark_main.ftr.qwen3_omni_instruct\endcsname{447}
\expandafter\gdef\csname odunum@val@benchmark_main.ftr.qwen3_omni_instruct.ao\endcsname{0.807}
\expandafter\gdef\csname odunum@n@benchmark_main.ftr.qwen3_omni_instruct.ao\endcsname{161}
\expandafter\gdef\csname odunum@val@benchmark_main.ftr.qwen3_omni_instruct.av\endcsname{0.934}
\expandafter\gdef\csname odunum@n@benchmark_main.ftr.qwen3_omni_instruct.av\endcsname{286}
\expandafter\gdef\csname odunum@val@benchmark_main.ftr.qwen3_omni_think\endcsname{0.749}
\expandafter\gdef\csname odunum@n@benchmark_main.ftr.qwen3_omni_think\endcsname{447}
\expandafter\gdef\csname odunum@val@benchmark_main.ftr.qwen3_omni_think.ao\endcsname{0.609}
\expandafter\gdef\csname odunum@n@benchmark_main.ftr.qwen3_omni_think.ao\endcsname{161}
\expandafter\gdef\csname odunum@val@benchmark_main.ftr.qwen3_omni_think.av\endcsname{0.829}
\expandafter\gdef\csname odunum@n@benchmark_main.ftr.qwen3_omni_think.av\endcsname{286}
\expandafter\gdef\csname odunum@val@benchmark_main.ftr.qwen_plus\endcsname{0.671}
\expandafter\gdef\csname odunum@n@benchmark_main.ftr.qwen_plus\endcsname{447}
\expandafter\gdef\csname odunum@val@benchmark_main.ftr.qwen_plus.ao\endcsname{0.553}
\expandafter\gdef\csname odunum@n@benchmark_main.ftr.qwen_plus.ao\endcsname{161}
\expandafter\gdef\csname odunum@val@benchmark_main.ftr.qwen_plus.av\endcsname{0.738}
\expandafter\gdef\csname odunum@n@benchmark_main.ftr.qwen_plus.av\endcsname{286}
\expandafter\gdef\csname odunum@val@benchmark_main.ftr.salmonn2_7b\endcsname{0.921}
\expandafter\gdef\csname odunum@n@benchmark_main.ftr.salmonn2_7b\endcsname{444}
\expandafter\gdef\csname odunum@val@benchmark_main.ftr.salmonn2_7b.ao\endcsname{0.975}
\expandafter\gdef\csname odunum@n@benchmark_main.ftr.salmonn2_7b.ao\endcsname{160}
\expandafter\gdef\csname odunum@val@benchmark_main.ftr.salmonn2_7b.av\endcsname{0.891}
\expandafter\gdef\csname odunum@n@benchmark_main.ftr.salmonn2_7b.av\endcsname{284}
\expandafter\gdef\csname odunum@val@benchmark_main.ftr.seed\endcsname{0.817}
\expandafter\gdef\csname odunum@n@benchmark_main.ftr.seed\endcsname{447}
\expandafter\gdef\csname odunum@val@benchmark_main.ftr.seed.ao\endcsname{0.807}
\expandafter\gdef\csname odunum@n@benchmark_main.ftr.seed.ao\endcsname{161}
\expandafter\gdef\csname odunum@val@benchmark_main.ftr.seed.av\endcsname{0.822}
\expandafter\gdef\csname odunum@n@benchmark_main.ftr.seed.av\endcsname{286}
\expandafter\gdef\csname odunum@val@benchmark_main.incomplete.cascade_asr\endcsname{0}
\expandafter\gdef\csname odunum@n@benchmark_main.incomplete.cascade_asr\endcsname{2\,078}
\expandafter\gdef\csname odunum@val@benchmark_main.incomplete.gemini\endcsname{0}
\expandafter\gdef\csname odunum@n@benchmark_main.incomplete.gemini\endcsname{2\,078}
\expandafter\gdef\csname odunum@val@benchmark_main.incomplete.gemini35_flash_lite\endcsname{0}
\expandafter\gdef\csname odunum@n@benchmark_main.incomplete.gemini35_flash_lite\endcsname{2\,078}
\expandafter\gdef\csname odunum@val@benchmark_main.incomplete.gemini37_flash\endcsname{0}
\expandafter\gdef\csname odunum@n@benchmark_main.incomplete.gemini37_flash\endcsname{2\,078}
\expandafter\gdef\csname odunum@val@benchmark_main.incomplete.gpt_realtime\endcsname{0}
\expandafter\gdef\csname odunum@n@benchmark_main.incomplete.gpt_realtime\endcsname{731}
\expandafter\gdef\csname odunum@val@benchmark_main.incomplete.kimi_audio_7b_instruct\endcsname{0}
\expandafter\gdef\csname odunum@n@benchmark_main.incomplete.kimi_audio_7b_instruct\endcsname{731}
\expandafter\gdef\csname odunum@val@benchmark_main.incomplete.ming\endcsname{0}
\expandafter\gdef\csname odunum@n@benchmark_main.incomplete.ming\endcsname{2\,078}
\expandafter\gdef\csname odunum@val@benchmark_main.incomplete.minicpm_o\endcsname{0}
\expandafter\gdef\csname odunum@n@benchmark_main.incomplete.minicpm_o\endcsname{2\,078}
\expandafter\gdef\csname odunum@val@benchmark_main.incomplete.nemotron\endcsname{0}
\expandafter\gdef\csname odunum@n@benchmark_main.incomplete.nemotron\endcsname{2\,078}
\expandafter\gdef\csname odunum@val@benchmark_main.incomplete.qwen25_omni\endcsname{0}
\expandafter\gdef\csname odunum@n@benchmark_main.incomplete.qwen25_omni\endcsname{2\,078}
\expandafter\gdef\csname odunum@val@benchmark_main.incomplete.qwen3_omni_instruct\endcsname{0}
\expandafter\gdef\csname odunum@n@benchmark_main.incomplete.qwen3_omni_instruct\endcsname{2\,078}
\expandafter\gdef\csname odunum@val@benchmark_main.incomplete.qwen3_omni_think\endcsname{0}
\expandafter\gdef\csname odunum@n@benchmark_main.incomplete.qwen3_omni_think\endcsname{2\,078}
\expandafter\gdef\csname odunum@val@benchmark_main.incomplete.qwen_plus\endcsname{0}
\expandafter\gdef\csname odunum@n@benchmark_main.incomplete.qwen_plus\endcsname{2\,078}
\expandafter\gdef\csname odunum@val@benchmark_main.incomplete.salmonn2_7b\endcsname{0}
\expandafter\gdef\csname odunum@n@benchmark_main.incomplete.salmonn2_7b\endcsname{2\,078}
\expandafter\gdef\csname odunum@val@benchmark_main.incomplete.seed\endcsname{0}
\expandafter\gdef\csname odunum@n@benchmark_main.incomplete.seed\endcsname{2\,078}
\expandafter\gdef\csname odunum@val@benchmark_main.judge\endcsname{dashscope.qwen3.6-flash-thinking}
\expandafter\gdef\csname odunum@n@benchmark_main.judge\endcsname{1}
\expandafter\gdef\csname odunum@val@benchmark_main.m1.cascade_asr\endcsname{0.677}
\expandafter\gdef\csname odunum@n@benchmark_main.m1.cascade_asr\endcsname{2\,078}
\expandafter\gdef\csname odunum@val@benchmark_main.m1.cascade_asr.ao\endcsname{0.789}
\expandafter\gdef\csname odunum@n@benchmark_main.m1.cascade_asr.ao\endcsname{731}
\expandafter\gdef\csname odunum@val@benchmark_main.m1.cascade_asr.av\endcsname{0.606}
\expandafter\gdef\csname odunum@n@benchmark_main.m1.cascade_asr.av\endcsname{1\,347}
\expandafter\gdef\csname odunum@val@benchmark_main.m1.gemini\endcsname{0.846}
\expandafter\gdef\csname odunum@n@benchmark_main.m1.gemini\endcsname{2\,078}
\expandafter\gdef\csname odunum@val@benchmark_main.m1.gemini.ao\endcsname{0.873}
\expandafter\gdef\csname odunum@n@benchmark_main.m1.gemini.ao\endcsname{731}
\expandafter\gdef\csname odunum@val@benchmark_main.m1.gemini.av\endcsname{0.830}
\expandafter\gdef\csname odunum@n@benchmark_main.m1.gemini.av\endcsname{1\,347}
\expandafter\gdef\csname odunum@val@benchmark_main.m1.gemini35_flash_lite\endcsname{0.677}
\expandafter\gdef\csname odunum@n@benchmark_main.m1.gemini35_flash_lite\endcsname{2\,078}
\expandafter\gdef\csname odunum@val@benchmark_main.m1.gemini35_flash_lite.ao\endcsname{0.745}
\expandafter\gdef\csname odunum@n@benchmark_main.m1.gemini35_flash_lite.ao\endcsname{731}
\expandafter\gdef\csname odunum@val@benchmark_main.m1.gemini35_flash_lite.av\endcsname{0.639}
\expandafter\gdef\csname odunum@n@benchmark_main.m1.gemini35_flash_lite.av\endcsname{1\,347}
\expandafter\gdef\csname odunum@val@benchmark_main.m1.gemini37_flash\endcsname{0.860}
\expandafter\gdef\csname odunum@n@benchmark_main.m1.gemini37_flash\endcsname{2\,067}
\expandafter\gdef\csname odunum@val@benchmark_main.m1.gemini37_flash.ao\endcsname{0.896}
\expandafter\gdef\csname odunum@n@benchmark_main.m1.gemini37_flash.ao\endcsname{723}
\expandafter\gdef\csname odunum@val@benchmark_main.m1.gemini37_flash.av\endcsname{0.839}
\expandafter\gdef\csname odunum@n@benchmark_main.m1.gemini37_flash.av\endcsname{1\,344}
\expandafter\gdef\csname odunum@val@benchmark_main.m1.gpt_realtime\endcsname{0.710}
\expandafter\gdef\csname odunum@n@benchmark_main.m1.gpt_realtime\endcsname{731}
\expandafter\gdef\csname odunum@val@benchmark_main.m1.gpt_realtime.ao\endcsname{0.710}
\expandafter\gdef\csname odunum@n@benchmark_main.m1.gpt_realtime.ao\endcsname{731}
\expandafter\gdef\csname odunum@val@benchmark_main.m1.kimi_audio_7b_instruct\endcsname{0.584}
\expandafter\gdef\csname odunum@n@benchmark_main.m1.kimi_audio_7b_instruct\endcsname{730}
\expandafter\gdef\csname odunum@val@benchmark_main.m1.kimi_audio_7b_instruct.ao\endcsname{0.584}
\expandafter\gdef\csname odunum@n@benchmark_main.m1.kimi_audio_7b_instruct.ao\endcsname{730}
\expandafter\gdef\csname odunum@val@benchmark_main.m1.ming\endcsname{0.554}
\expandafter\gdef\csname odunum@n@benchmark_main.m1.ming\endcsname{2\,078}
\expandafter\gdef\csname odunum@val@benchmark_main.m1.ming.ao\endcsname{0.630}
\expandafter\gdef\csname odunum@n@benchmark_main.m1.ming.ao\endcsname{731}
\expandafter\gdef\csname odunum@val@benchmark_main.m1.ming.av\endcsname{0.501}
\expandafter\gdef\csname odunum@n@benchmark_main.m1.ming.av\endcsname{1\,347}
\expandafter\gdef\csname odunum@val@benchmark_main.m1.minicpm_o\endcsname{0.582}
\expandafter\gdef\csname odunum@n@benchmark_main.m1.minicpm_o\endcsname{2\,075}
\expandafter\gdef\csname odunum@val@benchmark_main.m1.minicpm_o.ao\endcsname{0.651}
\expandafter\gdef\csname odunum@n@benchmark_main.m1.minicpm_o.ao\endcsname{730}
\expandafter\gdef\csname odunum@val@benchmark_main.m1.minicpm_o.av\endcsname{0.545}
\expandafter\gdef\csname odunum@n@benchmark_main.m1.minicpm_o.av\endcsname{1\,345}
\expandafter\gdef\csname odunum@val@benchmark_main.m1.nemotron\endcsname{0.571}
\expandafter\gdef\csname odunum@n@benchmark_main.m1.nemotron\endcsname{2\,078}
\expandafter\gdef\csname odunum@val@benchmark_main.m1.nemotron.ao\endcsname{0.594}
\expandafter\gdef\csname odunum@n@benchmark_main.m1.nemotron.ao\endcsname{731}
\expandafter\gdef\csname odunum@val@benchmark_main.m1.nemotron.av\endcsname{0.558}
\expandafter\gdef\csname odunum@n@benchmark_main.m1.nemotron.av\endcsname{1\,347}
\expandafter\gdef\csname odunum@val@benchmark_main.m1.qwen25_omni\endcsname{0.591}
\expandafter\gdef\csname odunum@n@benchmark_main.m1.qwen25_omni\endcsname{2\,078}
\expandafter\gdef\csname odunum@val@benchmark_main.m1.qwen25_omni.ao\endcsname{0.671}
\expandafter\gdef\csname odunum@n@benchmark_main.m1.qwen25_omni.ao\endcsname{731}
\expandafter\gdef\csname odunum@val@benchmark_main.m1.qwen25_omni.av\endcsname{0.538}
\expandafter\gdef\csname odunum@n@benchmark_main.m1.qwen25_omni.av\endcsname{1\,347}
\expandafter\gdef\csname odunum@val@benchmark_main.m1.qwen3_omni_instruct\endcsname{0.546}
\expandafter\gdef\csname odunum@n@benchmark_main.m1.qwen3_omni_instruct\endcsname{2\,078}
\expandafter\gdef\csname odunum@val@benchmark_main.m1.qwen3_omni_instruct.ao\endcsname{0.610}
\expandafter\gdef\csname odunum@n@benchmark_main.m1.qwen3_omni_instruct.ao\endcsname{731}
\expandafter\gdef\csname odunum@val@benchmark_main.m1.qwen3_omni_instruct.av\endcsname{0.506}
\expandafter\gdef\csname odunum@n@benchmark_main.m1.qwen3_omni_instruct.av\endcsname{1\,347}
\expandafter\gdef\csname odunum@val@benchmark_main.m1.qwen3_omni_think\endcsname{0.641}
\expandafter\gdef\csname odunum@n@benchmark_main.m1.qwen3_omni_think\endcsname{2\,078}
\expandafter\gdef\csname odunum@val@benchmark_main.m1.qwen3_omni_think.ao\endcsname{0.737}
\expandafter\gdef\csname odunum@n@benchmark_main.m1.qwen3_omni_think.ao\endcsname{731}
\expandafter\gdef\csname odunum@val@benchmark_main.m1.qwen3_omni_think.av\endcsname{0.582}
\expandafter\gdef\csname odunum@n@benchmark_main.m1.qwen3_omni_think.av\endcsname{1\,347}
\expandafter\gdef\csname odunum@val@benchmark_main.m1.qwen_plus\endcsname{0.685}
\expandafter\gdef\csname odunum@n@benchmark_main.m1.qwen_plus\endcsname{2\,077}
\expandafter\gdef\csname odunum@val@benchmark_main.m1.qwen_plus.ao\endcsname{0.762}
\expandafter\gdef\csname odunum@n@benchmark_main.m1.qwen_plus.ao\endcsname{731}
\expandafter\gdef\csname odunum@val@benchmark_main.m1.qwen_plus.av\endcsname{0.639}
\expandafter\gdef\csname odunum@n@benchmark_main.m1.qwen_plus.av\endcsname{1\,346}
\expandafter\gdef\csname odunum@val@benchmark_main.m1.salmonn2_7b\endcsname{0.493}
\expandafter\gdef\csname odunum@n@benchmark_main.m1.salmonn2_7b\endcsname{2\,058}
\expandafter\gdef\csname odunum@val@benchmark_main.m1.salmonn2_7b.ao\endcsname{0.464}
\expandafter\gdef\csname odunum@n@benchmark_main.m1.salmonn2_7b.ao\endcsname{728}
\expandafter\gdef\csname odunum@val@benchmark_main.m1.salmonn2_7b.av\endcsname{0.503}
\expandafter\gdef\csname odunum@n@benchmark_main.m1.salmonn2_7b.av\endcsname{1\,330}
\expandafter\gdef\csname odunum@val@benchmark_main.m1.seed\endcsname{0.591}
\expandafter\gdef\csname odunum@n@benchmark_main.m1.seed\endcsname{2\,078}
\expandafter\gdef\csname odunum@val@benchmark_main.m1.seed.ao\endcsname{0.608}
\expandafter\gdef\csname odunum@n@benchmark_main.m1.seed.ao\endcsname{731}
\expandafter\gdef\csname odunum@val@benchmark_main.m1.seed.av\endcsname{0.581}
\expandafter\gdef\csname odunum@n@benchmark_main.m1.seed.av\endcsname{1\,347}
\expandafter\gdef\csname odunum@val@benchmark_main.m2.cascade_asr\endcsname{0.600}
\expandafter\gdef\csname odunum@n@benchmark_main.m2.cascade_asr\endcsname{2\,078}
\expandafter\gdef\csname odunum@val@benchmark_main.m2.cascade_asr.ao\endcsname{0.683}
\expandafter\gdef\csname odunum@n@benchmark_main.m2.cascade_asr.ao\endcsname{731}
\expandafter\gdef\csname odunum@val@benchmark_main.m2.cascade_asr.av\endcsname{0.555}
\expandafter\gdef\csname odunum@n@benchmark_main.m2.cascade_asr.av\endcsname{1\,347}
\expandafter\gdef\csname odunum@val@benchmark_main.m2.gemini\endcsname{0.657}
\expandafter\gdef\csname odunum@n@benchmark_main.m2.gemini\endcsname{2\,078}
\expandafter\gdef\csname odunum@val@benchmark_main.m2.gemini.ao\endcsname{0.701}
\expandafter\gdef\csname odunum@n@benchmark_main.m2.gemini.ao\endcsname{731}
\expandafter\gdef\csname odunum@val@benchmark_main.m2.gemini.av\endcsname{0.634}
\expandafter\gdef\csname odunum@n@benchmark_main.m2.gemini.av\endcsname{1\,347}
\expandafter\gdef\csname odunum@val@benchmark_main.m2.gemini35_flash_lite\endcsname{0.522}
\expandafter\gdef\csname odunum@n@benchmark_main.m2.gemini35_flash_lite\endcsname{2\,078}
\expandafter\gdef\csname odunum@val@benchmark_main.m2.gemini35_flash_lite.ao\endcsname{0.564}
\expandafter\gdef\csname odunum@n@benchmark_main.m2.gemini35_flash_lite.ao\endcsname{731}
\expandafter\gdef\csname odunum@val@benchmark_main.m2.gemini35_flash_lite.av\endcsname{0.499}
\expandafter\gdef\csname odunum@n@benchmark_main.m2.gemini35_flash_lite.av\endcsname{1\,347}
\expandafter\gdef\csname odunum@val@benchmark_main.m2.gemini37_flash\endcsname{0.622}
\expandafter\gdef\csname odunum@n@benchmark_main.m2.gemini37_flash\endcsname{2\,067}
\expandafter\gdef\csname odunum@val@benchmark_main.m2.gemini37_flash.ao\endcsname{0.661}
\expandafter\gdef\csname odunum@n@benchmark_main.m2.gemini37_flash.ao\endcsname{723}
\expandafter\gdef\csname odunum@val@benchmark_main.m2.gemini37_flash.av\endcsname{0.602}
\expandafter\gdef\csname odunum@n@benchmark_main.m2.gemini37_flash.av\endcsname{1\,344}
\expandafter\gdef\csname odunum@val@benchmark_main.m2.gpt_realtime\endcsname{0.634}
\expandafter\gdef\csname odunum@n@benchmark_main.m2.gpt_realtime\endcsname{731}
\expandafter\gdef\csname odunum@val@benchmark_main.m2.gpt_realtime.ao\endcsname{0.634}
\expandafter\gdef\csname odunum@n@benchmark_main.m2.gpt_realtime.ao\endcsname{731}
\expandafter\gdef\csname odunum@val@benchmark_main.m2.kimi_audio_7b_instruct\endcsname{0.511}
\expandafter\gdef\csname odunum@n@benchmark_main.m2.kimi_audio_7b_instruct\endcsname{730}
\expandafter\gdef\csname odunum@val@benchmark_main.m2.kimi_audio_7b_instruct.ao\endcsname{0.511}
\expandafter\gdef\csname odunum@n@benchmark_main.m2.kimi_audio_7b_instruct.ao\endcsname{730}
\expandafter\gdef\csname odunum@val@benchmark_main.m2.ming\endcsname{0.499}
\expandafter\gdef\csname odunum@n@benchmark_main.m2.ming\endcsname{2\,078}
\expandafter\gdef\csname odunum@val@benchmark_main.m2.ming.ao\endcsname{0.505}
\expandafter\gdef\csname odunum@n@benchmark_main.m2.ming.ao\endcsname{731}
\expandafter\gdef\csname odunum@val@benchmark_main.m2.ming.av\endcsname{0.495}
\expandafter\gdef\csname odunum@n@benchmark_main.m2.ming.av\endcsname{1\,347}
\expandafter\gdef\csname odunum@val@benchmark_main.m2.minicpm_o\endcsname{0.398}
\expandafter\gdef\csname odunum@n@benchmark_main.m2.minicpm_o\endcsname{2\,075}
\expandafter\gdef\csname odunum@val@benchmark_main.m2.minicpm_o.ao\endcsname{0.479}
\expandafter\gdef\csname odunum@n@benchmark_main.m2.minicpm_o.ao\endcsname{730}
\expandafter\gdef\csname odunum@val@benchmark_main.m2.minicpm_o.av\endcsname{0.355}
\expandafter\gdef\csname odunum@n@benchmark_main.m2.minicpm_o.av\endcsname{1\,345}
\expandafter\gdef\csname odunum@val@benchmark_main.m2.nemotron\endcsname{0.292}
\expandafter\gdef\csname odunum@n@benchmark_main.m2.nemotron\endcsname{2\,078}
\expandafter\gdef\csname odunum@val@benchmark_main.m2.nemotron.ao\endcsname{0.338}
\expandafter\gdef\csname odunum@n@benchmark_main.m2.nemotron.ao\endcsname{731}
\expandafter\gdef\csname odunum@val@benchmark_main.m2.nemotron.av\endcsname{0.268}
\expandafter\gdef\csname odunum@n@benchmark_main.m2.nemotron.av\endcsname{1\,347}
\expandafter\gdef\csname odunum@val@benchmark_main.m2.qwen25_omni\endcsname{0.433}
\expandafter\gdef\csname odunum@n@benchmark_main.m2.qwen25_omni\endcsname{2\,078}
\expandafter\gdef\csname odunum@val@benchmark_main.m2.qwen25_omni.ao\endcsname{0.457}
\expandafter\gdef\csname odunum@n@benchmark_main.m2.qwen25_omni.ao\endcsname{731}
\expandafter\gdef\csname odunum@val@benchmark_main.m2.qwen25_omni.av\endcsname{0.421}
\expandafter\gdef\csname odunum@n@benchmark_main.m2.qwen25_omni.av\endcsname{1\,347}
\expandafter\gdef\csname odunum@val@benchmark_main.m2.qwen3_omni_instruct\endcsname{0.558}
\expandafter\gdef\csname odunum@n@benchmark_main.m2.qwen3_omni_instruct\endcsname{2\,078}
\expandafter\gdef\csname odunum@val@benchmark_main.m2.qwen3_omni_instruct.ao\endcsname{0.609}
\expandafter\gdef\csname odunum@n@benchmark_main.m2.qwen3_omni_instruct.ao\endcsname{731}
\expandafter\gdef\csname odunum@val@benchmark_main.m2.qwen3_omni_instruct.av\endcsname{0.531}
\expandafter\gdef\csname odunum@n@benchmark_main.m2.qwen3_omni_instruct.av\endcsname{1\,347}
\expandafter\gdef\csname odunum@val@benchmark_main.m2.qwen3_omni_think\endcsname{0.582}
\expandafter\gdef\csname odunum@n@benchmark_main.m2.qwen3_omni_think\endcsname{2\,078}
\expandafter\gdef\csname odunum@val@benchmark_main.m2.qwen3_omni_think.ao\endcsname{0.630}
\expandafter\gdef\csname odunum@n@benchmark_main.m2.qwen3_omni_think.ao\endcsname{731}
\expandafter\gdef\csname odunum@val@benchmark_main.m2.qwen3_omni_think.av\endcsname{0.556}
\expandafter\gdef\csname odunum@n@benchmark_main.m2.qwen3_omni_think.av\endcsname{1\,347}
\expandafter\gdef\csname odunum@val@benchmark_main.m2.qwen_plus\endcsname{0.678}
\expandafter\gdef\csname odunum@n@benchmark_main.m2.qwen_plus\endcsname{2\,077}
\expandafter\gdef\csname odunum@val@benchmark_main.m2.qwen_plus.ao\endcsname{0.715}
\expandafter\gdef\csname odunum@n@benchmark_main.m2.qwen_plus.ao\endcsname{731}
\expandafter\gdef\csname odunum@val@benchmark_main.m2.qwen_plus.av\endcsname{0.658}
\expandafter\gdef\csname odunum@n@benchmark_main.m2.qwen_plus.av\endcsname{1\,346}
\expandafter\gdef\csname odunum@val@benchmark_main.m2.salmonn2_7b\endcsname{0.108}
\expandafter\gdef\csname odunum@n@benchmark_main.m2.salmonn2_7b\endcsname{2\,058}
\expandafter\gdef\csname odunum@val@benchmark_main.m2.salmonn2_7b.ao\endcsname{0.184}
\expandafter\gdef\csname odunum@n@benchmark_main.m2.salmonn2_7b.ao\endcsname{728}
\expandafter\gdef\csname odunum@val@benchmark_main.m2.salmonn2_7b.av\endcsname{0.067}
\expandafter\gdef\csname odunum@n@benchmark_main.m2.salmonn2_7b.av\endcsname{1\,330}
\expandafter\gdef\csname odunum@val@benchmark_main.m2.seed\endcsname{0.690}
\expandafter\gdef\csname odunum@n@benchmark_main.m2.seed\endcsname{2\,078}
\expandafter\gdef\csname odunum@val@benchmark_main.m2.seed.ao\endcsname{0.780}
\expandafter\gdef\csname odunum@n@benchmark_main.m2.seed.ao\endcsname{731}
\expandafter\gdef\csname odunum@val@benchmark_main.m2.seed.av\endcsname{0.641}
\expandafter\gdef\csname odunum@n@benchmark_main.m2.seed.av\endcsname{1\,347}
\expandafter\gdef\csname odunum@val@benchmark_main.m3.cascade_asr\endcsname{0.837}
\expandafter\gdef\csname odunum@n@benchmark_main.m3.cascade_asr\endcsname{2\,078}
\expandafter\gdef\csname odunum@val@benchmark_main.m3.cascade_asr.ao\endcsname{0.861}
\expandafter\gdef\csname odunum@n@benchmark_main.m3.cascade_asr.ao\endcsname{731}
\expandafter\gdef\csname odunum@val@benchmark_main.m3.cascade_asr.av\endcsname{0.823}
\expandafter\gdef\csname odunum@n@benchmark_main.m3.cascade_asr.av\endcsname{1\,347}
\expandafter\gdef\csname odunum@val@benchmark_main.m3.gemini\endcsname{0.748}
\expandafter\gdef\csname odunum@n@benchmark_main.m3.gemini\endcsname{2\,078}
\expandafter\gdef\csname odunum@val@benchmark_main.m3.gemini.ao\endcsname{0.615}
\expandafter\gdef\csname odunum@n@benchmark_main.m3.gemini.ao\endcsname{731}
\expandafter\gdef\csname odunum@val@benchmark_main.m3.gemini.av\endcsname{0.819}
\expandafter\gdef\csname odunum@n@benchmark_main.m3.gemini.av\endcsname{1\,347}
\expandafter\gdef\csname odunum@val@benchmark_main.m3.gemini35_flash_lite\endcsname{0.641}
\expandafter\gdef\csname odunum@n@benchmark_main.m3.gemini35_flash_lite\endcsname{2\,078}
\expandafter\gdef\csname odunum@val@benchmark_main.m3.gemini35_flash_lite.ao\endcsname{0.567}
\expandafter\gdef\csname odunum@n@benchmark_main.m3.gemini35_flash_lite.ao\endcsname{731}
\expandafter\gdef\csname odunum@val@benchmark_main.m3.gemini35_flash_lite.av\endcsname{0.681}
\expandafter\gdef\csname odunum@n@benchmark_main.m3.gemini35_flash_lite.av\endcsname{1\,347}
\expandafter\gdef\csname odunum@val@benchmark_main.m3.gemini37_flash\endcsname{0.732}
\expandafter\gdef\csname odunum@n@benchmark_main.m3.gemini37_flash\endcsname{2\,067}
\expandafter\gdef\csname odunum@val@benchmark_main.m3.gemini37_flash.ao\endcsname{0.627}
\expandafter\gdef\csname odunum@n@benchmark_main.m3.gemini37_flash.ao\endcsname{723}
\expandafter\gdef\csname odunum@val@benchmark_main.m3.gemini37_flash.av\endcsname{0.787}
\expandafter\gdef\csname odunum@n@benchmark_main.m3.gemini37_flash.av\endcsname{1\,344}
\expandafter\gdef\csname odunum@val@benchmark_main.m3.gpt_realtime\endcsname{0.343}
\expandafter\gdef\csname odunum@n@benchmark_main.m3.gpt_realtime\endcsname{731}
\expandafter\gdef\csname odunum@val@benchmark_main.m3.gpt_realtime.ao\endcsname{0.343}
\expandafter\gdef\csname odunum@n@benchmark_main.m3.gpt_realtime.ao\endcsname{731}
\expandafter\gdef\csname odunum@val@benchmark_main.m3.kimi_audio_7b_instruct\endcsname{0.172}
\expandafter\gdef\csname odunum@n@benchmark_main.m3.kimi_audio_7b_instruct\endcsname{730}
\expandafter\gdef\csname odunum@val@benchmark_main.m3.kimi_audio_7b_instruct.ao\endcsname{0.172}
\expandafter\gdef\csname odunum@n@benchmark_main.m3.kimi_audio_7b_instruct.ao\endcsname{730}
\expandafter\gdef\csname odunum@val@benchmark_main.m3.ming\endcsname{0.218}
\expandafter\gdef\csname odunum@n@benchmark_main.m3.ming\endcsname{2\,078}
\expandafter\gdef\csname odunum@val@benchmark_main.m3.ming.ao\endcsname{0.073}
\expandafter\gdef\csname odunum@n@benchmark_main.m3.ming.ao\endcsname{731}
\expandafter\gdef\csname odunum@val@benchmark_main.m3.ming.av\endcsname{0.296}
\expandafter\gdef\csname odunum@n@benchmark_main.m3.ming.av\endcsname{1\,347}
\expandafter\gdef\csname odunum@val@benchmark_main.m3.minicpm_o\endcsname{0.033}
\expandafter\gdef\csname odunum@n@benchmark_main.m3.minicpm_o\endcsname{2\,075}
\expandafter\gdef\csname odunum@val@benchmark_main.m3.minicpm_o.ao\endcsname{0.062}
\expandafter\gdef\csname odunum@n@benchmark_main.m3.minicpm_o.ao\endcsname{730}
\expandafter\gdef\csname odunum@val@benchmark_main.m3.minicpm_o.av\endcsname{0.017}
\expandafter\gdef\csname odunum@n@benchmark_main.m3.minicpm_o.av\endcsname{1\,345}
\expandafter\gdef\csname odunum@val@benchmark_main.m3.nemotron\endcsname{0.371}
\expandafter\gdef\csname odunum@n@benchmark_main.m3.nemotron\endcsname{2\,078}
\expandafter\gdef\csname odunum@val@benchmark_main.m3.nemotron.ao\endcsname{0.393}
\expandafter\gdef\csname odunum@n@benchmark_main.m3.nemotron.ao\endcsname{731}
\expandafter\gdef\csname odunum@val@benchmark_main.m3.nemotron.av\endcsname{0.359}
\expandafter\gdef\csname odunum@n@benchmark_main.m3.nemotron.av\endcsname{1\,347}
\expandafter\gdef\csname odunum@val@benchmark_main.m3.qwen25_omni\endcsname{0.057}
\expandafter\gdef\csname odunum@n@benchmark_main.m3.qwen25_omni\endcsname{2\,078}
\expandafter\gdef\csname odunum@val@benchmark_main.m3.qwen25_omni.ao\endcsname{0.055}
\expandafter\gdef\csname odunum@n@benchmark_main.m3.qwen25_omni.ao\endcsname{731}
\expandafter\gdef\csname odunum@val@benchmark_main.m3.qwen25_omni.av\endcsname{0.058}
\expandafter\gdef\csname odunum@n@benchmark_main.m3.qwen25_omni.av\endcsname{1\,347}
\expandafter\gdef\csname odunum@val@benchmark_main.m3.qwen3_omni_instruct\endcsname{0.227}
\expandafter\gdef\csname odunum@n@benchmark_main.m3.qwen3_omni_instruct\endcsname{2\,078}
\expandafter\gdef\csname odunum@val@benchmark_main.m3.qwen3_omni_instruct.ao\endcsname{0.205}
\expandafter\gdef\csname odunum@n@benchmark_main.m3.qwen3_omni_instruct.ao\endcsname{731}
\expandafter\gdef\csname odunum@val@benchmark_main.m3.qwen3_omni_instruct.av\endcsname{0.239}
\expandafter\gdef\csname odunum@n@benchmark_main.m3.qwen3_omni_instruct.av\endcsname{1\,347}
\expandafter\gdef\csname odunum@val@benchmark_main.m3.qwen3_omni_think\endcsname{0.482}
\expandafter\gdef\csname odunum@n@benchmark_main.m3.qwen3_omni_think\endcsname{2\,078}
\expandafter\gdef\csname odunum@val@benchmark_main.m3.qwen3_omni_think.ao\endcsname{0.505}
\expandafter\gdef\csname odunum@n@benchmark_main.m3.qwen3_omni_think.ao\endcsname{731}
\expandafter\gdef\csname odunum@val@benchmark_main.m3.qwen3_omni_think.av\endcsname{0.470}
\expandafter\gdef\csname odunum@n@benchmark_main.m3.qwen3_omni_think.av\endcsname{1\,347}
\expandafter\gdef\csname odunum@val@benchmark_main.m3.qwen_plus\endcsname{0.724}
\expandafter\gdef\csname odunum@n@benchmark_main.m3.qwen_plus\endcsname{2\,077}
\expandafter\gdef\csname odunum@val@benchmark_main.m3.qwen_plus.ao\endcsname{0.661}
\expandafter\gdef\csname odunum@n@benchmark_main.m3.qwen_plus.ao\endcsname{731}
\expandafter\gdef\csname odunum@val@benchmark_main.m3.qwen_plus.av\endcsname{0.757}
\expandafter\gdef\csname odunum@n@benchmark_main.m3.qwen_plus.av\endcsname{1\,346}
\expandafter\gdef\csname odunum@val@benchmark_main.m3.salmonn2_7b\endcsname{0.054}
\expandafter\gdef\csname odunum@n@benchmark_main.m3.salmonn2_7b\endcsname{2\,058}
\expandafter\gdef\csname odunum@val@benchmark_main.m3.salmonn2_7b.ao\endcsname{0.100}
\expandafter\gdef\csname odunum@n@benchmark_main.m3.salmonn2_7b.ao\endcsname{728}
\expandafter\gdef\csname odunum@val@benchmark_main.m3.salmonn2_7b.av\endcsname{0.029}
\expandafter\gdef\csname odunum@n@benchmark_main.m3.salmonn2_7b.av\endcsname{1\,330}
\expandafter\gdef\csname odunum@val@benchmark_main.m3.seed\endcsname{0.788}
\expandafter\gdef\csname odunum@n@benchmark_main.m3.seed\endcsname{2\,078}
\expandafter\gdef\csname odunum@val@benchmark_main.m3.seed.ao\endcsname{0.714}
\expandafter\gdef\csname odunum@n@benchmark_main.m3.seed.ao\endcsname{731}
\expandafter\gdef\csname odunum@val@benchmark_main.m3.seed.av\endcsname{0.828}
\expandafter\gdef\csname odunum@n@benchmark_main.m3.seed.av\endcsname{1\,347}
\expandafter\gdef\csname odunum@val@benchmark_main.m4.cascade_asr\endcsname{0.830}
\expandafter\gdef\csname odunum@n@benchmark_main.m4.cascade_asr\endcsname{2\,078}
\expandafter\gdef\csname odunum@val@benchmark_main.m4.cascade_asr.ao\endcsname{0.851}
\expandafter\gdef\csname odunum@n@benchmark_main.m4.cascade_asr.ao\endcsname{731}
\expandafter\gdef\csname odunum@val@benchmark_main.m4.cascade_asr.av\endcsname{0.819}
\expandafter\gdef\csname odunum@n@benchmark_main.m4.cascade_asr.av\endcsname{1\,347}
\expandafter\gdef\csname odunum@val@benchmark_main.m4.gemini\endcsname{0.915}
\expandafter\gdef\csname odunum@n@benchmark_main.m4.gemini\endcsname{2\,078}
\expandafter\gdef\csname odunum@val@benchmark_main.m4.gemini.ao\endcsname{0.917}
\expandafter\gdef\csname odunum@n@benchmark_main.m4.gemini.ao\endcsname{731}
\expandafter\gdef\csname odunum@val@benchmark_main.m4.gemini.av\endcsname{0.913}
\expandafter\gdef\csname odunum@n@benchmark_main.m4.gemini.av\endcsname{1\,347}
\expandafter\gdef\csname odunum@val@benchmark_main.m4.gemini35_flash_lite\endcsname{0.725}
\expandafter\gdef\csname odunum@n@benchmark_main.m4.gemini35_flash_lite\endcsname{2\,078}
\expandafter\gdef\csname odunum@val@benchmark_main.m4.gemini35_flash_lite.ao\endcsname{0.743}
\expandafter\gdef\csname odunum@n@benchmark_main.m4.gemini35_flash_lite.ao\endcsname{731}
\expandafter\gdef\csname odunum@val@benchmark_main.m4.gemini35_flash_lite.av\endcsname{0.715}
\expandafter\gdef\csname odunum@n@benchmark_main.m4.gemini35_flash_lite.av\endcsname{1\,347}
\expandafter\gdef\csname odunum@val@benchmark_main.m4.gemini37_flash\endcsname{0.868}
\expandafter\gdef\csname odunum@n@benchmark_main.m4.gemini37_flash\endcsname{2\,067}
\expandafter\gdef\csname odunum@val@benchmark_main.m4.gemini37_flash.ao\endcsname{0.884}
\expandafter\gdef\csname odunum@n@benchmark_main.m4.gemini37_flash.ao\endcsname{723}
\expandafter\gdef\csname odunum@val@benchmark_main.m4.gemini37_flash.av\endcsname{0.859}
\expandafter\gdef\csname odunum@n@benchmark_main.m4.gemini37_flash.av\endcsname{1\,344}
\expandafter\gdef\csname odunum@val@benchmark_main.m4.gpt_realtime\endcsname{0.814}
\expandafter\gdef\csname odunum@n@benchmark_main.m4.gpt_realtime\endcsname{731}
\expandafter\gdef\csname odunum@val@benchmark_main.m4.gpt_realtime.ao\endcsname{0.814}
\expandafter\gdef\csname odunum@n@benchmark_main.m4.gpt_realtime.ao\endcsname{731}
\expandafter\gdef\csname odunum@val@benchmark_main.m4.kimi_audio_7b_instruct\endcsname{0.718}
\expandafter\gdef\csname odunum@n@benchmark_main.m4.kimi_audio_7b_instruct\endcsname{730}
\expandafter\gdef\csname odunum@val@benchmark_main.m4.kimi_audio_7b_instruct.ao\endcsname{0.718}
\expandafter\gdef\csname odunum@n@benchmark_main.m4.kimi_audio_7b_instruct.ao\endcsname{730}
\expandafter\gdef\csname odunum@val@benchmark_main.m4.ming\endcsname{0.841}
\expandafter\gdef\csname odunum@n@benchmark_main.m4.ming\endcsname{2\,078}
\expandafter\gdef\csname odunum@val@benchmark_main.m4.ming.ao\endcsname{0.861}
\expandafter\gdef\csname odunum@n@benchmark_main.m4.ming.ao\endcsname{731}
\expandafter\gdef\csname odunum@val@benchmark_main.m4.ming.av\endcsname{0.831}
\expandafter\gdef\csname odunum@n@benchmark_main.m4.ming.av\endcsname{1\,347}
\expandafter\gdef\csname odunum@val@benchmark_main.m4.minicpm_o\endcsname{0.607}
\expandafter\gdef\csname odunum@n@benchmark_main.m4.minicpm_o\endcsname{2\,075}
\expandafter\gdef\csname odunum@val@benchmark_main.m4.minicpm_o.ao\endcsname{0.615}
\expandafter\gdef\csname odunum@n@benchmark_main.m4.minicpm_o.ao\endcsname{730}
\expandafter\gdef\csname odunum@val@benchmark_main.m4.minicpm_o.av\endcsname{0.603}
\expandafter\gdef\csname odunum@n@benchmark_main.m4.minicpm_o.av\endcsname{1\,345}
\expandafter\gdef\csname odunum@val@benchmark_main.m4.nemotron\endcsname{0.354}
\expandafter\gdef\csname odunum@n@benchmark_main.m4.nemotron\endcsname{2\,078}
\expandafter\gdef\csname odunum@val@benchmark_main.m4.nemotron.ao\endcsname{0.376}
\expandafter\gdef\csname odunum@n@benchmark_main.m4.nemotron.ao\endcsname{731}
\expandafter\gdef\csname odunum@val@benchmark_main.m4.nemotron.av\endcsname{0.343}
\expandafter\gdef\csname odunum@n@benchmark_main.m4.nemotron.av\endcsname{1\,347}
\expandafter\gdef\csname odunum@val@benchmark_main.m4.qwen25_omni\endcsname{0.725}
\expandafter\gdef\csname odunum@n@benchmark_main.m4.qwen25_omni\endcsname{2\,078}
\expandafter\gdef\csname odunum@val@benchmark_main.m4.qwen25_omni.ao\endcsname{0.694}
\expandafter\gdef\csname odunum@n@benchmark_main.m4.qwen25_omni.ao\endcsname{731}
\expandafter\gdef\csname odunum@val@benchmark_main.m4.qwen25_omni.av\endcsname{0.742}
\expandafter\gdef\csname odunum@n@benchmark_main.m4.qwen25_omni.av\endcsname{1\,347}
\expandafter\gdef\csname odunum@val@benchmark_main.m4.qwen3_omni_instruct\endcsname{0.885}
\expandafter\gdef\csname odunum@n@benchmark_main.m4.qwen3_omni_instruct\endcsname{2\,078}
\expandafter\gdef\csname odunum@val@benchmark_main.m4.qwen3_omni_instruct.ao\endcsname{0.910}
\expandafter\gdef\csname odunum@n@benchmark_main.m4.qwen3_omni_instruct.ao\endcsname{731}
\expandafter\gdef\csname odunum@val@benchmark_main.m4.qwen3_omni_instruct.av\endcsname{0.871}
\expandafter\gdef\csname odunum@n@benchmark_main.m4.qwen3_omni_instruct.av\endcsname{1\,347}
\expandafter\gdef\csname odunum@val@benchmark_main.m4.qwen3_omni_think\endcsname{0.862}
\expandafter\gdef\csname odunum@n@benchmark_main.m4.qwen3_omni_think\endcsname{2\,078}
\expandafter\gdef\csname odunum@val@benchmark_main.m4.qwen3_omni_think.ao\endcsname{0.881}
\expandafter\gdef\csname odunum@n@benchmark_main.m4.qwen3_omni_think.ao\endcsname{731}
\expandafter\gdef\csname odunum@val@benchmark_main.m4.qwen3_omni_think.av\endcsname{0.852}
\expandafter\gdef\csname odunum@n@benchmark_main.m4.qwen3_omni_think.av\endcsname{1\,347}
\expandafter\gdef\csname odunum@val@benchmark_main.m4.qwen_plus\endcsname{0.854}
\expandafter\gdef\csname odunum@n@benchmark_main.m4.qwen_plus\endcsname{2\,077}
\expandafter\gdef\csname odunum@val@benchmark_main.m4.qwen_plus.ao\endcsname{0.893}
\expandafter\gdef\csname odunum@n@benchmark_main.m4.qwen_plus.ao\endcsname{731}
\expandafter\gdef\csname odunum@val@benchmark_main.m4.qwen_plus.av\endcsname{0.833}
\expandafter\gdef\csname odunum@n@benchmark_main.m4.qwen_plus.av\endcsname{1\,346}
\expandafter\gdef\csname odunum@val@benchmark_main.m4.salmonn2_7b\endcsname{0.172}
\expandafter\gdef\csname odunum@n@benchmark_main.m4.salmonn2_7b\endcsname{2\,058}
\expandafter\gdef\csname odunum@val@benchmark_main.m4.salmonn2_7b.ao\endcsname{0.263}
\expandafter\gdef\csname odunum@n@benchmark_main.m4.salmonn2_7b.ao\endcsname{728}
\expandafter\gdef\csname odunum@val@benchmark_main.m4.salmonn2_7b.av\endcsname{0.123}
\expandafter\gdef\csname odunum@n@benchmark_main.m4.salmonn2_7b.av\endcsname{1\,330}
\expandafter\gdef\csname odunum@val@benchmark_main.m4.seed\endcsname{0.831}
\expandafter\gdef\csname odunum@n@benchmark_main.m4.seed\endcsname{2\,078}
\expandafter\gdef\csname odunum@val@benchmark_main.m4.seed.ao\endcsname{0.932}
\expandafter\gdef\csname odunum@n@benchmark_main.m4.seed.ao\endcsname{731}
\expandafter\gdef\csname odunum@val@benchmark_main.m4.seed.av\endcsname{0.777}
\expandafter\gdef\csname odunum@n@benchmark_main.m4.seed.av\endcsname{1\,347}
\expandafter\gdef\csname odunum@val@benchmark_main.m5.cascade_asr\endcsname{0.124}
\expandafter\gdef\csname odunum@n@benchmark_main.m5.cascade_asr\endcsname{2\,078}
\expandafter\gdef\csname odunum@val@benchmark_main.m5.cascade_asr.ao\endcsname{0.306}
\expandafter\gdef\csname odunum@n@benchmark_main.m5.cascade_asr.ao\endcsname{731}
\expandafter\gdef\csname odunum@val@benchmark_main.m5.cascade_asr.av\endcsname{0.026}
\expandafter\gdef\csname odunum@n@benchmark_main.m5.cascade_asr.av\endcsname{1\,347}
\expandafter\gdef\csname odunum@val@benchmark_main.m5.gemini\endcsname{0.964}
\expandafter\gdef\csname odunum@n@benchmark_main.m5.gemini\endcsname{2\,078}
\expandafter\gdef\csname odunum@val@benchmark_main.m5.gemini.ao\endcsname{0.976}
\expandafter\gdef\csname odunum@n@benchmark_main.m5.gemini.ao\endcsname{731}
\expandafter\gdef\csname odunum@val@benchmark_main.m5.gemini.av\endcsname{0.957}
\expandafter\gdef\csname odunum@n@benchmark_main.m5.gemini.av\endcsname{1\,347}
\expandafter\gdef\csname odunum@val@benchmark_main.m5.gemini35_flash_lite\endcsname{0.804}
\expandafter\gdef\csname odunum@n@benchmark_main.m5.gemini35_flash_lite\endcsname{2\,078}
\expandafter\gdef\csname odunum@val@benchmark_main.m5.gemini35_flash_lite.ao\endcsname{0.824}
\expandafter\gdef\csname odunum@n@benchmark_main.m5.gemini35_flash_lite.ao\endcsname{731}
\expandafter\gdef\csname odunum@val@benchmark_main.m5.gemini35_flash_lite.av\endcsname{0.793}
\expandafter\gdef\csname odunum@n@benchmark_main.m5.gemini35_flash_lite.av\endcsname{1\,347}
\expandafter\gdef\csname odunum@val@benchmark_main.m5.gemini37_flash\endcsname{0.903}
\expandafter\gdef\csname odunum@n@benchmark_main.m5.gemini37_flash\endcsname{2\,067}
\expandafter\gdef\csname odunum@val@benchmark_main.m5.gemini37_flash.ao\endcsname{0.941}
\expandafter\gdef\csname odunum@n@benchmark_main.m5.gemini37_flash.ao\endcsname{723}
\expandafter\gdef\csname odunum@val@benchmark_main.m5.gemini37_flash.av\endcsname{0.883}
\expandafter\gdef\csname odunum@n@benchmark_main.m5.gemini37_flash.av\endcsname{1\,344}
\expandafter\gdef\csname odunum@val@benchmark_main.m5.gpt_realtime\endcsname{0.712}
\expandafter\gdef\csname odunum@n@benchmark_main.m5.gpt_realtime\endcsname{731}
\expandafter\gdef\csname odunum@val@benchmark_main.m5.gpt_realtime.ao\endcsname{0.712}
\expandafter\gdef\csname odunum@n@benchmark_main.m5.gpt_realtime.ao\endcsname{731}
\expandafter\gdef\csname odunum@val@benchmark_main.m5.kimi_audio_7b_instruct\endcsname{0.850}
\expandafter\gdef\csname odunum@n@benchmark_main.m5.kimi_audio_7b_instruct\endcsname{730}
\expandafter\gdef\csname odunum@val@benchmark_main.m5.kimi_audio_7b_instruct.ao\endcsname{0.850}
\expandafter\gdef\csname odunum@n@benchmark_main.m5.kimi_audio_7b_instruct.ao\endcsname{730}
\expandafter\gdef\csname odunum@val@benchmark_main.m5.ming\endcsname{0.779}
\expandafter\gdef\csname odunum@n@benchmark_main.m5.ming\endcsname{2\,078}
\expandafter\gdef\csname odunum@val@benchmark_main.m5.ming.ao\endcsname{0.696}
\expandafter\gdef\csname odunum@n@benchmark_main.m5.ming.ao\endcsname{731}
\expandafter\gdef\csname odunum@val@benchmark_main.m5.ming.av\endcsname{0.823}
\expandafter\gdef\csname odunum@n@benchmark_main.m5.ming.av\endcsname{1\,347}
\expandafter\gdef\csname odunum@val@benchmark_main.m5.minicpm_o\endcsname{0.708}
\expandafter\gdef\csname odunum@n@benchmark_main.m5.minicpm_o\endcsname{2\,075}
\expandafter\gdef\csname odunum@val@benchmark_main.m5.minicpm_o.ao\endcsname{0.759}
\expandafter\gdef\csname odunum@n@benchmark_main.m5.minicpm_o.ao\endcsname{730}
\expandafter\gdef\csname odunum@val@benchmark_main.m5.minicpm_o.av\endcsname{0.681}
\expandafter\gdef\csname odunum@n@benchmark_main.m5.minicpm_o.av\endcsname{1\,345}
\expandafter\gdef\csname odunum@val@benchmark_main.m5.nemotron\endcsname{0.677}
\expandafter\gdef\csname odunum@n@benchmark_main.m5.nemotron\endcsname{2\,078}
\expandafter\gdef\csname odunum@val@benchmark_main.m5.nemotron.ao\endcsname{0.641}
\expandafter\gdef\csname odunum@n@benchmark_main.m5.nemotron.ao\endcsname{731}
\expandafter\gdef\csname odunum@val@benchmark_main.m5.nemotron.av\endcsname{0.696}
\expandafter\gdef\csname odunum@n@benchmark_main.m5.nemotron.av\endcsname{1\,347}
\expandafter\gdef\csname odunum@val@benchmark_main.m5.qwen25_omni\endcsname{0.721}
\expandafter\gdef\csname odunum@n@benchmark_main.m5.qwen25_omni\endcsname{2\,078}
\expandafter\gdef\csname odunum@val@benchmark_main.m5.qwen25_omni.ao\endcsname{0.607}
\expandafter\gdef\csname odunum@n@benchmark_main.m5.qwen25_omni.ao\endcsname{731}
\expandafter\gdef\csname odunum@val@benchmark_main.m5.qwen25_omni.av\endcsname{0.782}
\expandafter\gdef\csname odunum@n@benchmark_main.m5.qwen25_omni.av\endcsname{1\,347}
\expandafter\gdef\csname odunum@val@benchmark_main.m5.qwen3_omni_instruct\endcsname{0.936}
\expandafter\gdef\csname odunum@n@benchmark_main.m5.qwen3_omni_instruct\endcsname{2\,078}
\expandafter\gdef\csname odunum@val@benchmark_main.m5.qwen3_omni_instruct.ao\endcsname{0.952}
\expandafter\gdef\csname odunum@n@benchmark_main.m5.qwen3_omni_instruct.ao\endcsname{731}
\expandafter\gdef\csname odunum@val@benchmark_main.m5.qwen3_omni_instruct.av\endcsname{0.928}
\expandafter\gdef\csname odunum@n@benchmark_main.m5.qwen3_omni_instruct.av\endcsname{1\,347}
\expandafter\gdef\csname odunum@val@benchmark_main.m5.qwen3_omni_think\endcsname{0.929}
\expandafter\gdef\csname odunum@n@benchmark_main.m5.qwen3_omni_think\endcsname{2\,078}
\expandafter\gdef\csname odunum@val@benchmark_main.m5.qwen3_omni_think.ao\endcsname{0.946}
\expandafter\gdef\csname odunum@n@benchmark_main.m5.qwen3_omni_think.ao\endcsname{731}
\expandafter\gdef\csname odunum@val@benchmark_main.m5.qwen3_omni_think.av\endcsname{0.921}
\expandafter\gdef\csname odunum@n@benchmark_main.m5.qwen3_omni_think.av\endcsname{1\,347}
\expandafter\gdef\csname odunum@val@benchmark_main.m5.qwen_plus\endcsname{0.943}
\expandafter\gdef\csname odunum@n@benchmark_main.m5.qwen_plus\endcsname{2\,077}
\expandafter\gdef\csname odunum@val@benchmark_main.m5.qwen_plus.ao\endcsname{0.958}
\expandafter\gdef\csname odunum@n@benchmark_main.m5.qwen_plus.ao\endcsname{731}
\expandafter\gdef\csname odunum@val@benchmark_main.m5.qwen_plus.av\endcsname{0.935}
\expandafter\gdef\csname odunum@n@benchmark_main.m5.qwen_plus.av\endcsname{1\,346}
\expandafter\gdef\csname odunum@val@benchmark_main.m5.salmonn2_7b\endcsname{0.302}
\expandafter\gdef\csname odunum@n@benchmark_main.m5.salmonn2_7b\endcsname{2\,058}
\expandafter\gdef\csname odunum@val@benchmark_main.m5.salmonn2_7b.ao\endcsname{0.426}
\expandafter\gdef\csname odunum@n@benchmark_main.m5.salmonn2_7b.ao\endcsname{728}
\expandafter\gdef\csname odunum@val@benchmark_main.m5.salmonn2_7b.av\endcsname{0.234}
\expandafter\gdef\csname odunum@n@benchmark_main.m5.salmonn2_7b.av\endcsname{1\,330}
\expandafter\gdef\csname odunum@val@benchmark_main.m5.seed\endcsname{0.942}
\expandafter\gdef\csname odunum@n@benchmark_main.m5.seed\endcsname{2\,078}
\expandafter\gdef\csname odunum@val@benchmark_main.m5.seed.ao\endcsname{0.975}
\expandafter\gdef\csname odunum@n@benchmark_main.m5.seed.ao\endcsname{731}
\expandafter\gdef\csname odunum@val@benchmark_main.m5.seed.av\endcsname{0.925}
\expandafter\gdef\csname odunum@n@benchmark_main.m5.seed.av\endcsname{1\,347}
\expandafter\gdef\csname odunum@val@benchmark_main.modalities.cascade_asr\endcsname{audio-visual, audio-only}
\expandafter\gdef\csname odunum@n@benchmark_main.modalities.cascade_asr\endcsname{2}
\expandafter\gdef\csname odunum@val@benchmark_main.modalities.gemini\endcsname{audio-visual, audio-only}
\expandafter\gdef\csname odunum@n@benchmark_main.modalities.gemini\endcsname{2}
\expandafter\gdef\csname odunum@val@benchmark_main.modalities.gemini35_flash_lite\endcsname{audio-visual, audio-only}
\expandafter\gdef\csname odunum@n@benchmark_main.modalities.gemini35_flash_lite\endcsname{2}
\expandafter\gdef\csname odunum@val@benchmark_main.modalities.gemini37_flash\endcsname{audio-visual, audio-only}
\expandafter\gdef\csname odunum@n@benchmark_main.modalities.gemini37_flash\endcsname{2}
\expandafter\gdef\csname odunum@val@benchmark_main.modalities.gpt_realtime\endcsname{audio-only}
\expandafter\gdef\csname odunum@n@benchmark_main.modalities.gpt_realtime\endcsname{1}
\expandafter\gdef\csname odunum@val@benchmark_main.modalities.kimi_audio_7b_instruct\endcsname{audio-only}
\expandafter\gdef\csname odunum@n@benchmark_main.modalities.kimi_audio_7b_instruct\endcsname{1}
\expandafter\gdef\csname odunum@val@benchmark_main.modalities.ming\endcsname{audio-visual, audio-only}
\expandafter\gdef\csname odunum@n@benchmark_main.modalities.ming\endcsname{2}
\expandafter\gdef\csname odunum@val@benchmark_main.modalities.minicpm_o\endcsname{audio-visual, audio-only}
\expandafter\gdef\csname odunum@n@benchmark_main.modalities.minicpm_o\endcsname{2}
\expandafter\gdef\csname odunum@val@benchmark_main.modalities.nemotron\endcsname{audio-visual, audio-only}
\expandafter\gdef\csname odunum@n@benchmark_main.modalities.nemotron\endcsname{2}
\expandafter\gdef\csname odunum@val@benchmark_main.modalities.qwen25_omni\endcsname{audio-visual, audio-only}
\expandafter\gdef\csname odunum@n@benchmark_main.modalities.qwen25_omni\endcsname{2}
\expandafter\gdef\csname odunum@val@benchmark_main.modalities.qwen3_omni_instruct\endcsname{audio-visual, audio-only}
\expandafter\gdef\csname odunum@n@benchmark_main.modalities.qwen3_omni_instruct\endcsname{2}
\expandafter\gdef\csname odunum@val@benchmark_main.modalities.qwen3_omni_think\endcsname{audio-visual, audio-only}
\expandafter\gdef\csname odunum@n@benchmark_main.modalities.qwen3_omni_think\endcsname{2}
\expandafter\gdef\csname odunum@val@benchmark_main.modalities.qwen_plus\endcsname{audio-visual, audio-only}
\expandafter\gdef\csname odunum@n@benchmark_main.modalities.qwen_plus\endcsname{2}
\expandafter\gdef\csname odunum@val@benchmark_main.modalities.salmonn2_7b\endcsname{audio-visual, audio-only}
\expandafter\gdef\csname odunum@n@benchmark_main.modalities.salmonn2_7b\endcsname{2}
\expandafter\gdef\csname odunum@val@benchmark_main.modalities.seed\endcsname{audio-visual, audio-only}
\expandafter\gdef\csname odunum@n@benchmark_main.modalities.seed\endcsname{2}
\expandafter\gdef\csname odunum@val@benchmark_main.n.cascade_asr\endcsname{2\,078}
\expandafter\gdef\csname odunum@n@benchmark_main.n.cascade_asr\endcsname{2\,078}
\expandafter\gdef\csname odunum@val@benchmark_main.n.cascade_asr.ao\endcsname{731}
\expandafter\gdef\csname odunum@n@benchmark_main.n.cascade_asr.ao\endcsname{731}
\expandafter\gdef\csname odunum@val@benchmark_main.n.cascade_asr.av\endcsname{1\,347}
\expandafter\gdef\csname odunum@n@benchmark_main.n.cascade_asr.av\endcsname{1\,347}
\expandafter\gdef\csname odunum@val@benchmark_main.n.gemini\endcsname{2\,078}
\expandafter\gdef\csname odunum@n@benchmark_main.n.gemini\endcsname{2\,078}
\expandafter\gdef\csname odunum@val@benchmark_main.n.gemini.ao\endcsname{731}
\expandafter\gdef\csname odunum@n@benchmark_main.n.gemini.ao\endcsname{731}
\expandafter\gdef\csname odunum@val@benchmark_main.n.gemini.av\endcsname{1\,347}
\expandafter\gdef\csname odunum@n@benchmark_main.n.gemini.av\endcsname{1\,347}
\expandafter\gdef\csname odunum@val@benchmark_main.n.gemini35_flash_lite\endcsname{2\,078}
\expandafter\gdef\csname odunum@n@benchmark_main.n.gemini35_flash_lite\endcsname{2\,078}
\expandafter\gdef\csname odunum@val@benchmark_main.n.gemini35_flash_lite.ao\endcsname{731}
\expandafter\gdef\csname odunum@n@benchmark_main.n.gemini35_flash_lite.ao\endcsname{731}
\expandafter\gdef\csname odunum@val@benchmark_main.n.gemini35_flash_lite.av\endcsname{1\,347}
\expandafter\gdef\csname odunum@n@benchmark_main.n.gemini35_flash_lite.av\endcsname{1\,347}
\expandafter\gdef\csname odunum@val@benchmark_main.n.gemini37_flash\endcsname{2\,067}
\expandafter\gdef\csname odunum@n@benchmark_main.n.gemini37_flash\endcsname{2\,078}
\expandafter\gdef\csname odunum@val@benchmark_main.n.gemini37_flash.ao\endcsname{723}
\expandafter\gdef\csname odunum@n@benchmark_main.n.gemini37_flash.ao\endcsname{731}
\expandafter\gdef\csname odunum@val@benchmark_main.n.gemini37_flash.av\endcsname{1\,344}
\expandafter\gdef\csname odunum@n@benchmark_main.n.gemini37_flash.av\endcsname{1\,347}
\expandafter\gdef\csname odunum@val@benchmark_main.n.gpt_realtime\endcsname{731}
\expandafter\gdef\csname odunum@n@benchmark_main.n.gpt_realtime\endcsname{731}
\expandafter\gdef\csname odunum@val@benchmark_main.n.gpt_realtime.ao\endcsname{731}
\expandafter\gdef\csname odunum@n@benchmark_main.n.gpt_realtime.ao\endcsname{731}
\expandafter\gdef\csname odunum@val@benchmark_main.n.kimi_audio_7b_instruct\endcsname{730}
\expandafter\gdef\csname odunum@n@benchmark_main.n.kimi_audio_7b_instruct\endcsname{731}
\expandafter\gdef\csname odunum@val@benchmark_main.n.kimi_audio_7b_instruct.ao\endcsname{730}
\expandafter\gdef\csname odunum@n@benchmark_main.n.kimi_audio_7b_instruct.ao\endcsname{731}
\expandafter\gdef\csname odunum@val@benchmark_main.n.ming\endcsname{2\,078}
\expandafter\gdef\csname odunum@n@benchmark_main.n.ming\endcsname{2\,078}
\expandafter\gdef\csname odunum@val@benchmark_main.n.ming.ao\endcsname{731}
\expandafter\gdef\csname odunum@n@benchmark_main.n.ming.ao\endcsname{731}
\expandafter\gdef\csname odunum@val@benchmark_main.n.ming.av\endcsname{1\,347}
\expandafter\gdef\csname odunum@n@benchmark_main.n.ming.av\endcsname{1\,347}
\expandafter\gdef\csname odunum@val@benchmark_main.n.minicpm_o\endcsname{2\,075}
\expandafter\gdef\csname odunum@n@benchmark_main.n.minicpm_o\endcsname{2\,078}
\expandafter\gdef\csname odunum@val@benchmark_main.n.minicpm_o.ao\endcsname{730}
\expandafter\gdef\csname odunum@n@benchmark_main.n.minicpm_o.ao\endcsname{731}
\expandafter\gdef\csname odunum@val@benchmark_main.n.minicpm_o.av\endcsname{1\,345}
\expandafter\gdef\csname odunum@n@benchmark_main.n.minicpm_o.av\endcsname{1\,347}
\expandafter\gdef\csname odunum@val@benchmark_main.n.nemotron\endcsname{2\,078}
\expandafter\gdef\csname odunum@n@benchmark_main.n.nemotron\endcsname{2\,078}
\expandafter\gdef\csname odunum@val@benchmark_main.n.nemotron.ao\endcsname{731}
\expandafter\gdef\csname odunum@n@benchmark_main.n.nemotron.ao\endcsname{731}
\expandafter\gdef\csname odunum@val@benchmark_main.n.nemotron.av\endcsname{1\,347}
\expandafter\gdef\csname odunum@n@benchmark_main.n.nemotron.av\endcsname{1\,347}
\expandafter\gdef\csname odunum@val@benchmark_main.n.qwen25_omni\endcsname{2\,078}
\expandafter\gdef\csname odunum@n@benchmark_main.n.qwen25_omni\endcsname{2\,078}
\expandafter\gdef\csname odunum@val@benchmark_main.n.qwen25_omni.ao\endcsname{731}
\expandafter\gdef\csname odunum@n@benchmark_main.n.qwen25_omni.ao\endcsname{731}
\expandafter\gdef\csname odunum@val@benchmark_main.n.qwen25_omni.av\endcsname{1\,347}
\expandafter\gdef\csname odunum@n@benchmark_main.n.qwen25_omni.av\endcsname{1\,347}
\expandafter\gdef\csname odunum@val@benchmark_main.n.qwen3_omni_instruct\endcsname{2\,078}
\expandafter\gdef\csname odunum@n@benchmark_main.n.qwen3_omni_instruct\endcsname{2\,078}
\expandafter\gdef\csname odunum@val@benchmark_main.n.qwen3_omni_instruct.ao\endcsname{731}
\expandafter\gdef\csname odunum@n@benchmark_main.n.qwen3_omni_instruct.ao\endcsname{731}
\expandafter\gdef\csname odunum@val@benchmark_main.n.qwen3_omni_instruct.av\endcsname{1\,347}
\expandafter\gdef\csname odunum@n@benchmark_main.n.qwen3_omni_instruct.av\endcsname{1\,347}
\expandafter\gdef\csname odunum@val@benchmark_main.n.qwen3_omni_think\endcsname{2\,078}
\expandafter\gdef\csname odunum@n@benchmark_main.n.qwen3_omni_think\endcsname{2\,078}
\expandafter\gdef\csname odunum@val@benchmark_main.n.qwen3_omni_think.ao\endcsname{731}
\expandafter\gdef\csname odunum@n@benchmark_main.n.qwen3_omni_think.ao\endcsname{731}
\expandafter\gdef\csname odunum@val@benchmark_main.n.qwen3_omni_think.av\endcsname{1\,347}
\expandafter\gdef\csname odunum@n@benchmark_main.n.qwen3_omni_think.av\endcsname{1\,347}
\expandafter\gdef\csname odunum@val@benchmark_main.n.qwen_plus\endcsname{2\,077}
\expandafter\gdef\csname odunum@n@benchmark_main.n.qwen_plus\endcsname{2\,078}
\expandafter\gdef\csname odunum@val@benchmark_main.n.qwen_plus.ao\endcsname{731}
\expandafter\gdef\csname odunum@n@benchmark_main.n.qwen_plus.ao\endcsname{731}
\expandafter\gdef\csname odunum@val@benchmark_main.n.qwen_plus.av\endcsname{1\,346}
\expandafter\gdef\csname odunum@n@benchmark_main.n.qwen_plus.av\endcsname{1\,347}
\expandafter\gdef\csname odunum@val@benchmark_main.n.salmonn2_7b\endcsname{2\,058}
\expandafter\gdef\csname odunum@n@benchmark_main.n.salmonn2_7b\endcsname{2\,078}
\expandafter\gdef\csname odunum@val@benchmark_main.n.salmonn2_7b.ao\endcsname{728}
\expandafter\gdef\csname odunum@n@benchmark_main.n.salmonn2_7b.ao\endcsname{731}
\expandafter\gdef\csname odunum@val@benchmark_main.n.salmonn2_7b.av\endcsname{1\,330}
\expandafter\gdef\csname odunum@n@benchmark_main.n.salmonn2_7b.av\endcsname{1\,347}
\expandafter\gdef\csname odunum@val@benchmark_main.n.seed\endcsname{2\,078}
\expandafter\gdef\csname odunum@n@benchmark_main.n.seed\endcsname{2\,078}
\expandafter\gdef\csname odunum@val@benchmark_main.n.seed.ao\endcsname{731}
\expandafter\gdef\csname odunum@n@benchmark_main.n.seed.ao\endcsname{731}
\expandafter\gdef\csname odunum@val@benchmark_main.n.seed.av\endcsname{1\,347}
\expandafter\gdef\csname odunum@n@benchmark_main.n.seed.av\endcsname{1\,347}
\expandafter\gdef\csname odunum@val@benchmark_main.name.cascade_asr\endcsname{GPT-5.4 + ASR transcript}
\expandafter\gdef\csname odunum@n@benchmark_main.name.cascade_asr\endcsname{1}
\expandafter\gdef\csname odunum@val@benchmark_main.name.gemini\endcsname{Gemini 3.1 Pro}
\expandafter\gdef\csname odunum@n@benchmark_main.name.gemini\endcsname{1}
\expandafter\gdef\csname odunum@val@benchmark_main.name.gemini35_flash_lite\endcsname{Gemini 3.5 Flash Lite}
\expandafter\gdef\csname odunum@n@benchmark_main.name.gemini35_flash_lite\endcsname{1}
\expandafter\gdef\csname odunum@val@benchmark_main.name.gemini37_flash\endcsname{Gemini 3.7 Flash}
\expandafter\gdef\csname odunum@n@benchmark_main.name.gemini37_flash\endcsname{1}
\expandafter\gdef\csname odunum@val@benchmark_main.name.gpt_realtime\endcsname{GPT-Realtime-2}
\expandafter\gdef\csname odunum@n@benchmark_main.name.gpt_realtime\endcsname{1}
\expandafter\gdef\csname odunum@val@benchmark_main.name.kimi_audio_7b_instruct\endcsname{Kimi-Audio-7B-Instruct}
\expandafter\gdef\csname odunum@n@benchmark_main.name.kimi_audio_7b_instruct\endcsname{1}
\expandafter\gdef\csname odunum@val@benchmark_main.name.ming\endcsname{Ming-Flash-Omni 2.0}
\expandafter\gdef\csname odunum@n@benchmark_main.name.ming\endcsname{1}
\expandafter\gdef\csname odunum@val@benchmark_main.name.minicpm_o\endcsname{MiniCPM-o 4.5}
\expandafter\gdef\csname odunum@n@benchmark_main.name.minicpm_o\endcsname{1}
\expandafter\gdef\csname odunum@val@benchmark_main.name.nemotron\endcsname{Nemotron 3 Nano Omni 30B-A3B}
\expandafter\gdef\csname odunum@n@benchmark_main.name.nemotron\endcsname{1}
\expandafter\gdef\csname odunum@val@benchmark_main.name.qwen25_omni\endcsname{Qwen2.5-Omni-7B}
\expandafter\gdef\csname odunum@n@benchmark_main.name.qwen25_omni\endcsname{1}
\expandafter\gdef\csname odunum@val@benchmark_main.name.qwen3_omni_instruct\endcsname{Qwen3-Omni-30B-A3B-Instruct}
\expandafter\gdef\csname odunum@n@benchmark_main.name.qwen3_omni_instruct\endcsname{1}
\expandafter\gdef\csname odunum@val@benchmark_main.name.qwen3_omni_think\endcsname{Qwen3-Omni-30B-A3B}
\expandafter\gdef\csname odunum@n@benchmark_main.name.qwen3_omni_think\endcsname{1}
\expandafter\gdef\csname odunum@val@benchmark_main.name.qwen_plus\endcsname{Qwen3.5-Omni-Plus}
\expandafter\gdef\csname odunum@n@benchmark_main.name.qwen_plus\endcsname{1}
\expandafter\gdef\csname odunum@val@benchmark_main.name.salmonn2_7b\endcsname{video-SALMONN2+ 7B}
\expandafter\gdef\csname odunum@n@benchmark_main.name.salmonn2_7b\endcsname{1}
\expandafter\gdef\csname odunum@val@benchmark_main.name.seed\endcsname{Seed 2.0 Lite}
\expandafter\gdef\csname odunum@n@benchmark_main.name.seed\endcsname{1}
\expandafter\gdef\csname odunum@val@benchmark_main.overall.cascade_asr\endcsname{0.635}
\expandafter\gdef\csname odunum@n@benchmark_main.overall.cascade_asr\endcsname{2\,078}
\expandafter\gdef\csname odunum@val@benchmark_main.overall.cascade_asr.ao\endcsname{0.715}
\expandafter\gdef\csname odunum@n@benchmark_main.overall.cascade_asr.ao\endcsname{731}
\expandafter\gdef\csname odunum@val@benchmark_main.overall.cascade_asr.av\endcsname{0.590}
\expandafter\gdef\csname odunum@n@benchmark_main.overall.cascade_asr.av\endcsname{1\,347}
\expandafter\gdef\csname odunum@val@benchmark_main.overall.gemini\endcsname{0.736}
\expandafter\gdef\csname odunum@n@benchmark_main.overall.gemini\endcsname{2\,078}
\expandafter\gdef\csname odunum@val@benchmark_main.overall.gemini.ao\endcsname{0.754}
\expandafter\gdef\csname odunum@n@benchmark_main.overall.gemini.ao\endcsname{731}
\expandafter\gdef\csname odunum@val@benchmark_main.overall.gemini.av\endcsname{0.726}
\expandafter\gdef\csname odunum@n@benchmark_main.overall.gemini.av\endcsname{1\,347}
\expandafter\gdef\csname odunum@val@benchmark_main.overall.gemini35_flash_lite\endcsname{0.591}
\expandafter\gdef\csname odunum@n@benchmark_main.overall.gemini35_flash_lite\endcsname{2\,078}
\expandafter\gdef\csname odunum@val@benchmark_main.overall.gemini35_flash_lite.ao\endcsname{0.622}
\expandafter\gdef\csname odunum@n@benchmark_main.overall.gemini35_flash_lite.ao\endcsname{731}
\expandafter\gdef\csname odunum@val@benchmark_main.overall.gemini35_flash_lite.av\endcsname{0.574}
\expandafter\gdef\csname odunum@n@benchmark_main.overall.gemini35_flash_lite.av\endcsname{1\,347}
\expandafter\gdef\csname odunum@val@benchmark_main.overall.gemini37_flash\endcsname{0.707}
\expandafter\gdef\csname odunum@n@benchmark_main.overall.gemini37_flash\endcsname{2\,067}
\expandafter\gdef\csname odunum@val@benchmark_main.overall.gemini37_flash.ao\endcsname{0.729}
\expandafter\gdef\csname odunum@n@benchmark_main.overall.gemini37_flash.ao\endcsname{723}
\expandafter\gdef\csname odunum@val@benchmark_main.overall.gemini37_flash.av\endcsname{0.696}
\expandafter\gdef\csname odunum@n@benchmark_main.overall.gemini37_flash.av\endcsname{1\,344}
\expandafter\gdef\csname odunum@val@benchmark_main.overall.gpt_realtime\endcsname{0.638}
\expandafter\gdef\csname odunum@n@benchmark_main.overall.gpt_realtime\endcsname{731}
\expandafter\gdef\csname odunum@val@benchmark_main.overall.gpt_realtime.ao\endcsname{0.638}
\expandafter\gdef\csname odunum@n@benchmark_main.overall.gpt_realtime.ao\endcsname{731}
\expandafter\gdef\csname odunum@val@benchmark_main.overall.kimi_audio_7b_instruct\endcsname{0.526}
\expandafter\gdef\csname odunum@n@benchmark_main.overall.kimi_audio_7b_instruct\endcsname{730}
\expandafter\gdef\csname odunum@val@benchmark_main.overall.kimi_audio_7b_instruct.ao\endcsname{0.526}
\expandafter\gdef\csname odunum@n@benchmark_main.overall.kimi_audio_7b_instruct.ao\endcsname{730}
\expandafter\gdef\csname odunum@val@benchmark_main.overall.ming\endcsname{0.527}
\expandafter\gdef\csname odunum@n@benchmark_main.overall.ming\endcsname{2\,078}
\expandafter\gdef\csname odunum@val@benchmark_main.overall.ming.ao\endcsname{0.526}
\expandafter\gdef\csname odunum@n@benchmark_main.overall.ming.ao\endcsname{731}
\expandafter\gdef\csname odunum@val@benchmark_main.overall.ming.av\endcsname{0.526}
\expandafter\gdef\csname odunum@n@benchmark_main.overall.ming.av\endcsname{1\,347}
\expandafter\gdef\csname odunum@val@benchmark_main.overall.minicpm_o\endcsname{0.426}
\expandafter\gdef\csname odunum@n@benchmark_main.overall.minicpm_o\endcsname{2\,075}
\expandafter\gdef\csname odunum@val@benchmark_main.overall.minicpm_o.ao\endcsname{0.491}
\expandafter\gdef\csname odunum@n@benchmark_main.overall.minicpm_o.ao\endcsname{730}
\expandafter\gdef\csname odunum@val@benchmark_main.overall.minicpm_o.av\endcsname{0.391}
\expandafter\gdef\csname odunum@n@benchmark_main.overall.minicpm_o.av\endcsname{1\,345}
\expandafter\gdef\csname odunum@val@benchmark_main.overall.nemotron\endcsname{0.367}
\expandafter\gdef\csname odunum@n@benchmark_main.overall.nemotron\endcsname{2\,078}
\expandafter\gdef\csname odunum@val@benchmark_main.overall.nemotron.ao\endcsname{0.401}
\expandafter\gdef\csname odunum@n@benchmark_main.overall.nemotron.ao\endcsname{731}
\expandafter\gdef\csname odunum@val@benchmark_main.overall.nemotron.av\endcsname{0.349}
\expandafter\gdef\csname odunum@n@benchmark_main.overall.nemotron.av\endcsname{1\,347}
\expandafter\gdef\csname odunum@val@benchmark_main.overall.qwen25_omni\endcsname{0.463}
\expandafter\gdef\csname odunum@n@benchmark_main.overall.qwen25_omni\endcsname{2\,078}
\expandafter\gdef\csname odunum@val@benchmark_main.overall.qwen25_omni.ao\endcsname{0.480}
\expandafter\gdef\csname odunum@n@benchmark_main.overall.qwen25_omni.ao\endcsname{731}
\expandafter\gdef\csname odunum@val@benchmark_main.overall.qwen25_omni.av\endcsname{0.452}
\expandafter\gdef\csname odunum@n@benchmark_main.overall.qwen25_omni.av\endcsname{1\,347}
\expandafter\gdef\csname odunum@val@benchmark_main.overall.qwen3_omni_instruct\endcsname{0.575}
\expandafter\gdef\csname odunum@n@benchmark_main.overall.qwen3_omni_instruct\endcsname{2\,078}
\expandafter\gdef\csname odunum@val@benchmark_main.overall.qwen3_omni_instruct.ao\endcsname{0.616}
\expandafter\gdef\csname odunum@n@benchmark_main.overall.qwen3_omni_instruct.ao\endcsname{731}
\expandafter\gdef\csname odunum@val@benchmark_main.overall.qwen3_omni_instruct.av\endcsname{0.552}
\expandafter\gdef\csname odunum@n@benchmark_main.overall.qwen3_omni_instruct.av\endcsname{1\,347}
\expandafter\gdef\csname odunum@val@benchmark_main.overall.qwen3_omni_think\endcsname{0.626}
\expandafter\gdef\csname odunum@n@benchmark_main.overall.qwen3_omni_think\endcsname{2\,078}
\expandafter\gdef\csname odunum@val@benchmark_main.overall.qwen3_omni_think.ao\endcsname{0.674}
\expandafter\gdef\csname odunum@n@benchmark_main.overall.qwen3_omni_think.ao\endcsname{731}
\expandafter\gdef\csname odunum@val@benchmark_main.overall.qwen3_omni_think.av\endcsname{0.599}
\expandafter\gdef\csname odunum@n@benchmark_main.overall.qwen3_omni_think.av\endcsname{1\,347}
\expandafter\gdef\csname odunum@val@benchmark_main.overall.qwen_plus\endcsname{0.714}
\expandafter\gdef\csname odunum@n@benchmark_main.overall.qwen_plus\endcsname{2\,077}
\expandafter\gdef\csname odunum@val@benchmark_main.overall.qwen_plus.ao\endcsname{0.747}
\expandafter\gdef\csname odunum@n@benchmark_main.overall.qwen_plus.ao\endcsname{731}
\expandafter\gdef\csname odunum@val@benchmark_main.overall.qwen_plus.av\endcsname{0.696}
\expandafter\gdef\csname odunum@n@benchmark_main.overall.qwen_plus.av\endcsname{1\,346}
\expandafter\gdef\csname odunum@val@benchmark_main.overall.salmonn2_7b\endcsname{0.176}
\expandafter\gdef\csname odunum@n@benchmark_main.overall.salmonn2_7b\endcsname{2\,058}
\expandafter\gdef\csname odunum@val@benchmark_main.overall.salmonn2_7b.ao\endcsname{0.238}
\expandafter\gdef\csname odunum@n@benchmark_main.overall.salmonn2_7b.ao\endcsname{728}
\expandafter\gdef\csname odunum@val@benchmark_main.overall.salmonn2_7b.av\endcsname{0.142}
\expandafter\gdef\csname odunum@n@benchmark_main.overall.salmonn2_7b.av\endcsname{1\,330}
\expandafter\gdef\csname odunum@val@benchmark_main.overall.seed\endcsname{0.712}
\expandafter\gdef\csname odunum@n@benchmark_main.overall.seed\endcsname{2\,078}
\expandafter\gdef\csname odunum@val@benchmark_main.overall.seed.ao\endcsname{0.773}
\expandafter\gdef\csname odunum@n@benchmark_main.overall.seed.ao\endcsname{731}
\expandafter\gdef\csname odunum@val@benchmark_main.overall.seed.av\endcsname{0.679}
\expandafter\gdef\csname odunum@n@benchmark_main.overall.seed.av\endcsname{1\,347}
\expandafter\gdef\csname odunum@val@benchmark_main.population.n_ao\endcsname{731}
\expandafter\gdef\csname odunum@n@benchmark_main.population.n_ao\endcsname{2\,078}
\expandafter\gdef\csname odunum@val@benchmark_main.population.n_ao_neg\endcsname{161}
\expandafter\gdef\csname odunum@n@benchmark_main.population.n_ao_neg\endcsname{2\,078}
\expandafter\gdef\csname odunum@val@benchmark_main.population.n_ao_pos\endcsname{570}
\expandafter\gdef\csname odunum@n@benchmark_main.population.n_ao_pos\endcsname{2\,078}
\expandafter\gdef\csname odunum@val@benchmark_main.population.n_av\endcsname{1\,347}
\expandafter\gdef\csname odunum@n@benchmark_main.population.n_av\endcsname{2\,078}
\expandafter\gdef\csname odunum@val@benchmark_main.population.n_av_neg\endcsname{286}
\expandafter\gdef\csname odunum@n@benchmark_main.population.n_av_neg\endcsname{2\,078}
\expandafter\gdef\csname odunum@val@benchmark_main.population.n_av_pos\endcsname{1\,061}
\expandafter\gdef\csname odunum@n@benchmark_main.population.n_av_pos\endcsname{2\,078}
\expandafter\gdef\csname odunum@val@benchmark_main.population.n_scenes\endcsname{2\,078}
\expandafter\gdef\csname odunum@n@benchmark_main.population.n_scenes\endcsname{2\,078}
\expandafter\gdef\csname odunum@val@benchmark_main.size.cascade_asr\endcsname{undisclosed}
\expandafter\gdef\csname odunum@n@benchmark_main.size.cascade_asr\endcsname{1}
\expandafter\gdef\csname odunum@val@benchmark_main.size.gemini\endcsname{undisclosed}
\expandafter\gdef\csname odunum@n@benchmark_main.size.gemini\endcsname{1}
\expandafter\gdef\csname odunum@val@benchmark_main.size.gemini35_flash_lite\endcsname{undisclosed}
\expandafter\gdef\csname odunum@n@benchmark_main.size.gemini35_flash_lite\endcsname{1}
\expandafter\gdef\csname odunum@val@benchmark_main.size.gemini37_flash\endcsname{undisclosed}
\expandafter\gdef\csname odunum@n@benchmark_main.size.gemini37_flash\endcsname{1}
\expandafter\gdef\csname odunum@val@benchmark_main.size.gpt_realtime\endcsname{undisclosed}
\expandafter\gdef\csname odunum@n@benchmark_main.size.gpt_realtime\endcsname{1}
\expandafter\gdef\csname odunum@val@benchmark_main.size.kimi_audio_7b_instruct\endcsname{7B}
\expandafter\gdef\csname odunum@n@benchmark_main.size.kimi_audio_7b_instruct\endcsname{1}
\expandafter\gdef\csname odunum@val@benchmark_main.size.ming\endcsname{104B-A6B}
\expandafter\gdef\csname odunum@n@benchmark_main.size.ming\endcsname{1}
\expandafter\gdef\csname odunum@val@benchmark_main.size.minicpm_o\endcsname{9B}
\expandafter\gdef\csname odunum@n@benchmark_main.size.minicpm_o\endcsname{1}
\expandafter\gdef\csname odunum@val@benchmark_main.size.nemotron\endcsname{30B-A3B}
\expandafter\gdef\csname odunum@n@benchmark_main.size.nemotron\endcsname{1}
\expandafter\gdef\csname odunum@val@benchmark_main.size.qwen25_omni\endcsname{7B}
\expandafter\gdef\csname odunum@n@benchmark_main.size.qwen25_omni\endcsname{1}
\expandafter\gdef\csname odunum@val@benchmark_main.size.qwen3_omni_instruct\endcsname{30B-A3B}
\expandafter\gdef\csname odunum@n@benchmark_main.size.qwen3_omni_instruct\endcsname{1}
\expandafter\gdef\csname odunum@val@benchmark_main.size.qwen3_omni_think\endcsname{30B-A3B}
\expandafter\gdef\csname odunum@n@benchmark_main.size.qwen3_omni_think\endcsname{1}
\expandafter\gdef\csname odunum@val@benchmark_main.size.qwen_plus\endcsname{undisclosed}
\expandafter\gdef\csname odunum@n@benchmark_main.size.qwen_plus\endcsname{1}
\expandafter\gdef\csname odunum@val@benchmark_main.size.salmonn2_7b\endcsname{7B}
\expandafter\gdef\csname odunum@n@benchmark_main.size.salmonn2_7b\endcsname{1}
\expandafter\gdef\csname odunum@val@benchmark_main.size.seed\endcsname{undisclosed}
\expandafter\gdef\csname odunum@n@benchmark_main.size.seed\endcsname{1}
\expandafter\gdef\csname odunum@val@benchmark_main.weight.m1\endcsname{0.15}
\expandafter\gdef\csname odunum@n@benchmark_main.weight.m1\endcsname{1}
\expandafter\gdef\csname odunum@val@benchmark_main.weight.m2\endcsname{0.60}
\expandafter\gdef\csname odunum@n@benchmark_main.weight.m2\endcsname{1}
\expandafter\gdef\csname odunum@val@benchmark_main.weight.m3\endcsname{0.10}
\expandafter\gdef\csname odunum@n@benchmark_main.weight.m3\endcsname{1}
\expandafter\gdef\csname odunum@val@benchmark_main.weight.m4\endcsname{0.10}
\expandafter\gdef\csname odunum@n@benchmark_main.weight.m4\endcsname{1}
\expandafter\gdef\csname odunum@val@benchmark_main.weight.m5\endcsname{0.05}
\expandafter\gdef\csname odunum@n@benchmark_main.weight.m5\endcsname{1}
\expandafter\gdef\csname odunum@val@benchmark_main.weights\endcsname{0.15/0.60/0.10/0.10/0.05}
\expandafter\gdef\csname odunum@n@benchmark_main.weights\endcsname{1}
\expandafter\gdef\csname odunum@val@benchmark_overview.ftr.above_half.ao\endcsname{11}
\expandafter\gdef\csname odunum@n@benchmark_overview.ftr.above_half.ao\endcsname{14}
\expandafter\gdef\csname odunum@val@benchmark_overview.ftr.above_half.av\endcsname{10}
\expandafter\gdef\csname odunum@n@benchmark_overview.ftr.above_half.av\endcsname{12}
\expandafter\gdef\csname odunum@val@benchmark_overview.ftr.above_half.pooled\endcsname{11}
\expandafter\gdef\csname odunum@n@benchmark_overview.ftr.above_half.pooled\endcsname{14}
\expandafter\gdef\csname odunum@val@benchmark_overview.native.n_models.ao\endcsname{14}
\expandafter\gdef\csname odunum@n@benchmark_overview.native.n_models.ao\endcsname{14}
\expandafter\gdef\csname odunum@val@benchmark_overview.native.n_models.av\endcsname{12}
\expandafter\gdef\csname odunum@n@benchmark_overview.native.n_models.av\endcsname{14}
\expandafter\gdef\csname odunum@val@benchmark_overview.native.n_models.pooled\endcsname{14}
\expandafter\gdef\csname odunum@n@benchmark_overview.native.n_models.pooled\endcsname{14}
\expandafter\gdef\csname odunum@val@bounds_asr_verbatim.asr.span_identical_to_gt\endcsname{0.753}
\expandafter\gdef\csname odunum@n@bounds_asr_verbatim.asr.span_identical_to_gt\endcsname{1\,630}
\expandafter\gdef\csname odunum@val@bounds_asr_verbatim.asr.span_identical_to_gt.gen\endcsname{0.794}
\expandafter\gdef\csname odunum@n@bounds_asr_verbatim.asr.span_identical_to_gt.gen\endcsname{1\,429}
\expandafter\gdef\csname odunum@val@bounds_asr_verbatim.asr.span_identical_to_gt.recorded\endcsname{0.463}
\expandafter\gdef\csname odunum@n@bounds_asr_verbatim.asr.span_identical_to_gt.recorded\endcsname{201}
\expandafter\gdef\csname odunum@val@bounds_asr_verbatim.asr.text_identical_to_gt\endcsname{0.582}
\expandafter\gdef\csname odunum@n@bounds_asr_verbatim.asr.text_identical_to_gt\endcsname{1\,630}
\expandafter\gdef\csname odunum@val@bounds_asr_verbatim.asr.text_identical_to_gt.gen\endcsname{0.628}
\expandafter\gdef\csname odunum@n@bounds_asr_verbatim.asr.text_identical_to_gt.gen\endcsname{1\,429}
\expandafter\gdef\csname odunum@val@bounds_asr_verbatim.asr.text_identical_to_gt.recorded\endcsname{0.254}
\expandafter\gdef\csname odunum@n@bounds_asr_verbatim.asr.text_identical_to_gt.recorded\endcsname{201}
\expandafter\gdef\csname odunum@val@bounds_asr_verbatim.asr.token_recall_of_gt\endcsname{0.979}
\expandafter\gdef\csname odunum@n@bounds_asr_verbatim.asr.token_recall_of_gt\endcsname{1\,629}
\expandafter\gdef\csname odunum@val@bounds_asr_verbatim.asr.token_recall_of_gt.gen\endcsname{0.983}
\expandafter\gdef\csname odunum@n@bounds_asr_verbatim.asr.token_recall_of_gt.gen\endcsname{1\,428}
\expandafter\gdef\csname odunum@val@bounds_asr_verbatim.asr.token_recall_of_gt.recorded\endcsname{0.957}
\expandafter\gdef\csname odunum@n@bounds_asr_verbatim.asr.token_recall_of_gt.recorded\endcsname{201}
\expandafter\gdef\csname odunum@val@bounds_asr_verbatim.asr.token_share_absent_from_gt\endcsname{0.081}
\expandafter\gdef\csname odunum@n@bounds_asr_verbatim.asr.token_share_absent_from_gt\endcsname{1\,629}
\expandafter\gdef\csname odunum@val@bounds_asr_verbatim.asr.token_share_absent_from_gt.gen\endcsname{0.059}
\expandafter\gdef\csname odunum@n@bounds_asr_verbatim.asr.token_share_absent_from_gt.gen\endcsname{1\,428}
\expandafter\gdef\csname odunum@val@bounds_asr_verbatim.asr.token_share_absent_from_gt.recorded\endcsname{0.233}
\expandafter\gdef\csname odunum@n@bounds_asr_verbatim.asr.token_share_absent_from_gt.recorded\endcsname{201}
\expandafter\gdef\csname odunum@val@bounds_asr_verbatim.audit.all.judged_points\endcsname{5\,859}
\expandafter\gdef\csname odunum@n@bounds_asr_verbatim.audit.all.judged_points\endcsname{5\,859}
\expandafter\gdef\csname odunum@val@bounds_asr_verbatim.audit.ao.judged_points\endcsname{1\,886}
\expandafter\gdef\csname odunum@n@bounds_asr_verbatim.audit.ao.judged_points\endcsname{1\,886}
\expandafter\gdef\csname odunum@val@bounds_asr_verbatim.audit.asr_empty_scenes\endcsname{1}
\expandafter\gdef\csname odunum@n@bounds_asr_verbatim.audit.asr_empty_scenes\endcsname{1\,631}
\expandafter\gdef\csname odunum@val@bounds_asr_verbatim.audit.asr_empty_scenes.gen\endcsname{0}
\expandafter\gdef\csname odunum@n@bounds_asr_verbatim.audit.asr_empty_scenes.gen\endcsname{1\,429}
\expandafter\gdef\csname odunum@val@bounds_asr_verbatim.audit.asr_empty_scenes.recorded\endcsname{1}
\expandafter\gdef\csname odunum@n@bounds_asr_verbatim.audit.asr_empty_scenes.recorded\endcsname{202}
\expandafter\gdef\csname odunum@val@bounds_asr_verbatim.audit.asr_sentences\endcsname{3\,572}
\expandafter\gdef\csname odunum@n@bounds_asr_verbatim.audit.asr_sentences\endcsname{1\,631}
\expandafter\gdef\csname odunum@val@bounds_asr_verbatim.audit.asr_sentences.gen\endcsname{3\,087}
\expandafter\gdef\csname odunum@n@bounds_asr_verbatim.audit.asr_sentences.gen\endcsname{1\,429}
\expandafter\gdef\csname odunum@val@bounds_asr_verbatim.audit.asr_sentences.recorded\endcsname{485}
\expandafter\gdef\csname odunum@n@bounds_asr_verbatim.audit.asr_sentences.recorded\endcsname{202}
\expandafter\gdef\csname odunum@val@bounds_asr_verbatim.audit.asr_status_ok\endcsname{1\,630}
\expandafter\gdef\csname odunum@n@bounds_asr_verbatim.audit.asr_status_ok\endcsname{1\,631}
\expandafter\gdef\csname odunum@val@bounds_asr_verbatim.audit.asr_status_ok.gen\endcsname{1\,429}
\expandafter\gdef\csname odunum@n@bounds_asr_verbatim.audit.asr_status_ok.gen\endcsname{1\,429}
\expandafter\gdef\csname odunum@val@bounds_asr_verbatim.audit.asr_status_ok.recorded\endcsname{201}
\expandafter\gdef\csname odunum@n@bounds_asr_verbatim.audit.asr_status_ok.recorded\endcsname{202}
\expandafter\gdef\csname odunum@val@bounds_asr_verbatim.audit.av.judged_points\endcsname{3\,973}
\expandafter\gdef\csname odunum@n@bounds_asr_verbatim.audit.av.judged_points\endcsname{3\,973}
\expandafter\gdef\csname odunum@val@bounds_asr_verbatim.audit.point_set_shared_with_other_arm\endcsname{5\,798}
\expandafter\gdef\csname odunum@n@bounds_asr_verbatim.audit.point_set_shared_with_other_arm\endcsname{5\,859}
\expandafter\gdef\csname odunum@val@bounds_asr_verbatim.audit.points_without_verdict\endcsname{0}
\expandafter\gdef\csname odunum@n@bounds_asr_verbatim.audit.points_without_verdict\endcsname{5\,859}
\expandafter\gdef\csname odunum@val@bounds_asr_verbatim.audit.reader_matches_keypoint_types\endcsname{5\,859}
\expandafter\gdef\csname odunum@n@bounds_asr_verbatim.audit.reader_matches_keypoint_types\endcsname{5\,859}
\expandafter\gdef\csname odunum@val@bounds_asr_verbatim.audit.scenes_incomplete\endcsname{0}
\expandafter\gdef\csname odunum@n@bounds_asr_verbatim.audit.scenes_incomplete\endcsname{1\,631}
\expandafter\gdef\csname odunum@val@bounds_asr_verbatim.audit.scenes_with_full_gt_denominator\endcsname{1\,631}
\expandafter\gdef\csname odunum@n@bounds_asr_verbatim.audit.scenes_with_full_gt_denominator\endcsname{1\,631}
\expandafter\gdef\csname odunum@val@bounds_asr_verbatim.audit.segments_missed\endcsname{1}
\expandafter\gdef\csname odunum@n@bounds_asr_verbatim.audit.segments_missed\endcsname{1\,679}
\expandafter\gdef\csname odunum@val@bounds_asr_verbatim.cov.all.audio\endcsname{0.547}
\expandafter\gdef\csname odunum@n@bounds_asr_verbatim.cov.all.audio\endcsname{899}
\expandafter\gdef\csname odunum@ci@bounds_asr_verbatim.cov.all.audio\endcsname{[0.514, 0.580]}
\expandafter\gdef\csname odunum@val@bounds_asr_verbatim.cov.all.context\endcsname{0.426}
\expandafter\gdef\csname odunum@n@bounds_asr_verbatim.cov.all.context\endcsname{2\,702}
\expandafter\gdef\csname odunum@ci@bounds_asr_verbatim.cov.all.context\endcsname{[0.405, 0.447]}
\expandafter\gdef\csname odunum@val@bounds_asr_verbatim.cov.all.context.gen\endcsname{0.408}
\expandafter\gdef\csname odunum@n@bounds_asr_verbatim.cov.all.context.gen\endcsname{2\,363}
\expandafter\gdef\csname odunum@ci@bounds_asr_verbatim.cov.all.context.gen\endcsname{[0.386, 0.431]}
\expandafter\gdef\csname odunum@val@bounds_asr_verbatim.cov.all.context.latin\endcsname{0.432}
\expandafter\gdef\csname odunum@n@bounds_asr_verbatim.cov.all.context.latin\endcsname{1\,180}
\expandafter\gdef\csname odunum@ci@bounds_asr_verbatim.cov.all.context.latin\endcsname{[0.400, 0.464]}
\expandafter\gdef\csname odunum@val@bounds_asr_verbatim.cov.all.context.lex_absent\endcsname{0.419}
\expandafter\gdef\csname odunum@n@bounds_asr_verbatim.cov.all.context.lex_absent\endcsname{2\,649}
\expandafter\gdef\csname odunum@ci@bounds_asr_verbatim.cov.all.context.lex_absent\endcsname{[0.398, 0.441]}
\expandafter\gdef\csname odunum@val@bounds_asr_verbatim.cov.all.context.lex_absent.latin\endcsname{0.417}
\expandafter\gdef\csname odunum@n@bounds_asr_verbatim.cov.all.context.lex_absent.latin\endcsname{1\,127}
\expandafter\gdef\csname odunum@ci@bounds_asr_verbatim.cov.all.context.lex_absent.latin\endcsname{[0.385, 0.449]}
\expandafter\gdef\csname odunum@val@bounds_asr_verbatim.cov.all.context.lex_present\endcsname{0.755}
\expandafter\gdef\csname odunum@n@bounds_asr_verbatim.cov.all.context.lex_present\endcsname{53}
\expandafter\gdef\csname odunum@ci@bounds_asr_verbatim.cov.all.context.lex_present\endcsname{[0.630, 0.870]}
\expandafter\gdef\csname odunum@val@bounds_asr_verbatim.cov.all.context.lex_present.latin\endcsname{0.755}
\expandafter\gdef\csname odunum@n@bounds_asr_verbatim.cov.all.context.lex_present.latin\endcsname{53}
\expandafter\gdef\csname odunum@ci@bounds_asr_verbatim.cov.all.context.lex_present.latin\endcsname{[0.630, 0.870]}
\expandafter\gdef\csname odunum@val@bounds_asr_verbatim.cov.all.context.recorded\endcsname{0.549}
\expandafter\gdef\csname odunum@n@bounds_asr_verbatim.cov.all.context.recorded\endcsname{339}
\expandafter\gdef\csname odunum@ci@bounds_asr_verbatim.cov.all.context.recorded\endcsname{[0.490, 0.608]}
\expandafter\gdef\csname odunum@val@bounds_asr_verbatim.cov.all.history\endcsname{0.384}
\expandafter\gdef\csname odunum@n@bounds_asr_verbatim.cov.all.history\endcsname{818}
\expandafter\gdef\csname odunum@ci@bounds_asr_verbatim.cov.all.history\endcsname{[0.346, 0.421]}
\expandafter\gdef\csname odunum@val@bounds_asr_verbatim.cov.all.intent\endcsname{0.958}
\expandafter\gdef\csname odunum@n@bounds_asr_verbatim.cov.all.intent\endcsname{3\,144}
\expandafter\gdef\csname odunum@ci@bounds_asr_verbatim.cov.all.intent\endcsname{[0.950, 0.965]}
\expandafter\gdef\csname odunum@val@bounds_asr_verbatim.cov.all.intent.gen\endcsname{0.965}
\expandafter\gdef\csname odunum@n@bounds_asr_verbatim.cov.all.intent.gen\endcsname{2\,782}
\expandafter\gdef\csname odunum@ci@bounds_asr_verbatim.cov.all.intent.gen\endcsname{[0.958, 0.972]}
\expandafter\gdef\csname odunum@val@bounds_asr_verbatim.cov.all.intent.latin\endcsname{0.965}
\expandafter\gdef\csname odunum@n@bounds_asr_verbatim.cov.all.intent.latin\endcsname{1\,444}
\expandafter\gdef\csname odunum@ci@bounds_asr_verbatim.cov.all.intent.latin\endcsname{[0.955, 0.974]}
\expandafter\gdef\csname odunum@val@bounds_asr_verbatim.cov.all.intent.recorded\endcsname{0.898}
\expandafter\gdef\csname odunum@n@bounds_asr_verbatim.cov.all.intent.recorded\endcsname{362}
\expandafter\gdef\csname odunum@ci@bounds_asr_verbatim.cov.all.intent.recorded\endcsname{[0.864, 0.929]}
\expandafter\gdef\csname odunum@val@bounds_asr_verbatim.cov.all.other_context\endcsname{0.453}
\expandafter\gdef\csname odunum@n@bounds_asr_verbatim.cov.all.other_context\endcsname{181}
\expandafter\gdef\csname odunum@ci@bounds_asr_verbatim.cov.all.other_context\endcsname{[0.379, 0.526]}
\expandafter\gdef\csname odunum@val@bounds_asr_verbatim.cov.all.overall\endcsname{0.711}
\expandafter\gdef\csname odunum@n@bounds_asr_verbatim.cov.all.overall\endcsname{5\,859}
\expandafter\gdef\csname odunum@ci@bounds_asr_verbatim.cov.all.overall\endcsname{[0.700, 0.723]}
\expandafter\gdef\csname odunum@val@bounds_asr_verbatim.cov.all.overall.gen\endcsname{0.710}
\expandafter\gdef\csname odunum@n@bounds_asr_verbatim.cov.all.overall.gen\endcsname{5\,145}
\expandafter\gdef\csname odunum@ci@bounds_asr_verbatim.cov.all.overall.gen\endcsname{[0.697, 0.722]}
\expandafter\gdef\csname odunum@val@bounds_asr_verbatim.cov.all.overall.recorded\endcsname{0.724}
\expandafter\gdef\csname odunum@n@bounds_asr_verbatim.cov.all.overall.recorded\endcsname{714}
\expandafter\gdef\csname odunum@ci@bounds_asr_verbatim.cov.all.overall.recorded\endcsname{[0.687, 0.762]}
\expandafter\gdef\csname odunum@val@bounds_asr_verbatim.cov.all.unattributed\endcsname{0.462}
\expandafter\gdef\csname odunum@n@bounds_asr_verbatim.cov.all.unattributed\endcsname{13}
\expandafter\gdef\csname odunum@ci@bounds_asr_verbatim.cov.all.unattributed\endcsname{[0.118, 0.900]}
\expandafter\gdef\csname odunum@val@bounds_asr_verbatim.cov.all.visual\endcsname{0.327}
\expandafter\gdef\csname odunum@n@bounds_asr_verbatim.cov.all.visual\endcsname{804}
\expandafter\gdef\csname odunum@ci@bounds_asr_verbatim.cov.all.visual\endcsname{[0.293, 0.362]}
\expandafter\gdef\csname odunum@val@bounds_asr_verbatim.cov.ao.audio\endcsname{0.516}
\expandafter\gdef\csname odunum@n@bounds_asr_verbatim.cov.ao.audio\endcsname{341}
\expandafter\gdef\csname odunum@ci@bounds_asr_verbatim.cov.ao.audio\endcsname{[0.460, 0.571]}
\expandafter\gdef\csname odunum@val@bounds_asr_verbatim.cov.ao.context\endcsname{0.451}
\expandafter\gdef\csname odunum@n@bounds_asr_verbatim.cov.ao.context\endcsname{728}
\expandafter\gdef\csname odunum@ci@bounds_asr_verbatim.cov.ao.context\endcsname{[0.410, 0.490]}
\expandafter\gdef\csname odunum@val@bounds_asr_verbatim.cov.ao.context.gen\endcsname{0.451}
\expandafter\gdef\csname odunum@n@bounds_asr_verbatim.cov.ao.context.gen\endcsname{728}
\expandafter\gdef\csname odunum@ci@bounds_asr_verbatim.cov.ao.context.gen\endcsname{[0.410, 0.490]}
\expandafter\gdef\csname odunum@val@bounds_asr_verbatim.cov.ao.context.latin\endcsname{0.477}
\expandafter\gdef\csname odunum@n@bounds_asr_verbatim.cov.ao.context.latin\endcsname{369}
\expandafter\gdef\csname odunum@ci@bounds_asr_verbatim.cov.ao.context.latin\endcsname{[0.418, 0.534]}
\expandafter\gdef\csname odunum@val@bounds_asr_verbatim.cov.ao.context.lex_absent\endcsname{0.442}
\expandafter\gdef\csname odunum@n@bounds_asr_verbatim.cov.ao.context.lex_absent\endcsname{713}
\expandafter\gdef\csname odunum@ci@bounds_asr_verbatim.cov.ao.context.lex_absent\endcsname{[0.401, 0.482]}
\expandafter\gdef\csname odunum@val@bounds_asr_verbatim.cov.ao.context.lex_absent.latin\endcsname{0.460}
\expandafter\gdef\csname odunum@n@bounds_asr_verbatim.cov.ao.context.lex_absent.latin\endcsname{354}
\expandafter\gdef\csname odunum@ci@bounds_asr_verbatim.cov.ao.context.lex_absent.latin\endcsname{[0.401, 0.518]}
\expandafter\gdef\csname odunum@val@bounds_asr_verbatim.cov.ao.context.lex_present\endcsname{0.867}
\expandafter\gdef\csname odunum@n@bounds_asr_verbatim.cov.ao.context.lex_present\endcsname{15}
\expandafter\gdef\csname odunum@ci@bounds_asr_verbatim.cov.ao.context.lex_present\endcsname{[0.667, 1.000]}
\expandafter\gdef\csname odunum@val@bounds_asr_verbatim.cov.ao.context.lex_present.latin\endcsname{0.867}
\expandafter\gdef\csname odunum@n@bounds_asr_verbatim.cov.ao.context.lex_present.latin\endcsname{15}
\expandafter\gdef\csname odunum@ci@bounds_asr_verbatim.cov.ao.context.lex_present.latin\endcsname{[0.667, 1.000]}
\expandafter\gdef\csname odunum@val@bounds_asr_verbatim.cov.ao.history\endcsname{0.380}
\expandafter\gdef\csname odunum@n@bounds_asr_verbatim.cov.ao.history\endcsname{342}
\expandafter\gdef\csname odunum@ci@bounds_asr_verbatim.cov.ao.history\endcsname{[0.320, 0.441]}
\expandafter\gdef\csname odunum@val@bounds_asr_verbatim.cov.ao.intent\endcsname{0.968}
\expandafter\gdef\csname odunum@n@bounds_asr_verbatim.cov.ao.intent\endcsname{1\,158}
\expandafter\gdef\csname odunum@ci@bounds_asr_verbatim.cov.ao.intent\endcsname{[0.957, 0.978]}
\expandafter\gdef\csname odunum@val@bounds_asr_verbatim.cov.ao.intent.gen\endcsname{0.968}
\expandafter\gdef\csname odunum@n@bounds_asr_verbatim.cov.ao.intent.gen\endcsname{1\,158}
\expandafter\gdef\csname odunum@ci@bounds_asr_verbatim.cov.ao.intent.gen\endcsname{[0.957, 0.978]}
\expandafter\gdef\csname odunum@val@bounds_asr_verbatim.cov.ao.intent.latin\endcsname{0.963}
\expandafter\gdef\csname odunum@n@bounds_asr_verbatim.cov.ao.intent.latin\endcsname{595}
\expandafter\gdef\csname odunum@ci@bounds_asr_verbatim.cov.ao.intent.latin\endcsname{[0.946, 0.978]}
\expandafter\gdef\csname odunum@val@bounds_asr_verbatim.cov.ao.other_context\endcsname{0.489}
\expandafter\gdef\csname odunum@n@bounds_asr_verbatim.cov.ao.other_context\endcsname{45}
\expandafter\gdef\csname odunum@ci@bounds_asr_verbatim.cov.ao.other_context\endcsname{[0.340, 0.638]}
\expandafter\gdef\csname odunum@val@bounds_asr_verbatim.cov.ao.overall\endcsname{0.768}
\expandafter\gdef\csname odunum@n@bounds_asr_verbatim.cov.ao.overall\endcsname{1\,886}
\expandafter\gdef\csname odunum@ci@bounds_asr_verbatim.cov.ao.overall\endcsname{[0.749, 0.788]}
\expandafter\gdef\csname odunum@val@bounds_asr_verbatim.cov.ao.overall.gen\endcsname{0.768}
\expandafter\gdef\csname odunum@n@bounds_asr_verbatim.cov.ao.overall.gen\endcsname{1\,886}
\expandafter\gdef\csname odunum@ci@bounds_asr_verbatim.cov.ao.overall.gen\endcsname{[0.749, 0.788]}
\expandafter\gdef\csname odunum@val@bounds_asr_verbatim.cov.av.audio\endcsname{0.566}
\expandafter\gdef\csname odunum@n@bounds_asr_verbatim.cov.av.audio\endcsname{558}
\expandafter\gdef\csname odunum@ci@bounds_asr_verbatim.cov.av.audio\endcsname{[0.524, 0.608]}
\expandafter\gdef\csname odunum@val@bounds_asr_verbatim.cov.av.context\endcsname{0.417}
\expandafter\gdef\csname odunum@n@bounds_asr_verbatim.cov.av.context\endcsname{1\,974}
\expandafter\gdef\csname odunum@ci@bounds_asr_verbatim.cov.av.context\endcsname{[0.393, 0.441]}
\expandafter\gdef\csname odunum@val@bounds_asr_verbatim.cov.av.context.gen\endcsname{0.390}
\expandafter\gdef\csname odunum@n@bounds_asr_verbatim.cov.av.context.gen\endcsname{1\,635}
\expandafter\gdef\csname odunum@ci@bounds_asr_verbatim.cov.av.context.gen\endcsname{[0.363, 0.416]}
\expandafter\gdef\csname odunum@val@bounds_asr_verbatim.cov.av.context.latin\endcsname{0.412}
\expandafter\gdef\csname odunum@n@bounds_asr_verbatim.cov.av.context.latin\endcsname{811}
\expandafter\gdef\csname odunum@ci@bounds_asr_verbatim.cov.av.context.latin\endcsname{[0.374, 0.450]}
\expandafter\gdef\csname odunum@val@bounds_asr_verbatim.cov.av.context.lex_absent\endcsname{0.411}
\expandafter\gdef\csname odunum@n@bounds_asr_verbatim.cov.av.context.lex_absent\endcsname{1\,936}
\expandafter\gdef\csname odunum@ci@bounds_asr_verbatim.cov.av.context.lex_absent\endcsname{[0.387, 0.436]}
\expandafter\gdef\csname odunum@val@bounds_asr_verbatim.cov.av.context.lex_absent.latin\endcsname{0.397}
\expandafter\gdef\csname odunum@n@bounds_asr_verbatim.cov.av.context.lex_absent.latin\endcsname{773}
\expandafter\gdef\csname odunum@ci@bounds_asr_verbatim.cov.av.context.lex_absent.latin\endcsname{[0.358, 0.436]}
\expandafter\gdef\csname odunum@val@bounds_asr_verbatim.cov.av.context.lex_present\endcsname{0.711}
\expandafter\gdef\csname odunum@n@bounds_asr_verbatim.cov.av.context.lex_present\endcsname{38}
\expandafter\gdef\csname odunum@ci@bounds_asr_verbatim.cov.av.context.lex_present\endcsname{[0.553, 0.865]}
\expandafter\gdef\csname odunum@val@bounds_asr_verbatim.cov.av.context.lex_present.latin\endcsname{0.711}
\expandafter\gdef\csname odunum@n@bounds_asr_verbatim.cov.av.context.lex_present.latin\endcsname{38}
\expandafter\gdef\csname odunum@ci@bounds_asr_verbatim.cov.av.context.lex_present.latin\endcsname{[0.553, 0.865]}
\expandafter\gdef\csname odunum@val@bounds_asr_verbatim.cov.av.context.recorded\endcsname{0.549}
\expandafter\gdef\csname odunum@n@bounds_asr_verbatim.cov.av.context.recorded\endcsname{339}
\expandafter\gdef\csname odunum@ci@bounds_asr_verbatim.cov.av.context.recorded\endcsname{[0.490, 0.608]}
\expandafter\gdef\csname odunum@val@bounds_asr_verbatim.cov.av.history\endcsname{0.387}
\expandafter\gdef\csname odunum@n@bounds_asr_verbatim.cov.av.history\endcsname{476}
\expandafter\gdef\csname odunum@ci@bounds_asr_verbatim.cov.av.history\endcsname{[0.340, 0.435]}
\expandafter\gdef\csname odunum@val@bounds_asr_verbatim.cov.av.intent\endcsname{0.952}
\expandafter\gdef\csname odunum@n@bounds_asr_verbatim.cov.av.intent\endcsname{1\,986}
\expandafter\gdef\csname odunum@ci@bounds_asr_verbatim.cov.av.intent\endcsname{[0.942, 0.962]}
\expandafter\gdef\csname odunum@val@bounds_asr_verbatim.cov.av.intent.gen\endcsname{0.964}
\expandafter\gdef\csname odunum@n@bounds_asr_verbatim.cov.av.intent.gen\endcsname{1\,624}
\expandafter\gdef\csname odunum@ci@bounds_asr_verbatim.cov.av.intent.gen\endcsname{[0.954, 0.973]}
\expandafter\gdef\csname odunum@val@bounds_asr_verbatim.cov.av.intent.latin\endcsname{0.966}
\expandafter\gdef\csname odunum@n@bounds_asr_verbatim.cov.av.intent.latin\endcsname{849}
\expandafter\gdef\csname odunum@ci@bounds_asr_verbatim.cov.av.intent.latin\endcsname{[0.954, 0.977]}
\expandafter\gdef\csname odunum@val@bounds_asr_verbatim.cov.av.intent.recorded\endcsname{0.898}
\expandafter\gdef\csname odunum@n@bounds_asr_verbatim.cov.av.intent.recorded\endcsname{362}
\expandafter\gdef\csname odunum@ci@bounds_asr_verbatim.cov.av.intent.recorded\endcsname{[0.864, 0.929]}
\expandafter\gdef\csname odunum@val@bounds_asr_verbatim.cov.av.other_context\endcsname{0.441}
\expandafter\gdef\csname odunum@n@bounds_asr_verbatim.cov.av.other_context\endcsname{136}
\expandafter\gdef\csname odunum@ci@bounds_asr_verbatim.cov.av.other_context\endcsname{[0.358, 0.526]}
\expandafter\gdef\csname odunum@val@bounds_asr_verbatim.cov.av.overall\endcsname{0.684}
\expandafter\gdef\csname odunum@n@bounds_asr_verbatim.cov.av.overall\endcsname{3\,973}
\expandafter\gdef\csname odunum@ci@bounds_asr_verbatim.cov.av.overall\endcsname{[0.670, 0.699]}
\expandafter\gdef\csname odunum@val@bounds_asr_verbatim.cov.av.overall.gen\endcsname{0.676}
\expandafter\gdef\csname odunum@n@bounds_asr_verbatim.cov.av.overall.gen\endcsname{3\,259}
\expandafter\gdef\csname odunum@ci@bounds_asr_verbatim.cov.av.overall.gen\endcsname{[0.660, 0.691]}
\expandafter\gdef\csname odunum@val@bounds_asr_verbatim.cov.av.overall.recorded\endcsname{0.724}
\expandafter\gdef\csname odunum@n@bounds_asr_verbatim.cov.av.overall.recorded\endcsname{714}
\expandafter\gdef\csname odunum@ci@bounds_asr_verbatim.cov.av.overall.recorded\endcsname{[0.687, 0.762]}
\expandafter\gdef\csname odunum@val@bounds_asr_verbatim.cov.av.unattributed\endcsname{0.462}
\expandafter\gdef\csname odunum@n@bounds_asr_verbatim.cov.av.unattributed\endcsname{13}
\expandafter\gdef\csname odunum@ci@bounds_asr_verbatim.cov.av.unattributed\endcsname{[0.118, 0.900]}
\expandafter\gdef\csname odunum@val@bounds_asr_verbatim.cov.av.visual\endcsname{0.327}
\expandafter\gdef\csname odunum@n@bounds_asr_verbatim.cov.av.visual\endcsname{804}
\expandafter\gdef\csname odunum@ci@bounds_asr_verbatim.cov.av.visual\endcsname{[0.293, 0.362]}
\expandafter\gdef\csname odunum@val@bounds_asr_verbatim.delta.all.context\endcsname{0.573}
\expandafter\gdef\csname odunum@n@bounds_asr_verbatim.delta.all.context\endcsname{2\,677}
\expandafter\gdef\csname odunum@ci@bounds_asr_verbatim.delta.all.context\endcsname{[0.551, 0.595]}
\expandafter\gdef\csname odunum@val@bounds_asr_verbatim.delta.all.intent\endcsname{0.042}
\expandafter\gdef\csname odunum@n@bounds_asr_verbatim.delta.all.intent\endcsname{3\,108}
\expandafter\gdef\csname odunum@ci@bounds_asr_verbatim.delta.all.intent\endcsname{[0.035, 0.050]}
\expandafter\gdef\csname odunum@val@bounds_asr_verbatim.delta.ao.context\endcsname{0.551}
\expandafter\gdef\csname odunum@n@bounds_asr_verbatim.delta.ao.context\endcsname{721}
\expandafter\gdef\csname odunum@ci@bounds_asr_verbatim.delta.ao.context\endcsname{[0.510, 0.592]}
\expandafter\gdef\csname odunum@val@bounds_asr_verbatim.delta.ao.intent\endcsname{0.032}
\expandafter\gdef\csname odunum@n@bounds_asr_verbatim.delta.ao.intent\endcsname{1\,144}
\expandafter\gdef\csname odunum@ci@bounds_asr_verbatim.delta.ao.intent\endcsname{[0.022, 0.043]}
\expandafter\gdef\csname odunum@val@bounds_asr_verbatim.delta.av.context\endcsname{0.581}
\expandafter\gdef\csname odunum@n@bounds_asr_verbatim.delta.av.context\endcsname{1\,956}
\expandafter\gdef\csname odunum@ci@bounds_asr_verbatim.delta.av.context\endcsname{[0.557, 0.606]}
\expandafter\gdef\csname odunum@val@bounds_asr_verbatim.delta.av.intent\endcsname{0.048}
\expandafter\gdef\csname odunum@n@bounds_asr_verbatim.delta.av.intent\endcsname{1\,964}
\expandafter\gdef\csname odunum@ci@bounds_asr_verbatim.delta.av.intent\endcsname{[0.039, 0.058]}
\expandafter\gdef\csname odunum@val@bounds_asr_verbatim.gap.all.intent_minus_context\endcsname{0.532}
\expandafter\gdef\csname odunum@n@bounds_asr_verbatim.gap.all.intent_minus_context\endcsname{5\,846}
\expandafter\gdef\csname odunum@val@bounds_asr_verbatim.gap.ao.intent_minus_context\endcsname{0.517}
\expandafter\gdef\csname odunum@n@bounds_asr_verbatim.gap.ao.intent_minus_context\endcsname{1\,886}
\expandafter\gdef\csname odunum@val@bounds_asr_verbatim.gap.av.intent_minus_context\endcsname{0.535}
\expandafter\gdef\csname odunum@n@bounds_asr_verbatim.gap.av.intent_minus_context\endcsname{3\,960}
\expandafter\gdef\csname odunum@val@bounds_asr_verbatim.judge.id\endcsname{mr\_ali / dashscope.qwen3.6-flash (thinking)}
\expandafter\gdef\csname odunum@n@bounds_asr_verbatim.judge.id\endcsname{1}
\expandafter\gdef\csname odunum@val@bounds_asr_verbatim.judge.scored_on\endcsname{2026-09-03}
\expandafter\gdef\csname odunum@n@bounds_asr_verbatim.judge.scored_on\endcsname{1}
\expandafter\gdef\csname odunum@val@bounds_asr_verbatim.lex.all.audio.share_half\endcsname{0.027}
\expandafter\gdef\csname odunum@n@bounds_asr_verbatim.lex.all.audio.share_half\endcsname{899}
\expandafter\gdef\csname odunum@val@bounds_asr_verbatim.lex.all.context.share_half\endcsname{0.020}
\expandafter\gdef\csname odunum@n@bounds_asr_verbatim.lex.all.context.share_half\endcsname{2\,702}
\expandafter\gdef\csname odunum@val@bounds_asr_verbatim.lex.all.history.share_half\endcsname{0.022}
\expandafter\gdef\csname odunum@n@bounds_asr_verbatim.lex.all.history.share_half\endcsname{818}
\expandafter\gdef\csname odunum@val@bounds_asr_verbatim.lex.all.intent.share_half\endcsname{0.212}
\expandafter\gdef\csname odunum@n@bounds_asr_verbatim.lex.all.intent.share_half\endcsname{3\,144}
\expandafter\gdef\csname odunum@val@bounds_asr_verbatim.lex.all.visual.share_half\endcsname{0.012}
\expandafter\gdef\csname odunum@n@bounds_asr_verbatim.lex.all.visual.share_half\endcsname{804}
\expandafter\gdef\csname odunum@val@bounds_asr_verbatim.lex.ao.audio.share_half\endcsname{0.026}
\expandafter\gdef\csname odunum@n@bounds_asr_verbatim.lex.ao.audio.share_half\endcsname{341}
\expandafter\gdef\csname odunum@val@bounds_asr_verbatim.lex.ao.context.share_half\endcsname{0.021}
\expandafter\gdef\csname odunum@n@bounds_asr_verbatim.lex.ao.context.share_half\endcsname{728}
\expandafter\gdef\csname odunum@val@bounds_asr_verbatim.lex.ao.history.share_half\endcsname{0.018}
\expandafter\gdef\csname odunum@n@bounds_asr_verbatim.lex.ao.history.share_half\endcsname{342}
\expandafter\gdef\csname odunum@val@bounds_asr_verbatim.lex.ao.intent.share_half\endcsname{0.286}
\expandafter\gdef\csname odunum@n@bounds_asr_verbatim.lex.ao.intent.share_half\endcsname{1\,158}
\expandafter\gdef\csname odunum@val@bounds_asr_verbatim.lex.av.audio.share_half\endcsname{0.027}
\expandafter\gdef\csname odunum@n@bounds_asr_verbatim.lex.av.audio.share_half\endcsname{558}
\expandafter\gdef\csname odunum@val@bounds_asr_verbatim.lex.av.context.share_half\endcsname{0.019}
\expandafter\gdef\csname odunum@n@bounds_asr_verbatim.lex.av.context.share_half\endcsname{1\,974}
\expandafter\gdef\csname odunum@val@bounds_asr_verbatim.lex.av.history.share_half\endcsname{0.025}
\expandafter\gdef\csname odunum@n@bounds_asr_verbatim.lex.av.history.share_half\endcsname{476}
\expandafter\gdef\csname odunum@val@bounds_asr_verbatim.lex.av.intent.share_half\endcsname{0.169}
\expandafter\gdef\csname odunum@n@bounds_asr_verbatim.lex.av.intent.share_half\endcsname{1\,986}
\expandafter\gdef\csname odunum@val@bounds_asr_verbatim.lex.av.visual.share_half\endcsname{0.012}
\expandafter\gdef\csname odunum@n@bounds_asr_verbatim.lex.av.visual.share_half\endcsname{804}
\expandafter\gdef\csname odunum@val@bounds_asr_verbatim.pop.all.n_points\endcsname{5\,859}
\expandafter\gdef\csname odunum@n@bounds_asr_verbatim.pop.all.n_points\endcsname{1\,631}
\expandafter\gdef\csname odunum@val@bounds_asr_verbatim.pop.all.n_scenes\endcsname{1\,631}
\expandafter\gdef\csname odunum@n@bounds_asr_verbatim.pop.all.n_scenes\endcsname{1\,631}
\expandafter\gdef\csname odunum@val@bounds_asr_verbatim.pop.ao.n_points\endcsname{1\,886}
\expandafter\gdef\csname odunum@n@bounds_asr_verbatim.pop.ao.n_points\endcsname{570}
\expandafter\gdef\csname odunum@val@bounds_asr_verbatim.pop.ao.n_scenes\endcsname{570}
\expandafter\gdef\csname odunum@n@bounds_asr_verbatim.pop.ao.n_scenes\endcsname{570}
\expandafter\gdef\csname odunum@val@bounds_asr_verbatim.pop.av.n_points\endcsname{3\,973}
\expandafter\gdef\csname odunum@n@bounds_asr_verbatim.pop.av.n_points\endcsname{1\,061}
\expandafter\gdef\csname odunum@val@bounds_asr_verbatim.pop.av.n_scenes\endcsname{1\,061}
\expandafter\gdef\csname odunum@n@bounds_asr_verbatim.pop.av.n_scenes\endcsname{1\,061}
\expandafter\gdef\csname odunum@val@bounds_oracle.audit.all.judged_points\endcsname{5\,859}
\expandafter\gdef\csname odunum@n@bounds_oracle.audit.all.judged_points\endcsname{5\,859}
\expandafter\gdef\csname odunum@val@bounds_oracle.audit.ao.judged_points\endcsname{1\,886}
\expandafter\gdef\csname odunum@n@bounds_oracle.audit.ao.judged_points\endcsname{1\,886}
\expandafter\gdef\csname odunum@val@bounds_oracle.audit.av.judged_points\endcsname{3\,973}
\expandafter\gdef\csname odunum@n@bounds_oracle.audit.av.judged_points\endcsname{3\,973}
\expandafter\gdef\csname odunum@val@bounds_oracle.audit.point_set_shared_with_other_arm\endcsname{5\,859}
\expandafter\gdef\csname odunum@n@bounds_oracle.audit.point_set_shared_with_other_arm\endcsname{5\,859}
\expandafter\gdef\csname odunum@val@bounds_oracle.audit.points_without_verdict\endcsname{0}
\expandafter\gdef\csname odunum@n@bounds_oracle.audit.points_without_verdict\endcsname{5\,859}
\expandafter\gdef\csname odunum@val@bounds_oracle.audit.reader_matches_keypoint_types\endcsname{5\,859}
\expandafter\gdef\csname odunum@n@bounds_oracle.audit.reader_matches_keypoint_types\endcsname{5\,859}
\expandafter\gdef\csname odunum@val@bounds_oracle.audit.scenes_incomplete\endcsname{0}
\expandafter\gdef\csname odunum@n@bounds_oracle.audit.scenes_incomplete\endcsname{1\,631}
\expandafter\gdef\csname odunum@val@bounds_oracle.audit.scenes_with_full_gt_denominator\endcsname{1\,631}
\expandafter\gdef\csname odunum@n@bounds_oracle.audit.scenes_with_full_gt_denominator\endcsname{1\,631}
\expandafter\gdef\csname odunum@val@bounds_oracle.audit.segments_missed\endcsname{0}
\expandafter\gdef\csname odunum@n@bounds_oracle.audit.segments_missed\endcsname{1\,679}
\expandafter\gdef\csname odunum@val@bounds_oracle.ceiling.all.audio\endcsname{0.998}
\expandafter\gdef\csname odunum@n@bounds_oracle.ceiling.all.audio\endcsname{899}
\expandafter\gdef\csname odunum@ci@bounds_oracle.ceiling.all.audio\endcsname{[0.994, 1.000]}
\expandafter\gdef\csname odunum@val@bounds_oracle.ceiling.all.context\endcsname{0.997}
\expandafter\gdef\csname odunum@n@bounds_oracle.ceiling.all.context\endcsname{2\,702}
\expandafter\gdef\csname odunum@ci@bounds_oracle.ceiling.all.context\endcsname{[0.995, 0.999]}
\expandafter\gdef\csname odunum@val@bounds_oracle.ceiling.all.history\endcsname{0.999}
\expandafter\gdef\csname odunum@n@bounds_oracle.ceiling.all.history\endcsname{818}
\expandafter\gdef\csname odunum@ci@bounds_oracle.ceiling.all.history\endcsname{[0.996, 1.000]}
\expandafter\gdef\csname odunum@val@bounds_oracle.ceiling.all.intent\endcsname{1.000}
\expandafter\gdef\csname odunum@n@bounds_oracle.ceiling.all.intent\endcsname{3\,144}
\expandafter\gdef\csname odunum@ci@bounds_oracle.ceiling.all.intent\endcsname{[0.999, 1.000]}
\expandafter\gdef\csname odunum@val@bounds_oracle.ceiling.all.other_context\endcsname{0.989}
\expandafter\gdef\csname odunum@n@bounds_oracle.ceiling.all.other_context\endcsname{181}
\expandafter\gdef\csname odunum@ci@bounds_oracle.ceiling.all.other_context\endcsname{[0.972, 1.000]}
\expandafter\gdef\csname odunum@val@bounds_oracle.ceiling.all.overall\endcsname{0.998}
\expandafter\gdef\csname odunum@n@bounds_oracle.ceiling.all.overall\endcsname{5\,859}
\expandafter\gdef\csname odunum@ci@bounds_oracle.ceiling.all.overall\endcsname{[0.997, 0.999]}
\expandafter\gdef\csname odunum@val@bounds_oracle.ceiling.all.unattributed\endcsname{1.000}
\expandafter\gdef\csname odunum@n@bounds_oracle.ceiling.all.unattributed\endcsname{13}
\expandafter\gdef\csname odunum@ci@bounds_oracle.ceiling.all.unattributed\endcsname{[1.000, 1.000]}
\expandafter\gdef\csname odunum@val@bounds_oracle.ceiling.all.visual\endcsname{0.996}
\expandafter\gdef\csname odunum@n@bounds_oracle.ceiling.all.visual\endcsname{804}
\expandafter\gdef\csname odunum@ci@bounds_oracle.ceiling.all.visual\endcsname{[0.991, 1.000]}
\expandafter\gdef\csname odunum@val@bounds_oracle.ceiling.ao.audio\endcsname{0.997}
\expandafter\gdef\csname odunum@n@bounds_oracle.ceiling.ao.audio\endcsname{341}
\expandafter\gdef\csname odunum@ci@bounds_oracle.ceiling.ao.audio\endcsname{[0.991, 1.000]}
\expandafter\gdef\csname odunum@val@bounds_oracle.ceiling.ao.context\endcsname{0.999}
\expandafter\gdef\csname odunum@n@bounds_oracle.ceiling.ao.context\endcsname{728}
\expandafter\gdef\csname odunum@ci@bounds_oracle.ceiling.ao.context\endcsname{[0.996, 1.000]}
\expandafter\gdef\csname odunum@val@bounds_oracle.ceiling.ao.history\endcsname{1.000}
\expandafter\gdef\csname odunum@n@bounds_oracle.ceiling.ao.history\endcsname{342}
\expandafter\gdef\csname odunum@ci@bounds_oracle.ceiling.ao.history\endcsname{[1.000, 1.000]}
\expandafter\gdef\csname odunum@val@bounds_oracle.ceiling.ao.intent\endcsname{1.000}
\expandafter\gdef\csname odunum@n@bounds_oracle.ceiling.ao.intent\endcsname{1\,158}
\expandafter\gdef\csname odunum@ci@bounds_oracle.ceiling.ao.intent\endcsname{[1.000, 1.000]}
\expandafter\gdef\csname odunum@val@bounds_oracle.ceiling.ao.other_context\endcsname{1.000}
\expandafter\gdef\csname odunum@n@bounds_oracle.ceiling.ao.other_context\endcsname{45}
\expandafter\gdef\csname odunum@ci@bounds_oracle.ceiling.ao.other_context\endcsname{[1.000, 1.000]}
\expandafter\gdef\csname odunum@val@bounds_oracle.ceiling.ao.overall\endcsname{0.999}
\expandafter\gdef\csname odunum@n@bounds_oracle.ceiling.ao.overall\endcsname{1\,886}
\expandafter\gdef\csname odunum@ci@bounds_oracle.ceiling.ao.overall\endcsname{[0.998, 1.000]}
\expandafter\gdef\csname odunum@val@bounds_oracle.ceiling.av.audio\endcsname{0.998}
\expandafter\gdef\csname odunum@n@bounds_oracle.ceiling.av.audio\endcsname{558}
\expandafter\gdef\csname odunum@ci@bounds_oracle.ceiling.av.audio\endcsname{[0.995, 1.000]}
\expandafter\gdef\csname odunum@val@bounds_oracle.ceiling.av.context\endcsname{0.996}
\expandafter\gdef\csname odunum@n@bounds_oracle.ceiling.av.context\endcsname{1\,974}
\expandafter\gdef\csname odunum@ci@bounds_oracle.ceiling.av.context\endcsname{[0.994, 0.999]}
\expandafter\gdef\csname odunum@val@bounds_oracle.ceiling.av.history\endcsname{0.998}
\expandafter\gdef\csname odunum@n@bounds_oracle.ceiling.av.history\endcsname{476}
\expandafter\gdef\csname odunum@ci@bounds_oracle.ceiling.av.history\endcsname{[0.994, 1.000]}
\expandafter\gdef\csname odunum@val@bounds_oracle.ceiling.av.intent\endcsname{0.999}
\expandafter\gdef\csname odunum@n@bounds_oracle.ceiling.av.intent\endcsname{1\,986}
\expandafter\gdef\csname odunum@ci@bounds_oracle.ceiling.av.intent\endcsname{[0.999, 1.000]}
\expandafter\gdef\csname odunum@val@bounds_oracle.ceiling.av.other_context\endcsname{0.985}
\expandafter\gdef\csname odunum@n@bounds_oracle.ceiling.av.other_context\endcsname{136}
\expandafter\gdef\csname odunum@ci@bounds_oracle.ceiling.av.other_context\endcsname{[0.962, 1.000]}
\expandafter\gdef\csname odunum@val@bounds_oracle.ceiling.av.overall\endcsname{0.998}
\expandafter\gdef\csname odunum@n@bounds_oracle.ceiling.av.overall\endcsname{3\,973}
\expandafter\gdef\csname odunum@ci@bounds_oracle.ceiling.av.overall\endcsname{[0.997, 0.999]}
\expandafter\gdef\csname odunum@val@bounds_oracle.ceiling.av.unattributed\endcsname{1.000}
\expandafter\gdef\csname odunum@n@bounds_oracle.ceiling.av.unattributed\endcsname{13}
\expandafter\gdef\csname odunum@ci@bounds_oracle.ceiling.av.unattributed\endcsname{[1.000, 1.000]}
\expandafter\gdef\csname odunum@val@bounds_oracle.ceiling.av.visual\endcsname{0.996}
\expandafter\gdef\csname odunum@n@bounds_oracle.ceiling.av.visual\endcsname{804}
\expandafter\gdef\csname odunum@ci@bounds_oracle.ceiling.av.visual\endcsname{[0.991, 1.000]}
\expandafter\gdef\csname odunum@val@bounds_oracle.fn_upper.all.audio\endcsname{0.002}
\expandafter\gdef\csname odunum@n@bounds_oracle.fn_upper.all.audio\endcsname{899}
\expandafter\gdef\csname odunum@ci@bounds_oracle.fn_upper.all.audio\endcsname{[0.000, 0.006]}
\expandafter\gdef\csname odunum@val@bounds_oracle.fn_upper.all.context\endcsname{0.003}
\expandafter\gdef\csname odunum@n@bounds_oracle.fn_upper.all.context\endcsname{2\,702}
\expandafter\gdef\csname odunum@ci@bounds_oracle.fn_upper.all.context\endcsname{[0.001, 0.005]}
\expandafter\gdef\csname odunum@val@bounds_oracle.fn_upper.all.history\endcsname{0.001}
\expandafter\gdef\csname odunum@n@bounds_oracle.fn_upper.all.history\endcsname{818}
\expandafter\gdef\csname odunum@ci@bounds_oracle.fn_upper.all.history\endcsname{[0.000, 0.004]}
\expandafter\gdef\csname odunum@val@bounds_oracle.fn_upper.all.intent\endcsname{3.2\ensuremath{\times 10^{-4}}}
\expandafter\gdef\csname odunum@n@bounds_oracle.fn_upper.all.intent\endcsname{3\,144}
\expandafter\gdef\csname odunum@ci@bounds_oracle.fn_upper.all.intent\endcsname{[0.000, 0.001]}
\expandafter\gdef\csname odunum@val@bounds_oracle.fn_upper.all.other_context\endcsname{0.011}
\expandafter\gdef\csname odunum@n@bounds_oracle.fn_upper.all.other_context\endcsname{181}
\expandafter\gdef\csname odunum@ci@bounds_oracle.fn_upper.all.other_context\endcsname{[0.000, 0.028]}
\expandafter\gdef\csname odunum@val@bounds_oracle.fn_upper.all.unattributed\endcsname{0.000}
\expandafter\gdef\csname odunum@n@bounds_oracle.fn_upper.all.unattributed\endcsname{13}
\expandafter\gdef\csname odunum@ci@bounds_oracle.fn_upper.all.unattributed\endcsname{[0.000, 0.000]}
\expandafter\gdef\csname odunum@val@bounds_oracle.fn_upper.all.visual\endcsname{0.004}
\expandafter\gdef\csname odunum@n@bounds_oracle.fn_upper.all.visual\endcsname{804}
\expandafter\gdef\csname odunum@ci@bounds_oracle.fn_upper.all.visual\endcsname{[0.000, 0.009]}
\expandafter\gdef\csname odunum@val@bounds_oracle.fn_upper.ao.audio\endcsname{0.003}
\expandafter\gdef\csname odunum@n@bounds_oracle.fn_upper.ao.audio\endcsname{341}
\expandafter\gdef\csname odunum@ci@bounds_oracle.fn_upper.ao.audio\endcsname{[0.000, 0.009]}
\expandafter\gdef\csname odunum@val@bounds_oracle.fn_upper.ao.context\endcsname{0.001}
\expandafter\gdef\csname odunum@n@bounds_oracle.fn_upper.ao.context\endcsname{728}
\expandafter\gdef\csname odunum@ci@bounds_oracle.fn_upper.ao.context\endcsname{[0.000, 0.004]}
\expandafter\gdef\csname odunum@val@bounds_oracle.fn_upper.ao.history\endcsname{0.000}
\expandafter\gdef\csname odunum@n@bounds_oracle.fn_upper.ao.history\endcsname{342}
\expandafter\gdef\csname odunum@ci@bounds_oracle.fn_upper.ao.history\endcsname{[0.000, 0.000]}
\expandafter\gdef\csname odunum@val@bounds_oracle.fn_upper.ao.intent\endcsname{0.000}
\expandafter\gdef\csname odunum@n@bounds_oracle.fn_upper.ao.intent\endcsname{1\,158}
\expandafter\gdef\csname odunum@ci@bounds_oracle.fn_upper.ao.intent\endcsname{[0.000, 0.000]}
\expandafter\gdef\csname odunum@val@bounds_oracle.fn_upper.ao.other_context\endcsname{0.000}
\expandafter\gdef\csname odunum@n@bounds_oracle.fn_upper.ao.other_context\endcsname{45}
\expandafter\gdef\csname odunum@ci@bounds_oracle.fn_upper.ao.other_context\endcsname{[0.000, 0.000]}
\expandafter\gdef\csname odunum@val@bounds_oracle.fn_upper.av.audio\endcsname{0.002}
\expandafter\gdef\csname odunum@n@bounds_oracle.fn_upper.av.audio\endcsname{558}
\expandafter\gdef\csname odunum@ci@bounds_oracle.fn_upper.av.audio\endcsname{[0.000, 0.005]}
\expandafter\gdef\csname odunum@val@bounds_oracle.fn_upper.av.context\endcsname{0.004}
\expandafter\gdef\csname odunum@n@bounds_oracle.fn_upper.av.context\endcsname{1\,974}
\expandafter\gdef\csname odunum@ci@bounds_oracle.fn_upper.av.context\endcsname{[0.001, 0.006]}
\expandafter\gdef\csname odunum@val@bounds_oracle.fn_upper.av.history\endcsname{0.002}
\expandafter\gdef\csname odunum@n@bounds_oracle.fn_upper.av.history\endcsname{476}
\expandafter\gdef\csname odunum@ci@bounds_oracle.fn_upper.av.history\endcsname{[0.000, 0.006]}
\expandafter\gdef\csname odunum@val@bounds_oracle.fn_upper.av.intent\endcsname{0.001}
\expandafter\gdef\csname odunum@n@bounds_oracle.fn_upper.av.intent\endcsname{1\,986}
\expandafter\gdef\csname odunum@ci@bounds_oracle.fn_upper.av.intent\endcsname{[0.000, 0.002]}
\expandafter\gdef\csname odunum@val@bounds_oracle.fn_upper.av.other_context\endcsname{0.015}
\expandafter\gdef\csname odunum@n@bounds_oracle.fn_upper.av.other_context\endcsname{136}
\expandafter\gdef\csname odunum@ci@bounds_oracle.fn_upper.av.other_context\endcsname{[0.000, 0.038]}
\expandafter\gdef\csname odunum@val@bounds_oracle.fn_upper.av.unattributed\endcsname{0.000}
\expandafter\gdef\csname odunum@n@bounds_oracle.fn_upper.av.unattributed\endcsname{13}
\expandafter\gdef\csname odunum@ci@bounds_oracle.fn_upper.av.unattributed\endcsname{[0.000, 0.000]}
\expandafter\gdef\csname odunum@val@bounds_oracle.fn_upper.av.visual\endcsname{0.004}
\expandafter\gdef\csname odunum@n@bounds_oracle.fn_upper.av.visual\endcsname{804}
\expandafter\gdef\csname odunum@ci@bounds_oracle.fn_upper.av.visual\endcsname{[0.000, 0.009]}
\expandafter\gdef\csname odunum@val@bounds_oracle.judge.id\endcsname{mr\_ali / dashscope.qwen3.6-flash (thinking)}
\expandafter\gdef\csname odunum@n@bounds_oracle.judge.id\endcsname{1}
\expandafter\gdef\csname odunum@val@bounds_oracle.judge.scored_on\endcsname{2026-09-03}
\expandafter\gdef\csname odunum@n@bounds_oracle.judge.scored_on\endcsname{1}
\expandafter\gdef\csname odunum@val@bounds_oracle.pop.all.n_points\endcsname{5\,859}
\expandafter\gdef\csname odunum@n@bounds_oracle.pop.all.n_points\endcsname{1\,631}
\expandafter\gdef\csname odunum@val@bounds_oracle.pop.all.n_scenes\endcsname{1\,631}
\expandafter\gdef\csname odunum@n@bounds_oracle.pop.all.n_scenes\endcsname{1\,631}
\expandafter\gdef\csname odunum@val@bounds_oracle.pop.ao.n_points\endcsname{1\,886}
\expandafter\gdef\csname odunum@n@bounds_oracle.pop.ao.n_points\endcsname{570}
\expandafter\gdef\csname odunum@val@bounds_oracle.pop.ao.n_scenes\endcsname{570}
\expandafter\gdef\csname odunum@n@bounds_oracle.pop.ao.n_scenes\endcsname{570}
\expandafter\gdef\csname odunum@val@bounds_oracle.pop.av.n_points\endcsname{3\,973}
\expandafter\gdef\csname odunum@n@bounds_oracle.pop.av.n_points\endcsname{1\,061}
\expandafter\gdef\csname odunum@val@bounds_oracle.pop.av.n_scenes\endcsname{1\,061}
\expandafter\gdef\csname odunum@n@bounds_oracle.pop.av.n_scenes\endcsname{1\,061}
\expandafter\gdef\csname odunum@val@bounds_trivial.avg.always_demand.all\endcsname{0.066}
\expandafter\gdef\csname odunum@n@bounds_trivial.avg.always_demand.all\endcsname{2\,078}
\expandafter\gdef\csname odunum@val@bounds_trivial.avg.always_demand.ao\endcsname{0.066}
\expandafter\gdef\csname odunum@n@bounds_trivial.avg.always_demand.ao\endcsname{731}
\expandafter\gdef\csname odunum@val@bounds_trivial.avg.always_demand.av\endcsname{0.066}
\expandafter\gdef\csname odunum@n@bounds_trivial.avg.always_demand.av\endcsname{1\,347}
\expandafter\gdef\csname odunum@val@bounds_trivial.avg.always_no_demand.all\endcsname{0.027}
\expandafter\gdef\csname odunum@n@bounds_trivial.avg.always_no_demand.all\endcsname{2\,078}
\expandafter\gdef\csname odunum@val@bounds_trivial.avg.always_no_demand.ao\endcsname{0.027}
\expandafter\gdef\csname odunum@n@bounds_trivial.avg.always_no_demand.ao\endcsname{731}
\expandafter\gdef\csname odunum@val@bounds_trivial.avg.always_no_demand.av\endcsname{0.026}
\expandafter\gdef\csname odunum@n@bounds_trivial.avg.always_no_demand.av\endcsname{1\,347}
\expandafter\gdef\csname odunum@val@bounds_trivial.ftr.always_demand.all\endcsname{1.000}
\expandafter\gdef\csname odunum@n@bounds_trivial.ftr.always_demand.all\endcsname{447}
\expandafter\gdef\csname odunum@val@bounds_trivial.ftr.always_demand.ao\endcsname{1.000}
\expandafter\gdef\csname odunum@n@bounds_trivial.ftr.always_demand.ao\endcsname{161}
\expandafter\gdef\csname odunum@val@bounds_trivial.ftr.always_demand.av\endcsname{1.000}
\expandafter\gdef\csname odunum@n@bounds_trivial.ftr.always_demand.av\endcsname{286}
\expandafter\gdef\csname odunum@val@bounds_trivial.ftr.always_no_demand.all\endcsname{0.000}
\expandafter\gdef\csname odunum@n@bounds_trivial.ftr.always_no_demand.all\endcsname{447}
\expandafter\gdef\csname odunum@val@bounds_trivial.ftr.always_no_demand.ao\endcsname{0.000}
\expandafter\gdef\csname odunum@n@bounds_trivial.ftr.always_no_demand.ao\endcsname{161}
\expandafter\gdef\csname odunum@val@bounds_trivial.ftr.always_no_demand.av\endcsname{0.000}
\expandafter\gdef\csname odunum@n@bounds_trivial.ftr.always_no_demand.av\endcsname{286}
\expandafter\gdef\csname odunum@val@bounds_trivial.m1.always_demand.all\endcsname{0.440}
\expandafter\gdef\csname odunum@n@bounds_trivial.m1.always_demand.all\endcsname{2\,078}
\expandafter\gdef\csname odunum@val@bounds_trivial.m1.always_demand.ao\endcsname{0.438}
\expandafter\gdef\csname odunum@n@bounds_trivial.m1.always_demand.ao\endcsname{731}
\expandafter\gdef\csname odunum@val@bounds_trivial.m1.always_demand.av\endcsname{0.441}
\expandafter\gdef\csname odunum@n@bounds_trivial.m1.always_demand.av\endcsname{1\,347}
\expandafter\gdef\csname odunum@val@bounds_trivial.m1.always_no_demand.all\endcsname{0.177}
\expandafter\gdef\csname odunum@n@bounds_trivial.m1.always_no_demand.all\endcsname{2\,078}
\expandafter\gdef\csname odunum@val@bounds_trivial.m1.always_no_demand.ao\endcsname{0.180}
\expandafter\gdef\csname odunum@n@bounds_trivial.m1.always_no_demand.ao\endcsname{731}
\expandafter\gdef\csname odunum@val@bounds_trivial.m1.always_no_demand.av\endcsname{0.175}
\expandafter\gdef\csname odunum@n@bounds_trivial.m1.always_no_demand.av\endcsname{1\,347}
\expandafter\gdef\csname odunum@val@bounds_trivial.m2.always_demand.all\endcsname{0.000}
\expandafter\gdef\csname odunum@n@bounds_trivial.m2.always_demand.all\endcsname{1\,631}
\expandafter\gdef\csname odunum@val@bounds_trivial.m2.always_demand.ao\endcsname{0.000}
\expandafter\gdef\csname odunum@n@bounds_trivial.m2.always_demand.ao\endcsname{570}
\expandafter\gdef\csname odunum@val@bounds_trivial.m2.always_demand.av\endcsname{0.000}
\expandafter\gdef\csname odunum@n@bounds_trivial.m2.always_demand.av\endcsname{1\,061}
\expandafter\gdef\csname odunum@val@bounds_trivial.m2.always_no_demand.all\endcsname{0.000}
\expandafter\gdef\csname odunum@n@bounds_trivial.m2.always_no_demand.all\endcsname{1\,631}
\expandafter\gdef\csname odunum@val@bounds_trivial.m2.always_no_demand.ao\endcsname{0.000}
\expandafter\gdef\csname odunum@n@bounds_trivial.m2.always_no_demand.ao\endcsname{570}
\expandafter\gdef\csname odunum@val@bounds_trivial.m2.always_no_demand.av\endcsname{0.000}
\expandafter\gdef\csname odunum@n@bounds_trivial.m2.always_no_demand.av\endcsname{1\,061}
\expandafter\gdef\csname odunum@val@bounds_trivial.m3.always_demand.all\endcsname{0.000}
\expandafter\gdef\csname odunum@n@bounds_trivial.m3.always_demand.all\endcsname{1\,631}
\expandafter\gdef\csname odunum@val@bounds_trivial.m3.always_demand.ao\endcsname{0.000}
\expandafter\gdef\csname odunum@n@bounds_trivial.m3.always_demand.ao\endcsname{570}
\expandafter\gdef\csname odunum@val@bounds_trivial.m3.always_demand.av\endcsname{0.000}
\expandafter\gdef\csname odunum@n@bounds_trivial.m3.always_demand.av\endcsname{1\,061}
\expandafter\gdef\csname odunum@val@bounds_trivial.m3.always_no_demand.all\endcsname{0.000}
\expandafter\gdef\csname odunum@n@bounds_trivial.m3.always_no_demand.all\endcsname{1\,631}
\expandafter\gdef\csname odunum@val@bounds_trivial.m3.always_no_demand.ao\endcsname{0.000}
\expandafter\gdef\csname odunum@n@bounds_trivial.m3.always_no_demand.ao\endcsname{570}
\expandafter\gdef\csname odunum@val@bounds_trivial.m3.always_no_demand.av\endcsname{0.000}
\expandafter\gdef\csname odunum@n@bounds_trivial.m3.always_no_demand.av\endcsname{1\,061}
\expandafter\gdef\csname odunum@val@bounds_trivial.m4.always_demand.all\endcsname{0.000}
\expandafter\gdef\csname odunum@n@bounds_trivial.m4.always_demand.all\endcsname{1\,631}
\expandafter\gdef\csname odunum@val@bounds_trivial.m4.always_demand.ao\endcsname{0.000}
\expandafter\gdef\csname odunum@n@bounds_trivial.m4.always_demand.ao\endcsname{570}
\expandafter\gdef\csname odunum@val@bounds_trivial.m4.always_demand.av\endcsname{0.000}
\expandafter\gdef\csname odunum@n@bounds_trivial.m4.always_demand.av\endcsname{1\,061}
\expandafter\gdef\csname odunum@val@bounds_trivial.m4.always_no_demand.all\endcsname{0.000}
\expandafter\gdef\csname odunum@n@bounds_trivial.m4.always_no_demand.all\endcsname{1\,631}
\expandafter\gdef\csname odunum@val@bounds_trivial.m4.always_no_demand.ao\endcsname{0.000}
\expandafter\gdef\csname odunum@n@bounds_trivial.m4.always_no_demand.ao\endcsname{570}
\expandafter\gdef\csname odunum@val@bounds_trivial.m4.always_no_demand.av\endcsname{0.000}
\expandafter\gdef\csname odunum@n@bounds_trivial.m4.always_no_demand.av\endcsname{1\,061}
\expandafter\gdef\csname odunum@val@bounds_trivial.m5.always_demand.all\endcsname{0.000}
\expandafter\gdef\csname odunum@n@bounds_trivial.m5.always_demand.all\endcsname{1\,631}
\expandafter\gdef\csname odunum@val@bounds_trivial.m5.always_demand.ao\endcsname{0.000}
\expandafter\gdef\csname odunum@n@bounds_trivial.m5.always_demand.ao\endcsname{570}
\expandafter\gdef\csname odunum@val@bounds_trivial.m5.always_demand.av\endcsname{0.000}
\expandafter\gdef\csname odunum@n@bounds_trivial.m5.always_demand.av\endcsname{1\,061}
\expandafter\gdef\csname odunum@val@bounds_trivial.m5.always_no_demand.all\endcsname{0.000}
\expandafter\gdef\csname odunum@n@bounds_trivial.m5.always_no_demand.all\endcsname{1\,631}
\expandafter\gdef\csname odunum@val@bounds_trivial.m5.always_no_demand.ao\endcsname{0.000}
\expandafter\gdef\csname odunum@n@bounds_trivial.m5.always_no_demand.ao\endcsname{570}
\expandafter\gdef\csname odunum@val@bounds_trivial.m5.always_no_demand.av\endcsname{0.000}
\expandafter\gdef\csname odunum@n@bounds_trivial.m5.always_no_demand.av\endcsname{1\,061}
\expandafter\gdef\csname odunum@val@bounds_trivial.pop.all.n\endcsname{2\,078}
\expandafter\gdef\csname odunum@n@bounds_trivial.pop.all.n\endcsname{2\,078}
\expandafter\gdef\csname odunum@val@bounds_trivial.pop.all.n_neg\endcsname{447}
\expandafter\gdef\csname odunum@n@bounds_trivial.pop.all.n_neg\endcsname{2\,078}
\expandafter\gdef\csname odunum@val@bounds_trivial.pop.all.n_pos\endcsname{1\,631}
\expandafter\gdef\csname odunum@n@bounds_trivial.pop.all.n_pos\endcsname{2\,078}
\expandafter\gdef\csname odunum@val@bounds_trivial.pop.ao.n\endcsname{731}
\expandafter\gdef\csname odunum@n@bounds_trivial.pop.ao.n\endcsname{731}
\expandafter\gdef\csname odunum@val@bounds_trivial.pop.ao.n_neg\endcsname{161}
\expandafter\gdef\csname odunum@n@bounds_trivial.pop.ao.n_neg\endcsname{731}
\expandafter\gdef\csname odunum@val@bounds_trivial.pop.ao.n_pos\endcsname{570}
\expandafter\gdef\csname odunum@n@bounds_trivial.pop.ao.n_pos\endcsname{731}
\expandafter\gdef\csname odunum@val@bounds_trivial.pop.av.n\endcsname{1\,347}
\expandafter\gdef\csname odunum@n@bounds_trivial.pop.av.n\endcsname{1\,347}
\expandafter\gdef\csname odunum@val@bounds_trivial.pop.av.n_neg\endcsname{286}
\expandafter\gdef\csname odunum@n@bounds_trivial.pop.av.n_neg\endcsname{1\,347}
\expandafter\gdef\csname odunum@val@bounds_trivial.pop.av.n_pos\endcsname{1\,061}
\expandafter\gdef\csname odunum@n@bounds_trivial.pop.av.n_pos\endcsname{1\,347}
\expandafter\gdef\csname odunum@val@bounds_utterance_copy.audit.all.judged_points\endcsname{5\,859}
\expandafter\gdef\csname odunum@n@bounds_utterance_copy.audit.all.judged_points\endcsname{5\,859}
\expandafter\gdef\csname odunum@val@bounds_utterance_copy.audit.ao.judged_points\endcsname{1\,886}
\expandafter\gdef\csname odunum@n@bounds_utterance_copy.audit.ao.judged_points\endcsname{1\,886}
\expandafter\gdef\csname odunum@val@bounds_utterance_copy.audit.av.judged_points\endcsname{3\,973}
\expandafter\gdef\csname odunum@n@bounds_utterance_copy.audit.av.judged_points\endcsname{3\,973}
\expandafter\gdef\csname odunum@val@bounds_utterance_copy.audit.point_set_shared_with_other_arm\endcsname{5\,859}
\expandafter\gdef\csname odunum@n@bounds_utterance_copy.audit.point_set_shared_with_other_arm\endcsname{5\,859}
\expandafter\gdef\csname odunum@val@bounds_utterance_copy.audit.points_without_verdict\endcsname{0}
\expandafter\gdef\csname odunum@n@bounds_utterance_copy.audit.points_without_verdict\endcsname{5\,859}
\expandafter\gdef\csname odunum@val@bounds_utterance_copy.audit.reader_matches_keypoint_types\endcsname{5\,859}
\expandafter\gdef\csname odunum@n@bounds_utterance_copy.audit.reader_matches_keypoint_types\endcsname{5\,859}
\expandafter\gdef\csname odunum@val@bounds_utterance_copy.audit.scenes_incomplete\endcsname{0}
\expandafter\gdef\csname odunum@n@bounds_utterance_copy.audit.scenes_incomplete\endcsname{1\,631}
\expandafter\gdef\csname odunum@val@bounds_utterance_copy.audit.scenes_with_full_gt_denominator\endcsname{1\,631}
\expandafter\gdef\csname odunum@n@bounds_utterance_copy.audit.scenes_with_full_gt_denominator\endcsname{1\,631}
\expandafter\gdef\csname odunum@val@bounds_utterance_copy.audit.segments_missed\endcsname{0}
\expandafter\gdef\csname odunum@n@bounds_utterance_copy.audit.segments_missed\endcsname{1\,679}
\expandafter\gdef\csname odunum@val@bounds_utterance_copy.cov.all.audio\endcsname{0.452}
\expandafter\gdef\csname odunum@n@bounds_utterance_copy.cov.all.audio\endcsname{899}
\expandafter\gdef\csname odunum@ci@bounds_utterance_copy.cov.all.audio\endcsname{[0.419, 0.485]}
\expandafter\gdef\csname odunum@val@bounds_utterance_copy.cov.all.context\endcsname{0.389}
\expandafter\gdef\csname odunum@n@bounds_utterance_copy.cov.all.context\endcsname{2\,702}
\expandafter\gdef\csname odunum@ci@bounds_utterance_copy.cov.all.context\endcsname{[0.369, 0.409]}
\expandafter\gdef\csname odunum@val@bounds_utterance_copy.cov.all.context.gen\endcsname{0.380}
\expandafter\gdef\csname odunum@n@bounds_utterance_copy.cov.all.context.gen\endcsname{2\,363}
\expandafter\gdef\csname odunum@ci@bounds_utterance_copy.cov.all.context.gen\endcsname{[0.358, 0.402]}
\expandafter\gdef\csname odunum@val@bounds_utterance_copy.cov.all.context.latin\endcsname{0.386}
\expandafter\gdef\csname odunum@n@bounds_utterance_copy.cov.all.context.latin\endcsname{1\,180}
\expandafter\gdef\csname odunum@ci@bounds_utterance_copy.cov.all.context.latin\endcsname{[0.354, 0.418]}
\expandafter\gdef\csname odunum@val@bounds_utterance_copy.cov.all.context.lex_absent\endcsname{0.384}
\expandafter\gdef\csname odunum@n@bounds_utterance_copy.cov.all.context.lex_absent\endcsname{2\,649}
\expandafter\gdef\csname odunum@ci@bounds_utterance_copy.cov.all.context.lex_absent\endcsname{[0.364, 0.404]}
\expandafter\gdef\csname odunum@val@bounds_utterance_copy.cov.all.context.lex_absent.latin\endcsname{0.373}
\expandafter\gdef\csname odunum@n@bounds_utterance_copy.cov.all.context.lex_absent.latin\endcsname{1\,127}
\expandafter\gdef\csname odunum@ci@bounds_utterance_copy.cov.all.context.lex_absent.latin\endcsname{[0.341, 0.404]}
\expandafter\gdef\csname odunum@val@bounds_utterance_copy.cov.all.context.lex_present\endcsname{0.660}
\expandafter\gdef\csname odunum@n@bounds_utterance_copy.cov.all.context.lex_present\endcsname{53}
\expandafter\gdef\csname odunum@ci@bounds_utterance_copy.cov.all.context.lex_present\endcsname{[0.527, 0.792]}
\expandafter\gdef\csname odunum@val@bounds_utterance_copy.cov.all.context.lex_present.latin\endcsname{0.660}
\expandafter\gdef\csname odunum@n@bounds_utterance_copy.cov.all.context.lex_present.latin\endcsname{53}
\expandafter\gdef\csname odunum@ci@bounds_utterance_copy.cov.all.context.lex_present.latin\endcsname{[0.527, 0.792]}
\expandafter\gdef\csname odunum@val@bounds_utterance_copy.cov.all.context.recorded\endcsname{0.454}
\expandafter\gdef\csname odunum@n@bounds_utterance_copy.cov.all.context.recorded\endcsname{339}
\expandafter\gdef\csname odunum@ci@bounds_utterance_copy.cov.all.context.recorded\endcsname{[0.395, 0.513]}
\expandafter\gdef\csname odunum@val@bounds_utterance_copy.cov.all.history\endcsname{0.367}
\expandafter\gdef\csname odunum@n@bounds_utterance_copy.cov.all.history\endcsname{818}
\expandafter\gdef\csname odunum@ci@bounds_utterance_copy.cov.all.history\endcsname{[0.330, 0.403]}
\expandafter\gdef\csname odunum@val@bounds_utterance_copy.cov.all.intent\endcsname{0.970}
\expandafter\gdef\csname odunum@n@bounds_utterance_copy.cov.all.intent\endcsname{3\,144}
\expandafter\gdef\csname odunum@ci@bounds_utterance_copy.cov.all.intent\endcsname{[0.964, 0.976]}
\expandafter\gdef\csname odunum@val@bounds_utterance_copy.cov.all.intent.gen\endcsname{0.975}
\expandafter\gdef\csname odunum@n@bounds_utterance_copy.cov.all.intent.gen\endcsname{2\,782}
\expandafter\gdef\csname odunum@ci@bounds_utterance_copy.cov.all.intent.gen\endcsname{[0.969, 0.981]}
\expandafter\gdef\csname odunum@val@bounds_utterance_copy.cov.all.intent.latin\endcsname{0.979}
\expandafter\gdef\csname odunum@n@bounds_utterance_copy.cov.all.intent.latin\endcsname{1\,444}
\expandafter\gdef\csname odunum@ci@bounds_utterance_copy.cov.all.intent.latin\endcsname{[0.971, 0.986]}
\expandafter\gdef\csname odunum@val@bounds_utterance_copy.cov.all.intent.recorded\endcsname{0.931}
\expandafter\gdef\csname odunum@n@bounds_utterance_copy.cov.all.intent.recorded\endcsname{362}
\expandafter\gdef\csname odunum@ci@bounds_utterance_copy.cov.all.intent.recorded\endcsname{[0.903, 0.956]}
\expandafter\gdef\csname odunum@val@bounds_utterance_copy.cov.all.other_context\endcsname{0.497}
\expandafter\gdef\csname odunum@n@bounds_utterance_copy.cov.all.other_context\endcsname{181}
\expandafter\gdef\csname odunum@ci@bounds_utterance_copy.cov.all.other_context\endcsname{[0.425, 0.570]}
\expandafter\gdef\csname odunum@val@bounds_utterance_copy.cov.all.overall\endcsname{0.701}
\expandafter\gdef\csname odunum@n@bounds_utterance_copy.cov.all.overall\endcsname{5\,859}
\expandafter\gdef\csname odunum@ci@bounds_utterance_copy.cov.all.overall\endcsname{[0.690, 0.712]}
\expandafter\gdef\csname odunum@val@bounds_utterance_copy.cov.all.overall.gen\endcsname{0.702}
\expandafter\gdef\csname odunum@n@bounds_utterance_copy.cov.all.overall.gen\endcsname{5\,145}
\expandafter\gdef\csname odunum@ci@bounds_utterance_copy.cov.all.overall.gen\endcsname{[0.690, 0.714]}
\expandafter\gdef\csname odunum@val@bounds_utterance_copy.cov.all.overall.recorded\endcsname{0.695}
\expandafter\gdef\csname odunum@n@bounds_utterance_copy.cov.all.overall.recorded\endcsname{714}
\expandafter\gdef\csname odunum@ci@bounds_utterance_copy.cov.all.overall.recorded\endcsname{[0.661, 0.729]}
\expandafter\gdef\csname odunum@val@bounds_utterance_copy.cov.all.unattributed\endcsname{0.385}
\expandafter\gdef\csname odunum@n@bounds_utterance_copy.cov.all.unattributed\endcsname{13}
\expandafter\gdef\csname odunum@ci@bounds_utterance_copy.cov.all.unattributed\endcsname{[0.050, 0.786]}
\expandafter\gdef\csname odunum@val@bounds_utterance_copy.cov.all.visual\endcsname{0.318}
\expandafter\gdef\csname odunum@n@bounds_utterance_copy.cov.all.visual\endcsname{804}
\expandafter\gdef\csname odunum@ci@bounds_utterance_copy.cov.all.visual\endcsname{[0.284, 0.353]}
\expandafter\gdef\csname odunum@val@bounds_utterance_copy.cov.ao.audio\endcsname{0.443}
\expandafter\gdef\csname odunum@n@bounds_utterance_copy.cov.ao.audio\endcsname{341}
\expandafter\gdef\csname odunum@ci@bounds_utterance_copy.cov.ao.audio\endcsname{[0.390, 0.497]}
\expandafter\gdef\csname odunum@val@bounds_utterance_copy.cov.ao.context\endcsname{0.426}
\expandafter\gdef\csname odunum@n@bounds_utterance_copy.cov.ao.context\endcsname{728}
\expandafter\gdef\csname odunum@ci@bounds_utterance_copy.cov.ao.context\endcsname{[0.387, 0.464]}
\expandafter\gdef\csname odunum@val@bounds_utterance_copy.cov.ao.context.gen\endcsname{0.426}
\expandafter\gdef\csname odunum@n@bounds_utterance_copy.cov.ao.context.gen\endcsname{728}
\expandafter\gdef\csname odunum@ci@bounds_utterance_copy.cov.ao.context.gen\endcsname{[0.387, 0.464]}
\expandafter\gdef\csname odunum@val@bounds_utterance_copy.cov.ao.context.latin\endcsname{0.423}
\expandafter\gdef\csname odunum@n@bounds_utterance_copy.cov.ao.context.latin\endcsname{369}
\expandafter\gdef\csname odunum@ci@bounds_utterance_copy.cov.ao.context.latin\endcsname{[0.367, 0.480]}
\expandafter\gdef\csname odunum@val@bounds_utterance_copy.cov.ao.context.lex_absent\endcsname{0.422}
\expandafter\gdef\csname odunum@n@bounds_utterance_copy.cov.ao.context.lex_absent\endcsname{713}
\expandafter\gdef\csname odunum@ci@bounds_utterance_copy.cov.ao.context.lex_absent\endcsname{[0.383, 0.461]}
\expandafter\gdef\csname odunum@val@bounds_utterance_copy.cov.ao.context.lex_absent.latin\endcsname{0.415}
\expandafter\gdef\csname odunum@n@bounds_utterance_copy.cov.ao.context.lex_absent.latin\endcsname{354}
\expandafter\gdef\csname odunum@ci@bounds_utterance_copy.cov.ao.context.lex_absent.latin\endcsname{[0.358, 0.473]}
\expandafter\gdef\csname odunum@val@bounds_utterance_copy.cov.ao.context.lex_present\endcsname{0.600}
\expandafter\gdef\csname odunum@n@bounds_utterance_copy.cov.ao.context.lex_present\endcsname{15}
\expandafter\gdef\csname odunum@ci@bounds_utterance_copy.cov.ao.context.lex_present\endcsname{[0.333, 0.867]}
\expandafter\gdef\csname odunum@val@bounds_utterance_copy.cov.ao.context.lex_present.latin\endcsname{0.600}
\expandafter\gdef\csname odunum@n@bounds_utterance_copy.cov.ao.context.lex_present.latin\endcsname{15}
\expandafter\gdef\csname odunum@ci@bounds_utterance_copy.cov.ao.context.lex_present.latin\endcsname{[0.333, 0.867]}
\expandafter\gdef\csname odunum@val@bounds_utterance_copy.cov.ao.history\endcsname{0.383}
\expandafter\gdef\csname odunum@n@bounds_utterance_copy.cov.ao.history\endcsname{342}
\expandafter\gdef\csname odunum@ci@bounds_utterance_copy.cov.ao.history\endcsname{[0.325, 0.443]}
\expandafter\gdef\csname odunum@val@bounds_utterance_copy.cov.ao.intent\endcsname{0.978}
\expandafter\gdef\csname odunum@n@bounds_utterance_copy.cov.ao.intent\endcsname{1\,158}
\expandafter\gdef\csname odunum@ci@bounds_utterance_copy.cov.ao.intent\endcsname{[0.969, 0.987]}
\expandafter\gdef\csname odunum@val@bounds_utterance_copy.cov.ao.intent.gen\endcsname{0.978}
\expandafter\gdef\csname odunum@n@bounds_utterance_copy.cov.ao.intent.gen\endcsname{1\,158}
\expandafter\gdef\csname odunum@ci@bounds_utterance_copy.cov.ao.intent.gen\endcsname{[0.969, 0.987]}
\expandafter\gdef\csname odunum@val@bounds_utterance_copy.cov.ao.intent.latin\endcsname{0.975}
\expandafter\gdef\csname odunum@n@bounds_utterance_copy.cov.ao.intent.latin\endcsname{595}
\expandafter\gdef\csname odunum@ci@bounds_utterance_copy.cov.ao.intent.latin\endcsname{[0.961, 0.987]}
\expandafter\gdef\csname odunum@val@bounds_utterance_copy.cov.ao.other_context\endcsname{0.622}
\expandafter\gdef\csname odunum@n@bounds_utterance_copy.cov.ao.other_context\endcsname{45}
\expandafter\gdef\csname odunum@ci@bounds_utterance_copy.cov.ao.other_context\endcsname{[0.477, 0.756]}
\expandafter\gdef\csname odunum@val@bounds_utterance_copy.cov.ao.overall\endcsname{0.765}
\expandafter\gdef\csname odunum@n@bounds_utterance_copy.cov.ao.overall\endcsname{1\,886}
\expandafter\gdef\csname odunum@ci@bounds_utterance_copy.cov.ao.overall\endcsname{[0.746, 0.783]}
\expandafter\gdef\csname odunum@val@bounds_utterance_copy.cov.ao.overall.gen\endcsname{0.765}
\expandafter\gdef\csname odunum@n@bounds_utterance_copy.cov.ao.overall.gen\endcsname{1\,886}
\expandafter\gdef\csname odunum@ci@bounds_utterance_copy.cov.ao.overall.gen\endcsname{[0.746, 0.783]}
\expandafter\gdef\csname odunum@val@bounds_utterance_copy.cov.av.audio\endcsname{0.457}
\expandafter\gdef\csname odunum@n@bounds_utterance_copy.cov.av.audio\endcsname{558}
\expandafter\gdef\csname odunum@ci@bounds_utterance_copy.cov.av.audio\endcsname{[0.414, 0.499]}
\expandafter\gdef\csname odunum@val@bounds_utterance_copy.cov.av.context\endcsname{0.376}
\expandafter\gdef\csname odunum@n@bounds_utterance_copy.cov.av.context\endcsname{1\,974}
\expandafter\gdef\csname odunum@ci@bounds_utterance_copy.cov.av.context\endcsname{[0.352, 0.400]}
\expandafter\gdef\csname odunum@val@bounds_utterance_copy.cov.av.context.gen\endcsname{0.360}
\expandafter\gdef\csname odunum@n@bounds_utterance_copy.cov.av.context.gen\endcsname{1\,635}
\expandafter\gdef\csname odunum@ci@bounds_utterance_copy.cov.av.context.gen\endcsname{[0.334, 0.385]}
\expandafter\gdef\csname odunum@val@bounds_utterance_copy.cov.av.context.latin\endcsname{0.369}
\expandafter\gdef\csname odunum@n@bounds_utterance_copy.cov.av.context.latin\endcsname{811}
\expandafter\gdef\csname odunum@ci@bounds_utterance_copy.cov.av.context.latin\endcsname{[0.330, 0.407]}
\expandafter\gdef\csname odunum@val@bounds_utterance_copy.cov.av.context.lex_absent\endcsname{0.370}
\expandafter\gdef\csname odunum@n@bounds_utterance_copy.cov.av.context.lex_absent\endcsname{1\,936}
\expandafter\gdef\csname odunum@ci@bounds_utterance_copy.cov.av.context.lex_absent\endcsname{[0.346, 0.394]}
\expandafter\gdef\csname odunum@val@bounds_utterance_copy.cov.av.context.lex_absent.latin\endcsname{0.353}
\expandafter\gdef\csname odunum@n@bounds_utterance_copy.cov.av.context.lex_absent.latin\endcsname{773}
\expandafter\gdef\csname odunum@ci@bounds_utterance_copy.cov.av.context.lex_absent.latin\endcsname{[0.315, 0.391]}
\expandafter\gdef\csname odunum@val@bounds_utterance_copy.cov.av.context.lex_present\endcsname{0.684}
\expandafter\gdef\csname odunum@n@bounds_utterance_copy.cov.av.context.lex_present\endcsname{38}
\expandafter\gdef\csname odunum@ci@bounds_utterance_copy.cov.av.context.lex_present\endcsname{[0.525, 0.838]}
\expandafter\gdef\csname odunum@val@bounds_utterance_copy.cov.av.context.lex_present.latin\endcsname{0.684}
\expandafter\gdef\csname odunum@n@bounds_utterance_copy.cov.av.context.lex_present.latin\endcsname{38}
\expandafter\gdef\csname odunum@ci@bounds_utterance_copy.cov.av.context.lex_present.latin\endcsname{[0.525, 0.838]}
\expandafter\gdef\csname odunum@val@bounds_utterance_copy.cov.av.context.recorded\endcsname{0.454}
\expandafter\gdef\csname odunum@n@bounds_utterance_copy.cov.av.context.recorded\endcsname{339}
\expandafter\gdef\csname odunum@ci@bounds_utterance_copy.cov.av.context.recorded\endcsname{[0.395, 0.513]}
\expandafter\gdef\csname odunum@val@bounds_utterance_copy.cov.av.history\endcsname{0.355}
\expandafter\gdef\csname odunum@n@bounds_utterance_copy.cov.av.history\endcsname{476}
\expandafter\gdef\csname odunum@ci@bounds_utterance_copy.cov.av.history\endcsname{[0.310, 0.404]}
\expandafter\gdef\csname odunum@val@bounds_utterance_copy.cov.av.intent\endcsname{0.965}
\expandafter\gdef\csname odunum@n@bounds_utterance_copy.cov.av.intent\endcsname{1\,986}
\expandafter\gdef\csname odunum@ci@bounds_utterance_copy.cov.av.intent\endcsname{[0.957, 0.974]}
\expandafter\gdef\csname odunum@val@bounds_utterance_copy.cov.av.intent.gen\endcsname{0.973}
\expandafter\gdef\csname odunum@n@bounds_utterance_copy.cov.av.intent.gen\endcsname{1\,624}
\expandafter\gdef\csname odunum@ci@bounds_utterance_copy.cov.av.intent.gen\endcsname{[0.964, 0.981]}
\expandafter\gdef\csname odunum@val@bounds_utterance_copy.cov.av.intent.latin\endcsname{0.982}
\expandafter\gdef\csname odunum@n@bounds_utterance_copy.cov.av.intent.latin\endcsname{849}
\expandafter\gdef\csname odunum@ci@bounds_utterance_copy.cov.av.intent.latin\endcsname{[0.973, 0.991]}
\expandafter\gdef\csname odunum@val@bounds_utterance_copy.cov.av.intent.recorded\endcsname{0.931}
\expandafter\gdef\csname odunum@n@bounds_utterance_copy.cov.av.intent.recorded\endcsname{362}
\expandafter\gdef\csname odunum@ci@bounds_utterance_copy.cov.av.intent.recorded\endcsname{[0.903, 0.956]}
\expandafter\gdef\csname odunum@val@bounds_utterance_copy.cov.av.other_context\endcsname{0.456}
\expandafter\gdef\csname odunum@n@bounds_utterance_copy.cov.av.other_context\endcsname{136}
\expandafter\gdef\csname odunum@ci@bounds_utterance_copy.cov.av.other_context\endcsname{[0.374, 0.538]}
\expandafter\gdef\csname odunum@val@bounds_utterance_copy.cov.av.overall\endcsname{0.671}
\expandafter\gdef\csname odunum@n@bounds_utterance_copy.cov.av.overall\endcsname{3\,973}
\expandafter\gdef\csname odunum@ci@bounds_utterance_copy.cov.av.overall\endcsname{[0.657, 0.684]}
\expandafter\gdef\csname odunum@val@bounds_utterance_copy.cov.av.overall.gen\endcsname{0.665}
\expandafter\gdef\csname odunum@n@bounds_utterance_copy.cov.av.overall.gen\endcsname{3\,259}
\expandafter\gdef\csname odunum@ci@bounds_utterance_copy.cov.av.overall.gen\endcsname{[0.650, 0.680]}
\expandafter\gdef\csname odunum@val@bounds_utterance_copy.cov.av.overall.recorded\endcsname{0.695}
\expandafter\gdef\csname odunum@n@bounds_utterance_copy.cov.av.overall.recorded\endcsname{714}
\expandafter\gdef\csname odunum@ci@bounds_utterance_copy.cov.av.overall.recorded\endcsname{[0.661, 0.729]}
\expandafter\gdef\csname odunum@val@bounds_utterance_copy.cov.av.unattributed\endcsname{0.385}
\expandafter\gdef\csname odunum@n@bounds_utterance_copy.cov.av.unattributed\endcsname{13}
\expandafter\gdef\csname odunum@ci@bounds_utterance_copy.cov.av.unattributed\endcsname{[0.050, 0.786]}
\expandafter\gdef\csname odunum@val@bounds_utterance_copy.cov.av.visual\endcsname{0.318}
\expandafter\gdef\csname odunum@n@bounds_utterance_copy.cov.av.visual\endcsname{804}
\expandafter\gdef\csname odunum@ci@bounds_utterance_copy.cov.av.visual\endcsname{[0.284, 0.353]}
\expandafter\gdef\csname odunum@val@bounds_utterance_copy.delta.all.context\endcsname{0.608}
\expandafter\gdef\csname odunum@n@bounds_utterance_copy.delta.all.context\endcsname{2\,702}
\expandafter\gdef\csname odunum@ci@bounds_utterance_copy.delta.all.context\endcsname{[0.588, 0.628]}
\expandafter\gdef\csname odunum@val@bounds_utterance_copy.delta.all.intent\endcsname{0.030}
\expandafter\gdef\csname odunum@n@bounds_utterance_copy.delta.all.intent\endcsname{3\,144}
\expandafter\gdef\csname odunum@ci@bounds_utterance_copy.delta.all.intent\endcsname{[0.024, 0.036]}
\expandafter\gdef\csname odunum@val@bounds_utterance_copy.delta.ao.context\endcsname{0.573}
\expandafter\gdef\csname odunum@n@bounds_utterance_copy.delta.ao.context\endcsname{728}
\expandafter\gdef\csname odunum@ci@bounds_utterance_copy.delta.ao.context\endcsname{[0.533, 0.612]}
\expandafter\gdef\csname odunum@val@bounds_utterance_copy.delta.ao.intent\endcsname{0.022}
\expandafter\gdef\csname odunum@n@bounds_utterance_copy.delta.ao.intent\endcsname{1\,158}
\expandafter\gdef\csname odunum@ci@bounds_utterance_copy.delta.ao.intent\endcsname{[0.013, 0.031]}
\expandafter\gdef\csname odunum@val@bounds_utterance_copy.delta.av.context\endcsname{0.621}
\expandafter\gdef\csname odunum@n@bounds_utterance_copy.delta.av.context\endcsname{1\,974}
\expandafter\gdef\csname odunum@ci@bounds_utterance_copy.delta.av.context\endcsname{[0.596, 0.644]}
\expandafter\gdef\csname odunum@val@bounds_utterance_copy.delta.av.intent\endcsname{0.034}
\expandafter\gdef\csname odunum@n@bounds_utterance_copy.delta.av.intent\endcsname{1\,986}
\expandafter\gdef\csname odunum@ci@bounds_utterance_copy.delta.av.intent\endcsname{[0.026, 0.043]}
\expandafter\gdef\csname odunum@val@bounds_utterance_copy.gap.all.intent_minus_context\endcsname{0.581}
\expandafter\gdef\csname odunum@n@bounds_utterance_copy.gap.all.intent_minus_context\endcsname{5\,846}
\expandafter\gdef\csname odunum@val@bounds_utterance_copy.gap.ao.intent_minus_context\endcsname{0.553}
\expandafter\gdef\csname odunum@n@bounds_utterance_copy.gap.ao.intent_minus_context\endcsname{1\,886}
\expandafter\gdef\csname odunum@val@bounds_utterance_copy.gap.av.intent_minus_context\endcsname{0.589}
\expandafter\gdef\csname odunum@n@bounds_utterance_copy.gap.av.intent_minus_context\endcsname{3\,960}
\expandafter\gdef\csname odunum@val@bounds_utterance_copy.judge.id\endcsname{mr\_ali / dashscope.qwen3.6-flash (thinking)}
\expandafter\gdef\csname odunum@n@bounds_utterance_copy.judge.id\endcsname{1}
\expandafter\gdef\csname odunum@val@bounds_utterance_copy.judge.scored_on\endcsname{2026-09-03}
\expandafter\gdef\csname odunum@n@bounds_utterance_copy.judge.scored_on\endcsname{1}
\expandafter\gdef\csname odunum@val@bounds_utterance_copy.lex.all.audio.share_half\endcsname{0.027}
\expandafter\gdef\csname odunum@n@bounds_utterance_copy.lex.all.audio.share_half\endcsname{899}
\expandafter\gdef\csname odunum@val@bounds_utterance_copy.lex.all.context.share_half\endcsname{0.020}
\expandafter\gdef\csname odunum@n@bounds_utterance_copy.lex.all.context.share_half\endcsname{2\,702}
\expandafter\gdef\csname odunum@val@bounds_utterance_copy.lex.all.history.share_half\endcsname{0.022}
\expandafter\gdef\csname odunum@n@bounds_utterance_copy.lex.all.history.share_half\endcsname{818}
\expandafter\gdef\csname odunum@val@bounds_utterance_copy.lex.all.intent.share_half\endcsname{0.214}
\expandafter\gdef\csname odunum@n@bounds_utterance_copy.lex.all.intent.share_half\endcsname{3\,144}
\expandafter\gdef\csname odunum@val@bounds_utterance_copy.lex.all.visual.share_half\endcsname{0.012}
\expandafter\gdef\csname odunum@n@bounds_utterance_copy.lex.all.visual.share_half\endcsname{804}
\expandafter\gdef\csname odunum@val@bounds_utterance_copy.lex.ao.audio.share_half\endcsname{0.026}
\expandafter\gdef\csname odunum@n@bounds_utterance_copy.lex.ao.audio.share_half\endcsname{341}
\expandafter\gdef\csname odunum@val@bounds_utterance_copy.lex.ao.context.share_half\endcsname{0.021}
\expandafter\gdef\csname odunum@n@bounds_utterance_copy.lex.ao.context.share_half\endcsname{728}
\expandafter\gdef\csname odunum@val@bounds_utterance_copy.lex.ao.history.share_half\endcsname{0.018}
\expandafter\gdef\csname odunum@n@bounds_utterance_copy.lex.ao.history.share_half\endcsname{342}
\expandafter\gdef\csname odunum@val@bounds_utterance_copy.lex.ao.intent.share_half\endcsname{0.286}
\expandafter\gdef\csname odunum@n@bounds_utterance_copy.lex.ao.intent.share_half\endcsname{1\,158}
\expandafter\gdef\csname odunum@val@bounds_utterance_copy.lex.av.audio.share_half\endcsname{0.027}
\expandafter\gdef\csname odunum@n@bounds_utterance_copy.lex.av.audio.share_half\endcsname{558}
\expandafter\gdef\csname odunum@val@bounds_utterance_copy.lex.av.context.share_half\endcsname{0.019}
\expandafter\gdef\csname odunum@n@bounds_utterance_copy.lex.av.context.share_half\endcsname{1\,974}
\expandafter\gdef\csname odunum@val@bounds_utterance_copy.lex.av.history.share_half\endcsname{0.025}
\expandafter\gdef\csname odunum@n@bounds_utterance_copy.lex.av.history.share_half\endcsname{476}
\expandafter\gdef\csname odunum@val@bounds_utterance_copy.lex.av.intent.share_half\endcsname{0.173}
\expandafter\gdef\csname odunum@n@bounds_utterance_copy.lex.av.intent.share_half\endcsname{1\,986}
\expandafter\gdef\csname odunum@val@bounds_utterance_copy.lex.av.visual.share_half\endcsname{0.012}
\expandafter\gdef\csname odunum@n@bounds_utterance_copy.lex.av.visual.share_half\endcsname{804}
\expandafter\gdef\csname odunum@val@bounds_utterance_copy.pop.all.n_points\endcsname{5\,859}
\expandafter\gdef\csname odunum@n@bounds_utterance_copy.pop.all.n_points\endcsname{1\,631}
\expandafter\gdef\csname odunum@val@bounds_utterance_copy.pop.all.n_scenes\endcsname{1\,631}
\expandafter\gdef\csname odunum@n@bounds_utterance_copy.pop.all.n_scenes\endcsname{1\,631}
\expandafter\gdef\csname odunum@val@bounds_utterance_copy.pop.ao.n_points\endcsname{1\,886}
\expandafter\gdef\csname odunum@n@bounds_utterance_copy.pop.ao.n_points\endcsname{570}
\expandafter\gdef\csname odunum@val@bounds_utterance_copy.pop.ao.n_scenes\endcsname{570}
\expandafter\gdef\csname odunum@n@bounds_utterance_copy.pop.ao.n_scenes\endcsname{570}
\expandafter\gdef\csname odunum@val@bounds_utterance_copy.pop.av.n_points\endcsname{3\,973}
\expandafter\gdef\csname odunum@n@bounds_utterance_copy.pop.av.n_points\endcsname{1\,061}
\expandafter\gdef\csname odunum@val@bounds_utterance_copy.pop.av.n_scenes\endcsname{1\,061}
\expandafter\gdef\csname odunum@n@bounds_utterance_copy.pop.av.n_scenes\endcsname{1\,061}
\expandafter\gdef\csname odunum@val@cascade_ladder.cov.caption.audio\endcsname{0.497}
\expandafter\gdef\csname odunum@n@cascade_ladder.cov.caption.audio\endcsname{547}
\expandafter\gdef\csname odunum@ci@cascade_ladder.cov.caption.audio\endcsname{[0.456, 0.539]}
\expandafter\gdef\csname odunum@val@cascade_ladder.cov.caption.history\endcsname{0.506}
\expandafter\gdef\csname odunum@n@cascade_ladder.cov.caption.history\endcsname{470}
\expandafter\gdef\csname odunum@ci@cascade_ladder.cov.caption.history\endcsname{[0.461, 0.553]}
\expandafter\gdef\csname odunum@val@cascade_ladder.cov.caption.intent\endcsname{0.814}
\expandafter\gdef\csname odunum@n@cascade_ladder.cov.caption.intent\endcsname{1\,947}
\expandafter\gdef\csname odunum@ci@cascade_ladder.cov.caption.intent\endcsname{[0.794, 0.834]}
\expandafter\gdef\csname odunum@val@cascade_ladder.cov.caption.other_context\endcsname{0.459}
\expandafter\gdef\csname odunum@n@cascade_ladder.cov.caption.other_context\endcsname{133}
\expandafter\gdef\csname odunum@ci@cascade_ladder.cov.caption.other_context\endcsname{[0.376, 0.541]}
\expandafter\gdef\csname odunum@val@cascade_ladder.cov.caption.visual\endcsname{0.455}
\expandafter\gdef\csname odunum@n@cascade_ladder.cov.caption.visual\endcsname{791}
\expandafter\gdef\csname odunum@ci@cascade_ladder.cov.caption.visual\endcsname{[0.419, 0.490]}
\expandafter\gdef\csname odunum@val@cascade_ladder.cov.gemini.audio\endcsname{0.473}
\expandafter\gdef\csname odunum@n@cascade_ladder.cov.gemini.audio\endcsname{547}
\expandafter\gdef\csname odunum@ci@cascade_ladder.cov.gemini.audio\endcsname{[0.431, 0.515]}
\expandafter\gdef\csname odunum@val@cascade_ladder.cov.gemini.history\endcsname{0.419}
\expandafter\gdef\csname odunum@n@cascade_ladder.cov.gemini.history\endcsname{470}
\expandafter\gdef\csname odunum@ci@cascade_ladder.cov.gemini.history\endcsname{[0.373, 0.466]}
\expandafter\gdef\csname odunum@val@cascade_ladder.cov.gemini.intent\endcsname{0.828}
\expandafter\gdef\csname odunum@n@cascade_ladder.cov.gemini.intent\endcsname{1\,947}
\expandafter\gdef\csname odunum@ci@cascade_ladder.cov.gemini.intent\endcsname{[0.810, 0.846]}
\expandafter\gdef\csname odunum@val@cascade_ladder.cov.gemini.other_context\endcsname{0.203}
\expandafter\gdef\csname odunum@n@cascade_ladder.cov.gemini.other_context\endcsname{133}
\expandafter\gdef\csname odunum@ci@cascade_ladder.cov.gemini.other_context\endcsname{[0.137, 0.272]}
\expandafter\gdef\csname odunum@val@cascade_ladder.cov.gemini.visual\endcsname{0.482}
\expandafter\gdef\csname odunum@n@cascade_ladder.cov.gemini.visual\endcsname{791}
\expandafter\gdef\csname odunum@ci@cascade_ladder.cov.gemini.visual\endcsname{[0.446, 0.517]}
\expandafter\gdef\csname odunum@val@cascade_ladder.cov.native_mean.audio\endcsname{0.465}
\expandafter\gdef\csname odunum@n@cascade_ladder.cov.native_mean.audio\endcsname{547}
\expandafter\gdef\csname odunum@val@cascade_ladder.cov.native_mean.history\endcsname{0.538}
\expandafter\gdef\csname odunum@n@cascade_ladder.cov.native_mean.history\endcsname{470}
\expandafter\gdef\csname odunum@val@cascade_ladder.cov.native_mean.intent\endcsname{0.808}
\expandafter\gdef\csname odunum@n@cascade_ladder.cov.native_mean.intent\endcsname{1\,947}
\expandafter\gdef\csname odunum@val@cascade_ladder.cov.native_mean.other_context\endcsname{0.326}
\expandafter\gdef\csname odunum@n@cascade_ladder.cov.native_mean.other_context\endcsname{133}
\expandafter\gdef\csname odunum@val@cascade_ladder.cov.native_mean.visual\endcsname{0.499}
\expandafter\gdef\csname odunum@n@cascade_ladder.cov.native_mean.visual\endcsname{791}
\expandafter\gdef\csname odunum@val@cascade_ladder.cov.qwen_plus.audio\endcsname{0.457}
\expandafter\gdef\csname odunum@n@cascade_ladder.cov.qwen_plus.audio\endcsname{547}
\expandafter\gdef\csname odunum@ci@cascade_ladder.cov.qwen_plus.audio\endcsname{[0.416, 0.499]}
\expandafter\gdef\csname odunum@val@cascade_ladder.cov.qwen_plus.history\endcsname{0.577}
\expandafter\gdef\csname odunum@n@cascade_ladder.cov.qwen_plus.history\endcsname{470}
\expandafter\gdef\csname odunum@ci@cascade_ladder.cov.qwen_plus.history\endcsname{[0.530, 0.623]}
\expandafter\gdef\csname odunum@val@cascade_ladder.cov.qwen_plus.intent\endcsname{0.838}
\expandafter\gdef\csname odunum@n@cascade_ladder.cov.qwen_plus.intent\endcsname{1\,947}
\expandafter\gdef\csname odunum@ci@cascade_ladder.cov.qwen_plus.intent\endcsname{[0.821, 0.854]}
\expandafter\gdef\csname odunum@val@cascade_ladder.cov.qwen_plus.other_context\endcsname{0.429}
\expandafter\gdef\csname odunum@n@cascade_ladder.cov.qwen_plus.other_context\endcsname{133}
\expandafter\gdef\csname odunum@ci@cascade_ladder.cov.qwen_plus.other_context\endcsname{[0.346, 0.511]}
\expandafter\gdef\csname odunum@val@cascade_ladder.cov.qwen_plus.visual\endcsname{0.473}
\expandafter\gdef\csname odunum@n@cascade_ladder.cov.qwen_plus.visual\endcsname{791}
\expandafter\gdef\csname odunum@ci@cascade_ladder.cov.qwen_plus.visual\endcsname{[0.437, 0.508]}
\expandafter\gdef\csname odunum@val@cascade_ladder.cov.seed.audio\endcsname{0.464}
\expandafter\gdef\csname odunum@n@cascade_ladder.cov.seed.audio\endcsname{547}
\expandafter\gdef\csname odunum@ci@cascade_ladder.cov.seed.audio\endcsname{[0.422, 0.507]}
\expandafter\gdef\csname odunum@val@cascade_ladder.cov.seed.history\endcsname{0.619}
\expandafter\gdef\csname odunum@n@cascade_ladder.cov.seed.history\endcsname{470}
\expandafter\gdef\csname odunum@ci@cascade_ladder.cov.seed.history\endcsname{[0.573, 0.665]}
\expandafter\gdef\csname odunum@val@cascade_ladder.cov.seed.intent\endcsname{0.759}
\expandafter\gdef\csname odunum@n@cascade_ladder.cov.seed.intent\endcsname{1\,947}
\expandafter\gdef\csname odunum@ci@cascade_ladder.cov.seed.intent\endcsname{[0.735, 0.782]}
\expandafter\gdef\csname odunum@val@cascade_ladder.cov.seed.other_context\endcsname{0.346}
\expandafter\gdef\csname odunum@n@cascade_ladder.cov.seed.other_context\endcsname{133}
\expandafter\gdef\csname odunum@ci@cascade_ladder.cov.seed.other_context\endcsname{[0.265, 0.427]}
\expandafter\gdef\csname odunum@val@cascade_ladder.cov.seed.visual\endcsname{0.542}
\expandafter\gdef\csname odunum@n@cascade_ladder.cov.seed.visual\endcsname{791}
\expandafter\gdef\csname odunum@ci@cascade_ladder.cov.seed.visual\endcsname{[0.506, 0.578]}
\expandafter\gdef\csname odunum@val@cascade_ladder.cov.transcript.audio\endcsname{0.336}
\expandafter\gdef\csname odunum@n@cascade_ladder.cov.transcript.audio\endcsname{547}
\expandafter\gdef\csname odunum@ci@cascade_ladder.cov.transcript.audio\endcsname{[0.297, 0.376]}
\expandafter\gdef\csname odunum@val@cascade_ladder.cov.transcript.history\endcsname{0.592}
\expandafter\gdef\csname odunum@n@cascade_ladder.cov.transcript.history\endcsname{470}
\expandafter\gdef\csname odunum@ci@cascade_ladder.cov.transcript.history\endcsname{[0.545, 0.636]}
\expandafter\gdef\csname odunum@val@cascade_ladder.cov.transcript.intent\endcsname{0.791}
\expandafter\gdef\csname odunum@n@cascade_ladder.cov.transcript.intent\endcsname{1\,947}
\expandafter\gdef\csname odunum@ci@cascade_ladder.cov.transcript.intent\endcsname{[0.771, 0.810]}
\expandafter\gdef\csname odunum@val@cascade_ladder.cov.transcript.other_context\endcsname{0.158}
\expandafter\gdef\csname odunum@n@cascade_ladder.cov.transcript.other_context\endcsname{133}
\expandafter\gdef\csname odunum@ci@cascade_ladder.cov.transcript.other_context\endcsname{[0.099, 0.223]}
\expandafter\gdef\csname odunum@val@cascade_ladder.cov.transcript.visual\endcsname{0.205}
\expandafter\gdef\csname odunum@n@cascade_ladder.cov.transcript.visual\endcsname{791}
\expandafter\gdef\csname odunum@ci@cascade_ladder.cov.transcript.visual\endcsname{[0.177, 0.234]}
\expandafter\gdef\csname odunum@val@cascade_ladder.gap_to_native.ao.caption.avg\endcsname{0.012}
\expandafter\gdef\csname odunum@n@cascade_ladder.gap_to_native.ao.caption.avg\endcsname{731}
\expandafter\gdef\csname odunum@val@cascade_ladder.gap_to_native.ao.caption.m2\endcsname{0.035}
\expandafter\gdef\csname odunum@n@cascade_ladder.gap_to_native.ao.caption.m2\endcsname{731}
\expandafter\gdef\csname odunum@val@cascade_ladder.gap_to_native.ao.transcript.avg\endcsname{0.043}
\expandafter\gdef\csname odunum@n@cascade_ladder.gap_to_native.ao.transcript.avg\endcsname{731}
\expandafter\gdef\csname odunum@val@cascade_ladder.gap_to_native.ao.transcript.m2\endcsname{0.049}
\expandafter\gdef\csname odunum@n@cascade_ladder.gap_to_native.ao.transcript.m2\endcsname{731}
\expandafter\gdef\csname odunum@val@cascade_ladder.gap_to_native.av.caption.avg\endcsname{0.017}
\expandafter\gdef\csname odunum@n@cascade_ladder.gap_to_native.av.caption.avg\endcsname{1\,346}
\expandafter\gdef\csname odunum@val@cascade_ladder.gap_to_native.av.caption.m2\endcsname{0.003}
\expandafter\gdef\csname odunum@n@cascade_ladder.gap_to_native.av.caption.m2\endcsname{1\,346}
\expandafter\gdef\csname odunum@val@cascade_ladder.gap_to_native.av.transcript.avg\endcsname{0.111}
\expandafter\gdef\csname odunum@n@cascade_ladder.gap_to_native.av.transcript.avg\endcsname{1\,346}
\expandafter\gdef\csname odunum@val@cascade_ladder.gap_to_native.av.transcript.m2\endcsname{0.089}
\expandafter\gdef\csname odunum@n@cascade_ladder.gap_to_native.av.transcript.m2\endcsname{1\,346}
\expandafter\gdef\csname odunum@val@cascade_ladder.native_mean.ao.avg\endcsname{0.758}
\expandafter\gdef\csname odunum@n@cascade_ladder.native_mean.ao.avg\endcsname{3}
\expandafter\gdef\csname odunum@val@cascade_ladder.native_mean.ao.m2\endcsname{0.732}
\expandafter\gdef\csname odunum@n@cascade_ladder.native_mean.ao.m2\endcsname{3}
\expandafter\gdef\csname odunum@val@cascade_ladder.native_mean.av.avg\endcsname{0.700}
\expandafter\gdef\csname odunum@n@cascade_ladder.native_mean.av.avg\endcsname{3}
\expandafter\gdef\csname odunum@val@cascade_ladder.native_mean.av.m2\endcsname{0.644}
\expandafter\gdef\csname odunum@n@cascade_ladder.native_mean.av.m2\endcsname{3}
\expandafter\gdef\csname odunum@val@cascade_ladder.overlap.caption.audio\endcsname{0.726}
\expandafter\gdef\csname odunum@n@cascade_ladder.overlap.caption.audio\endcsname{547}
\expandafter\gdef\csname odunum@val@cascade_ladder.overlap.caption.audio.share_half\endcsname{0.943}
\expandafter\gdef\csname odunum@n@cascade_ladder.overlap.caption.audio.share_half\endcsname{547}
\expandafter\gdef\csname odunum@val@cascade_ladder.overlap.caption.history\endcsname{0.698}
\expandafter\gdef\csname odunum@n@cascade_ladder.overlap.caption.history\endcsname{470}
\expandafter\gdef\csname odunum@val@cascade_ladder.overlap.caption.history.share_half\endcsname{0.900}
\expandafter\gdef\csname odunum@n@cascade_ladder.overlap.caption.history.share_half\endcsname{470}
\expandafter\gdef\csname odunum@val@cascade_ladder.overlap.caption.intent\endcsname{0.629}
\expandafter\gdef\csname odunum@n@cascade_ladder.overlap.caption.intent\endcsname{1\,947}
\expandafter\gdef\csname odunum@val@cascade_ladder.overlap.caption.intent.share_half\endcsname{0.832}
\expandafter\gdef\csname odunum@n@cascade_ladder.overlap.caption.intent.share_half\endcsname{1\,947}
\expandafter\gdef\csname odunum@val@cascade_ladder.overlap.caption.visual\endcsname{0.740}
\expandafter\gdef\csname odunum@n@cascade_ladder.overlap.caption.visual\endcsname{791}
\expandafter\gdef\csname odunum@val@cascade_ladder.overlap.caption.visual.share_half\endcsname{0.971}
\expandafter\gdef\csname odunum@n@cascade_ladder.overlap.caption.visual.share_half\endcsname{791}
\expandafter\gdef\csname odunum@val@cascade_ladder.overlap.transcript.audio\endcsname{0.409}
\expandafter\gdef\csname odunum@n@cascade_ladder.overlap.transcript.audio\endcsname{547}
\expandafter\gdef\csname odunum@val@cascade_ladder.overlap.transcript.audio.share_half\endcsname{0.305}
\expandafter\gdef\csname odunum@n@cascade_ladder.overlap.transcript.audio.share_half\endcsname{547}
\expandafter\gdef\csname odunum@val@cascade_ladder.overlap.transcript.history\endcsname{0.480}
\expandafter\gdef\csname odunum@n@cascade_ladder.overlap.transcript.history\endcsname{470}
\expandafter\gdef\csname odunum@val@cascade_ladder.overlap.transcript.history.share_half\endcsname{0.445}
\expandafter\gdef\csname odunum@n@cascade_ladder.overlap.transcript.history.share_half\endcsname{470}
\expandafter\gdef\csname odunum@val@cascade_ladder.overlap.transcript.intent\endcsname{0.434}
\expandafter\gdef\csname odunum@n@cascade_ladder.overlap.transcript.intent\endcsname{1\,947}
\expandafter\gdef\csname odunum@val@cascade_ladder.overlap.transcript.intent.share_half\endcsname{0.346}
\expandafter\gdef\csname odunum@n@cascade_ladder.overlap.transcript.intent.share_half\endcsname{1\,947}
\expandafter\gdef\csname odunum@val@cascade_ladder.overlap.transcript.visual\endcsname{0.418}
\expandafter\gdef\csname odunum@n@cascade_ladder.overlap.transcript.visual\endcsname{791}
\expandafter\gdef\csname odunum@val@cascade_ladder.overlap.transcript.visual.share_half\endcsname{0.250}
\expandafter\gdef\csname odunum@n@cascade_ladder.overlap.transcript.visual.share_half\endcsname{791}
\expandafter\gdef\csname odunum@val@cascade_ladder.pop.ao.n\endcsname{731}
\expandafter\gdef\csname odunum@n@cascade_ladder.pop.ao.n\endcsname{2\,077}
\expandafter\gdef\csname odunum@val@cascade_ladder.pop.av.n\endcsname{1\,346}
\expandafter\gdef\csname odunum@n@cascade_ladder.pop.av.n\endcsname{2\,077}
\expandafter\gdef\csname odunum@val@cascade_ladder.pop.common_points.n\endcsname{3\,901}
\expandafter\gdef\csname odunum@n@cascade_ladder.pop.common_points.n\endcsname{3\,973}
\expandafter\gdef\csname odunum@val@cascade_ladder.pop.common_scenes.n\endcsname{2\,077}
\expandafter\gdef\csname odunum@n@cascade_ladder.pop.common_scenes.n\endcsname{2\,078}
\expandafter\gdef\csname odunum@val@cascade_ladder.pop.dropped.caption.n\endcsname{1}
\expandafter\gdef\csname odunum@n@cascade_ladder.pop.dropped.caption.n\endcsname{2\,078}
\expandafter\gdef\csname odunum@val@cascade_ladder.pop.dropped.gemini.n\endcsname{1}
\expandafter\gdef\csname odunum@n@cascade_ladder.pop.dropped.gemini.n\endcsname{2\,078}
\expandafter\gdef\csname odunum@val@cascade_ladder.pop.dropped.qwen_plus.n\endcsname{0}
\expandafter\gdef\csname odunum@n@cascade_ladder.pop.dropped.qwen_plus.n\endcsname{2\,077}
\expandafter\gdef\csname odunum@val@cascade_ladder.pop.dropped.seed.n\endcsname{1}
\expandafter\gdef\csname odunum@n@cascade_ladder.pop.dropped.seed.n\endcsname{2\,078}
\expandafter\gdef\csname odunum@val@cascade_ladder.pop.dropped.transcript.n\endcsname{1}
\expandafter\gdef\csname odunum@n@cascade_ladder.pop.dropped.transcript.n\endcsname{2\,078}
\expandafter\gdef\csname odunum@val@cascade_ladder.pop.points.audio.n\endcsname{547}
\expandafter\gdef\csname odunum@n@cascade_ladder.pop.points.audio.n\endcsname{3\,901}
\expandafter\gdef\csname odunum@val@cascade_ladder.pop.points.history.n\endcsname{470}
\expandafter\gdef\csname odunum@n@cascade_ladder.pop.points.history.n\endcsname{3\,901}
\expandafter\gdef\csname odunum@val@cascade_ladder.pop.points.intent.n\endcsname{1\,947}
\expandafter\gdef\csname odunum@n@cascade_ladder.pop.points.intent.n\endcsname{3\,901}
\expandafter\gdef\csname odunum@val@cascade_ladder.pop.points.other_context.n\endcsname{133}
\expandafter\gdef\csname odunum@n@cascade_ladder.pop.points.other_context.n\endcsname{3\,901}
\expandafter\gdef\csname odunum@val@cascade_ladder.pop.points.unattributed.n\endcsname{13}
\expandafter\gdef\csname odunum@n@cascade_ladder.pop.points.unattributed.n\endcsname{3\,901}
\expandafter\gdef\csname odunum@val@cascade_ladder.pop.points.visual.n\endcsname{791}
\expandafter\gdef\csname odunum@n@cascade_ladder.pop.points.visual.n\endcsname{3\,901}
\expandafter\gdef\csname odunum@val@cascade_ladder.pop.points_source_disagreed.n\endcsname{0}
\expandafter\gdef\csname odunum@n@cascade_ladder.pop.points_source_disagreed.n\endcsname{3\,901}
\expandafter\gdef\csname odunum@val@cascade_ladder.pop.shown_text.caption.n\endcsname{2\,078}
\expandafter\gdef\csname odunum@n@cascade_ladder.pop.shown_text.caption.n\endcsname{2\,078}
\expandafter\gdef\csname odunum@val@cascade_ladder.pop.shown_text.transcript.n\endcsname{2\,078}
\expandafter\gdef\csname odunum@n@cascade_ladder.pop.shown_text.transcript.n\endcsname{2\,078}
\expandafter\gdef\csname odunum@val@cascade_ladder.prior.caption.audio.native0\endcsname{0.239}
\expandafter\gdef\csname odunum@n@cascade_ladder.prior.caption.audio.native0\endcsname{180}
\expandafter\gdef\csname odunum@val@cascade_ladder.prior.caption.audio.native1\endcsname{0.438}
\expandafter\gdef\csname odunum@n@cascade_ladder.prior.caption.audio.native1\endcsname{112}
\expandafter\gdef\csname odunum@val@cascade_ladder.prior.caption.audio.native2\endcsname{0.605}
\expandafter\gdef\csname odunum@n@cascade_ladder.prior.caption.audio.native2\endcsname{114}
\expandafter\gdef\csname odunum@val@cascade_ladder.prior.caption.audio.native3\endcsname{0.787}
\expandafter\gdef\csname odunum@n@cascade_ladder.prior.caption.audio.native3\endcsname{141}
\expandafter\gdef\csname odunum@val@cascade_ladder.prior.caption.history.native0\endcsname{0.135}
\expandafter\gdef\csname odunum@n@cascade_ladder.prior.caption.history.native0\endcsname{111}
\expandafter\gdef\csname odunum@val@cascade_ladder.prior.caption.history.native1\endcsname{0.408}
\expandafter\gdef\csname odunum@n@cascade_ladder.prior.caption.history.native1\endcsname{103}
\expandafter\gdef\csname odunum@val@cascade_ladder.prior.caption.history.native2\endcsname{0.527}
\expandafter\gdef\csname odunum@n@cascade_ladder.prior.caption.history.native2\endcsname{112}
\expandafter\gdef\csname odunum@val@cascade_ladder.prior.caption.history.native3\endcsname{0.847}
\expandafter\gdef\csname odunum@n@cascade_ladder.prior.caption.history.native3\endcsname{144}
\expandafter\gdef\csname odunum@val@cascade_ladder.prior.caption.intent.native0\endcsname{0.200}
\expandafter\gdef\csname odunum@n@cascade_ladder.prior.caption.intent.native0\endcsname{135}
\expandafter\gdef\csname odunum@val@cascade_ladder.prior.caption.intent.native1\endcsname{0.432}
\expandafter\gdef\csname odunum@n@cascade_ladder.prior.caption.intent.native1\endcsname{155}
\expandafter\gdef\csname odunum@val@cascade_ladder.prior.caption.intent.native2\endcsname{0.800}
\expandafter\gdef\csname odunum@n@cascade_ladder.prior.caption.intent.native2\endcsname{406}
\expandafter\gdef\csname odunum@val@cascade_ladder.prior.caption.intent.native3\endcsname{0.932}
\expandafter\gdef\csname odunum@n@cascade_ladder.prior.caption.intent.native3\endcsname{1\,251}
\expandafter\gdef\csname odunum@val@cascade_ladder.prior.caption.visual.native0\endcsname{0.156}
\expandafter\gdef\csname odunum@n@cascade_ladder.prior.caption.visual.native0\endcsname{237}
\expandafter\gdef\csname odunum@val@cascade_ladder.prior.caption.visual.native1\endcsname{0.344}
\expandafter\gdef\csname odunum@n@cascade_ladder.prior.caption.visual.native1\endcsname{160}
\expandafter\gdef\csname odunum@val@cascade_ladder.prior.caption.visual.native2\endcsname{0.525}
\expandafter\gdef\csname odunum@n@cascade_ladder.prior.caption.visual.native2\endcsname{158}
\expandafter\gdef\csname odunum@val@cascade_ladder.prior.caption.visual.native3\endcsname{0.784}
\expandafter\gdef\csname odunum@n@cascade_ladder.prior.caption.visual.native3\endcsname{236}
\expandafter\gdef\csname odunum@val@cascade_ladder.prior.transcript.audio.native0\endcsname{0.106}
\expandafter\gdef\csname odunum@n@cascade_ladder.prior.transcript.audio.native0\endcsname{180}
\expandafter\gdef\csname odunum@val@cascade_ladder.prior.transcript.audio.native1\endcsname{0.188}
\expandafter\gdef\csname odunum@n@cascade_ladder.prior.transcript.audio.native1\endcsname{112}
\expandafter\gdef\csname odunum@val@cascade_ladder.prior.transcript.audio.native2\endcsname{0.395}
\expandafter\gdef\csname odunum@n@cascade_ladder.prior.transcript.audio.native2\endcsname{114}
\expandafter\gdef\csname odunum@val@cascade_ladder.prior.transcript.audio.native3\endcsname{0.702}
\expandafter\gdef\csname odunum@n@cascade_ladder.prior.transcript.audio.native3\endcsname{141}
\expandafter\gdef\csname odunum@val@cascade_ladder.prior.transcript.history.native0\endcsname{0.207}
\expandafter\gdef\csname odunum@n@cascade_ladder.prior.transcript.history.native0\endcsname{111}
\expandafter\gdef\csname odunum@val@cascade_ladder.prior.transcript.history.native1\endcsname{0.553}
\expandafter\gdef\csname odunum@n@cascade_ladder.prior.transcript.history.native1\endcsname{103}
\expandafter\gdef\csname odunum@val@cascade_ladder.prior.transcript.history.native2\endcsname{0.607}
\expandafter\gdef\csname odunum@n@cascade_ladder.prior.transcript.history.native2\endcsname{112}
\expandafter\gdef\csname odunum@val@cascade_ladder.prior.transcript.history.native3\endcsname{0.903}
\expandafter\gdef\csname odunum@n@cascade_ladder.prior.transcript.history.native3\endcsname{144}
\expandafter\gdef\csname odunum@val@cascade_ladder.prior.transcript.intent.native0\endcsname{0.133}
\expandafter\gdef\csname odunum@n@cascade_ladder.prior.transcript.intent.native0\endcsname{135}
\expandafter\gdef\csname odunum@val@cascade_ladder.prior.transcript.intent.native1\endcsname{0.471}
\expandafter\gdef\csname odunum@n@cascade_ladder.prior.transcript.intent.native1\endcsname{155}
\expandafter\gdef\csname odunum@val@cascade_ladder.prior.transcript.intent.native2\endcsname{0.793}
\expandafter\gdef\csname odunum@n@cascade_ladder.prior.transcript.intent.native2\endcsname{406}
\expandafter\gdef\csname odunum@val@cascade_ladder.prior.transcript.intent.native3\endcsname{0.901}
\expandafter\gdef\csname odunum@n@cascade_ladder.prior.transcript.intent.native3\endcsname{1\,251}
\expandafter\gdef\csname odunum@val@cascade_ladder.prior.transcript.visual.native0\endcsname{0.051}
\expandafter\gdef\csname odunum@n@cascade_ladder.prior.transcript.visual.native0\endcsname{237}
\expandafter\gdef\csname odunum@val@cascade_ladder.prior.transcript.visual.native1\endcsname{0.106}
\expandafter\gdef\csname odunum@n@cascade_ladder.prior.transcript.visual.native1\endcsname{160}
\expandafter\gdef\csname odunum@val@cascade_ladder.prior.transcript.visual.native2\endcsname{0.266}
\expandafter\gdef\csname odunum@n@cascade_ladder.prior.transcript.visual.native2\endcsname{158}
\expandafter\gdef\csname odunum@val@cascade_ladder.prior.transcript.visual.native3\endcsname{0.386}
\expandafter\gdef\csname odunum@n@cascade_ladder.prior.transcript.visual.native3\endcsname{236}
\expandafter\gdef\csname odunum@val@cascade_ladder.provenance.caption_frontend.qwen3_5_omni_plus.n\endcsname{2\,078}
\expandafter\gdef\csname odunum@n@cascade_ladder.provenance.caption_frontend.qwen3_5_omni_plus.n\endcsname{2\,078}
\expandafter\gdef\csname odunum@val@cascade_ladder.restated.caption.audio.restated\endcsname{0.498}
\expandafter\gdef\csname odunum@n@cascade_ladder.restated.caption.audio.restated\endcsname{546}
\expandafter\gdef\csname odunum@val@cascade_ladder.restated.caption.history.restated\endcsname{0.506}
\expandafter\gdef\csname odunum@n@cascade_ladder.restated.caption.history.restated\endcsname{470}
\expandafter\gdef\csname odunum@val@cascade_ladder.restated.caption.intent.restated\endcsname{0.814}
\expandafter\gdef\csname odunum@n@cascade_ladder.restated.caption.intent.restated\endcsname{1\,942}
\expandafter\gdef\csname odunum@val@cascade_ladder.restated.caption.visual.restated\endcsname{0.455}
\expandafter\gdef\csname odunum@n@cascade_ladder.restated.caption.visual.restated\endcsname{791}
\expandafter\gdef\csname odunum@val@cascade_ladder.restated.transcript.audio.novel\endcsname{0.410}
\expandafter\gdef\csname odunum@n@cascade_ladder.restated.transcript.audio.novel\endcsname{39}
\expandafter\gdef\csname odunum@val@cascade_ladder.restated.transcript.audio.ratio\endcsname{0.81\ensuremath{\times}}
\expandafter\gdef\csname odunum@n@cascade_ladder.restated.transcript.audio.ratio\endcsname{547}
\expandafter\gdef\csname odunum@val@cascade_ladder.restated.transcript.audio.restated\endcsname{0.331}
\expandafter\gdef\csname odunum@n@cascade_ladder.restated.transcript.audio.restated\endcsname{508}
\expandafter\gdef\csname odunum@val@cascade_ladder.restated.transcript.history.novel\endcsname{0.524}
\expandafter\gdef\csname odunum@n@cascade_ladder.restated.transcript.history.novel\endcsname{21}
\expandafter\gdef\csname odunum@val@cascade_ladder.restated.transcript.history.ratio\endcsname{1.14\ensuremath{\times}}
\expandafter\gdef\csname odunum@n@cascade_ladder.restated.transcript.history.ratio\endcsname{470}
\expandafter\gdef\csname odunum@val@cascade_ladder.restated.transcript.history.restated\endcsname{0.595}
\expandafter\gdef\csname odunum@n@cascade_ladder.restated.transcript.history.restated\endcsname{449}
\expandafter\gdef\csname odunum@val@cascade_ladder.restated.transcript.intent.novel\endcsname{0.747}
\expandafter\gdef\csname odunum@n@cascade_ladder.restated.transcript.intent.novel\endcsname{75}
\expandafter\gdef\csname odunum@val@cascade_ladder.restated.transcript.intent.ratio\endcsname{1.06\ensuremath{\times}}
\expandafter\gdef\csname odunum@n@cascade_ladder.restated.transcript.intent.ratio\endcsname{1\,947}
\expandafter\gdef\csname odunum@val@cascade_ladder.restated.transcript.intent.restated\endcsname{0.793}
\expandafter\gdef\csname odunum@n@cascade_ladder.restated.transcript.intent.restated\endcsname{1\,872}
\expandafter\gdef\csname odunum@val@cascade_ladder.restated.transcript.visual.restated\endcsname{0.205}
\expandafter\gdef\csname odunum@n@cascade_ladder.restated.transcript.visual.restated\endcsname{780}
\expandafter\gdef\csname odunum@val@cascade_ladder.rung_delta.caption_minus_transcript.audio\endcsname{0.161}
\expandafter\gdef\csname odunum@n@cascade_ladder.rung_delta.caption_minus_transcript.audio\endcsname{547}
\expandafter\gdef\csname odunum@ci@cascade_ladder.rung_delta.caption_minus_transcript.audio\endcsname{[0.118, 0.206]}
\expandafter\gdef\csname odunum@val@cascade_ladder.rung_delta.caption_minus_transcript.audio.sign_p\endcsname{7.7\ensuremath{\times 10^{-13}}\ensuremath{\times}}
\expandafter\gdef\csname odunum@n@cascade_ladder.rung_delta.caption_minus_transcript.audio.sign_p\endcsname{156}
\expandafter\gdef\csname odunum@val@cascade_ladder.rung_delta.caption_minus_transcript.history\endcsname{-0.085}
\expandafter\gdef\csname odunum@n@cascade_ladder.rung_delta.caption_minus_transcript.history\endcsname{470}
\expandafter\gdef\csname odunum@ci@cascade_ladder.rung_delta.caption_minus_transcript.history\endcsname{[-0.135, -0.036]}
\expandafter\gdef\csname odunum@val@cascade_ladder.rung_delta.caption_minus_transcript.history.sign_p\endcsname{9.0\ensuremath{\times 10^{-4}}\ensuremath{\times}}
\expandafter\gdef\csname odunum@n@cascade_ladder.rung_delta.caption_minus_transcript.history.sign_p\endcsname{126}
\expandafter\gdef\csname odunum@val@cascade_ladder.rung_delta.caption_minus_transcript.intent\endcsname{0.023}
\expandafter\gdef\csname odunum@n@cascade_ladder.rung_delta.caption_minus_transcript.intent\endcsname{1\,947}
\expandafter\gdef\csname odunum@ci@cascade_ladder.rung_delta.caption_minus_transcript.intent\endcsname{[0.004, 0.042]}
\expandafter\gdef\csname odunum@val@cascade_ladder.rung_delta.caption_minus_transcript.intent.sign_p\endcsname{0.01\ensuremath{\times}}
\expandafter\gdef\csname odunum@n@cascade_ladder.rung_delta.caption_minus_transcript.intent.sign_p\endcsname{261}
\expandafter\gdef\csname odunum@val@cascade_ladder.rung_delta.caption_minus_transcript.other_context\endcsname{0.301}
\expandafter\gdef\csname odunum@n@cascade_ladder.rung_delta.caption_minus_transcript.other_context\endcsname{133}
\expandafter\gdef\csname odunum@ci@cascade_ladder.rung_delta.caption_minus_transcript.other_context\endcsname{[0.218, 0.384]}
\expandafter\gdef\csname odunum@val@cascade_ladder.rung_delta.caption_minus_transcript.other_context.sign_p\endcsname{8.7\ensuremath{\times 10^{-10}}\ensuremath{\times}}
\expandafter\gdef\csname odunum@n@cascade_ladder.rung_delta.caption_minus_transcript.other_context.sign_p\endcsname{45}
\expandafter\gdef\csname odunum@val@cascade_ladder.rung_delta.caption_minus_transcript.visual\endcsname{0.250}
\expandafter\gdef\csname odunum@n@cascade_ladder.rung_delta.caption_minus_transcript.visual\endcsname{791}
\expandafter\gdef\csname odunum@ci@cascade_ladder.rung_delta.caption_minus_transcript.visual\endcsname{[0.212, 0.288]}
\expandafter\gdef\csname odunum@val@cascade_ladder.rung_delta.caption_minus_transcript.visual.sign_p\endcsname{1.1\ensuremath{\times 10^{-35}}\ensuremath{\times}}
\expandafter\gdef\csname odunum@n@cascade_ladder.rung_delta.caption_minus_transcript.visual.sign_p\endcsname{252}
\expandafter\gdef\csname odunum@val@cascade_ladder.rung_delta.gemini_minus_caption.audio\endcsname{-0.024}
\expandafter\gdef\csname odunum@n@cascade_ladder.rung_delta.gemini_minus_caption.audio\endcsname{547}
\expandafter\gdef\csname odunum@ci@cascade_ladder.rung_delta.gemini_minus_caption.audio\endcsname{[-0.072, 0.024]}
\expandafter\gdef\csname odunum@val@cascade_ladder.rung_delta.gemini_minus_caption.audio.sign_p\endcsname{0.28\ensuremath{\times}}
\expandafter\gdef\csname odunum@n@cascade_ladder.rung_delta.gemini_minus_caption.audio.sign_p\endcsname{171}
\expandafter\gdef\csname odunum@val@cascade_ladder.rung_delta.gemini_minus_caption.history\endcsname{-0.087}
\expandafter\gdef\csname odunum@n@cascade_ladder.rung_delta.gemini_minus_caption.history\endcsname{470}
\expandafter\gdef\csname odunum@ci@cascade_ladder.rung_delta.gemini_minus_caption.history\endcsname{[-0.136, -0.038]}
\expandafter\gdef\csname odunum@val@cascade_ladder.rung_delta.gemini_minus_caption.history.sign_p\endcsname{4.3\ensuremath{\times 10^{-4}}\ensuremath{\times}}
\expandafter\gdef\csname odunum@n@cascade_ladder.rung_delta.gemini_minus_caption.history.sign_p\endcsname{131}
\expandafter\gdef\csname odunum@val@cascade_ladder.rung_delta.gemini_minus_caption.intent\endcsname{0.014}
\expandafter\gdef\csname odunum@n@cascade_ladder.rung_delta.gemini_minus_caption.intent\endcsname{1\,947}
\expandafter\gdef\csname odunum@ci@cascade_ladder.rung_delta.gemini_minus_caption.intent\endcsname{[-0.006, 0.033]}
\expandafter\gdef\csname odunum@val@cascade_ladder.rung_delta.gemini_minus_caption.intent.sign_p\endcsname{0.47\ensuremath{\times}}
\expandafter\gdef\csname odunum@n@cascade_ladder.rung_delta.gemini_minus_caption.intent.sign_p\endcsname{281}
\expandafter\gdef\csname odunum@val@cascade_ladder.rung_delta.gemini_minus_caption.other_context\endcsname{-0.256}
\expandafter\gdef\csname odunum@n@cascade_ladder.rung_delta.gemini_minus_caption.other_context\endcsname{133}
\expandafter\gdef\csname odunum@ci@cascade_ladder.rung_delta.gemini_minus_caption.other_context\endcsname{[-0.348, -0.160]}
\expandafter\gdef\csname odunum@val@cascade_ladder.rung_delta.gemini_minus_caption.other_context.sign_p\endcsname{2.0\ensuremath{\times 10^{-6}}\ensuremath{\times}}
\expandafter\gdef\csname odunum@n@cascade_ladder.rung_delta.gemini_minus_caption.other_context.sign_p\endcsname{52}
\expandafter\gdef\csname odunum@val@cascade_ladder.rung_delta.gemini_minus_caption.visual\endcsname{0.026}
\expandafter\gdef\csname odunum@n@cascade_ladder.rung_delta.gemini_minus_caption.visual\endcsname{791}
\expandafter\gdef\csname odunum@ci@cascade_ladder.rung_delta.gemini_minus_caption.visual\endcsname{[-0.010, 0.065]}
\expandafter\gdef\csname odunum@val@cascade_ladder.rung_delta.gemini_minus_caption.visual.sign_p\endcsname{0.31\ensuremath{\times}}
\expandafter\gdef\csname odunum@n@cascade_ladder.rung_delta.gemini_minus_caption.visual.sign_p\endcsname{216}
\expandafter\gdef\csname odunum@val@cascade_ladder.rung_delta.qwen_plus_minus_caption.audio\endcsname{-0.040}
\expandafter\gdef\csname odunum@n@cascade_ladder.rung_delta.qwen_plus_minus_caption.audio\endcsname{547}
\expandafter\gdef\csname odunum@ci@cascade_ladder.rung_delta.qwen_plus_minus_caption.audio\endcsname{[-0.088, 0.007]}
\expandafter\gdef\csname odunum@val@cascade_ladder.rung_delta.qwen_plus_minus_caption.audio.sign_p\endcsname{0.08\ensuremath{\times}}
\expandafter\gdef\csname odunum@n@cascade_ladder.rung_delta.qwen_plus_minus_caption.audio.sign_p\endcsname{168}
\expandafter\gdef\csname odunum@val@cascade_ladder.rung_delta.qwen_plus_minus_caption.history\endcsname{0.070}
\expandafter\gdef\csname odunum@n@cascade_ladder.rung_delta.qwen_plus_minus_caption.history\endcsname{470}
\expandafter\gdef\csname odunum@ci@cascade_ladder.rung_delta.qwen_plus_minus_caption.history\endcsname{[0.021, 0.119]}
\expandafter\gdef\csname odunum@val@cascade_ladder.rung_delta.qwen_plus_minus_caption.history.sign_p\endcsname{0.01\ensuremath{\times}}
\expandafter\gdef\csname odunum@n@cascade_ladder.rung_delta.qwen_plus_minus_caption.history.sign_p\endcsname{120}
\expandafter\gdef\csname odunum@val@cascade_ladder.rung_delta.qwen_plus_minus_caption.intent\endcsname{0.024}
\expandafter\gdef\csname odunum@n@cascade_ladder.rung_delta.qwen_plus_minus_caption.intent\endcsname{1\,947}
\expandafter\gdef\csname odunum@ci@cascade_ladder.rung_delta.qwen_plus_minus_caption.intent\endcsname{[0.005, 0.042]}
\expandafter\gdef\csname odunum@val@cascade_ladder.rung_delta.qwen_plus_minus_caption.intent.sign_p\endcsname{0.09\ensuremath{\times}}
\expandafter\gdef\csname odunum@n@cascade_ladder.rung_delta.qwen_plus_minus_caption.intent.sign_p\endcsname{235}
\expandafter\gdef\csname odunum@val@cascade_ladder.rung_delta.qwen_plus_minus_caption.other_context\endcsname{-0.030}
\expandafter\gdef\csname odunum@n@cascade_ladder.rung_delta.qwen_plus_minus_caption.other_context\endcsname{133}
\expandafter\gdef\csname odunum@ci@cascade_ladder.rung_delta.qwen_plus_minus_caption.other_context\endcsname{[-0.111, 0.054]}
\expandafter\gdef\csname odunum@val@cascade_ladder.rung_delta.qwen_plus_minus_caption.other_context.sign_p\endcsname{0.60\ensuremath{\times}}
\expandafter\gdef\csname odunum@n@cascade_ladder.rung_delta.qwen_plus_minus_caption.other_context.sign_p\endcsname{32}
\expandafter\gdef\csname odunum@val@cascade_ladder.rung_delta.qwen_plus_minus_caption.visual\endcsname{0.018}
\expandafter\gdef\csname odunum@n@cascade_ladder.rung_delta.qwen_plus_minus_caption.visual\endcsname{791}
\expandafter\gdef\csname odunum@ci@cascade_ladder.rung_delta.qwen_plus_minus_caption.visual\endcsname{[-0.019, 0.055]}
\expandafter\gdef\csname odunum@val@cascade_ladder.rung_delta.qwen_plus_minus_caption.visual.sign_p\endcsname{0.53\ensuremath{\times}}
\expandafter\gdef\csname odunum@n@cascade_ladder.rung_delta.qwen_plus_minus_caption.visual.sign_p\endcsname{200}
\expandafter\gdef\csname odunum@val@cascade_ladder.rung_delta.seed_minus_caption.audio\endcsname{-0.033}
\expandafter\gdef\csname odunum@n@cascade_ladder.rung_delta.seed_minus_caption.audio\endcsname{547}
\expandafter\gdef\csname odunum@ci@cascade_ladder.rung_delta.seed_minus_caption.audio\endcsname{[-0.083, 0.017]}
\expandafter\gdef\csname odunum@val@cascade_ladder.rung_delta.seed_minus_caption.audio.sign_p\endcsname{0.16\ensuremath{\times}}
\expandafter\gdef\csname odunum@n@cascade_ladder.rung_delta.seed_minus_caption.audio.sign_p\endcsname{184}
\expandafter\gdef\csname odunum@val@cascade_ladder.rung_delta.seed_minus_caption.history\endcsname{0.113}
\expandafter\gdef\csname odunum@n@cascade_ladder.rung_delta.seed_minus_caption.history\endcsname{470}
\expandafter\gdef\csname odunum@ci@cascade_ladder.rung_delta.seed_minus_caption.history\endcsname{[0.062, 0.164]}
\expandafter\gdef\csname odunum@val@cascade_ladder.rung_delta.seed_minus_caption.history.sign_p\endcsname{7.8\ensuremath{\times 10^{-5}}\ensuremath{\times}}
\expandafter\gdef\csname odunum@n@cascade_ladder.rung_delta.seed_minus_caption.history.sign_p\endcsname{144}
\expandafter\gdef\csname odunum@val@cascade_ladder.rung_delta.seed_minus_caption.intent\endcsname{-0.056}
\expandafter\gdef\csname odunum@n@cascade_ladder.rung_delta.seed_minus_caption.intent\endcsname{1\,947}
\expandafter\gdef\csname odunum@ci@cascade_ladder.rung_delta.seed_minus_caption.intent\endcsname{[-0.082, -0.029]}
\expandafter\gdef\csname odunum@val@cascade_ladder.rung_delta.seed_minus_caption.intent.sign_p\endcsname{2.0\ensuremath{\times 10^{-3}}\ensuremath{\times}}
\expandafter\gdef\csname odunum@n@cascade_ladder.rung_delta.seed_minus_caption.intent.sign_p\endcsname{329}
\expandafter\gdef\csname odunum@val@cascade_ladder.rung_delta.seed_minus_caption.other_context\endcsname{-0.113}
\expandafter\gdef\csname odunum@n@cascade_ladder.rung_delta.seed_minus_caption.other_context\endcsname{133}
\expandafter\gdef\csname odunum@ci@cascade_ladder.rung_delta.seed_minus_caption.other_context\endcsname{[-0.217, -0.008]}
\expandafter\gdef\csname odunum@val@cascade_ladder.rung_delta.seed_minus_caption.other_context.sign_p\endcsname{0.06\ensuremath{\times}}
\expandafter\gdef\csname odunum@n@cascade_ladder.rung_delta.seed_minus_caption.other_context.sign_p\endcsname{55}
\expandafter\gdef\csname odunum@val@cascade_ladder.rung_delta.seed_minus_caption.visual\endcsname{0.087}
\expandafter\gdef\csname odunum@n@cascade_ladder.rung_delta.seed_minus_caption.visual\endcsname{791}
\expandafter\gdef\csname odunum@ci@cascade_ladder.rung_delta.seed_minus_caption.visual\endcsname{[0.048, 0.127]}
\expandafter\gdef\csname odunum@val@cascade_ladder.rung_delta.seed_minus_caption.visual.sign_p\endcsname{3.4\ensuremath{\times 10^{-5}}\ensuremath{\times}}
\expandafter\gdef\csname odunum@n@cascade_ladder.rung_delta.seed_minus_caption.visual.sign_p\endcsname{241}
\expandafter\gdef\csname odunum@val@cascade_ladder.scene.ao.caption.avg\endcsname{0.746}
\expandafter\gdef\csname odunum@n@cascade_ladder.scene.ao.caption.avg\endcsname{731}
\expandafter\gdef\csname odunum@val@cascade_ladder.scene.ao.caption.ftr\endcsname{0.317}
\expandafter\gdef\csname odunum@n@cascade_ladder.scene.ao.caption.ftr\endcsname{161}
\expandafter\gdef\csname odunum@ci@cascade_ladder.scene.ao.caption.ftr\endcsname{[0.250, 0.392]}
\expandafter\gdef\csname odunum@val@cascade_ladder.scene.ao.caption.m1\endcsname{0.814}
\expandafter\gdef\csname odunum@n@cascade_ladder.scene.ao.caption.m1\endcsname{731}
\expandafter\gdef\csname odunum@val@cascade_ladder.scene.ao.caption.m2\endcsname{0.697}
\expandafter\gdef\csname odunum@n@cascade_ladder.scene.ao.caption.m2\endcsname{570}
\expandafter\gdef\csname odunum@ci@cascade_ladder.scene.ao.caption.m2\endcsname{[0.673, 0.721]}
\expandafter\gdef\csname odunum@val@cascade_ladder.scene.ao.caption.m3\endcsname{0.746}
\expandafter\gdef\csname odunum@n@cascade_ladder.scene.ao.caption.m3\endcsname{570}
\expandafter\gdef\csname odunum@val@cascade_ladder.scene.ao.caption.m4\endcsname{0.845}
\expandafter\gdef\csname odunum@n@cascade_ladder.scene.ao.caption.m4\endcsname{570}
\expandafter\gdef\csname odunum@val@cascade_ladder.scene.ao.caption.m5\endcsname{0.919}
\expandafter\gdef\csname odunum@n@cascade_ladder.scene.ao.caption.m5\endcsname{570}
\expandafter\gdef\csname odunum@val@cascade_ladder.scene.ao.gemini.avg\endcsname{0.754}
\expandafter\gdef\csname odunum@n@cascade_ladder.scene.ao.gemini.avg\endcsname{731}
\expandafter\gdef\csname odunum@val@cascade_ladder.scene.ao.gemini.ftr\endcsname{0.317}
\expandafter\gdef\csname odunum@n@cascade_ladder.scene.ao.gemini.ftr\endcsname{161}
\expandafter\gdef\csname odunum@ci@cascade_ladder.scene.ao.gemini.ftr\endcsname{[0.250, 0.392]}
\expandafter\gdef\csname odunum@val@cascade_ladder.scene.ao.gemini.m1\endcsname{0.873}
\expandafter\gdef\csname odunum@n@cascade_ladder.scene.ao.gemini.m1\endcsname{731}
\expandafter\gdef\csname odunum@val@cascade_ladder.scene.ao.gemini.m2\endcsname{0.701}
\expandafter\gdef\csname odunum@n@cascade_ladder.scene.ao.gemini.m2\endcsname{570}
\expandafter\gdef\csname odunum@ci@cascade_ladder.scene.ao.gemini.m2\endcsname{[0.680, 0.722]}
\expandafter\gdef\csname odunum@val@cascade_ladder.scene.ao.gemini.m3\endcsname{0.615}
\expandafter\gdef\csname odunum@n@cascade_ladder.scene.ao.gemini.m3\endcsname{570}
\expandafter\gdef\csname odunum@val@cascade_ladder.scene.ao.gemini.m4\endcsname{0.917}
\expandafter\gdef\csname odunum@n@cascade_ladder.scene.ao.gemini.m4\endcsname{570}
\expandafter\gdef\csname odunum@val@cascade_ladder.scene.ao.gemini.m5\endcsname{0.976}
\expandafter\gdef\csname odunum@n@cascade_ladder.scene.ao.gemini.m5\endcsname{570}
\expandafter\gdef\csname odunum@val@cascade_ladder.scene.ao.qwen_plus.avg\endcsname{0.747}
\expandafter\gdef\csname odunum@n@cascade_ladder.scene.ao.qwen_plus.avg\endcsname{731}
\expandafter\gdef\csname odunum@val@cascade_ladder.scene.ao.qwen_plus.ftr\endcsname{0.553}
\expandafter\gdef\csname odunum@n@cascade_ladder.scene.ao.qwen_plus.ftr\endcsname{161}
\expandafter\gdef\csname odunum@ci@cascade_ladder.scene.ao.qwen_plus.ftr\endcsname{[0.476, 0.627]}
\expandafter\gdef\csname odunum@val@cascade_ladder.scene.ao.qwen_plus.m1\endcsname{0.762}
\expandafter\gdef\csname odunum@n@cascade_ladder.scene.ao.qwen_plus.m1\endcsname{731}
\expandafter\gdef\csname odunum@val@cascade_ladder.scene.ao.qwen_plus.m2\endcsname{0.715}
\expandafter\gdef\csname odunum@n@cascade_ladder.scene.ao.qwen_plus.m2\endcsname{570}
\expandafter\gdef\csname odunum@ci@cascade_ladder.scene.ao.qwen_plus.m2\endcsname{[0.694, 0.737]}
\expandafter\gdef\csname odunum@val@cascade_ladder.scene.ao.qwen_plus.m3\endcsname{0.661}
\expandafter\gdef\csname odunum@n@cascade_ladder.scene.ao.qwen_plus.m3\endcsname{570}
\expandafter\gdef\csname odunum@val@cascade_ladder.scene.ao.qwen_plus.m4\endcsname{0.893}
\expandafter\gdef\csname odunum@n@cascade_ladder.scene.ao.qwen_plus.m4\endcsname{570}
\expandafter\gdef\csname odunum@val@cascade_ladder.scene.ao.qwen_plus.m5\endcsname{0.958}
\expandafter\gdef\csname odunum@n@cascade_ladder.scene.ao.qwen_plus.m5\endcsname{570}
\expandafter\gdef\csname odunum@val@cascade_ladder.scene.ao.seed.avg\endcsname{0.773}
\expandafter\gdef\csname odunum@n@cascade_ladder.scene.ao.seed.avg\endcsname{731}
\expandafter\gdef\csname odunum@val@cascade_ladder.scene.ao.seed.ftr\endcsname{0.807}
\expandafter\gdef\csname odunum@n@cascade_ladder.scene.ao.seed.ftr\endcsname{161}
\expandafter\gdef\csname odunum@ci@cascade_ladder.scene.ao.seed.ftr\endcsname{[0.740, 0.861]}
\expandafter\gdef\csname odunum@val@cascade_ladder.scene.ao.seed.m1\endcsname{0.608}
\expandafter\gdef\csname odunum@n@cascade_ladder.scene.ao.seed.m1\endcsname{731}
\expandafter\gdef\csname odunum@val@cascade_ladder.scene.ao.seed.m2\endcsname{0.780}
\expandafter\gdef\csname odunum@n@cascade_ladder.scene.ao.seed.m2\endcsname{570}
\expandafter\gdef\csname odunum@ci@cascade_ladder.scene.ao.seed.m2\endcsname{[0.761, 0.800]}
\expandafter\gdef\csname odunum@val@cascade_ladder.scene.ao.seed.m3\endcsname{0.715}
\expandafter\gdef\csname odunum@n@cascade_ladder.scene.ao.seed.m3\endcsname{570}
\expandafter\gdef\csname odunum@val@cascade_ladder.scene.ao.seed.m4\endcsname{0.932}
\expandafter\gdef\csname odunum@n@cascade_ladder.scene.ao.seed.m4\endcsname{570}
\expandafter\gdef\csname odunum@val@cascade_ladder.scene.ao.seed.m5\endcsname{0.975}
\expandafter\gdef\csname odunum@n@cascade_ladder.scene.ao.seed.m5\endcsname{570}
\expandafter\gdef\csname odunum@val@cascade_ladder.scene.ao.transcript.avg\endcsname{0.715}
\expandafter\gdef\csname odunum@n@cascade_ladder.scene.ao.transcript.avg\endcsname{731}
\expandafter\gdef\csname odunum@val@cascade_ladder.scene.ao.transcript.ftr\endcsname{0.404}
\expandafter\gdef\csname odunum@n@cascade_ladder.scene.ao.transcript.ftr\endcsname{161}
\expandafter\gdef\csname odunum@ci@cascade_ladder.scene.ao.transcript.ftr\endcsname{[0.331, 0.481]}
\expandafter\gdef\csname odunum@val@cascade_ladder.scene.ao.transcript.m1\endcsname{0.789}
\expandafter\gdef\csname odunum@n@cascade_ladder.scene.ao.transcript.m1\endcsname{731}
\expandafter\gdef\csname odunum@val@cascade_ladder.scene.ao.transcript.m2\endcsname{0.683}
\expandafter\gdef\csname odunum@n@cascade_ladder.scene.ao.transcript.m2\endcsname{570}
\expandafter\gdef\csname odunum@ci@cascade_ladder.scene.ao.transcript.m2\endcsname{[0.660, 0.707]}
\expandafter\gdef\csname odunum@val@cascade_ladder.scene.ao.transcript.m3\endcsname{0.862}
\expandafter\gdef\csname odunum@n@cascade_ladder.scene.ao.transcript.m3\endcsname{570}
\expandafter\gdef\csname odunum@val@cascade_ladder.scene.ao.transcript.m4\endcsname{0.851}
\expandafter\gdef\csname odunum@n@cascade_ladder.scene.ao.transcript.m4\endcsname{570}
\expandafter\gdef\csname odunum@val@cascade_ladder.scene.ao.transcript.m5\endcsname{0.306}
\expandafter\gdef\csname odunum@n@cascade_ladder.scene.ao.transcript.m5\endcsname{570}
\expandafter\gdef\csname odunum@val@cascade_ladder.scene.av.caption.avg\endcsname{0.684}
\expandafter\gdef\csname odunum@n@cascade_ladder.scene.av.caption.avg\endcsname{1\,346}
\expandafter\gdef\csname odunum@val@cascade_ladder.scene.av.caption.ftr\endcsname{0.619}
\expandafter\gdef\csname odunum@n@cascade_ladder.scene.av.caption.ftr\endcsname{286}
\expandafter\gdef\csname odunum@ci@cascade_ladder.scene.av.caption.ftr\endcsname{[0.561, 0.673]}
\expandafter\gdef\csname odunum@val@cascade_ladder.scene.av.caption.m1\endcsname{0.674}
\expandafter\gdef\csname odunum@n@cascade_ladder.scene.av.caption.m1\endcsname{1\,346}
\expandafter\gdef\csname odunum@val@cascade_ladder.scene.av.caption.m2\endcsname{0.641}
\expandafter\gdef\csname odunum@n@cascade_ladder.scene.av.caption.m2\endcsname{1\,060}
\expandafter\gdef\csname odunum@ci@cascade_ladder.scene.av.caption.m2\endcsname{[0.624, 0.659]}
\expandafter\gdef\csname odunum@val@cascade_ladder.scene.av.caption.m3\endcsname{0.698}
\expandafter\gdef\csname odunum@n@cascade_ladder.scene.av.caption.m3\endcsname{1\,060}
\expandafter\gdef\csname odunum@val@cascade_ladder.scene.av.caption.m4\endcsname{0.831}
\expandafter\gdef\csname odunum@n@cascade_ladder.scene.av.caption.m4\endcsname{1\,060}
\expandafter\gdef\csname odunum@val@cascade_ladder.scene.av.caption.m5\endcsname{0.891}
\expandafter\gdef\csname odunum@n@cascade_ladder.scene.av.caption.m5\endcsname{1\,060}
\expandafter\gdef\csname odunum@val@cascade_ladder.scene.av.gemini.avg\endcsname{0.726}
\expandafter\gdef\csname odunum@n@cascade_ladder.scene.av.gemini.avg\endcsname{1\,346}
\expandafter\gdef\csname odunum@val@cascade_ladder.scene.av.gemini.ftr\endcsname{0.406}
\expandafter\gdef\csname odunum@n@cascade_ladder.scene.av.gemini.ftr\endcsname{286}
\expandafter\gdef\csname odunum@ci@cascade_ladder.scene.av.gemini.ftr\endcsname{[0.350, 0.463]}
\expandafter\gdef\csname odunum@val@cascade_ladder.scene.av.gemini.m1\endcsname{0.830}
\expandafter\gdef\csname odunum@n@cascade_ladder.scene.av.gemini.m1\endcsname{1\,346}
\expandafter\gdef\csname odunum@val@cascade_ladder.scene.av.gemini.m2\endcsname{0.633}
\expandafter\gdef\csname odunum@n@cascade_ladder.scene.av.gemini.m2\endcsname{1\,060}
\expandafter\gdef\csname odunum@ci@cascade_ladder.scene.av.gemini.m2\endcsname{[0.619, 0.648]}
\expandafter\gdef\csname odunum@val@cascade_ladder.scene.av.gemini.m3\endcsname{0.819}
\expandafter\gdef\csname odunum@n@cascade_ladder.scene.av.gemini.m3\endcsname{1\,060}
\expandafter\gdef\csname odunum@val@cascade_ladder.scene.av.gemini.m4\endcsname{0.913}
\expandafter\gdef\csname odunum@n@cascade_ladder.scene.av.gemini.m4\endcsname{1\,060}
\expandafter\gdef\csname odunum@val@cascade_ladder.scene.av.gemini.m5\endcsname{0.957}
\expandafter\gdef\csname odunum@n@cascade_ladder.scene.av.gemini.m5\endcsname{1\,060}
\expandafter\gdef\csname odunum@val@cascade_ladder.scene.av.qwen_plus.avg\endcsname{0.696}
\expandafter\gdef\csname odunum@n@cascade_ladder.scene.av.qwen_plus.avg\endcsname{1\,346}
\expandafter\gdef\csname odunum@val@cascade_ladder.scene.av.qwen_plus.ftr\endcsname{0.738}
\expandafter\gdef\csname odunum@n@cascade_ladder.scene.av.qwen_plus.ftr\endcsname{286}
\expandafter\gdef\csname odunum@ci@cascade_ladder.scene.av.qwen_plus.ftr\endcsname{[0.684, 0.785]}
\expandafter\gdef\csname odunum@val@cascade_ladder.scene.av.qwen_plus.m1\endcsname{0.639}
\expandafter\gdef\csname odunum@n@cascade_ladder.scene.av.qwen_plus.m1\endcsname{1\,346}
\expandafter\gdef\csname odunum@val@cascade_ladder.scene.av.qwen_plus.m2\endcsname{0.658}
\expandafter\gdef\csname odunum@n@cascade_ladder.scene.av.qwen_plus.m2\endcsname{1\,060}
\expandafter\gdef\csname odunum@ci@cascade_ladder.scene.av.qwen_plus.m2\endcsname{[0.642, 0.673]}
\expandafter\gdef\csname odunum@val@cascade_ladder.scene.av.qwen_plus.m3\endcsname{0.757}
\expandafter\gdef\csname odunum@n@cascade_ladder.scene.av.qwen_plus.m3\endcsname{1\,060}
\expandafter\gdef\csname odunum@val@cascade_ladder.scene.av.qwen_plus.m4\endcsname{0.833}
\expandafter\gdef\csname odunum@n@cascade_ladder.scene.av.qwen_plus.m4\endcsname{1\,060}
\expandafter\gdef\csname odunum@val@cascade_ladder.scene.av.qwen_plus.m5\endcsname{0.935}
\expandafter\gdef\csname odunum@n@cascade_ladder.scene.av.qwen_plus.m5\endcsname{1\,060}
\expandafter\gdef\csname odunum@val@cascade_ladder.scene.av.seed.avg\endcsname{0.679}
\expandafter\gdef\csname odunum@n@cascade_ladder.scene.av.seed.avg\endcsname{1\,346}
\expandafter\gdef\csname odunum@val@cascade_ladder.scene.av.seed.ftr\endcsname{0.822}
\expandafter\gdef\csname odunum@n@cascade_ladder.scene.av.seed.ftr\endcsname{286}
\expandafter\gdef\csname odunum@ci@cascade_ladder.scene.av.seed.ftr\endcsname{[0.773, 0.862]}
\expandafter\gdef\csname odunum@val@cascade_ladder.scene.av.seed.m1\endcsname{0.581}
\expandafter\gdef\csname odunum@n@cascade_ladder.scene.av.seed.m1\endcsname{1\,346}
\expandafter\gdef\csname odunum@val@cascade_ladder.scene.av.seed.m2\endcsname{0.642}
\expandafter\gdef\csname odunum@n@cascade_ladder.scene.av.seed.m2\endcsname{1\,060}
\expandafter\gdef\csname odunum@ci@cascade_ladder.scene.av.seed.m2\endcsname{[0.623, 0.660]}
\expandafter\gdef\csname odunum@val@cascade_ladder.scene.av.seed.m3\endcsname{0.828}
\expandafter\gdef\csname odunum@n@cascade_ladder.scene.av.seed.m3\endcsname{1\,060}
\expandafter\gdef\csname odunum@val@cascade_ladder.scene.av.seed.m4\endcsname{0.776}
\expandafter\gdef\csname odunum@n@cascade_ladder.scene.av.seed.m4\endcsname{1\,060}
\expandafter\gdef\csname odunum@val@cascade_ladder.scene.av.seed.m5\endcsname{0.925}
\expandafter\gdef\csname odunum@n@cascade_ladder.scene.av.seed.m5\endcsname{1\,060}
\expandafter\gdef\csname odunum@val@cascade_ladder.scene.av.transcript.avg\endcsname{0.590}
\expandafter\gdef\csname odunum@n@cascade_ladder.scene.av.transcript.avg\endcsname{1\,346}
\expandafter\gdef\csname odunum@val@cascade_ladder.scene.av.transcript.ftr\endcsname{0.734}
\expandafter\gdef\csname odunum@n@cascade_ladder.scene.av.transcript.ftr\endcsname{286}
\expandafter\gdef\csname odunum@ci@cascade_ladder.scene.av.transcript.ftr\endcsname{[0.680, 0.782]}
\expandafter\gdef\csname odunum@val@cascade_ladder.scene.av.transcript.m1\endcsname{0.606}
\expandafter\gdef\csname odunum@n@cascade_ladder.scene.av.transcript.m1\endcsname{1\,346}
\expandafter\gdef\csname odunum@val@cascade_ladder.scene.av.transcript.m2\endcsname{0.556}
\expandafter\gdef\csname odunum@n@cascade_ladder.scene.av.transcript.m2\endcsname{1\,060}
\expandafter\gdef\csname odunum@ci@cascade_ladder.scene.av.transcript.m2\endcsname{[0.539, 0.573]}
\expandafter\gdef\csname odunum@val@cascade_ladder.scene.av.transcript.m3\endcsname{0.823}
\expandafter\gdef\csname odunum@n@cascade_ladder.scene.av.transcript.m3\endcsname{1\,060}
\expandafter\gdef\csname odunum@val@cascade_ladder.scene.av.transcript.m4\endcsname{0.819}
\expandafter\gdef\csname odunum@n@cascade_ladder.scene.av.transcript.m4\endcsname{1\,060}
\expandafter\gdef\csname odunum@val@cascade_ladder.scene.av.transcript.m5\endcsname{0.026}
\expandafter\gdef\csname odunum@n@cascade_ladder.scene.av.transcript.m5\endcsname{1\,060}
\expandafter\gdef\csname odunum@val@cascade_redundancy.band.cascade_asr.audio.hi\endcsname{0.464}
\expandafter\gdef\csname odunum@n@cascade_redundancy.band.cascade_asr.audio.hi\endcsname{56}
\expandafter\gdef\csname odunum@ci@cascade_redundancy.band.cascade_asr.audio.hi\endcsname{[0.339, 0.589]}
\expandafter\gdef\csname odunum@val@cascade_redundancy.band.cascade_asr.audio.hi.latin\endcsname{0.464}
\expandafter\gdef\csname odunum@n@cascade_redundancy.band.cascade_asr.audio.hi.latin\endcsname{56}
\expandafter\gdef\csname odunum@ci@cascade_redundancy.band.cascade_asr.audio.hi.latin\endcsname{[0.339, 0.589]}
\expandafter\gdef\csname odunum@val@cascade_redundancy.band.cascade_asr.audio.lo\endcsname{0.312}
\expandafter\gdef\csname odunum@n@cascade_redundancy.band.cascade_asr.audio.lo\endcsname{568}
\expandafter\gdef\csname odunum@ci@cascade_redundancy.band.cascade_asr.audio.lo\endcsname{[0.273, 0.350]}
\expandafter\gdef\csname odunum@val@cascade_redundancy.band.cascade_asr.audio.lo.latin\endcsname{0.156}
\expandafter\gdef\csname odunum@n@cascade_redundancy.band.cascade_asr.audio.lo.latin\endcsname{141}
\expandafter\gdef\csname odunum@ci@cascade_redundancy.band.cascade_asr.audio.lo.latin\endcsname{[0.099, 0.218]}
\expandafter\gdef\csname odunum@val@cascade_redundancy.band.cascade_asr.audio.mid1\endcsname{0.347}
\expandafter\gdef\csname odunum@n@cascade_redundancy.band.cascade_asr.audio.mid1\endcsname{167}
\expandafter\gdef\csname odunum@ci@cascade_redundancy.band.cascade_asr.audio.mid1\endcsname{[0.275, 0.419]}
\expandafter\gdef\csname odunum@val@cascade_redundancy.band.cascade_asr.audio.mid1.latin\endcsname{0.341}
\expandafter\gdef\csname odunum@n@cascade_redundancy.band.cascade_asr.audio.mid1.latin\endcsname{135}
\expandafter\gdef\csname odunum@ci@cascade_redundancy.band.cascade_asr.audio.mid1.latin\endcsname{[0.263, 0.419]}
\expandafter\gdef\csname odunum@val@cascade_redundancy.band.cascade_asr.audio.mid2\endcsname{0.407}
\expandafter\gdef\csname odunum@n@cascade_redundancy.band.cascade_asr.audio.mid2\endcsname{91}
\expandafter\gdef\csname odunum@ci@cascade_redundancy.band.cascade_asr.audio.mid2\endcsname{[0.308, 0.511]}
\expandafter\gdef\csname odunum@val@cascade_redundancy.band.cascade_asr.audio.mid2.latin\endcsname{0.388}
\expandafter\gdef\csname odunum@n@cascade_redundancy.band.cascade_asr.audio.mid2.latin\endcsname{85}
\expandafter\gdef\csname odunum@ci@cascade_redundancy.band.cascade_asr.audio.mid2.latin\endcsname{[0.286, 0.494]}
\expandafter\gdef\csname odunum@val@cascade_redundancy.band.cascade_asr.visual.hi\endcsname{0.562}
\expandafter\gdef\csname odunum@n@cascade_redundancy.band.cascade_asr.visual.hi\endcsname{16}
\expandafter\gdef\csname odunum@ci@cascade_redundancy.band.cascade_asr.visual.hi\endcsname{[0.312, 0.812]}
\expandafter\gdef\csname odunum@val@cascade_redundancy.band.cascade_asr.visual.hi.latin\endcsname{0.562}
\expandafter\gdef\csname odunum@n@cascade_redundancy.band.cascade_asr.visual.hi.latin\endcsname{16}
\expandafter\gdef\csname odunum@ci@cascade_redundancy.band.cascade_asr.visual.hi.latin\endcsname{[0.312, 0.812]}
\expandafter\gdef\csname odunum@val@cascade_redundancy.band.cascade_asr.visual.lo\endcsname{0.174}
\expandafter\gdef\csname odunum@n@cascade_redundancy.band.cascade_asr.visual.lo\endcsname{627}
\expandafter\gdef\csname odunum@ci@cascade_redundancy.band.cascade_asr.visual.lo\endcsname{[0.145, 0.204]}
\expandafter\gdef\csname odunum@val@cascade_redundancy.band.cascade_asr.visual.lo.latin\endcsname{0.107}
\expandafter\gdef\csname odunum@n@cascade_redundancy.band.cascade_asr.visual.lo.latin\endcsname{187}
\expandafter\gdef\csname odunum@ci@cascade_redundancy.band.cascade_asr.visual.lo.latin\endcsname{[0.064, 0.153]}
\expandafter\gdef\csname odunum@val@cascade_redundancy.band.cascade_asr.visual.mid1\endcsname{0.255}
\expandafter\gdef\csname odunum@n@cascade_redundancy.band.cascade_asr.visual.mid1\endcsname{110}
\expandafter\gdef\csname odunum@ci@cascade_redundancy.band.cascade_asr.visual.mid1\endcsname{[0.176, 0.336]}
\expandafter\gdef\csname odunum@val@cascade_redundancy.band.cascade_asr.visual.mid1.latin\endcsname{0.250}
\expandafter\gdef\csname odunum@n@cascade_redundancy.band.cascade_asr.visual.mid1.latin\endcsname{104}
\expandafter\gdef\csname odunum@ci@cascade_redundancy.band.cascade_asr.visual.mid1.latin\endcsname{[0.168, 0.333]}
\expandafter\gdef\csname odunum@val@cascade_redundancy.band.cascade_asr.visual.mid2\endcsname{0.421}
\expandafter\gdef\csname odunum@n@cascade_redundancy.band.cascade_asr.visual.mid2\endcsname{38}
\expandafter\gdef\csname odunum@ci@cascade_redundancy.band.cascade_asr.visual.mid2\endcsname{[0.263, 0.579]}
\expandafter\gdef\csname odunum@val@cascade_redundancy.band.cascade_asr.visual.mid2.latin\endcsname{0.421}
\expandafter\gdef\csname odunum@n@cascade_redundancy.band.cascade_asr.visual.mid2.latin\endcsname{38}
\expandafter\gdef\csname odunum@ci@cascade_redundancy.band.cascade_asr.visual.mid2.latin\endcsname{[0.263, 0.579]}
\expandafter\gdef\csname odunum@val@cascade_redundancy.band.gemini.audio.hi\endcsname{0.554}
\expandafter\gdef\csname odunum@n@cascade_redundancy.band.gemini.audio.hi\endcsname{56}
\expandafter\gdef\csname odunum@ci@cascade_redundancy.band.gemini.audio.hi\endcsname{[0.429, 0.679]}
\expandafter\gdef\csname odunum@val@cascade_redundancy.band.gemini.audio.hi.latin\endcsname{0.554}
\expandafter\gdef\csname odunum@n@cascade_redundancy.band.gemini.audio.hi.latin\endcsname{56}
\expandafter\gdef\csname odunum@ci@cascade_redundancy.band.gemini.audio.hi.latin\endcsname{[0.429, 0.679]}
\expandafter\gdef\csname odunum@val@cascade_redundancy.band.gemini.audio.lo\endcsname{0.461}
\expandafter\gdef\csname odunum@n@cascade_redundancy.band.gemini.audio.lo\endcsname{568}
\expandafter\gdef\csname odunum@ci@cascade_redundancy.band.gemini.audio.lo\endcsname{[0.420, 0.503]}
\expandafter\gdef\csname odunum@val@cascade_redundancy.band.gemini.audio.lo.latin\endcsname{0.433}
\expandafter\gdef\csname odunum@n@cascade_redundancy.band.gemini.audio.lo.latin\endcsname{141}
\expandafter\gdef\csname odunum@ci@cascade_redundancy.band.gemini.audio.lo.latin\endcsname{[0.352, 0.510]}
\expandafter\gdef\csname odunum@val@cascade_redundancy.band.gemini.audio.mid1\endcsname{0.515}
\expandafter\gdef\csname odunum@n@cascade_redundancy.band.gemini.audio.mid1\endcsname{167}
\expandafter\gdef\csname odunum@ci@cascade_redundancy.band.gemini.audio.mid1\endcsname{[0.439, 0.592]}
\expandafter\gdef\csname odunum@val@cascade_redundancy.band.gemini.audio.mid1.latin\endcsname{0.519}
\expandafter\gdef\csname odunum@n@cascade_redundancy.band.gemini.audio.mid1.latin\endcsname{135}
\expandafter\gdef\csname odunum@ci@cascade_redundancy.band.gemini.audio.mid1.latin\endcsname{[0.432, 0.602]}
\expandafter\gdef\csname odunum@val@cascade_redundancy.band.gemini.audio.mid2\endcsname{0.527}
\expandafter\gdef\csname odunum@n@cascade_redundancy.band.gemini.audio.mid2\endcsname{91}
\expandafter\gdef\csname odunum@ci@cascade_redundancy.band.gemini.audio.mid2\endcsname{[0.424, 0.626]}
\expandafter\gdef\csname odunum@val@cascade_redundancy.band.gemini.audio.mid2.latin\endcsname{0.529}
\expandafter\gdef\csname odunum@n@cascade_redundancy.band.gemini.audio.mid2.latin\endcsname{85}
\expandafter\gdef\csname odunum@ci@cascade_redundancy.band.gemini.audio.mid2.latin\endcsname{[0.423, 0.635]}
\expandafter\gdef\csname odunum@val@cascade_redundancy.band.gemini.visual.hi\endcsname{0.688}
\expandafter\gdef\csname odunum@n@cascade_redundancy.band.gemini.visual.hi\endcsname{16}
\expandafter\gdef\csname odunum@ci@cascade_redundancy.band.gemini.visual.hi\endcsname{[0.438, 0.875]}
\expandafter\gdef\csname odunum@val@cascade_redundancy.band.gemini.visual.hi.latin\endcsname{0.688}
\expandafter\gdef\csname odunum@n@cascade_redundancy.band.gemini.visual.hi.latin\endcsname{16}
\expandafter\gdef\csname odunum@ci@cascade_redundancy.band.gemini.visual.hi.latin\endcsname{[0.438, 0.875]}
\expandafter\gdef\csname odunum@val@cascade_redundancy.band.gemini.visual.lo\endcsname{0.453}
\expandafter\gdef\csname odunum@n@cascade_redundancy.band.gemini.visual.lo\endcsname{627}
\expandafter\gdef\csname odunum@ci@cascade_redundancy.band.gemini.visual.lo\endcsname{[0.412, 0.493]}
\expandafter\gdef\csname odunum@val@cascade_redundancy.band.gemini.visual.lo.latin\endcsname{0.369}
\expandafter\gdef\csname odunum@n@cascade_redundancy.band.gemini.visual.lo.latin\endcsname{187}
\expandafter\gdef\csname odunum@ci@cascade_redundancy.band.gemini.visual.lo.latin\endcsname{[0.303, 0.439]}
\expandafter\gdef\csname odunum@val@cascade_redundancy.band.gemini.visual.mid1\endcsname{0.573}
\expandafter\gdef\csname odunum@n@cascade_redundancy.band.gemini.visual.mid1\endcsname{110}
\expandafter\gdef\csname odunum@ci@cascade_redundancy.band.gemini.visual.mid1\endcsname{[0.476, 0.670]}
\expandafter\gdef\csname odunum@val@cascade_redundancy.band.gemini.visual.mid1.latin\endcsname{0.587}
\expandafter\gdef\csname odunum@n@cascade_redundancy.band.gemini.visual.mid1.latin\endcsname{104}
\expandafter\gdef\csname odunum@ci@cascade_redundancy.band.gemini.visual.mid1.latin\endcsname{[0.484, 0.685]}
\expandafter\gdef\csname odunum@val@cascade_redundancy.band.gemini.visual.mid2\endcsname{0.605}
\expandafter\gdef\csname odunum@n@cascade_redundancy.band.gemini.visual.mid2\endcsname{38}
\expandafter\gdef\csname odunum@ci@cascade_redundancy.band.gemini.visual.mid2\endcsname{[0.432, 0.762]}
\expandafter\gdef\csname odunum@val@cascade_redundancy.band.gemini.visual.mid2.latin\endcsname{0.605}
\expandafter\gdef\csname odunum@n@cascade_redundancy.band.gemini.visual.mid2.latin\endcsname{38}
\expandafter\gdef\csname odunum@ci@cascade_redundancy.band.gemini.visual.mid2.latin\endcsname{[0.432, 0.762]}
\expandafter\gdef\csname odunum@val@cascade_redundancy.band.qwen_plus.audio.hi\endcsname{0.518}
\expandafter\gdef\csname odunum@n@cascade_redundancy.band.qwen_plus.audio.hi\endcsname{56}
\expandafter\gdef\csname odunum@ci@cascade_redundancy.band.qwen_plus.audio.hi\endcsname{[0.393, 0.643]}
\expandafter\gdef\csname odunum@val@cascade_redundancy.band.qwen_plus.audio.hi.latin\endcsname{0.518}
\expandafter\gdef\csname odunum@n@cascade_redundancy.band.qwen_plus.audio.hi.latin\endcsname{56}
\expandafter\gdef\csname odunum@ci@cascade_redundancy.band.qwen_plus.audio.hi.latin\endcsname{[0.393, 0.643]}
\expandafter\gdef\csname odunum@val@cascade_redundancy.band.qwen_plus.audio.lo\endcsname{0.451}
\expandafter\gdef\csname odunum@n@cascade_redundancy.band.qwen_plus.audio.lo\endcsname{568}
\expandafter\gdef\csname odunum@ci@cascade_redundancy.band.qwen_plus.audio.lo\endcsname{[0.410, 0.492]}
\expandafter\gdef\csname odunum@val@cascade_redundancy.band.qwen_plus.audio.lo.latin\endcsname{0.468}
\expandafter\gdef\csname odunum@n@cascade_redundancy.band.qwen_plus.audio.lo.latin\endcsname{141}
\expandafter\gdef\csname odunum@ci@cascade_redundancy.band.qwen_plus.audio.lo.latin\endcsname{[0.387, 0.549]}
\expandafter\gdef\csname odunum@val@cascade_redundancy.band.qwen_plus.audio.mid1\endcsname{0.473}
\expandafter\gdef\csname odunum@n@cascade_redundancy.band.qwen_plus.audio.mid1\endcsname{167}
\expandafter\gdef\csname odunum@ci@cascade_redundancy.band.qwen_plus.audio.mid1\endcsname{[0.398, 0.547]}
\expandafter\gdef\csname odunum@val@cascade_redundancy.band.qwen_plus.audio.mid1.latin\endcsname{0.504}
\expandafter\gdef\csname odunum@n@cascade_redundancy.band.qwen_plus.audio.mid1.latin\endcsname{135}
\expandafter\gdef\csname odunum@ci@cascade_redundancy.band.qwen_plus.audio.mid1.latin\endcsname{[0.418, 0.585]}
\expandafter\gdef\csname odunum@val@cascade_redundancy.band.qwen_plus.audio.mid2\endcsname{0.549}
\expandafter\gdef\csname odunum@n@cascade_redundancy.band.qwen_plus.audio.mid2\endcsname{91}
\expandafter\gdef\csname odunum@ci@cascade_redundancy.band.qwen_plus.audio.mid2\endcsname{[0.444, 0.652]}
\expandafter\gdef\csname odunum@val@cascade_redundancy.band.qwen_plus.audio.mid2.latin\endcsname{0.553}
\expandafter\gdef\csname odunum@n@cascade_redundancy.band.qwen_plus.audio.mid2.latin\endcsname{85}
\expandafter\gdef\csname odunum@ci@cascade_redundancy.band.qwen_plus.audio.mid2.latin\endcsname{[0.447, 0.659]}
\expandafter\gdef\csname odunum@val@cascade_redundancy.band.qwen_plus.visual.hi\endcsname{0.875}
\expandafter\gdef\csname odunum@n@cascade_redundancy.band.qwen_plus.visual.hi\endcsname{16}
\expandafter\gdef\csname odunum@ci@cascade_redundancy.band.qwen_plus.visual.hi\endcsname{[0.688, 1.000]}
\expandafter\gdef\csname odunum@val@cascade_redundancy.band.qwen_plus.visual.hi.latin\endcsname{0.875}
\expandafter\gdef\csname odunum@n@cascade_redundancy.band.qwen_plus.visual.hi.latin\endcsname{16}
\expandafter\gdef\csname odunum@ci@cascade_redundancy.band.qwen_plus.visual.hi.latin\endcsname{[0.688, 1.000]}
\expandafter\gdef\csname odunum@val@cascade_redundancy.band.qwen_plus.visual.lo\endcsname{0.450}
\expandafter\gdef\csname odunum@n@cascade_redundancy.band.qwen_plus.visual.lo\endcsname{627}
\expandafter\gdef\csname odunum@ci@cascade_redundancy.band.qwen_plus.visual.lo\endcsname{[0.411, 0.489]}
\expandafter\gdef\csname odunum@val@cascade_redundancy.band.qwen_plus.visual.lo.latin\endcsname{0.428}
\expandafter\gdef\csname odunum@n@cascade_redundancy.band.qwen_plus.visual.lo.latin\endcsname{187}
\expandafter\gdef\csname odunum@ci@cascade_redundancy.band.qwen_plus.visual.lo.latin\endcsname{[0.358, 0.500]}
\expandafter\gdef\csname odunum@val@cascade_redundancy.band.qwen_plus.visual.mid1\endcsname{0.491}
\expandafter\gdef\csname odunum@n@cascade_redundancy.band.qwen_plus.visual.mid1\endcsname{110}
\expandafter\gdef\csname odunum@ci@cascade_redundancy.band.qwen_plus.visual.mid1\endcsname{[0.395, 0.586]}
\expandafter\gdef\csname odunum@val@cascade_redundancy.band.qwen_plus.visual.mid1.latin\endcsname{0.490}
\expandafter\gdef\csname odunum@n@cascade_redundancy.band.qwen_plus.visual.mid1.latin\endcsname{104}
\expandafter\gdef\csname odunum@ci@cascade_redundancy.band.qwen_plus.visual.mid1.latin\endcsname{[0.392, 0.588]}
\expandafter\gdef\csname odunum@val@cascade_redundancy.band.qwen_plus.visual.mid2\endcsname{0.632}
\expandafter\gdef\csname odunum@n@cascade_redundancy.band.qwen_plus.visual.mid2\endcsname{38}
\expandafter\gdef\csname odunum@ci@cascade_redundancy.band.qwen_plus.visual.mid2\endcsname{[0.474, 0.784]}
\expandafter\gdef\csname odunum@val@cascade_redundancy.band.qwen_plus.visual.mid2.latin\endcsname{0.632}
\expandafter\gdef\csname odunum@n@cascade_redundancy.band.qwen_plus.visual.mid2.latin\endcsname{38}
\expandafter\gdef\csname odunum@ci@cascade_redundancy.band.qwen_plus.visual.mid2.latin\endcsname{[0.474, 0.784]}
\expandafter\gdef\csname odunum@val@cascade_redundancy.band.seed.audio.hi\endcsname{0.500}
\expandafter\gdef\csname odunum@n@cascade_redundancy.band.seed.audio.hi\endcsname{56}
\expandafter\gdef\csname odunum@ci@cascade_redundancy.band.seed.audio.hi\endcsname{[0.375, 0.625]}
\expandafter\gdef\csname odunum@val@cascade_redundancy.band.seed.audio.hi.latin\endcsname{0.500}
\expandafter\gdef\csname odunum@n@cascade_redundancy.band.seed.audio.hi.latin\endcsname{56}
\expandafter\gdef\csname odunum@ci@cascade_redundancy.band.seed.audio.hi.latin\endcsname{[0.375, 0.625]}
\expandafter\gdef\csname odunum@val@cascade_redundancy.band.seed.audio.lo\endcsname{0.449}
\expandafter\gdef\csname odunum@n@cascade_redundancy.band.seed.audio.lo\endcsname{568}
\expandafter\gdef\csname odunum@ci@cascade_redundancy.band.seed.audio.lo\endcsname{[0.407, 0.491]}
\expandafter\gdef\csname odunum@val@cascade_redundancy.band.seed.audio.lo.latin\endcsname{0.333}
\expandafter\gdef\csname odunum@n@cascade_redundancy.band.seed.audio.lo.latin\endcsname{141}
\expandafter\gdef\csname odunum@ci@cascade_redundancy.band.seed.audio.lo.latin\endcsname{[0.255, 0.414]}
\expandafter\gdef\csname odunum@val@cascade_redundancy.band.seed.audio.mid1\endcsname{0.473}
\expandafter\gdef\csname odunum@n@cascade_redundancy.band.seed.audio.mid1\endcsname{167}
\expandafter\gdef\csname odunum@ci@cascade_redundancy.band.seed.audio.mid1\endcsname{[0.399, 0.548]}
\expandafter\gdef\csname odunum@val@cascade_redundancy.band.seed.audio.mid1.latin\endcsname{0.496}
\expandafter\gdef\csname odunum@n@cascade_redundancy.band.seed.audio.mid1.latin\endcsname{135}
\expandafter\gdef\csname odunum@ci@cascade_redundancy.band.seed.audio.mid1.latin\endcsname{[0.413, 0.578]}
\expandafter\gdef\csname odunum@val@cascade_redundancy.band.seed.audio.mid2\endcsname{0.538}
\expandafter\gdef\csname odunum@n@cascade_redundancy.band.seed.audio.mid2\endcsname{91}
\expandafter\gdef\csname odunum@ci@cascade_redundancy.band.seed.audio.mid2\endcsname{[0.433, 0.644]}
\expandafter\gdef\csname odunum@val@cascade_redundancy.band.seed.audio.mid2.latin\endcsname{0.529}
\expandafter\gdef\csname odunum@n@cascade_redundancy.band.seed.audio.mid2.latin\endcsname{85}
\expandafter\gdef\csname odunum@ci@cascade_redundancy.band.seed.audio.mid2.latin\endcsname{[0.423, 0.635]}
\expandafter\gdef\csname odunum@val@cascade_redundancy.band.seed.visual.hi\endcsname{0.688}
\expandafter\gdef\csname odunum@n@cascade_redundancy.band.seed.visual.hi\endcsname{16}
\expandafter\gdef\csname odunum@ci@cascade_redundancy.band.seed.visual.hi\endcsname{[0.438, 0.875]}
\expandafter\gdef\csname odunum@val@cascade_redundancy.band.seed.visual.hi.latin\endcsname{0.688}
\expandafter\gdef\csname odunum@n@cascade_redundancy.band.seed.visual.hi.latin\endcsname{16}
\expandafter\gdef\csname odunum@ci@cascade_redundancy.band.seed.visual.hi.latin\endcsname{[0.438, 0.875]}
\expandafter\gdef\csname odunum@val@cascade_redundancy.band.seed.visual.lo\endcsname{0.544}
\expandafter\gdef\csname odunum@n@cascade_redundancy.band.seed.visual.lo\endcsname{627}
\expandafter\gdef\csname odunum@ci@cascade_redundancy.band.seed.visual.lo\endcsname{[0.504, 0.582]}
\expandafter\gdef\csname odunum@val@cascade_redundancy.band.seed.visual.lo.latin\endcsname{0.529}
\expandafter\gdef\csname odunum@n@cascade_redundancy.band.seed.visual.lo.latin\endcsname{187}
\expandafter\gdef\csname odunum@ci@cascade_redundancy.band.seed.visual.lo.latin\endcsname{[0.456, 0.601]}
\expandafter\gdef\csname odunum@val@cascade_redundancy.band.seed.visual.mid1\endcsname{0.500}
\expandafter\gdef\csname odunum@n@cascade_redundancy.band.seed.visual.mid1\endcsname{110}
\expandafter\gdef\csname odunum@ci@cascade_redundancy.band.seed.visual.mid1\endcsname{[0.400, 0.600]}
\expandafter\gdef\csname odunum@val@cascade_redundancy.band.seed.visual.mid1.latin\endcsname{0.500}
\expandafter\gdef\csname odunum@n@cascade_redundancy.band.seed.visual.mid1.latin\endcsname{104}
\expandafter\gdef\csname odunum@ci@cascade_redundancy.band.seed.visual.mid1.latin\endcsname{[0.398, 0.602]}
\expandafter\gdef\csname odunum@val@cascade_redundancy.band.seed.visual.mid2\endcsname{0.579}
\expandafter\gdef\csname odunum@n@cascade_redundancy.band.seed.visual.mid2\endcsname{38}
\expandafter\gdef\csname odunum@ci@cascade_redundancy.band.seed.visual.mid2\endcsname{[0.410, 0.737]}
\expandafter\gdef\csname odunum@val@cascade_redundancy.band.seed.visual.mid2.latin\endcsname{0.579}
\expandafter\gdef\csname odunum@n@cascade_redundancy.band.seed.visual.mid2.latin\endcsname{38}
\expandafter\gdef\csname odunum@ci@cascade_redundancy.band.seed.visual.mid2.latin\endcsname{[0.410, 0.737]}
\expandafter\gdef\csname odunum@val@cascade_redundancy.null.cascade_asr.multispeaker_delta\endcsname{0.007}
\expandafter\gdef\csname odunum@n@cascade_redundancy.null.cascade_asr.multispeaker_delta\endcsname{882}
\expandafter\gdef\csname odunum@val@cascade_redundancy.null.cascade_asr.speech_content_delta\endcsname{-0.008}
\expandafter\gdef\csname odunum@n@cascade_redundancy.null.cascade_asr.speech_content_delta\endcsname{408}
\expandafter\gdef\csname odunum@val@cascade_redundancy.overlap.audio\endcsname{0.167}
\expandafter\gdef\csname odunum@n@cascade_redundancy.overlap.audio\endcsname{882}
\expandafter\gdef\csname odunum@val@cascade_redundancy.overlap.audio.share_half\endcsname{0.115}
\expandafter\gdef\csname odunum@n@cascade_redundancy.overlap.audio.share_half\endcsname{882}
\expandafter\gdef\csname odunum@val@cascade_redundancy.overlap.history\endcsname{0.260}
\expandafter\gdef\csname odunum@n@cascade_redundancy.overlap.history\endcsname{809}
\expandafter\gdef\csname odunum@val@cascade_redundancy.overlap.history.share_half\endcsname{0.283}
\expandafter\gdef\csname odunum@n@cascade_redundancy.overlap.history.share_half\endcsname{809}
\expandafter\gdef\csname odunum@val@cascade_redundancy.overlap.intent\endcsname{0.257}
\expandafter\gdef\csname odunum@n@cascade_redundancy.overlap.intent\endcsname{3\,083}
\expandafter\gdef\csname odunum@val@cascade_redundancy.overlap.intent.share_half\endcsname{0.277}
\expandafter\gdef\csname odunum@n@cascade_redundancy.overlap.intent.share_half\endcsname{3\,083}
\expandafter\gdef\csname odunum@val@cascade_redundancy.overlap.visual\endcsname{0.085}
\expandafter\gdef\csname odunum@n@cascade_redundancy.overlap.visual\endcsname{791}
\expandafter\gdef\csname odunum@val@cascade_redundancy.overlap.visual.share_half\endcsname{0.044}
\expandafter\gdef\csname odunum@n@cascade_redundancy.overlap.visual.share_half\endcsname{791}
\expandafter\gdef\csname odunum@val@cascade_redundancy.population.n_latin_scenes\endcsname{720}
\expandafter\gdef\csname odunum@n@cascade_redundancy.population.n_latin_scenes\endcsname{1\,612}
\expandafter\gdef\csname odunum@val@cascade_redundancy.population.n_points\endcsname{5\,754}
\expandafter\gdef\csname odunum@n@cascade_redundancy.population.n_points\endcsname{5\,859}
\expandafter\gdef\csname odunum@val@cascade_redundancy.population.n_scenes\endcsname{1\,612}
\expandafter\gdef\csname odunum@n@cascade_redundancy.population.n_scenes\endcsname{1\,631}
\expandafter\gdef\csname odunum@val@cascade_redundancy.prior.audio.cascade_only\endcsname{0.087}
\expandafter\gdef\csname odunum@n@cascade_redundancy.prior.audio.cascade_only\endcsname{298}
\expandafter\gdef\csname odunum@val@cascade_redundancy.prior.audio.native0\endcsname{0.089}
\expandafter\gdef\csname odunum@n@cascade_redundancy.prior.audio.native0\endcsname{291}
\expandafter\gdef\csname odunum@ci@cascade_redundancy.prior.audio.native0\endcsname{[0.058, 0.123]}
\expandafter\gdef\csname odunum@val@cascade_redundancy.prior.audio.native1\endcsname{0.194}
\expandafter\gdef\csname odunum@n@cascade_redundancy.prior.audio.native1\endcsname{165}
\expandafter\gdef\csname odunum@ci@cascade_redundancy.prior.audio.native1\endcsname{[0.134, 0.255]}
\expandafter\gdef\csname odunum@val@cascade_redundancy.prior.audio.native2\endcsname{0.372}
\expandafter\gdef\csname odunum@n@cascade_redundancy.prior.audio.native2\endcsname{191}
\expandafter\gdef\csname odunum@ci@cascade_redundancy.prior.audio.native2\endcsname{[0.304, 0.443]}
\expandafter\gdef\csname odunum@val@cascade_redundancy.prior.audio.native3\endcsname{0.719}
\expandafter\gdef\csname odunum@n@cascade_redundancy.prior.audio.native3\endcsname{235}
\expandafter\gdef\csname odunum@ci@cascade_redundancy.prior.audio.native3\endcsname{[0.658, 0.776]}
\expandafter\gdef\csname odunum@val@cascade_redundancy.prior.visual.cascade_only\endcsname{0.074}
\expandafter\gdef\csname odunum@n@cascade_redundancy.prior.visual.cascade_only\endcsname{162}
\expandafter\gdef\csname odunum@val@cascade_redundancy.prior.visual.native0\endcsname{0.051}
\expandafter\gdef\csname odunum@n@cascade_redundancy.prior.visual.native0\endcsname{237}
\expandafter\gdef\csname odunum@ci@cascade_redundancy.prior.visual.native0\endcsname{[0.022, 0.082]}
\expandafter\gdef\csname odunum@val@cascade_redundancy.prior.visual.native1\endcsname{0.106}
\expandafter\gdef\csname odunum@n@cascade_redundancy.prior.visual.native1\endcsname{160}
\expandafter\gdef\csname odunum@ci@cascade_redundancy.prior.visual.native1\endcsname{[0.057, 0.159]}
\expandafter\gdef\csname odunum@val@cascade_redundancy.prior.visual.native2\endcsname{0.266}
\expandafter\gdef\csname odunum@n@cascade_redundancy.prior.visual.native2\endcsname{158}
\expandafter\gdef\csname odunum@ci@cascade_redundancy.prior.visual.native2\endcsname{[0.200, 0.335]}
\expandafter\gdef\csname odunum@val@cascade_redundancy.prior.visual.native3\endcsname{0.386}
\expandafter\gdef\csname odunum@n@cascade_redundancy.prior.visual.native3\endcsname{236}
\expandafter\gdef\csname odunum@ci@cascade_redundancy.prior.visual.native3\endcsname{[0.324, 0.448]}
\expandafter\gdef\csname odunum@val@cascade_redundancy.ratio.cascade_asr.audio\endcsname{1.49\ensuremath{\times}}
\expandafter\gdef\csname odunum@n@cascade_redundancy.ratio.cascade_asr.audio\endcsname{624}
\expandafter\gdef\csname odunum@val@cascade_redundancy.ratio.cascade_asr.audio.latin\endcsname{2.98\ensuremath{\times}}
\expandafter\gdef\csname odunum@n@cascade_redundancy.ratio.cascade_asr.audio.latin\endcsname{197}
\expandafter\gdef\csname odunum@val@cascade_redundancy.ratio.cascade_asr.visual\endcsname{3.24\ensuremath{\times}}
\expandafter\gdef\csname odunum@n@cascade_redundancy.ratio.cascade_asr.visual\endcsname{643}
\expandafter\gdef\csname odunum@val@cascade_redundancy.ratio.cascade_asr.visual.latin\endcsname{5.26\ensuremath{\times}}
\expandafter\gdef\csname odunum@n@cascade_redundancy.ratio.cascade_asr.visual.latin\endcsname{203}
\expandafter\gdef\csname odunum@val@cascade_redundancy.ratio.gemini.audio\endcsname{1.20\ensuremath{\times}}
\expandafter\gdef\csname odunum@n@cascade_redundancy.ratio.gemini.audio\endcsname{624}
\expandafter\gdef\csname odunum@val@cascade_redundancy.ratio.gemini.audio.latin\endcsname{1.28\ensuremath{\times}}
\expandafter\gdef\csname odunum@n@cascade_redundancy.ratio.gemini.audio.latin\endcsname{197}
\expandafter\gdef\csname odunum@val@cascade_redundancy.ratio.gemini.visual\endcsname{1.52\ensuremath{\times}}
\expandafter\gdef\csname odunum@n@cascade_redundancy.ratio.gemini.visual\endcsname{643}
\expandafter\gdef\csname odunum@val@cascade_redundancy.ratio.gemini.visual.latin\endcsname{1.86\ensuremath{\times}}
\expandafter\gdef\csname odunum@n@cascade_redundancy.ratio.gemini.visual.latin\endcsname{203}
\expandafter\gdef\csname odunum@val@cascade_redundancy.ratio.qwen_plus.audio\endcsname{1.15\ensuremath{\times}}
\expandafter\gdef\csname odunum@n@cascade_redundancy.ratio.qwen_plus.audio\endcsname{624}
\expandafter\gdef\csname odunum@val@cascade_redundancy.ratio.qwen_plus.audio.latin\endcsname{1.11\ensuremath{\times}}
\expandafter\gdef\csname odunum@n@cascade_redundancy.ratio.qwen_plus.audio.latin\endcsname{197}
\expandafter\gdef\csname odunum@val@cascade_redundancy.ratio.qwen_plus.visual\endcsname{1.95\ensuremath{\times}}
\expandafter\gdef\csname odunum@n@cascade_redundancy.ratio.qwen_plus.visual\endcsname{643}
\expandafter\gdef\csname odunum@val@cascade_redundancy.ratio.qwen_plus.visual.latin\endcsname{2.05\ensuremath{\times}}
\expandafter\gdef\csname odunum@n@cascade_redundancy.ratio.qwen_plus.visual.latin\endcsname{203}
\expandafter\gdef\csname odunum@val@cascade_redundancy.ratio.seed.audio\endcsname{1.11\ensuremath{\times}}
\expandafter\gdef\csname odunum@n@cascade_redundancy.ratio.seed.audio\endcsname{624}
\expandafter\gdef\csname odunum@val@cascade_redundancy.ratio.seed.audio.latin\endcsname{1.50\ensuremath{\times}}
\expandafter\gdef\csname odunum@n@cascade_redundancy.ratio.seed.audio.latin\endcsname{197}
\expandafter\gdef\csname odunum@val@cascade_redundancy.ratio.seed.visual\endcsname{1.26\ensuremath{\times}}
\expandafter\gdef\csname odunum@n@cascade_redundancy.ratio.seed.visual\endcsname{643}
\expandafter\gdef\csname odunum@val@cascade_redundancy.ratio.seed.visual.latin\endcsname{1.30\ensuremath{\times}}
\expandafter\gdef\csname odunum@n@cascade_redundancy.ratio.seed.visual.latin\endcsname{203}
\expandafter\gdef\csname odunum@val@cascade_redundancy.reader\endcsname{GPT-5.4 reading FunASR transcripts, no audio or video}
\expandafter\gdef\csname odunum@n@cascade_redundancy.reader\endcsname{1}
\expandafter\gdef\csname odunum@val@cascade_redundancy.redundant.cascade_asr.audio.absent\endcsname{0.312}
\expandafter\gdef\csname odunum@n@cascade_redundancy.redundant.cascade_asr.audio.absent\endcsname{568}
\expandafter\gdef\csname odunum@ci@cascade_redundancy.redundant.cascade_asr.audio.absent\endcsname{[0.273, 0.350]}
\expandafter\gdef\csname odunum@val@cascade_redundancy.redundant.cascade_asr.audio.absent.latin\endcsname{0.156}
\expandafter\gdef\csname odunum@n@cascade_redundancy.redundant.cascade_asr.audio.absent.latin\endcsname{141}
\expandafter\gdef\csname odunum@ci@cascade_redundancy.redundant.cascade_asr.audio.absent.latin\endcsname{[0.099, 0.218]}
\expandafter\gdef\csname odunum@val@cascade_redundancy.redundant.cascade_asr.audio.present\endcsname{0.385}
\expandafter\gdef\csname odunum@n@cascade_redundancy.redundant.cascade_asr.audio.present\endcsname{314}
\expandafter\gdef\csname odunum@ci@cascade_redundancy.redundant.cascade_asr.audio.present\endcsname{[0.333, 0.439]}
\expandafter\gdef\csname odunum@val@cascade_redundancy.redundant.cascade_asr.audio.present.latin\endcsname{0.380}
\expandafter\gdef\csname odunum@n@cascade_redundancy.redundant.cascade_asr.audio.present.latin\endcsname{276}
\expandafter\gdef\csname odunum@ci@cascade_redundancy.redundant.cascade_asr.audio.present.latin\endcsname{[0.324, 0.438]}
\expandafter\gdef\csname odunum@val@cascade_redundancy.redundant.cascade_asr.audio.ratio\endcsname{1.24\ensuremath{\times}}
\expandafter\gdef\csname odunum@n@cascade_redundancy.redundant.cascade_asr.audio.ratio\endcsname{882}
\expandafter\gdef\csname odunum@val@cascade_redundancy.redundant.cascade_asr.audio.ratio.latin\endcsname{2.44\ensuremath{\times}}
\expandafter\gdef\csname odunum@n@cascade_redundancy.redundant.cascade_asr.audio.ratio.latin\endcsname{417}
\expandafter\gdef\csname odunum@val@cascade_redundancy.redundant.cascade_asr.visual.absent\endcsname{0.174}
\expandafter\gdef\csname odunum@n@cascade_redundancy.redundant.cascade_asr.visual.absent\endcsname{627}
\expandafter\gdef\csname odunum@ci@cascade_redundancy.redundant.cascade_asr.visual.absent\endcsname{[0.145, 0.204]}
\expandafter\gdef\csname odunum@val@cascade_redundancy.redundant.cascade_asr.visual.absent.latin\endcsname{0.107}
\expandafter\gdef\csname odunum@n@cascade_redundancy.redundant.cascade_asr.visual.absent.latin\endcsname{187}
\expandafter\gdef\csname odunum@ci@cascade_redundancy.redundant.cascade_asr.visual.absent.latin\endcsname{[0.064, 0.153]}
\expandafter\gdef\csname odunum@val@cascade_redundancy.redundant.cascade_asr.visual.present\endcsname{0.323}
\expandafter\gdef\csname odunum@n@cascade_redundancy.redundant.cascade_asr.visual.present\endcsname{164}
\expandafter\gdef\csname odunum@ci@cascade_redundancy.redundant.cascade_asr.visual.present\endcsname{[0.252, 0.395]}
\expandafter\gdef\csname odunum@val@cascade_redundancy.redundant.cascade_asr.visual.present.latin\endcsname{0.323}
\expandafter\gdef\csname odunum@n@cascade_redundancy.redundant.cascade_asr.visual.present.latin\endcsname{158}
\expandafter\gdef\csname odunum@ci@cascade_redundancy.redundant.cascade_asr.visual.present.latin\endcsname{[0.252, 0.396]}
\expandafter\gdef\csname odunum@val@cascade_redundancy.redundant.cascade_asr.visual.ratio\endcsname{1.86\ensuremath{\times}}
\expandafter\gdef\csname odunum@n@cascade_redundancy.redundant.cascade_asr.visual.ratio\endcsname{791}
\expandafter\gdef\csname odunum@val@cascade_redundancy.redundant.cascade_asr.visual.ratio.latin\endcsname{3.02\ensuremath{\times}}
\expandafter\gdef\csname odunum@n@cascade_redundancy.redundant.cascade_asr.visual.ratio.latin\endcsname{345}
\expandafter\gdef\csname odunum@val@cascade_redundancy.redundant.gemini.audio.absent\endcsname{0.461}
\expandafter\gdef\csname odunum@n@cascade_redundancy.redundant.gemini.audio.absent\endcsname{568}
\expandafter\gdef\csname odunum@ci@cascade_redundancy.redundant.gemini.audio.absent\endcsname{[0.420, 0.503]}
\expandafter\gdef\csname odunum@val@cascade_redundancy.redundant.gemini.audio.absent.latin\endcsname{0.433}
\expandafter\gdef\csname odunum@n@cascade_redundancy.redundant.gemini.audio.absent.latin\endcsname{141}
\expandafter\gdef\csname odunum@ci@cascade_redundancy.redundant.gemini.audio.absent.latin\endcsname{[0.352, 0.510]}
\expandafter\gdef\csname odunum@val@cascade_redundancy.redundant.gemini.audio.present\endcsname{0.525}
\expandafter\gdef\csname odunum@n@cascade_redundancy.redundant.gemini.audio.present\endcsname{314}
\expandafter\gdef\csname odunum@ci@cascade_redundancy.redundant.gemini.audio.present\endcsname{[0.469, 0.580]}
\expandafter\gdef\csname odunum@val@cascade_redundancy.redundant.gemini.audio.present.latin\endcsname{0.529}
\expandafter\gdef\csname odunum@n@cascade_redundancy.redundant.gemini.audio.present.latin\endcsname{276}
\expandafter\gdef\csname odunum@ci@cascade_redundancy.redundant.gemini.audio.present.latin\endcsname{[0.470, 0.588]}
\expandafter\gdef\csname odunum@val@cascade_redundancy.redundant.gemini.audio.ratio\endcsname{1.14\ensuremath{\times}}
\expandafter\gdef\csname odunum@n@cascade_redundancy.redundant.gemini.audio.ratio\endcsname{882}
\expandafter\gdef\csname odunum@val@cascade_redundancy.redundant.gemini.audio.ratio.latin\endcsname{1.22\ensuremath{\times}}
\expandafter\gdef\csname odunum@n@cascade_redundancy.redundant.gemini.audio.ratio.latin\endcsname{417}
\expandafter\gdef\csname odunum@val@cascade_redundancy.redundant.gemini.visual.absent\endcsname{0.453}
\expandafter\gdef\csname odunum@n@cascade_redundancy.redundant.gemini.visual.absent\endcsname{627}
\expandafter\gdef\csname odunum@ci@cascade_redundancy.redundant.gemini.visual.absent\endcsname{[0.412, 0.493]}
\expandafter\gdef\csname odunum@val@cascade_redundancy.redundant.gemini.visual.absent.latin\endcsname{0.369}
\expandafter\gdef\csname odunum@n@cascade_redundancy.redundant.gemini.visual.absent.latin\endcsname{187}
\expandafter\gdef\csname odunum@ci@cascade_redundancy.redundant.gemini.visual.absent.latin\endcsname{[0.303, 0.439]}
\expandafter\gdef\csname odunum@val@cascade_redundancy.redundant.gemini.visual.present\endcsname{0.591}
\expandafter\gdef\csname odunum@n@cascade_redundancy.redundant.gemini.visual.present\endcsname{164}
\expandafter\gdef\csname odunum@ci@cascade_redundancy.redundant.gemini.visual.present\endcsname{[0.515, 0.667]}
\expandafter\gdef\csname odunum@val@cascade_redundancy.redundant.gemini.visual.present.latin\endcsname{0.601}
\expandafter\gdef\csname odunum@n@cascade_redundancy.redundant.gemini.visual.present.latin\endcsname{158}
\expandafter\gdef\csname odunum@ci@cascade_redundancy.redundant.gemini.visual.present.latin\endcsname{[0.523, 0.679]}
\expandafter\gdef\csname odunum@val@cascade_redundancy.redundant.gemini.visual.ratio\endcsname{1.31\ensuremath{\times}}
\expandafter\gdef\csname odunum@n@cascade_redundancy.redundant.gemini.visual.ratio\endcsname{791}
\expandafter\gdef\csname odunum@val@cascade_redundancy.redundant.gemini.visual.ratio.latin\endcsname{1.63\ensuremath{\times}}
\expandafter\gdef\csname odunum@n@cascade_redundancy.redundant.gemini.visual.ratio.latin\endcsname{345}
\expandafter\gdef\csname odunum@val@cascade_redundancy.redundant.qwen_plus.audio.absent\endcsname{0.451}
\expandafter\gdef\csname odunum@n@cascade_redundancy.redundant.qwen_plus.audio.absent\endcsname{568}
\expandafter\gdef\csname odunum@ci@cascade_redundancy.redundant.qwen_plus.audio.absent\endcsname{[0.410, 0.492]}
\expandafter\gdef\csname odunum@val@cascade_redundancy.redundant.qwen_plus.audio.absent.latin\endcsname{0.468}
\expandafter\gdef\csname odunum@n@cascade_redundancy.redundant.qwen_plus.audio.absent.latin\endcsname{141}
\expandafter\gdef\csname odunum@ci@cascade_redundancy.redundant.qwen_plus.audio.absent.latin\endcsname{[0.387, 0.549]}
\expandafter\gdef\csname odunum@val@cascade_redundancy.redundant.qwen_plus.audio.present\endcsname{0.503}
\expandafter\gdef\csname odunum@n@cascade_redundancy.redundant.qwen_plus.audio.present\endcsname{314}
\expandafter\gdef\csname odunum@ci@cascade_redundancy.redundant.qwen_plus.audio.present\endcsname{[0.447, 0.556]}
\expandafter\gdef\csname odunum@val@cascade_redundancy.redundant.qwen_plus.audio.present.latin\endcsname{0.522}
\expandafter\gdef\csname odunum@n@cascade_redundancy.redundant.qwen_plus.audio.present.latin\endcsname{276}
\expandafter\gdef\csname odunum@ci@cascade_redundancy.redundant.qwen_plus.audio.present.latin\endcsname{[0.464, 0.580]}
\expandafter\gdef\csname odunum@val@cascade_redundancy.redundant.qwen_plus.audio.ratio\endcsname{1.12\ensuremath{\times}}
\expandafter\gdef\csname odunum@n@cascade_redundancy.redundant.qwen_plus.audio.ratio\endcsname{882}
\expandafter\gdef\csname odunum@val@cascade_redundancy.redundant.qwen_plus.audio.ratio.latin\endcsname{1.11\ensuremath{\times}}
\expandafter\gdef\csname odunum@n@cascade_redundancy.redundant.qwen_plus.audio.ratio.latin\endcsname{417}
\expandafter\gdef\csname odunum@val@cascade_redundancy.redundant.qwen_plus.visual.absent\endcsname{0.450}
\expandafter\gdef\csname odunum@n@cascade_redundancy.redundant.qwen_plus.visual.absent\endcsname{627}
\expandafter\gdef\csname odunum@ci@cascade_redundancy.redundant.qwen_plus.visual.absent\endcsname{[0.411, 0.489]}
\expandafter\gdef\csname odunum@val@cascade_redundancy.redundant.qwen_plus.visual.absent.latin\endcsname{0.428}
\expandafter\gdef\csname odunum@n@cascade_redundancy.redundant.qwen_plus.visual.absent.latin\endcsname{187}
\expandafter\gdef\csname odunum@ci@cascade_redundancy.redundant.qwen_plus.visual.absent.latin\endcsname{[0.358, 0.500]}
\expandafter\gdef\csname odunum@val@cascade_redundancy.redundant.qwen_plus.visual.present\endcsname{0.561}
\expandafter\gdef\csname odunum@n@cascade_redundancy.redundant.qwen_plus.visual.present\endcsname{164}
\expandafter\gdef\csname odunum@ci@cascade_redundancy.redundant.qwen_plus.visual.present\endcsname{[0.482, 0.637]}
\expandafter\gdef\csname odunum@val@cascade_redundancy.redundant.qwen_plus.visual.present.latin\endcsname{0.563}
\expandafter\gdef\csname odunum@n@cascade_redundancy.redundant.qwen_plus.visual.present.latin\endcsname{158}
\expandafter\gdef\csname odunum@ci@cascade_redundancy.redundant.qwen_plus.visual.present.latin\endcsname{[0.484, 0.640]}
\expandafter\gdef\csname odunum@val@cascade_redundancy.redundant.qwen_plus.visual.ratio\endcsname{1.25\ensuremath{\times}}
\expandafter\gdef\csname odunum@n@cascade_redundancy.redundant.qwen_plus.visual.ratio\endcsname{791}
\expandafter\gdef\csname odunum@val@cascade_redundancy.redundant.qwen_plus.visual.ratio.latin\endcsname{1.32\ensuremath{\times}}
\expandafter\gdef\csname odunum@n@cascade_redundancy.redundant.qwen_plus.visual.ratio.latin\endcsname{345}
\expandafter\gdef\csname odunum@val@cascade_redundancy.redundant.seed.audio.absent\endcsname{0.449}
\expandafter\gdef\csname odunum@n@cascade_redundancy.redundant.seed.audio.absent\endcsname{568}
\expandafter\gdef\csname odunum@ci@cascade_redundancy.redundant.seed.audio.absent\endcsname{[0.407, 0.491]}
\expandafter\gdef\csname odunum@val@cascade_redundancy.redundant.seed.audio.absent.latin\endcsname{0.333}
\expandafter\gdef\csname odunum@n@cascade_redundancy.redundant.seed.audio.absent.latin\endcsname{141}
\expandafter\gdef\csname odunum@ci@cascade_redundancy.redundant.seed.audio.absent.latin\endcsname{[0.255, 0.414]}
\expandafter\gdef\csname odunum@val@cascade_redundancy.redundant.seed.audio.present\endcsname{0.497}
\expandafter\gdef\csname odunum@n@cascade_redundancy.redundant.seed.audio.present\endcsname{314}
\expandafter\gdef\csname odunum@ci@cascade_redundancy.redundant.seed.audio.present\endcsname{[0.441, 0.551]}
\expandafter\gdef\csname odunum@val@cascade_redundancy.redundant.seed.audio.present.latin\endcsname{0.507}
\expandafter\gdef\csname odunum@n@cascade_redundancy.redundant.seed.audio.present.latin\endcsname{276}
\expandafter\gdef\csname odunum@ci@cascade_redundancy.redundant.seed.audio.present.latin\endcsname{[0.449, 0.566]}
\expandafter\gdef\csname odunum@val@cascade_redundancy.redundant.seed.audio.ratio\endcsname{1.11\ensuremath{\times}}
\expandafter\gdef\csname odunum@n@cascade_redundancy.redundant.seed.audio.ratio\endcsname{882}
\expandafter\gdef\csname odunum@val@cascade_redundancy.redundant.seed.audio.ratio.latin\endcsname{1.52\ensuremath{\times}}
\expandafter\gdef\csname odunum@n@cascade_redundancy.redundant.seed.audio.ratio.latin\endcsname{417}
\expandafter\gdef\csname odunum@val@cascade_redundancy.redundant.seed.visual.absent\endcsname{0.544}
\expandafter\gdef\csname odunum@n@cascade_redundancy.redundant.seed.visual.absent\endcsname{627}
\expandafter\gdef\csname odunum@ci@cascade_redundancy.redundant.seed.visual.absent\endcsname{[0.504, 0.582]}
\expandafter\gdef\csname odunum@val@cascade_redundancy.redundant.seed.visual.absent.latin\endcsname{0.529}
\expandafter\gdef\csname odunum@n@cascade_redundancy.redundant.seed.visual.absent.latin\endcsname{187}
\expandafter\gdef\csname odunum@ci@cascade_redundancy.redundant.seed.visual.absent.latin\endcsname{[0.456, 0.601]}
\expandafter\gdef\csname odunum@val@cascade_redundancy.redundant.seed.visual.present\endcsname{0.537}
\expandafter\gdef\csname odunum@n@cascade_redundancy.redundant.seed.visual.present\endcsname{164}
\expandafter\gdef\csname odunum@ci@cascade_redundancy.redundant.seed.visual.present\endcsname{[0.455, 0.616]}
\expandafter\gdef\csname odunum@val@cascade_redundancy.redundant.seed.visual.present.latin\endcsname{0.538}
\expandafter\gdef\csname odunum@n@cascade_redundancy.redundant.seed.visual.present.latin\endcsname{158}
\expandafter\gdef\csname odunum@ci@cascade_redundancy.redundant.seed.visual.present.latin\endcsname{[0.453, 0.620]}
\expandafter\gdef\csname odunum@val@cascade_redundancy.redundant.seed.visual.ratio\endcsname{0.99\ensuremath{\times}}
\expandafter\gdef\csname odunum@n@cascade_redundancy.redundant.seed.visual.ratio\endcsname{791}
\expandafter\gdef\csname odunum@val@cascade_redundancy.redundant.seed.visual.ratio.latin\endcsname{1.02\ensuremath{\times}}
\expandafter\gdef\csname odunum@n@cascade_redundancy.redundant.seed.visual.ratio.latin\endcsname{345}
\expandafter\gdef\csname odunum@val@cascade_redundancy.redundant_share.audio\endcsname{0.356}
\expandafter\gdef\csname odunum@n@cascade_redundancy.redundant_share.audio\endcsname{882}
\expandafter\gdef\csname odunum@val@cascade_redundancy.redundant_share.visual\endcsname{0.207}
\expandafter\gdef\csname odunum@n@cascade_redundancy.redundant_share.visual\endcsname{791}
\expandafter\gdef\csname odunum@val@cascade_redundancy.shape.channels_where_cascade_is_most_redundancy_dependent\endcsname{2}
\expandafter\gdef\csname odunum@n@cascade_redundancy.shape.channels_where_cascade_is_most_redundancy_dependent\endcsname{2}
\expandafter\gdef\csname odunum@val@channel_gaps.analysis.bootstrap_replicates\endcsname{10\,000}
\expandafter\gdef\csname odunum@n@channel_gaps.analysis.bootstrap_replicates\endcsname{1\,060}
\expandafter\gdef\csname odunum@val@channel_gaps.coverage.cascade_asr.av.audio\endcsname{0.336}
\expandafter\gdef\csname odunum@n@channel_gaps.coverage.cascade_asr.av.audio\endcsname{557}
\expandafter\gdef\csname odunum@val@channel_gaps.coverage.cascade_asr.av.context_all\endcsname{0.330}
\expandafter\gdef\csname odunum@n@channel_gaps.coverage.cascade_asr.av.context_all\endcsname{1\,972}
\expandafter\gdef\csname odunum@val@channel_gaps.coverage.cascade_asr.av.history\endcsname{0.587}
\expandafter\gdef\csname odunum@n@channel_gaps.coverage.cascade_asr.av.history\endcsname{475}
\expandafter\gdef\csname odunum@val@channel_gaps.coverage.cascade_asr.av.intent\endcsname{0.789}
\expandafter\gdef\csname odunum@n@channel_gaps.coverage.cascade_asr.av.intent\endcsname{1\,984}
\expandafter\gdef\csname odunum@val@channel_gaps.coverage.cascade_asr.av.visual\endcsname{0.203}
\expandafter\gdef\csname odunum@n@channel_gaps.coverage.cascade_asr.av.visual\endcsname{804}
\expandafter\gdef\csname odunum@val@channel_gaps.coverage.gemini.av.audio\endcsname{0.472}
\expandafter\gdef\csname odunum@n@channel_gaps.coverage.gemini.av.audio\endcsname{557}
\expandafter\gdef\csname odunum@val@channel_gaps.coverage.gemini.av.context_all\endcsname{0.447}
\expandafter\gdef\csname odunum@n@channel_gaps.coverage.gemini.av.context_all\endcsname{1\,972}
\expandafter\gdef\csname odunum@val@channel_gaps.coverage.gemini.av.history\endcsname{0.421}
\expandafter\gdef\csname odunum@n@channel_gaps.coverage.gemini.av.history\endcsname{475}
\expandafter\gdef\csname odunum@val@channel_gaps.coverage.gemini.av.intent\endcsname{0.829}
\expandafter\gdef\csname odunum@n@channel_gaps.coverage.gemini.av.intent\endcsname{1\,984}
\expandafter\gdef\csname odunum@val@channel_gaps.coverage.gemini.av.visual\endcsname{0.486}
\expandafter\gdef\csname odunum@n@channel_gaps.coverage.gemini.av.visual\endcsname{804}
\expandafter\gdef\csname odunum@val@channel_gaps.coverage.qwen_plus.av.audio\endcsname{0.456}
\expandafter\gdef\csname odunum@n@channel_gaps.coverage.qwen_plus.av.audio\endcsname{557}
\expandafter\gdef\csname odunum@val@channel_gaps.coverage.qwen_plus.av.context_all\endcsname{0.489}
\expandafter\gdef\csname odunum@n@channel_gaps.coverage.qwen_plus.av.context_all\endcsname{1\,972}
\expandafter\gdef\csname odunum@val@channel_gaps.coverage.qwen_plus.av.history\endcsname{0.577}
\expandafter\gdef\csname odunum@n@channel_gaps.coverage.qwen_plus.av.history\endcsname{475}
\expandafter\gdef\csname odunum@val@channel_gaps.coverage.qwen_plus.av.intent\endcsname{0.836}
\expandafter\gdef\csname odunum@n@channel_gaps.coverage.qwen_plus.av.intent\endcsname{1\,984}
\expandafter\gdef\csname odunum@val@channel_gaps.coverage.qwen_plus.av.visual\endcsname{0.473}
\expandafter\gdef\csname odunum@n@channel_gaps.coverage.qwen_plus.av.visual\endcsname{804}
\expandafter\gdef\csname odunum@val@channel_gaps.coverage.seed.av.audio\endcsname{0.465}
\expandafter\gdef\csname odunum@n@channel_gaps.coverage.seed.av.audio\endcsname{557}
\expandafter\gdef\csname odunum@val@channel_gaps.coverage.seed.av.context_all\endcsname{0.528}
\expandafter\gdef\csname odunum@n@channel_gaps.coverage.seed.av.context_all\endcsname{1\,972}
\expandafter\gdef\csname odunum@val@channel_gaps.coverage.seed.av.history\endcsname{0.623}
\expandafter\gdef\csname odunum@n@channel_gaps.coverage.seed.av.history\endcsname{475}
\expandafter\gdef\csname odunum@val@channel_gaps.coverage.seed.av.intent\endcsname{0.760}
\expandafter\gdef\csname odunum@n@channel_gaps.coverage.seed.av.intent\endcsname{1\,984}
\expandafter\gdef\csname odunum@val@channel_gaps.coverage.seed.av.visual\endcsname{0.547}
\expandafter\gdef\csname odunum@n@channel_gaps.coverage.seed.av.visual\endcsname{804}
\expandafter\gdef\csname odunum@val@channel_gaps.gap.cascade_asr.av.intent_minus_audio\endcsname{0.453}
\expandafter\gdef\csname odunum@n@channel_gaps.gap.cascade_asr.av.intent_minus_audio\endcsname{557}
\expandafter\gdef\csname odunum@ci@channel_gaps.gap.cascade_asr.av.intent_minus_audio\endcsname{[0.413, 0.494]}
\expandafter\gdef\csname odunum@val@channel_gaps.gap.cascade_asr.av.intent_minus_context_all\endcsname{0.459}
\expandafter\gdef\csname odunum@n@channel_gaps.gap.cascade_asr.av.intent_minus_context_all\endcsname{1\,972}
\expandafter\gdef\csname odunum@ci@channel_gaps.gap.cascade_asr.av.intent_minus_context_all\endcsname{[0.432, 0.486]}
\expandafter\gdef\csname odunum@val@channel_gaps.gap.cascade_asr.av.intent_minus_history\endcsname{0.201}
\expandafter\gdef\csname odunum@n@channel_gaps.gap.cascade_asr.av.intent_minus_history\endcsname{475}
\expandafter\gdef\csname odunum@ci@channel_gaps.gap.cascade_asr.av.intent_minus_history\endcsname{[0.154, 0.249]}
\expandafter\gdef\csname odunum@val@channel_gaps.gap.cascade_asr.av.intent_minus_visual\endcsname{0.586}
\expandafter\gdef\csname odunum@n@channel_gaps.gap.cascade_asr.av.intent_minus_visual\endcsname{804}
\expandafter\gdef\csname odunum@ci@channel_gaps.gap.cascade_asr.av.intent_minus_visual\endcsname{[0.552, 0.619]}
\expandafter\gdef\csname odunum@val@channel_gaps.gap.gemini.av.intent_minus_audio\endcsname{0.356}
\expandafter\gdef\csname odunum@n@channel_gaps.gap.gemini.av.intent_minus_audio\endcsname{557}
\expandafter\gdef\csname odunum@ci@channel_gaps.gap.gemini.av.intent_minus_audio\endcsname{[0.312, 0.402]}
\expandafter\gdef\csname odunum@val@channel_gaps.gap.gemini.av.intent_minus_context_all\endcsname{0.382}
\expandafter\gdef\csname odunum@n@channel_gaps.gap.gemini.av.intent_minus_context_all\endcsname{1\,972}
\expandafter\gdef\csname odunum@ci@channel_gaps.gap.gemini.av.intent_minus_context_all\endcsname{[0.354, 0.411]}
\expandafter\gdef\csname odunum@val@channel_gaps.gap.gemini.av.intent_minus_history\endcsname{0.408}
\expandafter\gdef\csname odunum@n@channel_gaps.gap.gemini.av.intent_minus_history\endcsname{475}
\expandafter\gdef\csname odunum@ci@channel_gaps.gap.gemini.av.intent_minus_history\endcsname{[0.359, 0.457]}
\expandafter\gdef\csname odunum@val@channel_gaps.gap.gemini.av.intent_minus_visual\endcsname{0.342}
\expandafter\gdef\csname odunum@n@channel_gaps.gap.gemini.av.intent_minus_visual\endcsname{804}
\expandafter\gdef\csname odunum@ci@channel_gaps.gap.gemini.av.intent_minus_visual\endcsname{[0.303, 0.383]}
\expandafter\gdef\csname odunum@val@channel_gaps.gap.qwen_plus.av.intent_minus_audio\endcsname{0.380}
\expandafter\gdef\csname odunum@n@channel_gaps.gap.qwen_plus.av.intent_minus_audio\endcsname{557}
\expandafter\gdef\csname odunum@ci@channel_gaps.gap.qwen_plus.av.intent_minus_audio\endcsname{[0.336, 0.423]}
\expandafter\gdef\csname odunum@val@channel_gaps.gap.qwen_plus.av.intent_minus_context_all\endcsname{0.346}
\expandafter\gdef\csname odunum@n@channel_gaps.gap.qwen_plus.av.intent_minus_context_all\endcsname{1\,972}
\expandafter\gdef\csname odunum@ci@channel_gaps.gap.qwen_plus.av.intent_minus_context_all\endcsname{[0.320, 0.373]}
\expandafter\gdef\csname odunum@val@channel_gaps.gap.qwen_plus.av.intent_minus_history\endcsname{0.259}
\expandafter\gdef\csname odunum@n@channel_gaps.gap.qwen_plus.av.intent_minus_history\endcsname{475}
\expandafter\gdef\csname odunum@ci@channel_gaps.gap.qwen_plus.av.intent_minus_history\endcsname{[0.211, 0.306]}
\expandafter\gdef\csname odunum@val@channel_gaps.gap.qwen_plus.av.intent_minus_visual\endcsname{0.363}
\expandafter\gdef\csname odunum@n@channel_gaps.gap.qwen_plus.av.intent_minus_visual\endcsname{804}
\expandafter\gdef\csname odunum@ci@channel_gaps.gap.qwen_plus.av.intent_minus_visual\endcsname{[0.326, 0.401]}
\expandafter\gdef\csname odunum@val@channel_gaps.gap.seed.av.intent_minus_audio\endcsname{0.295}
\expandafter\gdef\csname odunum@n@channel_gaps.gap.seed.av.intent_minus_audio\endcsname{557}
\expandafter\gdef\csname odunum@ci@channel_gaps.gap.seed.av.intent_minus_audio\endcsname{[0.252, 0.340]}
\expandafter\gdef\csname odunum@val@channel_gaps.gap.seed.av.intent_minus_context_all\endcsname{0.232}
\expandafter\gdef\csname odunum@n@channel_gaps.gap.seed.av.intent_minus_context_all\endcsname{1\,972}
\expandafter\gdef\csname odunum@ci@channel_gaps.gap.seed.av.intent_minus_context_all\endcsname{[0.204, 0.260]}
\expandafter\gdef\csname odunum@val@channel_gaps.gap.seed.av.intent_minus_history\endcsname{0.137}
\expandafter\gdef\csname odunum@n@channel_gaps.gap.seed.av.intent_minus_history\endcsname{475}
\expandafter\gdef\csname odunum@ci@channel_gaps.gap.seed.av.intent_minus_history\endcsname{[0.088, 0.185]}
\expandafter\gdef\csname odunum@val@channel_gaps.gap.seed.av.intent_minus_visual\endcsname{0.213}
\expandafter\gdef\csname odunum@n@channel_gaps.gap.seed.av.intent_minus_visual\endcsname{804}
\expandafter\gdef\csname odunum@ci@channel_gaps.gap.seed.av.intent_minus_visual\endcsname{[0.174, 0.251]}
\expandafter\gdef\csname odunum@val@channel_gaps.points.av.audio\endcsname{557}
\expandafter\gdef\csname odunum@n@channel_gaps.points.av.audio\endcsname{557}
\expandafter\gdef\csname odunum@val@channel_gaps.points.av.context_all\endcsname{1\,972}
\expandafter\gdef\csname odunum@n@channel_gaps.points.av.context_all\endcsname{1\,972}
\expandafter\gdef\csname odunum@val@channel_gaps.points.av.history\endcsname{475}
\expandafter\gdef\csname odunum@n@channel_gaps.points.av.history\endcsname{475}
\expandafter\gdef\csname odunum@val@channel_gaps.points.av.intent\endcsname{1\,984}
\expandafter\gdef\csname odunum@n@channel_gaps.points.av.intent\endcsname{1\,984}
\expandafter\gdef\csname odunum@val@channel_gaps.points.av.visual\endcsname{804}
\expandafter\gdef\csname odunum@n@channel_gaps.points.av.visual\endcsname{804}
\expandafter\gdef\csname odunum@val@channel_gaps.pop.av.n_scenes\endcsname{1\,060}
\expandafter\gdef\csname odunum@n@channel_gaps.pop.av.n_scenes\endcsname{1\,060}
\expandafter\gdef\csname odunum@val@controlled_evidence_removal.audit.visual.audio_hash_mismatches\endcsname{0}
\expandafter\gdef\csname odunum@n@controlled_evidence_removal.audit.visual.audio_hash_mismatches\endcsname{40}
\expandafter\gdef\csname odunum@val@controlled_evidence_removal.audit.visual.audio_hash_sample_n\endcsname{40}
\expandafter\gdef\csname odunum@n@controlled_evidence_removal.audit.visual.audio_hash_sample_n\endcsname{712}
\expandafter\gdef\csname odunum@val@controlled_evidence_removal.history_removed.doubao_seed_2_0_lite.ablated.audio\endcsname{0.534}
\expandafter\gdef\csname odunum@n@controlled_evidence_removal.history_removed.doubao_seed_2_0_lite.ablated.audio\endcsname{118}
\expandafter\gdef\csname odunum@val@controlled_evidence_removal.history_removed.doubao_seed_2_0_lite.ablated.history\endcsname{0.247}
\expandafter\gdef\csname odunum@n@controlled_evidence_removal.history_removed.doubao_seed_2_0_lite.ablated.history\endcsname{465}
\expandafter\gdef\csname odunum@val@controlled_evidence_removal.history_removed.doubao_seed_2_0_lite.ablated.intent\endcsname{0.823}
\expandafter\gdef\csname odunum@n@controlled_evidence_removal.history_removed.doubao_seed_2_0_lite.ablated.intent\endcsname{718}
\expandafter\gdef\csname odunum@val@controlled_evidence_removal.history_removed.doubao_seed_2_0_lite.ablated.other_context\endcsname{0.280}
\expandafter\gdef\csname odunum@n@controlled_evidence_removal.history_removed.doubao_seed_2_0_lite.ablated.other_context\endcsname{25}
\expandafter\gdef\csname odunum@val@controlled_evidence_removal.history_removed.doubao_seed_2_0_lite.ablated.visual\endcsname{0.576}
\expandafter\gdef\csname odunum@n@controlled_evidence_removal.history_removed.doubao_seed_2_0_lite.ablated.visual\endcsname{210}
\expandafter\gdef\csname odunum@val@controlled_evidence_removal.history_removed.doubao_seed_2_0_lite.drop.audio\endcsname{-0.025}
\expandafter\gdef\csname odunum@n@controlled_evidence_removal.history_removed.doubao_seed_2_0_lite.drop.audio\endcsname{118}
\expandafter\gdef\csname odunum@val@controlled_evidence_removal.history_removed.doubao_seed_2_0_lite.drop.history\endcsname{0.417}
\expandafter\gdef\csname odunum@n@controlled_evidence_removal.history_removed.doubao_seed_2_0_lite.drop.history\endcsname{465}
\expandafter\gdef\csname odunum@val@controlled_evidence_removal.history_removed.doubao_seed_2_0_lite.drop.intent\endcsname{0.072}
\expandafter\gdef\csname odunum@n@controlled_evidence_removal.history_removed.doubao_seed_2_0_lite.drop.intent\endcsname{718}
\expandafter\gdef\csname odunum@val@controlled_evidence_removal.history_removed.doubao_seed_2_0_lite.drop.other_context\endcsname{0.120}
\expandafter\gdef\csname odunum@n@controlled_evidence_removal.history_removed.doubao_seed_2_0_lite.drop.other_context\endcsname{25}
\expandafter\gdef\csname odunum@val@controlled_evidence_removal.history_removed.doubao_seed_2_0_lite.drop.visual\endcsname{0.005}
\expandafter\gdef\csname odunum@n@controlled_evidence_removal.history_removed.doubao_seed_2_0_lite.drop.visual\endcsname{210}
\expandafter\gdef\csname odunum@val@controlled_evidence_removal.history_removed.doubao_seed_2_0_lite.flips_false_to_true\endcsname{0}
\expandafter\gdef\csname odunum@n@controlled_evidence_removal.history_removed.doubao_seed_2_0_lite.flips_false_to_true\endcsname{391}
\expandafter\gdef\csname odunum@val@controlled_evidence_removal.history_removed.doubao_seed_2_0_lite.flips_true_to_false\endcsname{0}
\expandafter\gdef\csname odunum@n@controlled_evidence_removal.history_removed.doubao_seed_2_0_lite.flips_true_to_false\endcsname{391}
\expandafter\gdef\csname odunum@val@controlled_evidence_removal.history_removed.doubao_seed_2_0_lite.full.audio\endcsname{0.508}
\expandafter\gdef\csname odunum@n@controlled_evidence_removal.history_removed.doubao_seed_2_0_lite.full.audio\endcsname{118}
\expandafter\gdef\csname odunum@val@controlled_evidence_removal.history_removed.doubao_seed_2_0_lite.full.history\endcsname{0.665}
\expandafter\gdef\csname odunum@n@controlled_evidence_removal.history_removed.doubao_seed_2_0_lite.full.history\endcsname{465}
\expandafter\gdef\csname odunum@val@controlled_evidence_removal.history_removed.doubao_seed_2_0_lite.full.intent\endcsname{0.896}
\expandafter\gdef\csname odunum@n@controlled_evidence_removal.history_removed.doubao_seed_2_0_lite.full.intent\endcsname{718}
\expandafter\gdef\csname odunum@val@controlled_evidence_removal.history_removed.doubao_seed_2_0_lite.full.other_context\endcsname{0.400}
\expandafter\gdef\csname odunum@n@controlled_evidence_removal.history_removed.doubao_seed_2_0_lite.full.other_context\endcsname{25}
\expandafter\gdef\csname odunum@val@controlled_evidence_removal.history_removed.doubao_seed_2_0_lite.full.visual\endcsname{0.581}
\expandafter\gdef\csname odunum@n@controlled_evidence_removal.history_removed.doubao_seed_2_0_lite.full.visual\endcsname{210}
\expandafter\gdef\csname odunum@val@controlled_evidence_removal.history_removed.doubao_seed_2_0_lite.has_demand_ablated\endcsname{1.000}
\expandafter\gdef\csname odunum@n@controlled_evidence_removal.history_removed.doubao_seed_2_0_lite.has_demand_ablated\endcsname{391}
\expandafter\gdef\csname odunum@val@controlled_evidence_removal.history_removed.doubao_seed_2_0_lite.has_demand_full\endcsname{1.000}
\expandafter\gdef\csname odunum@n@controlled_evidence_removal.history_removed.doubao_seed_2_0_lite.has_demand_full\endcsname{391}
\expandafter\gdef\csname odunum@val@controlled_evidence_removal.history_removed.doubao_seed_2_0_lite.intent_drop\endcsname{0.072}
\expandafter\gdef\csname odunum@n@controlled_evidence_removal.history_removed.doubao_seed_2_0_lite.intent_drop\endcsname{718}
\expandafter\gdef\csname odunum@ci@controlled_evidence_removal.history_removed.doubao_seed_2_0_lite.intent_drop\endcsname{[0.046, 0.100]}
\expandafter\gdef\csname odunum@val@controlled_evidence_removal.history_removed.doubao_seed_2_0_lite.n_scenes\endcsname{391}
\expandafter\gdef\csname odunum@n@controlled_evidence_removal.history_removed.doubao_seed_2_0_lite.n_scenes\endcsname{391}
\expandafter\gdef\csname odunum@val@controlled_evidence_removal.history_removed.doubao_seed_2_0_lite.specificity\endcsname{0.345}
\expandafter\gdef\csname odunum@n@controlled_evidence_removal.history_removed.doubao_seed_2_0_lite.specificity\endcsname{391}
\expandafter\gdef\csname odunum@ci@controlled_evidence_removal.history_removed.doubao_seed_2_0_lite.specificity\endcsname{[0.292, 0.398]}
\expandafter\gdef\csname odunum@val@controlled_evidence_removal.history_removed.doubao_seed_2_0_lite.target_drop\endcsname{0.417}
\expandafter\gdef\csname odunum@n@controlled_evidence_removal.history_removed.doubao_seed_2_0_lite.target_drop\endcsname{465}
\expandafter\gdef\csname odunum@ci@controlled_evidence_removal.history_removed.doubao_seed_2_0_lite.target_drop\endcsname{[0.370, 0.466]}
\expandafter\gdef\csname odunum@val@controlled_evidence_removal.history_removed.gemini_3_1_pro.ablated.audio\endcsname{0.398}
\expandafter\gdef\csname odunum@n@controlled_evidence_removal.history_removed.gemini_3_1_pro.ablated.audio\endcsname{118}
\expandafter\gdef\csname odunum@val@controlled_evidence_removal.history_removed.gemini_3_1_pro.ablated.history\endcsname{0.166}
\expandafter\gdef\csname odunum@n@controlled_evidence_removal.history_removed.gemini_3_1_pro.ablated.history\endcsname{465}
\expandafter\gdef\csname odunum@val@controlled_evidence_removal.history_removed.gemini_3_1_pro.ablated.intent\endcsname{0.701}
\expandafter\gdef\csname odunum@n@controlled_evidence_removal.history_removed.gemini_3_1_pro.ablated.intent\endcsname{718}
\expandafter\gdef\csname odunum@val@controlled_evidence_removal.history_removed.gemini_3_1_pro.ablated.other_context\endcsname{0.120}
\expandafter\gdef\csname odunum@n@controlled_evidence_removal.history_removed.gemini_3_1_pro.ablated.other_context\endcsname{25}
\expandafter\gdef\csname odunum@val@controlled_evidence_removal.history_removed.gemini_3_1_pro.ablated.visual\endcsname{0.448}
\expandafter\gdef\csname odunum@n@controlled_evidence_removal.history_removed.gemini_3_1_pro.ablated.visual\endcsname{210}
\expandafter\gdef\csname odunum@val@controlled_evidence_removal.history_removed.gemini_3_1_pro.drop.audio\endcsname{0.051}
\expandafter\gdef\csname odunum@n@controlled_evidence_removal.history_removed.gemini_3_1_pro.drop.audio\endcsname{118}
\expandafter\gdef\csname odunum@val@controlled_evidence_removal.history_removed.gemini_3_1_pro.drop.history\endcsname{0.243}
\expandafter\gdef\csname odunum@n@controlled_evidence_removal.history_removed.gemini_3_1_pro.drop.history\endcsname{465}
\expandafter\gdef\csname odunum@val@controlled_evidence_removal.history_removed.gemini_3_1_pro.drop.intent\endcsname{0.159}
\expandafter\gdef\csname odunum@n@controlled_evidence_removal.history_removed.gemini_3_1_pro.drop.intent\endcsname{718}
\expandafter\gdef\csname odunum@val@controlled_evidence_removal.history_removed.gemini_3_1_pro.drop.other_context\endcsname{0.160}
\expandafter\gdef\csname odunum@n@controlled_evidence_removal.history_removed.gemini_3_1_pro.drop.other_context\endcsname{25}
\expandafter\gdef\csname odunum@val@controlled_evidence_removal.history_removed.gemini_3_1_pro.drop.visual\endcsname{0.062}
\expandafter\gdef\csname odunum@n@controlled_evidence_removal.history_removed.gemini_3_1_pro.drop.visual\endcsname{210}
\expandafter\gdef\csname odunum@val@controlled_evidence_removal.history_removed.gemini_3_1_pro.flips_false_to_true\endcsname{1}
\expandafter\gdef\csname odunum@n@controlled_evidence_removal.history_removed.gemini_3_1_pro.flips_false_to_true\endcsname{391}
\expandafter\gdef\csname odunum@val@controlled_evidence_removal.history_removed.gemini_3_1_pro.flips_true_to_false\endcsname{11}
\expandafter\gdef\csname odunum@n@controlled_evidence_removal.history_removed.gemini_3_1_pro.flips_true_to_false\endcsname{391}
\expandafter\gdef\csname odunum@val@controlled_evidence_removal.history_removed.gemini_3_1_pro.full.audio\endcsname{0.449}
\expandafter\gdef\csname odunum@n@controlled_evidence_removal.history_removed.gemini_3_1_pro.full.audio\endcsname{118}
\expandafter\gdef\csname odunum@val@controlled_evidence_removal.history_removed.gemini_3_1_pro.full.history\endcsname{0.409}
\expandafter\gdef\csname odunum@n@controlled_evidence_removal.history_removed.gemini_3_1_pro.full.history\endcsname{465}
\expandafter\gdef\csname odunum@val@controlled_evidence_removal.history_removed.gemini_3_1_pro.full.intent\endcsname{0.859}
\expandafter\gdef\csname odunum@n@controlled_evidence_removal.history_removed.gemini_3_1_pro.full.intent\endcsname{718}
\expandafter\gdef\csname odunum@val@controlled_evidence_removal.history_removed.gemini_3_1_pro.full.other_context\endcsname{0.280}
\expandafter\gdef\csname odunum@n@controlled_evidence_removal.history_removed.gemini_3_1_pro.full.other_context\endcsname{25}
\expandafter\gdef\csname odunum@val@controlled_evidence_removal.history_removed.gemini_3_1_pro.full.visual\endcsname{0.510}
\expandafter\gdef\csname odunum@n@controlled_evidence_removal.history_removed.gemini_3_1_pro.full.visual\endcsname{210}
\expandafter\gdef\csname odunum@val@controlled_evidence_removal.history_removed.gemini_3_1_pro.has_demand_ablated\endcsname{0.969}
\expandafter\gdef\csname odunum@n@controlled_evidence_removal.history_removed.gemini_3_1_pro.has_demand_ablated\endcsname{391}
\expandafter\gdef\csname odunum@val@controlled_evidence_removal.history_removed.gemini_3_1_pro.has_demand_full\endcsname{0.995}
\expandafter\gdef\csname odunum@n@controlled_evidence_removal.history_removed.gemini_3_1_pro.has_demand_full\endcsname{391}
\expandafter\gdef\csname odunum@val@controlled_evidence_removal.history_removed.gemini_3_1_pro.intent_drop\endcsname{0.159}
\expandafter\gdef\csname odunum@n@controlled_evidence_removal.history_removed.gemini_3_1_pro.intent_drop\endcsname{718}
\expandafter\gdef\csname odunum@ci@controlled_evidence_removal.history_removed.gemini_3_1_pro.intent_drop\endcsname{[0.122, 0.195]}
\expandafter\gdef\csname odunum@val@controlled_evidence_removal.history_removed.gemini_3_1_pro.n_scenes\endcsname{391}
\expandafter\gdef\csname odunum@n@controlled_evidence_removal.history_removed.gemini_3_1_pro.n_scenes\endcsname{391}
\expandafter\gdef\csname odunum@val@controlled_evidence_removal.history_removed.gemini_3_1_pro.specificity\endcsname{0.084}
\expandafter\gdef\csname odunum@n@controlled_evidence_removal.history_removed.gemini_3_1_pro.specificity\endcsname{391}
\expandafter\gdef\csname odunum@ci@controlled_evidence_removal.history_removed.gemini_3_1_pro.specificity\endcsname{[0.034, 0.136]}
\expandafter\gdef\csname odunum@val@controlled_evidence_removal.history_removed.gemini_3_1_pro.target_drop\endcsname{0.243}
\expandafter\gdef\csname odunum@n@controlled_evidence_removal.history_removed.gemini_3_1_pro.target_drop\endcsname{465}
\expandafter\gdef\csname odunum@ci@controlled_evidence_removal.history_removed.gemini_3_1_pro.target_drop\endcsname{[0.198, 0.290]}
\expandafter\gdef\csname odunum@val@controlled_evidence_removal.history_removed.qwen3_5_omni_plus.ablated.audio\endcsname{0.441}
\expandafter\gdef\csname odunum@n@controlled_evidence_removal.history_removed.qwen3_5_omni_plus.ablated.audio\endcsname{118}
\expandafter\gdef\csname odunum@val@controlled_evidence_removal.history_removed.qwen3_5_omni_plus.ablated.history\endcsname{0.185}
\expandafter\gdef\csname odunum@n@controlled_evidence_removal.history_removed.qwen3_5_omni_plus.ablated.history\endcsname{465}
\expandafter\gdef\csname odunum@val@controlled_evidence_removal.history_removed.qwen3_5_omni_plus.ablated.intent\endcsname{0.770}
\expandafter\gdef\csname odunum@n@controlled_evidence_removal.history_removed.qwen3_5_omni_plus.ablated.intent\endcsname{718}
\expandafter\gdef\csname odunum@val@controlled_evidence_removal.history_removed.qwen3_5_omni_plus.ablated.other_context\endcsname{0.560}
\expandafter\gdef\csname odunum@n@controlled_evidence_removal.history_removed.qwen3_5_omni_plus.ablated.other_context\endcsname{25}
\expandafter\gdef\csname odunum@val@controlled_evidence_removal.history_removed.qwen3_5_omni_plus.ablated.visual\endcsname{0.452}
\expandafter\gdef\csname odunum@n@controlled_evidence_removal.history_removed.qwen3_5_omni_plus.ablated.visual\endcsname{210}
\expandafter\gdef\csname odunum@val@controlled_evidence_removal.history_removed.qwen3_5_omni_plus.drop.audio\endcsname{0.093}
\expandafter\gdef\csname odunum@n@controlled_evidence_removal.history_removed.qwen3_5_omni_plus.drop.audio\endcsname{118}
\expandafter\gdef\csname odunum@val@controlled_evidence_removal.history_removed.qwen3_5_omni_plus.drop.history\endcsname{0.320}
\expandafter\gdef\csname odunum@n@controlled_evidence_removal.history_removed.qwen3_5_omni_plus.drop.history\endcsname{465}
\expandafter\gdef\csname odunum@val@controlled_evidence_removal.history_removed.qwen3_5_omni_plus.drop.intent\endcsname{0.091}
\expandafter\gdef\csname odunum@n@controlled_evidence_removal.history_removed.qwen3_5_omni_plus.drop.intent\endcsname{718}
\expandafter\gdef\csname odunum@val@controlled_evidence_removal.history_removed.qwen3_5_omni_plus.drop.other_context\endcsname{0.080}
\expandafter\gdef\csname odunum@n@controlled_evidence_removal.history_removed.qwen3_5_omni_plus.drop.other_context\endcsname{25}
\expandafter\gdef\csname odunum@val@controlled_evidence_removal.history_removed.qwen3_5_omni_plus.drop.visual\endcsname{0.062}
\expandafter\gdef\csname odunum@n@controlled_evidence_removal.history_removed.qwen3_5_omni_plus.drop.visual\endcsname{210}
\expandafter\gdef\csname odunum@val@controlled_evidence_removal.history_removed.qwen3_5_omni_plus.flips_false_to_true\endcsname{3}
\expandafter\gdef\csname odunum@n@controlled_evidence_removal.history_removed.qwen3_5_omni_plus.flips_false_to_true\endcsname{391}
\expandafter\gdef\csname odunum@val@controlled_evidence_removal.history_removed.qwen3_5_omni_plus.flips_true_to_false\endcsname{9}
\expandafter\gdef\csname odunum@n@controlled_evidence_removal.history_removed.qwen3_5_omni_plus.flips_true_to_false\endcsname{391}
\expandafter\gdef\csname odunum@val@controlled_evidence_removal.history_removed.qwen3_5_omni_plus.full.audio\endcsname{0.534}
\expandafter\gdef\csname odunum@n@controlled_evidence_removal.history_removed.qwen3_5_omni_plus.full.audio\endcsname{118}
\expandafter\gdef\csname odunum@val@controlled_evidence_removal.history_removed.qwen3_5_omni_plus.full.history\endcsname{0.505}
\expandafter\gdef\csname odunum@n@controlled_evidence_removal.history_removed.qwen3_5_omni_plus.full.history\endcsname{465}
\expandafter\gdef\csname odunum@val@controlled_evidence_removal.history_removed.qwen3_5_omni_plus.full.intent\endcsname{0.861}
\expandafter\gdef\csname odunum@n@controlled_evidence_removal.history_removed.qwen3_5_omni_plus.full.intent\endcsname{718}
\expandafter\gdef\csname odunum@val@controlled_evidence_removal.history_removed.qwen3_5_omni_plus.full.other_context\endcsname{0.640}
\expandafter\gdef\csname odunum@n@controlled_evidence_removal.history_removed.qwen3_5_omni_plus.full.other_context\endcsname{25}
\expandafter\gdef\csname odunum@val@controlled_evidence_removal.history_removed.qwen3_5_omni_plus.full.visual\endcsname{0.514}
\expandafter\gdef\csname odunum@n@controlled_evidence_removal.history_removed.qwen3_5_omni_plus.full.visual\endcsname{210}
\expandafter\gdef\csname odunum@val@controlled_evidence_removal.history_removed.qwen3_5_omni_plus.has_demand_ablated\endcsname{0.977}
\expandafter\gdef\csname odunum@n@controlled_evidence_removal.history_removed.qwen3_5_omni_plus.has_demand_ablated\endcsname{391}
\expandafter\gdef\csname odunum@val@controlled_evidence_removal.history_removed.qwen3_5_omni_plus.has_demand_full\endcsname{0.992}
\expandafter\gdef\csname odunum@n@controlled_evidence_removal.history_removed.qwen3_5_omni_plus.has_demand_full\endcsname{391}
\expandafter\gdef\csname odunum@val@controlled_evidence_removal.history_removed.qwen3_5_omni_plus.intent_drop\endcsname{0.091}
\expandafter\gdef\csname odunum@n@controlled_evidence_removal.history_removed.qwen3_5_omni_plus.intent_drop\endcsname{718}
\expandafter\gdef\csname odunum@ci@controlled_evidence_removal.history_removed.qwen3_5_omni_plus.intent_drop\endcsname{[0.059, 0.123]}
\expandafter\gdef\csname odunum@val@controlled_evidence_removal.history_removed.qwen3_5_omni_plus.n_scenes\endcsname{391}
\expandafter\gdef\csname odunum@n@controlled_evidence_removal.history_removed.qwen3_5_omni_plus.n_scenes\endcsname{391}
\expandafter\gdef\csname odunum@val@controlled_evidence_removal.history_removed.qwen3_5_omni_plus.specificity\endcsname{0.230}
\expandafter\gdef\csname odunum@n@controlled_evidence_removal.history_removed.qwen3_5_omni_plus.specificity\endcsname{391}
\expandafter\gdef\csname odunum@ci@controlled_evidence_removal.history_removed.qwen3_5_omni_plus.specificity\endcsname{[0.178, 0.283]}
\expandafter\gdef\csname odunum@val@controlled_evidence_removal.history_removed.qwen3_5_omni_plus.target_drop\endcsname{0.320}
\expandafter\gdef\csname odunum@n@controlled_evidence_removal.history_removed.qwen3_5_omni_plus.target_drop\endcsname{465}
\expandafter\gdef\csname odunum@ci@controlled_evidence_removal.history_removed.qwen3_5_omni_plus.target_drop\endcsname{[0.272, 0.369]}
\expandafter\gdef\csname odunum@val@controlled_evidence_removal.visual_masked.doubao_seed_2_0_lite.ablated.audio\endcsname{0.497}
\expandafter\gdef\csname odunum@n@controlled_evidence_removal.visual_masked.doubao_seed_2_0_lite.ablated.audio\endcsname{195}
\expandafter\gdef\csname odunum@val@controlled_evidence_removal.visual_masked.doubao_seed_2_0_lite.ablated.history\endcsname{0.595}
\expandafter\gdef\csname odunum@n@controlled_evidence_removal.visual_masked.doubao_seed_2_0_lite.ablated.history\endcsname{131}
\expandafter\gdef\csname odunum@val@controlled_evidence_removal.visual_masked.doubao_seed_2_0_lite.ablated.intent\endcsname{0.874}
\expandafter\gdef\csname odunum@n@controlled_evidence_removal.visual_masked.doubao_seed_2_0_lite.ablated.intent\endcsname{749}
\expandafter\gdef\csname odunum@val@controlled_evidence_removal.visual_masked.doubao_seed_2_0_lite.ablated.other_context\endcsname{0.167}
\expandafter\gdef\csname odunum@n@controlled_evidence_removal.visual_masked.doubao_seed_2_0_lite.ablated.other_context\endcsname{42}
\expandafter\gdef\csname odunum@val@controlled_evidence_removal.visual_masked.doubao_seed_2_0_lite.ablated.unattributed\endcsname{0.500}
\expandafter\gdef\csname odunum@n@controlled_evidence_removal.visual_masked.doubao_seed_2_0_lite.ablated.unattributed\endcsname{4}
\expandafter\gdef\csname odunum@val@controlled_evidence_removal.visual_masked.doubao_seed_2_0_lite.ablated.visual\endcsname{0.252}
\expandafter\gdef\csname odunum@n@controlled_evidence_removal.visual_masked.doubao_seed_2_0_lite.ablated.visual\endcsname{469}
\expandafter\gdef\csname odunum@val@controlled_evidence_removal.visual_masked.doubao_seed_2_0_lite.drop.audio\endcsname{0.026}
\expandafter\gdef\csname odunum@n@controlled_evidence_removal.visual_masked.doubao_seed_2_0_lite.drop.audio\endcsname{195}
\expandafter\gdef\csname odunum@val@controlled_evidence_removal.visual_masked.doubao_seed_2_0_lite.drop.history\endcsname{0.053}
\expandafter\gdef\csname odunum@n@controlled_evidence_removal.visual_masked.doubao_seed_2_0_lite.drop.history\endcsname{131}
\expandafter\gdef\csname odunum@val@controlled_evidence_removal.visual_masked.doubao_seed_2_0_lite.drop.intent\endcsname{0.003}
\expandafter\gdef\csname odunum@n@controlled_evidence_removal.visual_masked.doubao_seed_2_0_lite.drop.intent\endcsname{749}
\expandafter\gdef\csname odunum@val@controlled_evidence_removal.visual_masked.doubao_seed_2_0_lite.drop.other_context\endcsname{0.214}
\expandafter\gdef\csname odunum@n@controlled_evidence_removal.visual_masked.doubao_seed_2_0_lite.drop.other_context\endcsname{42}
\expandafter\gdef\csname odunum@val@controlled_evidence_removal.visual_masked.doubao_seed_2_0_lite.drop.unattributed\endcsname{0.000}
\expandafter\gdef\csname odunum@n@controlled_evidence_removal.visual_masked.doubao_seed_2_0_lite.drop.unattributed\endcsname{4}
\expandafter\gdef\csname odunum@val@controlled_evidence_removal.visual_masked.doubao_seed_2_0_lite.drop.visual\endcsname{0.330}
\expandafter\gdef\csname odunum@n@controlled_evidence_removal.visual_masked.doubao_seed_2_0_lite.drop.visual\endcsname{469}
\expandafter\gdef\csname odunum@val@controlled_evidence_removal.visual_masked.doubao_seed_2_0_lite.flips_false_to_true\endcsname{0}
\expandafter\gdef\csname odunum@n@controlled_evidence_removal.visual_masked.doubao_seed_2_0_lite.flips_false_to_true\endcsname{400}
\expandafter\gdef\csname odunum@val@controlled_evidence_removal.visual_masked.doubao_seed_2_0_lite.flips_true_to_false\endcsname{1}
\expandafter\gdef\csname odunum@n@controlled_evidence_removal.visual_masked.doubao_seed_2_0_lite.flips_true_to_false\endcsname{400}
\expandafter\gdef\csname odunum@val@controlled_evidence_removal.visual_masked.doubao_seed_2_0_lite.full.audio\endcsname{0.523}
\expandafter\gdef\csname odunum@n@controlled_evidence_removal.visual_masked.doubao_seed_2_0_lite.full.audio\endcsname{195}
\expandafter\gdef\csname odunum@val@controlled_evidence_removal.visual_masked.doubao_seed_2_0_lite.full.history\endcsname{0.649}
\expandafter\gdef\csname odunum@n@controlled_evidence_removal.visual_masked.doubao_seed_2_0_lite.full.history\endcsname{131}
\expandafter\gdef\csname odunum@val@controlled_evidence_removal.visual_masked.doubao_seed_2_0_lite.full.intent\endcsname{0.877}
\expandafter\gdef\csname odunum@n@controlled_evidence_removal.visual_masked.doubao_seed_2_0_lite.full.intent\endcsname{749}
\expandafter\gdef\csname odunum@val@controlled_evidence_removal.visual_masked.doubao_seed_2_0_lite.full.other_context\endcsname{0.381}
\expandafter\gdef\csname odunum@n@controlled_evidence_removal.visual_masked.doubao_seed_2_0_lite.full.other_context\endcsname{42}
\expandafter\gdef\csname odunum@val@controlled_evidence_removal.visual_masked.doubao_seed_2_0_lite.full.unattributed\endcsname{0.500}
\expandafter\gdef\csname odunum@n@controlled_evidence_removal.visual_masked.doubao_seed_2_0_lite.full.unattributed\endcsname{4}
\expandafter\gdef\csname odunum@val@controlled_evidence_removal.visual_masked.doubao_seed_2_0_lite.full.visual\endcsname{0.582}
\expandafter\gdef\csname odunum@n@controlled_evidence_removal.visual_masked.doubao_seed_2_0_lite.full.visual\endcsname{469}
\expandafter\gdef\csname odunum@val@controlled_evidence_removal.visual_masked.doubao_seed_2_0_lite.has_demand_ablated\endcsname{0.998}
\expandafter\gdef\csname odunum@n@controlled_evidence_removal.visual_masked.doubao_seed_2_0_lite.has_demand_ablated\endcsname{400}
\expandafter\gdef\csname odunum@val@controlled_evidence_removal.visual_masked.doubao_seed_2_0_lite.has_demand_full\endcsname{1.000}
\expandafter\gdef\csname odunum@n@controlled_evidence_removal.visual_masked.doubao_seed_2_0_lite.has_demand_full\endcsname{400}
\expandafter\gdef\csname odunum@val@controlled_evidence_removal.visual_masked.doubao_seed_2_0_lite.intent_drop\endcsname{0.003}
\expandafter\gdef\csname odunum@n@controlled_evidence_removal.visual_masked.doubao_seed_2_0_lite.intent_drop\endcsname{749}
\expandafter\gdef\csname odunum@ci@controlled_evidence_removal.visual_masked.doubao_seed_2_0_lite.intent_drop\endcsname{[-0.017, 0.023]}
\expandafter\gdef\csname odunum@val@controlled_evidence_removal.visual_masked.doubao_seed_2_0_lite.n_scenes\endcsname{400}
\expandafter\gdef\csname odunum@n@controlled_evidence_removal.visual_masked.doubao_seed_2_0_lite.n_scenes\endcsname{400}
\expandafter\gdef\csname odunum@val@controlled_evidence_removal.visual_masked.doubao_seed_2_0_lite.specificity\endcsname{0.328}
\expandafter\gdef\csname odunum@n@controlled_evidence_removal.visual_masked.doubao_seed_2_0_lite.specificity\endcsname{400}
\expandafter\gdef\csname odunum@ci@controlled_evidence_removal.visual_masked.doubao_seed_2_0_lite.specificity\endcsname{[0.276, 0.379]}
\expandafter\gdef\csname odunum@val@controlled_evidence_removal.visual_masked.doubao_seed_2_0_lite.target_drop\endcsname{0.330}
\expandafter\gdef\csname odunum@n@controlled_evidence_removal.visual_masked.doubao_seed_2_0_lite.target_drop\endcsname{469}
\expandafter\gdef\csname odunum@ci@controlled_evidence_removal.visual_masked.doubao_seed_2_0_lite.target_drop\endcsname{[0.281, 0.380]}
\expandafter\gdef\csname odunum@val@controlled_evidence_removal.visual_masked.gemini_3_1_pro.ablated.audio\endcsname{0.446}
\expandafter\gdef\csname odunum@n@controlled_evidence_removal.visual_masked.gemini_3_1_pro.ablated.audio\endcsname{195}
\expandafter\gdef\csname odunum@val@controlled_evidence_removal.visual_masked.gemini_3_1_pro.ablated.history\endcsname{0.405}
\expandafter\gdef\csname odunum@n@controlled_evidence_removal.visual_masked.gemini_3_1_pro.ablated.history\endcsname{131}
\expandafter\gdef\csname odunum@val@controlled_evidence_removal.visual_masked.gemini_3_1_pro.ablated.intent\endcsname{0.816}
\expandafter\gdef\csname odunum@n@controlled_evidence_removal.visual_masked.gemini_3_1_pro.ablated.intent\endcsname{749}
\expandafter\gdef\csname odunum@val@controlled_evidence_removal.visual_masked.gemini_3_1_pro.ablated.other_context\endcsname{0.214}
\expandafter\gdef\csname odunum@n@controlled_evidence_removal.visual_masked.gemini_3_1_pro.ablated.other_context\endcsname{42}
\expandafter\gdef\csname odunum@val@controlled_evidence_removal.visual_masked.gemini_3_1_pro.ablated.unattributed\endcsname{0.750}
\expandafter\gdef\csname odunum@n@controlled_evidence_removal.visual_masked.gemini_3_1_pro.ablated.unattributed\endcsname{4}
\expandafter\gdef\csname odunum@val@controlled_evidence_removal.visual_masked.gemini_3_1_pro.ablated.visual\endcsname{0.198}
\expandafter\gdef\csname odunum@n@controlled_evidence_removal.visual_masked.gemini_3_1_pro.ablated.visual\endcsname{469}
\expandafter\gdef\csname odunum@val@controlled_evidence_removal.visual_masked.gemini_3_1_pro.drop.audio\endcsname{0.041}
\expandafter\gdef\csname odunum@n@controlled_evidence_removal.visual_masked.gemini_3_1_pro.drop.audio\endcsname{195}
\expandafter\gdef\csname odunum@val@controlled_evidence_removal.visual_masked.gemini_3_1_pro.drop.history\endcsname{0.046}
\expandafter\gdef\csname odunum@n@controlled_evidence_removal.visual_masked.gemini_3_1_pro.drop.history\endcsname{131}
\expandafter\gdef\csname odunum@val@controlled_evidence_removal.visual_masked.gemini_3_1_pro.drop.intent\endcsname{0.033}
\expandafter\gdef\csname odunum@n@controlled_evidence_removal.visual_masked.gemini_3_1_pro.drop.intent\endcsname{749}
\expandafter\gdef\csname odunum@val@controlled_evidence_removal.visual_masked.gemini_3_1_pro.drop.other_context\endcsname{0.024}
\expandafter\gdef\csname odunum@n@controlled_evidence_removal.visual_masked.gemini_3_1_pro.drop.other_context\endcsname{42}
\expandafter\gdef\csname odunum@val@controlled_evidence_removal.visual_masked.gemini_3_1_pro.drop.unattributed\endcsname{-0.500}
\expandafter\gdef\csname odunum@n@controlled_evidence_removal.visual_masked.gemini_3_1_pro.drop.unattributed\endcsname{4}
\expandafter\gdef\csname odunum@val@controlled_evidence_removal.visual_masked.gemini_3_1_pro.drop.visual\endcsname{0.288}
\expandafter\gdef\csname odunum@n@controlled_evidence_removal.visual_masked.gemini_3_1_pro.drop.visual\endcsname{469}
\expandafter\gdef\csname odunum@val@controlled_evidence_removal.visual_masked.gemini_3_1_pro.flips_false_to_true\endcsname{4}
\expandafter\gdef\csname odunum@n@controlled_evidence_removal.visual_masked.gemini_3_1_pro.flips_false_to_true\endcsname{400}
\expandafter\gdef\csname odunum@val@controlled_evidence_removal.visual_masked.gemini_3_1_pro.flips_true_to_false\endcsname{9}
\expandafter\gdef\csname odunum@n@controlled_evidence_removal.visual_masked.gemini_3_1_pro.flips_true_to_false\endcsname{400}
\expandafter\gdef\csname odunum@val@controlled_evidence_removal.visual_masked.gemini_3_1_pro.full.audio\endcsname{0.487}
\expandafter\gdef\csname odunum@n@controlled_evidence_removal.visual_masked.gemini_3_1_pro.full.audio\endcsname{195}
\expandafter\gdef\csname odunum@val@controlled_evidence_removal.visual_masked.gemini_3_1_pro.full.history\endcsname{0.450}
\expandafter\gdef\csname odunum@n@controlled_evidence_removal.visual_masked.gemini_3_1_pro.full.history\endcsname{131}
\expandafter\gdef\csname odunum@val@controlled_evidence_removal.visual_masked.gemini_3_1_pro.full.intent\endcsname{0.849}
\expandafter\gdef\csname odunum@n@controlled_evidence_removal.visual_masked.gemini_3_1_pro.full.intent\endcsname{749}
\expandafter\gdef\csname odunum@val@controlled_evidence_removal.visual_masked.gemini_3_1_pro.full.other_context\endcsname{0.238}
\expandafter\gdef\csname odunum@n@controlled_evidence_removal.visual_masked.gemini_3_1_pro.full.other_context\endcsname{42}
\expandafter\gdef\csname odunum@val@controlled_evidence_removal.visual_masked.gemini_3_1_pro.full.unattributed\endcsname{0.250}
\expandafter\gdef\csname odunum@n@controlled_evidence_removal.visual_masked.gemini_3_1_pro.full.unattributed\endcsname{4}
\expandafter\gdef\csname odunum@val@controlled_evidence_removal.visual_masked.gemini_3_1_pro.full.visual\endcsname{0.486}
\expandafter\gdef\csname odunum@n@controlled_evidence_removal.visual_masked.gemini_3_1_pro.full.visual\endcsname{469}
\expandafter\gdef\csname odunum@val@controlled_evidence_removal.visual_masked.gemini_3_1_pro.has_demand_ablated\endcsname{0.970}
\expandafter\gdef\csname odunum@n@controlled_evidence_removal.visual_masked.gemini_3_1_pro.has_demand_ablated\endcsname{400}
\expandafter\gdef\csname odunum@val@controlled_evidence_removal.visual_masked.gemini_3_1_pro.has_demand_full\endcsname{0.983}
\expandafter\gdef\csname odunum@n@controlled_evidence_removal.visual_masked.gemini_3_1_pro.has_demand_full\endcsname{400}
\expandafter\gdef\csname odunum@val@controlled_evidence_removal.visual_masked.gemini_3_1_pro.intent_drop\endcsname{0.033}
\expandafter\gdef\csname odunum@n@controlled_evidence_removal.visual_masked.gemini_3_1_pro.intent_drop\endcsname{749}
\expandafter\gdef\csname odunum@ci@controlled_evidence_removal.visual_masked.gemini_3_1_pro.intent_drop\endcsname{[0.007, 0.060]}
\expandafter\gdef\csname odunum@val@controlled_evidence_removal.visual_masked.gemini_3_1_pro.n_scenes\endcsname{400}
\expandafter\gdef\csname odunum@n@controlled_evidence_removal.visual_masked.gemini_3_1_pro.n_scenes\endcsname{400}
\expandafter\gdef\csname odunum@val@controlled_evidence_removal.visual_masked.gemini_3_1_pro.specificity\endcsname{0.254}
\expandafter\gdef\csname odunum@n@controlled_evidence_removal.visual_masked.gemini_3_1_pro.specificity\endcsname{400}
\expandafter\gdef\csname odunum@ci@controlled_evidence_removal.visual_masked.gemini_3_1_pro.specificity\endcsname{[0.203, 0.307]}
\expandafter\gdef\csname odunum@val@controlled_evidence_removal.visual_masked.gemini_3_1_pro.target_drop\endcsname{0.288}
\expandafter\gdef\csname odunum@n@controlled_evidence_removal.visual_masked.gemini_3_1_pro.target_drop\endcsname{469}
\expandafter\gdef\csname odunum@ci@controlled_evidence_removal.visual_masked.gemini_3_1_pro.target_drop\endcsname{[0.241, 0.335]}
\expandafter\gdef\csname odunum@val@controlled_evidence_removal.visual_masked.qwen3_5_omni_plus.ablated.audio\endcsname{0.415}
\expandafter\gdef\csname odunum@n@controlled_evidence_removal.visual_masked.qwen3_5_omni_plus.ablated.audio\endcsname{195}
\expandafter\gdef\csname odunum@val@controlled_evidence_removal.visual_masked.qwen3_5_omni_plus.ablated.history\endcsname{0.489}
\expandafter\gdef\csname odunum@n@controlled_evidence_removal.visual_masked.qwen3_5_omni_plus.ablated.history\endcsname{131}
\expandafter\gdef\csname odunum@val@controlled_evidence_removal.visual_masked.qwen3_5_omni_plus.ablated.intent\endcsname{0.817}
\expandafter\gdef\csname odunum@n@controlled_evidence_removal.visual_masked.qwen3_5_omni_plus.ablated.intent\endcsname{749}
\expandafter\gdef\csname odunum@val@controlled_evidence_removal.visual_masked.qwen3_5_omni_plus.ablated.other_context\endcsname{0.190}
\expandafter\gdef\csname odunum@n@controlled_evidence_removal.visual_masked.qwen3_5_omni_plus.ablated.other_context\endcsname{42}
\expandafter\gdef\csname odunum@val@controlled_evidence_removal.visual_masked.qwen3_5_omni_plus.ablated.unattributed\endcsname{0.500}
\expandafter\gdef\csname odunum@n@controlled_evidence_removal.visual_masked.qwen3_5_omni_plus.ablated.unattributed\endcsname{4}
\expandafter\gdef\csname odunum@val@controlled_evidence_removal.visual_masked.qwen3_5_omni_plus.ablated.visual\endcsname{0.186}
\expandafter\gdef\csname odunum@n@controlled_evidence_removal.visual_masked.qwen3_5_omni_plus.ablated.visual\endcsname{469}
\expandafter\gdef\csname odunum@val@controlled_evidence_removal.visual_masked.qwen3_5_omni_plus.drop.audio\endcsname{0.046}
\expandafter\gdef\csname odunum@n@controlled_evidence_removal.visual_masked.qwen3_5_omni_plus.drop.audio\endcsname{195}
\expandafter\gdef\csname odunum@val@controlled_evidence_removal.visual_masked.qwen3_5_omni_plus.drop.history\endcsname{0.092}
\expandafter\gdef\csname odunum@n@controlled_evidence_removal.visual_masked.qwen3_5_omni_plus.drop.history\endcsname{131}
\expandafter\gdef\csname odunum@val@controlled_evidence_removal.visual_masked.qwen3_5_omni_plus.drop.intent\endcsname{-0.004}
\expandafter\gdef\csname odunum@n@controlled_evidence_removal.visual_masked.qwen3_5_omni_plus.drop.intent\endcsname{749}
\expandafter\gdef\csname odunum@val@controlled_evidence_removal.visual_masked.qwen3_5_omni_plus.drop.other_context\endcsname{0.190}
\expandafter\gdef\csname odunum@n@controlled_evidence_removal.visual_masked.qwen3_5_omni_plus.drop.other_context\endcsname{42}
\expandafter\gdef\csname odunum@val@controlled_evidence_removal.visual_masked.qwen3_5_omni_plus.drop.unattributed\endcsname{0.000}
\expandafter\gdef\csname odunum@n@controlled_evidence_removal.visual_masked.qwen3_5_omni_plus.drop.unattributed\endcsname{4}
\expandafter\gdef\csname odunum@val@controlled_evidence_removal.visual_masked.qwen3_5_omni_plus.drop.visual\endcsname{0.284}
\expandafter\gdef\csname odunum@n@controlled_evidence_removal.visual_masked.qwen3_5_omni_plus.drop.visual\endcsname{469}
\expandafter\gdef\csname odunum@val@controlled_evidence_removal.visual_masked.qwen3_5_omni_plus.flips_false_to_true\endcsname{6}
\expandafter\gdef\csname odunum@n@controlled_evidence_removal.visual_masked.qwen3_5_omni_plus.flips_false_to_true\endcsname{400}
\expandafter\gdef\csname odunum@val@controlled_evidence_removal.visual_masked.qwen3_5_omni_plus.flips_true_to_false\endcsname{4}
\expandafter\gdef\csname odunum@n@controlled_evidence_removal.visual_masked.qwen3_5_omni_plus.flips_true_to_false\endcsname{400}
\expandafter\gdef\csname odunum@val@controlled_evidence_removal.visual_masked.qwen3_5_omni_plus.full.audio\endcsname{0.462}
\expandafter\gdef\csname odunum@n@controlled_evidence_removal.visual_masked.qwen3_5_omni_plus.full.audio\endcsname{195}
\expandafter\gdef\csname odunum@val@controlled_evidence_removal.visual_masked.qwen3_5_omni_plus.full.history\endcsname{0.580}
\expandafter\gdef\csname odunum@n@controlled_evidence_removal.visual_masked.qwen3_5_omni_plus.full.history\endcsname{131}
\expandafter\gdef\csname odunum@val@controlled_evidence_removal.visual_masked.qwen3_5_omni_plus.full.intent\endcsname{0.813}
\expandafter\gdef\csname odunum@n@controlled_evidence_removal.visual_masked.qwen3_5_omni_plus.full.intent\endcsname{749}
\expandafter\gdef\csname odunum@val@controlled_evidence_removal.visual_masked.qwen3_5_omni_plus.full.other_context\endcsname{0.381}
\expandafter\gdef\csname odunum@n@controlled_evidence_removal.visual_masked.qwen3_5_omni_plus.full.other_context\endcsname{42}
\expandafter\gdef\csname odunum@val@controlled_evidence_removal.visual_masked.qwen3_5_omni_plus.full.unattributed\endcsname{0.500}
\expandafter\gdef\csname odunum@n@controlled_evidence_removal.visual_masked.qwen3_5_omni_plus.full.unattributed\endcsname{4}
\expandafter\gdef\csname odunum@val@controlled_evidence_removal.visual_masked.qwen3_5_omni_plus.full.visual\endcsname{0.469}
\expandafter\gdef\csname odunum@n@controlled_evidence_removal.visual_masked.qwen3_5_omni_plus.full.visual\endcsname{469}
\expandafter\gdef\csname odunum@val@controlled_evidence_removal.visual_masked.qwen3_5_omni_plus.has_demand_ablated\endcsname{0.973}
\expandafter\gdef\csname odunum@n@controlled_evidence_removal.visual_masked.qwen3_5_omni_plus.has_demand_ablated\endcsname{400}
\expandafter\gdef\csname odunum@val@controlled_evidence_removal.visual_masked.qwen3_5_omni_plus.has_demand_full\endcsname{0.968}
\expandafter\gdef\csname odunum@n@controlled_evidence_removal.visual_masked.qwen3_5_omni_plus.has_demand_full\endcsname{400}
\expandafter\gdef\csname odunum@val@controlled_evidence_removal.visual_masked.qwen3_5_omni_plus.intent_drop\endcsname{-0.004}
\expandafter\gdef\csname odunum@n@controlled_evidence_removal.visual_masked.qwen3_5_omni_plus.intent_drop\endcsname{749}
\expandafter\gdef\csname odunum@ci@controlled_evidence_removal.visual_masked.qwen3_5_omni_plus.intent_drop\endcsname{[-0.031, 0.023]}
\expandafter\gdef\csname odunum@val@controlled_evidence_removal.visual_masked.qwen3_5_omni_plus.n_scenes\endcsname{400}
\expandafter\gdef\csname odunum@n@controlled_evidence_removal.visual_masked.qwen3_5_omni_plus.n_scenes\endcsname{400}
\expandafter\gdef\csname odunum@val@controlled_evidence_removal.visual_masked.qwen3_5_omni_plus.specificity\endcsname{0.288}
\expandafter\gdef\csname odunum@n@controlled_evidence_removal.visual_masked.qwen3_5_omni_plus.specificity\endcsname{400}
\expandafter\gdef\csname odunum@ci@controlled_evidence_removal.visual_masked.qwen3_5_omni_plus.specificity\endcsname{[0.234, 0.342]}
\expandafter\gdef\csname odunum@val@controlled_evidence_removal.visual_masked.qwen3_5_omni_plus.target_drop\endcsname{0.284}
\expandafter\gdef\csname odunum@n@controlled_evidence_removal.visual_masked.qwen3_5_omni_plus.target_drop\endcsname{469}
\expandafter\gdef\csname odunum@ci@controlled_evidence_removal.visual_masked.qwen3_5_omni_plus.target_drop\endcsname{[0.233, 0.335]}
\expandafter\gdef\csname odunum@val@error_analysis_cases.media_playback.gemini.detection_correct\endcsname{0}
\expandafter\gdef\csname odunum@n@error_analysis_cases.media_playback.gemini.detection_correct\endcsname{1}
\expandafter\gdef\csname odunum@val@error_analysis_cases.media_playback.gemini.m2_status\endcsname{not applicable}
\expandafter\gdef\csname odunum@n@error_analysis_cases.media_playback.gemini.m2_status\endcsname{1}
\expandafter\gdef\csname odunum@val@error_analysis_cases.media_playback.gemini.predicted_demand\endcsname{true}
\expandafter\gdef\csname odunum@n@error_analysis_cases.media_playback.gemini.predicted_demand\endcsname{1}
\expandafter\gdef\csname odunum@val@error_analysis_cases.media_playback.qwen.detection_correct\endcsname{0}
\expandafter\gdef\csname odunum@n@error_analysis_cases.media_playback.qwen.detection_correct\endcsname{1}
\expandafter\gdef\csname odunum@val@error_analysis_cases.media_playback.qwen.m2_status\endcsname{not applicable}
\expandafter\gdef\csname odunum@n@error_analysis_cases.media_playback.qwen.m2_status\endcsname{1}
\expandafter\gdef\csname odunum@val@error_analysis_cases.media_playback.qwen.predicted_demand\endcsname{true}
\expandafter\gdef\csname odunum@n@error_analysis_cases.media_playback.qwen.predicted_demand\endcsname{1}
\expandafter\gdef\csname odunum@val@error_analysis_cases.media_playback.reference_demand\endcsname{false}
\expandafter\gdef\csname odunum@n@error_analysis_cases.media_playback.reference_demand\endcsname{1}
\expandafter\gdef\csname odunum@val@error_analysis_cases.media_playback.scene_id\endcsname{avn\_gen\_000043}
\expandafter\gdef\csname odunum@n@error_analysis_cases.media_playback.scene_id\endcsname{1}
\expandafter\gdef\csname odunum@val@error_analysis_cases.media_playback.seed.detection_correct\endcsname{0}
\expandafter\gdef\csname odunum@n@error_analysis_cases.media_playback.seed.detection_correct\endcsname{1}
\expandafter\gdef\csname odunum@val@error_analysis_cases.media_playback.seed.m2_status\endcsname{not applicable}
\expandafter\gdef\csname odunum@n@error_analysis_cases.media_playback.seed.m2_status\endcsname{1}
\expandafter\gdef\csname odunum@val@error_analysis_cases.media_playback.seed.predicted_demand\endcsname{true}
\expandafter\gdef\csname odunum@n@error_analysis_cases.media_playback.seed.predicted_demand\endcsname{1}
\expandafter\gdef\csname odunum@val@error_analysis_cases.missed_query.gemini.detection_correct\endcsname{1}
\expandafter\gdef\csname odunum@n@error_analysis_cases.missed_query.gemini.detection_correct\endcsname{1}
\expandafter\gdef\csname odunum@val@error_analysis_cases.missed_query.gemini.hits\endcsname{1}
\expandafter\gdef\csname odunum@n@error_analysis_cases.missed_query.gemini.hits\endcsname{1}
\expandafter\gdef\csname odunum@val@error_analysis_cases.missed_query.gemini.kp01\endcsname{1}
\expandafter\gdef\csname odunum@n@error_analysis_cases.missed_query.gemini.kp01\endcsname{1}
\expandafter\gdef\csname odunum@val@error_analysis_cases.missed_query.gemini.kp02\endcsname{0}
\expandafter\gdef\csname odunum@n@error_analysis_cases.missed_query.gemini.kp02\endcsname{1}
\expandafter\gdef\csname odunum@val@error_analysis_cases.missed_query.gemini.m2\endcsname{0.500}
\expandafter\gdef\csname odunum@n@error_analysis_cases.missed_query.gemini.m2\endcsname{2}
\expandafter\gdef\csname odunum@val@error_analysis_cases.missed_query.gemini.predicted_demand\endcsname{true}
\expandafter\gdef\csname odunum@n@error_analysis_cases.missed_query.gemini.predicted_demand\endcsname{1}
\expandafter\gdef\csname odunum@val@error_analysis_cases.missed_query.gemini.total\endcsname{2}
\expandafter\gdef\csname odunum@n@error_analysis_cases.missed_query.gemini.total\endcsname{1}
\expandafter\gdef\csname odunum@val@error_analysis_cases.missed_query.qwen.detection_correct\endcsname{0}
\expandafter\gdef\csname odunum@n@error_analysis_cases.missed_query.qwen.detection_correct\endcsname{1}
\expandafter\gdef\csname odunum@val@error_analysis_cases.missed_query.qwen.hits\endcsname{0}
\expandafter\gdef\csname odunum@n@error_analysis_cases.missed_query.qwen.hits\endcsname{1}
\expandafter\gdef\csname odunum@val@error_analysis_cases.missed_query.qwen.kp01\endcsname{0}
\expandafter\gdef\csname odunum@n@error_analysis_cases.missed_query.qwen.kp01\endcsname{1}
\expandafter\gdef\csname odunum@val@error_analysis_cases.missed_query.qwen.kp02\endcsname{0}
\expandafter\gdef\csname odunum@n@error_analysis_cases.missed_query.qwen.kp02\endcsname{1}
\expandafter\gdef\csname odunum@val@error_analysis_cases.missed_query.qwen.m2\endcsname{0.000}
\expandafter\gdef\csname odunum@n@error_analysis_cases.missed_query.qwen.m2\endcsname{2}
\expandafter\gdef\csname odunum@val@error_analysis_cases.missed_query.qwen.predicted_demand\endcsname{false}
\expandafter\gdef\csname odunum@n@error_analysis_cases.missed_query.qwen.predicted_demand\endcsname{1}
\expandafter\gdef\csname odunum@val@error_analysis_cases.missed_query.qwen.total\endcsname{2}
\expandafter\gdef\csname odunum@n@error_analysis_cases.missed_query.qwen.total\endcsname{1}
\expandafter\gdef\csname odunum@val@error_analysis_cases.missed_query.reference_demand\endcsname{true}
\expandafter\gdef\csname odunum@n@error_analysis_cases.missed_query.reference_demand\endcsname{1}
\expandafter\gdef\csname odunum@val@error_analysis_cases.missed_query.scene_id\endcsname{avp\_gen\_000277}
\expandafter\gdef\csname odunum@n@error_analysis_cases.missed_query.scene_id\endcsname{1}
\expandafter\gdef\csname odunum@val@error_analysis_cases.missed_query.seed.detection_correct\endcsname{1}
\expandafter\gdef\csname odunum@n@error_analysis_cases.missed_query.seed.detection_correct\endcsname{1}
\expandafter\gdef\csname odunum@val@error_analysis_cases.missed_query.seed.hits\endcsname{1}
\expandafter\gdef\csname odunum@n@error_analysis_cases.missed_query.seed.hits\endcsname{1}
\expandafter\gdef\csname odunum@val@error_analysis_cases.missed_query.seed.kp01\endcsname{1}
\expandafter\gdef\csname odunum@n@error_analysis_cases.missed_query.seed.kp01\endcsname{1}
\expandafter\gdef\csname odunum@val@error_analysis_cases.missed_query.seed.kp02\endcsname{0}
\expandafter\gdef\csname odunum@n@error_analysis_cases.missed_query.seed.kp02\endcsname{1}
\expandafter\gdef\csname odunum@val@error_analysis_cases.missed_query.seed.m2\endcsname{0.500}
\expandafter\gdef\csname odunum@n@error_analysis_cases.missed_query.seed.m2\endcsname{2}
\expandafter\gdef\csname odunum@val@error_analysis_cases.missed_query.seed.predicted_demand\endcsname{true}
\expandafter\gdef\csname odunum@n@error_analysis_cases.missed_query.seed.predicted_demand\endcsname{1}
\expandafter\gdef\csname odunum@val@error_analysis_cases.missed_query.seed.total\endcsname{2}
\expandafter\gdef\csname odunum@n@error_analysis_cases.missed_query.seed.total\endcsname{1}
\expandafter\gdef\csname odunum@val@error_analysis_cases.naming.gemini.detection_correct\endcsname{1}
\expandafter\gdef\csname odunum@n@error_analysis_cases.naming.gemini.detection_correct\endcsname{1}
\expandafter\gdef\csname odunum@val@error_analysis_cases.naming.gemini.hits\endcsname{2}
\expandafter\gdef\csname odunum@n@error_analysis_cases.naming.gemini.hits\endcsname{1}
\expandafter\gdef\csname odunum@val@error_analysis_cases.naming.gemini.kp01\endcsname{1}
\expandafter\gdef\csname odunum@n@error_analysis_cases.naming.gemini.kp01\endcsname{1}
\expandafter\gdef\csname odunum@val@error_analysis_cases.naming.gemini.kp02\endcsname{1}
\expandafter\gdef\csname odunum@n@error_analysis_cases.naming.gemini.kp02\endcsname{1}
\expandafter\gdef\csname odunum@val@error_analysis_cases.naming.gemini.kp03\endcsname{0}
\expandafter\gdef\csname odunum@n@error_analysis_cases.naming.gemini.kp03\endcsname{1}
\expandafter\gdef\csname odunum@val@error_analysis_cases.naming.gemini.m2\endcsname{0.667}
\expandafter\gdef\csname odunum@n@error_analysis_cases.naming.gemini.m2\endcsname{3}
\expandafter\gdef\csname odunum@val@error_analysis_cases.naming.gemini.predicted_demand\endcsname{true}
\expandafter\gdef\csname odunum@n@error_analysis_cases.naming.gemini.predicted_demand\endcsname{1}
\expandafter\gdef\csname odunum@val@error_analysis_cases.naming.gemini.total\endcsname{3}
\expandafter\gdef\csname odunum@n@error_analysis_cases.naming.gemini.total\endcsname{1}
\expandafter\gdef\csname odunum@val@error_analysis_cases.naming.qwen.detection_correct\endcsname{1}
\expandafter\gdef\csname odunum@n@error_analysis_cases.naming.qwen.detection_correct\endcsname{1}
\expandafter\gdef\csname odunum@val@error_analysis_cases.naming.qwen.hits\endcsname{3}
\expandafter\gdef\csname odunum@n@error_analysis_cases.naming.qwen.hits\endcsname{1}
\expandafter\gdef\csname odunum@val@error_analysis_cases.naming.qwen.kp01\endcsname{1}
\expandafter\gdef\csname odunum@n@error_analysis_cases.naming.qwen.kp01\endcsname{1}
\expandafter\gdef\csname odunum@val@error_analysis_cases.naming.qwen.kp02\endcsname{1}
\expandafter\gdef\csname odunum@n@error_analysis_cases.naming.qwen.kp02\endcsname{1}
\expandafter\gdef\csname odunum@val@error_analysis_cases.naming.qwen.kp03\endcsname{1}
\expandafter\gdef\csname odunum@n@error_analysis_cases.naming.qwen.kp03\endcsname{1}
\expandafter\gdef\csname odunum@val@error_analysis_cases.naming.qwen.m2\endcsname{1.000}
\expandafter\gdef\csname odunum@n@error_analysis_cases.naming.qwen.m2\endcsname{3}
\expandafter\gdef\csname odunum@val@error_analysis_cases.naming.qwen.predicted_demand\endcsname{true}
\expandafter\gdef\csname odunum@n@error_analysis_cases.naming.qwen.predicted_demand\endcsname{1}
\expandafter\gdef\csname odunum@val@error_analysis_cases.naming.qwen.total\endcsname{3}
\expandafter\gdef\csname odunum@n@error_analysis_cases.naming.qwen.total\endcsname{1}
\expandafter\gdef\csname odunum@val@error_analysis_cases.naming.reference_demand\endcsname{true}
\expandafter\gdef\csname odunum@n@error_analysis_cases.naming.reference_demand\endcsname{1}
\expandafter\gdef\csname odunum@val@error_analysis_cases.naming.scene_id\endcsname{avp\_gen\_000353}
\expandafter\gdef\csname odunum@n@error_analysis_cases.naming.scene_id\endcsname{1}
\expandafter\gdef\csname odunum@val@error_analysis_cases.naming.seed.detection_correct\endcsname{1}
\expandafter\gdef\csname odunum@n@error_analysis_cases.naming.seed.detection_correct\endcsname{1}
\expandafter\gdef\csname odunum@val@error_analysis_cases.naming.seed.hits\endcsname{0}
\expandafter\gdef\csname odunum@n@error_analysis_cases.naming.seed.hits\endcsname{1}
\expandafter\gdef\csname odunum@val@error_analysis_cases.naming.seed.kp01\endcsname{0}
\expandafter\gdef\csname odunum@n@error_analysis_cases.naming.seed.kp01\endcsname{1}
\expandafter\gdef\csname odunum@val@error_analysis_cases.naming.seed.kp02\endcsname{0}
\expandafter\gdef\csname odunum@n@error_analysis_cases.naming.seed.kp02\endcsname{1}
\expandafter\gdef\csname odunum@val@error_analysis_cases.naming.seed.kp03\endcsname{0}
\expandafter\gdef\csname odunum@n@error_analysis_cases.naming.seed.kp03\endcsname{1}
\expandafter\gdef\csname odunum@val@error_analysis_cases.naming.seed.m2\endcsname{0.000}
\expandafter\gdef\csname odunum@n@error_analysis_cases.naming.seed.m2\endcsname{3}
\expandafter\gdef\csname odunum@val@error_analysis_cases.naming.seed.predicted_demand\endcsname{true}
\expandafter\gdef\csname odunum@n@error_analysis_cases.naming.seed.predicted_demand\endcsname{1}
\expandafter\gdef\csname odunum@val@error_analysis_cases.naming.seed.total\endcsname{3}
\expandafter\gdef\csname odunum@n@error_analysis_cases.naming.seed.total\endcsname{1}
\expandafter\gdef\csname odunum@val@error_analysis_cases.negative.gemini.detection_correct\endcsname{0}
\expandafter\gdef\csname odunum@n@error_analysis_cases.negative.gemini.detection_correct\endcsname{1}
\expandafter\gdef\csname odunum@val@error_analysis_cases.negative.gemini.m2_status\endcsname{not applicable}
\expandafter\gdef\csname odunum@n@error_analysis_cases.negative.gemini.m2_status\endcsname{1}
\expandafter\gdef\csname odunum@val@error_analysis_cases.negative.gemini.predicted_demand\endcsname{true}
\expandafter\gdef\csname odunum@n@error_analysis_cases.negative.gemini.predicted_demand\endcsname{1}
\expandafter\gdef\csname odunum@val@error_analysis_cases.negative.qwen.detection_correct\endcsname{0}
\expandafter\gdef\csname odunum@n@error_analysis_cases.negative.qwen.detection_correct\endcsname{1}
\expandafter\gdef\csname odunum@val@error_analysis_cases.negative.qwen.m2_status\endcsname{not applicable}
\expandafter\gdef\csname odunum@n@error_analysis_cases.negative.qwen.m2_status\endcsname{1}
\expandafter\gdef\csname odunum@val@error_analysis_cases.negative.qwen.predicted_demand\endcsname{true}
\expandafter\gdef\csname odunum@n@error_analysis_cases.negative.qwen.predicted_demand\endcsname{1}
\expandafter\gdef\csname odunum@val@error_analysis_cases.negative.reference_demand\endcsname{false}
\expandafter\gdef\csname odunum@n@error_analysis_cases.negative.reference_demand\endcsname{1}
\expandafter\gdef\csname odunum@val@error_analysis_cases.negative.scene_id\endcsname{avn\_gen\_000103}
\expandafter\gdef\csname odunum@n@error_analysis_cases.negative.scene_id\endcsname{1}
\expandafter\gdef\csname odunum@val@error_analysis_cases.negative.seed.detection_correct\endcsname{0}
\expandafter\gdef\csname odunum@n@error_analysis_cases.negative.seed.detection_correct\endcsname{1}
\expandafter\gdef\csname odunum@val@error_analysis_cases.negative.seed.m2_status\endcsname{not applicable}
\expandafter\gdef\csname odunum@n@error_analysis_cases.negative.seed.m2_status\endcsname{1}
\expandafter\gdef\csname odunum@val@error_analysis_cases.negative.seed.predicted_demand\endcsname{true}
\expandafter\gdef\csname odunum@n@error_analysis_cases.negative.seed.predicted_demand\endcsname{1}
\expandafter\gdef\csname odunum@val@error_analysis_cases.positive.gemini.detection_correct\endcsname{1}
\expandafter\gdef\csname odunum@n@error_analysis_cases.positive.gemini.detection_correct\endcsname{1}
\expandafter\gdef\csname odunum@val@error_analysis_cases.positive.gemini.hits\endcsname{3}
\expandafter\gdef\csname odunum@n@error_analysis_cases.positive.gemini.hits\endcsname{1}
\expandafter\gdef\csname odunum@val@error_analysis_cases.positive.gemini.kp01\endcsname{1}
\expandafter\gdef\csname odunum@n@error_analysis_cases.positive.gemini.kp01\endcsname{1}
\expandafter\gdef\csname odunum@val@error_analysis_cases.positive.gemini.kp02\endcsname{1}
\expandafter\gdef\csname odunum@n@error_analysis_cases.positive.gemini.kp02\endcsname{1}
\expandafter\gdef\csname odunum@val@error_analysis_cases.positive.gemini.kp03\endcsname{0}
\expandafter\gdef\csname odunum@n@error_analysis_cases.positive.gemini.kp03\endcsname{1}
\expandafter\gdef\csname odunum@val@error_analysis_cases.positive.gemini.kp04\endcsname{1}
\expandafter\gdef\csname odunum@n@error_analysis_cases.positive.gemini.kp04\endcsname{1}
\expandafter\gdef\csname odunum@val@error_analysis_cases.positive.gemini.m2\endcsname{0.750}
\expandafter\gdef\csname odunum@n@error_analysis_cases.positive.gemini.m2\endcsname{4}
\expandafter\gdef\csname odunum@val@error_analysis_cases.positive.gemini.predicted_demand\endcsname{true}
\expandafter\gdef\csname odunum@n@error_analysis_cases.positive.gemini.predicted_demand\endcsname{1}
\expandafter\gdef\csname odunum@val@error_analysis_cases.positive.gemini.total\endcsname{4}
\expandafter\gdef\csname odunum@n@error_analysis_cases.positive.gemini.total\endcsname{1}
\expandafter\gdef\csname odunum@val@error_analysis_cases.positive.qwen.detection_correct\endcsname{1}
\expandafter\gdef\csname odunum@n@error_analysis_cases.positive.qwen.detection_correct\endcsname{1}
\expandafter\gdef\csname odunum@val@error_analysis_cases.positive.qwen.hits\endcsname{3}
\expandafter\gdef\csname odunum@n@error_analysis_cases.positive.qwen.hits\endcsname{1}
\expandafter\gdef\csname odunum@val@error_analysis_cases.positive.qwen.kp01\endcsname{1}
\expandafter\gdef\csname odunum@n@error_analysis_cases.positive.qwen.kp01\endcsname{1}
\expandafter\gdef\csname odunum@val@error_analysis_cases.positive.qwen.kp02\endcsname{1}
\expandafter\gdef\csname odunum@n@error_analysis_cases.positive.qwen.kp02\endcsname{1}
\expandafter\gdef\csname odunum@val@error_analysis_cases.positive.qwen.kp03\endcsname{0}
\expandafter\gdef\csname odunum@n@error_analysis_cases.positive.qwen.kp03\endcsname{1}
\expandafter\gdef\csname odunum@val@error_analysis_cases.positive.qwen.kp04\endcsname{1}
\expandafter\gdef\csname odunum@n@error_analysis_cases.positive.qwen.kp04\endcsname{1}
\expandafter\gdef\csname odunum@val@error_analysis_cases.positive.qwen.m2\endcsname{0.750}
\expandafter\gdef\csname odunum@n@error_analysis_cases.positive.qwen.m2\endcsname{4}
\expandafter\gdef\csname odunum@val@error_analysis_cases.positive.qwen.predicted_demand\endcsname{true}
\expandafter\gdef\csname odunum@n@error_analysis_cases.positive.qwen.predicted_demand\endcsname{1}
\expandafter\gdef\csname odunum@val@error_analysis_cases.positive.qwen.total\endcsname{4}
\expandafter\gdef\csname odunum@n@error_analysis_cases.positive.qwen.total\endcsname{1}
\expandafter\gdef\csname odunum@val@error_analysis_cases.positive.reference_demand\endcsname{true}
\expandafter\gdef\csname odunum@n@error_analysis_cases.positive.reference_demand\endcsname{1}
\expandafter\gdef\csname odunum@val@error_analysis_cases.positive.scene_id\endcsname{avp\_gen\_000272}
\expandafter\gdef\csname odunum@n@error_analysis_cases.positive.scene_id\endcsname{1}
\expandafter\gdef\csname odunum@val@error_analysis_cases.positive.seed.detection_correct\endcsname{1}
\expandafter\gdef\csname odunum@n@error_analysis_cases.positive.seed.detection_correct\endcsname{1}
\expandafter\gdef\csname odunum@val@error_analysis_cases.positive.seed.hits\endcsname{3}
\expandafter\gdef\csname odunum@n@error_analysis_cases.positive.seed.hits\endcsname{1}
\expandafter\gdef\csname odunum@val@error_analysis_cases.positive.seed.kp01\endcsname{1}
\expandafter\gdef\csname odunum@n@error_analysis_cases.positive.seed.kp01\endcsname{1}
\expandafter\gdef\csname odunum@val@error_analysis_cases.positive.seed.kp02\endcsname{1}
\expandafter\gdef\csname odunum@n@error_analysis_cases.positive.seed.kp02\endcsname{1}
\expandafter\gdef\csname odunum@val@error_analysis_cases.positive.seed.kp03\endcsname{0}
\expandafter\gdef\csname odunum@n@error_analysis_cases.positive.seed.kp03\endcsname{1}
\expandafter\gdef\csname odunum@val@error_analysis_cases.positive.seed.kp04\endcsname{1}
\expandafter\gdef\csname odunum@n@error_analysis_cases.positive.seed.kp04\endcsname{1}
\expandafter\gdef\csname odunum@val@error_analysis_cases.positive.seed.m2\endcsname{0.750}
\expandafter\gdef\csname odunum@n@error_analysis_cases.positive.seed.m2\endcsname{4}
\expandafter\gdef\csname odunum@val@error_analysis_cases.positive.seed.predicted_demand\endcsname{true}
\expandafter\gdef\csname odunum@n@error_analysis_cases.positive.seed.predicted_demand\endcsname{1}
\expandafter\gdef\csname odunum@val@error_analysis_cases.positive.seed.total\endcsname{4}
\expandafter\gdef\csname odunum@n@error_analysis_cases.positive.seed.total\endcsname{1}
\expandafter\gdef\csname odunum@val@external_correlation.pop.csv_rows_with_value.n\endcsname{0}
\expandafter\gdef\csname odunum@n@external_correlation.pop.csv_rows_with_value.n\endcsname{1}
\expandafter\gdef\csname odunum@val@external_correlation.pop.unmatched_model_tags.n\endcsname{0}
\expandafter\gdef\csname odunum@n@external_correlation.pop.unmatched_model_tags.n\endcsname{1}
\expandafter\gdef\csname odunum@val@format_compliance.copy.baichuan_omni.any\endcsname{0.054}
\expandafter\gdef\csname odunum@n@format_compliance.copy.baichuan_omni.any\endcsname{1\,021}
\expandafter\gdef\csname odunum@ci@format_compliance.copy.baichuan_omni.any\endcsname{[0.042, 0.069]}
\expandafter\gdef\csname odunum@val@format_compliance.copy.baichuan_omni.example_context\endcsname{0.016}
\expandafter\gdef\csname odunum@n@format_compliance.copy.baichuan_omni.example_context\endcsname{1\,021}
\expandafter\gdef\csname odunum@val@format_compliance.copy.baichuan_omni.example_timestamps\endcsname{0.007}
\expandafter\gdef\csname odunum@n@format_compliance.copy.baichuan_omni.example_timestamps\endcsname{1\,021}
\expandafter\gdef\csname odunum@val@format_compliance.copy.baichuan_omni.placeholder_intent\endcsname{0.006}
\expandafter\gdef\csname odunum@n@format_compliance.copy.baichuan_omni.placeholder_intent\endcsname{1\,021}
\expandafter\gdef\csname odunum@val@format_compliance.copy.baichuan_omni.placeholder_transcript\endcsname{0.041}
\expandafter\gdef\csname odunum@n@format_compliance.copy.baichuan_omni.placeholder_transcript\endcsname{1\,021}
\expandafter\gdef\csname odunum@val@format_compliance.copy.cascade_asr.any\endcsname{0.000}
\expandafter\gdef\csname odunum@n@format_compliance.copy.cascade_asr.any\endcsname{2\,078}
\expandafter\gdef\csname odunum@ci@format_compliance.copy.cascade_asr.any\endcsname{[0.000, 0.002]}
\expandafter\gdef\csname odunum@val@format_compliance.copy.cascade_asr.example_context\endcsname{0.000}
\expandafter\gdef\csname odunum@n@format_compliance.copy.cascade_asr.example_context\endcsname{2\,078}
\expandafter\gdef\csname odunum@val@format_compliance.copy.cascade_asr.example_timestamps\endcsname{0.000}
\expandafter\gdef\csname odunum@n@format_compliance.copy.cascade_asr.example_timestamps\endcsname{2\,078}
\expandafter\gdef\csname odunum@val@format_compliance.copy.cascade_asr.placeholder_intent\endcsname{0.000}
\expandafter\gdef\csname odunum@n@format_compliance.copy.cascade_asr.placeholder_intent\endcsname{2\,078}
\expandafter\gdef\csname odunum@val@format_compliance.copy.cascade_asr.placeholder_transcript\endcsname{0.000}
\expandafter\gdef\csname odunum@n@format_compliance.copy.cascade_asr.placeholder_transcript\endcsname{2\,078}
\expandafter\gdef\csname odunum@val@format_compliance.copy.cascade_caption.any\endcsname{0.000}
\expandafter\gdef\csname odunum@n@format_compliance.copy.cascade_caption.any\endcsname{2\,078}
\expandafter\gdef\csname odunum@ci@format_compliance.copy.cascade_caption.any\endcsname{[0.000, 0.002]}
\expandafter\gdef\csname odunum@val@format_compliance.copy.cascade_caption.example_context\endcsname{0.000}
\expandafter\gdef\csname odunum@n@format_compliance.copy.cascade_caption.example_context\endcsname{2\,078}
\expandafter\gdef\csname odunum@val@format_compliance.copy.cascade_caption.example_timestamps\endcsname{0.000}
\expandafter\gdef\csname odunum@n@format_compliance.copy.cascade_caption.example_timestamps\endcsname{2\,078}
\expandafter\gdef\csname odunum@val@format_compliance.copy.cascade_caption.placeholder_intent\endcsname{0.000}
\expandafter\gdef\csname odunum@n@format_compliance.copy.cascade_caption.placeholder_intent\endcsname{2\,078}
\expandafter\gdef\csname odunum@val@format_compliance.copy.cascade_caption.placeholder_transcript\endcsname{0.000}
\expandafter\gdef\csname odunum@n@format_compliance.copy.cascade_caption.placeholder_transcript\endcsname{2\,078}
\expandafter\gdef\csname odunum@val@format_compliance.copy.gemini.any\endcsname{0.000}
\expandafter\gdef\csname odunum@n@format_compliance.copy.gemini.any\endcsname{2\,078}
\expandafter\gdef\csname odunum@ci@format_compliance.copy.gemini.any\endcsname{[0.000, 0.002]}
\expandafter\gdef\csname odunum@val@format_compliance.copy.gemini.example_context\endcsname{0.000}
\expandafter\gdef\csname odunum@n@format_compliance.copy.gemini.example_context\endcsname{2\,078}
\expandafter\gdef\csname odunum@val@format_compliance.copy.gemini.example_timestamps\endcsname{0.000}
\expandafter\gdef\csname odunum@n@format_compliance.copy.gemini.example_timestamps\endcsname{2\,078}
\expandafter\gdef\csname odunum@val@format_compliance.copy.gemini.placeholder_intent\endcsname{0.000}
\expandafter\gdef\csname odunum@n@format_compliance.copy.gemini.placeholder_intent\endcsname{2\,078}
\expandafter\gdef\csname odunum@val@format_compliance.copy.gemini.placeholder_transcript\endcsname{0.000}
\expandafter\gdef\csname odunum@n@format_compliance.copy.gemini.placeholder_transcript\endcsname{2\,078}
\expandafter\gdef\csname odunum@val@format_compliance.copy.gemini35_flash_lite.any\endcsname{0.000}
\expandafter\gdef\csname odunum@n@format_compliance.copy.gemini35_flash_lite.any\endcsname{2\,078}
\expandafter\gdef\csname odunum@ci@format_compliance.copy.gemini35_flash_lite.any\endcsname{[0.000, 0.002]}
\expandafter\gdef\csname odunum@val@format_compliance.copy.gemini35_flash_lite.example_context\endcsname{0.000}
\expandafter\gdef\csname odunum@n@format_compliance.copy.gemini35_flash_lite.example_context\endcsname{2\,078}
\expandafter\gdef\csname odunum@val@format_compliance.copy.gemini35_flash_lite.example_timestamps\endcsname{0.000}
\expandafter\gdef\csname odunum@n@format_compliance.copy.gemini35_flash_lite.example_timestamps\endcsname{2\,078}
\expandafter\gdef\csname odunum@val@format_compliance.copy.gemini35_flash_lite.placeholder_intent\endcsname{0.000}
\expandafter\gdef\csname odunum@n@format_compliance.copy.gemini35_flash_lite.placeholder_intent\endcsname{2\,078}
\expandafter\gdef\csname odunum@val@format_compliance.copy.gemini35_flash_lite.placeholder_transcript\endcsname{0.000}
\expandafter\gdef\csname odunum@n@format_compliance.copy.gemini35_flash_lite.placeholder_transcript\endcsname{2\,078}
\expandafter\gdef\csname odunum@val@format_compliance.copy.gemini37_flash.any\endcsname{0.000}
\expandafter\gdef\csname odunum@n@format_compliance.copy.gemini37_flash.any\endcsname{2\,067}
\expandafter\gdef\csname odunum@ci@format_compliance.copy.gemini37_flash.any\endcsname{[0.000, 0.002]}
\expandafter\gdef\csname odunum@val@format_compliance.copy.gemini37_flash.example_context\endcsname{0.000}
\expandafter\gdef\csname odunum@n@format_compliance.copy.gemini37_flash.example_context\endcsname{2\,067}
\expandafter\gdef\csname odunum@val@format_compliance.copy.gemini37_flash.example_timestamps\endcsname{0.000}
\expandafter\gdef\csname odunum@n@format_compliance.copy.gemini37_flash.example_timestamps\endcsname{2\,067}
\expandafter\gdef\csname odunum@val@format_compliance.copy.gemini37_flash.placeholder_intent\endcsname{0.000}
\expandafter\gdef\csname odunum@n@format_compliance.copy.gemini37_flash.placeholder_intent\endcsname{2\,067}
\expandafter\gdef\csname odunum@val@format_compliance.copy.gemini37_flash.placeholder_transcript\endcsname{0.000}
\expandafter\gdef\csname odunum@n@format_compliance.copy.gemini37_flash.placeholder_transcript\endcsname{2\,067}
\expandafter\gdef\csname odunum@val@format_compliance.copy.humanomni_v2.any\endcsname{0.001}
\expandafter\gdef\csname odunum@n@format_compliance.copy.humanomni_v2.any\endcsname{1\,111}
\expandafter\gdef\csname odunum@ci@format_compliance.copy.humanomni_v2.any\endcsname{[1.6\ensuremath{\times 10^{-4}}, 0.005]}
\expandafter\gdef\csname odunum@val@format_compliance.copy.humanomni_v2.example_context\endcsname{0.000}
\expandafter\gdef\csname odunum@n@format_compliance.copy.humanomni_v2.example_context\endcsname{1\,111}
\expandafter\gdef\csname odunum@val@format_compliance.copy.humanomni_v2.example_timestamps\endcsname{0.001}
\expandafter\gdef\csname odunum@n@format_compliance.copy.humanomni_v2.example_timestamps\endcsname{1\,111}
\expandafter\gdef\csname odunum@val@format_compliance.copy.humanomni_v2.placeholder_intent\endcsname{0.000}
\expandafter\gdef\csname odunum@n@format_compliance.copy.humanomni_v2.placeholder_intent\endcsname{1\,111}
\expandafter\gdef\csname odunum@val@format_compliance.copy.humanomni_v2.placeholder_transcript\endcsname{0.000}
\expandafter\gdef\csname odunum@n@format_compliance.copy.humanomni_v2.placeholder_transcript\endcsname{1\,111}
\expandafter\gdef\csname odunum@val@format_compliance.copy.ming.any\endcsname{0.026}
\expandafter\gdef\csname odunum@n@format_compliance.copy.ming.any\endcsname{2\,078}
\expandafter\gdef\csname odunum@ci@format_compliance.copy.ming.any\endcsname{[0.020, 0.034]}
\expandafter\gdef\csname odunum@val@format_compliance.copy.ming.example_context\endcsname{0.001}
\expandafter\gdef\csname odunum@n@format_compliance.copy.ming.example_context\endcsname{2\,078}
\expandafter\gdef\csname odunum@val@format_compliance.copy.ming.example_timestamps\endcsname{0.026}
\expandafter\gdef\csname odunum@n@format_compliance.copy.ming.example_timestamps\endcsname{2\,078}
\expandafter\gdef\csname odunum@val@format_compliance.copy.ming.placeholder_intent\endcsname{0.000}
\expandafter\gdef\csname odunum@n@format_compliance.copy.ming.placeholder_intent\endcsname{2\,078}
\expandafter\gdef\csname odunum@val@format_compliance.copy.ming.placeholder_transcript\endcsname{0.001}
\expandafter\gdef\csname odunum@n@format_compliance.copy.ming.placeholder_transcript\endcsname{2\,078}
\expandafter\gdef\csname odunum@val@format_compliance.copy.minicpm_o.any\endcsname{0.100}
\expandafter\gdef\csname odunum@n@format_compliance.copy.minicpm_o.any\endcsname{2\,075}
\expandafter\gdef\csname odunum@ci@format_compliance.copy.minicpm_o.any\endcsname{[0.088, 0.114]}
\expandafter\gdef\csname odunum@val@format_compliance.copy.minicpm_o.example_context\endcsname{0.003}
\expandafter\gdef\csname odunum@n@format_compliance.copy.minicpm_o.example_context\endcsname{2\,075}
\expandafter\gdef\csname odunum@val@format_compliance.copy.minicpm_o.example_timestamps\endcsname{0.100}
\expandafter\gdef\csname odunum@n@format_compliance.copy.minicpm_o.example_timestamps\endcsname{2\,075}
\expandafter\gdef\csname odunum@val@format_compliance.copy.minicpm_o.placeholder_intent\endcsname{0.000}
\expandafter\gdef\csname odunum@n@format_compliance.copy.minicpm_o.placeholder_intent\endcsname{2\,075}
\expandafter\gdef\csname odunum@val@format_compliance.copy.minicpm_o.placeholder_transcript\endcsname{0.000}
\expandafter\gdef\csname odunum@n@format_compliance.copy.minicpm_o.placeholder_transcript\endcsname{2\,075}
\expandafter\gdef\csname odunum@val@format_compliance.copy.nemotron.any\endcsname{0.004}
\expandafter\gdef\csname odunum@n@format_compliance.copy.nemotron.any\endcsname{2\,078}
\expandafter\gdef\csname odunum@ci@format_compliance.copy.nemotron.any\endcsname{[0.002, 0.008]}
\expandafter\gdef\csname odunum@val@format_compliance.copy.nemotron.example_context\endcsname{0.000}
\expandafter\gdef\csname odunum@n@format_compliance.copy.nemotron.example_context\endcsname{2\,078}
\expandafter\gdef\csname odunum@val@format_compliance.copy.nemotron.example_timestamps\endcsname{0.004}
\expandafter\gdef\csname odunum@n@format_compliance.copy.nemotron.example_timestamps\endcsname{2\,078}
\expandafter\gdef\csname odunum@val@format_compliance.copy.nemotron.placeholder_intent\endcsname{0.000}
\expandafter\gdef\csname odunum@n@format_compliance.copy.nemotron.placeholder_intent\endcsname{2\,078}
\expandafter\gdef\csname odunum@val@format_compliance.copy.nemotron.placeholder_transcript\endcsname{0.000}
\expandafter\gdef\csname odunum@n@format_compliance.copy.nemotron.placeholder_transcript\endcsname{2\,078}
\expandafter\gdef\csname odunum@val@format_compliance.copy.qwen25_omni.any\endcsname{0.148}
\expandafter\gdef\csname odunum@n@format_compliance.copy.qwen25_omni.any\endcsname{2\,078}
\expandafter\gdef\csname odunum@ci@format_compliance.copy.qwen25_omni.any\endcsname{[0.133, 0.164]}
\expandafter\gdef\csname odunum@val@format_compliance.copy.qwen25_omni.example_context\endcsname{0.023}
\expandafter\gdef\csname odunum@n@format_compliance.copy.qwen25_omni.example_context\endcsname{2\,078}
\expandafter\gdef\csname odunum@val@format_compliance.copy.qwen25_omni.example_timestamps\endcsname{0.136}
\expandafter\gdef\csname odunum@n@format_compliance.copy.qwen25_omni.example_timestamps\endcsname{2\,078}
\expandafter\gdef\csname odunum@val@format_compliance.copy.qwen25_omni.placeholder_intent\endcsname{0.000}
\expandafter\gdef\csname odunum@n@format_compliance.copy.qwen25_omni.placeholder_intent\endcsname{2\,078}
\expandafter\gdef\csname odunum@val@format_compliance.copy.qwen25_omni.placeholder_transcript\endcsname{0.006}
\expandafter\gdef\csname odunum@n@format_compliance.copy.qwen25_omni.placeholder_transcript\endcsname{2\,078}
\expandafter\gdef\csname odunum@val@format_compliance.copy.qwen3_omni_instruct.any\endcsname{0.371}
\expandafter\gdef\csname odunum@n@format_compliance.copy.qwen3_omni_instruct.any\endcsname{2\,078}
\expandafter\gdef\csname odunum@ci@format_compliance.copy.qwen3_omni_instruct.any\endcsname{[0.351, 0.392]}
\expandafter\gdef\csname odunum@val@format_compliance.copy.qwen3_omni_instruct.example_context\endcsname{0.000}
\expandafter\gdef\csname odunum@n@format_compliance.copy.qwen3_omni_instruct.example_context\endcsname{2\,078}
\expandafter\gdef\csname odunum@val@format_compliance.copy.qwen3_omni_instruct.example_timestamps\endcsname{0.371}
\expandafter\gdef\csname odunum@n@format_compliance.copy.qwen3_omni_instruct.example_timestamps\endcsname{2\,078}
\expandafter\gdef\csname odunum@val@format_compliance.copy.qwen3_omni_instruct.placeholder_intent\endcsname{0.000}
\expandafter\gdef\csname odunum@n@format_compliance.copy.qwen3_omni_instruct.placeholder_intent\endcsname{2\,078}
\expandafter\gdef\csname odunum@val@format_compliance.copy.qwen3_omni_instruct.placeholder_transcript\endcsname{0.000}
\expandafter\gdef\csname odunum@n@format_compliance.copy.qwen3_omni_instruct.placeholder_transcript\endcsname{2\,078}
\expandafter\gdef\csname odunum@val@format_compliance.copy.qwen3_omni_think.any\endcsname{0.010}
\expandafter\gdef\csname odunum@n@format_compliance.copy.qwen3_omni_think.any\endcsname{2\,078}
\expandafter\gdef\csname odunum@ci@format_compliance.copy.qwen3_omni_think.any\endcsname{[0.007, 0.015]}
\expandafter\gdef\csname odunum@val@format_compliance.copy.qwen3_omni_think.example_context\endcsname{0.000}
\expandafter\gdef\csname odunum@n@format_compliance.copy.qwen3_omni_think.example_context\endcsname{2\,078}
\expandafter\gdef\csname odunum@val@format_compliance.copy.qwen3_omni_think.example_timestamps\endcsname{0.010}
\expandafter\gdef\csname odunum@n@format_compliance.copy.qwen3_omni_think.example_timestamps\endcsname{2\,078}
\expandafter\gdef\csname odunum@val@format_compliance.copy.qwen3_omni_think.placeholder_intent\endcsname{0.000}
\expandafter\gdef\csname odunum@n@format_compliance.copy.qwen3_omni_think.placeholder_intent\endcsname{2\,078}
\expandafter\gdef\csname odunum@val@format_compliance.copy.qwen3_omni_think.placeholder_transcript\endcsname{0.000}
\expandafter\gdef\csname odunum@n@format_compliance.copy.qwen3_omni_think.placeholder_transcript\endcsname{2\,078}
\expandafter\gdef\csname odunum@val@format_compliance.copy.qwen_plus.any\endcsname{0.000}
\expandafter\gdef\csname odunum@n@format_compliance.copy.qwen_plus.any\endcsname{2\,077}
\expandafter\gdef\csname odunum@ci@format_compliance.copy.qwen_plus.any\endcsname{[0.000, 0.002]}
\expandafter\gdef\csname odunum@val@format_compliance.copy.qwen_plus.example_context\endcsname{0.000}
\expandafter\gdef\csname odunum@n@format_compliance.copy.qwen_plus.example_context\endcsname{2\,077}
\expandafter\gdef\csname odunum@val@format_compliance.copy.qwen_plus.example_timestamps\endcsname{0.000}
\expandafter\gdef\csname odunum@n@format_compliance.copy.qwen_plus.example_timestamps\endcsname{2\,077}
\expandafter\gdef\csname odunum@val@format_compliance.copy.qwen_plus.placeholder_intent\endcsname{0.000}
\expandafter\gdef\csname odunum@n@format_compliance.copy.qwen_plus.placeholder_intent\endcsname{2\,077}
\expandafter\gdef\csname odunum@val@format_compliance.copy.qwen_plus.placeholder_transcript\endcsname{0.000}
\expandafter\gdef\csname odunum@n@format_compliance.copy.qwen_plus.placeholder_transcript\endcsname{2\,077}
\expandafter\gdef\csname odunum@val@format_compliance.copy.salmonn2_7b.any\endcsname{0.105}
\expandafter\gdef\csname odunum@n@format_compliance.copy.salmonn2_7b.any\endcsname{2\,058}
\expandafter\gdef\csname odunum@ci@format_compliance.copy.salmonn2_7b.any\endcsname{[0.092, 0.119]}
\expandafter\gdef\csname odunum@val@format_compliance.copy.salmonn2_7b.example_context\endcsname{0.005}
\expandafter\gdef\csname odunum@n@format_compliance.copy.salmonn2_7b.example_context\endcsname{2\,058}
\expandafter\gdef\csname odunum@val@format_compliance.copy.salmonn2_7b.example_timestamps\endcsname{0.104}
\expandafter\gdef\csname odunum@n@format_compliance.copy.salmonn2_7b.example_timestamps\endcsname{2\,058}
\expandafter\gdef\csname odunum@val@format_compliance.copy.salmonn2_7b.placeholder_intent\endcsname{0.000}
\expandafter\gdef\csname odunum@n@format_compliance.copy.salmonn2_7b.placeholder_intent\endcsname{2\,058}
\expandafter\gdef\csname odunum@val@format_compliance.copy.salmonn2_7b.placeholder_transcript\endcsname{0.001}
\expandafter\gdef\csname odunum@n@format_compliance.copy.salmonn2_7b.placeholder_transcript\endcsname{2\,058}
\expandafter\gdef\csname odunum@val@format_compliance.copy.seed.any\endcsname{0.000}
\expandafter\gdef\csname odunum@n@format_compliance.copy.seed.any\endcsname{2\,078}
\expandafter\gdef\csname odunum@ci@format_compliance.copy.seed.any\endcsname{[0.000, 0.002]}
\expandafter\gdef\csname odunum@val@format_compliance.copy.seed.example_context\endcsname{0.000}
\expandafter\gdef\csname odunum@n@format_compliance.copy.seed.example_context\endcsname{2\,078}
\expandafter\gdef\csname odunum@val@format_compliance.copy.seed.example_timestamps\endcsname{0.000}
\expandafter\gdef\csname odunum@n@format_compliance.copy.seed.example_timestamps\endcsname{2\,078}
\expandafter\gdef\csname odunum@val@format_compliance.copy.seed.placeholder_intent\endcsname{0.000}
\expandafter\gdef\csname odunum@n@format_compliance.copy.seed.placeholder_intent\endcsname{2\,078}
\expandafter\gdef\csname odunum@val@format_compliance.copy.seed.placeholder_transcript\endcsname{0.000}
\expandafter\gdef\csname odunum@n@format_compliance.copy.seed.placeholder_transcript\endcsname{2\,078}
\expandafter\gdef\csname odunum@val@format_compliance.copy.videollama2.any\endcsname{0.953}
\expandafter\gdef\csname odunum@n@format_compliance.copy.videollama2.any\endcsname{1\,605}
\expandafter\gdef\csname odunum@ci@format_compliance.copy.videollama2.any\endcsname{[0.942, 0.963]}
\expandafter\gdef\csname odunum@val@format_compliance.copy.videollama2.example_context\endcsname{0.913}
\expandafter\gdef\csname odunum@n@format_compliance.copy.videollama2.example_context\endcsname{1\,605}
\expandafter\gdef\csname odunum@val@format_compliance.copy.videollama2.example_timestamps\endcsname{0.946}
\expandafter\gdef\csname odunum@n@format_compliance.copy.videollama2.example_timestamps\endcsname{1\,605}
\expandafter\gdef\csname odunum@val@format_compliance.copy.videollama2.placeholder_intent\endcsname{0.881}
\expandafter\gdef\csname odunum@n@format_compliance.copy.videollama2.placeholder_intent\endcsname{1\,605}
\expandafter\gdef\csname odunum@val@format_compliance.copy.videollama2.placeholder_transcript\endcsname{0.893}
\expandafter\gdef\csname odunum@n@format_compliance.copy.videollama2.placeholder_transcript\endcsname{1\,605}
\expandafter\gdef\csname odunum@val@format_compliance.copy.vita.any\endcsname{0.006}
\expandafter\gdef\csname odunum@n@format_compliance.copy.vita.any\endcsname{1\,517}
\expandafter\gdef\csname odunum@ci@format_compliance.copy.vita.any\endcsname{[0.003, 0.011]}
\expandafter\gdef\csname odunum@val@format_compliance.copy.vita.example_context\endcsname{0.002}
\expandafter\gdef\csname odunum@n@format_compliance.copy.vita.example_context\endcsname{1\,517}
\expandafter\gdef\csname odunum@val@format_compliance.copy.vita.example_timestamps\endcsname{0.002}
\expandafter\gdef\csname odunum@n@format_compliance.copy.vita.example_timestamps\endcsname{1\,517}
\expandafter\gdef\csname odunum@val@format_compliance.copy.vita.placeholder_intent\endcsname{0.003}
\expandafter\gdef\csname odunum@n@format_compliance.copy.vita.placeholder_intent\endcsname{1\,517}
\expandafter\gdef\csname odunum@val@format_compliance.copy.vita.placeholder_transcript\endcsname{0.003}
\expandafter\gdef\csname odunum@n@format_compliance.copy.vita.placeholder_transcript\endcsname{1\,517}
\expandafter\gdef\csname odunum@val@format_compliance.group_mean.coverage_too_low.copy_any\endcsname{0.254}
\expandafter\gdef\csname odunum@n@format_compliance.group_mean.coverage_too_low.copy_any\endcsname{4}
\expandafter\gdef\csname odunum@ci@format_compliance.group_mean.coverage_too_low.copy_any\endcsname{[0.003, 0.716]}
\expandafter\gdef\csname odunum@val@format_compliance.group_mean.coverage_too_low.schema_ok\endcsname{0.765}
\expandafter\gdef\csname odunum@n@format_compliance.group_mean.coverage_too_low.schema_ok\endcsname{4}
\expandafter\gdef\csname odunum@ci@format_compliance.group_mean.coverage_too_low.schema_ok\endcsname{[0.302, 0.999]}
\expandafter\gdef\csname odunum@val@format_compliance.group_mean.panel.copy_any\endcsname{0.059}
\expandafter\gdef\csname odunum@n@format_compliance.group_mean.panel.copy_any\endcsname{13}
\expandafter\gdef\csname odunum@ci@format_compliance.group_mean.panel.copy_any\endcsname{[0.012, 0.125]}
\expandafter\gdef\csname odunum@val@format_compliance.group_mean.panel.schema_ok\endcsname{0.974}
\expandafter\gdef\csname odunum@n@format_compliance.group_mean.panel.schema_ok\endcsname{13}
\expandafter\gdef\csname odunum@ci@format_compliance.group_mean.panel.schema_ok\endcsname{[0.933, 0.998]}
\expandafter\gdef\csname odunum@val@format_compliance.parse.baichuan_omni.harness_pred_ok\endcsname{1.000}
\expandafter\gdef\csname odunum@n@format_compliance.parse.baichuan_omni.harness_pred_ok\endcsname{1\,021}
\expandafter\gdef\csname odunum@ci@format_compliance.parse.baichuan_omni.harness_pred_ok\endcsname{[0.996, 1.000]}
\expandafter\gdef\csname odunum@val@format_compliance.parse.baichuan_omni.recovered_beyond_ladder\endcsname{0.001}
\expandafter\gdef\csname odunum@n@format_compliance.parse.baichuan_omni.recovered_beyond_ladder\endcsname{1\,021}
\expandafter\gdef\csname odunum@ci@format_compliance.parse.baichuan_omni.recovered_beyond_ladder\endcsname{[1.7\ensuremath{\times 10^{-4}}, 0.006]}
\expandafter\gdef\csname odunum@val@format_compliance.parse.baichuan_omni.schema_ok\endcsname{0.999}
\expandafter\gdef\csname odunum@n@format_compliance.parse.baichuan_omni.schema_ok\endcsname{1\,021}
\expandafter\gdef\csname odunum@ci@format_compliance.parse.baichuan_omni.schema_ok\endcsname{[0.994, 1.000]}
\expandafter\gdef\csname odunum@val@format_compliance.parse.baichuan_omni.step.unparseable\endcsname{1}
\expandafter\gdef\csname odunum@n@format_compliance.parse.baichuan_omni.step.unparseable\endcsname{1\,021}
\expandafter\gdef\csname odunum@val@format_compliance.parse.baichuan_omni.strict\endcsname{0.999}
\expandafter\gdef\csname odunum@n@format_compliance.parse.baichuan_omni.strict\endcsname{1\,021}
\expandafter\gdef\csname odunum@ci@format_compliance.parse.baichuan_omni.strict\endcsname{[0.994, 1.000]}
\expandafter\gdef\csname odunum@val@format_compliance.parse.cascade_asr.harness_pred_ok\endcsname{1.000}
\expandafter\gdef\csname odunum@n@format_compliance.parse.cascade_asr.harness_pred_ok\endcsname{2\,078}
\expandafter\gdef\csname odunum@ci@format_compliance.parse.cascade_asr.harness_pred_ok\endcsname{[0.998, 1.000]}
\expandafter\gdef\csname odunum@val@format_compliance.parse.cascade_asr.recovered_beyond_ladder\endcsname{0.000}
\expandafter\gdef\csname odunum@n@format_compliance.parse.cascade_asr.recovered_beyond_ladder\endcsname{2\,078}
\expandafter\gdef\csname odunum@ci@format_compliance.parse.cascade_asr.recovered_beyond_ladder\endcsname{[0.000, 0.002]}
\expandafter\gdef\csname odunum@val@format_compliance.parse.cascade_asr.schema_ok\endcsname{1.000}
\expandafter\gdef\csname odunum@n@format_compliance.parse.cascade_asr.schema_ok\endcsname{2\,078}
\expandafter\gdef\csname odunum@ci@format_compliance.parse.cascade_asr.schema_ok\endcsname{[0.998, 1.000]}
\expandafter\gdef\csname odunum@val@format_compliance.parse.cascade_asr.strict\endcsname{1.000}
\expandafter\gdef\csname odunum@n@format_compliance.parse.cascade_asr.strict\endcsname{2\,078}
\expandafter\gdef\csname odunum@ci@format_compliance.parse.cascade_asr.strict\endcsname{[0.998, 1.000]}
\expandafter\gdef\csname odunum@val@format_compliance.parse.cascade_caption.harness_pred_ok\endcsname{1.000}
\expandafter\gdef\csname odunum@n@format_compliance.parse.cascade_caption.harness_pred_ok\endcsname{2\,078}
\expandafter\gdef\csname odunum@ci@format_compliance.parse.cascade_caption.harness_pred_ok\endcsname{[0.998, 1.000]}
\expandafter\gdef\csname odunum@val@format_compliance.parse.cascade_caption.recovered_beyond_ladder\endcsname{0.000}
\expandafter\gdef\csname odunum@n@format_compliance.parse.cascade_caption.recovered_beyond_ladder\endcsname{2\,078}
\expandafter\gdef\csname odunum@ci@format_compliance.parse.cascade_caption.recovered_beyond_ladder\endcsname{[0.000, 0.002]}
\expandafter\gdef\csname odunum@val@format_compliance.parse.cascade_caption.schema_ok\endcsname{1.000}
\expandafter\gdef\csname odunum@n@format_compliance.parse.cascade_caption.schema_ok\endcsname{2\,078}
\expandafter\gdef\csname odunum@ci@format_compliance.parse.cascade_caption.schema_ok\endcsname{[0.998, 1.000]}
\expandafter\gdef\csname odunum@val@format_compliance.parse.cascade_caption.strict\endcsname{1.000}
\expandafter\gdef\csname odunum@n@format_compliance.parse.cascade_caption.strict\endcsname{2\,078}
\expandafter\gdef\csname odunum@ci@format_compliance.parse.cascade_caption.strict\endcsname{[0.998, 1.000]}
\expandafter\gdef\csname odunum@val@format_compliance.parse.gemini.harness_pred_ok\endcsname{1.000}
\expandafter\gdef\csname odunum@n@format_compliance.parse.gemini.harness_pred_ok\endcsname{2\,078}
\expandafter\gdef\csname odunum@ci@format_compliance.parse.gemini.harness_pred_ok\endcsname{[0.998, 1.000]}
\expandafter\gdef\csname odunum@val@format_compliance.parse.gemini.recovered_beyond_ladder\endcsname{0.000}
\expandafter\gdef\csname odunum@n@format_compliance.parse.gemini.recovered_beyond_ladder\endcsname{2\,078}
\expandafter\gdef\csname odunum@ci@format_compliance.parse.gemini.recovered_beyond_ladder\endcsname{[0.000, 0.002]}
\expandafter\gdef\csname odunum@val@format_compliance.parse.gemini.schema_ok\endcsname{1.000}
\expandafter\gdef\csname odunum@n@format_compliance.parse.gemini.schema_ok\endcsname{2\,078}
\expandafter\gdef\csname odunum@ci@format_compliance.parse.gemini.schema_ok\endcsname{[0.998, 1.000]}
\expandafter\gdef\csname odunum@val@format_compliance.parse.gemini.step.fenced\endcsname{1\,105}
\expandafter\gdef\csname odunum@n@format_compliance.parse.gemini.step.fenced\endcsname{2\,078}
\expandafter\gdef\csname odunum@val@format_compliance.parse.gemini.strict\endcsname{0.468}
\expandafter\gdef\csname odunum@n@format_compliance.parse.gemini.strict\endcsname{2\,078}
\expandafter\gdef\csname odunum@ci@format_compliance.parse.gemini.strict\endcsname{[0.447, 0.490]}
\expandafter\gdef\csname odunum@val@format_compliance.parse.gemini35_flash_lite.harness_pred_ok\endcsname{1.000}
\expandafter\gdef\csname odunum@n@format_compliance.parse.gemini35_flash_lite.harness_pred_ok\endcsname{2\,078}
\expandafter\gdef\csname odunum@ci@format_compliance.parse.gemini35_flash_lite.harness_pred_ok\endcsname{[0.998, 1.000]}
\expandafter\gdef\csname odunum@val@format_compliance.parse.gemini35_flash_lite.recovered_beyond_ladder\endcsname{0.009}
\expandafter\gdef\csname odunum@n@format_compliance.parse.gemini35_flash_lite.recovered_beyond_ladder\endcsname{2\,078}
\expandafter\gdef\csname odunum@ci@format_compliance.parse.gemini35_flash_lite.recovered_beyond_ladder\endcsname{[0.005, 0.014]}
\expandafter\gdef\csname odunum@val@format_compliance.parse.gemini35_flash_lite.schema_ok\endcsname{0.991}
\expandafter\gdef\csname odunum@n@format_compliance.parse.gemini35_flash_lite.schema_ok\endcsname{2\,078}
\expandafter\gdef\csname odunum@ci@format_compliance.parse.gemini35_flash_lite.schema_ok\endcsname{[0.986, 0.995]}
\expandafter\gdef\csname odunum@val@format_compliance.parse.gemini35_flash_lite.step.fenced\endcsname{15}
\expandafter\gdef\csname odunum@n@format_compliance.parse.gemini35_flash_lite.step.fenced\endcsname{2\,078}
\expandafter\gdef\csname odunum@val@format_compliance.parse.gemini35_flash_lite.step.unparseable\endcsname{18}
\expandafter\gdef\csname odunum@n@format_compliance.parse.gemini35_flash_lite.step.unparseable\endcsname{2\,078}
\expandafter\gdef\csname odunum@val@format_compliance.parse.gemini35_flash_lite.strict\endcsname{0.984}
\expandafter\gdef\csname odunum@n@format_compliance.parse.gemini35_flash_lite.strict\endcsname{2\,078}
\expandafter\gdef\csname odunum@ci@format_compliance.parse.gemini35_flash_lite.strict\endcsname{[0.978, 0.989]}
\expandafter\gdef\csname odunum@val@format_compliance.parse.gemini37_flash.harness_pred_ok\endcsname{1.000}
\expandafter\gdef\csname odunum@n@format_compliance.parse.gemini37_flash.harness_pred_ok\endcsname{2\,067}
\expandafter\gdef\csname odunum@ci@format_compliance.parse.gemini37_flash.harness_pred_ok\endcsname{[0.998, 1.000]}
\expandafter\gdef\csname odunum@val@format_compliance.parse.gemini37_flash.recovered_beyond_ladder\endcsname{0.000}
\expandafter\gdef\csname odunum@n@format_compliance.parse.gemini37_flash.recovered_beyond_ladder\endcsname{2\,067}
\expandafter\gdef\csname odunum@ci@format_compliance.parse.gemini37_flash.recovered_beyond_ladder\endcsname{[0.000, 0.002]}
\expandafter\gdef\csname odunum@val@format_compliance.parse.gemini37_flash.schema_ok\endcsname{1.000}
\expandafter\gdef\csname odunum@n@format_compliance.parse.gemini37_flash.schema_ok\endcsname{2\,067}
\expandafter\gdef\csname odunum@ci@format_compliance.parse.gemini37_flash.schema_ok\endcsname{[0.998, 1.000]}
\expandafter\gdef\csname odunum@val@format_compliance.parse.gemini37_flash.step.fenced\endcsname{1\,049}
\expandafter\gdef\csname odunum@n@format_compliance.parse.gemini37_flash.step.fenced\endcsname{2\,067}
\expandafter\gdef\csname odunum@val@format_compliance.parse.gemini37_flash.strict\endcsname{0.492}
\expandafter\gdef\csname odunum@n@format_compliance.parse.gemini37_flash.strict\endcsname{2\,067}
\expandafter\gdef\csname odunum@ci@format_compliance.parse.gemini37_flash.strict\endcsname{[0.471, 0.514]}
\expandafter\gdef\csname odunum@val@format_compliance.parse.humanomni_v2.harness_pred_ok\endcsname{0.996}
\expandafter\gdef\csname odunum@n@format_compliance.parse.humanomni_v2.harness_pred_ok\endcsname{1\,111}
\expandafter\gdef\csname odunum@ci@format_compliance.parse.humanomni_v2.harness_pred_ok\endcsname{[0.990, 0.998]}
\expandafter\gdef\csname odunum@val@format_compliance.parse.humanomni_v2.recovered_beyond_ladder\endcsname{0.926}
\expandafter\gdef\csname odunum@n@format_compliance.parse.humanomni_v2.recovered_beyond_ladder\endcsname{1\,111}
\expandafter\gdef\csname odunum@ci@format_compliance.parse.humanomni_v2.recovered_beyond_ladder\endcsname{[0.909, 0.940]}
\expandafter\gdef\csname odunum@val@format_compliance.parse.humanomni_v2.schema_ok\endcsname{0.069}
\expandafter\gdef\csname odunum@n@format_compliance.parse.humanomni_v2.schema_ok\endcsname{1\,111}
\expandafter\gdef\csname odunum@ci@format_compliance.parse.humanomni_v2.schema_ok\endcsname{[0.056, 0.086]}
\expandafter\gdef\csname odunum@val@format_compliance.parse.humanomni_v2.step.brace_span\endcsname{77}
\expandafter\gdef\csname odunum@n@format_compliance.parse.humanomni_v2.step.brace_span\endcsname{1\,111}
\expandafter\gdef\csname odunum@val@format_compliance.parse.humanomni_v2.step.brace_span_bad_schema\endcsname{2}
\expandafter\gdef\csname odunum@n@format_compliance.parse.humanomni_v2.step.brace_span_bad_schema\endcsname{1\,111}
\expandafter\gdef\csname odunum@val@format_compliance.parse.humanomni_v2.step.unparseable\endcsname{1\,032}
\expandafter\gdef\csname odunum@n@format_compliance.parse.humanomni_v2.step.unparseable\endcsname{1\,111}
\expandafter\gdef\csname odunum@val@format_compliance.parse.humanomni_v2.strict\endcsname{0.000}
\expandafter\gdef\csname odunum@n@format_compliance.parse.humanomni_v2.strict\endcsname{1\,111}
\expandafter\gdef\csname odunum@ci@format_compliance.parse.humanomni_v2.strict\endcsname{[0.000, 0.003]}
\expandafter\gdef\csname odunum@val@format_compliance.parse.ming.harness_pred_ok\endcsname{1.000}
\expandafter\gdef\csname odunum@n@format_compliance.parse.ming.harness_pred_ok\endcsname{2\,078}
\expandafter\gdef\csname odunum@ci@format_compliance.parse.ming.harness_pred_ok\endcsname{[0.998, 1.000]}
\expandafter\gdef\csname odunum@val@format_compliance.parse.ming.recovered_beyond_ladder\endcsname{0.001}
\expandafter\gdef\csname odunum@n@format_compliance.parse.ming.recovered_beyond_ladder\endcsname{2\,078}
\expandafter\gdef\csname odunum@ci@format_compliance.parse.ming.recovered_beyond_ladder\endcsname{[8.5\ensuremath{\times 10^{-5}}, 0.003]}
\expandafter\gdef\csname odunum@val@format_compliance.parse.ming.schema_ok\endcsname{1.000}
\expandafter\gdef\csname odunum@n@format_compliance.parse.ming.schema_ok\endcsname{2\,078}
\expandafter\gdef\csname odunum@ci@format_compliance.parse.ming.schema_ok\endcsname{[0.997, 1.000]}
\expandafter\gdef\csname odunum@val@format_compliance.parse.ming.step.fenced\endcsname{18}
\expandafter\gdef\csname odunum@n@format_compliance.parse.ming.step.fenced\endcsname{2\,078}
\expandafter\gdef\csname odunum@val@format_compliance.parse.ming.step.unparseable\endcsname{1}
\expandafter\gdef\csname odunum@n@format_compliance.parse.ming.step.unparseable\endcsname{2\,078}
\expandafter\gdef\csname odunum@val@format_compliance.parse.ming.strict\endcsname{0.991}
\expandafter\gdef\csname odunum@n@format_compliance.parse.ming.strict\endcsname{2\,078}
\expandafter\gdef\csname odunum@ci@format_compliance.parse.ming.strict\endcsname{[0.986, 0.994]}
\expandafter\gdef\csname odunum@val@format_compliance.parse.minicpm_o.harness_pred_ok\endcsname{0.986}
\expandafter\gdef\csname odunum@n@format_compliance.parse.minicpm_o.harness_pred_ok\endcsname{2\,075}
\expandafter\gdef\csname odunum@ci@format_compliance.parse.minicpm_o.harness_pred_ok\endcsname{[0.980, 0.990]}
\expandafter\gdef\csname odunum@val@format_compliance.parse.minicpm_o.recovered_beyond_ladder\endcsname{0.007}
\expandafter\gdef\csname odunum@n@format_compliance.parse.minicpm_o.recovered_beyond_ladder\endcsname{2\,075}
\expandafter\gdef\csname odunum@ci@format_compliance.parse.minicpm_o.recovered_beyond_ladder\endcsname{[0.004, 0.012]}
\expandafter\gdef\csname odunum@val@format_compliance.parse.minicpm_o.schema_ok\endcsname{0.979}
\expandafter\gdef\csname odunum@n@format_compliance.parse.minicpm_o.schema_ok\endcsname{2\,075}
\expandafter\gdef\csname odunum@ci@format_compliance.parse.minicpm_o.schema_ok\endcsname{[0.972, 0.984]}
\expandafter\gdef\csname odunum@val@format_compliance.parse.minicpm_o.step.brace_span\endcsname{189}
\expandafter\gdef\csname odunum@n@format_compliance.parse.minicpm_o.step.brace_span\endcsname{2\,075}
\expandafter\gdef\csname odunum@val@format_compliance.parse.minicpm_o.step.fenced\endcsname{11}
\expandafter\gdef\csname odunum@n@format_compliance.parse.minicpm_o.step.fenced\endcsname{2\,075}
\expandafter\gdef\csname odunum@val@format_compliance.parse.minicpm_o.step.strict_bad_schema\endcsname{5}
\expandafter\gdef\csname odunum@n@format_compliance.parse.minicpm_o.step.strict_bad_schema\endcsname{2\,075}
\expandafter\gdef\csname odunum@val@format_compliance.parse.minicpm_o.step.unparseable\endcsname{39}
\expandafter\gdef\csname odunum@n@format_compliance.parse.minicpm_o.step.unparseable\endcsname{2\,075}
\expandafter\gdef\csname odunum@val@format_compliance.parse.minicpm_o.strict\endcsname{0.882}
\expandafter\gdef\csname odunum@n@format_compliance.parse.minicpm_o.strict\endcsname{2\,075}
\expandafter\gdef\csname odunum@ci@format_compliance.parse.minicpm_o.strict\endcsname{[0.868, 0.896]}
\expandafter\gdef\csname odunum@val@format_compliance.parse.nemotron.harness_pred_ok\endcsname{1.000}
\expandafter\gdef\csname odunum@n@format_compliance.parse.nemotron.harness_pred_ok\endcsname{2\,078}
\expandafter\gdef\csname odunum@ci@format_compliance.parse.nemotron.harness_pred_ok\endcsname{[0.998, 1.000]}
\expandafter\gdef\csname odunum@val@format_compliance.parse.nemotron.recovered_beyond_ladder\endcsname{0.001}
\expandafter\gdef\csname odunum@n@format_compliance.parse.nemotron.recovered_beyond_ladder\endcsname{2\,078}
\expandafter\gdef\csname odunum@ci@format_compliance.parse.nemotron.recovered_beyond_ladder\endcsname{[2.6\ensuremath{\times 10^{-4}}, 0.004]}
\expandafter\gdef\csname odunum@val@format_compliance.parse.nemotron.schema_ok\endcsname{0.999}
\expandafter\gdef\csname odunum@n@format_compliance.parse.nemotron.schema_ok\endcsname{2\,078}
\expandafter\gdef\csname odunum@ci@format_compliance.parse.nemotron.schema_ok\endcsname{[0.996, 1.000]}
\expandafter\gdef\csname odunum@val@format_compliance.parse.nemotron.step.unparseable\endcsname{2}
\expandafter\gdef\csname odunum@n@format_compliance.parse.nemotron.step.unparseable\endcsname{2\,078}
\expandafter\gdef\csname odunum@val@format_compliance.parse.nemotron.strict\endcsname{0.999}
\expandafter\gdef\csname odunum@n@format_compliance.parse.nemotron.strict\endcsname{2\,078}
\expandafter\gdef\csname odunum@ci@format_compliance.parse.nemotron.strict\endcsname{[0.996, 1.000]}
\expandafter\gdef\csname odunum@val@format_compliance.parse.qwen25_omni.harness_pred_ok\endcsname{1.000}
\expandafter\gdef\csname odunum@n@format_compliance.parse.qwen25_omni.harness_pred_ok\endcsname{2\,078}
\expandafter\gdef\csname odunum@ci@format_compliance.parse.qwen25_omni.harness_pred_ok\endcsname{[0.998, 1.000]}
\expandafter\gdef\csname odunum@val@format_compliance.parse.qwen25_omni.recovered_beyond_ladder\endcsname{0.001}
\expandafter\gdef\csname odunum@n@format_compliance.parse.qwen25_omni.recovered_beyond_ladder\endcsname{2\,078}
\expandafter\gdef\csname odunum@ci@format_compliance.parse.qwen25_omni.recovered_beyond_ladder\endcsname{[8.5\ensuremath{\times 10^{-5}}, 0.003]}
\expandafter\gdef\csname odunum@val@format_compliance.parse.qwen25_omni.schema_ok\endcsname{1.000}
\expandafter\gdef\csname odunum@n@format_compliance.parse.qwen25_omni.schema_ok\endcsname{2\,078}
\expandafter\gdef\csname odunum@ci@format_compliance.parse.qwen25_omni.schema_ok\endcsname{[0.997, 1.000]}
\expandafter\gdef\csname odunum@val@format_compliance.parse.qwen25_omni.step.fenced\endcsname{114}
\expandafter\gdef\csname odunum@n@format_compliance.parse.qwen25_omni.step.fenced\endcsname{2\,078}
\expandafter\gdef\csname odunum@val@format_compliance.parse.qwen25_omni.step.unparseable\endcsname{1}
\expandafter\gdef\csname odunum@n@format_compliance.parse.qwen25_omni.step.unparseable\endcsname{2\,078}
\expandafter\gdef\csname odunum@val@format_compliance.parse.qwen25_omni.strict\endcsname{0.945}
\expandafter\gdef\csname odunum@n@format_compliance.parse.qwen25_omni.strict\endcsname{2\,078}
\expandafter\gdef\csname odunum@ci@format_compliance.parse.qwen25_omni.strict\endcsname{[0.934, 0.954]}
\expandafter\gdef\csname odunum@val@format_compliance.parse.qwen3_omni_instruct.harness_pred_ok\endcsname{1.000}
\expandafter\gdef\csname odunum@n@format_compliance.parse.qwen3_omni_instruct.harness_pred_ok\endcsname{2\,078}
\expandafter\gdef\csname odunum@ci@format_compliance.parse.qwen3_omni_instruct.harness_pred_ok\endcsname{[0.998, 1.000]}
\expandafter\gdef\csname odunum@val@format_compliance.parse.qwen3_omni_instruct.recovered_beyond_ladder\endcsname{0.000}
\expandafter\gdef\csname odunum@n@format_compliance.parse.qwen3_omni_instruct.recovered_beyond_ladder\endcsname{2\,078}
\expandafter\gdef\csname odunum@ci@format_compliance.parse.qwen3_omni_instruct.recovered_beyond_ladder\endcsname{[0.000, 0.002]}
\expandafter\gdef\csname odunum@val@format_compliance.parse.qwen3_omni_instruct.schema_ok\endcsname{1.000}
\expandafter\gdef\csname odunum@n@format_compliance.parse.qwen3_omni_instruct.schema_ok\endcsname{2\,078}
\expandafter\gdef\csname odunum@ci@format_compliance.parse.qwen3_omni_instruct.schema_ok\endcsname{[0.998, 1.000]}
\expandafter\gdef\csname odunum@val@format_compliance.parse.qwen3_omni_instruct.step.fenced\endcsname{94}
\expandafter\gdef\csname odunum@n@format_compliance.parse.qwen3_omni_instruct.step.fenced\endcsname{2\,078}
\expandafter\gdef\csname odunum@val@format_compliance.parse.qwen3_omni_instruct.strict\endcsname{0.955}
\expandafter\gdef\csname odunum@n@format_compliance.parse.qwen3_omni_instruct.strict\endcsname{2\,078}
\expandafter\gdef\csname odunum@ci@format_compliance.parse.qwen3_omni_instruct.strict\endcsname{[0.945, 0.963]}
\expandafter\gdef\csname odunum@val@format_compliance.parse.qwen3_omni_think.harness_pred_ok\endcsname{1.000}
\expandafter\gdef\csname odunum@n@format_compliance.parse.qwen3_omni_think.harness_pred_ok\endcsname{2\,078}
\expandafter\gdef\csname odunum@ci@format_compliance.parse.qwen3_omni_think.harness_pred_ok\endcsname{[0.998, 1.000]}
\expandafter\gdef\csname odunum@val@format_compliance.parse.qwen3_omni_think.recovered_beyond_ladder\endcsname{0.000}
\expandafter\gdef\csname odunum@n@format_compliance.parse.qwen3_omni_think.recovered_beyond_ladder\endcsname{2\,078}
\expandafter\gdef\csname odunum@ci@format_compliance.parse.qwen3_omni_think.recovered_beyond_ladder\endcsname{[0.000, 0.002]}
\expandafter\gdef\csname odunum@val@format_compliance.parse.qwen3_omni_think.schema_ok\endcsname{1.000}
\expandafter\gdef\csname odunum@n@format_compliance.parse.qwen3_omni_think.schema_ok\endcsname{2\,078}
\expandafter\gdef\csname odunum@ci@format_compliance.parse.qwen3_omni_think.schema_ok\endcsname{[0.998, 1.000]}
\expandafter\gdef\csname odunum@val@format_compliance.parse.qwen3_omni_think.strict\endcsname{1.000}
\expandafter\gdef\csname odunum@n@format_compliance.parse.qwen3_omni_think.strict\endcsname{2\,078}
\expandafter\gdef\csname odunum@ci@format_compliance.parse.qwen3_omni_think.strict\endcsname{[0.998, 1.000]}
\expandafter\gdef\csname odunum@val@format_compliance.parse.qwen_plus.harness_pred_ok\endcsname{1.000}
\expandafter\gdef\csname odunum@n@format_compliance.parse.qwen_plus.harness_pred_ok\endcsname{2\,077}
\expandafter\gdef\csname odunum@ci@format_compliance.parse.qwen_plus.harness_pred_ok\endcsname{[0.998, 1.000]}
\expandafter\gdef\csname odunum@val@format_compliance.parse.qwen_plus.recovered_beyond_ladder\endcsname{0.254}
\expandafter\gdef\csname odunum@n@format_compliance.parse.qwen_plus.recovered_beyond_ladder\endcsname{2\,077}
\expandafter\gdef\csname odunum@ci@format_compliance.parse.qwen_plus.recovered_beyond_ladder\endcsname{[0.236, 0.273]}
\expandafter\gdef\csname odunum@val@format_compliance.parse.qwen_plus.schema_ok\endcsname{0.746}
\expandafter\gdef\csname odunum@n@format_compliance.parse.qwen_plus.schema_ok\endcsname{2\,077}
\expandafter\gdef\csname odunum@ci@format_compliance.parse.qwen_plus.schema_ok\endcsname{[0.727, 0.764]}
\expandafter\gdef\csname odunum@val@format_compliance.parse.qwen_plus.step.unparseable\endcsname{528}
\expandafter\gdef\csname odunum@n@format_compliance.parse.qwen_plus.step.unparseable\endcsname{2\,077}
\expandafter\gdef\csname odunum@val@format_compliance.parse.qwen_plus.strict\endcsname{0.746}
\expandafter\gdef\csname odunum@n@format_compliance.parse.qwen_plus.strict\endcsname{2\,077}
\expandafter\gdef\csname odunum@ci@format_compliance.parse.qwen_plus.strict\endcsname{[0.727, 0.764]}
\expandafter\gdef\csname odunum@val@format_compliance.parse.salmonn2_7b.harness_pred_ok\endcsname{1.000}
\expandafter\gdef\csname odunum@n@format_compliance.parse.salmonn2_7b.harness_pred_ok\endcsname{2\,058}
\expandafter\gdef\csname odunum@ci@format_compliance.parse.salmonn2_7b.harness_pred_ok\endcsname{[0.997, 1.000]}
\expandafter\gdef\csname odunum@val@format_compliance.parse.salmonn2_7b.recovered_beyond_ladder\endcsname{0.046}
\expandafter\gdef\csname odunum@n@format_compliance.parse.salmonn2_7b.recovered_beyond_ladder\endcsname{2\,058}
\expandafter\gdef\csname odunum@ci@format_compliance.parse.salmonn2_7b.recovered_beyond_ladder\endcsname{[0.037, 0.056]}
\expandafter\gdef\csname odunum@val@format_compliance.parse.salmonn2_7b.schema_ok\endcsname{0.954}
\expandafter\gdef\csname odunum@n@format_compliance.parse.salmonn2_7b.schema_ok\endcsname{2\,058}
\expandafter\gdef\csname odunum@ci@format_compliance.parse.salmonn2_7b.schema_ok\endcsname{[0.944, 0.962]}
\expandafter\gdef\csname odunum@val@format_compliance.parse.salmonn2_7b.step.brace_span\endcsname{1\,581}
\expandafter\gdef\csname odunum@n@format_compliance.parse.salmonn2_7b.step.brace_span\endcsname{2\,058}
\expandafter\gdef\csname odunum@val@format_compliance.parse.salmonn2_7b.step.strict_bad_schema\endcsname{1}
\expandafter\gdef\csname odunum@n@format_compliance.parse.salmonn2_7b.step.strict_bad_schema\endcsname{2\,058}
\expandafter\gdef\csname odunum@val@format_compliance.parse.salmonn2_7b.step.unparseable\endcsname{94}
\expandafter\gdef\csname odunum@n@format_compliance.parse.salmonn2_7b.step.unparseable\endcsname{2\,058}
\expandafter\gdef\csname odunum@val@format_compliance.parse.salmonn2_7b.strict\endcsname{0.186}
\expandafter\gdef\csname odunum@n@format_compliance.parse.salmonn2_7b.strict\endcsname{2\,058}
\expandafter\gdef\csname odunum@ci@format_compliance.parse.salmonn2_7b.strict\endcsname{[0.169, 0.203]}
\expandafter\gdef\csname odunum@val@format_compliance.parse.seed.harness_pred_ok\endcsname{1.000}
\expandafter\gdef\csname odunum@n@format_compliance.parse.seed.harness_pred_ok\endcsname{2\,078}
\expandafter\gdef\csname odunum@ci@format_compliance.parse.seed.harness_pred_ok\endcsname{[0.998, 1.000]}
\expandafter\gdef\csname odunum@val@format_compliance.parse.seed.recovered_beyond_ladder\endcsname{0.001}
\expandafter\gdef\csname odunum@n@format_compliance.parse.seed.recovered_beyond_ladder\endcsname{2\,078}
\expandafter\gdef\csname odunum@ci@format_compliance.parse.seed.recovered_beyond_ladder\endcsname{[4.9\ensuremath{\times 10^{-4}}, 0.004]}
\expandafter\gdef\csname odunum@val@format_compliance.parse.seed.schema_ok\endcsname{0.999}
\expandafter\gdef\csname odunum@n@format_compliance.parse.seed.schema_ok\endcsname{2\,078}
\expandafter\gdef\csname odunum@ci@format_compliance.parse.seed.schema_ok\endcsname{[0.996, 1.000]}
\expandafter\gdef\csname odunum@val@format_compliance.parse.seed.step.unparseable\endcsname{3}
\expandafter\gdef\csname odunum@n@format_compliance.parse.seed.step.unparseable\endcsname{2\,078}
\expandafter\gdef\csname odunum@val@format_compliance.parse.seed.strict\endcsname{0.999}
\expandafter\gdef\csname odunum@n@format_compliance.parse.seed.strict\endcsname{2\,078}
\expandafter\gdef\csname odunum@ci@format_compliance.parse.seed.strict\endcsname{[0.996, 1.000]}
\expandafter\gdef\csname odunum@val@format_compliance.parse.videollama2.harness_pred_ok\endcsname{0.998}
\expandafter\gdef\csname odunum@n@format_compliance.parse.videollama2.harness_pred_ok\endcsname{1\,605}
\expandafter\gdef\csname odunum@ci@format_compliance.parse.videollama2.harness_pred_ok\endcsname{[0.994, 0.999]}
\expandafter\gdef\csname odunum@val@format_compliance.parse.videollama2.recovered_beyond_ladder\endcsname{0.003}
\expandafter\gdef\csname odunum@n@format_compliance.parse.videollama2.recovered_beyond_ladder\endcsname{1\,605}
\expandafter\gdef\csname odunum@ci@format_compliance.parse.videollama2.recovered_beyond_ladder\endcsname{[0.001, 0.007]}
\expandafter\gdef\csname odunum@val@format_compliance.parse.videollama2.schema_ok\endcsname{0.994}
\expandafter\gdef\csname odunum@n@format_compliance.parse.videollama2.schema_ok\endcsname{1\,605}
\expandafter\gdef\csname odunum@ci@format_compliance.parse.videollama2.schema_ok\endcsname{[0.989, 0.997]}
\expandafter\gdef\csname odunum@val@format_compliance.parse.videollama2.step.brace_span\endcsname{450}
\expandafter\gdef\csname odunum@n@format_compliance.parse.videollama2.step.brace_span\endcsname{1\,605}
\expandafter\gdef\csname odunum@val@format_compliance.parse.videollama2.step.brace_span_bad_schema\endcsname{3}
\expandafter\gdef\csname odunum@n@format_compliance.parse.videollama2.step.brace_span_bad_schema\endcsname{1\,605}
\expandafter\gdef\csname odunum@val@format_compliance.parse.videollama2.step.fenced\endcsname{15}
\expandafter\gdef\csname odunum@n@format_compliance.parse.videollama2.step.fenced\endcsname{1\,605}
\expandafter\gdef\csname odunum@val@format_compliance.parse.videollama2.step.unparseable\endcsname{6}
\expandafter\gdef\csname odunum@n@format_compliance.parse.videollama2.step.unparseable\endcsname{1\,605}
\expandafter\gdef\csname odunum@val@format_compliance.parse.videollama2.strict\endcsname{0.705}
\expandafter\gdef\csname odunum@n@format_compliance.parse.videollama2.strict\endcsname{1\,605}
\expandafter\gdef\csname odunum@ci@format_compliance.parse.videollama2.strict\endcsname{[0.682, 0.726]}
\expandafter\gdef\csname odunum@val@format_compliance.parse.vita.harness_pred_ok\endcsname{0.999}
\expandafter\gdef\csname odunum@n@format_compliance.parse.vita.harness_pred_ok\endcsname{1\,517}
\expandafter\gdef\csname odunum@ci@format_compliance.parse.vita.harness_pred_ok\endcsname{[0.996, 1.000]}
\expandafter\gdef\csname odunum@val@format_compliance.parse.vita.recovered_beyond_ladder\endcsname{0.001}
\expandafter\gdef\csname odunum@n@format_compliance.parse.vita.recovered_beyond_ladder\endcsname{1\,517}
\expandafter\gdef\csname odunum@ci@format_compliance.parse.vita.recovered_beyond_ladder\endcsname{[1.2\ensuremath{\times 10^{-4}}, 0.004]}
\expandafter\gdef\csname odunum@val@format_compliance.parse.vita.schema_ok\endcsname{0.999}
\expandafter\gdef\csname odunum@n@format_compliance.parse.vita.schema_ok\endcsname{1\,517}
\expandafter\gdef\csname odunum@ci@format_compliance.parse.vita.schema_ok\endcsname{[0.995, 1.000]}
\expandafter\gdef\csname odunum@val@format_compliance.parse.vita.step.brace_span\endcsname{1\,515}
\expandafter\gdef\csname odunum@n@format_compliance.parse.vita.step.brace_span\endcsname{1\,517}
\expandafter\gdef\csname odunum@val@format_compliance.parse.vita.step.unparseable\endcsname{2}
\expandafter\gdef\csname odunum@n@format_compliance.parse.vita.step.unparseable\endcsname{1\,517}
\expandafter\gdef\csname odunum@val@format_compliance.parse.vita.strict\endcsname{0.000}
\expandafter\gdef\csname odunum@n@format_compliance.parse.vita.strict\endcsname{1\,517}
\expandafter\gdef\csname odunum@ci@format_compliance.parse.vita.strict\endcsname{[0.000, 0.003]}
\expandafter\gdef\csname odunum@val@format_compliance.pop.baichuan_omni.n\endcsname{1\,021}
\expandafter\gdef\csname odunum@n@format_compliance.pop.baichuan_omni.n\endcsname{2\,078}
\expandafter\gdef\csname odunum@val@format_compliance.pop.cascade_asr.n\endcsname{2\,078}
\expandafter\gdef\csname odunum@n@format_compliance.pop.cascade_asr.n\endcsname{2\,078}
\expandafter\gdef\csname odunum@val@format_compliance.pop.cascade_caption.n\endcsname{2\,078}
\expandafter\gdef\csname odunum@n@format_compliance.pop.cascade_caption.n\endcsname{2\,078}
\expandafter\gdef\csname odunum@val@format_compliance.pop.gemini.n\endcsname{2\,078}
\expandafter\gdef\csname odunum@n@format_compliance.pop.gemini.n\endcsname{2\,078}
\expandafter\gdef\csname odunum@val@format_compliance.pop.gemini35_flash_lite.n\endcsname{2\,078}
\expandafter\gdef\csname odunum@n@format_compliance.pop.gemini35_flash_lite.n\endcsname{2\,078}
\expandafter\gdef\csname odunum@val@format_compliance.pop.gemini37_flash.n\endcsname{2\,067}
\expandafter\gdef\csname odunum@n@format_compliance.pop.gemini37_flash.n\endcsname{2\,078}
\expandafter\gdef\csname odunum@val@format_compliance.pop.humanomni_v2.n\endcsname{1\,111}
\expandafter\gdef\csname odunum@n@format_compliance.pop.humanomni_v2.n\endcsname{2\,078}
\expandafter\gdef\csname odunum@val@format_compliance.pop.ming.n\endcsname{2\,078}
\expandafter\gdef\csname odunum@n@format_compliance.pop.ming.n\endcsname{2\,078}
\expandafter\gdef\csname odunum@val@format_compliance.pop.minicpm_o.n\endcsname{2\,075}
\expandafter\gdef\csname odunum@n@format_compliance.pop.minicpm_o.n\endcsname{2\,078}
\expandafter\gdef\csname odunum@val@format_compliance.pop.nemotron.n\endcsname{2\,078}
\expandafter\gdef\csname odunum@n@format_compliance.pop.nemotron.n\endcsname{2\,078}
\expandafter\gdef\csname odunum@val@format_compliance.pop.qwen25_omni.n\endcsname{2\,078}
\expandafter\gdef\csname odunum@n@format_compliance.pop.qwen25_omni.n\endcsname{2\,078}
\expandafter\gdef\csname odunum@val@format_compliance.pop.qwen3_omni_instruct.n\endcsname{2\,078}
\expandafter\gdef\csname odunum@n@format_compliance.pop.qwen3_omni_instruct.n\endcsname{2\,078}
\expandafter\gdef\csname odunum@val@format_compliance.pop.qwen3_omni_think.n\endcsname{2\,078}
\expandafter\gdef\csname odunum@n@format_compliance.pop.qwen3_omni_think.n\endcsname{2\,078}
\expandafter\gdef\csname odunum@val@format_compliance.pop.qwen_plus.n\endcsname{2\,077}
\expandafter\gdef\csname odunum@n@format_compliance.pop.qwen_plus.n\endcsname{2\,078}
\expandafter\gdef\csname odunum@val@format_compliance.pop.release.audio_only.n\endcsname{731}
\expandafter\gdef\csname odunum@n@format_compliance.pop.release.audio_only.n\endcsname{2\,078}
\expandafter\gdef\csname odunum@val@format_compliance.pop.release.n\endcsname{2\,078}
\expandafter\gdef\csname odunum@n@format_compliance.pop.release.n\endcsname{2\,078}
\expandafter\gdef\csname odunum@val@format_compliance.pop.salmonn2_7b.n\endcsname{2\,058}
\expandafter\gdef\csname odunum@n@format_compliance.pop.salmonn2_7b.n\endcsname{2\,078}
\expandafter\gdef\csname odunum@val@format_compliance.pop.seed.n\endcsname{2\,078}
\expandafter\gdef\csname odunum@n@format_compliance.pop.seed.n\endcsname{2\,078}
\expandafter\gdef\csname odunum@val@format_compliance.pop.videollama2.n\endcsname{1\,605}
\expandafter\gdef\csname odunum@n@format_compliance.pop.videollama2.n\endcsname{2\,078}
\expandafter\gdef\csname odunum@val@format_compliance.pop.vita.n\endcsname{1\,517}
\expandafter\gdef\csname odunum@n@format_compliance.pop.vita.n\endcsname{2\,078}
\expandafter\gdef\csname odunum@val@format_compliance.segments.baichuan_omni.per_scene\endcsname{1.347}
\expandafter\gdef\csname odunum@n@format_compliance.segments.baichuan_omni.per_scene\endcsname{1\,021}
\expandafter\gdef\csname odunum@val@format_compliance.segments.cascade_asr.per_scene\endcsname{0.926}
\expandafter\gdef\csname odunum@n@format_compliance.segments.cascade_asr.per_scene\endcsname{2\,078}
\expandafter\gdef\csname odunum@val@format_compliance.segments.cascade_caption.per_scene\endcsname{0.871}
\expandafter\gdef\csname odunum@n@format_compliance.segments.cascade_caption.per_scene\endcsname{2\,078}
\expandafter\gdef\csname odunum@val@format_compliance.segments.gemini.per_scene\endcsname{0.879}
\expandafter\gdef\csname odunum@n@format_compliance.segments.gemini.per_scene\endcsname{2\,078}
\expandafter\gdef\csname odunum@val@format_compliance.segments.gemini35_flash_lite.per_scene\endcsname{0.786}
\expandafter\gdef\csname odunum@n@format_compliance.segments.gemini35_flash_lite.per_scene\endcsname{2\,078}
\expandafter\gdef\csname odunum@val@format_compliance.segments.gemini37_flash.per_scene\endcsname{0.809}
\expandafter\gdef\csname odunum@n@format_compliance.segments.gemini37_flash.per_scene\endcsname{2\,067}
\expandafter\gdef\csname odunum@val@format_compliance.segments.humanomni_v2.per_scene\endcsname{1.293}
\expandafter\gdef\csname odunum@n@format_compliance.segments.humanomni_v2.per_scene\endcsname{1\,111}
\expandafter\gdef\csname odunum@val@format_compliance.segments.ming.per_scene\endcsname{1.026}
\expandafter\gdef\csname odunum@n@format_compliance.segments.ming.per_scene\endcsname{2\,078}
\expandafter\gdef\csname odunum@val@format_compliance.segments.minicpm_o.per_scene\endcsname{0.949}
\expandafter\gdef\csname odunum@n@format_compliance.segments.minicpm_o.per_scene\endcsname{2\,075}
\expandafter\gdef\csname odunum@val@format_compliance.segments.nemotron.per_scene\endcsname{0.825}
\expandafter\gdef\csname odunum@n@format_compliance.segments.nemotron.per_scene\endcsname{2\,078}
\expandafter\gdef\csname odunum@val@format_compliance.segments.qwen25_omni.per_scene\endcsname{1.007}
\expandafter\gdef\csname odunum@n@format_compliance.segments.qwen25_omni.per_scene\endcsname{2\,078}
\expandafter\gdef\csname odunum@val@format_compliance.segments.qwen3_omni_instruct.per_scene\endcsname{0.989}
\expandafter\gdef\csname odunum@n@format_compliance.segments.qwen3_omni_instruct.per_scene\endcsname{2\,078}
\expandafter\gdef\csname odunum@val@format_compliance.segments.qwen3_omni_think.per_scene\endcsname{0.935}
\expandafter\gdef\csname odunum@n@format_compliance.segments.qwen3_omni_think.per_scene\endcsname{2\,078}
\expandafter\gdef\csname odunum@val@format_compliance.segments.qwen_plus.per_scene\endcsname{0.936}
\expandafter\gdef\csname odunum@n@format_compliance.segments.qwen_plus.per_scene\endcsname{2\,077}
\expandafter\gdef\csname odunum@val@format_compliance.segments.salmonn2_7b.per_scene\endcsname{1.047}
\expandafter\gdef\csname odunum@n@format_compliance.segments.salmonn2_7b.per_scene\endcsname{2\,058}
\expandafter\gdef\csname odunum@val@format_compliance.segments.seed.per_scene\endcsname{0.956}
\expandafter\gdef\csname odunum@n@format_compliance.segments.seed.per_scene\endcsname{2\,078}
\expandafter\gdef\csname odunum@val@format_compliance.segments.videollama2.per_scene\endcsname{0.978}
\expandafter\gdef\csname odunum@n@format_compliance.segments.videollama2.per_scene\endcsname{1\,605}
\expandafter\gdef\csname odunum@val@format_compliance.segments.vita.per_scene\endcsname{0.959}
\expandafter\gdef\csname odunum@n@format_compliance.segments.vita.per_scene\endcsname{1\,517}
\expandafter\gdef\csname odunum@val@human_ab.blind_better\endcsname{5}
\expandafter\gdef\csname odunum@n@human_ab.blind_better\endcsname{65}
\expandafter\gdef\csname odunum@val@human_ab.blind_better_frac\endcsname{0.077}
\expandafter\gdef\csname odunum@n@human_ab.blind_better_frac\endcsname{65}
\expandafter\gdef\csname odunum@val@human_ab.discordant\endcsname{25}
\expandafter\gdef\csname odunum@n@human_ab.discordant\endcsname{65}
\expandafter\gdef\csname odunum@val@human_ab.discordant_oracle_frac\endcsname{0.800}
\expandafter\gdef\csname odunum@n@human_ab.discordant_oracle_frac\endcsname{25}
\expandafter\gdef\csname odunum@ci@human_ab.discordant_oracle_frac\endcsname{[0.593, 0.932]}
\expandafter\gdef\csname odunum@val@human_ab.gt_correct_frac\endcsname{0.953}
\expandafter\gdef\csname odunum@n@human_ab.gt_correct_frac\endcsname{64}
\expandafter\gdef\csname odunum@ci@human_ab.gt_correct_frac\endcsname{[0.869, 0.990]}
\expandafter\gdef\csname odunum@val@human_ab.gt_correct_rated\endcsname{64}
\expandafter\gdef\csname odunum@n@human_ab.gt_correct_rated\endcsname{70}
\expandafter\gdef\csname odunum@val@human_ab.n_rated\endcsname{65}
\expandafter\gdef\csname odunum@n@human_ab.n_rated\endcsname{70}
\expandafter\gdef\csname odunum@val@human_ab.n_sampled\endcsname{70}
\expandafter\gdef\csname odunum@n@human_ab.n_sampled\endcsname{70}
\expandafter\gdef\csname odunum@val@human_ab.oracle_better\endcsname{20}
\expandafter\gdef\csname odunum@n@human_ab.oracle_better\endcsname{65}
\expandafter\gdef\csname odunum@val@human_ab.oracle_better_frac\endcsname{0.308}
\expandafter\gdef\csname odunum@n@human_ab.oracle_better_frac\endcsname{65}
\expandafter\gdef\csname odunum@val@human_ab.sign_test_p\endcsname{0.0041}
\expandafter\gdef\csname odunum@n@human_ab.sign_test_p\endcsname{25}
\expandafter\gdef\csname odunum@val@human_ab.tie\endcsname{40}
\expandafter\gdef\csname odunum@n@human_ab.tie\endcsname{65}
\expandafter\gdef\csname odunum@val@human_ab.tie_frac\endcsname{0.615}
\expandafter\gdef\csname odunum@n@human_ab.tie_frac\endcsname{65}
\expandafter\gdef\csname odunum@val@judge_stability.a.agreement.cur_ds\endcsname{0.996}
\expandafter\gdef\csname odunum@n@judge_stability.a.agreement.cur_ds\endcsname{3\,956}
\expandafter\gdef\csname odunum@val@judge_stability.a.agreement.cur_gpt54\endcsname{0.996}
\expandafter\gdef\csname odunum@n@judge_stability.a.agreement.cur_gpt54\endcsname{3\,956}
\expandafter\gdef\csname odunum@val@judge_stability.a.agreement.ds_gpt54\endcsname{0.993}
\expandafter\gdef\csname odunum@n@judge_stability.a.agreement.ds_gpt54\endcsname{3\,956}
\expandafter\gdef\csname odunum@val@judge_stability.a.completion.cur\endcsname{1.000}
\expandafter\gdef\csname odunum@n@judge_stability.a.completion.cur\endcsname{1\,091}
\expandafter\gdef\csname odunum@val@judge_stability.a.completion.ds\endcsname{1.000}
\expandafter\gdef\csname odunum@n@judge_stability.a.completion.ds\endcsname{1\,091}
\expandafter\gdef\csname odunum@val@judge_stability.a.completion.gpt54\endcsname{1.000}
\expandafter\gdef\csname odunum@n@judge_stability.a.completion.gpt54\endcsname{1\,091}
\expandafter\gdef\csname odunum@val@judge_stability.a.fleiss_kappa\endcsname{0.259}
\expandafter\gdef\csname odunum@n@judge_stability.a.fleiss_kappa\endcsname{3\,956}
\expandafter\gdef\csname odunum@val@judge_stability.a.hit.cur.audio\endcsname{1.000}
\expandafter\gdef\csname odunum@n@judge_stability.a.hit.cur.audio\endcsname{557}
\expandafter\gdef\csname odunum@val@judge_stability.a.hit.cur.context_all\endcsname{0.996}
\expandafter\gdef\csname odunum@n@judge_stability.a.hit.cur.context_all\endcsname{1\,972}
\expandafter\gdef\csname odunum@val@judge_stability.a.hit.cur.history\endcsname{0.992}
\expandafter\gdef\csname odunum@n@judge_stability.a.hit.cur.history\endcsname{475}
\expandafter\gdef\csname odunum@val@judge_stability.a.hit.cur.intent\endcsname{1.000}
\expandafter\gdef\csname odunum@n@judge_stability.a.hit.cur.intent\endcsname{1\,984}
\expandafter\gdef\csname odunum@val@judge_stability.a.hit.cur.overall\endcsname{0.998}
\expandafter\gdef\csname odunum@n@judge_stability.a.hit.cur.overall\endcsname{3\,956}
\expandafter\gdef\csname odunum@ci@judge_stability.a.hit.cur.overall\endcsname{[0.996, 0.999]}
\expandafter\gdef\csname odunum@val@judge_stability.a.hit.cur.user_state\endcsname{1.000}
\expandafter\gdef\csname odunum@n@judge_stability.a.hit.cur.user_state\endcsname{136}
\expandafter\gdef\csname odunum@val@judge_stability.a.hit.cur.visual\endcsname{0.995}
\expandafter\gdef\csname odunum@n@judge_stability.a.hit.cur.visual\endcsname{804}
\expandafter\gdef\csname odunum@val@judge_stability.a.hit.ds.audio\endcsname{0.996}
\expandafter\gdef\csname odunum@n@judge_stability.a.hit.ds.audio\endcsname{557}
\expandafter\gdef\csname odunum@val@judge_stability.a.hit.ds.context_all\endcsname{0.993}
\expandafter\gdef\csname odunum@n@judge_stability.a.hit.ds.context_all\endcsname{1\,972}
\expandafter\gdef\csname odunum@val@judge_stability.a.hit.ds.history\endcsname{0.992}
\expandafter\gdef\csname odunum@n@judge_stability.a.hit.ds.history\endcsname{475}
\expandafter\gdef\csname odunum@val@judge_stability.a.hit.ds.intent\endcsname{0.998}
\expandafter\gdef\csname odunum@n@judge_stability.a.hit.ds.intent\endcsname{1\,984}
\expandafter\gdef\csname odunum@val@judge_stability.a.hit.ds.overall\endcsname{0.996}
\expandafter\gdef\csname odunum@n@judge_stability.a.hit.ds.overall\endcsname{3\,956}
\expandafter\gdef\csname odunum@ci@judge_stability.a.hit.ds.overall\endcsname{[0.993, 0.998]}
\expandafter\gdef\csname odunum@val@judge_stability.a.hit.ds.user_state\endcsname{0.993}
\expandafter\gdef\csname odunum@n@judge_stability.a.hit.ds.user_state\endcsname{136}
\expandafter\gdef\csname odunum@val@judge_stability.a.hit.ds.visual\endcsname{0.993}
\expandafter\gdef\csname odunum@n@judge_stability.a.hit.ds.visual\endcsname{804}
\expandafter\gdef\csname odunum@val@judge_stability.a.hit.gpt54.audio\endcsname{0.996}
\expandafter\gdef\csname odunum@n@judge_stability.a.hit.gpt54.audio\endcsname{557}
\expandafter\gdef\csname odunum@val@judge_stability.a.hit.gpt54.context_all\endcsname{0.992}
\expandafter\gdef\csname odunum@n@judge_stability.a.hit.gpt54.context_all\endcsname{1\,972}
\expandafter\gdef\csname odunum@val@judge_stability.a.hit.gpt54.history\endcsname{0.992}
\expandafter\gdef\csname odunum@n@judge_stability.a.hit.gpt54.history\endcsname{475}
\expandafter\gdef\csname odunum@val@judge_stability.a.hit.gpt54.intent\endcsname{0.999}
\expandafter\gdef\csname odunum@n@judge_stability.a.hit.gpt54.intent\endcsname{1\,984}
\expandafter\gdef\csname odunum@val@judge_stability.a.hit.gpt54.overall\endcsname{0.996}
\expandafter\gdef\csname odunum@n@judge_stability.a.hit.gpt54.overall\endcsname{3\,956}
\expandafter\gdef\csname odunum@ci@judge_stability.a.hit.gpt54.overall\endcsname{[0.993, 0.998]}
\expandafter\gdef\csname odunum@val@judge_stability.a.hit.gpt54.user_state\endcsname{0.993}
\expandafter\gdef\csname odunum@n@judge_stability.a.hit.gpt54.user_state\endcsname{136}
\expandafter\gdef\csname odunum@val@judge_stability.a.hit.gpt54.visual\endcsname{0.989}
\expandafter\gdef\csname odunum@n@judge_stability.a.hit.gpt54.visual\endcsname{804}
\expandafter\gdef\csname odunum@val@judge_stability.a.kappa.cur_ds\endcsname{0.318}
\expandafter\gdef\csname odunum@n@judge_stability.a.kappa.cur_ds\endcsname{3\,956}
\expandafter\gdef\csname odunum@val@judge_stability.a.kappa.cur_gpt54\endcsname{0.318}
\expandafter\gdef\csname odunum@n@judge_stability.a.kappa.cur_gpt54\endcsname{3\,956}
\expandafter\gdef\csname odunum@val@judge_stability.a.kappa.ds_gpt54\endcsname{0.173}
\expandafter\gdef\csname odunum@n@judge_stability.a.kappa.ds_gpt54\endcsname{3\,956}
\expandafter\gdef\csname odunum@val@judge_stability.a.missing_verdicts.cur\endcsname{0}
\expandafter\gdef\csname odunum@n@judge_stability.a.missing_verdicts.cur\endcsname{1\,091}
\expandafter\gdef\csname odunum@val@judge_stability.a.missing_verdicts.ds\endcsname{0}
\expandafter\gdef\csname odunum@n@judge_stability.a.missing_verdicts.ds\endcsname{1\,091}
\expandafter\gdef\csname odunum@val@judge_stability.a.missing_verdicts.gpt54\endcsname{0}
\expandafter\gdef\csname odunum@n@judge_stability.a.missing_verdicts.gpt54\endcsname{1\,091}
\expandafter\gdef\csname odunum@val@judge_stability.a.orphan.cur.overall\endcsname{0.002}
\expandafter\gdef\csname odunum@n@judge_stability.a.orphan.cur.overall\endcsname{3\,956}
\expandafter\gdef\csname odunum@val@judge_stability.a.orphan.ds.overall\endcsname{0.004}
\expandafter\gdef\csname odunum@n@judge_stability.a.orphan.ds.overall\endcsname{3\,956}
\expandafter\gdef\csname odunum@val@judge_stability.a.orphan.gpt54.overall\endcsname{0.004}
\expandafter\gdef\csname odunum@n@judge_stability.a.orphan.gpt54.overall\endcsname{3\,956}
\expandafter\gdef\csname odunum@val@judge_stability.b.completion.cur.gemini\endcsname{1.000}
\expandafter\gdef\csname odunum@n@judge_stability.b.completion.cur.gemini\endcsname{1\,068}
\expandafter\gdef\csname odunum@val@judge_stability.b.completion.cur.qwen_plus\endcsname{1.000}
\expandafter\gdef\csname odunum@n@judge_stability.b.completion.cur.qwen_plus\endcsname{1\,048}
\expandafter\gdef\csname odunum@val@judge_stability.b.completion.cur.seed\endcsname{1.000}
\expandafter\gdef\csname odunum@n@judge_stability.b.completion.cur.seed\endcsname{1\,051}
\expandafter\gdef\csname odunum@val@judge_stability.b.completion.ds.gemini\endcsname{1.000}
\expandafter\gdef\csname odunum@n@judge_stability.b.completion.ds.gemini\endcsname{1\,068}
\expandafter\gdef\csname odunum@val@judge_stability.b.completion.ds.qwen_plus\endcsname{1.000}
\expandafter\gdef\csname odunum@n@judge_stability.b.completion.ds.qwen_plus\endcsname{1\,048}
\expandafter\gdef\csname odunum@val@judge_stability.b.completion.ds.seed\endcsname{1.000}
\expandafter\gdef\csname odunum@n@judge_stability.b.completion.ds.seed\endcsname{1\,051}
\expandafter\gdef\csname odunum@val@judge_stability.b.completion.gpt54.gemini\endcsname{1.000}
\expandafter\gdef\csname odunum@n@judge_stability.b.completion.gpt54.gemini\endcsname{1\,068}
\expandafter\gdef\csname odunum@val@judge_stability.b.completion.gpt54.qwen_plus\endcsname{1.000}
\expandafter\gdef\csname odunum@n@judge_stability.b.completion.gpt54.qwen_plus\endcsname{1\,048}
\expandafter\gdef\csname odunum@val@judge_stability.b.completion.gpt54.seed\endcsname{1.000}
\expandafter\gdef\csname odunum@n@judge_stability.b.completion.gpt54.seed\endcsname{1\,051}
\expandafter\gdef\csname odunum@val@judge_stability.b.cov.cur.gemini\endcsname{0.648}
\expandafter\gdef\csname odunum@n@judge_stability.b.cov.cur.gemini\endcsname{3\,910}
\expandafter\gdef\csname odunum@ci@judge_stability.b.cov.cur.gemini\endcsname{[0.633, 0.662]}
\expandafter\gdef\csname odunum@val@judge_stability.b.cov.cur.gemini.audio\endcsname{0.483}
\expandafter\gdef\csname odunum@n@judge_stability.b.cov.cur.gemini.audio\endcsname{549}
\expandafter\gdef\csname odunum@val@judge_stability.b.cov.cur.gemini.context_all\endcsname{0.451}
\expandafter\gdef\csname odunum@n@judge_stability.b.cov.cur.gemini.context_all\endcsname{1\,945}
\expandafter\gdef\csname odunum@val@judge_stability.b.cov.cur.gemini.history\endcsname{0.446}
\expandafter\gdef\csname odunum@n@judge_stability.b.cov.cur.gemini.history\endcsname{469}
\expandafter\gdef\csname odunum@val@judge_stability.b.cov.cur.gemini.intent\endcsname{0.846}
\expandafter\gdef\csname odunum@n@judge_stability.b.cov.cur.gemini.intent\endcsname{1\,952}
\expandafter\gdef\csname odunum@val@judge_stability.b.cov.cur.gemini.user_state\endcsname{0.207}
\expandafter\gdef\csname odunum@n@judge_stability.b.cov.cur.gemini.user_state\endcsname{135}
\expandafter\gdef\csname odunum@val@judge_stability.b.cov.cur.gemini.visual\endcsname{0.475}
\expandafter\gdef\csname odunum@n@judge_stability.b.cov.cur.gemini.visual\endcsname{792}
\expandafter\gdef\csname odunum@val@judge_stability.b.cov.cur.qwen_plus\endcsname{0.682}
\expandafter\gdef\csname odunum@n@judge_stability.b.cov.cur.qwen_plus\endcsname{3\,842}
\expandafter\gdef\csname odunum@ci@judge_stability.b.cov.cur.qwen_plus\endcsname{[0.667, 0.696]}
\expandafter\gdef\csname odunum@val@judge_stability.b.cov.cur.qwen_plus.audio\endcsname{0.468}
\expandafter\gdef\csname odunum@n@judge_stability.b.cov.cur.qwen_plus.audio\endcsname{534}
\expandafter\gdef\csname odunum@val@judge_stability.b.cov.cur.qwen_plus.context_all\endcsname{0.499}
\expandafter\gdef\csname odunum@n@judge_stability.b.cov.cur.qwen_plus.context_all\endcsname{1\,903}
\expandafter\gdef\csname odunum@val@judge_stability.b.cov.cur.qwen_plus.history\endcsname{0.574}
\expandafter\gdef\csname odunum@n@judge_stability.b.cov.cur.qwen_plus.history\endcsname{470}
\expandafter\gdef\csname odunum@val@judge_stability.b.cov.cur.qwen_plus.intent\endcsname{0.866}
\expandafter\gdef\csname odunum@n@judge_stability.b.cov.cur.qwen_plus.intent\endcsname{1\,926}
\expandafter\gdef\csname odunum@val@judge_stability.b.cov.cur.qwen_plus.user_state\endcsname{0.427}
\expandafter\gdef\csname odunum@n@judge_stability.b.cov.cur.qwen_plus.user_state\endcsname{131}
\expandafter\gdef\csname odunum@val@judge_stability.b.cov.cur.qwen_plus.visual\endcsname{0.486}
\expandafter\gdef\csname odunum@n@judge_stability.b.cov.cur.qwen_plus.visual\endcsname{768}
\expandafter\gdef\csname odunum@val@judge_stability.b.cov.cur.seed\endcsname{0.667}
\expandafter\gdef\csname odunum@n@judge_stability.b.cov.cur.seed\endcsname{3\,847}
\expandafter\gdef\csname odunum@ci@judge_stability.b.cov.cur.seed\endcsname{[0.649, 0.685]}
\expandafter\gdef\csname odunum@val@judge_stability.b.cov.cur.seed.audio\endcsname{0.507}
\expandafter\gdef\csname odunum@n@judge_stability.b.cov.cur.seed.audio\endcsname{534}
\expandafter\gdef\csname odunum@val@judge_stability.b.cov.cur.seed.context_all\endcsname{0.551}
\expandafter\gdef\csname odunum@n@judge_stability.b.cov.cur.seed.context_all\endcsname{1\,912}
\expandafter\gdef\csname odunum@val@judge_stability.b.cov.cur.seed.history\endcsname{0.633}
\expandafter\gdef\csname odunum@n@judge_stability.b.cov.cur.seed.history\endcsname{471}
\expandafter\gdef\csname odunum@val@judge_stability.b.cov.cur.seed.intent\endcsname{0.784}
\expandafter\gdef\csname odunum@n@judge_stability.b.cov.cur.seed.intent\endcsname{1\,922}
\expandafter\gdef\csname odunum@val@judge_stability.b.cov.cur.seed.user_state\endcsname{0.362}
\expandafter\gdef\csname odunum@n@judge_stability.b.cov.cur.seed.user_state\endcsname{130}
\expandafter\gdef\csname odunum@val@judge_stability.b.cov.cur.seed.visual\endcsname{0.564}
\expandafter\gdef\csname odunum@n@judge_stability.b.cov.cur.seed.visual\endcsname{777}
\expandafter\gdef\csname odunum@val@judge_stability.b.cov.ds.gemini\endcsname{0.639}
\expandafter\gdef\csname odunum@n@judge_stability.b.cov.ds.gemini\endcsname{3\,910}
\expandafter\gdef\csname odunum@ci@judge_stability.b.cov.ds.gemini\endcsname{[0.623, 0.654]}
\expandafter\gdef\csname odunum@val@judge_stability.b.cov.ds.gemini.audio\endcsname{0.479}
\expandafter\gdef\csname odunum@n@judge_stability.b.cov.ds.gemini.audio\endcsname{549}
\expandafter\gdef\csname odunum@val@judge_stability.b.cov.ds.gemini.context_all\endcsname{0.456}
\expandafter\gdef\csname odunum@n@judge_stability.b.cov.ds.gemini.context_all\endcsname{1\,945}
\expandafter\gdef\csname odunum@val@judge_stability.b.cov.ds.gemini.history\endcsname{0.452}
\expandafter\gdef\csname odunum@n@judge_stability.b.cov.ds.gemini.history\endcsname{469}
\expandafter\gdef\csname odunum@val@judge_stability.b.cov.ds.gemini.intent\endcsname{0.821}
\expandafter\gdef\csname odunum@n@judge_stability.b.cov.ds.gemini.intent\endcsname{1\,952}
\expandafter\gdef\csname odunum@val@judge_stability.b.cov.ds.gemini.user_state\endcsname{0.207}
\expandafter\gdef\csname odunum@n@judge_stability.b.cov.ds.gemini.user_state\endcsname{135}
\expandafter\gdef\csname odunum@val@judge_stability.b.cov.ds.gemini.visual\endcsname{0.484}
\expandafter\gdef\csname odunum@n@judge_stability.b.cov.ds.gemini.visual\endcsname{792}
\expandafter\gdef\csname odunum@val@judge_stability.b.cov.ds.qwen_plus\endcsname{0.668}
\expandafter\gdef\csname odunum@n@judge_stability.b.cov.ds.qwen_plus\endcsname{3\,842}
\expandafter\gdef\csname odunum@ci@judge_stability.b.cov.ds.qwen_plus\endcsname{[0.653, 0.683]}
\expandafter\gdef\csname odunum@val@judge_stability.b.cov.ds.qwen_plus.audio\endcsname{0.466}
\expandafter\gdef\csname odunum@n@judge_stability.b.cov.ds.qwen_plus.audio\endcsname{534}
\expandafter\gdef\csname odunum@val@judge_stability.b.cov.ds.qwen_plus.context_all\endcsname{0.491}
\expandafter\gdef\csname odunum@n@judge_stability.b.cov.ds.qwen_plus.context_all\endcsname{1\,903}
\expandafter\gdef\csname odunum@val@judge_stability.b.cov.ds.qwen_plus.history\endcsname{0.579}
\expandafter\gdef\csname odunum@n@judge_stability.b.cov.ds.qwen_plus.history\endcsname{470}
\expandafter\gdef\csname odunum@val@judge_stability.b.cov.ds.qwen_plus.intent\endcsname{0.845}
\expandafter\gdef\csname odunum@n@judge_stability.b.cov.ds.qwen_plus.intent\endcsname{1\,926}
\expandafter\gdef\csname odunum@val@judge_stability.b.cov.ds.qwen_plus.user_state\endcsname{0.382}
\expandafter\gdef\csname odunum@n@judge_stability.b.cov.ds.qwen_plus.user_state\endcsname{131}
\expandafter\gdef\csname odunum@val@judge_stability.b.cov.ds.qwen_plus.visual\endcsname{0.474}
\expandafter\gdef\csname odunum@n@judge_stability.b.cov.ds.qwen_plus.visual\endcsname{768}
\expandafter\gdef\csname odunum@val@judge_stability.b.cov.ds.seed\endcsname{0.656}
\expandafter\gdef\csname odunum@n@judge_stability.b.cov.ds.seed\endcsname{3\,847}
\expandafter\gdef\csname odunum@ci@judge_stability.b.cov.ds.seed\endcsname{[0.638, 0.674]}
\expandafter\gdef\csname odunum@val@judge_stability.b.cov.ds.seed.audio\endcsname{0.496}
\expandafter\gdef\csname odunum@n@judge_stability.b.cov.ds.seed.audio\endcsname{534}
\expandafter\gdef\csname odunum@val@judge_stability.b.cov.ds.seed.context_all\endcsname{0.541}
\expandafter\gdef\csname odunum@n@judge_stability.b.cov.ds.seed.context_all\endcsname{1\,912}
\expandafter\gdef\csname odunum@val@judge_stability.b.cov.ds.seed.history\endcsname{0.620}
\expandafter\gdef\csname odunum@n@judge_stability.b.cov.ds.seed.history\endcsname{471}
\expandafter\gdef\csname odunum@val@judge_stability.b.cov.ds.seed.intent\endcsname{0.773}
\expandafter\gdef\csname odunum@n@judge_stability.b.cov.ds.seed.intent\endcsname{1\,922}
\expandafter\gdef\csname odunum@val@judge_stability.b.cov.ds.seed.user_state\endcsname{0.362}
\expandafter\gdef\csname odunum@n@judge_stability.b.cov.ds.seed.user_state\endcsname{130}
\expandafter\gdef\csname odunum@val@judge_stability.b.cov.ds.seed.visual\endcsname{0.553}
\expandafter\gdef\csname odunum@n@judge_stability.b.cov.ds.seed.visual\endcsname{777}
\expandafter\gdef\csname odunum@val@judge_stability.b.cov.gpt54.gemini\endcsname{0.609}
\expandafter\gdef\csname odunum@n@judge_stability.b.cov.gpt54.gemini\endcsname{3\,910}
\expandafter\gdef\csname odunum@ci@judge_stability.b.cov.gpt54.gemini\endcsname{[0.596, 0.623]}
\expandafter\gdef\csname odunum@val@judge_stability.b.cov.gpt54.gemini.audio\endcsname{0.457}
\expandafter\gdef\csname odunum@n@judge_stability.b.cov.gpt54.gemini.audio\endcsname{549}
\expandafter\gdef\csname odunum@val@judge_stability.b.cov.gpt54.gemini.context_all\endcsname{0.396}
\expandafter\gdef\csname odunum@n@judge_stability.b.cov.gpt54.gemini.context_all\endcsname{1\,945}
\expandafter\gdef\csname odunum@val@judge_stability.b.cov.gpt54.gemini.history\endcsname{0.388}
\expandafter\gdef\csname odunum@n@judge_stability.b.cov.gpt54.gemini.history\endcsname{469}
\expandafter\gdef\csname odunum@val@judge_stability.b.cov.gpt54.gemini.intent\endcsname{0.824}
\expandafter\gdef\csname odunum@n@judge_stability.b.cov.gpt54.gemini.intent\endcsname{1\,952}
\expandafter\gdef\csname odunum@val@judge_stability.b.cov.gpt54.gemini.user_state\endcsname{0.193}
\expandafter\gdef\csname odunum@n@judge_stability.b.cov.gpt54.gemini.user_state\endcsname{135}
\expandafter\gdef\csname odunum@val@judge_stability.b.cov.gpt54.gemini.visual\endcsname{0.394}
\expandafter\gdef\csname odunum@n@judge_stability.b.cov.gpt54.gemini.visual\endcsname{792}
\expandafter\gdef\csname odunum@val@judge_stability.b.cov.gpt54.qwen_plus\endcsname{0.641}
\expandafter\gdef\csname odunum@n@judge_stability.b.cov.gpt54.qwen_plus\endcsname{3\,842}
\expandafter\gdef\csname odunum@ci@judge_stability.b.cov.gpt54.qwen_plus\endcsname{[0.627, 0.655]}
\expandafter\gdef\csname odunum@val@judge_stability.b.cov.gpt54.qwen_plus.audio\endcsname{0.466}
\expandafter\gdef\csname odunum@n@judge_stability.b.cov.gpt54.qwen_plus.audio\endcsname{534}
\expandafter\gdef\csname odunum@val@judge_stability.b.cov.gpt54.qwen_plus.context_all\endcsname{0.445}
\expandafter\gdef\csname odunum@n@judge_stability.b.cov.gpt54.qwen_plus.context_all\endcsname{1\,903}
\expandafter\gdef\csname odunum@val@judge_stability.b.cov.gpt54.qwen_plus.history\endcsname{0.509}
\expandafter\gdef\csname odunum@n@judge_stability.b.cov.gpt54.qwen_plus.history\endcsname{470}
\expandafter\gdef\csname odunum@val@judge_stability.b.cov.gpt54.qwen_plus.intent\endcsname{0.837}
\expandafter\gdef\csname odunum@n@judge_stability.b.cov.gpt54.qwen_plus.intent\endcsname{1\,926}
\expandafter\gdef\csname odunum@val@judge_stability.b.cov.gpt54.qwen_plus.user_state\endcsname{0.412}
\expandafter\gdef\csname odunum@n@judge_stability.b.cov.gpt54.qwen_plus.user_state\endcsname{131}
\expandafter\gdef\csname odunum@val@judge_stability.b.cov.gpt54.qwen_plus.visual\endcsname{0.396}
\expandafter\gdef\csname odunum@n@judge_stability.b.cov.gpt54.qwen_plus.visual\endcsname{768}
\expandafter\gdef\csname odunum@val@judge_stability.b.cov.gpt54.seed\endcsname{0.638}
\expandafter\gdef\csname odunum@n@judge_stability.b.cov.gpt54.seed\endcsname{3\,847}
\expandafter\gdef\csname odunum@ci@judge_stability.b.cov.gpt54.seed\endcsname{[0.621, 0.655]}
\expandafter\gdef\csname odunum@val@judge_stability.b.cov.gpt54.seed.audio\endcsname{0.487}
\expandafter\gdef\csname odunum@n@judge_stability.b.cov.gpt54.seed.audio\endcsname{534}
\expandafter\gdef\csname odunum@val@judge_stability.b.cov.gpt54.seed.context_all\endcsname{0.515}
\expandafter\gdef\csname odunum@n@judge_stability.b.cov.gpt54.seed.context_all\endcsname{1\,912}
\expandafter\gdef\csname odunum@val@judge_stability.b.cov.gpt54.seed.history\endcsname{0.605}
\expandafter\gdef\csname odunum@n@judge_stability.b.cov.gpt54.seed.history\endcsname{471}
\expandafter\gdef\csname odunum@val@judge_stability.b.cov.gpt54.seed.intent\endcsname{0.762}
\expandafter\gdef\csname odunum@n@judge_stability.b.cov.gpt54.seed.intent\endcsname{1\,922}
\expandafter\gdef\csname odunum@val@judge_stability.b.cov.gpt54.seed.user_state\endcsname{0.331}
\expandafter\gdef\csname odunum@n@judge_stability.b.cov.gpt54.seed.user_state\endcsname{130}
\expandafter\gdef\csname odunum@val@judge_stability.b.cov.gpt54.seed.visual\endcsname{0.511}
\expandafter\gdef\csname odunum@n@judge_stability.b.cov.gpt54.seed.visual\endcsname{777}
\expandafter\gdef\csname odunum@val@judge_stability.b.delta.ds.gemini\endcsname{-0.009}
\expandafter\gdef\csname odunum@n@judge_stability.b.delta.ds.gemini\endcsname{3\,910}
\expandafter\gdef\csname odunum@ci@judge_stability.b.delta.ds.gemini\endcsname{[-0.020, 0.002]}
\expandafter\gdef\csname odunum@val@judge_stability.b.delta.ds.qwen_plus\endcsname{-0.014}
\expandafter\gdef\csname odunum@n@judge_stability.b.delta.ds.qwen_plus\endcsname{3\,842}
\expandafter\gdef\csname odunum@ci@judge_stability.b.delta.ds.qwen_plus\endcsname{[-0.024, -0.004]}
\expandafter\gdef\csname odunum@val@judge_stability.b.delta.ds.seed\endcsname{-0.011}
\expandafter\gdef\csname odunum@n@judge_stability.b.delta.ds.seed\endcsname{3\,847}
\expandafter\gdef\csname odunum@ci@judge_stability.b.delta.ds.seed\endcsname{[-0.020, -0.002]}
\expandafter\gdef\csname odunum@val@judge_stability.b.delta.gpt54.gemini\endcsname{-0.039}
\expandafter\gdef\csname odunum@n@judge_stability.b.delta.gpt54.gemini\endcsname{3\,910}
\expandafter\gdef\csname odunum@ci@judge_stability.b.delta.gpt54.gemini\endcsname{[-0.049, -0.029]}
\expandafter\gdef\csname odunum@val@judge_stability.b.delta.gpt54.qwen_plus\endcsname{-0.041}
\expandafter\gdef\csname odunum@n@judge_stability.b.delta.gpt54.qwen_plus\endcsname{3\,842}
\expandafter\gdef\csname odunum@ci@judge_stability.b.delta.gpt54.qwen_plus\endcsname{[-0.051, -0.031]}
\expandafter\gdef\csname odunum@val@judge_stability.b.delta.gpt54.seed\endcsname{-0.029}
\expandafter\gdef\csname odunum@n@judge_stability.b.delta.gpt54.seed\endcsname{3\,847}
\expandafter\gdef\csname odunum@ci@judge_stability.b.delta.gpt54.seed\endcsname{[-0.039, -0.020]}
\expandafter\gdef\csname odunum@val@judge_stability.b.gap.cur.gemini\endcsname{0.394}
\expandafter\gdef\csname odunum@n@judge_stability.b.gap.cur.gemini\endcsname{1\,945}
\expandafter\gdef\csname odunum@val@judge_stability.b.gap.cur.qwen_plus\endcsname{0.367}
\expandafter\gdef\csname odunum@n@judge_stability.b.gap.cur.qwen_plus\endcsname{1\,903}
\expandafter\gdef\csname odunum@val@judge_stability.b.gap.cur.seed\endcsname{0.232}
\expandafter\gdef\csname odunum@n@judge_stability.b.gap.cur.seed\endcsname{1\,912}
\expandafter\gdef\csname odunum@val@judge_stability.b.gap.ds.gemini\endcsname{0.366}
\expandafter\gdef\csname odunum@n@judge_stability.b.gap.ds.gemini\endcsname{1\,945}
\expandafter\gdef\csname odunum@val@judge_stability.b.gap.ds.qwen_plus\endcsname{0.353}
\expandafter\gdef\csname odunum@n@judge_stability.b.gap.ds.qwen_plus\endcsname{1\,903}
\expandafter\gdef\csname odunum@val@judge_stability.b.gap.ds.seed\endcsname{0.232}
\expandafter\gdef\csname odunum@n@judge_stability.b.gap.ds.seed\endcsname{1\,912}
\expandafter\gdef\csname odunum@val@judge_stability.b.gap.gpt54.gemini\endcsname{0.427}
\expandafter\gdef\csname odunum@n@judge_stability.b.gap.gpt54.gemini\endcsname{1\,945}
\expandafter\gdef\csname odunum@val@judge_stability.b.gap.gpt54.qwen_plus\endcsname{0.393}
\expandafter\gdef\csname odunum@n@judge_stability.b.gap.gpt54.qwen_plus\endcsname{1\,903}
\expandafter\gdef\csname odunum@val@judge_stability.b.gap.gpt54.seed\endcsname{0.247}
\expandafter\gdef\csname odunum@n@judge_stability.b.gap.gpt54.seed\endcsname{1\,912}
\expandafter\gdef\csname odunum@val@judge_stability.b.gap_direction_preserved\endcsname{3}
\expandafter\gdef\csname odunum@n@judge_stability.b.gap_direction_preserved\endcsname{3}
\expandafter\gdef\csname odunum@val@judge_stability.b.gemini.agreement.cur_ds\endcsname{0.906}
\expandafter\gdef\csname odunum@n@judge_stability.b.gemini.agreement.cur_ds\endcsname{3\,910}
\expandafter\gdef\csname odunum@val@judge_stability.b.gemini.agreement.cur_gpt54\endcsname{0.911}
\expandafter\gdef\csname odunum@n@judge_stability.b.gemini.agreement.cur_gpt54\endcsname{3\,910}
\expandafter\gdef\csname odunum@val@judge_stability.b.gemini.agreement.ds_gpt54\endcsname{0.887}
\expandafter\gdef\csname odunum@n@judge_stability.b.gemini.agreement.ds_gpt54\endcsname{3\,910}
\expandafter\gdef\csname odunum@val@judge_stability.b.gemini.fleiss_kappa\endcsname{0.788}
\expandafter\gdef\csname odunum@n@judge_stability.b.gemini.fleiss_kappa\endcsname{3\,910}
\expandafter\gdef\csname odunum@val@judge_stability.b.gemini.kappa.cur_ds\endcsname{0.796}
\expandafter\gdef\csname odunum@n@judge_stability.b.gemini.kappa.cur_ds\endcsname{3\,910}
\expandafter\gdef\csname odunum@val@judge_stability.b.gemini.kappa.cur_gpt54\endcsname{0.809}
\expandafter\gdef\csname odunum@n@judge_stability.b.gemini.kappa.cur_gpt54\endcsname{3\,910}
\expandafter\gdef\csname odunum@val@judge_stability.b.gemini.kappa.ds_gpt54\endcsname{0.759}
\expandafter\gdef\csname odunum@n@judge_stability.b.gemini.kappa.ds_gpt54\endcsname{3\,910}
\expandafter\gdef\csname odunum@val@judge_stability.b.n_points.gemini\endcsname{3\,910}
\expandafter\gdef\csname odunum@n@judge_stability.b.n_points.gemini\endcsname{3\,910}
\expandafter\gdef\csname odunum@val@judge_stability.b.n_points.qwen_plus\endcsname{3\,842}
\expandafter\gdef\csname odunum@n@judge_stability.b.n_points.qwen_plus\endcsname{3\,842}
\expandafter\gdef\csname odunum@val@judge_stability.b.n_points.seed\endcsname{3\,847}
\expandafter\gdef\csname odunum@n@judge_stability.b.n_points.seed\endcsname{3\,847}
\expandafter\gdef\csname odunum@val@judge_stability.b.qwen_plus.agreement.cur_ds\endcsname{0.919}
\expandafter\gdef\csname odunum@n@judge_stability.b.qwen_plus.agreement.cur_ds\endcsname{3\,842}
\expandafter\gdef\csname odunum@val@judge_stability.b.qwen_plus.agreement.cur_gpt54\endcsname{0.910}
\expandafter\gdef\csname odunum@n@judge_stability.b.qwen_plus.agreement.cur_gpt54\endcsname{3\,842}
\expandafter\gdef\csname odunum@val@judge_stability.b.qwen_plus.agreement.ds_gpt54\endcsname{0.905}
\expandafter\gdef\csname odunum@n@judge_stability.b.qwen_plus.agreement.ds_gpt54\endcsname{3\,842}
\expandafter\gdef\csname odunum@val@judge_stability.b.qwen_plus.fleiss_kappa\endcsname{0.802}
\expandafter\gdef\csname odunum@n@judge_stability.b.qwen_plus.fleiss_kappa\endcsname{3\,842}
\expandafter\gdef\csname odunum@val@judge_stability.b.qwen_plus.kappa.cur_ds\endcsname{0.816}
\expandafter\gdef\csname odunum@n@judge_stability.b.qwen_plus.kappa.cur_ds\endcsname{3\,842}
\expandafter\gdef\csname odunum@val@judge_stability.b.qwen_plus.kappa.cur_gpt54\endcsname{0.800}
\expandafter\gdef\csname odunum@n@judge_stability.b.qwen_plus.kappa.cur_gpt54\endcsname{3\,842}
\expandafter\gdef\csname odunum@val@judge_stability.b.qwen_plus.kappa.ds_gpt54\endcsname{0.790}
\expandafter\gdef\csname odunum@n@judge_stability.b.qwen_plus.kappa.ds_gpt54\endcsname{3\,842}
\expandafter\gdef\csname odunum@val@judge_stability.b.range.gemini\endcsname{0.039}
\expandafter\gdef\csname odunum@n@judge_stability.b.range.gemini\endcsname{3\,910}
\expandafter\gdef\csname odunum@val@judge_stability.b.range.qwen_plus\endcsname{0.041}
\expandafter\gdef\csname odunum@n@judge_stability.b.range.qwen_plus\endcsname{3\,842}
\expandafter\gdef\csname odunum@val@judge_stability.b.range.seed\endcsname{0.029}
\expandafter\gdef\csname odunum@n@judge_stability.b.range.seed\endcsname{3\,847}
\expandafter\gdef\csname odunum@val@judge_stability.b.rank_bottom_preserved\endcsname{3}
\expandafter\gdef\csname odunum@n@judge_stability.b.rank_bottom_preserved\endcsname{3}
\expandafter\gdef\csname odunum@val@judge_stability.b.rank_order\endcsname{Qwen3.5-Omni-Plus \textgreater{} Seed 2.0 Lite \textgreater{} Gemini 3.1 Pro}
\expandafter\gdef\csname odunum@n@judge_stability.b.rank_order\endcsname{3}
\expandafter\gdef\csname odunum@val@judge_stability.b.rank_preserved\endcsname{3}
\expandafter\gdef\csname odunum@n@judge_stability.b.rank_preserved\endcsname{3}
\expandafter\gdef\csname odunum@val@judge_stability.b.seed.agreement.cur_ds\endcsname{0.926}
\expandafter\gdef\csname odunum@n@judge_stability.b.seed.agreement.cur_ds\endcsname{3\,847}
\expandafter\gdef\csname odunum@val@judge_stability.b.seed.agreement.cur_gpt54\endcsname{0.923}
\expandafter\gdef\csname odunum@n@judge_stability.b.seed.agreement.cur_gpt54\endcsname{3\,847}
\expandafter\gdef\csname odunum@val@judge_stability.b.seed.agreement.ds_gpt54\endcsname{0.919}
\expandafter\gdef\csname odunum@n@judge_stability.b.seed.agreement.ds_gpt54\endcsname{3\,847}
\expandafter\gdef\csname odunum@val@judge_stability.b.seed.fleiss_kappa\endcsname{0.829}
\expandafter\gdef\csname odunum@n@judge_stability.b.seed.fleiss_kappa\endcsname{3\,847}
\expandafter\gdef\csname odunum@val@judge_stability.b.seed.kappa.cur_ds\endcsname{0.834}
\expandafter\gdef\csname odunum@n@judge_stability.b.seed.kappa.cur_ds\endcsname{3\,847}
\expandafter\gdef\csname odunum@val@judge_stability.b.seed.kappa.cur_gpt54\endcsname{0.830}
\expandafter\gdef\csname odunum@n@judge_stability.b.seed.kappa.cur_gpt54\endcsname{3\,847}
\expandafter\gdef\csname odunum@val@judge_stability.b.seed.kappa.ds_gpt54\endcsname{0.824}
\expandafter\gdef\csname odunum@n@judge_stability.b.seed.kappa.ds_gpt54\endcsname{3\,847}
\expandafter\gdef\csname odunum@val@judge_stability.b.top2_margin.cur\endcsname{0.015}
\expandafter\gdef\csname odunum@n@judge_stability.b.top2_margin.cur\endcsname{3}
\expandafter\gdef\csname odunum@val@judge_stability.b.top2_margin.ds\endcsname{0.012}
\expandafter\gdef\csname odunum@n@judge_stability.b.top2_margin.ds\endcsname{3}
\expandafter\gdef\csname odunum@val@judge_stability.b.top2_margin.gpt54\endcsname{0.003}
\expandafter\gdef\csname odunum@n@judge_stability.b.top2_margin.gpt54\endcsname{3}
\expandafter\gdef\csname odunum@val@judge_stability.judge.cur.config\endcsname{mr\_ali/dashscope.qwen3.6-flash thinking}
\expandafter\gdef\csname odunum@n@judge_stability.judge.cur.config\endcsname{1}
\expandafter\gdef\csname odunum@val@judge_stability.judge.cur.name\endcsname{Qwen3.6-Flash}
\expandafter\gdef\csname odunum@n@judge_stability.judge.cur.name\endcsname{1}
\expandafter\gdef\csname odunum@val@judge_stability.judge.ds.config\endcsname{mr/deepseek-v4-flash}
\expandafter\gdef\csname odunum@n@judge_stability.judge.ds.config\endcsname{1}
\expandafter\gdef\csname odunum@val@judge_stability.judge.ds.name\endcsname{DeepSeek-V4-Flash}
\expandafter\gdef\csname odunum@n@judge_stability.judge.ds.name\endcsname{1}
\expandafter\gdef\csname odunum@val@judge_stability.judge.gpt54.config\endcsname{gpt56/openai.gpt-5.4}
\expandafter\gdef\csname odunum@n@judge_stability.judge.gpt54.config\endcsname{1}
\expandafter\gdef\csname odunum@val@judge_stability.judge.gpt54.name\endcsname{GPT-5.4}
\expandafter\gdef\csname odunum@n@judge_stability.judge.gpt54.name\endcsname{1}
\expandafter\gdef\csname odunum@val@judge_stability.pop.n_points_a\endcsname{3\,956}
\expandafter\gdef\csname odunum@n@judge_stability.pop.n_points_a\endcsname{3\,956}
\expandafter\gdef\csname odunum@val@judge_stability.pop.n_scenes\endcsname{1\,060}
\expandafter\gdef\csname odunum@n@judge_stability.pop.n_scenes\endcsname{1\,060}
\expandafter\gdef\csname odunum@val@keypoint_types.audit.cascade_asr.scenes_with_full_gt_denominator\endcsname{1\,630}
\expandafter\gdef\csname odunum@n@keypoint_types.audit.cascade_asr.scenes_with_full_gt_denominator\endcsname{1\,630}
\expandafter\gdef\csname odunum@val@keypoint_types.audit.gemini.scenes_with_full_gt_denominator\endcsname{1\,630}
\expandafter\gdef\csname odunum@n@keypoint_types.audit.gemini.scenes_with_full_gt_denominator\endcsname{1\,630}
\expandafter\gdef\csname odunum@val@keypoint_types.audit.qwen_plus.scenes_with_full_gt_denominator\endcsname{1\,630}
\expandafter\gdef\csname odunum@n@keypoint_types.audit.qwen_plus.scenes_with_full_gt_denominator\endcsname{1\,630}
\expandafter\gdef\csname odunum@val@keypoint_types.audit.seed.scenes_with_full_gt_denominator\endcsname{1\,630}
\expandafter\gdef\csname odunum@n@keypoint_types.audit.seed.scenes_with_full_gt_denominator\endcsname{1\,630}
\expandafter\gdef\csname odunum@val@keypoint_types.cov.cascade_asr.all.audio\endcsname{0.337}
\expandafter\gdef\csname odunum@n@keypoint_types.cov.cascade_asr.all.audio\endcsname{898}
\expandafter\gdef\csname odunum@ci@keypoint_types.cov.cascade_asr.all.audio\endcsname{[0.306, 0.369]}
\expandafter\gdef\csname odunum@val@keypoint_types.cov.cascade_asr.all.context_all\endcsname{0.364}
\expandafter\gdef\csname odunum@n@keypoint_types.cov.cascade_asr.all.context_all\endcsname{2\,700}
\expandafter\gdef\csname odunum@ci@keypoint_types.cov.cascade_asr.all.context_all\endcsname{[0.345, 0.384]}
\expandafter\gdef\csname odunum@val@keypoint_types.cov.cascade_asr.all.history\endcsname{0.592}
\expandafter\gdef\csname odunum@n@keypoint_types.cov.cascade_asr.all.history\endcsname{817}
\expandafter\gdef\csname odunum@ci@keypoint_types.cov.cascade_asr.all.history\endcsname{[0.558, 0.627]}
\expandafter\gdef\csname odunum@val@keypoint_types.cov.cascade_asr.all.intent\endcsname{0.801}
\expandafter\gdef\csname odunum@n@keypoint_types.cov.cascade_asr.all.intent\endcsname{3\,142}
\expandafter\gdef\csname odunum@ci@keypoint_types.cov.cascade_asr.all.intent\endcsname{[0.786, 0.817]}
\expandafter\gdef\csname odunum@val@keypoint_types.cov.cascade_asr.all.other_context\endcsname{0.188}
\expandafter\gdef\csname odunum@n@keypoint_types.cov.cascade_asr.all.other_context\endcsname{181}
\expandafter\gdef\csname odunum@ci@keypoint_types.cov.cascade_asr.all.other_context\endcsname{[0.133, 0.247]}
\expandafter\gdef\csname odunum@val@keypoint_types.cov.cascade_asr.all.overall\endcsname{0.598}
\expandafter\gdef\csname odunum@n@keypoint_types.cov.cascade_asr.all.overall\endcsname{5\,855}
\expandafter\gdef\csname odunum@ci@keypoint_types.cov.cascade_asr.all.overall\endcsname{[0.584, 0.613]}
\expandafter\gdef\csname odunum@val@keypoint_types.cov.cascade_asr.all.unattributed\endcsname{0.231}
\expandafter\gdef\csname odunum@n@keypoint_types.cov.cascade_asr.all.unattributed\endcsname{13}
\expandafter\gdef\csname odunum@ci@keypoint_types.cov.cascade_asr.all.unattributed\endcsname{[0.000, 0.368]}
\expandafter\gdef\csname odunum@val@keypoint_types.cov.cascade_asr.all.visual\endcsname{0.203}
\expandafter\gdef\csname odunum@n@keypoint_types.cov.cascade_asr.all.visual\endcsname{804}
\expandafter\gdef\csname odunum@ci@keypoint_types.cov.cascade_asr.all.visual\endcsname{[0.175, 0.232]}
\expandafter\gdef\csname odunum@val@keypoint_types.cov.cascade_asr.ao.audio\endcsname{0.340}
\expandafter\gdef\csname odunum@n@keypoint_types.cov.cascade_asr.ao.audio\endcsname{341}
\expandafter\gdef\csname odunum@ci@keypoint_types.cov.cascade_asr.ao.audio\endcsname{[0.289, 0.392]}
\expandafter\gdef\csname odunum@val@keypoint_types.cov.cascade_asr.ao.context_all\endcsname{0.459}
\expandafter\gdef\csname odunum@n@keypoint_types.cov.cascade_asr.ao.context_all\endcsname{728}
\expandafter\gdef\csname odunum@ci@keypoint_types.cov.cascade_asr.ao.context_all\endcsname{[0.421, 0.497]}
\expandafter\gdef\csname odunum@val@keypoint_types.cov.cascade_asr.ao.history\endcsname{0.599}
\expandafter\gdef\csname odunum@n@keypoint_types.cov.cascade_asr.ao.history\endcsname{342}
\expandafter\gdef\csname odunum@ci@keypoint_types.cov.cascade_asr.ao.history\endcsname{[0.545, 0.653]}
\expandafter\gdef\csname odunum@val@keypoint_types.cov.cascade_asr.ao.intent\endcsname{0.822}
\expandafter\gdef\csname odunum@n@keypoint_types.cov.cascade_asr.ao.intent\endcsname{1\,158}
\expandafter\gdef\csname odunum@ci@keypoint_types.cov.cascade_asr.ao.intent\endcsname{[0.797, 0.847]}
\expandafter\gdef\csname odunum@val@keypoint_types.cov.cascade_asr.ao.other_context\endcsname{0.289}
\expandafter\gdef\csname odunum@n@keypoint_types.cov.cascade_asr.ao.other_context\endcsname{45}
\expandafter\gdef\csname odunum@ci@keypoint_types.cov.cascade_asr.ao.other_context\endcsname{[0.163, 0.422]}
\expandafter\gdef\csname odunum@val@keypoint_types.cov.cascade_asr.ao.overall\endcsname{0.682}
\expandafter\gdef\csname odunum@n@keypoint_types.cov.cascade_asr.ao.overall\endcsname{1\,886}
\expandafter\gdef\csname odunum@ci@keypoint_types.cov.cascade_asr.ao.overall\endcsname{[0.657, 0.706]}
\expandafter\gdef\csname odunum@val@keypoint_types.cov.cascade_asr.av.audio\endcsname{0.336}
\expandafter\gdef\csname odunum@n@keypoint_types.cov.cascade_asr.av.audio\endcsname{557}
\expandafter\gdef\csname odunum@ci@keypoint_types.cov.cascade_asr.av.audio\endcsname{[0.297, 0.375]}
\expandafter\gdef\csname odunum@val@keypoint_types.cov.cascade_asr.av.context_all\endcsname{0.330}
\expandafter\gdef\csname odunum@n@keypoint_types.cov.cascade_asr.av.context_all\endcsname{1\,972}
\expandafter\gdef\csname odunum@ci@keypoint_types.cov.cascade_asr.av.context_all\endcsname{[0.308, 0.352]}
\expandafter\gdef\csname odunum@val@keypoint_types.cov.cascade_asr.av.history\endcsname{0.587}
\expandafter\gdef\csname odunum@n@keypoint_types.cov.cascade_asr.av.history\endcsname{475}
\expandafter\gdef\csname odunum@ci@keypoint_types.cov.cascade_asr.av.history\endcsname{[0.542, 0.633]}
\expandafter\gdef\csname odunum@val@keypoint_types.cov.cascade_asr.av.intent\endcsname{0.789}
\expandafter\gdef\csname odunum@n@keypoint_types.cov.cascade_asr.av.intent\endcsname{1\,984}
\expandafter\gdef\csname odunum@ci@keypoint_types.cov.cascade_asr.av.intent\endcsname{[0.769, 0.809]}
\expandafter\gdef\csname odunum@val@keypoint_types.cov.cascade_asr.av.other_context\endcsname{0.154}
\expandafter\gdef\csname odunum@n@keypoint_types.cov.cascade_asr.av.other_context\endcsname{136}
\expandafter\gdef\csname odunum@ci@keypoint_types.cov.cascade_asr.av.other_context\endcsname{[0.096, 0.218]}
\expandafter\gdef\csname odunum@val@keypoint_types.cov.cascade_asr.av.overall\endcsname{0.559}
\expandafter\gdef\csname odunum@n@keypoint_types.cov.cascade_asr.av.overall\endcsname{3\,969}
\expandafter\gdef\csname odunum@ci@keypoint_types.cov.cascade_asr.av.overall\endcsname{[0.542, 0.576]}
\expandafter\gdef\csname odunum@val@keypoint_types.cov.cascade_asr.av.unattributed\endcsname{0.231}
\expandafter\gdef\csname odunum@n@keypoint_types.cov.cascade_asr.av.unattributed\endcsname{13}
\expandafter\gdef\csname odunum@ci@keypoint_types.cov.cascade_asr.av.unattributed\endcsname{[0.000, 0.368]}
\expandafter\gdef\csname odunum@val@keypoint_types.cov.cascade_asr.av.visual\endcsname{0.203}
\expandafter\gdef\csname odunum@n@keypoint_types.cov.cascade_asr.av.visual\endcsname{804}
\expandafter\gdef\csname odunum@ci@keypoint_types.cov.cascade_asr.av.visual\endcsname{[0.175, 0.232]}
\expandafter\gdef\csname odunum@val@keypoint_types.cov.gemini.all.audio\endcsname{0.482}
\expandafter\gdef\csname odunum@n@keypoint_types.cov.gemini.all.audio\endcsname{898}
\expandafter\gdef\csname odunum@ci@keypoint_types.cov.gemini.all.audio\endcsname{[0.449, 0.515]}
\expandafter\gdef\csname odunum@val@keypoint_types.cov.gemini.all.context_all\endcsname{0.456}
\expandafter\gdef\csname odunum@n@keypoint_types.cov.gemini.all.context_all\endcsname{2\,700}
\expandafter\gdef\csname odunum@ci@keypoint_types.cov.gemini.all.context_all\endcsname{[0.436, 0.475]}
\expandafter\gdef\csname odunum@val@keypoint_types.cov.gemini.all.history\endcsname{0.443}
\expandafter\gdef\csname odunum@n@keypoint_types.cov.gemini.all.history\endcsname{817}
\expandafter\gdef\csname odunum@ci@keypoint_types.cov.gemini.all.history\endcsname{[0.408, 0.480]}
\expandafter\gdef\csname odunum@val@keypoint_types.cov.gemini.all.intent\endcsname{0.831}
\expandafter\gdef\csname odunum@n@keypoint_types.cov.gemini.all.intent\endcsname{3\,142}
\expandafter\gdef\csname odunum@ci@keypoint_types.cov.gemini.all.intent\endcsname{[0.817, 0.845]}
\expandafter\gdef\csname odunum@val@keypoint_types.cov.gemini.all.other_context\endcsname{0.249}
\expandafter\gdef\csname odunum@n@keypoint_types.cov.gemini.all.other_context\endcsname{181}
\expandafter\gdef\csname odunum@ci@keypoint_types.cov.gemini.all.other_context\endcsname{[0.188, 0.312]}
\expandafter\gdef\csname odunum@val@keypoint_types.cov.gemini.all.overall\endcsname{0.657}
\expandafter\gdef\csname odunum@n@keypoint_types.cov.gemini.all.overall\endcsname{5\,855}
\expandafter\gdef\csname odunum@ci@keypoint_types.cov.gemini.all.overall\endcsname{[0.645, 0.669]}
\expandafter\gdef\csname odunum@val@keypoint_types.cov.gemini.all.unattributed\endcsname{0.308}
\expandafter\gdef\csname odunum@n@keypoint_types.cov.gemini.all.unattributed\endcsname{13}
\expandafter\gdef\csname odunum@ci@keypoint_types.cov.gemini.all.unattributed\endcsname{[0.000, 0.706]}
\expandafter\gdef\csname odunum@val@keypoint_types.cov.gemini.all.visual\endcsname{0.486}
\expandafter\gdef\csname odunum@n@keypoint_types.cov.gemini.all.visual\endcsname{804}
\expandafter\gdef\csname odunum@ci@keypoint_types.cov.gemini.all.visual\endcsname{[0.451, 0.522]}
\expandafter\gdef\csname odunum@val@keypoint_types.cov.gemini.ao.audio\endcsname{0.499}
\expandafter\gdef\csname odunum@n@keypoint_types.cov.gemini.ao.audio\endcsname{341}
\expandafter\gdef\csname odunum@ci@keypoint_types.cov.gemini.ao.audio\endcsname{[0.446, 0.555]}
\expandafter\gdef\csname odunum@val@keypoint_types.cov.gemini.ao.context_all\endcsname{0.481}
\expandafter\gdef\csname odunum@n@keypoint_types.cov.gemini.ao.context_all\endcsname{728}
\expandafter\gdef\csname odunum@ci@keypoint_types.cov.gemini.ao.context_all\endcsname{[0.444, 0.518]}
\expandafter\gdef\csname odunum@val@keypoint_types.cov.gemini.ao.history\endcsname{0.474}
\expandafter\gdef\csname odunum@n@keypoint_types.cov.gemini.ao.history\endcsname{342}
\expandafter\gdef\csname odunum@ci@keypoint_types.cov.gemini.ao.history\endcsname{[0.418, 0.530]}
\expandafter\gdef\csname odunum@val@keypoint_types.cov.gemini.ao.intent\endcsname{0.835}
\expandafter\gdef\csname odunum@n@keypoint_types.cov.gemini.ao.intent\endcsname{1\,158}
\expandafter\gdef\csname odunum@ci@keypoint_types.cov.gemini.ao.intent\endcsname{[0.812, 0.858]}
\expandafter\gdef\csname odunum@val@keypoint_types.cov.gemini.ao.other_context\endcsname{0.400}
\expandafter\gdef\csname odunum@n@keypoint_types.cov.gemini.ao.other_context\endcsname{45}
\expandafter\gdef\csname odunum@ci@keypoint_types.cov.gemini.ao.other_context\endcsname{[0.267, 0.535]}
\expandafter\gdef\csname odunum@val@keypoint_types.cov.gemini.ao.overall\endcsname{0.698}
\expandafter\gdef\csname odunum@n@keypoint_types.cov.gemini.ao.overall\endcsname{1\,886}
\expandafter\gdef\csname odunum@ci@keypoint_types.cov.gemini.ao.overall\endcsname{[0.677, 0.719]}
\expandafter\gdef\csname odunum@val@keypoint_types.cov.gemini.av.audio\endcsname{0.472}
\expandafter\gdef\csname odunum@n@keypoint_types.cov.gemini.av.audio\endcsname{557}
\expandafter\gdef\csname odunum@ci@keypoint_types.cov.gemini.av.audio\endcsname{[0.430, 0.514]}
\expandafter\gdef\csname odunum@val@keypoint_types.cov.gemini.av.context_all\endcsname{0.447}
\expandafter\gdef\csname odunum@n@keypoint_types.cov.gemini.av.context_all\endcsname{1\,972}
\expandafter\gdef\csname odunum@ci@keypoint_types.cov.gemini.av.context_all\endcsname{[0.424, 0.469]}
\expandafter\gdef\csname odunum@val@keypoint_types.cov.gemini.av.history\endcsname{0.421}
\expandafter\gdef\csname odunum@n@keypoint_types.cov.gemini.av.history\endcsname{475}
\expandafter\gdef\csname odunum@ci@keypoint_types.cov.gemini.av.history\endcsname{[0.376, 0.468]}
\expandafter\gdef\csname odunum@val@keypoint_types.cov.gemini.av.intent\endcsname{0.829}
\expandafter\gdef\csname odunum@n@keypoint_types.cov.gemini.av.intent\endcsname{1\,984}
\expandafter\gdef\csname odunum@ci@keypoint_types.cov.gemini.av.intent\endcsname{[0.811, 0.846]}
\expandafter\gdef\csname odunum@val@keypoint_types.cov.gemini.av.other_context\endcsname{0.199}
\expandafter\gdef\csname odunum@n@keypoint_types.cov.gemini.av.other_context\endcsname{136}
\expandafter\gdef\csname odunum@ci@keypoint_types.cov.gemini.av.other_context\endcsname{[0.133, 0.266]}
\expandafter\gdef\csname odunum@val@keypoint_types.cov.gemini.av.overall\endcsname{0.637}
\expandafter\gdef\csname odunum@n@keypoint_types.cov.gemini.av.overall\endcsname{3\,969}
\expandafter\gdef\csname odunum@ci@keypoint_types.cov.gemini.av.overall\endcsname{[0.623, 0.652]}
\expandafter\gdef\csname odunum@val@keypoint_types.cov.gemini.av.unattributed\endcsname{0.308}
\expandafter\gdef\csname odunum@n@keypoint_types.cov.gemini.av.unattributed\endcsname{13}
\expandafter\gdef\csname odunum@ci@keypoint_types.cov.gemini.av.unattributed\endcsname{[0.000, 0.706]}
\expandafter\gdef\csname odunum@val@keypoint_types.cov.gemini.av.visual\endcsname{0.486}
\expandafter\gdef\csname odunum@n@keypoint_types.cov.gemini.av.visual\endcsname{804}
\expandafter\gdef\csname odunum@ci@keypoint_types.cov.gemini.av.visual\endcsname{[0.451, 0.522]}
\expandafter\gdef\csname odunum@val@keypoint_types.cov.qwen_plus.all.audio\endcsname{0.470}
\expandafter\gdef\csname odunum@n@keypoint_types.cov.qwen_plus.all.audio\endcsname{898}
\expandafter\gdef\csname odunum@ci@keypoint_types.cov.qwen_plus.all.audio\endcsname{[0.438, 0.502]}
\expandafter\gdef\csname odunum@val@keypoint_types.cov.qwen_plus.all.context_all\endcsname{0.505}
\expandafter\gdef\csname odunum@n@keypoint_types.cov.qwen_plus.all.context_all\endcsname{2\,700}
\expandafter\gdef\csname odunum@ci@keypoint_types.cov.qwen_plus.all.context_all\endcsname{[0.485, 0.525]}
\expandafter\gdef\csname odunum@val@keypoint_types.cov.qwen_plus.all.history\endcsname{0.581}
\expandafter\gdef\csname odunum@n@keypoint_types.cov.qwen_plus.all.history\endcsname{817}
\expandafter\gdef\csname odunum@ci@keypoint_types.cov.qwen_plus.all.history\endcsname{[0.546, 0.617]}
\expandafter\gdef\csname odunum@val@keypoint_types.cov.qwen_plus.all.intent\endcsname{0.830}
\expandafter\gdef\csname odunum@n@keypoint_types.cov.qwen_plus.all.intent\endcsname{3\,142}
\expandafter\gdef\csname odunum@ci@keypoint_types.cov.qwen_plus.all.intent\endcsname{[0.816, 0.844]}
\expandafter\gdef\csname odunum@val@keypoint_types.cov.qwen_plus.all.other_context\endcsname{0.481}
\expandafter\gdef\csname odunum@n@keypoint_types.cov.qwen_plus.all.other_context\endcsname{181}
\expandafter\gdef\csname odunum@ci@keypoint_types.cov.qwen_plus.all.other_context\endcsname{[0.409, 0.555]}
\expandafter\gdef\csname odunum@val@keypoint_types.cov.qwen_plus.all.overall\endcsname{0.679}
\expandafter\gdef\csname odunum@n@keypoint_types.cov.qwen_plus.all.overall\endcsname{5\,855}
\expandafter\gdef\csname odunum@ci@keypoint_types.cov.qwen_plus.all.overall\endcsname{[0.667, 0.691]}
\expandafter\gdef\csname odunum@val@keypoint_types.cov.qwen_plus.all.unattributed\endcsname{0.231}
\expandafter\gdef\csname odunum@n@keypoint_types.cov.qwen_plus.all.unattributed\endcsname{13}
\expandafter\gdef\csname odunum@ci@keypoint_types.cov.qwen_plus.all.unattributed\endcsname{[0.000, 0.500]}
\expandafter\gdef\csname odunum@val@keypoint_types.cov.qwen_plus.all.visual\endcsname{0.473}
\expandafter\gdef\csname odunum@n@keypoint_types.cov.qwen_plus.all.visual\endcsname{804}
\expandafter\gdef\csname odunum@ci@keypoint_types.cov.qwen_plus.all.visual\endcsname{[0.437, 0.507]}
\expandafter\gdef\csname odunum@val@keypoint_types.cov.qwen_plus.ao.audio\endcsname{0.493}
\expandafter\gdef\csname odunum@n@keypoint_types.cov.qwen_plus.ao.audio\endcsname{341}
\expandafter\gdef\csname odunum@ci@keypoint_types.cov.qwen_plus.ao.audio\endcsname{[0.440, 0.545]}
\expandafter\gdef\csname odunum@val@keypoint_types.cov.qwen_plus.ao.context_all\endcsname{0.548}
\expandafter\gdef\csname odunum@n@keypoint_types.cov.qwen_plus.ao.context_all\endcsname{728}
\expandafter\gdef\csname odunum@ci@keypoint_types.cov.qwen_plus.ao.context_all\endcsname{[0.511, 0.585]}
\expandafter\gdef\csname odunum@val@keypoint_types.cov.qwen_plus.ao.history\endcsname{0.588}
\expandafter\gdef\csname odunum@n@keypoint_types.cov.qwen_plus.ao.history\endcsname{342}
\expandafter\gdef\csname odunum@ci@keypoint_types.cov.qwen_plus.ao.history\endcsname{[0.533, 0.641]}
\expandafter\gdef\csname odunum@val@keypoint_types.cov.qwen_plus.ao.intent\endcsname{0.820}
\expandafter\gdef\csname odunum@n@keypoint_types.cov.qwen_plus.ao.intent\endcsname{1\,158}
\expandafter\gdef\csname odunum@ci@keypoint_types.cov.qwen_plus.ao.intent\endcsname{[0.796, 0.843]}
\expandafter\gdef\csname odunum@val@keypoint_types.cov.qwen_plus.ao.other_context\endcsname{0.667}
\expandafter\gdef\csname odunum@n@keypoint_types.cov.qwen_plus.ao.other_context\endcsname{45}
\expandafter\gdef\csname odunum@ci@keypoint_types.cov.qwen_plus.ao.other_context\endcsname{[0.533, 0.795]}
\expandafter\gdef\csname odunum@val@keypoint_types.cov.qwen_plus.ao.overall\endcsname{0.715}
\expandafter\gdef\csname odunum@n@keypoint_types.cov.qwen_plus.ao.overall\endcsname{1\,886}
\expandafter\gdef\csname odunum@ci@keypoint_types.cov.qwen_plus.ao.overall\endcsname{[0.694, 0.736]}
\expandafter\gdef\csname odunum@val@keypoint_types.cov.qwen_plus.av.audio\endcsname{0.456}
\expandafter\gdef\csname odunum@n@keypoint_types.cov.qwen_plus.av.audio\endcsname{557}
\expandafter\gdef\csname odunum@ci@keypoint_types.cov.qwen_plus.av.audio\endcsname{[0.414, 0.498]}
\expandafter\gdef\csname odunum@val@keypoint_types.cov.qwen_plus.av.context_all\endcsname{0.489}
\expandafter\gdef\csname odunum@n@keypoint_types.cov.qwen_plus.av.context_all\endcsname{1\,972}
\expandafter\gdef\csname odunum@ci@keypoint_types.cov.qwen_plus.av.context_all\endcsname{[0.466, 0.512]}
\expandafter\gdef\csname odunum@val@keypoint_types.cov.qwen_plus.av.history\endcsname{0.577}
\expandafter\gdef\csname odunum@n@keypoint_types.cov.qwen_plus.av.history\endcsname{475}
\expandafter\gdef\csname odunum@ci@keypoint_types.cov.qwen_plus.av.history\endcsname{[0.532, 0.622]}
\expandafter\gdef\csname odunum@val@keypoint_types.cov.qwen_plus.av.intent\endcsname{0.836}
\expandafter\gdef\csname odunum@n@keypoint_types.cov.qwen_plus.av.intent\endcsname{1\,984}
\expandafter\gdef\csname odunum@ci@keypoint_types.cov.qwen_plus.av.intent\endcsname{[0.819, 0.852]}
\expandafter\gdef\csname odunum@val@keypoint_types.cov.qwen_plus.av.other_context\endcsname{0.419}
\expandafter\gdef\csname odunum@n@keypoint_types.cov.qwen_plus.av.other_context\endcsname{136}
\expandafter\gdef\csname odunum@ci@keypoint_types.cov.qwen_plus.av.other_context\endcsname{[0.338, 0.504]}
\expandafter\gdef\csname odunum@val@keypoint_types.cov.qwen_plus.av.overall\endcsname{0.662}
\expandafter\gdef\csname odunum@n@keypoint_types.cov.qwen_plus.av.overall\endcsname{3\,969}
\expandafter\gdef\csname odunum@ci@keypoint_types.cov.qwen_plus.av.overall\endcsname{[0.647, 0.677]}
\expandafter\gdef\csname odunum@val@keypoint_types.cov.qwen_plus.av.unattributed\endcsname{0.231}
\expandafter\gdef\csname odunum@n@keypoint_types.cov.qwen_plus.av.unattributed\endcsname{13}
\expandafter\gdef\csname odunum@ci@keypoint_types.cov.qwen_plus.av.unattributed\endcsname{[0.000, 0.500]}
\expandafter\gdef\csname odunum@val@keypoint_types.cov.qwen_plus.av.visual\endcsname{0.473}
\expandafter\gdef\csname odunum@n@keypoint_types.cov.qwen_plus.av.visual\endcsname{804}
\expandafter\gdef\csname odunum@ci@keypoint_types.cov.qwen_plus.av.visual\endcsname{[0.437, 0.507]}
\expandafter\gdef\csname odunum@val@keypoint_types.cov.seed.all.audio\endcsname{0.467}
\expandafter\gdef\csname odunum@n@keypoint_types.cov.seed.all.audio\endcsname{898}
\expandafter\gdef\csname odunum@ci@keypoint_types.cov.seed.all.audio\endcsname{[0.434, 0.499]}
\expandafter\gdef\csname odunum@val@keypoint_types.cov.seed.all.context_all\endcsname{0.548}
\expandafter\gdef\csname odunum@n@keypoint_types.cov.seed.all.context_all\endcsname{2\,700}
\expandafter\gdef\csname odunum@ci@keypoint_types.cov.seed.all.context_all\endcsname{[0.528, 0.567]}
\expandafter\gdef\csname odunum@val@keypoint_types.cov.seed.all.history\endcsname{0.687}
\expandafter\gdef\csname odunum@n@keypoint_types.cov.seed.all.history\endcsname{817}
\expandafter\gdef\csname odunum@ci@keypoint_types.cov.seed.all.history\endcsname{[0.654, 0.719]}
\expandafter\gdef\csname odunum@val@keypoint_types.cov.seed.all.intent\endcsname{0.803}
\expandafter\gdef\csname odunum@n@keypoint_types.cov.seed.all.intent\endcsname{3\,142}
\expandafter\gdef\csname odunum@ci@keypoint_types.cov.seed.all.intent\endcsname{[0.786, 0.820]}
\expandafter\gdef\csname odunum@val@keypoint_types.cov.seed.all.other_context\endcsname{0.326}
\expandafter\gdef\csname odunum@n@keypoint_types.cov.seed.all.other_context\endcsname{181}
\expandafter\gdef\csname odunum@ci@keypoint_types.cov.seed.all.other_context\endcsname{[0.258, 0.394]}
\expandafter\gdef\csname odunum@val@keypoint_types.cov.seed.all.overall\endcsname{0.684}
\expandafter\gdef\csname odunum@n@keypoint_types.cov.seed.all.overall\endcsname{5\,855}
\expandafter\gdef\csname odunum@ci@keypoint_types.cov.seed.all.overall\endcsname{[0.669, 0.699]}
\expandafter\gdef\csname odunum@val@keypoint_types.cov.seed.all.unattributed\endcsname{0.462}
\expandafter\gdef\csname odunum@n@keypoint_types.cov.seed.all.unattributed\endcsname{13}
\expandafter\gdef\csname odunum@ci@keypoint_types.cov.seed.all.unattributed\endcsname{[0.118, 0.875]}
\expandafter\gdef\csname odunum@val@keypoint_types.cov.seed.all.visual\endcsname{0.547}
\expandafter\gdef\csname odunum@n@keypoint_types.cov.seed.all.visual\endcsname{804}
\expandafter\gdef\csname odunum@ci@keypoint_types.cov.seed.all.visual\endcsname{[0.512, 0.582]}
\expandafter\gdef\csname odunum@val@keypoint_types.cov.seed.ao.audio\endcsname{0.469}
\expandafter\gdef\csname odunum@n@keypoint_types.cov.seed.ao.audio\endcsname{341}
\expandafter\gdef\csname odunum@ci@keypoint_types.cov.seed.ao.audio\endcsname{[0.417, 0.523]}
\expandafter\gdef\csname odunum@val@keypoint_types.cov.seed.ao.context_all\endcsname{0.600}
\expandafter\gdef\csname odunum@n@keypoint_types.cov.seed.ao.context_all\endcsname{728}
\expandafter\gdef\csname odunum@ci@keypoint_types.cov.seed.ao.context_all\endcsname{[0.564, 0.636]}
\expandafter\gdef\csname odunum@val@keypoint_types.cov.seed.ao.history\endcsname{0.775}
\expandafter\gdef\csname odunum@n@keypoint_types.cov.seed.ao.history\endcsname{342}
\expandafter\gdef\csname odunum@ci@keypoint_types.cov.seed.ao.history\endcsname{[0.729, 0.820]}
\expandafter\gdef\csname odunum@val@keypoint_types.cov.seed.ao.intent\endcsname{0.876}
\expandafter\gdef\csname odunum@n@keypoint_types.cov.seed.ao.intent\endcsname{1\,158}
\expandafter\gdef\csname odunum@ci@keypoint_types.cov.seed.ao.intent\endcsname{[0.853, 0.897]}
\expandafter\gdef\csname odunum@val@keypoint_types.cov.seed.ao.other_context\endcsname{0.267}
\expandafter\gdef\csname odunum@n@keypoint_types.cov.seed.ao.other_context\endcsname{45}
\expandafter\gdef\csname odunum@ci@keypoint_types.cov.seed.ao.other_context\endcsname{[0.149, 0.404]}
\expandafter\gdef\csname odunum@val@keypoint_types.cov.seed.ao.overall\endcsname{0.769}
\expandafter\gdef\csname odunum@n@keypoint_types.cov.seed.ao.overall\endcsname{1\,886}
\expandafter\gdef\csname odunum@ci@keypoint_types.cov.seed.ao.overall\endcsname{[0.749, 0.790]}
\expandafter\gdef\csname odunum@val@keypoint_types.cov.seed.av.audio\endcsname{0.465}
\expandafter\gdef\csname odunum@n@keypoint_types.cov.seed.av.audio\endcsname{557}
\expandafter\gdef\csname odunum@ci@keypoint_types.cov.seed.av.audio\endcsname{[0.423, 0.506]}
\expandafter\gdef\csname odunum@val@keypoint_types.cov.seed.av.context_all\endcsname{0.528}
\expandafter\gdef\csname odunum@n@keypoint_types.cov.seed.av.context_all\endcsname{1\,972}
\expandafter\gdef\csname odunum@ci@keypoint_types.cov.seed.av.context_all\endcsname{[0.506, 0.551]}
\expandafter\gdef\csname odunum@val@keypoint_types.cov.seed.av.history\endcsname{0.623}
\expandafter\gdef\csname odunum@n@keypoint_types.cov.seed.av.history\endcsname{475}
\expandafter\gdef\csname odunum@ci@keypoint_types.cov.seed.av.history\endcsname{[0.576, 0.668]}
\expandafter\gdef\csname odunum@val@keypoint_types.cov.seed.av.intent\endcsname{0.760}
\expandafter\gdef\csname odunum@n@keypoint_types.cov.seed.av.intent\endcsname{1\,984}
\expandafter\gdef\csname odunum@ci@keypoint_types.cov.seed.av.intent\endcsname{[0.737, 0.783]}
\expandafter\gdef\csname odunum@val@keypoint_types.cov.seed.av.other_context\endcsname{0.346}
\expandafter\gdef\csname odunum@n@keypoint_types.cov.seed.av.other_context\endcsname{136}
\expandafter\gdef\csname odunum@ci@keypoint_types.cov.seed.av.other_context\endcsname{[0.268, 0.429]}
\expandafter\gdef\csname odunum@val@keypoint_types.cov.seed.av.overall\endcsname{0.644}
\expandafter\gdef\csname odunum@n@keypoint_types.cov.seed.av.overall\endcsname{3\,969}
\expandafter\gdef\csname odunum@ci@keypoint_types.cov.seed.av.overall\endcsname{[0.625, 0.662]}
\expandafter\gdef\csname odunum@val@keypoint_types.cov.seed.av.unattributed\endcsname{0.462}
\expandafter\gdef\csname odunum@n@keypoint_types.cov.seed.av.unattributed\endcsname{13}
\expandafter\gdef\csname odunum@ci@keypoint_types.cov.seed.av.unattributed\endcsname{[0.118, 0.875]}
\expandafter\gdef\csname odunum@val@keypoint_types.cov.seed.av.visual\endcsname{0.547}
\expandafter\gdef\csname odunum@n@keypoint_types.cov.seed.av.visual\endcsname{804}
\expandafter\gdef\csname odunum@ci@keypoint_types.cov.seed.av.visual\endcsname{[0.512, 0.582]}
\expandafter\gdef\csname odunum@val@keypoint_types.gap.cascade_asr.all.intent_minus_audio\endcsname{0.448}
\expandafter\gdef\csname odunum@n@keypoint_types.gap.cascade_asr.all.intent_minus_audio\endcsname{812}
\expandafter\gdef\csname odunum@ci@keypoint_types.gap.cascade_asr.all.intent_minus_audio\endcsname{[0.413, 0.482]}
\expandafter\gdef\csname odunum@val@keypoint_types.gap.cascade_asr.all.intent_minus_context\endcsname{0.442}
\expandafter\gdef\csname odunum@n@keypoint_types.gap.cascade_asr.all.intent_minus_context\endcsname{1\,525}
\expandafter\gdef\csname odunum@ci@keypoint_types.gap.cascade_asr.all.intent_minus_context\endcsname{[0.418, 0.465]}
\expandafter\gdef\csname odunum@val@keypoint_types.gap.cascade_asr.all.intent_minus_history\endcsname{0.279}
\expandafter\gdef\csname odunum@n@keypoint_types.gap.cascade_asr.all.intent_minus_history\endcsname{673}
\expandafter\gdef\csname odunum@ci@keypoint_types.gap.cascade_asr.all.intent_minus_history\endcsname{[0.239, 0.318]}
\expandafter\gdef\csname odunum@val@keypoint_types.gap.cascade_asr.all.intent_minus_visual\endcsname{0.587}
\expandafter\gdef\csname odunum@n@keypoint_types.gap.cascade_asr.all.intent_minus_visual\endcsname{675}
\expandafter\gdef\csname odunum@ci@keypoint_types.gap.cascade_asr.all.intent_minus_visual\endcsname{[0.550, 0.623]}
\expandafter\gdef\csname odunum@val@keypoint_types.gap.cascade_asr.ao.intent_minus_audio\endcsname{0.463}
\expandafter\gdef\csname odunum@n@keypoint_types.gap.cascade_asr.ao.intent_minus_audio\endcsname{298}
\expandafter\gdef\csname odunum@ci@keypoint_types.gap.cascade_asr.ao.intent_minus_audio\endcsname{[0.407, 0.521]}
\expandafter\gdef\csname odunum@val@keypoint_types.gap.cascade_asr.ao.intent_minus_context\endcsname{0.388}
\expandafter\gdef\csname odunum@n@keypoint_types.gap.cascade_asr.ao.intent_minus_context\endcsname{496}
\expandafter\gdef\csname odunum@ci@keypoint_types.gap.cascade_asr.ao.intent_minus_context\endcsname{[0.345, 0.431]}
\expandafter\gdef\csname odunum@val@keypoint_types.gap.cascade_asr.ao.intent_minus_history\endcsname{0.299}
\expandafter\gdef\csname odunum@n@keypoint_types.gap.cascade_asr.ao.intent_minus_history\endcsname{274}
\expandafter\gdef\csname odunum@ci@keypoint_types.gap.cascade_asr.ao.intent_minus_history\endcsname{[0.239, 0.360]}
\expandafter\gdef\csname odunum@val@keypoint_types.gap.cascade_asr.av.intent_minus_audio\endcsname{0.439}
\expandafter\gdef\csname odunum@n@keypoint_types.gap.cascade_asr.av.intent_minus_audio\endcsname{514}
\expandafter\gdef\csname odunum@ci@keypoint_types.gap.cascade_asr.av.intent_minus_audio\endcsname{[0.394, 0.482]}
\expandafter\gdef\csname odunum@val@keypoint_types.gap.cascade_asr.av.intent_minus_context\endcsname{0.468}
\expandafter\gdef\csname odunum@n@keypoint_types.gap.cascade_asr.av.intent_minus_context\endcsname{1\,029}
\expandafter\gdef\csname odunum@ci@keypoint_types.gap.cascade_asr.av.intent_minus_context\endcsname{[0.440, 0.496]}
\expandafter\gdef\csname odunum@val@keypoint_types.gap.cascade_asr.av.intent_minus_history\endcsname{0.265}
\expandafter\gdef\csname odunum@n@keypoint_types.gap.cascade_asr.av.intent_minus_history\endcsname{399}
\expandafter\gdef\csname odunum@ci@keypoint_types.gap.cascade_asr.av.intent_minus_history\endcsname{[0.214, 0.316]}
\expandafter\gdef\csname odunum@val@keypoint_types.gap.cascade_asr.av.intent_minus_visual\endcsname{0.587}
\expandafter\gdef\csname odunum@n@keypoint_types.gap.cascade_asr.av.intent_minus_visual\endcsname{675}
\expandafter\gdef\csname odunum@ci@keypoint_types.gap.cascade_asr.av.intent_minus_visual\endcsname{[0.549, 0.624]}
\expandafter\gdef\csname odunum@val@keypoint_types.gap.gemini.all.intent_minus_audio\endcsname{0.366}
\expandafter\gdef\csname odunum@n@keypoint_types.gap.gemini.all.intent_minus_audio\endcsname{812}
\expandafter\gdef\csname odunum@ci@keypoint_types.gap.gemini.all.intent_minus_audio\endcsname{[0.328, 0.403]}
\expandafter\gdef\csname odunum@val@keypoint_types.gap.gemini.all.intent_minus_context\endcsname{0.402}
\expandafter\gdef\csname odunum@n@keypoint_types.gap.gemini.all.intent_minus_context\endcsname{1\,525}
\expandafter\gdef\csname odunum@ci@keypoint_types.gap.gemini.all.intent_minus_context\endcsname{[0.377, 0.426]}
\expandafter\gdef\csname odunum@val@keypoint_types.gap.gemini.all.intent_minus_history\endcsname{0.427}
\expandafter\gdef\csname odunum@n@keypoint_types.gap.gemini.all.intent_minus_history\endcsname{673}
\expandafter\gdef\csname odunum@ci@keypoint_types.gap.gemini.all.intent_minus_history\endcsname{[0.386, 0.468]}
\expandafter\gdef\csname odunum@val@keypoint_types.gap.gemini.all.intent_minus_visual\endcsname{0.357}
\expandafter\gdef\csname odunum@n@keypoint_types.gap.gemini.all.intent_minus_visual\endcsname{675}
\expandafter\gdef\csname odunum@ci@keypoint_types.gap.gemini.all.intent_minus_visual\endcsname{[0.315, 0.400]}
\expandafter\gdef\csname odunum@val@keypoint_types.gap.gemini.ao.intent_minus_audio\endcsname{0.352}
\expandafter\gdef\csname odunum@n@keypoint_types.gap.gemini.ao.intent_minus_audio\endcsname{298}
\expandafter\gdef\csname odunum@ci@keypoint_types.gap.gemini.ao.intent_minus_audio\endcsname{[0.291, 0.412]}
\expandafter\gdef\csname odunum@val@keypoint_types.gap.gemini.ao.intent_minus_context\endcsname{0.396}
\expandafter\gdef\csname odunum@n@keypoint_types.gap.gemini.ao.intent_minus_context\endcsname{496}
\expandafter\gdef\csname odunum@ci@keypoint_types.gap.gemini.ao.intent_minus_context\endcsname{[0.352, 0.440]}
\expandafter\gdef\csname odunum@val@keypoint_types.gap.gemini.ao.intent_minus_history\endcsname{0.412}
\expandafter\gdef\csname odunum@n@keypoint_types.gap.gemini.ao.intent_minus_history\endcsname{274}
\expandafter\gdef\csname odunum@ci@keypoint_types.gap.gemini.ao.intent_minus_history\endcsname{[0.349, 0.475]}
\expandafter\gdef\csname odunum@val@keypoint_types.gap.gemini.av.intent_minus_audio\endcsname{0.374}
\expandafter\gdef\csname odunum@n@keypoint_types.gap.gemini.av.intent_minus_audio\endcsname{514}
\expandafter\gdef\csname odunum@ci@keypoint_types.gap.gemini.av.intent_minus_audio\endcsname{[0.327, 0.423]}
\expandafter\gdef\csname odunum@val@keypoint_types.gap.gemini.av.intent_minus_context\endcsname{0.404}
\expandafter\gdef\csname odunum@n@keypoint_types.gap.gemini.av.intent_minus_context\endcsname{1\,029}
\expandafter\gdef\csname odunum@ci@keypoint_types.gap.gemini.av.intent_minus_context\endcsname{[0.374, 0.434]}
\expandafter\gdef\csname odunum@val@keypoint_types.gap.gemini.av.intent_minus_history\endcsname{0.438}
\expandafter\gdef\csname odunum@n@keypoint_types.gap.gemini.av.intent_minus_history\endcsname{399}
\expandafter\gdef\csname odunum@ci@keypoint_types.gap.gemini.av.intent_minus_history\endcsname{[0.382, 0.492]}
\expandafter\gdef\csname odunum@val@keypoint_types.gap.gemini.av.intent_minus_visual\endcsname{0.357}
\expandafter\gdef\csname odunum@n@keypoint_types.gap.gemini.av.intent_minus_visual\endcsname{675}
\expandafter\gdef\csname odunum@ci@keypoint_types.gap.gemini.av.intent_minus_visual\endcsname{[0.313, 0.400]}
\expandafter\gdef\csname odunum@val@keypoint_types.gap.qwen_plus.all.intent_minus_audio\endcsname{0.365}
\expandafter\gdef\csname odunum@n@keypoint_types.gap.qwen_plus.all.intent_minus_audio\endcsname{812}
\expandafter\gdef\csname odunum@ci@keypoint_types.gap.qwen_plus.all.intent_minus_audio\endcsname{[0.327, 0.402]}
\expandafter\gdef\csname odunum@val@keypoint_types.gap.qwen_plus.all.intent_minus_context\endcsname{0.349}
\expandafter\gdef\csname odunum@n@keypoint_types.gap.qwen_plus.all.intent_minus_context\endcsname{1\,525}
\expandafter\gdef\csname odunum@ci@keypoint_types.gap.qwen_plus.all.intent_minus_context\endcsname{[0.325, 0.372]}
\expandafter\gdef\csname odunum@val@keypoint_types.gap.qwen_plus.all.intent_minus_history\endcsname{0.301}
\expandafter\gdef\csname odunum@n@keypoint_types.gap.qwen_plus.all.intent_minus_history\endcsname{673}
\expandafter\gdef\csname odunum@ci@keypoint_types.gap.qwen_plus.all.intent_minus_history\endcsname{[0.263, 0.339]}
\expandafter\gdef\csname odunum@val@keypoint_types.gap.qwen_plus.all.intent_minus_visual\endcsname{0.364}
\expandafter\gdef\csname odunum@n@keypoint_types.gap.qwen_plus.all.intent_minus_visual\endcsname{675}
\expandafter\gdef\csname odunum@ci@keypoint_types.gap.qwen_plus.all.intent_minus_visual\endcsname{[0.325, 0.405]}
\expandafter\gdef\csname odunum@val@keypoint_types.gap.qwen_plus.ao.intent_minus_audio\endcsname{0.326}
\expandafter\gdef\csname odunum@n@keypoint_types.gap.qwen_plus.ao.intent_minus_audio\endcsname{298}
\expandafter\gdef\csname odunum@ci@keypoint_types.gap.qwen_plus.ao.intent_minus_audio\endcsname{[0.266, 0.387]}
\expandafter\gdef\csname odunum@val@keypoint_types.gap.qwen_plus.ao.intent_minus_context\endcsname{0.310}
\expandafter\gdef\csname odunum@n@keypoint_types.gap.qwen_plus.ao.intent_minus_context\endcsname{496}
\expandafter\gdef\csname odunum@ci@keypoint_types.gap.qwen_plus.ao.intent_minus_context\endcsname{[0.267, 0.354]}
\expandafter\gdef\csname odunum@val@keypoint_types.gap.qwen_plus.ao.intent_minus_history\endcsname{0.292}
\expandafter\gdef\csname odunum@n@keypoint_types.gap.qwen_plus.ao.intent_minus_history\endcsname{274}
\expandafter\gdef\csname odunum@ci@keypoint_types.gap.qwen_plus.ao.intent_minus_history\endcsname{[0.232, 0.352]}
\expandafter\gdef\csname odunum@val@keypoint_types.gap.qwen_plus.av.intent_minus_audio\endcsname{0.387}
\expandafter\gdef\csname odunum@n@keypoint_types.gap.qwen_plus.av.intent_minus_audio\endcsname{514}
\expandafter\gdef\csname odunum@ci@keypoint_types.gap.qwen_plus.av.intent_minus_audio\endcsname{[0.342, 0.433]}
\expandafter\gdef\csname odunum@val@keypoint_types.gap.qwen_plus.av.intent_minus_context\endcsname{0.367}
\expandafter\gdef\csname odunum@n@keypoint_types.gap.qwen_plus.av.intent_minus_context\endcsname{1\,029}
\expandafter\gdef\csname odunum@ci@keypoint_types.gap.qwen_plus.av.intent_minus_context\endcsname{[0.339, 0.395]}
\expandafter\gdef\csname odunum@val@keypoint_types.gap.qwen_plus.av.intent_minus_history\endcsname{0.307}
\expandafter\gdef\csname odunum@n@keypoint_types.gap.qwen_plus.av.intent_minus_history\endcsname{399}
\expandafter\gdef\csname odunum@ci@keypoint_types.gap.qwen_plus.av.intent_minus_history\endcsname{[0.257, 0.357]}
\expandafter\gdef\csname odunum@val@keypoint_types.gap.qwen_plus.av.intent_minus_visual\endcsname{0.364}
\expandafter\gdef\csname odunum@n@keypoint_types.gap.qwen_plus.av.intent_minus_visual\endcsname{675}
\expandafter\gdef\csname odunum@ci@keypoint_types.gap.qwen_plus.av.intent_minus_visual\endcsname{[0.324, 0.405]}
\expandafter\gdef\csname odunum@val@keypoint_types.gap.seed.all.intent_minus_audio\endcsname{0.352}
\expandafter\gdef\csname odunum@n@keypoint_types.gap.seed.all.intent_minus_audio\endcsname{812}
\expandafter\gdef\csname odunum@ci@keypoint_types.gap.seed.all.intent_minus_audio\endcsname{[0.316, 0.387]}
\expandafter\gdef\csname odunum@val@keypoint_types.gap.seed.all.intent_minus_context\endcsname{0.254}
\expandafter\gdef\csname odunum@n@keypoint_types.gap.seed.all.intent_minus_context\endcsname{1\,525}
\expandafter\gdef\csname odunum@ci@keypoint_types.gap.seed.all.intent_minus_context\endcsname{[0.230, 0.278]}
\expandafter\gdef\csname odunum@val@keypoint_types.gap.seed.all.intent_minus_history\endcsname{0.143}
\expandafter\gdef\csname odunum@n@keypoint_types.gap.seed.all.intent_minus_history\endcsname{673}
\expandafter\gdef\csname odunum@ci@keypoint_types.gap.seed.all.intent_minus_history\endcsname{[0.107, 0.180]}
\expandafter\gdef\csname odunum@val@keypoint_types.gap.seed.all.intent_minus_visual\endcsname{0.212}
\expandafter\gdef\csname odunum@n@keypoint_types.gap.seed.all.intent_minus_visual\endcsname{675}
\expandafter\gdef\csname odunum@ci@keypoint_types.gap.seed.all.intent_minus_visual\endcsname{[0.171, 0.254]}
\expandafter\gdef\csname odunum@val@keypoint_types.gap.seed.ao.intent_minus_audio\endcsname{0.409}
\expandafter\gdef\csname odunum@n@keypoint_types.gap.seed.ao.intent_minus_audio\endcsname{298}
\expandafter\gdef\csname odunum@ci@keypoint_types.gap.seed.ao.intent_minus_audio\endcsname{[0.351, 0.467]}
\expandafter\gdef\csname odunum@val@keypoint_types.gap.seed.ao.intent_minus_context\endcsname{0.290}
\expandafter\gdef\csname odunum@n@keypoint_types.gap.seed.ao.intent_minus_context\endcsname{496}
\expandafter\gdef\csname odunum@ci@keypoint_types.gap.seed.ao.intent_minus_context\endcsname{[0.248, 0.332]}
\expandafter\gdef\csname odunum@val@keypoint_types.gap.seed.ao.intent_minus_history\endcsname{0.124}
\expandafter\gdef\csname odunum@n@keypoint_types.gap.seed.ao.intent_minus_history\endcsname{274}
\expandafter\gdef\csname odunum@ci@keypoint_types.gap.seed.ao.intent_minus_history\endcsname{[0.073, 0.176]}
\expandafter\gdef\csname odunum@val@keypoint_types.gap.seed.av.intent_minus_audio\endcsname{0.320}
\expandafter\gdef\csname odunum@n@keypoint_types.gap.seed.av.intent_minus_audio\endcsname{514}
\expandafter\gdef\csname odunum@ci@keypoint_types.gap.seed.av.intent_minus_audio\endcsname{[0.275, 0.364]}
\expandafter\gdef\csname odunum@val@keypoint_types.gap.seed.av.intent_minus_context\endcsname{0.237}
\expandafter\gdef\csname odunum@n@keypoint_types.gap.seed.av.intent_minus_context\endcsname{1\,029}
\expandafter\gdef\csname odunum@ci@keypoint_types.gap.seed.av.intent_minus_context\endcsname{[0.209, 0.265]}
\expandafter\gdef\csname odunum@val@keypoint_types.gap.seed.av.intent_minus_history\endcsname{0.156}
\expandafter\gdef\csname odunum@n@keypoint_types.gap.seed.av.intent_minus_history\endcsname{399}
\expandafter\gdef\csname odunum@ci@keypoint_types.gap.seed.av.intent_minus_history\endcsname{[0.107, 0.207]}
\expandafter\gdef\csname odunum@val@keypoint_types.gap.seed.av.intent_minus_visual\endcsname{0.212}
\expandafter\gdef\csname odunum@n@keypoint_types.gap.seed.av.intent_minus_visual\endcsname{675}
\expandafter\gdef\csname odunum@ci@keypoint_types.gap.seed.av.intent_minus_visual\endcsname{[0.170, 0.255]}
\expandafter\gdef\csname odunum@val@keypoint_types.points.av.audio\endcsname{557}
\expandafter\gdef\csname odunum@n@keypoint_types.points.av.audio\endcsname{557}
\expandafter\gdef\csname odunum@val@keypoint_types.points.av.history\endcsname{475}
\expandafter\gdef\csname odunum@n@keypoint_types.points.av.history\endcsname{475}
\expandafter\gdef\csname odunum@val@keypoint_types.points.av.intent\endcsname{1\,984}
\expandafter\gdef\csname odunum@n@keypoint_types.points.av.intent\endcsname{1\,984}
\expandafter\gdef\csname odunum@val@keypoint_types.points.av.other_context\endcsname{136}
\expandafter\gdef\csname odunum@n@keypoint_types.points.av.other_context\endcsname{136}
\expandafter\gdef\csname odunum@val@keypoint_types.points.av.total\endcsname{3\,969}
\expandafter\gdef\csname odunum@n@keypoint_types.points.av.total\endcsname{3\,969}
\expandafter\gdef\csname odunum@val@keypoint_types.points.av.unattributed\endcsname{13}
\expandafter\gdef\csname odunum@n@keypoint_types.points.av.unattributed\endcsname{13}
\expandafter\gdef\csname odunum@val@keypoint_types.points.av.visual\endcsname{804}
\expandafter\gdef\csname odunum@n@keypoint_types.points.av.visual\endcsname{804}
\expandafter\gdef\csname odunum@val@keypoint_types.pop.all.n_scenes\endcsname{1\,630}
\expandafter\gdef\csname odunum@n@keypoint_types.pop.all.n_scenes\endcsname{1\,630}
\expandafter\gdef\csname odunum@val@keypoint_types.pop.ao.n_scenes\endcsname{570}
\expandafter\gdef\csname odunum@n@keypoint_types.pop.ao.n_scenes\endcsname{570}
\expandafter\gdef\csname odunum@val@keypoint_types.pop.av.n_scenes\endcsname{1\,060}
\expandafter\gdef\csname odunum@n@keypoint_types.pop.av.n_scenes\endcsname{1\,060}
\expandafter\gdef\csname odunum@val@keypoint_types.shape.models_with_gap_ci_above_zero\endcsname{4}
\expandafter\gdef\csname odunum@n@keypoint_types.shape.models_with_gap_ci_above_zero\endcsname{4}
\expandafter\gdef\csname odunum@val@keypoint_types.shape.models_with_request_above_every_context_channel\endcsname{4}
\expandafter\gdef\csname odunum@n@keypoint_types.shape.models_with_request_above_every_context_channel\endcsname{4}
\expandafter\gdef\csname odunum@val@keypoint_types.tier.cascade_asr.all.shared\endcsname{0.380}
\expandafter\gdef\csname odunum@n@keypoint_types.tier.cascade_asr.all.shared\endcsname{2\,521}
\expandafter\gdef\csname odunum@ci@keypoint_types.tier.cascade_asr.all.shared\endcsname{[0.360, 0.400]}
\expandafter\gdef\csname odunum@val@keypoint_types.tier.cascade_asr.all.t1_only\endcsname{0.799}
\expandafter\gdef\csname odunum@n@keypoint_types.tier.cascade_asr.all.t1_only\endcsname{3\,155}
\expandafter\gdef\csname odunum@ci@keypoint_types.tier.cascade_asr.all.t1_only\endcsname{[0.783, 0.814]}
\expandafter\gdef\csname odunum@val@keypoint_types.tier.cascade_asr.all.t2_only\endcsname{0.151}
\expandafter\gdef\csname odunum@n@keypoint_types.tier.cascade_asr.all.t2_only\endcsname{179}
\expandafter\gdef\csname odunum@ci@keypoint_types.tier.cascade_asr.all.t2_only\endcsname{[0.099, 0.208]}
\expandafter\gdef\csname odunum@val@keypoint_types.tier.cascade_asr.ao.shared\endcsname{0.482}
\expandafter\gdef\csname odunum@n@keypoint_types.tier.cascade_asr.ao.shared\endcsname{677}
\expandafter\gdef\csname odunum@ci@keypoint_types.tier.cascade_asr.ao.shared\endcsname{[0.443, 0.521]}
\expandafter\gdef\csname odunum@val@keypoint_types.tier.cascade_asr.ao.t1_only\endcsname{0.822}
\expandafter\gdef\csname odunum@n@keypoint_types.tier.cascade_asr.ao.t1_only\endcsname{1\,158}
\expandafter\gdef\csname odunum@ci@keypoint_types.tier.cascade_asr.ao.t1_only\endcsname{[0.797, 0.847]}
\expandafter\gdef\csname odunum@val@keypoint_types.tier.cascade_asr.ao.t2_only\endcsname{0.157}
\expandafter\gdef\csname odunum@n@keypoint_types.tier.cascade_asr.ao.t2_only\endcsname{51}
\expandafter\gdef\csname odunum@ci@keypoint_types.tier.cascade_asr.ao.t2_only\endcsname{[0.059, 0.264]}
\expandafter\gdef\csname odunum@val@keypoint_types.tier.cascade_asr.av.shared\endcsname{0.342}
\expandafter\gdef\csname odunum@n@keypoint_types.tier.cascade_asr.av.shared\endcsname{1\,844}
\expandafter\gdef\csname odunum@ci@keypoint_types.tier.cascade_asr.av.shared\endcsname{[0.319, 0.366]}
\expandafter\gdef\csname odunum@val@keypoint_types.tier.cascade_asr.av.t1_only\endcsname{0.785}
\expandafter\gdef\csname odunum@n@keypoint_types.tier.cascade_asr.av.t1_only\endcsname{1\,997}
\expandafter\gdef\csname odunum@ci@keypoint_types.tier.cascade_asr.av.t1_only\endcsname{[0.765, 0.805]}
\expandafter\gdef\csname odunum@val@keypoint_types.tier.cascade_asr.av.t2_only\endcsname{0.148}
\expandafter\gdef\csname odunum@n@keypoint_types.tier.cascade_asr.av.t2_only\endcsname{128}
\expandafter\gdef\csname odunum@ci@keypoint_types.tier.cascade_asr.av.t2_only\endcsname{[0.089, 0.216]}
\expandafter\gdef\csname odunum@val@keypoint_types.tier.gemini.all.shared\endcsname{0.473}
\expandafter\gdef\csname odunum@n@keypoint_types.tier.gemini.all.shared\endcsname{2\,521}
\expandafter\gdef\csname odunum@ci@keypoint_types.tier.gemini.all.shared\endcsname{[0.453, 0.493]}
\expandafter\gdef\csname odunum@val@keypoint_types.tier.gemini.all.t1_only\endcsname{0.829}
\expandafter\gdef\csname odunum@n@keypoint_types.tier.gemini.all.t1_only\endcsname{3\,155}
\expandafter\gdef\csname odunum@ci@keypoint_types.tier.gemini.all.t1_only\endcsname{[0.815, 0.843]}
\expandafter\gdef\csname odunum@val@keypoint_types.tier.gemini.all.t2_only\endcsname{0.218}
\expandafter\gdef\csname odunum@n@keypoint_types.tier.gemini.all.t2_only\endcsname{179}
\expandafter\gdef\csname odunum@ci@keypoint_types.tier.gemini.all.t2_only\endcsname{[0.160, 0.278]}
\expandafter\gdef\csname odunum@val@keypoint_types.tier.gemini.ao.shared\endcsname{0.496}
\expandafter\gdef\csname odunum@n@keypoint_types.tier.gemini.ao.shared\endcsname{677}
\expandafter\gdef\csname odunum@ci@keypoint_types.tier.gemini.ao.shared\endcsname{[0.457, 0.536]}
\expandafter\gdef\csname odunum@val@keypoint_types.tier.gemini.ao.t1_only\endcsname{0.835}
\expandafter\gdef\csname odunum@n@keypoint_types.tier.gemini.ao.t1_only\endcsname{1\,158}
\expandafter\gdef\csname odunum@ci@keypoint_types.tier.gemini.ao.t1_only\endcsname{[0.812, 0.858]}
\expandafter\gdef\csname odunum@val@keypoint_types.tier.gemini.ao.t2_only\endcsname{0.275}
\expandafter\gdef\csname odunum@n@keypoint_types.tier.gemini.ao.t2_only\endcsname{51}
\expandafter\gdef\csname odunum@ci@keypoint_types.tier.gemini.ao.t2_only\endcsname{[0.157, 0.404]}
\expandafter\gdef\csname odunum@val@keypoint_types.tier.gemini.av.shared\endcsname{0.464}
\expandafter\gdef\csname odunum@n@keypoint_types.tier.gemini.av.shared\endcsname{1\,844}
\expandafter\gdef\csname odunum@ci@keypoint_types.tier.gemini.av.shared\endcsname{[0.440, 0.488]}
\expandafter\gdef\csname odunum@val@keypoint_types.tier.gemini.av.t1_only\endcsname{0.825}
\expandafter\gdef\csname odunum@n@keypoint_types.tier.gemini.av.t1_only\endcsname{1\,997}
\expandafter\gdef\csname odunum@ci@keypoint_types.tier.gemini.av.t1_only\endcsname{[0.808, 0.843]}
\expandafter\gdef\csname odunum@val@keypoint_types.tier.gemini.av.t2_only\endcsname{0.195}
\expandafter\gdef\csname odunum@n@keypoint_types.tier.gemini.av.t2_only\endcsname{128}
\expandafter\gdef\csname odunum@ci@keypoint_types.tier.gemini.av.t2_only\endcsname{[0.131, 0.267]}
\expandafter\gdef\csname odunum@val@keypoint_types.tier.qwen_plus.all.shared\endcsname{0.512}
\expandafter\gdef\csname odunum@n@keypoint_types.tier.qwen_plus.all.shared\endcsname{2\,521}
\expandafter\gdef\csname odunum@ci@keypoint_types.tier.qwen_plus.all.shared\endcsname{[0.492, 0.533]}
\expandafter\gdef\csname odunum@val@keypoint_types.tier.qwen_plus.all.t1_only\endcsname{0.828}
\expandafter\gdef\csname odunum@n@keypoint_types.tier.qwen_plus.all.t1_only\endcsname{3\,155}
\expandafter\gdef\csname odunum@ci@keypoint_types.tier.qwen_plus.all.t1_only\endcsname{[0.814, 0.841]}
\expandafter\gdef\csname odunum@val@keypoint_types.tier.qwen_plus.all.t2_only\endcsname{0.402}
\expandafter\gdef\csname odunum@n@keypoint_types.tier.qwen_plus.all.t2_only\endcsname{179}
\expandafter\gdef\csname odunum@ci@keypoint_types.tier.qwen_plus.all.t2_only\endcsname{[0.331, 0.476]}
\expandafter\gdef\csname odunum@val@keypoint_types.tier.qwen_plus.ao.shared\endcsname{0.548}
\expandafter\gdef\csname odunum@n@keypoint_types.tier.qwen_plus.ao.shared\endcsname{677}
\expandafter\gdef\csname odunum@ci@keypoint_types.tier.qwen_plus.ao.shared\endcsname{[0.510, 0.586]}
\expandafter\gdef\csname odunum@val@keypoint_types.tier.qwen_plus.ao.t1_only\endcsname{0.820}
\expandafter\gdef\csname odunum@n@keypoint_types.tier.qwen_plus.ao.t1_only\endcsname{1\,158}
\expandafter\gdef\csname odunum@ci@keypoint_types.tier.qwen_plus.ao.t1_only\endcsname{[0.796, 0.843]}
\expandafter\gdef\csname odunum@val@keypoint_types.tier.qwen_plus.ao.t2_only\endcsname{0.549}
\expandafter\gdef\csname odunum@n@keypoint_types.tier.qwen_plus.ao.t2_only\endcsname{51}
\expandafter\gdef\csname odunum@ci@keypoint_types.tier.qwen_plus.ao.t2_only\endcsname{[0.415, 0.686]}
\expandafter\gdef\csname odunum@val@keypoint_types.tier.qwen_plus.av.shared\endcsname{0.499}
\expandafter\gdef\csname odunum@n@keypoint_types.tier.qwen_plus.av.shared\endcsname{1\,844}
\expandafter\gdef\csname odunum@ci@keypoint_types.tier.qwen_plus.av.shared\endcsname{[0.476, 0.523]}
\expandafter\gdef\csname odunum@val@keypoint_types.tier.qwen_plus.av.t1_only\endcsname{0.832}
\expandafter\gdef\csname odunum@n@keypoint_types.tier.qwen_plus.av.t1_only\endcsname{1\,997}
\expandafter\gdef\csname odunum@ci@keypoint_types.tier.qwen_plus.av.t1_only\endcsname{[0.815, 0.849]}
\expandafter\gdef\csname odunum@val@keypoint_types.tier.qwen_plus.av.t2_only\endcsname{0.344}
\expandafter\gdef\csname odunum@n@keypoint_types.tier.qwen_plus.av.t2_only\endcsname{128}
\expandafter\gdef\csname odunum@ci@keypoint_types.tier.qwen_plus.av.t2_only\endcsname{[0.261, 0.427]}
\expandafter\gdef\csname odunum@val@keypoint_types.tier.seed.all.shared\endcsname{0.570}
\expandafter\gdef\csname odunum@n@keypoint_types.tier.seed.all.shared\endcsname{2\,521}
\expandafter\gdef\csname odunum@ci@keypoint_types.tier.seed.all.shared\endcsname{[0.549, 0.590]}
\expandafter\gdef\csname odunum@val@keypoint_types.tier.seed.all.t1_only\endcsname{0.801}
\expandafter\gdef\csname odunum@n@keypoint_types.tier.seed.all.t1_only\endcsname{3\,155}
\expandafter\gdef\csname odunum@ci@keypoint_types.tier.seed.all.t1_only\endcsname{[0.784, 0.818]}
\expandafter\gdef\csname odunum@val@keypoint_types.tier.seed.all.t2_only\endcsname{0.240}
\expandafter\gdef\csname odunum@n@keypoint_types.tier.seed.all.t2_only\endcsname{179}
\expandafter\gdef\csname odunum@ci@keypoint_types.tier.seed.all.t2_only\endcsname{[0.180, 0.304]}
\expandafter\gdef\csname odunum@val@keypoint_types.tier.seed.ao.shared\endcsname{0.634}
\expandafter\gdef\csname odunum@n@keypoint_types.tier.seed.ao.shared\endcsname{677}
\expandafter\gdef\csname odunum@ci@keypoint_types.tier.seed.ao.shared\endcsname{[0.595, 0.671]}
\expandafter\gdef\csname odunum@val@keypoint_types.tier.seed.ao.t1_only\endcsname{0.876}
\expandafter\gdef\csname odunum@n@keypoint_types.tier.seed.ao.t1_only\endcsname{1\,158}
\expandafter\gdef\csname odunum@ci@keypoint_types.tier.seed.ao.t1_only\endcsname{[0.853, 0.897]}
\expandafter\gdef\csname odunum@val@keypoint_types.tier.seed.ao.t2_only\endcsname{0.157}
\expandafter\gdef\csname odunum@n@keypoint_types.tier.seed.ao.t2_only\endcsname{51}
\expandafter\gdef\csname odunum@ci@keypoint_types.tier.seed.ao.t2_only\endcsname{[0.061, 0.265]}
\expandafter\gdef\csname odunum@val@keypoint_types.tier.seed.av.shared\endcsname{0.546}
\expandafter\gdef\csname odunum@n@keypoint_types.tier.seed.av.shared\endcsname{1\,844}
\expandafter\gdef\csname odunum@ci@keypoint_types.tier.seed.av.shared\endcsname{[0.522, 0.570]}
\expandafter\gdef\csname odunum@val@keypoint_types.tier.seed.av.t1_only\endcsname{0.758}
\expandafter\gdef\csname odunum@n@keypoint_types.tier.seed.av.t1_only\endcsname{1\,997}
\expandafter\gdef\csname odunum@ci@keypoint_types.tier.seed.av.t1_only\endcsname{[0.735, 0.781]}
\expandafter\gdef\csname odunum@val@keypoint_types.tier.seed.av.t2_only\endcsname{0.273}
\expandafter\gdef\csname odunum@n@keypoint_types.tier.seed.av.t2_only\endcsname{128}
\expandafter\gdef\csname odunum@ci@keypoint_types.tier.seed.av.t2_only\endcsname{[0.197, 0.355]}
\expandafter\gdef\csname odunum@val@language_split.avg.generated.cascade_asr.en\endcsname{0.629}
\expandafter\gdef\csname odunum@n@language_split.avg.generated.cascade_asr.en\endcsname{907}
\expandafter\gdef\csname odunum@val@language_split.avg.generated.cascade_asr.zh\endcsname{0.663}
\expandafter\gdef\csname odunum@n@language_split.avg.generated.cascade_asr.zh\endcsname{894}
\expandafter\gdef\csname odunum@val@language_split.avg.generated.gemini.en\endcsname{0.740}
\expandafter\gdef\csname odunum@n@language_split.avg.generated.gemini.en\endcsname{907}
\expandafter\gdef\csname odunum@val@language_split.avg.generated.gemini.zh\endcsname{0.729}
\expandafter\gdef\csname odunum@n@language_split.avg.generated.gemini.zh\endcsname{894}
\expandafter\gdef\csname odunum@val@language_split.avg.generated.gemini35_flash_lite.en\endcsname{0.633}
\expandafter\gdef\csname odunum@n@language_split.avg.generated.gemini35_flash_lite.en\endcsname{907}
\expandafter\gdef\csname odunum@val@language_split.avg.generated.gemini35_flash_lite.zh\endcsname{0.584}
\expandafter\gdef\csname odunum@n@language_split.avg.generated.gemini35_flash_lite.zh\endcsname{894}
\expandafter\gdef\csname odunum@val@language_split.avg.generated.gemini37_flash.en\endcsname{0.714}
\expandafter\gdef\csname odunum@n@language_split.avg.generated.gemini37_flash.en\endcsname{905}
\expandafter\gdef\csname odunum@val@language_split.avg.generated.gemini37_flash.zh\endcsname{0.703}
\expandafter\gdef\csname odunum@n@language_split.avg.generated.gemini37_flash.zh\endcsname{887}
\expandafter\gdef\csname odunum@val@language_split.avg.generated.ming.en\endcsname{0.536}
\expandafter\gdef\csname odunum@n@language_split.avg.generated.ming.en\endcsname{907}
\expandafter\gdef\csname odunum@val@language_split.avg.generated.ming.zh\endcsname{0.529}
\expandafter\gdef\csname odunum@n@language_split.avg.generated.ming.zh\endcsname{894}
\expandafter\gdef\csname odunum@val@language_split.avg.generated.minicpm_o.en\endcsname{0.461}
\expandafter\gdef\csname odunum@n@language_split.avg.generated.minicpm_o.en\endcsname{907}
\expandafter\gdef\csname odunum@val@language_split.avg.generated.minicpm_o.zh\endcsname{0.431}
\expandafter\gdef\csname odunum@n@language_split.avg.generated.minicpm_o.zh\endcsname{891}
\expandafter\gdef\csname odunum@val@language_split.avg.generated.nemotron.en\endcsname{0.580}
\expandafter\gdef\csname odunum@n@language_split.avg.generated.nemotron.en\endcsname{907}
\expandafter\gdef\csname odunum@val@language_split.avg.generated.nemotron.zh\endcsname{0.203}
\expandafter\gdef\csname odunum@n@language_split.avg.generated.nemotron.zh\endcsname{894}
\expandafter\gdef\csname odunum@val@language_split.avg.generated.qwen25_omni.en\endcsname{0.501}
\expandafter\gdef\csname odunum@n@language_split.avg.generated.qwen25_omni.en\endcsname{907}
\expandafter\gdef\csname odunum@val@language_split.avg.generated.qwen25_omni.zh\endcsname{0.462}
\expandafter\gdef\csname odunum@n@language_split.avg.generated.qwen25_omni.zh\endcsname{894}
\expandafter\gdef\csname odunum@val@language_split.avg.generated.qwen3_omni_instruct.en\endcsname{0.584}
\expandafter\gdef\csname odunum@n@language_split.avg.generated.qwen3_omni_instruct.en\endcsname{907}
\expandafter\gdef\csname odunum@val@language_split.avg.generated.qwen3_omni_instruct.zh\endcsname{0.584}
\expandafter\gdef\csname odunum@n@language_split.avg.generated.qwen3_omni_instruct.zh\endcsname{894}
\expandafter\gdef\csname odunum@val@language_split.avg.generated.qwen3_omni_think.en\endcsname{0.644}
\expandafter\gdef\csname odunum@n@language_split.avg.generated.qwen3_omni_think.en\endcsname{907}
\expandafter\gdef\csname odunum@val@language_split.avg.generated.qwen3_omni_think.zh\endcsname{0.635}
\expandafter\gdef\csname odunum@n@language_split.avg.generated.qwen3_omni_think.zh\endcsname{894}
\expandafter\gdef\csname odunum@val@language_split.avg.generated.qwen_plus.en\endcsname{0.739}
\expandafter\gdef\csname odunum@n@language_split.avg.generated.qwen_plus.en\endcsname{907}
\expandafter\gdef\csname odunum@val@language_split.avg.generated.qwen_plus.zh\endcsname{0.706}
\expandafter\gdef\csname odunum@n@language_split.avg.generated.qwen_plus.zh\endcsname{894}
\expandafter\gdef\csname odunum@val@language_split.avg.generated.salmonn2_7b.en\endcsname{0.279}
\expandafter\gdef\csname odunum@n@language_split.avg.generated.salmonn2_7b.en\endcsname{896}
\expandafter\gdef\csname odunum@val@language_split.avg.generated.salmonn2_7b.zh\endcsname{0.099}
\expandafter\gdef\csname odunum@n@language_split.avg.generated.salmonn2_7b.zh\endcsname{889}
\expandafter\gdef\csname odunum@val@language_split.avg.generated.seed.en\endcsname{0.721}
\expandafter\gdef\csname odunum@n@language_split.avg.generated.seed.en\endcsname{907}
\expandafter\gdef\csname odunum@val@language_split.avg.generated.seed.zh\endcsname{0.712}
\expandafter\gdef\csname odunum@n@language_split.avg.generated.seed.zh\endcsname{894}
\expandafter\gdef\csname odunum@val@language_split.avg.marginal.cascade_asr.en\endcsname{0.629}
\expandafter\gdef\csname odunum@n@language_split.avg.marginal.cascade_asr.en\endcsname{908}
\expandafter\gdef\csname odunum@val@language_split.avg.marginal.cascade_asr.zh\endcsname{0.639}
\expandafter\gdef\csname odunum@n@language_split.avg.marginal.cascade_asr.zh\endcsname{1\,170}
\expandafter\gdef\csname odunum@val@language_split.avg.marginal.gemini.en\endcsname{0.740}
\expandafter\gdef\csname odunum@n@language_split.avg.marginal.gemini.en\endcsname{908}
\expandafter\gdef\csname odunum@val@language_split.avg.marginal.gemini.zh\endcsname{0.732}
\expandafter\gdef\csname odunum@n@language_split.avg.marginal.gemini.zh\endcsname{1\,170}
\expandafter\gdef\csname odunum@val@language_split.avg.marginal.gemini35_flash_lite.en\endcsname{0.633}
\expandafter\gdef\csname odunum@n@language_split.avg.marginal.gemini35_flash_lite.en\endcsname{908}
\expandafter\gdef\csname odunum@val@language_split.avg.marginal.gemini35_flash_lite.zh\endcsname{0.558}
\expandafter\gdef\csname odunum@n@language_split.avg.marginal.gemini35_flash_lite.zh\endcsname{1\,170}
\expandafter\gdef\csname odunum@val@language_split.avg.marginal.gemini37_flash.en\endcsname{0.714}
\expandafter\gdef\csname odunum@n@language_split.avg.marginal.gemini37_flash.en\endcsname{906}
\expandafter\gdef\csname odunum@val@language_split.avg.marginal.gemini37_flash.zh\endcsname{0.702}
\expandafter\gdef\csname odunum@n@language_split.avg.marginal.gemini37_flash.zh\endcsname{1\,161}
\expandafter\gdef\csname odunum@val@language_split.avg.marginal.ming.en\endcsname{0.536}
\expandafter\gdef\csname odunum@n@language_split.avg.marginal.ming.en\endcsname{908}
\expandafter\gdef\csname odunum@val@language_split.avg.marginal.ming.zh\endcsname{0.521}
\expandafter\gdef\csname odunum@n@language_split.avg.marginal.ming.zh\endcsname{1\,170}
\expandafter\gdef\csname odunum@val@language_split.avg.marginal.minicpm_o.en\endcsname{0.461}
\expandafter\gdef\csname odunum@n@language_split.avg.marginal.minicpm_o.en\endcsname{908}
\expandafter\gdef\csname odunum@val@language_split.avg.marginal.minicpm_o.zh\endcsname{0.398}
\expandafter\gdef\csname odunum@n@language_split.avg.marginal.minicpm_o.zh\endcsname{1\,167}
\expandafter\gdef\csname odunum@val@language_split.avg.marginal.nemotron.en\endcsname{0.580}
\expandafter\gdef\csname odunum@n@language_split.avg.marginal.nemotron.en\endcsname{908}
\expandafter\gdef\csname odunum@val@language_split.avg.marginal.nemotron.zh\endcsname{0.194}
\expandafter\gdef\csname odunum@n@language_split.avg.marginal.nemotron.zh\endcsname{1\,170}
\expandafter\gdef\csname odunum@val@language_split.avg.marginal.qwen25_omni.en\endcsname{0.501}
\expandafter\gdef\csname odunum@n@language_split.avg.marginal.qwen25_omni.en\endcsname{908}
\expandafter\gdef\csname odunum@val@language_split.avg.marginal.qwen25_omni.zh\endcsname{0.433}
\expandafter\gdef\csname odunum@n@language_split.avg.marginal.qwen25_omni.zh\endcsname{1\,170}
\expandafter\gdef\csname odunum@val@language_split.avg.marginal.qwen3_omni_instruct.en\endcsname{0.584}
\expandafter\gdef\csname odunum@n@language_split.avg.marginal.qwen3_omni_instruct.en\endcsname{908}
\expandafter\gdef\csname odunum@val@language_split.avg.marginal.qwen3_omni_instruct.zh\endcsname{0.568}
\expandafter\gdef\csname odunum@n@language_split.avg.marginal.qwen3_omni_instruct.zh\endcsname{1\,170}
\expandafter\gdef\csname odunum@val@language_split.avg.marginal.qwen3_omni_think.en\endcsname{0.644}
\expandafter\gdef\csname odunum@n@language_split.avg.marginal.qwen3_omni_think.en\endcsname{908}
\expandafter\gdef\csname odunum@val@language_split.avg.marginal.qwen3_omni_think.zh\endcsname{0.612}
\expandafter\gdef\csname odunum@n@language_split.avg.marginal.qwen3_omni_think.zh\endcsname{1\,170}
\expandafter\gdef\csname odunum@val@language_split.avg.marginal.qwen_plus.en\endcsname{0.739}
\expandafter\gdef\csname odunum@n@language_split.avg.marginal.qwen_plus.en\endcsname{908}
\expandafter\gdef\csname odunum@val@language_split.avg.marginal.qwen_plus.zh\endcsname{0.694}
\expandafter\gdef\csname odunum@n@language_split.avg.marginal.qwen_plus.zh\endcsname{1\,169}
\expandafter\gdef\csname odunum@val@language_split.avg.marginal.salmonn2_7b.en\endcsname{0.278}
\expandafter\gdef\csname odunum@n@language_split.avg.marginal.salmonn2_7b.en\endcsname{897}
\expandafter\gdef\csname odunum@val@language_split.avg.marginal.salmonn2_7b.zh\endcsname{0.095}
\expandafter\gdef\csname odunum@n@language_split.avg.marginal.salmonn2_7b.zh\endcsname{1\,161}
\expandafter\gdef\csname odunum@val@language_split.avg.marginal.seed.en\endcsname{0.721}
\expandafter\gdef\csname odunum@n@language_split.avg.marginal.seed.en\endcsname{908}
\expandafter\gdef\csname odunum@val@language_split.avg.marginal.seed.zh\endcsname{0.704}
\expandafter\gdef\csname odunum@n@language_split.avg.marginal.seed.zh\endcsname{1\,170}
\expandafter\gdef\csname odunum@val@language_split.cov.cascade_asr.en.audio\endcsname{0.297}
\expandafter\gdef\csname odunum@n@language_split.cov.cascade_asr.en.audio\endcsname{239}
\expandafter\gdef\csname odunum@ci@language_split.cov.cascade_asr.en.audio\endcsname{[0.238, 0.356]}
\expandafter\gdef\csname odunum@val@language_split.cov.cascade_asr.en.context\endcsname{0.303}
\expandafter\gdef\csname odunum@n@language_split.cov.cascade_asr.en.context\endcsname{811}
\expandafter\gdef\csname odunum@ci@language_split.cov.cascade_asr.en.context\endcsname{[0.270, 0.339]}
\expandafter\gdef\csname odunum@val@language_split.cov.cascade_asr.en.intent\endcsname{0.793}
\expandafter\gdef\csname odunum@n@language_split.cov.cascade_asr.en.intent\endcsname{849}
\expandafter\gdef\csname odunum@ci@language_split.cov.cascade_asr.en.intent\endcsname{[0.762, 0.822]}
\expandafter\gdef\csname odunum@val@language_split.cov.cascade_asr.en.visual\endcsname{0.204}
\expandafter\gdef\csname odunum@n@language_split.cov.cascade_asr.en.visual\endcsname{348}
\expandafter\gdef\csname odunum@ci@language_split.cov.cascade_asr.en.visual\endcsname{[0.161, 0.250]}
\expandafter\gdef\csname odunum@val@language_split.cov.cascade_asr.zh.audio\endcsname{0.387}
\expandafter\gdef\csname odunum@n@language_split.cov.cascade_asr.zh.audio\endcsname{212}
\expandafter\gdef\csname odunum@ci@language_split.cov.cascade_asr.zh.audio\endcsname{[0.321, 0.454]}
\expandafter\gdef\csname odunum@val@language_split.cov.cascade_asr.zh.context\endcsname{0.358}
\expandafter\gdef\csname odunum@n@language_split.cov.cascade_asr.zh.context\endcsname{824}
\expandafter\gdef\csname odunum@ci@language_split.cov.cascade_asr.zh.context\endcsname{[0.324, 0.392]}
\expandafter\gdef\csname odunum@val@language_split.cov.cascade_asr.zh.intent\endcsname{0.823}
\expandafter\gdef\csname odunum@n@language_split.cov.cascade_asr.zh.intent\endcsname{775}
\expandafter\gdef\csname odunum@ci@language_split.cov.cascade_asr.zh.intent\endcsname{[0.794, 0.853]}
\expandafter\gdef\csname odunum@val@language_split.cov.cascade_asr.zh.visual\endcsname{0.186}
\expandafter\gdef\csname odunum@n@language_split.cov.cascade_asr.zh.visual\endcsname{333}
\expandafter\gdef\csname odunum@ci@language_split.cov.cascade_asr.zh.visual\endcsname{[0.145, 0.230]}
\expandafter\gdef\csname odunum@val@language_split.cov.gemini.en.audio\endcsname{0.489}
\expandafter\gdef\csname odunum@n@language_split.cov.gemini.en.audio\endcsname{239}
\expandafter\gdef\csname odunum@ci@language_split.cov.gemini.en.audio\endcsname{[0.426, 0.551]}
\expandafter\gdef\csname odunum@val@language_split.cov.gemini.en.context\endcsname{0.449}
\expandafter\gdef\csname odunum@n@language_split.cov.gemini.en.context\endcsname{811}
\expandafter\gdef\csname odunum@ci@language_split.cov.gemini.en.context\endcsname{[0.414, 0.485]}
\expandafter\gdef\csname odunum@val@language_split.cov.gemini.en.intent\endcsname{0.830}
\expandafter\gdef\csname odunum@n@language_split.cov.gemini.en.intent\endcsname{849}
\expandafter\gdef\csname odunum@ci@language_split.cov.gemini.en.intent\endcsname{[0.804, 0.856]}
\expandafter\gdef\csname odunum@val@language_split.cov.gemini.en.visual\endcsname{0.486}
\expandafter\gdef\csname odunum@n@language_split.cov.gemini.en.visual\endcsname{348}
\expandafter\gdef\csname odunum@ci@language_split.cov.gemini.en.visual\endcsname{[0.431, 0.540]}
\expandafter\gdef\csname odunum@val@language_split.cov.gemini.zh.audio\endcsname{0.453}
\expandafter\gdef\csname odunum@n@language_split.cov.gemini.zh.audio\endcsname{212}
\expandafter\gdef\csname odunum@ci@language_split.cov.gemini.zh.audio\endcsname{[0.386, 0.519]}
\expandafter\gdef\csname odunum@val@language_split.cov.gemini.zh.context\endcsname{0.422}
\expandafter\gdef\csname odunum@n@language_split.cov.gemini.zh.context\endcsname{824}
\expandafter\gdef\csname odunum@ci@language_split.cov.gemini.zh.context\endcsname{[0.387, 0.458]}
\expandafter\gdef\csname odunum@val@language_split.cov.gemini.zh.intent\endcsname{0.849}
\expandafter\gdef\csname odunum@n@language_split.cov.gemini.zh.intent\endcsname{775}
\expandafter\gdef\csname odunum@ci@language_split.cov.gemini.zh.intent\endcsname{[0.823, 0.874]}
\expandafter\gdef\csname odunum@val@language_split.cov.gemini.zh.visual\endcsname{0.417}
\expandafter\gdef\csname odunum@n@language_split.cov.gemini.zh.visual\endcsname{333}
\expandafter\gdef\csname odunum@ci@language_split.cov.gemini.zh.visual\endcsname{[0.363, 0.474]}
\expandafter\gdef\csname odunum@val@language_split.cov.qwen_plus.en.audio\endcsname{0.485}
\expandafter\gdef\csname odunum@n@language_split.cov.qwen_plus.en.audio\endcsname{239}
\expandafter\gdef\csname odunum@ci@language_split.cov.qwen_plus.en.audio\endcsname{[0.422, 0.549]}
\expandafter\gdef\csname odunum@val@language_split.cov.qwen_plus.en.context\endcsname{0.509}
\expandafter\gdef\csname odunum@n@language_split.cov.qwen_plus.en.context\endcsname{811}
\expandafter\gdef\csname odunum@ci@language_split.cov.qwen_plus.en.context\endcsname{[0.473, 0.545]}
\expandafter\gdef\csname odunum@val@language_split.cov.qwen_plus.en.intent\endcsname{0.838}
\expandafter\gdef\csname odunum@n@language_split.cov.qwen_plus.en.intent\endcsname{849}
\expandafter\gdef\csname odunum@ci@language_split.cov.qwen_plus.en.intent\endcsname{[0.812, 0.862]}
\expandafter\gdef\csname odunum@val@language_split.cov.qwen_plus.en.visual\endcsname{0.491}
\expandafter\gdef\csname odunum@n@language_split.cov.qwen_plus.en.visual\endcsname{348}
\expandafter\gdef\csname odunum@ci@language_split.cov.qwen_plus.en.visual\endcsname{[0.437, 0.545]}
\expandafter\gdef\csname odunum@val@language_split.cov.qwen_plus.zh.audio\endcsname{0.462}
\expandafter\gdef\csname odunum@n@language_split.cov.qwen_plus.zh.audio\endcsname{212}
\expandafter\gdef\csname odunum@ci@language_split.cov.qwen_plus.zh.audio\endcsname{[0.394, 0.529]}
\expandafter\gdef\csname odunum@val@language_split.cov.qwen_plus.zh.context\endcsname{0.472}
\expandafter\gdef\csname odunum@n@language_split.cov.qwen_plus.zh.context\endcsname{824}
\expandafter\gdef\csname odunum@ci@language_split.cov.qwen_plus.zh.context\endcsname{[0.434, 0.509]}
\expandafter\gdef\csname odunum@val@language_split.cov.qwen_plus.zh.intent\endcsname{0.852}
\expandafter\gdef\csname odunum@n@language_split.cov.qwen_plus.zh.intent\endcsname{775}
\expandafter\gdef\csname odunum@ci@language_split.cov.qwen_plus.zh.intent\endcsname{[0.826, 0.876]}
\expandafter\gdef\csname odunum@val@language_split.cov.qwen_plus.zh.visual\endcsname{0.408}
\expandafter\gdef\csname odunum@n@language_split.cov.qwen_plus.zh.visual\endcsname{333}
\expandafter\gdef\csname odunum@ci@language_split.cov.qwen_plus.zh.visual\endcsname{[0.355, 0.463]}
\expandafter\gdef\csname odunum@val@language_split.cov.seed.en.audio\endcsname{0.444}
\expandafter\gdef\csname odunum@n@language_split.cov.seed.en.audio\endcsname{239}
\expandafter\gdef\csname odunum@ci@language_split.cov.seed.en.audio\endcsname{[0.378, 0.508]}
\expandafter\gdef\csname odunum@val@language_split.cov.seed.en.context\endcsname{0.520}
\expandafter\gdef\csname odunum@n@language_split.cov.seed.en.context\endcsname{811}
\expandafter\gdef\csname odunum@ci@language_split.cov.seed.en.context\endcsname{[0.483, 0.557]}
\expandafter\gdef\csname odunum@val@language_split.cov.seed.en.intent\endcsname{0.761}
\expandafter\gdef\csname odunum@n@language_split.cov.seed.en.intent\endcsname{849}
\expandafter\gdef\csname odunum@ci@language_split.cov.seed.en.intent\endcsname{[0.724, 0.796]}
\expandafter\gdef\csname odunum@val@language_split.cov.seed.en.visual\endcsname{0.540}
\expandafter\gdef\csname odunum@n@language_split.cov.seed.en.visual\endcsname{348}
\expandafter\gdef\csname odunum@ci@language_split.cov.seed.en.visual\endcsname{[0.485, 0.595]}
\expandafter\gdef\csname odunum@val@language_split.cov.seed.zh.audio\endcsname{0.495}
\expandafter\gdef\csname odunum@n@language_split.cov.seed.zh.audio\endcsname{212}
\expandafter\gdef\csname odunum@ci@language_split.cov.seed.zh.audio\endcsname{[0.428, 0.564]}
\expandafter\gdef\csname odunum@val@language_split.cov.seed.zh.context\endcsname{0.533}
\expandafter\gdef\csname odunum@n@language_split.cov.seed.zh.context\endcsname{824}
\expandafter\gdef\csname odunum@ci@language_split.cov.seed.zh.context\endcsname{[0.498, 0.567]}
\expandafter\gdef\csname odunum@val@language_split.cov.seed.zh.intent\endcsname{0.768}
\expandafter\gdef\csname odunum@n@language_split.cov.seed.zh.intent\endcsname{775}
\expandafter\gdef\csname odunum@ci@language_split.cov.seed.zh.intent\endcsname{[0.731, 0.803]}
\expandafter\gdef\csname odunum@val@language_split.cov.seed.zh.visual\endcsname{0.507}
\expandafter\gdef\csname odunum@n@language_split.cov.seed.zh.visual\endcsname{333}
\expandafter\gdef\csname odunum@ci@language_split.cov.seed.zh.visual\endcsname{[0.454, 0.560]}
\expandafter\gdef\csname odunum@val@language_split.delta_zh_minus_en.generated.avg.cascade_asr\endcsname{0.034}
\expandafter\gdef\csname odunum@n@language_split.delta_zh_minus_en.generated.avg.cascade_asr\endcsname{894}
\expandafter\gdef\csname odunum@val@language_split.delta_zh_minus_en.generated.avg.gemini\endcsname{-0.011}
\expandafter\gdef\csname odunum@n@language_split.delta_zh_minus_en.generated.avg.gemini\endcsname{894}
\expandafter\gdef\csname odunum@val@language_split.delta_zh_minus_en.generated.avg.gemini35_flash_lite\endcsname{-0.050}
\expandafter\gdef\csname odunum@n@language_split.delta_zh_minus_en.generated.avg.gemini35_flash_lite\endcsname{894}
\expandafter\gdef\csname odunum@val@language_split.delta_zh_minus_en.generated.avg.gemini37_flash\endcsname{-0.011}
\expandafter\gdef\csname odunum@n@language_split.delta_zh_minus_en.generated.avg.gemini37_flash\endcsname{887}
\expandafter\gdef\csname odunum@val@language_split.delta_zh_minus_en.generated.avg.ming\endcsname{-0.006}
\expandafter\gdef\csname odunum@n@language_split.delta_zh_minus_en.generated.avg.ming\endcsname{894}
\expandafter\gdef\csname odunum@val@language_split.delta_zh_minus_en.generated.avg.minicpm_o\endcsname{-0.030}
\expandafter\gdef\csname odunum@n@language_split.delta_zh_minus_en.generated.avg.minicpm_o\endcsname{891}
\expandafter\gdef\csname odunum@val@language_split.delta_zh_minus_en.generated.avg.nemotron\endcsname{-0.377}
\expandafter\gdef\csname odunum@n@language_split.delta_zh_minus_en.generated.avg.nemotron\endcsname{894}
\expandafter\gdef\csname odunum@val@language_split.delta_zh_minus_en.generated.avg.qwen25_omni\endcsname{-0.040}
\expandafter\gdef\csname odunum@n@language_split.delta_zh_minus_en.generated.avg.qwen25_omni\endcsname{894}
\expandafter\gdef\csname odunum@val@language_split.delta_zh_minus_en.generated.avg.qwen3_omni_instruct\endcsname{1.0\ensuremath{\times 10^{-4}}}
\expandafter\gdef\csname odunum@n@language_split.delta_zh_minus_en.generated.avg.qwen3_omni_instruct\endcsname{894}
\expandafter\gdef\csname odunum@val@language_split.delta_zh_minus_en.generated.avg.qwen3_omni_think\endcsname{-0.009}
\expandafter\gdef\csname odunum@n@language_split.delta_zh_minus_en.generated.avg.qwen3_omni_think\endcsname{894}
\expandafter\gdef\csname odunum@val@language_split.delta_zh_minus_en.generated.avg.qwen_plus\endcsname{-0.034}
\expandafter\gdef\csname odunum@n@language_split.delta_zh_minus_en.generated.avg.qwen_plus\endcsname{894}
\expandafter\gdef\csname odunum@val@language_split.delta_zh_minus_en.generated.avg.salmonn2_7b\endcsname{-0.180}
\expandafter\gdef\csname odunum@n@language_split.delta_zh_minus_en.generated.avg.salmonn2_7b\endcsname{889}
\expandafter\gdef\csname odunum@val@language_split.delta_zh_minus_en.generated.avg.seed\endcsname{-0.009}
\expandafter\gdef\csname odunum@n@language_split.delta_zh_minus_en.generated.avg.seed\endcsname{894}
\expandafter\gdef\csname odunum@val@language_split.delta_zh_minus_en.generated.m1.cascade_asr\endcsname{0.038}
\expandafter\gdef\csname odunum@n@language_split.delta_zh_minus_en.generated.m1.cascade_asr\endcsname{894}
\expandafter\gdef\csname odunum@val@language_split.delta_zh_minus_en.generated.m1.gemini\endcsname{0.002}
\expandafter\gdef\csname odunum@n@language_split.delta_zh_minus_en.generated.m1.gemini\endcsname{894}
\expandafter\gdef\csname odunum@val@language_split.delta_zh_minus_en.generated.m1.gemini35_flash_lite\endcsname{-0.035}
\expandafter\gdef\csname odunum@n@language_split.delta_zh_minus_en.generated.m1.gemini35_flash_lite\endcsname{894}
\expandafter\gdef\csname odunum@val@language_split.delta_zh_minus_en.generated.m1.gemini37_flash\endcsname{0.026}
\expandafter\gdef\csname odunum@n@language_split.delta_zh_minus_en.generated.m1.gemini37_flash\endcsname{887}
\expandafter\gdef\csname odunum@val@language_split.delta_zh_minus_en.generated.m1.ming\endcsname{-0.059}
\expandafter\gdef\csname odunum@n@language_split.delta_zh_minus_en.generated.m1.ming\endcsname{894}
\expandafter\gdef\csname odunum@val@language_split.delta_zh_minus_en.generated.m1.minicpm_o\endcsname{-0.112}
\expandafter\gdef\csname odunum@n@language_split.delta_zh_minus_en.generated.m1.minicpm_o\endcsname{891}
\expandafter\gdef\csname odunum@val@language_split.delta_zh_minus_en.generated.m1.nemotron\endcsname{-0.007}
\expandafter\gdef\csname odunum@n@language_split.delta_zh_minus_en.generated.m1.nemotron\endcsname{894}
\expandafter\gdef\csname odunum@val@language_split.delta_zh_minus_en.generated.m1.qwen25_omni\endcsname{-0.047}
\expandafter\gdef\csname odunum@n@language_split.delta_zh_minus_en.generated.m1.qwen25_omni\endcsname{894}
\expandafter\gdef\csname odunum@val@language_split.delta_zh_minus_en.generated.m1.qwen3_omni_instruct\endcsname{-0.067}
\expandafter\gdef\csname odunum@n@language_split.delta_zh_minus_en.generated.m1.qwen3_omni_instruct\endcsname{894}
\expandafter\gdef\csname odunum@val@language_split.delta_zh_minus_en.generated.m1.qwen3_omni_think\endcsname{-0.021}
\expandafter\gdef\csname odunum@n@language_split.delta_zh_minus_en.generated.m1.qwen3_omni_think\endcsname{894}
\expandafter\gdef\csname odunum@val@language_split.delta_zh_minus_en.generated.m1.qwen_plus\endcsname{-0.124}
\expandafter\gdef\csname odunum@n@language_split.delta_zh_minus_en.generated.m1.qwen_plus\endcsname{894}
\expandafter\gdef\csname odunum@val@language_split.delta_zh_minus_en.generated.m1.salmonn2_7b\endcsname{-0.057}
\expandafter\gdef\csname odunum@n@language_split.delta_zh_minus_en.generated.m1.salmonn2_7b\endcsname{889}
\expandafter\gdef\csname odunum@val@language_split.delta_zh_minus_en.generated.m1.seed\endcsname{-0.029}
\expandafter\gdef\csname odunum@n@language_split.delta_zh_minus_en.generated.m1.seed\endcsname{894}
\expandafter\gdef\csname odunum@val@language_split.delta_zh_minus_en.generated.m2.cascade_asr\endcsname{0.027}
\expandafter\gdef\csname odunum@n@language_split.delta_zh_minus_en.generated.m2.cascade_asr\endcsname{894}
\expandafter\gdef\csname odunum@val@language_split.delta_zh_minus_en.generated.m2.gemini\endcsname{-0.021}
\expandafter\gdef\csname odunum@n@language_split.delta_zh_minus_en.generated.m2.gemini\endcsname{894}
\expandafter\gdef\csname odunum@val@language_split.delta_zh_minus_en.generated.m2.gemini35_flash_lite\endcsname{-0.061}
\expandafter\gdef\csname odunum@n@language_split.delta_zh_minus_en.generated.m2.gemini35_flash_lite\endcsname{894}
\expandafter\gdef\csname odunum@val@language_split.delta_zh_minus_en.generated.m2.gemini37_flash\endcsname{-0.022}
\expandafter\gdef\csname odunum@n@language_split.delta_zh_minus_en.generated.m2.gemini37_flash\endcsname{887}
\expandafter\gdef\csname odunum@val@language_split.delta_zh_minus_en.generated.m2.ming\endcsname{-0.011}
\expandafter\gdef\csname odunum@n@language_split.delta_zh_minus_en.generated.m2.ming\endcsname{894}
\expandafter\gdef\csname odunum@val@language_split.delta_zh_minus_en.generated.m2.minicpm_o\endcsname{-0.010}
\expandafter\gdef\csname odunum@n@language_split.delta_zh_minus_en.generated.m2.minicpm_o\endcsname{891}
\expandafter\gdef\csname odunum@val@language_split.delta_zh_minus_en.generated.m2.nemotron\endcsname{-0.470}
\expandafter\gdef\csname odunum@n@language_split.delta_zh_minus_en.generated.m2.nemotron\endcsname{894}
\expandafter\gdef\csname odunum@val@language_split.delta_zh_minus_en.generated.m2.qwen25_omni\endcsname{-0.063}
\expandafter\gdef\csname odunum@n@language_split.delta_zh_minus_en.generated.m2.qwen25_omni\endcsname{894}
\expandafter\gdef\csname odunum@val@language_split.delta_zh_minus_en.generated.m2.qwen3_omni_instruct\endcsname{0.005}
\expandafter\gdef\csname odunum@n@language_split.delta_zh_minus_en.generated.m2.qwen3_omni_instruct\endcsname{894}
\expandafter\gdef\csname odunum@val@language_split.delta_zh_minus_en.generated.m2.qwen3_omni_think\endcsname{-0.021}
\expandafter\gdef\csname odunum@n@language_split.delta_zh_minus_en.generated.m2.qwen3_omni_think\endcsname{894}
\expandafter\gdef\csname odunum@val@language_split.delta_zh_minus_en.generated.m2.qwen_plus\endcsname{-0.022}
\expandafter\gdef\csname odunum@n@language_split.delta_zh_minus_en.generated.m2.qwen_plus\endcsname{894}
\expandafter\gdef\csname odunum@val@language_split.delta_zh_minus_en.generated.m2.salmonn2_7b\endcsname{-0.195}
\expandafter\gdef\csname odunum@n@language_split.delta_zh_minus_en.generated.m2.salmonn2_7b\endcsname{889}
\expandafter\gdef\csname odunum@val@language_split.delta_zh_minus_en.generated.m2.seed\endcsname{-0.008}
\expandafter\gdef\csname odunum@n@language_split.delta_zh_minus_en.generated.m2.seed\endcsname{894}
\expandafter\gdef\csname odunum@val@language_split.delta_zh_minus_en.generated.m3.cascade_asr\endcsname{0.048}
\expandafter\gdef\csname odunum@n@language_split.delta_zh_minus_en.generated.m3.cascade_asr\endcsname{894}
\expandafter\gdef\csname odunum@val@language_split.delta_zh_minus_en.generated.m3.gemini\endcsname{0.017}
\expandafter\gdef\csname odunum@n@language_split.delta_zh_minus_en.generated.m3.gemini\endcsname{894}
\expandafter\gdef\csname odunum@val@language_split.delta_zh_minus_en.generated.m3.gemini35_flash_lite\endcsname{0.001}
\expandafter\gdef\csname odunum@n@language_split.delta_zh_minus_en.generated.m3.gemini35_flash_lite\endcsname{894}
\expandafter\gdef\csname odunum@val@language_split.delta_zh_minus_en.generated.m3.gemini37_flash\endcsname{0.006}
\expandafter\gdef\csname odunum@n@language_split.delta_zh_minus_en.generated.m3.gemini37_flash\endcsname{887}
\expandafter\gdef\csname odunum@val@language_split.delta_zh_minus_en.generated.m3.ming\endcsname{0.029}
\expandafter\gdef\csname odunum@n@language_split.delta_zh_minus_en.generated.m3.ming\endcsname{894}
\expandafter\gdef\csname odunum@val@language_split.delta_zh_minus_en.generated.m3.minicpm_o\endcsname{-0.010}
\expandafter\gdef\csname odunum@n@language_split.delta_zh_minus_en.generated.m3.minicpm_o\endcsname{891}
\expandafter\gdef\csname odunum@val@language_split.delta_zh_minus_en.generated.m3.nemotron\endcsname{-0.103}
\expandafter\gdef\csname odunum@n@language_split.delta_zh_minus_en.generated.m3.nemotron\endcsname{894}
\expandafter\gdef\csname odunum@val@language_split.delta_zh_minus_en.generated.m3.qwen25_omni\endcsname{0.012}
\expandafter\gdef\csname odunum@n@language_split.delta_zh_minus_en.generated.m3.qwen25_omni\endcsname{894}
\expandafter\gdef\csname odunum@val@language_split.delta_zh_minus_en.generated.m3.qwen3_omni_instruct\endcsname{0.032}
\expandafter\gdef\csname odunum@n@language_split.delta_zh_minus_en.generated.m3.qwen3_omni_instruct\endcsname{894}
\expandafter\gdef\csname odunum@val@language_split.delta_zh_minus_en.generated.m3.qwen3_omni_think\endcsname{0.021}
\expandafter\gdef\csname odunum@n@language_split.delta_zh_minus_en.generated.m3.qwen3_omni_think\endcsname{894}
\expandafter\gdef\csname odunum@val@language_split.delta_zh_minus_en.generated.m3.qwen_plus\endcsname{0.022}
\expandafter\gdef\csname odunum@n@language_split.delta_zh_minus_en.generated.m3.qwen_plus\endcsname{894}
\expandafter\gdef\csname odunum@val@language_split.delta_zh_minus_en.generated.m3.salmonn2_7b\endcsname{0.018}
\expandafter\gdef\csname odunum@n@language_split.delta_zh_minus_en.generated.m3.salmonn2_7b\endcsname{889}
\expandafter\gdef\csname odunum@val@language_split.delta_zh_minus_en.generated.m3.seed\endcsname{0.009}
\expandafter\gdef\csname odunum@n@language_split.delta_zh_minus_en.generated.m3.seed\endcsname{894}
\expandafter\gdef\csname odunum@val@language_split.delta_zh_minus_en.generated.m5.cascade_asr\endcsname{0.009}
\expandafter\gdef\csname odunum@n@language_split.delta_zh_minus_en.generated.m5.cascade_asr\endcsname{894}
\expandafter\gdef\csname odunum@val@language_split.delta_zh_minus_en.generated.m5.gemini\endcsname{0.003}
\expandafter\gdef\csname odunum@n@language_split.delta_zh_minus_en.generated.m5.gemini\endcsname{894}
\expandafter\gdef\csname odunum@val@language_split.delta_zh_minus_en.generated.m5.gemini35_flash_lite\endcsname{-0.051}
\expandafter\gdef\csname odunum@n@language_split.delta_zh_minus_en.generated.m5.gemini35_flash_lite\endcsname{894}
\expandafter\gdef\csname odunum@val@language_split.delta_zh_minus_en.generated.m5.gemini37_flash\endcsname{-0.023}
\expandafter\gdef\csname odunum@n@language_split.delta_zh_minus_en.generated.m5.gemini37_flash\endcsname{887}
\expandafter\gdef\csname odunum@val@language_split.delta_zh_minus_en.generated.m5.ming\endcsname{0.057}
\expandafter\gdef\csname odunum@n@language_split.delta_zh_minus_en.generated.m5.ming\endcsname{894}
\expandafter\gdef\csname odunum@val@language_split.delta_zh_minus_en.generated.m5.minicpm_o\endcsname{-0.028}
\expandafter\gdef\csname odunum@n@language_split.delta_zh_minus_en.generated.m5.minicpm_o\endcsname{891}
\expandafter\gdef\csname odunum@val@language_split.delta_zh_minus_en.generated.m5.nemotron\endcsname{-0.296}
\expandafter\gdef\csname odunum@n@language_split.delta_zh_minus_en.generated.m5.nemotron\endcsname{894}
\expandafter\gdef\csname odunum@val@language_split.delta_zh_minus_en.generated.m5.qwen25_omni\endcsname{0.063}
\expandafter\gdef\csname odunum@n@language_split.delta_zh_minus_en.generated.m5.qwen25_omni\endcsname{894}
\expandafter\gdef\csname odunum@val@language_split.delta_zh_minus_en.generated.m5.qwen3_omni_instruct\endcsname{0.019}
\expandafter\gdef\csname odunum@n@language_split.delta_zh_minus_en.generated.m5.qwen3_omni_instruct\endcsname{894}
\expandafter\gdef\csname odunum@val@language_split.delta_zh_minus_en.generated.m5.qwen3_omni_think\endcsname{0.007}
\expandafter\gdef\csname odunum@n@language_split.delta_zh_minus_en.generated.m5.qwen3_omni_think\endcsname{894}
\expandafter\gdef\csname odunum@val@language_split.delta_zh_minus_en.generated.m5.qwen_plus\endcsname{0.002}
\expandafter\gdef\csname odunum@n@language_split.delta_zh_minus_en.generated.m5.qwen_plus\endcsname{894}
\expandafter\gdef\csname odunum@val@language_split.delta_zh_minus_en.generated.m5.salmonn2_7b\endcsname{-0.361}
\expandafter\gdef\csname odunum@n@language_split.delta_zh_minus_en.generated.m5.salmonn2_7b\endcsname{889}
\expandafter\gdef\csname odunum@val@language_split.delta_zh_minus_en.generated.m5.seed\endcsname{-0.010}
\expandafter\gdef\csname odunum@n@language_split.delta_zh_minus_en.generated.m5.seed\endcsname{894}
\expandafter\gdef\csname odunum@val@language_split.delta_zh_minus_en.marginal.avg.cascade_asr\endcsname{0.010}
\expandafter\gdef\csname odunum@n@language_split.delta_zh_minus_en.marginal.avg.cascade_asr\endcsname{908}
\expandafter\gdef\csname odunum@val@language_split.delta_zh_minus_en.marginal.avg.gemini\endcsname{-0.008}
\expandafter\gdef\csname odunum@n@language_split.delta_zh_minus_en.marginal.avg.gemini\endcsname{908}
\expandafter\gdef\csname odunum@val@language_split.delta_zh_minus_en.marginal.avg.gemini35_flash_lite\endcsname{-0.075}
\expandafter\gdef\csname odunum@n@language_split.delta_zh_minus_en.marginal.avg.gemini35_flash_lite\endcsname{908}
\expandafter\gdef\csname odunum@val@language_split.delta_zh_minus_en.marginal.avg.gemini37_flash\endcsname{-0.012}
\expandafter\gdef\csname odunum@n@language_split.delta_zh_minus_en.marginal.avg.gemini37_flash\endcsname{906}
\expandafter\gdef\csname odunum@val@language_split.delta_zh_minus_en.marginal.avg.ming\endcsname{-0.015}
\expandafter\gdef\csname odunum@n@language_split.delta_zh_minus_en.marginal.avg.ming\endcsname{908}
\expandafter\gdef\csname odunum@val@language_split.delta_zh_minus_en.marginal.avg.minicpm_o\endcsname{-0.064}
\expandafter\gdef\csname odunum@n@language_split.delta_zh_minus_en.marginal.avg.minicpm_o\endcsname{908}
\expandafter\gdef\csname odunum@val@language_split.delta_zh_minus_en.marginal.avg.nemotron\endcsname{-0.386}
\expandafter\gdef\csname odunum@n@language_split.delta_zh_minus_en.marginal.avg.nemotron\endcsname{908}
\expandafter\gdef\csname odunum@val@language_split.delta_zh_minus_en.marginal.avg.qwen25_omni\endcsname{-0.068}
\expandafter\gdef\csname odunum@n@language_split.delta_zh_minus_en.marginal.avg.qwen25_omni\endcsname{908}
\expandafter\gdef\csname odunum@val@language_split.delta_zh_minus_en.marginal.avg.qwen3_omni_instruct\endcsname{-0.017}
\expandafter\gdef\csname odunum@n@language_split.delta_zh_minus_en.marginal.avg.qwen3_omni_instruct\endcsname{908}
\expandafter\gdef\csname odunum@val@language_split.delta_zh_minus_en.marginal.avg.qwen3_omni_think\endcsname{-0.032}
\expandafter\gdef\csname odunum@n@language_split.delta_zh_minus_en.marginal.avg.qwen3_omni_think\endcsname{908}
\expandafter\gdef\csname odunum@val@language_split.delta_zh_minus_en.marginal.avg.qwen_plus\endcsname{-0.045}
\expandafter\gdef\csname odunum@n@language_split.delta_zh_minus_en.marginal.avg.qwen_plus\endcsname{908}
\expandafter\gdef\csname odunum@val@language_split.delta_zh_minus_en.marginal.avg.salmonn2_7b\endcsname{-0.183}
\expandafter\gdef\csname odunum@n@language_split.delta_zh_minus_en.marginal.avg.salmonn2_7b\endcsname{897}
\expandafter\gdef\csname odunum@val@language_split.delta_zh_minus_en.marginal.avg.seed\endcsname{-0.017}
\expandafter\gdef\csname odunum@n@language_split.delta_zh_minus_en.marginal.avg.seed\endcsname{908}
\expandafter\gdef\csname odunum@val@language_split.delta_zh_minus_en.marginal.m1.cascade_asr\endcsname{0.014}
\expandafter\gdef\csname odunum@n@language_split.delta_zh_minus_en.marginal.m1.cascade_asr\endcsname{908}
\expandafter\gdef\csname odunum@val@language_split.delta_zh_minus_en.marginal.m1.gemini\endcsname{0.007}
\expandafter\gdef\csname odunum@n@language_split.delta_zh_minus_en.marginal.m1.gemini\endcsname{908}
\expandafter\gdef\csname odunum@val@language_split.delta_zh_minus_en.marginal.m1.gemini35_flash_lite\endcsname{-0.056}
\expandafter\gdef\csname odunum@n@language_split.delta_zh_minus_en.marginal.m1.gemini35_flash_lite\endcsname{908}
\expandafter\gdef\csname odunum@val@language_split.delta_zh_minus_en.marginal.m1.gemini37_flash\endcsname{0.032}
\expandafter\gdef\csname odunum@n@language_split.delta_zh_minus_en.marginal.m1.gemini37_flash\endcsname{906}
\expandafter\gdef\csname odunum@val@language_split.delta_zh_minus_en.marginal.m1.ming\endcsname{-0.086}
\expandafter\gdef\csname odunum@n@language_split.delta_zh_minus_en.marginal.m1.ming\endcsname{908}
\expandafter\gdef\csname odunum@val@language_split.delta_zh_minus_en.marginal.m1.minicpm_o\endcsname{-0.137}
\expandafter\gdef\csname odunum@n@language_split.delta_zh_minus_en.marginal.m1.minicpm_o\endcsname{908}
\expandafter\gdef\csname odunum@val@language_split.delta_zh_minus_en.marginal.m1.nemotron\endcsname{-0.011}
\expandafter\gdef\csname odunum@n@language_split.delta_zh_minus_en.marginal.m1.nemotron\endcsname{908}
\expandafter\gdef\csname odunum@val@language_split.delta_zh_minus_en.marginal.m1.qwen25_omni\endcsname{-0.055}
\expandafter\gdef\csname odunum@n@language_split.delta_zh_minus_en.marginal.m1.qwen25_omni\endcsname{908}
\expandafter\gdef\csname odunum@val@language_split.delta_zh_minus_en.marginal.m1.qwen3_omni_instruct\endcsname{-0.092}
\expandafter\gdef\csname odunum@n@language_split.delta_zh_minus_en.marginal.m1.qwen3_omni_instruct\endcsname{908}
\expandafter\gdef\csname odunum@val@language_split.delta_zh_minus_en.marginal.m1.qwen3_omni_think\endcsname{-0.066}
\expandafter\gdef\csname odunum@n@language_split.delta_zh_minus_en.marginal.m1.qwen3_omni_think\endcsname{908}
\expandafter\gdef\csname odunum@val@language_split.delta_zh_minus_en.marginal.m1.qwen_plus\endcsname{-0.151}
\expandafter\gdef\csname odunum@n@language_split.delta_zh_minus_en.marginal.m1.qwen_plus\endcsname{908}
\expandafter\gdef\csname odunum@val@language_split.delta_zh_minus_en.marginal.m1.salmonn2_7b\endcsname{-0.062}
\expandafter\gdef\csname odunum@n@language_split.delta_zh_minus_en.marginal.m1.salmonn2_7b\endcsname{897}
\expandafter\gdef\csname odunum@val@language_split.delta_zh_minus_en.marginal.m1.seed\endcsname{-0.035}
\expandafter\gdef\csname odunum@n@language_split.delta_zh_minus_en.marginal.m1.seed\endcsname{908}
\expandafter\gdef\csname odunum@val@language_split.delta_zh_minus_en.marginal.m2.cascade_asr\endcsname{0.003}
\expandafter\gdef\csname odunum@n@language_split.delta_zh_minus_en.marginal.m2.cascade_asr\endcsname{908}
\expandafter\gdef\csname odunum@val@language_split.delta_zh_minus_en.marginal.m2.gemini\endcsname{-0.020}
\expandafter\gdef\csname odunum@n@language_split.delta_zh_minus_en.marginal.m2.gemini\endcsname{908}
\expandafter\gdef\csname odunum@val@language_split.delta_zh_minus_en.marginal.m2.gemini35_flash_lite\endcsname{-0.092}
\expandafter\gdef\csname odunum@n@language_split.delta_zh_minus_en.marginal.m2.gemini35_flash_lite\endcsname{908}
\expandafter\gdef\csname odunum@val@language_split.delta_zh_minus_en.marginal.m2.gemini37_flash\endcsname{-0.026}
\expandafter\gdef\csname odunum@n@language_split.delta_zh_minus_en.marginal.m2.gemini37_flash\endcsname{906}
\expandafter\gdef\csname odunum@val@language_split.delta_zh_minus_en.marginal.m2.ming\endcsname{-0.016}
\expandafter\gdef\csname odunum@n@language_split.delta_zh_minus_en.marginal.m2.ming\endcsname{908}
\expandafter\gdef\csname odunum@val@language_split.delta_zh_minus_en.marginal.m2.minicpm_o\endcsname{-0.048}
\expandafter\gdef\csname odunum@n@language_split.delta_zh_minus_en.marginal.m2.minicpm_o\endcsname{908}
\expandafter\gdef\csname odunum@val@language_split.delta_zh_minus_en.marginal.m2.nemotron\endcsname{-0.481}
\expandafter\gdef\csname odunum@n@language_split.delta_zh_minus_en.marginal.m2.nemotron\endcsname{908}
\expandafter\gdef\csname odunum@val@language_split.delta_zh_minus_en.marginal.m2.qwen25_omni\endcsname{-0.096}
\expandafter\gdef\csname odunum@n@language_split.delta_zh_minus_en.marginal.m2.qwen25_omni\endcsname{908}
\expandafter\gdef\csname odunum@val@language_split.delta_zh_minus_en.marginal.m2.qwen3_omni_instruct\endcsname{-0.012}
\expandafter\gdef\csname odunum@n@language_split.delta_zh_minus_en.marginal.m2.qwen3_omni_instruct\endcsname{908}
\expandafter\gdef\csname odunum@val@language_split.delta_zh_minus_en.marginal.m2.qwen3_omni_think\endcsname{-0.040}
\expandafter\gdef\csname odunum@n@language_split.delta_zh_minus_en.marginal.m2.qwen3_omni_think\endcsname{908}
\expandafter\gdef\csname odunum@val@language_split.delta_zh_minus_en.marginal.m2.qwen_plus\endcsname{-0.029}
\expandafter\gdef\csname odunum@n@language_split.delta_zh_minus_en.marginal.m2.qwen_plus\endcsname{908}
\expandafter\gdef\csname odunum@val@language_split.delta_zh_minus_en.marginal.m2.salmonn2_7b\endcsname{-0.198}
\expandafter\gdef\csname odunum@n@language_split.delta_zh_minus_en.marginal.m2.salmonn2_7b\endcsname{897}
\expandafter\gdef\csname odunum@val@language_split.delta_zh_minus_en.marginal.m2.seed\endcsname{-0.021}
\expandafter\gdef\csname odunum@n@language_split.delta_zh_minus_en.marginal.m2.seed\endcsname{908}
\expandafter\gdef\csname odunum@val@language_split.delta_zh_minus_en.marginal.m3.cascade_asr\endcsname{0.028}
\expandafter\gdef\csname odunum@n@language_split.delta_zh_minus_en.marginal.m3.cascade_asr\endcsname{908}
\expandafter\gdef\csname odunum@val@language_split.delta_zh_minus_en.marginal.m3.gemini\endcsname{0.040}
\expandafter\gdef\csname odunum@n@language_split.delta_zh_minus_en.marginal.m3.gemini\endcsname{908}
\expandafter\gdef\csname odunum@val@language_split.delta_zh_minus_en.marginal.m3.gemini35_flash_lite\endcsname{-0.001}
\expandafter\gdef\csname odunum@n@language_split.delta_zh_minus_en.marginal.m3.gemini35_flash_lite\endcsname{908}
\expandafter\gdef\csname odunum@val@language_split.delta_zh_minus_en.marginal.m3.gemini37_flash\endcsname{0.026}
\expandafter\gdef\csname odunum@n@language_split.delta_zh_minus_en.marginal.m3.gemini37_flash\endcsname{906}
\expandafter\gdef\csname odunum@val@language_split.delta_zh_minus_en.marginal.m3.ming\endcsname{0.065}
\expandafter\gdef\csname odunum@n@language_split.delta_zh_minus_en.marginal.m3.ming\endcsname{908}
\expandafter\gdef\csname odunum@val@language_split.delta_zh_minus_en.marginal.m3.minicpm_o\endcsname{-0.016}
\expandafter\gdef\csname odunum@n@language_split.delta_zh_minus_en.marginal.m3.minicpm_o\endcsname{908}
\expandafter\gdef\csname odunum@val@language_split.delta_zh_minus_en.marginal.m3.nemotron\endcsname{-0.108}
\expandafter\gdef\csname odunum@n@language_split.delta_zh_minus_en.marginal.m3.nemotron\endcsname{908}
\expandafter\gdef\csname odunum@val@language_split.delta_zh_minus_en.marginal.m3.qwen25_omni\endcsname{0.011}
\expandafter\gdef\csname odunum@n@language_split.delta_zh_minus_en.marginal.m3.qwen25_omni\endcsname{908}
\expandafter\gdef\csname odunum@val@language_split.delta_zh_minus_en.marginal.m3.qwen3_omni_instruct\endcsname{0.043}
\expandafter\gdef\csname odunum@n@language_split.delta_zh_minus_en.marginal.m3.qwen3_omni_instruct\endcsname{908}
\expandafter\gdef\csname odunum@val@language_split.delta_zh_minus_en.marginal.m3.qwen3_omni_think\endcsname{0.023}
\expandafter\gdef\csname odunum@n@language_split.delta_zh_minus_en.marginal.m3.qwen3_omni_think\endcsname{908}
\expandafter\gdef\csname odunum@val@language_split.delta_zh_minus_en.marginal.m3.qwen_plus\endcsname{0.021}
\expandafter\gdef\csname odunum@n@language_split.delta_zh_minus_en.marginal.m3.qwen_plus\endcsname{908}
\expandafter\gdef\csname odunum@val@language_split.delta_zh_minus_en.marginal.m3.salmonn2_7b\endcsname{0.012}
\expandafter\gdef\csname odunum@n@language_split.delta_zh_minus_en.marginal.m3.salmonn2_7b\endcsname{897}
\expandafter\gdef\csname odunum@val@language_split.delta_zh_minus_en.marginal.m3.seed\endcsname{0.018}
\expandafter\gdef\csname odunum@n@language_split.delta_zh_minus_en.marginal.m3.seed\endcsname{908}
\expandafter\gdef\csname odunum@val@language_split.delta_zh_minus_en.marginal.m5.cascade_asr\endcsname{-0.017}
\expandafter\gdef\csname odunum@n@language_split.delta_zh_minus_en.marginal.m5.cascade_asr\endcsname{908}
\expandafter\gdef\csname odunum@val@language_split.delta_zh_minus_en.marginal.m5.gemini\endcsname{0.007}
\expandafter\gdef\csname odunum@n@language_split.delta_zh_minus_en.marginal.m5.gemini\endcsname{908}
\expandafter\gdef\csname odunum@val@language_split.delta_zh_minus_en.marginal.m5.gemini35_flash_lite\endcsname{-0.054}
\expandafter\gdef\csname odunum@n@language_split.delta_zh_minus_en.marginal.m5.gemini35_flash_lite\endcsname{908}
\expandafter\gdef\csname odunum@val@language_split.delta_zh_minus_en.marginal.m5.gemini37_flash\endcsname{-0.019}
\expandafter\gdef\csname odunum@n@language_split.delta_zh_minus_en.marginal.m5.gemini37_flash\endcsname{906}
\expandafter\gdef\csname odunum@val@language_split.delta_zh_minus_en.marginal.m5.ming\endcsname{0.069}
\expandafter\gdef\csname odunum@n@language_split.delta_zh_minus_en.marginal.m5.ming\endcsname{908}
\expandafter\gdef\csname odunum@val@language_split.delta_zh_minus_en.marginal.m5.minicpm_o\endcsname{-0.075}
\expandafter\gdef\csname odunum@n@language_split.delta_zh_minus_en.marginal.m5.minicpm_o\endcsname{908}
\expandafter\gdef\csname odunum@val@language_split.delta_zh_minus_en.marginal.m5.nemotron\endcsname{-0.297}
\expandafter\gdef\csname odunum@n@language_split.delta_zh_minus_en.marginal.m5.nemotron\endcsname{908}
\expandafter\gdef\csname odunum@val@language_split.delta_zh_minus_en.marginal.m5.qwen25_omni\endcsname{0.045}
\expandafter\gdef\csname odunum@n@language_split.delta_zh_minus_en.marginal.m5.qwen25_omni\endcsname{908}
\expandafter\gdef\csname odunum@val@language_split.delta_zh_minus_en.marginal.m5.qwen3_omni_instruct\endcsname{0.011}
\expandafter\gdef\csname odunum@n@language_split.delta_zh_minus_en.marginal.m5.qwen3_omni_instruct\endcsname{908}
\expandafter\gdef\csname odunum@val@language_split.delta_zh_minus_en.marginal.m5.qwen3_omni_think\endcsname{0.001}
\expandafter\gdef\csname odunum@n@language_split.delta_zh_minus_en.marginal.m5.qwen3_omni_think\endcsname{908}
\expandafter\gdef\csname odunum@val@language_split.delta_zh_minus_en.marginal.m5.qwen_plus\endcsname{-0.001}
\expandafter\gdef\csname odunum@n@language_split.delta_zh_minus_en.marginal.m5.qwen_plus\endcsname{908}
\expandafter\gdef\csname odunum@val@language_split.delta_zh_minus_en.marginal.m5.salmonn2_7b\endcsname{-0.369}
\expandafter\gdef\csname odunum@n@language_split.delta_zh_minus_en.marginal.m5.salmonn2_7b\endcsname{897}
\expandafter\gdef\csname odunum@val@language_split.delta_zh_minus_en.marginal.m5.seed\endcsname{-0.007}
\expandafter\gdef\csname odunum@n@language_split.delta_zh_minus_en.marginal.m5.seed\endcsname{908}
\expandafter\gdef\csname odunum@val@language_split.ftr.generated.cascade_asr.en\endcsname{0.613}
\expandafter\gdef\csname odunum@n@language_split.ftr.generated.cascade_asr.en\endcsname{181}
\expandafter\gdef\csname odunum@ci@language_split.ftr.generated.cascade_asr.en\endcsname{[0.541, 0.681]}
\expandafter\gdef\csname odunum@val@language_split.ftr.generated.cascade_asr.zh\endcsname{0.581}
\expandafter\gdef\csname odunum@n@language_split.ftr.generated.cascade_asr.zh\endcsname{191}
\expandafter\gdef\csname odunum@ci@language_split.ftr.generated.cascade_asr.zh\endcsname{[0.510, 0.649]}
\expandafter\gdef\csname odunum@val@language_split.ftr.generated.gemini.en\endcsname{0.370}
\expandafter\gdef\csname odunum@n@language_split.ftr.generated.gemini.en\endcsname{181}
\expandafter\gdef\csname odunum@ci@language_split.ftr.generated.gemini.en\endcsname{[0.303, 0.443]}
\expandafter\gdef\csname odunum@val@language_split.ftr.generated.gemini.zh\endcsname{0.387}
\expandafter\gdef\csname odunum@n@language_split.ftr.generated.gemini.zh\endcsname{191}
\expandafter\gdef\csname odunum@ci@language_split.ftr.generated.gemini.zh\endcsname{[0.321, 0.458]}
\expandafter\gdef\csname odunum@val@language_split.ftr.generated.gemini35_flash_lite.en\endcsname{0.459}
\expandafter\gdef\csname odunum@n@language_split.ftr.generated.gemini35_flash_lite.en\endcsname{181}
\expandafter\gdef\csname odunum@ci@language_split.ftr.generated.gemini35_flash_lite.en\endcsname{[0.388, 0.531]}
\expandafter\gdef\csname odunum@val@language_split.ftr.generated.gemini35_flash_lite.zh\endcsname{0.482}
\expandafter\gdef\csname odunum@n@language_split.ftr.generated.gemini35_flash_lite.zh\endcsname{191}
\expandafter\gdef\csname odunum@ci@language_split.ftr.generated.gemini35_flash_lite.zh\endcsname{[0.412, 0.552]}
\expandafter\gdef\csname odunum@val@language_split.ftr.generated.gemini37_flash.en\endcsname{0.278}
\expandafter\gdef\csname odunum@n@language_split.ftr.generated.gemini37_flash.en\endcsname{180}
\expandafter\gdef\csname odunum@ci@language_split.ftr.generated.gemini37_flash.en\endcsname{[0.218, 0.347]}
\expandafter\gdef\csname odunum@val@language_split.ftr.generated.gemini37_flash.zh\endcsname{0.215}
\expandafter\gdef\csname odunum@n@language_split.ftr.generated.gemini37_flash.zh\endcsname{186}
\expandafter\gdef\csname odunum@ci@language_split.ftr.generated.gemini37_flash.zh\endcsname{[0.162, 0.280]}
\expandafter\gdef\csname odunum@val@language_split.ftr.generated.ming.en\endcsname{0.812}
\expandafter\gdef\csname odunum@n@language_split.ftr.generated.ming.en\endcsname{181}
\expandafter\gdef\csname odunum@ci@language_split.ftr.generated.ming.en\endcsname{[0.749, 0.862]}
\expandafter\gdef\csname odunum@val@language_split.ftr.generated.ming.zh\endcsname{0.885}
\expandafter\gdef\csname odunum@n@language_split.ftr.generated.ming.zh\endcsname{191}
\expandafter\gdef\csname odunum@ci@language_split.ftr.generated.ming.zh\endcsname{[0.832, 0.923]}
\expandafter\gdef\csname odunum@val@language_split.ftr.generated.minicpm_o.en\endcsname{0.713}
\expandafter\gdef\csname odunum@n@language_split.ftr.generated.minicpm_o.en\endcsname{181}
\expandafter\gdef\csname odunum@ci@language_split.ftr.generated.minicpm_o.en\endcsname{[0.643, 0.774]}
\expandafter\gdef\csname odunum@val@language_split.ftr.generated.minicpm_o.zh\endcsname{0.858}
\expandafter\gdef\csname odunum@n@language_split.ftr.generated.minicpm_o.zh\endcsname{190}
\expandafter\gdef\csname odunum@ci@language_split.ftr.generated.minicpm_o.zh\endcsname{[0.801, 0.900]}
\expandafter\gdef\csname odunum@val@language_split.ftr.generated.nemotron.en\endcsname{0.845}
\expandafter\gdef\csname odunum@n@language_split.ftr.generated.nemotron.en\endcsname{181}
\expandafter\gdef\csname odunum@ci@language_split.ftr.generated.nemotron.en\endcsname{[0.786, 0.891]}
\expandafter\gdef\csname odunum@val@language_split.ftr.generated.nemotron.zh\endcsname{0.649}
\expandafter\gdef\csname odunum@n@language_split.ftr.generated.nemotron.zh\endcsname{191}
\expandafter\gdef\csname odunum@ci@language_split.ftr.generated.nemotron.zh\endcsname{[0.579, 0.713]}
\expandafter\gdef\csname odunum@val@language_split.ftr.generated.qwen25_omni.en\endcsname{0.757}
\expandafter\gdef\csname odunum@n@language_split.ftr.generated.qwen25_omni.en\endcsname{181}
\expandafter\gdef\csname odunum@ci@language_split.ftr.generated.qwen25_omni.en\endcsname{[0.689, 0.814]}
\expandafter\gdef\csname odunum@val@language_split.ftr.generated.qwen25_omni.zh\endcsname{0.827}
\expandafter\gdef\csname odunum@n@language_split.ftr.generated.qwen25_omni.zh\endcsname{191}
\expandafter\gdef\csname odunum@ci@language_split.ftr.generated.qwen25_omni.zh\endcsname{[0.767, 0.874]}
\expandafter\gdef\csname odunum@val@language_split.ftr.generated.qwen3_omni_instruct.en\endcsname{0.829}
\expandafter\gdef\csname odunum@n@language_split.ftr.generated.qwen3_omni_instruct.en\endcsname{181}
\expandafter\gdef\csname odunum@ci@language_split.ftr.generated.qwen3_omni_instruct.en\endcsname{[0.767, 0.877]}
\expandafter\gdef\csname odunum@val@language_split.ftr.generated.qwen3_omni_instruct.zh\endcsname{0.906}
\expandafter\gdef\csname odunum@n@language_split.ftr.generated.qwen3_omni_instruct.zh\endcsname{191}
\expandafter\gdef\csname odunum@ci@language_split.ftr.generated.qwen3_omni_instruct.zh\endcsname{[0.856, 0.940]}
\expandafter\gdef\csname odunum@val@language_split.ftr.generated.qwen3_omni_think.en\endcsname{0.691}
\expandafter\gdef\csname odunum@n@language_split.ftr.generated.qwen3_omni_think.en\endcsname{181}
\expandafter\gdef\csname odunum@ci@language_split.ftr.generated.qwen3_omni_think.en\endcsname{[0.620, 0.753]}
\expandafter\gdef\csname odunum@val@language_split.ftr.generated.qwen3_omni_think.zh\endcsname{0.738}
\expandafter\gdef\csname odunum@n@language_split.ftr.generated.qwen3_omni_think.zh\endcsname{191}
\expandafter\gdef\csname odunum@ci@language_split.ftr.generated.qwen3_omni_think.zh\endcsname{[0.672, 0.795]}
\expandafter\gdef\csname odunum@val@language_split.ftr.generated.qwen_plus.en\endcsname{0.514}
\expandafter\gdef\csname odunum@n@language_split.ftr.generated.qwen_plus.en\endcsname{181}
\expandafter\gdef\csname odunum@ci@language_split.ftr.generated.qwen_plus.en\endcsname{[0.441, 0.586]}
\expandafter\gdef\csname odunum@val@language_split.ftr.generated.qwen_plus.zh\endcsname{0.744}
\expandafter\gdef\csname odunum@n@language_split.ftr.generated.qwen_plus.zh\endcsname{191}
\expandafter\gdef\csname odunum@ci@language_split.ftr.generated.qwen_plus.zh\endcsname{[0.677, 0.800]}
\expandafter\gdef\csname odunum@val@language_split.ftr.generated.salmonn2_7b.en\endcsname{0.883}
\expandafter\gdef\csname odunum@n@language_split.ftr.generated.salmonn2_7b.en\endcsname{179}
\expandafter\gdef\csname odunum@ci@language_split.ftr.generated.salmonn2_7b.en\endcsname{[0.827, 0.922]}
\expandafter\gdef\csname odunum@val@language_split.ftr.generated.salmonn2_7b.zh\endcsname{0.948}
\expandafter\gdef\csname odunum@n@language_split.ftr.generated.salmonn2_7b.zh\endcsname{191}
\expandafter\gdef\csname odunum@ci@language_split.ftr.generated.salmonn2_7b.zh\endcsname{[0.906, 0.971]}
\expandafter\gdef\csname odunum@val@language_split.ftr.generated.seed.en\endcsname{0.796}
\expandafter\gdef\csname odunum@n@language_split.ftr.generated.seed.en\endcsname{181}
\expandafter\gdef\csname odunum@ci@language_split.ftr.generated.seed.en\endcsname{[0.731, 0.848]}
\expandafter\gdef\csname odunum@val@language_split.ftr.generated.seed.zh\endcsname{0.827}
\expandafter\gdef\csname odunum@n@language_split.ftr.generated.seed.zh\endcsname{191}
\expandafter\gdef\csname odunum@ci@language_split.ftr.generated.seed.zh\endcsname{[0.767, 0.874]}
\expandafter\gdef\csname odunum@val@language_split.ftr.marginal.cascade_asr.en\endcsname{0.613}
\expandafter\gdef\csname odunum@n@language_split.ftr.marginal.cascade_asr.en\endcsname{181}
\expandafter\gdef\csname odunum@ci@language_split.ftr.marginal.cascade_asr.en\endcsname{[0.541, 0.681]}
\expandafter\gdef\csname odunum@val@language_split.ftr.marginal.cascade_asr.zh\endcsname{0.617}
\expandafter\gdef\csname odunum@n@language_split.ftr.marginal.cascade_asr.zh\endcsname{266}
\expandafter\gdef\csname odunum@ci@language_split.ftr.marginal.cascade_asr.zh\endcsname{[0.557, 0.673]}
\expandafter\gdef\csname odunum@val@language_split.ftr.marginal.gemini.en\endcsname{0.370}
\expandafter\gdef\csname odunum@n@language_split.ftr.marginal.gemini.en\endcsname{181}
\expandafter\gdef\csname odunum@ci@language_split.ftr.marginal.gemini.en\endcsname{[0.303, 0.443]}
\expandafter\gdef\csname odunum@val@language_split.ftr.marginal.gemini.zh\endcsname{0.376}
\expandafter\gdef\csname odunum@n@language_split.ftr.marginal.gemini.zh\endcsname{266}
\expandafter\gdef\csname odunum@ci@language_split.ftr.marginal.gemini.zh\endcsname{[0.320, 0.436]}
\expandafter\gdef\csname odunum@val@language_split.ftr.marginal.gemini35_flash_lite.en\endcsname{0.459}
\expandafter\gdef\csname odunum@n@language_split.ftr.marginal.gemini35_flash_lite.en\endcsname{181}
\expandafter\gdef\csname odunum@ci@language_split.ftr.marginal.gemini35_flash_lite.en\endcsname{[0.388, 0.531]}
\expandafter\gdef\csname odunum@val@language_split.ftr.marginal.gemini35_flash_lite.zh\endcsname{0.519}
\expandafter\gdef\csname odunum@n@language_split.ftr.marginal.gemini35_flash_lite.zh\endcsname{266}
\expandafter\gdef\csname odunum@ci@language_split.ftr.marginal.gemini35_flash_lite.zh\endcsname{[0.459, 0.578]}
\expandafter\gdef\csname odunum@val@language_split.ftr.marginal.gemini37_flash.en\endcsname{0.278}
\expandafter\gdef\csname odunum@n@language_split.ftr.marginal.gemini37_flash.en\endcsname{180}
\expandafter\gdef\csname odunum@ci@language_split.ftr.marginal.gemini37_flash.en\endcsname{[0.218, 0.347]}
\expandafter\gdef\csname odunum@val@language_split.ftr.marginal.gemini37_flash.zh\endcsname{0.204}
\expandafter\gdef\csname odunum@n@language_split.ftr.marginal.gemini37_flash.zh\endcsname{260}
\expandafter\gdef\csname odunum@ci@language_split.ftr.marginal.gemini37_flash.zh\endcsname{[0.159, 0.257]}
\expandafter\gdef\csname odunum@val@language_split.ftr.marginal.ming.en\endcsname{0.812}
\expandafter\gdef\csname odunum@n@language_split.ftr.marginal.ming.en\endcsname{181}
\expandafter\gdef\csname odunum@ci@language_split.ftr.marginal.ming.en\endcsname{[0.749, 0.862]}
\expandafter\gdef\csname odunum@val@language_split.ftr.marginal.ming.zh\endcsname{0.913}
\expandafter\gdef\csname odunum@n@language_split.ftr.marginal.ming.zh\endcsname{266}
\expandafter\gdef\csname odunum@ci@language_split.ftr.marginal.ming.zh\endcsname{[0.874, 0.942]}
\expandafter\gdef\csname odunum@val@language_split.ftr.marginal.minicpm_o.en\endcsname{0.713}
\expandafter\gdef\csname odunum@n@language_split.ftr.marginal.minicpm_o.en\endcsname{181}
\expandafter\gdef\csname odunum@ci@language_split.ftr.marginal.minicpm_o.en\endcsname{[0.643, 0.774]}
\expandafter\gdef\csname odunum@val@language_split.ftr.marginal.minicpm_o.zh\endcsname{0.883}
\expandafter\gdef\csname odunum@n@language_split.ftr.marginal.minicpm_o.zh\endcsname{265}
\expandafter\gdef\csname odunum@ci@language_split.ftr.marginal.minicpm_o.zh\endcsname{[0.839, 0.916]}
\expandafter\gdef\csname odunum@val@language_split.ftr.marginal.nemotron.en\endcsname{0.845}
\expandafter\gdef\csname odunum@n@language_split.ftr.marginal.nemotron.en\endcsname{181}
\expandafter\gdef\csname odunum@ci@language_split.ftr.marginal.nemotron.en\endcsname{[0.786, 0.891]}
\expandafter\gdef\csname odunum@val@language_split.ftr.marginal.nemotron.zh\endcsname{0.635}
\expandafter\gdef\csname odunum@n@language_split.ftr.marginal.nemotron.zh\endcsname{266}
\expandafter\gdef\csname odunum@ci@language_split.ftr.marginal.nemotron.zh\endcsname{[0.576, 0.691]}
\expandafter\gdef\csname odunum@val@language_split.ftr.marginal.qwen25_omni.en\endcsname{0.757}
\expandafter\gdef\csname odunum@n@language_split.ftr.marginal.qwen25_omni.en\endcsname{181}
\expandafter\gdef\csname odunum@ci@language_split.ftr.marginal.qwen25_omni.en\endcsname{[0.689, 0.814]}
\expandafter\gdef\csname odunum@val@language_split.ftr.marginal.qwen25_omni.zh\endcsname{0.831}
\expandafter\gdef\csname odunum@n@language_split.ftr.marginal.qwen25_omni.zh\endcsname{266}
\expandafter\gdef\csname odunum@ci@language_split.ftr.marginal.qwen25_omni.zh\endcsname{[0.781, 0.871]}
\expandafter\gdef\csname odunum@val@language_split.ftr.marginal.qwen3_omni_instruct.en\endcsname{0.829}
\expandafter\gdef\csname odunum@n@language_split.ftr.marginal.qwen3_omni_instruct.en\endcsname{181}
\expandafter\gdef\csname odunum@ci@language_split.ftr.marginal.qwen3_omni_instruct.en\endcsname{[0.767, 0.877]}
\expandafter\gdef\csname odunum@val@language_split.ftr.marginal.qwen3_omni_instruct.zh\endcsname{0.929}
\expandafter\gdef\csname odunum@n@language_split.ftr.marginal.qwen3_omni_instruct.zh\endcsname{266}
\expandafter\gdef\csname odunum@ci@language_split.ftr.marginal.qwen3_omni_instruct.zh\endcsname{[0.891, 0.954]}
\expandafter\gdef\csname odunum@val@language_split.ftr.marginal.qwen3_omni_think.en\endcsname{0.691}
\expandafter\gdef\csname odunum@n@language_split.ftr.marginal.qwen3_omni_think.en\endcsname{181}
\expandafter\gdef\csname odunum@ci@language_split.ftr.marginal.qwen3_omni_think.en\endcsname{[0.620, 0.753]}
\expandafter\gdef\csname odunum@val@language_split.ftr.marginal.qwen3_omni_think.zh\endcsname{0.789}
\expandafter\gdef\csname odunum@n@language_split.ftr.marginal.qwen3_omni_think.zh\endcsname{266}
\expandafter\gdef\csname odunum@ci@language_split.ftr.marginal.qwen3_omni_think.zh\endcsname{[0.737, 0.834]}
\expandafter\gdef\csname odunum@val@language_split.ftr.marginal.qwen_plus.en\endcsname{0.514}
\expandafter\gdef\csname odunum@n@language_split.ftr.marginal.qwen_plus.en\endcsname{181}
\expandafter\gdef\csname odunum@ci@language_split.ftr.marginal.qwen_plus.en\endcsname{[0.441, 0.586]}
\expandafter\gdef\csname odunum@val@language_split.ftr.marginal.qwen_plus.zh\endcsname{0.778}
\expandafter\gdef\csname odunum@n@language_split.ftr.marginal.qwen_plus.zh\endcsname{266}
\expandafter\gdef\csname odunum@ci@language_split.ftr.marginal.qwen_plus.zh\endcsname{[0.725, 0.824]}
\expandafter\gdef\csname odunum@val@language_split.ftr.marginal.salmonn2_7b.en\endcsname{0.883}
\expandafter\gdef\csname odunum@n@language_split.ftr.marginal.salmonn2_7b.en\endcsname{179}
\expandafter\gdef\csname odunum@ci@language_split.ftr.marginal.salmonn2_7b.en\endcsname{[0.827, 0.922]}
\expandafter\gdef\csname odunum@val@language_split.ftr.marginal.salmonn2_7b.zh\endcsname{0.947}
\expandafter\gdef\csname odunum@n@language_split.ftr.marginal.salmonn2_7b.zh\endcsname{265}
\expandafter\gdef\csname odunum@ci@language_split.ftr.marginal.salmonn2_7b.zh\endcsname{[0.913, 0.968]}
\expandafter\gdef\csname odunum@val@language_split.ftr.marginal.seed.en\endcsname{0.796}
\expandafter\gdef\csname odunum@n@language_split.ftr.marginal.seed.en\endcsname{181}
\expandafter\gdef\csname odunum@ci@language_split.ftr.marginal.seed.en\endcsname{[0.731, 0.848]}
\expandafter\gdef\csname odunum@val@language_split.ftr.marginal.seed.zh\endcsname{0.831}
\expandafter\gdef\csname odunum@n@language_split.ftr.marginal.seed.zh\endcsname{266}
\expandafter\gdef\csname odunum@ci@language_split.ftr.marginal.seed.zh\endcsname{[0.781, 0.871]}
\expandafter\gdef\csname odunum@val@language_split.gap.cascade_asr.en.intent_minus_context\endcsname{0.489}
\expandafter\gdef\csname odunum@n@language_split.gap.cascade_asr.en.intent_minus_context\endcsname{1\,660}
\expandafter\gdef\csname odunum@val@language_split.gap.cascade_asr.zh.intent_minus_context\endcsname{0.465}
\expandafter\gdef\csname odunum@n@language_split.gap.cascade_asr.zh.intent_minus_context\endcsname{1\,599}
\expandafter\gdef\csname odunum@val@language_split.gap.gemini.en.intent_minus_context\endcsname{0.382}
\expandafter\gdef\csname odunum@n@language_split.gap.gemini.en.intent_minus_context\endcsname{1\,660}
\expandafter\gdef\csname odunum@val@language_split.gap.gemini.zh.intent_minus_context\endcsname{0.427}
\expandafter\gdef\csname odunum@n@language_split.gap.gemini.zh.intent_minus_context\endcsname{1\,599}
\expandafter\gdef\csname odunum@val@language_split.gap.qwen_plus.en.intent_minus_context\endcsname{0.328}
\expandafter\gdef\csname odunum@n@language_split.gap.qwen_plus.en.intent_minus_context\endcsname{1\,660}
\expandafter\gdef\csname odunum@val@language_split.gap.qwen_plus.zh.intent_minus_context\endcsname{0.380}
\expandafter\gdef\csname odunum@n@language_split.gap.qwen_plus.zh.intent_minus_context\endcsname{1\,599}
\expandafter\gdef\csname odunum@val@language_split.gap.seed.en.intent_minus_context\endcsname{0.240}
\expandafter\gdef\csname odunum@n@language_split.gap.seed.en.intent_minus_context\endcsname{1\,660}
\expandafter\gdef\csname odunum@val@language_split.gap.seed.zh.intent_minus_context\endcsname{0.235}
\expandafter\gdef\csname odunum@n@language_split.gap.seed.zh.intent_minus_context\endcsname{1\,599}
\expandafter\gdef\csname odunum@val@language_split.m1.generated.cascade_asr.en\endcsname{0.669}
\expandafter\gdef\csname odunum@n@language_split.m1.generated.cascade_asr.en\endcsname{907}
\expandafter\gdef\csname odunum@val@language_split.m1.generated.cascade_asr.zh\endcsname{0.707}
\expandafter\gdef\csname odunum@n@language_split.m1.generated.cascade_asr.zh\endcsname{894}
\expandafter\gdef\csname odunum@val@language_split.m1.generated.gemini.en\endcsname{0.842}
\expandafter\gdef\csname odunum@n@language_split.m1.generated.gemini.en\endcsname{907}
\expandafter\gdef\csname odunum@val@language_split.m1.generated.gemini.zh\endcsname{0.844}
\expandafter\gdef\csname odunum@n@language_split.m1.generated.gemini.zh\endcsname{894}
\expandafter\gdef\csname odunum@val@language_split.m1.generated.gemini35_flash_lite.en\endcsname{0.709}
\expandafter\gdef\csname odunum@n@language_split.m1.generated.gemini35_flash_lite.en\endcsname{907}
\expandafter\gdef\csname odunum@val@language_split.m1.generated.gemini35_flash_lite.zh\endcsname{0.674}
\expandafter\gdef\csname odunum@n@language_split.m1.generated.gemini35_flash_lite.zh\endcsname{894}
\expandafter\gdef\csname odunum@val@language_split.m1.generated.gemini37_flash.en\endcsname{0.841}
\expandafter\gdef\csname odunum@n@language_split.m1.generated.gemini37_flash.en\endcsname{905}
\expandafter\gdef\csname odunum@val@language_split.m1.generated.gemini37_flash.zh\endcsname{0.867}
\expandafter\gdef\csname odunum@n@language_split.m1.generated.gemini37_flash.zh\endcsname{887}
\expandafter\gdef\csname odunum@val@language_split.m1.generated.ming.en\endcsname{0.603}
\expandafter\gdef\csname odunum@n@language_split.m1.generated.ming.en\endcsname{907}
\expandafter\gdef\csname odunum@val@language_split.m1.generated.ming.zh\endcsname{0.544}
\expandafter\gdef\csname odunum@n@language_split.m1.generated.ming.zh\endcsname{894}
\expandafter\gdef\csname odunum@val@language_split.m1.generated.minicpm_o.en\endcsname{0.660}
\expandafter\gdef\csname odunum@n@language_split.m1.generated.minicpm_o.en\endcsname{907}
\expandafter\gdef\csname odunum@val@language_split.m1.generated.minicpm_o.zh\endcsname{0.548}
\expandafter\gdef\csname odunum@n@language_split.m1.generated.minicpm_o.zh\endcsname{891}
\expandafter\gdef\csname odunum@val@language_split.m1.generated.nemotron.en\endcsname{0.568}
\expandafter\gdef\csname odunum@n@language_split.m1.generated.nemotron.en\endcsname{907}
\expandafter\gdef\csname odunum@val@language_split.m1.generated.nemotron.zh\endcsname{0.561}
\expandafter\gdef\csname odunum@n@language_split.m1.generated.nemotron.zh\endcsname{894}
\expandafter\gdef\csname odunum@val@language_split.m1.generated.qwen25_omni.en\endcsname{0.623}
\expandafter\gdef\csname odunum@n@language_split.m1.generated.qwen25_omni.en\endcsname{907}
\expandafter\gdef\csname odunum@val@language_split.m1.generated.qwen25_omni.zh\endcsname{0.576}
\expandafter\gdef\csname odunum@n@language_split.m1.generated.qwen25_omni.zh\endcsname{894}
\expandafter\gdef\csname odunum@val@language_split.m1.generated.qwen3_omni_instruct.en\endcsname{0.598}
\expandafter\gdef\csname odunum@n@language_split.m1.generated.qwen3_omni_instruct.en\endcsname{907}
\expandafter\gdef\csname odunum@val@language_split.m1.generated.qwen3_omni_instruct.zh\endcsname{0.531}
\expandafter\gdef\csname odunum@n@language_split.m1.generated.qwen3_omni_instruct.zh\endcsname{894}
\expandafter\gdef\csname odunum@val@language_split.m1.generated.qwen3_omni_think.en\endcsname{0.679}
\expandafter\gdef\csname odunum@n@language_split.m1.generated.qwen3_omni_think.en\endcsname{907}
\expandafter\gdef\csname odunum@val@language_split.m1.generated.qwen3_omni_think.zh\endcsname{0.657}
\expandafter\gdef\csname odunum@n@language_split.m1.generated.qwen3_omni_think.zh\endcsname{894}
\expandafter\gdef\csname odunum@val@language_split.m1.generated.qwen_plus.en\endcsname{0.767}
\expandafter\gdef\csname odunum@n@language_split.m1.generated.qwen_plus.en\endcsname{907}
\expandafter\gdef\csname odunum@val@language_split.m1.generated.qwen_plus.zh\endcsname{0.643}
\expandafter\gdef\csname odunum@n@language_split.m1.generated.qwen_plus.zh\endcsname{894}
\expandafter\gdef\csname odunum@val@language_split.m1.generated.salmonn2_7b.en\endcsname{0.529}
\expandafter\gdef\csname odunum@n@language_split.m1.generated.salmonn2_7b.en\endcsname{896}
\expandafter\gdef\csname odunum@val@language_split.m1.generated.salmonn2_7b.zh\endcsname{0.472}
\expandafter\gdef\csname odunum@n@language_split.m1.generated.salmonn2_7b.zh\endcsname{889}
\expandafter\gdef\csname odunum@val@language_split.m1.generated.seed.en\endcsname{0.611}
\expandafter\gdef\csname odunum@n@language_split.m1.generated.seed.en\endcsname{907}
\expandafter\gdef\csname odunum@val@language_split.m1.generated.seed.zh\endcsname{0.581}
\expandafter\gdef\csname odunum@n@language_split.m1.generated.seed.zh\endcsname{894}
\expandafter\gdef\csname odunum@val@language_split.m1.marginal.cascade_asr.en\endcsname{0.669}
\expandafter\gdef\csname odunum@n@language_split.m1.marginal.cascade_asr.en\endcsname{908}
\expandafter\gdef\csname odunum@val@language_split.m1.marginal.cascade_asr.zh\endcsname{0.683}
\expandafter\gdef\csname odunum@n@language_split.m1.marginal.cascade_asr.zh\endcsname{1\,170}
\expandafter\gdef\csname odunum@val@language_split.m1.marginal.gemini.en\endcsname{0.842}
\expandafter\gdef\csname odunum@n@language_split.m1.marginal.gemini.en\endcsname{908}
\expandafter\gdef\csname odunum@val@language_split.m1.marginal.gemini.zh\endcsname{0.849}
\expandafter\gdef\csname odunum@n@language_split.m1.marginal.gemini.zh\endcsname{1\,170}
\expandafter\gdef\csname odunum@val@language_split.m1.marginal.gemini35_flash_lite.en\endcsname{0.709}
\expandafter\gdef\csname odunum@n@language_split.m1.marginal.gemini35_flash_lite.en\endcsname{908}
\expandafter\gdef\csname odunum@val@language_split.m1.marginal.gemini35_flash_lite.zh\endcsname{0.654}
\expandafter\gdef\csname odunum@n@language_split.m1.marginal.gemini35_flash_lite.zh\endcsname{1\,170}
\expandafter\gdef\csname odunum@val@language_split.m1.marginal.gemini37_flash.en\endcsname{0.841}
\expandafter\gdef\csname odunum@n@language_split.m1.marginal.gemini37_flash.en\endcsname{906}
\expandafter\gdef\csname odunum@val@language_split.m1.marginal.gemini37_flash.zh\endcsname{0.873}
\expandafter\gdef\csname odunum@n@language_split.m1.marginal.gemini37_flash.zh\endcsname{1\,161}
\expandafter\gdef\csname odunum@val@language_split.m1.marginal.ming.en\endcsname{0.603}
\expandafter\gdef\csname odunum@n@language_split.m1.marginal.ming.en\endcsname{908}
\expandafter\gdef\csname odunum@val@language_split.m1.marginal.ming.zh\endcsname{0.516}
\expandafter\gdef\csname odunum@n@language_split.m1.marginal.ming.zh\endcsname{1\,170}
\expandafter\gdef\csname odunum@val@language_split.m1.marginal.minicpm_o.en\endcsname{0.660}
\expandafter\gdef\csname odunum@n@language_split.m1.marginal.minicpm_o.en\endcsname{908}
\expandafter\gdef\csname odunum@val@language_split.m1.marginal.minicpm_o.zh\endcsname{0.524}
\expandafter\gdef\csname odunum@n@language_split.m1.marginal.minicpm_o.zh\endcsname{1\,167}
\expandafter\gdef\csname odunum@val@language_split.m1.marginal.nemotron.en\endcsname{0.568}
\expandafter\gdef\csname odunum@n@language_split.m1.marginal.nemotron.en\endcsname{908}
\expandafter\gdef\csname odunum@val@language_split.m1.marginal.nemotron.zh\endcsname{0.557}
\expandafter\gdef\csname odunum@n@language_split.m1.marginal.nemotron.zh\endcsname{1\,170}
\expandafter\gdef\csname odunum@val@language_split.m1.marginal.qwen25_omni.en\endcsname{0.623}
\expandafter\gdef\csname odunum@n@language_split.m1.marginal.qwen25_omni.en\endcsname{908}
\expandafter\gdef\csname odunum@val@language_split.m1.marginal.qwen25_omni.zh\endcsname{0.568}
\expandafter\gdef\csname odunum@n@language_split.m1.marginal.qwen25_omni.zh\endcsname{1\,170}
\expandafter\gdef\csname odunum@val@language_split.m1.marginal.qwen3_omni_instruct.en\endcsname{0.598}
\expandafter\gdef\csname odunum@n@language_split.m1.marginal.qwen3_omni_instruct.en\endcsname{908}
\expandafter\gdef\csname odunum@val@language_split.m1.marginal.qwen3_omni_instruct.zh\endcsname{0.507}
\expandafter\gdef\csname odunum@n@language_split.m1.marginal.qwen3_omni_instruct.zh\endcsname{1\,170}
\expandafter\gdef\csname odunum@val@language_split.m1.marginal.qwen3_omni_think.en\endcsname{0.679}
\expandafter\gdef\csname odunum@n@language_split.m1.marginal.qwen3_omni_think.en\endcsname{908}
\expandafter\gdef\csname odunum@val@language_split.m1.marginal.qwen3_omni_think.zh\endcsname{0.613}
\expandafter\gdef\csname odunum@n@language_split.m1.marginal.qwen3_omni_think.zh\endcsname{1\,170}
\expandafter\gdef\csname odunum@val@language_split.m1.marginal.qwen_plus.en\endcsname{0.767}
\expandafter\gdef\csname odunum@n@language_split.m1.marginal.qwen_plus.en\endcsname{908}
\expandafter\gdef\csname odunum@val@language_split.m1.marginal.qwen_plus.zh\endcsname{0.616}
\expandafter\gdef\csname odunum@n@language_split.m1.marginal.qwen_plus.zh\endcsname{1\,169}
\expandafter\gdef\csname odunum@val@language_split.m1.marginal.salmonn2_7b.en\endcsname{0.529}
\expandafter\gdef\csname odunum@n@language_split.m1.marginal.salmonn2_7b.en\endcsname{897}
\expandafter\gdef\csname odunum@val@language_split.m1.marginal.salmonn2_7b.zh\endcsname{0.467}
\expandafter\gdef\csname odunum@n@language_split.m1.marginal.salmonn2_7b.zh\endcsname{1\,161}
\expandafter\gdef\csname odunum@val@language_split.m1.marginal.seed.en\endcsname{0.611}
\expandafter\gdef\csname odunum@n@language_split.m1.marginal.seed.en\endcsname{908}
\expandafter\gdef\csname odunum@val@language_split.m1.marginal.seed.zh\endcsname{0.576}
\expandafter\gdef\csname odunum@n@language_split.m1.marginal.seed.zh\endcsname{1\,170}
\expandafter\gdef\csname odunum@val@language_split.m2.generated.cascade_asr.en\endcsname{0.599}
\expandafter\gdef\csname odunum@n@language_split.m2.generated.cascade_asr.en\endcsname{726}
\expandafter\gdef\csname odunum@ci@language_split.m2.generated.cascade_asr.en\endcsname{[0.577, 0.621]}
\expandafter\gdef\csname odunum@val@language_split.m2.generated.cascade_asr.zh\endcsname{0.626}
\expandafter\gdef\csname odunum@n@language_split.m2.generated.cascade_asr.zh\endcsname{703}
\expandafter\gdef\csname odunum@ci@language_split.m2.generated.cascade_asr.zh\endcsname{[0.605, 0.647]}
\expandafter\gdef\csname odunum@val@language_split.m2.generated.gemini.en\endcsname{0.668}
\expandafter\gdef\csname odunum@n@language_split.m2.generated.gemini.en\endcsname{726}
\expandafter\gdef\csname odunum@ci@language_split.m2.generated.gemini.en\endcsname{[0.650, 0.686]}
\expandafter\gdef\csname odunum@val@language_split.m2.generated.gemini.zh\endcsname{0.647}
\expandafter\gdef\csname odunum@n@language_split.m2.generated.gemini.zh\endcsname{703}
\expandafter\gdef\csname odunum@ci@language_split.m2.generated.gemini.zh\endcsname{[0.629, 0.666]}
\expandafter\gdef\csname odunum@val@language_split.m2.generated.gemini35_flash_lite.en\endcsname{0.573}
\expandafter\gdef\csname odunum@n@language_split.m2.generated.gemini35_flash_lite.en\endcsname{726}
\expandafter\gdef\csname odunum@ci@language_split.m2.generated.gemini35_flash_lite.en\endcsname{[0.549, 0.596]}
\expandafter\gdef\csname odunum@val@language_split.m2.generated.gemini35_flash_lite.zh\endcsname{0.511}
\expandafter\gdef\csname odunum@n@language_split.m2.generated.gemini35_flash_lite.zh\endcsname{703}
\expandafter\gdef\csname odunum@ci@language_split.m2.generated.gemini35_flash_lite.zh\endcsname{[0.487, 0.536]}
\expandafter\gdef\csname odunum@val@language_split.m2.generated.gemini37_flash.en\endcsname{0.637}
\expandafter\gdef\csname odunum@n@language_split.m2.generated.gemini37_flash.en\endcsname{725}
\expandafter\gdef\csname odunum@ci@language_split.m2.generated.gemini37_flash.en\endcsname{[0.617, 0.657]}
\expandafter\gdef\csname odunum@val@language_split.m2.generated.gemini37_flash.zh\endcsname{0.615}
\expandafter\gdef\csname odunum@n@language_split.m2.generated.gemini37_flash.zh\endcsname{701}
\expandafter\gdef\csname odunum@ci@language_split.m2.generated.gemini37_flash.zh\endcsname{[0.595, 0.636]}
\expandafter\gdef\csname odunum@val@language_split.m2.generated.ming.en\endcsname{0.508}
\expandafter\gdef\csname odunum@n@language_split.m2.generated.ming.en\endcsname{726}
\expandafter\gdef\csname odunum@ci@language_split.m2.generated.ming.en\endcsname{[0.488, 0.528]}
\expandafter\gdef\csname odunum@val@language_split.m2.generated.ming.zh\endcsname{0.496}
\expandafter\gdef\csname odunum@n@language_split.m2.generated.ming.zh\endcsname{703}
\expandafter\gdef\csname odunum@ci@language_split.m2.generated.ming.zh\endcsname{[0.476, 0.517]}
\expandafter\gdef\csname odunum@val@language_split.m2.generated.minicpm_o.en\endcsname{0.425}
\expandafter\gdef\csname odunum@n@language_split.m2.generated.minicpm_o.en\endcsname{726}
\expandafter\gdef\csname odunum@ci@language_split.m2.generated.minicpm_o.en\endcsname{[0.402, 0.447]}
\expandafter\gdef\csname odunum@val@language_split.m2.generated.minicpm_o.zh\endcsname{0.415}
\expandafter\gdef\csname odunum@n@language_split.m2.generated.minicpm_o.zh\endcsname{701}
\expandafter\gdef\csname odunum@ci@language_split.m2.generated.minicpm_o.zh\endcsname{[0.390, 0.439]}
\expandafter\gdef\csname odunum@val@language_split.m2.generated.nemotron.en\endcsname{0.560}
\expandafter\gdef\csname odunum@n@language_split.m2.generated.nemotron.en\endcsname{726}
\expandafter\gdef\csname odunum@ci@language_split.m2.generated.nemotron.en\endcsname{[0.538, 0.581]}
\expandafter\gdef\csname odunum@val@language_split.m2.generated.nemotron.zh\endcsname{0.090}
\expandafter\gdef\csname odunum@n@language_split.m2.generated.nemotron.zh\endcsname{703}
\expandafter\gdef\csname odunum@ci@language_split.m2.generated.nemotron.zh\endcsname{[0.076, 0.104]}
\expandafter\gdef\csname odunum@val@language_split.m2.generated.qwen25_omni.en\endcsname{0.487}
\expandafter\gdef\csname odunum@n@language_split.m2.generated.qwen25_omni.en\endcsname{726}
\expandafter\gdef\csname odunum@ci@language_split.m2.generated.qwen25_omni.en\endcsname{[0.465, 0.509]}
\expandafter\gdef\csname odunum@val@language_split.m2.generated.qwen25_omni.zh\endcsname{0.424}
\expandafter\gdef\csname odunum@n@language_split.m2.generated.qwen25_omni.zh\endcsname{703}
\expandafter\gdef\csname odunum@ci@language_split.m2.generated.qwen25_omni.zh\endcsname{[0.402, 0.446]}
\expandafter\gdef\csname odunum@val@language_split.m2.generated.qwen3_omni_instruct.en\endcsname{0.565}
\expandafter\gdef\csname odunum@n@language_split.m2.generated.qwen3_omni_instruct.en\endcsname{726}
\expandafter\gdef\csname odunum@ci@language_split.m2.generated.qwen3_omni_instruct.en\endcsname{[0.545, 0.585]}
\expandafter\gdef\csname odunum@val@language_split.m2.generated.qwen3_omni_instruct.zh\endcsname{0.571}
\expandafter\gdef\csname odunum@n@language_split.m2.generated.qwen3_omni_instruct.zh\endcsname{703}
\expandafter\gdef\csname odunum@ci@language_split.m2.generated.qwen3_omni_instruct.zh\endcsname{[0.552, 0.589]}
\expandafter\gdef\csname odunum@val@language_split.m2.generated.qwen3_omni_think.en\endcsname{0.605}
\expandafter\gdef\csname odunum@n@language_split.m2.generated.qwen3_omni_think.en\endcsname{726}
\expandafter\gdef\csname odunum@ci@language_split.m2.generated.qwen3_omni_think.en\endcsname{[0.585, 0.624]}
\expandafter\gdef\csname odunum@val@language_split.m2.generated.qwen3_omni_think.zh\endcsname{0.584}
\expandafter\gdef\csname odunum@n@language_split.m2.generated.qwen3_omni_think.zh\endcsname{703}
\expandafter\gdef\csname odunum@ci@language_split.m2.generated.qwen3_omni_think.zh\endcsname{[0.565, 0.603]}
\expandafter\gdef\csname odunum@val@language_split.m2.generated.qwen_plus.en\endcsname{0.694}
\expandafter\gdef\csname odunum@n@language_split.m2.generated.qwen_plus.en\endcsname{726}
\expandafter\gdef\csname odunum@ci@language_split.m2.generated.qwen_plus.en\endcsname{[0.675, 0.713]}
\expandafter\gdef\csname odunum@val@language_split.m2.generated.qwen_plus.zh\endcsname{0.672}
\expandafter\gdef\csname odunum@n@language_split.m2.generated.qwen_plus.zh\endcsname{703}
\expandafter\gdef\csname odunum@ci@language_split.m2.generated.qwen_plus.zh\endcsname{[0.653, 0.690]}
\expandafter\gdef\csname odunum@val@language_split.m2.generated.salmonn2_7b.en\endcsname{0.218}
\expandafter\gdef\csname odunum@n@language_split.m2.generated.salmonn2_7b.en\endcsname{717}
\expandafter\gdef\csname odunum@ci@language_split.m2.generated.salmonn2_7b.en\endcsname{[0.199, 0.237]}
\expandafter\gdef\csname odunum@val@language_split.m2.generated.salmonn2_7b.zh\endcsname{0.022}
\expandafter\gdef\csname odunum@n@language_split.m2.generated.salmonn2_7b.zh\endcsname{698}
\expandafter\gdef\csname odunum@ci@language_split.m2.generated.salmonn2_7b.zh\endcsname{[0.015, 0.031]}
\expandafter\gdef\csname odunum@val@language_split.m2.generated.seed.en\endcsname{0.702}
\expandafter\gdef\csname odunum@n@language_split.m2.generated.seed.en\endcsname{726}
\expandafter\gdef\csname odunum@ci@language_split.m2.generated.seed.en\endcsname{[0.680, 0.722]}
\expandafter\gdef\csname odunum@val@language_split.m2.generated.seed.zh\endcsname{0.693}
\expandafter\gdef\csname odunum@n@language_split.m2.generated.seed.zh\endcsname{703}
\expandafter\gdef\csname odunum@ci@language_split.m2.generated.seed.zh\endcsname{[0.672, 0.714]}
\expandafter\gdef\csname odunum@val@language_split.m2.marginal.cascade_asr.en\endcsname{0.599}
\expandafter\gdef\csname odunum@n@language_split.m2.marginal.cascade_asr.en\endcsname{727}
\expandafter\gdef\csname odunum@ci@language_split.m2.marginal.cascade_asr.en\endcsname{[0.577, 0.620]}
\expandafter\gdef\csname odunum@val@language_split.m2.marginal.cascade_asr.zh\endcsname{0.601}
\expandafter\gdef\csname odunum@n@language_split.m2.marginal.cascade_asr.zh\endcsname{904}
\expandafter\gdef\csname odunum@ci@language_split.m2.marginal.cascade_asr.zh\endcsname{[0.582, 0.621]}
\expandafter\gdef\csname odunum@val@language_split.m2.marginal.gemini.en\endcsname{0.668}
\expandafter\gdef\csname odunum@n@language_split.m2.marginal.gemini.en\endcsname{727}
\expandafter\gdef\csname odunum@ci@language_split.m2.marginal.gemini.en\endcsname{[0.650, 0.686]}
\expandafter\gdef\csname odunum@val@language_split.m2.marginal.gemini.zh\endcsname{0.648}
\expandafter\gdef\csname odunum@n@language_split.m2.marginal.gemini.zh\endcsname{904}
\expandafter\gdef\csname odunum@ci@language_split.m2.marginal.gemini.zh\endcsname{[0.632, 0.665]}
\expandafter\gdef\csname odunum@val@language_split.m2.marginal.gemini35_flash_lite.en\endcsname{0.573}
\expandafter\gdef\csname odunum@n@language_split.m2.marginal.gemini35_flash_lite.en\endcsname{727}
\expandafter\gdef\csname odunum@ci@language_split.m2.marginal.gemini35_flash_lite.en\endcsname{[0.550, 0.596]}
\expandafter\gdef\csname odunum@val@language_split.m2.marginal.gemini35_flash_lite.zh\endcsname{0.480}
\expandafter\gdef\csname odunum@n@language_split.m2.marginal.gemini35_flash_lite.zh\endcsname{904}
\expandafter\gdef\csname odunum@ci@language_split.m2.marginal.gemini35_flash_lite.zh\endcsname{[0.459, 0.502]}
\expandafter\gdef\csname odunum@val@language_split.m2.marginal.gemini37_flash.en\endcsname{0.637}
\expandafter\gdef\csname odunum@n@language_split.m2.marginal.gemini37_flash.en\endcsname{726}
\expandafter\gdef\csname odunum@ci@language_split.m2.marginal.gemini37_flash.en\endcsname{[0.617, 0.657]}
\expandafter\gdef\csname odunum@val@language_split.m2.marginal.gemini37_flash.zh\endcsname{0.611}
\expandafter\gdef\csname odunum@n@language_split.m2.marginal.gemini37_flash.zh\endcsname{901}
\expandafter\gdef\csname odunum@ci@language_split.m2.marginal.gemini37_flash.zh\endcsname{[0.592, 0.629]}
\expandafter\gdef\csname odunum@val@language_split.m2.marginal.ming.en\endcsname{0.508}
\expandafter\gdef\csname odunum@n@language_split.m2.marginal.ming.en\endcsname{727}
\expandafter\gdef\csname odunum@ci@language_split.m2.marginal.ming.en\endcsname{[0.488, 0.528]}
\expandafter\gdef\csname odunum@val@language_split.m2.marginal.ming.zh\endcsname{0.492}
\expandafter\gdef\csname odunum@n@language_split.m2.marginal.ming.zh\endcsname{904}
\expandafter\gdef\csname odunum@ci@language_split.m2.marginal.ming.zh\endcsname{[0.474, 0.510]}
\expandafter\gdef\csname odunum@val@language_split.m2.marginal.minicpm_o.en\endcsname{0.425}
\expandafter\gdef\csname odunum@n@language_split.m2.marginal.minicpm_o.en\endcsname{727}
\expandafter\gdef\csname odunum@ci@language_split.m2.marginal.minicpm_o.en\endcsname{[0.402, 0.447]}
\expandafter\gdef\csname odunum@val@language_split.m2.marginal.minicpm_o.zh\endcsname{0.377}
\expandafter\gdef\csname odunum@n@language_split.m2.marginal.minicpm_o.zh\endcsname{902}
\expandafter\gdef\csname odunum@ci@language_split.m2.marginal.minicpm_o.zh\endcsname{[0.355, 0.399]}
\expandafter\gdef\csname odunum@val@language_split.m2.marginal.nemotron.en\endcsname{0.559}
\expandafter\gdef\csname odunum@n@language_split.m2.marginal.nemotron.en\endcsname{727}
\expandafter\gdef\csname odunum@ci@language_split.m2.marginal.nemotron.en\endcsname{[0.538, 0.580]}
\expandafter\gdef\csname odunum@val@language_split.m2.marginal.nemotron.zh\endcsname{0.078}
\expandafter\gdef\csname odunum@n@language_split.m2.marginal.nemotron.zh\endcsname{904}
\expandafter\gdef\csname odunum@ci@language_split.m2.marginal.nemotron.zh\endcsname{[0.066, 0.090]}
\expandafter\gdef\csname odunum@val@language_split.m2.marginal.qwen25_omni.en\endcsname{0.486}
\expandafter\gdef\csname odunum@n@language_split.m2.marginal.qwen25_omni.en\endcsname{727}
\expandafter\gdef\csname odunum@ci@language_split.m2.marginal.qwen25_omni.en\endcsname{[0.465, 0.509]}
\expandafter\gdef\csname odunum@val@language_split.m2.marginal.qwen25_omni.zh\endcsname{0.390}
\expandafter\gdef\csname odunum@n@language_split.m2.marginal.qwen25_omni.zh\endcsname{904}
\expandafter\gdef\csname odunum@ci@language_split.m2.marginal.qwen25_omni.zh\endcsname{[0.371, 0.410]}
\expandafter\gdef\csname odunum@val@language_split.m2.marginal.qwen3_omni_instruct.en\endcsname{0.565}
\expandafter\gdef\csname odunum@n@language_split.m2.marginal.qwen3_omni_instruct.en\endcsname{727}
\expandafter\gdef\csname odunum@ci@language_split.m2.marginal.qwen3_omni_instruct.en\endcsname{[0.545, 0.585]}
\expandafter\gdef\csname odunum@val@language_split.m2.marginal.qwen3_omni_instruct.zh\endcsname{0.553}
\expandafter\gdef\csname odunum@n@language_split.m2.marginal.qwen3_omni_instruct.zh\endcsname{904}
\expandafter\gdef\csname odunum@ci@language_split.m2.marginal.qwen3_omni_instruct.zh\endcsname{[0.535, 0.570]}
\expandafter\gdef\csname odunum@val@language_split.m2.marginal.qwen3_omni_think.en\endcsname{0.604}
\expandafter\gdef\csname odunum@n@language_split.m2.marginal.qwen3_omni_think.en\endcsname{727}
\expandafter\gdef\csname odunum@ci@language_split.m2.marginal.qwen3_omni_think.en\endcsname{[0.585, 0.624]}
\expandafter\gdef\csname odunum@val@language_split.m2.marginal.qwen3_omni_think.zh\endcsname{0.564}
\expandafter\gdef\csname odunum@n@language_split.m2.marginal.qwen3_omni_think.zh\endcsname{904}
\expandafter\gdef\csname odunum@ci@language_split.m2.marginal.qwen3_omni_think.zh\endcsname{[0.547, 0.582]}
\expandafter\gdef\csname odunum@val@language_split.m2.marginal.qwen_plus.en\endcsname{0.694}
\expandafter\gdef\csname odunum@n@language_split.m2.marginal.qwen_plus.en\endcsname{727}
\expandafter\gdef\csname odunum@ci@language_split.m2.marginal.qwen_plus.en\endcsname{[0.675, 0.713]}
\expandafter\gdef\csname odunum@val@language_split.m2.marginal.qwen_plus.zh\endcsname{0.665}
\expandafter\gdef\csname odunum@n@language_split.m2.marginal.qwen_plus.zh\endcsname{903}
\expandafter\gdef\csname odunum@ci@language_split.m2.marginal.qwen_plus.zh\endcsname{[0.648, 0.682]}
\expandafter\gdef\csname odunum@val@language_split.m2.marginal.salmonn2_7b.en\endcsname{0.218}
\expandafter\gdef\csname odunum@n@language_split.m2.marginal.salmonn2_7b.en\endcsname{718}
\expandafter\gdef\csname odunum@ci@language_split.m2.marginal.salmonn2_7b.en\endcsname{[0.198, 0.237]}
\expandafter\gdef\csname odunum@val@language_split.m2.marginal.salmonn2_7b.zh\endcsname{0.020}
\expandafter\gdef\csname odunum@n@language_split.m2.marginal.salmonn2_7b.zh\endcsname{896}
\expandafter\gdef\csname odunum@ci@language_split.m2.marginal.salmonn2_7b.zh\endcsname{[0.014, 0.027]}
\expandafter\gdef\csname odunum@val@language_split.m2.marginal.seed.en\endcsname{0.702}
\expandafter\gdef\csname odunum@n@language_split.m2.marginal.seed.en\endcsname{727}
\expandafter\gdef\csname odunum@ci@language_split.m2.marginal.seed.en\endcsname{[0.681, 0.722]}
\expandafter\gdef\csname odunum@val@language_split.m2.marginal.seed.zh\endcsname{0.681}
\expandafter\gdef\csname odunum@n@language_split.m2.marginal.seed.zh\endcsname{904}
\expandafter\gdef\csname odunum@ci@language_split.m2.marginal.seed.zh\endcsname{[0.661, 0.700]}
\expandafter\gdef\csname odunum@val@language_split.m3.generated.cascade_asr.en\endcsname{0.821}
\expandafter\gdef\csname odunum@n@language_split.m3.generated.cascade_asr.en\endcsname{726}
\expandafter\gdef\csname odunum@val@language_split.m3.generated.cascade_asr.zh\endcsname{0.869}
\expandafter\gdef\csname odunum@n@language_split.m3.generated.cascade_asr.zh\endcsname{703}
\expandafter\gdef\csname odunum@val@language_split.m3.generated.gemini.en\endcsname{0.726}
\expandafter\gdef\csname odunum@n@language_split.m3.generated.gemini.en\endcsname{726}
\expandafter\gdef\csname odunum@val@language_split.m3.generated.gemini.zh\endcsname{0.742}
\expandafter\gdef\csname odunum@n@language_split.m3.generated.gemini.zh\endcsname{703}
\expandafter\gdef\csname odunum@val@language_split.m3.generated.gemini35_flash_lite.en\endcsname{0.641}
\expandafter\gdef\csname odunum@n@language_split.m3.generated.gemini35_flash_lite.en\endcsname{726}
\expandafter\gdef\csname odunum@val@language_split.m3.generated.gemini35_flash_lite.zh\endcsname{0.642}
\expandafter\gdef\csname odunum@n@language_split.m3.generated.gemini35_flash_lite.zh\endcsname{703}
\expandafter\gdef\csname odunum@val@language_split.m3.generated.gemini37_flash.en\endcsname{0.717}
\expandafter\gdef\csname odunum@n@language_split.m3.generated.gemini37_flash.en\endcsname{725}
\expandafter\gdef\csname odunum@val@language_split.m3.generated.gemini37_flash.zh\endcsname{0.723}
\expandafter\gdef\csname odunum@n@language_split.m3.generated.gemini37_flash.zh\endcsname{701}
\expandafter\gdef\csname odunum@val@language_split.m3.generated.ming.en\endcsname{0.182}
\expandafter\gdef\csname odunum@n@language_split.m3.generated.ming.en\endcsname{726}
\expandafter\gdef\csname odunum@val@language_split.m3.generated.ming.zh\endcsname{0.210}
\expandafter\gdef\csname odunum@n@language_split.m3.generated.ming.zh\endcsname{703}
\expandafter\gdef\csname odunum@val@language_split.m3.generated.minicpm_o.en\endcsname{0.041}
\expandafter\gdef\csname odunum@n@language_split.m3.generated.minicpm_o.en\endcsname{726}
\expandafter\gdef\csname odunum@val@language_split.m3.generated.minicpm_o.zh\endcsname{0.031}
\expandafter\gdef\csname odunum@n@language_split.m3.generated.minicpm_o.zh\endcsname{701}
\expandafter\gdef\csname odunum@val@language_split.m3.generated.nemotron.en\endcsname{0.430}
\expandafter\gdef\csname odunum@n@language_split.m3.generated.nemotron.en\endcsname{726}
\expandafter\gdef\csname odunum@val@language_split.m3.generated.nemotron.zh\endcsname{0.328}
\expandafter\gdef\csname odunum@n@language_split.m3.generated.nemotron.zh\endcsname{703}
\expandafter\gdef\csname odunum@val@language_split.m3.generated.qwen25_omni.en\endcsname{0.051}
\expandafter\gdef\csname odunum@n@language_split.m3.generated.qwen25_omni.en\endcsname{726}
\expandafter\gdef\csname odunum@val@language_split.m3.generated.qwen25_omni.zh\endcsname{0.063}
\expandafter\gdef\csname odunum@n@language_split.m3.generated.qwen25_omni.zh\endcsname{703}
\expandafter\gdef\csname odunum@val@language_split.m3.generated.qwen3_omni_instruct.en\endcsname{0.203}
\expandafter\gdef\csname odunum@n@language_split.m3.generated.qwen3_omni_instruct.en\endcsname{726}
\expandafter\gdef\csname odunum@val@language_split.m3.generated.qwen3_omni_instruct.zh\endcsname{0.235}
\expandafter\gdef\csname odunum@n@language_split.m3.generated.qwen3_omni_instruct.zh\endcsname{703}
\expandafter\gdef\csname odunum@val@language_split.m3.generated.qwen3_omni_think.en\endcsname{0.469}
\expandafter\gdef\csname odunum@n@language_split.m3.generated.qwen3_omni_think.en\endcsname{726}
\expandafter\gdef\csname odunum@val@language_split.m3.generated.qwen3_omni_think.zh\endcsname{0.491}
\expandafter\gdef\csname odunum@n@language_split.m3.generated.qwen3_omni_think.zh\endcsname{703}
\expandafter\gdef\csname odunum@val@language_split.m3.generated.qwen_plus.en\endcsname{0.712}
\expandafter\gdef\csname odunum@n@language_split.m3.generated.qwen_plus.en\endcsname{726}
\expandafter\gdef\csname odunum@val@language_split.m3.generated.qwen_plus.zh\endcsname{0.734}
\expandafter\gdef\csname odunum@n@language_split.m3.generated.qwen_plus.zh\endcsname{703}
\expandafter\gdef\csname odunum@val@language_split.m3.generated.salmonn2_7b.en\endcsname{0.048}
\expandafter\gdef\csname odunum@n@language_split.m3.generated.salmonn2_7b.en\endcsname{717}
\expandafter\gdef\csname odunum@val@language_split.m3.generated.salmonn2_7b.zh\endcsname{0.066}
\expandafter\gdef\csname odunum@n@language_split.m3.generated.salmonn2_7b.zh\endcsname{698}
\expandafter\gdef\csname odunum@val@language_split.m3.generated.seed.en\endcsname{0.778}
\expandafter\gdef\csname odunum@n@language_split.m3.generated.seed.en\endcsname{726}
\expandafter\gdef\csname odunum@val@language_split.m3.generated.seed.zh\endcsname{0.787}
\expandafter\gdef\csname odunum@n@language_split.m3.generated.seed.zh\endcsname{703}
\expandafter\gdef\csname odunum@val@language_split.m3.marginal.cascade_asr.en\endcsname{0.821}
\expandafter\gdef\csname odunum@n@language_split.m3.marginal.cascade_asr.en\endcsname{727}
\expandafter\gdef\csname odunum@val@language_split.m3.marginal.cascade_asr.zh\endcsname{0.849}
\expandafter\gdef\csname odunum@n@language_split.m3.marginal.cascade_asr.zh\endcsname{904}
\expandafter\gdef\csname odunum@val@language_split.m3.marginal.gemini.en\endcsname{0.726}
\expandafter\gdef\csname odunum@n@language_split.m3.marginal.gemini.en\endcsname{727}
\expandafter\gdef\csname odunum@val@language_split.m3.marginal.gemini.zh\endcsname{0.766}
\expandafter\gdef\csname odunum@n@language_split.m3.marginal.gemini.zh\endcsname{904}
\expandafter\gdef\csname odunum@val@language_split.m3.marginal.gemini35_flash_lite.en\endcsname{0.642}
\expandafter\gdef\csname odunum@n@language_split.m3.marginal.gemini35_flash_lite.en\endcsname{727}
\expandafter\gdef\csname odunum@val@language_split.m3.marginal.gemini35_flash_lite.zh\endcsname{0.640}
\expandafter\gdef\csname odunum@n@language_split.m3.marginal.gemini35_flash_lite.zh\endcsname{904}
\expandafter\gdef\csname odunum@val@language_split.m3.marginal.gemini37_flash.en\endcsname{0.717}
\expandafter\gdef\csname odunum@n@language_split.m3.marginal.gemini37_flash.en\endcsname{726}
\expandafter\gdef\csname odunum@val@language_split.m3.marginal.gemini37_flash.zh\endcsname{0.743}
\expandafter\gdef\csname odunum@n@language_split.m3.marginal.gemini37_flash.zh\endcsname{901}
\expandafter\gdef\csname odunum@val@language_split.m3.marginal.ming.en\endcsname{0.182}
\expandafter\gdef\csname odunum@n@language_split.m3.marginal.ming.en\endcsname{727}
\expandafter\gdef\csname odunum@val@language_split.m3.marginal.ming.zh\endcsname{0.247}
\expandafter\gdef\csname odunum@n@language_split.m3.marginal.ming.zh\endcsname{904}
\expandafter\gdef\csname odunum@val@language_split.m3.marginal.minicpm_o.en\endcsname{0.041}
\expandafter\gdef\csname odunum@n@language_split.m3.marginal.minicpm_o.en\endcsname{727}
\expandafter\gdef\csname odunum@val@language_split.m3.marginal.minicpm_o.zh\endcsname{0.026}
\expandafter\gdef\csname odunum@n@language_split.m3.marginal.minicpm_o.zh\endcsname{902}
\expandafter\gdef\csname odunum@val@language_split.m3.marginal.nemotron.en\endcsname{0.430}
\expandafter\gdef\csname odunum@n@language_split.m3.marginal.nemotron.en\endcsname{727}
\expandafter\gdef\csname odunum@val@language_split.m3.marginal.nemotron.zh\endcsname{0.323}
\expandafter\gdef\csname odunum@n@language_split.m3.marginal.nemotron.zh\endcsname{904}
\expandafter\gdef\csname odunum@val@language_split.m3.marginal.qwen25_omni.en\endcsname{0.051}
\expandafter\gdef\csname odunum@n@language_split.m3.marginal.qwen25_omni.en\endcsname{727}
\expandafter\gdef\csname odunum@val@language_split.m3.marginal.qwen25_omni.zh\endcsname{0.062}
\expandafter\gdef\csname odunum@n@language_split.m3.marginal.qwen25_omni.zh\endcsname{904}
\expandafter\gdef\csname odunum@val@language_split.m3.marginal.qwen3_omni_instruct.en\endcsname{0.204}
\expandafter\gdef\csname odunum@n@language_split.m3.marginal.qwen3_omni_instruct.en\endcsname{727}
\expandafter\gdef\csname odunum@val@language_split.m3.marginal.qwen3_omni_instruct.zh\endcsname{0.246}
\expandafter\gdef\csname odunum@n@language_split.m3.marginal.qwen3_omni_instruct.zh\endcsname{904}
\expandafter\gdef\csname odunum@val@language_split.m3.marginal.qwen3_omni_think.en\endcsname{0.470}
\expandafter\gdef\csname odunum@n@language_split.m3.marginal.qwen3_omni_think.en\endcsname{727}
\expandafter\gdef\csname odunum@val@language_split.m3.marginal.qwen3_omni_think.zh\endcsname{0.492}
\expandafter\gdef\csname odunum@n@language_split.m3.marginal.qwen3_omni_think.zh\endcsname{904}
\expandafter\gdef\csname odunum@val@language_split.m3.marginal.qwen_plus.en\endcsname{0.712}
\expandafter\gdef\csname odunum@n@language_split.m3.marginal.qwen_plus.en\endcsname{727}
\expandafter\gdef\csname odunum@val@language_split.m3.marginal.qwen_plus.zh\endcsname{0.733}
\expandafter\gdef\csname odunum@n@language_split.m3.marginal.qwen_plus.zh\endcsname{903}
\expandafter\gdef\csname odunum@val@language_split.m3.marginal.salmonn2_7b.en\endcsname{0.048}
\expandafter\gdef\csname odunum@n@language_split.m3.marginal.salmonn2_7b.en\endcsname{718}
\expandafter\gdef\csname odunum@val@language_split.m3.marginal.salmonn2_7b.zh\endcsname{0.060}
\expandafter\gdef\csname odunum@n@language_split.m3.marginal.salmonn2_7b.zh\endcsname{896}
\expandafter\gdef\csname odunum@val@language_split.m3.marginal.seed.en\endcsname{0.778}
\expandafter\gdef\csname odunum@n@language_split.m3.marginal.seed.en\endcsname{727}
\expandafter\gdef\csname odunum@val@language_split.m3.marginal.seed.zh\endcsname{0.796}
\expandafter\gdef\csname odunum@n@language_split.m3.marginal.seed.zh\endcsname{904}
\expandafter\gdef\csname odunum@val@language_split.m4.generated.cascade_asr.en\endcsname{0.807}
\expandafter\gdef\csname odunum@n@language_split.m4.generated.cascade_asr.en\endcsname{726}
\expandafter\gdef\csname odunum@val@language_split.m4.generated.cascade_asr.zh\endcsname{0.872}
\expandafter\gdef\csname odunum@n@language_split.m4.generated.cascade_asr.zh\endcsname{703}
\expandafter\gdef\csname odunum@val@language_split.m4.generated.gemini.en\endcsname{0.922}
\expandafter\gdef\csname odunum@n@language_split.m4.generated.gemini.en\endcsname{726}
\expandafter\gdef\csname odunum@val@language_split.m4.generated.gemini.zh\endcsname{0.915}
\expandafter\gdef\csname odunum@n@language_split.m4.generated.gemini.zh\endcsname{703}
\expandafter\gdef\csname odunum@val@language_split.m4.generated.gemini35_flash_lite.en\endcsname{0.773}
\expandafter\gdef\csname odunum@n@language_split.m4.generated.gemini35_flash_lite.en\endcsname{726}
\expandafter\gdef\csname odunum@val@language_split.m4.generated.gemini35_flash_lite.zh\endcsname{0.724}
\expandafter\gdef\csname odunum@n@language_split.m4.generated.gemini35_flash_lite.zh\endcsname{703}
\expandafter\gdef\csname odunum@val@language_split.m4.generated.gemini37_flash.en\endcsname{0.883}
\expandafter\gdef\csname odunum@n@language_split.m4.generated.gemini37_flash.en\endcsname{725}
\expandafter\gdef\csname odunum@val@language_split.m4.generated.gemini37_flash.zh\endcsname{0.865}
\expandafter\gdef\csname odunum@n@language_split.m4.generated.gemini37_flash.zh\endcsname{701}
\expandafter\gdef\csname odunum@val@language_split.m4.generated.ming.en\endcsname{0.854}
\expandafter\gdef\csname odunum@n@language_split.m4.generated.ming.en\endcsname{726}
\expandafter\gdef\csname odunum@val@language_split.m4.generated.ming.zh\endcsname{0.891}
\expandafter\gdef\csname odunum@n@language_split.m4.generated.ming.zh\endcsname{703}
\expandafter\gdef\csname odunum@val@language_split.m4.generated.minicpm_o.en\endcsname{0.657}
\expandafter\gdef\csname odunum@n@language_split.m4.generated.minicpm_o.en\endcsname{726}
\expandafter\gdef\csname odunum@val@language_split.m4.generated.minicpm_o.zh\endcsname{0.606}
\expandafter\gdef\csname odunum@n@language_split.m4.generated.minicpm_o.zh\endcsname{701}
\expandafter\gdef\csname odunum@val@language_split.m4.generated.nemotron.en\endcsname{0.741}
\expandafter\gdef\csname odunum@n@language_split.m4.generated.nemotron.en\endcsname{726}
\expandafter\gdef\csname odunum@val@language_split.m4.generated.nemotron.zh\endcsname{0.051}
\expandafter\gdef\csname odunum@n@language_split.m4.generated.nemotron.zh\endcsname{703}
\expandafter\gdef\csname odunum@val@language_split.m4.generated.qwen25_omni.en\endcsname{0.756}
\expandafter\gdef\csname odunum@n@language_split.m4.generated.qwen25_omni.en\endcsname{726}
\expandafter\gdef\csname odunum@val@language_split.m4.generated.qwen25_omni.zh\endcsname{0.768}
\expandafter\gdef\csname odunum@n@language_split.m4.generated.qwen25_omni.zh\endcsname{703}
\expandafter\gdef\csname odunum@val@language_split.m4.generated.qwen3_omni_instruct.en\endcsname{0.885}
\expandafter\gdef\csname odunum@n@language_split.m4.generated.qwen3_omni_instruct.en\endcsname{726}
\expandafter\gdef\csname odunum@val@language_split.m4.generated.qwen3_omni_instruct.zh\endcsname{0.913}
\expandafter\gdef\csname odunum@n@language_split.m4.generated.qwen3_omni_instruct.zh\endcsname{703}
\expandafter\gdef\csname odunum@val@language_split.m4.generated.qwen3_omni_think.en\endcsname{0.861}
\expandafter\gdef\csname odunum@n@language_split.m4.generated.qwen3_omni_think.en\endcsname{726}
\expandafter\gdef\csname odunum@val@language_split.m4.generated.qwen3_omni_think.zh\endcsname{0.905}
\expandafter\gdef\csname odunum@n@language_split.m4.generated.qwen3_omni_think.zh\endcsname{703}
\expandafter\gdef\csname odunum@val@language_split.m4.generated.qwen_plus.en\endcsname{0.891}
\expandafter\gdef\csname odunum@n@language_split.m4.generated.qwen_plus.en\endcsname{726}
\expandafter\gdef\csname odunum@val@language_split.m4.generated.qwen_plus.zh\endcsname{0.854}
\expandafter\gdef\csname odunum@n@language_split.m4.generated.qwen_plus.zh\endcsname{703}
\expandafter\gdef\csname odunum@val@language_split.m4.generated.salmonn2_7b.en\endcsname{0.384}
\expandafter\gdef\csname odunum@n@language_split.m4.generated.salmonn2_7b.en\endcsname{717}
\expandafter\gdef\csname odunum@val@language_split.m4.generated.salmonn2_7b.zh\endcsname{0.003}
\expandafter\gdef\csname odunum@n@language_split.m4.generated.salmonn2_7b.zh\endcsname{698}
\expandafter\gdef\csname odunum@val@language_split.m4.generated.seed.en\endcsname{0.838}
\expandafter\gdef\csname odunum@n@language_split.m4.generated.seed.en\endcsname{726}
\expandafter\gdef\csname odunum@val@language_split.m4.generated.seed.zh\endcsname{0.833}
\expandafter\gdef\csname odunum@n@language_split.m4.generated.seed.zh\endcsname{703}
\expandafter\gdef\csname odunum@val@language_split.m4.marginal.cascade_asr.en\endcsname{0.807}
\expandafter\gdef\csname odunum@n@language_split.m4.marginal.cascade_asr.en\endcsname{727}
\expandafter\gdef\csname odunum@val@language_split.m4.marginal.cascade_asr.zh\endcsname{0.849}
\expandafter\gdef\csname odunum@n@language_split.m4.marginal.cascade_asr.zh\endcsname{904}
\expandafter\gdef\csname odunum@val@language_split.m4.marginal.gemini.en\endcsname{0.922}
\expandafter\gdef\csname odunum@n@language_split.m4.marginal.gemini.en\endcsname{727}
\expandafter\gdef\csname odunum@val@language_split.m4.marginal.gemini.zh\endcsname{0.908}
\expandafter\gdef\csname odunum@n@language_split.m4.marginal.gemini.zh\endcsname{904}
\expandafter\gdef\csname odunum@val@language_split.m4.marginal.gemini35_flash_lite.en\endcsname{0.773}
\expandafter\gdef\csname odunum@n@language_split.m4.marginal.gemini35_flash_lite.en\endcsname{727}
\expandafter\gdef\csname odunum@val@language_split.m4.marginal.gemini35_flash_lite.zh\endcsname{0.686}
\expandafter\gdef\csname odunum@n@language_split.m4.marginal.gemini35_flash_lite.zh\endcsname{904}
\expandafter\gdef\csname odunum@val@language_split.m4.marginal.gemini37_flash.en\endcsname{0.883}
\expandafter\gdef\csname odunum@n@language_split.m4.marginal.gemini37_flash.en\endcsname{726}
\expandafter\gdef\csname odunum@val@language_split.m4.marginal.gemini37_flash.zh\endcsname{0.855}
\expandafter\gdef\csname odunum@n@language_split.m4.marginal.gemini37_flash.zh\endcsname{901}
\expandafter\gdef\csname odunum@val@language_split.m4.marginal.ming.en\endcsname{0.854}
\expandafter\gdef\csname odunum@n@language_split.m4.marginal.ming.en\endcsname{727}
\expandafter\gdef\csname odunum@val@language_split.m4.marginal.ming.zh\endcsname{0.831}
\expandafter\gdef\csname odunum@n@language_split.m4.marginal.ming.zh\endcsname{904}
\expandafter\gdef\csname odunum@val@language_split.m4.marginal.minicpm_o.en\endcsname{0.657}
\expandafter\gdef\csname odunum@n@language_split.m4.marginal.minicpm_o.en\endcsname{727}
\expandafter\gdef\csname odunum@val@language_split.m4.marginal.minicpm_o.zh\endcsname{0.566}
\expandafter\gdef\csname odunum@n@language_split.m4.marginal.minicpm_o.zh\endcsname{902}
\expandafter\gdef\csname odunum@val@language_split.m4.marginal.nemotron.en\endcsname{0.741}
\expandafter\gdef\csname odunum@n@language_split.m4.marginal.nemotron.en\endcsname{727}
\expandafter\gdef\csname odunum@val@language_split.m4.marginal.nemotron.zh\endcsname{0.043}
\expandafter\gdef\csname odunum@n@language_split.m4.marginal.nemotron.zh\endcsname{904}
\expandafter\gdef\csname odunum@val@language_split.m4.marginal.qwen25_omni.en\endcsname{0.756}
\expandafter\gdef\csname odunum@n@language_split.m4.marginal.qwen25_omni.en\endcsname{727}
\expandafter\gdef\csname odunum@val@language_split.m4.marginal.qwen25_omni.zh\endcsname{0.700}
\expandafter\gdef\csname odunum@n@language_split.m4.marginal.qwen25_omni.zh\endcsname{904}
\expandafter\gdef\csname odunum@val@language_split.m4.marginal.qwen3_omni_instruct.en\endcsname{0.885}
\expandafter\gdef\csname odunum@n@language_split.m4.marginal.qwen3_omni_instruct.en\endcsname{727}
\expandafter\gdef\csname odunum@val@language_split.m4.marginal.qwen3_omni_instruct.zh\endcsname{0.884}
\expandafter\gdef\csname odunum@n@language_split.m4.marginal.qwen3_omni_instruct.zh\endcsname{904}
\expandafter\gdef\csname odunum@val@language_split.m4.marginal.qwen3_omni_think.en\endcsname{0.862}
\expandafter\gdef\csname odunum@n@language_split.m4.marginal.qwen3_omni_think.en\endcsname{727}
\expandafter\gdef\csname odunum@val@language_split.m4.marginal.qwen3_omni_think.zh\endcsname{0.863}
\expandafter\gdef\csname odunum@n@language_split.m4.marginal.qwen3_omni_think.zh\endcsname{904}
\expandafter\gdef\csname odunum@val@language_split.m4.marginal.qwen_plus.en\endcsname{0.891}
\expandafter\gdef\csname odunum@n@language_split.m4.marginal.qwen_plus.en\endcsname{727}
\expandafter\gdef\csname odunum@val@language_split.m4.marginal.qwen_plus.zh\endcsname{0.823}
\expandafter\gdef\csname odunum@n@language_split.m4.marginal.qwen_plus.zh\endcsname{903}
\expandafter\gdef\csname odunum@val@language_split.m4.marginal.salmonn2_7b.en\endcsname{0.383}
\expandafter\gdef\csname odunum@n@language_split.m4.marginal.salmonn2_7b.en\endcsname{718}
\expandafter\gdef\csname odunum@val@language_split.m4.marginal.salmonn2_7b.zh\endcsname{0.003}
\expandafter\gdef\csname odunum@n@language_split.m4.marginal.salmonn2_7b.zh\endcsname{896}
\expandafter\gdef\csname odunum@val@language_split.m4.marginal.seed.en\endcsname{0.838}
\expandafter\gdef\csname odunum@n@language_split.m4.marginal.seed.en\endcsname{727}
\expandafter\gdef\csname odunum@val@language_split.m4.marginal.seed.zh\endcsname{0.826}
\expandafter\gdef\csname odunum@n@language_split.m4.marginal.seed.zh\endcsname{904}
\expandafter\gdef\csname odunum@val@language_split.m5.generated.cascade_asr.en\endcsname{0.133}
\expandafter\gdef\csname odunum@n@language_split.m5.generated.cascade_asr.en\endcsname{726}
\expandafter\gdef\csname odunum@val@language_split.m5.generated.cascade_asr.zh\endcsname{0.142}
\expandafter\gdef\csname odunum@n@language_split.m5.generated.cascade_asr.zh\endcsname{703}
\expandafter\gdef\csname odunum@val@language_split.m5.generated.gemini.en\endcsname{0.959}
\expandafter\gdef\csname odunum@n@language_split.m5.generated.gemini.en\endcsname{726}
\expandafter\gdef\csname odunum@val@language_split.m5.generated.gemini.zh\endcsname{0.962}
\expandafter\gdef\csname odunum@n@language_split.m5.generated.gemini.zh\endcsname{703}
\expandafter\gdef\csname odunum@val@language_split.m5.generated.gemini35_flash_lite.en\endcsname{0.834}
\expandafter\gdef\csname odunum@n@language_split.m5.generated.gemini35_flash_lite.en\endcsname{726}
\expandafter\gdef\csname odunum@val@language_split.m5.generated.gemini35_flash_lite.zh\endcsname{0.782}
\expandafter\gdef\csname odunum@n@language_split.m5.generated.gemini35_flash_lite.zh\endcsname{703}
\expandafter\gdef\csname odunum@val@language_split.m5.generated.gemini37_flash.en\endcsname{0.914}
\expandafter\gdef\csname odunum@n@language_split.m5.generated.gemini37_flash.en\endcsname{725}
\expandafter\gdef\csname odunum@val@language_split.m5.generated.gemini37_flash.zh\endcsname{0.891}
\expandafter\gdef\csname odunum@n@language_split.m5.generated.gemini37_flash.zh\endcsname{701}
\expandafter\gdef\csname odunum@val@language_split.m5.generated.ming.en\endcsname{0.741}
\expandafter\gdef\csname odunum@n@language_split.m5.generated.ming.en\endcsname{726}
\expandafter\gdef\csname odunum@val@language_split.m5.generated.ming.zh\endcsname{0.797}
\expandafter\gdef\csname odunum@n@language_split.m5.generated.ming.zh\endcsname{703}
\expandafter\gdef\csname odunum@val@language_split.m5.generated.minicpm_o.en\endcsname{0.750}
\expandafter\gdef\csname odunum@n@language_split.m5.generated.minicpm_o.en\endcsname{726}
\expandafter\gdef\csname odunum@val@language_split.m5.generated.minicpm_o.zh\endcsname{0.721}
\expandafter\gdef\csname odunum@n@language_split.m5.generated.minicpm_o.zh\endcsname{701}
\expandafter\gdef\csname odunum@val@language_split.m5.generated.nemotron.en\endcsname{0.842}
\expandafter\gdef\csname odunum@n@language_split.m5.generated.nemotron.en\endcsname{726}
\expandafter\gdef\csname odunum@val@language_split.m5.generated.nemotron.zh\endcsname{0.546}
\expandafter\gdef\csname odunum@n@language_split.m5.generated.nemotron.zh\endcsname{703}
\expandafter\gdef\csname odunum@val@language_split.m5.generated.qwen25_omni.en\endcsname{0.696}
\expandafter\gdef\csname odunum@n@language_split.m5.generated.qwen25_omni.en\endcsname{726}
\expandafter\gdef\csname odunum@val@language_split.m5.generated.qwen25_omni.zh\endcsname{0.758}
\expandafter\gdef\csname odunum@n@language_split.m5.generated.qwen25_omni.zh\endcsname{703}
\expandafter\gdef\csname odunum@val@language_split.m5.generated.qwen3_omni_instruct.en\endcsname{0.931}
\expandafter\gdef\csname odunum@n@language_split.m5.generated.qwen3_omni_instruct.en\endcsname{726}
\expandafter\gdef\csname odunum@val@language_split.m5.generated.qwen3_omni_instruct.zh\endcsname{0.950}
\expandafter\gdef\csname odunum@n@language_split.m5.generated.qwen3_omni_instruct.zh\endcsname{703}
\expandafter\gdef\csname odunum@val@language_split.m5.generated.qwen3_omni_think.en\endcsname{0.929}
\expandafter\gdef\csname odunum@n@language_split.m5.generated.qwen3_omni_think.en\endcsname{726}
\expandafter\gdef\csname odunum@val@language_split.m5.generated.qwen3_omni_think.zh\endcsname{0.937}
\expandafter\gdef\csname odunum@n@language_split.m5.generated.qwen3_omni_think.zh\endcsname{703}
\expandafter\gdef\csname odunum@val@language_split.m5.generated.qwen_plus.en\endcsname{0.943}
\expandafter\gdef\csname odunum@n@language_split.m5.generated.qwen_plus.en\endcsname{726}
\expandafter\gdef\csname odunum@val@language_split.m5.generated.qwen_plus.zh\endcsname{0.945}
\expandafter\gdef\csname odunum@n@language_split.m5.generated.qwen_plus.zh\endcsname{703}
\expandafter\gdef\csname odunum@val@language_split.m5.generated.salmonn2_7b.en\endcsname{0.508}
\expandafter\gdef\csname odunum@n@language_split.m5.generated.salmonn2_7b.en\endcsname{717}
\expandafter\gdef\csname odunum@val@language_split.m5.generated.salmonn2_7b.zh\endcsname{0.146}
\expandafter\gdef\csname odunum@n@language_split.m5.generated.salmonn2_7b.zh\endcsname{698}
\expandafter\gdef\csname odunum@val@language_split.m5.generated.seed.en\endcsname{0.947}
\expandafter\gdef\csname odunum@n@language_split.m5.generated.seed.en\endcsname{726}
\expandafter\gdef\csname odunum@val@language_split.m5.generated.seed.zh\endcsname{0.936}
\expandafter\gdef\csname odunum@n@language_split.m5.generated.seed.zh\endcsname{703}
\expandafter\gdef\csname odunum@val@language_split.m5.marginal.cascade_asr.en\endcsname{0.133}
\expandafter\gdef\csname odunum@n@language_split.m5.marginal.cascade_asr.en\endcsname{727}
\expandafter\gdef\csname odunum@val@language_split.m5.marginal.cascade_asr.zh\endcsname{0.117}
\expandafter\gdef\csname odunum@n@language_split.m5.marginal.cascade_asr.zh\endcsname{904}
\expandafter\gdef\csname odunum@val@language_split.m5.marginal.gemini.en\endcsname{0.960}
\expandafter\gdef\csname odunum@n@language_split.m5.marginal.gemini.en\endcsname{727}
\expandafter\gdef\csname odunum@val@language_split.m5.marginal.gemini.zh\endcsname{0.967}
\expandafter\gdef\csname odunum@n@language_split.m5.marginal.gemini.zh\endcsname{904}
\expandafter\gdef\csname odunum@val@language_split.m5.marginal.gemini35_flash_lite.en\endcsname{0.834}
\expandafter\gdef\csname odunum@n@language_split.m5.marginal.gemini35_flash_lite.en\endcsname{727}
\expandafter\gdef\csname odunum@val@language_split.m5.marginal.gemini35_flash_lite.zh\endcsname{0.780}
\expandafter\gdef\csname odunum@n@language_split.m5.marginal.gemini35_flash_lite.zh\endcsname{904}
\expandafter\gdef\csname odunum@val@language_split.m5.marginal.gemini37_flash.en\endcsname{0.914}
\expandafter\gdef\csname odunum@n@language_split.m5.marginal.gemini37_flash.en\endcsname{726}
\expandafter\gdef\csname odunum@val@language_split.m5.marginal.gemini37_flash.zh\endcsname{0.895}
\expandafter\gdef\csname odunum@n@language_split.m5.marginal.gemini37_flash.zh\endcsname{901}
\expandafter\gdef\csname odunum@val@language_split.m5.marginal.ming.en\endcsname{0.741}
\expandafter\gdef\csname odunum@n@language_split.m5.marginal.ming.en\endcsname{727}
\expandafter\gdef\csname odunum@val@language_split.m5.marginal.ming.zh\endcsname{0.809}
\expandafter\gdef\csname odunum@n@language_split.m5.marginal.ming.zh\endcsname{904}
\expandafter\gdef\csname odunum@val@language_split.m5.marginal.minicpm_o.en\endcsname{0.749}
\expandafter\gdef\csname odunum@n@language_split.m5.marginal.minicpm_o.en\endcsname{727}
\expandafter\gdef\csname odunum@val@language_split.m5.marginal.minicpm_o.zh\endcsname{0.674}
\expandafter\gdef\csname odunum@n@language_split.m5.marginal.minicpm_o.zh\endcsname{902}
\expandafter\gdef\csname odunum@val@language_split.m5.marginal.nemotron.en\endcsname{0.842}
\expandafter\gdef\csname odunum@n@language_split.m5.marginal.nemotron.en\endcsname{727}
\expandafter\gdef\csname odunum@val@language_split.m5.marginal.nemotron.zh\endcsname{0.545}
\expandafter\gdef\csname odunum@n@language_split.m5.marginal.nemotron.zh\endcsname{904}
\expandafter\gdef\csname odunum@val@language_split.m5.marginal.qwen25_omni.en\endcsname{0.696}
\expandafter\gdef\csname odunum@n@language_split.m5.marginal.qwen25_omni.en\endcsname{727}
\expandafter\gdef\csname odunum@val@language_split.m5.marginal.qwen25_omni.zh\endcsname{0.741}
\expandafter\gdef\csname odunum@n@language_split.m5.marginal.qwen25_omni.zh\endcsname{904}
\expandafter\gdef\csname odunum@val@language_split.m5.marginal.qwen3_omni_instruct.en\endcsname{0.930}
\expandafter\gdef\csname odunum@n@language_split.m5.marginal.qwen3_omni_instruct.en\endcsname{727}
\expandafter\gdef\csname odunum@val@language_split.m5.marginal.qwen3_omni_instruct.zh\endcsname{0.941}
\expandafter\gdef\csname odunum@n@language_split.m5.marginal.qwen3_omni_instruct.zh\endcsname{904}
\expandafter\gdef\csname odunum@val@language_split.m5.marginal.qwen3_omni_think.en\endcsname{0.929}
\expandafter\gdef\csname odunum@n@language_split.m5.marginal.qwen3_omni_think.en\endcsname{727}
\expandafter\gdef\csname odunum@val@language_split.m5.marginal.qwen3_omni_think.zh\endcsname{0.930}
\expandafter\gdef\csname odunum@n@language_split.m5.marginal.qwen3_omni_think.zh\endcsname{904}
\expandafter\gdef\csname odunum@val@language_split.m5.marginal.qwen_plus.en\endcsname{0.944}
\expandafter\gdef\csname odunum@n@language_split.m5.marginal.qwen_plus.en\endcsname{727}
\expandafter\gdef\csname odunum@val@language_split.m5.marginal.qwen_plus.zh\endcsname{0.943}
\expandafter\gdef\csname odunum@n@language_split.m5.marginal.qwen_plus.zh\endcsname{903}
\expandafter\gdef\csname odunum@val@language_split.m5.marginal.salmonn2_7b.en\endcsname{0.507}
\expandafter\gdef\csname odunum@n@language_split.m5.marginal.salmonn2_7b.en\endcsname{718}
\expandafter\gdef\csname odunum@val@language_split.m5.marginal.salmonn2_7b.zh\endcsname{0.138}
\expandafter\gdef\csname odunum@n@language_split.m5.marginal.salmonn2_7b.zh\endcsname{896}
\expandafter\gdef\csname odunum@val@language_split.m5.marginal.seed.en\endcsname{0.947}
\expandafter\gdef\csname odunum@n@language_split.m5.marginal.seed.en\endcsname{727}
\expandafter\gdef\csname odunum@val@language_split.m5.marginal.seed.zh\endcsname{0.939}
\expandafter\gdef\csname odunum@n@language_split.m5.marginal.seed.zh\endcsname{904}
\expandafter\gdef\csname odunum@val@language_split.panel_mean_delta.generated.avg\endcsname{-0.056}
\expandafter\gdef\csname odunum@n@language_split.panel_mean_delta.generated.avg\endcsname{13}
\expandafter\gdef\csname odunum@ci@language_split.panel_mean_delta.generated.avg\endcsname{[-0.121, -0.011]}
\expandafter\gdef\csname odunum@val@language_split.panel_mean_delta.generated.m1\endcsname{-0.038}
\expandafter\gdef\csname odunum@n@language_split.panel_mean_delta.generated.m1\endcsname{13}
\expandafter\gdef\csname odunum@ci@language_split.panel_mean_delta.generated.m1\endcsname{[-0.064, -0.013]}
\expandafter\gdef\csname odunum@val@language_split.panel_mean_delta.generated.m2\endcsname{-0.067}
\expandafter\gdef\csname odunum@n@language_split.panel_mean_delta.generated.m2\endcsname{13}
\expandafter\gdef\csname odunum@ci@language_split.panel_mean_delta.generated.m2\endcsname{[-0.147, -0.014]}
\expandafter\gdef\csname odunum@val@language_split.panel_mean_delta.generated.m3\endcsname{0.008}
\expandafter\gdef\csname odunum@n@language_split.panel_mean_delta.generated.m3\endcsname{13}
\expandafter\gdef\csname odunum@ci@language_split.panel_mean_delta.generated.m3\endcsname{[-0.014, 0.023]}
\expandafter\gdef\csname odunum@val@language_split.panel_mean_delta.generated.m5\endcsname{-0.047}
\expandafter\gdef\csname odunum@n@language_split.panel_mean_delta.generated.m5\endcsname{13}
\expandafter\gdef\csname odunum@ci@language_split.panel_mean_delta.generated.m5\endcsname{[-0.123, 0.011]}
\expandafter\gdef\csname odunum@val@language_split.panel_mean_delta.marginal.avg\endcsname{-0.070}
\expandafter\gdef\csname odunum@n@language_split.panel_mean_delta.marginal.avg\endcsname{13}
\expandafter\gdef\csname odunum@ci@language_split.panel_mean_delta.marginal.avg\endcsname{[-0.134, -0.025]}
\expandafter\gdef\csname odunum@val@language_split.panel_mean_delta.marginal.m1\endcsname{-0.054}
\expandafter\gdef\csname odunum@n@language_split.panel_mean_delta.marginal.m1\endcsname{13}
\expandafter\gdef\csname odunum@ci@language_split.panel_mean_delta.marginal.m1\endcsname{[-0.083, -0.024]}
\expandafter\gdef\csname odunum@val@language_split.panel_mean_delta.marginal.m2\endcsname{-0.083}
\expandafter\gdef\csname odunum@n@language_split.panel_mean_delta.marginal.m2\endcsname{13}
\expandafter\gdef\csname odunum@ci@language_split.panel_mean_delta.marginal.m2\endcsname{[-0.162, -0.030]}
\expandafter\gdef\csname odunum@val@language_split.panel_mean_delta.marginal.m3\endcsname{0.013}
\expandafter\gdef\csname odunum@n@language_split.panel_mean_delta.marginal.m3\endcsname{13}
\expandafter\gdef\csname odunum@ci@language_split.panel_mean_delta.marginal.m3\endcsname{[-0.012, 0.031]}
\expandafter\gdef\csname odunum@val@language_split.panel_mean_delta.marginal.m5\endcsname{-0.054}
\expandafter\gdef\csname odunum@n@language_split.panel_mean_delta.marginal.m5\endcsname{13}
\expandafter\gdef\csname odunum@ci@language_split.panel_mean_delta.marginal.m5\endcsname{[-0.129, 0.005]}
\expandafter\gdef\csname odunum@val@language_split.panel_sign.generated.avg\endcsname{2}
\expandafter\gdef\csname odunum@n@language_split.panel_sign.generated.avg\endcsname{13}
\expandafter\gdef\csname odunum@val@language_split.panel_sign.generated.m1\endcsname{3}
\expandafter\gdef\csname odunum@n@language_split.panel_sign.generated.m1\endcsname{13}
\expandafter\gdef\csname odunum@val@language_split.panel_sign.generated.m2\endcsname{2}
\expandafter\gdef\csname odunum@n@language_split.panel_sign.generated.m2\endcsname{13}
\expandafter\gdef\csname odunum@val@language_split.panel_sign.generated.m3\endcsname{11}
\expandafter\gdef\csname odunum@n@language_split.panel_sign.generated.m3\endcsname{13}
\expandafter\gdef\csname odunum@val@language_split.panel_sign.generated.m5\endcsname{7}
\expandafter\gdef\csname odunum@n@language_split.panel_sign.generated.m5\endcsname{13}
\expandafter\gdef\csname odunum@val@language_split.panel_sign.marginal.avg\endcsname{1}
\expandafter\gdef\csname odunum@n@language_split.panel_sign.marginal.avg\endcsname{13}
\expandafter\gdef\csname odunum@val@language_split.panel_sign.marginal.m1\endcsname{3}
\expandafter\gdef\csname odunum@n@language_split.panel_sign.marginal.m1\endcsname{13}
\expandafter\gdef\csname odunum@val@language_split.panel_sign.marginal.m2\endcsname{1}
\expandafter\gdef\csname odunum@n@language_split.panel_sign.marginal.m2\endcsname{13}
\expandafter\gdef\csname odunum@val@language_split.panel_sign.marginal.m3\endcsname{10}
\expandafter\gdef\csname odunum@n@language_split.panel_sign.marginal.m3\endcsname{13}
\expandafter\gdef\csname odunum@val@language_split.panel_sign.marginal.m5\endcsname{5}
\expandafter\gdef\csname odunum@n@language_split.panel_sign.marginal.m5\endcsname{13}
\expandafter\gdef\csname odunum@val@language_split.pop.en.audio_only.n\endcsname{366}
\expandafter\gdef\csname odunum@n@language_split.pop.en.audio_only.n\endcsname{908}
\expandafter\gdef\csname odunum@val@language_split.pop.en.audio_visual.n\endcsname{542}
\expandafter\gdef\csname odunum@n@language_split.pop.en.audio_visual.n\endcsname{908}
\expandafter\gdef\csname odunum@val@language_split.pop.en.n\endcsname{908}
\expandafter\gdef\csname odunum@n@language_split.pop.en.n\endcsname{2\,078}
\expandafter\gdef\csname odunum@val@language_split.pop.generated.en.n\endcsname{907}
\expandafter\gdef\csname odunum@n@language_split.pop.generated.en.n\endcsname{2\,078}
\expandafter\gdef\csname odunum@val@language_split.pop.generated.en.n_pos\endcsname{726}
\expandafter\gdef\csname odunum@n@language_split.pop.generated.en.n_pos\endcsname{907}
\expandafter\gdef\csname odunum@val@language_split.pop.generated.zh.n\endcsname{894}
\expandafter\gdef\csname odunum@n@language_split.pop.generated.zh.n\endcsname{2\,078}
\expandafter\gdef\csname odunum@val@language_split.pop.generated.zh.n_pos\endcsname{703}
\expandafter\gdef\csname odunum@n@language_split.pop.generated.zh.n_pos\endcsname{894}
\expandafter\gdef\csname odunum@val@language_split.pop.marginal.en.n\endcsname{908}
\expandafter\gdef\csname odunum@n@language_split.pop.marginal.en.n\endcsname{2\,078}
\expandafter\gdef\csname odunum@val@language_split.pop.marginal.en.n_pos\endcsname{727}
\expandafter\gdef\csname odunum@n@language_split.pop.marginal.en.n_pos\endcsname{908}
\expandafter\gdef\csname odunum@val@language_split.pop.marginal.zh.n\endcsname{1\,170}
\expandafter\gdef\csname odunum@n@language_split.pop.marginal.zh.n\endcsname{2\,078}
\expandafter\gdef\csname odunum@val@language_split.pop.marginal.zh.n_pos\endcsname{904}
\expandafter\gdef\csname odunum@n@language_split.pop.marginal.zh.n_pos\endcsname{1\,170}
\expandafter\gdef\csname odunum@val@language_split.pop.zh.audio_only.n\endcsname{365}
\expandafter\gdef\csname odunum@n@language_split.pop.zh.audio_only.n\endcsname{1\,170}
\expandafter\gdef\csname odunum@val@language_split.pop.zh.audio_visual.n\endcsname{805}
\expandafter\gdef\csname odunum@n@language_split.pop.zh.audio_visual.n\endcsname{1\,170}
\expandafter\gdef\csname odunum@val@language_split.pop.zh.n\endcsname{1\,170}
\expandafter\gdef\csname odunum@n@language_split.pop.zh.n\endcsname{2\,078}
\expandafter\gdef\csname odunum@val@language_split.rank.generated.avg.kendall_tau\endcsname{0.821}
\expandafter\gdef\csname odunum@n@language_split.rank.generated.avg.kendall_tau\endcsname{13}
\expandafter\gdef\csname odunum@val@language_split.rank.generated.avg.spearman\endcsname{0.934}
\expandafter\gdef\csname odunum@n@language_split.rank.generated.avg.spearman\endcsname{13}
\expandafter\gdef\csname odunum@val@language_split.rank.generated.m2.kendall_tau\endcsname{0.846}
\expandafter\gdef\csname odunum@n@language_split.rank.generated.m2.kendall_tau\endcsname{13}
\expandafter\gdef\csname odunum@val@language_split.rank.generated.m2.spearman\endcsname{0.945}
\expandafter\gdef\csname odunum@n@language_split.rank.generated.m2.spearman\endcsname{13}
\expandafter\gdef\csname odunum@val@language_split.rank.marginal.avg.kendall_tau\endcsname{0.795}
\expandafter\gdef\csname odunum@n@language_split.rank.marginal.avg.kendall_tau\endcsname{13}
\expandafter\gdef\csname odunum@val@language_split.rank.marginal.avg.spearman\endcsname{0.923}
\expandafter\gdef\csname odunum@n@language_split.rank.marginal.avg.spearman\endcsname{13}
\expandafter\gdef\csname odunum@val@language_split.rank.marginal.m2.kendall_tau\endcsname{0.846}
\expandafter\gdef\csname odunum@n@language_split.rank.marginal.m2.kendall_tau\endcsname{13}
\expandafter\gdef\csname odunum@val@language_split.rank.marginal.m2.spearman\endcsname{0.940}
\expandafter\gdef\csname odunum@n@language_split.rank.marginal.m2.spearman\endcsname{13}
\expandafter\gdef\csname odunum@val@metric_structure.agg.drop_m1.cascade_asr\endcsname{0.627}
\expandafter\gdef\csname odunum@n@metric_structure.agg.drop_m1.cascade_asr\endcsname{2\,078}
\expandafter\gdef\csname odunum@val@metric_structure.agg.drop_m1.gemini\endcsname{0.716}
\expandafter\gdef\csname odunum@n@metric_structure.agg.drop_m1.gemini\endcsname{2\,078}
\expandafter\gdef\csname odunum@val@metric_structure.agg.drop_m1.gemini35_flash_lite\endcsname{0.576}
\expandafter\gdef\csname odunum@n@metric_structure.agg.drop_m1.gemini35_flash_lite\endcsname{2\,078}
\expandafter\gdef\csname odunum@val@metric_structure.agg.drop_m1.gemini37_flash\endcsname{0.681}
\expandafter\gdef\csname odunum@n@metric_structure.agg.drop_m1.gemini37_flash\endcsname{2\,067}
\expandafter\gdef\csname odunum@val@metric_structure.agg.drop_m1.ming\endcsname{0.523}
\expandafter\gdef\csname odunum@n@metric_structure.agg.drop_m1.ming\endcsname{2\,078}
\expandafter\gdef\csname odunum@val@metric_structure.agg.drop_m1.minicpm_o\endcsname{0.398}
\expandafter\gdef\csname odunum@n@metric_structure.agg.drop_m1.minicpm_o\endcsname{2\,075}
\expandafter\gdef\csname odunum@val@metric_structure.agg.drop_m1.nemotron\endcsname{0.331}
\expandafter\gdef\csname odunum@n@metric_structure.agg.drop_m1.nemotron\endcsname{2\,078}
\expandafter\gdef\csname odunum@val@metric_structure.agg.drop_m1.qwen25_omni\endcsname{0.440}
\expandafter\gdef\csname odunum@n@metric_structure.agg.drop_m1.qwen25_omni\endcsname{2\,078}
\expandafter\gdef\csname odunum@val@metric_structure.agg.drop_m1.qwen3_omni_instruct\endcsname{0.580}
\expandafter\gdef\csname odunum@n@metric_structure.agg.drop_m1.qwen3_omni_instruct\endcsname{2\,078}
\expandafter\gdef\csname odunum@val@metric_structure.agg.drop_m1.qwen3_omni_think\endcsname{0.624}
\expandafter\gdef\csname odunum@n@metric_structure.agg.drop_m1.qwen3_omni_think\endcsname{2\,078}
\expandafter\gdef\csname odunum@val@metric_structure.agg.drop_m1.qwen_plus\endcsname{0.720}
\expandafter\gdef\csname odunum@n@metric_structure.agg.drop_m1.qwen_plus\endcsname{2\,077}
\expandafter\gdef\csname odunum@val@metric_structure.agg.drop_m1.salmonn2_7b\endcsname{0.120}
\expandafter\gdef\csname odunum@n@metric_structure.agg.drop_m1.salmonn2_7b\endcsname{2\,058}
\expandafter\gdef\csname odunum@val@metric_structure.agg.drop_m1.seed\endcsname{0.733}
\expandafter\gdef\csname odunum@n@metric_structure.agg.drop_m1.seed\endcsname{2\,078}
\expandafter\gdef\csname odunum@val@metric_structure.agg.drop_m4.cascade_asr\endcsname{0.613}
\expandafter\gdef\csname odunum@n@metric_structure.agg.drop_m4.cascade_asr\endcsname{2\,078}
\expandafter\gdef\csname odunum@val@metric_structure.agg.drop_m4.gemini\endcsname{0.716}
\expandafter\gdef\csname odunum@n@metric_structure.agg.drop_m4.gemini\endcsname{2\,078}
\expandafter\gdef\csname odunum@val@metric_structure.agg.drop_m4.gemini35_flash_lite\endcsname{0.577}
\expandafter\gdef\csname odunum@n@metric_structure.agg.drop_m4.gemini35_flash_lite\endcsname{2\,078}
\expandafter\gdef\csname odunum@val@metric_structure.agg.drop_m4.gemini37_flash\endcsname{0.690}
\expandafter\gdef\csname odunum@n@metric_structure.agg.drop_m4.gemini37_flash\endcsname{2\,067}
\expandafter\gdef\csname odunum@val@metric_structure.agg.drop_m4.ming\endcsname{0.492}
\expandafter\gdef\csname odunum@n@metric_structure.agg.drop_m4.ming\endcsname{2\,078}
\expandafter\gdef\csname odunum@val@metric_structure.agg.drop_m4.minicpm_o\endcsname{0.406}
\expandafter\gdef\csname odunum@n@metric_structure.agg.drop_m4.minicpm_o\endcsname{2\,075}
\expandafter\gdef\csname odunum@val@metric_structure.agg.drop_m4.nemotron\endcsname{0.369}
\expandafter\gdef\csname odunum@n@metric_structure.agg.drop_m4.nemotron\endcsname{2\,078}
\expandafter\gdef\csname odunum@val@metric_structure.agg.drop_m4.qwen25_omni\endcsname{0.434}
\expandafter\gdef\csname odunum@n@metric_structure.agg.drop_m4.qwen25_omni\endcsname{2\,078}
\expandafter\gdef\csname odunum@val@metric_structure.agg.drop_m4.qwen3_omni_instruct\endcsname{0.540}
\expandafter\gdef\csname odunum@n@metric_structure.agg.drop_m4.qwen3_omni_instruct\endcsname{2\,078}
\expandafter\gdef\csname odunum@val@metric_structure.agg.drop_m4.qwen3_omni_think\endcsname{0.600}
\expandafter\gdef\csname odunum@n@metric_structure.agg.drop_m4.qwen3_omni_think\endcsname{2\,078}
\expandafter\gdef\csname odunum@val@metric_structure.agg.drop_m4.qwen_plus\endcsname{0.699}
\expandafter\gdef\csname odunum@n@metric_structure.agg.drop_m4.qwen_plus\endcsname{2\,077}
\expandafter\gdef\csname odunum@val@metric_structure.agg.drop_m4.salmonn2_7b\endcsname{0.177}
\expandafter\gdef\csname odunum@n@metric_structure.agg.drop_m4.salmonn2_7b\endcsname{2\,058}
\expandafter\gdef\csname odunum@val@metric_structure.agg.drop_m4.seed\endcsname{0.698}
\expandafter\gdef\csname odunum@n@metric_structure.agg.drop_m4.seed\endcsname{2\,078}
\expandafter\gdef\csname odunum@val@metric_structure.agg.equal.cascade_asr\endcsname{0.614}
\expandafter\gdef\csname odunum@n@metric_structure.agg.equal.cascade_asr\endcsname{2\,078}
\expandafter\gdef\csname odunum@val@metric_structure.agg.equal.gemini\endcsname{0.826}
\expandafter\gdef\csname odunum@n@metric_structure.agg.equal.gemini\endcsname{2\,078}
\expandafter\gdef\csname odunum@val@metric_structure.agg.equal.gemini35_flash_lite\endcsname{0.674}
\expandafter\gdef\csname odunum@n@metric_structure.agg.equal.gemini35_flash_lite\endcsname{2\,078}
\expandafter\gdef\csname odunum@val@metric_structure.agg.equal.gemini37_flash\endcsname{0.797}
\expandafter\gdef\csname odunum@n@metric_structure.agg.equal.gemini37_flash\endcsname{2\,067}
\expandafter\gdef\csname odunum@val@metric_structure.agg.equal.ming\endcsname{0.578}
\expandafter\gdef\csname odunum@n@metric_structure.agg.equal.ming\endcsname{2\,078}
\expandafter\gdef\csname odunum@val@metric_structure.agg.equal.minicpm_o\endcsname{0.466}
\expandafter\gdef\csname odunum@n@metric_structure.agg.equal.minicpm_o\endcsname{2\,075}
\expandafter\gdef\csname odunum@val@metric_structure.agg.equal.nemotron\endcsname{0.453}
\expandafter\gdef\csname odunum@n@metric_structure.agg.equal.nemotron\endcsname{2\,078}
\expandafter\gdef\csname odunum@val@metric_structure.agg.equal.qwen25_omni\endcsname{0.505}
\expandafter\gdef\csname odunum@n@metric_structure.agg.equal.qwen25_omni\endcsname{2\,078}
\expandafter\gdef\csname odunum@val@metric_structure.agg.equal.qwen3_omni_instruct\endcsname{0.630}
\expandafter\gdef\csname odunum@n@metric_structure.agg.equal.qwen3_omni_instruct\endcsname{2\,078}
\expandafter\gdef\csname odunum@val@metric_structure.agg.equal.qwen3_omni_think\endcsname{0.699}
\expandafter\gdef\csname odunum@n@metric_structure.agg.equal.qwen3_omni_think\endcsname{2\,078}
\expandafter\gdef\csname odunum@val@metric_structure.agg.equal.qwen_plus\endcsname{0.777}
\expandafter\gdef\csname odunum@n@metric_structure.agg.equal.qwen_plus\endcsname{2\,077}
\expandafter\gdef\csname odunum@val@metric_structure.agg.equal.salmonn2_7b\endcsname{0.226}
\expandafter\gdef\csname odunum@n@metric_structure.agg.equal.salmonn2_7b\endcsname{2\,058}
\expandafter\gdef\csname odunum@val@metric_structure.agg.equal.seed\endcsname{0.768}
\expandafter\gdef\csname odunum@n@metric_structure.agg.equal.seed\endcsname{2\,078}
\expandafter\gdef\csname odunum@val@metric_structure.agg.m2_only.cascade_asr\endcsname{0.600}
\expandafter\gdef\csname odunum@n@metric_structure.agg.m2_only.cascade_asr\endcsname{2\,078}
\expandafter\gdef\csname odunum@val@metric_structure.agg.m2_only.gemini\endcsname{0.657}
\expandafter\gdef\csname odunum@n@metric_structure.agg.m2_only.gemini\endcsname{2\,078}
\expandafter\gdef\csname odunum@val@metric_structure.agg.m2_only.gemini35_flash_lite\endcsname{0.522}
\expandafter\gdef\csname odunum@n@metric_structure.agg.m2_only.gemini35_flash_lite\endcsname{2\,078}
\expandafter\gdef\csname odunum@val@metric_structure.agg.m2_only.gemini37_flash\endcsname{0.622}
\expandafter\gdef\csname odunum@n@metric_structure.agg.m2_only.gemini37_flash\endcsname{2\,067}
\expandafter\gdef\csname odunum@val@metric_structure.agg.m2_only.ming\endcsname{0.499}
\expandafter\gdef\csname odunum@n@metric_structure.agg.m2_only.ming\endcsname{2\,078}
\expandafter\gdef\csname odunum@val@metric_structure.agg.m2_only.minicpm_o\endcsname{0.398}
\expandafter\gdef\csname odunum@n@metric_structure.agg.m2_only.minicpm_o\endcsname{2\,075}
\expandafter\gdef\csname odunum@val@metric_structure.agg.m2_only.nemotron\endcsname{0.292}
\expandafter\gdef\csname odunum@n@metric_structure.agg.m2_only.nemotron\endcsname{2\,078}
\expandafter\gdef\csname odunum@val@metric_structure.agg.m2_only.qwen25_omni\endcsname{0.433}
\expandafter\gdef\csname odunum@n@metric_structure.agg.m2_only.qwen25_omni\endcsname{2\,078}
\expandafter\gdef\csname odunum@val@metric_structure.agg.m2_only.qwen3_omni_instruct\endcsname{0.558}
\expandafter\gdef\csname odunum@n@metric_structure.agg.m2_only.qwen3_omni_instruct\endcsname{2\,078}
\expandafter\gdef\csname odunum@val@metric_structure.agg.m2_only.qwen3_omni_think\endcsname{0.582}
\expandafter\gdef\csname odunum@n@metric_structure.agg.m2_only.qwen3_omni_think\endcsname{2\,078}
\expandafter\gdef\csname odunum@val@metric_structure.agg.m2_only.qwen_plus\endcsname{0.678}
\expandafter\gdef\csname odunum@n@metric_structure.agg.m2_only.qwen_plus\endcsname{2\,077}
\expandafter\gdef\csname odunum@val@metric_structure.agg.m2_only.salmonn2_7b\endcsname{0.108}
\expandafter\gdef\csname odunum@n@metric_structure.agg.m2_only.salmonn2_7b\endcsname{2\,058}
\expandafter\gdef\csname odunum@val@metric_structure.agg.m2_only.seed\endcsname{0.690}
\expandafter\gdef\csname odunum@n@metric_structure.agg.m2_only.seed\endcsname{2\,078}
\expandafter\gdef\csname odunum@val@metric_structure.agg.official.cascade_asr\endcsname{0.634}
\expandafter\gdef\csname odunum@n@metric_structure.agg.official.cascade_asr\endcsname{2\,078}
\expandafter\gdef\csname odunum@val@metric_structure.agg.official.gemini\endcsname{0.736}
\expandafter\gdef\csname odunum@n@metric_structure.agg.official.gemini\endcsname{2\,078}
\expandafter\gdef\csname odunum@val@metric_structure.agg.official.gemini35_flash_lite\endcsname{0.591}
\expandafter\gdef\csname odunum@n@metric_structure.agg.official.gemini35_flash_lite\endcsname{2\,078}
\expandafter\gdef\csname odunum@val@metric_structure.agg.official.gemini37_flash\endcsname{0.708}
\expandafter\gdef\csname odunum@n@metric_structure.agg.official.gemini37_flash\endcsname{2\,067}
\expandafter\gdef\csname odunum@val@metric_structure.agg.official.ming\endcsname{0.527}
\expandafter\gdef\csname odunum@n@metric_structure.agg.official.ming\endcsname{2\,078}
\expandafter\gdef\csname odunum@val@metric_structure.agg.official.minicpm_o\endcsname{0.426}
\expandafter\gdef\csname odunum@n@metric_structure.agg.official.minicpm_o\endcsname{2\,075}
\expandafter\gdef\csname odunum@val@metric_structure.agg.official.nemotron\endcsname{0.367}
\expandafter\gdef\csname odunum@n@metric_structure.agg.official.nemotron\endcsname{2\,078}
\expandafter\gdef\csname odunum@val@metric_structure.agg.official.qwen25_omni\endcsname{0.463}
\expandafter\gdef\csname odunum@n@metric_structure.agg.official.qwen25_omni\endcsname{2\,078}
\expandafter\gdef\csname odunum@val@metric_structure.agg.official.qwen3_omni_instruct\endcsname{0.575}
\expandafter\gdef\csname odunum@n@metric_structure.agg.official.qwen3_omni_instruct\endcsname{2\,078}
\expandafter\gdef\csname odunum@val@metric_structure.agg.official.qwen3_omni_think\endcsname{0.626}
\expandafter\gdef\csname odunum@n@metric_structure.agg.official.qwen3_omni_think\endcsname{2\,078}
\expandafter\gdef\csname odunum@val@metric_structure.agg.official.qwen_plus\endcsname{0.714}
\expandafter\gdef\csname odunum@n@metric_structure.agg.official.qwen_plus\endcsname{2\,077}
\expandafter\gdef\csname odunum@val@metric_structure.agg.official.salmonn2_7b\endcsname{0.176}
\expandafter\gdef\csname odunum@n@metric_structure.agg.official.salmonn2_7b\endcsname{2\,058}
\expandafter\gdef\csname odunum@val@metric_structure.agg.official.seed\endcsname{0.712}
\expandafter\gdef\csname odunum@n@metric_structure.agg.official.seed\endcsname{2\,078}
\expandafter\gdef\csname odunum@val@metric_structure.corr.cascade_asr.demand_correct_m2\endcsname{0.46\ensuremath{\times}}
\expandafter\gdef\csname odunum@n@metric_structure.corr.cascade_asr.demand_correct_m2\endcsname{1\,631}
\expandafter\gdef\csname odunum@val@metric_structure.corr.cascade_asr.demand_correct_m3\endcsname{0.68\ensuremath{\times}}
\expandafter\gdef\csname odunum@n@metric_structure.corr.cascade_asr.demand_correct_m3\endcsname{1\,631}
\expandafter\gdef\csname odunum@val@metric_structure.corr.cascade_asr.demand_correct_m4\endcsname{0.45\ensuremath{\times}}
\expandafter\gdef\csname odunum@n@metric_structure.corr.cascade_asr.demand_correct_m4\endcsname{1\,631}
\expandafter\gdef\csname odunum@val@metric_structure.corr.cascade_asr.demand_correct_m5\endcsname{0.81\ensuremath{\times}}
\expandafter\gdef\csname odunum@n@metric_structure.corr.cascade_asr.demand_correct_m5\endcsname{1\,631}
\expandafter\gdef\csname odunum@val@metric_structure.corr.cascade_asr.m2_m3\endcsname{0.50\ensuremath{\times}}
\expandafter\gdef\csname odunum@n@metric_structure.corr.cascade_asr.m2_m3\endcsname{1\,631}
\expandafter\gdef\csname odunum@val@metric_structure.corr.cascade_asr.m2_m4\endcsname{0.43\ensuremath{\times}}
\expandafter\gdef\csname odunum@n@metric_structure.corr.cascade_asr.m2_m4\endcsname{1\,631}
\expandafter\gdef\csname odunum@val@metric_structure.corr.cascade_asr.m2_m5\endcsname{0.32\ensuremath{\times}}
\expandafter\gdef\csname odunum@n@metric_structure.corr.cascade_asr.m2_m5\endcsname{1\,631}
\expandafter\gdef\csname odunum@val@metric_structure.corr.cascade_asr.m3_m4\endcsname{0.78\ensuremath{\times}}
\expandafter\gdef\csname odunum@n@metric_structure.corr.cascade_asr.m3_m4\endcsname{1\,631}
\expandafter\gdef\csname odunum@val@metric_structure.corr.cascade_asr.m3_m5\endcsname{0.51\ensuremath{\times}}
\expandafter\gdef\csname odunum@n@metric_structure.corr.cascade_asr.m3_m5\endcsname{1\,631}
\expandafter\gdef\csname odunum@val@metric_structure.corr.cascade_asr.m4_m5\endcsname{0.29\ensuremath{\times}}
\expandafter\gdef\csname odunum@n@metric_structure.corr.cascade_asr.m4_m5\endcsname{1\,631}
\expandafter\gdef\csname odunum@val@metric_structure.corr.gemini.demand_correct_m2\endcsname{0.29\ensuremath{\times}}
\expandafter\gdef\csname odunum@n@metric_structure.corr.gemini.demand_correct_m2\endcsname{1\,631}
\expandafter\gdef\csname odunum@val@metric_structure.corr.gemini.demand_correct_m3\endcsname{-0.40\ensuremath{\times}}
\expandafter\gdef\csname odunum@n@metric_structure.corr.gemini.demand_correct_m3\endcsname{1\,631}
\expandafter\gdef\csname odunum@val@metric_structure.corr.gemini.demand_correct_m4\endcsname{0.33\ensuremath{\times}}
\expandafter\gdef\csname odunum@n@metric_structure.corr.gemini.demand_correct_m4\endcsname{1\,631}
\expandafter\gdef\csname odunum@val@metric_structure.corr.gemini.demand_correct_m5\endcsname{0.92\ensuremath{\times}}
\expandafter\gdef\csname odunum@n@metric_structure.corr.gemini.demand_correct_m5\endcsname{1\,631}
\expandafter\gdef\csname odunum@val@metric_structure.corr.gemini.m2_m3\endcsname{-0.04\ensuremath{\times}}
\expandafter\gdef\csname odunum@n@metric_structure.corr.gemini.m2_m3\endcsname{1\,631}
\expandafter\gdef\csname odunum@val@metric_structure.corr.gemini.m2_m4\endcsname{0.16\ensuremath{\times}}
\expandafter\gdef\csname odunum@n@metric_structure.corr.gemini.m2_m4\endcsname{1\,631}
\expandafter\gdef\csname odunum@val@metric_structure.corr.gemini.m2_m5\endcsname{0.30\ensuremath{\times}}
\expandafter\gdef\csname odunum@n@metric_structure.corr.gemini.m2_m5\endcsname{1\,631}
\expandafter\gdef\csname odunum@val@metric_structure.corr.gemini.m3_m4\endcsname{-0.02\ensuremath{\times}}
\expandafter\gdef\csname odunum@n@metric_structure.corr.gemini.m3_m4\endcsname{1\,631}
\expandafter\gdef\csname odunum@val@metric_structure.corr.gemini.m3_m5\endcsname{-0.36\ensuremath{\times}}
\expandafter\gdef\csname odunum@n@metric_structure.corr.gemini.m3_m5\endcsname{1\,631}
\expandafter\gdef\csname odunum@val@metric_structure.corr.gemini.m4_m5\endcsname{0.35\ensuremath{\times}}
\expandafter\gdef\csname odunum@n@metric_structure.corr.gemini.m4_m5\endcsname{1\,631}
\expandafter\gdef\csname odunum@val@metric_structure.corr.gemini35_flash_lite.demand_correct_m2\endcsname{0.54\ensuremath{\times}}
\expandafter\gdef\csname odunum@n@metric_structure.corr.gemini35_flash_lite.demand_correct_m2\endcsname{1\,631}
\expandafter\gdef\csname odunum@val@metric_structure.corr.gemini35_flash_lite.demand_correct_m3\endcsname{0.14\ensuremath{\times}}
\expandafter\gdef\csname odunum@n@metric_structure.corr.gemini35_flash_lite.demand_correct_m3\endcsname{1\,631}
\expandafter\gdef\csname odunum@val@metric_structure.corr.gemini35_flash_lite.demand_correct_m4\endcsname{0.51\ensuremath{\times}}
\expandafter\gdef\csname odunum@n@metric_structure.corr.gemini35_flash_lite.demand_correct_m4\endcsname{1\,631}
\expandafter\gdef\csname odunum@val@metric_structure.corr.gemini35_flash_lite.demand_correct_m5\endcsname{0.85\ensuremath{\times}}
\expandafter\gdef\csname odunum@n@metric_structure.corr.gemini35_flash_lite.demand_correct_m5\endcsname{1\,631}
\expandafter\gdef\csname odunum@val@metric_structure.corr.gemini35_flash_lite.m2_m3\endcsname{0.36\ensuremath{\times}}
\expandafter\gdef\csname odunum@n@metric_structure.corr.gemini35_flash_lite.m2_m3\endcsname{1\,631}
\expandafter\gdef\csname odunum@val@metric_structure.corr.gemini35_flash_lite.m2_m4\endcsname{0.49\ensuremath{\times}}
\expandafter\gdef\csname odunum@n@metric_structure.corr.gemini35_flash_lite.m2_m4\endcsname{1\,631}
\expandafter\gdef\csname odunum@val@metric_structure.corr.gemini35_flash_lite.m2_m5\endcsname{0.56\ensuremath{\times}}
\expandafter\gdef\csname odunum@n@metric_structure.corr.gemini35_flash_lite.m2_m5\endcsname{1\,631}
\expandafter\gdef\csname odunum@val@metric_structure.corr.gemini35_flash_lite.m3_m4\endcsname{0.47\ensuremath{\times}}
\expandafter\gdef\csname odunum@n@metric_structure.corr.gemini35_flash_lite.m3_m4\endcsname{1\,631}
\expandafter\gdef\csname odunum@val@metric_structure.corr.gemini35_flash_lite.m3_m5\endcsname{0.22\ensuremath{\times}}
\expandafter\gdef\csname odunum@n@metric_structure.corr.gemini35_flash_lite.m3_m5\endcsname{1\,631}
\expandafter\gdef\csname odunum@val@metric_structure.corr.gemini35_flash_lite.m4_m5\endcsname{0.54\ensuremath{\times}}
\expandafter\gdef\csname odunum@n@metric_structure.corr.gemini35_flash_lite.m4_m5\endcsname{1\,631}
\expandafter\gdef\csname odunum@val@metric_structure.corr.gemini37_flash.demand_correct_m2\endcsname{0.36\ensuremath{\times}}
\expandafter\gdef\csname odunum@n@metric_structure.corr.gemini37_flash.demand_correct_m2\endcsname{1\,627}
\expandafter\gdef\csname odunum@val@metric_structure.corr.gemini37_flash.demand_correct_m3\endcsname{-0.19\ensuremath{\times}}
\expandafter\gdef\csname odunum@n@metric_structure.corr.gemini37_flash.demand_correct_m3\endcsname{1\,627}
\expandafter\gdef\csname odunum@val@metric_structure.corr.gemini37_flash.demand_correct_m4\endcsname{0.39\ensuremath{\times}}
\expandafter\gdef\csname odunum@n@metric_structure.corr.gemini37_flash.demand_correct_m4\endcsname{1\,627}
\expandafter\gdef\csname odunum@val@metric_structure.corr.gemini37_flash.demand_correct_m5\endcsname{0.90\ensuremath{\times}}
\expandafter\gdef\csname odunum@n@metric_structure.corr.gemini37_flash.demand_correct_m5\endcsname{1\,627}
\expandafter\gdef\csname odunum@val@metric_structure.corr.gemini37_flash.m2_m3\endcsname{0.14\ensuremath{\times}}
\expandafter\gdef\csname odunum@n@metric_structure.corr.gemini37_flash.m2_m3\endcsname{1\,627}
\expandafter\gdef\csname odunum@val@metric_structure.corr.gemini37_flash.m2_m4\endcsname{0.25\ensuremath{\times}}
\expandafter\gdef\csname odunum@n@metric_structure.corr.gemini37_flash.m2_m4\endcsname{1\,627}
\expandafter\gdef\csname odunum@val@metric_structure.corr.gemini37_flash.m2_m5\endcsname{0.35\ensuremath{\times}}
\expandafter\gdef\csname odunum@n@metric_structure.corr.gemini37_flash.m2_m5\endcsname{1\,627}
\expandafter\gdef\csname odunum@val@metric_structure.corr.gemini37_flash.m3_m4\endcsname{0.16\ensuremath{\times}}
\expandafter\gdef\csname odunum@n@metric_structure.corr.gemini37_flash.m3_m4\endcsname{1\,627}
\expandafter\gdef\csname odunum@val@metric_structure.corr.gemini37_flash.m3_m5\endcsname{-0.14\ensuremath{\times}}
\expandafter\gdef\csname odunum@n@metric_structure.corr.gemini37_flash.m3_m5\endcsname{1\,627}
\expandafter\gdef\csname odunum@val@metric_structure.corr.gemini37_flash.m4_m5\endcsname{0.39\ensuremath{\times}}
\expandafter\gdef\csname odunum@n@metric_structure.corr.gemini37_flash.m4_m5\endcsname{1\,627}
\expandafter\gdef\csname odunum@val@metric_structure.corr.ming.demand_correct_m2\endcsname{0.17\ensuremath{\times}}
\expandafter\gdef\csname odunum@n@metric_structure.corr.ming.demand_correct_m2\endcsname{1\,631}
\expandafter\gdef\csname odunum@val@metric_structure.corr.ming.demand_correct_m3\endcsname{-0.08\ensuremath{\times}}
\expandafter\gdef\csname odunum@n@metric_structure.corr.ming.demand_correct_m3\endcsname{1\,631}
\expandafter\gdef\csname odunum@val@metric_structure.corr.ming.demand_correct_m4\endcsname{0.32\ensuremath{\times}}
\expandafter\gdef\csname odunum@n@metric_structure.corr.ming.demand_correct_m4\endcsname{1\,631}
\expandafter\gdef\csname odunum@val@metric_structure.corr.ming.demand_correct_m5\endcsname{0.47\ensuremath{\times}}
\expandafter\gdef\csname odunum@n@metric_structure.corr.ming.demand_correct_m5\endcsname{1\,631}
\expandafter\gdef\csname odunum@val@metric_structure.corr.ming.m2_m3\endcsname{0.06\ensuremath{\times}}
\expandafter\gdef\csname odunum@n@metric_structure.corr.ming.m2_m3\endcsname{1\,631}
\expandafter\gdef\csname odunum@val@metric_structure.corr.ming.m2_m4\endcsname{0.21\ensuremath{\times}}
\expandafter\gdef\csname odunum@n@metric_structure.corr.ming.m2_m4\endcsname{1\,631}
\expandafter\gdef\csname odunum@val@metric_structure.corr.ming.m2_m5\endcsname{0.21\ensuremath{\times}}
\expandafter\gdef\csname odunum@n@metric_structure.corr.ming.m2_m5\endcsname{1\,631}
\expandafter\gdef\csname odunum@val@metric_structure.corr.ming.m3_m4\endcsname{-0.05\ensuremath{\times}}
\expandafter\gdef\csname odunum@n@metric_structure.corr.ming.m3_m4\endcsname{1\,631}
\expandafter\gdef\csname odunum@val@metric_structure.corr.ming.m3_m5\endcsname{0.02\ensuremath{\times}}
\expandafter\gdef\csname odunum@n@metric_structure.corr.ming.m3_m5\endcsname{1\,631}
\expandafter\gdef\csname odunum@val@metric_structure.corr.ming.m4_m5\endcsname{0.28\ensuremath{\times}}
\expandafter\gdef\csname odunum@n@metric_structure.corr.ming.m4_m5\endcsname{1\,631}
\expandafter\gdef\csname odunum@val@metric_structure.corr.minicpm_o.demand_correct_m2\endcsname{0.38\ensuremath{\times}}
\expandafter\gdef\csname odunum@n@metric_structure.corr.minicpm_o.demand_correct_m2\endcsname{1\,629}
\expandafter\gdef\csname odunum@val@metric_structure.corr.minicpm_o.demand_correct_m3\endcsname{0.81\ensuremath{\times}}
\expandafter\gdef\csname odunum@n@metric_structure.corr.minicpm_o.demand_correct_m3\endcsname{1\,629}
\expandafter\gdef\csname odunum@val@metric_structure.corr.minicpm_o.demand_correct_m4\endcsname{0.27\ensuremath{\times}}
\expandafter\gdef\csname odunum@n@metric_structure.corr.minicpm_o.demand_correct_m4\endcsname{1\,629}
\expandafter\gdef\csname odunum@val@metric_structure.corr.minicpm_o.demand_correct_m5\endcsname{0.62\ensuremath{\times}}
\expandafter\gdef\csname odunum@n@metric_structure.corr.minicpm_o.demand_correct_m5\endcsname{1\,629}
\expandafter\gdef\csname odunum@val@metric_structure.corr.minicpm_o.m2_m3\endcsname{0.39\ensuremath{\times}}
\expandafter\gdef\csname odunum@n@metric_structure.corr.minicpm_o.m2_m3\endcsname{1\,629}
\expandafter\gdef\csname odunum@val@metric_structure.corr.minicpm_o.m2_m4\endcsname{0.62\ensuremath{\times}}
\expandafter\gdef\csname odunum@n@metric_structure.corr.minicpm_o.m2_m4\endcsname{1\,629}
\expandafter\gdef\csname odunum@val@metric_structure.corr.minicpm_o.m2_m5\endcsname{0.66\ensuremath{\times}}
\expandafter\gdef\csname odunum@n@metric_structure.corr.minicpm_o.m2_m5\endcsname{1\,629}
\expandafter\gdef\csname odunum@val@metric_structure.corr.minicpm_o.m3_m4\endcsname{0.24\ensuremath{\times}}
\expandafter\gdef\csname odunum@n@metric_structure.corr.minicpm_o.m3_m4\endcsname{1\,629}
\expandafter\gdef\csname odunum@val@metric_structure.corr.minicpm_o.m3_m5\endcsname{0.58\ensuremath{\times}}
\expandafter\gdef\csname odunum@n@metric_structure.corr.minicpm_o.m3_m5\endcsname{1\,629}
\expandafter\gdef\csname odunum@val@metric_structure.corr.minicpm_o.m4_m5\endcsname{0.61\ensuremath{\times}}
\expandafter\gdef\csname odunum@n@metric_structure.corr.minicpm_o.m4_m5\endcsname{1\,629}
\expandafter\gdef\csname odunum@val@metric_structure.corr.nemotron.demand_correct_m2\endcsname{0.68\ensuremath{\times}}
\expandafter\gdef\csname odunum@n@metric_structure.corr.nemotron.demand_correct_m2\endcsname{1\,631}
\expandafter\gdef\csname odunum@val@metric_structure.corr.nemotron.demand_correct_m3\endcsname{0.38\ensuremath{\times}}
\expandafter\gdef\csname odunum@n@metric_structure.corr.nemotron.demand_correct_m3\endcsname{1\,631}
\expandafter\gdef\csname odunum@val@metric_structure.corr.nemotron.demand_correct_m4\endcsname{0.64\ensuremath{\times}}
\expandafter\gdef\csname odunum@n@metric_structure.corr.nemotron.demand_correct_m4\endcsname{1\,631}
\expandafter\gdef\csname odunum@val@metric_structure.corr.nemotron.demand_correct_m5\endcsname{0.66\ensuremath{\times}}
\expandafter\gdef\csname odunum@n@metric_structure.corr.nemotron.demand_correct_m5\endcsname{1\,631}
\expandafter\gdef\csname odunum@val@metric_structure.corr.nemotron.m2_m3\endcsname{0.32\ensuremath{\times}}
\expandafter\gdef\csname odunum@n@metric_structure.corr.nemotron.m2_m3\endcsname{1\,631}
\expandafter\gdef\csname odunum@val@metric_structure.corr.nemotron.m2_m4\endcsname{0.76\ensuremath{\times}}
\expandafter\gdef\csname odunum@n@metric_structure.corr.nemotron.m2_m4\endcsname{1\,631}
\expandafter\gdef\csname odunum@val@metric_structure.corr.nemotron.m2_m5\endcsname{0.60\ensuremath{\times}}
\expandafter\gdef\csname odunum@n@metric_structure.corr.nemotron.m2_m5\endcsname{1\,631}
\expandafter\gdef\csname odunum@val@metric_structure.corr.nemotron.m3_m4\endcsname{0.40\ensuremath{\times}}
\expandafter\gdef\csname odunum@n@metric_structure.corr.nemotron.m3_m4\endcsname{1\,631}
\expandafter\gdef\csname odunum@val@metric_structure.corr.nemotron.m3_m5\endcsname{0.54\ensuremath{\times}}
\expandafter\gdef\csname odunum@n@metric_structure.corr.nemotron.m3_m5\endcsname{1\,631}
\expandafter\gdef\csname odunum@val@metric_structure.corr.nemotron.m4_m5\endcsname{0.62\ensuremath{\times}}
\expandafter\gdef\csname odunum@n@metric_structure.corr.nemotron.m4_m5\endcsname{1\,631}
\expandafter\gdef\csname odunum@val@metric_structure.corr.qwen25_omni.demand_correct_m2\endcsname{0.33\ensuremath{\times}}
\expandafter\gdef\csname odunum@n@metric_structure.corr.qwen25_omni.demand_correct_m2\endcsname{1\,631}
\expandafter\gdef\csname odunum@val@metric_structure.corr.qwen25_omni.demand_correct_m3\endcsname{0.31\ensuremath{\times}}
\expandafter\gdef\csname odunum@n@metric_structure.corr.qwen25_omni.demand_correct_m3\endcsname{1\,631}
\expandafter\gdef\csname odunum@val@metric_structure.corr.qwen25_omni.demand_correct_m4\endcsname{0.25\ensuremath{\times}}
\expandafter\gdef\csname odunum@n@metric_structure.corr.qwen25_omni.demand_correct_m4\endcsname{1\,631}
\expandafter\gdef\csname odunum@val@metric_structure.corr.qwen25_omni.demand_correct_m5\endcsname{0.45\ensuremath{\times}}
\expandafter\gdef\csname odunum@n@metric_structure.corr.qwen25_omni.demand_correct_m5\endcsname{1\,631}
\expandafter\gdef\csname odunum@val@metric_structure.corr.qwen25_omni.m2_m3\endcsname{0.25\ensuremath{\times}}
\expandafter\gdef\csname odunum@n@metric_structure.corr.qwen25_omni.m2_m3\endcsname{1\,631}
\expandafter\gdef\csname odunum@val@metric_structure.corr.qwen25_omni.m2_m4\endcsname{0.44\ensuremath{\times}}
\expandafter\gdef\csname odunum@n@metric_structure.corr.qwen25_omni.m2_m4\endcsname{1\,631}
\expandafter\gdef\csname odunum@val@metric_structure.corr.qwen25_omni.m2_m5\endcsname{0.39\ensuremath{\times}}
\expandafter\gdef\csname odunum@n@metric_structure.corr.qwen25_omni.m2_m5\endcsname{1\,631}
\expandafter\gdef\csname odunum@val@metric_structure.corr.qwen25_omni.m3_m4\endcsname{0.12\ensuremath{\times}}
\expandafter\gdef\csname odunum@n@metric_structure.corr.qwen25_omni.m3_m4\endcsname{1\,631}
\expandafter\gdef\csname odunum@val@metric_structure.corr.qwen25_omni.m3_m5\endcsname{0.27\ensuremath{\times}}
\expandafter\gdef\csname odunum@n@metric_structure.corr.qwen25_omni.m3_m5\endcsname{1\,631}
\expandafter\gdef\csname odunum@val@metric_structure.corr.qwen25_omni.m4_m5\endcsname{0.42\ensuremath{\times}}
\expandafter\gdef\csname odunum@n@metric_structure.corr.qwen25_omni.m4_m5\endcsname{1\,631}
\expandafter\gdef\csname odunum@val@metric_structure.corr.qwen3_omni_instruct.demand_correct_m2\endcsname{0.24\ensuremath{\times}}
\expandafter\gdef\csname odunum@n@metric_structure.corr.qwen3_omni_instruct.demand_correct_m2\endcsname{1\,631}
\expandafter\gdef\csname odunum@val@metric_structure.corr.qwen3_omni_instruct.demand_correct_m3\endcsname{0.06\ensuremath{\times}}
\expandafter\gdef\csname odunum@n@metric_structure.corr.qwen3_omni_instruct.demand_correct_m3\endcsname{1\,631}
\expandafter\gdef\csname odunum@val@metric_structure.corr.qwen3_omni_instruct.demand_correct_m4\endcsname{0.31\ensuremath{\times}}
\expandafter\gdef\csname odunum@n@metric_structure.corr.qwen3_omni_instruct.demand_correct_m4\endcsname{1\,631}
\expandafter\gdef\csname odunum@val@metric_structure.corr.qwen3_omni_instruct.demand_correct_m5\endcsname{0.82\ensuremath{\times}}
\expandafter\gdef\csname odunum@n@metric_structure.corr.qwen3_omni_instruct.demand_correct_m5\endcsname{1\,631}
\expandafter\gdef\csname odunum@val@metric_structure.corr.qwen3_omni_instruct.m2_m3\endcsname{0.11\ensuremath{\times}}
\expandafter\gdef\csname odunum@n@metric_structure.corr.qwen3_omni_instruct.m2_m3\endcsname{1\,631}
\expandafter\gdef\csname odunum@val@metric_structure.corr.qwen3_omni_instruct.m2_m4\endcsname{0.23\ensuremath{\times}}
\expandafter\gdef\csname odunum@n@metric_structure.corr.qwen3_omni_instruct.m2_m4\endcsname{1\,631}
\expandafter\gdef\csname odunum@val@metric_structure.corr.qwen3_omni_instruct.m2_m5\endcsname{0.31\ensuremath{\times}}
\expandafter\gdef\csname odunum@n@metric_structure.corr.qwen3_omni_instruct.m2_m5\endcsname{1\,631}
\expandafter\gdef\csname odunum@val@metric_structure.corr.qwen3_omni_instruct.m3_m4\endcsname{0.02\ensuremath{\times}}
\expandafter\gdef\csname odunum@n@metric_structure.corr.qwen3_omni_instruct.m3_m4\endcsname{1\,631}
\expandafter\gdef\csname odunum@val@metric_structure.corr.qwen3_omni_instruct.m3_m5\endcsname{0.10\ensuremath{\times}}
\expandafter\gdef\csname odunum@n@metric_structure.corr.qwen3_omni_instruct.m3_m5\endcsname{1\,631}
\expandafter\gdef\csname odunum@val@metric_structure.corr.qwen3_omni_instruct.m4_m5\endcsname{0.41\ensuremath{\times}}
\expandafter\gdef\csname odunum@n@metric_structure.corr.qwen3_omni_instruct.m4_m5\endcsname{1\,631}
\expandafter\gdef\csname odunum@val@metric_structure.corr.qwen3_omni_think.demand_correct_m2\endcsname{0.29\ensuremath{\times}}
\expandafter\gdef\csname odunum@n@metric_structure.corr.qwen3_omni_think.demand_correct_m2\endcsname{1\,631}
\expandafter\gdef\csname odunum@val@metric_structure.corr.qwen3_omni_think.demand_correct_m3\endcsname{0.06\ensuremath{\times}}
\expandafter\gdef\csname odunum@n@metric_structure.corr.qwen3_omni_think.demand_correct_m3\endcsname{1\,631}
\expandafter\gdef\csname odunum@val@metric_structure.corr.qwen3_omni_think.demand_correct_m4\endcsname{0.31\ensuremath{\times}}
\expandafter\gdef\csname odunum@n@metric_structure.corr.qwen3_omni_think.demand_correct_m4\endcsname{1\,631}
\expandafter\gdef\csname odunum@val@metric_structure.corr.qwen3_omni_think.demand_correct_m5\endcsname{0.83\ensuremath{\times}}
\expandafter\gdef\csname odunum@n@metric_structure.corr.qwen3_omni_think.demand_correct_m5\endcsname{1\,631}
\expandafter\gdef\csname odunum@val@metric_structure.corr.qwen3_omni_think.m2_m3\endcsname{0.20\ensuremath{\times}}
\expandafter\gdef\csname odunum@n@metric_structure.corr.qwen3_omni_think.m2_m3\endcsname{1\,631}
\expandafter\gdef\csname odunum@val@metric_structure.corr.qwen3_omni_think.m2_m4\endcsname{0.21\ensuremath{\times}}
\expandafter\gdef\csname odunum@n@metric_structure.corr.qwen3_omni_think.m2_m4\endcsname{1\,631}
\expandafter\gdef\csname odunum@val@metric_structure.corr.qwen3_omni_think.m2_m5\endcsname{0.34\ensuremath{\times}}
\expandafter\gdef\csname odunum@n@metric_structure.corr.qwen3_omni_think.m2_m5\endcsname{1\,631}
\expandafter\gdef\csname odunum@val@metric_structure.corr.qwen3_omni_think.m3_m4\endcsname{0.20\ensuremath{\times}}
\expandafter\gdef\csname odunum@n@metric_structure.corr.qwen3_omni_think.m3_m4\endcsname{1\,631}
\expandafter\gdef\csname odunum@val@metric_structure.corr.qwen3_omni_think.m3_m5\endcsname{0.14\ensuremath{\times}}
\expandafter\gdef\csname odunum@n@metric_structure.corr.qwen3_omni_think.m3_m5\endcsname{1\,631}
\expandafter\gdef\csname odunum@val@metric_structure.corr.qwen3_omni_think.m4_m5\endcsname{0.36\ensuremath{\times}}
\expandafter\gdef\csname odunum@n@metric_structure.corr.qwen3_omni_think.m4_m5\endcsname{1\,631}
\expandafter\gdef\csname odunum@val@metric_structure.corr.qwen_plus.demand_correct_m2\endcsname{0.31\ensuremath{\times}}
\expandafter\gdef\csname odunum@n@metric_structure.corr.qwen_plus.demand_correct_m2\endcsname{1\,630}
\expandafter\gdef\csname odunum@val@metric_structure.corr.qwen_plus.demand_correct_m3\endcsname{-0.16\ensuremath{\times}}
\expandafter\gdef\csname odunum@n@metric_structure.corr.qwen_plus.demand_correct_m3\endcsname{1\,630}
\expandafter\gdef\csname odunum@val@metric_structure.corr.qwen_plus.demand_correct_m4\endcsname{0.39\ensuremath{\times}}
\expandafter\gdef\csname odunum@n@metric_structure.corr.qwen_plus.demand_correct_m4\endcsname{1\,630}
\expandafter\gdef\csname odunum@val@metric_structure.corr.qwen_plus.demand_correct_m5\endcsname{0.90\ensuremath{\times}}
\expandafter\gdef\csname odunum@n@metric_structure.corr.qwen_plus.demand_correct_m5\endcsname{1\,630}
\expandafter\gdef\csname odunum@val@metric_structure.corr.qwen_plus.m2_m3\endcsname{0.13\ensuremath{\times}}
\expandafter\gdef\csname odunum@n@metric_structure.corr.qwen_plus.m2_m3\endcsname{1\,630}
\expandafter\gdef\csname odunum@val@metric_structure.corr.qwen_plus.m2_m4\endcsname{0.25\ensuremath{\times}}
\expandafter\gdef\csname odunum@n@metric_structure.corr.qwen_plus.m2_m4\endcsname{1\,630}
\expandafter\gdef\csname odunum@val@metric_structure.corr.qwen_plus.m2_m5\endcsname{0.34\ensuremath{\times}}
\expandafter\gdef\csname odunum@n@metric_structure.corr.qwen_plus.m2_m5\endcsname{1\,630}
\expandafter\gdef\csname odunum@val@metric_structure.corr.qwen_plus.m3_m4\endcsname{0.14\ensuremath{\times}}
\expandafter\gdef\csname odunum@n@metric_structure.corr.qwen_plus.m3_m4\endcsname{1\,630}
\expandafter\gdef\csname odunum@val@metric_structure.corr.qwen_plus.m3_m5\endcsname{-0.09\ensuremath{\times}}
\expandafter\gdef\csname odunum@n@metric_structure.corr.qwen_plus.m3_m5\endcsname{1\,630}
\expandafter\gdef\csname odunum@val@metric_structure.corr.qwen_plus.m4_m5\endcsname{0.41\ensuremath{\times}}
\expandafter\gdef\csname odunum@n@metric_structure.corr.qwen_plus.m4_m5\endcsname{1\,630}
\expandafter\gdef\csname odunum@val@metric_structure.corr.salmonn2_7b.demand_correct_m2\endcsname{0.82\ensuremath{\times}}
\expandafter\gdef\csname odunum@n@metric_structure.corr.salmonn2_7b.demand_correct_m2\endcsname{1\,614}
\expandafter\gdef\csname odunum@val@metric_structure.corr.salmonn2_7b.demand_correct_m3\endcsname{0.72\ensuremath{\times}}
\expandafter\gdef\csname odunum@n@metric_structure.corr.salmonn2_7b.demand_correct_m3\endcsname{1\,614}
\expandafter\gdef\csname odunum@val@metric_structure.corr.salmonn2_7b.demand_correct_m4\endcsname{0.81\ensuremath{\times}}
\expandafter\gdef\csname odunum@n@metric_structure.corr.salmonn2_7b.demand_correct_m4\endcsname{1\,614}
\expandafter\gdef\csname odunum@val@metric_structure.corr.salmonn2_7b.demand_correct_m5\endcsname{0.69\ensuremath{\times}}
\expandafter\gdef\csname odunum@n@metric_structure.corr.salmonn2_7b.demand_correct_m5\endcsname{1\,614}
\expandafter\gdef\csname odunum@val@metric_structure.corr.salmonn2_7b.m2_m3\endcsname{0.73\ensuremath{\times}}
\expandafter\gdef\csname odunum@n@metric_structure.corr.salmonn2_7b.m2_m3\endcsname{1\,614}
\expandafter\gdef\csname odunum@val@metric_structure.corr.salmonn2_7b.m2_m4\endcsname{0.92\ensuremath{\times}}
\expandafter\gdef\csname odunum@n@metric_structure.corr.salmonn2_7b.m2_m4\endcsname{1\,614}
\expandafter\gdef\csname odunum@val@metric_structure.corr.salmonn2_7b.m2_m5\endcsname{0.84\ensuremath{\times}}
\expandafter\gdef\csname odunum@n@metric_structure.corr.salmonn2_7b.m2_m5\endcsname{1\,614}
\expandafter\gdef\csname odunum@val@metric_structure.corr.salmonn2_7b.m3_m4\endcsname{0.66\ensuremath{\times}}
\expandafter\gdef\csname odunum@n@metric_structure.corr.salmonn2_7b.m3_m4\endcsname{1\,614}
\expandafter\gdef\csname odunum@val@metric_structure.corr.salmonn2_7b.m3_m5\endcsname{0.85\ensuremath{\times}}
\expandafter\gdef\csname odunum@n@metric_structure.corr.salmonn2_7b.m3_m5\endcsname{1\,614}
\expandafter\gdef\csname odunum@val@metric_structure.corr.salmonn2_7b.m4_m5\endcsname{0.85\ensuremath{\times}}
\expandafter\gdef\csname odunum@n@metric_structure.corr.salmonn2_7b.m4_m5\endcsname{1\,614}
\expandafter\gdef\csname odunum@val@metric_structure.corr.seed.demand_correct_m2\endcsname{0.37\ensuremath{\times}}
\expandafter\gdef\csname odunum@n@metric_structure.corr.seed.demand_correct_m2\endcsname{1\,631}
\expandafter\gdef\csname odunum@val@metric_structure.corr.seed.demand_correct_m3\endcsname{-0.27\ensuremath{\times}}
\expandafter\gdef\csname odunum@n@metric_structure.corr.seed.demand_correct_m3\endcsname{1\,631}
\expandafter\gdef\csname odunum@val@metric_structure.corr.seed.demand_correct_m4\endcsname{0.43\ensuremath{\times}}
\expandafter\gdef\csname odunum@n@metric_structure.corr.seed.demand_correct_m4\endcsname{1\,631}
\expandafter\gdef\csname odunum@val@metric_structure.corr.seed.demand_correct_m5\endcsname{0.89\ensuremath{\times}}
\expandafter\gdef\csname odunum@n@metric_structure.corr.seed.demand_correct_m5\endcsname{1\,631}
\expandafter\gdef\csname odunum@val@metric_structure.corr.seed.m2_m3\endcsname{0.05\ensuremath{\times}}
\expandafter\gdef\csname odunum@n@metric_structure.corr.seed.m2_m3\endcsname{1\,631}
\expandafter\gdef\csname odunum@val@metric_structure.corr.seed.m2_m4\endcsname{0.34\ensuremath{\times}}
\expandafter\gdef\csname odunum@n@metric_structure.corr.seed.m2_m4\endcsname{1\,631}
\expandafter\gdef\csname odunum@val@metric_structure.corr.seed.m2_m5\endcsname{0.39\ensuremath{\times}}
\expandafter\gdef\csname odunum@n@metric_structure.corr.seed.m2_m5\endcsname{1\,631}
\expandafter\gdef\csname odunum@val@metric_structure.corr.seed.m3_m4\endcsname{0.10\ensuremath{\times}}
\expandafter\gdef\csname odunum@n@metric_structure.corr.seed.m3_m4\endcsname{1\,631}
\expandafter\gdef\csname odunum@val@metric_structure.corr.seed.m3_m5\endcsname{-0.22\ensuremath{\times}}
\expandafter\gdef\csname odunum@n@metric_structure.corr.seed.m3_m5\endcsname{1\,631}
\expandafter\gdef\csname odunum@val@metric_structure.corr.seed.m4_m5\endcsname{0.44\ensuremath{\times}}
\expandafter\gdef\csname odunum@n@metric_structure.corr.seed.m4_m5\endcsname{1\,631}
\expandafter\gdef\csname odunum@val@metric_structure.panel_corr.demand_correct_m2\endcsname{0.40\ensuremath{\times}}
\expandafter\gdef\csname odunum@n@metric_structure.panel_corr.demand_correct_m2\endcsname{13}
\expandafter\gdef\csname odunum@ci@metric_structure.panel_corr.demand_correct_m2\endcsname{[0.31\ensuremath{\times}, 0.50\ensuremath{\times}]}
\expandafter\gdef\csname odunum@val@metric_structure.panel_corr.demand_correct_m3\endcsname{0.16\ensuremath{\times}}
\expandafter\gdef\csname odunum@n@metric_structure.panel_corr.demand_correct_m3\endcsname{13}
\expandafter\gdef\csname odunum@ci@metric_structure.panel_corr.demand_correct_m3\endcsname{[-0.04\ensuremath{\times}, 0.37\ensuremath{\times}]}
\expandafter\gdef\csname odunum@val@metric_structure.panel_corr.demand_correct_m4\endcsname{0.42\ensuremath{\times}}
\expandafter\gdef\csname odunum@n@metric_structure.panel_corr.demand_correct_m4\endcsname{13}
\expandafter\gdef\csname odunum@ci@metric_structure.panel_corr.demand_correct_m4\endcsname{[0.34\ensuremath{\times}, 0.51\ensuremath{\times}]}
\expandafter\gdef\csname odunum@val@metric_structure.panel_corr.demand_correct_m5\endcsname{0.75\ensuremath{\times}}
\expandafter\gdef\csname odunum@n@metric_structure.panel_corr.demand_correct_m5\endcsname{13}
\expandafter\gdef\csname odunum@ci@metric_structure.panel_corr.demand_correct_m5\endcsname{[0.67\ensuremath{\times}, 0.84\ensuremath{\times}]}
\expandafter\gdef\csname odunum@val@metric_structure.panel_corr.m2_m3\endcsname{0.25\ensuremath{\times}}
\expandafter\gdef\csname odunum@n@metric_structure.panel_corr.m2_m3\endcsname{13}
\expandafter\gdef\csname odunum@ci@metric_structure.panel_corr.m2_m3\endcsname{[0.14\ensuremath{\times}, 0.36\ensuremath{\times}]}
\expandafter\gdef\csname odunum@val@metric_structure.panel_corr.m2_m4\endcsname{0.41\ensuremath{\times}}
\expandafter\gdef\csname odunum@n@metric_structure.panel_corr.m2_m4\endcsname{13}
\expandafter\gdef\csname odunum@ci@metric_structure.panel_corr.m2_m4\endcsname{[0.29\ensuremath{\times}, 0.54\ensuremath{\times}]}
\expandafter\gdef\csname odunum@val@metric_structure.panel_corr.m2_m5\endcsname{0.43\ensuremath{\times}}
\expandafter\gdef\csname odunum@n@metric_structure.panel_corr.m2_m5\endcsname{13}
\expandafter\gdef\csname odunum@ci@metric_structure.panel_corr.m2_m5\endcsname{[0.35\ensuremath{\times}, 0.53\ensuremath{\times}]}
\expandafter\gdef\csname odunum@val@metric_structure.panel_corr.m3_m4\endcsname{0.25\ensuremath{\times}}
\expandafter\gdef\csname odunum@n@metric_structure.panel_corr.m3_m4\endcsname{13}
\expandafter\gdef\csname odunum@ci@metric_structure.panel_corr.m3_m4\endcsname{[0.12\ensuremath{\times}, 0.39\ensuremath{\times}]}
\expandafter\gdef\csname odunum@val@metric_structure.panel_corr.m3_m5\endcsname{0.19\ensuremath{\times}}
\expandafter\gdef\csname odunum@n@metric_structure.panel_corr.m3_m5\endcsname{13}
\expandafter\gdef\csname odunum@ci@metric_structure.panel_corr.m3_m5\endcsname{[0.01\ensuremath{\times}, 0.38\ensuremath{\times}]}
\expandafter\gdef\csname odunum@val@metric_structure.panel_corr.m4_m5\endcsname{0.46\ensuremath{\times}}
\expandafter\gdef\csname odunum@n@metric_structure.panel_corr.m4_m5\endcsname{13}
\expandafter\gdef\csname odunum@ci@metric_structure.panel_corr.m4_m5\endcsname{[0.38\ensuremath{\times}, 0.55\ensuremath{\times}]}
\expandafter\gdef\csname odunum@val@metric_structure.panel_corr.mean_abs_offdiagonal\endcsname{0.40\ensuremath{\times}}
\expandafter\gdef\csname odunum@n@metric_structure.panel_corr.mean_abs_offdiagonal\endcsname{130}
\expandafter\gdef\csname odunum@val@metric_structure.panel_corr.mean_abs_with_m2\endcsname{0.37\ensuremath{\times}}
\expandafter\gdef\csname odunum@n@metric_structure.panel_corr.mean_abs_with_m2\endcsname{52}
\expandafter\gdef\csname odunum@val@metric_structure.pop.cascade_asr.n_scenes\endcsname{1\,631}
\expandafter\gdef\csname odunum@n@metric_structure.pop.cascade_asr.n_scenes\endcsname{2\,078}
\expandafter\gdef\csname odunum@val@metric_structure.pop.gemini.n_scenes\endcsname{1\,631}
\expandafter\gdef\csname odunum@n@metric_structure.pop.gemini.n_scenes\endcsname{2\,078}
\expandafter\gdef\csname odunum@val@metric_structure.pop.gemini35_flash_lite.n_scenes\endcsname{1\,631}
\expandafter\gdef\csname odunum@n@metric_structure.pop.gemini35_flash_lite.n_scenes\endcsname{2\,078}
\expandafter\gdef\csname odunum@val@metric_structure.pop.gemini37_flash.n_scenes\endcsname{1\,627}
\expandafter\gdef\csname odunum@n@metric_structure.pop.gemini37_flash.n_scenes\endcsname{2\,067}
\expandafter\gdef\csname odunum@val@metric_structure.pop.ming.n_scenes\endcsname{1\,631}
\expandafter\gdef\csname odunum@n@metric_structure.pop.ming.n_scenes\endcsname{2\,078}
\expandafter\gdef\csname odunum@val@metric_structure.pop.minicpm_o.n_scenes\endcsname{1\,629}
\expandafter\gdef\csname odunum@n@metric_structure.pop.minicpm_o.n_scenes\endcsname{2\,075}
\expandafter\gdef\csname odunum@val@metric_structure.pop.models.n\endcsname{13}
\expandafter\gdef\csname odunum@n@metric_structure.pop.models.n\endcsname{13}
\expandafter\gdef\csname odunum@val@metric_structure.pop.nemotron.n_scenes\endcsname{1\,631}
\expandafter\gdef\csname odunum@n@metric_structure.pop.nemotron.n_scenes\endcsname{2\,078}
\expandafter\gdef\csname odunum@val@metric_structure.pop.qwen25_omni.n_scenes\endcsname{1\,631}
\expandafter\gdef\csname odunum@n@metric_structure.pop.qwen25_omni.n_scenes\endcsname{2\,078}
\expandafter\gdef\csname odunum@val@metric_structure.pop.qwen3_omni_instruct.n_scenes\endcsname{1\,631}
\expandafter\gdef\csname odunum@n@metric_structure.pop.qwen3_omni_instruct.n_scenes\endcsname{2\,078}
\expandafter\gdef\csname odunum@val@metric_structure.pop.qwen3_omni_think.n_scenes\endcsname{1\,631}
\expandafter\gdef\csname odunum@n@metric_structure.pop.qwen3_omni_think.n_scenes\endcsname{2\,078}
\expandafter\gdef\csname odunum@val@metric_structure.pop.qwen_plus.n_scenes\endcsname{1\,630}
\expandafter\gdef\csname odunum@n@metric_structure.pop.qwen_plus.n_scenes\endcsname{2\,077}
\expandafter\gdef\csname odunum@val@metric_structure.pop.salmonn2_7b.n_scenes\endcsname{1\,614}
\expandafter\gdef\csname odunum@n@metric_structure.pop.salmonn2_7b.n_scenes\endcsname{2\,058}
\expandafter\gdef\csname odunum@val@metric_structure.pop.seed.n_scenes\endcsname{1\,631}
\expandafter\gdef\csname odunum@n@metric_structure.pop.seed.n_scenes\endcsname{2\,078}
\expandafter\gdef\csname odunum@val@metric_structure.rank_stability.drop_m1.kendall_tau\endcsname{0.897}
\expandafter\gdef\csname odunum@n@metric_structure.rank_stability.drop_m1.kendall_tau\endcsname{13}
\expandafter\gdef\csname odunum@val@metric_structure.rank_stability.drop_m1.max_abs_delta\endcsname{0.056}
\expandafter\gdef\csname odunum@n@metric_structure.rank_stability.drop_m1.max_abs_delta\endcsname{13}
\expandafter\gdef\csname odunum@val@metric_structure.rank_stability.drop_m1.max_shift\endcsname{2}
\expandafter\gdef\csname odunum@n@metric_structure.rank_stability.drop_m1.max_shift\endcsname{13}
\expandafter\gdef\csname odunum@val@metric_structure.rank_stability.drop_m1.n_moved\endcsname{4}
\expandafter\gdef\csname odunum@n@metric_structure.rank_stability.drop_m1.n_moved\endcsname{13}
\expandafter\gdef\csname odunum@val@metric_structure.rank_stability.drop_m1.spearman\endcsname{0.973}
\expandafter\gdef\csname odunum@n@metric_structure.rank_stability.drop_m1.spearman\endcsname{13}
\expandafter\gdef\csname odunum@val@metric_structure.rank_stability.drop_m1.top1_unchanged\endcsname{0}
\expandafter\gdef\csname odunum@n@metric_structure.rank_stability.drop_m1.top1_unchanged\endcsname{13}
\expandafter\gdef\csname odunum@val@metric_structure.rank_stability.drop_m4.kendall_tau\endcsname{1.000}
\expandafter\gdef\csname odunum@n@metric_structure.rank_stability.drop_m4.kendall_tau\endcsname{13}
\expandafter\gdef\csname odunum@val@metric_structure.rank_stability.drop_m4.max_abs_delta\endcsname{0.035}
\expandafter\gdef\csname odunum@n@metric_structure.rank_stability.drop_m4.max_abs_delta\endcsname{13}
\expandafter\gdef\csname odunum@val@metric_structure.rank_stability.drop_m4.max_shift\endcsname{0}
\expandafter\gdef\csname odunum@n@metric_structure.rank_stability.drop_m4.max_shift\endcsname{13}
\expandafter\gdef\csname odunum@val@metric_structure.rank_stability.drop_m4.n_moved\endcsname{0}
\expandafter\gdef\csname odunum@n@metric_structure.rank_stability.drop_m4.n_moved\endcsname{13}
\expandafter\gdef\csname odunum@val@metric_structure.rank_stability.drop_m4.spearman\endcsname{1.000}
\expandafter\gdef\csname odunum@n@metric_structure.rank_stability.drop_m4.spearman\endcsname{13}
\expandafter\gdef\csname odunum@val@metric_structure.rank_stability.drop_m4.top1_unchanged\endcsname{1}
\expandafter\gdef\csname odunum@n@metric_structure.rank_stability.drop_m4.top1_unchanged\endcsname{13}
\expandafter\gdef\csname odunum@val@metric_structure.rank_stability.equal.kendall_tau\endcsname{0.872}
\expandafter\gdef\csname odunum@n@metric_structure.rank_stability.equal.kendall_tau\endcsname{13}
\expandafter\gdef\csname odunum@val@metric_structure.rank_stability.equal.max_abs_delta\endcsname{0.090}
\expandafter\gdef\csname odunum@n@metric_structure.rank_stability.equal.max_abs_delta\endcsname{13}
\expandafter\gdef\csname odunum@val@metric_structure.rank_stability.equal.max_shift\endcsname{3}
\expandafter\gdef\csname odunum@n@metric_structure.rank_stability.equal.max_shift\endcsname{13}
\expandafter\gdef\csname odunum@val@metric_structure.rank_stability.equal.n_moved\endcsname{7}
\expandafter\gdef\csname odunum@n@metric_structure.rank_stability.equal.n_moved\endcsname{13}
\expandafter\gdef\csname odunum@val@metric_structure.rank_stability.equal.spearman\endcsname{0.951}
\expandafter\gdef\csname odunum@n@metric_structure.rank_stability.equal.spearman\endcsname{13}
\expandafter\gdef\csname odunum@val@metric_structure.rank_stability.equal.top1_unchanged\endcsname{1}
\expandafter\gdef\csname odunum@n@metric_structure.rank_stability.equal.top1_unchanged\endcsname{13}
\expandafter\gdef\csname odunum@val@metric_structure.rank_stability.m2_only.kendall_tau\endcsname{0.897}
\expandafter\gdef\csname odunum@n@metric_structure.rank_stability.m2_only.kendall_tau\endcsname{13}
\expandafter\gdef\csname odunum@val@metric_structure.rank_stability.m2_only.max_abs_delta\endcsname{0.085}
\expandafter\gdef\csname odunum@n@metric_structure.rank_stability.m2_only.max_abs_delta\endcsname{13}
\expandafter\gdef\csname odunum@val@metric_structure.rank_stability.m2_only.max_shift\endcsname{2}
\expandafter\gdef\csname odunum@n@metric_structure.rank_stability.m2_only.max_shift\endcsname{13}
\expandafter\gdef\csname odunum@val@metric_structure.rank_stability.m2_only.n_moved\endcsname{4}
\expandafter\gdef\csname odunum@n@metric_structure.rank_stability.m2_only.n_moved\endcsname{13}
\expandafter\gdef\csname odunum@val@metric_structure.rank_stability.m2_only.spearman\endcsname{0.973}
\expandafter\gdef\csname odunum@n@metric_structure.rank_stability.m2_only.spearman\endcsname{13}
\expandafter\gdef\csname odunum@val@metric_structure.rank_stability.m2_only.top1_unchanged\endcsname{0}
\expandafter\gdef\csname odunum@n@metric_structure.rank_stability.m2_only.top1_unchanged\endcsname{13}
\expandafter\gdef\csname odunum@val@modality_ablation.audit.novideo.doubao_seed_2_0_lite.judged_points\endcsname{3\,973}
\expandafter\gdef\csname odunum@n@modality_ablation.audit.novideo.doubao_seed_2_0_lite.judged_points\endcsname{3\,973}
\expandafter\gdef\csname odunum@val@modality_ablation.audit.novideo.doubao_seed_2_0_lite.scenes_incomplete\endcsname{0}
\expandafter\gdef\csname odunum@n@modality_ablation.audit.novideo.doubao_seed_2_0_lite.scenes_incomplete\endcsname{1\,061}
\expandafter\gdef\csname odunum@val@modality_ablation.audit.novideo.gemini_3_1_pro.judged_points\endcsname{3\,973}
\expandafter\gdef\csname odunum@n@modality_ablation.audit.novideo.gemini_3_1_pro.judged_points\endcsname{3\,973}
\expandafter\gdef\csname odunum@val@modality_ablation.audit.novideo.gemini_3_1_pro.scenes_incomplete\endcsname{0}
\expandafter\gdef\csname odunum@n@modality_ablation.audit.novideo.gemini_3_1_pro.scenes_incomplete\endcsname{1\,061}
\expandafter\gdef\csname odunum@val@modality_ablation.audit.novideo.qwen3_5_omni_plus.judged_points\endcsname{3\,973}
\expandafter\gdef\csname odunum@n@modality_ablation.audit.novideo.qwen3_5_omni_plus.judged_points\endcsname{3\,973}
\expandafter\gdef\csname odunum@val@modality_ablation.audit.novideo.qwen3_5_omni_plus.scenes_incomplete\endcsname{0}
\expandafter\gdef\csname odunum@n@modality_ablation.audit.novideo.qwen3_5_omni_plus.scenes_incomplete\endcsname{1\,061}
\expandafter\gdef\csname odunum@val@modality_ablation.audit.paired.doubao_seed_2_0_lite\endcsname{3\,959}
\expandafter\gdef\csname odunum@n@modality_ablation.audit.paired.doubao_seed_2_0_lite\endcsname{3\,973}
\expandafter\gdef\csname odunum@val@modality_ablation.audit.paired.gemini_3_1_pro\endcsname{3\,943}
\expandafter\gdef\csname odunum@n@modality_ablation.audit.paired.gemini_3_1_pro\endcsname{3\,973}
\expandafter\gdef\csname odunum@val@modality_ablation.audit.paired.qwen3_5_omni_plus\endcsname{3\,928}
\expandafter\gdef\csname odunum@n@modality_ablation.audit.paired.qwen3_5_omni_plus\endcsname{3\,973}
\expandafter\gdef\csname odunum@val@modality_ablation.audit.reader_matches_keypoint_types\endcsname{3\,973}
\expandafter\gdef\csname odunum@n@modality_ablation.audit.reader_matches_keypoint_types\endcsname{3\,973}
\expandafter\gdef\csname odunum@val@modality_ablation.audit.withvideo.doubao_seed_2_0_lite.judged_points\endcsname{3\,973}
\expandafter\gdef\csname odunum@n@modality_ablation.audit.withvideo.doubao_seed_2_0_lite.judged_points\endcsname{3\,973}
\expandafter\gdef\csname odunum@val@modality_ablation.audit.withvideo.doubao_seed_2_0_lite.scenes_incomplete\endcsname{0}
\expandafter\gdef\csname odunum@n@modality_ablation.audit.withvideo.doubao_seed_2_0_lite.scenes_incomplete\endcsname{1\,061}
\expandafter\gdef\csname odunum@val@modality_ablation.audit.withvideo.gemini_3_1_pro.judged_points\endcsname{3\,973}
\expandafter\gdef\csname odunum@n@modality_ablation.audit.withvideo.gemini_3_1_pro.judged_points\endcsname{3\,973}
\expandafter\gdef\csname odunum@val@modality_ablation.audit.withvideo.gemini_3_1_pro.scenes_incomplete\endcsname{0}
\expandafter\gdef\csname odunum@n@modality_ablation.audit.withvideo.gemini_3_1_pro.scenes_incomplete\endcsname{1\,061}
\expandafter\gdef\csname odunum@val@modality_ablation.audit.withvideo.qwen3_5_omni_plus.judged_points\endcsname{3\,969}
\expandafter\gdef\csname odunum@n@modality_ablation.audit.withvideo.qwen3_5_omni_plus.judged_points\endcsname{3\,973}
\expandafter\gdef\csname odunum@val@modality_ablation.audit.withvideo.qwen3_5_omni_plus.scenes_incomplete\endcsname{0}
\expandafter\gdef\csname odunum@n@modality_ablation.audit.withvideo.qwen3_5_omni_plus.scenes_incomplete\endcsname{1\,061}
\expandafter\gdef\csname odunum@val@modality_ablation.delta.doubao_seed_2_0_lite.audio\endcsname{0.011}
\expandafter\gdef\csname odunum@n@modality_ablation.delta.doubao_seed_2_0_lite.audio\endcsname{557}
\expandafter\gdef\csname odunum@ci@modality_ablation.delta.doubao_seed_2_0_lite.audio\endcsname{[-0.033, 0.055]}
\expandafter\gdef\csname odunum@val@modality_ablation.delta.doubao_seed_2_0_lite.context\endcsname{0.157}
\expandafter\gdef\csname odunum@n@modality_ablation.delta.doubao_seed_2_0_lite.context\endcsname{1\,969}
\expandafter\gdef\csname odunum@ci@modality_ablation.delta.doubao_seed_2_0_lite.context\endcsname{[0.131, 0.183]}
\expandafter\gdef\csname odunum@val@modality_ablation.delta.doubao_seed_2_0_lite.history\endcsname{-0.002}
\expandafter\gdef\csname odunum@n@modality_ablation.delta.doubao_seed_2_0_lite.history\endcsname{474}
\expandafter\gdef\csname odunum@ci@modality_ablation.delta.doubao_seed_2_0_lite.history\endcsname{[-0.049, 0.045]}
\expandafter\gdef\csname odunum@val@modality_ablation.delta.doubao_seed_2_0_lite.intent\endcsname{-0.102}
\expandafter\gdef\csname odunum@n@modality_ablation.delta.doubao_seed_2_0_lite.intent\endcsname{1\,977}
\expandafter\gdef\csname odunum@ci@modality_ablation.delta.doubao_seed_2_0_lite.intent\endcsname{[-0.126, -0.079]}
\expandafter\gdef\csname odunum@val@modality_ablation.delta.doubao_seed_2_0_lite.other_context\endcsname{0.154}
\expandafter\gdef\csname odunum@n@modality_ablation.delta.doubao_seed_2_0_lite.other_context\endcsname{136}
\expandafter\gdef\csname odunum@ci@modality_ablation.delta.doubao_seed_2_0_lite.other_context\endcsname{[0.070, 0.241]}
\expandafter\gdef\csname odunum@val@modality_ablation.delta.doubao_seed_2_0_lite.overall\endcsname{0.028}
\expandafter\gdef\csname odunum@n@modality_ablation.delta.doubao_seed_2_0_lite.overall\endcsname{3\,959}
\expandafter\gdef\csname odunum@ci@modality_ablation.delta.doubao_seed_2_0_lite.overall\endcsname{[0.008, 0.047]}
\expandafter\gdef\csname odunum@val@modality_ablation.delta.doubao_seed_2_0_lite.unattributed\endcsname{0.231}
\expandafter\gdef\csname odunum@n@modality_ablation.delta.doubao_seed_2_0_lite.unattributed\endcsname{13}
\expandafter\gdef\csname odunum@ci@modality_ablation.delta.doubao_seed_2_0_lite.unattributed\endcsname{[-0.125, 0.600]}
\expandafter\gdef\csname odunum@val@modality_ablation.delta.doubao_seed_2_0_lite.visual\endcsname{0.354}
\expandafter\gdef\csname odunum@n@modality_ablation.delta.doubao_seed_2_0_lite.visual\endcsname{802}
\expandafter\gdef\csname odunum@ci@modality_ablation.delta.doubao_seed_2_0_lite.visual\endcsname{[0.315, 0.394]}
\expandafter\gdef\csname odunum@val@modality_ablation.delta.gemini_3_1_pro.audio\endcsname{0.043}
\expandafter\gdef\csname odunum@n@modality_ablation.delta.gemini_3_1_pro.audio\endcsname{553}
\expandafter\gdef\csname odunum@ci@modality_ablation.delta.gemini_3_1_pro.audio\endcsname{[0.002, 0.085]}
\expandafter\gdef\csname odunum@val@modality_ablation.delta.gemini_3_1_pro.context\endcsname{0.143}
\expandafter\gdef\csname odunum@n@modality_ablation.delta.gemini_3_1_pro.context\endcsname{1\,960}
\expandafter\gdef\csname odunum@ci@modality_ablation.delta.gemini_3_1_pro.context\endcsname{[0.120, 0.166]}
\expandafter\gdef\csname odunum@val@modality_ablation.delta.gemini_3_1_pro.history\endcsname{0.038}
\expandafter\gdef\csname odunum@n@modality_ablation.delta.gemini_3_1_pro.history\endcsname{475}
\expandafter\gdef\csname odunum@ci@modality_ablation.delta.gemini_3_1_pro.history\endcsname{[-0.002, 0.077]}
\expandafter\gdef\csname odunum@val@modality_ablation.delta.gemini_3_1_pro.intent\endcsname{0.006}
\expandafter\gdef\csname odunum@n@modality_ablation.delta.gemini_3_1_pro.intent\endcsname{1\,970}
\expandafter\gdef\csname odunum@ci@modality_ablation.delta.gemini_3_1_pro.intent\endcsname{[-0.012, 0.024]}
\expandafter\gdef\csname odunum@val@modality_ablation.delta.gemini_3_1_pro.other_context\endcsname{-0.007}
\expandafter\gdef\csname odunum@n@modality_ablation.delta.gemini_3_1_pro.other_context\endcsname{136}
\expandafter\gdef\csname odunum@ci@modality_ablation.delta.gemini_3_1_pro.other_context\endcsname{[-0.089, 0.073]}
\expandafter\gdef\csname odunum@val@modality_ablation.delta.gemini_3_1_pro.overall\endcsname{0.074}
\expandafter\gdef\csname odunum@n@modality_ablation.delta.gemini_3_1_pro.overall\endcsname{3\,943}
\expandafter\gdef\csname odunum@ci@modality_ablation.delta.gemini_3_1_pro.overall\endcsname{[0.058, 0.090]}
\expandafter\gdef\csname odunum@val@modality_ablation.delta.gemini_3_1_pro.unattributed\endcsname{-0.077}
\expandafter\gdef\csname odunum@n@modality_ablation.delta.gemini_3_1_pro.unattributed\endcsname{13}
\expandafter\gdef\csname odunum@ci@modality_ablation.delta.gemini_3_1_pro.unattributed\endcsname{[-0.300, 0.000]}
\expandafter\gdef\csname odunum@val@modality_ablation.delta.gemini_3_1_pro.visual\endcsname{0.300}
\expandafter\gdef\csname odunum@n@modality_ablation.delta.gemini_3_1_pro.visual\endcsname{796}
\expandafter\gdef\csname odunum@ci@modality_ablation.delta.gemini_3_1_pro.visual\endcsname{[0.266, 0.337]}
\expandafter\gdef\csname odunum@val@modality_ablation.delta.qwen3_5_omni_plus.audio\endcsname{0.047}
\expandafter\gdef\csname odunum@n@modality_ablation.delta.qwen3_5_omni_plus.audio\endcsname{549}
\expandafter\gdef\csname odunum@ci@modality_ablation.delta.qwen3_5_omni_plus.audio\endcsname{[0.007, 0.088]}
\expandafter\gdef\csname odunum@val@modality_ablation.delta.qwen3_5_omni_plus.context\endcsname{0.175}
\expandafter\gdef\csname odunum@n@modality_ablation.delta.qwen3_5_omni_plus.context\endcsname{1\,953}
\expandafter\gdef\csname odunum@ci@modality_ablation.delta.qwen3_5_omni_plus.context\endcsname{[0.151, 0.199]}
\expandafter\gdef\csname odunum@val@modality_ablation.delta.qwen3_5_omni_plus.history\endcsname{0.089}
\expandafter\gdef\csname odunum@n@modality_ablation.delta.qwen3_5_omni_plus.history\endcsname{474}
\expandafter\gdef\csname odunum@ci@modality_ablation.delta.qwen3_5_omni_plus.history\endcsname{[0.040, 0.137]}
\expandafter\gdef\csname odunum@val@modality_ablation.delta.qwen3_5_omni_plus.intent\endcsname{0.048}
\expandafter\gdef\csname odunum@n@modality_ablation.delta.qwen3_5_omni_plus.intent\endcsname{1\,962}
\expandafter\gdef\csname odunum@ci@modality_ablation.delta.qwen3_5_omni_plus.intent\endcsname{[0.031, 0.066]}
\expandafter\gdef\csname odunum@val@modality_ablation.delta.qwen3_5_omni_plus.other_context\endcsname{0.075}
\expandafter\gdef\csname odunum@n@modality_ablation.delta.qwen3_5_omni_plus.other_context\endcsname{134}
\expandafter\gdef\csname odunum@ci@modality_ablation.delta.qwen3_5_omni_plus.other_context\endcsname{[-0.007, 0.157]}
\expandafter\gdef\csname odunum@val@modality_ablation.delta.qwen3_5_omni_plus.overall\endcsname{0.111}
\expandafter\gdef\csname odunum@n@modality_ablation.delta.qwen3_5_omni_plus.overall\endcsname{3\,928}
\expandafter\gdef\csname odunum@ci@modality_ablation.delta.qwen3_5_omni_plus.overall\endcsname{[0.095, 0.127]}
\expandafter\gdef\csname odunum@val@modality_ablation.delta.qwen3_5_omni_plus.unattributed\endcsname{0.000}
\expandafter\gdef\csname odunum@n@modality_ablation.delta.qwen3_5_omni_plus.unattributed\endcsname{13}
\expandafter\gdef\csname odunum@ci@modality_ablation.delta.qwen3_5_omni_plus.unattributed\endcsname{[-0.250, 0.250]}
\expandafter\gdef\csname odunum@val@modality_ablation.delta.qwen3_5_omni_plus.visual\endcsname{0.332}
\expandafter\gdef\csname odunum@n@modality_ablation.delta.qwen3_5_omni_plus.visual\endcsname{796}
\expandafter\gdef\csname odunum@ci@modality_ablation.delta.qwen3_5_omni_plus.visual\endcsname{[0.293, 0.370]}
\expandafter\gdef\csname odunum@val@modality_ablation.drift.doubao_seed_2_0_lite.overall\endcsname{0.001}
\expandafter\gdef\csname odunum@n@modality_ablation.drift.doubao_seed_2_0_lite.overall\endcsname{3\,959}
\expandafter\gdef\csname odunum@ci@modality_ablation.drift.doubao_seed_2_0_lite.overall\endcsname{[-0.006, 0.008]}
\expandafter\gdef\csname odunum@val@modality_ablation.drift.gemini_3_1_pro.overall\endcsname{-0.001}
\expandafter\gdef\csname odunum@n@modality_ablation.drift.gemini_3_1_pro.overall\endcsname{3\,943}
\expandafter\gdef\csname odunum@ci@modality_ablation.drift.gemini_3_1_pro.overall\endcsname{[-0.008, 0.007]}
\expandafter\gdef\csname odunum@val@modality_ablation.drift.qwen3_5_omni_plus.overall\endcsname{-0.002}
\expandafter\gdef\csname odunum@n@modality_ablation.drift.qwen3_5_omni_plus.overall\endcsname{3\,928}
\expandafter\gdef\csname odunum@ci@modality_ablation.drift.qwen3_5_omni_plus.overall\endcsname{[-0.009, 0.006]}
\expandafter\gdef\csname odunum@val@modality_ablation.judge.id\endcsname{mr\_ali / dashscope.qwen3.6-flash (thinking)}
\expandafter\gdef\csname odunum@n@modality_ablation.judge.id\endcsname{1}
\expandafter\gdef\csname odunum@val@modality_ablation.judge.scored_on\endcsname{2026-09-03}
\expandafter\gdef\csname odunum@n@modality_ablation.judge.scored_on\endcsname{1}
\expandafter\gdef\csname odunum@val@modality_ablation.novideo.doubao_seed_2_0_lite.audio\endcsname{0.470}
\expandafter\gdef\csname odunum@n@modality_ablation.novideo.doubao_seed_2_0_lite.audio\endcsname{557}
\expandafter\gdef\csname odunum@ci@modality_ablation.novideo.doubao_seed_2_0_lite.audio\endcsname{[0.429, 0.512]}
\expandafter\gdef\csname odunum@val@modality_ablation.novideo.doubao_seed_2_0_lite.context\endcsname{0.376}
\expandafter\gdef\csname odunum@n@modality_ablation.novideo.doubao_seed_2_0_lite.context\endcsname{1\,969}
\expandafter\gdef\csname odunum@ci@modality_ablation.novideo.doubao_seed_2_0_lite.context\endcsname{[0.354, 0.398]}
\expandafter\gdef\csname odunum@val@modality_ablation.novideo.doubao_seed_2_0_lite.history\endcsname{0.622}
\expandafter\gdef\csname odunum@n@modality_ablation.novideo.doubao_seed_2_0_lite.history\endcsname{474}
\expandafter\gdef\csname odunum@ci@modality_ablation.novideo.doubao_seed_2_0_lite.history\endcsname{[0.578, 0.667]}
\expandafter\gdef\csname odunum@val@modality_ablation.novideo.doubao_seed_2_0_lite.intent\endcsname{0.858}
\expandafter\gdef\csname odunum@n@modality_ablation.novideo.doubao_seed_2_0_lite.intent\endcsname{1\,977}
\expandafter\gdef\csname odunum@ci@modality_ablation.novideo.doubao_seed_2_0_lite.intent\endcsname{[0.842, 0.874]}
\expandafter\gdef\csname odunum@val@modality_ablation.novideo.doubao_seed_2_0_lite.other_context\endcsname{0.199}
\expandafter\gdef\csname odunum@n@modality_ablation.novideo.doubao_seed_2_0_lite.other_context\endcsname{136}
\expandafter\gdef\csname odunum@ci@modality_ablation.novideo.doubao_seed_2_0_lite.other_context\endcsname{[0.134, 0.267]}
\expandafter\gdef\csname odunum@val@modality_ablation.novideo.doubao_seed_2_0_lite.overall\endcsname{0.617}
\expandafter\gdef\csname odunum@n@modality_ablation.novideo.doubao_seed_2_0_lite.overall\endcsname{3\,959}
\expandafter\gdef\csname odunum@ci@modality_ablation.novideo.doubao_seed_2_0_lite.overall\endcsname{[0.602, 0.631]}
\expandafter\gdef\csname odunum@val@modality_ablation.novideo.doubao_seed_2_0_lite.unattributed\endcsname{0.231}
\expandafter\gdef\csname odunum@n@modality_ablation.novideo.doubao_seed_2_0_lite.unattributed\endcsname{13}
\expandafter\gdef\csname odunum@ci@modality_ablation.novideo.doubao_seed_2_0_lite.unattributed\endcsname{[0.045, 0.500]}
\expandafter\gdef\csname odunum@val@modality_ablation.novideo.doubao_seed_2_0_lite.visual\endcsname{0.196}
\expandafter\gdef\csname odunum@n@modality_ablation.novideo.doubao_seed_2_0_lite.visual\endcsname{802}
\expandafter\gdef\csname odunum@ci@modality_ablation.novideo.doubao_seed_2_0_lite.visual\endcsname{[0.167, 0.226]}
\expandafter\gdef\csname odunum@val@modality_ablation.novideo.gemini_3_1_pro.audio\endcsname{0.429}
\expandafter\gdef\csname odunum@n@modality_ablation.novideo.gemini_3_1_pro.audio\endcsname{553}
\expandafter\gdef\csname odunum@ci@modality_ablation.novideo.gemini_3_1_pro.audio\endcsname{[0.387, 0.470]}
\expandafter\gdef\csname odunum@val@modality_ablation.novideo.gemini_3_1_pro.context\endcsname{0.301}
\expandafter\gdef\csname odunum@n@modality_ablation.novideo.gemini_3_1_pro.context\endcsname{1\,960}
\expandafter\gdef\csname odunum@ci@modality_ablation.novideo.gemini_3_1_pro.context\endcsname{[0.280, 0.323]}
\expandafter\gdef\csname odunum@val@modality_ablation.novideo.gemini_3_1_pro.history\endcsname{0.389}
\expandafter\gdef\csname odunum@n@modality_ablation.novideo.gemini_3_1_pro.history\endcsname{475}
\expandafter\gdef\csname odunum@ci@modality_ablation.novideo.gemini_3_1_pro.history\endcsname{[0.344, 0.435]}
\expandafter\gdef\csname odunum@val@modality_ablation.novideo.gemini_3_1_pro.intent\endcsname{0.824}
\expandafter\gdef\csname odunum@n@modality_ablation.novideo.gemini_3_1_pro.intent\endcsname{1\,970}
\expandafter\gdef\csname odunum@ci@modality_ablation.novideo.gemini_3_1_pro.intent\endcsname{[0.806, 0.843]}
\expandafter\gdef\csname odunum@val@modality_ablation.novideo.gemini_3_1_pro.other_context\endcsname{0.206}
\expandafter\gdef\csname odunum@n@modality_ablation.novideo.gemini_3_1_pro.other_context\endcsname{136}
\expandafter\gdef\csname odunum@ci@modality_ablation.novideo.gemini_3_1_pro.other_context\endcsname{[0.141, 0.275]}
\expandafter\gdef\csname odunum@val@modality_ablation.novideo.gemini_3_1_pro.overall\endcsname{0.563}
\expandafter\gdef\csname odunum@n@modality_ablation.novideo.gemini_3_1_pro.overall\endcsname{3\,943}
\expandafter\gdef\csname odunum@ci@modality_ablation.novideo.gemini_3_1_pro.overall\endcsname{[0.547, 0.578]}
\expandafter\gdef\csname odunum@val@modality_ablation.novideo.gemini_3_1_pro.unattributed\endcsname{0.385}
\expandafter\gdef\csname odunum@n@modality_ablation.novideo.gemini_3_1_pro.unattributed\endcsname{13}
\expandafter\gdef\csname odunum@ci@modality_ablation.novideo.gemini_3_1_pro.unattributed\endcsname{[0.050, 0.789]}
\expandafter\gdef\csname odunum@val@modality_ablation.novideo.gemini_3_1_pro.visual\endcsname{0.176}
\expandafter\gdef\csname odunum@n@modality_ablation.novideo.gemini_3_1_pro.visual\endcsname{796}
\expandafter\gdef\csname odunum@ci@modality_ablation.novideo.gemini_3_1_pro.visual\endcsname{[0.150, 0.204]}
\expandafter\gdef\csname odunum@val@modality_ablation.novideo.qwen3_5_omni_plus.audio\endcsname{0.404}
\expandafter\gdef\csname odunum@n@modality_ablation.novideo.qwen3_5_omni_plus.audio\endcsname{549}
\expandafter\gdef\csname odunum@ci@modality_ablation.novideo.qwen3_5_omni_plus.audio\endcsname{[0.363, 0.447]}
\expandafter\gdef\csname odunum@val@modality_ablation.novideo.qwen3_5_omni_plus.context\endcsname{0.308}
\expandafter\gdef\csname odunum@n@modality_ablation.novideo.qwen3_5_omni_plus.context\endcsname{1\,953}
\expandafter\gdef\csname odunum@ci@modality_ablation.novideo.qwen3_5_omni_plus.context\endcsname{[0.286, 0.329]}
\expandafter\gdef\csname odunum@val@modality_ablation.novideo.qwen3_5_omni_plus.history\endcsname{0.481}
\expandafter\gdef\csname odunum@n@modality_ablation.novideo.qwen3_5_omni_plus.history\endcsname{474}
\expandafter\gdef\csname odunum@ci@modality_ablation.novideo.qwen3_5_omni_plus.history\endcsname{[0.435, 0.527]}
\expandafter\gdef\csname odunum@val@modality_ablation.novideo.qwen3_5_omni_plus.intent\endcsname{0.791}
\expandafter\gdef\csname odunum@n@modality_ablation.novideo.qwen3_5_omni_plus.intent\endcsname{1\,962}
\expandafter\gdef\csname odunum@ci@modality_ablation.novideo.qwen3_5_omni_plus.intent\endcsname{[0.771, 0.810]}
\expandafter\gdef\csname odunum@val@modality_ablation.novideo.qwen3_5_omni_plus.other_context\endcsname{0.351}
\expandafter\gdef\csname odunum@n@modality_ablation.novideo.qwen3_5_omni_plus.other_context\endcsname{134}
\expandafter\gdef\csname odunum@ci@modality_ablation.novideo.qwen3_5_omni_plus.other_context\endcsname{[0.273, 0.432]}
\expandafter\gdef\csname odunum@val@modality_ablation.novideo.qwen3_5_omni_plus.overall\endcsname{0.549}
\expandafter\gdef\csname odunum@n@modality_ablation.novideo.qwen3_5_omni_plus.overall\endcsname{3\,928}
\expandafter\gdef\csname odunum@ci@modality_ablation.novideo.qwen3_5_omni_plus.overall\endcsname{[0.533, 0.565]}
\expandafter\gdef\csname odunum@val@modality_ablation.novideo.qwen3_5_omni_plus.unattributed\endcsname{0.308}
\expandafter\gdef\csname odunum@n@modality_ablation.novideo.qwen3_5_omni_plus.unattributed\endcsname{13}
\expandafter\gdef\csname odunum@ci@modality_ablation.novideo.qwen3_5_omni_plus.unattributed\endcsname{[0.000, 0.625]}
\expandafter\gdef\csname odunum@val@modality_ablation.novideo.qwen3_5_omni_plus.visual\endcsname{0.131}
\expandafter\gdef\csname odunum@n@modality_ablation.novideo.qwen3_5_omni_plus.visual\endcsname{796}
\expandafter\gdef\csname odunum@ci@modality_ablation.novideo.qwen3_5_omni_plus.visual\endcsname{[0.106, 0.157]}
\expandafter\gdef\csname odunum@val@modality_ablation.pop.n_points_common\endcsname{3\,901}
\expandafter\gdef\csname odunum@n@modality_ablation.pop.n_points_common\endcsname{3\,973}
\expandafter\gdef\csname odunum@val@modality_ablation.pop.n_scenes\endcsname{1\,049}
\expandafter\gdef\csname odunum@n@modality_ablation.pop.n_scenes\endcsname{1\,061}
\expandafter\gdef\csname odunum@val@modality_ablation.withvideo.doubao_seed_2_0_lite.audio\endcsname{0.481}
\expandafter\gdef\csname odunum@n@modality_ablation.withvideo.doubao_seed_2_0_lite.audio\endcsname{557}
\expandafter\gdef\csname odunum@ci@modality_ablation.withvideo.doubao_seed_2_0_lite.audio\endcsname{[0.439, 0.522]}
\expandafter\gdef\csname odunum@val@modality_ablation.withvideo.doubao_seed_2_0_lite.context\endcsname{0.534}
\expandafter\gdef\csname odunum@n@modality_ablation.withvideo.doubao_seed_2_0_lite.context\endcsname{1\,969}
\expandafter\gdef\csname odunum@ci@modality_ablation.withvideo.doubao_seed_2_0_lite.context\endcsname{[0.510, 0.557]}
\expandafter\gdef\csname odunum@val@modality_ablation.withvideo.doubao_seed_2_0_lite.history\endcsname{0.620}
\expandafter\gdef\csname odunum@n@modality_ablation.withvideo.doubao_seed_2_0_lite.history\endcsname{474}
\expandafter\gdef\csname odunum@ci@modality_ablation.withvideo.doubao_seed_2_0_lite.history\endcsname{[0.574, 0.665]}
\expandafter\gdef\csname odunum@val@modality_ablation.withvideo.doubao_seed_2_0_lite.intent\endcsname{0.756}
\expandafter\gdef\csname odunum@n@modality_ablation.withvideo.doubao_seed_2_0_lite.intent\endcsname{1\,977}
\expandafter\gdef\csname odunum@ci@modality_ablation.withvideo.doubao_seed_2_0_lite.intent\endcsname{[0.733, 0.779]}
\expandafter\gdef\csname odunum@val@modality_ablation.withvideo.doubao_seed_2_0_lite.other_context\endcsname{0.353}
\expandafter\gdef\csname odunum@n@modality_ablation.withvideo.doubao_seed_2_0_lite.other_context\endcsname{136}
\expandafter\gdef\csname odunum@ci@modality_ablation.withvideo.doubao_seed_2_0_lite.other_context\endcsname{[0.272, 0.437]}
\expandafter\gdef\csname odunum@val@modality_ablation.withvideo.doubao_seed_2_0_lite.overall\endcsname{0.645}
\expandafter\gdef\csname odunum@n@modality_ablation.withvideo.doubao_seed_2_0_lite.overall\endcsname{3\,959}
\expandafter\gdef\csname odunum@ci@modality_ablation.withvideo.doubao_seed_2_0_lite.overall\endcsname{[0.626, 0.663]}
\expandafter\gdef\csname odunum@val@modality_ablation.withvideo.doubao_seed_2_0_lite.unattributed\endcsname{0.462}
\expandafter\gdef\csname odunum@n@modality_ablation.withvideo.doubao_seed_2_0_lite.unattributed\endcsname{13}
\expandafter\gdef\csname odunum@ci@modality_ablation.withvideo.doubao_seed_2_0_lite.unattributed\endcsname{[0.100, 0.900]}
\expandafter\gdef\csname odunum@val@modality_ablation.withvideo.doubao_seed_2_0_lite.visual\endcsname{0.550}
\expandafter\gdef\csname odunum@n@modality_ablation.withvideo.doubao_seed_2_0_lite.visual\endcsname{802}
\expandafter\gdef\csname odunum@ci@modality_ablation.withvideo.doubao_seed_2_0_lite.visual\endcsname{[0.515, 0.586]}
\expandafter\gdef\csname odunum@val@modality_ablation.withvideo.gemini_3_1_pro.audio\endcsname{0.472}
\expandafter\gdef\csname odunum@n@modality_ablation.withvideo.gemini_3_1_pro.audio\endcsname{553}
\expandafter\gdef\csname odunum@ci@modality_ablation.withvideo.gemini_3_1_pro.audio\endcsname{[0.430, 0.514]}
\expandafter\gdef\csname odunum@val@modality_ablation.withvideo.gemini_3_1_pro.context\endcsname{0.444}
\expandafter\gdef\csname odunum@n@modality_ablation.withvideo.gemini_3_1_pro.context\endcsname{1\,960}
\expandafter\gdef\csname odunum@ci@modality_ablation.withvideo.gemini_3_1_pro.context\endcsname{[0.422, 0.467]}
\expandafter\gdef\csname odunum@val@modality_ablation.withvideo.gemini_3_1_pro.history\endcsname{0.427}
\expandafter\gdef\csname odunum@n@modality_ablation.withvideo.gemini_3_1_pro.history\endcsname{475}
\expandafter\gdef\csname odunum@ci@modality_ablation.withvideo.gemini_3_1_pro.history\endcsname{[0.381, 0.474]}
\expandafter\gdef\csname odunum@val@modality_ablation.withvideo.gemini_3_1_pro.intent\endcsname{0.830}
\expandafter\gdef\csname odunum@n@modality_ablation.withvideo.gemini_3_1_pro.intent\endcsname{1\,970}
\expandafter\gdef\csname odunum@ci@modality_ablation.withvideo.gemini_3_1_pro.intent\endcsname{[0.813, 0.848]}
\expandafter\gdef\csname odunum@val@modality_ablation.withvideo.gemini_3_1_pro.other_context\endcsname{0.199}
\expandafter\gdef\csname odunum@n@modality_ablation.withvideo.gemini_3_1_pro.other_context\endcsname{136}
\expandafter\gdef\csname odunum@ci@modality_ablation.withvideo.gemini_3_1_pro.other_context\endcsname{[0.134, 0.267]}
\expandafter\gdef\csname odunum@val@modality_ablation.withvideo.gemini_3_1_pro.overall\endcsname{0.637}
\expandafter\gdef\csname odunum@n@modality_ablation.withvideo.gemini_3_1_pro.overall\endcsname{3\,943}
\expandafter\gdef\csname odunum@ci@modality_ablation.withvideo.gemini_3_1_pro.overall\endcsname{[0.622, 0.651]}
\expandafter\gdef\csname odunum@val@modality_ablation.withvideo.gemini_3_1_pro.unattributed\endcsname{0.308}
\expandafter\gdef\csname odunum@n@modality_ablation.withvideo.gemini_3_1_pro.unattributed\endcsname{13}
\expandafter\gdef\csname odunum@ci@modality_ablation.withvideo.gemini_3_1_pro.unattributed\endcsname{[0.000, 0.714]}
\expandafter\gdef\csname odunum@val@modality_ablation.withvideo.gemini_3_1_pro.visual\endcsname{0.476}
\expandafter\gdef\csname odunum@n@modality_ablation.withvideo.gemini_3_1_pro.visual\endcsname{796}
\expandafter\gdef\csname odunum@ci@modality_ablation.withvideo.gemini_3_1_pro.visual\endcsname{[0.442, 0.512]}
\expandafter\gdef\csname odunum@val@modality_ablation.withvideo.qwen3_5_omni_plus.audio\endcsname{0.452}
\expandafter\gdef\csname odunum@n@modality_ablation.withvideo.qwen3_5_omni_plus.audio\endcsname{549}
\expandafter\gdef\csname odunum@ci@modality_ablation.withvideo.qwen3_5_omni_plus.audio\endcsname{[0.410, 0.494]}
\expandafter\gdef\csname odunum@val@modality_ablation.withvideo.qwen3_5_omni_plus.context\endcsname{0.483}
\expandafter\gdef\csname odunum@n@modality_ablation.withvideo.qwen3_5_omni_plus.context\endcsname{1\,953}
\expandafter\gdef\csname odunum@ci@modality_ablation.withvideo.qwen3_5_omni_plus.context\endcsname{[0.459, 0.506]}
\expandafter\gdef\csname odunum@val@modality_ablation.withvideo.qwen3_5_omni_plus.history\endcsname{0.570}
\expandafter\gdef\csname odunum@n@modality_ablation.withvideo.qwen3_5_omni_plus.history\endcsname{474}
\expandafter\gdef\csname odunum@ci@modality_ablation.withvideo.qwen3_5_omni_plus.history\endcsname{[0.523, 0.614]}
\expandafter\gdef\csname odunum@val@modality_ablation.withvideo.qwen3_5_omni_plus.intent\endcsname{0.839}
\expandafter\gdef\csname odunum@n@modality_ablation.withvideo.qwen3_5_omni_plus.intent\endcsname{1\,962}
\expandafter\gdef\csname odunum@ci@modality_ablation.withvideo.qwen3_5_omni_plus.intent\endcsname{[0.822, 0.856]}
\expandafter\gdef\csname odunum@val@modality_ablation.withvideo.qwen3_5_omni_plus.other_context\endcsname{0.425}
\expandafter\gdef\csname odunum@n@modality_ablation.withvideo.qwen3_5_omni_plus.other_context\endcsname{134}
\expandafter\gdef\csname odunum@ci@modality_ablation.withvideo.qwen3_5_omni_plus.other_context\endcsname{[0.341, 0.508]}
\expandafter\gdef\csname odunum@val@modality_ablation.withvideo.qwen3_5_omni_plus.overall\endcsname{0.660}
\expandafter\gdef\csname odunum@n@modality_ablation.withvideo.qwen3_5_omni_plus.overall\endcsname{3\,928}
\expandafter\gdef\csname odunum@ci@modality_ablation.withvideo.qwen3_5_omni_plus.overall\endcsname{[0.645, 0.676]}
\expandafter\gdef\csname odunum@val@modality_ablation.withvideo.qwen3_5_omni_plus.unattributed\endcsname{0.308}
\expandafter\gdef\csname odunum@n@modality_ablation.withvideo.qwen3_5_omni_plus.unattributed\endcsname{13}
\expandafter\gdef\csname odunum@ci@modality_ablation.withvideo.qwen3_5_omni_plus.unattributed\endcsname{[0.050, 0.625]}
\expandafter\gdef\csname odunum@val@modality_ablation.withvideo.qwen3_5_omni_plus.visual\endcsname{0.462}
\expandafter\gdef\csname odunum@n@modality_ablation.withvideo.qwen3_5_omni_plus.visual\endcsname{796}
\expandafter\gdef\csname odunum@ci@modality_ablation.withvideo.qwen3_5_omni_plus.visual\endcsname{[0.426, 0.499]}
\expandafter\gdef\csname odunum@val@negative_patterns.coverage.cascade_asr.n\endcsname{372}
\expandafter\gdef\csname odunum@n@negative_patterns.coverage.cascade_asr.n\endcsname{372}
\expandafter\gdef\csname odunum@val@negative_patterns.coverage.gemini.n\endcsname{372}
\expandafter\gdef\csname odunum@n@negative_patterns.coverage.gemini.n\endcsname{372}
\expandafter\gdef\csname odunum@val@negative_patterns.coverage.gemini35_flash_lite.n\endcsname{372}
\expandafter\gdef\csname odunum@n@negative_patterns.coverage.gemini35_flash_lite.n\endcsname{372}
\expandafter\gdef\csname odunum@val@negative_patterns.coverage.gemini37_flash.n\endcsname{366}
\expandafter\gdef\csname odunum@n@negative_patterns.coverage.gemini37_flash.n\endcsname{372}
\expandafter\gdef\csname odunum@val@negative_patterns.coverage.ming.n\endcsname{372}
\expandafter\gdef\csname odunum@n@negative_patterns.coverage.ming.n\endcsname{372}
\expandafter\gdef\csname odunum@val@negative_patterns.coverage.minicpm_o.n\endcsname{371}
\expandafter\gdef\csname odunum@n@negative_patterns.coverage.minicpm_o.n\endcsname{372}
\expandafter\gdef\csname odunum@val@negative_patterns.coverage.nemotron.n\endcsname{372}
\expandafter\gdef\csname odunum@n@negative_patterns.coverage.nemotron.n\endcsname{372}
\expandafter\gdef\csname odunum@val@negative_patterns.coverage.qwen25_omni.n\endcsname{372}
\expandafter\gdef\csname odunum@n@negative_patterns.coverage.qwen25_omni.n\endcsname{372}
\expandafter\gdef\csname odunum@val@negative_patterns.coverage.qwen3_omni_instruct.n\endcsname{372}
\expandafter\gdef\csname odunum@n@negative_patterns.coverage.qwen3_omni_instruct.n\endcsname{372}
\expandafter\gdef\csname odunum@val@negative_patterns.coverage.qwen3_omni_think.n\endcsname{372}
\expandafter\gdef\csname odunum@n@negative_patterns.coverage.qwen3_omni_think.n\endcsname{372}
\expandafter\gdef\csname odunum@val@negative_patterns.coverage.qwen_plus.n\endcsname{372}
\expandafter\gdef\csname odunum@n@negative_patterns.coverage.qwen_plus.n\endcsname{372}
\expandafter\gdef\csname odunum@val@negative_patterns.coverage.salmonn2_7b.n\endcsname{370}
\expandafter\gdef\csname odunum@n@negative_patterns.coverage.salmonn2_7b.n\endcsname{372}
\expandafter\gdef\csname odunum@val@negative_patterns.coverage.seed.n\endcsname{372}
\expandafter\gdef\csname odunum@n@negative_patterns.coverage.seed.n\endcsname{372}
\expandafter\gdef\csname odunum@val@negative_patterns.delta_own_mean.cascade_asr.invalidation_reason.demand_incomplete\endcsname{0.096}
\expandafter\gdef\csname odunum@n@negative_patterns.delta_own_mean.cascade_asr.invalidation_reason.demand_incomplete\endcsname{78}
\expandafter\gdef\csname odunum@val@negative_patterns.delta_own_mean.cascade_asr.invalidation_reason.function_not_directed\endcsname{-0.126}
\expandafter\gdef\csname odunum@n@negative_patterns.delta_own_mean.cascade_asr.invalidation_reason.function_not_directed\endcsname{85}
\expandafter\gdef\csname odunum@val@negative_patterns.delta_own_mean.cascade_asr.invalidation_reason.intent_not_real\endcsname{-0.073}
\expandafter\gdef\csname odunum@n@negative_patterns.delta_own_mean.cascade_asr.invalidation_reason.intent_not_real\endcsname{84}
\expandafter\gdef\csname odunum@val@negative_patterns.delta_own_mean.cascade_asr.invalidation_reason.source_mismatch\endcsname{0.082}
\expandafter\gdef\csname odunum@n@negative_patterns.delta_own_mean.cascade_asr.invalidation_reason.source_mismatch\endcsname{81}
\expandafter\gdef\csname odunum@val@negative_patterns.delta_own_mean.cascade_asr.invalidation_reason.target_mismatch\endcsname{0.062}
\expandafter\gdef\csname odunum@n@negative_patterns.delta_own_mean.cascade_asr.invalidation_reason.target_mismatch\endcsname{44}
\expandafter\gdef\csname odunum@val@negative_patterns.delta_own_mean.cascade_asr.signal_type.contextual_signal\endcsname{-0.050}
\expandafter\gdef\csname odunum@n@negative_patterns.delta_own_mean.cascade_asr.signal_type.contextual_signal\endcsname{86}
\expandafter\gdef\csname odunum@val@negative_patterns.delta_own_mean.cascade_asr.signal_type.interactive_signal\endcsname{-0.056}
\expandafter\gdef\csname odunum@n@negative_patterns.delta_own_mean.cascade_asr.signal_type.interactive_signal\endcsname{37}
\expandafter\gdef\csname odunum@val@negative_patterns.delta_own_mean.cascade_asr.signal_type.linguistic_signal\endcsname{0.229}
\expandafter\gdef\csname odunum@n@negative_patterns.delta_own_mean.cascade_asr.signal_type.linguistic_signal\endcsname{69}
\expandafter\gdef\csname odunum@val@negative_patterns.delta_own_mean.cascade_asr.signal_type.prosodic_signal\endcsname{-0.030}
\expandafter\gdef\csname odunum@n@negative_patterns.delta_own_mean.cascade_asr.signal_type.prosodic_signal\endcsname{67}
\expandafter\gdef\csname odunum@val@negative_patterns.delta_own_mean.cascade_asr.signal_type.semantic_signal\endcsname{-0.066}
\expandafter\gdef\csname odunum@n@negative_patterns.delta_own_mean.cascade_asr.signal_type.semantic_signal\endcsname{113}
\expandafter\gdef\csname odunum@val@negative_patterns.delta_own_mean.gemini.invalidation_reason.demand_incomplete\endcsname{0.018}
\expandafter\gdef\csname odunum@n@negative_patterns.delta_own_mean.gemini.invalidation_reason.demand_incomplete\endcsname{78}
\expandafter\gdef\csname odunum@val@negative_patterns.delta_own_mean.gemini.invalidation_reason.function_not_directed\endcsname{-0.120}
\expandafter\gdef\csname odunum@n@negative_patterns.delta_own_mean.gemini.invalidation_reason.function_not_directed\endcsname{85}
\expandafter\gdef\csname odunum@val@negative_patterns.delta_own_mean.gemini.invalidation_reason.intent_not_real\endcsname{-0.034}
\expandafter\gdef\csname odunum@n@negative_patterns.delta_own_mean.gemini.invalidation_reason.intent_not_real\endcsname{84}
\expandafter\gdef\csname odunum@val@negative_patterns.delta_own_mean.gemini.invalidation_reason.source_mismatch\endcsname{0.115}
\expandafter\gdef\csname odunum@n@negative_patterns.delta_own_mean.gemini.invalidation_reason.source_mismatch\endcsname{81}
\expandafter\gdef\csname odunum@val@negative_patterns.delta_own_mean.gemini.invalidation_reason.target_mismatch\endcsname{0.053}
\expandafter\gdef\csname odunum@n@negative_patterns.delta_own_mean.gemini.invalidation_reason.target_mismatch\endcsname{44}
\expandafter\gdef\csname odunum@val@negative_patterns.delta_own_mean.gemini.signal_type.contextual_signal\endcsname{-0.019}
\expandafter\gdef\csname odunum@n@negative_patterns.delta_own_mean.gemini.signal_type.contextual_signal\endcsname{86}
\expandafter\gdef\csname odunum@val@negative_patterns.delta_own_mean.gemini.signal_type.interactive_signal\endcsname{0.080}
\expandafter\gdef\csname odunum@n@negative_patterns.delta_own_mean.gemini.signal_type.interactive_signal\endcsname{37}
\expandafter\gdef\csname odunum@val@negative_patterns.delta_own_mean.gemini.signal_type.linguistic_signal\endcsname{0.070}
\expandafter\gdef\csname odunum@n@negative_patterns.delta_own_mean.gemini.signal_type.linguistic_signal\endcsname{69}
\expandafter\gdef\csname odunum@val@negative_patterns.delta_own_mean.gemini.signal_type.prosodic_signal\endcsname{-0.006}
\expandafter\gdef\csname odunum@n@negative_patterns.delta_own_mean.gemini.signal_type.prosodic_signal\endcsname{67}
\expandafter\gdef\csname odunum@val@negative_patterns.delta_own_mean.gemini.signal_type.semantic_signal\endcsname{-0.052}
\expandafter\gdef\csname odunum@n@negative_patterns.delta_own_mean.gemini.signal_type.semantic_signal\endcsname{113}
\expandafter\gdef\csname odunum@val@negative_patterns.delta_own_mean.gemini35_flash_lite.invalidation_reason.demand_incomplete\endcsname{-0.124}
\expandafter\gdef\csname odunum@n@negative_patterns.delta_own_mean.gemini35_flash_lite.invalidation_reason.demand_incomplete\endcsname{78}
\expandafter\gdef\csname odunum@val@negative_patterns.delta_own_mean.gemini35_flash_lite.invalidation_reason.function_not_directed\endcsname{-0.023}
\expandafter\gdef\csname odunum@n@negative_patterns.delta_own_mean.gemini35_flash_lite.invalidation_reason.function_not_directed\endcsname{85}
\expandafter\gdef\csname odunum@val@negative_patterns.delta_own_mean.gemini35_flash_lite.invalidation_reason.intent_not_real\endcsname{-0.054}
\expandafter\gdef\csname odunum@n@negative_patterns.delta_own_mean.gemini35_flash_lite.invalidation_reason.intent_not_real\endcsname{84}
\expandafter\gdef\csname odunum@val@negative_patterns.delta_own_mean.gemini35_flash_lite.invalidation_reason.source_mismatch\endcsname{0.073}
\expandafter\gdef\csname odunum@n@negative_patterns.delta_own_mean.gemini35_flash_lite.invalidation_reason.source_mismatch\endcsname{81}
\expandafter\gdef\csname odunum@val@negative_patterns.delta_own_mean.gemini35_flash_lite.invalidation_reason.target_mismatch\endcsname{0.234}
\expandafter\gdef\csname odunum@n@negative_patterns.delta_own_mean.gemini35_flash_lite.invalidation_reason.target_mismatch\endcsname{44}
\expandafter\gdef\csname odunum@val@negative_patterns.delta_own_mean.gemini35_flash_lite.signal_type.contextual_signal\endcsname{-0.052}
\expandafter\gdef\csname odunum@n@negative_patterns.delta_own_mean.gemini35_flash_lite.signal_type.contextual_signal\endcsname{86}
\expandafter\gdef\csname odunum@val@negative_patterns.delta_own_mean.gemini35_flash_lite.signal_type.interactive_signal\endcsname{0.043}
\expandafter\gdef\csname odunum@n@negative_patterns.delta_own_mean.gemini35_flash_lite.signal_type.interactive_signal\endcsname{37}
\expandafter\gdef\csname odunum@val@negative_patterns.delta_own_mean.gemini35_flash_lite.signal_type.linguistic_signal\endcsname{0.240}
\expandafter\gdef\csname odunum@n@negative_patterns.delta_own_mean.gemini35_flash_lite.signal_type.linguistic_signal\endcsname{69}
\expandafter\gdef\csname odunum@val@negative_patterns.delta_own_mean.gemini35_flash_lite.signal_type.prosodic_signal\endcsname{-0.112}
\expandafter\gdef\csname odunum@n@negative_patterns.delta_own_mean.gemini35_flash_lite.signal_type.prosodic_signal\endcsname{67}
\expandafter\gdef\csname odunum@val@negative_patterns.delta_own_mean.gemini35_flash_lite.signal_type.semantic_signal\endcsname{-0.055}
\expandafter\gdef\csname odunum@n@negative_patterns.delta_own_mean.gemini35_flash_lite.signal_type.semantic_signal\endcsname{113}
\expandafter\gdef\csname odunum@val@negative_patterns.delta_own_mean.gemini37_flash.invalidation_reason.demand_incomplete\endcsname{-0.084}
\expandafter\gdef\csname odunum@n@negative_patterns.delta_own_mean.gemini37_flash.invalidation_reason.demand_incomplete\endcsname{74}
\expandafter\gdef\csname odunum@val@negative_patterns.delta_own_mean.gemini37_flash.invalidation_reason.function_not_directed\endcsname{-0.069}
\expandafter\gdef\csname odunum@n@negative_patterns.delta_own_mean.gemini37_flash.invalidation_reason.function_not_directed\endcsname{85}
\expandafter\gdef\csname odunum@val@negative_patterns.delta_own_mean.gemini37_flash.invalidation_reason.intent_not_real\endcsname{-0.100}
\expandafter\gdef\csname odunum@n@negative_patterns.delta_own_mean.gemini37_flash.invalidation_reason.intent_not_real\endcsname{82}
\expandafter\gdef\csname odunum@val@negative_patterns.delta_own_mean.gemini37_flash.invalidation_reason.source_mismatch\endcsname{0.260}
\expandafter\gdef\csname odunum@n@negative_patterns.delta_own_mean.gemini37_flash.invalidation_reason.source_mismatch\endcsname{81}
\expandafter\gdef\csname odunum@val@negative_patterns.delta_own_mean.gemini37_flash.invalidation_reason.target_mismatch\endcsname{-0.019}
\expandafter\gdef\csname odunum@n@negative_patterns.delta_own_mean.gemini37_flash.invalidation_reason.target_mismatch\endcsname{44}
\expandafter\gdef\csname odunum@val@negative_patterns.delta_own_mean.gemini37_flash.signal_type.contextual_signal\endcsname{0.045}
\expandafter\gdef\csname odunum@n@negative_patterns.delta_own_mean.gemini37_flash.signal_type.contextual_signal\endcsname{86}
\expandafter\gdef\csname odunum@val@negative_patterns.delta_own_mean.gemini37_flash.signal_type.interactive_signal\endcsname{0.032}
\expandafter\gdef\csname odunum@n@negative_patterns.delta_own_mean.gemini37_flash.signal_type.interactive_signal\endcsname{36}
\expandafter\gdef\csname odunum@val@negative_patterns.delta_own_mean.gemini37_flash.signal_type.linguistic_signal\endcsname{0.073}
\expandafter\gdef\csname odunum@n@negative_patterns.delta_own_mean.gemini37_flash.signal_type.linguistic_signal\endcsname{69}
\expandafter\gdef\csname odunum@val@negative_patterns.delta_own_mean.gemini37_flash.signal_type.prosodic_signal\endcsname{-0.107}
\expandafter\gdef\csname odunum@n@negative_patterns.delta_own_mean.gemini37_flash.signal_type.prosodic_signal\endcsname{65}
\expandafter\gdef\csname odunum@val@negative_patterns.delta_own_mean.gemini37_flash.signal_type.semantic_signal\endcsname{-0.028}
\expandafter\gdef\csname odunum@n@negative_patterns.delta_own_mean.gemini37_flash.signal_type.semantic_signal\endcsname{110}
\expandafter\gdef\csname odunum@val@negative_patterns.delta_own_mean.ming.invalidation_reason.demand_incomplete\endcsname{-0.157}
\expandafter\gdef\csname odunum@n@negative_patterns.delta_own_mean.ming.invalidation_reason.demand_incomplete\endcsname{78}
\expandafter\gdef\csname odunum@val@negative_patterns.delta_own_mean.ming.invalidation_reason.function_not_directed\endcsname{0.009}
\expandafter\gdef\csname odunum@n@negative_patterns.delta_own_mean.ming.invalidation_reason.function_not_directed\endcsname{85}
\expandafter\gdef\csname odunum@val@negative_patterns.delta_own_mean.ming.invalidation_reason.intent_not_real\endcsname{0.031}
\expandafter\gdef\csname odunum@n@negative_patterns.delta_own_mean.ming.invalidation_reason.intent_not_real\endcsname{84}
\expandafter\gdef\csname odunum@val@negative_patterns.delta_own_mean.ming.invalidation_reason.source_mismatch\endcsname{0.064}
\expandafter\gdef\csname odunum@n@negative_patterns.delta_own_mean.ming.invalidation_reason.source_mismatch\endcsname{81}
\expandafter\gdef\csname odunum@val@negative_patterns.delta_own_mean.ming.invalidation_reason.target_mismatch\endcsname{0.082}
\expandafter\gdef\csname odunum@n@negative_patterns.delta_own_mean.ming.invalidation_reason.target_mismatch\endcsname{44}
\expandafter\gdef\csname odunum@val@negative_patterns.delta_own_mean.ming.signal_type.contextual_signal\endcsname{0.011}
\expandafter\gdef\csname odunum@n@negative_patterns.delta_own_mean.ming.signal_type.contextual_signal\endcsname{86}
\expandafter\gdef\csname odunum@val@negative_patterns.delta_own_mean.ming.signal_type.interactive_signal\endcsname{0.015}
\expandafter\gdef\csname odunum@n@negative_patterns.delta_own_mean.ming.signal_type.interactive_signal\endcsname{37}
\expandafter\gdef\csname odunum@val@negative_patterns.delta_own_mean.ming.signal_type.linguistic_signal\endcsname{0.093}
\expandafter\gdef\csname odunum@n@negative_patterns.delta_own_mean.ming.signal_type.linguistic_signal\endcsname{69}
\expandafter\gdef\csname odunum@val@negative_patterns.delta_own_mean.ming.signal_type.prosodic_signal\endcsname{0.001}
\expandafter\gdef\csname odunum@n@negative_patterns.delta_own_mean.ming.signal_type.prosodic_signal\endcsname{67}
\expandafter\gdef\csname odunum@val@negative_patterns.delta_own_mean.ming.signal_type.semantic_signal\endcsname{-0.071}
\expandafter\gdef\csname odunum@n@negative_patterns.delta_own_mean.ming.signal_type.semantic_signal\endcsname{113}
\expandafter\gdef\csname odunum@val@negative_patterns.delta_own_mean.minicpm_o.invalidation_reason.demand_incomplete\endcsname{0.033}
\expandafter\gdef\csname odunum@n@negative_patterns.delta_own_mean.minicpm_o.invalidation_reason.demand_incomplete\endcsname{78}
\expandafter\gdef\csname odunum@val@negative_patterns.delta_own_mean.minicpm_o.invalidation_reason.function_not_directed\endcsname{-0.061}
\expandafter\gdef\csname odunum@n@negative_patterns.delta_own_mean.minicpm_o.invalidation_reason.function_not_directed\endcsname{84}
\expandafter\gdef\csname odunum@val@negative_patterns.delta_own_mean.minicpm_o.invalidation_reason.intent_not_real\endcsname{-0.025}
\expandafter\gdef\csname odunum@n@negative_patterns.delta_own_mean.minicpm_o.invalidation_reason.intent_not_real\endcsname{84}
\expandafter\gdef\csname odunum@val@negative_patterns.delta_own_mean.minicpm_o.invalidation_reason.source_mismatch\endcsname{-0.022}
\expandafter\gdef\csname odunum@n@negative_patterns.delta_own_mean.minicpm_o.invalidation_reason.source_mismatch\endcsname{81}
\expandafter\gdef\csname odunum@val@negative_patterns.delta_own_mean.minicpm_o.invalidation_reason.target_mismatch\endcsname{0.145}
\expandafter\gdef\csname odunum@n@negative_patterns.delta_own_mean.minicpm_o.invalidation_reason.target_mismatch\endcsname{44}
\expandafter\gdef\csname odunum@val@negative_patterns.delta_own_mean.minicpm_o.signal_type.contextual_signal\endcsname{0.060}
\expandafter\gdef\csname odunum@n@negative_patterns.delta_own_mean.minicpm_o.signal_type.contextual_signal\endcsname{85}
\expandafter\gdef\csname odunum@val@negative_patterns.delta_own_mean.minicpm_o.signal_type.interactive_signal\endcsname{-0.030}
\expandafter\gdef\csname odunum@n@negative_patterns.delta_own_mean.minicpm_o.signal_type.interactive_signal\endcsname{37}
\expandafter\gdef\csname odunum@val@negative_patterns.delta_own_mean.minicpm_o.signal_type.linguistic_signal\endcsname{0.097}
\expandafter\gdef\csname odunum@n@negative_patterns.delta_own_mean.minicpm_o.signal_type.linguistic_signal\endcsname{69}
\expandafter\gdef\csname odunum@val@negative_patterns.delta_own_mean.minicpm_o.signal_type.prosodic_signal\endcsname{-0.026}
\expandafter\gdef\csname odunum@n@negative_patterns.delta_own_mean.minicpm_o.signal_type.prosodic_signal\endcsname{67}
\expandafter\gdef\csname odunum@val@negative_patterns.delta_own_mean.minicpm_o.signal_type.semantic_signal\endcsname{-0.079}
\expandafter\gdef\csname odunum@n@negative_patterns.delta_own_mean.minicpm_o.signal_type.semantic_signal\endcsname{113}
\expandafter\gdef\csname odunum@val@negative_patterns.delta_own_mean.nemotron.invalidation_reason.demand_incomplete\endcsname{-0.027}
\expandafter\gdef\csname odunum@n@negative_patterns.delta_own_mean.nemotron.invalidation_reason.demand_incomplete\endcsname{78}
\expandafter\gdef\csname odunum@val@negative_patterns.delta_own_mean.nemotron.invalidation_reason.function_not_directed\endcsname{-0.003}
\expandafter\gdef\csname odunum@n@negative_patterns.delta_own_mean.nemotron.invalidation_reason.function_not_directed\endcsname{85}
\expandafter\gdef\csname odunum@val@negative_patterns.delta_own_mean.nemotron.invalidation_reason.intent_not_real\endcsname{-0.007}
\expandafter\gdef\csname odunum@n@negative_patterns.delta_own_mean.nemotron.invalidation_reason.intent_not_real\endcsname{84}
\expandafter\gdef\csname odunum@val@negative_patterns.delta_own_mean.nemotron.invalidation_reason.source_mismatch\endcsname{0.033}
\expandafter\gdef\csname odunum@n@negative_patterns.delta_own_mean.nemotron.invalidation_reason.source_mismatch\endcsname{81}
\expandafter\gdef\csname odunum@val@negative_patterns.delta_own_mean.nemotron.invalidation_reason.target_mismatch\endcsname{0.005}
\expandafter\gdef\csname odunum@n@negative_patterns.delta_own_mean.nemotron.invalidation_reason.target_mismatch\endcsname{44}
\expandafter\gdef\csname odunum@val@negative_patterns.delta_own_mean.nemotron.signal_type.contextual_signal\endcsname{0.046}
\expandafter\gdef\csname odunum@n@negative_patterns.delta_own_mean.nemotron.signal_type.contextual_signal\endcsname{86}
\expandafter\gdef\csname odunum@val@negative_patterns.delta_own_mean.nemotron.signal_type.interactive_signal\endcsname{-0.042}
\expandafter\gdef\csname odunum@n@negative_patterns.delta_own_mean.nemotron.signal_type.interactive_signal\endcsname{37}
\expandafter\gdef\csname odunum@val@negative_patterns.delta_own_mean.nemotron.signal_type.linguistic_signal\endcsname{0.023}
\expandafter\gdef\csname odunum@n@negative_patterns.delta_own_mean.nemotron.signal_type.linguistic_signal\endcsname{69}
\expandafter\gdef\csname odunum@val@negative_patterns.delta_own_mean.nemotron.signal_type.prosodic_signal\endcsname{0.002}
\expandafter\gdef\csname odunum@n@negative_patterns.delta_own_mean.nemotron.signal_type.prosodic_signal\endcsname{67}
\expandafter\gdef\csname odunum@val@negative_patterns.delta_own_mean.nemotron.signal_type.semantic_signal\endcsname{-0.037}
\expandafter\gdef\csname odunum@n@negative_patterns.delta_own_mean.nemotron.signal_type.semantic_signal\endcsname{113}
\expandafter\gdef\csname odunum@val@negative_patterns.delta_own_mean.qwen25_omni.invalidation_reason.demand_incomplete\endcsname{0.053}
\expandafter\gdef\csname odunum@n@negative_patterns.delta_own_mean.qwen25_omni.invalidation_reason.demand_incomplete\endcsname{78}
\expandafter\gdef\csname odunum@val@negative_patterns.delta_own_mean.qwen25_omni.invalidation_reason.function_not_directed\endcsname{-0.052}
\expandafter\gdef\csname odunum@n@negative_patterns.delta_own_mean.qwen25_omni.invalidation_reason.function_not_directed\endcsname{85}
\expandafter\gdef\csname odunum@val@negative_patterns.delta_own_mean.qwen25_omni.invalidation_reason.intent_not_real\endcsname{-0.031}
\expandafter\gdef\csname odunum@n@negative_patterns.delta_own_mean.qwen25_omni.invalidation_reason.intent_not_real\endcsname{84}
\expandafter\gdef\csname odunum@val@negative_patterns.delta_own_mean.qwen25_omni.invalidation_reason.source_mismatch\endcsname{-0.003}
\expandafter\gdef\csname odunum@n@negative_patterns.delta_own_mean.qwen25_omni.invalidation_reason.source_mismatch\endcsname{81}
\expandafter\gdef\csname odunum@val@negative_patterns.delta_own_mean.qwen25_omni.invalidation_reason.target_mismatch\endcsname{0.071}
\expandafter\gdef\csname odunum@n@negative_patterns.delta_own_mean.qwen25_omni.invalidation_reason.target_mismatch\endcsname{44}
\expandafter\gdef\csname odunum@val@negative_patterns.delta_own_mean.qwen25_omni.signal_type.contextual_signal\endcsname{-0.014}
\expandafter\gdef\csname odunum@n@negative_patterns.delta_own_mean.qwen25_omni.signal_type.contextual_signal\endcsname{86}
\expandafter\gdef\csname odunum@val@negative_patterns.delta_own_mean.qwen25_omni.signal_type.interactive_signal\endcsname{-0.117}
\expandafter\gdef\csname odunum@n@negative_patterns.delta_own_mean.qwen25_omni.signal_type.interactive_signal\endcsname{37}
\expandafter\gdef\csname odunum@val@negative_patterns.delta_own_mean.qwen25_omni.signal_type.linguistic_signal\endcsname{0.135}
\expandafter\gdef\csname odunum@n@negative_patterns.delta_own_mean.qwen25_omni.signal_type.linguistic_signal\endcsname{69}
\expandafter\gdef\csname odunum@val@negative_patterns.delta_own_mean.qwen25_omni.signal_type.prosodic_signal\endcsname{-0.032}
\expandafter\gdef\csname odunum@n@negative_patterns.delta_own_mean.qwen25_omni.signal_type.prosodic_signal\endcsname{67}
\expandafter\gdef\csname odunum@val@negative_patterns.delta_own_mean.qwen25_omni.signal_type.semantic_signal\endcsname{-0.014}
\expandafter\gdef\csname odunum@n@negative_patterns.delta_own_mean.qwen25_omni.signal_type.semantic_signal\endcsname{113}
\expandafter\gdef\csname odunum@val@negative_patterns.delta_own_mean.qwen3_omni_instruct.invalidation_reason.demand_incomplete\endcsname{0.093}
\expandafter\gdef\csname odunum@n@negative_patterns.delta_own_mean.qwen3_omni_instruct.invalidation_reason.demand_incomplete\endcsname{78}
\expandafter\gdef\csname odunum@val@negative_patterns.delta_own_mean.qwen3_omni_instruct.invalidation_reason.function_not_directed\endcsname{-0.057}
\expandafter\gdef\csname odunum@n@negative_patterns.delta_own_mean.qwen3_omni_instruct.invalidation_reason.function_not_directed\endcsname{85}
\expandafter\gdef\csname odunum@val@negative_patterns.delta_own_mean.qwen3_omni_instruct.invalidation_reason.intent_not_real\endcsname{-0.047}
\expandafter\gdef\csname odunum@n@negative_patterns.delta_own_mean.qwen3_omni_instruct.invalidation_reason.intent_not_real\endcsname{84}
\expandafter\gdef\csname odunum@val@negative_patterns.delta_own_mean.qwen3_omni_instruct.invalidation_reason.source_mismatch\endcsname{-0.029}
\expandafter\gdef\csname odunum@n@negative_patterns.delta_own_mean.qwen3_omni_instruct.invalidation_reason.source_mismatch\endcsname{81}
\expandafter\gdef\csname odunum@val@negative_patterns.delta_own_mean.qwen3_omni_instruct.invalidation_reason.target_mismatch\endcsname{0.086}
\expandafter\gdef\csname odunum@n@negative_patterns.delta_own_mean.qwen3_omni_instruct.invalidation_reason.target_mismatch\endcsname{44}
\expandafter\gdef\csname odunum@val@negative_patterns.delta_own_mean.qwen3_omni_instruct.signal_type.contextual_signal\endcsname{0.015}
\expandafter\gdef\csname odunum@n@negative_patterns.delta_own_mean.qwen3_omni_instruct.signal_type.contextual_signal\endcsname{86}
\expandafter\gdef\csname odunum@val@negative_patterns.delta_own_mean.qwen3_omni_instruct.signal_type.interactive_signal\endcsname{0.024}
\expandafter\gdef\csname odunum@n@negative_patterns.delta_own_mean.qwen3_omni_instruct.signal_type.interactive_signal\endcsname{37}
\expandafter\gdef\csname odunum@val@negative_patterns.delta_own_mean.qwen3_omni_instruct.signal_type.linguistic_signal\endcsname{0.088}
\expandafter\gdef\csname odunum@n@negative_patterns.delta_own_mean.qwen3_omni_instruct.signal_type.linguistic_signal\endcsname{69}
\expandafter\gdef\csname odunum@val@negative_patterns.delta_own_mean.qwen3_omni_instruct.signal_type.prosodic_signal\endcsname{-0.077}
\expandafter\gdef\csname odunum@n@negative_patterns.delta_own_mean.qwen3_omni_instruct.signal_type.prosodic_signal\endcsname{67}
\expandafter\gdef\csname odunum@val@negative_patterns.delta_own_mean.qwen3_omni_instruct.signal_type.semantic_signal\endcsname{-0.028}
\expandafter\gdef\csname odunum@n@negative_patterns.delta_own_mean.qwen3_omni_instruct.signal_type.semantic_signal\endcsname{113}
\expandafter\gdef\csname odunum@val@negative_patterns.delta_own_mean.qwen3_omni_think.invalidation_reason.demand_incomplete\endcsname{-0.125}
\expandafter\gdef\csname odunum@n@negative_patterns.delta_own_mean.qwen3_omni_think.invalidation_reason.demand_incomplete\endcsname{78}
\expandafter\gdef\csname odunum@val@negative_patterns.delta_own_mean.qwen3_omni_think.invalidation_reason.function_not_directed\endcsname{-0.009}
\expandafter\gdef\csname odunum@n@negative_patterns.delta_own_mean.qwen3_omni_think.invalidation_reason.function_not_directed\endcsname{85}
\expandafter\gdef\csname odunum@val@negative_patterns.delta_own_mean.qwen3_omni_think.invalidation_reason.intent_not_real\endcsname{-0.048}
\expandafter\gdef\csname odunum@n@negative_patterns.delta_own_mean.qwen3_omni_think.invalidation_reason.intent_not_real\endcsname{84}
\expandafter\gdef\csname odunum@val@negative_patterns.delta_own_mean.qwen3_omni_think.invalidation_reason.source_mismatch\endcsname{0.112}
\expandafter\gdef\csname odunum@n@negative_patterns.delta_own_mean.qwen3_omni_think.invalidation_reason.source_mismatch\endcsname{81}
\expandafter\gdef\csname odunum@val@negative_patterns.delta_own_mean.qwen3_omni_think.invalidation_reason.target_mismatch\endcsname{0.126}
\expandafter\gdef\csname odunum@n@negative_patterns.delta_own_mean.qwen3_omni_think.invalidation_reason.target_mismatch\endcsname{44}
\expandafter\gdef\csname odunum@val@negative_patterns.delta_own_mean.qwen3_omni_think.signal_type.contextual_signal\endcsname{0.006}
\expandafter\gdef\csname odunum@n@negative_patterns.delta_own_mean.qwen3_omni_think.signal_type.contextual_signal\endcsname{86}
\expandafter\gdef\csname odunum@val@negative_patterns.delta_own_mean.qwen3_omni_think.signal_type.interactive_signal\endcsname{-0.066}
\expandafter\gdef\csname odunum@n@negative_patterns.delta_own_mean.qwen3_omni_think.signal_type.interactive_signal\endcsname{37}
\expandafter\gdef\csname odunum@val@negative_patterns.delta_own_mean.qwen3_omni_think.signal_type.linguistic_signal\endcsname{0.183}
\expandafter\gdef\csname odunum@n@negative_patterns.delta_own_mean.qwen3_omni_think.signal_type.linguistic_signal\endcsname{69}
\expandafter\gdef\csname odunum@val@negative_patterns.delta_own_mean.qwen3_omni_think.signal_type.prosodic_signal\endcsname{-0.028}
\expandafter\gdef\csname odunum@n@negative_patterns.delta_own_mean.qwen3_omni_think.signal_type.prosodic_signal\endcsname{67}
\expandafter\gdef\csname odunum@val@negative_patterns.delta_own_mean.qwen3_omni_think.signal_type.semantic_signal\endcsname{-0.078}
\expandafter\gdef\csname odunum@n@negative_patterns.delta_own_mean.qwen3_omni_think.signal_type.semantic_signal\endcsname{113}
\expandafter\gdef\csname odunum@val@negative_patterns.delta_own_mean.qwen_plus.invalidation_reason.demand_incomplete\endcsname{0.009}
\expandafter\gdef\csname odunum@n@negative_patterns.delta_own_mean.qwen_plus.invalidation_reason.demand_incomplete\endcsname{78}
\expandafter\gdef\csname odunum@val@negative_patterns.delta_own_mean.qwen_plus.invalidation_reason.function_not_directed\endcsname{-0.043}
\expandafter\gdef\csname odunum@n@negative_patterns.delta_own_mean.qwen_plus.invalidation_reason.function_not_directed\endcsname{85}
\expandafter\gdef\csname odunum@val@negative_patterns.delta_own_mean.qwen_plus.invalidation_reason.intent_not_real\endcsname{-0.096}
\expandafter\gdef\csname odunum@n@negative_patterns.delta_own_mean.qwen_plus.invalidation_reason.intent_not_real\endcsname{84}
\expandafter\gdef\csname odunum@val@negative_patterns.delta_own_mean.qwen_plus.invalidation_reason.source_mismatch\endcsname{0.023}
\expandafter\gdef\csname odunum@n@negative_patterns.delta_own_mean.qwen_plus.invalidation_reason.source_mismatch\endcsname{81}
\expandafter\gdef\csname odunum@val@negative_patterns.delta_own_mean.qwen_plus.invalidation_reason.target_mismatch\endcsname{0.209}
\expandafter\gdef\csname odunum@n@negative_patterns.delta_own_mean.qwen_plus.invalidation_reason.target_mismatch\endcsname{44}
\expandafter\gdef\csname odunum@val@negative_patterns.delta_own_mean.qwen_plus.signal_type.contextual_signal\endcsname{0.043}
\expandafter\gdef\csname odunum@n@negative_patterns.delta_own_mean.qwen_plus.signal_type.contextual_signal\endcsname{86}
\expandafter\gdef\csname odunum@val@negative_patterns.delta_own_mean.qwen_plus.signal_type.interactive_signal\endcsname{0.044}
\expandafter\gdef\csname odunum@n@negative_patterns.delta_own_mean.qwen_plus.signal_type.interactive_signal\endcsname{37}
\expandafter\gdef\csname odunum@val@negative_patterns.delta_own_mean.qwen_plus.signal_type.linguistic_signal\endcsname{0.107}
\expandafter\gdef\csname odunum@n@negative_patterns.delta_own_mean.qwen_plus.signal_type.linguistic_signal\endcsname{69}
\expandafter\gdef\csname odunum@val@negative_patterns.delta_own_mean.qwen_plus.signal_type.prosodic_signal\endcsname{-0.094}
\expandafter\gdef\csname odunum@n@negative_patterns.delta_own_mean.qwen_plus.signal_type.prosodic_signal\endcsname{67}
\expandafter\gdef\csname odunum@val@negative_patterns.delta_own_mean.qwen_plus.signal_type.semantic_signal\endcsname{-0.056}
\expandafter\gdef\csname odunum@n@negative_patterns.delta_own_mean.qwen_plus.signal_type.semantic_signal\endcsname{113}
\expandafter\gdef\csname odunum@val@negative_patterns.delta_own_mean.salmonn2_7b.invalidation_reason.demand_incomplete\endcsname{0.045}
\expandafter\gdef\csname odunum@n@negative_patterns.delta_own_mean.salmonn2_7b.invalidation_reason.demand_incomplete\endcsname{78}
\expandafter\gdef\csname odunum@val@negative_patterns.delta_own_mean.salmonn2_7b.invalidation_reason.function_not_directed\endcsname{-0.022}
\expandafter\gdef\csname odunum@n@negative_patterns.delta_own_mean.salmonn2_7b.invalidation_reason.function_not_directed\endcsname{85}
\expandafter\gdef\csname odunum@val@negative_patterns.delta_own_mean.salmonn2_7b.invalidation_reason.intent_not_real\endcsname{-0.002}
\expandafter\gdef\csname odunum@n@negative_patterns.delta_own_mean.salmonn2_7b.invalidation_reason.intent_not_real\endcsname{82}
\expandafter\gdef\csname odunum@val@negative_patterns.delta_own_mean.salmonn2_7b.invalidation_reason.source_mismatch\endcsname{-0.003}
\expandafter\gdef\csname odunum@n@negative_patterns.delta_own_mean.salmonn2_7b.invalidation_reason.source_mismatch\endcsname{81}
\expandafter\gdef\csname odunum@val@negative_patterns.delta_own_mean.salmonn2_7b.invalidation_reason.target_mismatch\endcsname{-0.030}
\expandafter\gdef\csname odunum@n@negative_patterns.delta_own_mean.salmonn2_7b.invalidation_reason.target_mismatch\endcsname{44}
\expandafter\gdef\csname odunum@val@negative_patterns.delta_own_mean.salmonn2_7b.signal_type.contextual_signal\endcsname{0.014}
\expandafter\gdef\csname odunum@n@negative_patterns.delta_own_mean.salmonn2_7b.signal_type.contextual_signal\endcsname{86}
\expandafter\gdef\csname odunum@val@negative_patterns.delta_own_mean.salmonn2_7b.signal_type.interactive_signal\endcsname{-0.078}
\expandafter\gdef\csname odunum@n@negative_patterns.delta_own_mean.salmonn2_7b.signal_type.interactive_signal\endcsname{37}
\expandafter\gdef\csname odunum@val@negative_patterns.delta_own_mean.salmonn2_7b.signal_type.linguistic_signal\endcsname{0.055}
\expandafter\gdef\csname odunum@n@negative_patterns.delta_own_mean.salmonn2_7b.signal_type.linguistic_signal\endcsname{69}
\expandafter\gdef\csname odunum@val@negative_patterns.delta_own_mean.salmonn2_7b.signal_type.prosodic_signal\endcsname{-0.022}
\expandafter\gdef\csname odunum@n@negative_patterns.delta_own_mean.salmonn2_7b.signal_type.prosodic_signal\endcsname{66}
\expandafter\gdef\csname odunum@val@negative_patterns.delta_own_mean.salmonn2_7b.signal_type.semantic_signal\endcsname{-0.006}
\expandafter\gdef\csname odunum@n@negative_patterns.delta_own_mean.salmonn2_7b.signal_type.semantic_signal\endcsname{112}
\expandafter\gdef\csname odunum@val@negative_patterns.delta_own_mean.seed.invalidation_reason.demand_incomplete\endcsname{-0.043}
\expandafter\gdef\csname odunum@n@negative_patterns.delta_own_mean.seed.invalidation_reason.demand_incomplete\endcsname{78}
\expandafter\gdef\csname odunum@val@negative_patterns.delta_own_mean.seed.invalidation_reason.function_not_directed\endcsname{-0.024}
\expandafter\gdef\csname odunum@n@negative_patterns.delta_own_mean.seed.invalidation_reason.function_not_directed\endcsname{85}
\expandafter\gdef\csname odunum@val@negative_patterns.delta_own_mean.seed.invalidation_reason.intent_not_real\endcsname{-0.002}
\expandafter\gdef\csname odunum@n@negative_patterns.delta_own_mean.seed.invalidation_reason.intent_not_real\endcsname{84}
\expandafter\gdef\csname odunum@val@negative_patterns.delta_own_mean.seed.invalidation_reason.source_mismatch\endcsname{0.052}
\expandafter\gdef\csname odunum@n@negative_patterns.delta_own_mean.seed.invalidation_reason.source_mismatch\endcsname{81}
\expandafter\gdef\csname odunum@val@negative_patterns.delta_own_mean.seed.invalidation_reason.target_mismatch\endcsname{0.029}
\expandafter\gdef\csname odunum@n@negative_patterns.delta_own_mean.seed.invalidation_reason.target_mismatch\endcsname{44}
\expandafter\gdef\csname odunum@val@negative_patterns.delta_own_mean.seed.signal_type.contextual_signal\endcsname{-0.068}
\expandafter\gdef\csname odunum@n@negative_patterns.delta_own_mean.seed.signal_type.contextual_signal\endcsname{86}
\expandafter\gdef\csname odunum@val@negative_patterns.delta_own_mean.seed.signal_type.interactive_signal\endcsname{-0.028}
\expandafter\gdef\csname odunum@n@negative_patterns.delta_own_mean.seed.signal_type.interactive_signal\endcsname{37}
\expandafter\gdef\csname odunum@val@negative_patterns.delta_own_mean.seed.signal_type.linguistic_signal\endcsname{0.101}
\expandafter\gdef\csname odunum@n@negative_patterns.delta_own_mean.seed.signal_type.linguistic_signal\endcsname{69}
\expandafter\gdef\csname odunum@val@negative_patterns.delta_own_mean.seed.signal_type.prosodic_signal\endcsname{0.009}
\expandafter\gdef\csname odunum@n@negative_patterns.delta_own_mean.seed.signal_type.prosodic_signal\endcsname{67}
\expandafter\gdef\csname odunum@val@negative_patterns.delta_own_mean.seed.signal_type.semantic_signal\endcsname{-0.007}
\expandafter\gdef\csname odunum@n@negative_patterns.delta_own_mean.seed.signal_type.semantic_signal\endcsname{113}
\expandafter\gdef\csname odunum@val@negative_patterns.ftr.cascade_asr.invalidation_reason.demand_incomplete\endcsname{0.692}
\expandafter\gdef\csname odunum@n@negative_patterns.ftr.cascade_asr.invalidation_reason.demand_incomplete\endcsname{78}
\expandafter\gdef\csname odunum@ci@negative_patterns.ftr.cascade_asr.invalidation_reason.demand_incomplete\endcsname{[0.583, 0.784]}
\expandafter\gdef\csname odunum@val@negative_patterns.ftr.cascade_asr.invalidation_reason.function_not_directed\endcsname{0.471}
\expandafter\gdef\csname odunum@n@negative_patterns.ftr.cascade_asr.invalidation_reason.function_not_directed\endcsname{85}
\expandafter\gdef\csname odunum@ci@negative_patterns.ftr.cascade_asr.invalidation_reason.function_not_directed\endcsname{[0.368, 0.576]}
\expandafter\gdef\csname odunum@val@negative_patterns.ftr.cascade_asr.invalidation_reason.intent_not_real\endcsname{0.524}
\expandafter\gdef\csname odunum@n@negative_patterns.ftr.cascade_asr.invalidation_reason.intent_not_real\endcsname{84}
\expandafter\gdef\csname odunum@ci@negative_patterns.ftr.cascade_asr.invalidation_reason.intent_not_real\endcsname{[0.418, 0.627]}
\expandafter\gdef\csname odunum@val@negative_patterns.ftr.cascade_asr.invalidation_reason.source_mismatch\endcsname{0.679}
\expandafter\gdef\csname odunum@n@negative_patterns.ftr.cascade_asr.invalidation_reason.source_mismatch\endcsname{81}
\expandafter\gdef\csname odunum@ci@negative_patterns.ftr.cascade_asr.invalidation_reason.source_mismatch\endcsname{[0.571, 0.771]}
\expandafter\gdef\csname odunum@val@negative_patterns.ftr.cascade_asr.invalidation_reason.target_mismatch\endcsname{0.659}
\expandafter\gdef\csname odunum@n@negative_patterns.ftr.cascade_asr.invalidation_reason.target_mismatch\endcsname{44}
\expandafter\gdef\csname odunum@ci@negative_patterns.ftr.cascade_asr.invalidation_reason.target_mismatch\endcsname{[0.511, 0.781]}
\expandafter\gdef\csname odunum@val@negative_patterns.ftr.cascade_asr.signal_type.contextual_signal\endcsname{0.547}
\expandafter\gdef\csname odunum@n@negative_patterns.ftr.cascade_asr.signal_type.contextual_signal\endcsname{86}
\expandafter\gdef\csname odunum@ci@negative_patterns.ftr.cascade_asr.signal_type.contextual_signal\endcsname{[0.442, 0.647]}
\expandafter\gdef\csname odunum@val@negative_patterns.ftr.cascade_asr.signal_type.interactive_signal\endcsname{0.541}
\expandafter\gdef\csname odunum@n@negative_patterns.ftr.cascade_asr.signal_type.interactive_signal\endcsname{37}
\expandafter\gdef\csname odunum@ci@negative_patterns.ftr.cascade_asr.signal_type.interactive_signal\endcsname{[0.384, 0.690]}
\expandafter\gdef\csname odunum@val@negative_patterns.ftr.cascade_asr.signal_type.linguistic_signal\endcsname{0.826}
\expandafter\gdef\csname odunum@n@negative_patterns.ftr.cascade_asr.signal_type.linguistic_signal\endcsname{69}
\expandafter\gdef\csname odunum@ci@negative_patterns.ftr.cascade_asr.signal_type.linguistic_signal\endcsname{[0.720, 0.898]}
\expandafter\gdef\csname odunum@val@negative_patterns.ftr.cascade_asr.signal_type.prosodic_signal\endcsname{0.567}
\expandafter\gdef\csname odunum@n@negative_patterns.ftr.cascade_asr.signal_type.prosodic_signal\endcsname{67}
\expandafter\gdef\csname odunum@ci@negative_patterns.ftr.cascade_asr.signal_type.prosodic_signal\endcsname{[0.448, 0.679]}
\expandafter\gdef\csname odunum@val@negative_patterns.ftr.cascade_asr.signal_type.semantic_signal\endcsname{0.531}
\expandafter\gdef\csname odunum@n@negative_patterns.ftr.cascade_asr.signal_type.semantic_signal\endcsname{113}
\expandafter\gdef\csname odunum@ci@negative_patterns.ftr.cascade_asr.signal_type.semantic_signal\endcsname{[0.439, 0.620]}
\expandafter\gdef\csname odunum@val@negative_patterns.ftr.gemini.invalidation_reason.demand_incomplete\endcsname{0.397}
\expandafter\gdef\csname odunum@n@negative_patterns.ftr.gemini.invalidation_reason.demand_incomplete\endcsname{78}
\expandafter\gdef\csname odunum@ci@negative_patterns.ftr.gemini.invalidation_reason.demand_incomplete\endcsname{[0.296, 0.508]}
\expandafter\gdef\csname odunum@val@negative_patterns.ftr.gemini.invalidation_reason.function_not_directed\endcsname{0.259}
\expandafter\gdef\csname odunum@n@negative_patterns.ftr.gemini.invalidation_reason.function_not_directed\endcsname{85}
\expandafter\gdef\csname odunum@ci@negative_patterns.ftr.gemini.invalidation_reason.function_not_directed\endcsname{[0.178, 0.361]}
\expandafter\gdef\csname odunum@val@negative_patterns.ftr.gemini.invalidation_reason.intent_not_real\endcsname{0.345}
\expandafter\gdef\csname odunum@n@negative_patterns.ftr.gemini.invalidation_reason.intent_not_real\endcsname{84}
\expandafter\gdef\csname odunum@ci@negative_patterns.ftr.gemini.invalidation_reason.intent_not_real\endcsname{[0.252, 0.452]}
\expandafter\gdef\csname odunum@val@negative_patterns.ftr.gemini.invalidation_reason.source_mismatch\endcsname{0.494}
\expandafter\gdef\csname odunum@n@negative_patterns.ftr.gemini.invalidation_reason.source_mismatch\endcsname{81}
\expandafter\gdef\csname odunum@ci@negative_patterns.ftr.gemini.invalidation_reason.source_mismatch\endcsname{[0.388, 0.600]}
\expandafter\gdef\csname odunum@val@negative_patterns.ftr.gemini.invalidation_reason.target_mismatch\endcsname{0.432}
\expandafter\gdef\csname odunum@n@negative_patterns.ftr.gemini.invalidation_reason.target_mismatch\endcsname{44}
\expandafter\gdef\csname odunum@ci@negative_patterns.ftr.gemini.invalidation_reason.target_mismatch\endcsname{[0.297, 0.578]}
\expandafter\gdef\csname odunum@val@negative_patterns.ftr.gemini.signal_type.contextual_signal\endcsname{0.360}
\expandafter\gdef\csname odunum@n@negative_patterns.ftr.gemini.signal_type.contextual_signal\endcsname{86}
\expandafter\gdef\csname odunum@ci@negative_patterns.ftr.gemini.signal_type.contextual_signal\endcsname{[0.267, 0.466]}
\expandafter\gdef\csname odunum@val@negative_patterns.ftr.gemini.signal_type.interactive_signal\endcsname{0.459}
\expandafter\gdef\csname odunum@n@negative_patterns.ftr.gemini.signal_type.interactive_signal\endcsname{37}
\expandafter\gdef\csname odunum@ci@negative_patterns.ftr.gemini.signal_type.interactive_signal\endcsname{[0.310, 0.616]}
\expandafter\gdef\csname odunum@val@negative_patterns.ftr.gemini.signal_type.linguistic_signal\endcsname{0.449}
\expandafter\gdef\csname odunum@n@negative_patterns.ftr.gemini.signal_type.linguistic_signal\endcsname{69}
\expandafter\gdef\csname odunum@ci@negative_patterns.ftr.gemini.signal_type.linguistic_signal\endcsname{[0.338, 0.566]}
\expandafter\gdef\csname odunum@val@negative_patterns.ftr.gemini.signal_type.prosodic_signal\endcsname{0.373}
\expandafter\gdef\csname odunum@n@negative_patterns.ftr.gemini.signal_type.prosodic_signal\endcsname{67}
\expandafter\gdef\csname odunum@ci@negative_patterns.ftr.gemini.signal_type.prosodic_signal\endcsname{[0.267, 0.493]}
\expandafter\gdef\csname odunum@val@negative_patterns.ftr.gemini.signal_type.semantic_signal\endcsname{0.327}
\expandafter\gdef\csname odunum@n@negative_patterns.ftr.gemini.signal_type.semantic_signal\endcsname{113}
\expandafter\gdef\csname odunum@ci@negative_patterns.ftr.gemini.signal_type.semantic_signal\endcsname{[0.248, 0.418]}
\expandafter\gdef\csname odunum@val@negative_patterns.ftr.gemini35_flash_lite.invalidation_reason.demand_incomplete\endcsname{0.346}
\expandafter\gdef\csname odunum@n@negative_patterns.ftr.gemini35_flash_lite.invalidation_reason.demand_incomplete\endcsname{78}
\expandafter\gdef\csname odunum@ci@negative_patterns.ftr.gemini35_flash_lite.invalidation_reason.demand_incomplete\endcsname{[0.250, 0.457]}
\expandafter\gdef\csname odunum@val@negative_patterns.ftr.gemini35_flash_lite.invalidation_reason.function_not_directed\endcsname{0.447}
\expandafter\gdef\csname odunum@n@negative_patterns.ftr.gemini35_flash_lite.invalidation_reason.function_not_directed\endcsname{85}
\expandafter\gdef\csname odunum@ci@negative_patterns.ftr.gemini35_flash_lite.invalidation_reason.function_not_directed\endcsname{[0.346, 0.553]}
\expandafter\gdef\csname odunum@val@negative_patterns.ftr.gemini35_flash_lite.invalidation_reason.intent_not_real\endcsname{0.417}
\expandafter\gdef\csname odunum@n@negative_patterns.ftr.gemini35_flash_lite.invalidation_reason.intent_not_real\endcsname{84}
\expandafter\gdef\csname odunum@ci@negative_patterns.ftr.gemini35_flash_lite.invalidation_reason.intent_not_real\endcsname{[0.317, 0.523]}
\expandafter\gdef\csname odunum@val@negative_patterns.ftr.gemini35_flash_lite.invalidation_reason.source_mismatch\endcsname{0.543}
\expandafter\gdef\csname odunum@n@negative_patterns.ftr.gemini35_flash_lite.invalidation_reason.source_mismatch\endcsname{81}
\expandafter\gdef\csname odunum@ci@negative_patterns.ftr.gemini35_flash_lite.invalidation_reason.source_mismatch\endcsname{[0.435, 0.647]}
\expandafter\gdef\csname odunum@val@negative_patterns.ftr.gemini35_flash_lite.invalidation_reason.target_mismatch\endcsname{0.705}
\expandafter\gdef\csname odunum@n@negative_patterns.ftr.gemini35_flash_lite.invalidation_reason.target_mismatch\endcsname{44}
\expandafter\gdef\csname odunum@ci@negative_patterns.ftr.gemini35_flash_lite.invalidation_reason.target_mismatch\endcsname{[0.558, 0.818]}
\expandafter\gdef\csname odunum@val@negative_patterns.ftr.gemini35_flash_lite.signal_type.contextual_signal\endcsname{0.419}
\expandafter\gdef\csname odunum@n@negative_patterns.ftr.gemini35_flash_lite.signal_type.contextual_signal\endcsname{86}
\expandafter\gdef\csname odunum@ci@negative_patterns.ftr.gemini35_flash_lite.signal_type.contextual_signal\endcsname{[0.320, 0.524]}
\expandafter\gdef\csname odunum@val@negative_patterns.ftr.gemini35_flash_lite.signal_type.interactive_signal\endcsname{0.514}
\expandafter\gdef\csname odunum@n@negative_patterns.ftr.gemini35_flash_lite.signal_type.interactive_signal\endcsname{37}
\expandafter\gdef\csname odunum@ci@negative_patterns.ftr.gemini35_flash_lite.signal_type.interactive_signal\endcsname{[0.359, 0.666]}
\expandafter\gdef\csname odunum@val@negative_patterns.ftr.gemini35_flash_lite.signal_type.linguistic_signal\endcsname{0.710}
\expandafter\gdef\csname odunum@n@negative_patterns.ftr.gemini35_flash_lite.signal_type.linguistic_signal\endcsname{69}
\expandafter\gdef\csname odunum@ci@negative_patterns.ftr.gemini35_flash_lite.signal_type.linguistic_signal\endcsname{[0.594, 0.804]}
\expandafter\gdef\csname odunum@val@negative_patterns.ftr.gemini35_flash_lite.signal_type.prosodic_signal\endcsname{0.358}
\expandafter\gdef\csname odunum@n@negative_patterns.ftr.gemini35_flash_lite.signal_type.prosodic_signal\endcsname{67}
\expandafter\gdef\csname odunum@ci@negative_patterns.ftr.gemini35_flash_lite.signal_type.prosodic_signal\endcsname{[0.254, 0.478]}
\expandafter\gdef\csname odunum@val@negative_patterns.ftr.gemini35_flash_lite.signal_type.semantic_signal\endcsname{0.416}
\expandafter\gdef\csname odunum@n@negative_patterns.ftr.gemini35_flash_lite.signal_type.semantic_signal\endcsname{113}
\expandafter\gdef\csname odunum@ci@negative_patterns.ftr.gemini35_flash_lite.signal_type.semantic_signal\endcsname{[0.329, 0.508]}
\expandafter\gdef\csname odunum@val@negative_patterns.ftr.gemini37_flash.invalidation_reason.demand_incomplete\endcsname{0.162}
\expandafter\gdef\csname odunum@n@negative_patterns.ftr.gemini37_flash.invalidation_reason.demand_incomplete\endcsname{74}
\expandafter\gdef\csname odunum@ci@negative_patterns.ftr.gemini37_flash.invalidation_reason.demand_incomplete\endcsname{[0.095, 0.262]}
\expandafter\gdef\csname odunum@val@negative_patterns.ftr.gemini37_flash.invalidation_reason.function_not_directed\endcsname{0.176}
\expandafter\gdef\csname odunum@n@negative_patterns.ftr.gemini37_flash.invalidation_reason.function_not_directed\endcsname{85}
\expandafter\gdef\csname odunum@ci@negative_patterns.ftr.gemini37_flash.invalidation_reason.function_not_directed\endcsname{[0.110, 0.271]}
\expandafter\gdef\csname odunum@val@negative_patterns.ftr.gemini37_flash.invalidation_reason.intent_not_real\endcsname{0.146}
\expandafter\gdef\csname odunum@n@negative_patterns.ftr.gemini37_flash.invalidation_reason.intent_not_real\endcsname{82}
\expandafter\gdef\csname odunum@ci@negative_patterns.ftr.gemini37_flash.invalidation_reason.intent_not_real\endcsname{[0.086, 0.239]}
\expandafter\gdef\csname odunum@val@negative_patterns.ftr.gemini37_flash.invalidation_reason.source_mismatch\endcsname{0.506}
\expandafter\gdef\csname odunum@n@negative_patterns.ftr.gemini37_flash.invalidation_reason.source_mismatch\endcsname{81}
\expandafter\gdef\csname odunum@ci@negative_patterns.ftr.gemini37_flash.invalidation_reason.source_mismatch\endcsname{[0.400, 0.612]}
\expandafter\gdef\csname odunum@val@negative_patterns.ftr.gemini37_flash.invalidation_reason.target_mismatch\endcsname{0.227}
\expandafter\gdef\csname odunum@n@negative_patterns.ftr.gemini37_flash.invalidation_reason.target_mismatch\endcsname{44}
\expandafter\gdef\csname odunum@ci@negative_patterns.ftr.gemini37_flash.invalidation_reason.target_mismatch\endcsname{[0.128, 0.370]}
\expandafter\gdef\csname odunum@val@negative_patterns.ftr.gemini37_flash.signal_type.contextual_signal\endcsname{0.291}
\expandafter\gdef\csname odunum@n@negative_patterns.ftr.gemini37_flash.signal_type.contextual_signal\endcsname{86}
\expandafter\gdef\csname odunum@ci@negative_patterns.ftr.gemini37_flash.signal_type.contextual_signal\endcsname{[0.205, 0.394]}
\expandafter\gdef\csname odunum@val@negative_patterns.ftr.gemini37_flash.signal_type.interactive_signal\endcsname{0.278}
\expandafter\gdef\csname odunum@n@negative_patterns.ftr.gemini37_flash.signal_type.interactive_signal\endcsname{36}
\expandafter\gdef\csname odunum@ci@negative_patterns.ftr.gemini37_flash.signal_type.interactive_signal\endcsname{[0.158, 0.440]}
\expandafter\gdef\csname odunum@val@negative_patterns.ftr.gemini37_flash.signal_type.linguistic_signal\endcsname{0.319}
\expandafter\gdef\csname odunum@n@negative_patterns.ftr.gemini37_flash.signal_type.linguistic_signal\endcsname{69}
\expandafter\gdef\csname odunum@ci@negative_patterns.ftr.gemini37_flash.signal_type.linguistic_signal\endcsname{[0.221, 0.436]}
\expandafter\gdef\csname odunum@val@negative_patterns.ftr.gemini37_flash.signal_type.prosodic_signal\endcsname{0.138}
\expandafter\gdef\csname odunum@n@negative_patterns.ftr.gemini37_flash.signal_type.prosodic_signal\endcsname{65}
\expandafter\gdef\csname odunum@ci@negative_patterns.ftr.gemini37_flash.signal_type.prosodic_signal\endcsname{[0.075, 0.243]}
\expandafter\gdef\csname odunum@val@negative_patterns.ftr.gemini37_flash.signal_type.semantic_signal\endcsname{0.218}
\expandafter\gdef\csname odunum@n@negative_patterns.ftr.gemini37_flash.signal_type.semantic_signal\endcsname{110}
\expandafter\gdef\csname odunum@ci@negative_patterns.ftr.gemini37_flash.signal_type.semantic_signal\endcsname{[0.151, 0.304]}
\expandafter\gdef\csname odunum@val@negative_patterns.ftr.ming.invalidation_reason.demand_incomplete\endcsname{0.692}
\expandafter\gdef\csname odunum@n@negative_patterns.ftr.ming.invalidation_reason.demand_incomplete\endcsname{78}
\expandafter\gdef\csname odunum@ci@negative_patterns.ftr.ming.invalidation_reason.demand_incomplete\endcsname{[0.583, 0.784]}
\expandafter\gdef\csname odunum@val@negative_patterns.ftr.ming.invalidation_reason.function_not_directed\endcsname{0.859}
\expandafter\gdef\csname odunum@n@negative_patterns.ftr.ming.invalidation_reason.function_not_directed\endcsname{85}
\expandafter\gdef\csname odunum@ci@negative_patterns.ftr.ming.invalidation_reason.function_not_directed\endcsname{[0.769, 0.917]}
\expandafter\gdef\csname odunum@val@negative_patterns.ftr.ming.invalidation_reason.intent_not_real\endcsname{0.881}
\expandafter\gdef\csname odunum@n@negative_patterns.ftr.ming.invalidation_reason.intent_not_real\endcsname{84}
\expandafter\gdef\csname odunum@ci@negative_patterns.ftr.ming.invalidation_reason.intent_not_real\endcsname{[0.795, 0.934]}
\expandafter\gdef\csname odunum@val@negative_patterns.ftr.ming.invalidation_reason.source_mismatch\endcsname{0.914}
\expandafter\gdef\csname odunum@n@negative_patterns.ftr.ming.invalidation_reason.source_mismatch\endcsname{81}
\expandafter\gdef\csname odunum@ci@negative_patterns.ftr.ming.invalidation_reason.source_mismatch\endcsname{[0.832, 0.958]}
\expandafter\gdef\csname odunum@val@negative_patterns.ftr.ming.invalidation_reason.target_mismatch\endcsname{0.932}
\expandafter\gdef\csname odunum@n@negative_patterns.ftr.ming.invalidation_reason.target_mismatch\endcsname{44}
\expandafter\gdef\csname odunum@ci@negative_patterns.ftr.ming.invalidation_reason.target_mismatch\endcsname{[0.818, 0.977]}
\expandafter\gdef\csname odunum@val@negative_patterns.ftr.ming.signal_type.contextual_signal\endcsname{0.860}
\expandafter\gdef\csname odunum@n@negative_patterns.ftr.ming.signal_type.contextual_signal\endcsname{86}
\expandafter\gdef\csname odunum@ci@negative_patterns.ftr.ming.signal_type.contextual_signal\endcsname{[0.772, 0.918]}
\expandafter\gdef\csname odunum@val@negative_patterns.ftr.ming.signal_type.interactive_signal\endcsname{0.865}
\expandafter\gdef\csname odunum@n@negative_patterns.ftr.ming.signal_type.interactive_signal\endcsname{37}
\expandafter\gdef\csname odunum@ci@negative_patterns.ftr.ming.signal_type.interactive_signal\endcsname{[0.720, 0.941]}
\expandafter\gdef\csname odunum@val@negative_patterns.ftr.ming.signal_type.linguistic_signal\endcsname{0.942}
\expandafter\gdef\csname odunum@n@negative_patterns.ftr.ming.signal_type.linguistic_signal\endcsname{69}
\expandafter\gdef\csname odunum@ci@negative_patterns.ftr.ming.signal_type.linguistic_signal\endcsname{[0.860, 0.977]}
\expandafter\gdef\csname odunum@val@negative_patterns.ftr.ming.signal_type.prosodic_signal\endcsname{0.851}
\expandafter\gdef\csname odunum@n@negative_patterns.ftr.ming.signal_type.prosodic_signal\endcsname{67}
\expandafter\gdef\csname odunum@ci@negative_patterns.ftr.ming.signal_type.prosodic_signal\endcsname{[0.747, 0.917]}
\expandafter\gdef\csname odunum@val@negative_patterns.ftr.ming.signal_type.semantic_signal\endcsname{0.779}
\expandafter\gdef\csname odunum@n@negative_patterns.ftr.ming.signal_type.semantic_signal\endcsname{113}
\expandafter\gdef\csname odunum@ci@negative_patterns.ftr.ming.signal_type.semantic_signal\endcsname{[0.694, 0.845]}
\expandafter\gdef\csname odunum@val@negative_patterns.ftr.minicpm_o.invalidation_reason.demand_incomplete\endcsname{0.821}
\expandafter\gdef\csname odunum@n@negative_patterns.ftr.minicpm_o.invalidation_reason.demand_incomplete\endcsname{78}
\expandafter\gdef\csname odunum@ci@negative_patterns.ftr.minicpm_o.invalidation_reason.demand_incomplete\endcsname{[0.721, 0.890]}
\expandafter\gdef\csname odunum@val@negative_patterns.ftr.minicpm_o.invalidation_reason.function_not_directed\endcsname{0.726}
\expandafter\gdef\csname odunum@n@negative_patterns.ftr.minicpm_o.invalidation_reason.function_not_directed\endcsname{84}
\expandafter\gdef\csname odunum@ci@negative_patterns.ftr.minicpm_o.invalidation_reason.function_not_directed\endcsname{[0.623, 0.810]}
\expandafter\gdef\csname odunum@val@negative_patterns.ftr.minicpm_o.invalidation_reason.intent_not_real\endcsname{0.762}
\expandafter\gdef\csname odunum@n@negative_patterns.ftr.minicpm_o.invalidation_reason.intent_not_real\endcsname{84}
\expandafter\gdef\csname odunum@ci@negative_patterns.ftr.minicpm_o.invalidation_reason.intent_not_real\endcsname{[0.661, 0.840]}
\expandafter\gdef\csname odunum@val@negative_patterns.ftr.minicpm_o.invalidation_reason.source_mismatch\endcsname{0.765}
\expandafter\gdef\csname odunum@n@negative_patterns.ftr.minicpm_o.invalidation_reason.source_mismatch\endcsname{81}
\expandafter\gdef\csname odunum@ci@negative_patterns.ftr.minicpm_o.invalidation_reason.source_mismatch\endcsname{[0.662, 0.844]}
\expandafter\gdef\csname odunum@val@negative_patterns.ftr.minicpm_o.invalidation_reason.target_mismatch\endcsname{0.932}
\expandafter\gdef\csname odunum@n@negative_patterns.ftr.minicpm_o.invalidation_reason.target_mismatch\endcsname{44}
\expandafter\gdef\csname odunum@ci@negative_patterns.ftr.minicpm_o.invalidation_reason.target_mismatch\endcsname{[0.818, 0.977]}
\expandafter\gdef\csname odunum@val@negative_patterns.ftr.minicpm_o.signal_type.contextual_signal\endcsname{0.847}
\expandafter\gdef\csname odunum@n@negative_patterns.ftr.minicpm_o.signal_type.contextual_signal\endcsname{85}
\expandafter\gdef\csname odunum@ci@negative_patterns.ftr.minicpm_o.signal_type.contextual_signal\endcsname{[0.756, 0.908]}
\expandafter\gdef\csname odunum@val@negative_patterns.ftr.minicpm_o.signal_type.interactive_signal\endcsname{0.757}
\expandafter\gdef\csname odunum@n@negative_patterns.ftr.minicpm_o.signal_type.interactive_signal\endcsname{37}
\expandafter\gdef\csname odunum@ci@negative_patterns.ftr.minicpm_o.signal_type.interactive_signal\endcsname{[0.599, 0.866]}
\expandafter\gdef\csname odunum@val@negative_patterns.ftr.minicpm_o.signal_type.linguistic_signal\endcsname{0.884}
\expandafter\gdef\csname odunum@n@negative_patterns.ftr.minicpm_o.signal_type.linguistic_signal\endcsname{69}
\expandafter\gdef\csname odunum@ci@negative_patterns.ftr.minicpm_o.signal_type.linguistic_signal\endcsname{[0.788, 0.940]}
\expandafter\gdef\csname odunum@val@negative_patterns.ftr.minicpm_o.signal_type.prosodic_signal\endcsname{0.761}
\expandafter\gdef\csname odunum@n@negative_patterns.ftr.minicpm_o.signal_type.prosodic_signal\endcsname{67}
\expandafter\gdef\csname odunum@ci@negative_patterns.ftr.minicpm_o.signal_type.prosodic_signal\endcsname{[0.647, 0.847]}
\expandafter\gdef\csname odunum@val@negative_patterns.ftr.minicpm_o.signal_type.semantic_signal\endcsname{0.708}
\expandafter\gdef\csname odunum@n@negative_patterns.ftr.minicpm_o.signal_type.semantic_signal\endcsname{113}
\expandafter\gdef\csname odunum@ci@negative_patterns.ftr.minicpm_o.signal_type.semantic_signal\endcsname{[0.618, 0.784]}
\expandafter\gdef\csname odunum@val@negative_patterns.ftr.nemotron.invalidation_reason.demand_incomplete\endcsname{0.718}
\expandafter\gdef\csname odunum@n@negative_patterns.ftr.nemotron.invalidation_reason.demand_incomplete\endcsname{78}
\expandafter\gdef\csname odunum@ci@negative_patterns.ftr.nemotron.invalidation_reason.demand_incomplete\endcsname{[0.610, 0.806]}
\expandafter\gdef\csname odunum@val@negative_patterns.ftr.nemotron.invalidation_reason.function_not_directed\endcsname{0.741}
\expandafter\gdef\csname odunum@n@negative_patterns.ftr.nemotron.invalidation_reason.function_not_directed\endcsname{85}
\expandafter\gdef\csname odunum@ci@negative_patterns.ftr.nemotron.invalidation_reason.function_not_directed\endcsname{[0.639, 0.822]}
\expandafter\gdef\csname odunum@val@negative_patterns.ftr.nemotron.invalidation_reason.intent_not_real\endcsname{0.738}
\expandafter\gdef\csname odunum@n@negative_patterns.ftr.nemotron.invalidation_reason.intent_not_real\endcsname{84}
\expandafter\gdef\csname odunum@ci@negative_patterns.ftr.nemotron.invalidation_reason.intent_not_real\endcsname{[0.635, 0.820]}
\expandafter\gdef\csname odunum@val@negative_patterns.ftr.nemotron.invalidation_reason.source_mismatch\endcsname{0.778}
\expandafter\gdef\csname odunum@n@negative_patterns.ftr.nemotron.invalidation_reason.source_mismatch\endcsname{81}
\expandafter\gdef\csname odunum@ci@negative_patterns.ftr.nemotron.invalidation_reason.source_mismatch\endcsname{[0.676, 0.855]}
\expandafter\gdef\csname odunum@val@negative_patterns.ftr.nemotron.invalidation_reason.target_mismatch\endcsname{0.750}
\expandafter\gdef\csname odunum@n@negative_patterns.ftr.nemotron.invalidation_reason.target_mismatch\endcsname{44}
\expandafter\gdef\csname odunum@ci@negative_patterns.ftr.nemotron.invalidation_reason.target_mismatch\endcsname{[0.606, 0.854]}
\expandafter\gdef\csname odunum@val@negative_patterns.ftr.nemotron.signal_type.contextual_signal\endcsname{0.791}
\expandafter\gdef\csname odunum@n@negative_patterns.ftr.nemotron.signal_type.contextual_signal\endcsname{86}
\expandafter\gdef\csname odunum@ci@negative_patterns.ftr.nemotron.signal_type.contextual_signal\endcsname{[0.693, 0.863]}
\expandafter\gdef\csname odunum@val@negative_patterns.ftr.nemotron.signal_type.interactive_signal\endcsname{0.703}
\expandafter\gdef\csname odunum@n@negative_patterns.ftr.nemotron.signal_type.interactive_signal\endcsname{37}
\expandafter\gdef\csname odunum@ci@negative_patterns.ftr.nemotron.signal_type.interactive_signal\endcsname{[0.542, 0.825]}
\expandafter\gdef\csname odunum@val@negative_patterns.ftr.nemotron.signal_type.linguistic_signal\endcsname{0.768}
\expandafter\gdef\csname odunum@n@negative_patterns.ftr.nemotron.signal_type.linguistic_signal\endcsname{69}
\expandafter\gdef\csname odunum@ci@negative_patterns.ftr.nemotron.signal_type.linguistic_signal\endcsname{[0.656, 0.852]}
\expandafter\gdef\csname odunum@val@negative_patterns.ftr.nemotron.signal_type.prosodic_signal\endcsname{0.746}
\expandafter\gdef\csname odunum@n@negative_patterns.ftr.nemotron.signal_type.prosodic_signal\endcsname{67}
\expandafter\gdef\csname odunum@ci@negative_patterns.ftr.nemotron.signal_type.prosodic_signal\endcsname{[0.631, 0.835]}
\expandafter\gdef\csname odunum@val@negative_patterns.ftr.nemotron.signal_type.semantic_signal\endcsname{0.708}
\expandafter\gdef\csname odunum@n@negative_patterns.ftr.nemotron.signal_type.semantic_signal\endcsname{113}
\expandafter\gdef\csname odunum@ci@negative_patterns.ftr.nemotron.signal_type.semantic_signal\endcsname{[0.618, 0.784]}
\expandafter\gdef\csname odunum@val@negative_patterns.ftr.qwen25_omni.invalidation_reason.demand_incomplete\endcsname{0.846}
\expandafter\gdef\csname odunum@n@negative_patterns.ftr.qwen25_omni.invalidation_reason.demand_incomplete\endcsname{78}
\expandafter\gdef\csname odunum@ci@negative_patterns.ftr.qwen25_omni.invalidation_reason.demand_incomplete\endcsname{[0.750, 0.910]}
\expandafter\gdef\csname odunum@val@negative_patterns.ftr.qwen25_omni.invalidation_reason.function_not_directed\endcsname{0.741}
\expandafter\gdef\csname odunum@n@negative_patterns.ftr.qwen25_omni.invalidation_reason.function_not_directed\endcsname{85}
\expandafter\gdef\csname odunum@ci@negative_patterns.ftr.qwen25_omni.invalidation_reason.function_not_directed\endcsname{[0.639, 0.822]}
\expandafter\gdef\csname odunum@val@negative_patterns.ftr.qwen25_omni.invalidation_reason.intent_not_real\endcsname{0.762}
\expandafter\gdef\csname odunum@n@negative_patterns.ftr.qwen25_omni.invalidation_reason.intent_not_real\endcsname{84}
\expandafter\gdef\csname odunum@ci@negative_patterns.ftr.qwen25_omni.invalidation_reason.intent_not_real\endcsname{[0.661, 0.840]}
\expandafter\gdef\csname odunum@val@negative_patterns.ftr.qwen25_omni.invalidation_reason.source_mismatch\endcsname{0.790}
\expandafter\gdef\csname odunum@n@negative_patterns.ftr.qwen25_omni.invalidation_reason.source_mismatch\endcsname{81}
\expandafter\gdef\csname odunum@ci@negative_patterns.ftr.qwen25_omni.invalidation_reason.source_mismatch\endcsname{[0.689, 0.865]}
\expandafter\gdef\csname odunum@val@negative_patterns.ftr.qwen25_omni.invalidation_reason.target_mismatch\endcsname{0.864}
\expandafter\gdef\csname odunum@n@negative_patterns.ftr.qwen25_omni.invalidation_reason.target_mismatch\endcsname{44}
\expandafter\gdef\csname odunum@ci@negative_patterns.ftr.qwen25_omni.invalidation_reason.target_mismatch\endcsname{[0.733, 0.936]}
\expandafter\gdef\csname odunum@val@negative_patterns.ftr.qwen25_omni.signal_type.contextual_signal\endcsname{0.779}
\expandafter\gdef\csname odunum@n@negative_patterns.ftr.qwen25_omni.signal_type.contextual_signal\endcsname{86}
\expandafter\gdef\csname odunum@ci@negative_patterns.ftr.qwen25_omni.signal_type.contextual_signal\endcsname{[0.681, 0.854]}
\expandafter\gdef\csname odunum@val@negative_patterns.ftr.qwen25_omni.signal_type.interactive_signal\endcsname{0.676}
\expandafter\gdef\csname odunum@n@negative_patterns.ftr.qwen25_omni.signal_type.interactive_signal\endcsname{37}
\expandafter\gdef\csname odunum@ci@negative_patterns.ftr.qwen25_omni.signal_type.interactive_signal\endcsname{[0.515, 0.804]}
\expandafter\gdef\csname odunum@val@negative_patterns.ftr.qwen25_omni.signal_type.linguistic_signal\endcsname{0.928}
\expandafter\gdef\csname odunum@n@negative_patterns.ftr.qwen25_omni.signal_type.linguistic_signal\endcsname{69}
\expandafter\gdef\csname odunum@ci@negative_patterns.ftr.qwen25_omni.signal_type.linguistic_signal\endcsname{[0.841, 0.969]}
\expandafter\gdef\csname odunum@val@negative_patterns.ftr.qwen25_omni.signal_type.prosodic_signal\endcsname{0.761}
\expandafter\gdef\csname odunum@n@negative_patterns.ftr.qwen25_omni.signal_type.prosodic_signal\endcsname{67}
\expandafter\gdef\csname odunum@ci@negative_patterns.ftr.qwen25_omni.signal_type.prosodic_signal\endcsname{[0.647, 0.847]}
\expandafter\gdef\csname odunum@val@negative_patterns.ftr.qwen25_omni.signal_type.semantic_signal\endcsname{0.779}
\expandafter\gdef\csname odunum@n@negative_patterns.ftr.qwen25_omni.signal_type.semantic_signal\endcsname{113}
\expandafter\gdef\csname odunum@ci@negative_patterns.ftr.qwen25_omni.signal_type.semantic_signal\endcsname{[0.694, 0.845]}
\expandafter\gdef\csname odunum@val@negative_patterns.ftr.qwen3_omni_instruct.invalidation_reason.demand_incomplete\endcsname{0.962}
\expandafter\gdef\csname odunum@n@negative_patterns.ftr.qwen3_omni_instruct.invalidation_reason.demand_incomplete\endcsname{78}
\expandafter\gdef\csname odunum@ci@negative_patterns.ftr.qwen3_omni_instruct.invalidation_reason.demand_incomplete\endcsname{[0.893, 0.987]}
\expandafter\gdef\csname odunum@val@negative_patterns.ftr.qwen3_omni_instruct.invalidation_reason.function_not_directed\endcsname{0.812}
\expandafter\gdef\csname odunum@n@negative_patterns.ftr.qwen3_omni_instruct.invalidation_reason.function_not_directed\endcsname{85}
\expandafter\gdef\csname odunum@ci@negative_patterns.ftr.qwen3_omni_instruct.invalidation_reason.function_not_directed\endcsname{[0.716, 0.881]}
\expandafter\gdef\csname odunum@val@negative_patterns.ftr.qwen3_omni_instruct.invalidation_reason.intent_not_real\endcsname{0.821}
\expandafter\gdef\csname odunum@n@negative_patterns.ftr.qwen3_omni_instruct.invalidation_reason.intent_not_real\endcsname{84}
\expandafter\gdef\csname odunum@ci@negative_patterns.ftr.qwen3_omni_instruct.invalidation_reason.intent_not_real\endcsname{[0.726, 0.889]}
\expandafter\gdef\csname odunum@val@negative_patterns.ftr.qwen3_omni_instruct.invalidation_reason.source_mismatch\endcsname{0.840}
\expandafter\gdef\csname odunum@n@negative_patterns.ftr.qwen3_omni_instruct.invalidation_reason.source_mismatch\endcsname{81}
\expandafter\gdef\csname odunum@ci@negative_patterns.ftr.qwen3_omni_instruct.invalidation_reason.source_mismatch\endcsname{[0.745, 0.904]}
\expandafter\gdef\csname odunum@val@negative_patterns.ftr.qwen3_omni_instruct.invalidation_reason.target_mismatch\endcsname{0.955}
\expandafter\gdef\csname odunum@n@negative_patterns.ftr.qwen3_omni_instruct.invalidation_reason.target_mismatch\endcsname{44}
\expandafter\gdef\csname odunum@ci@negative_patterns.ftr.qwen3_omni_instruct.invalidation_reason.target_mismatch\endcsname{[0.849, 0.987]}
\expandafter\gdef\csname odunum@val@negative_patterns.ftr.qwen3_omni_instruct.signal_type.contextual_signal\endcsname{0.884}
\expandafter\gdef\csname odunum@n@negative_patterns.ftr.qwen3_omni_instruct.signal_type.contextual_signal\endcsname{86}
\expandafter\gdef\csname odunum@ci@negative_patterns.ftr.qwen3_omni_instruct.signal_type.contextual_signal\endcsname{[0.799, 0.936]}
\expandafter\gdef\csname odunum@val@negative_patterns.ftr.qwen3_omni_instruct.signal_type.interactive_signal\endcsname{0.892}
\expandafter\gdef\csname odunum@n@negative_patterns.ftr.qwen3_omni_instruct.signal_type.interactive_signal\endcsname{37}
\expandafter\gdef\csname odunum@ci@negative_patterns.ftr.qwen3_omni_instruct.signal_type.interactive_signal\endcsname{[0.753, 0.957]}
\expandafter\gdef\csname odunum@val@negative_patterns.ftr.qwen3_omni_instruct.signal_type.linguistic_signal\endcsname{0.957}
\expandafter\gdef\csname odunum@n@negative_patterns.ftr.qwen3_omni_instruct.signal_type.linguistic_signal\endcsname{69}
\expandafter\gdef\csname odunum@ci@negative_patterns.ftr.qwen3_omni_instruct.signal_type.linguistic_signal\endcsname{[0.880, 0.985]}
\expandafter\gdef\csname odunum@val@negative_patterns.ftr.qwen3_omni_instruct.signal_type.prosodic_signal\endcsname{0.791}
\expandafter\gdef\csname odunum@n@negative_patterns.ftr.qwen3_omni_instruct.signal_type.prosodic_signal\endcsname{67}
\expandafter\gdef\csname odunum@ci@negative_patterns.ftr.qwen3_omni_instruct.signal_type.prosodic_signal\endcsname{[0.679, 0.871]}
\expandafter\gdef\csname odunum@val@negative_patterns.ftr.qwen3_omni_instruct.signal_type.semantic_signal\endcsname{0.841}
\expandafter\gdef\csname odunum@n@negative_patterns.ftr.qwen3_omni_instruct.signal_type.semantic_signal\endcsname{113}
\expandafter\gdef\csname odunum@ci@negative_patterns.ftr.qwen3_omni_instruct.signal_type.semantic_signal\endcsname{[0.762, 0.897]}
\expandafter\gdef\csname odunum@val@negative_patterns.ftr.qwen3_omni_think.invalidation_reason.demand_incomplete\endcsname{0.590}
\expandafter\gdef\csname odunum@n@negative_patterns.ftr.qwen3_omni_think.invalidation_reason.demand_incomplete\endcsname{78}
\expandafter\gdef\csname odunum@ci@negative_patterns.ftr.qwen3_omni_think.invalidation_reason.demand_incomplete\endcsname{[0.479, 0.692]}
\expandafter\gdef\csname odunum@val@negative_patterns.ftr.qwen3_omni_think.invalidation_reason.function_not_directed\endcsname{0.706}
\expandafter\gdef\csname odunum@n@negative_patterns.ftr.qwen3_omni_think.invalidation_reason.function_not_directed\endcsname{85}
\expandafter\gdef\csname odunum@ci@negative_patterns.ftr.qwen3_omni_think.invalidation_reason.function_not_directed\endcsname{[0.602, 0.792]}
\expandafter\gdef\csname odunum@val@negative_patterns.ftr.qwen3_omni_think.invalidation_reason.intent_not_real\endcsname{0.667}
\expandafter\gdef\csname odunum@n@negative_patterns.ftr.qwen3_omni_think.invalidation_reason.intent_not_real\endcsname{84}
\expandafter\gdef\csname odunum@ci@negative_patterns.ftr.qwen3_omni_think.invalidation_reason.intent_not_real\endcsname{[0.561, 0.758]}
\expandafter\gdef\csname odunum@val@negative_patterns.ftr.qwen3_omni_think.invalidation_reason.source_mismatch\endcsname{0.827}
\expandafter\gdef\csname odunum@n@negative_patterns.ftr.qwen3_omni_think.invalidation_reason.source_mismatch\endcsname{81}
\expandafter\gdef\csname odunum@ci@negative_patterns.ftr.qwen3_omni_think.invalidation_reason.source_mismatch\endcsname{[0.731, 0.894]}
\expandafter\gdef\csname odunum@val@negative_patterns.ftr.qwen3_omni_think.invalidation_reason.target_mismatch\endcsname{0.841}
\expandafter\gdef\csname odunum@n@negative_patterns.ftr.qwen3_omni_think.invalidation_reason.target_mismatch\endcsname{44}
\expandafter\gdef\csname odunum@ci@negative_patterns.ftr.qwen3_omni_think.invalidation_reason.target_mismatch\endcsname{[0.706, 0.921]}
\expandafter\gdef\csname odunum@val@negative_patterns.ftr.qwen3_omni_think.signal_type.contextual_signal\endcsname{0.721}
\expandafter\gdef\csname odunum@n@negative_patterns.ftr.qwen3_omni_think.signal_type.contextual_signal\endcsname{86}
\expandafter\gdef\csname odunum@ci@negative_patterns.ftr.qwen3_omni_think.signal_type.contextual_signal\endcsname{[0.618, 0.805]}
\expandafter\gdef\csname odunum@val@negative_patterns.ftr.qwen3_omni_think.signal_type.interactive_signal\endcsname{0.649}
\expandafter\gdef\csname odunum@n@negative_patterns.ftr.qwen3_omni_think.signal_type.interactive_signal\endcsname{37}
\expandafter\gdef\csname odunum@ci@negative_patterns.ftr.qwen3_omni_think.signal_type.interactive_signal\endcsname{[0.488, 0.782]}
\expandafter\gdef\csname odunum@val@negative_patterns.ftr.qwen3_omni_think.signal_type.linguistic_signal\endcsname{0.899}
\expandafter\gdef\csname odunum@n@negative_patterns.ftr.qwen3_omni_think.signal_type.linguistic_signal\endcsname{69}
\expandafter\gdef\csname odunum@ci@negative_patterns.ftr.qwen3_omni_think.signal_type.linguistic_signal\endcsname{[0.805, 0.950]}
\expandafter\gdef\csname odunum@val@negative_patterns.ftr.qwen3_omni_think.signal_type.prosodic_signal\endcsname{0.687}
\expandafter\gdef\csname odunum@n@negative_patterns.ftr.qwen3_omni_think.signal_type.prosodic_signal\endcsname{67}
\expandafter\gdef\csname odunum@ci@negative_patterns.ftr.qwen3_omni_think.signal_type.prosodic_signal\endcsname{[0.568, 0.785]}
\expandafter\gdef\csname odunum@val@negative_patterns.ftr.qwen3_omni_think.signal_type.semantic_signal\endcsname{0.637}
\expandafter\gdef\csname odunum@n@negative_patterns.ftr.qwen3_omni_think.signal_type.semantic_signal\endcsname{113}
\expandafter\gdef\csname odunum@ci@negative_patterns.ftr.qwen3_omni_think.signal_type.semantic_signal\endcsname{[0.545, 0.720]}
\expandafter\gdef\csname odunum@val@negative_patterns.ftr.qwen_plus.invalidation_reason.demand_incomplete\endcsname{0.641}
\expandafter\gdef\csname odunum@n@negative_patterns.ftr.qwen_plus.invalidation_reason.demand_incomplete\endcsname{78}
\expandafter\gdef\csname odunum@ci@negative_patterns.ftr.qwen_plus.invalidation_reason.demand_incomplete\endcsname{[0.530, 0.739]}
\expandafter\gdef\csname odunum@val@negative_patterns.ftr.qwen_plus.invalidation_reason.function_not_directed\endcsname{0.588}
\expandafter\gdef\csname odunum@n@negative_patterns.ftr.qwen_plus.invalidation_reason.function_not_directed\endcsname{85}
\expandafter\gdef\csname odunum@ci@negative_patterns.ftr.qwen_plus.invalidation_reason.function_not_directed\endcsname{[0.482, 0.687]}
\expandafter\gdef\csname odunum@val@negative_patterns.ftr.qwen_plus.invalidation_reason.intent_not_real\endcsname{0.536}
\expandafter\gdef\csname odunum@n@negative_patterns.ftr.qwen_plus.invalidation_reason.intent_not_real\endcsname{84}
\expandafter\gdef\csname odunum@ci@negative_patterns.ftr.qwen_plus.invalidation_reason.intent_not_real\endcsname{[0.430, 0.638]}
\expandafter\gdef\csname odunum@val@negative_patterns.ftr.qwen_plus.invalidation_reason.source_mismatch\endcsname{0.654}
\expandafter\gdef\csname odunum@n@negative_patterns.ftr.qwen_plus.invalidation_reason.source_mismatch\endcsname{81}
\expandafter\gdef\csname odunum@ci@negative_patterns.ftr.qwen_plus.invalidation_reason.source_mismatch\endcsname{[0.546, 0.749]}
\expandafter\gdef\csname odunum@val@negative_patterns.ftr.qwen_plus.invalidation_reason.target_mismatch\endcsname{0.841}
\expandafter\gdef\csname odunum@n@negative_patterns.ftr.qwen_plus.invalidation_reason.target_mismatch\endcsname{44}
\expandafter\gdef\csname odunum@ci@negative_patterns.ftr.qwen_plus.invalidation_reason.target_mismatch\endcsname{[0.706, 0.921]}
\expandafter\gdef\csname odunum@val@negative_patterns.ftr.qwen_plus.signal_type.contextual_signal\endcsname{0.674}
\expandafter\gdef\csname odunum@n@negative_patterns.ftr.qwen_plus.signal_type.contextual_signal\endcsname{86}
\expandafter\gdef\csname odunum@ci@negative_patterns.ftr.qwen_plus.signal_type.contextual_signal\endcsname{[0.570, 0.764]}
\expandafter\gdef\csname odunum@val@negative_patterns.ftr.qwen_plus.signal_type.interactive_signal\endcsname{0.676}
\expandafter\gdef\csname odunum@n@negative_patterns.ftr.qwen_plus.signal_type.interactive_signal\endcsname{37}
\expandafter\gdef\csname odunum@ci@negative_patterns.ftr.qwen_plus.signal_type.interactive_signal\endcsname{[0.515, 0.804]}
\expandafter\gdef\csname odunum@val@negative_patterns.ftr.qwen_plus.signal_type.linguistic_signal\endcsname{0.739}
\expandafter\gdef\csname odunum@n@negative_patterns.ftr.qwen_plus.signal_type.linguistic_signal\endcsname{69}
\expandafter\gdef\csname odunum@ci@negative_patterns.ftr.qwen_plus.signal_type.linguistic_signal\endcsname{[0.625, 0.828]}
\expandafter\gdef\csname odunum@val@negative_patterns.ftr.qwen_plus.signal_type.prosodic_signal\endcsname{0.537}
\expandafter\gdef\csname odunum@n@negative_patterns.ftr.qwen_plus.signal_type.prosodic_signal\endcsname{67}
\expandafter\gdef\csname odunum@ci@negative_patterns.ftr.qwen_plus.signal_type.prosodic_signal\endcsname{[0.419, 0.651]}
\expandafter\gdef\csname odunum@val@negative_patterns.ftr.qwen_plus.signal_type.semantic_signal\endcsname{0.575}
\expandafter\gdef\csname odunum@n@negative_patterns.ftr.qwen_plus.signal_type.semantic_signal\endcsname{113}
\expandafter\gdef\csname odunum@ci@negative_patterns.ftr.qwen_plus.signal_type.semantic_signal\endcsname{[0.483, 0.662]}
\expandafter\gdef\csname odunum@val@negative_patterns.ftr.salmonn2_7b.invalidation_reason.demand_incomplete\endcsname{0.962}
\expandafter\gdef\csname odunum@n@negative_patterns.ftr.salmonn2_7b.invalidation_reason.demand_incomplete\endcsname{78}
\expandafter\gdef\csname odunum@ci@negative_patterns.ftr.salmonn2_7b.invalidation_reason.demand_incomplete\endcsname{[0.893, 0.987]}
\expandafter\gdef\csname odunum@val@negative_patterns.ftr.salmonn2_7b.invalidation_reason.function_not_directed\endcsname{0.894}
\expandafter\gdef\csname odunum@n@negative_patterns.ftr.salmonn2_7b.invalidation_reason.function_not_directed\endcsname{85}
\expandafter\gdef\csname odunum@ci@negative_patterns.ftr.salmonn2_7b.invalidation_reason.function_not_directed\endcsname{[0.811, 0.943]}
\expandafter\gdef\csname odunum@val@negative_patterns.ftr.salmonn2_7b.invalidation_reason.intent_not_real\endcsname{0.915}
\expandafter\gdef\csname odunum@n@negative_patterns.ftr.salmonn2_7b.invalidation_reason.intent_not_real\endcsname{82}
\expandafter\gdef\csname odunum@ci@negative_patterns.ftr.salmonn2_7b.invalidation_reason.intent_not_real\endcsname{[0.834, 0.958]}
\expandafter\gdef\csname odunum@val@negative_patterns.ftr.salmonn2_7b.invalidation_reason.source_mismatch\endcsname{0.914}
\expandafter\gdef\csname odunum@n@negative_patterns.ftr.salmonn2_7b.invalidation_reason.source_mismatch\endcsname{81}
\expandafter\gdef\csname odunum@ci@negative_patterns.ftr.salmonn2_7b.invalidation_reason.source_mismatch\endcsname{[0.832, 0.958]}
\expandafter\gdef\csname odunum@val@negative_patterns.ftr.salmonn2_7b.invalidation_reason.target_mismatch\endcsname{0.886}
\expandafter\gdef\csname odunum@n@negative_patterns.ftr.salmonn2_7b.invalidation_reason.target_mismatch\endcsname{44}
\expandafter\gdef\csname odunum@ci@negative_patterns.ftr.salmonn2_7b.invalidation_reason.target_mismatch\endcsname{[0.760, 0.950]}
\expandafter\gdef\csname odunum@val@negative_patterns.ftr.salmonn2_7b.signal_type.contextual_signal\endcsname{0.930}
\expandafter\gdef\csname odunum@n@negative_patterns.ftr.salmonn2_7b.signal_type.contextual_signal\endcsname{86}
\expandafter\gdef\csname odunum@ci@negative_patterns.ftr.salmonn2_7b.signal_type.contextual_signal\endcsname{[0.856, 0.968]}
\expandafter\gdef\csname odunum@val@negative_patterns.ftr.salmonn2_7b.signal_type.interactive_signal\endcsname{0.838}
\expandafter\gdef\csname odunum@n@negative_patterns.ftr.salmonn2_7b.signal_type.interactive_signal\endcsname{37}
\expandafter\gdef\csname odunum@ci@negative_patterns.ftr.salmonn2_7b.signal_type.interactive_signal\endcsname{[0.689, 0.923]}
\expandafter\gdef\csname odunum@val@negative_patterns.ftr.salmonn2_7b.signal_type.linguistic_signal\endcsname{0.971}
\expandafter\gdef\csname odunum@n@negative_patterns.ftr.salmonn2_7b.signal_type.linguistic_signal\endcsname{69}
\expandafter\gdef\csname odunum@ci@negative_patterns.ftr.salmonn2_7b.signal_type.linguistic_signal\endcsname{[0.900, 0.992]}
\expandafter\gdef\csname odunum@val@negative_patterns.ftr.salmonn2_7b.signal_type.prosodic_signal\endcsname{0.894}
\expandafter\gdef\csname odunum@n@negative_patterns.ftr.salmonn2_7b.signal_type.prosodic_signal\endcsname{66}
\expandafter\gdef\csname odunum@ci@negative_patterns.ftr.salmonn2_7b.signal_type.prosodic_signal\endcsname{[0.797, 0.948]}
\expandafter\gdef\csname odunum@val@negative_patterns.ftr.salmonn2_7b.signal_type.semantic_signal\endcsname{0.911}
\expandafter\gdef\csname odunum@n@negative_patterns.ftr.salmonn2_7b.signal_type.semantic_signal\endcsname{112}
\expandafter\gdef\csname odunum@ci@negative_patterns.ftr.salmonn2_7b.signal_type.semantic_signal\endcsname{[0.843, 0.951]}
\expandafter\gdef\csname odunum@val@negative_patterns.ftr.seed.invalidation_reason.demand_incomplete\endcsname{0.769}
\expandafter\gdef\csname odunum@n@negative_patterns.ftr.seed.invalidation_reason.demand_incomplete\endcsname{78}
\expandafter\gdef\csname odunum@ci@negative_patterns.ftr.seed.invalidation_reason.demand_incomplete\endcsname{[0.664, 0.849]}
\expandafter\gdef\csname odunum@val@negative_patterns.ftr.seed.invalidation_reason.function_not_directed\endcsname{0.788}
\expandafter\gdef\csname odunum@n@negative_patterns.ftr.seed.invalidation_reason.function_not_directed\endcsname{85}
\expandafter\gdef\csname odunum@ci@negative_patterns.ftr.seed.invalidation_reason.function_not_directed\endcsname{[0.690, 0.862]}
\expandafter\gdef\csname odunum@val@negative_patterns.ftr.seed.invalidation_reason.intent_not_real\endcsname{0.810}
\expandafter\gdef\csname odunum@n@negative_patterns.ftr.seed.invalidation_reason.intent_not_real\endcsname{84}
\expandafter\gdef\csname odunum@ci@negative_patterns.ftr.seed.invalidation_reason.intent_not_real\endcsname{[0.713, 0.879]}
\expandafter\gdef\csname odunum@val@negative_patterns.ftr.seed.invalidation_reason.source_mismatch\endcsname{0.864}
\expandafter\gdef\csname odunum@n@negative_patterns.ftr.seed.invalidation_reason.source_mismatch\endcsname{81}
\expandafter\gdef\csname odunum@ci@negative_patterns.ftr.seed.invalidation_reason.source_mismatch\endcsname{[0.773, 0.922]}
\expandafter\gdef\csname odunum@val@negative_patterns.ftr.seed.invalidation_reason.target_mismatch\endcsname{0.841}
\expandafter\gdef\csname odunum@n@negative_patterns.ftr.seed.invalidation_reason.target_mismatch\endcsname{44}
\expandafter\gdef\csname odunum@ci@negative_patterns.ftr.seed.invalidation_reason.target_mismatch\endcsname{[0.706, 0.921]}
\expandafter\gdef\csname odunum@val@negative_patterns.ftr.seed.signal_type.contextual_signal\endcsname{0.744}
\expandafter\gdef\csname odunum@n@negative_patterns.ftr.seed.signal_type.contextual_signal\endcsname{86}
\expandafter\gdef\csname odunum@ci@negative_patterns.ftr.seed.signal_type.contextual_signal\endcsname{[0.643, 0.825]}
\expandafter\gdef\csname odunum@val@negative_patterns.ftr.seed.signal_type.interactive_signal\endcsname{0.784}
\expandafter\gdef\csname odunum@n@negative_patterns.ftr.seed.signal_type.interactive_signal\endcsname{37}
\expandafter\gdef\csname odunum@ci@negative_patterns.ftr.seed.signal_type.interactive_signal\endcsname{[0.628, 0.886]}
\expandafter\gdef\csname odunum@val@negative_patterns.ftr.seed.signal_type.linguistic_signal\endcsname{0.913}
\expandafter\gdef\csname odunum@n@negative_patterns.ftr.seed.signal_type.linguistic_signal\endcsname{69}
\expandafter\gdef\csname odunum@ci@negative_patterns.ftr.seed.signal_type.linguistic_signal\endcsname{[0.823, 0.960]}
\expandafter\gdef\csname odunum@val@negative_patterns.ftr.seed.signal_type.prosodic_signal\endcsname{0.821}
\expandafter\gdef\csname odunum@n@negative_patterns.ftr.seed.signal_type.prosodic_signal\endcsname{67}
\expandafter\gdef\csname odunum@ci@negative_patterns.ftr.seed.signal_type.prosodic_signal\endcsname{[0.713, 0.894]}
\expandafter\gdef\csname odunum@val@negative_patterns.ftr.seed.signal_type.semantic_signal\endcsname{0.805}
\expandafter\gdef\csname odunum@n@negative_patterns.ftr.seed.signal_type.semantic_signal\endcsname{113}
\expandafter\gdef\csname odunum@ci@negative_patterns.ftr.seed.signal_type.semantic_signal\endcsname{[0.723, 0.868]}
\expandafter\gdef\csname odunum@val@negative_patterns.overall_ftr.cascade_asr\endcsname{0.597}
\expandafter\gdef\csname odunum@n@negative_patterns.overall_ftr.cascade_asr\endcsname{372}
\expandafter\gdef\csname odunum@val@negative_patterns.overall_ftr.gemini\endcsname{0.379}
\expandafter\gdef\csname odunum@n@negative_patterns.overall_ftr.gemini\endcsname{372}
\expandafter\gdef\csname odunum@val@negative_patterns.overall_ftr.gemini35_flash_lite\endcsname{0.470}
\expandafter\gdef\csname odunum@n@negative_patterns.overall_ftr.gemini35_flash_lite\endcsname{372}
\expandafter\gdef\csname odunum@val@negative_patterns.overall_ftr.gemini37_flash\endcsname{0.246}
\expandafter\gdef\csname odunum@n@negative_patterns.overall_ftr.gemini37_flash\endcsname{366}
\expandafter\gdef\csname odunum@val@negative_patterns.overall_ftr.ming\endcsname{0.849}
\expandafter\gdef\csname odunum@n@negative_patterns.overall_ftr.ming\endcsname{372}
\expandafter\gdef\csname odunum@val@negative_patterns.overall_ftr.minicpm_o\endcsname{0.787}
\expandafter\gdef\csname odunum@n@negative_patterns.overall_ftr.minicpm_o\endcsname{371}
\expandafter\gdef\csname odunum@val@negative_patterns.overall_ftr.nemotron\endcsname{0.745}
\expandafter\gdef\csname odunum@n@negative_patterns.overall_ftr.nemotron\endcsname{372}
\expandafter\gdef\csname odunum@val@negative_patterns.overall_ftr.qwen25_omni\endcsname{0.793}
\expandafter\gdef\csname odunum@n@negative_patterns.overall_ftr.qwen25_omni\endcsname{372}
\expandafter\gdef\csname odunum@val@negative_patterns.overall_ftr.qwen3_omni_instruct\endcsname{0.868}
\expandafter\gdef\csname odunum@n@negative_patterns.overall_ftr.qwen3_omni_instruct\endcsname{372}
\expandafter\gdef\csname odunum@val@negative_patterns.overall_ftr.qwen3_omni_think\endcsname{0.715}
\expandafter\gdef\csname odunum@n@negative_patterns.overall_ftr.qwen3_omni_think\endcsname{372}
\expandafter\gdef\csname odunum@val@negative_patterns.overall_ftr.qwen_plus\endcsname{0.632}
\expandafter\gdef\csname odunum@n@negative_patterns.overall_ftr.qwen_plus\endcsname{372}
\expandafter\gdef\csname odunum@val@negative_patterns.overall_ftr.salmonn2_7b\endcsname{0.916}
\expandafter\gdef\csname odunum@n@negative_patterns.overall_ftr.salmonn2_7b\endcsname{370}
\expandafter\gdef\csname odunum@val@negative_patterns.overall_ftr.seed\endcsname{0.812}
\expandafter\gdef\csname odunum@n@negative_patterns.overall_ftr.seed\endcsname{372}
\expandafter\gdef\csname odunum@val@negative_patterns.panel_mean_delta.invalidation_reason.demand_incomplete\endcsname{-0.016}
\expandafter\gdef\csname odunum@n@negative_patterns.panel_mean_delta.invalidation_reason.demand_incomplete\endcsname{13}
\expandafter\gdef\csname odunum@val@negative_patterns.panel_mean_delta.invalidation_reason.function_not_directed\endcsname{-0.046}
\expandafter\gdef\csname odunum@n@negative_patterns.panel_mean_delta.invalidation_reason.function_not_directed\endcsname{13}
\expandafter\gdef\csname odunum@val@negative_patterns.panel_mean_delta.invalidation_reason.intent_not_real\endcsname{-0.037}
\expandafter\gdef\csname odunum@n@negative_patterns.panel_mean_delta.invalidation_reason.intent_not_real\endcsname{13}
\expandafter\gdef\csname odunum@val@negative_patterns.panel_mean_delta.invalidation_reason.source_mismatch\endcsname{0.058}
\expandafter\gdef\csname odunum@n@negative_patterns.panel_mean_delta.invalidation_reason.source_mismatch\endcsname{13}
\expandafter\gdef\csname odunum@val@negative_patterns.panel_mean_delta.invalidation_reason.target_mismatch\endcsname{0.081}
\expandafter\gdef\csname odunum@n@negative_patterns.panel_mean_delta.invalidation_reason.target_mismatch\endcsname{13}
\expandafter\gdef\csname odunum@val@negative_patterns.panel_mean_delta.signal_type.contextual_signal\endcsname{0.003}
\expandafter\gdef\csname odunum@n@negative_patterns.panel_mean_delta.signal_type.contextual_signal\endcsname{13}
\expandafter\gdef\csname odunum@val@negative_patterns.panel_mean_delta.signal_type.interactive_signal\endcsname{-0.014}
\expandafter\gdef\csname odunum@n@negative_patterns.panel_mean_delta.signal_type.interactive_signal\endcsname{13}
\expandafter\gdef\csname odunum@val@negative_patterns.panel_mean_delta.signal_type.linguistic_signal\endcsname{0.115}
\expandafter\gdef\csname odunum@n@negative_patterns.panel_mean_delta.signal_type.linguistic_signal\endcsname{13}
\expandafter\gdef\csname odunum@val@negative_patterns.panel_mean_delta.signal_type.prosodic_signal\endcsname{-0.040}
\expandafter\gdef\csname odunum@n@negative_patterns.panel_mean_delta.signal_type.prosodic_signal\endcsname{13}
\expandafter\gdef\csname odunum@val@negative_patterns.panel_mean_delta.signal_type.semantic_signal\endcsname{-0.044}
\expandafter\gdef\csname odunum@n@negative_patterns.panel_mean_delta.signal_type.semantic_signal\endcsname{13}
\expandafter\gdef\csname odunum@val@negative_patterns.panel_mean_ftr.invalidation_reason.demand_incomplete\endcsname{0.661}
\expandafter\gdef\csname odunum@n@negative_patterns.panel_mean_ftr.invalidation_reason.demand_incomplete\endcsname{13}
\expandafter\gdef\csname odunum@val@negative_patterns.panel_mean_ftr.invalidation_reason.function_not_directed\endcsname{0.631}
\expandafter\gdef\csname odunum@n@negative_patterns.panel_mean_ftr.invalidation_reason.function_not_directed\endcsname{13}
\expandafter\gdef\csname odunum@val@negative_patterns.panel_mean_ftr.invalidation_reason.intent_not_real\endcsname{0.640}
\expandafter\gdef\csname odunum@n@negative_patterns.panel_mean_ftr.invalidation_reason.intent_not_real\endcsname{13}
\expandafter\gdef\csname odunum@val@negative_patterns.panel_mean_ftr.invalidation_reason.source_mismatch\endcsname{0.736}
\expandafter\gdef\csname odunum@n@negative_patterns.panel_mean_ftr.invalidation_reason.source_mismatch\endcsname{13}
\expandafter\gdef\csname odunum@val@negative_patterns.panel_mean_ftr.invalidation_reason.target_mismatch\endcsname{0.759}
\expandafter\gdef\csname odunum@n@negative_patterns.panel_mean_ftr.invalidation_reason.target_mismatch\endcsname{13}
\expandafter\gdef\csname odunum@val@negative_patterns.panel_mean_ftr.signal_type.contextual_signal\endcsname{0.681}
\expandafter\gdef\csname odunum@n@negative_patterns.panel_mean_ftr.signal_type.contextual_signal\endcsname{13}
\expandafter\gdef\csname odunum@val@negative_patterns.panel_mean_ftr.signal_type.interactive_signal\endcsname{0.664}
\expandafter\gdef\csname odunum@n@negative_patterns.panel_mean_ftr.signal_type.interactive_signal\endcsname{13}
\expandafter\gdef\csname odunum@val@negative_patterns.panel_mean_ftr.signal_type.linguistic_signal\endcsname{0.793}
\expandafter\gdef\csname odunum@n@negative_patterns.panel_mean_ftr.signal_type.linguistic_signal\endcsname{13}
\expandafter\gdef\csname odunum@val@negative_patterns.panel_mean_ftr.signal_type.prosodic_signal\endcsname{0.637}
\expandafter\gdef\csname odunum@n@negative_patterns.panel_mean_ftr.signal_type.prosodic_signal\endcsname{13}
\expandafter\gdef\csname odunum@val@negative_patterns.panel_mean_ftr.signal_type.semantic_signal\endcsname{0.633}
\expandafter\gdef\csname odunum@n@negative_patterns.panel_mean_ftr.signal_type.semantic_signal\endcsname{13}
\expandafter\gdef\csname odunum@val@negative_patterns.panel_sign_positive.invalidation_reason.demand_incomplete\endcsname{7}
\expandafter\gdef\csname odunum@n@negative_patterns.panel_sign_positive.invalidation_reason.demand_incomplete\endcsname{13}
\expandafter\gdef\csname odunum@val@negative_patterns.panel_sign_positive.invalidation_reason.function_not_directed\endcsname{1}
\expandafter\gdef\csname odunum@n@negative_patterns.panel_sign_positive.invalidation_reason.function_not_directed\endcsname{13}
\expandafter\gdef\csname odunum@val@negative_patterns.panel_sign_positive.invalidation_reason.intent_not_real\endcsname{1}
\expandafter\gdef\csname odunum@n@negative_patterns.panel_sign_positive.invalidation_reason.intent_not_real\endcsname{13}
\expandafter\gdef\csname odunum@val@negative_patterns.panel_sign_positive.invalidation_reason.source_mismatch\endcsname{9}
\expandafter\gdef\csname odunum@n@negative_patterns.panel_sign_positive.invalidation_reason.source_mismatch\endcsname{13}
\expandafter\gdef\csname odunum@val@negative_patterns.panel_sign_positive.invalidation_reason.target_mismatch\endcsname{11}
\expandafter\gdef\csname odunum@n@negative_patterns.panel_sign_positive.invalidation_reason.target_mismatch\endcsname{13}
\expandafter\gdef\csname odunum@val@negative_patterns.panel_sign_positive.signal_type.contextual_signal\endcsname{8}
\expandafter\gdef\csname odunum@n@negative_patterns.panel_sign_positive.signal_type.contextual_signal\endcsname{13}
\expandafter\gdef\csname odunum@val@negative_patterns.panel_sign_positive.signal_type.interactive_signal\endcsname{6}
\expandafter\gdef\csname odunum@n@negative_patterns.panel_sign_positive.signal_type.interactive_signal\endcsname{13}
\expandafter\gdef\csname odunum@val@negative_patterns.panel_sign_positive.signal_type.linguistic_signal\endcsname{13}
\expandafter\gdef\csname odunum@n@negative_patterns.panel_sign_positive.signal_type.linguistic_signal\endcsname{13}
\expandafter\gdef\csname odunum@val@negative_patterns.panel_sign_positive.signal_type.prosodic_signal\endcsname{3}
\expandafter\gdef\csname odunum@n@negative_patterns.panel_sign_positive.signal_type.prosodic_signal\endcsname{13}
\expandafter\gdef\csname odunum@val@negative_patterns.panel_sign_positive.signal_type.semantic_signal\endcsname{0}
\expandafter\gdef\csname odunum@n@negative_patterns.panel_sign_positive.signal_type.semantic_signal\endcsname{13}
\expandafter\gdef\csname odunum@val@negative_patterns.pop.invalidation_reason.demand_incomplete.n\endcsname{78}
\expandafter\gdef\csname odunum@n@negative_patterns.pop.invalidation_reason.demand_incomplete.n\endcsname{372}
\expandafter\gdef\csname odunum@val@negative_patterns.pop.invalidation_reason.function_not_directed.n\endcsname{85}
\expandafter\gdef\csname odunum@n@negative_patterns.pop.invalidation_reason.function_not_directed.n\endcsname{372}
\expandafter\gdef\csname odunum@val@negative_patterns.pop.invalidation_reason.intent_not_real.n\endcsname{84}
\expandafter\gdef\csname odunum@n@negative_patterns.pop.invalidation_reason.intent_not_real.n\endcsname{372}
\expandafter\gdef\csname odunum@val@negative_patterns.pop.invalidation_reason.source_mismatch.n\endcsname{81}
\expandafter\gdef\csname odunum@n@negative_patterns.pop.invalidation_reason.source_mismatch.n\endcsname{372}
\expandafter\gdef\csname odunum@val@negative_patterns.pop.invalidation_reason.target_mismatch.n\endcsname{44}
\expandafter\gdef\csname odunum@n@negative_patterns.pop.invalidation_reason.target_mismatch.n\endcsname{372}
\expandafter\gdef\csname odunum@val@negative_patterns.pop.n_models\endcsname{13}
\expandafter\gdef\csname odunum@n@negative_patterns.pop.n_models\endcsname{13}
\expandafter\gdef\csname odunum@val@negative_patterns.pop.negative.n\endcsname{372}
\expandafter\gdef\csname odunum@n@negative_patterns.pop.negative.n\endcsname{447}
\expandafter\gdef\csname odunum@val@negative_patterns.pop.signal_type.contextual_signal.n\endcsname{86}
\expandafter\gdef\csname odunum@n@negative_patterns.pop.signal_type.contextual_signal.n\endcsname{372}
\expandafter\gdef\csname odunum@val@negative_patterns.pop.signal_type.interactive_signal.n\endcsname{37}
\expandafter\gdef\csname odunum@n@negative_patterns.pop.signal_type.interactive_signal.n\endcsname{372}
\expandafter\gdef\csname odunum@val@negative_patterns.pop.signal_type.linguistic_signal.n\endcsname{69}
\expandafter\gdef\csname odunum@n@negative_patterns.pop.signal_type.linguistic_signal.n\endcsname{372}
\expandafter\gdef\csname odunum@val@negative_patterns.pop.signal_type.prosodic_signal.n\endcsname{67}
\expandafter\gdef\csname odunum@n@negative_patterns.pop.signal_type.prosodic_signal.n\endcsname{372}
\expandafter\gdef\csname odunum@val@negative_patterns.pop.signal_type.semantic_signal.n\endcsname{113}
\expandafter\gdef\csname odunum@n@negative_patterns.pop.signal_type.semantic_signal.n\endcsname{372}
\expandafter\gdef\csname odunum@val@negative_patterns.pop.unlabelled.n\endcsname{75}
\expandafter\gdef\csname odunum@n@negative_patterns.pop.unlabelled.n\endcsname{447}
\expandafter\gdef\csname odunum@val@negative_subtypes.ftr.cascade_asr.all\endcsname{0.615}
\expandafter\gdef\csname odunum@n@negative_subtypes.ftr.cascade_asr.all\endcsname{447}
\expandafter\gdef\csname odunum@ci@negative_subtypes.ftr.cascade_asr.all\endcsname{[0.569, 0.659]}
\expandafter\gdef\csname odunum@val@negative_subtypes.ftr.cascade_asr.invalidation_reason.demand_incomplete\endcsname{0.692}
\expandafter\gdef\csname odunum@n@negative_subtypes.ftr.cascade_asr.invalidation_reason.demand_incomplete\endcsname{78}
\expandafter\gdef\csname odunum@ci@negative_subtypes.ftr.cascade_asr.invalidation_reason.demand_incomplete\endcsname{[0.583, 0.784]}
\expandafter\gdef\csname odunum@val@negative_subtypes.ftr.cascade_asr.invalidation_reason.function_not_directed\endcsname{0.471}
\expandafter\gdef\csname odunum@n@negative_subtypes.ftr.cascade_asr.invalidation_reason.function_not_directed\endcsname{85}
\expandafter\gdef\csname odunum@ci@negative_subtypes.ftr.cascade_asr.invalidation_reason.function_not_directed\endcsname{[0.368, 0.576]}
\expandafter\gdef\csname odunum@val@negative_subtypes.ftr.cascade_asr.invalidation_reason.intent_not_real\endcsname{0.524}
\expandafter\gdef\csname odunum@n@negative_subtypes.ftr.cascade_asr.invalidation_reason.intent_not_real\endcsname{84}
\expandafter\gdef\csname odunum@ci@negative_subtypes.ftr.cascade_asr.invalidation_reason.intent_not_real\endcsname{[0.418, 0.627]}
\expandafter\gdef\csname odunum@val@negative_subtypes.ftr.cascade_asr.invalidation_reason.source_mismatch\endcsname{0.679}
\expandafter\gdef\csname odunum@n@negative_subtypes.ftr.cascade_asr.invalidation_reason.source_mismatch\endcsname{81}
\expandafter\gdef\csname odunum@ci@negative_subtypes.ftr.cascade_asr.invalidation_reason.source_mismatch\endcsname{[0.571, 0.771]}
\expandafter\gdef\csname odunum@val@negative_subtypes.ftr.cascade_asr.invalidation_reason.target_mismatch\endcsname{0.659}
\expandafter\gdef\csname odunum@n@negative_subtypes.ftr.cascade_asr.invalidation_reason.target_mismatch\endcsname{44}
\expandafter\gdef\csname odunum@ci@negative_subtypes.ftr.cascade_asr.invalidation_reason.target_mismatch\endcsname{[0.511, 0.781]}
\expandafter\gdef\csname odunum@val@negative_subtypes.ftr.cascade_asr.invalidation_reason.unlabelled_recorded\endcsname{0.707}
\expandafter\gdef\csname odunum@n@negative_subtypes.ftr.cascade_asr.invalidation_reason.unlabelled_recorded\endcsname{75}
\expandafter\gdef\csname odunum@ci@negative_subtypes.ftr.cascade_asr.invalidation_reason.unlabelled_recorded\endcsname{[0.596, 0.798]}
\expandafter\gdef\csname odunum@val@negative_subtypes.ftr.cascade_asr.signal_type.contextual_signal\endcsname{0.546}
\expandafter\gdef\csname odunum@n@negative_subtypes.ftr.cascade_asr.signal_type.contextual_signal\endcsname{86}
\expandafter\gdef\csname odunum@ci@negative_subtypes.ftr.cascade_asr.signal_type.contextual_signal\endcsname{[0.442, 0.647]}
\expandafter\gdef\csname odunum@val@negative_subtypes.ftr.cascade_asr.signal_type.interactive_signal\endcsname{0.540}
\expandafter\gdef\csname odunum@n@negative_subtypes.ftr.cascade_asr.signal_type.interactive_signal\endcsname{37}
\expandafter\gdef\csname odunum@ci@negative_subtypes.ftr.cascade_asr.signal_type.interactive_signal\endcsname{[0.384, 0.690]}
\expandafter\gdef\csname odunum@val@negative_subtypes.ftr.cascade_asr.signal_type.linguistic_signal\endcsname{0.826}
\expandafter\gdef\csname odunum@n@negative_subtypes.ftr.cascade_asr.signal_type.linguistic_signal\endcsname{69}
\expandafter\gdef\csname odunum@ci@negative_subtypes.ftr.cascade_asr.signal_type.linguistic_signal\endcsname{[0.720, 0.898]}
\expandafter\gdef\csname odunum@val@negative_subtypes.ftr.cascade_asr.signal_type.prosodic_signal\endcsname{0.567}
\expandafter\gdef\csname odunum@n@negative_subtypes.ftr.cascade_asr.signal_type.prosodic_signal\endcsname{67}
\expandafter\gdef\csname odunum@ci@negative_subtypes.ftr.cascade_asr.signal_type.prosodic_signal\endcsname{[0.448, 0.679]}
\expandafter\gdef\csname odunum@val@negative_subtypes.ftr.cascade_asr.signal_type.semantic_signal\endcsname{0.531}
\expandafter\gdef\csname odunum@n@negative_subtypes.ftr.cascade_asr.signal_type.semantic_signal\endcsname{113}
\expandafter\gdef\csname odunum@ci@negative_subtypes.ftr.cascade_asr.signal_type.semantic_signal\endcsname{[0.439, 0.620]}
\expandafter\gdef\csname odunum@val@negative_subtypes.ftr.cascade_asr.signal_type.unlabelled_recorded\endcsname{0.707}
\expandafter\gdef\csname odunum@n@negative_subtypes.ftr.cascade_asr.signal_type.unlabelled_recorded\endcsname{75}
\expandafter\gdef\csname odunum@ci@negative_subtypes.ftr.cascade_asr.signal_type.unlabelled_recorded\endcsname{[0.596, 0.798]}
\expandafter\gdef\csname odunum@val@negative_subtypes.ftr.gemini.all\endcsname{0.374}
\expandafter\gdef\csname odunum@n@negative_subtypes.ftr.gemini.all\endcsname{447}
\expandafter\gdef\csname odunum@ci@negative_subtypes.ftr.gemini.all\endcsname{[0.330, 0.419]}
\expandafter\gdef\csname odunum@val@negative_subtypes.ftr.gemini.invalidation_reason.demand_incomplete\endcsname{0.397}
\expandafter\gdef\csname odunum@n@negative_subtypes.ftr.gemini.invalidation_reason.demand_incomplete\endcsname{78}
\expandafter\gdef\csname odunum@ci@negative_subtypes.ftr.gemini.invalidation_reason.demand_incomplete\endcsname{[0.296, 0.508]}
\expandafter\gdef\csname odunum@val@negative_subtypes.ftr.gemini.invalidation_reason.function_not_directed\endcsname{0.259}
\expandafter\gdef\csname odunum@n@negative_subtypes.ftr.gemini.invalidation_reason.function_not_directed\endcsname{85}
\expandafter\gdef\csname odunum@ci@negative_subtypes.ftr.gemini.invalidation_reason.function_not_directed\endcsname{[0.178, 0.361]}
\expandafter\gdef\csname odunum@val@negative_subtypes.ftr.gemini.invalidation_reason.intent_not_real\endcsname{0.345}
\expandafter\gdef\csname odunum@n@negative_subtypes.ftr.gemini.invalidation_reason.intent_not_real\endcsname{84}
\expandafter\gdef\csname odunum@ci@negative_subtypes.ftr.gemini.invalidation_reason.intent_not_real\endcsname{[0.252, 0.452]}
\expandafter\gdef\csname odunum@val@negative_subtypes.ftr.gemini.invalidation_reason.source_mismatch\endcsname{0.494}
\expandafter\gdef\csname odunum@n@negative_subtypes.ftr.gemini.invalidation_reason.source_mismatch\endcsname{81}
\expandafter\gdef\csname odunum@ci@negative_subtypes.ftr.gemini.invalidation_reason.source_mismatch\endcsname{[0.388, 0.600]}
\expandafter\gdef\csname odunum@val@negative_subtypes.ftr.gemini.invalidation_reason.target_mismatch\endcsname{0.432}
\expandafter\gdef\csname odunum@n@negative_subtypes.ftr.gemini.invalidation_reason.target_mismatch\endcsname{44}
\expandafter\gdef\csname odunum@ci@negative_subtypes.ftr.gemini.invalidation_reason.target_mismatch\endcsname{[0.297, 0.578]}
\expandafter\gdef\csname odunum@val@negative_subtypes.ftr.gemini.invalidation_reason.unlabelled_recorded\endcsname{0.347}
\expandafter\gdef\csname odunum@n@negative_subtypes.ftr.gemini.invalidation_reason.unlabelled_recorded\endcsname{75}
\expandafter\gdef\csname odunum@ci@negative_subtypes.ftr.gemini.invalidation_reason.unlabelled_recorded\endcsname{[0.249, 0.459]}
\expandafter\gdef\csname odunum@val@negative_subtypes.ftr.gemini.signal_type.contextual_signal\endcsname{0.360}
\expandafter\gdef\csname odunum@n@negative_subtypes.ftr.gemini.signal_type.contextual_signal\endcsname{86}
\expandafter\gdef\csname odunum@ci@negative_subtypes.ftr.gemini.signal_type.contextual_signal\endcsname{[0.267, 0.466]}
\expandafter\gdef\csname odunum@val@negative_subtypes.ftr.gemini.signal_type.interactive_signal\endcsname{0.460}
\expandafter\gdef\csname odunum@n@negative_subtypes.ftr.gemini.signal_type.interactive_signal\endcsname{37}
\expandafter\gdef\csname odunum@ci@negative_subtypes.ftr.gemini.signal_type.interactive_signal\endcsname{[0.310, 0.616]}
\expandafter\gdef\csname odunum@val@negative_subtypes.ftr.gemini.signal_type.linguistic_signal\endcsname{0.449}
\expandafter\gdef\csname odunum@n@negative_subtypes.ftr.gemini.signal_type.linguistic_signal\endcsname{69}
\expandafter\gdef\csname odunum@ci@negative_subtypes.ftr.gemini.signal_type.linguistic_signal\endcsname{[0.338, 0.566]}
\expandafter\gdef\csname odunum@val@negative_subtypes.ftr.gemini.signal_type.prosodic_signal\endcsname{0.373}
\expandafter\gdef\csname odunum@n@negative_subtypes.ftr.gemini.signal_type.prosodic_signal\endcsname{67}
\expandafter\gdef\csname odunum@ci@negative_subtypes.ftr.gemini.signal_type.prosodic_signal\endcsname{[0.267, 0.493]}
\expandafter\gdef\csname odunum@val@negative_subtypes.ftr.gemini.signal_type.semantic_signal\endcsname{0.327}
\expandafter\gdef\csname odunum@n@negative_subtypes.ftr.gemini.signal_type.semantic_signal\endcsname{113}
\expandafter\gdef\csname odunum@ci@negative_subtypes.ftr.gemini.signal_type.semantic_signal\endcsname{[0.248, 0.418]}
\expandafter\gdef\csname odunum@val@negative_subtypes.ftr.gemini.signal_type.unlabelled_recorded\endcsname{0.347}
\expandafter\gdef\csname odunum@n@negative_subtypes.ftr.gemini.signal_type.unlabelled_recorded\endcsname{75}
\expandafter\gdef\csname odunum@ci@negative_subtypes.ftr.gemini.signal_type.unlabelled_recorded\endcsname{[0.249, 0.459]}
\expandafter\gdef\csname odunum@val@negative_subtypes.ftr.gemini35_flash_lite.all\endcsname{0.494}
\expandafter\gdef\csname odunum@n@negative_subtypes.ftr.gemini35_flash_lite.all\endcsname{447}
\expandafter\gdef\csname odunum@ci@negative_subtypes.ftr.gemini35_flash_lite.all\endcsname{[0.448, 0.541]}
\expandafter\gdef\csname odunum@val@negative_subtypes.ftr.gemini35_flash_lite.invalidation_reason.demand_incomplete\endcsname{0.346}
\expandafter\gdef\csname odunum@n@negative_subtypes.ftr.gemini35_flash_lite.invalidation_reason.demand_incomplete\endcsname{78}
\expandafter\gdef\csname odunum@ci@negative_subtypes.ftr.gemini35_flash_lite.invalidation_reason.demand_incomplete\endcsname{[0.250, 0.457]}
\expandafter\gdef\csname odunum@val@negative_subtypes.ftr.gemini35_flash_lite.invalidation_reason.function_not_directed\endcsname{0.447}
\expandafter\gdef\csname odunum@n@negative_subtypes.ftr.gemini35_flash_lite.invalidation_reason.function_not_directed\endcsname{85}
\expandafter\gdef\csname odunum@ci@negative_subtypes.ftr.gemini35_flash_lite.invalidation_reason.function_not_directed\endcsname{[0.346, 0.553]}
\expandafter\gdef\csname odunum@val@negative_subtypes.ftr.gemini35_flash_lite.invalidation_reason.intent_not_real\endcsname{0.417}
\expandafter\gdef\csname odunum@n@negative_subtypes.ftr.gemini35_flash_lite.invalidation_reason.intent_not_real\endcsname{84}
\expandafter\gdef\csname odunum@ci@negative_subtypes.ftr.gemini35_flash_lite.invalidation_reason.intent_not_real\endcsname{[0.317, 0.523]}
\expandafter\gdef\csname odunum@val@negative_subtypes.ftr.gemini35_flash_lite.invalidation_reason.source_mismatch\endcsname{0.543}
\expandafter\gdef\csname odunum@n@negative_subtypes.ftr.gemini35_flash_lite.invalidation_reason.source_mismatch\endcsname{81}
\expandafter\gdef\csname odunum@ci@negative_subtypes.ftr.gemini35_flash_lite.invalidation_reason.source_mismatch\endcsname{[0.435, 0.647]}
\expandafter\gdef\csname odunum@val@negative_subtypes.ftr.gemini35_flash_lite.invalidation_reason.target_mismatch\endcsname{0.705}
\expandafter\gdef\csname odunum@n@negative_subtypes.ftr.gemini35_flash_lite.invalidation_reason.target_mismatch\endcsname{44}
\expandafter\gdef\csname odunum@ci@negative_subtypes.ftr.gemini35_flash_lite.invalidation_reason.target_mismatch\endcsname{[0.558, 0.818]}
\expandafter\gdef\csname odunum@val@negative_subtypes.ftr.gemini35_flash_lite.invalidation_reason.unlabelled_recorded\endcsname{0.613}
\expandafter\gdef\csname odunum@n@negative_subtypes.ftr.gemini35_flash_lite.invalidation_reason.unlabelled_recorded\endcsname{75}
\expandafter\gdef\csname odunum@ci@negative_subtypes.ftr.gemini35_flash_lite.invalidation_reason.unlabelled_recorded\endcsname{[0.500, 0.715]}
\expandafter\gdef\csname odunum@val@negative_subtypes.ftr.gemini35_flash_lite.signal_type.contextual_signal\endcsname{0.419}
\expandafter\gdef\csname odunum@n@negative_subtypes.ftr.gemini35_flash_lite.signal_type.contextual_signal\endcsname{86}
\expandafter\gdef\csname odunum@ci@negative_subtypes.ftr.gemini35_flash_lite.signal_type.contextual_signal\endcsname{[0.320, 0.524]}
\expandafter\gdef\csname odunum@val@negative_subtypes.ftr.gemini35_flash_lite.signal_type.interactive_signal\endcsname{0.513}
\expandafter\gdef\csname odunum@n@negative_subtypes.ftr.gemini35_flash_lite.signal_type.interactive_signal\endcsname{37}
\expandafter\gdef\csname odunum@ci@negative_subtypes.ftr.gemini35_flash_lite.signal_type.interactive_signal\endcsname{[0.359, 0.666]}
\expandafter\gdef\csname odunum@val@negative_subtypes.ftr.gemini35_flash_lite.signal_type.linguistic_signal\endcsname{0.710}
\expandafter\gdef\csname odunum@n@negative_subtypes.ftr.gemini35_flash_lite.signal_type.linguistic_signal\endcsname{69}
\expandafter\gdef\csname odunum@ci@negative_subtypes.ftr.gemini35_flash_lite.signal_type.linguistic_signal\endcsname{[0.594, 0.804]}
\expandafter\gdef\csname odunum@val@negative_subtypes.ftr.gemini35_flash_lite.signal_type.prosodic_signal\endcsname{0.358}
\expandafter\gdef\csname odunum@n@negative_subtypes.ftr.gemini35_flash_lite.signal_type.prosodic_signal\endcsname{67}
\expandafter\gdef\csname odunum@ci@negative_subtypes.ftr.gemini35_flash_lite.signal_type.prosodic_signal\endcsname{[0.254, 0.478]}
\expandafter\gdef\csname odunum@val@negative_subtypes.ftr.gemini35_flash_lite.signal_type.semantic_signal\endcsname{0.416}
\expandafter\gdef\csname odunum@n@negative_subtypes.ftr.gemini35_flash_lite.signal_type.semantic_signal\endcsname{113}
\expandafter\gdef\csname odunum@ci@negative_subtypes.ftr.gemini35_flash_lite.signal_type.semantic_signal\endcsname{[0.329, 0.508]}
\expandafter\gdef\csname odunum@val@negative_subtypes.ftr.gemini35_flash_lite.signal_type.unlabelled_recorded\endcsname{0.613}
\expandafter\gdef\csname odunum@n@negative_subtypes.ftr.gemini35_flash_lite.signal_type.unlabelled_recorded\endcsname{75}
\expandafter\gdef\csname odunum@ci@negative_subtypes.ftr.gemini35_flash_lite.signal_type.unlabelled_recorded\endcsname{[0.500, 0.715]}
\expandafter\gdef\csname odunum@val@negative_subtypes.ftr.gemini37_flash.all\endcsname{0.234}
\expandafter\gdef\csname odunum@n@negative_subtypes.ftr.gemini37_flash.all\endcsname{440}
\expandafter\gdef\csname odunum@ci@negative_subtypes.ftr.gemini37_flash.all\endcsname{[0.197, 0.276]}
\expandafter\gdef\csname odunum@val@negative_subtypes.ftr.gemini37_flash.invalidation_reason.demand_incomplete\endcsname{0.162}
\expandafter\gdef\csname odunum@n@negative_subtypes.ftr.gemini37_flash.invalidation_reason.demand_incomplete\endcsname{74}
\expandafter\gdef\csname odunum@ci@negative_subtypes.ftr.gemini37_flash.invalidation_reason.demand_incomplete\endcsname{[0.095, 0.262]}
\expandafter\gdef\csname odunum@val@negative_subtypes.ftr.gemini37_flash.invalidation_reason.function_not_directed\endcsname{0.176}
\expandafter\gdef\csname odunum@n@negative_subtypes.ftr.gemini37_flash.invalidation_reason.function_not_directed\endcsname{85}
\expandafter\gdef\csname odunum@ci@negative_subtypes.ftr.gemini37_flash.invalidation_reason.function_not_directed\endcsname{[0.110, 0.271]}
\expandafter\gdef\csname odunum@val@negative_subtypes.ftr.gemini37_flash.invalidation_reason.intent_not_real\endcsname{0.146}
\expandafter\gdef\csname odunum@n@negative_subtypes.ftr.gemini37_flash.invalidation_reason.intent_not_real\endcsname{82}
\expandafter\gdef\csname odunum@ci@negative_subtypes.ftr.gemini37_flash.invalidation_reason.intent_not_real\endcsname{[0.086, 0.239]}
\expandafter\gdef\csname odunum@val@negative_subtypes.ftr.gemini37_flash.invalidation_reason.source_mismatch\endcsname{0.506}
\expandafter\gdef\csname odunum@n@negative_subtypes.ftr.gemini37_flash.invalidation_reason.source_mismatch\endcsname{81}
\expandafter\gdef\csname odunum@ci@negative_subtypes.ftr.gemini37_flash.invalidation_reason.source_mismatch\endcsname{[0.400, 0.612]}
\expandafter\gdef\csname odunum@val@negative_subtypes.ftr.gemini37_flash.invalidation_reason.target_mismatch\endcsname{0.227}
\expandafter\gdef\csname odunum@n@negative_subtypes.ftr.gemini37_flash.invalidation_reason.target_mismatch\endcsname{44}
\expandafter\gdef\csname odunum@ci@negative_subtypes.ftr.gemini37_flash.invalidation_reason.target_mismatch\endcsname{[0.128, 0.370]}
\expandafter\gdef\csname odunum@val@negative_subtypes.ftr.gemini37_flash.invalidation_reason.unlabelled_recorded\endcsname{0.176}
\expandafter\gdef\csname odunum@n@negative_subtypes.ftr.gemini37_flash.invalidation_reason.unlabelled_recorded\endcsname{74}
\expandafter\gdef\csname odunum@ci@negative_subtypes.ftr.gemini37_flash.invalidation_reason.unlabelled_recorded\endcsname{[0.106, 0.278]}
\expandafter\gdef\csname odunum@val@negative_subtypes.ftr.gemini37_flash.signal_type.contextual_signal\endcsname{0.291}
\expandafter\gdef\csname odunum@n@negative_subtypes.ftr.gemini37_flash.signal_type.contextual_signal\endcsname{86}
\expandafter\gdef\csname odunum@ci@negative_subtypes.ftr.gemini37_flash.signal_type.contextual_signal\endcsname{[0.205, 0.394]}
\expandafter\gdef\csname odunum@val@negative_subtypes.ftr.gemini37_flash.signal_type.interactive_signal\endcsname{0.278}
\expandafter\gdef\csname odunum@n@negative_subtypes.ftr.gemini37_flash.signal_type.interactive_signal\endcsname{36}
\expandafter\gdef\csname odunum@ci@negative_subtypes.ftr.gemini37_flash.signal_type.interactive_signal\endcsname{[0.158, 0.440]}
\expandafter\gdef\csname odunum@val@negative_subtypes.ftr.gemini37_flash.signal_type.linguistic_signal\endcsname{0.319}
\expandafter\gdef\csname odunum@n@negative_subtypes.ftr.gemini37_flash.signal_type.linguistic_signal\endcsname{69}
\expandafter\gdef\csname odunum@ci@negative_subtypes.ftr.gemini37_flash.signal_type.linguistic_signal\endcsname{[0.221, 0.436]}
\expandafter\gdef\csname odunum@val@negative_subtypes.ftr.gemini37_flash.signal_type.prosodic_signal\endcsname{0.139}
\expandafter\gdef\csname odunum@n@negative_subtypes.ftr.gemini37_flash.signal_type.prosodic_signal\endcsname{65}
\expandafter\gdef\csname odunum@ci@negative_subtypes.ftr.gemini37_flash.signal_type.prosodic_signal\endcsname{[0.075, 0.243]}
\expandafter\gdef\csname odunum@val@negative_subtypes.ftr.gemini37_flash.signal_type.semantic_signal\endcsname{0.218}
\expandafter\gdef\csname odunum@n@negative_subtypes.ftr.gemini37_flash.signal_type.semantic_signal\endcsname{110}
\expandafter\gdef\csname odunum@ci@negative_subtypes.ftr.gemini37_flash.signal_type.semantic_signal\endcsname{[0.151, 0.304]}
\expandafter\gdef\csname odunum@val@negative_subtypes.ftr.gemini37_flash.signal_type.unlabelled_recorded\endcsname{0.176}
\expandafter\gdef\csname odunum@n@negative_subtypes.ftr.gemini37_flash.signal_type.unlabelled_recorded\endcsname{74}
\expandafter\gdef\csname odunum@ci@negative_subtypes.ftr.gemini37_flash.signal_type.unlabelled_recorded\endcsname{[0.106, 0.278]}
\expandafter\gdef\csname odunum@val@negative_subtypes.ftr.ming.all\endcsname{0.873}
\expandafter\gdef\csname odunum@n@negative_subtypes.ftr.ming.all\endcsname{447}
\expandafter\gdef\csname odunum@ci@negative_subtypes.ftr.ming.all\endcsname{[0.838, 0.900]}
\expandafter\gdef\csname odunum@val@negative_subtypes.ftr.ming.invalidation_reason.demand_incomplete\endcsname{0.692}
\expandafter\gdef\csname odunum@n@negative_subtypes.ftr.ming.invalidation_reason.demand_incomplete\endcsname{78}
\expandafter\gdef\csname odunum@ci@negative_subtypes.ftr.ming.invalidation_reason.demand_incomplete\endcsname{[0.583, 0.784]}
\expandafter\gdef\csname odunum@val@negative_subtypes.ftr.ming.invalidation_reason.function_not_directed\endcsname{0.859}
\expandafter\gdef\csname odunum@n@negative_subtypes.ftr.ming.invalidation_reason.function_not_directed\endcsname{85}
\expandafter\gdef\csname odunum@ci@negative_subtypes.ftr.ming.invalidation_reason.function_not_directed\endcsname{[0.769, 0.917]}
\expandafter\gdef\csname odunum@val@negative_subtypes.ftr.ming.invalidation_reason.intent_not_real\endcsname{0.881}
\expandafter\gdef\csname odunum@n@negative_subtypes.ftr.ming.invalidation_reason.intent_not_real\endcsname{84}
\expandafter\gdef\csname odunum@ci@negative_subtypes.ftr.ming.invalidation_reason.intent_not_real\endcsname{[0.795, 0.934]}
\expandafter\gdef\csname odunum@val@negative_subtypes.ftr.ming.invalidation_reason.source_mismatch\endcsname{0.914}
\expandafter\gdef\csname odunum@n@negative_subtypes.ftr.ming.invalidation_reason.source_mismatch\endcsname{81}
\expandafter\gdef\csname odunum@ci@negative_subtypes.ftr.ming.invalidation_reason.source_mismatch\endcsname{[0.832, 0.958]}
\expandafter\gdef\csname odunum@val@negative_subtypes.ftr.ming.invalidation_reason.target_mismatch\endcsname{0.932}
\expandafter\gdef\csname odunum@n@negative_subtypes.ftr.ming.invalidation_reason.target_mismatch\endcsname{44}
\expandafter\gdef\csname odunum@ci@negative_subtypes.ftr.ming.invalidation_reason.target_mismatch\endcsname{[0.818, 0.977]}
\expandafter\gdef\csname odunum@val@negative_subtypes.ftr.ming.invalidation_reason.unlabelled_recorded\endcsname{0.987}
\expandafter\gdef\csname odunum@n@negative_subtypes.ftr.ming.invalidation_reason.unlabelled_recorded\endcsname{75}
\expandafter\gdef\csname odunum@ci@negative_subtypes.ftr.ming.invalidation_reason.unlabelled_recorded\endcsname{[0.928, 0.998]}
\expandafter\gdef\csname odunum@val@negative_subtypes.ftr.ming.signal_type.contextual_signal\endcsname{0.861}
\expandafter\gdef\csname odunum@n@negative_subtypes.ftr.ming.signal_type.contextual_signal\endcsname{86}
\expandafter\gdef\csname odunum@ci@negative_subtypes.ftr.ming.signal_type.contextual_signal\endcsname{[0.772, 0.918]}
\expandafter\gdef\csname odunum@val@negative_subtypes.ftr.ming.signal_type.interactive_signal\endcsname{0.865}
\expandafter\gdef\csname odunum@n@negative_subtypes.ftr.ming.signal_type.interactive_signal\endcsname{37}
\expandafter\gdef\csname odunum@ci@negative_subtypes.ftr.ming.signal_type.interactive_signal\endcsname{[0.720, 0.941]}
\expandafter\gdef\csname odunum@val@negative_subtypes.ftr.ming.signal_type.linguistic_signal\endcsname{0.942}
\expandafter\gdef\csname odunum@n@negative_subtypes.ftr.ming.signal_type.linguistic_signal\endcsname{69}
\expandafter\gdef\csname odunum@ci@negative_subtypes.ftr.ming.signal_type.linguistic_signal\endcsname{[0.860, 0.977]}
\expandafter\gdef\csname odunum@val@negative_subtypes.ftr.ming.signal_type.prosodic_signal\endcsname{0.851}
\expandafter\gdef\csname odunum@n@negative_subtypes.ftr.ming.signal_type.prosodic_signal\endcsname{67}
\expandafter\gdef\csname odunum@ci@negative_subtypes.ftr.ming.signal_type.prosodic_signal\endcsname{[0.747, 0.917]}
\expandafter\gdef\csname odunum@val@negative_subtypes.ftr.ming.signal_type.semantic_signal\endcsname{0.779}
\expandafter\gdef\csname odunum@n@negative_subtypes.ftr.ming.signal_type.semantic_signal\endcsname{113}
\expandafter\gdef\csname odunum@ci@negative_subtypes.ftr.ming.signal_type.semantic_signal\endcsname{[0.694, 0.845]}
\expandafter\gdef\csname odunum@val@negative_subtypes.ftr.ming.signal_type.unlabelled_recorded\endcsname{0.987}
\expandafter\gdef\csname odunum@n@negative_subtypes.ftr.ming.signal_type.unlabelled_recorded\endcsname{75}
\expandafter\gdef\csname odunum@ci@negative_subtypes.ftr.ming.signal_type.unlabelled_recorded\endcsname{[0.928, 0.998]}
\expandafter\gdef\csname odunum@val@negative_subtypes.ftr.minicpm_o.all\endcsname{0.814}
\expandafter\gdef\csname odunum@n@negative_subtypes.ftr.minicpm_o.all\endcsname{446}
\expandafter\gdef\csname odunum@ci@negative_subtypes.ftr.minicpm_o.all\endcsname{[0.775, 0.847]}
\expandafter\gdef\csname odunum@val@negative_subtypes.ftr.minicpm_o.invalidation_reason.demand_incomplete\endcsname{0.821}
\expandafter\gdef\csname odunum@n@negative_subtypes.ftr.minicpm_o.invalidation_reason.demand_incomplete\endcsname{78}
\expandafter\gdef\csname odunum@ci@negative_subtypes.ftr.minicpm_o.invalidation_reason.demand_incomplete\endcsname{[0.721, 0.890]}
\expandafter\gdef\csname odunum@val@negative_subtypes.ftr.minicpm_o.invalidation_reason.function_not_directed\endcsname{0.726}
\expandafter\gdef\csname odunum@n@negative_subtypes.ftr.minicpm_o.invalidation_reason.function_not_directed\endcsname{84}
\expandafter\gdef\csname odunum@ci@negative_subtypes.ftr.minicpm_o.invalidation_reason.function_not_directed\endcsname{[0.623, 0.810]}
\expandafter\gdef\csname odunum@val@negative_subtypes.ftr.minicpm_o.invalidation_reason.intent_not_real\endcsname{0.762}
\expandafter\gdef\csname odunum@n@negative_subtypes.ftr.minicpm_o.invalidation_reason.intent_not_real\endcsname{84}
\expandafter\gdef\csname odunum@ci@negative_subtypes.ftr.minicpm_o.invalidation_reason.intent_not_real\endcsname{[0.661, 0.840]}
\expandafter\gdef\csname odunum@val@negative_subtypes.ftr.minicpm_o.invalidation_reason.source_mismatch\endcsname{0.765}
\expandafter\gdef\csname odunum@n@negative_subtypes.ftr.minicpm_o.invalidation_reason.source_mismatch\endcsname{81}
\expandafter\gdef\csname odunum@ci@negative_subtypes.ftr.minicpm_o.invalidation_reason.source_mismatch\endcsname{[0.662, 0.844]}
\expandafter\gdef\csname odunum@val@negative_subtypes.ftr.minicpm_o.invalidation_reason.target_mismatch\endcsname{0.932}
\expandafter\gdef\csname odunum@n@negative_subtypes.ftr.minicpm_o.invalidation_reason.target_mismatch\endcsname{44}
\expandafter\gdef\csname odunum@ci@negative_subtypes.ftr.minicpm_o.invalidation_reason.target_mismatch\endcsname{[0.818, 0.977]}
\expandafter\gdef\csname odunum@val@negative_subtypes.ftr.minicpm_o.invalidation_reason.unlabelled_recorded\endcsname{0.947}
\expandafter\gdef\csname odunum@n@negative_subtypes.ftr.minicpm_o.invalidation_reason.unlabelled_recorded\endcsname{75}
\expandafter\gdef\csname odunum@ci@negative_subtypes.ftr.minicpm_o.invalidation_reason.unlabelled_recorded\endcsname{[0.871, 0.979]}
\expandafter\gdef\csname odunum@val@negative_subtypes.ftr.minicpm_o.signal_type.contextual_signal\endcsname{0.847}
\expandafter\gdef\csname odunum@n@negative_subtypes.ftr.minicpm_o.signal_type.contextual_signal\endcsname{85}
\expandafter\gdef\csname odunum@ci@negative_subtypes.ftr.minicpm_o.signal_type.contextual_signal\endcsname{[0.756, 0.908]}
\expandafter\gdef\csname odunum@val@negative_subtypes.ftr.minicpm_o.signal_type.interactive_signal\endcsname{0.757}
\expandafter\gdef\csname odunum@n@negative_subtypes.ftr.minicpm_o.signal_type.interactive_signal\endcsname{37}
\expandafter\gdef\csname odunum@ci@negative_subtypes.ftr.minicpm_o.signal_type.interactive_signal\endcsname{[0.599, 0.866]}
\expandafter\gdef\csname odunum@val@negative_subtypes.ftr.minicpm_o.signal_type.linguistic_signal\endcsname{0.884}
\expandafter\gdef\csname odunum@n@negative_subtypes.ftr.minicpm_o.signal_type.linguistic_signal\endcsname{69}
\expandafter\gdef\csname odunum@ci@negative_subtypes.ftr.minicpm_o.signal_type.linguistic_signal\endcsname{[0.788, 0.940]}
\expandafter\gdef\csname odunum@val@negative_subtypes.ftr.minicpm_o.signal_type.prosodic_signal\endcsname{0.761}
\expandafter\gdef\csname odunum@n@negative_subtypes.ftr.minicpm_o.signal_type.prosodic_signal\endcsname{67}
\expandafter\gdef\csname odunum@ci@negative_subtypes.ftr.minicpm_o.signal_type.prosodic_signal\endcsname{[0.647, 0.847]}
\expandafter\gdef\csname odunum@val@negative_subtypes.ftr.minicpm_o.signal_type.semantic_signal\endcsname{0.708}
\expandafter\gdef\csname odunum@n@negative_subtypes.ftr.minicpm_o.signal_type.semantic_signal\endcsname{113}
\expandafter\gdef\csname odunum@ci@negative_subtypes.ftr.minicpm_o.signal_type.semantic_signal\endcsname{[0.618, 0.784]}
\expandafter\gdef\csname odunum@val@negative_subtypes.ftr.minicpm_o.signal_type.unlabelled_recorded\endcsname{0.947}
\expandafter\gdef\csname odunum@n@negative_subtypes.ftr.minicpm_o.signal_type.unlabelled_recorded\endcsname{75}
\expandafter\gdef\csname odunum@ci@negative_subtypes.ftr.minicpm_o.signal_type.unlabelled_recorded\endcsname{[0.871, 0.979]}
\expandafter\gdef\csname odunum@val@negative_subtypes.ftr.nemotron.all\endcsname{0.720}
\expandafter\gdef\csname odunum@n@negative_subtypes.ftr.nemotron.all\endcsname{447}
\expandafter\gdef\csname odunum@ci@negative_subtypes.ftr.nemotron.all\endcsname{[0.677, 0.760]}
\expandafter\gdef\csname odunum@val@negative_subtypes.ftr.nemotron.invalidation_reason.demand_incomplete\endcsname{0.718}
\expandafter\gdef\csname odunum@n@negative_subtypes.ftr.nemotron.invalidation_reason.demand_incomplete\endcsname{78}
\expandafter\gdef\csname odunum@ci@negative_subtypes.ftr.nemotron.invalidation_reason.demand_incomplete\endcsname{[0.610, 0.806]}
\expandafter\gdef\csname odunum@val@negative_subtypes.ftr.nemotron.invalidation_reason.function_not_directed\endcsname{0.741}
\expandafter\gdef\csname odunum@n@negative_subtypes.ftr.nemotron.invalidation_reason.function_not_directed\endcsname{85}
\expandafter\gdef\csname odunum@ci@negative_subtypes.ftr.nemotron.invalidation_reason.function_not_directed\endcsname{[0.639, 0.822]}
\expandafter\gdef\csname odunum@val@negative_subtypes.ftr.nemotron.invalidation_reason.intent_not_real\endcsname{0.738}
\expandafter\gdef\csname odunum@n@negative_subtypes.ftr.nemotron.invalidation_reason.intent_not_real\endcsname{84}
\expandafter\gdef\csname odunum@ci@negative_subtypes.ftr.nemotron.invalidation_reason.intent_not_real\endcsname{[0.635, 0.820]}
\expandafter\gdef\csname odunum@val@negative_subtypes.ftr.nemotron.invalidation_reason.source_mismatch\endcsname{0.778}
\expandafter\gdef\csname odunum@n@negative_subtypes.ftr.nemotron.invalidation_reason.source_mismatch\endcsname{81}
\expandafter\gdef\csname odunum@ci@negative_subtypes.ftr.nemotron.invalidation_reason.source_mismatch\endcsname{[0.676, 0.855]}
\expandafter\gdef\csname odunum@val@negative_subtypes.ftr.nemotron.invalidation_reason.target_mismatch\endcsname{0.750}
\expandafter\gdef\csname odunum@n@negative_subtypes.ftr.nemotron.invalidation_reason.target_mismatch\endcsname{44}
\expandafter\gdef\csname odunum@ci@negative_subtypes.ftr.nemotron.invalidation_reason.target_mismatch\endcsname{[0.606, 0.854]}
\expandafter\gdef\csname odunum@val@negative_subtypes.ftr.nemotron.invalidation_reason.unlabelled_recorded\endcsname{0.600}
\expandafter\gdef\csname odunum@n@negative_subtypes.ftr.nemotron.invalidation_reason.unlabelled_recorded\endcsname{75}
\expandafter\gdef\csname odunum@ci@negative_subtypes.ftr.nemotron.invalidation_reason.unlabelled_recorded\endcsname{[0.487, 0.703]}
\expandafter\gdef\csname odunum@val@negative_subtypes.ftr.nemotron.signal_type.contextual_signal\endcsname{0.791}
\expandafter\gdef\csname odunum@n@negative_subtypes.ftr.nemotron.signal_type.contextual_signal\endcsname{86}
\expandafter\gdef\csname odunum@ci@negative_subtypes.ftr.nemotron.signal_type.contextual_signal\endcsname{[0.693, 0.863]}
\expandafter\gdef\csname odunum@val@negative_subtypes.ftr.nemotron.signal_type.interactive_signal\endcsname{0.703}
\expandafter\gdef\csname odunum@n@negative_subtypes.ftr.nemotron.signal_type.interactive_signal\endcsname{37}
\expandafter\gdef\csname odunum@ci@negative_subtypes.ftr.nemotron.signal_type.interactive_signal\endcsname{[0.542, 0.825]}
\expandafter\gdef\csname odunum@val@negative_subtypes.ftr.nemotron.signal_type.linguistic_signal\endcsname{0.768}
\expandafter\gdef\csname odunum@n@negative_subtypes.ftr.nemotron.signal_type.linguistic_signal\endcsname{69}
\expandafter\gdef\csname odunum@ci@negative_subtypes.ftr.nemotron.signal_type.linguistic_signal\endcsname{[0.656, 0.852]}
\expandafter\gdef\csname odunum@val@negative_subtypes.ftr.nemotron.signal_type.prosodic_signal\endcsname{0.746}
\expandafter\gdef\csname odunum@n@negative_subtypes.ftr.nemotron.signal_type.prosodic_signal\endcsname{67}
\expandafter\gdef\csname odunum@ci@negative_subtypes.ftr.nemotron.signal_type.prosodic_signal\endcsname{[0.631, 0.835]}
\expandafter\gdef\csname odunum@val@negative_subtypes.ftr.nemotron.signal_type.semantic_signal\endcsname{0.708}
\expandafter\gdef\csname odunum@n@negative_subtypes.ftr.nemotron.signal_type.semantic_signal\endcsname{113}
\expandafter\gdef\csname odunum@ci@negative_subtypes.ftr.nemotron.signal_type.semantic_signal\endcsname{[0.618, 0.784]}
\expandafter\gdef\csname odunum@val@negative_subtypes.ftr.nemotron.signal_type.unlabelled_recorded\endcsname{0.600}
\expandafter\gdef\csname odunum@n@negative_subtypes.ftr.nemotron.signal_type.unlabelled_recorded\endcsname{75}
\expandafter\gdef\csname odunum@ci@negative_subtypes.ftr.nemotron.signal_type.unlabelled_recorded\endcsname{[0.487, 0.703]}
\expandafter\gdef\csname odunum@val@negative_subtypes.ftr.qwen25_omni.all\endcsname{0.801}
\expandafter\gdef\csname odunum@n@negative_subtypes.ftr.qwen25_omni.all\endcsname{447}
\expandafter\gdef\csname odunum@ci@negative_subtypes.ftr.qwen25_omni.all\endcsname{[0.761, 0.835]}
\expandafter\gdef\csname odunum@val@negative_subtypes.ftr.qwen25_omni.invalidation_reason.demand_incomplete\endcsname{0.846}
\expandafter\gdef\csname odunum@n@negative_subtypes.ftr.qwen25_omni.invalidation_reason.demand_incomplete\endcsname{78}
\expandafter\gdef\csname odunum@ci@negative_subtypes.ftr.qwen25_omni.invalidation_reason.demand_incomplete\endcsname{[0.750, 0.910]}
\expandafter\gdef\csname odunum@val@negative_subtypes.ftr.qwen25_omni.invalidation_reason.function_not_directed\endcsname{0.741}
\expandafter\gdef\csname odunum@n@negative_subtypes.ftr.qwen25_omni.invalidation_reason.function_not_directed\endcsname{85}
\expandafter\gdef\csname odunum@ci@negative_subtypes.ftr.qwen25_omni.invalidation_reason.function_not_directed\endcsname{[0.639, 0.822]}
\expandafter\gdef\csname odunum@val@negative_subtypes.ftr.qwen25_omni.invalidation_reason.intent_not_real\endcsname{0.762}
\expandafter\gdef\csname odunum@n@negative_subtypes.ftr.qwen25_omni.invalidation_reason.intent_not_real\endcsname{84}
\expandafter\gdef\csname odunum@ci@negative_subtypes.ftr.qwen25_omni.invalidation_reason.intent_not_real\endcsname{[0.661, 0.840]}
\expandafter\gdef\csname odunum@val@negative_subtypes.ftr.qwen25_omni.invalidation_reason.source_mismatch\endcsname{0.790}
\expandafter\gdef\csname odunum@n@negative_subtypes.ftr.qwen25_omni.invalidation_reason.source_mismatch\endcsname{81}
\expandafter\gdef\csname odunum@ci@negative_subtypes.ftr.qwen25_omni.invalidation_reason.source_mismatch\endcsname{[0.689, 0.865]}
\expandafter\gdef\csname odunum@val@negative_subtypes.ftr.qwen25_omni.invalidation_reason.target_mismatch\endcsname{0.864}
\expandafter\gdef\csname odunum@n@negative_subtypes.ftr.qwen25_omni.invalidation_reason.target_mismatch\endcsname{44}
\expandafter\gdef\csname odunum@ci@negative_subtypes.ftr.qwen25_omni.invalidation_reason.target_mismatch\endcsname{[0.733, 0.936]}
\expandafter\gdef\csname odunum@val@negative_subtypes.ftr.qwen25_omni.invalidation_reason.unlabelled_recorded\endcsname{0.840}
\expandafter\gdef\csname odunum@n@negative_subtypes.ftr.qwen25_omni.invalidation_reason.unlabelled_recorded\endcsname{75}
\expandafter\gdef\csname odunum@ci@negative_subtypes.ftr.qwen25_omni.invalidation_reason.unlabelled_recorded\endcsname{[0.741, 0.906]}
\expandafter\gdef\csname odunum@val@negative_subtypes.ftr.qwen25_omni.signal_type.contextual_signal\endcsname{0.779}
\expandafter\gdef\csname odunum@n@negative_subtypes.ftr.qwen25_omni.signal_type.contextual_signal\endcsname{86}
\expandafter\gdef\csname odunum@ci@negative_subtypes.ftr.qwen25_omni.signal_type.contextual_signal\endcsname{[0.681, 0.854]}
\expandafter\gdef\csname odunum@val@negative_subtypes.ftr.qwen25_omni.signal_type.interactive_signal\endcsname{0.676}
\expandafter\gdef\csname odunum@n@negative_subtypes.ftr.qwen25_omni.signal_type.interactive_signal\endcsname{37}
\expandafter\gdef\csname odunum@ci@negative_subtypes.ftr.qwen25_omni.signal_type.interactive_signal\endcsname{[0.515, 0.804]}
\expandafter\gdef\csname odunum@val@negative_subtypes.ftr.qwen25_omni.signal_type.linguistic_signal\endcsname{0.927}
\expandafter\gdef\csname odunum@n@negative_subtypes.ftr.qwen25_omni.signal_type.linguistic_signal\endcsname{69}
\expandafter\gdef\csname odunum@ci@negative_subtypes.ftr.qwen25_omni.signal_type.linguistic_signal\endcsname{[0.841, 0.969]}
\expandafter\gdef\csname odunum@val@negative_subtypes.ftr.qwen25_omni.signal_type.prosodic_signal\endcsname{0.761}
\expandafter\gdef\csname odunum@n@negative_subtypes.ftr.qwen25_omni.signal_type.prosodic_signal\endcsname{67}
\expandafter\gdef\csname odunum@ci@negative_subtypes.ftr.qwen25_omni.signal_type.prosodic_signal\endcsname{[0.647, 0.847]}
\expandafter\gdef\csname odunum@val@negative_subtypes.ftr.qwen25_omni.signal_type.semantic_signal\endcsname{0.779}
\expandafter\gdef\csname odunum@n@negative_subtypes.ftr.qwen25_omni.signal_type.semantic_signal\endcsname{113}
\expandafter\gdef\csname odunum@ci@negative_subtypes.ftr.qwen25_omni.signal_type.semantic_signal\endcsname{[0.694, 0.845]}
\expandafter\gdef\csname odunum@val@negative_subtypes.ftr.qwen25_omni.signal_type.unlabelled_recorded\endcsname{0.840}
\expandafter\gdef\csname odunum@n@negative_subtypes.ftr.qwen25_omni.signal_type.unlabelled_recorded\endcsname{75}
\expandafter\gdef\csname odunum@ci@negative_subtypes.ftr.qwen25_omni.signal_type.unlabelled_recorded\endcsname{[0.741, 0.906]}
\expandafter\gdef\csname odunum@val@negative_subtypes.ftr.qwen3_omni_instruct.all\endcsname{0.888}
\expandafter\gdef\csname odunum@n@negative_subtypes.ftr.qwen3_omni_instruct.all\endcsname{447}
\expandafter\gdef\csname odunum@ci@negative_subtypes.ftr.qwen3_omni_instruct.all\endcsname{[0.856, 0.914]}
\expandafter\gdef\csname odunum@val@negative_subtypes.ftr.qwen3_omni_instruct.invalidation_reason.demand_incomplete\endcsname{0.962}
\expandafter\gdef\csname odunum@n@negative_subtypes.ftr.qwen3_omni_instruct.invalidation_reason.demand_incomplete\endcsname{78}
\expandafter\gdef\csname odunum@ci@negative_subtypes.ftr.qwen3_omni_instruct.invalidation_reason.demand_incomplete\endcsname{[0.893, 0.987]}
\expandafter\gdef\csname odunum@val@negative_subtypes.ftr.qwen3_omni_instruct.invalidation_reason.function_not_directed\endcsname{0.812}
\expandafter\gdef\csname odunum@n@negative_subtypes.ftr.qwen3_omni_instruct.invalidation_reason.function_not_directed\endcsname{85}
\expandafter\gdef\csname odunum@ci@negative_subtypes.ftr.qwen3_omni_instruct.invalidation_reason.function_not_directed\endcsname{[0.716, 0.881]}
\expandafter\gdef\csname odunum@val@negative_subtypes.ftr.qwen3_omni_instruct.invalidation_reason.intent_not_real\endcsname{0.821}
\expandafter\gdef\csname odunum@n@negative_subtypes.ftr.qwen3_omni_instruct.invalidation_reason.intent_not_real\endcsname{84}
\expandafter\gdef\csname odunum@ci@negative_subtypes.ftr.qwen3_omni_instruct.invalidation_reason.intent_not_real\endcsname{[0.726, 0.889]}
\expandafter\gdef\csname odunum@val@negative_subtypes.ftr.qwen3_omni_instruct.invalidation_reason.source_mismatch\endcsname{0.840}
\expandafter\gdef\csname odunum@n@negative_subtypes.ftr.qwen3_omni_instruct.invalidation_reason.source_mismatch\endcsname{81}
\expandafter\gdef\csname odunum@ci@negative_subtypes.ftr.qwen3_omni_instruct.invalidation_reason.source_mismatch\endcsname{[0.745, 0.904]}
\expandafter\gdef\csname odunum@val@negative_subtypes.ftr.qwen3_omni_instruct.invalidation_reason.target_mismatch\endcsname{0.955}
\expandafter\gdef\csname odunum@n@negative_subtypes.ftr.qwen3_omni_instruct.invalidation_reason.target_mismatch\endcsname{44}
\expandafter\gdef\csname odunum@ci@negative_subtypes.ftr.qwen3_omni_instruct.invalidation_reason.target_mismatch\endcsname{[0.849, 0.987]}
\expandafter\gdef\csname odunum@val@negative_subtypes.ftr.qwen3_omni_instruct.invalidation_reason.unlabelled_recorded\endcsname{0.987}
\expandafter\gdef\csname odunum@n@negative_subtypes.ftr.qwen3_omni_instruct.invalidation_reason.unlabelled_recorded\endcsname{75}
\expandafter\gdef\csname odunum@ci@negative_subtypes.ftr.qwen3_omni_instruct.invalidation_reason.unlabelled_recorded\endcsname{[0.928, 0.998]}
\expandafter\gdef\csname odunum@val@negative_subtypes.ftr.qwen3_omni_instruct.signal_type.contextual_signal\endcsname{0.884}
\expandafter\gdef\csname odunum@n@negative_subtypes.ftr.qwen3_omni_instruct.signal_type.contextual_signal\endcsname{86}
\expandafter\gdef\csname odunum@ci@negative_subtypes.ftr.qwen3_omni_instruct.signal_type.contextual_signal\endcsname{[0.799, 0.936]}
\expandafter\gdef\csname odunum@val@negative_subtypes.ftr.qwen3_omni_instruct.signal_type.interactive_signal\endcsname{0.892}
\expandafter\gdef\csname odunum@n@negative_subtypes.ftr.qwen3_omni_instruct.signal_type.interactive_signal\endcsname{37}
\expandafter\gdef\csname odunum@ci@negative_subtypes.ftr.qwen3_omni_instruct.signal_type.interactive_signal\endcsname{[0.753, 0.957]}
\expandafter\gdef\csname odunum@val@negative_subtypes.ftr.qwen3_omni_instruct.signal_type.linguistic_signal\endcsname{0.957}
\expandafter\gdef\csname odunum@n@negative_subtypes.ftr.qwen3_omni_instruct.signal_type.linguistic_signal\endcsname{69}
\expandafter\gdef\csname odunum@ci@negative_subtypes.ftr.qwen3_omni_instruct.signal_type.linguistic_signal\endcsname{[0.880, 0.985]}
\expandafter\gdef\csname odunum@val@negative_subtypes.ftr.qwen3_omni_instruct.signal_type.prosodic_signal\endcsname{0.791}
\expandafter\gdef\csname odunum@n@negative_subtypes.ftr.qwen3_omni_instruct.signal_type.prosodic_signal\endcsname{67}
\expandafter\gdef\csname odunum@ci@negative_subtypes.ftr.qwen3_omni_instruct.signal_type.prosodic_signal\endcsname{[0.679, 0.871]}
\expandafter\gdef\csname odunum@val@negative_subtypes.ftr.qwen3_omni_instruct.signal_type.semantic_signal\endcsname{0.841}
\expandafter\gdef\csname odunum@n@negative_subtypes.ftr.qwen3_omni_instruct.signal_type.semantic_signal\endcsname{113}
\expandafter\gdef\csname odunum@ci@negative_subtypes.ftr.qwen3_omni_instruct.signal_type.semantic_signal\endcsname{[0.762, 0.897]}
\expandafter\gdef\csname odunum@val@negative_subtypes.ftr.qwen3_omni_instruct.signal_type.unlabelled_recorded\endcsname{0.987}
\expandafter\gdef\csname odunum@n@negative_subtypes.ftr.qwen3_omni_instruct.signal_type.unlabelled_recorded\endcsname{75}
\expandafter\gdef\csname odunum@ci@negative_subtypes.ftr.qwen3_omni_instruct.signal_type.unlabelled_recorded\endcsname{[0.928, 0.998]}
\expandafter\gdef\csname odunum@val@negative_subtypes.ftr.qwen3_omni_think.all\endcsname{0.749}
\expandafter\gdef\csname odunum@n@negative_subtypes.ftr.qwen3_omni_think.all\endcsname{447}
\expandafter\gdef\csname odunum@ci@negative_subtypes.ftr.qwen3_omni_think.all\endcsname{[0.707, 0.787]}
\expandafter\gdef\csname odunum@val@negative_subtypes.ftr.qwen3_omni_think.invalidation_reason.demand_incomplete\endcsname{0.590}
\expandafter\gdef\csname odunum@n@negative_subtypes.ftr.qwen3_omni_think.invalidation_reason.demand_incomplete\endcsname{78}
\expandafter\gdef\csname odunum@ci@negative_subtypes.ftr.qwen3_omni_think.invalidation_reason.demand_incomplete\endcsname{[0.479, 0.692]}
\expandafter\gdef\csname odunum@val@negative_subtypes.ftr.qwen3_omni_think.invalidation_reason.function_not_directed\endcsname{0.706}
\expandafter\gdef\csname odunum@n@negative_subtypes.ftr.qwen3_omni_think.invalidation_reason.function_not_directed\endcsname{85}
\expandafter\gdef\csname odunum@ci@negative_subtypes.ftr.qwen3_omni_think.invalidation_reason.function_not_directed\endcsname{[0.602, 0.792]}
\expandafter\gdef\csname odunum@val@negative_subtypes.ftr.qwen3_omni_think.invalidation_reason.intent_not_real\endcsname{0.667}
\expandafter\gdef\csname odunum@n@negative_subtypes.ftr.qwen3_omni_think.invalidation_reason.intent_not_real\endcsname{84}
\expandafter\gdef\csname odunum@ci@negative_subtypes.ftr.qwen3_omni_think.invalidation_reason.intent_not_real\endcsname{[0.561, 0.758]}
\expandafter\gdef\csname odunum@val@negative_subtypes.ftr.qwen3_omni_think.invalidation_reason.source_mismatch\endcsname{0.827}
\expandafter\gdef\csname odunum@n@negative_subtypes.ftr.qwen3_omni_think.invalidation_reason.source_mismatch\endcsname{81}
\expandafter\gdef\csname odunum@ci@negative_subtypes.ftr.qwen3_omni_think.invalidation_reason.source_mismatch\endcsname{[0.731, 0.894]}
\expandafter\gdef\csname odunum@val@negative_subtypes.ftr.qwen3_omni_think.invalidation_reason.target_mismatch\endcsname{0.841}
\expandafter\gdef\csname odunum@n@negative_subtypes.ftr.qwen3_omni_think.invalidation_reason.target_mismatch\endcsname{44}
\expandafter\gdef\csname odunum@ci@negative_subtypes.ftr.qwen3_omni_think.invalidation_reason.target_mismatch\endcsname{[0.706, 0.921]}
\expandafter\gdef\csname odunum@val@negative_subtypes.ftr.qwen3_omni_think.invalidation_reason.unlabelled_recorded\endcsname{0.920}
\expandafter\gdef\csname odunum@n@negative_subtypes.ftr.qwen3_omni_think.invalidation_reason.unlabelled_recorded\endcsname{75}
\expandafter\gdef\csname odunum@ci@negative_subtypes.ftr.qwen3_omni_think.invalidation_reason.unlabelled_recorded\endcsname{[0.836, 0.963]}
\expandafter\gdef\csname odunum@val@negative_subtypes.ftr.qwen3_omni_think.signal_type.contextual_signal\endcsname{0.721}
\expandafter\gdef\csname odunum@n@negative_subtypes.ftr.qwen3_omni_think.signal_type.contextual_signal\endcsname{86}
\expandafter\gdef\csname odunum@ci@negative_subtypes.ftr.qwen3_omni_think.signal_type.contextual_signal\endcsname{[0.618, 0.805]}
\expandafter\gdef\csname odunum@val@negative_subtypes.ftr.qwen3_omni_think.signal_type.interactive_signal\endcsname{0.649}
\expandafter\gdef\csname odunum@n@negative_subtypes.ftr.qwen3_omni_think.signal_type.interactive_signal\endcsname{37}
\expandafter\gdef\csname odunum@ci@negative_subtypes.ftr.qwen3_omni_think.signal_type.interactive_signal\endcsname{[0.488, 0.782]}
\expandafter\gdef\csname odunum@val@negative_subtypes.ftr.qwen3_omni_think.signal_type.linguistic_signal\endcsname{0.899}
\expandafter\gdef\csname odunum@n@negative_subtypes.ftr.qwen3_omni_think.signal_type.linguistic_signal\endcsname{69}
\expandafter\gdef\csname odunum@ci@negative_subtypes.ftr.qwen3_omni_think.signal_type.linguistic_signal\endcsname{[0.805, 0.950]}
\expandafter\gdef\csname odunum@val@negative_subtypes.ftr.qwen3_omni_think.signal_type.prosodic_signal\endcsname{0.687}
\expandafter\gdef\csname odunum@n@negative_subtypes.ftr.qwen3_omni_think.signal_type.prosodic_signal\endcsname{67}
\expandafter\gdef\csname odunum@ci@negative_subtypes.ftr.qwen3_omni_think.signal_type.prosodic_signal\endcsname{[0.568, 0.785]}
\expandafter\gdef\csname odunum@val@negative_subtypes.ftr.qwen3_omni_think.signal_type.semantic_signal\endcsname{0.637}
\expandafter\gdef\csname odunum@n@negative_subtypes.ftr.qwen3_omni_think.signal_type.semantic_signal\endcsname{113}
\expandafter\gdef\csname odunum@ci@negative_subtypes.ftr.qwen3_omni_think.signal_type.semantic_signal\endcsname{[0.545, 0.720]}
\expandafter\gdef\csname odunum@val@negative_subtypes.ftr.qwen3_omni_think.signal_type.unlabelled_recorded\endcsname{0.920}
\expandafter\gdef\csname odunum@n@negative_subtypes.ftr.qwen3_omni_think.signal_type.unlabelled_recorded\endcsname{75}
\expandafter\gdef\csname odunum@ci@negative_subtypes.ftr.qwen3_omni_think.signal_type.unlabelled_recorded\endcsname{[0.836, 0.963]}
\expandafter\gdef\csname odunum@val@negative_subtypes.ftr.qwen_plus.all\endcsname{0.671}
\expandafter\gdef\csname odunum@n@negative_subtypes.ftr.qwen_plus.all\endcsname{447}
\expandafter\gdef\csname odunum@ci@negative_subtypes.ftr.qwen_plus.all\endcsname{[0.626, 0.713]}
\expandafter\gdef\csname odunum@val@negative_subtypes.ftr.qwen_plus.invalidation_reason.demand_incomplete\endcsname{0.641}
\expandafter\gdef\csname odunum@n@negative_subtypes.ftr.qwen_plus.invalidation_reason.demand_incomplete\endcsname{78}
\expandafter\gdef\csname odunum@ci@negative_subtypes.ftr.qwen_plus.invalidation_reason.demand_incomplete\endcsname{[0.530, 0.739]}
\expandafter\gdef\csname odunum@val@negative_subtypes.ftr.qwen_plus.invalidation_reason.function_not_directed\endcsname{0.588}
\expandafter\gdef\csname odunum@n@negative_subtypes.ftr.qwen_plus.invalidation_reason.function_not_directed\endcsname{85}
\expandafter\gdef\csname odunum@ci@negative_subtypes.ftr.qwen_plus.invalidation_reason.function_not_directed\endcsname{[0.482, 0.687]}
\expandafter\gdef\csname odunum@val@negative_subtypes.ftr.qwen_plus.invalidation_reason.intent_not_real\endcsname{0.536}
\expandafter\gdef\csname odunum@n@negative_subtypes.ftr.qwen_plus.invalidation_reason.intent_not_real\endcsname{84}
\expandafter\gdef\csname odunum@ci@negative_subtypes.ftr.qwen_plus.invalidation_reason.intent_not_real\endcsname{[0.430, 0.638]}
\expandafter\gdef\csname odunum@val@negative_subtypes.ftr.qwen_plus.invalidation_reason.source_mismatch\endcsname{0.654}
\expandafter\gdef\csname odunum@n@negative_subtypes.ftr.qwen_plus.invalidation_reason.source_mismatch\endcsname{81}
\expandafter\gdef\csname odunum@ci@negative_subtypes.ftr.qwen_plus.invalidation_reason.source_mismatch\endcsname{[0.546, 0.749]}
\expandafter\gdef\csname odunum@val@negative_subtypes.ftr.qwen_plus.invalidation_reason.target_mismatch\endcsname{0.841}
\expandafter\gdef\csname odunum@n@negative_subtypes.ftr.qwen_plus.invalidation_reason.target_mismatch\endcsname{44}
\expandafter\gdef\csname odunum@ci@negative_subtypes.ftr.qwen_plus.invalidation_reason.target_mismatch\endcsname{[0.706, 0.921]}
\expandafter\gdef\csname odunum@val@negative_subtypes.ftr.qwen_plus.invalidation_reason.unlabelled_recorded\endcsname{0.867}
\expandafter\gdef\csname odunum@n@negative_subtypes.ftr.qwen_plus.invalidation_reason.unlabelled_recorded\endcsname{75}
\expandafter\gdef\csname odunum@ci@negative_subtypes.ftr.qwen_plus.invalidation_reason.unlabelled_recorded\endcsname{[0.772, 0.926]}
\expandafter\gdef\csname odunum@val@negative_subtypes.ftr.qwen_plus.signal_type.contextual_signal\endcsname{0.674}
\expandafter\gdef\csname odunum@n@negative_subtypes.ftr.qwen_plus.signal_type.contextual_signal\endcsname{86}
\expandafter\gdef\csname odunum@ci@negative_subtypes.ftr.qwen_plus.signal_type.contextual_signal\endcsname{[0.570, 0.764]}
\expandafter\gdef\csname odunum@val@negative_subtypes.ftr.qwen_plus.signal_type.interactive_signal\endcsname{0.676}
\expandafter\gdef\csname odunum@n@negative_subtypes.ftr.qwen_plus.signal_type.interactive_signal\endcsname{37}
\expandafter\gdef\csname odunum@ci@negative_subtypes.ftr.qwen_plus.signal_type.interactive_signal\endcsname{[0.515, 0.804]}
\expandafter\gdef\csname odunum@val@negative_subtypes.ftr.qwen_plus.signal_type.linguistic_signal\endcsname{0.739}
\expandafter\gdef\csname odunum@n@negative_subtypes.ftr.qwen_plus.signal_type.linguistic_signal\endcsname{69}
\expandafter\gdef\csname odunum@ci@negative_subtypes.ftr.qwen_plus.signal_type.linguistic_signal\endcsname{[0.625, 0.828]}
\expandafter\gdef\csname odunum@val@negative_subtypes.ftr.qwen_plus.signal_type.prosodic_signal\endcsname{0.537}
\expandafter\gdef\csname odunum@n@negative_subtypes.ftr.qwen_plus.signal_type.prosodic_signal\endcsname{67}
\expandafter\gdef\csname odunum@ci@negative_subtypes.ftr.qwen_plus.signal_type.prosodic_signal\endcsname{[0.419, 0.651]}
\expandafter\gdef\csname odunum@val@negative_subtypes.ftr.qwen_plus.signal_type.semantic_signal\endcsname{0.575}
\expandafter\gdef\csname odunum@n@negative_subtypes.ftr.qwen_plus.signal_type.semantic_signal\endcsname{113}
\expandafter\gdef\csname odunum@ci@negative_subtypes.ftr.qwen_plus.signal_type.semantic_signal\endcsname{[0.483, 0.662]}
\expandafter\gdef\csname odunum@val@negative_subtypes.ftr.qwen_plus.signal_type.unlabelled_recorded\endcsname{0.867}
\expandafter\gdef\csname odunum@n@negative_subtypes.ftr.qwen_plus.signal_type.unlabelled_recorded\endcsname{75}
\expandafter\gdef\csname odunum@ci@negative_subtypes.ftr.qwen_plus.signal_type.unlabelled_recorded\endcsname{[0.772, 0.926]}
\expandafter\gdef\csname odunum@val@negative_subtypes.ftr.salmonn2_7b.all\endcsname{0.921}
\expandafter\gdef\csname odunum@n@negative_subtypes.ftr.salmonn2_7b.all\endcsname{444}
\expandafter\gdef\csname odunum@ci@negative_subtypes.ftr.salmonn2_7b.all\endcsname{[0.892, 0.943]}
\expandafter\gdef\csname odunum@val@negative_subtypes.ftr.salmonn2_7b.invalidation_reason.demand_incomplete\endcsname{0.962}
\expandafter\gdef\csname odunum@n@negative_subtypes.ftr.salmonn2_7b.invalidation_reason.demand_incomplete\endcsname{78}
\expandafter\gdef\csname odunum@ci@negative_subtypes.ftr.salmonn2_7b.invalidation_reason.demand_incomplete\endcsname{[0.893, 0.987]}
\expandafter\gdef\csname odunum@val@negative_subtypes.ftr.salmonn2_7b.invalidation_reason.function_not_directed\endcsname{0.894}
\expandafter\gdef\csname odunum@n@negative_subtypes.ftr.salmonn2_7b.invalidation_reason.function_not_directed\endcsname{85}
\expandafter\gdef\csname odunum@ci@negative_subtypes.ftr.salmonn2_7b.invalidation_reason.function_not_directed\endcsname{[0.811, 0.943]}
\expandafter\gdef\csname odunum@val@negative_subtypes.ftr.salmonn2_7b.invalidation_reason.intent_not_real\endcsname{0.915}
\expandafter\gdef\csname odunum@n@negative_subtypes.ftr.salmonn2_7b.invalidation_reason.intent_not_real\endcsname{82}
\expandafter\gdef\csname odunum@ci@negative_subtypes.ftr.salmonn2_7b.invalidation_reason.intent_not_real\endcsname{[0.834, 0.958]}
\expandafter\gdef\csname odunum@val@negative_subtypes.ftr.salmonn2_7b.invalidation_reason.source_mismatch\endcsname{0.914}
\expandafter\gdef\csname odunum@n@negative_subtypes.ftr.salmonn2_7b.invalidation_reason.source_mismatch\endcsname{81}
\expandafter\gdef\csname odunum@ci@negative_subtypes.ftr.salmonn2_7b.invalidation_reason.source_mismatch\endcsname{[0.832, 0.958]}
\expandafter\gdef\csname odunum@val@negative_subtypes.ftr.salmonn2_7b.invalidation_reason.target_mismatch\endcsname{0.886}
\expandafter\gdef\csname odunum@n@negative_subtypes.ftr.salmonn2_7b.invalidation_reason.target_mismatch\endcsname{44}
\expandafter\gdef\csname odunum@ci@negative_subtypes.ftr.salmonn2_7b.invalidation_reason.target_mismatch\endcsname{[0.760, 0.950]}
\expandafter\gdef\csname odunum@val@negative_subtypes.ftr.salmonn2_7b.invalidation_reason.unlabelled_recorded\endcsname{0.946}
\expandafter\gdef\csname odunum@n@negative_subtypes.ftr.salmonn2_7b.invalidation_reason.unlabelled_recorded\endcsname{74}
\expandafter\gdef\csname odunum@ci@negative_subtypes.ftr.salmonn2_7b.invalidation_reason.unlabelled_recorded\endcsname{[0.869, 0.979]}
\expandafter\gdef\csname odunum@val@negative_subtypes.ftr.salmonn2_7b.signal_type.contextual_signal\endcsname{0.930}
\expandafter\gdef\csname odunum@n@negative_subtypes.ftr.salmonn2_7b.signal_type.contextual_signal\endcsname{86}
\expandafter\gdef\csname odunum@ci@negative_subtypes.ftr.salmonn2_7b.signal_type.contextual_signal\endcsname{[0.856, 0.968]}
\expandafter\gdef\csname odunum@val@negative_subtypes.ftr.salmonn2_7b.signal_type.interactive_signal\endcsname{0.838}
\expandafter\gdef\csname odunum@n@negative_subtypes.ftr.salmonn2_7b.signal_type.interactive_signal\endcsname{37}
\expandafter\gdef\csname odunum@ci@negative_subtypes.ftr.salmonn2_7b.signal_type.interactive_signal\endcsname{[0.689, 0.923]}
\expandafter\gdef\csname odunum@val@negative_subtypes.ftr.salmonn2_7b.signal_type.linguistic_signal\endcsname{0.971}
\expandafter\gdef\csname odunum@n@negative_subtypes.ftr.salmonn2_7b.signal_type.linguistic_signal\endcsname{69}
\expandafter\gdef\csname odunum@ci@negative_subtypes.ftr.salmonn2_7b.signal_type.linguistic_signal\endcsname{[0.900, 0.992]}
\expandafter\gdef\csname odunum@val@negative_subtypes.ftr.salmonn2_7b.signal_type.prosodic_signal\endcsname{0.894}
\expandafter\gdef\csname odunum@n@negative_subtypes.ftr.salmonn2_7b.signal_type.prosodic_signal\endcsname{66}
\expandafter\gdef\csname odunum@ci@negative_subtypes.ftr.salmonn2_7b.signal_type.prosodic_signal\endcsname{[0.797, 0.948]}
\expandafter\gdef\csname odunum@val@negative_subtypes.ftr.salmonn2_7b.signal_type.semantic_signal\endcsname{0.911}
\expandafter\gdef\csname odunum@n@negative_subtypes.ftr.salmonn2_7b.signal_type.semantic_signal\endcsname{112}
\expandafter\gdef\csname odunum@ci@negative_subtypes.ftr.salmonn2_7b.signal_type.semantic_signal\endcsname{[0.843, 0.951]}
\expandafter\gdef\csname odunum@val@negative_subtypes.ftr.salmonn2_7b.signal_type.unlabelled_recorded\endcsname{0.946}
\expandafter\gdef\csname odunum@n@negative_subtypes.ftr.salmonn2_7b.signal_type.unlabelled_recorded\endcsname{74}
\expandafter\gdef\csname odunum@ci@negative_subtypes.ftr.salmonn2_7b.signal_type.unlabelled_recorded\endcsname{[0.869, 0.979]}
\expandafter\gdef\csname odunum@val@negative_subtypes.ftr.seed.all\endcsname{0.817}
\expandafter\gdef\csname odunum@n@negative_subtypes.ftr.seed.all\endcsname{447}
\expandafter\gdef\csname odunum@ci@negative_subtypes.ftr.seed.all\endcsname{[0.778, 0.850]}
\expandafter\gdef\csname odunum@val@negative_subtypes.ftr.seed.invalidation_reason.demand_incomplete\endcsname{0.769}
\expandafter\gdef\csname odunum@n@negative_subtypes.ftr.seed.invalidation_reason.demand_incomplete\endcsname{78}
\expandafter\gdef\csname odunum@ci@negative_subtypes.ftr.seed.invalidation_reason.demand_incomplete\endcsname{[0.664, 0.849]}
\expandafter\gdef\csname odunum@val@negative_subtypes.ftr.seed.invalidation_reason.function_not_directed\endcsname{0.788}
\expandafter\gdef\csname odunum@n@negative_subtypes.ftr.seed.invalidation_reason.function_not_directed\endcsname{85}
\expandafter\gdef\csname odunum@ci@negative_subtypes.ftr.seed.invalidation_reason.function_not_directed\endcsname{[0.690, 0.862]}
\expandafter\gdef\csname odunum@val@negative_subtypes.ftr.seed.invalidation_reason.intent_not_real\endcsname{0.809}
\expandafter\gdef\csname odunum@n@negative_subtypes.ftr.seed.invalidation_reason.intent_not_real\endcsname{84}
\expandafter\gdef\csname odunum@ci@negative_subtypes.ftr.seed.invalidation_reason.intent_not_real\endcsname{[0.713, 0.879]}
\expandafter\gdef\csname odunum@val@negative_subtypes.ftr.seed.invalidation_reason.source_mismatch\endcsname{0.864}
\expandafter\gdef\csname odunum@n@negative_subtypes.ftr.seed.invalidation_reason.source_mismatch\endcsname{81}
\expandafter\gdef\csname odunum@ci@negative_subtypes.ftr.seed.invalidation_reason.source_mismatch\endcsname{[0.773, 0.922]}
\expandafter\gdef\csname odunum@val@negative_subtypes.ftr.seed.invalidation_reason.target_mismatch\endcsname{0.841}
\expandafter\gdef\csname odunum@n@negative_subtypes.ftr.seed.invalidation_reason.target_mismatch\endcsname{44}
\expandafter\gdef\csname odunum@ci@negative_subtypes.ftr.seed.invalidation_reason.target_mismatch\endcsname{[0.706, 0.921]}
\expandafter\gdef\csname odunum@val@negative_subtypes.ftr.seed.invalidation_reason.unlabelled_recorded\endcsname{0.840}
\expandafter\gdef\csname odunum@n@negative_subtypes.ftr.seed.invalidation_reason.unlabelled_recorded\endcsname{75}
\expandafter\gdef\csname odunum@ci@negative_subtypes.ftr.seed.invalidation_reason.unlabelled_recorded\endcsname{[0.741, 0.906]}
\expandafter\gdef\csname odunum@val@negative_subtypes.ftr.seed.signal_type.contextual_signal\endcsname{0.744}
\expandafter\gdef\csname odunum@n@negative_subtypes.ftr.seed.signal_type.contextual_signal\endcsname{86}
\expandafter\gdef\csname odunum@ci@negative_subtypes.ftr.seed.signal_type.contextual_signal\endcsname{[0.643, 0.825]}
\expandafter\gdef\csname odunum@val@negative_subtypes.ftr.seed.signal_type.interactive_signal\endcsname{0.784}
\expandafter\gdef\csname odunum@n@negative_subtypes.ftr.seed.signal_type.interactive_signal\endcsname{37}
\expandafter\gdef\csname odunum@ci@negative_subtypes.ftr.seed.signal_type.interactive_signal\endcsname{[0.628, 0.886]}
\expandafter\gdef\csname odunum@val@negative_subtypes.ftr.seed.signal_type.linguistic_signal\endcsname{0.913}
\expandafter\gdef\csname odunum@n@negative_subtypes.ftr.seed.signal_type.linguistic_signal\endcsname{69}
\expandafter\gdef\csname odunum@ci@negative_subtypes.ftr.seed.signal_type.linguistic_signal\endcsname{[0.823, 0.960]}
\expandafter\gdef\csname odunum@val@negative_subtypes.ftr.seed.signal_type.prosodic_signal\endcsname{0.821}
\expandafter\gdef\csname odunum@n@negative_subtypes.ftr.seed.signal_type.prosodic_signal\endcsname{67}
\expandafter\gdef\csname odunum@ci@negative_subtypes.ftr.seed.signal_type.prosodic_signal\endcsname{[0.713, 0.894]}
\expandafter\gdef\csname odunum@val@negative_subtypes.ftr.seed.signal_type.semantic_signal\endcsname{0.805}
\expandafter\gdef\csname odunum@n@negative_subtypes.ftr.seed.signal_type.semantic_signal\endcsname{113}
\expandafter\gdef\csname odunum@ci@negative_subtypes.ftr.seed.signal_type.semantic_signal\endcsname{[0.723, 0.868]}
\expandafter\gdef\csname odunum@val@negative_subtypes.ftr.seed.signal_type.unlabelled_recorded\endcsname{0.840}
\expandafter\gdef\csname odunum@n@negative_subtypes.ftr.seed.signal_type.unlabelled_recorded\endcsname{75}
\expandafter\gdef\csname odunum@ci@negative_subtypes.ftr.seed.signal_type.unlabelled_recorded\endcsname{[0.741, 0.906]}
\expandafter\gdef\csname odunum@val@negative_subtypes.panel_mean_ftr.invalidation_reason.demand_incomplete\endcsname{0.661}
\expandafter\gdef\csname odunum@n@negative_subtypes.panel_mean_ftr.invalidation_reason.demand_incomplete\endcsname{13}
\expandafter\gdef\csname odunum@ci@negative_subtypes.panel_mean_ftr.invalidation_reason.demand_incomplete\endcsname{[0.533, 0.780]}
\expandafter\gdef\csname odunum@val@negative_subtypes.panel_mean_ftr.invalidation_reason.function_not_directed\endcsname{0.631}
\expandafter\gdef\csname odunum@n@negative_subtypes.panel_mean_ftr.invalidation_reason.function_not_directed\endcsname{13}
\expandafter\gdef\csname odunum@ci@negative_subtypes.panel_mean_ftr.invalidation_reason.function_not_directed\endcsname{[0.506, 0.743]}
\expandafter\gdef\csname odunum@val@negative_subtypes.panel_mean_ftr.invalidation_reason.intent_not_real\endcsname{0.640}
\expandafter\gdef\csname odunum@n@negative_subtypes.panel_mean_ftr.invalidation_reason.intent_not_real\endcsname{13}
\expandafter\gdef\csname odunum@ci@negative_subtypes.panel_mean_ftr.invalidation_reason.intent_not_real\endcsname{[0.510, 0.754]}
\expandafter\gdef\csname odunum@val@negative_subtypes.panel_mean_ftr.invalidation_reason.source_mismatch\endcsname{0.736}
\expandafter\gdef\csname odunum@n@negative_subtypes.panel_mean_ftr.invalidation_reason.source_mismatch\endcsname{13}
\expandafter\gdef\csname odunum@ci@negative_subtypes.panel_mean_ftr.invalidation_reason.source_mismatch\endcsname{[0.658, 0.810]}
\expandafter\gdef\csname odunum@val@negative_subtypes.panel_mean_ftr.invalidation_reason.target_mismatch\endcsname{0.759}
\expandafter\gdef\csname odunum@n@negative_subtypes.panel_mean_ftr.invalidation_reason.target_mismatch\endcsname{13}
\expandafter\gdef\csname odunum@ci@negative_subtypes.panel_mean_ftr.invalidation_reason.target_mismatch\endcsname{[0.636, 0.858]}
\expandafter\gdef\csname odunum@val@negative_subtypes.panel_mean_ftr.invalidation_reason.unlabelled_recorded\endcsname{0.752}
\expandafter\gdef\csname odunum@n@negative_subtypes.panel_mean_ftr.invalidation_reason.unlabelled_recorded\endcsname{13}
\expandafter\gdef\csname odunum@ci@negative_subtypes.panel_mean_ftr.invalidation_reason.unlabelled_recorded\endcsname{[0.607, 0.873]}
\expandafter\gdef\csname odunum@val@negative_subtypes.panel_mean_ftr.signal_type.contextual_signal\endcsname{0.680}
\expandafter\gdef\csname odunum@n@negative_subtypes.panel_mean_ftr.signal_type.contextual_signal\endcsname{13}
\expandafter\gdef\csname odunum@ci@negative_subtypes.panel_mean_ftr.signal_type.contextual_signal\endcsname{[0.566, 0.784]}
\expandafter\gdef\csname odunum@val@negative_subtypes.panel_mean_ftr.signal_type.interactive_signal\endcsname{0.664}
\expandafter\gdef\csname odunum@n@negative_subtypes.panel_mean_ftr.signal_type.interactive_signal\endcsname{13}
\expandafter\gdef\csname odunum@ci@negative_subtypes.panel_mean_ftr.signal_type.interactive_signal\endcsname{[0.567, 0.753]}
\expandafter\gdef\csname odunum@val@negative_subtypes.panel_mean_ftr.signal_type.linguistic_signal\endcsname{0.793}
\expandafter\gdef\csname odunum@n@negative_subtypes.panel_mean_ftr.signal_type.linguistic_signal\endcsname{13}
\expandafter\gdef\csname odunum@ci@negative_subtypes.panel_mean_ftr.signal_type.linguistic_signal\endcsname{[0.677, 0.887]}
\expandafter\gdef\csname odunum@val@negative_subtypes.panel_mean_ftr.signal_type.prosodic_signal\endcsname{0.637}
\expandafter\gdef\csname odunum@n@negative_subtypes.panel_mean_ftr.signal_type.prosodic_signal\endcsname{13}
\expandafter\gdef\csname odunum@ci@negative_subtypes.panel_mean_ftr.signal_type.prosodic_signal\endcsname{[0.508, 0.747]}
\expandafter\gdef\csname odunum@val@negative_subtypes.panel_mean_ftr.signal_type.semantic_signal\endcsname{0.633}
\expandafter\gdef\csname odunum@n@negative_subtypes.panel_mean_ftr.signal_type.semantic_signal\endcsname{13}
\expandafter\gdef\csname odunum@ci@negative_subtypes.panel_mean_ftr.signal_type.semantic_signal\endcsname{[0.518, 0.741]}
\expandafter\gdef\csname odunum@val@negative_subtypes.panel_mean_ftr.signal_type.unlabelled_recorded\endcsname{0.752}
\expandafter\gdef\csname odunum@n@negative_subtypes.panel_mean_ftr.signal_type.unlabelled_recorded\endcsname{13}
\expandafter\gdef\csname odunum@ci@negative_subtypes.panel_mean_ftr.signal_type.unlabelled_recorded\endcsname{[0.607, 0.873]}
\expandafter\gdef\csname odunum@val@negative_subtypes.panel_sign_above_own_ftr.invalidation_reason.demand_incomplete\endcsname{6}
\expandafter\gdef\csname odunum@n@negative_subtypes.panel_sign_above_own_ftr.invalidation_reason.demand_incomplete\endcsname{13}
\expandafter\gdef\csname odunum@val@negative_subtypes.panel_sign_above_own_ftr.invalidation_reason.function_not_directed\endcsname{1}
\expandafter\gdef\csname odunum@n@negative_subtypes.panel_sign_above_own_ftr.invalidation_reason.function_not_directed\endcsname{13}
\expandafter\gdef\csname odunum@val@negative_subtypes.panel_sign_above_own_ftr.invalidation_reason.intent_not_real\endcsname{2}
\expandafter\gdef\csname odunum@n@negative_subtypes.panel_sign_above_own_ftr.invalidation_reason.intent_not_real\endcsname{13}
\expandafter\gdef\csname odunum@val@negative_subtypes.panel_sign_above_own_ftr.invalidation_reason.source_mismatch\endcsname{8}
\expandafter\gdef\csname odunum@n@negative_subtypes.panel_sign_above_own_ftr.invalidation_reason.source_mismatch\endcsname{13}
\expandafter\gdef\csname odunum@val@negative_subtypes.panel_sign_above_own_ftr.invalidation_reason.target_mismatch\endcsname{11}
\expandafter\gdef\csname odunum@n@negative_subtypes.panel_sign_above_own_ftr.invalidation_reason.target_mismatch\endcsname{13}
\expandafter\gdef\csname odunum@val@negative_subtypes.panel_sign_above_own_ftr.invalidation_reason.unlabelled_recorded\endcsname{10}
\expandafter\gdef\csname odunum@n@negative_subtypes.panel_sign_above_own_ftr.invalidation_reason.unlabelled_recorded\endcsname{13}
\expandafter\gdef\csname odunum@val@negative_subtypes.panel_sign_above_own_ftr.signal_type.contextual_signal\endcsname{5}
\expandafter\gdef\csname odunum@n@negative_subtypes.panel_sign_above_own_ftr.signal_type.contextual_signal\endcsname{13}
\expandafter\gdef\csname odunum@val@negative_subtypes.panel_sign_above_own_ftr.signal_type.interactive_signal\endcsname{5}
\expandafter\gdef\csname odunum@n@negative_subtypes.panel_sign_above_own_ftr.signal_type.interactive_signal\endcsname{13}
\expandafter\gdef\csname odunum@val@negative_subtypes.panel_sign_above_own_ftr.signal_type.linguistic_signal\endcsname{13}
\expandafter\gdef\csname odunum@n@negative_subtypes.panel_sign_above_own_ftr.signal_type.linguistic_signal\endcsname{13}
\expandafter\gdef\csname odunum@val@negative_subtypes.panel_sign_above_own_ftr.signal_type.prosodic_signal\endcsname{2}
\expandafter\gdef\csname odunum@n@negative_subtypes.panel_sign_above_own_ftr.signal_type.prosodic_signal\endcsname{13}
\expandafter\gdef\csname odunum@val@negative_subtypes.panel_sign_above_own_ftr.signal_type.semantic_signal\endcsname{0}
\expandafter\gdef\csname odunum@n@negative_subtypes.panel_sign_above_own_ftr.signal_type.semantic_signal\endcsname{13}
\expandafter\gdef\csname odunum@val@negative_subtypes.panel_sign_above_own_ftr.signal_type.unlabelled_recorded\endcsname{10}
\expandafter\gdef\csname odunum@n@negative_subtypes.panel_sign_above_own_ftr.signal_type.unlabelled_recorded\endcsname{13}
\expandafter\gdef\csname odunum@val@negative_subtypes.pop.invalidation_reason.demand_incomplete.n\endcsname{78}
\expandafter\gdef\csname odunum@n@negative_subtypes.pop.invalidation_reason.demand_incomplete.n\endcsname{447}
\expandafter\gdef\csname odunum@val@negative_subtypes.pop.invalidation_reason.function_not_directed.n\endcsname{85}
\expandafter\gdef\csname odunum@n@negative_subtypes.pop.invalidation_reason.function_not_directed.n\endcsname{447}
\expandafter\gdef\csname odunum@val@negative_subtypes.pop.invalidation_reason.intent_not_real.n\endcsname{84}
\expandafter\gdef\csname odunum@n@negative_subtypes.pop.invalidation_reason.intent_not_real.n\endcsname{447}
\expandafter\gdef\csname odunum@val@negative_subtypes.pop.invalidation_reason.source_mismatch.n\endcsname{81}
\expandafter\gdef\csname odunum@n@negative_subtypes.pop.invalidation_reason.source_mismatch.n\endcsname{447}
\expandafter\gdef\csname odunum@val@negative_subtypes.pop.invalidation_reason.target_mismatch.n\endcsname{44}
\expandafter\gdef\csname odunum@n@negative_subtypes.pop.invalidation_reason.target_mismatch.n\endcsname{447}
\expandafter\gdef\csname odunum@val@negative_subtypes.pop.invalidation_reason.unlabelled_recorded.n\endcsname{75}
\expandafter\gdef\csname odunum@n@negative_subtypes.pop.invalidation_reason.unlabelled_recorded.n\endcsname{447}
\expandafter\gdef\csname odunum@val@negative_subtypes.pop.negative.audio_only.n\endcsname{161}
\expandafter\gdef\csname odunum@n@negative_subtypes.pop.negative.audio_only.n\endcsname{447}
\expandafter\gdef\csname odunum@val@negative_subtypes.pop.negative.audio_visual.n\endcsname{286}
\expandafter\gdef\csname odunum@n@negative_subtypes.pop.negative.audio_visual.n\endcsname{447}
\expandafter\gdef\csname odunum@val@negative_subtypes.pop.negative.n\endcsname{447}
\expandafter\gdef\csname odunum@n@negative_subtypes.pop.negative.n\endcsname{2\,078}
\expandafter\gdef\csname odunum@val@negative_subtypes.pop.negative.unlabelled.n\endcsname{75}
\expandafter\gdef\csname odunum@n@negative_subtypes.pop.negative.unlabelled.n\endcsname{447}
\expandafter\gdef\csname odunum@val@negative_subtypes.pop.negative_type.median_group_size\endcsname{6}
\expandafter\gdef\csname odunum@n@negative_subtypes.pop.negative_type.median_group_size\endcsname{58}
\expandafter\gdef\csname odunum@val@negative_subtypes.pop.negative_type.n_distinct\endcsname{58}
\expandafter\gdef\csname odunum@n@negative_subtypes.pop.negative_type.n_distinct\endcsname{372}
\expandafter\gdef\csname odunum@val@negative_subtypes.pop.signal_type.contextual_signal.n\endcsname{86}
\expandafter\gdef\csname odunum@n@negative_subtypes.pop.signal_type.contextual_signal.n\endcsname{447}
\expandafter\gdef\csname odunum@val@negative_subtypes.pop.signal_type.interactive_signal.n\endcsname{37}
\expandafter\gdef\csname odunum@n@negative_subtypes.pop.signal_type.interactive_signal.n\endcsname{447}
\expandafter\gdef\csname odunum@val@negative_subtypes.pop.signal_type.linguistic_signal.n\endcsname{69}
\expandafter\gdef\csname odunum@n@negative_subtypes.pop.signal_type.linguistic_signal.n\endcsname{447}
\expandafter\gdef\csname odunum@val@negative_subtypes.pop.signal_type.prosodic_signal.n\endcsname{67}
\expandafter\gdef\csname odunum@n@negative_subtypes.pop.signal_type.prosodic_signal.n\endcsname{447}
\expandafter\gdef\csname odunum@val@negative_subtypes.pop.signal_type.semantic_signal.n\endcsname{113}
\expandafter\gdef\csname odunum@n@negative_subtypes.pop.signal_type.semantic_signal.n\endcsname{447}
\expandafter\gdef\csname odunum@val@negative_subtypes.pop.signal_type.unlabelled_recorded.n\endcsname{75}
\expandafter\gdef\csname odunum@n@negative_subtypes.pop.signal_type.unlabelled_recorded.n\endcsname{447}
\expandafter\gdef\csname odunum@val@negative_subtypes.spread.cascade_asr.invalidation_reason\endcsname{0.236}
\expandafter\gdef\csname odunum@n@negative_subtypes.spread.cascade_asr.invalidation_reason\endcsname{6}
\expandafter\gdef\csname odunum@val@negative_subtypes.spread.cascade_asr.signal_type\endcsname{0.295}
\expandafter\gdef\csname odunum@n@negative_subtypes.spread.cascade_asr.signal_type\endcsname{6}
\expandafter\gdef\csname odunum@val@negative_subtypes.spread.gemini.invalidation_reason\endcsname{0.235}
\expandafter\gdef\csname odunum@n@negative_subtypes.spread.gemini.invalidation_reason\endcsname{6}
\expandafter\gdef\csname odunum@val@negative_subtypes.spread.gemini.signal_type\endcsname{0.132}
\expandafter\gdef\csname odunum@n@negative_subtypes.spread.gemini.signal_type\endcsname{6}
\expandafter\gdef\csname odunum@val@negative_subtypes.spread.gemini35_flash_lite.invalidation_reason\endcsname{0.358}
\expandafter\gdef\csname odunum@n@negative_subtypes.spread.gemini35_flash_lite.invalidation_reason\endcsname{6}
\expandafter\gdef\csname odunum@val@negative_subtypes.spread.gemini35_flash_lite.signal_type\endcsname{0.352}
\expandafter\gdef\csname odunum@n@negative_subtypes.spread.gemini35_flash_lite.signal_type\endcsname{6}
\expandafter\gdef\csname odunum@val@negative_subtypes.spread.gemini37_flash.invalidation_reason\endcsname{0.360}
\expandafter\gdef\csname odunum@n@negative_subtypes.spread.gemini37_flash.invalidation_reason\endcsname{6}
\expandafter\gdef\csname odunum@val@negative_subtypes.spread.gemini37_flash.signal_type\endcsname{0.180}
\expandafter\gdef\csname odunum@n@negative_subtypes.spread.gemini37_flash.signal_type\endcsname{6}
\expandafter\gdef\csname odunum@val@negative_subtypes.spread.ming.invalidation_reason\endcsname{0.294}
\expandafter\gdef\csname odunum@n@negative_subtypes.spread.ming.invalidation_reason\endcsname{6}
\expandafter\gdef\csname odunum@val@negative_subtypes.spread.ming.signal_type\endcsname{0.208}
\expandafter\gdef\csname odunum@n@negative_subtypes.spread.ming.signal_type\endcsname{6}
\expandafter\gdef\csname odunum@val@negative_subtypes.spread.minicpm_o.invalidation_reason\endcsname{0.221}
\expandafter\gdef\csname odunum@n@negative_subtypes.spread.minicpm_o.invalidation_reason\endcsname{6}
\expandafter\gdef\csname odunum@val@negative_subtypes.spread.minicpm_o.signal_type\endcsname{0.239}
\expandafter\gdef\csname odunum@n@negative_subtypes.spread.minicpm_o.signal_type\endcsname{6}
\expandafter\gdef\csname odunum@val@negative_subtypes.spread.nemotron.invalidation_reason\endcsname{0.178}
\expandafter\gdef\csname odunum@n@negative_subtypes.spread.nemotron.invalidation_reason\endcsname{6}
\expandafter\gdef\csname odunum@val@negative_subtypes.spread.nemotron.signal_type\endcsname{0.191}
\expandafter\gdef\csname odunum@n@negative_subtypes.spread.nemotron.signal_type\endcsname{6}
\expandafter\gdef\csname odunum@val@negative_subtypes.spread.qwen25_omni.invalidation_reason\endcsname{0.122}
\expandafter\gdef\csname odunum@n@negative_subtypes.spread.qwen25_omni.invalidation_reason\endcsname{6}
\expandafter\gdef\csname odunum@val@negative_subtypes.spread.qwen25_omni.signal_type\endcsname{0.252}
\expandafter\gdef\csname odunum@n@negative_subtypes.spread.qwen25_omni.signal_type\endcsname{6}
\expandafter\gdef\csname odunum@val@negative_subtypes.spread.qwen3_omni_instruct.invalidation_reason\endcsname{0.175}
\expandafter\gdef\csname odunum@n@negative_subtypes.spread.qwen3_omni_instruct.invalidation_reason\endcsname{6}
\expandafter\gdef\csname odunum@val@negative_subtypes.spread.qwen3_omni_instruct.signal_type\endcsname{0.196}
\expandafter\gdef\csname odunum@n@negative_subtypes.spread.qwen3_omni_instruct.signal_type\endcsname{6}
\expandafter\gdef\csname odunum@val@negative_subtypes.spread.qwen3_omni_think.invalidation_reason\endcsname{0.330}
\expandafter\gdef\csname odunum@n@negative_subtypes.spread.qwen3_omni_think.invalidation_reason\endcsname{6}
\expandafter\gdef\csname odunum@val@negative_subtypes.spread.qwen3_omni_think.signal_type\endcsname{0.283}
\expandafter\gdef\csname odunum@n@negative_subtypes.spread.qwen3_omni_think.signal_type\endcsname{6}
\expandafter\gdef\csname odunum@val@negative_subtypes.spread.qwen_plus.invalidation_reason\endcsname{0.331}
\expandafter\gdef\csname odunum@n@negative_subtypes.spread.qwen_plus.invalidation_reason\endcsname{6}
\expandafter\gdef\csname odunum@val@negative_subtypes.spread.qwen_plus.signal_type\endcsname{0.329}
\expandafter\gdef\csname odunum@n@negative_subtypes.spread.qwen_plus.signal_type\endcsname{6}
\expandafter\gdef\csname odunum@val@negative_subtypes.spread.salmonn2_7b.invalidation_reason\endcsname{0.075}
\expandafter\gdef\csname odunum@n@negative_subtypes.spread.salmonn2_7b.invalidation_reason\endcsname{6}
\expandafter\gdef\csname odunum@val@negative_subtypes.spread.salmonn2_7b.signal_type\endcsname{0.133}
\expandafter\gdef\csname odunum@n@negative_subtypes.spread.salmonn2_7b.signal_type\endcsname{6}
\expandafter\gdef\csname odunum@val@negative_subtypes.spread.seed.invalidation_reason\endcsname{0.095}
\expandafter\gdef\csname odunum@n@negative_subtypes.spread.seed.invalidation_reason\endcsname{6}
\expandafter\gdef\csname odunum@val@negative_subtypes.spread.seed.signal_type\endcsname{0.169}
\expandafter\gdef\csname odunum@n@negative_subtypes.spread.seed.signal_type\endcsname{6}
\expandafter\gdef\csname odunum@val@perception_probe.audit.doubao_seed_2_0_lite.targets\endcsname{449}
\expandafter\gdef\csname odunum@n@perception_probe.audit.doubao_seed_2_0_lite.targets\endcsname{450}
\expandafter\gdef\csname odunum@val@perception_probe.audit.doubao_seed_2_0_lite.unanswered\endcsname{24}
\expandafter\gdef\csname odunum@n@perception_probe.audit.doubao_seed_2_0_lite.unanswered\endcsname{449}
\expandafter\gdef\csname odunum@val@perception_probe.audit.doubao_seed_2_0_lite.unanswered_share.gen\endcsname{0.000}
\expandafter\gdef\csname odunum@n@perception_probe.audit.doubao_seed_2_0_lite.unanswered_share.gen\endcsname{362}
\expandafter\gdef\csname odunum@val@perception_probe.audit.doubao_seed_2_0_lite.unanswered_share.recorded\endcsname{0.276}
\expandafter\gdef\csname odunum@n@perception_probe.audit.doubao_seed_2_0_lite.unanswered_share.recorded\endcsname{87}
\expandafter\gdef\csname odunum@val@perception_probe.audit.doubao_seed_2_0_lite.unjudged\endcsname{24}
\expandafter\gdef\csname odunum@n@perception_probe.audit.doubao_seed_2_0_lite.unjudged\endcsname{449}
\expandafter\gdef\csname odunum@val@perception_probe.audit.gemini_3_1_pro.targets\endcsname{450}
\expandafter\gdef\csname odunum@n@perception_probe.audit.gemini_3_1_pro.targets\endcsname{450}
\expandafter\gdef\csname odunum@val@perception_probe.audit.gemini_3_1_pro.unanswered\endcsname{0}
\expandafter\gdef\csname odunum@n@perception_probe.audit.gemini_3_1_pro.unanswered\endcsname{450}
\expandafter\gdef\csname odunum@val@perception_probe.audit.gemini_3_1_pro.unanswered_share.gen\endcsname{0.000}
\expandafter\gdef\csname odunum@n@perception_probe.audit.gemini_3_1_pro.unanswered_share.gen\endcsname{382}
\expandafter\gdef\csname odunum@val@perception_probe.audit.gemini_3_1_pro.unanswered_share.recorded\endcsname{0.000}
\expandafter\gdef\csname odunum@n@perception_probe.audit.gemini_3_1_pro.unanswered_share.recorded\endcsname{68}
\expandafter\gdef\csname odunum@val@perception_probe.audit.gemini_3_1_pro.unjudged\endcsname{0}
\expandafter\gdef\csname odunum@n@perception_probe.audit.gemini_3_1_pro.unjudged\endcsname{450}
\expandafter\gdef\csname odunum@val@perception_probe.audit.nomedia_unanswered\endcsname{0}
\expandafter\gdef\csname odunum@n@perception_probe.audit.nomedia_unanswered\endcsname{1\,057}
\expandafter\gdef\csname odunum@val@perception_probe.audit.nomedia_unjudged\endcsname{0}
\expandafter\gdef\csname odunum@n@perception_probe.audit.nomedia_unjudged\endcsname{1\,057}
\expandafter\gdef\csname odunum@val@perception_probe.audit.qwen3_5_omni_plus.targets\endcsname{449}
\expandafter\gdef\csname odunum@n@perception_probe.audit.qwen3_5_omni_plus.targets\endcsname{450}
\expandafter\gdef\csname odunum@val@perception_probe.audit.qwen3_5_omni_plus.unanswered\endcsname{0}
\expandafter\gdef\csname odunum@n@perception_probe.audit.qwen3_5_omni_plus.unanswered\endcsname{449}
\expandafter\gdef\csname odunum@val@perception_probe.audit.qwen3_5_omni_plus.unanswered_share.gen\endcsname{0.000}
\expandafter\gdef\csname odunum@n@perception_probe.audit.qwen3_5_omni_plus.unanswered_share.gen\endcsname{386}
\expandafter\gdef\csname odunum@val@perception_probe.audit.qwen3_5_omni_plus.unanswered_share.recorded\endcsname{0.000}
\expandafter\gdef\csname odunum@n@perception_probe.audit.qwen3_5_omni_plus.unanswered_share.recorded\endcsname{63}
\expandafter\gdef\csname odunum@val@perception_probe.audit.qwen3_5_omni_plus.unjudged\endcsname{0}
\expandafter\gdef\csname odunum@n@perception_probe.audit.qwen3_5_omni_plus.unjudged\endcsname{449}
\expandafter\gdef\csname odunum@val@perception_probe.calib.doubao_seed_2_0_lite.audio\endcsname{0.750}
\expandafter\gdef\csname odunum@n@perception_probe.calib.doubao_seed_2_0_lite.audio\endcsname{36}
\expandafter\gdef\csname odunum@ci@perception_probe.calib.doubao_seed_2_0_lite.audio\endcsname{[0.611, 0.889]}
\expandafter\gdef\csname odunum@val@perception_probe.calib.doubao_seed_2_0_lite.history\endcsname{0.756}
\expandafter\gdef\csname odunum@n@perception_probe.calib.doubao_seed_2_0_lite.history\endcsname{41}
\expandafter\gdef\csname odunum@ci@perception_probe.calib.doubao_seed_2_0_lite.history\endcsname{[0.610, 0.878]}
\expandafter\gdef\csname odunum@val@perception_probe.calib.doubao_seed_2_0_lite.other_context\endcsname{0.833}
\expandafter\gdef\csname odunum@n@perception_probe.calib.doubao_seed_2_0_lite.other_context\endcsname{6}
\expandafter\gdef\csname odunum@ci@perception_probe.calib.doubao_seed_2_0_lite.other_context\endcsname{[0.400, 1.000]}
\expandafter\gdef\csname odunum@val@perception_probe.calib.doubao_seed_2_0_lite.overall\endcsname{0.780}
\expandafter\gdef\csname odunum@n@perception_probe.calib.doubao_seed_2_0_lite.overall\endcsname{141}
\expandafter\gdef\csname odunum@ci@perception_probe.calib.doubao_seed_2_0_lite.overall\endcsname{[0.710, 0.846]}
\expandafter\gdef\csname odunum@val@perception_probe.calib.doubao_seed_2_0_lite.visual\endcsname{0.810}
\expandafter\gdef\csname odunum@n@perception_probe.calib.doubao_seed_2_0_lite.visual\endcsname{58}
\expandafter\gdef\csname odunum@ci@perception_probe.calib.doubao_seed_2_0_lite.visual\endcsname{[0.707, 0.897]}
\expandafter\gdef\csname odunum@val@perception_probe.calib.gemini_3_1_pro.audio\endcsname{0.727}
\expandafter\gdef\csname odunum@n@perception_probe.calib.gemini_3_1_pro.audio\endcsname{44}
\expandafter\gdef\csname odunum@ci@perception_probe.calib.gemini_3_1_pro.audio\endcsname{[0.591, 0.864]}
\expandafter\gdef\csname odunum@val@perception_probe.calib.gemini_3_1_pro.history\endcsname{0.743}
\expandafter\gdef\csname odunum@n@perception_probe.calib.gemini_3_1_pro.history\endcsname{35}
\expandafter\gdef\csname odunum@ci@perception_probe.calib.gemini_3_1_pro.history\endcsname{[0.600, 0.886]}
\expandafter\gdef\csname odunum@val@perception_probe.calib.gemini_3_1_pro.other_context\endcsname{0.800}
\expandafter\gdef\csname odunum@n@perception_probe.calib.gemini_3_1_pro.other_context\endcsname{5}
\expandafter\gdef\csname odunum@ci@perception_probe.calib.gemini_3_1_pro.other_context\endcsname{[0.400, 1.000]}
\expandafter\gdef\csname odunum@val@perception_probe.calib.gemini_3_1_pro.overall\endcsname{0.733}
\expandafter\gdef\csname odunum@n@perception_probe.calib.gemini_3_1_pro.overall\endcsname{150}
\expandafter\gdef\csname odunum@ci@perception_probe.calib.gemini_3_1_pro.overall\endcsname{[0.662, 0.804]}
\expandafter\gdef\csname odunum@val@perception_probe.calib.gemini_3_1_pro.visual\endcsname{0.727}
\expandafter\gdef\csname odunum@n@perception_probe.calib.gemini_3_1_pro.visual\endcsname{66}
\expandafter\gdef\csname odunum@ci@perception_probe.calib.gemini_3_1_pro.visual\endcsname{[0.621, 0.833]}
\expandafter\gdef\csname odunum@val@perception_probe.calib.qwen3_5_omni_plus.audio\endcsname{0.775}
\expandafter\gdef\csname odunum@n@perception_probe.calib.qwen3_5_omni_plus.audio\endcsname{40}
\expandafter\gdef\csname odunum@ci@perception_probe.calib.qwen3_5_omni_plus.audio\endcsname{[0.650, 0.900]}
\expandafter\gdef\csname odunum@val@perception_probe.calib.qwen3_5_omni_plus.history\endcsname{0.558}
\expandafter\gdef\csname odunum@n@perception_probe.calib.qwen3_5_omni_plus.history\endcsname{43}
\expandafter\gdef\csname odunum@ci@perception_probe.calib.qwen3_5_omni_plus.history\endcsname{[0.405, 0.705]}
\expandafter\gdef\csname odunum@val@perception_probe.calib.qwen3_5_omni_plus.other_context\endcsname{0.667}
\expandafter\gdef\csname odunum@n@perception_probe.calib.qwen3_5_omni_plus.other_context\endcsname{9}
\expandafter\gdef\csname odunum@ci@perception_probe.calib.qwen3_5_omni_plus.other_context\endcsname{[0.333, 0.889]}
\expandafter\gdef\csname odunum@val@perception_probe.calib.qwen3_5_omni_plus.overall\endcsname{0.733}
\expandafter\gdef\csname odunum@n@perception_probe.calib.qwen3_5_omni_plus.overall\endcsname{150}
\expandafter\gdef\csname odunum@ci@perception_probe.calib.qwen3_5_omni_plus.overall\endcsname{[0.660, 0.803]}
\expandafter\gdef\csname odunum@val@perception_probe.calib.qwen3_5_omni_plus.visual\endcsname{0.845}
\expandafter\gdef\csname odunum@n@perception_probe.calib.qwen3_5_omni_plus.visual\endcsname{58}
\expandafter\gdef\csname odunum@ci@perception_probe.calib.qwen3_5_omni_plus.visual\endcsname{[0.741, 0.931]}
\expandafter\gdef\csname odunum@val@perception_probe.judge.id\endcsname{mr\_ali / dashscope.qwen3.6-flash (thinking)}
\expandafter\gdef\csname odunum@n@perception_probe.judge.id\endcsname{1}
\expandafter\gdef\csname odunum@val@perception_probe.leak.nomedia.audio\endcsname{0.472}
\expandafter\gdef\csname odunum@n@perception_probe.leak.nomedia.audio\endcsname{307}
\expandafter\gdef\csname odunum@ci@perception_probe.leak.nomedia.audio\endcsname{[0.416, 0.528]}
\expandafter\gdef\csname odunum@val@perception_probe.leak.nomedia.history\endcsname{0.710}
\expandafter\gdef\csname odunum@n@perception_probe.leak.nomedia.history\endcsname{238}
\expandafter\gdef\csname odunum@ci@perception_probe.leak.nomedia.history\endcsname{[0.650, 0.766]}
\expandafter\gdef\csname odunum@val@perception_probe.leak.nomedia.other_context\endcsname{0.309}
\expandafter\gdef\csname odunum@n@perception_probe.leak.nomedia.other_context\endcsname{81}
\expandafter\gdef\csname odunum@ci@perception_probe.leak.nomedia.other_context\endcsname{[0.214, 0.412]}
\expandafter\gdef\csname odunum@val@perception_probe.leak.nomedia.overall\endcsname{0.429}
\expandafter\gdef\csname odunum@n@perception_probe.leak.nomedia.overall\endcsname{1\,057}
\expandafter\gdef\csname odunum@ci@perception_probe.leak.nomedia.overall\endcsname{[0.397, 0.460]}
\expandafter\gdef\csname odunum@val@perception_probe.leak.nomedia.visual\endcsname{0.265}
\expandafter\gdef\csname odunum@n@perception_probe.leak.nomedia.visual\endcsname{431}
\expandafter\gdef\csname odunum@ci@perception_probe.leak.nomedia.visual\endcsname{[0.223, 0.307]}
\expandafter\gdef\csname odunum@val@perception_probe.net.doubao_seed_2_0_lite.hit\endcsname{0.255}
\expandafter\gdef\csname odunum@n@perception_probe.net.doubao_seed_2_0_lite.hit\endcsname{141}
\expandafter\gdef\csname odunum@ci@perception_probe.net.doubao_seed_2_0_lite.hit\endcsname{[0.150, 0.357]}
\expandafter\gdef\csname odunum@val@perception_probe.net.doubao_seed_2_0_lite.missed\endcsname{0.324}
\expandafter\gdef\csname odunum@n@perception_probe.net.doubao_seed_2_0_lite.missed\endcsname{284}
\expandafter\gdef\csname odunum@ci@perception_probe.net.doubao_seed_2_0_lite.missed\endcsname{[0.255, 0.392]}
\expandafter\gdef\csname odunum@val@perception_probe.net.doubao_seed_2_0_lite.missed_gen\endcsname{0.325}
\expandafter\gdef\csname odunum@n@perception_probe.net.doubao_seed_2_0_lite.missed_gen\endcsname{252}
\expandafter\gdef\csname odunum@ci@perception_probe.net.doubao_seed_2_0_lite.missed_gen\endcsname{[0.252, 0.398]}
\expandafter\gdef\csname odunum@val@perception_probe.net.doubao_seed_2_0_lite.missed_recorded\endcsname{0.312}
\expandafter\gdef\csname odunum@n@perception_probe.net.doubao_seed_2_0_lite.missed_recorded\endcsname{32}
\expandafter\gdef\csname odunum@ci@perception_probe.net.doubao_seed_2_0_lite.missed_recorded\endcsname{[0.100, 0.516]}
\expandafter\gdef\csname odunum@val@perception_probe.net.gemini_3_1_pro.hit\endcsname{0.140}
\expandafter\gdef\csname odunum@n@perception_probe.net.gemini_3_1_pro.hit\endcsname{150}
\expandafter\gdef\csname odunum@ci@perception_probe.net.gemini_3_1_pro.hit\endcsname{[0.047, 0.238]}
\expandafter\gdef\csname odunum@val@perception_probe.net.gemini_3_1_pro.missed\endcsname{0.210}
\expandafter\gdef\csname odunum@n@perception_probe.net.gemini_3_1_pro.missed\endcsname{300}
\expandafter\gdef\csname odunum@ci@perception_probe.net.gemini_3_1_pro.missed\endcsname{[0.140, 0.283]}
\expandafter\gdef\csname odunum@val@perception_probe.net.gemini_3_1_pro.missed_gen\endcsname{0.239}
\expandafter\gdef\csname odunum@n@perception_probe.net.gemini_3_1_pro.missed_gen\endcsname{255}
\expandafter\gdef\csname odunum@ci@perception_probe.net.gemini_3_1_pro.missed_gen\endcsname{[0.161, 0.318]}
\expandafter\gdef\csname odunum@val@perception_probe.net.gemini_3_1_pro.missed_recorded\endcsname{0.044}
\expandafter\gdef\csname odunum@n@perception_probe.net.gemini_3_1_pro.missed_recorded\endcsname{45}
\expandafter\gdef\csname odunum@ci@perception_probe.net.gemini_3_1_pro.missed_recorded\endcsname{[-0.116, 0.204]}
\expandafter\gdef\csname odunum@val@perception_probe.net.qwen3_5_omni_plus.hit\endcsname{0.133}
\expandafter\gdef\csname odunum@n@perception_probe.net.qwen3_5_omni_plus.hit\endcsname{150}
\expandafter\gdef\csname odunum@ci@perception_probe.net.qwen3_5_omni_plus.hit\endcsname{[0.033, 0.235]}
\expandafter\gdef\csname odunum@val@perception_probe.net.qwen3_5_omni_plus.missed\endcsname{0.254}
\expandafter\gdef\csname odunum@n@perception_probe.net.qwen3_5_omni_plus.missed\endcsname{299}
\expandafter\gdef\csname odunum@ci@perception_probe.net.qwen3_5_omni_plus.missed\endcsname{[0.183, 0.324]}
\expandafter\gdef\csname odunum@val@perception_probe.net.qwen3_5_omni_plus.missed_gen\endcsname{0.234}
\expandafter\gdef\csname odunum@n@perception_probe.net.qwen3_5_omni_plus.missed_gen\endcsname{265}
\expandafter\gdef\csname odunum@ci@perception_probe.net.qwen3_5_omni_plus.missed_gen\endcsname{[0.158, 0.311]}
\expandafter\gdef\csname odunum@val@perception_probe.net.qwen3_5_omni_plus.missed_recorded\endcsname{0.412}
\expandafter\gdef\csname odunum@n@perception_probe.net.qwen3_5_omni_plus.missed_recorded\endcsname{34}
\expandafter\gdef\csname odunum@ci@perception_probe.net.qwen3_5_omni_plus.missed_recorded\endcsname{[0.222, 0.606]}
\expandafter\gdef\csname odunum@val@perception_probe.probe.doubao_seed_2_0_lite.audio\endcsname{0.678}
\expandafter\gdef\csname odunum@n@perception_probe.probe.doubao_seed_2_0_lite.audio\endcsname{90}
\expandafter\gdef\csname odunum@ci@perception_probe.probe.doubao_seed_2_0_lite.audio\endcsname{[0.578, 0.769]}
\expandafter\gdef\csname odunum@val@perception_probe.probe.doubao_seed_2_0_lite.history\endcsname{0.569}
\expandafter\gdef\csname odunum@n@perception_probe.probe.doubao_seed_2_0_lite.history\endcsname{51}
\expandafter\gdef\csname odunum@ci@perception_probe.probe.doubao_seed_2_0_lite.history\endcsname{[0.423, 0.712]}
\expandafter\gdef\csname odunum@val@perception_probe.probe.doubao_seed_2_0_lite.other_context\endcsname{0.750}
\expandafter\gdef\csname odunum@n@perception_probe.probe.doubao_seed_2_0_lite.other_context\endcsname{28}
\expandafter\gdef\csname odunum@ci@perception_probe.probe.doubao_seed_2_0_lite.other_context\endcsname{[0.571, 0.893]}
\expandafter\gdef\csname odunum@val@perception_probe.probe.doubao_seed_2_0_lite.overall\endcsname{0.655}
\expandafter\gdef\csname odunum@n@perception_probe.probe.doubao_seed_2_0_lite.overall\endcsname{284}
\expandafter\gdef\csname odunum@ci@perception_probe.probe.doubao_seed_2_0_lite.overall\endcsname{[0.599, 0.708]}
\expandafter\gdef\csname odunum@val@perception_probe.probe.doubao_seed_2_0_lite.visual\endcsname{0.652}
\expandafter\gdef\csname odunum@n@perception_probe.probe.doubao_seed_2_0_lite.visual\endcsname{115}
\expandafter\gdef\csname odunum@ci@perception_probe.probe.doubao_seed_2_0_lite.visual\endcsname{[0.561, 0.735]}
\expandafter\gdef\csname odunum@val@perception_probe.probe.gemini_3_1_pro.audio\endcsname{0.605}
\expandafter\gdef\csname odunum@n@perception_probe.probe.gemini_3_1_pro.audio\endcsname{81}
\expandafter\gdef\csname odunum@ci@perception_probe.probe.gemini_3_1_pro.audio\endcsname{[0.494, 0.713]}
\expandafter\gdef\csname odunum@val@perception_probe.probe.gemini_3_1_pro.history\endcsname{0.573}
\expandafter\gdef\csname odunum@n@perception_probe.probe.gemini_3_1_pro.history\endcsname{75}
\expandafter\gdef\csname odunum@ci@perception_probe.probe.gemini_3_1_pro.history\endcsname{[0.461, 0.684]}
\expandafter\gdef\csname odunum@val@perception_probe.probe.gemini_3_1_pro.other_context\endcsname{0.633}
\expandafter\gdef\csname odunum@n@perception_probe.probe.gemini_3_1_pro.other_context\endcsname{30}
\expandafter\gdef\csname odunum@ci@perception_probe.probe.gemini_3_1_pro.other_context\endcsname{[0.467, 0.800]}
\expandafter\gdef\csname odunum@val@perception_probe.probe.gemini_3_1_pro.overall\endcsname{0.603}
\expandafter\gdef\csname odunum@n@perception_probe.probe.gemini_3_1_pro.overall\endcsname{300}
\expandafter\gdef\csname odunum@ci@perception_probe.probe.gemini_3_1_pro.overall\endcsname{[0.550, 0.659]}
\expandafter\gdef\csname odunum@val@perception_probe.probe.gemini_3_1_pro.visual\endcsname{0.614}
\expandafter\gdef\csname odunum@n@perception_probe.probe.gemini_3_1_pro.visual\endcsname{114}
\expandafter\gdef\csname odunum@ci@perception_probe.probe.gemini_3_1_pro.visual\endcsname{[0.522, 0.702]}
\expandafter\gdef\csname odunum@val@perception_probe.probe.generator\endcsname{mr\_ali / dashscope.qwen3.7-max}
\expandafter\gdef\csname odunum@n@perception_probe.probe.generator\endcsname{1}
\expandafter\gdef\csname odunum@val@perception_probe.probe.n_dropped\endcsname{2}
\expandafter\gdef\csname odunum@n@perception_probe.probe.n_dropped\endcsname{1\,059}
\expandafter\gdef\csname odunum@val@perception_probe.probe.n_usable\endcsname{1\,057}
\expandafter\gdef\csname odunum@n@perception_probe.probe.n_usable\endcsname{1\,059}
\expandafter\gdef\csname odunum@val@perception_probe.probe.qwen3_5_omni_plus.audio\endcsname{0.596}
\expandafter\gdef\csname odunum@n@perception_probe.probe.qwen3_5_omni_plus.audio\endcsname{89}
\expandafter\gdef\csname odunum@ci@perception_probe.probe.qwen3_5_omni_plus.audio\endcsname{[0.494, 0.697]}
\expandafter\gdef\csname odunum@val@perception_probe.probe.qwen3_5_omni_plus.history\endcsname{0.667}
\expandafter\gdef\csname odunum@n@perception_probe.probe.qwen3_5_omni_plus.history\endcsname{60}
\expandafter\gdef\csname odunum@ci@perception_probe.probe.qwen3_5_omni_plus.history\endcsname{[0.541, 0.787]}
\expandafter\gdef\csname odunum@val@perception_probe.probe.qwen3_5_omni_plus.other_context\endcsname{0.565}
\expandafter\gdef\csname odunum@n@perception_probe.probe.qwen3_5_omni_plus.other_context\endcsname{23}
\expandafter\gdef\csname odunum@ci@perception_probe.probe.qwen3_5_omni_plus.other_context\endcsname{[0.364, 0.750]}
\expandafter\gdef\csname odunum@val@perception_probe.probe.qwen3_5_omni_plus.overall\endcsname{0.612}
\expandafter\gdef\csname odunum@n@perception_probe.probe.qwen3_5_omni_plus.overall\endcsname{299}
\expandafter\gdef\csname odunum@ci@perception_probe.probe.qwen3_5_omni_plus.overall\endcsname{[0.557, 0.665]}
\expandafter\gdef\csname odunum@val@perception_probe.probe.qwen3_5_omni_plus.visual\endcsname{0.606}
\expandafter\gdef\csname odunum@n@perception_probe.probe.qwen3_5_omni_plus.visual\endcsname{127}
\expandafter\gdef\csname odunum@ci@perception_probe.probe.qwen3_5_omni_plus.visual\endcsname{[0.520, 0.691]}
\expandafter\gdef\csname odunum@val@pipeline_stats.boundary.n_positive\endcsname{1\,429}
\expandafter\gdef\csname odunum@n@pipeline_stats.boundary.n_positive\endcsname{1\,429}
\expandafter\gdef\csname odunum@val@pipeline_stats.boundary.n_unjoined\endcsname{277}
\expandafter\gdef\csname odunum@n@pipeline_stats.boundary.n_unjoined\endcsname{2\,078}
\expandafter\gdef\csname odunum@val@pipeline_stats.boundary.positive.match\endcsname{0}
\expandafter\gdef\csname odunum@n@pipeline_stats.boundary.positive.match\endcsname{1\,429}
\expandafter\gdef\csname odunum@val@pipeline_stats.boundary.positive.match_pct\endcsname{0.0\%}
\expandafter\gdef\csname odunum@n@pipeline_stats.boundary.positive.match_pct\endcsname{1\,429}
\expandafter\gdef\csname odunum@val@pipeline_stats.boundary.positive.mismatch\endcsname{187}
\expandafter\gdef\csname odunum@n@pipeline_stats.boundary.positive.mismatch\endcsname{1\,429}
\expandafter\gdef\csname odunum@val@pipeline_stats.boundary.positive.mismatch_pct\endcsname{13.1\%}
\expandafter\gdef\csname odunum@n@pipeline_stats.boundary.positive.mismatch_pct\endcsname{1\,429}
\expandafter\gdef\csname odunum@val@pipeline_stats.boundary.positive.not_recorded\endcsname{45}
\expandafter\gdef\csname odunum@n@pipeline_stats.boundary.positive.not_recorded\endcsname{1\,429}
\expandafter\gdef\csname odunum@val@pipeline_stats.boundary.positive.not_recorded_pct\endcsname{3.1\%}
\expandafter\gdef\csname odunum@n@pipeline_stats.boundary.positive.not_recorded_pct\endcsname{1\,429}
\expandafter\gdef\csname odunum@val@pipeline_stats.boundary.positive.partial\endcsname{1\,197}
\expandafter\gdef\csname odunum@n@pipeline_stats.boundary.positive.partial\endcsname{1\,429}
\expandafter\gdef\csname odunum@val@pipeline_stats.boundary.positive.partial_pct\endcsname{83.8\%}
\expandafter\gdef\csname odunum@n@pipeline_stats.boundary.positive.partial_pct\endcsname{1\,429}
\expandafter\gdef\csname odunum@val@pipeline_stats.curation.n_candidates\endcsname{51\,756}
\expandafter\gdef\csname odunum@n@pipeline_stats.curation.n_candidates\endcsname{51\,756}
\expandafter\gdef\csname odunum@val@pipeline_stats.curation.n_gated\endcsname{9\,217}
\expandafter\gdef\csname odunum@n@pipeline_stats.curation.n_gated\endcsname{51\,756}
\expandafter\gdef\csname odunum@val@pipeline_stats.curation.n_usable\endcsname{31\,606}
\expandafter\gdef\csname odunum@n@pipeline_stats.curation.n_usable\endcsname{51\,756}
\expandafter\gdef\csname odunum@val@pipeline_stats.gate.ao.n_judged\endcsname{29\,487}
\expandafter\gdef\csname odunum@n@pipeline_stats.gate.ao.n_judged\endcsname{29\,487}
\expandafter\gdef\csname odunum@val@pipeline_stats.gate.ao.negative.retention_pct\endcsname{65.2\%}
\expandafter\gdef\csname odunum@n@pipeline_stats.gate.ao.negative.retention_pct\endcsname{5\,000}
\expandafter\gdef\csname odunum@val@pipeline_stats.gate.ao.positive.retention_pct\endcsname{83.8\%}
\expandafter\gdef\csname odunum@n@pipeline_stats.gate.ao.positive.retention_pct\endcsname{24\,487}
\expandafter\gdef\csname odunum@val@pipeline_stats.gate.ao.retention_pct\endcsname{80.7\%}
\expandafter\gdef\csname odunum@n@pipeline_stats.gate.ao.retention_pct\endcsname{29\,487}
\expandafter\gdef\csname odunum@val@pipeline_stats.gate.av.n_judged\endcsname{40\,731}
\expandafter\gdef\csname odunum@n@pipeline_stats.gate.av.n_judged\endcsname{40\,731}
\expandafter\gdef\csname odunum@val@pipeline_stats.gate.av.negative.retention_pct\endcsname{84.5\%}
\expandafter\gdef\csname odunum@n@pipeline_stats.gate.av.negative.retention_pct\endcsname{7\,576}
\expandafter\gdef\csname odunum@val@pipeline_stats.gate.av.positive.retention_pct\endcsname{89.4\%}
\expandafter\gdef\csname odunum@n@pipeline_stats.gate.av.positive.retention_pct\endcsname{33\,155}
\expandafter\gdef\csname odunum@val@pipeline_stats.gate.av.retention_pct\endcsname{88.5\%}
\expandafter\gdef\csname odunum@n@pipeline_stats.gate.av.retention_pct\endcsname{40\,731}
\expandafter\gdef\csname odunum@val@pipeline_stats.gate.n_judged\endcsname{70\,218}
\expandafter\gdef\csname odunum@n@pipeline_stats.gate.n_judged\endcsname{70\,218}
\expandafter\gdef\csname odunum@val@pipeline_stats.gate.n_passed\endcsname{59\,824}
\expandafter\gdef\csname odunum@n@pipeline_stats.gate.n_passed\endcsname{70\,218}
\expandafter\gdef\csname odunum@val@pipeline_stats.gate.n_rejected\endcsname{10\,394}
\expandafter\gdef\csname odunum@n@pipeline_stats.gate.n_rejected\endcsname{70\,218}
\expandafter\gdef\csname odunum@val@pipeline_stats.gate.reject.evidence_grounded\endcsname{3\,100}
\expandafter\gdef\csname odunum@n@pipeline_stats.gate.reject.evidence_grounded\endcsname{10\,394}
\expandafter\gdef\csname odunum@val@pipeline_stats.gate.reject.intent_clear\endcsname{2\,646}
\expandafter\gdef\csname odunum@n@pipeline_stats.gate.reject.intent_clear\endcsname{10\,394}
\expandafter\gdef\csname odunum@val@pipeline_stats.gate.reject.multiturn_coherent\endcsname{318}
\expandafter\gdef\csname odunum@n@pipeline_stats.gate.reject.multiturn_coherent\endcsname{10\,394}
\expandafter\gdef\csname odunum@val@pipeline_stats.gate.reject.polarity_clear\endcsname{6\,795}
\expandafter\gdef\csname odunum@n@pipeline_stats.gate.reject.polarity_clear\endcsname{10\,394}
\expandafter\gdef\csname odunum@val@pipeline_stats.gate.reject.realistic_consistent\endcsname{2\,884}
\expandafter\gdef\csname odunum@n@pipeline_stats.gate.reject.realistic_consistent\endcsname{10\,394}
\expandafter\gdef\csname odunum@val@pipeline_stats.gate.retention_pct\endcsname{85.2\%}
\expandafter\gdef\csname odunum@n@pipeline_stats.gate.retention_pct\endcsname{70\,218}
\expandafter\gdef\csname odunum@val@positive_taxonomy_m2.analysis.bootstrap_replicates\endcsname{10\,000}
\expandafter\gdef\csname odunum@n@positive_taxonomy_m2.analysis.bootstrap_replicates\endcsname{10\,000}
\expandafter\gdef\csname odunum@val@positive_taxonomy_m2.analysis.thin_group_threshold\endcsname{40}
\expandafter\gdef\csname odunum@n@positive_taxonomy_m2.analysis.thin_group_threshold\endcsname{1\,631}
\expandafter\gdef\csname odunum@val@positive_taxonomy_m2.ao.count.axis1.1_1\endcsname{156}
\expandafter\gdef\csname odunum@n@positive_taxonomy_m2.ao.count.axis1.1_1\endcsname{563}
\expandafter\gdef\csname odunum@val@positive_taxonomy_m2.ao.count.axis1.1_2\endcsname{407}
\expandafter\gdef\csname odunum@n@positive_taxonomy_m2.ao.count.axis1.1_2\endcsname{563}
\expandafter\gdef\csname odunum@val@positive_taxonomy_m2.ao.count.axis2.2_1\endcsname{21}
\expandafter\gdef\csname odunum@n@positive_taxonomy_m2.ao.count.axis2.2_1\endcsname{563}
\expandafter\gdef\csname odunum@val@positive_taxonomy_m2.ao.count.axis2.2_2\endcsname{169}
\expandafter\gdef\csname odunum@n@positive_taxonomy_m2.ao.count.axis2.2_2\endcsname{563}
\expandafter\gdef\csname odunum@val@positive_taxonomy_m2.ao.count.axis2.2_3\endcsname{248}
\expandafter\gdef\csname odunum@n@positive_taxonomy_m2.ao.count.axis2.2_3\endcsname{563}
\expandafter\gdef\csname odunum@val@positive_taxonomy_m2.ao.count.axis2.2_4\endcsname{34}
\expandafter\gdef\csname odunum@n@positive_taxonomy_m2.ao.count.axis2.2_4\endcsname{563}
\expandafter\gdef\csname odunum@val@positive_taxonomy_m2.ao.count.axis2.2_5\endcsname{91}
\expandafter\gdef\csname odunum@n@positive_taxonomy_m2.ao.count.axis2.2_5\endcsname{563}
\expandafter\gdef\csname odunum@val@positive_taxonomy_m2.ao.count.axis3.3_1\endcsname{18}
\expandafter\gdef\csname odunum@n@positive_taxonomy_m2.ao.count.axis3.3_1\endcsname{563}
\expandafter\gdef\csname odunum@val@positive_taxonomy_m2.ao.count.axis3.3_2\endcsname{30}
\expandafter\gdef\csname odunum@n@positive_taxonomy_m2.ao.count.axis3.3_2\endcsname{563}
\expandafter\gdef\csname odunum@val@positive_taxonomy_m2.ao.count.axis3.3_3\endcsname{67}
\expandafter\gdef\csname odunum@n@positive_taxonomy_m2.ao.count.axis3.3_3\endcsname{563}
\expandafter\gdef\csname odunum@val@positive_taxonomy_m2.ao.count.axis3.3_4\endcsname{141}
\expandafter\gdef\csname odunum@n@positive_taxonomy_m2.ao.count.axis3.3_4\endcsname{563}
\expandafter\gdef\csname odunum@val@positive_taxonomy_m2.ao.count.axis3.3_5\endcsname{105}
\expandafter\gdef\csname odunum@n@positive_taxonomy_m2.ao.count.axis3.3_5\endcsname{563}
\expandafter\gdef\csname odunum@val@positive_taxonomy_m2.ao.count.axis3.3_6\endcsname{202}
\expandafter\gdef\csname odunum@n@positive_taxonomy_m2.ao.count.axis3.3_6\endcsname{563}
\expandafter\gdef\csname odunum@val@positive_taxonomy_m2.ao.count.axis4.4_1\endcsname{63}
\expandafter\gdef\csname odunum@n@positive_taxonomy_m2.ao.count.axis4.4_1\endcsname{563}
\expandafter\gdef\csname odunum@val@positive_taxonomy_m2.ao.count.axis4.4_2\endcsname{161}
\expandafter\gdef\csname odunum@n@positive_taxonomy_m2.ao.count.axis4.4_2\endcsname{563}
\expandafter\gdef\csname odunum@val@positive_taxonomy_m2.ao.count.axis4.4_3\endcsname{43}
\expandafter\gdef\csname odunum@n@positive_taxonomy_m2.ao.count.axis4.4_3\endcsname{563}
\expandafter\gdef\csname odunum@val@positive_taxonomy_m2.ao.count.axis4.4_4\endcsname{172}
\expandafter\gdef\csname odunum@n@positive_taxonomy_m2.ao.count.axis4.4_4\endcsname{563}
\expandafter\gdef\csname odunum@val@positive_taxonomy_m2.ao.count.axis4.4_5\endcsname{124}
\expandafter\gdef\csname odunum@n@positive_taxonomy_m2.ao.count.axis4.4_5\endcsname{563}
\expandafter\gdef\csname odunum@val@positive_taxonomy_m2.ao.delta.cascade_asr.axis1.1_1\endcsname{0.085}
\expandafter\gdef\csname odunum@n@positive_taxonomy_m2.ao.delta.cascade_asr.axis1.1_1\endcsname{156}
\expandafter\gdef\csname odunum@ci@positive_taxonomy_m2.ao.delta.cascade_asr.axis1.1_1\endcsname{[0.050, 0.120]}
\expandafter\gdef\csname odunum@val@positive_taxonomy_m2.ao.delta.cascade_asr.axis1.1_2\endcsname{-0.032}
\expandafter\gdef\csname odunum@n@positive_taxonomy_m2.ao.delta.cascade_asr.axis1.1_2\endcsname{407}
\expandafter\gdef\csname odunum@ci@positive_taxonomy_m2.ao.delta.cascade_asr.axis1.1_2\endcsname{[-0.047, -0.019]}
\expandafter\gdef\csname odunum@val@positive_taxonomy_m2.ao.delta.cascade_asr.axis2.2_1\endcsname{-0.186}
\expandafter\gdef\csname odunum@n@positive_taxonomy_m2.ao.delta.cascade_asr.axis2.2_1\endcsname{21}
\expandafter\gdef\csname odunum@ci@positive_taxonomy_m2.ao.delta.cascade_asr.axis2.2_1\endcsname{[-0.348, -0.030]}
\expandafter\gdef\csname odunum@val@positive_taxonomy_m2.ao.delta.cascade_asr.axis2.2_2\endcsname{0.015}
\expandafter\gdef\csname odunum@n@positive_taxonomy_m2.ao.delta.cascade_asr.axis2.2_2\endcsname{169}
\expandafter\gdef\csname odunum@ci@positive_taxonomy_m2.ao.delta.cascade_asr.axis2.2_2\endcsname{[-0.019, 0.050]}
\expandafter\gdef\csname odunum@val@positive_taxonomy_m2.ao.delta.cascade_asr.axis2.2_3\endcsname{0.049}
\expandafter\gdef\csname odunum@n@positive_taxonomy_m2.ao.delta.cascade_asr.axis2.2_3\endcsname{248}
\expandafter\gdef\csname odunum@ci@positive_taxonomy_m2.ao.delta.cascade_asr.axis2.2_3\endcsname{[0.024, 0.075]}
\expandafter\gdef\csname odunum@val@positive_taxonomy_m2.ao.delta.cascade_asr.axis2.2_4\endcsname{-0.069}
\expandafter\gdef\csname odunum@n@positive_taxonomy_m2.ao.delta.cascade_asr.axis2.2_4\endcsname{34}
\expandafter\gdef\csname odunum@ci@positive_taxonomy_m2.ao.delta.cascade_asr.axis2.2_4\endcsname{[-0.184, 0.043]}
\expandafter\gdef\csname odunum@val@positive_taxonomy_m2.ao.delta.cascade_asr.axis2.2_5\endcsname{-0.094}
\expandafter\gdef\csname odunum@n@positive_taxonomy_m2.ao.delta.cascade_asr.axis2.2_5\endcsname{91}
\expandafter\gdef\csname odunum@ci@positive_taxonomy_m2.ao.delta.cascade_asr.axis2.2_5\endcsname{[-0.159, -0.031]}
\expandafter\gdef\csname odunum@val@positive_taxonomy_m2.ao.delta.cascade_asr.axis3.3_1\endcsname{0.076}
\expandafter\gdef\csname odunum@n@positive_taxonomy_m2.ao.delta.cascade_asr.axis3.3_1\endcsname{18}
\expandafter\gdef\csname odunum@ci@positive_taxonomy_m2.ao.delta.cascade_asr.axis3.3_1\endcsname{[-0.066, 0.203]}
\expandafter\gdef\csname odunum@val@positive_taxonomy_m2.ao.delta.cascade_asr.axis3.3_2\endcsname{-0.031}
\expandafter\gdef\csname odunum@n@positive_taxonomy_m2.ao.delta.cascade_asr.axis3.3_2\endcsname{30}
\expandafter\gdef\csname odunum@ci@positive_taxonomy_m2.ao.delta.cascade_asr.axis3.3_2\endcsname{[-0.103, 0.042]}
\expandafter\gdef\csname odunum@val@positive_taxonomy_m2.ao.delta.cascade_asr.axis3.3_3\endcsname{-0.128}
\expandafter\gdef\csname odunum@n@positive_taxonomy_m2.ao.delta.cascade_asr.axis3.3_3\endcsname{67}
\expandafter\gdef\csname odunum@ci@positive_taxonomy_m2.ao.delta.cascade_asr.axis3.3_3\endcsname{[-0.214, -0.045]}
\expandafter\gdef\csname odunum@val@positive_taxonomy_m2.ao.delta.cascade_asr.axis3.3_4\endcsname{0.044}
\expandafter\gdef\csname odunum@n@positive_taxonomy_m2.ao.delta.cascade_asr.axis3.3_4\endcsname{141}
\expandafter\gdef\csname odunum@ci@positive_taxonomy_m2.ao.delta.cascade_asr.axis3.3_4\endcsname{[0.004, 0.083]}
\expandafter\gdef\csname odunum@val@positive_taxonomy_m2.ao.delta.cascade_asr.axis3.3_5\endcsname{-0.105}
\expandafter\gdef\csname odunum@n@positive_taxonomy_m2.ao.delta.cascade_asr.axis3.3_5\endcsname{105}
\expandafter\gdef\csname odunum@ci@positive_taxonomy_m2.ao.delta.cascade_asr.axis3.3_5\endcsname{[-0.156, -0.054]}
\expandafter\gdef\csname odunum@val@positive_taxonomy_m2.ao.delta.cascade_asr.axis3.3_6\endcsname{0.065}
\expandafter\gdef\csname odunum@n@positive_taxonomy_m2.ao.delta.cascade_asr.axis3.3_6\endcsname{202}
\expandafter\gdef\csname odunum@ci@positive_taxonomy_m2.ao.delta.cascade_asr.axis3.3_6\endcsname{[0.035, 0.095]}
\expandafter\gdef\csname odunum@val@positive_taxonomy_m2.ao.delta.cascade_asr.axis4.4_1\endcsname{-1.0\ensuremath{\times 10^{-4}}}
\expandafter\gdef\csname odunum@n@positive_taxonomy_m2.ao.delta.cascade_asr.axis4.4_1\endcsname{63}
\expandafter\gdef\csname odunum@ci@positive_taxonomy_m2.ao.delta.cascade_asr.axis4.4_1\endcsname{[-0.061, 0.058]}
\expandafter\gdef\csname odunum@val@positive_taxonomy_m2.ao.delta.cascade_asr.axis4.4_2\endcsname{0.006}
\expandafter\gdef\csname odunum@n@positive_taxonomy_m2.ao.delta.cascade_asr.axis4.4_2\endcsname{161}
\expandafter\gdef\csname odunum@ci@positive_taxonomy_m2.ao.delta.cascade_asr.axis4.4_2\endcsname{[-0.033, 0.042]}
\expandafter\gdef\csname odunum@val@positive_taxonomy_m2.ao.delta.cascade_asr.axis4.4_3\endcsname{-0.006}
\expandafter\gdef\csname odunum@n@positive_taxonomy_m2.ao.delta.cascade_asr.axis4.4_3\endcsname{43}
\expandafter\gdef\csname odunum@ci@positive_taxonomy_m2.ao.delta.cascade_asr.axis4.4_3\endcsname{[-0.099, 0.081]}
\expandafter\gdef\csname odunum@val@positive_taxonomy_m2.ao.delta.cascade_asr.axis4.4_4\endcsname{-0.017}
\expandafter\gdef\csname odunum@n@positive_taxonomy_m2.ao.delta.cascade_asr.axis4.4_4\endcsname{172}
\expandafter\gdef\csname odunum@ci@positive_taxonomy_m2.ao.delta.cascade_asr.axis4.4_4\endcsname{[-0.055, 0.020]}
\expandafter\gdef\csname odunum@val@positive_taxonomy_m2.ao.delta.cascade_asr.axis4.4_5\endcsname{0.018}
\expandafter\gdef\csname odunum@n@positive_taxonomy_m2.ao.delta.cascade_asr.axis4.4_5\endcsname{124}
\expandafter\gdef\csname odunum@ci@positive_taxonomy_m2.ao.delta.cascade_asr.axis4.4_5\endcsname{[-0.026, 0.062]}
\expandafter\gdef\csname odunum@val@positive_taxonomy_m2.ao.delta.gemini.axis1.1_1\endcsname{0.007}
\expandafter\gdef\csname odunum@n@positive_taxonomy_m2.ao.delta.gemini.axis1.1_1\endcsname{156}
\expandafter\gdef\csname odunum@ci@positive_taxonomy_m2.ao.delta.gemini.axis1.1_1\endcsname{[-0.027, 0.041]}
\expandafter\gdef\csname odunum@val@positive_taxonomy_m2.ao.delta.gemini.axis1.1_2\endcsname{-0.003}
\expandafter\gdef\csname odunum@n@positive_taxonomy_m2.ao.delta.gemini.axis1.1_2\endcsname{407}
\expandafter\gdef\csname odunum@ci@positive_taxonomy_m2.ao.delta.gemini.axis1.1_2\endcsname{[-0.016, 0.010]}
\expandafter\gdef\csname odunum@val@positive_taxonomy_m2.ao.delta.gemini.axis2.2_1\endcsname{-0.005}
\expandafter\gdef\csname odunum@n@positive_taxonomy_m2.ao.delta.gemini.axis2.2_1\endcsname{21}
\expandafter\gdef\csname odunum@ci@positive_taxonomy_m2.ao.delta.gemini.axis2.2_1\endcsname{[-0.145, 0.127]}
\expandafter\gdef\csname odunum@val@positive_taxonomy_m2.ao.delta.gemini.axis2.2_2\endcsname{0.011}
\expandafter\gdef\csname odunum@n@positive_taxonomy_m2.ao.delta.gemini.axis2.2_2\endcsname{169}
\expandafter\gdef\csname odunum@ci@positive_taxonomy_m2.ao.delta.gemini.axis2.2_2\endcsname{[-0.020, 0.042]}
\expandafter\gdef\csname odunum@val@positive_taxonomy_m2.ao.delta.gemini.axis2.2_3\endcsname{0.001}
\expandafter\gdef\csname odunum@n@positive_taxonomy_m2.ao.delta.gemini.axis2.2_3\endcsname{248}
\expandafter\gdef\csname odunum@ci@positive_taxonomy_m2.ao.delta.gemini.axis2.2_3\endcsname{[-0.023, 0.024]}
\expandafter\gdef\csname odunum@val@positive_taxonomy_m2.ao.delta.gemini.axis2.2_4\endcsname{-0.051}
\expandafter\gdef\csname odunum@n@positive_taxonomy_m2.ao.delta.gemini.axis2.2_4\endcsname{34}
\expandafter\gdef\csname odunum@ci@positive_taxonomy_m2.ao.delta.gemini.axis2.2_4\endcsname{[-0.146, 0.036]}
\expandafter\gdef\csname odunum@val@positive_taxonomy_m2.ao.delta.gemini.axis2.2_5\endcsname{-0.002}
\expandafter\gdef\csname odunum@n@positive_taxonomy_m2.ao.delta.gemini.axis2.2_5\endcsname{91}
\expandafter\gdef\csname odunum@ci@positive_taxonomy_m2.ao.delta.gemini.axis2.2_5\endcsname{[-0.054, 0.051]}
\expandafter\gdef\csname odunum@val@positive_taxonomy_m2.ao.delta.gemini.axis3.3_1\endcsname{0.064}
\expandafter\gdef\csname odunum@n@positive_taxonomy_m2.ao.delta.gemini.axis3.3_1\endcsname{18}
\expandafter\gdef\csname odunum@ci@positive_taxonomy_m2.ao.delta.gemini.axis3.3_1\endcsname{[-0.034, 0.162]}
\expandafter\gdef\csname odunum@val@positive_taxonomy_m2.ao.delta.gemini.axis3.3_2\endcsname{0.015}
\expandafter\gdef\csname odunum@n@positive_taxonomy_m2.ao.delta.gemini.axis3.3_2\endcsname{30}
\expandafter\gdef\csname odunum@ci@positive_taxonomy_m2.ao.delta.gemini.axis3.3_2\endcsname{[-0.077, 0.100]}
\expandafter\gdef\csname odunum@val@positive_taxonomy_m2.ao.delta.gemini.axis3.3_3\endcsname{-0.014}
\expandafter\gdef\csname odunum@n@positive_taxonomy_m2.ao.delta.gemini.axis3.3_3\endcsname{67}
\expandafter\gdef\csname odunum@ci@positive_taxonomy_m2.ao.delta.gemini.axis3.3_3\endcsname{[-0.075, 0.046]}
\expandafter\gdef\csname odunum@val@positive_taxonomy_m2.ao.delta.gemini.axis3.3_4\endcsname{-0.037}
\expandafter\gdef\csname odunum@n@positive_taxonomy_m2.ao.delta.gemini.axis3.3_4\endcsname{141}
\expandafter\gdef\csname odunum@ci@positive_taxonomy_m2.ao.delta.gemini.axis3.3_4\endcsname{[-0.076, 0.001]}
\expandafter\gdef\csname odunum@val@positive_taxonomy_m2.ao.delta.gemini.axis3.3_5\endcsname{-0.075}
\expandafter\gdef\csname odunum@n@positive_taxonomy_m2.ao.delta.gemini.axis3.3_5\endcsname{105}
\expandafter\gdef\csname odunum@ci@positive_taxonomy_m2.ao.delta.gemini.axis3.3_5\endcsname{[-0.120, -0.030]}
\expandafter\gdef\csname odunum@val@positive_taxonomy_m2.ao.delta.gemini.axis3.3_6\endcsname{0.062}
\expandafter\gdef\csname odunum@n@positive_taxonomy_m2.ao.delta.gemini.axis3.3_6\endcsname{202}
\expandafter\gdef\csname odunum@ci@positive_taxonomy_m2.ao.delta.gemini.axis3.3_6\endcsname{[0.035, 0.089]}
\expandafter\gdef\csname odunum@val@positive_taxonomy_m2.ao.delta.gemini.axis4.4_1\endcsname{-0.008}
\expandafter\gdef\csname odunum@n@positive_taxonomy_m2.ao.delta.gemini.axis4.4_1\endcsname{63}
\expandafter\gdef\csname odunum@ci@positive_taxonomy_m2.ao.delta.gemini.axis4.4_1\endcsname{[-0.058, 0.042]}
\expandafter\gdef\csname odunum@val@positive_taxonomy_m2.ao.delta.gemini.axis4.4_2\endcsname{0.001}
\expandafter\gdef\csname odunum@n@positive_taxonomy_m2.ao.delta.gemini.axis4.4_2\endcsname{161}
\expandafter\gdef\csname odunum@ci@positive_taxonomy_m2.ao.delta.gemini.axis4.4_2\endcsname{[-0.033, 0.034]}
\expandafter\gdef\csname odunum@val@positive_taxonomy_m2.ao.delta.gemini.axis4.4_3\endcsname{-0.011}
\expandafter\gdef\csname odunum@n@positive_taxonomy_m2.ao.delta.gemini.axis4.4_3\endcsname{43}
\expandafter\gdef\csname odunum@ci@positive_taxonomy_m2.ao.delta.gemini.axis4.4_3\endcsname{[-0.092, 0.067]}
\expandafter\gdef\csname odunum@val@positive_taxonomy_m2.ao.delta.gemini.axis4.4_4\endcsname{0.026}
\expandafter\gdef\csname odunum@n@positive_taxonomy_m2.ao.delta.gemini.axis4.4_4\endcsname{172}
\expandafter\gdef\csname odunum@ci@positive_taxonomy_m2.ao.delta.gemini.axis4.4_4\endcsname{[-0.006, 0.056]}
\expandafter\gdef\csname odunum@val@positive_taxonomy_m2.ao.delta.gemini.axis4.4_5\endcsname{-0.029}
\expandafter\gdef\csname odunum@n@positive_taxonomy_m2.ao.delta.gemini.axis4.4_5\endcsname{124}
\expandafter\gdef\csname odunum@ci@positive_taxonomy_m2.ao.delta.gemini.axis4.4_5\endcsname{[-0.070, 0.012]}
\expandafter\gdef\csname odunum@val@positive_taxonomy_m2.ao.delta.gemini35_flash_lite.axis1.1_1\endcsname{0.089}
\expandafter\gdef\csname odunum@n@positive_taxonomy_m2.ao.delta.gemini35_flash_lite.axis1.1_1\endcsname{156}
\expandafter\gdef\csname odunum@ci@positive_taxonomy_m2.ao.delta.gemini35_flash_lite.axis1.1_1\endcsname{[0.045, 0.133]}
\expandafter\gdef\csname odunum@val@positive_taxonomy_m2.ao.delta.gemini35_flash_lite.axis1.1_2\endcsname{-0.034}
\expandafter\gdef\csname odunum@n@positive_taxonomy_m2.ao.delta.gemini35_flash_lite.axis1.1_2\endcsname{407}
\expandafter\gdef\csname odunum@ci@positive_taxonomy_m2.ao.delta.gemini35_flash_lite.axis1.1_2\endcsname{[-0.052, -0.017]}
\expandafter\gdef\csname odunum@val@positive_taxonomy_m2.ao.delta.gemini35_flash_lite.axis2.2_1\endcsname{-0.148}
\expandafter\gdef\csname odunum@n@positive_taxonomy_m2.ao.delta.gemini35_flash_lite.axis2.2_1\endcsname{21}
\expandafter\gdef\csname odunum@ci@positive_taxonomy_m2.ao.delta.gemini35_flash_lite.axis2.2_1\endcsname{[-0.313, 0.023]}
\expandafter\gdef\csname odunum@val@positive_taxonomy_m2.ao.delta.gemini35_flash_lite.axis2.2_2\endcsname{0.042}
\expandafter\gdef\csname odunum@n@positive_taxonomy_m2.ao.delta.gemini35_flash_lite.axis2.2_2\endcsname{169}
\expandafter\gdef\csname odunum@ci@positive_taxonomy_m2.ao.delta.gemini35_flash_lite.axis2.2_2\endcsname{[0.001, 0.082]}
\expandafter\gdef\csname odunum@val@positive_taxonomy_m2.ao.delta.gemini35_flash_lite.axis2.2_3\endcsname{0.080}
\expandafter\gdef\csname odunum@n@positive_taxonomy_m2.ao.delta.gemini35_flash_lite.axis2.2_3\endcsname{248}
\expandafter\gdef\csname odunum@ci@positive_taxonomy_m2.ao.delta.gemini35_flash_lite.axis2.2_3\endcsname{[0.051, 0.110]}
\expandafter\gdef\csname odunum@val@positive_taxonomy_m2.ao.delta.gemini35_flash_lite.axis2.2_4\endcsname{-0.218}
\expandafter\gdef\csname odunum@n@positive_taxonomy_m2.ao.delta.gemini35_flash_lite.axis2.2_4\endcsname{34}
\expandafter\gdef\csname odunum@ci@positive_taxonomy_m2.ao.delta.gemini35_flash_lite.axis2.2_4\endcsname{[-0.331, -0.100]}
\expandafter\gdef\csname odunum@val@positive_taxonomy_m2.ao.delta.gemini35_flash_lite.axis2.2_5\endcsname{-0.180}
\expandafter\gdef\csname odunum@n@positive_taxonomy_m2.ao.delta.gemini35_flash_lite.axis2.2_5\endcsname{91}
\expandafter\gdef\csname odunum@ci@positive_taxonomy_m2.ao.delta.gemini35_flash_lite.axis2.2_5\endcsname{[-0.248, -0.110]}
\expandafter\gdef\csname odunum@val@positive_taxonomy_m2.ao.delta.gemini35_flash_lite.axis3.3_1\endcsname{0.149}
\expandafter\gdef\csname odunum@n@positive_taxonomy_m2.ao.delta.gemini35_flash_lite.axis3.3_1\endcsname{18}
\expandafter\gdef\csname odunum@ci@positive_taxonomy_m2.ao.delta.gemini35_flash_lite.axis3.3_1\endcsname{[0.009, 0.278]}
\expandafter\gdef\csname odunum@val@positive_taxonomy_m2.ao.delta.gemini35_flash_lite.axis3.3_2\endcsname{0.018}
\expandafter\gdef\csname odunum@n@positive_taxonomy_m2.ao.delta.gemini35_flash_lite.axis3.3_2\endcsname{30}
\expandafter\gdef\csname odunum@ci@positive_taxonomy_m2.ao.delta.gemini35_flash_lite.axis3.3_2\endcsname{[-0.078, 0.117]}
\expandafter\gdef\csname odunum@val@positive_taxonomy_m2.ao.delta.gemini35_flash_lite.axis3.3_3\endcsname{-0.201}
\expandafter\gdef\csname odunum@n@positive_taxonomy_m2.ao.delta.gemini35_flash_lite.axis3.3_3\endcsname{67}
\expandafter\gdef\csname odunum@ci@positive_taxonomy_m2.ao.delta.gemini35_flash_lite.axis3.3_3\endcsname{[-0.284, -0.115]}
\expandafter\gdef\csname odunum@val@positive_taxonomy_m2.ao.delta.gemini35_flash_lite.axis3.3_4\endcsname{0.033}
\expandafter\gdef\csname odunum@n@positive_taxonomy_m2.ao.delta.gemini35_flash_lite.axis3.3_4\endcsname{141}
\expandafter\gdef\csname odunum@ci@positive_taxonomy_m2.ao.delta.gemini35_flash_lite.axis3.3_4\endcsname{[-0.011, 0.077]}
\expandafter\gdef\csname odunum@val@positive_taxonomy_m2.ao.delta.gemini35_flash_lite.axis3.3_5\endcsname{-0.061}
\expandafter\gdef\csname odunum@n@positive_taxonomy_m2.ao.delta.gemini35_flash_lite.axis3.3_5\endcsname{105}
\expandafter\gdef\csname odunum@ci@positive_taxonomy_m2.ao.delta.gemini35_flash_lite.axis3.3_5\endcsname{[-0.121, -0.002]}
\expandafter\gdef\csname odunum@val@positive_taxonomy_m2.ao.delta.gemini35_flash_lite.axis3.3_6\endcsname{0.060}
\expandafter\gdef\csname odunum@n@positive_taxonomy_m2.ao.delta.gemini35_flash_lite.axis3.3_6\endcsname{202}
\expandafter\gdef\csname odunum@ci@positive_taxonomy_m2.ao.delta.gemini35_flash_lite.axis3.3_6\endcsname{[0.025, 0.095]}
\expandafter\gdef\csname odunum@val@positive_taxonomy_m2.ao.delta.gemini35_flash_lite.axis4.4_1\endcsname{-0.015}
\expandafter\gdef\csname odunum@n@positive_taxonomy_m2.ao.delta.gemini35_flash_lite.axis4.4_1\endcsname{63}
\expandafter\gdef\csname odunum@ci@positive_taxonomy_m2.ao.delta.gemini35_flash_lite.axis4.4_1\endcsname{[-0.088, 0.055]}
\expandafter\gdef\csname odunum@val@positive_taxonomy_m2.ao.delta.gemini35_flash_lite.axis4.4_2\endcsname{0.022}
\expandafter\gdef\csname odunum@n@positive_taxonomy_m2.ao.delta.gemini35_flash_lite.axis4.4_2\endcsname{161}
\expandafter\gdef\csname odunum@ci@positive_taxonomy_m2.ao.delta.gemini35_flash_lite.axis4.4_2\endcsname{[-0.020, 0.063]}
\expandafter\gdef\csname odunum@val@positive_taxonomy_m2.ao.delta.gemini35_flash_lite.axis4.4_3\endcsname{-0.103}
\expandafter\gdef\csname odunum@n@positive_taxonomy_m2.ao.delta.gemini35_flash_lite.axis4.4_3\endcsname{43}
\expandafter\gdef\csname odunum@ci@positive_taxonomy_m2.ao.delta.gemini35_flash_lite.axis4.4_3\endcsname{[-0.210, 0.001]}
\expandafter\gdef\csname odunum@val@positive_taxonomy_m2.ao.delta.gemini35_flash_lite.axis4.4_4\endcsname{-0.025}
\expandafter\gdef\csname odunum@n@positive_taxonomy_m2.ao.delta.gemini35_flash_lite.axis4.4_4\endcsname{172}
\expandafter\gdef\csname odunum@ci@positive_taxonomy_m2.ao.delta.gemini35_flash_lite.axis4.4_4\endcsname{[-0.066, 0.017]}
\expandafter\gdef\csname odunum@val@positive_taxonomy_m2.ao.delta.gemini35_flash_lite.axis4.4_5\endcsname{0.050}
\expandafter\gdef\csname odunum@n@positive_taxonomy_m2.ao.delta.gemini35_flash_lite.axis4.4_5\endcsname{124}
\expandafter\gdef\csname odunum@ci@positive_taxonomy_m2.ao.delta.gemini35_flash_lite.axis4.4_5\endcsname{[-0.003, 0.102]}
\expandafter\gdef\csname odunum@val@positive_taxonomy_m2.ao.delta.gemini37_flash.axis1.1_1\endcsname{0.058}
\expandafter\gdef\csname odunum@n@positive_taxonomy_m2.ao.delta.gemini37_flash.axis1.1_1\endcsname{156}
\expandafter\gdef\csname odunum@ci@positive_taxonomy_m2.ao.delta.gemini37_flash.axis1.1_1\endcsname{[0.022, 0.095]}
\expandafter\gdef\csname odunum@val@positive_taxonomy_m2.ao.delta.gemini37_flash.axis1.1_2\endcsname{-0.022}
\expandafter\gdef\csname odunum@n@positive_taxonomy_m2.ao.delta.gemini37_flash.axis1.1_2\endcsname{407}
\expandafter\gdef\csname odunum@ci@positive_taxonomy_m2.ao.delta.gemini37_flash.axis1.1_2\endcsname{[-0.037, -0.008]}
\expandafter\gdef\csname odunum@val@positive_taxonomy_m2.ao.delta.gemini37_flash.axis2.2_1\endcsname{-0.029}
\expandafter\gdef\csname odunum@n@positive_taxonomy_m2.ao.delta.gemini37_flash.axis2.2_1\endcsname{21}
\expandafter\gdef\csname odunum@ci@positive_taxonomy_m2.ao.delta.gemini37_flash.axis2.2_1\endcsname{[-0.172, 0.107]}
\expandafter\gdef\csname odunum@val@positive_taxonomy_m2.ao.delta.gemini37_flash.axis2.2_2\endcsname{0.005}
\expandafter\gdef\csname odunum@n@positive_taxonomy_m2.ao.delta.gemini37_flash.axis2.2_2\endcsname{169}
\expandafter\gdef\csname odunum@ci@positive_taxonomy_m2.ao.delta.gemini37_flash.axis2.2_2\endcsname{[-0.028, 0.039]}
\expandafter\gdef\csname odunum@val@positive_taxonomy_m2.ao.delta.gemini37_flash.axis2.2_3\endcsname{0.033}
\expandafter\gdef\csname odunum@n@positive_taxonomy_m2.ao.delta.gemini37_flash.axis2.2_3\endcsname{248}
\expandafter\gdef\csname odunum@ci@positive_taxonomy_m2.ao.delta.gemini37_flash.axis2.2_3\endcsname{[0.009, 0.059]}
\expandafter\gdef\csname odunum@val@positive_taxonomy_m2.ao.delta.gemini37_flash.axis2.2_4\endcsname{-0.038}
\expandafter\gdef\csname odunum@n@positive_taxonomy_m2.ao.delta.gemini37_flash.axis2.2_4\endcsname{34}
\expandafter\gdef\csname odunum@ci@positive_taxonomy_m2.ao.delta.gemini37_flash.axis2.2_4\endcsname{[-0.151, 0.067]}
\expandafter\gdef\csname odunum@val@positive_taxonomy_m2.ao.delta.gemini37_flash.axis2.2_5\endcsname{-0.079}
\expandafter\gdef\csname odunum@n@positive_taxonomy_m2.ao.delta.gemini37_flash.axis2.2_5\endcsname{91}
\expandafter\gdef\csname odunum@ci@positive_taxonomy_m2.ao.delta.gemini37_flash.axis2.2_5\endcsname{[-0.141, -0.019]}
\expandafter\gdef\csname odunum@val@positive_taxonomy_m2.ao.delta.gemini37_flash.axis3.3_1\endcsname{0.057}
\expandafter\gdef\csname odunum@n@positive_taxonomy_m2.ao.delta.gemini37_flash.axis3.3_1\endcsname{18}
\expandafter\gdef\csname odunum@ci@positive_taxonomy_m2.ao.delta.gemini37_flash.axis3.3_1\endcsname{[-0.092, 0.194]}
\expandafter\gdef\csname odunum@val@positive_taxonomy_m2.ao.delta.gemini37_flash.axis3.3_2\endcsname{-0.028}
\expandafter\gdef\csname odunum@n@positive_taxonomy_m2.ao.delta.gemini37_flash.axis3.3_2\endcsname{30}
\expandafter\gdef\csname odunum@ci@positive_taxonomy_m2.ao.delta.gemini37_flash.axis3.3_2\endcsname{[-0.115, 0.060]}
\expandafter\gdef\csname odunum@val@positive_taxonomy_m2.ao.delta.gemini37_flash.axis3.3_3\endcsname{-0.100}
\expandafter\gdef\csname odunum@n@positive_taxonomy_m2.ao.delta.gemini37_flash.axis3.3_3\endcsname{67}
\expandafter\gdef\csname odunum@ci@positive_taxonomy_m2.ao.delta.gemini37_flash.axis3.3_3\endcsname{[-0.174, -0.029]}
\expandafter\gdef\csname odunum@val@positive_taxonomy_m2.ao.delta.gemini37_flash.axis3.3_4\endcsname{0.004}
\expandafter\gdef\csname odunum@n@positive_taxonomy_m2.ao.delta.gemini37_flash.axis3.3_4\endcsname{141}
\expandafter\gdef\csname odunum@ci@positive_taxonomy_m2.ao.delta.gemini37_flash.axis3.3_4\endcsname{[-0.037, 0.043]}
\expandafter\gdef\csname odunum@val@positive_taxonomy_m2.ao.delta.gemini37_flash.axis3.3_5\endcsname{-0.090}
\expandafter\gdef\csname odunum@n@positive_taxonomy_m2.ao.delta.gemini37_flash.axis3.3_5\endcsname{105}
\expandafter\gdef\csname odunum@ci@positive_taxonomy_m2.ao.delta.gemini37_flash.axis3.3_5\endcsname{[-0.137, -0.043]}
\expandafter\gdef\csname odunum@val@positive_taxonomy_m2.ao.delta.gemini37_flash.axis3.3_6\endcsname{0.077}
\expandafter\gdef\csname odunum@n@positive_taxonomy_m2.ao.delta.gemini37_flash.axis3.3_6\endcsname{202}
\expandafter\gdef\csname odunum@ci@positive_taxonomy_m2.ao.delta.gemini37_flash.axis3.3_6\endcsname{[0.048, 0.105]}
\expandafter\gdef\csname odunum@val@positive_taxonomy_m2.ao.delta.gemini37_flash.axis4.4_1\endcsname{0.032}
\expandafter\gdef\csname odunum@n@positive_taxonomy_m2.ao.delta.gemini37_flash.axis4.4_1\endcsname{63}
\expandafter\gdef\csname odunum@ci@positive_taxonomy_m2.ao.delta.gemini37_flash.axis4.4_1\endcsname{[-0.021, 0.085]}
\expandafter\gdef\csname odunum@val@positive_taxonomy_m2.ao.delta.gemini37_flash.axis4.4_2\endcsname{-0.008}
\expandafter\gdef\csname odunum@n@positive_taxonomy_m2.ao.delta.gemini37_flash.axis4.4_2\endcsname{161}
\expandafter\gdef\csname odunum@ci@positive_taxonomy_m2.ao.delta.gemini37_flash.axis4.4_2\endcsname{[-0.043, 0.027]}
\expandafter\gdef\csname odunum@val@positive_taxonomy_m2.ao.delta.gemini37_flash.axis4.4_3\endcsname{0.021}
\expandafter\gdef\csname odunum@n@positive_taxonomy_m2.ao.delta.gemini37_flash.axis4.4_3\endcsname{43}
\expandafter\gdef\csname odunum@ci@positive_taxonomy_m2.ao.delta.gemini37_flash.axis4.4_3\endcsname{[-0.064, 0.101]}
\expandafter\gdef\csname odunum@val@positive_taxonomy_m2.ao.delta.gemini37_flash.axis4.4_4\endcsname{-0.010}
\expandafter\gdef\csname odunum@n@positive_taxonomy_m2.ao.delta.gemini37_flash.axis4.4_4\endcsname{172}
\expandafter\gdef\csname odunum@ci@positive_taxonomy_m2.ao.delta.gemini37_flash.axis4.4_4\endcsname{[-0.048, 0.027]}
\expandafter\gdef\csname odunum@val@positive_taxonomy_m2.ao.delta.gemini37_flash.axis4.4_5\endcsname{3.6\ensuremath{\times 10^{-4}}}
\expandafter\gdef\csname odunum@n@positive_taxonomy_m2.ao.delta.gemini37_flash.axis4.4_5\endcsname{124}
\expandafter\gdef\csname odunum@ci@positive_taxonomy_m2.ao.delta.gemini37_flash.axis4.4_5\endcsname{[-0.044, 0.045]}
\expandafter\gdef\csname odunum@val@positive_taxonomy_m2.ao.delta.gpt_realtime.axis1.1_1\endcsname{0.058}
\expandafter\gdef\csname odunum@n@positive_taxonomy_m2.ao.delta.gpt_realtime.axis1.1_1\endcsname{156}
\expandafter\gdef\csname odunum@ci@positive_taxonomy_m2.ao.delta.gpt_realtime.axis1.1_1\endcsname{[0.020, 0.096]}
\expandafter\gdef\csname odunum@val@positive_taxonomy_m2.ao.delta.gpt_realtime.axis1.1_2\endcsname{-0.022}
\expandafter\gdef\csname odunum@n@positive_taxonomy_m2.ao.delta.gpt_realtime.axis1.1_2\endcsname{407}
\expandafter\gdef\csname odunum@ci@positive_taxonomy_m2.ao.delta.gpt_realtime.axis1.1_2\endcsname{[-0.038, -0.007]}
\expandafter\gdef\csname odunum@val@positive_taxonomy_m2.ao.delta.gpt_realtime.axis2.2_1\endcsname{-0.128}
\expandafter\gdef\csname odunum@n@positive_taxonomy_m2.ao.delta.gpt_realtime.axis2.2_1\endcsname{21}
\expandafter\gdef\csname odunum@ci@positive_taxonomy_m2.ao.delta.gpt_realtime.axis2.2_1\endcsname{[-0.281, 0.019]}
\expandafter\gdef\csname odunum@val@positive_taxonomy_m2.ao.delta.gpt_realtime.axis2.2_2\endcsname{0.017}
\expandafter\gdef\csname odunum@n@positive_taxonomy_m2.ao.delta.gpt_realtime.axis2.2_2\endcsname{169}
\expandafter\gdef\csname odunum@ci@positive_taxonomy_m2.ao.delta.gpt_realtime.axis2.2_2\endcsname{[-0.017, 0.051]}
\expandafter\gdef\csname odunum@val@positive_taxonomy_m2.ao.delta.gpt_realtime.axis2.2_3\endcsname{0.048}
\expandafter\gdef\csname odunum@n@positive_taxonomy_m2.ao.delta.gpt_realtime.axis2.2_3\endcsname{248}
\expandafter\gdef\csname odunum@ci@positive_taxonomy_m2.ao.delta.gpt_realtime.axis2.2_3\endcsname{[0.022, 0.075]}
\expandafter\gdef\csname odunum@val@positive_taxonomy_m2.ao.delta.gpt_realtime.axis2.2_4\endcsname{-0.175}
\expandafter\gdef\csname odunum@n@positive_taxonomy_m2.ao.delta.gpt_realtime.axis2.2_4\endcsname{34}
\expandafter\gdef\csname odunum@ci@positive_taxonomy_m2.ao.delta.gpt_realtime.axis2.2_4\endcsname{[-0.300, -0.053]}
\expandafter\gdef\csname odunum@val@positive_taxonomy_m2.ao.delta.gpt_realtime.axis2.2_5\endcsname{-0.068}
\expandafter\gdef\csname odunum@n@positive_taxonomy_m2.ao.delta.gpt_realtime.axis2.2_5\endcsname{91}
\expandafter\gdef\csname odunum@ci@positive_taxonomy_m2.ao.delta.gpt_realtime.axis2.2_5\endcsname{[-0.133, -0.006]}
\expandafter\gdef\csname odunum@val@positive_taxonomy_m2.ao.delta.gpt_realtime.axis3.3_1\endcsname{0.096}
\expandafter\gdef\csname odunum@n@positive_taxonomy_m2.ao.delta.gpt_realtime.axis3.3_1\endcsname{18}
\expandafter\gdef\csname odunum@ci@positive_taxonomy_m2.ao.delta.gpt_realtime.axis3.3_1\endcsname{[-0.038, 0.221]}
\expandafter\gdef\csname odunum@val@positive_taxonomy_m2.ao.delta.gpt_realtime.axis3.3_2\endcsname{-0.051}
\expandafter\gdef\csname odunum@n@positive_taxonomy_m2.ao.delta.gpt_realtime.axis3.3_2\endcsname{30}
\expandafter\gdef\csname odunum@ci@positive_taxonomy_m2.ao.delta.gpt_realtime.axis3.3_2\endcsname{[-0.156, 0.052]}
\expandafter\gdef\csname odunum@val@positive_taxonomy_m2.ao.delta.gpt_realtime.axis3.3_3\endcsname{-0.067}
\expandafter\gdef\csname odunum@n@positive_taxonomy_m2.ao.delta.gpt_realtime.axis3.3_3\endcsname{67}
\expandafter\gdef\csname odunum@ci@positive_taxonomy_m2.ao.delta.gpt_realtime.axis3.3_3\endcsname{[-0.150, 0.016]}
\expandafter\gdef\csname odunum@val@positive_taxonomy_m2.ao.delta.gpt_realtime.axis3.3_4\endcsname{-0.002}
\expandafter\gdef\csname odunum@n@positive_taxonomy_m2.ao.delta.gpt_realtime.axis3.3_4\endcsname{141}
\expandafter\gdef\csname odunum@ci@positive_taxonomy_m2.ao.delta.gpt_realtime.axis3.3_4\endcsname{[-0.043, 0.039]}
\expandafter\gdef\csname odunum@val@positive_taxonomy_m2.ao.delta.gpt_realtime.axis3.3_5\endcsname{-0.053}
\expandafter\gdef\csname odunum@n@positive_taxonomy_m2.ao.delta.gpt_realtime.axis3.3_5\endcsname{105}
\expandafter\gdef\csname odunum@ci@positive_taxonomy_m2.ao.delta.gpt_realtime.axis3.3_5\endcsname{[-0.100, -0.006]}
\expandafter\gdef\csname odunum@val@positive_taxonomy_m2.ao.delta.gpt_realtime.axis3.3_6\endcsname{0.050}
\expandafter\gdef\csname odunum@n@positive_taxonomy_m2.ao.delta.gpt_realtime.axis3.3_6\endcsname{202}
\expandafter\gdef\csname odunum@ci@positive_taxonomy_m2.ao.delta.gpt_realtime.axis3.3_6\endcsname{[0.018, 0.082]}
\expandafter\gdef\csname odunum@val@positive_taxonomy_m2.ao.delta.gpt_realtime.axis4.4_1\endcsname{0.001}
\expandafter\gdef\csname odunum@n@positive_taxonomy_m2.ao.delta.gpt_realtime.axis4.4_1\endcsname{63}
\expandafter\gdef\csname odunum@ci@positive_taxonomy_m2.ao.delta.gpt_realtime.axis4.4_1\endcsname{[-0.067, 0.066]}
\expandafter\gdef\csname odunum@val@positive_taxonomy_m2.ao.delta.gpt_realtime.axis4.4_2\endcsname{0.005}
\expandafter\gdef\csname odunum@n@positive_taxonomy_m2.ao.delta.gpt_realtime.axis4.4_2\endcsname{161}
\expandafter\gdef\csname odunum@ci@positive_taxonomy_m2.ao.delta.gpt_realtime.axis4.4_2\endcsname{[-0.033, 0.042]}
\expandafter\gdef\csname odunum@val@positive_taxonomy_m2.ao.delta.gpt_realtime.axis4.4_3\endcsname{-0.054}
\expandafter\gdef\csname odunum@n@positive_taxonomy_m2.ao.delta.gpt_realtime.axis4.4_3\endcsname{43}
\expandafter\gdef\csname odunum@ci@positive_taxonomy_m2.ao.delta.gpt_realtime.axis4.4_3\endcsname{[-0.157, 0.045]}
\expandafter\gdef\csname odunum@val@positive_taxonomy_m2.ao.delta.gpt_realtime.axis4.4_4\endcsname{0.007}
\expandafter\gdef\csname odunum@n@positive_taxonomy_m2.ao.delta.gpt_realtime.axis4.4_4\endcsname{172}
\expandafter\gdef\csname odunum@ci@positive_taxonomy_m2.ao.delta.gpt_realtime.axis4.4_4\endcsname{[-0.029, 0.044]}
\expandafter\gdef\csname odunum@val@positive_taxonomy_m2.ao.delta.gpt_realtime.axis4.4_5\endcsname{0.002}
\expandafter\gdef\csname odunum@n@positive_taxonomy_m2.ao.delta.gpt_realtime.axis4.4_5\endcsname{124}
\expandafter\gdef\csname odunum@ci@positive_taxonomy_m2.ao.delta.gpt_realtime.axis4.4_5\endcsname{[-0.045, 0.049]}
\expandafter\gdef\csname odunum@val@positive_taxonomy_m2.ao.delta.kimi_audio_7b_instruct.axis1.1_1\endcsname{0.055}
\expandafter\gdef\csname odunum@n@positive_taxonomy_m2.ao.delta.kimi_audio_7b_instruct.axis1.1_1\endcsname{156}
\expandafter\gdef\csname odunum@ci@positive_taxonomy_m2.ao.delta.kimi_audio_7b_instruct.axis1.1_1\endcsname{[0.013, 0.099]}
\expandafter\gdef\csname odunum@val@positive_taxonomy_m2.ao.delta.kimi_audio_7b_instruct.axis1.1_2\endcsname{-0.021}
\expandafter\gdef\csname odunum@n@positive_taxonomy_m2.ao.delta.kimi_audio_7b_instruct.axis1.1_2\endcsname{407}
\expandafter\gdef\csname odunum@ci@positive_taxonomy_m2.ao.delta.kimi_audio_7b_instruct.axis1.1_2\endcsname{[-0.038, -0.005]}
\expandafter\gdef\csname odunum@val@positive_taxonomy_m2.ao.delta.kimi_audio_7b_instruct.axis2.2_1\endcsname{-0.115}
\expandafter\gdef\csname odunum@n@positive_taxonomy_m2.ao.delta.kimi_audio_7b_instruct.axis2.2_1\endcsname{21}
\expandafter\gdef\csname odunum@ci@positive_taxonomy_m2.ao.delta.kimi_audio_7b_instruct.axis2.2_1\endcsname{[-0.273, 0.057]}
\expandafter\gdef\csname odunum@val@positive_taxonomy_m2.ao.delta.kimi_audio_7b_instruct.axis2.2_2\endcsname{0.034}
\expandafter\gdef\csname odunum@n@positive_taxonomy_m2.ao.delta.kimi_audio_7b_instruct.axis2.2_2\endcsname{169}
\expandafter\gdef\csname odunum@ci@positive_taxonomy_m2.ao.delta.kimi_audio_7b_instruct.axis2.2_2\endcsname{[-0.005, 0.072]}
\expandafter\gdef\csname odunum@val@positive_taxonomy_m2.ao.delta.kimi_audio_7b_instruct.axis2.2_3\endcsname{0.060}
\expandafter\gdef\csname odunum@n@positive_taxonomy_m2.ao.delta.kimi_audio_7b_instruct.axis2.2_3\endcsname{248}
\expandafter\gdef\csname odunum@ci@positive_taxonomy_m2.ao.delta.kimi_audio_7b_instruct.axis2.2_3\endcsname{[0.031, 0.088]}
\expandafter\gdef\csname odunum@val@positive_taxonomy_m2.ao.delta.kimi_audio_7b_instruct.axis2.2_4\endcsname{-0.169}
\expandafter\gdef\csname odunum@n@positive_taxonomy_m2.ao.delta.kimi_audio_7b_instruct.axis2.2_4\endcsname{34}
\expandafter\gdef\csname odunum@ci@positive_taxonomy_m2.ao.delta.kimi_audio_7b_instruct.axis2.2_4\endcsname{[-0.269, -0.065]}
\expandafter\gdef\csname odunum@val@positive_taxonomy_m2.ao.delta.kimi_audio_7b_instruct.axis2.2_5\endcsname{-0.136}
\expandafter\gdef\csname odunum@n@positive_taxonomy_m2.ao.delta.kimi_audio_7b_instruct.axis2.2_5\endcsname{91}
\expandafter\gdef\csname odunum@ci@positive_taxonomy_m2.ao.delta.kimi_audio_7b_instruct.axis2.2_5\endcsname{[-0.200, -0.071]}
\expandafter\gdef\csname odunum@val@positive_taxonomy_m2.ao.delta.kimi_audio_7b_instruct.axis3.3_1\endcsname{0.077}
\expandafter\gdef\csname odunum@n@positive_taxonomy_m2.ao.delta.kimi_audio_7b_instruct.axis3.3_1\endcsname{18}
\expandafter\gdef\csname odunum@ci@positive_taxonomy_m2.ao.delta.kimi_audio_7b_instruct.axis3.3_1\endcsname{[-0.074, 0.226]}
\expandafter\gdef\csname odunum@val@positive_taxonomy_m2.ao.delta.kimi_audio_7b_instruct.axis3.3_2\endcsname{0.018}
\expandafter\gdef\csname odunum@n@positive_taxonomy_m2.ao.delta.kimi_audio_7b_instruct.axis3.3_2\endcsname{30}
\expandafter\gdef\csname odunum@ci@positive_taxonomy_m2.ao.delta.kimi_audio_7b_instruct.axis3.3_2\endcsname{[-0.087, 0.124]}
\expandafter\gdef\csname odunum@val@positive_taxonomy_m2.ao.delta.kimi_audio_7b_instruct.axis3.3_3\endcsname{-0.034}
\expandafter\gdef\csname odunum@n@positive_taxonomy_m2.ao.delta.kimi_audio_7b_instruct.axis3.3_3\endcsname{67}
\expandafter\gdef\csname odunum@ci@positive_taxonomy_m2.ao.delta.kimi_audio_7b_instruct.axis3.3_3\endcsname{[-0.110, 0.041]}
\expandafter\gdef\csname odunum@val@positive_taxonomy_m2.ao.delta.kimi_audio_7b_instruct.axis3.3_4\endcsname{0.013}
\expandafter\gdef\csname odunum@n@positive_taxonomy_m2.ao.delta.kimi_audio_7b_instruct.axis3.3_4\endcsname{141}
\expandafter\gdef\csname odunum@ci@positive_taxonomy_m2.ao.delta.kimi_audio_7b_instruct.axis3.3_4\endcsname{[-0.033, 0.060]}
\expandafter\gdef\csname odunum@val@positive_taxonomy_m2.ao.delta.kimi_audio_7b_instruct.axis3.3_5\endcsname{-0.057}
\expandafter\gdef\csname odunum@n@positive_taxonomy_m2.ao.delta.kimi_audio_7b_instruct.axis3.3_5\endcsname{105}
\expandafter\gdef\csname odunum@ci@positive_taxonomy_m2.ao.delta.kimi_audio_7b_instruct.axis3.3_5\endcsname{[-0.112, -0.004]}
\expandafter\gdef\csname odunum@val@positive_taxonomy_m2.ao.delta.kimi_audio_7b_instruct.axis3.3_6\endcsname{0.022}
\expandafter\gdef\csname odunum@n@positive_taxonomy_m2.ao.delta.kimi_audio_7b_instruct.axis3.3_6\endcsname{202}
\expandafter\gdef\csname odunum@ci@positive_taxonomy_m2.ao.delta.kimi_audio_7b_instruct.axis3.3_6\endcsname{[-0.012, 0.056]}
\expandafter\gdef\csname odunum@val@positive_taxonomy_m2.ao.delta.kimi_audio_7b_instruct.axis4.4_1\endcsname{0.013}
\expandafter\gdef\csname odunum@n@positive_taxonomy_m2.ao.delta.kimi_audio_7b_instruct.axis4.4_1\endcsname{63}
\expandafter\gdef\csname odunum@ci@positive_taxonomy_m2.ao.delta.kimi_audio_7b_instruct.axis4.4_1\endcsname{[-0.060, 0.085]}
\expandafter\gdef\csname odunum@val@positive_taxonomy_m2.ao.delta.kimi_audio_7b_instruct.axis4.4_2\endcsname{-0.031}
\expandafter\gdef\csname odunum@n@positive_taxonomy_m2.ao.delta.kimi_audio_7b_instruct.axis4.4_2\endcsname{161}
\expandafter\gdef\csname odunum@ci@positive_taxonomy_m2.ao.delta.kimi_audio_7b_instruct.axis4.4_2\endcsname{[-0.071, 0.009]}
\expandafter\gdef\csname odunum@val@positive_taxonomy_m2.ao.delta.kimi_audio_7b_instruct.axis4.4_3\endcsname{0.005}
\expandafter\gdef\csname odunum@n@positive_taxonomy_m2.ao.delta.kimi_audio_7b_instruct.axis4.4_3\endcsname{43}
\expandafter\gdef\csname odunum@ci@positive_taxonomy_m2.ao.delta.kimi_audio_7b_instruct.axis4.4_3\endcsname{[-0.087, 0.095]}
\expandafter\gdef\csname odunum@val@positive_taxonomy_m2.ao.delta.kimi_audio_7b_instruct.axis4.4_4\endcsname{-0.001}
\expandafter\gdef\csname odunum@n@positive_taxonomy_m2.ao.delta.kimi_audio_7b_instruct.axis4.4_4\endcsname{172}
\expandafter\gdef\csname odunum@ci@positive_taxonomy_m2.ao.delta.kimi_audio_7b_instruct.axis4.4_4\endcsname{[-0.041, 0.039]}
\expandafter\gdef\csname odunum@val@positive_taxonomy_m2.ao.delta.kimi_audio_7b_instruct.axis4.4_5\endcsname{0.033}
\expandafter\gdef\csname odunum@n@positive_taxonomy_m2.ao.delta.kimi_audio_7b_instruct.axis4.4_5\endcsname{124}
\expandafter\gdef\csname odunum@ci@positive_taxonomy_m2.ao.delta.kimi_audio_7b_instruct.axis4.4_5\endcsname{[-0.019, 0.085]}
\expandafter\gdef\csname odunum@val@positive_taxonomy_m2.ao.delta.ming.axis1.1_1\endcsname{0.038}
\expandafter\gdef\csname odunum@n@positive_taxonomy_m2.ao.delta.ming.axis1.1_1\endcsname{156}
\expandafter\gdef\csname odunum@ci@positive_taxonomy_m2.ao.delta.ming.axis1.1_1\endcsname{[-0.003, 0.080]}
\expandafter\gdef\csname odunum@val@positive_taxonomy_m2.ao.delta.ming.axis1.1_2\endcsname{-0.015}
\expandafter\gdef\csname odunum@n@positive_taxonomy_m2.ao.delta.ming.axis1.1_2\endcsname{407}
\expandafter\gdef\csname odunum@ci@positive_taxonomy_m2.ao.delta.ming.axis1.1_2\endcsname{[-0.031, 0.001]}
\expandafter\gdef\csname odunum@val@positive_taxonomy_m2.ao.delta.ming.axis2.2_1\endcsname{-0.047}
\expandafter\gdef\csname odunum@n@positive_taxonomy_m2.ao.delta.ming.axis2.2_1\endcsname{21}
\expandafter\gdef\csname odunum@ci@positive_taxonomy_m2.ao.delta.ming.axis2.2_1\endcsname{[-0.201, 0.113]}
\expandafter\gdef\csname odunum@val@positive_taxonomy_m2.ao.delta.ming.axis2.2_2\endcsname{0.020}
\expandafter\gdef\csname odunum@n@positive_taxonomy_m2.ao.delta.ming.axis2.2_2\endcsname{169}
\expandafter\gdef\csname odunum@ci@positive_taxonomy_m2.ao.delta.ming.axis2.2_2\endcsname{[-0.016, 0.056]}
\expandafter\gdef\csname odunum@val@positive_taxonomy_m2.ao.delta.ming.axis2.2_3\endcsname{0.041}
\expandafter\gdef\csname odunum@n@positive_taxonomy_m2.ao.delta.ming.axis2.2_3\endcsname{248}
\expandafter\gdef\csname odunum@ci@positive_taxonomy_m2.ao.delta.ming.axis2.2_3\endcsname{[0.013, 0.069]}
\expandafter\gdef\csname odunum@val@positive_taxonomy_m2.ao.delta.ming.axis2.2_4\endcsname{-0.133}
\expandafter\gdef\csname odunum@n@positive_taxonomy_m2.ao.delta.ming.axis2.2_4\endcsname{34}
\expandafter\gdef\csname odunum@ci@positive_taxonomy_m2.ao.delta.ming.axis2.2_4\endcsname{[-0.238, -0.027]}
\expandafter\gdef\csname odunum@val@positive_taxonomy_m2.ao.delta.ming.axis2.2_5\endcsname{-0.088}
\expandafter\gdef\csname odunum@n@positive_taxonomy_m2.ao.delta.ming.axis2.2_5\endcsname{91}
\expandafter\gdef\csname odunum@ci@positive_taxonomy_m2.ao.delta.ming.axis2.2_5\endcsname{[-0.147, -0.028]}
\expandafter\gdef\csname odunum@val@positive_taxonomy_m2.ao.delta.ming.axis3.3_1\endcsname{0.057}
\expandafter\gdef\csname odunum@n@positive_taxonomy_m2.ao.delta.ming.axis3.3_1\endcsname{18}
\expandafter\gdef\csname odunum@ci@positive_taxonomy_m2.ao.delta.ming.axis3.3_1\endcsname{[-0.086, 0.203]}
\expandafter\gdef\csname odunum@val@positive_taxonomy_m2.ao.delta.ming.axis3.3_2\endcsname{-0.041}
\expandafter\gdef\csname odunum@n@positive_taxonomy_m2.ao.delta.ming.axis3.3_2\endcsname{30}
\expandafter\gdef\csname odunum@ci@positive_taxonomy_m2.ao.delta.ming.axis3.3_2\endcsname{[-0.135, 0.052]}
\expandafter\gdef\csname odunum@val@positive_taxonomy_m2.ao.delta.ming.axis3.3_3\endcsname{-0.137}
\expandafter\gdef\csname odunum@n@positive_taxonomy_m2.ao.delta.ming.axis3.3_3\endcsname{67}
\expandafter\gdef\csname odunum@ci@positive_taxonomy_m2.ao.delta.ming.axis3.3_3\endcsname{[-0.204, -0.071]}
\expandafter\gdef\csname odunum@val@positive_taxonomy_m2.ao.delta.ming.axis3.3_4\endcsname{0.018}
\expandafter\gdef\csname odunum@n@positive_taxonomy_m2.ao.delta.ming.axis3.3_4\endcsname{141}
\expandafter\gdef\csname odunum@ci@positive_taxonomy_m2.ao.delta.ming.axis3.3_4\endcsname{[-0.025, 0.064]}
\expandafter\gdef\csname odunum@val@positive_taxonomy_m2.ao.delta.ming.axis3.3_5\endcsname{-0.040}
\expandafter\gdef\csname odunum@n@positive_taxonomy_m2.ao.delta.ming.axis3.3_5\endcsname{105}
\expandafter\gdef\csname odunum@ci@positive_taxonomy_m2.ao.delta.ming.axis3.3_5\endcsname{[-0.089, 0.010]}
\expandafter\gdef\csname odunum@val@positive_taxonomy_m2.ao.delta.ming.axis3.3_6\endcsname{0.054}
\expandafter\gdef\csname odunum@n@positive_taxonomy_m2.ao.delta.ming.axis3.3_6\endcsname{202}
\expandafter\gdef\csname odunum@ci@positive_taxonomy_m2.ao.delta.ming.axis3.3_6\endcsname{[0.020, 0.088]}
\expandafter\gdef\csname odunum@val@positive_taxonomy_m2.ao.delta.ming.axis4.4_1\endcsname{0.056}
\expandafter\gdef\csname odunum@n@positive_taxonomy_m2.ao.delta.ming.axis4.4_1\endcsname{63}
\expandafter\gdef\csname odunum@ci@positive_taxonomy_m2.ao.delta.ming.axis4.4_1\endcsname{[-0.012, 0.122]}
\expandafter\gdef\csname odunum@val@positive_taxonomy_m2.ao.delta.ming.axis4.4_2\endcsname{-0.033}
\expandafter\gdef\csname odunum@n@positive_taxonomy_m2.ao.delta.ming.axis4.4_2\endcsname{161}
\expandafter\gdef\csname odunum@ci@positive_taxonomy_m2.ao.delta.ming.axis4.4_2\endcsname{[-0.071, 0.006]}
\expandafter\gdef\csname odunum@val@positive_taxonomy_m2.ao.delta.ming.axis4.4_3\endcsname{-0.010}
\expandafter\gdef\csname odunum@n@positive_taxonomy_m2.ao.delta.ming.axis4.4_3\endcsname{43}
\expandafter\gdef\csname odunum@ci@positive_taxonomy_m2.ao.delta.ming.axis4.4_3\endcsname{[-0.100, 0.082]}
\expandafter\gdef\csname odunum@val@positive_taxonomy_m2.ao.delta.ming.axis4.4_4\endcsname{-0.014}
\expandafter\gdef\csname odunum@n@positive_taxonomy_m2.ao.delta.ming.axis4.4_4\endcsname{172}
\expandafter\gdef\csname odunum@ci@positive_taxonomy_m2.ao.delta.ming.axis4.4_4\endcsname{[-0.052, 0.024]}
\expandafter\gdef\csname odunum@val@positive_taxonomy_m2.ao.delta.ming.axis4.4_5\endcsname{0.037}
\expandafter\gdef\csname odunum@n@positive_taxonomy_m2.ao.delta.ming.axis4.4_5\endcsname{124}
\expandafter\gdef\csname odunum@ci@positive_taxonomy_m2.ao.delta.ming.axis4.4_5\endcsname{[-0.013, 0.088]}
\expandafter\gdef\csname odunum@val@positive_taxonomy_m2.ao.delta.minicpm_o.axis1.1_1\endcsname{0.030}
\expandafter\gdef\csname odunum@n@positive_taxonomy_m2.ao.delta.minicpm_o.axis1.1_1\endcsname{156}
\expandafter\gdef\csname odunum@ci@positive_taxonomy_m2.ao.delta.minicpm_o.axis1.1_1\endcsname{[-0.020, 0.079]}
\expandafter\gdef\csname odunum@val@positive_taxonomy_m2.ao.delta.minicpm_o.axis1.1_2\endcsname{-0.011}
\expandafter\gdef\csname odunum@n@positive_taxonomy_m2.ao.delta.minicpm_o.axis1.1_2\endcsname{407}
\expandafter\gdef\csname odunum@ci@positive_taxonomy_m2.ao.delta.minicpm_o.axis1.1_2\endcsname{[-0.030, 0.007]}
\expandafter\gdef\csname odunum@val@positive_taxonomy_m2.ao.delta.minicpm_o.axis2.2_1\endcsname{0.160}
\expandafter\gdef\csname odunum@n@positive_taxonomy_m2.ao.delta.minicpm_o.axis2.2_1\endcsname{21}
\expandafter\gdef\csname odunum@ci@positive_taxonomy_m2.ao.delta.minicpm_o.axis2.2_1\endcsname{[0.004, 0.312]}
\expandafter\gdef\csname odunum@val@positive_taxonomy_m2.ao.delta.minicpm_o.axis2.2_2\endcsname{-0.022}
\expandafter\gdef\csname odunum@n@positive_taxonomy_m2.ao.delta.minicpm_o.axis2.2_2\endcsname{169}
\expandafter\gdef\csname odunum@ci@positive_taxonomy_m2.ao.delta.minicpm_o.axis2.2_2\endcsname{[-0.064, 0.022]}
\expandafter\gdef\csname odunum@val@positive_taxonomy_m2.ao.delta.minicpm_o.axis2.2_3\endcsname{0.008}
\expandafter\gdef\csname odunum@n@positive_taxonomy_m2.ao.delta.minicpm_o.axis2.2_3\endcsname{248}
\expandafter\gdef\csname odunum@ci@positive_taxonomy_m2.ao.delta.minicpm_o.axis2.2_3\endcsname{[-0.024, 0.041]}
\expandafter\gdef\csname odunum@val@positive_taxonomy_m2.ao.delta.minicpm_o.axis2.2_4\endcsname{-0.048}
\expandafter\gdef\csname odunum@n@positive_taxonomy_m2.ao.delta.minicpm_o.axis2.2_4\endcsname{34}
\expandafter\gdef\csname odunum@ci@positive_taxonomy_m2.ao.delta.minicpm_o.axis2.2_4\endcsname{[-0.155, 0.061]}
\expandafter\gdef\csname odunum@val@positive_taxonomy_m2.ao.delta.minicpm_o.axis2.2_5\endcsname{-0.001}
\expandafter\gdef\csname odunum@n@positive_taxonomy_m2.ao.delta.minicpm_o.axis2.2_5\endcsname{91}
\expandafter\gdef\csname odunum@ci@positive_taxonomy_m2.ao.delta.minicpm_o.axis2.2_5\endcsname{[-0.064, 0.062]}
\expandafter\gdef\csname odunum@val@positive_taxonomy_m2.ao.delta.minicpm_o.axis3.3_1\endcsname{0.005}
\expandafter\gdef\csname odunum@n@positive_taxonomy_m2.ao.delta.minicpm_o.axis3.3_1\endcsname{18}
\expandafter\gdef\csname odunum@ci@positive_taxonomy_m2.ao.delta.minicpm_o.axis3.3_1\endcsname{[-0.167, 0.177]}
\expandafter\gdef\csname odunum@val@positive_taxonomy_m2.ao.delta.minicpm_o.axis3.3_2\endcsname{0.052}
\expandafter\gdef\csname odunum@n@positive_taxonomy_m2.ao.delta.minicpm_o.axis3.3_2\endcsname{30}
\expandafter\gdef\csname odunum@ci@positive_taxonomy_m2.ao.delta.minicpm_o.axis3.3_2\endcsname{[-0.049, 0.152]}
\expandafter\gdef\csname odunum@val@positive_taxonomy_m2.ao.delta.minicpm_o.axis3.3_3\endcsname{-0.056}
\expandafter\gdef\csname odunum@n@positive_taxonomy_m2.ao.delta.minicpm_o.axis3.3_3\endcsname{67}
\expandafter\gdef\csname odunum@ci@positive_taxonomy_m2.ao.delta.minicpm_o.axis3.3_3\endcsname{[-0.143, 0.032]}
\expandafter\gdef\csname odunum@val@positive_taxonomy_m2.ao.delta.minicpm_o.axis3.3_4\endcsname{0.021}
\expandafter\gdef\csname odunum@n@positive_taxonomy_m2.ao.delta.minicpm_o.axis3.3_4\endcsname{141}
\expandafter\gdef\csname odunum@ci@positive_taxonomy_m2.ao.delta.minicpm_o.axis3.3_4\endcsname{[-0.030, 0.072]}
\expandafter\gdef\csname odunum@val@positive_taxonomy_m2.ao.delta.minicpm_o.axis3.3_5\endcsname{-0.062}
\expandafter\gdef\csname odunum@n@positive_taxonomy_m2.ao.delta.minicpm_o.axis3.3_5\endcsname{105}
\expandafter\gdef\csname odunum@ci@positive_taxonomy_m2.ao.delta.minicpm_o.axis3.3_5\endcsname{[-0.118, -0.007]}
\expandafter\gdef\csname odunum@val@positive_taxonomy_m2.ao.delta.minicpm_o.axis3.3_6\endcsname{0.028}
\expandafter\gdef\csname odunum@n@positive_taxonomy_m2.ao.delta.minicpm_o.axis3.3_6\endcsname{202}
\expandafter\gdef\csname odunum@ci@positive_taxonomy_m2.ao.delta.minicpm_o.axis3.3_6\endcsname{[-0.009, 0.066]}
\expandafter\gdef\csname odunum@val@positive_taxonomy_m2.ao.delta.minicpm_o.axis4.4_1\endcsname{-0.029}
\expandafter\gdef\csname odunum@n@positive_taxonomy_m2.ao.delta.minicpm_o.axis4.4_1\endcsname{63}
\expandafter\gdef\csname odunum@ci@positive_taxonomy_m2.ao.delta.minicpm_o.axis4.4_1\endcsname{[-0.106, 0.046]}
\expandafter\gdef\csname odunum@val@positive_taxonomy_m2.ao.delta.minicpm_o.axis4.4_2\endcsname{0.005}
\expandafter\gdef\csname odunum@n@positive_taxonomy_m2.ao.delta.minicpm_o.axis4.4_2\endcsname{161}
\expandafter\gdef\csname odunum@ci@positive_taxonomy_m2.ao.delta.minicpm_o.axis4.4_2\endcsname{[-0.040, 0.050]}
\expandafter\gdef\csname odunum@val@positive_taxonomy_m2.ao.delta.minicpm_o.axis4.4_3\endcsname{-0.048}
\expandafter\gdef\csname odunum@n@positive_taxonomy_m2.ao.delta.minicpm_o.axis4.4_3\endcsname{43}
\expandafter\gdef\csname odunum@ci@positive_taxonomy_m2.ao.delta.minicpm_o.axis4.4_3\endcsname{[-0.145, 0.047]}
\expandafter\gdef\csname odunum@val@positive_taxonomy_m2.ao.delta.minicpm_o.axis4.4_4\endcsname{-0.042}
\expandafter\gdef\csname odunum@n@positive_taxonomy_m2.ao.delta.minicpm_o.axis4.4_4\endcsname{172}
\expandafter\gdef\csname odunum@ci@positive_taxonomy_m2.ao.delta.minicpm_o.axis4.4_4\endcsname{[-0.085, 0.001]}
\expandafter\gdef\csname odunum@val@positive_taxonomy_m2.ao.delta.minicpm_o.axis4.4_5\endcsname{0.083}
\expandafter\gdef\csname odunum@n@positive_taxonomy_m2.ao.delta.minicpm_o.axis4.4_5\endcsname{124}
\expandafter\gdef\csname odunum@ci@positive_taxonomy_m2.ao.delta.minicpm_o.axis4.4_5\endcsname{[0.026, 0.140]}
\expandafter\gdef\csname odunum@val@positive_taxonomy_m2.ao.delta.nemotron.axis1.1_1\endcsname{0.003}
\expandafter\gdef\csname odunum@n@positive_taxonomy_m2.ao.delta.nemotron.axis1.1_1\endcsname{156}
\expandafter\gdef\csname odunum@ci@positive_taxonomy_m2.ao.delta.nemotron.axis1.1_1\endcsname{[-0.047, 0.053]}
\expandafter\gdef\csname odunum@val@positive_taxonomy_m2.ao.delta.nemotron.axis1.1_2\endcsname{-0.001}
\expandafter\gdef\csname odunum@n@positive_taxonomy_m2.ao.delta.nemotron.axis1.1_2\endcsname{407}
\expandafter\gdef\csname odunum@ci@positive_taxonomy_m2.ao.delta.nemotron.axis1.1_2\endcsname{[-0.020, 0.018]}
\expandafter\gdef\csname odunum@val@positive_taxonomy_m2.ao.delta.nemotron.axis2.2_1\endcsname{0.090}
\expandafter\gdef\csname odunum@n@positive_taxonomy_m2.ao.delta.nemotron.axis2.2_1\endcsname{21}
\expandafter\gdef\csname odunum@ci@positive_taxonomy_m2.ao.delta.nemotron.axis2.2_1\endcsname{[-0.077, 0.258]}
\expandafter\gdef\csname odunum@val@positive_taxonomy_m2.ao.delta.nemotron.axis2.2_2\endcsname{0.038}
\expandafter\gdef\csname odunum@n@positive_taxonomy_m2.ao.delta.nemotron.axis2.2_2\endcsname{169}
\expandafter\gdef\csname odunum@ci@positive_taxonomy_m2.ao.delta.nemotron.axis2.2_2\endcsname{[-0.007, 0.085]}
\expandafter\gdef\csname odunum@val@positive_taxonomy_m2.ao.delta.nemotron.axis2.2_3\endcsname{-0.010}
\expandafter\gdef\csname odunum@n@positive_taxonomy_m2.ao.delta.nemotron.axis2.2_3\endcsname{248}
\expandafter\gdef\csname odunum@ci@positive_taxonomy_m2.ao.delta.nemotron.axis2.2_3\endcsname{[-0.044, 0.023]}
\expandafter\gdef\csname odunum@val@positive_taxonomy_m2.ao.delta.nemotron.axis2.2_4\endcsname{0.052}
\expandafter\gdef\csname odunum@n@positive_taxonomy_m2.ao.delta.nemotron.axis2.2_4\endcsname{34}
\expandafter\gdef\csname odunum@ci@positive_taxonomy_m2.ao.delta.nemotron.axis2.2_4\endcsname{[-0.077, 0.185]}
\expandafter\gdef\csname odunum@val@positive_taxonomy_m2.ao.delta.nemotron.axis2.2_5\endcsname{-0.083}
\expandafter\gdef\csname odunum@n@positive_taxonomy_m2.ao.delta.nemotron.axis2.2_5\endcsname{91}
\expandafter\gdef\csname odunum@ci@positive_taxonomy_m2.ao.delta.nemotron.axis2.2_5\endcsname{[-0.147, -0.018]}
\expandafter\gdef\csname odunum@val@positive_taxonomy_m2.ao.delta.nemotron.axis3.3_1\endcsname{0.037}
\expandafter\gdef\csname odunum@n@positive_taxonomy_m2.ao.delta.nemotron.axis3.3_1\endcsname{18}
\expandafter\gdef\csname odunum@ci@positive_taxonomy_m2.ao.delta.nemotron.axis3.3_1\endcsname{[-0.123, 0.203]}
\expandafter\gdef\csname odunum@val@positive_taxonomy_m2.ao.delta.nemotron.axis3.3_2\endcsname{0.023}
\expandafter\gdef\csname odunum@n@positive_taxonomy_m2.ao.delta.nemotron.axis3.3_2\endcsname{30}
\expandafter\gdef\csname odunum@ci@positive_taxonomy_m2.ao.delta.nemotron.axis3.3_2\endcsname{[-0.088, 0.135]}
\expandafter\gdef\csname odunum@val@positive_taxonomy_m2.ao.delta.nemotron.axis3.3_3\endcsname{-0.174}
\expandafter\gdef\csname odunum@n@positive_taxonomy_m2.ao.delta.nemotron.axis3.3_3\endcsname{67}
\expandafter\gdef\csname odunum@ci@positive_taxonomy_m2.ao.delta.nemotron.axis3.3_3\endcsname{[-0.244, -0.097]}
\expandafter\gdef\csname odunum@val@positive_taxonomy_m2.ao.delta.nemotron.axis3.3_4\endcsname{0.024}
\expandafter\gdef\csname odunum@n@positive_taxonomy_m2.ao.delta.nemotron.axis3.3_4\endcsname{141}
\expandafter\gdef\csname odunum@ci@positive_taxonomy_m2.ao.delta.nemotron.axis3.3_4\endcsname{[-0.029, 0.076]}
\expandafter\gdef\csname odunum@val@positive_taxonomy_m2.ao.delta.nemotron.axis3.3_5\endcsname{0.052}
\expandafter\gdef\csname odunum@n@positive_taxonomy_m2.ao.delta.nemotron.axis3.3_5\endcsname{105}
\expandafter\gdef\csname odunum@ci@positive_taxonomy_m2.ao.delta.nemotron.axis3.3_5\endcsname{[-0.009, 0.115]}
\expandafter\gdef\csname odunum@val@positive_taxonomy_m2.ao.delta.nemotron.axis3.3_6\endcsname{0.007}
\expandafter\gdef\csname odunum@n@positive_taxonomy_m2.ao.delta.nemotron.axis3.3_6\endcsname{202}
\expandafter\gdef\csname odunum@ci@positive_taxonomy_m2.ao.delta.nemotron.axis3.3_6\endcsname{[-0.032, 0.049]}
\expandafter\gdef\csname odunum@val@positive_taxonomy_m2.ao.delta.nemotron.axis4.4_1\endcsname{0.082}
\expandafter\gdef\csname odunum@n@positive_taxonomy_m2.ao.delta.nemotron.axis4.4_1\endcsname{63}
\expandafter\gdef\csname odunum@ci@positive_taxonomy_m2.ao.delta.nemotron.axis4.4_1\endcsname{[-0.002, 0.166]}
\expandafter\gdef\csname odunum@val@positive_taxonomy_m2.ao.delta.nemotron.axis4.4_2\endcsname{-0.005}
\expandafter\gdef\csname odunum@n@positive_taxonomy_m2.ao.delta.nemotron.axis4.4_2\endcsname{161}
\expandafter\gdef\csname odunum@ci@positive_taxonomy_m2.ao.delta.nemotron.axis4.4_2\endcsname{[-0.052, 0.043]}
\expandafter\gdef\csname odunum@val@positive_taxonomy_m2.ao.delta.nemotron.axis4.4_3\endcsname{0.009}
\expandafter\gdef\csname odunum@n@positive_taxonomy_m2.ao.delta.nemotron.axis4.4_3\endcsname{43}
\expandafter\gdef\csname odunum@ci@positive_taxonomy_m2.ao.delta.nemotron.axis4.4_3\endcsname{[-0.098, 0.121]}
\expandafter\gdef\csname odunum@val@positive_taxonomy_m2.ao.delta.nemotron.axis4.4_4\endcsname{-0.015}
\expandafter\gdef\csname odunum@n@positive_taxonomy_m2.ao.delta.nemotron.axis4.4_4\endcsname{172}
\expandafter\gdef\csname odunum@ci@positive_taxonomy_m2.ao.delta.nemotron.axis4.4_4\endcsname{[-0.060, 0.032]}
\expandafter\gdef\csname odunum@val@positive_taxonomy_m2.ao.delta.nemotron.axis4.4_5\endcsname{-0.017}
\expandafter\gdef\csname odunum@n@positive_taxonomy_m2.ao.delta.nemotron.axis4.4_5\endcsname{124}
\expandafter\gdef\csname odunum@ci@positive_taxonomy_m2.ao.delta.nemotron.axis4.4_5\endcsname{[-0.072, 0.040]}
\expandafter\gdef\csname odunum@val@positive_taxonomy_m2.ao.delta.qwen25_omni.axis1.1_1\endcsname{0.053}
\expandafter\gdef\csname odunum@n@positive_taxonomy_m2.ao.delta.qwen25_omni.axis1.1_1\endcsname{156}
\expandafter\gdef\csname odunum@ci@positive_taxonomy_m2.ao.delta.qwen25_omni.axis1.1_1\endcsname{[0.007, 0.099]}
\expandafter\gdef\csname odunum@val@positive_taxonomy_m2.ao.delta.qwen25_omni.axis1.1_2\endcsname{-0.020}
\expandafter\gdef\csname odunum@n@positive_taxonomy_m2.ao.delta.qwen25_omni.axis1.1_2\endcsname{407}
\expandafter\gdef\csname odunum@ci@positive_taxonomy_m2.ao.delta.qwen25_omni.axis1.1_2\endcsname{[-0.038, -0.003]}
\expandafter\gdef\csname odunum@val@positive_taxonomy_m2.ao.delta.qwen25_omni.axis2.2_1\endcsname{-0.143}
\expandafter\gdef\csname odunum@n@positive_taxonomy_m2.ao.delta.qwen25_omni.axis2.2_1\endcsname{21}
\expandafter\gdef\csname odunum@ci@positive_taxonomy_m2.ao.delta.qwen25_omni.axis2.2_1\endcsname{[-0.288, 0.017]}
\expandafter\gdef\csname odunum@val@positive_taxonomy_m2.ao.delta.qwen25_omni.axis2.2_2\endcsname{0.052}
\expandafter\gdef\csname odunum@n@positive_taxonomy_m2.ao.delta.qwen25_omni.axis2.2_2\endcsname{169}
\expandafter\gdef\csname odunum@ci@positive_taxonomy_m2.ao.delta.qwen25_omni.axis2.2_2\endcsname{[0.011, 0.094]}
\expandafter\gdef\csname odunum@val@positive_taxonomy_m2.ao.delta.qwen25_omni.axis2.2_3\endcsname{0.047}
\expandafter\gdef\csname odunum@n@positive_taxonomy_m2.ao.delta.qwen25_omni.axis2.2_3\endcsname{248}
\expandafter\gdef\csname odunum@ci@positive_taxonomy_m2.ao.delta.qwen25_omni.axis2.2_3\endcsname{[0.017, 0.078]}
\expandafter\gdef\csname odunum@val@positive_taxonomy_m2.ao.delta.qwen25_omni.axis2.2_4\endcsname{-0.161}
\expandafter\gdef\csname odunum@n@positive_taxonomy_m2.ao.delta.qwen25_omni.axis2.2_4\endcsname{34}
\expandafter\gdef\csname odunum@ci@positive_taxonomy_m2.ao.delta.qwen25_omni.axis2.2_4\endcsname{[-0.265, -0.055]}
\expandafter\gdef\csname odunum@val@positive_taxonomy_m2.ao.delta.qwen25_omni.axis2.2_5\endcsname{-0.132}
\expandafter\gdef\csname odunum@n@positive_taxonomy_m2.ao.delta.qwen25_omni.axis2.2_5\endcsname{91}
\expandafter\gdef\csname odunum@ci@positive_taxonomy_m2.ao.delta.qwen25_omni.axis2.2_5\endcsname{[-0.194, -0.071]}
\expandafter\gdef\csname odunum@val@positive_taxonomy_m2.ao.delta.qwen25_omni.axis3.3_1\endcsname{0.090}
\expandafter\gdef\csname odunum@n@positive_taxonomy_m2.ao.delta.qwen25_omni.axis3.3_1\endcsname{18}
\expandafter\gdef\csname odunum@ci@positive_taxonomy_m2.ao.delta.qwen25_omni.axis3.3_1\endcsname{[-0.048, 0.230]}
\expandafter\gdef\csname odunum@val@positive_taxonomy_m2.ao.delta.qwen25_omni.axis3.3_2\endcsname{0.059}
\expandafter\gdef\csname odunum@n@positive_taxonomy_m2.ao.delta.qwen25_omni.axis3.3_2\endcsname{30}
\expandafter\gdef\csname odunum@ci@positive_taxonomy_m2.ao.delta.qwen25_omni.axis3.3_2\endcsname{[-0.039, 0.161]}
\expandafter\gdef\csname odunum@val@positive_taxonomy_m2.ao.delta.qwen25_omni.axis3.3_3\endcsname{-0.112}
\expandafter\gdef\csname odunum@n@positive_taxonomy_m2.ao.delta.qwen25_omni.axis3.3_3\endcsname{67}
\expandafter\gdef\csname odunum@ci@positive_taxonomy_m2.ao.delta.qwen25_omni.axis3.3_3\endcsname{[-0.189, -0.035]}
\expandafter\gdef\csname odunum@val@positive_taxonomy_m2.ao.delta.qwen25_omni.axis3.3_4\endcsname{-0.036}
\expandafter\gdef\csname odunum@n@positive_taxonomy_m2.ao.delta.qwen25_omni.axis3.3_4\endcsname{141}
\expandafter\gdef\csname odunum@ci@positive_taxonomy_m2.ao.delta.qwen25_omni.axis3.3_4\endcsname{[-0.086, 0.014]}
\expandafter\gdef\csname odunum@val@positive_taxonomy_m2.ao.delta.qwen25_omni.axis3.3_5\endcsname{0.059}
\expandafter\gdef\csname odunum@n@positive_taxonomy_m2.ao.delta.qwen25_omni.axis3.3_5\endcsname{105}
\expandafter\gdef\csname odunum@ci@positive_taxonomy_m2.ao.delta.qwen25_omni.axis3.3_5\endcsname{[0.006, 0.113]}
\expandafter\gdef\csname odunum@val@positive_taxonomy_m2.ao.delta.qwen25_omni.axis3.3_6\endcsname{0.015}
\expandafter\gdef\csname odunum@n@positive_taxonomy_m2.ao.delta.qwen25_omni.axis3.3_6\endcsname{202}
\expandafter\gdef\csname odunum@ci@positive_taxonomy_m2.ao.delta.qwen25_omni.axis3.3_6\endcsname{[-0.020, 0.051]}
\expandafter\gdef\csname odunum@val@positive_taxonomy_m2.ao.delta.qwen25_omni.axis4.4_1\endcsname{0.057}
\expandafter\gdef\csname odunum@n@positive_taxonomy_m2.ao.delta.qwen25_omni.axis4.4_1\endcsname{63}
\expandafter\gdef\csname odunum@ci@positive_taxonomy_m2.ao.delta.qwen25_omni.axis4.4_1\endcsname{[-0.011, 0.122]}
\expandafter\gdef\csname odunum@val@positive_taxonomy_m2.ao.delta.qwen25_omni.axis4.4_2\endcsname{-0.015}
\expandafter\gdef\csname odunum@n@positive_taxonomy_m2.ao.delta.qwen25_omni.axis4.4_2\endcsname{161}
\expandafter\gdef\csname odunum@ci@positive_taxonomy_m2.ao.delta.qwen25_omni.axis4.4_2\endcsname{[-0.058, 0.028]}
\expandafter\gdef\csname odunum@val@positive_taxonomy_m2.ao.delta.qwen25_omni.axis4.4_3\endcsname{-0.011}
\expandafter\gdef\csname odunum@n@positive_taxonomy_m2.ao.delta.qwen25_omni.axis4.4_3\endcsname{43}
\expandafter\gdef\csname odunum@ci@positive_taxonomy_m2.ao.delta.qwen25_omni.axis4.4_3\endcsname{[-0.121, 0.101]}
\expandafter\gdef\csname odunum@val@positive_taxonomy_m2.ao.delta.qwen25_omni.axis4.4_4\endcsname{0.022}
\expandafter\gdef\csname odunum@n@positive_taxonomy_m2.ao.delta.qwen25_omni.axis4.4_4\endcsname{172}
\expandafter\gdef\csname odunum@ci@positive_taxonomy_m2.ao.delta.qwen25_omni.axis4.4_4\endcsname{[-0.019, 0.065]}
\expandafter\gdef\csname odunum@val@positive_taxonomy_m2.ao.delta.qwen25_omni.axis4.4_5\endcsname{-0.036}
\expandafter\gdef\csname odunum@n@positive_taxonomy_m2.ao.delta.qwen25_omni.axis4.4_5\endcsname{124}
\expandafter\gdef\csname odunum@ci@positive_taxonomy_m2.ao.delta.qwen25_omni.axis4.4_5\endcsname{[-0.088, 0.017]}
\expandafter\gdef\csname odunum@val@positive_taxonomy_m2.ao.delta.qwen3_omni_instruct.axis1.1_1\endcsname{0.030}
\expandafter\gdef\csname odunum@n@positive_taxonomy_m2.ao.delta.qwen3_omni_instruct.axis1.1_1\endcsname{156}
\expandafter\gdef\csname odunum@ci@positive_taxonomy_m2.ao.delta.qwen3_omni_instruct.axis1.1_1\endcsname{[-0.008, 0.068]}
\expandafter\gdef\csname odunum@val@positive_taxonomy_m2.ao.delta.qwen3_omni_instruct.axis1.1_2\endcsname{-0.011}
\expandafter\gdef\csname odunum@n@positive_taxonomy_m2.ao.delta.qwen3_omni_instruct.axis1.1_2\endcsname{407}
\expandafter\gdef\csname odunum@ci@positive_taxonomy_m2.ao.delta.qwen3_omni_instruct.axis1.1_2\endcsname{[-0.026, 0.003]}
\expandafter\gdef\csname odunum@val@positive_taxonomy_m2.ao.delta.qwen3_omni_instruct.axis2.2_1\endcsname{0.054}
\expandafter\gdef\csname odunum@n@positive_taxonomy_m2.ao.delta.qwen3_omni_instruct.axis2.2_1\endcsname{21}
\expandafter\gdef\csname odunum@ci@positive_taxonomy_m2.ao.delta.qwen3_omni_instruct.axis2.2_1\endcsname{[-0.074, 0.177]}
\expandafter\gdef\csname odunum@val@positive_taxonomy_m2.ao.delta.qwen3_omni_instruct.axis2.2_2\endcsname{0.010}
\expandafter\gdef\csname odunum@n@positive_taxonomy_m2.ao.delta.qwen3_omni_instruct.axis2.2_2\endcsname{169}
\expandafter\gdef\csname odunum@ci@positive_taxonomy_m2.ao.delta.qwen3_omni_instruct.axis2.2_2\endcsname{[-0.024, 0.044]}
\expandafter\gdef\csname odunum@val@positive_taxonomy_m2.ao.delta.qwen3_omni_instruct.axis2.2_3\endcsname{0.002}
\expandafter\gdef\csname odunum@n@positive_taxonomy_m2.ao.delta.qwen3_omni_instruct.axis2.2_3\endcsname{248}
\expandafter\gdef\csname odunum@ci@positive_taxonomy_m2.ao.delta.qwen3_omni_instruct.axis2.2_3\endcsname{[-0.023, 0.026]}
\expandafter\gdef\csname odunum@val@positive_taxonomy_m2.ao.delta.qwen3_omni_instruct.axis2.2_4\endcsname{-0.072}
\expandafter\gdef\csname odunum@n@positive_taxonomy_m2.ao.delta.qwen3_omni_instruct.axis2.2_4\endcsname{34}
\expandafter\gdef\csname odunum@ci@positive_taxonomy_m2.ao.delta.qwen3_omni_instruct.axis2.2_4\endcsname{[-0.170, 0.022]}
\expandafter\gdef\csname odunum@val@positive_taxonomy_m2.ao.delta.qwen3_omni_instruct.axis2.2_5\endcsname{-0.008}
\expandafter\gdef\csname odunum@n@positive_taxonomy_m2.ao.delta.qwen3_omni_instruct.axis2.2_5\endcsname{91}
\expandafter\gdef\csname odunum@ci@positive_taxonomy_m2.ao.delta.qwen3_omni_instruct.axis2.2_5\endcsname{[-0.058, 0.042]}
\expandafter\gdef\csname odunum@val@positive_taxonomy_m2.ao.delta.qwen3_omni_instruct.axis3.3_1\endcsname{0.029}
\expandafter\gdef\csname odunum@n@positive_taxonomy_m2.ao.delta.qwen3_omni_instruct.axis3.3_1\endcsname{18}
\expandafter\gdef\csname odunum@ci@positive_taxonomy_m2.ao.delta.qwen3_omni_instruct.axis3.3_1\endcsname{[-0.083, 0.148]}
\expandafter\gdef\csname odunum@val@positive_taxonomy_m2.ao.delta.qwen3_omni_instruct.axis3.3_2\endcsname{0.002}
\expandafter\gdef\csname odunum@n@positive_taxonomy_m2.ao.delta.qwen3_omni_instruct.axis3.3_2\endcsname{30}
\expandafter\gdef\csname odunum@ci@positive_taxonomy_m2.ao.delta.qwen3_omni_instruct.axis3.3_2\endcsname{[-0.086, 0.091]}
\expandafter\gdef\csname odunum@val@positive_taxonomy_m2.ao.delta.qwen3_omni_instruct.axis3.3_3\endcsname{-0.036}
\expandafter\gdef\csname odunum@n@positive_taxonomy_m2.ao.delta.qwen3_omni_instruct.axis3.3_3\endcsname{67}
\expandafter\gdef\csname odunum@ci@positive_taxonomy_m2.ao.delta.qwen3_omni_instruct.axis3.3_3\endcsname{[-0.101, 0.031]}
\expandafter\gdef\csname odunum@val@positive_taxonomy_m2.ao.delta.qwen3_omni_instruct.axis3.3_4\endcsname{-0.009}
\expandafter\gdef\csname odunum@n@positive_taxonomy_m2.ao.delta.qwen3_omni_instruct.axis3.3_4\endcsname{141}
\expandafter\gdef\csname odunum@ci@positive_taxonomy_m2.ao.delta.qwen3_omni_instruct.axis3.3_4\endcsname{[-0.050, 0.031]}
\expandafter\gdef\csname odunum@val@positive_taxonomy_m2.ao.delta.qwen3_omni_instruct.axis3.3_5\endcsname{-0.051}
\expandafter\gdef\csname odunum@n@positive_taxonomy_m2.ao.delta.qwen3_omni_instruct.axis3.3_5\endcsname{105}
\expandafter\gdef\csname odunum@ci@positive_taxonomy_m2.ao.delta.qwen3_omni_instruct.axis3.3_5\endcsname{[-0.096, -0.005]}
\expandafter\gdef\csname odunum@val@positive_taxonomy_m2.ao.delta.qwen3_omni_instruct.axis3.3_6\endcsname{0.042}
\expandafter\gdef\csname odunum@n@positive_taxonomy_m2.ao.delta.qwen3_omni_instruct.axis3.3_6\endcsname{202}
\expandafter\gdef\csname odunum@ci@positive_taxonomy_m2.ao.delta.qwen3_omni_instruct.axis3.3_6\endcsname{[0.014, 0.071]}
\expandafter\gdef\csname odunum@val@positive_taxonomy_m2.ao.delta.qwen3_omni_instruct.axis4.4_1\endcsname{-0.022}
\expandafter\gdef\csname odunum@n@positive_taxonomy_m2.ao.delta.qwen3_omni_instruct.axis4.4_1\endcsname{63}
\expandafter\gdef\csname odunum@ci@positive_taxonomy_m2.ao.delta.qwen3_omni_instruct.axis4.4_1\endcsname{[-0.083, 0.037]}
\expandafter\gdef\csname odunum@val@positive_taxonomy_m2.ao.delta.qwen3_omni_instruct.axis4.4_2\endcsname{-0.018}
\expandafter\gdef\csname odunum@n@positive_taxonomy_m2.ao.delta.qwen3_omni_instruct.axis4.4_2\endcsname{161}
\expandafter\gdef\csname odunum@ci@positive_taxonomy_m2.ao.delta.qwen3_omni_instruct.axis4.4_2\endcsname{[-0.053, 0.017]}
\expandafter\gdef\csname odunum@val@positive_taxonomy_m2.ao.delta.qwen3_omni_instruct.axis4.4_3\endcsname{-0.004}
\expandafter\gdef\csname odunum@n@positive_taxonomy_m2.ao.delta.qwen3_omni_instruct.axis4.4_3\endcsname{43}
\expandafter\gdef\csname odunum@ci@positive_taxonomy_m2.ao.delta.qwen3_omni_instruct.axis4.4_3\endcsname{[-0.096, 0.086]}
\expandafter\gdef\csname odunum@val@positive_taxonomy_m2.ao.delta.qwen3_omni_instruct.axis4.4_4\endcsname{0.023}
\expandafter\gdef\csname odunum@n@positive_taxonomy_m2.ao.delta.qwen3_omni_instruct.axis4.4_4\endcsname{172}
\expandafter\gdef\csname odunum@ci@positive_taxonomy_m2.ao.delta.qwen3_omni_instruct.axis4.4_4\endcsname{[-0.010, 0.057]}
\expandafter\gdef\csname odunum@val@positive_taxonomy_m2.ao.delta.qwen3_omni_instruct.axis4.4_5\endcsname{0.005}
\expandafter\gdef\csname odunum@n@positive_taxonomy_m2.ao.delta.qwen3_omni_instruct.axis4.4_5\endcsname{124}
\expandafter\gdef\csname odunum@ci@positive_taxonomy_m2.ao.delta.qwen3_omni_instruct.axis4.4_5\endcsname{[-0.037, 0.047]}
\expandafter\gdef\csname odunum@val@positive_taxonomy_m2.ao.delta.qwen3_omni_think.axis1.1_1\endcsname{0.012}
\expandafter\gdef\csname odunum@n@positive_taxonomy_m2.ao.delta.qwen3_omni_think.axis1.1_1\endcsname{156}
\expandafter\gdef\csname odunum@ci@positive_taxonomy_m2.ao.delta.qwen3_omni_think.axis1.1_1\endcsname{[-0.026, 0.050]}
\expandafter\gdef\csname odunum@val@positive_taxonomy_m2.ao.delta.qwen3_omni_think.axis1.1_2\endcsname{-0.005}
\expandafter\gdef\csname odunum@n@positive_taxonomy_m2.ao.delta.qwen3_omni_think.axis1.1_2\endcsname{407}
\expandafter\gdef\csname odunum@ci@positive_taxonomy_m2.ao.delta.qwen3_omni_think.axis1.1_2\endcsname{[-0.019, 0.010]}
\expandafter\gdef\csname odunum@val@positive_taxonomy_m2.ao.delta.qwen3_omni_think.axis2.2_1\endcsname{-0.073}
\expandafter\gdef\csname odunum@n@positive_taxonomy_m2.ao.delta.qwen3_omni_think.axis2.2_1\endcsname{21}
\expandafter\gdef\csname odunum@ci@positive_taxonomy_m2.ao.delta.qwen3_omni_think.axis2.2_1\endcsname{[-0.214, 0.064]}
\expandafter\gdef\csname odunum@val@positive_taxonomy_m2.ao.delta.qwen3_omni_think.axis2.2_2\endcsname{0.013}
\expandafter\gdef\csname odunum@n@positive_taxonomy_m2.ao.delta.qwen3_omni_think.axis2.2_2\endcsname{169}
\expandafter\gdef\csname odunum@ci@positive_taxonomy_m2.ao.delta.qwen3_omni_think.axis2.2_2\endcsname{[-0.019, 0.046]}
\expandafter\gdef\csname odunum@val@positive_taxonomy_m2.ao.delta.qwen3_omni_think.axis2.2_3\endcsname{0.019}
\expandafter\gdef\csname odunum@n@positive_taxonomy_m2.ao.delta.qwen3_omni_think.axis2.2_3\endcsname{248}
\expandafter\gdef\csname odunum@ci@positive_taxonomy_m2.ao.delta.qwen3_omni_think.axis2.2_3\endcsname{[-0.007, 0.046]}
\expandafter\gdef\csname odunum@val@positive_taxonomy_m2.ao.delta.qwen3_omni_think.axis2.2_4\endcsname{-0.042}
\expandafter\gdef\csname odunum@n@positive_taxonomy_m2.ao.delta.qwen3_omni_think.axis2.2_4\endcsname{34}
\expandafter\gdef\csname odunum@ci@positive_taxonomy_m2.ao.delta.qwen3_omni_think.axis2.2_4\endcsname{[-0.145, 0.057]}
\expandafter\gdef\csname odunum@val@positive_taxonomy_m2.ao.delta.qwen3_omni_think.axis2.2_5\endcsname{-0.045}
\expandafter\gdef\csname odunum@n@positive_taxonomy_m2.ao.delta.qwen3_omni_think.axis2.2_5\endcsname{91}
\expandafter\gdef\csname odunum@ci@positive_taxonomy_m2.ao.delta.qwen3_omni_think.axis2.2_5\endcsname{[-0.100, 0.010]}
\expandafter\gdef\csname odunum@val@positive_taxonomy_m2.ao.delta.qwen3_omni_think.axis3.3_1\endcsname{0.095}
\expandafter\gdef\csname odunum@n@positive_taxonomy_m2.ao.delta.qwen3_omni_think.axis3.3_1\endcsname{18}
\expandafter\gdef\csname odunum@ci@positive_taxonomy_m2.ao.delta.qwen3_omni_think.axis3.3_1\endcsname{[-0.018, 0.208]}
\expandafter\gdef\csname odunum@val@positive_taxonomy_m2.ao.delta.qwen3_omni_think.axis3.3_2\endcsname{0.072}
\expandafter\gdef\csname odunum@n@positive_taxonomy_m2.ao.delta.qwen3_omni_think.axis3.3_2\endcsname{30}
\expandafter\gdef\csname odunum@ci@positive_taxonomy_m2.ao.delta.qwen3_omni_think.axis3.3_2\endcsname{[-0.016, 0.157]}
\expandafter\gdef\csname odunum@val@positive_taxonomy_m2.ao.delta.qwen3_omni_think.axis3.3_3\endcsname{-0.066}
\expandafter\gdef\csname odunum@n@positive_taxonomy_m2.ao.delta.qwen3_omni_think.axis3.3_3\endcsname{67}
\expandafter\gdef\csname odunum@ci@positive_taxonomy_m2.ao.delta.qwen3_omni_think.axis3.3_3\endcsname{[-0.136, 0.003]}
\expandafter\gdef\csname odunum@val@positive_taxonomy_m2.ao.delta.qwen3_omni_think.axis3.3_4\endcsname{-0.008}
\expandafter\gdef\csname odunum@n@positive_taxonomy_m2.ao.delta.qwen3_omni_think.axis3.3_4\endcsname{141}
\expandafter\gdef\csname odunum@ci@positive_taxonomy_m2.ao.delta.qwen3_omni_think.axis3.3_4\endcsname{[-0.049, 0.032]}
\expandafter\gdef\csname odunum@val@positive_taxonomy_m2.ao.delta.qwen3_omni_think.axis3.3_5\endcsname{-0.056}
\expandafter\gdef\csname odunum@n@positive_taxonomy_m2.ao.delta.qwen3_omni_think.axis3.3_5\endcsname{105}
\expandafter\gdef\csname odunum@ci@positive_taxonomy_m2.ao.delta.qwen3_omni_think.axis3.3_5\endcsname{[-0.101, -0.009]}
\expandafter\gdef\csname odunum@val@positive_taxonomy_m2.ao.delta.qwen3_omni_think.axis3.3_6\endcsname{0.037}
\expandafter\gdef\csname odunum@n@positive_taxonomy_m2.ao.delta.qwen3_omni_think.axis3.3_6\endcsname{202}
\expandafter\gdef\csname odunum@ci@positive_taxonomy_m2.ao.delta.qwen3_omni_think.axis3.3_6\endcsname{[0.008, 0.068]}
\expandafter\gdef\csname odunum@val@positive_taxonomy_m2.ao.delta.qwen3_omni_think.axis4.4_1\endcsname{0.041}
\expandafter\gdef\csname odunum@n@positive_taxonomy_m2.ao.delta.qwen3_omni_think.axis4.4_1\endcsname{63}
\expandafter\gdef\csname odunum@ci@positive_taxonomy_m2.ao.delta.qwen3_omni_think.axis4.4_1\endcsname{[-0.019, 0.101]}
\expandafter\gdef\csname odunum@val@positive_taxonomy_m2.ao.delta.qwen3_omni_think.axis4.4_2\endcsname{-0.010}
\expandafter\gdef\csname odunum@n@positive_taxonomy_m2.ao.delta.qwen3_omni_think.axis4.4_2\endcsname{161}
\expandafter\gdef\csname odunum@ci@positive_taxonomy_m2.ao.delta.qwen3_omni_think.axis4.4_2\endcsname{[-0.046, 0.027]}
\expandafter\gdef\csname odunum@val@positive_taxonomy_m2.ao.delta.qwen3_omni_think.axis4.4_3\endcsname{0.010}
\expandafter\gdef\csname odunum@n@positive_taxonomy_m2.ao.delta.qwen3_omni_think.axis4.4_3\endcsname{43}
\expandafter\gdef\csname odunum@ci@positive_taxonomy_m2.ao.delta.qwen3_omni_think.axis4.4_3\endcsname{[-0.081, 0.101]}
\expandafter\gdef\csname odunum@val@positive_taxonomy_m2.ao.delta.qwen3_omni_think.axis4.4_4\endcsname{0.017}
\expandafter\gdef\csname odunum@n@positive_taxonomy_m2.ao.delta.qwen3_omni_think.axis4.4_4\endcsname{172}
\expandafter\gdef\csname odunum@ci@positive_taxonomy_m2.ao.delta.qwen3_omni_think.axis4.4_4\endcsname{[-0.017, 0.051]}
\expandafter\gdef\csname odunum@val@positive_taxonomy_m2.ao.delta.qwen3_omni_think.axis4.4_5\endcsname{-0.035}
\expandafter\gdef\csname odunum@n@positive_taxonomy_m2.ao.delta.qwen3_omni_think.axis4.4_5\endcsname{124}
\expandafter\gdef\csname odunum@ci@positive_taxonomy_m2.ao.delta.qwen3_omni_think.axis4.4_5\endcsname{[-0.077, 0.009]}
\expandafter\gdef\csname odunum@val@positive_taxonomy_m2.ao.delta.qwen_plus.axis1.1_1\endcsname{0.033}
\expandafter\gdef\csname odunum@n@positive_taxonomy_m2.ao.delta.qwen_plus.axis1.1_1\endcsname{156}
\expandafter\gdef\csname odunum@ci@positive_taxonomy_m2.ao.delta.qwen_plus.axis1.1_1\endcsname{[-0.001, 0.067]}
\expandafter\gdef\csname odunum@val@positive_taxonomy_m2.ao.delta.qwen_plus.axis1.1_2\endcsname{-0.013}
\expandafter\gdef\csname odunum@n@positive_taxonomy_m2.ao.delta.qwen_plus.axis1.1_2\endcsname{407}
\expandafter\gdef\csname odunum@ci@positive_taxonomy_m2.ao.delta.qwen_plus.axis1.1_2\endcsname{[-0.026, 2.6\ensuremath{\times 10^{-4}}]}
\expandafter\gdef\csname odunum@val@positive_taxonomy_m2.ao.delta.qwen_plus.axis2.2_1\endcsname{-0.073}
\expandafter\gdef\csname odunum@n@positive_taxonomy_m2.ao.delta.qwen_plus.axis2.2_1\endcsname{21}
\expandafter\gdef\csname odunum@ci@positive_taxonomy_m2.ao.delta.qwen_plus.axis2.2_1\endcsname{[-0.204, 0.049]}
\expandafter\gdef\csname odunum@val@positive_taxonomy_m2.ao.delta.qwen_plus.axis2.2_2\endcsname{0.017}
\expandafter\gdef\csname odunum@n@positive_taxonomy_m2.ao.delta.qwen_plus.axis2.2_2\endcsname{169}
\expandafter\gdef\csname odunum@ci@positive_taxonomy_m2.ao.delta.qwen_plus.axis2.2_2\endcsname{[-0.015, 0.048]}
\expandafter\gdef\csname odunum@val@positive_taxonomy_m2.ao.delta.qwen_plus.axis2.2_3\endcsname{0.016}
\expandafter\gdef\csname odunum@n@positive_taxonomy_m2.ao.delta.qwen_plus.axis2.2_3\endcsname{248}
\expandafter\gdef\csname odunum@ci@positive_taxonomy_m2.ao.delta.qwen_plus.axis2.2_3\endcsname{[-0.007, 0.040]}
\expandafter\gdef\csname odunum@val@positive_taxonomy_m2.ao.delta.qwen_plus.axis2.2_4\endcsname{-0.102}
\expandafter\gdef\csname odunum@n@positive_taxonomy_m2.ao.delta.qwen_plus.axis2.2_4\endcsname{34}
\expandafter\gdef\csname odunum@ci@positive_taxonomy_m2.ao.delta.qwen_plus.axis2.2_4\endcsname{[-0.209, 0.004]}
\expandafter\gdef\csname odunum@val@positive_taxonomy_m2.ao.delta.qwen_plus.axis2.2_5\endcsname{-0.021}
\expandafter\gdef\csname odunum@n@positive_taxonomy_m2.ao.delta.qwen_plus.axis2.2_5\endcsname{91}
\expandafter\gdef\csname odunum@ci@positive_taxonomy_m2.ao.delta.qwen_plus.axis2.2_5\endcsname{[-0.075, 0.032]}
\expandafter\gdef\csname odunum@val@positive_taxonomy_m2.ao.delta.qwen_plus.axis3.3_1\endcsname{-0.016}
\expandafter\gdef\csname odunum@n@positive_taxonomy_m2.ao.delta.qwen_plus.axis3.3_1\endcsname{18}
\expandafter\gdef\csname odunum@ci@positive_taxonomy_m2.ao.delta.qwen_plus.axis3.3_1\endcsname{[-0.156, 0.116]}
\expandafter\gdef\csname odunum@val@positive_taxonomy_m2.ao.delta.qwen_plus.axis3.3_2\endcsname{0.012}
\expandafter\gdef\csname odunum@n@positive_taxonomy_m2.ao.delta.qwen_plus.axis3.3_2\endcsname{30}
\expandafter\gdef\csname odunum@ci@positive_taxonomy_m2.ao.delta.qwen_plus.axis3.3_2\endcsname{[-0.078, 0.099]}
\expandafter\gdef\csname odunum@val@positive_taxonomy_m2.ao.delta.qwen_plus.axis3.3_3\endcsname{-0.031}
\expandafter\gdef\csname odunum@n@positive_taxonomy_m2.ao.delta.qwen_plus.axis3.3_3\endcsname{67}
\expandafter\gdef\csname odunum@ci@positive_taxonomy_m2.ao.delta.qwen_plus.axis3.3_3\endcsname{[-0.094, 0.031]}
\expandafter\gdef\csname odunum@val@positive_taxonomy_m2.ao.delta.qwen_plus.axis3.3_4\endcsname{-0.028}
\expandafter\gdef\csname odunum@n@positive_taxonomy_m2.ao.delta.qwen_plus.axis3.3_4\endcsname{141}
\expandafter\gdef\csname odunum@ci@positive_taxonomy_m2.ao.delta.qwen_plus.axis3.3_4\endcsname{[-0.067, 0.010]}
\expandafter\gdef\csname odunum@val@positive_taxonomy_m2.ao.delta.qwen_plus.axis3.3_5\endcsname{-0.015}
\expandafter\gdef\csname odunum@n@positive_taxonomy_m2.ao.delta.qwen_plus.axis3.3_5\endcsname{105}
\expandafter\gdef\csname odunum@ci@positive_taxonomy_m2.ao.delta.qwen_plus.axis3.3_5\endcsname{[-0.061, 0.031]}
\expandafter\gdef\csname odunum@val@positive_taxonomy_m2.ao.delta.qwen_plus.axis3.3_6\endcsname{0.037}
\expandafter\gdef\csname odunum@n@positive_taxonomy_m2.ao.delta.qwen_plus.axis3.3_6\endcsname{202}
\expandafter\gdef\csname odunum@ci@positive_taxonomy_m2.ao.delta.qwen_plus.axis3.3_6\endcsname{[0.010, 0.065]}
\expandafter\gdef\csname odunum@val@positive_taxonomy_m2.ao.delta.qwen_plus.axis4.4_1\endcsname{0.038}
\expandafter\gdef\csname odunum@n@positive_taxonomy_m2.ao.delta.qwen_plus.axis4.4_1\endcsname{63}
\expandafter\gdef\csname odunum@ci@positive_taxonomy_m2.ao.delta.qwen_plus.axis4.4_1\endcsname{[-0.011, 0.089]}
\expandafter\gdef\csname odunum@val@positive_taxonomy_m2.ao.delta.qwen_plus.axis4.4_2\endcsname{-0.020}
\expandafter\gdef\csname odunum@n@positive_taxonomy_m2.ao.delta.qwen_plus.axis4.4_2\endcsname{161}
\expandafter\gdef\csname odunum@ci@positive_taxonomy_m2.ao.delta.qwen_plus.axis4.4_2\endcsname{[-0.054, 0.014]}
\expandafter\gdef\csname odunum@val@positive_taxonomy_m2.ao.delta.qwen_plus.axis4.4_3\endcsname{-0.075}
\expandafter\gdef\csname odunum@n@positive_taxonomy_m2.ao.delta.qwen_plus.axis4.4_3\endcsname{43}
\expandafter\gdef\csname odunum@ci@positive_taxonomy_m2.ao.delta.qwen_plus.axis4.4_3\endcsname{[-0.165, 0.011]}
\expandafter\gdef\csname odunum@val@positive_taxonomy_m2.ao.delta.qwen_plus.axis4.4_4\endcsname{0.023}
\expandafter\gdef\csname odunum@n@positive_taxonomy_m2.ao.delta.qwen_plus.axis4.4_4\endcsname{172}
\expandafter\gdef\csname odunum@ci@positive_taxonomy_m2.ao.delta.qwen_plus.axis4.4_4\endcsname{[-0.011, 0.055]}
\expandafter\gdef\csname odunum@val@positive_taxonomy_m2.ao.delta.qwen_plus.axis4.4_5\endcsname{0.001}
\expandafter\gdef\csname odunum@n@positive_taxonomy_m2.ao.delta.qwen_plus.axis4.4_5\endcsname{124}
\expandafter\gdef\csname odunum@ci@positive_taxonomy_m2.ao.delta.qwen_plus.axis4.4_5\endcsname{[-0.040, 0.043]}
\expandafter\gdef\csname odunum@val@positive_taxonomy_m2.ao.delta.salmonn2_7b.axis1.1_1\endcsname{0.027}
\expandafter\gdef\csname odunum@n@positive_taxonomy_m2.ao.delta.salmonn2_7b.axis1.1_1\endcsname{156}
\expandafter\gdef\csname odunum@ci@positive_taxonomy_m2.ao.delta.salmonn2_7b.axis1.1_1\endcsname{[-0.012, 0.069]}
\expandafter\gdef\csname odunum@val@positive_taxonomy_m2.ao.delta.salmonn2_7b.axis1.1_2\endcsname{-0.010}
\expandafter\gdef\csname odunum@n@positive_taxonomy_m2.ao.delta.salmonn2_7b.axis1.1_2\endcsname{407}
\expandafter\gdef\csname odunum@ci@positive_taxonomy_m2.ao.delta.salmonn2_7b.axis1.1_2\endcsname{[-0.026, 0.005]}
\expandafter\gdef\csname odunum@val@positive_taxonomy_m2.ao.delta.salmonn2_7b.axis2.2_1\endcsname{0.049}
\expandafter\gdef\csname odunum@n@positive_taxonomy_m2.ao.delta.salmonn2_7b.axis2.2_1\endcsname{21}
\expandafter\gdef\csname odunum@ci@positive_taxonomy_m2.ao.delta.salmonn2_7b.axis2.2_1\endcsname{[-0.067, 0.178]}
\expandafter\gdef\csname odunum@val@positive_taxonomy_m2.ao.delta.salmonn2_7b.axis2.2_2\endcsname{0.013}
\expandafter\gdef\csname odunum@n@positive_taxonomy_m2.ao.delta.salmonn2_7b.axis2.2_2\endcsname{169}
\expandafter\gdef\csname odunum@ci@positive_taxonomy_m2.ao.delta.salmonn2_7b.axis2.2_2\endcsname{[-0.022, 0.048]}
\expandafter\gdef\csname odunum@val@positive_taxonomy_m2.ao.delta.salmonn2_7b.axis2.2_3\endcsname{-0.057}
\expandafter\gdef\csname odunum@n@positive_taxonomy_m2.ao.delta.salmonn2_7b.axis2.2_3\endcsname{248}
\expandafter\gdef\csname odunum@ci@positive_taxonomy_m2.ao.delta.salmonn2_7b.axis2.2_3\endcsname{[-0.082, -0.031]}
\expandafter\gdef\csname odunum@val@positive_taxonomy_m2.ao.delta.salmonn2_7b.axis2.2_4\endcsname{0.204}
\expandafter\gdef\csname odunum@n@positive_taxonomy_m2.ao.delta.salmonn2_7b.axis2.2_4\endcsname{34}
\expandafter\gdef\csname odunum@ci@positive_taxonomy_m2.ao.delta.salmonn2_7b.axis2.2_4\endcsname{[0.093, 0.319]}
\expandafter\gdef\csname odunum@val@positive_taxonomy_m2.ao.delta.salmonn2_7b.axis2.2_5\endcsname{0.042}
\expandafter\gdef\csname odunum@n@positive_taxonomy_m2.ao.delta.salmonn2_7b.axis2.2_5\endcsname{91}
\expandafter\gdef\csname odunum@ci@positive_taxonomy_m2.ao.delta.salmonn2_7b.axis2.2_5\endcsname{[-0.007, 0.096]}
\expandafter\gdef\csname odunum@val@positive_taxonomy_m2.ao.delta.salmonn2_7b.axis3.3_1\endcsname{0.024}
\expandafter\gdef\csname odunum@n@positive_taxonomy_m2.ao.delta.salmonn2_7b.axis3.3_1\endcsname{18}
\expandafter\gdef\csname odunum@ci@positive_taxonomy_m2.ao.delta.salmonn2_7b.axis3.3_1\endcsname{[-0.115, 0.179]}
\expandafter\gdef\csname odunum@val@positive_taxonomy_m2.ao.delta.salmonn2_7b.axis3.3_2\endcsname{0.006}
\expandafter\gdef\csname odunum@n@positive_taxonomy_m2.ao.delta.salmonn2_7b.axis3.3_2\endcsname{30}
\expandafter\gdef\csname odunum@ci@positive_taxonomy_m2.ao.delta.salmonn2_7b.axis3.3_2\endcsname{[-0.076, 0.093]}
\expandafter\gdef\csname odunum@val@positive_taxonomy_m2.ao.delta.salmonn2_7b.axis3.3_3\endcsname{-0.057}
\expandafter\gdef\csname odunum@n@positive_taxonomy_m2.ao.delta.salmonn2_7b.axis3.3_3\endcsname{67}
\expandafter\gdef\csname odunum@ci@positive_taxonomy_m2.ao.delta.salmonn2_7b.axis3.3_3\endcsname{[-0.112, 0.002]}
\expandafter\gdef\csname odunum@val@positive_taxonomy_m2.ao.delta.salmonn2_7b.axis3.3_4\endcsname{-0.003}
\expandafter\gdef\csname odunum@n@positive_taxonomy_m2.ao.delta.salmonn2_7b.axis3.3_4\endcsname{141}
\expandafter\gdef\csname odunum@ci@positive_taxonomy_m2.ao.delta.salmonn2_7b.axis3.3_4\endcsname{[-0.043, 0.036]}
\expandafter\gdef\csname odunum@val@positive_taxonomy_m2.ao.delta.salmonn2_7b.axis3.3_5\endcsname{0.034}
\expandafter\gdef\csname odunum@n@positive_taxonomy_m2.ao.delta.salmonn2_7b.axis3.3_5\endcsname{105}
\expandafter\gdef\csname odunum@ci@positive_taxonomy_m2.ao.delta.salmonn2_7b.axis3.3_5\endcsname{[-0.013, 0.082]}
\expandafter\gdef\csname odunum@val@positive_taxonomy_m2.ao.delta.salmonn2_7b.axis3.3_6\endcsname{0.001}
\expandafter\gdef\csname odunum@n@positive_taxonomy_m2.ao.delta.salmonn2_7b.axis3.3_6\endcsname{202}
\expandafter\gdef\csname odunum@ci@positive_taxonomy_m2.ao.delta.salmonn2_7b.axis3.3_6\endcsname{[-0.028, 0.032]}
\expandafter\gdef\csname odunum@val@positive_taxonomy_m2.ao.delta.salmonn2_7b.axis4.4_1\endcsname{-0.002}
\expandafter\gdef\csname odunum@n@positive_taxonomy_m2.ao.delta.salmonn2_7b.axis4.4_1\endcsname{63}
\expandafter\gdef\csname odunum@ci@positive_taxonomy_m2.ao.delta.salmonn2_7b.axis4.4_1\endcsname{[-0.062, 0.064]}
\expandafter\gdef\csname odunum@val@positive_taxonomy_m2.ao.delta.salmonn2_7b.axis4.4_2\endcsname{0.004}
\expandafter\gdef\csname odunum@n@positive_taxonomy_m2.ao.delta.salmonn2_7b.axis4.4_2\endcsname{161}
\expandafter\gdef\csname odunum@ci@positive_taxonomy_m2.ao.delta.salmonn2_7b.axis4.4_2\endcsname{[-0.032, 0.040]}
\expandafter\gdef\csname odunum@val@positive_taxonomy_m2.ao.delta.salmonn2_7b.axis4.4_3\endcsname{0.038}
\expandafter\gdef\csname odunum@n@positive_taxonomy_m2.ao.delta.salmonn2_7b.axis4.4_3\endcsname{43}
\expandafter\gdef\csname odunum@ci@positive_taxonomy_m2.ao.delta.salmonn2_7b.axis4.4_3\endcsname{[-0.041, 0.121]}
\expandafter\gdef\csname odunum@val@positive_taxonomy_m2.ao.delta.salmonn2_7b.axis4.4_4\endcsname{-0.019}
\expandafter\gdef\csname odunum@n@positive_taxonomy_m2.ao.delta.salmonn2_7b.axis4.4_4\endcsname{172}
\expandafter\gdef\csname odunum@ci@positive_taxonomy_m2.ao.delta.salmonn2_7b.axis4.4_4\endcsname{[-0.052, 0.015]}
\expandafter\gdef\csname odunum@val@positive_taxonomy_m2.ao.delta.salmonn2_7b.axis4.4_5\endcsname{0.009}
\expandafter\gdef\csname odunum@n@positive_taxonomy_m2.ao.delta.salmonn2_7b.axis4.4_5\endcsname{124}
\expandafter\gdef\csname odunum@ci@positive_taxonomy_m2.ao.delta.salmonn2_7b.axis4.4_5\endcsname{[-0.035, 0.057]}
\expandafter\gdef\csname odunum@val@positive_taxonomy_m2.ao.delta.seed.axis1.1_1\endcsname{0.019}
\expandafter\gdef\csname odunum@n@positive_taxonomy_m2.ao.delta.seed.axis1.1_1\endcsname{156}
\expandafter\gdef\csname odunum@ci@positive_taxonomy_m2.ao.delta.seed.axis1.1_1\endcsname{[-0.014, 0.053]}
\expandafter\gdef\csname odunum@val@positive_taxonomy_m2.ao.delta.seed.axis1.1_2\endcsname{-0.007}
\expandafter\gdef\csname odunum@n@positive_taxonomy_m2.ao.delta.seed.axis1.1_2\endcsname{407}
\expandafter\gdef\csname odunum@ci@positive_taxonomy_m2.ao.delta.seed.axis1.1_2\endcsname{[-0.020, 0.005]}
\expandafter\gdef\csname odunum@val@positive_taxonomy_m2.ao.delta.seed.axis2.2_1\endcsname{0.034}
\expandafter\gdef\csname odunum@n@positive_taxonomy_m2.ao.delta.seed.axis2.2_1\endcsname{21}
\expandafter\gdef\csname odunum@ci@positive_taxonomy_m2.ao.delta.seed.axis2.2_1\endcsname{[-0.045, 0.112]}
\expandafter\gdef\csname odunum@val@positive_taxonomy_m2.ao.delta.seed.axis2.2_2\endcsname{-0.017}
\expandafter\gdef\csname odunum@n@positive_taxonomy_m2.ao.delta.seed.axis2.2_2\endcsname{169}
\expandafter\gdef\csname odunum@ci@positive_taxonomy_m2.ao.delta.seed.axis2.2_2\endcsname{[-0.048, 0.014]}
\expandafter\gdef\csname odunum@val@positive_taxonomy_m2.ao.delta.seed.axis2.2_3\endcsname{0.025}
\expandafter\gdef\csname odunum@n@positive_taxonomy_m2.ao.delta.seed.axis2.2_3\endcsname{248}
\expandafter\gdef\csname odunum@ci@positive_taxonomy_m2.ao.delta.seed.axis2.2_3\endcsname{[0.003, 0.048]}
\expandafter\gdef\csname odunum@val@positive_taxonomy_m2.ao.delta.seed.axis2.2_4\endcsname{-0.075}
\expandafter\gdef\csname odunum@n@positive_taxonomy_m2.ao.delta.seed.axis2.2_4\endcsname{34}
\expandafter\gdef\csname odunum@ci@positive_taxonomy_m2.ao.delta.seed.axis2.2_4\endcsname{[-0.170, 0.013]}
\expandafter\gdef\csname odunum@val@positive_taxonomy_m2.ao.delta.seed.axis2.2_5\endcsname{-0.016}
\expandafter\gdef\csname odunum@n@positive_taxonomy_m2.ao.delta.seed.axis2.2_5\endcsname{91}
\expandafter\gdef\csname odunum@ci@positive_taxonomy_m2.ao.delta.seed.axis2.2_5\endcsname{[-0.064, 0.030]}
\expandafter\gdef\csname odunum@val@positive_taxonomy_m2.ao.delta.seed.axis3.3_1\endcsname{0.091}
\expandafter\gdef\csname odunum@n@positive_taxonomy_m2.ao.delta.seed.axis3.3_1\endcsname{18}
\expandafter\gdef\csname odunum@ci@positive_taxonomy_m2.ao.delta.seed.axis3.3_1\endcsname{[-0.008, 0.178]}
\expandafter\gdef\csname odunum@val@positive_taxonomy_m2.ao.delta.seed.axis3.3_2\endcsname{-0.069}
\expandafter\gdef\csname odunum@n@positive_taxonomy_m2.ao.delta.seed.axis3.3_2\endcsname{30}
\expandafter\gdef\csname odunum@ci@positive_taxonomy_m2.ao.delta.seed.axis3.3_2\endcsname{[-0.144, 0.008]}
\expandafter\gdef\csname odunum@val@positive_taxonomy_m2.ao.delta.seed.axis3.3_3\endcsname{-0.011}
\expandafter\gdef\csname odunum@n@positive_taxonomy_m2.ao.delta.seed.axis3.3_3\endcsname{67}
\expandafter\gdef\csname odunum@ci@positive_taxonomy_m2.ao.delta.seed.axis3.3_3\endcsname{[-0.068, 0.045]}
\expandafter\gdef\csname odunum@val@positive_taxonomy_m2.ao.delta.seed.axis3.3_4\endcsname{0.010}
\expandafter\gdef\csname odunum@n@positive_taxonomy_m2.ao.delta.seed.axis3.3_4\endcsname{141}
\expandafter\gdef\csname odunum@ci@positive_taxonomy_m2.ao.delta.seed.axis3.3_4\endcsname{[-0.024, 0.044]}
\expandafter\gdef\csname odunum@val@positive_taxonomy_m2.ao.delta.seed.axis3.3_5\endcsname{-0.073}
\expandafter\gdef\csname odunum@n@positive_taxonomy_m2.ao.delta.seed.axis3.3_5\endcsname{105}
\expandafter\gdef\csname odunum@ci@positive_taxonomy_m2.ao.delta.seed.axis3.3_5\endcsname{[-0.121, -0.027]}
\expandafter\gdef\csname odunum@val@positive_taxonomy_m2.ao.delta.seed.axis3.3_6\endcsname{0.037}
\expandafter\gdef\csname odunum@n@positive_taxonomy_m2.ao.delta.seed.axis3.3_6\endcsname{202}
\expandafter\gdef\csname odunum@ci@positive_taxonomy_m2.ao.delta.seed.axis3.3_6\endcsname{[0.012, 0.062]}
\expandafter\gdef\csname odunum@val@positive_taxonomy_m2.ao.delta.seed.axis4.4_1\endcsname{0.018}
\expandafter\gdef\csname odunum@n@positive_taxonomy_m2.ao.delta.seed.axis4.4_1\endcsname{63}
\expandafter\gdef\csname odunum@ci@positive_taxonomy_m2.ao.delta.seed.axis4.4_1\endcsname{[-0.032, 0.066]}
\expandafter\gdef\csname odunum@val@positive_taxonomy_m2.ao.delta.seed.axis4.4_2\endcsname{-0.025}
\expandafter\gdef\csname odunum@n@positive_taxonomy_m2.ao.delta.seed.axis4.4_2\endcsname{161}
\expandafter\gdef\csname odunum@ci@positive_taxonomy_m2.ao.delta.seed.axis4.4_2\endcsname{[-0.057, 0.006]}
\expandafter\gdef\csname odunum@val@positive_taxonomy_m2.ao.delta.seed.axis4.4_3\endcsname{-0.012}
\expandafter\gdef\csname odunum@n@positive_taxonomy_m2.ao.delta.seed.axis4.4_3\endcsname{43}
\expandafter\gdef\csname odunum@ci@positive_taxonomy_m2.ao.delta.seed.axis4.4_3\endcsname{[-0.090, 0.062]}
\expandafter\gdef\csname odunum@val@positive_taxonomy_m2.ao.delta.seed.axis4.4_4\endcsname{-0.001}
\expandafter\gdef\csname odunum@n@positive_taxonomy_m2.ao.delta.seed.axis4.4_4\endcsname{172}
\expandafter\gdef\csname odunum@ci@positive_taxonomy_m2.ao.delta.seed.axis4.4_4\endcsname{[-0.031, 0.029]}
\expandafter\gdef\csname odunum@val@positive_taxonomy_m2.ao.delta.seed.axis4.4_5\endcsname{0.029}
\expandafter\gdef\csname odunum@n@positive_taxonomy_m2.ao.delta.seed.axis4.4_5\endcsname{124}
\expandafter\gdef\csname odunum@ci@positive_taxonomy_m2.ao.delta.seed.axis4.4_5\endcsname{[-0.008, 0.065]}
\expandafter\gdef\csname odunum@val@positive_taxonomy_m2.ao.mean.cascade_asr.axis1.1_1\endcsname{0.768}
\expandafter\gdef\csname odunum@n@positive_taxonomy_m2.ao.mean.cascade_asr.axis1.1_1\endcsname{156}
\expandafter\gdef\csname odunum@ci@positive_taxonomy_m2.ao.mean.cascade_asr.axis1.1_1\endcsname{[0.729, 0.805]}
\expandafter\gdef\csname odunum@val@positive_taxonomy_m2.ao.mean.cascade_asr.axis1.1_2\endcsname{0.651}
\expandafter\gdef\csname odunum@n@positive_taxonomy_m2.ao.mean.cascade_asr.axis1.1_2\endcsname{407}
\expandafter\gdef\csname odunum@ci@positive_taxonomy_m2.ao.mean.cascade_asr.axis1.1_2\endcsname{[0.621, 0.679]}
\expandafter\gdef\csname odunum@val@positive_taxonomy_m2.ao.mean.cascade_asr.axis2.2_1\endcsname{0.497}
\expandafter\gdef\csname odunum@n@positive_taxonomy_m2.ao.mean.cascade_asr.axis2.2_1\endcsname{21}
\expandafter\gdef\csname odunum@ci@positive_taxonomy_m2.ao.mean.cascade_asr.axis2.2_1\endcsname{[0.332, 0.657]}
\expandafter\gdef\csname odunum@val@positive_taxonomy_m2.ao.mean.cascade_asr.axis2.2_2\endcsname{0.698}
\expandafter\gdef\csname odunum@n@positive_taxonomy_m2.ao.mean.cascade_asr.axis2.2_2\endcsname{169}
\expandafter\gdef\csname odunum@ci@positive_taxonomy_m2.ao.mean.cascade_asr.axis2.2_2\endcsname{[0.659, 0.736]}
\expandafter\gdef\csname odunum@val@positive_taxonomy_m2.ao.mean.cascade_asr.axis2.2_3\endcsname{0.732}
\expandafter\gdef\csname odunum@n@positive_taxonomy_m2.ao.mean.cascade_asr.axis2.2_3\endcsname{248}
\expandafter\gdef\csname odunum@ci@positive_taxonomy_m2.ao.mean.cascade_asr.axis2.2_3\endcsname{[0.702, 0.762]}
\expandafter\gdef\csname odunum@val@positive_taxonomy_m2.ao.mean.cascade_asr.axis2.2_4\endcsname{0.614}
\expandafter\gdef\csname odunum@n@positive_taxonomy_m2.ao.mean.cascade_asr.axis2.2_4\endcsname{34}
\expandafter\gdef\csname odunum@ci@positive_taxonomy_m2.ao.mean.cascade_asr.axis2.2_4\endcsname{[0.494, 0.731]}
\expandafter\gdef\csname odunum@val@positive_taxonomy_m2.ao.mean.cascade_asr.axis2.2_5\endcsname{0.589}
\expandafter\gdef\csname odunum@n@positive_taxonomy_m2.ao.mean.cascade_asr.axis2.2_5\endcsname{91}
\expandafter\gdef\csname odunum@ci@positive_taxonomy_m2.ao.mean.cascade_asr.axis2.2_5\endcsname{[0.515, 0.660]}
\expandafter\gdef\csname odunum@val@positive_taxonomy_m2.ao.mean.cascade_asr.axis3.3_1\endcsname{0.759}
\expandafter\gdef\csname odunum@n@positive_taxonomy_m2.ao.mean.cascade_asr.axis3.3_1\endcsname{18}
\expandafter\gdef\csname odunum@ci@positive_taxonomy_m2.ao.mean.cascade_asr.axis3.3_1\endcsname{[0.614, 0.887]}
\expandafter\gdef\csname odunum@val@positive_taxonomy_m2.ao.mean.cascade_asr.axis3.3_2\endcsname{0.652}
\expandafter\gdef\csname odunum@n@positive_taxonomy_m2.ao.mean.cascade_asr.axis3.3_2\endcsname{30}
\expandafter\gdef\csname odunum@ci@positive_taxonomy_m2.ao.mean.cascade_asr.axis3.3_2\endcsname{[0.580, 0.726]}
\expandafter\gdef\csname odunum@val@positive_taxonomy_m2.ao.mean.cascade_asr.axis3.3_3\endcsname{0.555}
\expandafter\gdef\csname odunum@n@positive_taxonomy_m2.ao.mean.cascade_asr.axis3.3_3\endcsname{67}
\expandafter\gdef\csname odunum@ci@positive_taxonomy_m2.ao.mean.cascade_asr.axis3.3_3\endcsname{[0.461, 0.646]}
\expandafter\gdef\csname odunum@val@positive_taxonomy_m2.ao.mean.cascade_asr.axis3.3_4\endcsname{0.727}
\expandafter\gdef\csname odunum@n@positive_taxonomy_m2.ao.mean.cascade_asr.axis3.3_4\endcsname{141}
\expandafter\gdef\csname odunum@ci@positive_taxonomy_m2.ao.mean.cascade_asr.axis3.3_4\endcsname{[0.681, 0.770]}
\expandafter\gdef\csname odunum@val@positive_taxonomy_m2.ao.mean.cascade_asr.axis3.3_5\endcsname{0.578}
\expandafter\gdef\csname odunum@n@positive_taxonomy_m2.ao.mean.cascade_asr.axis3.3_5\endcsname{105}
\expandafter\gdef\csname odunum@ci@positive_taxonomy_m2.ao.mean.cascade_asr.axis3.3_5\endcsname{[0.522, 0.633]}
\expandafter\gdef\csname odunum@val@positive_taxonomy_m2.ao.mean.cascade_asr.axis3.3_6\endcsname{0.748}
\expandafter\gdef\csname odunum@n@positive_taxonomy_m2.ao.mean.cascade_asr.axis3.3_6\endcsname{202}
\expandafter\gdef\csname odunum@ci@positive_taxonomy_m2.ao.mean.cascade_asr.axis3.3_6\endcsname{[0.713, 0.781]}
\expandafter\gdef\csname odunum@val@positive_taxonomy_m2.ao.mean.cascade_asr.axis4.4_1\endcsname{0.683}
\expandafter\gdef\csname odunum@n@positive_taxonomy_m2.ao.mean.cascade_asr.axis4.4_1\endcsname{63}
\expandafter\gdef\csname odunum@ci@positive_taxonomy_m2.ao.mean.cascade_asr.axis4.4_1\endcsname{[0.619, 0.744]}
\expandafter\gdef\csname odunum@val@positive_taxonomy_m2.ao.mean.cascade_asr.axis4.4_2\endcsname{0.689}
\expandafter\gdef\csname odunum@n@positive_taxonomy_m2.ao.mean.cascade_asr.axis4.4_2\endcsname{161}
\expandafter\gdef\csname odunum@ci@positive_taxonomy_m2.ao.mean.cascade_asr.axis4.4_2\endcsname{[0.644, 0.731]}
\expandafter\gdef\csname odunum@val@positive_taxonomy_m2.ao.mean.cascade_asr.axis4.4_3\endcsname{0.677}
\expandafter\gdef\csname odunum@n@positive_taxonomy_m2.ao.mean.cascade_asr.axis4.4_3\endcsname{43}
\expandafter\gdef\csname odunum@ci@positive_taxonomy_m2.ao.mean.cascade_asr.axis4.4_3\endcsname{[0.580, 0.765]}
\expandafter\gdef\csname odunum@val@positive_taxonomy_m2.ao.mean.cascade_asr.axis4.4_4\endcsname{0.666}
\expandafter\gdef\csname odunum@n@positive_taxonomy_m2.ao.mean.cascade_asr.axis4.4_4\endcsname{172}
\expandafter\gdef\csname odunum@ci@positive_taxonomy_m2.ao.mean.cascade_asr.axis4.4_4\endcsname{[0.618, 0.712]}
\expandafter\gdef\csname odunum@val@positive_taxonomy_m2.ao.mean.cascade_asr.axis4.4_5\endcsname{0.701}
\expandafter\gdef\csname odunum@n@positive_taxonomy_m2.ao.mean.cascade_asr.axis4.4_5\endcsname{124}
\expandafter\gdef\csname odunum@ci@positive_taxonomy_m2.ao.mean.cascade_asr.axis4.4_5\endcsname{[0.651, 0.748]}
\expandafter\gdef\csname odunum@val@positive_taxonomy_m2.ao.mean.gemini.axis1.1_1\endcsname{0.708}
\expandafter\gdef\csname odunum@n@positive_taxonomy_m2.ao.mean.gemini.axis1.1_1\endcsname{156}
\expandafter\gdef\csname odunum@ci@positive_taxonomy_m2.ao.mean.gemini.axis1.1_1\endcsname{[0.667, 0.747]}
\expandafter\gdef\csname odunum@val@positive_taxonomy_m2.ao.mean.gemini.axis1.1_2\endcsname{0.697}
\expandafter\gdef\csname odunum@n@positive_taxonomy_m2.ao.mean.gemini.axis1.1_2\endcsname{407}
\expandafter\gdef\csname odunum@ci@positive_taxonomy_m2.ao.mean.gemini.axis1.1_2\endcsname{[0.673, 0.722]}
\expandafter\gdef\csname odunum@val@positive_taxonomy_m2.ao.mean.gemini.axis2.2_1\endcsname{0.695}
\expandafter\gdef\csname odunum@n@positive_taxonomy_m2.ao.mean.gemini.axis2.2_1\endcsname{21}
\expandafter\gdef\csname odunum@ci@positive_taxonomy_m2.ao.mean.gemini.axis2.2_1\endcsname{[0.551, 0.829]}
\expandafter\gdef\csname odunum@val@positive_taxonomy_m2.ao.mean.gemini.axis2.2_2\endcsname{0.711}
\expandafter\gdef\csname odunum@n@positive_taxonomy_m2.ao.mean.gemini.axis2.2_2\endcsname{169}
\expandafter\gdef\csname odunum@ci@positive_taxonomy_m2.ao.mean.gemini.axis2.2_2\endcsname{[0.675, 0.747]}
\expandafter\gdef\csname odunum@val@positive_taxonomy_m2.ao.mean.gemini.axis2.2_3\endcsname{0.701}
\expandafter\gdef\csname odunum@n@positive_taxonomy_m2.ao.mean.gemini.axis2.2_3\endcsname{248}
\expandafter\gdef\csname odunum@ci@positive_taxonomy_m2.ao.mean.gemini.axis2.2_3\endcsname{[0.671, 0.730]}
\expandafter\gdef\csname odunum@val@positive_taxonomy_m2.ao.mean.gemini.axis2.2_4\endcsname{0.649}
\expandafter\gdef\csname odunum@n@positive_taxonomy_m2.ao.mean.gemini.axis2.2_4\endcsname{34}
\expandafter\gdef\csname odunum@ci@positive_taxonomy_m2.ao.mean.gemini.axis2.2_4\endcsname{[0.551, 0.741]}
\expandafter\gdef\csname odunum@val@positive_taxonomy_m2.ao.mean.gemini.axis2.2_5\endcsname{0.699}
\expandafter\gdef\csname odunum@n@positive_taxonomy_m2.ao.mean.gemini.axis2.2_5\endcsname{91}
\expandafter\gdef\csname odunum@ci@positive_taxonomy_m2.ao.mean.gemini.axis2.2_5\endcsname{[0.640, 0.757]}
\expandafter\gdef\csname odunum@val@positive_taxonomy_m2.ao.mean.gemini.axis3.3_1\endcsname{0.764}
\expandafter\gdef\csname odunum@n@positive_taxonomy_m2.ao.mean.gemini.axis3.3_1\endcsname{18}
\expandafter\gdef\csname odunum@ci@positive_taxonomy_m2.ao.mean.gemini.axis3.3_1\endcsname{[0.667, 0.864]}
\expandafter\gdef\csname odunum@val@positive_taxonomy_m2.ao.mean.gemini.axis3.3_2\endcsname{0.715}
\expandafter\gdef\csname odunum@n@positive_taxonomy_m2.ao.mean.gemini.axis3.3_2\endcsname{30}
\expandafter\gdef\csname odunum@ci@positive_taxonomy_m2.ao.mean.gemini.axis3.3_2\endcsname{[0.620, 0.803]}
\expandafter\gdef\csname odunum@val@positive_taxonomy_m2.ao.mean.gemini.axis3.3_3\endcsname{0.686}
\expandafter\gdef\csname odunum@n@positive_taxonomy_m2.ao.mean.gemini.axis3.3_3\endcsname{67}
\expandafter\gdef\csname odunum@ci@positive_taxonomy_m2.ao.mean.gemini.axis3.3_3\endcsname{[0.620, 0.751]}
\expandafter\gdef\csname odunum@val@positive_taxonomy_m2.ao.mean.gemini.axis3.3_4\endcsname{0.663}
\expandafter\gdef\csname odunum@n@positive_taxonomy_m2.ao.mean.gemini.axis3.3_4\endcsname{141}
\expandafter\gdef\csname odunum@ci@positive_taxonomy_m2.ao.mean.gemini.axis3.3_4\endcsname{[0.617, 0.708]}
\expandafter\gdef\csname odunum@val@positive_taxonomy_m2.ao.mean.gemini.axis3.3_5\endcsname{0.625}
\expandafter\gdef\csname odunum@n@positive_taxonomy_m2.ao.mean.gemini.axis3.3_5\endcsname{105}
\expandafter\gdef\csname odunum@ci@positive_taxonomy_m2.ao.mean.gemini.axis3.3_5\endcsname{[0.575, 0.676]}
\expandafter\gdef\csname odunum@val@positive_taxonomy_m2.ao.mean.gemini.axis3.3_6\endcsname{0.762}
\expandafter\gdef\csname odunum@n@positive_taxonomy_m2.ao.mean.gemini.axis3.3_6\endcsname{202}
\expandafter\gdef\csname odunum@ci@positive_taxonomy_m2.ao.mean.gemini.axis3.3_6\endcsname{[0.732, 0.791]}
\expandafter\gdef\csname odunum@val@positive_taxonomy_m2.ao.mean.gemini.axis4.4_1\endcsname{0.692}
\expandafter\gdef\csname odunum@n@positive_taxonomy_m2.ao.mean.gemini.axis4.4_1\endcsname{63}
\expandafter\gdef\csname odunum@ci@positive_taxonomy_m2.ao.mean.gemini.axis4.4_1\endcsname{[0.641, 0.743]}
\expandafter\gdef\csname odunum@val@positive_taxonomy_m2.ao.mean.gemini.axis4.4_2\endcsname{0.701}
\expandafter\gdef\csname odunum@n@positive_taxonomy_m2.ao.mean.gemini.axis4.4_2\endcsname{161}
\expandafter\gdef\csname odunum@ci@positive_taxonomy_m2.ao.mean.gemini.axis4.4_2\endcsname{[0.661, 0.741]}
\expandafter\gdef\csname odunum@val@positive_taxonomy_m2.ao.mean.gemini.axis4.4_3\endcsname{0.690}
\expandafter\gdef\csname odunum@n@positive_taxonomy_m2.ao.mean.gemini.axis4.4_3\endcsname{43}
\expandafter\gdef\csname odunum@ci@positive_taxonomy_m2.ao.mean.gemini.axis4.4_3\endcsname{[0.604, 0.770]}
\expandafter\gdef\csname odunum@val@positive_taxonomy_m2.ao.mean.gemini.axis4.4_4\endcsname{0.726}
\expandafter\gdef\csname odunum@n@positive_taxonomy_m2.ao.mean.gemini.axis4.4_4\endcsname{172}
\expandafter\gdef\csname odunum@ci@positive_taxonomy_m2.ao.mean.gemini.axis4.4_4\endcsname{[0.688, 0.762]}
\expandafter\gdef\csname odunum@val@positive_taxonomy_m2.ao.mean.gemini.axis4.4_5\endcsname{0.672}
\expandafter\gdef\csname odunum@n@positive_taxonomy_m2.ao.mean.gemini.axis4.4_5\endcsname{124}
\expandafter\gdef\csname odunum@ci@positive_taxonomy_m2.ao.mean.gemini.axis4.4_5\endcsname{[0.624, 0.718]}
\expandafter\gdef\csname odunum@val@positive_taxonomy_m2.ao.mean.gemini35_flash_lite.axis1.1_1\endcsname{0.655}
\expandafter\gdef\csname odunum@n@positive_taxonomy_m2.ao.mean.gemini35_flash_lite.axis1.1_1\endcsname{156}
\expandafter\gdef\csname odunum@ci@positive_taxonomy_m2.ao.mean.gemini35_flash_lite.axis1.1_1\endcsname{[0.603, 0.704]}
\expandafter\gdef\csname odunum@val@positive_taxonomy_m2.ao.mean.gemini35_flash_lite.axis1.1_2\endcsname{0.531}
\expandafter\gdef\csname odunum@n@positive_taxonomy_m2.ao.mean.gemini35_flash_lite.axis1.1_2\endcsname{407}
\expandafter\gdef\csname odunum@ci@positive_taxonomy_m2.ao.mean.gemini35_flash_lite.axis1.1_2\endcsname{[0.499, 0.563]}
\expandafter\gdef\csname odunum@val@positive_taxonomy_m2.ao.mean.gemini35_flash_lite.axis2.2_1\endcsname{0.417}
\expandafter\gdef\csname odunum@n@positive_taxonomy_m2.ao.mean.gemini35_flash_lite.axis2.2_1\endcsname{21}
\expandafter\gdef\csname odunum@ci@positive_taxonomy_m2.ao.mean.gemini35_flash_lite.axis2.2_1\endcsname{[0.248, 0.593]}
\expandafter\gdef\csname odunum@val@positive_taxonomy_m2.ao.mean.gemini35_flash_lite.axis2.2_2\endcsname{0.607}
\expandafter\gdef\csname odunum@n@positive_taxonomy_m2.ao.mean.gemini35_flash_lite.axis2.2_2\endcsname{169}
\expandafter\gdef\csname odunum@ci@positive_taxonomy_m2.ao.mean.gemini35_flash_lite.axis2.2_2\endcsname{[0.559, 0.654]}
\expandafter\gdef\csname odunum@val@positive_taxonomy_m2.ao.mean.gemini35_flash_lite.axis2.2_3\endcsname{0.645}
\expandafter\gdef\csname odunum@n@positive_taxonomy_m2.ao.mean.gemini35_flash_lite.axis2.2_3\endcsname{248}
\expandafter\gdef\csname odunum@ci@positive_taxonomy_m2.ao.mean.gemini35_flash_lite.axis2.2_3\endcsname{[0.610, 0.678]}
\expandafter\gdef\csname odunum@val@positive_taxonomy_m2.ao.mean.gemini35_flash_lite.axis2.2_4\endcsname{0.347}
\expandafter\gdef\csname odunum@n@positive_taxonomy_m2.ao.mean.gemini35_flash_lite.axis2.2_4\endcsname{34}
\expandafter\gdef\csname odunum@ci@positive_taxonomy_m2.ao.mean.gemini35_flash_lite.axis2.2_4\endcsname{[0.231, 0.469]}
\expandafter\gdef\csname odunum@val@positive_taxonomy_m2.ao.mean.gemini35_flash_lite.axis2.2_5\endcsname{0.385}
\expandafter\gdef\csname odunum@n@positive_taxonomy_m2.ao.mean.gemini35_flash_lite.axis2.2_5\endcsname{91}
\expandafter\gdef\csname odunum@ci@positive_taxonomy_m2.ao.mean.gemini35_flash_lite.axis2.2_5\endcsname{[0.310, 0.465]}
\expandafter\gdef\csname odunum@val@positive_taxonomy_m2.ao.mean.gemini35_flash_lite.axis3.3_1\endcsname{0.714}
\expandafter\gdef\csname odunum@n@positive_taxonomy_m2.ao.mean.gemini35_flash_lite.axis3.3_1\endcsname{18}
\expandafter\gdef\csname odunum@ci@positive_taxonomy_m2.ao.mean.gemini35_flash_lite.axis3.3_1\endcsname{[0.571, 0.843]}
\expandafter\gdef\csname odunum@val@positive_taxonomy_m2.ao.mean.gemini35_flash_lite.axis3.3_2\endcsname{0.583}
\expandafter\gdef\csname odunum@n@positive_taxonomy_m2.ao.mean.gemini35_flash_lite.axis3.3_2\endcsname{30}
\expandafter\gdef\csname odunum@ci@positive_taxonomy_m2.ao.mean.gemini35_flash_lite.axis3.3_2\endcsname{[0.485, 0.682]}
\expandafter\gdef\csname odunum@val@positive_taxonomy_m2.ao.mean.gemini35_flash_lite.axis3.3_3\endcsname{0.364}
\expandafter\gdef\csname odunum@n@positive_taxonomy_m2.ao.mean.gemini35_flash_lite.axis3.3_3\endcsname{67}
\expandafter\gdef\csname odunum@ci@positive_taxonomy_m2.ao.mean.gemini35_flash_lite.axis3.3_3\endcsname{[0.274, 0.457]}
\expandafter\gdef\csname odunum@val@positive_taxonomy_m2.ao.mean.gemini35_flash_lite.axis3.3_4\endcsname{0.598}
\expandafter\gdef\csname odunum@n@positive_taxonomy_m2.ao.mean.gemini35_flash_lite.axis3.3_4\endcsname{141}
\expandafter\gdef\csname odunum@ci@positive_taxonomy_m2.ao.mean.gemini35_flash_lite.axis3.3_4\endcsname{[0.547, 0.648]}
\expandafter\gdef\csname odunum@val@positive_taxonomy_m2.ao.mean.gemini35_flash_lite.axis3.3_5\endcsname{0.504}
\expandafter\gdef\csname odunum@n@positive_taxonomy_m2.ao.mean.gemini35_flash_lite.axis3.3_5\endcsname{105}
\expandafter\gdef\csname odunum@ci@positive_taxonomy_m2.ao.mean.gemini35_flash_lite.axis3.3_5\endcsname{[0.439, 0.570]}
\expandafter\gdef\csname odunum@val@positive_taxonomy_m2.ao.mean.gemini35_flash_lite.axis3.3_6\endcsname{0.625}
\expandafter\gdef\csname odunum@n@positive_taxonomy_m2.ao.mean.gemini35_flash_lite.axis3.3_6\endcsname{202}
\expandafter\gdef\csname odunum@ci@positive_taxonomy_m2.ao.mean.gemini35_flash_lite.axis3.3_6\endcsname{[0.582, 0.667]}
\expandafter\gdef\csname odunum@val@positive_taxonomy_m2.ao.mean.gemini35_flash_lite.axis4.4_1\endcsname{0.550}
\expandafter\gdef\csname odunum@n@positive_taxonomy_m2.ao.mean.gemini35_flash_lite.axis4.4_1\endcsname{63}
\expandafter\gdef\csname odunum@ci@positive_taxonomy_m2.ao.mean.gemini35_flash_lite.axis4.4_1\endcsname{[0.473, 0.623]}
\expandafter\gdef\csname odunum@val@positive_taxonomy_m2.ao.mean.gemini35_flash_lite.axis4.4_2\endcsname{0.587}
\expandafter\gdef\csname odunum@n@positive_taxonomy_m2.ao.mean.gemini35_flash_lite.axis4.4_2\endcsname{161}
\expandafter\gdef\csname odunum@ci@positive_taxonomy_m2.ao.mean.gemini35_flash_lite.axis4.4_2\endcsname{[0.537, 0.634]}
\expandafter\gdef\csname odunum@val@positive_taxonomy_m2.ao.mean.gemini35_flash_lite.axis4.4_3\endcsname{0.462}
\expandafter\gdef\csname odunum@n@positive_taxonomy_m2.ao.mean.gemini35_flash_lite.axis4.4_3\endcsname{43}
\expandafter\gdef\csname odunum@ci@positive_taxonomy_m2.ao.mean.gemini35_flash_lite.axis4.4_3\endcsname{[0.350, 0.572]}
\expandafter\gdef\csname odunum@val@positive_taxonomy_m2.ao.mean.gemini35_flash_lite.axis4.4_4\endcsname{0.540}
\expandafter\gdef\csname odunum@n@positive_taxonomy_m2.ao.mean.gemini35_flash_lite.axis4.4_4\endcsname{172}
\expandafter\gdef\csname odunum@ci@positive_taxonomy_m2.ao.mean.gemini35_flash_lite.axis4.4_4\endcsname{[0.490, 0.590]}
\expandafter\gdef\csname odunum@val@positive_taxonomy_m2.ao.mean.gemini35_flash_lite.axis4.4_5\endcsname{0.615}
\expandafter\gdef\csname odunum@n@positive_taxonomy_m2.ao.mean.gemini35_flash_lite.axis4.4_5\endcsname{124}
\expandafter\gdef\csname odunum@ci@positive_taxonomy_m2.ao.mean.gemini35_flash_lite.axis4.4_5\endcsname{[0.554, 0.674]}
\expandafter\gdef\csname odunum@val@positive_taxonomy_m2.ao.mean.gemini37_flash.axis1.1_1\endcsname{0.718}
\expandafter\gdef\csname odunum@n@positive_taxonomy_m2.ao.mean.gemini37_flash.axis1.1_1\endcsname{156}
\expandafter\gdef\csname odunum@ci@positive_taxonomy_m2.ao.mean.gemini37_flash.axis1.1_1\endcsname{[0.676, 0.760]}
\expandafter\gdef\csname odunum@val@positive_taxonomy_m2.ao.mean.gemini37_flash.axis1.1_2\endcsname{0.637}
\expandafter\gdef\csname odunum@n@positive_taxonomy_m2.ao.mean.gemini37_flash.axis1.1_2\endcsname{407}
\expandafter\gdef\csname odunum@ci@positive_taxonomy_m2.ao.mean.gemini37_flash.axis1.1_2\endcsname{[0.609, 0.664]}
\expandafter\gdef\csname odunum@val@positive_taxonomy_m2.ao.mean.gemini37_flash.axis2.2_1\endcsname{0.630}
\expandafter\gdef\csname odunum@n@positive_taxonomy_m2.ao.mean.gemini37_flash.axis2.2_1\endcsname{21}
\expandafter\gdef\csname odunum@ci@positive_taxonomy_m2.ao.mean.gemini37_flash.axis2.2_1\endcsname{[0.484, 0.768]}
\expandafter\gdef\csname odunum@val@positive_taxonomy_m2.ao.mean.gemini37_flash.axis2.2_2\endcsname{0.665}
\expandafter\gdef\csname odunum@n@positive_taxonomy_m2.ao.mean.gemini37_flash.axis2.2_2\endcsname{169}
\expandafter\gdef\csname odunum@ci@positive_taxonomy_m2.ao.mean.gemini37_flash.axis2.2_2\endcsname{[0.626, 0.703]}
\expandafter\gdef\csname odunum@val@positive_taxonomy_m2.ao.mean.gemini37_flash.axis2.2_3\endcsname{0.693}
\expandafter\gdef\csname odunum@n@positive_taxonomy_m2.ao.mean.gemini37_flash.axis2.2_3\endcsname{248}
\expandafter\gdef\csname odunum@ci@positive_taxonomy_m2.ao.mean.gemini37_flash.axis2.2_3\endcsname{[0.661, 0.724]}
\expandafter\gdef\csname odunum@val@positive_taxonomy_m2.ao.mean.gemini37_flash.axis2.2_4\endcsname{0.621}
\expandafter\gdef\csname odunum@n@positive_taxonomy_m2.ao.mean.gemini37_flash.axis2.2_4\endcsname{34}
\expandafter\gdef\csname odunum@ci@positive_taxonomy_m2.ao.mean.gemini37_flash.axis2.2_4\endcsname{[0.504, 0.730]}
\expandafter\gdef\csname odunum@val@positive_taxonomy_m2.ao.mean.gemini37_flash.axis2.2_5\endcsname{0.580}
\expandafter\gdef\csname odunum@n@positive_taxonomy_m2.ao.mean.gemini37_flash.axis2.2_5\endcsname{91}
\expandafter\gdef\csname odunum@ci@positive_taxonomy_m2.ao.mean.gemini37_flash.axis2.2_5\endcsname{[0.512, 0.647]}
\expandafter\gdef\csname odunum@val@positive_taxonomy_m2.ao.mean.gemini37_flash.axis3.3_1\endcsname{0.717}
\expandafter\gdef\csname odunum@n@positive_taxonomy_m2.ao.mean.gemini37_flash.axis3.3_1\endcsname{18}
\expandafter\gdef\csname odunum@ci@positive_taxonomy_m2.ao.mean.gemini37_flash.axis3.3_1\endcsname{[0.563, 0.854]}
\expandafter\gdef\csname odunum@val@positive_taxonomy_m2.ao.mean.gemini37_flash.axis3.3_2\endcsname{0.632}
\expandafter\gdef\csname odunum@n@positive_taxonomy_m2.ao.mean.gemini37_flash.axis3.3_2\endcsname{30}
\expandafter\gdef\csname odunum@ci@positive_taxonomy_m2.ao.mean.gemini37_flash.axis3.3_2\endcsname{[0.544, 0.722]}
\expandafter\gdef\csname odunum@val@positive_taxonomy_m2.ao.mean.gemini37_flash.axis3.3_3\endcsname{0.559}
\expandafter\gdef\csname odunum@n@positive_taxonomy_m2.ao.mean.gemini37_flash.axis3.3_3\endcsname{67}
\expandafter\gdef\csname odunum@ci@positive_taxonomy_m2.ao.mean.gemini37_flash.axis3.3_3\endcsname{[0.480, 0.636]}
\expandafter\gdef\csname odunum@val@positive_taxonomy_m2.ao.mean.gemini37_flash.axis3.3_4\endcsname{0.663}
\expandafter\gdef\csname odunum@n@positive_taxonomy_m2.ao.mean.gemini37_flash.axis3.3_4\endcsname{141}
\expandafter\gdef\csname odunum@ci@positive_taxonomy_m2.ao.mean.gemini37_flash.axis3.3_4\endcsname{[0.617, 0.709]}
\expandafter\gdef\csname odunum@val@positive_taxonomy_m2.ao.mean.gemini37_flash.axis3.3_5\endcsname{0.570}
\expandafter\gdef\csname odunum@n@positive_taxonomy_m2.ao.mean.gemini37_flash.axis3.3_5\endcsname{105}
\expandafter\gdef\csname odunum@ci@positive_taxonomy_m2.ao.mean.gemini37_flash.axis3.3_5\endcsname{[0.518, 0.622]}
\expandafter\gdef\csname odunum@val@positive_taxonomy_m2.ao.mean.gemini37_flash.axis3.3_6\endcsname{0.736}
\expandafter\gdef\csname odunum@n@positive_taxonomy_m2.ao.mean.gemini37_flash.axis3.3_6\endcsname{202}
\expandafter\gdef\csname odunum@ci@positive_taxonomy_m2.ao.mean.gemini37_flash.axis3.3_6\endcsname{[0.702, 0.769]}
\expandafter\gdef\csname odunum@val@positive_taxonomy_m2.ao.mean.gemini37_flash.axis4.4_1\endcsname{0.692}
\expandafter\gdef\csname odunum@n@positive_taxonomy_m2.ao.mean.gemini37_flash.axis4.4_1\endcsname{63}
\expandafter\gdef\csname odunum@ci@positive_taxonomy_m2.ao.mean.gemini37_flash.axis4.4_1\endcsname{[0.637, 0.746]}
\expandafter\gdef\csname odunum@val@positive_taxonomy_m2.ao.mean.gemini37_flash.axis4.4_2\endcsname{0.652}
\expandafter\gdef\csname odunum@n@positive_taxonomy_m2.ao.mean.gemini37_flash.axis4.4_2\endcsname{161}
\expandafter\gdef\csname odunum@ci@positive_taxonomy_m2.ao.mean.gemini37_flash.axis4.4_2\endcsname{[0.611, 0.692]}
\expandafter\gdef\csname odunum@val@positive_taxonomy_m2.ao.mean.gemini37_flash.axis4.4_3\endcsname{0.681}
\expandafter\gdef\csname odunum@n@positive_taxonomy_m2.ao.mean.gemini37_flash.axis4.4_3\endcsname{43}
\expandafter\gdef\csname odunum@ci@positive_taxonomy_m2.ao.mean.gemini37_flash.axis4.4_3\endcsname{[0.591, 0.764]}
\expandafter\gdef\csname odunum@val@positive_taxonomy_m2.ao.mean.gemini37_flash.axis4.4_4\endcsname{0.649}
\expandafter\gdef\csname odunum@n@positive_taxonomy_m2.ao.mean.gemini37_flash.axis4.4_4\endcsname{172}
\expandafter\gdef\csname odunum@ci@positive_taxonomy_m2.ao.mean.gemini37_flash.axis4.4_4\endcsname{[0.603, 0.694]}
\expandafter\gdef\csname odunum@val@positive_taxonomy_m2.ao.mean.gemini37_flash.axis4.4_5\endcsname{0.660}
\expandafter\gdef\csname odunum@n@positive_taxonomy_m2.ao.mean.gemini37_flash.axis4.4_5\endcsname{124}
\expandafter\gdef\csname odunum@ci@positive_taxonomy_m2.ao.mean.gemini37_flash.axis4.4_5\endcsname{[0.608, 0.712]}
\expandafter\gdef\csname odunum@val@positive_taxonomy_m2.ao.mean.gpt_realtime.axis1.1_1\endcsname{0.691}
\expandafter\gdef\csname odunum@n@positive_taxonomy_m2.ao.mean.gpt_realtime.axis1.1_1\endcsname{156}
\expandafter\gdef\csname odunum@ci@positive_taxonomy_m2.ao.mean.gpt_realtime.axis1.1_1\endcsname{[0.647, 0.735]}
\expandafter\gdef\csname odunum@val@positive_taxonomy_m2.ao.mean.gpt_realtime.axis1.1_2\endcsname{0.611}
\expandafter\gdef\csname odunum@n@positive_taxonomy_m2.ao.mean.gpt_realtime.axis1.1_2\endcsname{407}
\expandafter\gdef\csname odunum@ci@positive_taxonomy_m2.ao.mean.gpt_realtime.axis1.1_2\endcsname{[0.582, 0.639]}
\expandafter\gdef\csname odunum@val@positive_taxonomy_m2.ao.mean.gpt_realtime.axis2.2_1\endcsname{0.505}
\expandafter\gdef\csname odunum@n@positive_taxonomy_m2.ao.mean.gpt_realtime.axis2.2_1\endcsname{21}
\expandafter\gdef\csname odunum@ci@positive_taxonomy_m2.ao.mean.gpt_realtime.axis2.2_1\endcsname{[0.347, 0.654]}
\expandafter\gdef\csname odunum@val@positive_taxonomy_m2.ao.mean.gpt_realtime.axis2.2_2\endcsname{0.650}
\expandafter\gdef\csname odunum@n@positive_taxonomy_m2.ao.mean.gpt_realtime.axis2.2_2\endcsname{169}
\expandafter\gdef\csname odunum@ci@positive_taxonomy_m2.ao.mean.gpt_realtime.axis2.2_2\endcsname{[0.612, 0.688]}
\expandafter\gdef\csname odunum@val@positive_taxonomy_m2.ao.mean.gpt_realtime.axis2.2_3\endcsname{0.681}
\expandafter\gdef\csname odunum@n@positive_taxonomy_m2.ao.mean.gpt_realtime.axis2.2_3\endcsname{248}
\expandafter\gdef\csname odunum@ci@positive_taxonomy_m2.ao.mean.gpt_realtime.axis2.2_3\endcsname{[0.648, 0.713]}
\expandafter\gdef\csname odunum@val@positive_taxonomy_m2.ao.mean.gpt_realtime.axis2.2_4\endcsname{0.459}
\expandafter\gdef\csname odunum@n@positive_taxonomy_m2.ao.mean.gpt_realtime.axis2.2_4\endcsname{34}
\expandafter\gdef\csname odunum@ci@positive_taxonomy_m2.ao.mean.gpt_realtime.axis2.2_4\endcsname{[0.329, 0.587]}
\expandafter\gdef\csname odunum@val@positive_taxonomy_m2.ao.mean.gpt_realtime.axis2.2_5\endcsname{0.566}
\expandafter\gdef\csname odunum@n@positive_taxonomy_m2.ao.mean.gpt_realtime.axis2.2_5\endcsname{91}
\expandafter\gdef\csname odunum@ci@positive_taxonomy_m2.ao.mean.gpt_realtime.axis2.2_5\endcsname{[0.493, 0.636]}
\expandafter\gdef\csname odunum@val@positive_taxonomy_m2.ao.mean.gpt_realtime.axis3.3_1\endcsname{0.730}
\expandafter\gdef\csname odunum@n@positive_taxonomy_m2.ao.mean.gpt_realtime.axis3.3_1\endcsname{18}
\expandafter\gdef\csname odunum@ci@positive_taxonomy_m2.ao.mean.gpt_realtime.axis3.3_1\endcsname{[0.594, 0.853]}
\expandafter\gdef\csname odunum@val@positive_taxonomy_m2.ao.mean.gpt_realtime.axis3.3_2\endcsname{0.582}
\expandafter\gdef\csname odunum@n@positive_taxonomy_m2.ao.mean.gpt_realtime.axis3.3_2\endcsname{30}
\expandafter\gdef\csname odunum@ci@positive_taxonomy_m2.ao.mean.gpt_realtime.axis3.3_2\endcsname{[0.474, 0.688]}
\expandafter\gdef\csname odunum@val@positive_taxonomy_m2.ao.mean.gpt_realtime.axis3.3_3\endcsname{0.566}
\expandafter\gdef\csname odunum@n@positive_taxonomy_m2.ao.mean.gpt_realtime.axis3.3_3\endcsname{67}
\expandafter\gdef\csname odunum@ci@positive_taxonomy_m2.ao.mean.gpt_realtime.axis3.3_3\endcsname{[0.475, 0.657]}
\expandafter\gdef\csname odunum@val@positive_taxonomy_m2.ao.mean.gpt_realtime.axis3.3_4\endcsname{0.631}
\expandafter\gdef\csname odunum@n@positive_taxonomy_m2.ao.mean.gpt_realtime.axis3.3_4\endcsname{141}
\expandafter\gdef\csname odunum@ci@positive_taxonomy_m2.ao.mean.gpt_realtime.axis3.3_4\endcsname{[0.584, 0.678]}
\expandafter\gdef\csname odunum@val@positive_taxonomy_m2.ao.mean.gpt_realtime.axis3.3_5\endcsname{0.580}
\expandafter\gdef\csname odunum@n@positive_taxonomy_m2.ao.mean.gpt_realtime.axis3.3_5\endcsname{105}
\expandafter\gdef\csname odunum@ci@positive_taxonomy_m2.ao.mean.gpt_realtime.axis3.3_5\endcsname{[0.530, 0.631]}
\expandafter\gdef\csname odunum@val@positive_taxonomy_m2.ao.mean.gpt_realtime.axis3.3_6\endcsname{0.683}
\expandafter\gdef\csname odunum@n@positive_taxonomy_m2.ao.mean.gpt_realtime.axis3.3_6\endcsname{202}
\expandafter\gdef\csname odunum@ci@positive_taxonomy_m2.ao.mean.gpt_realtime.axis3.3_6\endcsname{[0.645, 0.720]}
\expandafter\gdef\csname odunum@val@positive_taxonomy_m2.ao.mean.gpt_realtime.axis4.4_1\endcsname{0.634}
\expandafter\gdef\csname odunum@n@positive_taxonomy_m2.ao.mean.gpt_realtime.axis4.4_1\endcsname{63}
\expandafter\gdef\csname odunum@ci@positive_taxonomy_m2.ao.mean.gpt_realtime.axis4.4_1\endcsname{[0.563, 0.702]}
\expandafter\gdef\csname odunum@val@positive_taxonomy_m2.ao.mean.gpt_realtime.axis4.4_2\endcsname{0.638}
\expandafter\gdef\csname odunum@n@positive_taxonomy_m2.ao.mean.gpt_realtime.axis4.4_2\endcsname{161}
\expandafter\gdef\csname odunum@ci@positive_taxonomy_m2.ao.mean.gpt_realtime.axis4.4_2\endcsname{[0.594, 0.679]}
\expandafter\gdef\csname odunum@val@positive_taxonomy_m2.ao.mean.gpt_realtime.axis4.4_3\endcsname{0.579}
\expandafter\gdef\csname odunum@n@positive_taxonomy_m2.ao.mean.gpt_realtime.axis4.4_3\endcsname{43}
\expandafter\gdef\csname odunum@ci@positive_taxonomy_m2.ao.mean.gpt_realtime.axis4.4_3\endcsname{[0.470, 0.684]}
\expandafter\gdef\csname odunum@val@positive_taxonomy_m2.ao.mean.gpt_realtime.axis4.4_4\endcsname{0.640}
\expandafter\gdef\csname odunum@n@positive_taxonomy_m2.ao.mean.gpt_realtime.axis4.4_4\endcsname{172}
\expandafter\gdef\csname odunum@ci@positive_taxonomy_m2.ao.mean.gpt_realtime.axis4.4_4\endcsname{[0.598, 0.683]}
\expandafter\gdef\csname odunum@val@positive_taxonomy_m2.ao.mean.gpt_realtime.axis4.4_5\endcsname{0.635}
\expandafter\gdef\csname odunum@n@positive_taxonomy_m2.ao.mean.gpt_realtime.axis4.4_5\endcsname{124}
\expandafter\gdef\csname odunum@ci@positive_taxonomy_m2.ao.mean.gpt_realtime.axis4.4_5\endcsname{[0.581, 0.689]}
\expandafter\gdef\csname odunum@val@positive_taxonomy_m2.ao.mean.kimi_audio_7b_instruct.axis1.1_1\endcsname{0.565}
\expandafter\gdef\csname odunum@n@positive_taxonomy_m2.ao.mean.kimi_audio_7b_instruct.axis1.1_1\endcsname{156}
\expandafter\gdef\csname odunum@ci@positive_taxonomy_m2.ao.mean.kimi_audio_7b_instruct.axis1.1_1\endcsname{[0.515, 0.615]}
\expandafter\gdef\csname odunum@val@positive_taxonomy_m2.ao.mean.kimi_audio_7b_instruct.axis1.1_2\endcsname{0.489}
\expandafter\gdef\csname odunum@n@positive_taxonomy_m2.ao.mean.kimi_audio_7b_instruct.axis1.1_2\endcsname{407}
\expandafter\gdef\csname odunum@ci@positive_taxonomy_m2.ao.mean.kimi_audio_7b_instruct.axis1.1_2\endcsname{[0.458, 0.518]}
\expandafter\gdef\csname odunum@val@positive_taxonomy_m2.ao.mean.kimi_audio_7b_instruct.axis2.2_1\endcsname{0.395}
\expandafter\gdef\csname odunum@n@positive_taxonomy_m2.ao.mean.kimi_audio_7b_instruct.axis2.2_1\endcsname{21}
\expandafter\gdef\csname odunum@ci@positive_taxonomy_m2.ao.mean.kimi_audio_7b_instruct.axis2.2_1\endcsname{[0.233, 0.569]}
\expandafter\gdef\csname odunum@val@positive_taxonomy_m2.ao.mean.kimi_audio_7b_instruct.axis2.2_2\endcsname{0.544}
\expandafter\gdef\csname odunum@n@positive_taxonomy_m2.ao.mean.kimi_audio_7b_instruct.axis2.2_2\endcsname{169}
\expandafter\gdef\csname odunum@ci@positive_taxonomy_m2.ao.mean.kimi_audio_7b_instruct.axis2.2_2\endcsname{[0.500, 0.586]}
\expandafter\gdef\csname odunum@val@positive_taxonomy_m2.ao.mean.kimi_audio_7b_instruct.axis2.2_3\endcsname{0.570}
\expandafter\gdef\csname odunum@n@positive_taxonomy_m2.ao.mean.kimi_audio_7b_instruct.axis2.2_3\endcsname{248}
\expandafter\gdef\csname odunum@ci@positive_taxonomy_m2.ao.mean.kimi_audio_7b_instruct.axis2.2_3\endcsname{[0.532, 0.606]}
\expandafter\gdef\csname odunum@val@positive_taxonomy_m2.ao.mean.kimi_audio_7b_instruct.axis2.2_4\endcsname{0.341}
\expandafter\gdef\csname odunum@n@positive_taxonomy_m2.ao.mean.kimi_audio_7b_instruct.axis2.2_4\endcsname{34}
\expandafter\gdef\csname odunum@ci@positive_taxonomy_m2.ao.mean.kimi_audio_7b_instruct.axis2.2_4\endcsname{[0.238, 0.448]}
\expandafter\gdef\csname odunum@val@positive_taxonomy_m2.ao.mean.kimi_audio_7b_instruct.axis2.2_5\endcsname{0.374}
\expandafter\gdef\csname odunum@n@positive_taxonomy_m2.ao.mean.kimi_audio_7b_instruct.axis2.2_5\endcsname{91}
\expandafter\gdef\csname odunum@ci@positive_taxonomy_m2.ao.mean.kimi_audio_7b_instruct.axis2.2_5\endcsname{[0.305, 0.445]}
\expandafter\gdef\csname odunum@val@positive_taxonomy_m2.ao.mean.kimi_audio_7b_instruct.axis3.3_1\endcsname{0.587}
\expandafter\gdef\csname odunum@n@positive_taxonomy_m2.ao.mean.kimi_audio_7b_instruct.axis3.3_1\endcsname{18}
\expandafter\gdef\csname odunum@ci@positive_taxonomy_m2.ao.mean.kimi_audio_7b_instruct.axis3.3_1\endcsname{[0.433, 0.737]}
\expandafter\gdef\csname odunum@val@positive_taxonomy_m2.ao.mean.kimi_audio_7b_instruct.axis3.3_2\endcsname{0.528}
\expandafter\gdef\csname odunum@n@positive_taxonomy_m2.ao.mean.kimi_audio_7b_instruct.axis3.3_2\endcsname{30}
\expandafter\gdef\csname odunum@ci@positive_taxonomy_m2.ao.mean.kimi_audio_7b_instruct.axis3.3_2\endcsname{[0.418, 0.636]}
\expandafter\gdef\csname odunum@val@positive_taxonomy_m2.ao.mean.kimi_audio_7b_instruct.axis3.3_3\endcsname{0.476}
\expandafter\gdef\csname odunum@n@positive_taxonomy_m2.ao.mean.kimi_audio_7b_instruct.axis3.3_3\endcsname{67}
\expandafter\gdef\csname odunum@ci@positive_taxonomy_m2.ao.mean.kimi_audio_7b_instruct.axis3.3_3\endcsname{[0.396, 0.557]}
\expandafter\gdef\csname odunum@val@positive_taxonomy_m2.ao.mean.kimi_audio_7b_instruct.axis3.3_4\endcsname{0.523}
\expandafter\gdef\csname odunum@n@positive_taxonomy_m2.ao.mean.kimi_audio_7b_instruct.axis3.3_4\endcsname{141}
\expandafter\gdef\csname odunum@ci@positive_taxonomy_m2.ao.mean.kimi_audio_7b_instruct.axis3.3_4\endcsname{[0.469, 0.577]}
\expandafter\gdef\csname odunum@val@positive_taxonomy_m2.ao.mean.kimi_audio_7b_instruct.axis3.3_5\endcsname{0.453}
\expandafter\gdef\csname odunum@n@positive_taxonomy_m2.ao.mean.kimi_audio_7b_instruct.axis3.3_5\endcsname{105}
\expandafter\gdef\csname odunum@ci@positive_taxonomy_m2.ao.mean.kimi_audio_7b_instruct.axis3.3_5\endcsname{[0.393, 0.512]}
\expandafter\gdef\csname odunum@val@positive_taxonomy_m2.ao.mean.kimi_audio_7b_instruct.axis3.3_6\endcsname{0.532}
\expandafter\gdef\csname odunum@n@positive_taxonomy_m2.ao.mean.kimi_audio_7b_instruct.axis3.3_6\endcsname{202}
\expandafter\gdef\csname odunum@ci@positive_taxonomy_m2.ao.mean.kimi_audio_7b_instruct.axis3.3_6\endcsname{[0.490, 0.573]}
\expandafter\gdef\csname odunum@val@positive_taxonomy_m2.ao.mean.kimi_audio_7b_instruct.axis4.4_1\endcsname{0.523}
\expandafter\gdef\csname odunum@n@positive_taxonomy_m2.ao.mean.kimi_audio_7b_instruct.axis4.4_1\endcsname{63}
\expandafter\gdef\csname odunum@ci@positive_taxonomy_m2.ao.mean.kimi_audio_7b_instruct.axis4.4_1\endcsname{[0.446, 0.598]}
\expandafter\gdef\csname odunum@val@positive_taxonomy_m2.ao.mean.kimi_audio_7b_instruct.axis4.4_2\endcsname{0.479}
\expandafter\gdef\csname odunum@n@positive_taxonomy_m2.ao.mean.kimi_audio_7b_instruct.axis4.4_2\endcsname{161}
\expandafter\gdef\csname odunum@ci@positive_taxonomy_m2.ao.mean.kimi_audio_7b_instruct.axis4.4_2\endcsname{[0.432, 0.524]}
\expandafter\gdef\csname odunum@val@positive_taxonomy_m2.ao.mean.kimi_audio_7b_instruct.axis4.4_3\endcsname{0.515}
\expandafter\gdef\csname odunum@n@positive_taxonomy_m2.ao.mean.kimi_audio_7b_instruct.axis4.4_3\endcsname{43}
\expandafter\gdef\csname odunum@ci@positive_taxonomy_m2.ao.mean.kimi_audio_7b_instruct.axis4.4_3\endcsname{[0.420, 0.609]}
\expandafter\gdef\csname odunum@val@positive_taxonomy_m2.ao.mean.kimi_audio_7b_instruct.axis4.4_4\endcsname{0.509}
\expandafter\gdef\csname odunum@n@positive_taxonomy_m2.ao.mean.kimi_audio_7b_instruct.axis4.4_4\endcsname{172}
\expandafter\gdef\csname odunum@ci@positive_taxonomy_m2.ao.mean.kimi_audio_7b_instruct.axis4.4_4\endcsname{[0.462, 0.557]}
\expandafter\gdef\csname odunum@val@positive_taxonomy_m2.ao.mean.kimi_audio_7b_instruct.axis4.4_5\endcsname{0.543}
\expandafter\gdef\csname odunum@n@positive_taxonomy_m2.ao.mean.kimi_audio_7b_instruct.axis4.4_5\endcsname{124}
\expandafter\gdef\csname odunum@ci@positive_taxonomy_m2.ao.mean.kimi_audio_7b_instruct.axis4.4_5\endcsname{[0.482, 0.601]}
\expandafter\gdef\csname odunum@val@positive_taxonomy_m2.ao.mean.ming.axis1.1_1\endcsname{0.542}
\expandafter\gdef\csname odunum@n@positive_taxonomy_m2.ao.mean.ming.axis1.1_1\endcsname{156}
\expandafter\gdef\csname odunum@ci@positive_taxonomy_m2.ao.mean.ming.axis1.1_1\endcsname{[0.491, 0.593]}
\expandafter\gdef\csname odunum@val@positive_taxonomy_m2.ao.mean.ming.axis1.1_2\endcsname{0.489}
\expandafter\gdef\csname odunum@n@positive_taxonomy_m2.ao.mean.ming.axis1.1_2\endcsname{407}
\expandafter\gdef\csname odunum@ci@positive_taxonomy_m2.ao.mean.ming.axis1.1_2\endcsname{[0.460, 0.517]}
\expandafter\gdef\csname odunum@val@positive_taxonomy_m2.ao.mean.ming.axis2.2_1\endcsname{0.457}
\expandafter\gdef\csname odunum@n@positive_taxonomy_m2.ao.mean.ming.axis2.2_1\endcsname{21}
\expandafter\gdef\csname odunum@ci@positive_taxonomy_m2.ao.mean.ming.axis2.2_1\endcsname{[0.299, 0.618]}
\expandafter\gdef\csname odunum@val@positive_taxonomy_m2.ao.mean.ming.axis2.2_2\endcsname{0.524}
\expandafter\gdef\csname odunum@n@positive_taxonomy_m2.ao.mean.ming.axis2.2_2\endcsname{169}
\expandafter\gdef\csname odunum@ci@positive_taxonomy_m2.ao.mean.ming.axis2.2_2\endcsname{[0.482, 0.565]}
\expandafter\gdef\csname odunum@val@positive_taxonomy_m2.ao.mean.ming.axis2.2_3\endcsname{0.545}
\expandafter\gdef\csname odunum@n@positive_taxonomy_m2.ao.mean.ming.axis2.2_3\endcsname{248}
\expandafter\gdef\csname odunum@ci@positive_taxonomy_m2.ao.mean.ming.axis2.2_3\endcsname{[0.508, 0.582]}
\expandafter\gdef\csname odunum@val@positive_taxonomy_m2.ao.mean.ming.axis2.2_4\endcsname{0.371}
\expandafter\gdef\csname odunum@n@positive_taxonomy_m2.ao.mean.ming.axis2.2_4\endcsname{34}
\expandafter\gdef\csname odunum@ci@positive_taxonomy_m2.ao.mean.ming.axis2.2_4\endcsname{[0.263, 0.480]}
\expandafter\gdef\csname odunum@val@positive_taxonomy_m2.ao.mean.ming.axis2.2_5\endcsname{0.416}
\expandafter\gdef\csname odunum@n@positive_taxonomy_m2.ao.mean.ming.axis2.2_5\endcsname{91}
\expandafter\gdef\csname odunum@ci@positive_taxonomy_m2.ao.mean.ming.axis2.2_5\endcsname{[0.350, 0.482]}
\expandafter\gdef\csname odunum@val@positive_taxonomy_m2.ao.mean.ming.axis3.3_1\endcsname{0.561}
\expandafter\gdef\csname odunum@n@positive_taxonomy_m2.ao.mean.ming.axis3.3_1\endcsname{18}
\expandafter\gdef\csname odunum@ci@positive_taxonomy_m2.ao.mean.ming.axis3.3_1\endcsname{[0.413, 0.705]}
\expandafter\gdef\csname odunum@val@positive_taxonomy_m2.ao.mean.ming.axis3.3_2\endcsname{0.463}
\expandafter\gdef\csname odunum@n@positive_taxonomy_m2.ao.mean.ming.axis3.3_2\endcsname{30}
\expandafter\gdef\csname odunum@ci@positive_taxonomy_m2.ao.mean.ming.axis3.3_2\endcsname{[0.369, 0.559]}
\expandafter\gdef\csname odunum@val@positive_taxonomy_m2.ao.mean.ming.axis3.3_3\endcsname{0.367}
\expandafter\gdef\csname odunum@n@positive_taxonomy_m2.ao.mean.ming.axis3.3_3\endcsname{67}
\expandafter\gdef\csname odunum@ci@positive_taxonomy_m2.ao.mean.ming.axis3.3_3\endcsname{[0.298, 0.439]}
\expandafter\gdef\csname odunum@val@positive_taxonomy_m2.ao.mean.ming.axis3.3_4\endcsname{0.522}
\expandafter\gdef\csname odunum@n@positive_taxonomy_m2.ao.mean.ming.axis3.3_4\endcsname{141}
\expandafter\gdef\csname odunum@ci@positive_taxonomy_m2.ao.mean.ming.axis3.3_4\endcsname{[0.472, 0.574]}
\expandafter\gdef\csname odunum@val@positive_taxonomy_m2.ao.mean.ming.axis3.3_5\endcsname{0.464}
\expandafter\gdef\csname odunum@n@positive_taxonomy_m2.ao.mean.ming.axis3.3_5\endcsname{105}
\expandafter\gdef\csname odunum@ci@positive_taxonomy_m2.ao.mean.ming.axis3.3_5\endcsname{[0.411, 0.518]}
\expandafter\gdef\csname odunum@val@positive_taxonomy_m2.ao.mean.ming.axis3.3_6\endcsname{0.558}
\expandafter\gdef\csname odunum@n@positive_taxonomy_m2.ao.mean.ming.axis3.3_6\endcsname{202}
\expandafter\gdef\csname odunum@ci@positive_taxonomy_m2.ao.mean.ming.axis3.3_6\endcsname{[0.515, 0.601]}
\expandafter\gdef\csname odunum@val@positive_taxonomy_m2.ao.mean.ming.axis4.4_1\endcsname{0.560}
\expandafter\gdef\csname odunum@n@positive_taxonomy_m2.ao.mean.ming.axis4.4_1\endcsname{63}
\expandafter\gdef\csname odunum@ci@positive_taxonomy_m2.ao.mean.ming.axis4.4_1\endcsname{[0.489, 0.628]}
\expandafter\gdef\csname odunum@val@positive_taxonomy_m2.ao.mean.ming.axis4.4_2\endcsname{0.471}
\expandafter\gdef\csname odunum@n@positive_taxonomy_m2.ao.mean.ming.axis4.4_2\endcsname{161}
\expandafter\gdef\csname odunum@ci@positive_taxonomy_m2.ao.mean.ming.axis4.4_2\endcsname{[0.426, 0.515]}
\expandafter\gdef\csname odunum@val@positive_taxonomy_m2.ao.mean.ming.axis4.4_3\endcsname{0.494}
\expandafter\gdef\csname odunum@n@positive_taxonomy_m2.ao.mean.ming.axis4.4_3\endcsname{43}
\expandafter\gdef\csname odunum@ci@positive_taxonomy_m2.ao.mean.ming.axis4.4_3\endcsname{[0.399, 0.589]}
\expandafter\gdef\csname odunum@val@positive_taxonomy_m2.ao.mean.ming.axis4.4_4\endcsname{0.490}
\expandafter\gdef\csname odunum@n@positive_taxonomy_m2.ao.mean.ming.axis4.4_4\endcsname{172}
\expandafter\gdef\csname odunum@ci@positive_taxonomy_m2.ao.mean.ming.axis4.4_4\endcsname{[0.444, 0.535]}
\expandafter\gdef\csname odunum@val@positive_taxonomy_m2.ao.mean.ming.axis4.4_5\endcsname{0.541}
\expandafter\gdef\csname odunum@n@positive_taxonomy_m2.ao.mean.ming.axis4.4_5\endcsname{124}
\expandafter\gdef\csname odunum@ci@positive_taxonomy_m2.ao.mean.ming.axis4.4_5\endcsname{[0.482, 0.599]}
\expandafter\gdef\csname odunum@val@positive_taxonomy_m2.ao.mean.minicpm_o.axis1.1_1\endcsname{0.509}
\expandafter\gdef\csname odunum@n@positive_taxonomy_m2.ao.mean.minicpm_o.axis1.1_1\endcsname{156}
\expandafter\gdef\csname odunum@ci@positive_taxonomy_m2.ao.mean.minicpm_o.axis1.1_1\endcsname{[0.448, 0.570]}
\expandafter\gdef\csname odunum@val@positive_taxonomy_m2.ao.mean.minicpm_o.axis1.1_2\endcsname{0.468}
\expandafter\gdef\csname odunum@n@positive_taxonomy_m2.ao.mean.minicpm_o.axis1.1_2\endcsname{407}
\expandafter\gdef\csname odunum@ci@positive_taxonomy_m2.ao.mean.minicpm_o.axis1.1_2\endcsname{[0.436, 0.499]}
\expandafter\gdef\csname odunum@val@positive_taxonomy_m2.ao.mean.minicpm_o.axis2.2_1\endcsname{0.640}
\expandafter\gdef\csname odunum@n@positive_taxonomy_m2.ao.mean.minicpm_o.axis2.2_1\endcsname{21}
\expandafter\gdef\csname odunum@ci@positive_taxonomy_m2.ao.mean.minicpm_o.axis2.2_1\endcsname{[0.483, 0.793]}
\expandafter\gdef\csname odunum@val@positive_taxonomy_m2.ao.mean.minicpm_o.axis2.2_2\endcsname{0.458}
\expandafter\gdef\csname odunum@n@positive_taxonomy_m2.ao.mean.minicpm_o.axis2.2_2\endcsname{169}
\expandafter\gdef\csname odunum@ci@positive_taxonomy_m2.ao.mean.minicpm_o.axis2.2_2\endcsname{[0.407, 0.509]}
\expandafter\gdef\csname odunum@val@positive_taxonomy_m2.ao.mean.minicpm_o.axis2.2_3\endcsname{0.488}
\expandafter\gdef\csname odunum@n@positive_taxonomy_m2.ao.mean.minicpm_o.axis2.2_3\endcsname{248}
\expandafter\gdef\csname odunum@ci@positive_taxonomy_m2.ao.mean.minicpm_o.axis2.2_3\endcsname{[0.444, 0.532]}
\expandafter\gdef\csname odunum@val@positive_taxonomy_m2.ao.mean.minicpm_o.axis2.2_4\endcsname{0.431}
\expandafter\gdef\csname odunum@n@positive_taxonomy_m2.ao.mean.minicpm_o.axis2.2_4\endcsname{34}
\expandafter\gdef\csname odunum@ci@positive_taxonomy_m2.ao.mean.minicpm_o.axis2.2_4\endcsname{[0.322, 0.541]}
\expandafter\gdef\csname odunum@val@positive_taxonomy_m2.ao.mean.minicpm_o.axis2.2_5\endcsname{0.479}
\expandafter\gdef\csname odunum@n@positive_taxonomy_m2.ao.mean.minicpm_o.axis2.2_5\endcsname{91}
\expandafter\gdef\csname odunum@ci@positive_taxonomy_m2.ao.mean.minicpm_o.axis2.2_5\endcsname{[0.410, 0.545]}
\expandafter\gdef\csname odunum@val@positive_taxonomy_m2.ao.mean.minicpm_o.axis3.3_1\endcsname{0.485}
\expandafter\gdef\csname odunum@n@positive_taxonomy_m2.ao.mean.minicpm_o.axis3.3_1\endcsname{18}
\expandafter\gdef\csname odunum@ci@positive_taxonomy_m2.ao.mean.minicpm_o.axis3.3_1\endcsname{[0.310, 0.658]}
\expandafter\gdef\csname odunum@val@positive_taxonomy_m2.ao.mean.minicpm_o.axis3.3_2\endcsname{0.532}
\expandafter\gdef\csname odunum@n@positive_taxonomy_m2.ao.mean.minicpm_o.axis3.3_2\endcsname{30}
\expandafter\gdef\csname odunum@ci@positive_taxonomy_m2.ao.mean.minicpm_o.axis3.3_2\endcsname{[0.430, 0.633]}
\expandafter\gdef\csname odunum@val@positive_taxonomy_m2.ao.mean.minicpm_o.axis3.3_3\endcsname{0.424}
\expandafter\gdef\csname odunum@n@positive_taxonomy_m2.ao.mean.minicpm_o.axis3.3_3\endcsname{67}
\expandafter\gdef\csname odunum@ci@positive_taxonomy_m2.ao.mean.minicpm_o.axis3.3_3\endcsname{[0.331, 0.517]}
\expandafter\gdef\csname odunum@val@positive_taxonomy_m2.ao.mean.minicpm_o.axis3.3_4\endcsname{0.501}
\expandafter\gdef\csname odunum@n@positive_taxonomy_m2.ao.mean.minicpm_o.axis3.3_4\endcsname{141}
\expandafter\gdef\csname odunum@ci@positive_taxonomy_m2.ao.mean.minicpm_o.axis3.3_4\endcsname{[0.440, 0.560]}
\expandafter\gdef\csname odunum@val@positive_taxonomy_m2.ao.mean.minicpm_o.axis3.3_5\endcsname{0.418}
\expandafter\gdef\csname odunum@n@positive_taxonomy_m2.ao.mean.minicpm_o.axis3.3_5\endcsname{105}
\expandafter\gdef\csname odunum@ci@positive_taxonomy_m2.ao.mean.minicpm_o.axis3.3_5\endcsname{[0.357, 0.479]}
\expandafter\gdef\csname odunum@val@positive_taxonomy_m2.ao.mean.minicpm_o.axis3.3_6\endcsname{0.507}
\expandafter\gdef\csname odunum@n@positive_taxonomy_m2.ao.mean.minicpm_o.axis3.3_6\endcsname{202}
\expandafter\gdef\csname odunum@ci@positive_taxonomy_m2.ao.mean.minicpm_o.axis3.3_6\endcsname{[0.461, 0.553]}
\expandafter\gdef\csname odunum@val@positive_taxonomy_m2.ao.mean.minicpm_o.axis4.4_1\endcsname{0.451}
\expandafter\gdef\csname odunum@n@positive_taxonomy_m2.ao.mean.minicpm_o.axis4.4_1\endcsname{63}
\expandafter\gdef\csname odunum@ci@positive_taxonomy_m2.ao.mean.minicpm_o.axis4.4_1\endcsname{[0.371, 0.529]}
\expandafter\gdef\csname odunum@val@positive_taxonomy_m2.ao.mean.minicpm_o.axis4.4_2\endcsname{0.485}
\expandafter\gdef\csname odunum@n@positive_taxonomy_m2.ao.mean.minicpm_o.axis4.4_2\endcsname{161}
\expandafter\gdef\csname odunum@ci@positive_taxonomy_m2.ao.mean.minicpm_o.axis4.4_2\endcsname{[0.432, 0.537]}
\expandafter\gdef\csname odunum@val@positive_taxonomy_m2.ao.mean.minicpm_o.axis4.4_3\endcsname{0.432}
\expandafter\gdef\csname odunum@n@positive_taxonomy_m2.ao.mean.minicpm_o.axis4.4_3\endcsname{43}
\expandafter\gdef\csname odunum@ci@positive_taxonomy_m2.ao.mean.minicpm_o.axis4.4_3\endcsname{[0.331, 0.530]}
\expandafter\gdef\csname odunum@val@positive_taxonomy_m2.ao.mean.minicpm_o.axis4.4_4\endcsname{0.438}
\expandafter\gdef\csname odunum@n@positive_taxonomy_m2.ao.mean.minicpm_o.axis4.4_4\endcsname{172}
\expandafter\gdef\csname odunum@ci@positive_taxonomy_m2.ao.mean.minicpm_o.axis4.4_4\endcsname{[0.386, 0.489]}
\expandafter\gdef\csname odunum@val@positive_taxonomy_m2.ao.mean.minicpm_o.axis4.4_5\endcsname{0.563}
\expandafter\gdef\csname odunum@n@positive_taxonomy_m2.ao.mean.minicpm_o.axis4.4_5\endcsname{124}
\expandafter\gdef\csname odunum@ci@positive_taxonomy_m2.ao.mean.minicpm_o.axis4.4_5\endcsname{[0.499, 0.625]}
\expandafter\gdef\csname odunum@val@positive_taxonomy_m2.ao.mean.nemotron.axis1.1_1\endcsname{0.344}
\expandafter\gdef\csname odunum@n@positive_taxonomy_m2.ao.mean.nemotron.axis1.1_1\endcsname{156}
\expandafter\gdef\csname odunum@ci@positive_taxonomy_m2.ao.mean.nemotron.axis1.1_1\endcsname{[0.286, 0.406]}
\expandafter\gdef\csname odunum@val@positive_taxonomy_m2.ao.mean.nemotron.axis1.1_2\endcsname{0.340}
\expandafter\gdef\csname odunum@n@positive_taxonomy_m2.ao.mean.nemotron.axis1.1_2\endcsname{407}
\expandafter\gdef\csname odunum@ci@positive_taxonomy_m2.ao.mean.nemotron.axis1.1_2\endcsname{[0.306, 0.375]}
\expandafter\gdef\csname odunum@val@positive_taxonomy_m2.ao.mean.nemotron.axis2.2_1\endcsname{0.431}
\expandafter\gdef\csname odunum@n@positive_taxonomy_m2.ao.mean.nemotron.axis2.2_1\endcsname{21}
\expandafter\gdef\csname odunum@ci@positive_taxonomy_m2.ao.mean.nemotron.axis2.2_1\endcsname{[0.263, 0.602]}
\expandafter\gdef\csname odunum@val@positive_taxonomy_m2.ao.mean.nemotron.axis2.2_2\endcsname{0.380}
\expandafter\gdef\csname odunum@n@positive_taxonomy_m2.ao.mean.nemotron.axis2.2_2\endcsname{169}
\expandafter\gdef\csname odunum@ci@positive_taxonomy_m2.ao.mean.nemotron.axis2.2_2\endcsname{[0.325, 0.435]}
\expandafter\gdef\csname odunum@val@positive_taxonomy_m2.ao.mean.nemotron.axis2.2_3\endcsname{0.331}
\expandafter\gdef\csname odunum@n@positive_taxonomy_m2.ao.mean.nemotron.axis2.2_3\endcsname{248}
\expandafter\gdef\csname odunum@ci@positive_taxonomy_m2.ao.mean.nemotron.axis2.2_3\endcsname{[0.287, 0.375]}
\expandafter\gdef\csname odunum@val@positive_taxonomy_m2.ao.mean.nemotron.axis2.2_4\endcsname{0.394}
\expandafter\gdef\csname odunum@n@positive_taxonomy_m2.ao.mean.nemotron.axis2.2_4\endcsname{34}
\expandafter\gdef\csname odunum@ci@positive_taxonomy_m2.ao.mean.nemotron.axis2.2_4\endcsname{[0.259, 0.531]}
\expandafter\gdef\csname odunum@val@positive_taxonomy_m2.ao.mean.nemotron.axis2.2_5\endcsname{0.258}
\expandafter\gdef\csname odunum@n@positive_taxonomy_m2.ao.mean.nemotron.axis2.2_5\endcsname{91}
\expandafter\gdef\csname odunum@ci@positive_taxonomy_m2.ao.mean.nemotron.axis2.2_5\endcsname{[0.190, 0.329]}
\expandafter\gdef\csname odunum@val@positive_taxonomy_m2.ao.mean.nemotron.axis3.3_1\endcsname{0.379}
\expandafter\gdef\csname odunum@n@positive_taxonomy_m2.ao.mean.nemotron.axis3.3_1\endcsname{18}
\expandafter\gdef\csname odunum@ci@positive_taxonomy_m2.ao.mean.nemotron.axis3.3_1\endcsname{[0.215, 0.546]}
\expandafter\gdef\csname odunum@val@positive_taxonomy_m2.ao.mean.nemotron.axis3.3_2\endcsname{0.364}
\expandafter\gdef\csname odunum@n@positive_taxonomy_m2.ao.mean.nemotron.axis3.3_2\endcsname{30}
\expandafter\gdef\csname odunum@ci@positive_taxonomy_m2.ao.mean.nemotron.axis3.3_2\endcsname{[0.253, 0.479]}
\expandafter\gdef\csname odunum@val@positive_taxonomy_m2.ao.mean.nemotron.axis3.3_3\endcsname{0.167}
\expandafter\gdef\csname odunum@n@positive_taxonomy_m2.ao.mean.nemotron.axis3.3_3\endcsname{67}
\expandafter\gdef\csname odunum@ci@positive_taxonomy_m2.ao.mean.nemotron.axis3.3_3\endcsname{[0.096, 0.247]}
\expandafter\gdef\csname odunum@val@positive_taxonomy_m2.ao.mean.nemotron.axis3.3_4\endcsname{0.365}
\expandafter\gdef\csname odunum@n@positive_taxonomy_m2.ao.mean.nemotron.axis3.3_4\endcsname{141}
\expandafter\gdef\csname odunum@ci@positive_taxonomy_m2.ao.mean.nemotron.axis3.3_4\endcsname{[0.304, 0.428]}
\expandafter\gdef\csname odunum@val@positive_taxonomy_m2.ao.mean.nemotron.axis3.3_5\endcsname{0.393}
\expandafter\gdef\csname odunum@n@positive_taxonomy_m2.ao.mean.nemotron.axis3.3_5\endcsname{105}
\expandafter\gdef\csname odunum@ci@positive_taxonomy_m2.ao.mean.nemotron.axis3.3_5\endcsname{[0.327, 0.462]}
\expandafter\gdef\csname odunum@val@positive_taxonomy_m2.ao.mean.nemotron.axis3.3_6\endcsname{0.348}
\expandafter\gdef\csname odunum@n@positive_taxonomy_m2.ao.mean.nemotron.axis3.3_6\endcsname{202}
\expandafter\gdef\csname odunum@ci@positive_taxonomy_m2.ao.mean.nemotron.axis3.3_6\endcsname{[0.298, 0.399]}
\expandafter\gdef\csname odunum@val@positive_taxonomy_m2.ao.mean.nemotron.axis4.4_1\endcsname{0.424}
\expandafter\gdef\csname odunum@n@positive_taxonomy_m2.ao.mean.nemotron.axis4.4_1\endcsname{63}
\expandafter\gdef\csname odunum@ci@positive_taxonomy_m2.ao.mean.nemotron.axis4.4_1\endcsname{[0.334, 0.511]}
\expandafter\gdef\csname odunum@val@positive_taxonomy_m2.ao.mean.nemotron.axis4.4_2\endcsname{0.336}
\expandafter\gdef\csname odunum@n@positive_taxonomy_m2.ao.mean.nemotron.axis4.4_2\endcsname{161}
\expandafter\gdef\csname odunum@ci@positive_taxonomy_m2.ao.mean.nemotron.axis4.4_2\endcsname{[0.281, 0.392]}
\expandafter\gdef\csname odunum@val@positive_taxonomy_m2.ao.mean.nemotron.axis4.4_3\endcsname{0.350}
\expandafter\gdef\csname odunum@n@positive_taxonomy_m2.ao.mean.nemotron.axis4.4_3\endcsname{43}
\expandafter\gdef\csname odunum@ci@positive_taxonomy_m2.ao.mean.nemotron.axis4.4_3\endcsname{[0.239, 0.467]}
\expandafter\gdef\csname odunum@val@positive_taxonomy_m2.ao.mean.nemotron.axis4.4_4\endcsname{0.326}
\expandafter\gdef\csname odunum@n@positive_taxonomy_m2.ao.mean.nemotron.axis4.4_4\endcsname{172}
\expandafter\gdef\csname odunum@ci@positive_taxonomy_m2.ao.mean.nemotron.axis4.4_4\endcsname{[0.273, 0.382]}
\expandafter\gdef\csname odunum@val@positive_taxonomy_m2.ao.mean.nemotron.axis4.4_5\endcsname{0.324}
\expandafter\gdef\csname odunum@n@positive_taxonomy_m2.ao.mean.nemotron.axis4.4_5\endcsname{124}
\expandafter\gdef\csname odunum@ci@positive_taxonomy_m2.ao.mean.nemotron.axis4.4_5\endcsname{[0.262, 0.390]}
\expandafter\gdef\csname odunum@val@positive_taxonomy_m2.ao.mean.qwen25_omni.axis1.1_1\endcsname{0.511}
\expandafter\gdef\csname odunum@n@positive_taxonomy_m2.ao.mean.qwen25_omni.axis1.1_1\endcsname{156}
\expandafter\gdef\csname odunum@ci@positive_taxonomy_m2.ao.mean.qwen25_omni.axis1.1_1\endcsname{[0.455, 0.566]}
\expandafter\gdef\csname odunum@val@positive_taxonomy_m2.ao.mean.qwen25_omni.axis1.1_2\endcsname{0.437}
\expandafter\gdef\csname odunum@n@positive_taxonomy_m2.ao.mean.qwen25_omni.axis1.1_2\endcsname{407}
\expandafter\gdef\csname odunum@ci@positive_taxonomy_m2.ao.mean.qwen25_omni.axis1.1_2\endcsname{[0.406, 0.469]}
\expandafter\gdef\csname odunum@val@positive_taxonomy_m2.ao.mean.qwen25_omni.axis2.2_1\endcsname{0.315}
\expandafter\gdef\csname odunum@n@positive_taxonomy_m2.ao.mean.qwen25_omni.axis2.2_1\endcsname{21}
\expandafter\gdef\csname odunum@ci@positive_taxonomy_m2.ao.mean.qwen25_omni.axis2.2_1\endcsname{[0.169, 0.478]}
\expandafter\gdef\csname odunum@val@positive_taxonomy_m2.ao.mean.qwen25_omni.axis2.2_2\endcsname{0.510}
\expandafter\gdef\csname odunum@n@positive_taxonomy_m2.ao.mean.qwen25_omni.axis2.2_2\endcsname{169}
\expandafter\gdef\csname odunum@ci@positive_taxonomy_m2.ao.mean.qwen25_omni.axis2.2_2\endcsname{[0.461, 0.558]}
\expandafter\gdef\csname odunum@val@positive_taxonomy_m2.ao.mean.qwen25_omni.axis2.2_3\endcsname{0.505}
\expandafter\gdef\csname odunum@n@positive_taxonomy_m2.ao.mean.qwen25_omni.axis2.2_3\endcsname{248}
\expandafter\gdef\csname odunum@ci@positive_taxonomy_m2.ao.mean.qwen25_omni.axis2.2_3\endcsname{[0.464, 0.545]}
\expandafter\gdef\csname odunum@val@positive_taxonomy_m2.ao.mean.qwen25_omni.axis2.2_4\endcsname{0.297}
\expandafter\gdef\csname odunum@n@positive_taxonomy_m2.ao.mean.qwen25_omni.axis2.2_4\endcsname{34}
\expandafter\gdef\csname odunum@ci@positive_taxonomy_m2.ao.mean.qwen25_omni.axis2.2_4\endcsname{[0.190, 0.405]}
\expandafter\gdef\csname odunum@val@positive_taxonomy_m2.ao.mean.qwen25_omni.axis2.2_5\endcsname{0.326}
\expandafter\gdef\csname odunum@n@positive_taxonomy_m2.ao.mean.qwen25_omni.axis2.2_5\endcsname{91}
\expandafter\gdef\csname odunum@ci@positive_taxonomy_m2.ao.mean.qwen25_omni.axis2.2_5\endcsname{[0.258, 0.392]}
\expandafter\gdef\csname odunum@val@positive_taxonomy_m2.ao.mean.qwen25_omni.axis3.3_1\endcsname{0.548}
\expandafter\gdef\csname odunum@n@positive_taxonomy_m2.ao.mean.qwen25_omni.axis3.3_1\endcsname{18}
\expandafter\gdef\csname odunum@ci@positive_taxonomy_m2.ao.mean.qwen25_omni.axis3.3_1\endcsname{[0.406, 0.687]}
\expandafter\gdef\csname odunum@val@positive_taxonomy_m2.ao.mean.qwen25_omni.axis3.3_2\endcsname{0.517}
\expandafter\gdef\csname odunum@n@positive_taxonomy_m2.ao.mean.qwen25_omni.axis3.3_2\endcsname{30}
\expandafter\gdef\csname odunum@ci@positive_taxonomy_m2.ao.mean.qwen25_omni.axis3.3_2\endcsname{[0.416, 0.620]}
\expandafter\gdef\csname odunum@val@positive_taxonomy_m2.ao.mean.qwen25_omni.axis3.3_3\endcsname{0.346}
\expandafter\gdef\csname odunum@n@positive_taxonomy_m2.ao.mean.qwen25_omni.axis3.3_3\endcsname{67}
\expandafter\gdef\csname odunum@ci@positive_taxonomy_m2.ao.mean.qwen25_omni.axis3.3_3\endcsname{[0.265, 0.429]}
\expandafter\gdef\csname odunum@val@positive_taxonomy_m2.ao.mean.qwen25_omni.axis3.3_4\endcsname{0.422}
\expandafter\gdef\csname odunum@n@positive_taxonomy_m2.ao.mean.qwen25_omni.axis3.3_4\endcsname{141}
\expandafter\gdef\csname odunum@ci@positive_taxonomy_m2.ao.mean.qwen25_omni.axis3.3_4\endcsname{[0.363, 0.480]}
\expandafter\gdef\csname odunum@val@positive_taxonomy_m2.ao.mean.qwen25_omni.axis3.3_5\endcsname{0.517}
\expandafter\gdef\csname odunum@n@positive_taxonomy_m2.ao.mean.qwen25_omni.axis3.3_5\endcsname{105}
\expandafter\gdef\csname odunum@ci@positive_taxonomy_m2.ao.mean.qwen25_omni.axis3.3_5\endcsname{[0.459, 0.575]}
\expandafter\gdef\csname odunum@val@positive_taxonomy_m2.ao.mean.qwen25_omni.axis3.3_6\endcsname{0.473}
\expandafter\gdef\csname odunum@n@positive_taxonomy_m2.ao.mean.qwen25_omni.axis3.3_6\endcsname{202}
\expandafter\gdef\csname odunum@ci@positive_taxonomy_m2.ao.mean.qwen25_omni.axis3.3_6\endcsname{[0.429, 0.517]}
\expandafter\gdef\csname odunum@val@positive_taxonomy_m2.ao.mean.qwen25_omni.axis4.4_1\endcsname{0.515}
\expandafter\gdef\csname odunum@n@positive_taxonomy_m2.ao.mean.qwen25_omni.axis4.4_1\endcsname{63}
\expandafter\gdef\csname odunum@ci@positive_taxonomy_m2.ao.mean.qwen25_omni.axis4.4_1\endcsname{[0.446, 0.583]}
\expandafter\gdef\csname odunum@val@positive_taxonomy_m2.ao.mean.qwen25_omni.axis4.4_2\endcsname{0.443}
\expandafter\gdef\csname odunum@n@positive_taxonomy_m2.ao.mean.qwen25_omni.axis4.4_2\endcsname{161}
\expandafter\gdef\csname odunum@ci@positive_taxonomy_m2.ao.mean.qwen25_omni.axis4.4_2\endcsname{[0.392, 0.494]}
\expandafter\gdef\csname odunum@val@positive_taxonomy_m2.ao.mean.qwen25_omni.axis4.4_3\endcsname{0.447}
\expandafter\gdef\csname odunum@n@positive_taxonomy_m2.ao.mean.qwen25_omni.axis4.4_3\endcsname{43}
\expandafter\gdef\csname odunum@ci@positive_taxonomy_m2.ao.mean.qwen25_omni.axis4.4_3\endcsname{[0.329, 0.566]}
\expandafter\gdef\csname odunum@val@positive_taxonomy_m2.ao.mean.qwen25_omni.axis4.4_4\endcsname{0.480}
\expandafter\gdef\csname odunum@n@positive_taxonomy_m2.ao.mean.qwen25_omni.axis4.4_4\endcsname{172}
\expandafter\gdef\csname odunum@ci@positive_taxonomy_m2.ao.mean.qwen25_omni.axis4.4_4\endcsname{[0.431, 0.531]}
\expandafter\gdef\csname odunum@val@positive_taxonomy_m2.ao.mean.qwen25_omni.axis4.4_5\endcsname{0.422}
\expandafter\gdef\csname odunum@n@positive_taxonomy_m2.ao.mean.qwen25_omni.axis4.4_5\endcsname{124}
\expandafter\gdef\csname odunum@ci@positive_taxonomy_m2.ao.mean.qwen25_omni.axis4.4_5\endcsname{[0.364, 0.481]}
\expandafter\gdef\csname odunum@val@positive_taxonomy_m2.ao.mean.qwen3_omni_instruct.axis1.1_1\endcsname{0.639}
\expandafter\gdef\csname odunum@n@positive_taxonomy_m2.ao.mean.qwen3_omni_instruct.axis1.1_1\endcsname{156}
\expandafter\gdef\csname odunum@ci@positive_taxonomy_m2.ao.mean.qwen3_omni_instruct.axis1.1_1\endcsname{[0.592, 0.686]}
\expandafter\gdef\csname odunum@val@positive_taxonomy_m2.ao.mean.qwen3_omni_instruct.axis1.1_2\endcsname{0.598}
\expandafter\gdef\csname odunum@n@positive_taxonomy_m2.ao.mean.qwen3_omni_instruct.axis1.1_2\endcsname{407}
\expandafter\gdef\csname odunum@ci@positive_taxonomy_m2.ao.mean.qwen3_omni_instruct.axis1.1_2\endcsname{[0.572, 0.623]}
\expandafter\gdef\csname odunum@val@positive_taxonomy_m2.ao.mean.qwen3_omni_instruct.axis2.2_1\endcsname{0.663}
\expandafter\gdef\csname odunum@n@positive_taxonomy_m2.ao.mean.qwen3_omni_instruct.axis2.2_1\endcsname{21}
\expandafter\gdef\csname odunum@ci@positive_taxonomy_m2.ao.mean.qwen3_omni_instruct.axis2.2_1\endcsname{[0.532, 0.788]}
\expandafter\gdef\csname odunum@val@positive_taxonomy_m2.ao.mean.qwen3_omni_instruct.axis2.2_2\endcsname{0.619}
\expandafter\gdef\csname odunum@n@positive_taxonomy_m2.ao.mean.qwen3_omni_instruct.axis2.2_2\endcsname{169}
\expandafter\gdef\csname odunum@ci@positive_taxonomy_m2.ao.mean.qwen3_omni_instruct.axis2.2_2\endcsname{[0.578, 0.660]}
\expandafter\gdef\csname odunum@val@positive_taxonomy_m2.ao.mean.qwen3_omni_instruct.axis2.2_3\endcsname{0.611}
\expandafter\gdef\csname odunum@n@positive_taxonomy_m2.ao.mean.qwen3_omni_instruct.axis2.2_3\endcsname{248}
\expandafter\gdef\csname odunum@ci@positive_taxonomy_m2.ao.mean.qwen3_omni_instruct.axis2.2_3\endcsname{[0.579, 0.643]}
\expandafter\gdef\csname odunum@val@positive_taxonomy_m2.ao.mean.qwen3_omni_instruct.axis2.2_4\endcsname{0.537}
\expandafter\gdef\csname odunum@n@positive_taxonomy_m2.ao.mean.qwen3_omni_instruct.axis2.2_4\endcsname{34}
\expandafter\gdef\csname odunum@ci@positive_taxonomy_m2.ao.mean.qwen3_omni_instruct.axis2.2_4\endcsname{[0.437, 0.636]}
\expandafter\gdef\csname odunum@val@positive_taxonomy_m2.ao.mean.qwen3_omni_instruct.axis2.2_5\endcsname{0.601}
\expandafter\gdef\csname odunum@n@positive_taxonomy_m2.ao.mean.qwen3_omni_instruct.axis2.2_5\endcsname{91}
\expandafter\gdef\csname odunum@ci@positive_taxonomy_m2.ao.mean.qwen3_omni_instruct.axis2.2_5\endcsname{[0.545, 0.656]}
\expandafter\gdef\csname odunum@val@positive_taxonomy_m2.ao.mean.qwen3_omni_instruct.axis3.3_1\endcsname{0.638}
\expandafter\gdef\csname odunum@n@positive_taxonomy_m2.ao.mean.qwen3_omni_instruct.axis3.3_1\endcsname{18}
\expandafter\gdef\csname odunum@ci@positive_taxonomy_m2.ao.mean.qwen3_omni_instruct.axis3.3_1\endcsname{[0.526, 0.759]}
\expandafter\gdef\csname odunum@val@positive_taxonomy_m2.ao.mean.qwen3_omni_instruct.axis3.3_2\endcsname{0.611}
\expandafter\gdef\csname odunum@n@positive_taxonomy_m2.ao.mean.qwen3_omni_instruct.axis3.3_2\endcsname{30}
\expandafter\gdef\csname odunum@ci@positive_taxonomy_m2.ao.mean.qwen3_omni_instruct.axis3.3_2\endcsname{[0.521, 0.703]}
\expandafter\gdef\csname odunum@val@positive_taxonomy_m2.ao.mean.qwen3_omni_instruct.axis3.3_3\endcsname{0.573}
\expandafter\gdef\csname odunum@n@positive_taxonomy_m2.ao.mean.qwen3_omni_instruct.axis3.3_3\endcsname{67}
\expandafter\gdef\csname odunum@ci@positive_taxonomy_m2.ao.mean.qwen3_omni_instruct.axis3.3_3\endcsname{[0.502, 0.644]}
\expandafter\gdef\csname odunum@val@positive_taxonomy_m2.ao.mean.qwen3_omni_instruct.axis3.3_4\endcsname{0.600}
\expandafter\gdef\csname odunum@n@positive_taxonomy_m2.ao.mean.qwen3_omni_instruct.axis3.3_4\endcsname{141}
\expandafter\gdef\csname odunum@ci@positive_taxonomy_m2.ao.mean.qwen3_omni_instruct.axis3.3_4\endcsname{[0.552, 0.646]}
\expandafter\gdef\csname odunum@val@positive_taxonomy_m2.ao.mean.qwen3_omni_instruct.axis3.3_5\endcsname{0.558}
\expandafter\gdef\csname odunum@n@positive_taxonomy_m2.ao.mean.qwen3_omni_instruct.axis3.3_5\endcsname{105}
\expandafter\gdef\csname odunum@ci@positive_taxonomy_m2.ao.mean.qwen3_omni_instruct.axis3.3_5\endcsname{[0.509, 0.609]}
\expandafter\gdef\csname odunum@val@positive_taxonomy_m2.ao.mean.qwen3_omni_instruct.axis3.3_6\endcsname{0.651}
\expandafter\gdef\csname odunum@n@positive_taxonomy_m2.ao.mean.qwen3_omni_instruct.axis3.3_6\endcsname{202}
\expandafter\gdef\csname odunum@ci@positive_taxonomy_m2.ao.mean.qwen3_omni_instruct.axis3.3_6\endcsname{[0.617, 0.686]}
\expandafter\gdef\csname odunum@val@positive_taxonomy_m2.ao.mean.qwen3_omni_instruct.axis4.4_1\endcsname{0.587}
\expandafter\gdef\csname odunum@n@positive_taxonomy_m2.ao.mean.qwen3_omni_instruct.axis4.4_1\endcsname{63}
\expandafter\gdef\csname odunum@ci@positive_taxonomy_m2.ao.mean.qwen3_omni_instruct.axis4.4_1\endcsname{[0.524, 0.648]}
\expandafter\gdef\csname odunum@val@positive_taxonomy_m2.ao.mean.qwen3_omni_instruct.axis4.4_2\endcsname{0.591}
\expandafter\gdef\csname odunum@n@positive_taxonomy_m2.ao.mean.qwen3_omni_instruct.axis4.4_2\endcsname{161}
\expandafter\gdef\csname odunum@ci@positive_taxonomy_m2.ao.mean.qwen3_omni_instruct.axis4.4_2\endcsname{[0.549, 0.632]}
\expandafter\gdef\csname odunum@val@positive_taxonomy_m2.ao.mean.qwen3_omni_instruct.axis4.4_3\endcsname{0.605}
\expandafter\gdef\csname odunum@n@positive_taxonomy_m2.ao.mean.qwen3_omni_instruct.axis4.4_3\endcsname{43}
\expandafter\gdef\csname odunum@ci@positive_taxonomy_m2.ao.mean.qwen3_omni_instruct.axis4.4_3\endcsname{[0.509, 0.700]}
\expandafter\gdef\csname odunum@val@positive_taxonomy_m2.ao.mean.qwen3_omni_instruct.axis4.4_4\endcsname{0.632}
\expandafter\gdef\csname odunum@n@positive_taxonomy_m2.ao.mean.qwen3_omni_instruct.axis4.4_4\endcsname{172}
\expandafter\gdef\csname odunum@ci@positive_taxonomy_m2.ao.mean.qwen3_omni_instruct.axis4.4_4\endcsname{[0.593, 0.673]}
\expandafter\gdef\csname odunum@val@positive_taxonomy_m2.ao.mean.qwen3_omni_instruct.axis4.4_5\endcsname{0.614}
\expandafter\gdef\csname odunum@n@positive_taxonomy_m2.ao.mean.qwen3_omni_instruct.axis4.4_5\endcsname{124}
\expandafter\gdef\csname odunum@ci@positive_taxonomy_m2.ao.mean.qwen3_omni_instruct.axis4.4_5\endcsname{[0.567, 0.661]}
\expandafter\gdef\csname odunum@val@positive_taxonomy_m2.ao.mean.qwen3_omni_think.axis1.1_1\endcsname{0.640}
\expandafter\gdef\csname odunum@n@positive_taxonomy_m2.ao.mean.qwen3_omni_think.axis1.1_1\endcsname{156}
\expandafter\gdef\csname odunum@ci@positive_taxonomy_m2.ao.mean.qwen3_omni_think.axis1.1_1\endcsname{[0.596, 0.685]}
\expandafter\gdef\csname odunum@val@positive_taxonomy_m2.ao.mean.qwen3_omni_think.axis1.1_2\endcsname{0.624}
\expandafter\gdef\csname odunum@n@positive_taxonomy_m2.ao.mean.qwen3_omni_think.axis1.1_2\endcsname{407}
\expandafter\gdef\csname odunum@ci@positive_taxonomy_m2.ao.mean.qwen3_omni_think.axis1.1_2\endcsname{[0.597, 0.650]}
\expandafter\gdef\csname odunum@val@positive_taxonomy_m2.ao.mean.qwen3_omni_think.axis2.2_1\endcsname{0.556}
\expandafter\gdef\csname odunum@n@positive_taxonomy_m2.ao.mean.qwen3_omni_think.axis2.2_1\endcsname{21}
\expandafter\gdef\csname odunum@ci@positive_taxonomy_m2.ao.mean.qwen3_omni_think.axis2.2_1\endcsname{[0.413, 0.695]}
\expandafter\gdef\csname odunum@val@positive_taxonomy_m2.ao.mean.qwen3_omni_think.axis2.2_2\endcsname{0.641}
\expandafter\gdef\csname odunum@n@positive_taxonomy_m2.ao.mean.qwen3_omni_think.axis2.2_2\endcsname{169}
\expandafter\gdef\csname odunum@ci@positive_taxonomy_m2.ao.mean.qwen3_omni_think.axis2.2_2\endcsname{[0.606, 0.678]}
\expandafter\gdef\csname odunum@val@positive_taxonomy_m2.ao.mean.qwen3_omni_think.axis2.2_3\endcsname{0.648}
\expandafter\gdef\csname odunum@n@positive_taxonomy_m2.ao.mean.qwen3_omni_think.axis2.2_3\endcsname{248}
\expandafter\gdef\csname odunum@ci@positive_taxonomy_m2.ao.mean.qwen3_omni_think.axis2.2_3\endcsname{[0.612, 0.682]}
\expandafter\gdef\csname odunum@val@positive_taxonomy_m2.ao.mean.qwen3_omni_think.axis2.2_4\endcsname{0.587}
\expandafter\gdef\csname odunum@n@positive_taxonomy_m2.ao.mean.qwen3_omni_think.axis2.2_4\endcsname{34}
\expandafter\gdef\csname odunum@ci@positive_taxonomy_m2.ao.mean.qwen3_omni_think.axis2.2_4\endcsname{[0.480, 0.690]}
\expandafter\gdef\csname odunum@val@positive_taxonomy_m2.ao.mean.qwen3_omni_think.axis2.2_5\endcsname{0.583}
\expandafter\gdef\csname odunum@n@positive_taxonomy_m2.ao.mean.qwen3_omni_think.axis2.2_5\endcsname{91}
\expandafter\gdef\csname odunum@ci@positive_taxonomy_m2.ao.mean.qwen3_omni_think.axis2.2_5\endcsname{[0.524, 0.644]}
\expandafter\gdef\csname odunum@val@positive_taxonomy_m2.ao.mean.qwen3_omni_think.axis3.3_1\endcsname{0.723}
\expandafter\gdef\csname odunum@n@positive_taxonomy_m2.ao.mean.qwen3_omni_think.axis3.3_1\endcsname{18}
\expandafter\gdef\csname odunum@ci@positive_taxonomy_m2.ao.mean.qwen3_omni_think.axis3.3_1\endcsname{[0.611, 0.838]}
\expandafter\gdef\csname odunum@val@positive_taxonomy_m2.ao.mean.qwen3_omni_think.axis3.3_2\endcsname{0.700}
\expandafter\gdef\csname odunum@n@positive_taxonomy_m2.ao.mean.qwen3_omni_think.axis3.3_2\endcsname{30}
\expandafter\gdef\csname odunum@ci@positive_taxonomy_m2.ao.mean.qwen3_omni_think.axis3.3_2\endcsname{[0.611, 0.787]}
\expandafter\gdef\csname odunum@val@positive_taxonomy_m2.ao.mean.qwen3_omni_think.axis3.3_3\endcsname{0.563}
\expandafter\gdef\csname odunum@n@positive_taxonomy_m2.ao.mean.qwen3_omni_think.axis3.3_3\endcsname{67}
\expandafter\gdef\csname odunum@ci@positive_taxonomy_m2.ao.mean.qwen3_omni_think.axis3.3_3\endcsname{[0.488, 0.637]}
\expandafter\gdef\csname odunum@val@positive_taxonomy_m2.ao.mean.qwen3_omni_think.axis3.3_4\endcsname{0.620}
\expandafter\gdef\csname odunum@n@positive_taxonomy_m2.ao.mean.qwen3_omni_think.axis3.3_4\endcsname{141}
\expandafter\gdef\csname odunum@ci@positive_taxonomy_m2.ao.mean.qwen3_omni_think.axis3.3_4\endcsname{[0.571, 0.668]}
\expandafter\gdef\csname odunum@val@positive_taxonomy_m2.ao.mean.qwen3_omni_think.axis3.3_5\endcsname{0.573}
\expandafter\gdef\csname odunum@n@positive_taxonomy_m2.ao.mean.qwen3_omni_think.axis3.3_5\endcsname{105}
\expandafter\gdef\csname odunum@ci@positive_taxonomy_m2.ao.mean.qwen3_omni_think.axis3.3_5\endcsname{[0.521, 0.623]}
\expandafter\gdef\csname odunum@val@positive_taxonomy_m2.ao.mean.qwen3_omni_think.axis3.3_6\endcsname{0.666}
\expandafter\gdef\csname odunum@n@positive_taxonomy_m2.ao.mean.qwen3_omni_think.axis3.3_6\endcsname{202}
\expandafter\gdef\csname odunum@ci@positive_taxonomy_m2.ao.mean.qwen3_omni_think.axis3.3_6\endcsname{[0.629, 0.702]}
\expandafter\gdef\csname odunum@val@positive_taxonomy_m2.ao.mean.qwen3_omni_think.axis4.4_1\endcsname{0.670}
\expandafter\gdef\csname odunum@n@positive_taxonomy_m2.ao.mean.qwen3_omni_think.axis4.4_1\endcsname{63}
\expandafter\gdef\csname odunum@ci@positive_taxonomy_m2.ao.mean.qwen3_omni_think.axis4.4_1\endcsname{[0.606, 0.732]}
\expandafter\gdef\csname odunum@val@positive_taxonomy_m2.ao.mean.qwen3_omni_think.axis4.4_2\endcsname{0.619}
\expandafter\gdef\csname odunum@n@positive_taxonomy_m2.ao.mean.qwen3_omni_think.axis4.4_2\endcsname{161}
\expandafter\gdef\csname odunum@ci@positive_taxonomy_m2.ao.mean.qwen3_omni_think.axis4.4_2\endcsname{[0.575, 0.662]}
\expandafter\gdef\csname odunum@val@positive_taxonomy_m2.ao.mean.qwen3_omni_think.axis4.4_3\endcsname{0.638}
\expandafter\gdef\csname odunum@n@positive_taxonomy_m2.ao.mean.qwen3_omni_think.axis4.4_3\endcsname{43}
\expandafter\gdef\csname odunum@ci@positive_taxonomy_m2.ao.mean.qwen3_omni_think.axis4.4_3\endcsname{[0.542, 0.733]}
\expandafter\gdef\csname odunum@val@positive_taxonomy_m2.ao.mean.qwen3_omni_think.axis4.4_4\endcsname{0.645}
\expandafter\gdef\csname odunum@n@positive_taxonomy_m2.ao.mean.qwen3_omni_think.axis4.4_4\endcsname{172}
\expandafter\gdef\csname odunum@ci@positive_taxonomy_m2.ao.mean.qwen3_omni_think.axis4.4_4\endcsname{[0.604, 0.686]}
\expandafter\gdef\csname odunum@val@positive_taxonomy_m2.ao.mean.qwen3_omni_think.axis4.4_5\endcsname{0.593}
\expandafter\gdef\csname odunum@n@positive_taxonomy_m2.ao.mean.qwen3_omni_think.axis4.4_5\endcsname{124}
\expandafter\gdef\csname odunum@ci@positive_taxonomy_m2.ao.mean.qwen3_omni_think.axis4.4_5\endcsname{[0.546, 0.642]}
\expandafter\gdef\csname odunum@val@positive_taxonomy_m2.ao.mean.qwen_plus.axis1.1_1\endcsname{0.748}
\expandafter\gdef\csname odunum@n@positive_taxonomy_m2.ao.mean.qwen_plus.axis1.1_1\endcsname{156}
\expandafter\gdef\csname odunum@ci@positive_taxonomy_m2.ao.mean.qwen_plus.axis1.1_1\endcsname{[0.708, 0.786]}
\expandafter\gdef\csname odunum@val@positive_taxonomy_m2.ao.mean.qwen_plus.axis1.1_2\endcsname{0.702}
\expandafter\gdef\csname odunum@n@positive_taxonomy_m2.ao.mean.qwen_plus.axis1.1_2\endcsname{407}
\expandafter\gdef\csname odunum@ci@positive_taxonomy_m2.ao.mean.qwen_plus.axis1.1_2\endcsname{[0.676, 0.727]}
\expandafter\gdef\csname odunum@val@positive_taxonomy_m2.ao.mean.qwen_plus.axis2.2_1\endcsname{0.642}
\expandafter\gdef\csname odunum@n@positive_taxonomy_m2.ao.mean.qwen_plus.axis2.2_1\endcsname{21}
\expandafter\gdef\csname odunum@ci@positive_taxonomy_m2.ao.mean.qwen_plus.axis2.2_1\endcsname{[0.507, 0.767]}
\expandafter\gdef\csname odunum@val@positive_taxonomy_m2.ao.mean.qwen_plus.axis2.2_2\endcsname{0.731}
\expandafter\gdef\csname odunum@n@positive_taxonomy_m2.ao.mean.qwen_plus.axis2.2_2\endcsname{169}
\expandafter\gdef\csname odunum@ci@positive_taxonomy_m2.ao.mean.qwen_plus.axis2.2_2\endcsname{[0.695, 0.767]}
\expandafter\gdef\csname odunum@val@positive_taxonomy_m2.ao.mean.qwen_plus.axis2.2_3\endcsname{0.731}
\expandafter\gdef\csname odunum@n@positive_taxonomy_m2.ao.mean.qwen_plus.axis2.2_3\endcsname{248}
\expandafter\gdef\csname odunum@ci@positive_taxonomy_m2.ao.mean.qwen_plus.axis2.2_3\endcsname{[0.701, 0.760]}
\expandafter\gdef\csname odunum@val@positive_taxonomy_m2.ao.mean.qwen_plus.axis2.2_4\endcsname{0.613}
\expandafter\gdef\csname odunum@n@positive_taxonomy_m2.ao.mean.qwen_plus.axis2.2_4\endcsname{34}
\expandafter\gdef\csname odunum@ci@positive_taxonomy_m2.ao.mean.qwen_plus.axis2.2_4\endcsname{[0.500, 0.722]}
\expandafter\gdef\csname odunum@val@positive_taxonomy_m2.ao.mean.qwen_plus.axis2.2_5\endcsname{0.694}
\expandafter\gdef\csname odunum@n@positive_taxonomy_m2.ao.mean.qwen_plus.axis2.2_5\endcsname{91}
\expandafter\gdef\csname odunum@ci@positive_taxonomy_m2.ao.mean.qwen_plus.axis2.2_5\endcsname{[0.633, 0.751]}
\expandafter\gdef\csname odunum@val@positive_taxonomy_m2.ao.mean.qwen_plus.axis3.3_1\endcsname{0.699}
\expandafter\gdef\csname odunum@n@positive_taxonomy_m2.ao.mean.qwen_plus.axis3.3_1\endcsname{18}
\expandafter\gdef\csname odunum@ci@positive_taxonomy_m2.ao.mean.qwen_plus.axis3.3_1\endcsname{[0.556, 0.833]}
\expandafter\gdef\csname odunum@val@positive_taxonomy_m2.ao.mean.qwen_plus.axis3.3_2\endcsname{0.727}
\expandafter\gdef\csname odunum@n@positive_taxonomy_m2.ao.mean.qwen_plus.axis3.3_2\endcsname{30}
\expandafter\gdef\csname odunum@ci@positive_taxonomy_m2.ao.mean.qwen_plus.axis3.3_2\endcsname{[0.635, 0.814]}
\expandafter\gdef\csname odunum@val@positive_taxonomy_m2.ao.mean.qwen_plus.axis3.3_3\endcsname{0.684}
\expandafter\gdef\csname odunum@n@positive_taxonomy_m2.ao.mean.qwen_plus.axis3.3_3\endcsname{67}
\expandafter\gdef\csname odunum@ci@positive_taxonomy_m2.ao.mean.qwen_plus.axis3.3_3\endcsname{[0.616, 0.750]}
\expandafter\gdef\csname odunum@val@positive_taxonomy_m2.ao.mean.qwen_plus.axis3.3_4\endcsname{0.687}
\expandafter\gdef\csname odunum@n@positive_taxonomy_m2.ao.mean.qwen_plus.axis3.3_4\endcsname{141}
\expandafter\gdef\csname odunum@ci@positive_taxonomy_m2.ao.mean.qwen_plus.axis3.3_4\endcsname{[0.641, 0.730]}
\expandafter\gdef\csname odunum@val@positive_taxonomy_m2.ao.mean.qwen_plus.axis3.3_5\endcsname{0.700}
\expandafter\gdef\csname odunum@n@positive_taxonomy_m2.ao.mean.qwen_plus.axis3.3_5\endcsname{105}
\expandafter\gdef\csname odunum@ci@positive_taxonomy_m2.ao.mean.qwen_plus.axis3.3_5\endcsname{[0.648, 0.750]}
\expandafter\gdef\csname odunum@val@positive_taxonomy_m2.ao.mean.qwen_plus.axis3.3_6\endcsname{0.752}
\expandafter\gdef\csname odunum@n@positive_taxonomy_m2.ao.mean.qwen_plus.axis3.3_6\endcsname{202}
\expandafter\gdef\csname odunum@ci@positive_taxonomy_m2.ao.mean.qwen_plus.axis3.3_6\endcsname{[0.719, 0.784]}
\expandafter\gdef\csname odunum@val@positive_taxonomy_m2.ao.mean.qwen_plus.axis4.4_1\endcsname{0.753}
\expandafter\gdef\csname odunum@n@positive_taxonomy_m2.ao.mean.qwen_plus.axis4.4_1\endcsname{63}
\expandafter\gdef\csname odunum@ci@positive_taxonomy_m2.ao.mean.qwen_plus.axis4.4_1\endcsname{[0.703, 0.804]}
\expandafter\gdef\csname odunum@val@positive_taxonomy_m2.ao.mean.qwen_plus.axis4.4_2\endcsname{0.694}
\expandafter\gdef\csname odunum@n@positive_taxonomy_m2.ao.mean.qwen_plus.axis4.4_2\endcsname{161}
\expandafter\gdef\csname odunum@ci@positive_taxonomy_m2.ao.mean.qwen_plus.axis4.4_2\endcsname{[0.654, 0.733]}
\expandafter\gdef\csname odunum@val@positive_taxonomy_m2.ao.mean.qwen_plus.axis4.4_3\endcsname{0.640}
\expandafter\gdef\csname odunum@n@positive_taxonomy_m2.ao.mean.qwen_plus.axis4.4_3\endcsname{43}
\expandafter\gdef\csname odunum@ci@positive_taxonomy_m2.ao.mean.qwen_plus.axis4.4_3\endcsname{[0.545, 0.730]}
\expandafter\gdef\csname odunum@val@positive_taxonomy_m2.ao.mean.qwen_plus.axis4.4_4\endcsname{0.738}
\expandafter\gdef\csname odunum@n@positive_taxonomy_m2.ao.mean.qwen_plus.axis4.4_4\endcsname{172}
\expandafter\gdef\csname odunum@ci@positive_taxonomy_m2.ao.mean.qwen_plus.axis4.4_4\endcsname{[0.697, 0.775]}
\expandafter\gdef\csname odunum@val@positive_taxonomy_m2.ao.mean.qwen_plus.axis4.4_5\endcsname{0.716}
\expandafter\gdef\csname odunum@n@positive_taxonomy_m2.ao.mean.qwen_plus.axis4.4_5\endcsname{124}
\expandafter\gdef\csname odunum@ci@positive_taxonomy_m2.ao.mean.qwen_plus.axis4.4_5\endcsname{[0.667, 0.764]}
\expandafter\gdef\csname odunum@val@positive_taxonomy_m2.ao.mean.salmonn2_7b.axis1.1_1\endcsname{0.212}
\expandafter\gdef\csname odunum@n@positive_taxonomy_m2.ao.mean.salmonn2_7b.axis1.1_1\endcsname{156}
\expandafter\gdef\csname odunum@ci@positive_taxonomy_m2.ao.mean.salmonn2_7b.axis1.1_1\endcsname{[0.164, 0.263]}
\expandafter\gdef\csname odunum@val@positive_taxonomy_m2.ao.mean.salmonn2_7b.axis1.1_2\endcsname{0.174}
\expandafter\gdef\csname odunum@n@positive_taxonomy_m2.ao.mean.salmonn2_7b.axis1.1_2\endcsname{407}
\expandafter\gdef\csname odunum@ci@positive_taxonomy_m2.ao.mean.salmonn2_7b.axis1.1_2\endcsname{[0.150, 0.200]}
\expandafter\gdef\csname odunum@val@positive_taxonomy_m2.ao.mean.salmonn2_7b.axis2.2_1\endcsname{0.233}
\expandafter\gdef\csname odunum@n@positive_taxonomy_m2.ao.mean.salmonn2_7b.axis2.2_1\endcsname{21}
\expandafter\gdef\csname odunum@ci@positive_taxonomy_m2.ao.mean.salmonn2_7b.axis2.2_1\endcsname{[0.118, 0.364]}
\expandafter\gdef\csname odunum@val@positive_taxonomy_m2.ao.mean.salmonn2_7b.axis2.2_2\endcsname{0.198}
\expandafter\gdef\csname odunum@n@positive_taxonomy_m2.ao.mean.salmonn2_7b.axis2.2_2\endcsname{169}
\expandafter\gdef\csname odunum@ci@positive_taxonomy_m2.ao.mean.salmonn2_7b.axis2.2_2\endcsname{[0.157, 0.240]}
\expandafter\gdef\csname odunum@val@positive_taxonomy_m2.ao.mean.salmonn2_7b.axis2.2_3\endcsname{0.128}
\expandafter\gdef\csname odunum@n@positive_taxonomy_m2.ao.mean.salmonn2_7b.axis2.2_3\endcsname{248}
\expandafter\gdef\csname odunum@ci@positive_taxonomy_m2.ao.mean.salmonn2_7b.axis2.2_3\endcsname{[0.099, 0.160]}
\expandafter\gdef\csname odunum@val@positive_taxonomy_m2.ao.mean.salmonn2_7b.axis2.2_4\endcsname{0.388}
\expandafter\gdef\csname odunum@n@positive_taxonomy_m2.ao.mean.salmonn2_7b.axis2.2_4\endcsname{34}
\expandafter\gdef\csname odunum@ci@positive_taxonomy_m2.ao.mean.salmonn2_7b.axis2.2_4\endcsname{[0.273, 0.509]}
\expandafter\gdef\csname odunum@val@positive_taxonomy_m2.ao.mean.salmonn2_7b.axis2.2_5\endcsname{0.227}
\expandafter\gdef\csname odunum@n@positive_taxonomy_m2.ao.mean.salmonn2_7b.axis2.2_5\endcsname{91}
\expandafter\gdef\csname odunum@ci@positive_taxonomy_m2.ao.mean.salmonn2_7b.axis2.2_5\endcsname{[0.175, 0.284]}
\expandafter\gdef\csname odunum@val@positive_taxonomy_m2.ao.mean.salmonn2_7b.axis3.3_1\endcsname{0.208}
\expandafter\gdef\csname odunum@n@positive_taxonomy_m2.ao.mean.salmonn2_7b.axis3.3_1\endcsname{18}
\expandafter\gdef\csname odunum@ci@positive_taxonomy_m2.ao.mean.salmonn2_7b.axis3.3_1\endcsname{[0.069, 0.368]}
\expandafter\gdef\csname odunum@val@positive_taxonomy_m2.ao.mean.salmonn2_7b.axis3.3_2\endcsname{0.191}
\expandafter\gdef\csname odunum@n@positive_taxonomy_m2.ao.mean.salmonn2_7b.axis3.3_2\endcsname{30}
\expandafter\gdef\csname odunum@ci@positive_taxonomy_m2.ao.mean.salmonn2_7b.axis3.3_2\endcsname{[0.108, 0.280]}
\expandafter\gdef\csname odunum@val@positive_taxonomy_m2.ao.mean.salmonn2_7b.axis3.3_3\endcsname{0.127}
\expandafter\gdef\csname odunum@n@positive_taxonomy_m2.ao.mean.salmonn2_7b.axis3.3_3\endcsname{67}
\expandafter\gdef\csname odunum@ci@positive_taxonomy_m2.ao.mean.salmonn2_7b.axis3.3_3\endcsname{[0.071, 0.192]}
\expandafter\gdef\csname odunum@val@positive_taxonomy_m2.ao.mean.salmonn2_7b.axis3.3_4\endcsname{0.181}
\expandafter\gdef\csname odunum@n@positive_taxonomy_m2.ao.mean.salmonn2_7b.axis3.3_4\endcsname{141}
\expandafter\gdef\csname odunum@ci@positive_taxonomy_m2.ao.mean.salmonn2_7b.axis3.3_4\endcsname{[0.137, 0.229]}
\expandafter\gdef\csname odunum@val@positive_taxonomy_m2.ao.mean.salmonn2_7b.axis3.3_5\endcsname{0.218}
\expandafter\gdef\csname odunum@n@positive_taxonomy_m2.ao.mean.salmonn2_7b.axis3.3_5\endcsname{105}
\expandafter\gdef\csname odunum@ci@positive_taxonomy_m2.ao.mean.salmonn2_7b.axis3.3_5\endcsname{[0.168, 0.272]}
\expandafter\gdef\csname odunum@val@positive_taxonomy_m2.ao.mean.salmonn2_7b.axis3.3_6\endcsname{0.185}
\expandafter\gdef\csname odunum@n@positive_taxonomy_m2.ao.mean.salmonn2_7b.axis3.3_6\endcsname{202}
\expandafter\gdef\csname odunum@ci@positive_taxonomy_m2.ao.mean.salmonn2_7b.axis3.3_6\endcsname{[0.149, 0.223]}
\expandafter\gdef\csname odunum@val@positive_taxonomy_m2.ao.mean.salmonn2_7b.axis4.4_1\endcsname{0.183}
\expandafter\gdef\csname odunum@n@positive_taxonomy_m2.ao.mean.salmonn2_7b.axis4.4_1\endcsname{63}
\expandafter\gdef\csname odunum@ci@positive_taxonomy_m2.ao.mean.salmonn2_7b.axis4.4_1\endcsname{[0.120, 0.252]}
\expandafter\gdef\csname odunum@val@positive_taxonomy_m2.ao.mean.salmonn2_7b.axis4.4_2\endcsname{0.188}
\expandafter\gdef\csname odunum@n@positive_taxonomy_m2.ao.mean.salmonn2_7b.axis4.4_2\endcsname{161}
\expandafter\gdef\csname odunum@ci@positive_taxonomy_m2.ao.mean.salmonn2_7b.axis4.4_2\endcsname{[0.147, 0.232]}
\expandafter\gdef\csname odunum@val@positive_taxonomy_m2.ao.mean.salmonn2_7b.axis4.4_3\endcsname{0.223}
\expandafter\gdef\csname odunum@n@positive_taxonomy_m2.ao.mean.salmonn2_7b.axis4.4_3\endcsname{43}
\expandafter\gdef\csname odunum@ci@positive_taxonomy_m2.ao.mean.salmonn2_7b.axis4.4_3\endcsname{[0.141, 0.308]}
\expandafter\gdef\csname odunum@val@positive_taxonomy_m2.ao.mean.salmonn2_7b.axis4.4_4\endcsname{0.166}
\expandafter\gdef\csname odunum@n@positive_taxonomy_m2.ao.mean.salmonn2_7b.axis4.4_4\endcsname{172}
\expandafter\gdef\csname odunum@ci@positive_taxonomy_m2.ao.mean.salmonn2_7b.axis4.4_4\endcsname{[0.128, 0.206]}
\expandafter\gdef\csname odunum@val@positive_taxonomy_m2.ao.mean.salmonn2_7b.axis4.4_5\endcsname{0.194}
\expandafter\gdef\csname odunum@n@positive_taxonomy_m2.ao.mean.salmonn2_7b.axis4.4_5\endcsname{124}
\expandafter\gdef\csname odunum@ci@positive_taxonomy_m2.ao.mean.salmonn2_7b.axis4.4_5\endcsname{[0.142, 0.249]}
\expandafter\gdef\csname odunum@val@positive_taxonomy_m2.ao.mean.seed.axis1.1_1\endcsname{0.799}
\expandafter\gdef\csname odunum@n@positive_taxonomy_m2.ao.mean.seed.axis1.1_1\endcsname{156}
\expandafter\gdef\csname odunum@ci@positive_taxonomy_m2.ao.mean.seed.axis1.1_1\endcsname{[0.758, 0.839]}
\expandafter\gdef\csname odunum@val@positive_taxonomy_m2.ao.mean.seed.axis1.1_2\endcsname{0.772}
\expandafter\gdef\csname odunum@n@positive_taxonomy_m2.ao.mean.seed.axis1.1_2\endcsname{407}
\expandafter\gdef\csname odunum@ci@positive_taxonomy_m2.ao.mean.seed.axis1.1_2\endcsname{[0.749, 0.795]}
\expandafter\gdef\csname odunum@val@positive_taxonomy_m2.ao.mean.seed.axis2.2_1\endcsname{0.814}
\expandafter\gdef\csname odunum@n@positive_taxonomy_m2.ao.mean.seed.axis2.2_1\endcsname{21}
\expandafter\gdef\csname odunum@ci@positive_taxonomy_m2.ao.mean.seed.axis2.2_1\endcsname{[0.735, 0.893]}
\expandafter\gdef\csname odunum@val@positive_taxonomy_m2.ao.mean.seed.axis2.2_2\endcsname{0.763}
\expandafter\gdef\csname odunum@n@positive_taxonomy_m2.ao.mean.seed.axis2.2_2\endcsname{169}
\expandafter\gdef\csname odunum@ci@positive_taxonomy_m2.ao.mean.seed.axis2.2_2\endcsname{[0.725, 0.799]}
\expandafter\gdef\csname odunum@val@positive_taxonomy_m2.ao.mean.seed.axis2.2_3\endcsname{0.805}
\expandafter\gdef\csname odunum@n@positive_taxonomy_m2.ao.mean.seed.axis2.2_3\endcsname{248}
\expandafter\gdef\csname odunum@ci@positive_taxonomy_m2.ao.mean.seed.axis2.2_3\endcsname{[0.776, 0.832]}
\expandafter\gdef\csname odunum@val@positive_taxonomy_m2.ao.mean.seed.axis2.2_4\endcsname{0.705}
\expandafter\gdef\csname odunum@n@positive_taxonomy_m2.ao.mean.seed.axis2.2_4\endcsname{34}
\expandafter\gdef\csname odunum@ci@positive_taxonomy_m2.ao.mean.seed.axis2.2_4\endcsname{[0.606, 0.796]}
\expandafter\gdef\csname odunum@val@positive_taxonomy_m2.ao.mean.seed.axis2.2_5\endcsname{0.763}
\expandafter\gdef\csname odunum@n@positive_taxonomy_m2.ao.mean.seed.axis2.2_5\endcsname{91}
\expandafter\gdef\csname odunum@ci@positive_taxonomy_m2.ao.mean.seed.axis2.2_5\endcsname{[0.710, 0.813]}
\expandafter\gdef\csname odunum@val@positive_taxonomy_m2.ao.mean.seed.axis3.3_1\endcsname{0.870}
\expandafter\gdef\csname odunum@n@positive_taxonomy_m2.ao.mean.seed.axis3.3_1\endcsname{18}
\expandafter\gdef\csname odunum@ci@positive_taxonomy_m2.ao.mean.seed.axis3.3_1\endcsname{[0.771, 0.958]}
\expandafter\gdef\csname odunum@val@positive_taxonomy_m2.ao.mean.seed.axis3.3_2\endcsname{0.711}
\expandafter\gdef\csname odunum@n@positive_taxonomy_m2.ao.mean.seed.axis3.3_2\endcsname{30}
\expandafter\gdef\csname odunum@ci@positive_taxonomy_m2.ao.mean.seed.axis3.3_2\endcsname{[0.635, 0.789]}
\expandafter\gdef\csname odunum@val@positive_taxonomy_m2.ao.mean.seed.axis3.3_3\endcsname{0.769}
\expandafter\gdef\csname odunum@n@positive_taxonomy_m2.ao.mean.seed.axis3.3_3\endcsname{67}
\expandafter\gdef\csname odunum@ci@positive_taxonomy_m2.ao.mean.seed.axis3.3_3\endcsname{[0.708, 0.829]}
\expandafter\gdef\csname odunum@val@positive_taxonomy_m2.ao.mean.seed.axis3.3_4\endcsname{0.790}
\expandafter\gdef\csname odunum@n@positive_taxonomy_m2.ao.mean.seed.axis3.3_4\endcsname{141}
\expandafter\gdef\csname odunum@ci@positive_taxonomy_m2.ao.mean.seed.axis3.3_4\endcsname{[0.751, 0.828]}
\expandafter\gdef\csname odunum@val@positive_taxonomy_m2.ao.mean.seed.axis3.3_5\endcsname{0.707}
\expandafter\gdef\csname odunum@n@positive_taxonomy_m2.ao.mean.seed.axis3.3_5\endcsname{105}
\expandafter\gdef\csname odunum@ci@positive_taxonomy_m2.ao.mean.seed.axis3.3_5\endcsname{[0.651, 0.759]}
\expandafter\gdef\csname odunum@val@positive_taxonomy_m2.ao.mean.seed.axis3.3_6\endcsname{0.816}
\expandafter\gdef\csname odunum@n@positive_taxonomy_m2.ao.mean.seed.axis3.3_6\endcsname{202}
\expandafter\gdef\csname odunum@ci@positive_taxonomy_m2.ao.mean.seed.axis3.3_6\endcsname{[0.787, 0.845]}
\expandafter\gdef\csname odunum@val@positive_taxonomy_m2.ao.mean.seed.axis4.4_1\endcsname{0.798}
\expandafter\gdef\csname odunum@n@positive_taxonomy_m2.ao.mean.seed.axis4.4_1\endcsname{63}
\expandafter\gdef\csname odunum@ci@positive_taxonomy_m2.ao.mean.seed.axis4.4_1\endcsname{[0.746, 0.847]}
\expandafter\gdef\csname odunum@val@positive_taxonomy_m2.ao.mean.seed.axis4.4_2\endcsname{0.755}
\expandafter\gdef\csname odunum@n@positive_taxonomy_m2.ao.mean.seed.axis4.4_2\endcsname{161}
\expandafter\gdef\csname odunum@ci@positive_taxonomy_m2.ao.mean.seed.axis4.4_2\endcsname{[0.717, 0.790]}
\expandafter\gdef\csname odunum@val@positive_taxonomy_m2.ao.mean.seed.axis4.4_3\endcsname{0.768}
\expandafter\gdef\csname odunum@n@positive_taxonomy_m2.ao.mean.seed.axis4.4_3\endcsname{43}
\expandafter\gdef\csname odunum@ci@positive_taxonomy_m2.ao.mean.seed.axis4.4_3\endcsname{[0.685, 0.845]}
\expandafter\gdef\csname odunum@val@positive_taxonomy_m2.ao.mean.seed.axis4.4_4\endcsname{0.779}
\expandafter\gdef\csname odunum@n@positive_taxonomy_m2.ao.mean.seed.axis4.4_4\endcsname{172}
\expandafter\gdef\csname odunum@ci@positive_taxonomy_m2.ao.mean.seed.axis4.4_4\endcsname{[0.741, 0.815]}
\expandafter\gdef\csname odunum@val@positive_taxonomy_m2.ao.mean.seed.axis4.4_5\endcsname{0.808}
\expandafter\gdef\csname odunum@n@positive_taxonomy_m2.ao.mean.seed.axis4.4_5\endcsname{124}
\expandafter\gdef\csname odunum@ci@positive_taxonomy_m2.ao.mean.seed.axis4.4_5\endcsname{[0.767, 0.849]}
\expandafter\gdef\csname odunum@val@positive_taxonomy_m2.ao.n_models\endcsname{15}
\expandafter\gdef\csname odunum@n@positive_taxonomy_m2.ao.n_models\endcsname{15}
\expandafter\gdef\csname odunum@val@positive_taxonomy_m2.ao.n_scenes\endcsname{563}
\expandafter\gdef\csname odunum@n@positive_taxonomy_m2.ao.n_scenes\endcsname{570}
\expandafter\gdef\csname odunum@val@positive_taxonomy_m2.ao.overall.cascade_asr\endcsname{0.683}
\expandafter\gdef\csname odunum@n@positive_taxonomy_m2.ao.overall.cascade_asr\endcsname{563}
\expandafter\gdef\csname odunum@ci@positive_taxonomy_m2.ao.overall.cascade_asr\endcsname{[0.659, 0.706]}
\expandafter\gdef\csname odunum@val@positive_taxonomy_m2.ao.overall.gemini\endcsname{0.700}
\expandafter\gdef\csname odunum@n@positive_taxonomy_m2.ao.overall.gemini\endcsname{563}
\expandafter\gdef\csname odunum@ci@positive_taxonomy_m2.ao.overall.gemini\endcsname{[0.679, 0.721]}
\expandafter\gdef\csname odunum@val@positive_taxonomy_m2.ao.overall.gemini35_flash_lite\endcsname{0.565}
\expandafter\gdef\csname odunum@n@positive_taxonomy_m2.ao.overall.gemini35_flash_lite\endcsname{563}
\expandafter\gdef\csname odunum@ci@positive_taxonomy_m2.ao.overall.gemini35_flash_lite\endcsname{[0.538, 0.593]}
\expandafter\gdef\csname odunum@val@positive_taxonomy_m2.ao.overall.gemini37_flash\endcsname{0.660}
\expandafter\gdef\csname odunum@n@positive_taxonomy_m2.ao.overall.gemini37_flash\endcsname{563}
\expandafter\gdef\csname odunum@ci@positive_taxonomy_m2.ao.overall.gemini37_flash\endcsname{[0.636, 0.683]}
\expandafter\gdef\csname odunum@val@positive_taxonomy_m2.ao.overall.gpt_realtime\endcsname{0.633}
\expandafter\gdef\csname odunum@n@positive_taxonomy_m2.ao.overall.gpt_realtime\endcsname{563}
\expandafter\gdef\csname odunum@ci@positive_taxonomy_m2.ao.overall.gpt_realtime\endcsname{[0.609, 0.657]}
\expandafter\gdef\csname odunum@val@positive_taxonomy_m2.ao.overall.kimi_audio_7b_instruct\endcsname{0.510}
\expandafter\gdef\csname odunum@n@positive_taxonomy_m2.ao.overall.kimi_audio_7b_instruct\endcsname{563}
\expandafter\gdef\csname odunum@ci@positive_taxonomy_m2.ao.overall.kimi_audio_7b_instruct\endcsname{[0.484, 0.535]}
\expandafter\gdef\csname odunum@val@positive_taxonomy_m2.ao.overall.ming\endcsname{0.504}
\expandafter\gdef\csname odunum@n@positive_taxonomy_m2.ao.overall.ming\endcsname{563}
\expandafter\gdef\csname odunum@ci@positive_taxonomy_m2.ao.overall.ming\endcsname{[0.478, 0.529]}
\expandafter\gdef\csname odunum@val@positive_taxonomy_m2.ao.overall.minicpm_o\endcsname{0.480}
\expandafter\gdef\csname odunum@n@positive_taxonomy_m2.ao.overall.minicpm_o\endcsname{563}
\expandafter\gdef\csname odunum@ci@positive_taxonomy_m2.ao.overall.minicpm_o\endcsname{[0.451, 0.509]}
\expandafter\gdef\csname odunum@val@positive_taxonomy_m2.ao.overall.nemotron\endcsname{0.341}
\expandafter\gdef\csname odunum@n@positive_taxonomy_m2.ao.overall.nemotron\endcsname{563}
\expandafter\gdef\csname odunum@ci@positive_taxonomy_m2.ao.overall.nemotron\endcsname{[0.312, 0.372]}
\expandafter\gdef\csname odunum@val@positive_taxonomy_m2.ao.overall.qwen25_omni\endcsname{0.458}
\expandafter\gdef\csname odunum@n@positive_taxonomy_m2.ao.overall.qwen25_omni\endcsname{563}
\expandafter\gdef\csname odunum@ci@positive_taxonomy_m2.ao.overall.qwen25_omni\endcsname{[0.430, 0.485]}
\expandafter\gdef\csname odunum@val@positive_taxonomy_m2.ao.overall.qwen3_omni_instruct\endcsname{0.609}
\expandafter\gdef\csname odunum@n@positive_taxonomy_m2.ao.overall.qwen3_omni_instruct\endcsname{563}
\expandafter\gdef\csname odunum@ci@positive_taxonomy_m2.ao.overall.qwen3_omni_instruct\endcsname{[0.587, 0.631]}
\expandafter\gdef\csname odunum@val@positive_taxonomy_m2.ao.overall.qwen3_omni_think\endcsname{0.628}
\expandafter\gdef\csname odunum@n@positive_taxonomy_m2.ao.overall.qwen3_omni_think\endcsname{563}
\expandafter\gdef\csname odunum@ci@positive_taxonomy_m2.ao.overall.qwen3_omni_think\endcsname{[0.605, 0.651]}
\expandafter\gdef\csname odunum@val@positive_taxonomy_m2.ao.overall.qwen_plus\endcsname{0.715}
\expandafter\gdef\csname odunum@n@positive_taxonomy_m2.ao.overall.qwen_plus\endcsname{563}
\expandafter\gdef\csname odunum@ci@positive_taxonomy_m2.ao.overall.qwen_plus\endcsname{[0.693, 0.736]}
\expandafter\gdef\csname odunum@val@positive_taxonomy_m2.ao.overall.salmonn2_7b\endcsname{0.185}
\expandafter\gdef\csname odunum@n@positive_taxonomy_m2.ao.overall.salmonn2_7b\endcsname{563}
\expandafter\gdef\csname odunum@ci@positive_taxonomy_m2.ao.overall.salmonn2_7b\endcsname{[0.162, 0.208]}
\expandafter\gdef\csname odunum@val@positive_taxonomy_m2.ao.overall.seed\endcsname{0.780}
\expandafter\gdef\csname odunum@n@positive_taxonomy_m2.ao.overall.seed\endcsname{563}
\expandafter\gdef\csname odunum@ci@positive_taxonomy_m2.ao.overall.seed\endcsname{[0.760, 0.799]}
\expandafter\gdef\csname odunum@val@positive_taxonomy_m2.ao.panel_delta.axis1.1_1\endcsname{0.040}
\expandafter\gdef\csname odunum@n@positive_taxonomy_m2.ao.panel_delta.axis1.1_1\endcsname{156}
\expandafter\gdef\csname odunum@ci@positive_taxonomy_m2.ao.panel_delta.axis1.1_1\endcsname{[0.016, 0.064]}
\expandafter\gdef\csname odunum@val@positive_taxonomy_m2.ao.panel_delta.axis1.1_2\endcsname{-0.015}
\expandafter\gdef\csname odunum@n@positive_taxonomy_m2.ao.panel_delta.axis1.1_2\endcsname{407}
\expandafter\gdef\csname odunum@ci@positive_taxonomy_m2.ao.panel_delta.axis1.1_2\endcsname{[-0.025, -0.006]}
\expandafter\gdef\csname odunum@val@positive_taxonomy_m2.ao.panel_delta.axis2.2_1\endcsname{-0.037}
\expandafter\gdef\csname odunum@n@positive_taxonomy_m2.ao.panel_delta.axis2.2_1\endcsname{21}
\expandafter\gdef\csname odunum@ci@positive_taxonomy_m2.ao.panel_delta.axis2.2_1\endcsname{[-0.119, 0.052]}
\expandafter\gdef\csname odunum@val@positive_taxonomy_m2.ao.panel_delta.axis2.2_2\endcsname{0.017}
\expandafter\gdef\csname odunum@n@positive_taxonomy_m2.ao.panel_delta.axis2.2_2\endcsname{169}
\expandafter\gdef\csname odunum@ci@positive_taxonomy_m2.ao.panel_delta.axis2.2_2\endcsname{[-0.006, 0.039]}
\expandafter\gdef\csname odunum@val@positive_taxonomy_m2.ao.panel_delta.axis2.2_3\endcsname{0.024}
\expandafter\gdef\csname odunum@n@positive_taxonomy_m2.ao.panel_delta.axis2.2_3\endcsname{248}
\expandafter\gdef\csname odunum@ci@positive_taxonomy_m2.ao.panel_delta.axis2.2_3\endcsname{[0.008, 0.041]}
\expandafter\gdef\csname odunum@val@positive_taxonomy_m2.ao.panel_delta.axis2.2_4\endcsname{-0.073}
\expandafter\gdef\csname odunum@n@positive_taxonomy_m2.ao.panel_delta.axis2.2_4\endcsname{34}
\expandafter\gdef\csname odunum@ci@positive_taxonomy_m2.ao.panel_delta.axis2.2_4\endcsname{[-0.134, -0.010]}
\expandafter\gdef\csname odunum@val@positive_taxonomy_m2.ao.panel_delta.axis2.2_5\endcsname{-0.061}
\expandafter\gdef\csname odunum@n@positive_taxonomy_m2.ao.panel_delta.axis2.2_5\endcsname{91}
\expandafter\gdef\csname odunum@ci@positive_taxonomy_m2.ao.panel_delta.axis2.2_5\endcsname{[-0.093, -0.028]}
\expandafter\gdef\csname odunum@val@positive_taxonomy_m2.ao.panel_delta.axis3.3_1\endcsname{0.062}
\expandafter\gdef\csname odunum@n@positive_taxonomy_m2.ao.panel_delta.axis3.3_1\endcsname{18}
\expandafter\gdef\csname odunum@ci@positive_taxonomy_m2.ao.panel_delta.axis3.3_1\endcsname{[-0.015, 0.141]}
\expandafter\gdef\csname odunum@val@positive_taxonomy_m2.ao.panel_delta.axis3.3_2\endcsname{0.004}
\expandafter\gdef\csname odunum@n@positive_taxonomy_m2.ao.panel_delta.axis3.3_2\endcsname{30}
\expandafter\gdef\csname odunum@ci@positive_taxonomy_m2.ao.panel_delta.axis3.3_2\endcsname{[-0.049, 0.058]}
\expandafter\gdef\csname odunum@val@positive_taxonomy_m2.ao.panel_delta.axis3.3_3\endcsname{-0.082}
\expandafter\gdef\csname odunum@n@positive_taxonomy_m2.ao.panel_delta.axis3.3_3\endcsname{67}
\expandafter\gdef\csname odunum@ci@positive_taxonomy_m2.ao.panel_delta.axis3.3_3\endcsname{[-0.120, -0.044]}
\expandafter\gdef\csname odunum@val@positive_taxonomy_m2.ao.panel_delta.axis3.3_4\endcsname{0.003}
\expandafter\gdef\csname odunum@n@positive_taxonomy_m2.ao.panel_delta.axis3.3_4\endcsname{141}
\expandafter\gdef\csname odunum@ci@positive_taxonomy_m2.ao.panel_delta.axis3.3_4\endcsname{[-0.022, 0.028]}
\expandafter\gdef\csname odunum@val@positive_taxonomy_m2.ao.panel_delta.axis3.3_5\endcsname{-0.039}
\expandafter\gdef\csname odunum@n@positive_taxonomy_m2.ao.panel_delta.axis3.3_5\endcsname{105}
\expandafter\gdef\csname odunum@ci@positive_taxonomy_m2.ao.panel_delta.axis3.3_5\endcsname{[-0.071, -0.007]}
\expandafter\gdef\csname odunum@val@positive_taxonomy_m2.ao.panel_delta.axis3.3_6\endcsname{0.040}
\expandafter\gdef\csname odunum@n@positive_taxonomy_m2.ao.panel_delta.axis3.3_6\endcsname{202}
\expandafter\gdef\csname odunum@ci@positive_taxonomy_m2.ao.panel_delta.axis3.3_6\endcsname{[0.022, 0.058]}
\expandafter\gdef\csname odunum@val@positive_taxonomy_m2.ao.panel_delta.axis4.4_1\endcsname{0.017}
\expandafter\gdef\csname odunum@n@positive_taxonomy_m2.ao.panel_delta.axis4.4_1\endcsname{63}
\expandafter\gdef\csname odunum@ci@positive_taxonomy_m2.ao.panel_delta.axis4.4_1\endcsname{[-0.021, 0.056]}
\expandafter\gdef\csname odunum@val@positive_taxonomy_m2.ao.panel_delta.axis4.4_2\endcsname{-0.008}
\expandafter\gdef\csname odunum@n@positive_taxonomy_m2.ao.panel_delta.axis4.4_2\endcsname{161}
\expandafter\gdef\csname odunum@ci@positive_taxonomy_m2.ao.panel_delta.axis4.4_2\endcsname{[-0.030, 0.014]}
\expandafter\gdef\csname odunum@val@positive_taxonomy_m2.ao.panel_delta.axis4.4_3\endcsname{-0.017}
\expandafter\gdef\csname odunum@n@positive_taxonomy_m2.ao.panel_delta.axis4.4_3\endcsname{43}
\expandafter\gdef\csname odunum@ci@positive_taxonomy_m2.ao.panel_delta.axis4.4_3\endcsname{[-0.072, 0.038]}
\expandafter\gdef\csname odunum@val@positive_taxonomy_m2.ao.panel_delta.axis4.4_4\endcsname{-0.002}
\expandafter\gdef\csname odunum@n@positive_taxonomy_m2.ao.panel_delta.axis4.4_4\endcsname{172}
\expandafter\gdef\csname odunum@ci@positive_taxonomy_m2.ao.panel_delta.axis4.4_4\endcsname{[-0.024, 0.021]}
\expandafter\gdef\csname odunum@val@positive_taxonomy_m2.ao.panel_delta.axis4.4_5\endcsname{0.010}
\expandafter\gdef\csname odunum@n@positive_taxonomy_m2.ao.panel_delta.axis4.4_5\endcsname{124}
\expandafter\gdef\csname odunum@ci@positive_taxonomy_m2.ao.panel_delta.axis4.4_5\endcsname{[-0.018, 0.039]}
\expandafter\gdef\csname odunum@val@positive_taxonomy_m2.audit.audio_axis1_corrections\endcsname{33}
\expandafter\gdef\csname odunum@n@positive_taxonomy_m2.audit.audio_axis1_corrections\endcsname{570}
\expandafter\gdef\csname odunum@val@positive_taxonomy_m2.av.count.axis1.1_1\endcsname{92}
\expandafter\gdef\csname odunum@n@positive_taxonomy_m2.av.count.axis1.1_1\endcsname{1\,043}
\expandafter\gdef\csname odunum@val@positive_taxonomy_m2.av.count.axis1.1_2\endcsname{288}
\expandafter\gdef\csname odunum@n@positive_taxonomy_m2.av.count.axis1.1_2\endcsname{1\,043}
\expandafter\gdef\csname odunum@val@positive_taxonomy_m2.av.count.axis1.1_3\endcsname{286}
\expandafter\gdef\csname odunum@n@positive_taxonomy_m2.av.count.axis1.1_3\endcsname{1\,043}
\expandafter\gdef\csname odunum@val@positive_taxonomy_m2.av.count.axis1.1_4\endcsname{377}
\expandafter\gdef\csname odunum@n@positive_taxonomy_m2.av.count.axis1.1_4\endcsname{1\,043}
\expandafter\gdef\csname odunum@val@positive_taxonomy_m2.av.count.axis2.2_1\endcsname{118}
\expandafter\gdef\csname odunum@n@positive_taxonomy_m2.av.count.axis2.2_1\endcsname{1\,043}
\expandafter\gdef\csname odunum@val@positive_taxonomy_m2.av.count.axis2.2_2\endcsname{266}
\expandafter\gdef\csname odunum@n@positive_taxonomy_m2.av.count.axis2.2_2\endcsname{1\,043}
\expandafter\gdef\csname odunum@val@positive_taxonomy_m2.av.count.axis2.2_3\endcsname{274}
\expandafter\gdef\csname odunum@n@positive_taxonomy_m2.av.count.axis2.2_3\endcsname{1\,043}
\expandafter\gdef\csname odunum@val@positive_taxonomy_m2.av.count.axis2.2_4\endcsname{226}
\expandafter\gdef\csname odunum@n@positive_taxonomy_m2.av.count.axis2.2_4\endcsname{1\,043}
\expandafter\gdef\csname odunum@val@positive_taxonomy_m2.av.count.axis2.2_5\endcsname{159}
\expandafter\gdef\csname odunum@n@positive_taxonomy_m2.av.count.axis2.2_5\endcsname{1\,043}
\expandafter\gdef\csname odunum@val@positive_taxonomy_m2.av.count.axis3.3_1\endcsname{91}
\expandafter\gdef\csname odunum@n@positive_taxonomy_m2.av.count.axis3.3_1\endcsname{1\,043}
\expandafter\gdef\csname odunum@val@positive_taxonomy_m2.av.count.axis3.3_2\endcsname{85}
\expandafter\gdef\csname odunum@n@positive_taxonomy_m2.av.count.axis3.3_2\endcsname{1\,043}
\expandafter\gdef\csname odunum@val@positive_taxonomy_m2.av.count.axis3.3_3\endcsname{137}
\expandafter\gdef\csname odunum@n@positive_taxonomy_m2.av.count.axis3.3_3\endcsname{1\,043}
\expandafter\gdef\csname odunum@val@positive_taxonomy_m2.av.count.axis3.3_4\endcsname{256}
\expandafter\gdef\csname odunum@n@positive_taxonomy_m2.av.count.axis3.3_4\endcsname{1\,043}
\expandafter\gdef\csname odunum@val@positive_taxonomy_m2.av.count.axis3.3_5\endcsname{185}
\expandafter\gdef\csname odunum@n@positive_taxonomy_m2.av.count.axis3.3_5\endcsname{1\,043}
\expandafter\gdef\csname odunum@val@positive_taxonomy_m2.av.count.axis3.3_6\endcsname{289}
\expandafter\gdef\csname odunum@n@positive_taxonomy_m2.av.count.axis3.3_6\endcsname{1\,043}
\expandafter\gdef\csname odunum@val@positive_taxonomy_m2.av.count.axis4.4_1\endcsname{137}
\expandafter\gdef\csname odunum@n@positive_taxonomy_m2.av.count.axis4.4_1\endcsname{1\,043}
\expandafter\gdef\csname odunum@val@positive_taxonomy_m2.av.count.axis4.4_2\endcsname{289}
\expandafter\gdef\csname odunum@n@positive_taxonomy_m2.av.count.axis4.4_2\endcsname{1\,043}
\expandafter\gdef\csname odunum@val@positive_taxonomy_m2.av.count.axis4.4_3\endcsname{120}
\expandafter\gdef\csname odunum@n@positive_taxonomy_m2.av.count.axis4.4_3\endcsname{1\,043}
\expandafter\gdef\csname odunum@val@positive_taxonomy_m2.av.count.axis4.4_4\endcsname{324}
\expandafter\gdef\csname odunum@n@positive_taxonomy_m2.av.count.axis4.4_4\endcsname{1\,043}
\expandafter\gdef\csname odunum@val@positive_taxonomy_m2.av.count.axis4.4_5\endcsname{173}
\expandafter\gdef\csname odunum@n@positive_taxonomy_m2.av.count.axis4.4_5\endcsname{1\,043}
\expandafter\gdef\csname odunum@val@positive_taxonomy_m2.av.delta.cascade_asr.axis1.1_1\endcsname{0.116}
\expandafter\gdef\csname odunum@n@positive_taxonomy_m2.av.delta.cascade_asr.axis1.1_1\endcsname{92}
\expandafter\gdef\csname odunum@ci@positive_taxonomy_m2.av.delta.cascade_asr.axis1.1_1\endcsname{[0.063, 0.169]}
\expandafter\gdef\csname odunum@val@positive_taxonomy_m2.av.delta.cascade_asr.axis1.1_2\endcsname{-0.044}
\expandafter\gdef\csname odunum@n@positive_taxonomy_m2.av.delta.cascade_asr.axis1.1_2\endcsname{288}
\expandafter\gdef\csname odunum@ci@positive_taxonomy_m2.av.delta.cascade_asr.axis1.1_2\endcsname{[-0.072, -0.017]}
\expandafter\gdef\csname odunum@val@positive_taxonomy_m2.av.delta.cascade_asr.axis1.1_3\endcsname{0.023}
\expandafter\gdef\csname odunum@n@positive_taxonomy_m2.av.delta.cascade_asr.axis1.1_3\endcsname{286}
\expandafter\gdef\csname odunum@ci@positive_taxonomy_m2.av.delta.cascade_asr.axis1.1_3\endcsname{[-0.005, 0.052]}
\expandafter\gdef\csname odunum@val@positive_taxonomy_m2.av.delta.cascade_asr.axis1.1_4\endcsname{-0.012}
\expandafter\gdef\csname odunum@n@positive_taxonomy_m2.av.delta.cascade_asr.axis1.1_4\endcsname{377}
\expandafter\gdef\csname odunum@ci@positive_taxonomy_m2.av.delta.cascade_asr.axis1.1_4\endcsname{[-0.035, 0.010]}
\expandafter\gdef\csname odunum@val@positive_taxonomy_m2.av.delta.cascade_asr.axis2.2_1\endcsname{-0.135}
\expandafter\gdef\csname odunum@n@positive_taxonomy_m2.av.delta.cascade_asr.axis2.2_1\endcsname{118}
\expandafter\gdef\csname odunum@ci@positive_taxonomy_m2.av.delta.cascade_asr.axis2.2_1\endcsname{[-0.185, -0.084]}
\expandafter\gdef\csname odunum@val@positive_taxonomy_m2.av.delta.cascade_asr.axis2.2_2\endcsname{0.072}
\expandafter\gdef\csname odunum@n@positive_taxonomy_m2.av.delta.cascade_asr.axis2.2_2\endcsname{266}
\expandafter\gdef\csname odunum@ci@positive_taxonomy_m2.av.delta.cascade_asr.axis2.2_2\endcsname{[0.045, 0.098]}
\expandafter\gdef\csname odunum@val@positive_taxonomy_m2.av.delta.cascade_asr.axis2.2_3\endcsname{0.092}
\expandafter\gdef\csname odunum@n@positive_taxonomy_m2.av.delta.cascade_asr.axis2.2_3\endcsname{274}
\expandafter\gdef\csname odunum@ci@positive_taxonomy_m2.av.delta.cascade_asr.axis2.2_3\endcsname{[0.066, 0.119]}
\expandafter\gdef\csname odunum@val@positive_taxonomy_m2.av.delta.cascade_asr.axis2.2_4\endcsname{-0.067}
\expandafter\gdef\csname odunum@n@positive_taxonomy_m2.av.delta.cascade_asr.axis2.2_4\endcsname{226}
\expandafter\gdef\csname odunum@ci@positive_taxonomy_m2.av.delta.cascade_asr.axis2.2_4\endcsname{[-0.101, -0.034]}
\expandafter\gdef\csname odunum@val@positive_taxonomy_m2.av.delta.cascade_asr.axis2.2_5\endcsname{-0.084}
\expandafter\gdef\csname odunum@n@positive_taxonomy_m2.av.delta.cascade_asr.axis2.2_5\endcsname{159}
\expandafter\gdef\csname odunum@ci@positive_taxonomy_m2.av.delta.cascade_asr.axis2.2_5\endcsname{[-0.127, -0.042]}
\expandafter\gdef\csname odunum@val@positive_taxonomy_m2.av.delta.cascade_asr.axis3.3_1\endcsname{0.004}
\expandafter\gdef\csname odunum@n@positive_taxonomy_m2.av.delta.cascade_asr.axis3.3_1\endcsname{91}
\expandafter\gdef\csname odunum@ci@positive_taxonomy_m2.av.delta.cascade_asr.axis3.3_1\endcsname{[-0.050, 0.057]}
\expandafter\gdef\csname odunum@val@positive_taxonomy_m2.av.delta.cascade_asr.axis3.3_2\endcsname{-0.001}
\expandafter\gdef\csname odunum@n@positive_taxonomy_m2.av.delta.cascade_asr.axis3.3_2\endcsname{85}
\expandafter\gdef\csname odunum@ci@positive_taxonomy_m2.av.delta.cascade_asr.axis3.3_2\endcsname{[-0.058, 0.057]}
\expandafter\gdef\csname odunum@val@positive_taxonomy_m2.av.delta.cascade_asr.axis3.3_3\endcsname{-0.124}
\expandafter\gdef\csname odunum@n@positive_taxonomy_m2.av.delta.cascade_asr.axis3.3_3\endcsname{137}
\expandafter\gdef\csname odunum@ci@positive_taxonomy_m2.av.delta.cascade_asr.axis3.3_3\endcsname{[-0.173, -0.075]}
\expandafter\gdef\csname odunum@val@positive_taxonomy_m2.av.delta.cascade_asr.axis3.3_4\endcsname{0.042}
\expandafter\gdef\csname odunum@n@positive_taxonomy_m2.av.delta.cascade_asr.axis3.3_4\endcsname{256}
\expandafter\gdef\csname odunum@ci@positive_taxonomy_m2.av.delta.cascade_asr.axis3.3_4\endcsname{[0.012, 0.071]}
\expandafter\gdef\csname odunum@val@positive_taxonomy_m2.av.delta.cascade_asr.axis3.3_5\endcsname{-0.087}
\expandafter\gdef\csname odunum@n@positive_taxonomy_m2.av.delta.cascade_asr.axis3.3_5\endcsname{185}
\expandafter\gdef\csname odunum@ci@positive_taxonomy_m2.av.delta.cascade_asr.axis3.3_5\endcsname{[-0.126, -0.049]}
\expandafter\gdef\csname odunum@val@positive_taxonomy_m2.av.delta.cascade_asr.axis3.3_6\endcsname{0.076}
\expandafter\gdef\csname odunum@n@positive_taxonomy_m2.av.delta.cascade_asr.axis3.3_6\endcsname{289}
\expandafter\gdef\csname odunum@ci@positive_taxonomy_m2.av.delta.cascade_asr.axis3.3_6\endcsname{[0.051, 0.102]}
\expandafter\gdef\csname odunum@val@positive_taxonomy_m2.av.delta.cascade_asr.axis4.4_1\endcsname{0.026}
\expandafter\gdef\csname odunum@n@positive_taxonomy_m2.av.delta.cascade_asr.axis4.4_1\endcsname{137}
\expandafter\gdef\csname odunum@ci@positive_taxonomy_m2.av.delta.cascade_asr.axis4.4_1\endcsname{[-0.017, 0.069]}
\expandafter\gdef\csname odunum@val@positive_taxonomy_m2.av.delta.cascade_asr.axis4.4_2\endcsname{0.008}
\expandafter\gdef\csname odunum@n@positive_taxonomy_m2.av.delta.cascade_asr.axis4.4_2\endcsname{289}
\expandafter\gdef\csname odunum@ci@positive_taxonomy_m2.av.delta.cascade_asr.axis4.4_2\endcsname{[-0.020, 0.035]}
\expandafter\gdef\csname odunum@val@positive_taxonomy_m2.av.delta.cascade_asr.axis4.4_3\endcsname{-0.055}
\expandafter\gdef\csname odunum@n@positive_taxonomy_m2.av.delta.cascade_asr.axis4.4_3\endcsname{120}
\expandafter\gdef\csname odunum@ci@positive_taxonomy_m2.av.delta.cascade_asr.axis4.4_3\endcsname{[-0.110, -0.001]}
\expandafter\gdef\csname odunum@val@positive_taxonomy_m2.av.delta.cascade_asr.axis4.4_4\endcsname{-0.016}
\expandafter\gdef\csname odunum@n@positive_taxonomy_m2.av.delta.cascade_asr.axis4.4_4\endcsname{324}
\expandafter\gdef\csname odunum@ci@positive_taxonomy_m2.av.delta.cascade_asr.axis4.4_4\endcsname{[-0.041, 0.010]}
\expandafter\gdef\csname odunum@val@positive_taxonomy_m2.av.delta.cascade_asr.axis4.4_5\endcsname{0.033}
\expandafter\gdef\csname odunum@n@positive_taxonomy_m2.av.delta.cascade_asr.axis4.4_5\endcsname{173}
\expandafter\gdef\csname odunum@ci@positive_taxonomy_m2.av.delta.cascade_asr.axis4.4_5\endcsname{[-0.004, 0.071]}
\expandafter\gdef\csname odunum@val@positive_taxonomy_m2.av.delta.gemini.axis1.1_1\endcsname{-0.002}
\expandafter\gdef\csname odunum@n@positive_taxonomy_m2.av.delta.gemini.axis1.1_1\endcsname{92}
\expandafter\gdef\csname odunum@ci@positive_taxonomy_m2.av.delta.gemini.axis1.1_1\endcsname{[-0.054, 0.049]}
\expandafter\gdef\csname odunum@val@positive_taxonomy_m2.av.delta.gemini.axis1.1_2\endcsname{-0.021}
\expandafter\gdef\csname odunum@n@positive_taxonomy_m2.av.delta.gemini.axis1.1_2\endcsname{288}
\expandafter\gdef\csname odunum@ci@positive_taxonomy_m2.av.delta.gemini.axis1.1_2\endcsname{[-0.047, 0.004]}
\expandafter\gdef\csname odunum@val@positive_taxonomy_m2.av.delta.gemini.axis1.1_3\endcsname{0.020}
\expandafter\gdef\csname odunum@n@positive_taxonomy_m2.av.delta.gemini.axis1.1_3\endcsname{286}
\expandafter\gdef\csname odunum@ci@positive_taxonomy_m2.av.delta.gemini.axis1.1_3\endcsname{[-0.004, 0.044]}
\expandafter\gdef\csname odunum@val@positive_taxonomy_m2.av.delta.gemini.axis1.1_4\endcsname{0.001}
\expandafter\gdef\csname odunum@n@positive_taxonomy_m2.av.delta.gemini.axis1.1_4\endcsname{377}
\expandafter\gdef\csname odunum@ci@positive_taxonomy_m2.av.delta.gemini.axis1.1_4\endcsname{[-0.018, 0.021]}
\expandafter\gdef\csname odunum@val@positive_taxonomy_m2.av.delta.gemini.axis2.2_1\endcsname{-0.014}
\expandafter\gdef\csname odunum@n@positive_taxonomy_m2.av.delta.gemini.axis2.2_1\endcsname{118}
\expandafter\gdef\csname odunum@ci@positive_taxonomy_m2.av.delta.gemini.axis2.2_1\endcsname{[-0.063, 0.036]}
\expandafter\gdef\csname odunum@val@positive_taxonomy_m2.av.delta.gemini.axis2.2_2\endcsname{-0.008}
\expandafter\gdef\csname odunum@n@positive_taxonomy_m2.av.delta.gemini.axis2.2_2\endcsname{266}
\expandafter\gdef\csname odunum@ci@positive_taxonomy_m2.av.delta.gemini.axis2.2_2\endcsname{[-0.032, 0.016]}
\expandafter\gdef\csname odunum@val@positive_taxonomy_m2.av.delta.gemini.axis2.2_3\endcsname{0.004}
\expandafter\gdef\csname odunum@n@positive_taxonomy_m2.av.delta.gemini.axis2.2_3\endcsname{274}
\expandafter\gdef\csname odunum@ci@positive_taxonomy_m2.av.delta.gemini.axis2.2_3\endcsname{[-0.020, 0.029]}
\expandafter\gdef\csname odunum@val@positive_taxonomy_m2.av.delta.gemini.axis2.2_4\endcsname{0.011}
\expandafter\gdef\csname odunum@n@positive_taxonomy_m2.av.delta.gemini.axis2.2_4\endcsname{226}
\expandafter\gdef\csname odunum@ci@positive_taxonomy_m2.av.delta.gemini.axis2.2_4\endcsname{[-0.018, 0.040]}
\expandafter\gdef\csname odunum@val@positive_taxonomy_m2.av.delta.gemini.axis2.2_5\endcsname{-3.6\ensuremath{\times 10^{-5}}}
\expandafter\gdef\csname odunum@n@positive_taxonomy_m2.av.delta.gemini.axis2.2_5\endcsname{159}
\expandafter\gdef\csname odunum@ci@positive_taxonomy_m2.av.delta.gemini.axis2.2_5\endcsname{[-0.034, 0.034]}
\expandafter\gdef\csname odunum@val@positive_taxonomy_m2.av.delta.gemini.axis3.3_1\endcsname{0.042}
\expandafter\gdef\csname odunum@n@positive_taxonomy_m2.av.delta.gemini.axis3.3_1\endcsname{91}
\expandafter\gdef\csname odunum@ci@positive_taxonomy_m2.av.delta.gemini.axis3.3_1\endcsname{[-0.003, 0.087]}
\expandafter\gdef\csname odunum@val@positive_taxonomy_m2.av.delta.gemini.axis3.3_2\endcsname{-0.025}
\expandafter\gdef\csname odunum@n@positive_taxonomy_m2.av.delta.gemini.axis3.3_2\endcsname{85}
\expandafter\gdef\csname odunum@ci@positive_taxonomy_m2.av.delta.gemini.axis3.3_2\endcsname{[-0.069, 0.020]}
\expandafter\gdef\csname odunum@val@positive_taxonomy_m2.av.delta.gemini.axis3.3_3\endcsname{-0.073}
\expandafter\gdef\csname odunum@n@positive_taxonomy_m2.av.delta.gemini.axis3.3_3\endcsname{137}
\expandafter\gdef\csname odunum@ci@positive_taxonomy_m2.av.delta.gemini.axis3.3_3\endcsname{[-0.117, -0.028]}
\expandafter\gdef\csname odunum@val@positive_taxonomy_m2.av.delta.gemini.axis3.3_4\endcsname{0.018}
\expandafter\gdef\csname odunum@n@positive_taxonomy_m2.av.delta.gemini.axis3.3_4\endcsname{256}
\expandafter\gdef\csname odunum@ci@positive_taxonomy_m2.av.delta.gemini.axis3.3_4\endcsname{[-0.007, 0.042]}
\expandafter\gdef\csname odunum@val@positive_taxonomy_m2.av.delta.gemini.axis3.3_5\endcsname{-0.076}
\expandafter\gdef\csname odunum@n@positive_taxonomy_m2.av.delta.gemini.axis3.3_5\endcsname{185}
\expandafter\gdef\csname odunum@ci@positive_taxonomy_m2.av.delta.gemini.axis3.3_5\endcsname{[-0.109, -0.042]}
\expandafter\gdef\csname odunum@val@positive_taxonomy_m2.av.delta.gemini.axis3.3_6\endcsname{0.062}
\expandafter\gdef\csname odunum@n@positive_taxonomy_m2.av.delta.gemini.axis3.3_6\endcsname{289}
\expandafter\gdef\csname odunum@ci@positive_taxonomy_m2.av.delta.gemini.axis3.3_6\endcsname{[0.039, 0.084]}
\expandafter\gdef\csname odunum@val@positive_taxonomy_m2.av.delta.gemini.axis4.4_1\endcsname{-0.014}
\expandafter\gdef\csname odunum@n@positive_taxonomy_m2.av.delta.gemini.axis4.4_1\endcsname{137}
\expandafter\gdef\csname odunum@ci@positive_taxonomy_m2.av.delta.gemini.axis4.4_1\endcsname{[-0.052, 0.024]}
\expandafter\gdef\csname odunum@val@positive_taxonomy_m2.av.delta.gemini.axis4.4_2\endcsname{-0.004}
\expandafter\gdef\csname odunum@n@positive_taxonomy_m2.av.delta.gemini.axis4.4_2\endcsname{289}
\expandafter\gdef\csname odunum@ci@positive_taxonomy_m2.av.delta.gemini.axis4.4_2\endcsname{[-0.028, 0.020]}
\expandafter\gdef\csname odunum@val@positive_taxonomy_m2.av.delta.gemini.axis4.4_3\endcsname{0.005}
\expandafter\gdef\csname odunum@n@positive_taxonomy_m2.av.delta.gemini.axis4.4_3\endcsname{120}
\expandafter\gdef\csname odunum@ci@positive_taxonomy_m2.av.delta.gemini.axis4.4_3\endcsname{[-0.039, 0.051]}
\expandafter\gdef\csname odunum@val@positive_taxonomy_m2.av.delta.gemini.axis4.4_4\endcsname{0.005}
\expandafter\gdef\csname odunum@n@positive_taxonomy_m2.av.delta.gemini.axis4.4_4\endcsname{324}
\expandafter\gdef\csname odunum@ci@positive_taxonomy_m2.av.delta.gemini.axis4.4_4\endcsname{[-0.018, 0.027]}
\expandafter\gdef\csname odunum@val@positive_taxonomy_m2.av.delta.gemini.axis4.4_5\endcsname{0.004}
\expandafter\gdef\csname odunum@n@positive_taxonomy_m2.av.delta.gemini.axis4.4_5\endcsname{173}
\expandafter\gdef\csname odunum@ci@positive_taxonomy_m2.av.delta.gemini.axis4.4_5\endcsname{[-0.028, 0.036]}
\expandafter\gdef\csname odunum@val@positive_taxonomy_m2.av.delta.gemini35_flash_lite.axis1.1_1\endcsname{0.009}
\expandafter\gdef\csname odunum@n@positive_taxonomy_m2.av.delta.gemini35_flash_lite.axis1.1_1\endcsname{92}
\expandafter\gdef\csname odunum@ci@positive_taxonomy_m2.av.delta.gemini35_flash_lite.axis1.1_1\endcsname{[-0.053, 0.071]}
\expandafter\gdef\csname odunum@val@positive_taxonomy_m2.av.delta.gemini35_flash_lite.axis1.1_2\endcsname{-0.014}
\expandafter\gdef\csname odunum@n@positive_taxonomy_m2.av.delta.gemini35_flash_lite.axis1.1_2\endcsname{288}
\expandafter\gdef\csname odunum@ci@positive_taxonomy_m2.av.delta.gemini35_flash_lite.axis1.1_2\endcsname{[-0.046, 0.017]}
\expandafter\gdef\csname odunum@val@positive_taxonomy_m2.av.delta.gemini35_flash_lite.axis1.1_3\endcsname{0.008}
\expandafter\gdef\csname odunum@n@positive_taxonomy_m2.av.delta.gemini35_flash_lite.axis1.1_3\endcsname{286}
\expandafter\gdef\csname odunum@ci@positive_taxonomy_m2.av.delta.gemini35_flash_lite.axis1.1_3\endcsname{[-0.024, 0.040]}
\expandafter\gdef\csname odunum@val@positive_taxonomy_m2.av.delta.gemini35_flash_lite.axis1.1_4\endcsname{0.003}
\expandafter\gdef\csname odunum@n@positive_taxonomy_m2.av.delta.gemini35_flash_lite.axis1.1_4\endcsname{377}
\expandafter\gdef\csname odunum@ci@positive_taxonomy_m2.av.delta.gemini35_flash_lite.axis1.1_4\endcsname{[-0.023, 0.029]}
\expandafter\gdef\csname odunum@val@positive_taxonomy_m2.av.delta.gemini35_flash_lite.axis2.2_1\endcsname{-0.083}
\expandafter\gdef\csname odunum@n@positive_taxonomy_m2.av.delta.gemini35_flash_lite.axis2.2_1\endcsname{118}
\expandafter\gdef\csname odunum@ci@positive_taxonomy_m2.av.delta.gemini35_flash_lite.axis2.2_1\endcsname{[-0.146, -0.019]}
\expandafter\gdef\csname odunum@val@positive_taxonomy_m2.av.delta.gemini35_flash_lite.axis2.2_2\endcsname{0.052}
\expandafter\gdef\csname odunum@n@positive_taxonomy_m2.av.delta.gemini35_flash_lite.axis2.2_2\endcsname{266}
\expandafter\gdef\csname odunum@ci@positive_taxonomy_m2.av.delta.gemini35_flash_lite.axis2.2_2\endcsname{[0.021, 0.084]}
\expandafter\gdef\csname odunum@val@positive_taxonomy_m2.av.delta.gemini35_flash_lite.axis2.2_3\endcsname{0.096}
\expandafter\gdef\csname odunum@n@positive_taxonomy_m2.av.delta.gemini35_flash_lite.axis2.2_3\endcsname{274}
\expandafter\gdef\csname odunum@ci@positive_taxonomy_m2.av.delta.gemini35_flash_lite.axis2.2_3\endcsname{[0.065, 0.126]}
\expandafter\gdef\csname odunum@val@positive_taxonomy_m2.av.delta.gemini35_flash_lite.axis2.2_4\endcsname{-0.093}
\expandafter\gdef\csname odunum@n@positive_taxonomy_m2.av.delta.gemini35_flash_lite.axis2.2_4\endcsname{226}
\expandafter\gdef\csname odunum@ci@positive_taxonomy_m2.av.delta.gemini35_flash_lite.axis2.2_4\endcsname{[-0.131, -0.055]}
\expandafter\gdef\csname odunum@val@positive_taxonomy_m2.av.delta.gemini35_flash_lite.axis2.2_5\endcsname{-0.060}
\expandafter\gdef\csname odunum@n@positive_taxonomy_m2.av.delta.gemini35_flash_lite.axis2.2_5\endcsname{159}
\expandafter\gdef\csname odunum@ci@positive_taxonomy_m2.av.delta.gemini35_flash_lite.axis2.2_5\endcsname{[-0.107, -0.014]}
\expandafter\gdef\csname odunum@val@positive_taxonomy_m2.av.delta.gemini35_flash_lite.axis3.3_1\endcsname{0.025}
\expandafter\gdef\csname odunum@n@positive_taxonomy_m2.av.delta.gemini35_flash_lite.axis3.3_1\endcsname{91}
\expandafter\gdef\csname odunum@ci@positive_taxonomy_m2.av.delta.gemini35_flash_lite.axis3.3_1\endcsname{[-0.032, 0.081]}
\expandafter\gdef\csname odunum@val@positive_taxonomy_m2.av.delta.gemini35_flash_lite.axis3.3_2\endcsname{0.034}
\expandafter\gdef\csname odunum@n@positive_taxonomy_m2.av.delta.gemini35_flash_lite.axis3.3_2\endcsname{85}
\expandafter\gdef\csname odunum@ci@positive_taxonomy_m2.av.delta.gemini35_flash_lite.axis3.3_2\endcsname{[-0.028, 0.094]}
\expandafter\gdef\csname odunum@val@positive_taxonomy_m2.av.delta.gemini35_flash_lite.axis3.3_3\endcsname{-0.169}
\expandafter\gdef\csname odunum@n@positive_taxonomy_m2.av.delta.gemini35_flash_lite.axis3.3_3\endcsname{137}
\expandafter\gdef\csname odunum@ci@positive_taxonomy_m2.av.delta.gemini35_flash_lite.axis3.3_3\endcsname{[-0.223, -0.113]}
\expandafter\gdef\csname odunum@val@positive_taxonomy_m2.av.delta.gemini35_flash_lite.axis3.3_4\endcsname{-0.001}
\expandafter\gdef\csname odunum@n@positive_taxonomy_m2.av.delta.gemini35_flash_lite.axis3.3_4\endcsname{256}
\expandafter\gdef\csname odunum@ci@positive_taxonomy_m2.av.delta.gemini35_flash_lite.axis3.3_4\endcsname{[-0.036, 0.033]}
\expandafter\gdef\csname odunum@val@positive_taxonomy_m2.av.delta.gemini35_flash_lite.axis3.3_5\endcsname{-0.009}
\expandafter\gdef\csname odunum@n@positive_taxonomy_m2.av.delta.gemini35_flash_lite.axis3.3_5\endcsname{185}
\expandafter\gdef\csname odunum@ci@positive_taxonomy_m2.av.delta.gemini35_flash_lite.axis3.3_5\endcsname{[-0.051, 0.034]}
\expandafter\gdef\csname odunum@val@positive_taxonomy_m2.av.delta.gemini35_flash_lite.axis3.3_6\endcsname{0.069}
\expandafter\gdef\csname odunum@n@positive_taxonomy_m2.av.delta.gemini35_flash_lite.axis3.3_6\endcsname{289}
\expandafter\gdef\csname odunum@ci@positive_taxonomy_m2.av.delta.gemini35_flash_lite.axis3.3_6\endcsname{[0.040, 0.100]}
\expandafter\gdef\csname odunum@val@positive_taxonomy_m2.av.delta.gemini35_flash_lite.axis4.4_1\endcsname{0.024}
\expandafter\gdef\csname odunum@n@positive_taxonomy_m2.av.delta.gemini35_flash_lite.axis4.4_1\endcsname{137}
\expandafter\gdef\csname odunum@ci@positive_taxonomy_m2.av.delta.gemini35_flash_lite.axis4.4_1\endcsname{[-0.024, 0.071]}
\expandafter\gdef\csname odunum@val@positive_taxonomy_m2.av.delta.gemini35_flash_lite.axis4.4_2\endcsname{0.036}
\expandafter\gdef\csname odunum@n@positive_taxonomy_m2.av.delta.gemini35_flash_lite.axis4.4_2\endcsname{289}
\expandafter\gdef\csname odunum@ci@positive_taxonomy_m2.av.delta.gemini35_flash_lite.axis4.4_2\endcsname{[0.004, 0.067]}
\expandafter\gdef\csname odunum@val@positive_taxonomy_m2.av.delta.gemini35_flash_lite.axis4.4_3\endcsname{-0.045}
\expandafter\gdef\csname odunum@n@positive_taxonomy_m2.av.delta.gemini35_flash_lite.axis4.4_3\endcsname{120}
\expandafter\gdef\csname odunum@ci@positive_taxonomy_m2.av.delta.gemini35_flash_lite.axis4.4_3\endcsname{[-0.102, 0.013]}
\expandafter\gdef\csname odunum@val@positive_taxonomy_m2.av.delta.gemini35_flash_lite.axis4.4_4\endcsname{-0.036}
\expandafter\gdef\csname odunum@n@positive_taxonomy_m2.av.delta.gemini35_flash_lite.axis4.4_4\endcsname{324}
\expandafter\gdef\csname odunum@ci@positive_taxonomy_m2.av.delta.gemini35_flash_lite.axis4.4_4\endcsname{[-0.066, -0.006]}
\expandafter\gdef\csname odunum@val@positive_taxonomy_m2.av.delta.gemini35_flash_lite.axis4.4_5\endcsname{0.020}
\expandafter\gdef\csname odunum@n@positive_taxonomy_m2.av.delta.gemini35_flash_lite.axis4.4_5\endcsname{173}
\expandafter\gdef\csname odunum@ci@positive_taxonomy_m2.av.delta.gemini35_flash_lite.axis4.4_5\endcsname{[-0.022, 0.061]}
\expandafter\gdef\csname odunum@val@positive_taxonomy_m2.av.delta.gemini37_flash.axis1.1_1\endcsname{0.021}
\expandafter\gdef\csname odunum@n@positive_taxonomy_m2.av.delta.gemini37_flash.axis1.1_1\endcsname{92}
\expandafter\gdef\csname odunum@ci@positive_taxonomy_m2.av.delta.gemini37_flash.axis1.1_1\endcsname{[-0.034, 0.075]}
\expandafter\gdef\csname odunum@val@positive_taxonomy_m2.av.delta.gemini37_flash.axis1.1_2\endcsname{-0.006}
\expandafter\gdef\csname odunum@n@positive_taxonomy_m2.av.delta.gemini37_flash.axis1.1_2\endcsname{288}
\expandafter\gdef\csname odunum@ci@positive_taxonomy_m2.av.delta.gemini37_flash.axis1.1_2\endcsname{[-0.034, 0.023]}
\expandafter\gdef\csname odunum@val@positive_taxonomy_m2.av.delta.gemini37_flash.axis1.1_3\endcsname{0.010}
\expandafter\gdef\csname odunum@n@positive_taxonomy_m2.av.delta.gemini37_flash.axis1.1_3\endcsname{286}
\expandafter\gdef\csname odunum@ci@positive_taxonomy_m2.av.delta.gemini37_flash.axis1.1_3\endcsname{[-0.017, 0.039]}
\expandafter\gdef\csname odunum@val@positive_taxonomy_m2.av.delta.gemini37_flash.axis1.1_4\endcsname{-0.009}
\expandafter\gdef\csname odunum@n@positive_taxonomy_m2.av.delta.gemini37_flash.axis1.1_4\endcsname{377}
\expandafter\gdef\csname odunum@ci@positive_taxonomy_m2.av.delta.gemini37_flash.axis1.1_4\endcsname{[-0.031, 0.013]}
\expandafter\gdef\csname odunum@val@positive_taxonomy_m2.av.delta.gemini37_flash.axis2.2_1\endcsname{-0.034}
\expandafter\gdef\csname odunum@n@positive_taxonomy_m2.av.delta.gemini37_flash.axis2.2_1\endcsname{118}
\expandafter\gdef\csname odunum@ci@positive_taxonomy_m2.av.delta.gemini37_flash.axis2.2_1\endcsname{[-0.090, 0.021]}
\expandafter\gdef\csname odunum@val@positive_taxonomy_m2.av.delta.gemini37_flash.axis2.2_2\endcsname{0.022}
\expandafter\gdef\csname odunum@n@positive_taxonomy_m2.av.delta.gemini37_flash.axis2.2_2\endcsname{266}
\expandafter\gdef\csname odunum@ci@positive_taxonomy_m2.av.delta.gemini37_flash.axis2.2_2\endcsname{[-0.006, 0.049]}
\expandafter\gdef\csname odunum@val@positive_taxonomy_m2.av.delta.gemini37_flash.axis2.2_3\endcsname{0.037}
\expandafter\gdef\csname odunum@n@positive_taxonomy_m2.av.delta.gemini37_flash.axis2.2_3\endcsname{274}
\expandafter\gdef\csname odunum@ci@positive_taxonomy_m2.av.delta.gemini37_flash.axis2.2_3\endcsname{[0.011, 0.064]}
\expandafter\gdef\csname odunum@val@positive_taxonomy_m2.av.delta.gemini37_flash.axis2.2_4\endcsname{-0.020}
\expandafter\gdef\csname odunum@n@positive_taxonomy_m2.av.delta.gemini37_flash.axis2.2_4\endcsname{226}
\expandafter\gdef\csname odunum@ci@positive_taxonomy_m2.av.delta.gemini37_flash.axis2.2_4\endcsname{[-0.054, 0.015]}
\expandafter\gdef\csname odunum@val@positive_taxonomy_m2.av.delta.gemini37_flash.axis2.2_5\endcsname{-0.047}
\expandafter\gdef\csname odunum@n@positive_taxonomy_m2.av.delta.gemini37_flash.axis2.2_5\endcsname{159}
\expandafter\gdef\csname odunum@ci@positive_taxonomy_m2.av.delta.gemini37_flash.axis2.2_5\endcsname{[-0.089, -0.006]}
\expandafter\gdef\csname odunum@val@positive_taxonomy_m2.av.delta.gemini37_flash.axis3.3_1\endcsname{0.034}
\expandafter\gdef\csname odunum@n@positive_taxonomy_m2.av.delta.gemini37_flash.axis3.3_1\endcsname{91}
\expandafter\gdef\csname odunum@ci@positive_taxonomy_m2.av.delta.gemini37_flash.axis3.3_1\endcsname{[-0.019, 0.087]}
\expandafter\gdef\csname odunum@val@positive_taxonomy_m2.av.delta.gemini37_flash.axis3.3_2\endcsname{0.017}
\expandafter\gdef\csname odunum@n@positive_taxonomy_m2.av.delta.gemini37_flash.axis3.3_2\endcsname{85}
\expandafter\gdef\csname odunum@ci@positive_taxonomy_m2.av.delta.gemini37_flash.axis3.3_2\endcsname{[-0.040, 0.071]}
\expandafter\gdef\csname odunum@val@positive_taxonomy_m2.av.delta.gemini37_flash.axis3.3_3\endcsname{-0.093}
\expandafter\gdef\csname odunum@n@positive_taxonomy_m2.av.delta.gemini37_flash.axis3.3_3\endcsname{137}
\expandafter\gdef\csname odunum@ci@positive_taxonomy_m2.av.delta.gemini37_flash.axis3.3_3\endcsname{[-0.141, -0.044]}
\expandafter\gdef\csname odunum@val@positive_taxonomy_m2.av.delta.gemini37_flash.axis3.3_4\endcsname{0.025}
\expandafter\gdef\csname odunum@n@positive_taxonomy_m2.av.delta.gemini37_flash.axis3.3_4\endcsname{256}
\expandafter\gdef\csname odunum@ci@positive_taxonomy_m2.av.delta.gemini37_flash.axis3.3_4\endcsname{[-0.005, 0.055]}
\expandafter\gdef\csname odunum@val@positive_taxonomy_m2.av.delta.gemini37_flash.axis3.3_5\endcsname{-0.093}
\expandafter\gdef\csname odunum@n@positive_taxonomy_m2.av.delta.gemini37_flash.axis3.3_5\endcsname{185}
\expandafter\gdef\csname odunum@ci@positive_taxonomy_m2.av.delta.gemini37_flash.axis3.3_5\endcsname{[-0.130, -0.057]}
\expandafter\gdef\csname odunum@val@positive_taxonomy_m2.av.delta.gemini37_flash.axis3.3_6\endcsname{0.066}
\expandafter\gdef\csname odunum@n@positive_taxonomy_m2.av.delta.gemini37_flash.axis3.3_6\endcsname{289}
\expandafter\gdef\csname odunum@ci@positive_taxonomy_m2.av.delta.gemini37_flash.axis3.3_6\endcsname{[0.040, 0.091]}
\expandafter\gdef\csname odunum@val@positive_taxonomy_m2.av.delta.gemini37_flash.axis4.4_1\endcsname{0.010}
\expandafter\gdef\csname odunum@n@positive_taxonomy_m2.av.delta.gemini37_flash.axis4.4_1\endcsname{137}
\expandafter\gdef\csname odunum@ci@positive_taxonomy_m2.av.delta.gemini37_flash.axis4.4_1\endcsname{[-0.031, 0.051]}
\expandafter\gdef\csname odunum@val@positive_taxonomy_m2.av.delta.gemini37_flash.axis4.4_2\endcsname{-0.011}
\expandafter\gdef\csname odunum@n@positive_taxonomy_m2.av.delta.gemini37_flash.axis4.4_2\endcsname{289}
\expandafter\gdef\csname odunum@ci@positive_taxonomy_m2.av.delta.gemini37_flash.axis4.4_2\endcsname{[-0.039, 0.015]}
\expandafter\gdef\csname odunum@val@positive_taxonomy_m2.av.delta.gemini37_flash.axis4.4_3\endcsname{-0.014}
\expandafter\gdef\csname odunum@n@positive_taxonomy_m2.av.delta.gemini37_flash.axis4.4_3\endcsname{120}
\expandafter\gdef\csname odunum@ci@positive_taxonomy_m2.av.delta.gemini37_flash.axis4.4_3\endcsname{[-0.063, 0.035]}
\expandafter\gdef\csname odunum@val@positive_taxonomy_m2.av.delta.gemini37_flash.axis4.4_4\endcsname{-0.011}
\expandafter\gdef\csname odunum@n@positive_taxonomy_m2.av.delta.gemini37_flash.axis4.4_4\endcsname{324}
\expandafter\gdef\csname odunum@ci@positive_taxonomy_m2.av.delta.gemini37_flash.axis4.4_4\endcsname{[-0.036, 0.014]}
\expandafter\gdef\csname odunum@val@positive_taxonomy_m2.av.delta.gemini37_flash.axis4.4_5\endcsname{0.042}
\expandafter\gdef\csname odunum@n@positive_taxonomy_m2.av.delta.gemini37_flash.axis4.4_5\endcsname{173}
\expandafter\gdef\csname odunum@ci@positive_taxonomy_m2.av.delta.gemini37_flash.axis4.4_5\endcsname{[0.003, 0.081]}
\expandafter\gdef\csname odunum@val@positive_taxonomy_m2.av.delta.ming.axis1.1_1\endcsname{0.066}
\expandafter\gdef\csname odunum@n@positive_taxonomy_m2.av.delta.ming.axis1.1_1\endcsname{92}
\expandafter\gdef\csname odunum@ci@positive_taxonomy_m2.av.delta.ming.axis1.1_1\endcsname{[0.014, 0.120]}
\expandafter\gdef\csname odunum@val@positive_taxonomy_m2.av.delta.ming.axis1.1_2\endcsname{-0.002}
\expandafter\gdef\csname odunum@n@positive_taxonomy_m2.av.delta.ming.axis1.1_2\endcsname{288}
\expandafter\gdef\csname odunum@ci@positive_taxonomy_m2.av.delta.ming.axis1.1_2\endcsname{[-0.028, 0.023]}
\expandafter\gdef\csname odunum@val@positive_taxonomy_m2.av.delta.ming.axis1.1_3\endcsname{-0.007}
\expandafter\gdef\csname odunum@n@positive_taxonomy_m2.av.delta.ming.axis1.1_3\endcsname{286}
\expandafter\gdef\csname odunum@ci@positive_taxonomy_m2.av.delta.ming.axis1.1_3\endcsname{[-0.034, 0.020]}
\expandafter\gdef\csname odunum@val@positive_taxonomy_m2.av.delta.ming.axis1.1_4\endcsname{-0.009}
\expandafter\gdef\csname odunum@n@positive_taxonomy_m2.av.delta.ming.axis1.1_4\endcsname{377}
\expandafter\gdef\csname odunum@ci@positive_taxonomy_m2.av.delta.ming.axis1.1_4\endcsname{[-0.031, 0.012]}
\expandafter\gdef\csname odunum@val@positive_taxonomy_m2.av.delta.ming.axis2.2_1\endcsname{0.003}
\expandafter\gdef\csname odunum@n@positive_taxonomy_m2.av.delta.ming.axis2.2_1\endcsname{118}
\expandafter\gdef\csname odunum@ci@positive_taxonomy_m2.av.delta.ming.axis2.2_1\endcsname{[-0.046, 0.053]}
\expandafter\gdef\csname odunum@val@positive_taxonomy_m2.av.delta.ming.axis2.2_2\endcsname{0.033}
\expandafter\gdef\csname odunum@n@positive_taxonomy_m2.av.delta.ming.axis2.2_2\endcsname{266}
\expandafter\gdef\csname odunum@ci@positive_taxonomy_m2.av.delta.ming.axis2.2_2\endcsname{[0.007, 0.060]}
\expandafter\gdef\csname odunum@val@positive_taxonomy_m2.av.delta.ming.axis2.2_3\endcsname{-2.8\ensuremath{\times 10^{-4}}}
\expandafter\gdef\csname odunum@n@positive_taxonomy_m2.av.delta.ming.axis2.2_3\endcsname{274}
\expandafter\gdef\csname odunum@ci@positive_taxonomy_m2.av.delta.ming.axis2.2_3\endcsname{[-0.027, 0.027]}
\expandafter\gdef\csname odunum@val@positive_taxonomy_m2.av.delta.ming.axis2.2_4\endcsname{-0.022}
\expandafter\gdef\csname odunum@n@positive_taxonomy_m2.av.delta.ming.axis2.2_4\endcsname{226}
\expandafter\gdef\csname odunum@ci@positive_taxonomy_m2.av.delta.ming.axis2.2_4\endcsname{[-0.053, 0.009]}
\expandafter\gdef\csname odunum@val@positive_taxonomy_m2.av.delta.ming.axis2.2_5\endcsname{-0.025}
\expandafter\gdef\csname odunum@n@positive_taxonomy_m2.av.delta.ming.axis2.2_5\endcsname{159}
\expandafter\gdef\csname odunum@ci@positive_taxonomy_m2.av.delta.ming.axis2.2_5\endcsname{[-0.064, 0.012]}
\expandafter\gdef\csname odunum@val@positive_taxonomy_m2.av.delta.ming.axis3.3_1\endcsname{0.009}
\expandafter\gdef\csname odunum@n@positive_taxonomy_m2.av.delta.ming.axis3.3_1\endcsname{91}
\expandafter\gdef\csname odunum@ci@positive_taxonomy_m2.av.delta.ming.axis3.3_1\endcsname{[-0.048, 0.065]}
\expandafter\gdef\csname odunum@val@positive_taxonomy_m2.av.delta.ming.axis3.3_2\endcsname{-0.024}
\expandafter\gdef\csname odunum@n@positive_taxonomy_m2.av.delta.ming.axis3.3_2\endcsname{85}
\expandafter\gdef\csname odunum@ci@positive_taxonomy_m2.av.delta.ming.axis3.3_2\endcsname{[-0.076, 0.028]}
\expandafter\gdef\csname odunum@val@positive_taxonomy_m2.av.delta.ming.axis3.3_3\endcsname{-0.035}
\expandafter\gdef\csname odunum@n@positive_taxonomy_m2.av.delta.ming.axis3.3_3\endcsname{137}
\expandafter\gdef\csname odunum@ci@positive_taxonomy_m2.av.delta.ming.axis3.3_3\endcsname{[-0.079, 0.009]}
\expandafter\gdef\csname odunum@val@positive_taxonomy_m2.av.delta.ming.axis3.3_4\endcsname{-0.020}
\expandafter\gdef\csname odunum@n@positive_taxonomy_m2.av.delta.ming.axis3.3_4\endcsname{256}
\expandafter\gdef\csname odunum@ci@positive_taxonomy_m2.av.delta.ming.axis3.3_4\endcsname{[-0.048, 0.008]}
\expandafter\gdef\csname odunum@val@positive_taxonomy_m2.av.delta.ming.axis3.3_5\endcsname{-0.041}
\expandafter\gdef\csname odunum@n@positive_taxonomy_m2.av.delta.ming.axis3.3_5\endcsname{185}
\expandafter\gdef\csname odunum@ci@positive_taxonomy_m2.av.delta.ming.axis3.3_5\endcsname{[-0.072, -0.009]}
\expandafter\gdef\csname odunum@val@positive_taxonomy_m2.av.delta.ming.axis3.3_6\endcsname{0.064}
\expandafter\gdef\csname odunum@n@positive_taxonomy_m2.av.delta.ming.axis3.3_6\endcsname{289}
\expandafter\gdef\csname odunum@ci@positive_taxonomy_m2.av.delta.ming.axis3.3_6\endcsname{[0.039, 0.090]}
\expandafter\gdef\csname odunum@val@positive_taxonomy_m2.av.delta.ming.axis4.4_1\endcsname{0.001}
\expandafter\gdef\csname odunum@n@positive_taxonomy_m2.av.delta.ming.axis4.4_1\endcsname{137}
\expandafter\gdef\csname odunum@ci@positive_taxonomy_m2.av.delta.ming.axis4.4_1\endcsname{[-0.037, 0.041]}
\expandafter\gdef\csname odunum@val@positive_taxonomy_m2.av.delta.ming.axis4.4_2\endcsname{0.007}
\expandafter\gdef\csname odunum@n@positive_taxonomy_m2.av.delta.ming.axis4.4_2\endcsname{289}
\expandafter\gdef\csname odunum@ci@positive_taxonomy_m2.av.delta.ming.axis4.4_2\endcsname{[-0.019, 0.033]}
\expandafter\gdef\csname odunum@val@positive_taxonomy_m2.av.delta.ming.axis4.4_3\endcsname{0.001}
\expandafter\gdef\csname odunum@n@positive_taxonomy_m2.av.delta.ming.axis4.4_3\endcsname{120}
\expandafter\gdef\csname odunum@ci@positive_taxonomy_m2.av.delta.ming.axis4.4_3\endcsname{[-0.046, 0.048]}
\expandafter\gdef\csname odunum@val@positive_taxonomy_m2.av.delta.ming.axis4.4_4\endcsname{-0.007}
\expandafter\gdef\csname odunum@n@positive_taxonomy_m2.av.delta.ming.axis4.4_4\endcsname{324}
\expandafter\gdef\csname odunum@ci@positive_taxonomy_m2.av.delta.ming.axis4.4_4\endcsname{[-0.031, 0.017]}
\expandafter\gdef\csname odunum@val@positive_taxonomy_m2.av.delta.ming.axis4.4_5\endcsname{0.001}
\expandafter\gdef\csname odunum@n@positive_taxonomy_m2.av.delta.ming.axis4.4_5\endcsname{173}
\expandafter\gdef\csname odunum@ci@positive_taxonomy_m2.av.delta.ming.axis4.4_5\endcsname{[-0.036, 0.036]}
\expandafter\gdef\csname odunum@val@positive_taxonomy_m2.av.delta.minicpm_o.axis1.1_1\endcsname{-0.003}
\expandafter\gdef\csname odunum@n@positive_taxonomy_m2.av.delta.minicpm_o.axis1.1_1\endcsname{92}
\expandafter\gdef\csname odunum@ci@positive_taxonomy_m2.av.delta.minicpm_o.axis1.1_1\endcsname{[-0.066, 0.061]}
\expandafter\gdef\csname odunum@val@positive_taxonomy_m2.av.delta.minicpm_o.axis1.1_2\endcsname{0.025}
\expandafter\gdef\csname odunum@n@positive_taxonomy_m2.av.delta.minicpm_o.axis1.1_2\endcsname{288}
\expandafter\gdef\csname odunum@ci@positive_taxonomy_m2.av.delta.minicpm_o.axis1.1_2\endcsname{[-0.006, 0.055]}
\expandafter\gdef\csname odunum@val@positive_taxonomy_m2.av.delta.minicpm_o.axis1.1_3\endcsname{-0.007}
\expandafter\gdef\csname odunum@n@positive_taxonomy_m2.av.delta.minicpm_o.axis1.1_3\endcsname{286}
\expandafter\gdef\csname odunum@ci@positive_taxonomy_m2.av.delta.minicpm_o.axis1.1_3\endcsname{[-0.037, 0.022]}
\expandafter\gdef\csname odunum@val@positive_taxonomy_m2.av.delta.minicpm_o.axis1.1_4\endcsname{-0.013}
\expandafter\gdef\csname odunum@n@positive_taxonomy_m2.av.delta.minicpm_o.axis1.1_4\endcsname{377}
\expandafter\gdef\csname odunum@ci@positive_taxonomy_m2.av.delta.minicpm_o.axis1.1_4\endcsname{[-0.037, 0.011]}
\expandafter\gdef\csname odunum@val@positive_taxonomy_m2.av.delta.minicpm_o.axis2.2_1\endcsname{-0.003}
\expandafter\gdef\csname odunum@n@positive_taxonomy_m2.av.delta.minicpm_o.axis2.2_1\endcsname{118}
\expandafter\gdef\csname odunum@ci@positive_taxonomy_m2.av.delta.minicpm_o.axis2.2_1\endcsname{[-0.052, 0.049]}
\expandafter\gdef\csname odunum@val@positive_taxonomy_m2.av.delta.minicpm_o.axis2.2_2\endcsname{0.009}
\expandafter\gdef\csname odunum@n@positive_taxonomy_m2.av.delta.minicpm_o.axis2.2_2\endcsname{266}
\expandafter\gdef\csname odunum@ci@positive_taxonomy_m2.av.delta.minicpm_o.axis2.2_2\endcsname{[-0.022, 0.040]}
\expandafter\gdef\csname odunum@val@positive_taxonomy_m2.av.delta.minicpm_o.axis2.2_3\endcsname{-0.047}
\expandafter\gdef\csname odunum@n@positive_taxonomy_m2.av.delta.minicpm_o.axis2.2_3\endcsname{274}
\expandafter\gdef\csname odunum@ci@positive_taxonomy_m2.av.delta.minicpm_o.axis2.2_3\endcsname{[-0.079, -0.015]}
\expandafter\gdef\csname odunum@val@positive_taxonomy_m2.av.delta.minicpm_o.axis2.2_4\endcsname{0.027}
\expandafter\gdef\csname odunum@n@positive_taxonomy_m2.av.delta.minicpm_o.axis2.2_4\endcsname{226}
\expandafter\gdef\csname odunum@ci@positive_taxonomy_m2.av.delta.minicpm_o.axis2.2_4\endcsname{[-0.008, 0.061]}
\expandafter\gdef\csname odunum@val@positive_taxonomy_m2.av.delta.minicpm_o.axis2.2_5\endcsname{0.030}
\expandafter\gdef\csname odunum@n@positive_taxonomy_m2.av.delta.minicpm_o.axis2.2_5\endcsname{159}
\expandafter\gdef\csname odunum@ci@positive_taxonomy_m2.av.delta.minicpm_o.axis2.2_5\endcsname{[-0.010, 0.069]}
\expandafter\gdef\csname odunum@val@positive_taxonomy_m2.av.delta.minicpm_o.axis3.3_1\endcsname{0.056}
\expandafter\gdef\csname odunum@n@positive_taxonomy_m2.av.delta.minicpm_o.axis3.3_1\endcsname{91}
\expandafter\gdef\csname odunum@ci@positive_taxonomy_m2.av.delta.minicpm_o.axis3.3_1\endcsname{[-0.005, 0.120]}
\expandafter\gdef\csname odunum@val@positive_taxonomy_m2.av.delta.minicpm_o.axis3.3_2\endcsname{-0.018}
\expandafter\gdef\csname odunum@n@positive_taxonomy_m2.av.delta.minicpm_o.axis3.3_2\endcsname{85}
\expandafter\gdef\csname odunum@ci@positive_taxonomy_m2.av.delta.minicpm_o.axis3.3_2\endcsname{[-0.085, 0.048]}
\expandafter\gdef\csname odunum@val@positive_taxonomy_m2.av.delta.minicpm_o.axis3.3_3\endcsname{-0.095}
\expandafter\gdef\csname odunum@n@positive_taxonomy_m2.av.delta.minicpm_o.axis3.3_3\endcsname{137}
\expandafter\gdef\csname odunum@ci@positive_taxonomy_m2.av.delta.minicpm_o.axis3.3_3\endcsname{[-0.139, -0.050]}
\expandafter\gdef\csname odunum@val@positive_taxonomy_m2.av.delta.minicpm_o.axis3.3_4\endcsname{0.055}
\expandafter\gdef\csname odunum@n@positive_taxonomy_m2.av.delta.minicpm_o.axis3.3_4\endcsname{256}
\expandafter\gdef\csname odunum@ci@positive_taxonomy_m2.av.delta.minicpm_o.axis3.3_4\endcsname{[0.021, 0.089]}
\expandafter\gdef\csname odunum@val@positive_taxonomy_m2.av.delta.minicpm_o.axis3.3_5\endcsname{0.007}
\expandafter\gdef\csname odunum@n@positive_taxonomy_m2.av.delta.minicpm_o.axis3.3_5\endcsname{185}
\expandafter\gdef\csname odunum@ci@positive_taxonomy_m2.av.delta.minicpm_o.axis3.3_5\endcsname{[-0.031, 0.045]}
\expandafter\gdef\csname odunum@val@positive_taxonomy_m2.av.delta.minicpm_o.axis3.3_6\endcsname{-0.021}
\expandafter\gdef\csname odunum@n@positive_taxonomy_m2.av.delta.minicpm_o.axis3.3_6\endcsname{289}
\expandafter\gdef\csname odunum@ci@positive_taxonomy_m2.av.delta.minicpm_o.axis3.3_6\endcsname{[-0.049, 0.008]}
\expandafter\gdef\csname odunum@val@positive_taxonomy_m2.av.delta.minicpm_o.axis4.4_1\endcsname{-0.043}
\expandafter\gdef\csname odunum@n@positive_taxonomy_m2.av.delta.minicpm_o.axis4.4_1\endcsname{137}
\expandafter\gdef\csname odunum@ci@positive_taxonomy_m2.av.delta.minicpm_o.axis4.4_1\endcsname{[-0.088, 0.003]}
\expandafter\gdef\csname odunum@val@positive_taxonomy_m2.av.delta.minicpm_o.axis4.4_2\endcsname{0.025}
\expandafter\gdef\csname odunum@n@positive_taxonomy_m2.av.delta.minicpm_o.axis4.4_2\endcsname{289}
\expandafter\gdef\csname odunum@ci@positive_taxonomy_m2.av.delta.minicpm_o.axis4.4_2\endcsname{[-0.005, 0.055]}
\expandafter\gdef\csname odunum@val@positive_taxonomy_m2.av.delta.minicpm_o.axis4.4_3\endcsname{-0.063}
\expandafter\gdef\csname odunum@n@positive_taxonomy_m2.av.delta.minicpm_o.axis4.4_3\endcsname{120}
\expandafter\gdef\csname odunum@ci@positive_taxonomy_m2.av.delta.minicpm_o.axis4.4_3\endcsname{[-0.111, -0.014]}
\expandafter\gdef\csname odunum@val@positive_taxonomy_m2.av.delta.minicpm_o.axis4.4_4\endcsname{0.006}
\expandafter\gdef\csname odunum@n@positive_taxonomy_m2.av.delta.minicpm_o.axis4.4_4\endcsname{324}
\expandafter\gdef\csname odunum@ci@positive_taxonomy_m2.av.delta.minicpm_o.axis4.4_4\endcsname{[-0.023, 0.033]}
\expandafter\gdef\csname odunum@val@positive_taxonomy_m2.av.delta.minicpm_o.axis4.4_5\endcsname{0.025}
\expandafter\gdef\csname odunum@n@positive_taxonomy_m2.av.delta.minicpm_o.axis4.4_5\endcsname{173}
\expandafter\gdef\csname odunum@ci@positive_taxonomy_m2.av.delta.minicpm_o.axis4.4_5\endcsname{[-0.017, 0.067]}
\expandafter\gdef\csname odunum@val@positive_taxonomy_m2.av.delta.nemotron.axis1.1_1\endcsname{-0.052}
\expandafter\gdef\csname odunum@n@positive_taxonomy_m2.av.delta.nemotron.axis1.1_1\endcsname{92}
\expandafter\gdef\csname odunum@ci@positive_taxonomy_m2.av.delta.nemotron.axis1.1_1\endcsname{[-0.111, 0.010]}
\expandafter\gdef\csname odunum@val@positive_taxonomy_m2.av.delta.nemotron.axis1.1_2\endcsname{0.014}
\expandafter\gdef\csname odunum@n@positive_taxonomy_m2.av.delta.nemotron.axis1.1_2\endcsname{288}
\expandafter\gdef\csname odunum@ci@positive_taxonomy_m2.av.delta.nemotron.axis1.1_2\endcsname{[-0.018, 0.045]}
\expandafter\gdef\csname odunum@val@positive_taxonomy_m2.av.delta.nemotron.axis1.1_3\endcsname{-0.034}
\expandafter\gdef\csname odunum@n@positive_taxonomy_m2.av.delta.nemotron.axis1.1_3\endcsname{286}
\expandafter\gdef\csname odunum@ci@positive_taxonomy_m2.av.delta.nemotron.axis1.1_3\endcsname{[-0.064, -0.003]}
\expandafter\gdef\csname odunum@val@positive_taxonomy_m2.av.delta.nemotron.axis1.1_4\endcsname{0.028}
\expandafter\gdef\csname odunum@n@positive_taxonomy_m2.av.delta.nemotron.axis1.1_4\endcsname{377}
\expandafter\gdef\csname odunum@ci@positive_taxonomy_m2.av.delta.nemotron.axis1.1_4\endcsname{[0.002, 0.054]}
\expandafter\gdef\csname odunum@val@positive_taxonomy_m2.av.delta.nemotron.axis2.2_1\endcsname{-0.001}
\expandafter\gdef\csname odunum@n@positive_taxonomy_m2.av.delta.nemotron.axis2.2_1\endcsname{118}
\expandafter\gdef\csname odunum@ci@positive_taxonomy_m2.av.delta.nemotron.axis2.2_1\endcsname{[-0.054, 0.054]}
\expandafter\gdef\csname odunum@val@positive_taxonomy_m2.av.delta.nemotron.axis2.2_2\endcsname{0.009}
\expandafter\gdef\csname odunum@n@positive_taxonomy_m2.av.delta.nemotron.axis2.2_2\endcsname{266}
\expandafter\gdef\csname odunum@ci@positive_taxonomy_m2.av.delta.nemotron.axis2.2_2\endcsname{[-0.025, 0.042]}
\expandafter\gdef\csname odunum@val@positive_taxonomy_m2.av.delta.nemotron.axis2.2_3\endcsname{0.008}
\expandafter\gdef\csname odunum@n@positive_taxonomy_m2.av.delta.nemotron.axis2.2_3\endcsname{274}
\expandafter\gdef\csname odunum@ci@positive_taxonomy_m2.av.delta.nemotron.axis2.2_3\endcsname{[-0.025, 0.042]}
\expandafter\gdef\csname odunum@val@positive_taxonomy_m2.av.delta.nemotron.axis2.2_4\endcsname{-0.011}
\expandafter\gdef\csname odunum@n@positive_taxonomy_m2.av.delta.nemotron.axis2.2_4\endcsname{226}
\expandafter\gdef\csname odunum@ci@positive_taxonomy_m2.av.delta.nemotron.axis2.2_4\endcsname{[-0.048, 0.026]}
\expandafter\gdef\csname odunum@val@positive_taxonomy_m2.av.delta.nemotron.axis2.2_5\endcsname{-0.010}
\expandafter\gdef\csname odunum@n@positive_taxonomy_m2.av.delta.nemotron.axis2.2_5\endcsname{159}
\expandafter\gdef\csname odunum@ci@positive_taxonomy_m2.av.delta.nemotron.axis2.2_5\endcsname{[-0.053, 0.033]}
\expandafter\gdef\csname odunum@val@positive_taxonomy_m2.av.delta.nemotron.axis3.3_1\endcsname{0.022}
\expandafter\gdef\csname odunum@n@positive_taxonomy_m2.av.delta.nemotron.axis3.3_1\endcsname{91}
\expandafter\gdef\csname odunum@ci@positive_taxonomy_m2.av.delta.nemotron.axis3.3_1\endcsname{[-0.041, 0.090]}
\expandafter\gdef\csname odunum@val@positive_taxonomy_m2.av.delta.nemotron.axis3.3_2\endcsname{0.015}
\expandafter\gdef\csname odunum@n@positive_taxonomy_m2.av.delta.nemotron.axis3.3_2\endcsname{85}
\expandafter\gdef\csname odunum@ci@positive_taxonomy_m2.av.delta.nemotron.axis3.3_2\endcsname{[-0.050, 0.081]}
\expandafter\gdef\csname odunum@val@positive_taxonomy_m2.av.delta.nemotron.axis3.3_3\endcsname{-0.078}
\expandafter\gdef\csname odunum@n@positive_taxonomy_m2.av.delta.nemotron.axis3.3_3\endcsname{137}
\expandafter\gdef\csname odunum@ci@positive_taxonomy_m2.av.delta.nemotron.axis3.3_3\endcsname{[-0.123, -0.032]}
\expandafter\gdef\csname odunum@val@positive_taxonomy_m2.av.delta.nemotron.axis3.3_4\endcsname{0.022}
\expandafter\gdef\csname odunum@n@positive_taxonomy_m2.av.delta.nemotron.axis3.3_4\endcsname{256}
\expandafter\gdef\csname odunum@ci@positive_taxonomy_m2.av.delta.nemotron.axis3.3_4\endcsname{[-0.015, 0.058]}
\expandafter\gdef\csname odunum@val@positive_taxonomy_m2.av.delta.nemotron.axis3.3_5\endcsname{0.028}
\expandafter\gdef\csname odunum@n@positive_taxonomy_m2.av.delta.nemotron.axis3.3_5\endcsname{185}
\expandafter\gdef\csname odunum@ci@positive_taxonomy_m2.av.delta.nemotron.axis3.3_5\endcsname{[-0.014, 0.069]}
\expandafter\gdef\csname odunum@val@positive_taxonomy_m2.av.delta.nemotron.axis3.3_6\endcsname{-0.011}
\expandafter\gdef\csname odunum@n@positive_taxonomy_m2.av.delta.nemotron.axis3.3_6\endcsname{289}
\expandafter\gdef\csname odunum@ci@positive_taxonomy_m2.av.delta.nemotron.axis3.3_6\endcsname{[-0.041, 0.019]}
\expandafter\gdef\csname odunum@val@positive_taxonomy_m2.av.delta.nemotron.axis4.4_1\endcsname{-0.020}
\expandafter\gdef\csname odunum@n@positive_taxonomy_m2.av.delta.nemotron.axis4.4_1\endcsname{137}
\expandafter\gdef\csname odunum@ci@positive_taxonomy_m2.av.delta.nemotron.axis4.4_1\endcsname{[-0.070, 0.030]}
\expandafter\gdef\csname odunum@val@positive_taxonomy_m2.av.delta.nemotron.axis4.4_2\endcsname{0.039}
\expandafter\gdef\csname odunum@n@positive_taxonomy_m2.av.delta.nemotron.axis4.4_2\endcsname{289}
\expandafter\gdef\csname odunum@ci@positive_taxonomy_m2.av.delta.nemotron.axis4.4_2\endcsname{[0.008, 0.071]}
\expandafter\gdef\csname odunum@val@positive_taxonomy_m2.av.delta.nemotron.axis4.4_3\endcsname{-0.045}
\expandafter\gdef\csname odunum@n@positive_taxonomy_m2.av.delta.nemotron.axis4.4_3\endcsname{120}
\expandafter\gdef\csname odunum@ci@positive_taxonomy_m2.av.delta.nemotron.axis4.4_3\endcsname{[-0.097, 0.011]}
\expandafter\gdef\csname odunum@val@positive_taxonomy_m2.av.delta.nemotron.axis4.4_4\endcsname{-0.006}
\expandafter\gdef\csname odunum@n@positive_taxonomy_m2.av.delta.nemotron.axis4.4_4\endcsname{324}
\expandafter\gdef\csname odunum@ci@positive_taxonomy_m2.av.delta.nemotron.axis4.4_4\endcsname{[-0.034, 0.024]}
\expandafter\gdef\csname odunum@val@positive_taxonomy_m2.av.delta.nemotron.axis4.4_5\endcsname{-0.007}
\expandafter\gdef\csname odunum@n@positive_taxonomy_m2.av.delta.nemotron.axis4.4_5\endcsname{173}
\expandafter\gdef\csname odunum@ci@positive_taxonomy_m2.av.delta.nemotron.axis4.4_5\endcsname{[-0.049, 0.036]}
\expandafter\gdef\csname odunum@val@positive_taxonomy_m2.av.delta.qwen25_omni.axis1.1_1\endcsname{0.033}
\expandafter\gdef\csname odunum@n@positive_taxonomy_m2.av.delta.qwen25_omni.axis1.1_1\endcsname{92}
\expandafter\gdef\csname odunum@ci@positive_taxonomy_m2.av.delta.qwen25_omni.axis1.1_1\endcsname{[-0.025, 0.094]}
\expandafter\gdef\csname odunum@val@positive_taxonomy_m2.av.delta.qwen25_omni.axis1.1_2\endcsname{-0.005}
\expandafter\gdef\csname odunum@n@positive_taxonomy_m2.av.delta.qwen25_omni.axis1.1_2\endcsname{288}
\expandafter\gdef\csname odunum@ci@positive_taxonomy_m2.av.delta.qwen25_omni.axis1.1_2\endcsname{[-0.034, 0.023]}
\expandafter\gdef\csname odunum@val@positive_taxonomy_m2.av.delta.qwen25_omni.axis1.1_3\endcsname{0.018}
\expandafter\gdef\csname odunum@n@positive_taxonomy_m2.av.delta.qwen25_omni.axis1.1_3\endcsname{286}
\expandafter\gdef\csname odunum@ci@positive_taxonomy_m2.av.delta.qwen25_omni.axis1.1_3\endcsname{[-0.012, 0.048]}
\expandafter\gdef\csname odunum@val@positive_taxonomy_m2.av.delta.qwen25_omni.axis1.1_4\endcsname{-0.018}
\expandafter\gdef\csname odunum@n@positive_taxonomy_m2.av.delta.qwen25_omni.axis1.1_4\endcsname{377}
\expandafter\gdef\csname odunum@ci@positive_taxonomy_m2.av.delta.qwen25_omni.axis1.1_4\endcsname{[-0.041, 0.005]}
\expandafter\gdef\csname odunum@val@positive_taxonomy_m2.av.delta.qwen25_omni.axis2.2_1\endcsname{-0.074}
\expandafter\gdef\csname odunum@n@positive_taxonomy_m2.av.delta.qwen25_omni.axis2.2_1\endcsname{118}
\expandafter\gdef\csname odunum@ci@positive_taxonomy_m2.av.delta.qwen25_omni.axis2.2_1\endcsname{[-0.123, -0.024]}
\expandafter\gdef\csname odunum@val@positive_taxonomy_m2.av.delta.qwen25_omni.axis2.2_2\endcsname{0.041}
\expandafter\gdef\csname odunum@n@positive_taxonomy_m2.av.delta.qwen25_omni.axis2.2_2\endcsname{266}
\expandafter\gdef\csname odunum@ci@positive_taxonomy_m2.av.delta.qwen25_omni.axis2.2_2\endcsname{[0.010, 0.070]}
\expandafter\gdef\csname odunum@val@positive_taxonomy_m2.av.delta.qwen25_omni.axis2.2_3\endcsname{0.022}
\expandafter\gdef\csname odunum@n@positive_taxonomy_m2.av.delta.qwen25_omni.axis2.2_3\endcsname{274}
\expandafter\gdef\csname odunum@ci@positive_taxonomy_m2.av.delta.qwen25_omni.axis2.2_3\endcsname{[-0.007, 0.053]}
\expandafter\gdef\csname odunum@val@positive_taxonomy_m2.av.delta.qwen25_omni.axis2.2_4\endcsname{-0.011}
\expandafter\gdef\csname odunum@n@positive_taxonomy_m2.av.delta.qwen25_omni.axis2.2_4\endcsname{226}
\expandafter\gdef\csname odunum@ci@positive_taxonomy_m2.av.delta.qwen25_omni.axis2.2_4\endcsname{[-0.044, 0.021]}
\expandafter\gdef\csname odunum@val@positive_taxonomy_m2.av.delta.qwen25_omni.axis2.2_5\endcsname{-0.035}
\expandafter\gdef\csname odunum@n@positive_taxonomy_m2.av.delta.qwen25_omni.axis2.2_5\endcsname{159}
\expandafter\gdef\csname odunum@ci@positive_taxonomy_m2.av.delta.qwen25_omni.axis2.2_5\endcsname{[-0.075, 0.004]}
\expandafter\gdef\csname odunum@val@positive_taxonomy_m2.av.delta.qwen25_omni.axis3.3_1\endcsname{0.017}
\expandafter\gdef\csname odunum@n@positive_taxonomy_m2.av.delta.qwen25_omni.axis3.3_1\endcsname{91}
\expandafter\gdef\csname odunum@ci@positive_taxonomy_m2.av.delta.qwen25_omni.axis3.3_1\endcsname{[-0.039, 0.074]}
\expandafter\gdef\csname odunum@val@positive_taxonomy_m2.av.delta.qwen25_omni.axis3.3_2\endcsname{0.034}
\expandafter\gdef\csname odunum@n@positive_taxonomy_m2.av.delta.qwen25_omni.axis3.3_2\endcsname{85}
\expandafter\gdef\csname odunum@ci@positive_taxonomy_m2.av.delta.qwen25_omni.axis3.3_2\endcsname{[-0.027, 0.096]}
\expandafter\gdef\csname odunum@val@positive_taxonomy_m2.av.delta.qwen25_omni.axis3.3_3\endcsname{-0.110}
\expandafter\gdef\csname odunum@n@positive_taxonomy_m2.av.delta.qwen25_omni.axis3.3_3\endcsname{137}
\expandafter\gdef\csname odunum@ci@positive_taxonomy_m2.av.delta.qwen25_omni.axis3.3_3\endcsname{[-0.156, -0.065]}
\expandafter\gdef\csname odunum@val@positive_taxonomy_m2.av.delta.qwen25_omni.axis3.3_4\endcsname{0.023}
\expandafter\gdef\csname odunum@n@positive_taxonomy_m2.av.delta.qwen25_omni.axis3.3_4\endcsname{256}
\expandafter\gdef\csname odunum@ci@positive_taxonomy_m2.av.delta.qwen25_omni.axis3.3_4\endcsname{[-0.009, 0.053]}
\expandafter\gdef\csname odunum@val@positive_taxonomy_m2.av.delta.qwen25_omni.axis3.3_5\endcsname{-0.007}
\expandafter\gdef\csname odunum@n@positive_taxonomy_m2.av.delta.qwen25_omni.axis3.3_5\endcsname{185}
\expandafter\gdef\csname odunum@ci@positive_taxonomy_m2.av.delta.qwen25_omni.axis3.3_5\endcsname{[-0.044, 0.030]}
\expandafter\gdef\csname odunum@val@positive_taxonomy_m2.av.delta.qwen25_omni.axis3.3_6\endcsname{0.021}
\expandafter\gdef\csname odunum@n@positive_taxonomy_m2.av.delta.qwen25_omni.axis3.3_6\endcsname{289}
\expandafter\gdef\csname odunum@ci@positive_taxonomy_m2.av.delta.qwen25_omni.axis3.3_6\endcsname{[-0.007, 0.050]}
\expandafter\gdef\csname odunum@val@positive_taxonomy_m2.av.delta.qwen25_omni.axis4.4_1\endcsname{0.014}
\expandafter\gdef\csname odunum@n@positive_taxonomy_m2.av.delta.qwen25_omni.axis4.4_1\endcsname{137}
\expandafter\gdef\csname odunum@ci@positive_taxonomy_m2.av.delta.qwen25_omni.axis4.4_1\endcsname{[-0.030, 0.058]}
\expandafter\gdef\csname odunum@val@positive_taxonomy_m2.av.delta.qwen25_omni.axis4.4_2\endcsname{0.028}
\expandafter\gdef\csname odunum@n@positive_taxonomy_m2.av.delta.qwen25_omni.axis4.4_2\endcsname{289}
\expandafter\gdef\csname odunum@ci@positive_taxonomy_m2.av.delta.qwen25_omni.axis4.4_2\endcsname{[4.0\ensuremath{\times 10^{-4}}, 0.056]}
\expandafter\gdef\csname odunum@val@positive_taxonomy_m2.av.delta.qwen25_omni.axis4.4_3\endcsname{-0.076}
\expandafter\gdef\csname odunum@n@positive_taxonomy_m2.av.delta.qwen25_omni.axis4.4_3\endcsname{120}
\expandafter\gdef\csname odunum@ci@positive_taxonomy_m2.av.delta.qwen25_omni.axis4.4_3\endcsname{[-0.126, -0.024]}
\expandafter\gdef\csname odunum@val@positive_taxonomy_m2.av.delta.qwen25_omni.axis4.4_4\endcsname{-0.003}
\expandafter\gdef\csname odunum@n@positive_taxonomy_m2.av.delta.qwen25_omni.axis4.4_4\endcsname{324}
\expandafter\gdef\csname odunum@ci@positive_taxonomy_m2.av.delta.qwen25_omni.axis4.4_4\endcsname{[-0.029, 0.023]}
\expandafter\gdef\csname odunum@val@positive_taxonomy_m2.av.delta.qwen25_omni.axis4.4_5\endcsname{-4.1\ensuremath{\times 10^{-4}}}
\expandafter\gdef\csname odunum@n@positive_taxonomy_m2.av.delta.qwen25_omni.axis4.4_5\endcsname{173}
\expandafter\gdef\csname odunum@ci@positive_taxonomy_m2.av.delta.qwen25_omni.axis4.4_5\endcsname{[-0.040, 0.040]}
\expandafter\gdef\csname odunum@val@positive_taxonomy_m2.av.delta.qwen3_omni_instruct.axis1.1_1\endcsname{-0.008}
\expandafter\gdef\csname odunum@n@positive_taxonomy_m2.av.delta.qwen3_omni_instruct.axis1.1_1\endcsname{92}
\expandafter\gdef\csname odunum@ci@positive_taxonomy_m2.av.delta.qwen3_omni_instruct.axis1.1_1\endcsname{[-0.065, 0.048]}
\expandafter\gdef\csname odunum@val@positive_taxonomy_m2.av.delta.qwen3_omni_instruct.axis1.1_2\endcsname{0.017}
\expandafter\gdef\csname odunum@n@positive_taxonomy_m2.av.delta.qwen3_omni_instruct.axis1.1_2\endcsname{288}
\expandafter\gdef\csname odunum@ci@positive_taxonomy_m2.av.delta.qwen3_omni_instruct.axis1.1_2\endcsname{[-0.008, 0.042]}
\expandafter\gdef\csname odunum@val@positive_taxonomy_m2.av.delta.qwen3_omni_instruct.axis1.1_3\endcsname{0.007}
\expandafter\gdef\csname odunum@n@positive_taxonomy_m2.av.delta.qwen3_omni_instruct.axis1.1_3\endcsname{286}
\expandafter\gdef\csname odunum@ci@positive_taxonomy_m2.av.delta.qwen3_omni_instruct.axis1.1_3\endcsname{[-0.019, 0.034]}
\expandafter\gdef\csname odunum@val@positive_taxonomy_m2.av.delta.qwen3_omni_instruct.axis1.1_4\endcsname{-0.017}
\expandafter\gdef\csname odunum@n@positive_taxonomy_m2.av.delta.qwen3_omni_instruct.axis1.1_4\endcsname{377}
\expandafter\gdef\csname odunum@ci@positive_taxonomy_m2.av.delta.qwen3_omni_instruct.axis1.1_4\endcsname{[-0.037, 0.004]}
\expandafter\gdef\csname odunum@val@positive_taxonomy_m2.av.delta.qwen3_omni_instruct.axis2.2_1\endcsname{0.017}
\expandafter\gdef\csname odunum@n@positive_taxonomy_m2.av.delta.qwen3_omni_instruct.axis2.2_1\endcsname{118}
\expandafter\gdef\csname odunum@ci@positive_taxonomy_m2.av.delta.qwen3_omni_instruct.axis2.2_1\endcsname{[-0.029, 0.062]}
\expandafter\gdef\csname odunum@val@positive_taxonomy_m2.av.delta.qwen3_omni_instruct.axis2.2_2\endcsname{-0.018}
\expandafter\gdef\csname odunum@n@positive_taxonomy_m2.av.delta.qwen3_omni_instruct.axis2.2_2\endcsname{266}
\expandafter\gdef\csname odunum@ci@positive_taxonomy_m2.av.delta.qwen3_omni_instruct.axis2.2_2\endcsname{[-0.045, 0.009]}
\expandafter\gdef\csname odunum@val@positive_taxonomy_m2.av.delta.qwen3_omni_instruct.axis2.2_3\endcsname{-0.008}
\expandafter\gdef\csname odunum@n@positive_taxonomy_m2.av.delta.qwen3_omni_instruct.axis2.2_3\endcsname{274}
\expandafter\gdef\csname odunum@ci@positive_taxonomy_m2.av.delta.qwen3_omni_instruct.axis2.2_3\endcsname{[-0.035, 0.018]}
\expandafter\gdef\csname odunum@val@positive_taxonomy_m2.av.delta.qwen3_omni_instruct.axis2.2_4\endcsname{0.005}
\expandafter\gdef\csname odunum@n@positive_taxonomy_m2.av.delta.qwen3_omni_instruct.axis2.2_4\endcsname{226}
\expandafter\gdef\csname odunum@ci@positive_taxonomy_m2.av.delta.qwen3_omni_instruct.axis2.2_4\endcsname{[-0.028, 0.036]}
\expandafter\gdef\csname odunum@val@positive_taxonomy_m2.av.delta.qwen3_omni_instruct.axis2.2_5\endcsname{0.025}
\expandafter\gdef\csname odunum@n@positive_taxonomy_m2.av.delta.qwen3_omni_instruct.axis2.2_5\endcsname{159}
\expandafter\gdef\csname odunum@ci@positive_taxonomy_m2.av.delta.qwen3_omni_instruct.axis2.2_5\endcsname{[-0.011, 0.061]}
\expandafter\gdef\csname odunum@val@positive_taxonomy_m2.av.delta.qwen3_omni_instruct.axis3.3_1\endcsname{0.006}
\expandafter\gdef\csname odunum@n@positive_taxonomy_m2.av.delta.qwen3_omni_instruct.axis3.3_1\endcsname{91}
\expandafter\gdef\csname odunum@ci@positive_taxonomy_m2.av.delta.qwen3_omni_instruct.axis3.3_1\endcsname{[-0.048, 0.060]}
\expandafter\gdef\csname odunum@val@positive_taxonomy_m2.av.delta.qwen3_omni_instruct.axis3.3_2\endcsname{0.006}
\expandafter\gdef\csname odunum@n@positive_taxonomy_m2.av.delta.qwen3_omni_instruct.axis3.3_2\endcsname{85}
\expandafter\gdef\csname odunum@ci@positive_taxonomy_m2.av.delta.qwen3_omni_instruct.axis3.3_2\endcsname{[-0.047, 0.059]}
\expandafter\gdef\csname odunum@val@positive_taxonomy_m2.av.delta.qwen3_omni_instruct.axis3.3_3\endcsname{-0.047}
\expandafter\gdef\csname odunum@n@positive_taxonomy_m2.av.delta.qwen3_omni_instruct.axis3.3_3\endcsname{137}
\expandafter\gdef\csname odunum@ci@positive_taxonomy_m2.av.delta.qwen3_omni_instruct.axis3.3_3\endcsname{[-0.088, -0.006]}
\expandafter\gdef\csname odunum@val@positive_taxonomy_m2.av.delta.qwen3_omni_instruct.axis3.3_4\endcsname{-0.007}
\expandafter\gdef\csname odunum@n@positive_taxonomy_m2.av.delta.qwen3_omni_instruct.axis3.3_4\endcsname{256}
\expandafter\gdef\csname odunum@ci@positive_taxonomy_m2.av.delta.qwen3_omni_instruct.axis3.3_4\endcsname{[-0.035, 0.022]}
\expandafter\gdef\csname odunum@val@positive_taxonomy_m2.av.delta.qwen3_omni_instruct.axis3.3_5\endcsname{-0.024}
\expandafter\gdef\csname odunum@n@positive_taxonomy_m2.av.delta.qwen3_omni_instruct.axis3.3_5\endcsname{185}
\expandafter\gdef\csname odunum@ci@positive_taxonomy_m2.av.delta.qwen3_omni_instruct.axis3.3_5\endcsname{[-0.058, 0.011]}
\expandafter\gdef\csname odunum@val@positive_taxonomy_m2.av.delta.qwen3_omni_instruct.axis3.3_6\endcsname{0.039}
\expandafter\gdef\csname odunum@n@positive_taxonomy_m2.av.delta.qwen3_omni_instruct.axis3.3_6\endcsname{289}
\expandafter\gdef\csname odunum@ci@positive_taxonomy_m2.av.delta.qwen3_omni_instruct.axis3.3_6\endcsname{[0.015, 0.065]}
\expandafter\gdef\csname odunum@val@positive_taxonomy_m2.av.delta.qwen3_omni_instruct.axis4.4_1\endcsname{-0.031}
\expandafter\gdef\csname odunum@n@positive_taxonomy_m2.av.delta.qwen3_omni_instruct.axis4.4_1\endcsname{137}
\expandafter\gdef\csname odunum@ci@positive_taxonomy_m2.av.delta.qwen3_omni_instruct.axis4.4_1\endcsname{[-0.070, 0.008]}
\expandafter\gdef\csname odunum@val@positive_taxonomy_m2.av.delta.qwen3_omni_instruct.axis4.4_2\endcsname{0.030}
\expandafter\gdef\csname odunum@n@positive_taxonomy_m2.av.delta.qwen3_omni_instruct.axis4.4_2\endcsname{289}
\expandafter\gdef\csname odunum@ci@positive_taxonomy_m2.av.delta.qwen3_omni_instruct.axis4.4_2\endcsname{[0.005, 0.055]}
\expandafter\gdef\csname odunum@val@positive_taxonomy_m2.av.delta.qwen3_omni_instruct.axis4.4_3\endcsname{-0.035}
\expandafter\gdef\csname odunum@n@positive_taxonomy_m2.av.delta.qwen3_omni_instruct.axis4.4_3\endcsname{120}
\expandafter\gdef\csname odunum@ci@positive_taxonomy_m2.av.delta.qwen3_omni_instruct.axis4.4_3\endcsname{[-0.086, 0.015]}
\expandafter\gdef\csname odunum@val@positive_taxonomy_m2.av.delta.qwen3_omni_instruct.axis4.4_4\endcsname{-0.008}
\expandafter\gdef\csname odunum@n@positive_taxonomy_m2.av.delta.qwen3_omni_instruct.axis4.4_4\endcsname{324}
\expandafter\gdef\csname odunum@ci@positive_taxonomy_m2.av.delta.qwen3_omni_instruct.axis4.4_4\endcsname{[-0.032, 0.017]}
\expandafter\gdef\csname odunum@val@positive_taxonomy_m2.av.delta.qwen3_omni_instruct.axis4.4_5\endcsname{0.013}
\expandafter\gdef\csname odunum@n@positive_taxonomy_m2.av.delta.qwen3_omni_instruct.axis4.4_5\endcsname{173}
\expandafter\gdef\csname odunum@ci@positive_taxonomy_m2.av.delta.qwen3_omni_instruct.axis4.4_5\endcsname{[-0.019, 0.046]}
\expandafter\gdef\csname odunum@val@positive_taxonomy_m2.av.delta.qwen3_omni_think.axis1.1_1\endcsname{0.044}
\expandafter\gdef\csname odunum@n@positive_taxonomy_m2.av.delta.qwen3_omni_think.axis1.1_1\endcsname{92}
\expandafter\gdef\csname odunum@ci@positive_taxonomy_m2.av.delta.qwen3_omni_think.axis1.1_1\endcsname{[-0.009, 0.097]}
\expandafter\gdef\csname odunum@val@positive_taxonomy_m2.av.delta.qwen3_omni_think.axis1.1_2\endcsname{0.009}
\expandafter\gdef\csname odunum@n@positive_taxonomy_m2.av.delta.qwen3_omni_think.axis1.1_2\endcsname{288}
\expandafter\gdef\csname odunum@ci@positive_taxonomy_m2.av.delta.qwen3_omni_think.axis1.1_2\endcsname{[-0.017, 0.034]}
\expandafter\gdef\csname odunum@val@positive_taxonomy_m2.av.delta.qwen3_omni_think.axis1.1_3\endcsname{-0.019}
\expandafter\gdef\csname odunum@n@positive_taxonomy_m2.av.delta.qwen3_omni_think.axis1.1_3\endcsname{286}
\expandafter\gdef\csname odunum@ci@positive_taxonomy_m2.av.delta.qwen3_omni_think.axis1.1_3\endcsname{[-0.044, 0.007]}
\expandafter\gdef\csname odunum@val@positive_taxonomy_m2.av.delta.qwen3_omni_think.axis1.1_4\endcsname{-0.003}
\expandafter\gdef\csname odunum@n@positive_taxonomy_m2.av.delta.qwen3_omni_think.axis1.1_4\endcsname{377}
\expandafter\gdef\csname odunum@ci@positive_taxonomy_m2.av.delta.qwen3_omni_think.axis1.1_4\endcsname{[-0.024, 0.018]}
\expandafter\gdef\csname odunum@val@positive_taxonomy_m2.av.delta.qwen3_omni_think.axis2.2_1\endcsname{0.017}
\expandafter\gdef\csname odunum@n@positive_taxonomy_m2.av.delta.qwen3_omni_think.axis2.2_1\endcsname{118}
\expandafter\gdef\csname odunum@ci@positive_taxonomy_m2.av.delta.qwen3_omni_think.axis2.2_1\endcsname{[-0.032, 0.067]}
\expandafter\gdef\csname odunum@val@positive_taxonomy_m2.av.delta.qwen3_omni_think.axis2.2_2\endcsname{-0.014}
\expandafter\gdef\csname odunum@n@positive_taxonomy_m2.av.delta.qwen3_omni_think.axis2.2_2\endcsname{266}
\expandafter\gdef\csname odunum@ci@positive_taxonomy_m2.av.delta.qwen3_omni_think.axis2.2_2\endcsname{[-0.040, 0.013]}
\expandafter\gdef\csname odunum@val@positive_taxonomy_m2.av.delta.qwen3_omni_think.axis2.2_3\endcsname{0.019}
\expandafter\gdef\csname odunum@n@positive_taxonomy_m2.av.delta.qwen3_omni_think.axis2.2_3\endcsname{274}
\expandafter\gdef\csname odunum@ci@positive_taxonomy_m2.av.delta.qwen3_omni_think.axis2.2_3\endcsname{[-0.006, 0.045]}
\expandafter\gdef\csname odunum@val@positive_taxonomy_m2.av.delta.qwen3_omni_think.axis2.2_4\endcsname{-0.020}
\expandafter\gdef\csname odunum@n@positive_taxonomy_m2.av.delta.qwen3_omni_think.axis2.2_4\endcsname{226}
\expandafter\gdef\csname odunum@ci@positive_taxonomy_m2.av.delta.qwen3_omni_think.axis2.2_4\endcsname{[-0.052, 0.012]}
\expandafter\gdef\csname odunum@val@positive_taxonomy_m2.av.delta.qwen3_omni_think.axis2.2_5\endcsname{0.006}
\expandafter\gdef\csname odunum@n@positive_taxonomy_m2.av.delta.qwen3_omni_think.axis2.2_5\endcsname{159}
\expandafter\gdef\csname odunum@ci@positive_taxonomy_m2.av.delta.qwen3_omni_think.axis2.2_5\endcsname{[-0.029, 0.040]}
\expandafter\gdef\csname odunum@val@positive_taxonomy_m2.av.delta.qwen3_omni_think.axis3.3_1\endcsname{-0.002}
\expandafter\gdef\csname odunum@n@positive_taxonomy_m2.av.delta.qwen3_omni_think.axis3.3_1\endcsname{91}
\expandafter\gdef\csname odunum@ci@positive_taxonomy_m2.av.delta.qwen3_omni_think.axis3.3_1\endcsname{[-0.051, 0.048]}
\expandafter\gdef\csname odunum@val@positive_taxonomy_m2.av.delta.qwen3_omni_think.axis3.3_2\endcsname{-0.039}
\expandafter\gdef\csname odunum@n@positive_taxonomy_m2.av.delta.qwen3_omni_think.axis3.3_2\endcsname{85}
\expandafter\gdef\csname odunum@ci@positive_taxonomy_m2.av.delta.qwen3_omni_think.axis3.3_2\endcsname{[-0.089, 0.011]}
\expandafter\gdef\csname odunum@val@positive_taxonomy_m2.av.delta.qwen3_omni_think.axis3.3_3\endcsname{-0.024}
\expandafter\gdef\csname odunum@n@positive_taxonomy_m2.av.delta.qwen3_omni_think.axis3.3_3\endcsname{137}
\expandafter\gdef\csname odunum@ci@positive_taxonomy_m2.av.delta.qwen3_omni_think.axis3.3_3\endcsname{[-0.065, 0.018]}
\expandafter\gdef\csname odunum@val@positive_taxonomy_m2.av.delta.qwen3_omni_think.axis3.3_4\endcsname{0.014}
\expandafter\gdef\csname odunum@n@positive_taxonomy_m2.av.delta.qwen3_omni_think.axis3.3_4\endcsname{256}
\expandafter\gdef\csname odunum@ci@positive_taxonomy_m2.av.delta.qwen3_omni_think.axis3.3_4\endcsname{[-0.014, 0.042]}
\expandafter\gdef\csname odunum@val@positive_taxonomy_m2.av.delta.qwen3_omni_think.axis3.3_5\endcsname{-0.062}
\expandafter\gdef\csname odunum@n@positive_taxonomy_m2.av.delta.qwen3_omni_think.axis3.3_5\endcsname{185}
\expandafter\gdef\csname odunum@ci@positive_taxonomy_m2.av.delta.qwen3_omni_think.axis3.3_5\endcsname{[-0.096, -0.027]}
\expandafter\gdef\csname odunum@val@positive_taxonomy_m2.av.delta.qwen3_omni_think.axis3.3_6\endcsname{0.050}
\expandafter\gdef\csname odunum@n@positive_taxonomy_m2.av.delta.qwen3_omni_think.axis3.3_6\endcsname{289}
\expandafter\gdef\csname odunum@ci@positive_taxonomy_m2.av.delta.qwen3_omni_think.axis3.3_6\endcsname{[0.025, 0.075]}
\expandafter\gdef\csname odunum@val@positive_taxonomy_m2.av.delta.qwen3_omni_think.axis4.4_1\endcsname{-0.036}
\expandafter\gdef\csname odunum@n@positive_taxonomy_m2.av.delta.qwen3_omni_think.axis4.4_1\endcsname{137}
\expandafter\gdef\csname odunum@ci@positive_taxonomy_m2.av.delta.qwen3_omni_think.axis4.4_1\endcsname{[-0.075, 0.002]}
\expandafter\gdef\csname odunum@val@positive_taxonomy_m2.av.delta.qwen3_omni_think.axis4.4_2\endcsname{0.033}
\expandafter\gdef\csname odunum@n@positive_taxonomy_m2.av.delta.qwen3_omni_think.axis4.4_2\endcsname{289}
\expandafter\gdef\csname odunum@ci@positive_taxonomy_m2.av.delta.qwen3_omni_think.axis4.4_2\endcsname{[0.007, 0.058]}
\expandafter\gdef\csname odunum@val@positive_taxonomy_m2.av.delta.qwen3_omni_think.axis4.4_3\endcsname{0.004}
\expandafter\gdef\csname odunum@n@positive_taxonomy_m2.av.delta.qwen3_omni_think.axis4.4_3\endcsname{120}
\expandafter\gdef\csname odunum@ci@positive_taxonomy_m2.av.delta.qwen3_omni_think.axis4.4_3\endcsname{[-0.045, 0.052]}
\expandafter\gdef\csname odunum@val@positive_taxonomy_m2.av.delta.qwen3_omni_think.axis4.4_4\endcsname{-0.018}
\expandafter\gdef\csname odunum@n@positive_taxonomy_m2.av.delta.qwen3_omni_think.axis4.4_4\endcsname{324}
\expandafter\gdef\csname odunum@ci@positive_taxonomy_m2.av.delta.qwen3_omni_think.axis4.4_4\endcsname{[-0.042, 0.005]}
\expandafter\gdef\csname odunum@val@positive_taxonomy_m2.av.delta.qwen3_omni_think.axis4.4_5\endcsname{0.005}
\expandafter\gdef\csname odunum@n@positive_taxonomy_m2.av.delta.qwen3_omni_think.axis4.4_5\endcsname{173}
\expandafter\gdef\csname odunum@ci@positive_taxonomy_m2.av.delta.qwen3_omni_think.axis4.4_5\endcsname{[-0.028, 0.039]}
\expandafter\gdef\csname odunum@val@positive_taxonomy_m2.av.delta.qwen_plus.axis1.1_1\endcsname{0.071}
\expandafter\gdef\csname odunum@n@positive_taxonomy_m2.av.delta.qwen_plus.axis1.1_1\endcsname{92}
\expandafter\gdef\csname odunum@ci@positive_taxonomy_m2.av.delta.qwen_plus.axis1.1_1\endcsname{[0.021, 0.120]}
\expandafter\gdef\csname odunum@val@positive_taxonomy_m2.av.delta.qwen_plus.axis1.1_2\endcsname{-0.024}
\expandafter\gdef\csname odunum@n@positive_taxonomy_m2.av.delta.qwen_plus.axis1.1_2\endcsname{288}
\expandafter\gdef\csname odunum@ci@positive_taxonomy_m2.av.delta.qwen_plus.axis1.1_2\endcsname{[-0.050, 0.002]}
\expandafter\gdef\csname odunum@val@positive_taxonomy_m2.av.delta.qwen_plus.axis1.1_3\endcsname{0.016}
\expandafter\gdef\csname odunum@n@positive_taxonomy_m2.av.delta.qwen_plus.axis1.1_3\endcsname{286}
\expandafter\gdef\csname odunum@ci@positive_taxonomy_m2.av.delta.qwen_plus.axis1.1_3\endcsname{[-0.010, 0.041]}
\expandafter\gdef\csname odunum@val@positive_taxonomy_m2.av.delta.qwen_plus.axis1.1_4\endcsname{-0.011}
\expandafter\gdef\csname odunum@n@positive_taxonomy_m2.av.delta.qwen_plus.axis1.1_4\endcsname{377}
\expandafter\gdef\csname odunum@ci@positive_taxonomy_m2.av.delta.qwen_plus.axis1.1_4\endcsname{[-0.032, 0.009]}
\expandafter\gdef\csname odunum@val@positive_taxonomy_m2.av.delta.qwen_plus.axis2.2_1\endcsname{-0.040}
\expandafter\gdef\csname odunum@n@positive_taxonomy_m2.av.delta.qwen_plus.axis2.2_1\endcsname{118}
\expandafter\gdef\csname odunum@ci@positive_taxonomy_m2.av.delta.qwen_plus.axis2.2_1\endcsname{[-0.092, 0.011]}
\expandafter\gdef\csname odunum@val@positive_taxonomy_m2.av.delta.qwen_plus.axis2.2_2\endcsname{0.024}
\expandafter\gdef\csname odunum@n@positive_taxonomy_m2.av.delta.qwen_plus.axis2.2_2\endcsname{266}
\expandafter\gdef\csname odunum@ci@positive_taxonomy_m2.av.delta.qwen_plus.axis2.2_2\endcsname{[-0.002, 0.049]}
\expandafter\gdef\csname odunum@val@positive_taxonomy_m2.av.delta.qwen_plus.axis2.2_3\endcsname{0.058}
\expandafter\gdef\csname odunum@n@positive_taxonomy_m2.av.delta.qwen_plus.axis2.2_3\endcsname{274}
\expandafter\gdef\csname odunum@ci@positive_taxonomy_m2.av.delta.qwen_plus.axis2.2_3\endcsname{[0.034, 0.083]}
\expandafter\gdef\csname odunum@val@positive_taxonomy_m2.av.delta.qwen_plus.axis2.2_4\endcsname{-0.071}
\expandafter\gdef\csname odunum@n@positive_taxonomy_m2.av.delta.qwen_plus.axis2.2_4\endcsname{226}
\expandafter\gdef\csname odunum@ci@positive_taxonomy_m2.av.delta.qwen_plus.axis2.2_4\endcsname{[-0.104, -0.038]}
\expandafter\gdef\csname odunum@val@positive_taxonomy_m2.av.delta.qwen_plus.axis2.2_5\endcsname{-0.009}
\expandafter\gdef\csname odunum@n@positive_taxonomy_m2.av.delta.qwen_plus.axis2.2_5\endcsname{159}
\expandafter\gdef\csname odunum@ci@positive_taxonomy_m2.av.delta.qwen_plus.axis2.2_5\endcsname{[-0.045, 0.025]}
\expandafter\gdef\csname odunum@val@positive_taxonomy_m2.av.delta.qwen_plus.axis3.3_1\endcsname{-0.022}
\expandafter\gdef\csname odunum@n@positive_taxonomy_m2.av.delta.qwen_plus.axis3.3_1\endcsname{91}
\expandafter\gdef\csname odunum@ci@positive_taxonomy_m2.av.delta.qwen_plus.axis3.3_1\endcsname{[-0.078, 0.032]}
\expandafter\gdef\csname odunum@val@positive_taxonomy_m2.av.delta.qwen_plus.axis3.3_2\endcsname{0.029}
\expandafter\gdef\csname odunum@n@positive_taxonomy_m2.av.delta.qwen_plus.axis3.3_2\endcsname{85}
\expandafter\gdef\csname odunum@ci@positive_taxonomy_m2.av.delta.qwen_plus.axis3.3_2\endcsname{[-0.021, 0.078]}
\expandafter\gdef\csname odunum@val@positive_taxonomy_m2.av.delta.qwen_plus.axis3.3_3\endcsname{-0.037}
\expandafter\gdef\csname odunum@n@positive_taxonomy_m2.av.delta.qwen_plus.axis3.3_3\endcsname{137}
\expandafter\gdef\csname odunum@ci@positive_taxonomy_m2.av.delta.qwen_plus.axis3.3_3\endcsname{[-0.082, 0.008]}
\expandafter\gdef\csname odunum@val@positive_taxonomy_m2.av.delta.qwen_plus.axis3.3_4\endcsname{0.003}
\expandafter\gdef\csname odunum@n@positive_taxonomy_m2.av.delta.qwen_plus.axis3.3_4\endcsname{256}
\expandafter\gdef\csname odunum@ci@positive_taxonomy_m2.av.delta.qwen_plus.axis3.3_4\endcsname{[-0.024, 0.030]}
\expandafter\gdef\csname odunum@val@positive_taxonomy_m2.av.delta.qwen_plus.axis3.3_5\endcsname{-0.061}
\expandafter\gdef\csname odunum@n@positive_taxonomy_m2.av.delta.qwen_plus.axis3.3_5\endcsname{185}
\expandafter\gdef\csname odunum@ci@positive_taxonomy_m2.av.delta.qwen_plus.axis3.3_5\endcsname{[-0.097, -0.026]}
\expandafter\gdef\csname odunum@val@positive_taxonomy_m2.av.delta.qwen_plus.axis3.3_6\endcsname{0.053}
\expandafter\gdef\csname odunum@n@positive_taxonomy_m2.av.delta.qwen_plus.axis3.3_6\endcsname{289}
\expandafter\gdef\csname odunum@ci@positive_taxonomy_m2.av.delta.qwen_plus.axis3.3_6\endcsname{[0.029, 0.076]}
\expandafter\gdef\csname odunum@val@positive_taxonomy_m2.av.delta.qwen_plus.axis4.4_1\endcsname{-4.4\ensuremath{\times 10^{-4}}}
\expandafter\gdef\csname odunum@n@positive_taxonomy_m2.av.delta.qwen_plus.axis4.4_1\endcsname{137}
\expandafter\gdef\csname odunum@ci@positive_taxonomy_m2.av.delta.qwen_plus.axis4.4_1\endcsname{[-0.041, 0.039]}
\expandafter\gdef\csname odunum@val@positive_taxonomy_m2.av.delta.qwen_plus.axis4.4_2\endcsname{0.002}
\expandafter\gdef\csname odunum@n@positive_taxonomy_m2.av.delta.qwen_plus.axis4.4_2\endcsname{289}
\expandafter\gdef\csname odunum@ci@positive_taxonomy_m2.av.delta.qwen_plus.axis4.4_2\endcsname{[-0.023, 0.026]}
\expandafter\gdef\csname odunum@val@positive_taxonomy_m2.av.delta.qwen_plus.axis4.4_3\endcsname{-0.004}
\expandafter\gdef\csname odunum@n@positive_taxonomy_m2.av.delta.qwen_plus.axis4.4_3\endcsname{120}
\expandafter\gdef\csname odunum@ci@positive_taxonomy_m2.av.delta.qwen_plus.axis4.4_3\endcsname{[-0.053, 0.045]}
\expandafter\gdef\csname odunum@val@positive_taxonomy_m2.av.delta.qwen_plus.axis4.4_4\endcsname{-0.011}
\expandafter\gdef\csname odunum@n@positive_taxonomy_m2.av.delta.qwen_plus.axis4.4_4\endcsname{324}
\expandafter\gdef\csname odunum@ci@positive_taxonomy_m2.av.delta.qwen_plus.axis4.4_4\endcsname{[-0.034, 0.013]}
\expandafter\gdef\csname odunum@val@positive_taxonomy_m2.av.delta.qwen_plus.axis4.4_5\endcsname{0.019}
\expandafter\gdef\csname odunum@n@positive_taxonomy_m2.av.delta.qwen_plus.axis4.4_5\endcsname{173}
\expandafter\gdef\csname odunum@ci@positive_taxonomy_m2.av.delta.qwen_plus.axis4.4_5\endcsname{[-0.015, 0.053]}
\expandafter\gdef\csname odunum@val@positive_taxonomy_m2.av.delta.salmonn2_7b.axis1.1_1\endcsname{-0.016}
\expandafter\gdef\csname odunum@n@positive_taxonomy_m2.av.delta.salmonn2_7b.axis1.1_1\endcsname{92}
\expandafter\gdef\csname odunum@ci@positive_taxonomy_m2.av.delta.salmonn2_7b.axis1.1_1\endcsname{[-0.042, 0.013]}
\expandafter\gdef\csname odunum@val@positive_taxonomy_m2.av.delta.salmonn2_7b.axis1.1_2\endcsname{0.013}
\expandafter\gdef\csname odunum@n@positive_taxonomy_m2.av.delta.salmonn2_7b.axis1.1_2\endcsname{288}
\expandafter\gdef\csname odunum@ci@positive_taxonomy_m2.av.delta.salmonn2_7b.axis1.1_2\endcsname{[-0.003, 0.030]}
\expandafter\gdef\csname odunum@val@positive_taxonomy_m2.av.delta.salmonn2_7b.axis1.1_3\endcsname{-0.011}
\expandafter\gdef\csname odunum@n@positive_taxonomy_m2.av.delta.salmonn2_7b.axis1.1_3\endcsname{286}
\expandafter\gdef\csname odunum@ci@positive_taxonomy_m2.av.delta.salmonn2_7b.axis1.1_3\endcsname{[-0.025, 0.004]}
\expandafter\gdef\csname odunum@val@positive_taxonomy_m2.av.delta.salmonn2_7b.axis1.1_4\endcsname{0.002}
\expandafter\gdef\csname odunum@n@positive_taxonomy_m2.av.delta.salmonn2_7b.axis1.1_4\endcsname{377}
\expandafter\gdef\csname odunum@ci@positive_taxonomy_m2.av.delta.salmonn2_7b.axis1.1_4\endcsname{[-0.011, 0.016]}
\expandafter\gdef\csname odunum@val@positive_taxonomy_m2.av.delta.salmonn2_7b.axis2.2_1\endcsname{0.009}
\expandafter\gdef\csname odunum@n@positive_taxonomy_m2.av.delta.salmonn2_7b.axis2.2_1\endcsname{118}
\expandafter\gdef\csname odunum@ci@positive_taxonomy_m2.av.delta.salmonn2_7b.axis2.2_1\endcsname{[-0.017, 0.036]}
\expandafter\gdef\csname odunum@val@positive_taxonomy_m2.av.delta.salmonn2_7b.axis2.2_2\endcsname{-0.020}
\expandafter\gdef\csname odunum@n@positive_taxonomy_m2.av.delta.salmonn2_7b.axis2.2_2\endcsname{266}
\expandafter\gdef\csname odunum@ci@positive_taxonomy_m2.av.delta.salmonn2_7b.axis2.2_2\endcsname{[-0.034, -0.004]}
\expandafter\gdef\csname odunum@val@positive_taxonomy_m2.av.delta.salmonn2_7b.axis2.2_3\endcsname{-0.034}
\expandafter\gdef\csname odunum@n@positive_taxonomy_m2.av.delta.salmonn2_7b.axis2.2_3\endcsname{274}
\expandafter\gdef\csname odunum@ci@positive_taxonomy_m2.av.delta.salmonn2_7b.axis2.2_3\endcsname{[-0.047, -0.021]}
\expandafter\gdef\csname odunum@val@positive_taxonomy_m2.av.delta.salmonn2_7b.axis2.2_4\endcsname{0.042}
\expandafter\gdef\csname odunum@n@positive_taxonomy_m2.av.delta.salmonn2_7b.axis2.2_4\endcsname{226}
\expandafter\gdef\csname odunum@ci@positive_taxonomy_m2.av.delta.salmonn2_7b.axis2.2_4\endcsname{[0.019, 0.066]}
\expandafter\gdef\csname odunum@val@positive_taxonomy_m2.av.delta.salmonn2_7b.axis2.2_5\endcsname{0.026}
\expandafter\gdef\csname odunum@n@positive_taxonomy_m2.av.delta.salmonn2_7b.axis2.2_5\endcsname{159}
\expandafter\gdef\csname odunum@ci@positive_taxonomy_m2.av.delta.salmonn2_7b.axis2.2_5\endcsname{[0.002, 0.052]}
\expandafter\gdef\csname odunum@val@positive_taxonomy_m2.av.delta.salmonn2_7b.axis3.3_1\endcsname{0.034}
\expandafter\gdef\csname odunum@n@positive_taxonomy_m2.av.delta.salmonn2_7b.axis3.3_1\endcsname{91}
\expandafter\gdef\csname odunum@ci@positive_taxonomy_m2.av.delta.salmonn2_7b.axis3.3_1\endcsname{[-0.004, 0.078]}
\expandafter\gdef\csname odunum@val@positive_taxonomy_m2.av.delta.salmonn2_7b.axis3.3_2\endcsname{-0.009}
\expandafter\gdef\csname odunum@n@positive_taxonomy_m2.av.delta.salmonn2_7b.axis3.3_2\endcsname{85}
\expandafter\gdef\csname odunum@ci@positive_taxonomy_m2.av.delta.salmonn2_7b.axis3.3_2\endcsname{[-0.037, 0.020]}
\expandafter\gdef\csname odunum@val@positive_taxonomy_m2.av.delta.salmonn2_7b.axis3.3_3\endcsname{-0.023}
\expandafter\gdef\csname odunum@n@positive_taxonomy_m2.av.delta.salmonn2_7b.axis3.3_3\endcsname{137}
\expandafter\gdef\csname odunum@ci@positive_taxonomy_m2.av.delta.salmonn2_7b.axis3.3_3\endcsname{[-0.044, 2.1\ensuremath{\times 10^{-4}}]}
\expandafter\gdef\csname odunum@val@positive_taxonomy_m2.av.delta.salmonn2_7b.axis3.3_4\endcsname{-0.024}
\expandafter\gdef\csname odunum@n@positive_taxonomy_m2.av.delta.salmonn2_7b.axis3.3_4\endcsname{256}
\expandafter\gdef\csname odunum@ci@positive_taxonomy_m2.av.delta.salmonn2_7b.axis3.3_4\endcsname{[-0.039, -0.008]}
\expandafter\gdef\csname odunum@val@positive_taxonomy_m2.av.delta.salmonn2_7b.axis3.3_5\endcsname{0.055}
\expandafter\gdef\csname odunum@n@positive_taxonomy_m2.av.delta.salmonn2_7b.axis3.3_5\endcsname{185}
\expandafter\gdef\csname odunum@ci@positive_taxonomy_m2.av.delta.salmonn2_7b.axis3.3_5\endcsname{[0.031, 0.081]}
\expandafter\gdef\csname odunum@val@positive_taxonomy_m2.av.delta.salmonn2_7b.axis3.3_6\endcsname{-0.011}
\expandafter\gdef\csname odunum@n@positive_taxonomy_m2.av.delta.salmonn2_7b.axis3.3_6\endcsname{289}
\expandafter\gdef\csname odunum@ci@positive_taxonomy_m2.av.delta.salmonn2_7b.axis3.3_6\endcsname{[-0.025, 0.003]}
\expandafter\gdef\csname odunum@val@positive_taxonomy_m2.av.delta.salmonn2_7b.axis4.4_1\endcsname{-0.034}
\expandafter\gdef\csname odunum@n@positive_taxonomy_m2.av.delta.salmonn2_7b.axis4.4_1\endcsname{137}
\expandafter\gdef\csname odunum@ci@positive_taxonomy_m2.av.delta.salmonn2_7b.axis4.4_1\endcsname{[-0.052, -0.015]}
\expandafter\gdef\csname odunum@val@positive_taxonomy_m2.av.delta.salmonn2_7b.axis4.4_2\endcsname{0.025}
\expandafter\gdef\csname odunum@n@positive_taxonomy_m2.av.delta.salmonn2_7b.axis4.4_2\endcsname{289}
\expandafter\gdef\csname odunum@ci@positive_taxonomy_m2.av.delta.salmonn2_7b.axis4.4_2\endcsname{[0.009, 0.043]}
\expandafter\gdef\csname odunum@val@positive_taxonomy_m2.av.delta.salmonn2_7b.axis4.4_3\endcsname{-0.009}
\expandafter\gdef\csname odunum@n@positive_taxonomy_m2.av.delta.salmonn2_7b.axis4.4_3\endcsname{120}
\expandafter\gdef\csname odunum@ci@positive_taxonomy_m2.av.delta.salmonn2_7b.axis4.4_3\endcsname{[-0.034, 0.018]}
\expandafter\gdef\csname odunum@val@positive_taxonomy_m2.av.delta.salmonn2_7b.axis4.4_4\endcsname{-0.009}
\expandafter\gdef\csname odunum@n@positive_taxonomy_m2.av.delta.salmonn2_7b.axis4.4_4\endcsname{324}
\expandafter\gdef\csname odunum@ci@positive_taxonomy_m2.av.delta.salmonn2_7b.axis4.4_4\endcsname{[-0.023, 0.006]}
\expandafter\gdef\csname odunum@val@positive_taxonomy_m2.av.delta.salmonn2_7b.axis4.4_5\endcsname{0.008}
\expandafter\gdef\csname odunum@n@positive_taxonomy_m2.av.delta.salmonn2_7b.axis4.4_5\endcsname{173}
\expandafter\gdef\csname odunum@ci@positive_taxonomy_m2.av.delta.salmonn2_7b.axis4.4_5\endcsname{[-0.015, 0.032]}
\expandafter\gdef\csname odunum@val@positive_taxonomy_m2.av.delta.seed.axis1.1_1\endcsname{0.001}
\expandafter\gdef\csname odunum@n@positive_taxonomy_m2.av.delta.seed.axis1.1_1\endcsname{92}
\expandafter\gdef\csname odunum@ci@positive_taxonomy_m2.av.delta.seed.axis1.1_1\endcsname{[-0.065, 0.065]}
\expandafter\gdef\csname odunum@val@positive_taxonomy_m2.av.delta.seed.axis1.1_2\endcsname{-0.008}
\expandafter\gdef\csname odunum@n@positive_taxonomy_m2.av.delta.seed.axis1.1_2\endcsname{288}
\expandafter\gdef\csname odunum@ci@positive_taxonomy_m2.av.delta.seed.axis1.1_2\endcsname{[-0.038, 0.022]}
\expandafter\gdef\csname odunum@val@positive_taxonomy_m2.av.delta.seed.axis1.1_3\endcsname{0.014}
\expandafter\gdef\csname odunum@n@positive_taxonomy_m2.av.delta.seed.axis1.1_3\endcsname{286}
\expandafter\gdef\csname odunum@ci@positive_taxonomy_m2.av.delta.seed.axis1.1_3\endcsname{[-0.018, 0.045]}
\expandafter\gdef\csname odunum@val@positive_taxonomy_m2.av.delta.seed.axis1.1_4\endcsname{-0.005}
\expandafter\gdef\csname odunum@n@positive_taxonomy_m2.av.delta.seed.axis1.1_4\endcsname{377}
\expandafter\gdef\csname odunum@ci@positive_taxonomy_m2.av.delta.seed.axis1.1_4\endcsname{[-0.029, 0.020]}
\expandafter\gdef\csname odunum@val@positive_taxonomy_m2.av.delta.seed.axis2.2_1\endcsname{-0.003}
\expandafter\gdef\csname odunum@n@positive_taxonomy_m2.av.delta.seed.axis2.2_1\endcsname{118}
\expandafter\gdef\csname odunum@ci@positive_taxonomy_m2.av.delta.seed.axis2.2_1\endcsname{[-0.059, 0.053]}
\expandafter\gdef\csname odunum@val@positive_taxonomy_m2.av.delta.seed.axis2.2_2\endcsname{0.004}
\expandafter\gdef\csname odunum@n@positive_taxonomy_m2.av.delta.seed.axis2.2_2\endcsname{266}
\expandafter\gdef\csname odunum@ci@positive_taxonomy_m2.av.delta.seed.axis2.2_2\endcsname{[-0.029, 0.035]}
\expandafter\gdef\csname odunum@val@positive_taxonomy_m2.av.delta.seed.axis2.2_3\endcsname{0.016}
\expandafter\gdef\csname odunum@n@positive_taxonomy_m2.av.delta.seed.axis2.2_3\endcsname{274}
\expandafter\gdef\csname odunum@ci@positive_taxonomy_m2.av.delta.seed.axis2.2_3\endcsname{[-0.014, 0.047]}
\expandafter\gdef\csname odunum@val@positive_taxonomy_m2.av.delta.seed.axis2.2_4\endcsname{-0.029}
\expandafter\gdef\csname odunum@n@positive_taxonomy_m2.av.delta.seed.axis2.2_4\endcsname{226}
\expandafter\gdef\csname odunum@ci@positive_taxonomy_m2.av.delta.seed.axis2.2_4\endcsname{[-0.068, 0.010]}
\expandafter\gdef\csname odunum@val@positive_taxonomy_m2.av.delta.seed.axis2.2_5\endcsname{0.009}
\expandafter\gdef\csname odunum@n@positive_taxonomy_m2.av.delta.seed.axis2.2_5\endcsname{159}
\expandafter\gdef\csname odunum@ci@positive_taxonomy_m2.av.delta.seed.axis2.2_5\endcsname{[-0.034, 0.053]}
\expandafter\gdef\csname odunum@val@positive_taxonomy_m2.av.delta.seed.axis3.3_1\endcsname{-0.009}
\expandafter\gdef\csname odunum@n@positive_taxonomy_m2.av.delta.seed.axis3.3_1\endcsname{91}
\expandafter\gdef\csname odunum@ci@positive_taxonomy_m2.av.delta.seed.axis3.3_1\endcsname{[-0.074, 0.057]}
\expandafter\gdef\csname odunum@val@positive_taxonomy_m2.av.delta.seed.axis3.3_2\endcsname{0.007}
\expandafter\gdef\csname odunum@n@positive_taxonomy_m2.av.delta.seed.axis3.3_2\endcsname{85}
\expandafter\gdef\csname odunum@ci@positive_taxonomy_m2.av.delta.seed.axis3.3_2\endcsname{[-0.059, 0.070]}
\expandafter\gdef\csname odunum@val@positive_taxonomy_m2.av.delta.seed.axis3.3_3\endcsname{-0.022}
\expandafter\gdef\csname odunum@n@positive_taxonomy_m2.av.delta.seed.axis3.3_3\endcsname{137}
\expandafter\gdef\csname odunum@ci@positive_taxonomy_m2.av.delta.seed.axis3.3_3\endcsname{[-0.070, 0.026]}
\expandafter\gdef\csname odunum@val@positive_taxonomy_m2.av.delta.seed.axis3.3_4\endcsname{-0.017}
\expandafter\gdef\csname odunum@n@positive_taxonomy_m2.av.delta.seed.axis3.3_4\endcsname{256}
\expandafter\gdef\csname odunum@ci@positive_taxonomy_m2.av.delta.seed.axis3.3_4\endcsname{[-0.051, 0.017]}
\expandafter\gdef\csname odunum@val@positive_taxonomy_m2.av.delta.seed.axis3.3_5\endcsname{-0.045}
\expandafter\gdef\csname odunum@n@positive_taxonomy_m2.av.delta.seed.axis3.3_5\endcsname{185}
\expandafter\gdef\csname odunum@ci@positive_taxonomy_m2.av.delta.seed.axis3.3_5\endcsname{[-0.084, -0.006]}
\expandafter\gdef\csname odunum@val@positive_taxonomy_m2.av.delta.seed.axis3.3_6\endcsname{0.056}
\expandafter\gdef\csname odunum@n@positive_taxonomy_m2.av.delta.seed.axis3.3_6\endcsname{289}
\expandafter\gdef\csname odunum@ci@positive_taxonomy_m2.av.delta.seed.axis3.3_6\endcsname{[0.026, 0.086]}
\expandafter\gdef\csname odunum@val@positive_taxonomy_m2.av.delta.seed.axis4.4_1\endcsname{-0.053}
\expandafter\gdef\csname odunum@n@positive_taxonomy_m2.av.delta.seed.axis4.4_1\endcsname{137}
\expandafter\gdef\csname odunum@ci@positive_taxonomy_m2.av.delta.seed.axis4.4_1\endcsname{[-0.103, -0.004]}
\expandafter\gdef\csname odunum@val@positive_taxonomy_m2.av.delta.seed.axis4.4_2\endcsname{0.024}
\expandafter\gdef\csname odunum@n@positive_taxonomy_m2.av.delta.seed.axis4.4_2\endcsname{289}
\expandafter\gdef\csname odunum@ci@positive_taxonomy_m2.av.delta.seed.axis4.4_2\endcsname{[-0.005, 0.054]}
\expandafter\gdef\csname odunum@val@positive_taxonomy_m2.av.delta.seed.axis4.4_3\endcsname{0.058}
\expandafter\gdef\csname odunum@n@positive_taxonomy_m2.av.delta.seed.axis4.4_3\endcsname{120}
\expandafter\gdef\csname odunum@ci@positive_taxonomy_m2.av.delta.seed.axis4.4_3\endcsname{[0.004, 0.113]}
\expandafter\gdef\csname odunum@val@positive_taxonomy_m2.av.delta.seed.axis4.4_4\endcsname{-0.025}
\expandafter\gdef\csname odunum@n@positive_taxonomy_m2.av.delta.seed.axis4.4_4\endcsname{324}
\expandafter\gdef\csname odunum@ci@positive_taxonomy_m2.av.delta.seed.axis4.4_4\endcsname{[-0.054, 0.004]}
\expandafter\gdef\csname odunum@val@positive_taxonomy_m2.av.delta.seed.axis4.4_5\endcsname{0.007}
\expandafter\gdef\csname odunum@n@positive_taxonomy_m2.av.delta.seed.axis4.4_5\endcsname{173}
\expandafter\gdef\csname odunum@ci@positive_taxonomy_m2.av.delta.seed.axis4.4_5\endcsname{[-0.037, 0.050]}
\expandafter\gdef\csname odunum@val@positive_taxonomy_m2.av.mean.cascade_asr.axis1.1_1\endcsname{0.670}
\expandafter\gdef\csname odunum@n@positive_taxonomy_m2.av.mean.cascade_asr.axis1.1_1\endcsname{92}
\expandafter\gdef\csname odunum@ci@positive_taxonomy_m2.av.mean.cascade_asr.axis1.1_1\endcsname{[0.615, 0.725]}
\expandafter\gdef\csname odunum@val@positive_taxonomy_m2.av.mean.cascade_asr.axis1.1_2\endcsname{0.510}
\expandafter\gdef\csname odunum@n@positive_taxonomy_m2.av.mean.cascade_asr.axis1.1_2\endcsname{288}
\expandafter\gdef\csname odunum@ci@positive_taxonomy_m2.av.mean.cascade_asr.axis1.1_2\endcsname{[0.477, 0.541]}
\expandafter\gdef\csname odunum@val@positive_taxonomy_m2.av.mean.cascade_asr.axis1.1_3\endcsname{0.577}
\expandafter\gdef\csname odunum@n@positive_taxonomy_m2.av.mean.cascade_asr.axis1.1_3\endcsname{286}
\expandafter\gdef\csname odunum@ci@positive_taxonomy_m2.av.mean.cascade_asr.axis1.1_3\endcsname{[0.543, 0.612]}
\expandafter\gdef\csname odunum@val@positive_taxonomy_m2.av.mean.cascade_asr.axis1.1_4\endcsname{0.542}
\expandafter\gdef\csname odunum@n@positive_taxonomy_m2.av.mean.cascade_asr.axis1.1_4\endcsname{377}
\expandafter\gdef\csname odunum@ci@positive_taxonomy_m2.av.mean.cascade_asr.axis1.1_4\endcsname{[0.514, 0.568]}
\expandafter\gdef\csname odunum@val@positive_taxonomy_m2.av.mean.cascade_asr.axis2.2_1\endcsname{0.419}
\expandafter\gdef\csname odunum@n@positive_taxonomy_m2.av.mean.cascade_asr.axis2.2_1\endcsname{118}
\expandafter\gdef\csname odunum@ci@positive_taxonomy_m2.av.mean.cascade_asr.axis2.2_1\endcsname{[0.366, 0.473]}
\expandafter\gdef\csname odunum@val@positive_taxonomy_m2.av.mean.cascade_asr.axis2.2_2\endcsname{0.626}
\expandafter\gdef\csname odunum@n@positive_taxonomy_m2.av.mean.cascade_asr.axis2.2_2\endcsname{266}
\expandafter\gdef\csname odunum@ci@positive_taxonomy_m2.av.mean.cascade_asr.axis2.2_2\endcsname{[0.596, 0.656]}
\expandafter\gdef\csname odunum@val@positive_taxonomy_m2.av.mean.cascade_asr.axis2.2_3\endcsname{0.646}
\expandafter\gdef\csname odunum@n@positive_taxonomy_m2.av.mean.cascade_asr.axis2.2_3\endcsname{274}
\expandafter\gdef\csname odunum@ci@positive_taxonomy_m2.av.mean.cascade_asr.axis2.2_3\endcsname{[0.616, 0.676]}
\expandafter\gdef\csname odunum@val@positive_taxonomy_m2.av.mean.cascade_asr.axis2.2_4\endcsname{0.487}
\expandafter\gdef\csname odunum@n@positive_taxonomy_m2.av.mean.cascade_asr.axis2.2_4\endcsname{226}
\expandafter\gdef\csname odunum@ci@positive_taxonomy_m2.av.mean.cascade_asr.axis2.2_4\endcsname{[0.449, 0.524]}
\expandafter\gdef\csname odunum@val@positive_taxonomy_m2.av.mean.cascade_asr.axis2.2_5\endcsname{0.470}
\expandafter\gdef\csname odunum@n@positive_taxonomy_m2.av.mean.cascade_asr.axis2.2_5\endcsname{159}
\expandafter\gdef\csname odunum@ci@positive_taxonomy_m2.av.mean.cascade_asr.axis2.2_5\endcsname{[0.423, 0.517]}
\expandafter\gdef\csname odunum@val@positive_taxonomy_m2.av.mean.cascade_asr.axis3.3_1\endcsname{0.558}
\expandafter\gdef\csname odunum@n@positive_taxonomy_m2.av.mean.cascade_asr.axis3.3_1\endcsname{91}
\expandafter\gdef\csname odunum@ci@positive_taxonomy_m2.av.mean.cascade_asr.axis3.3_1\endcsname{[0.501, 0.612]}
\expandafter\gdef\csname odunum@val@positive_taxonomy_m2.av.mean.cascade_asr.axis3.3_2\endcsname{0.553}
\expandafter\gdef\csname odunum@n@positive_taxonomy_m2.av.mean.cascade_asr.axis3.3_2\endcsname{85}
\expandafter\gdef\csname odunum@ci@positive_taxonomy_m2.av.mean.cascade_asr.axis3.3_2\endcsname{[0.493, 0.613]}
\expandafter\gdef\csname odunum@val@positive_taxonomy_m2.av.mean.cascade_asr.axis3.3_3\endcsname{0.430}
\expandafter\gdef\csname odunum@n@positive_taxonomy_m2.av.mean.cascade_asr.axis3.3_3\endcsname{137}
\expandafter\gdef\csname odunum@ci@positive_taxonomy_m2.av.mean.cascade_asr.axis3.3_3\endcsname{[0.376, 0.482]}
\expandafter\gdef\csname odunum@val@positive_taxonomy_m2.av.mean.cascade_asr.axis3.3_4\endcsname{0.596}
\expandafter\gdef\csname odunum@n@positive_taxonomy_m2.av.mean.cascade_asr.axis3.3_4\endcsname{256}
\expandafter\gdef\csname odunum@ci@positive_taxonomy_m2.av.mean.cascade_asr.axis3.3_4\endcsname{[0.561, 0.629]}
\expandafter\gdef\csname odunum@val@positive_taxonomy_m2.av.mean.cascade_asr.axis3.3_5\endcsname{0.467}
\expandafter\gdef\csname odunum@n@positive_taxonomy_m2.av.mean.cascade_asr.axis3.3_5\endcsname{185}
\expandafter\gdef\csname odunum@ci@positive_taxonomy_m2.av.mean.cascade_asr.axis3.3_5\endcsname{[0.424, 0.509]}
\expandafter\gdef\csname odunum@val@positive_taxonomy_m2.av.mean.cascade_asr.axis3.3_6\endcsname{0.630}
\expandafter\gdef\csname odunum@n@positive_taxonomy_m2.av.mean.cascade_asr.axis3.3_6\endcsname{289}
\expandafter\gdef\csname odunum@ci@positive_taxonomy_m2.av.mean.cascade_asr.axis3.3_6\endcsname{[0.603, 0.657]}
\expandafter\gdef\csname odunum@val@positive_taxonomy_m2.av.mean.cascade_asr.axis4.4_1\endcsname{0.580}
\expandafter\gdef\csname odunum@n@positive_taxonomy_m2.av.mean.cascade_asr.axis4.4_1\endcsname{137}
\expandafter\gdef\csname odunum@ci@positive_taxonomy_m2.av.mean.cascade_asr.axis4.4_1\endcsname{[0.534, 0.627]}
\expandafter\gdef\csname odunum@val@positive_taxonomy_m2.av.mean.cascade_asr.axis4.4_2\endcsname{0.562}
\expandafter\gdef\csname odunum@n@positive_taxonomy_m2.av.mean.cascade_asr.axis4.4_2\endcsname{289}
\expandafter\gdef\csname odunum@ci@positive_taxonomy_m2.av.mean.cascade_asr.axis4.4_2\endcsname{[0.530, 0.594]}
\expandafter\gdef\csname odunum@val@positive_taxonomy_m2.av.mean.cascade_asr.axis4.4_3\endcsname{0.499}
\expandafter\gdef\csname odunum@n@positive_taxonomy_m2.av.mean.cascade_asr.axis4.4_3\endcsname{120}
\expandafter\gdef\csname odunum@ci@positive_taxonomy_m2.av.mean.cascade_asr.axis4.4_3\endcsname{[0.440, 0.556]}
\expandafter\gdef\csname odunum@val@positive_taxonomy_m2.av.mean.cascade_asr.axis4.4_4\endcsname{0.538}
\expandafter\gdef\csname odunum@n@positive_taxonomy_m2.av.mean.cascade_asr.axis4.4_4\endcsname{324}
\expandafter\gdef\csname odunum@ci@positive_taxonomy_m2.av.mean.cascade_asr.axis4.4_4\endcsname{[0.507, 0.569]}
\expandafter\gdef\csname odunum@val@positive_taxonomy_m2.av.mean.cascade_asr.axis4.4_5\endcsname{0.587}
\expandafter\gdef\csname odunum@n@positive_taxonomy_m2.av.mean.cascade_asr.axis4.4_5\endcsname{173}
\expandafter\gdef\csname odunum@ci@positive_taxonomy_m2.av.mean.cascade_asr.axis4.4_5\endcsname{[0.547, 0.629]}
\expandafter\gdef\csname odunum@val@positive_taxonomy_m2.av.mean.gemini.axis1.1_1\endcsname{0.632}
\expandafter\gdef\csname odunum@n@positive_taxonomy_m2.av.mean.gemini.axis1.1_1\endcsname{92}
\expandafter\gdef\csname odunum@ci@positive_taxonomy_m2.av.mean.gemini.axis1.1_1\endcsname{[0.579, 0.686]}
\expandafter\gdef\csname odunum@val@positive_taxonomy_m2.av.mean.gemini.axis1.1_2\endcsname{0.613}
\expandafter\gdef\csname odunum@n@positive_taxonomy_m2.av.mean.gemini.axis1.1_2\endcsname{288}
\expandafter\gdef\csname odunum@ci@positive_taxonomy_m2.av.mean.gemini.axis1.1_2\endcsname{[0.583, 0.642]}
\expandafter\gdef\csname odunum@val@positive_taxonomy_m2.av.mean.gemini.axis1.1_3\endcsname{0.654}
\expandafter\gdef\csname odunum@n@positive_taxonomy_m2.av.mean.gemini.axis1.1_3\endcsname{286}
\expandafter\gdef\csname odunum@ci@positive_taxonomy_m2.av.mean.gemini.axis1.1_3\endcsname{[0.626, 0.683]}
\expandafter\gdef\csname odunum@val@positive_taxonomy_m2.av.mean.gemini.axis1.1_4\endcsname{0.636}
\expandafter\gdef\csname odunum@n@positive_taxonomy_m2.av.mean.gemini.axis1.1_4\endcsname{377}
\expandafter\gdef\csname odunum@ci@positive_taxonomy_m2.av.mean.gemini.axis1.1_4\endcsname{[0.612, 0.660]}
\expandafter\gdef\csname odunum@val@positive_taxonomy_m2.av.mean.gemini.axis2.2_1\endcsname{0.621}
\expandafter\gdef\csname odunum@n@positive_taxonomy_m2.av.mean.gemini.axis2.2_1\endcsname{118}
\expandafter\gdef\csname odunum@ci@positive_taxonomy_m2.av.mean.gemini.axis2.2_1\endcsname{[0.568, 0.674]}
\expandafter\gdef\csname odunum@val@positive_taxonomy_m2.av.mean.gemini.axis2.2_2\endcsname{0.627}
\expandafter\gdef\csname odunum@n@positive_taxonomy_m2.av.mean.gemini.axis2.2_2\endcsname{266}
\expandafter\gdef\csname odunum@ci@positive_taxonomy_m2.av.mean.gemini.axis2.2_2\endcsname{[0.599, 0.654]}
\expandafter\gdef\csname odunum@val@positive_taxonomy_m2.av.mean.gemini.axis2.2_3\endcsname{0.638}
\expandafter\gdef\csname odunum@n@positive_taxonomy_m2.av.mean.gemini.axis2.2_3\endcsname{274}
\expandafter\gdef\csname odunum@ci@positive_taxonomy_m2.av.mean.gemini.axis2.2_3\endcsname{[0.610, 0.667]}
\expandafter\gdef\csname odunum@val@positive_taxonomy_m2.av.mean.gemini.axis2.2_4\endcsname{0.646}
\expandafter\gdef\csname odunum@n@positive_taxonomy_m2.av.mean.gemini.axis2.2_4\endcsname{226}
\expandafter\gdef\csname odunum@ci@positive_taxonomy_m2.av.mean.gemini.axis2.2_4\endcsname{[0.612, 0.678]}
\expandafter\gdef\csname odunum@val@positive_taxonomy_m2.av.mean.gemini.axis2.2_5\endcsname{0.634}
\expandafter\gdef\csname odunum@n@positive_taxonomy_m2.av.mean.gemini.axis2.2_5\endcsname{159}
\expandafter\gdef\csname odunum@ci@positive_taxonomy_m2.av.mean.gemini.axis2.2_5\endcsname{[0.598, 0.670]}
\expandafter\gdef\csname odunum@val@positive_taxonomy_m2.av.mean.gemini.axis3.3_1\endcsname{0.677}
\expandafter\gdef\csname odunum@n@positive_taxonomy_m2.av.mean.gemini.axis3.3_1\endcsname{91}
\expandafter\gdef\csname odunum@ci@positive_taxonomy_m2.av.mean.gemini.axis3.3_1\endcsname{[0.630, 0.724]}
\expandafter\gdef\csname odunum@val@positive_taxonomy_m2.av.mean.gemini.axis3.3_2\endcsname{0.609}
\expandafter\gdef\csname odunum@n@positive_taxonomy_m2.av.mean.gemini.axis3.3_2\endcsname{85}
\expandafter\gdef\csname odunum@ci@positive_taxonomy_m2.av.mean.gemini.axis3.3_2\endcsname{[0.563, 0.655]}
\expandafter\gdef\csname odunum@val@positive_taxonomy_m2.av.mean.gemini.axis3.3_3\endcsname{0.561}
\expandafter\gdef\csname odunum@n@positive_taxonomy_m2.av.mean.gemini.axis3.3_3\endcsname{137}
\expandafter\gdef\csname odunum@ci@positive_taxonomy_m2.av.mean.gemini.axis3.3_3\endcsname{[0.512, 0.611]}
\expandafter\gdef\csname odunum@val@positive_taxonomy_m2.av.mean.gemini.axis3.3_4\endcsname{0.652}
\expandafter\gdef\csname odunum@n@positive_taxonomy_m2.av.mean.gemini.axis3.3_4\endcsname{256}
\expandafter\gdef\csname odunum@ci@positive_taxonomy_m2.av.mean.gemini.axis3.3_4\endcsname{[0.625, 0.679]}
\expandafter\gdef\csname odunum@val@positive_taxonomy_m2.av.mean.gemini.axis3.3_5\endcsname{0.559}
\expandafter\gdef\csname odunum@n@positive_taxonomy_m2.av.mean.gemini.axis3.3_5\endcsname{185}
\expandafter\gdef\csname odunum@ci@positive_taxonomy_m2.av.mean.gemini.axis3.3_5\endcsname{[0.521, 0.596]}
\expandafter\gdef\csname odunum@val@positive_taxonomy_m2.av.mean.gemini.axis3.3_6\endcsname{0.696}
\expandafter\gdef\csname odunum@n@positive_taxonomy_m2.av.mean.gemini.axis3.3_6\endcsname{289}
\expandafter\gdef\csname odunum@ci@positive_taxonomy_m2.av.mean.gemini.axis3.3_6\endcsname{[0.670, 0.721]}
\expandafter\gdef\csname odunum@val@positive_taxonomy_m2.av.mean.gemini.axis4.4_1\endcsname{0.621}
\expandafter\gdef\csname odunum@n@positive_taxonomy_m2.av.mean.gemini.axis4.4_1\endcsname{137}
\expandafter\gdef\csname odunum@ci@positive_taxonomy_m2.av.mean.gemini.axis4.4_1\endcsname{[0.579, 0.662]}
\expandafter\gdef\csname odunum@val@positive_taxonomy_m2.av.mean.gemini.axis4.4_2\endcsname{0.630}
\expandafter\gdef\csname odunum@n@positive_taxonomy_m2.av.mean.gemini.axis4.4_2\endcsname{289}
\expandafter\gdef\csname odunum@ci@positive_taxonomy_m2.av.mean.gemini.axis4.4_2\endcsname{[0.603, 0.657]}
\expandafter\gdef\csname odunum@val@positive_taxonomy_m2.av.mean.gemini.axis4.4_3\endcsname{0.640}
\expandafter\gdef\csname odunum@n@positive_taxonomy_m2.av.mean.gemini.axis4.4_3\endcsname{120}
\expandafter\gdef\csname odunum@ci@positive_taxonomy_m2.av.mean.gemini.axis4.4_3\endcsname{[0.592, 0.687]}
\expandafter\gdef\csname odunum@val@positive_taxonomy_m2.av.mean.gemini.axis4.4_4\endcsname{0.639}
\expandafter\gdef\csname odunum@n@positive_taxonomy_m2.av.mean.gemini.axis4.4_4\endcsname{324}
\expandafter\gdef\csname odunum@ci@positive_taxonomy_m2.av.mean.gemini.axis4.4_4\endcsname{[0.612, 0.666]}
\expandafter\gdef\csname odunum@val@positive_taxonomy_m2.av.mean.gemini.axis4.4_5\endcsname{0.638}
\expandafter\gdef\csname odunum@n@positive_taxonomy_m2.av.mean.gemini.axis4.4_5\endcsname{173}
\expandafter\gdef\csname odunum@ci@positive_taxonomy_m2.av.mean.gemini.axis4.4_5\endcsname{[0.603, 0.673]}
\expandafter\gdef\csname odunum@val@positive_taxonomy_m2.av.mean.gemini35_flash_lite.axis1.1_1\endcsname{0.508}
\expandafter\gdef\csname odunum@n@positive_taxonomy_m2.av.mean.gemini35_flash_lite.axis1.1_1\endcsname{92}
\expandafter\gdef\csname odunum@ci@positive_taxonomy_m2.av.mean.gemini35_flash_lite.axis1.1_1\endcsname{[0.444, 0.573]}
\expandafter\gdef\csname odunum@val@positive_taxonomy_m2.av.mean.gemini35_flash_lite.axis1.1_2\endcsname{0.485}
\expandafter\gdef\csname odunum@n@positive_taxonomy_m2.av.mean.gemini35_flash_lite.axis1.1_2\endcsname{288}
\expandafter\gdef\csname odunum@ci@positive_taxonomy_m2.av.mean.gemini35_flash_lite.axis1.1_2\endcsname{[0.447, 0.523]}
\expandafter\gdef\csname odunum@val@positive_taxonomy_m2.av.mean.gemini35_flash_lite.axis1.1_3\endcsname{0.507}
\expandafter\gdef\csname odunum@n@positive_taxonomy_m2.av.mean.gemini35_flash_lite.axis1.1_3\endcsname{286}
\expandafter\gdef\csname odunum@ci@positive_taxonomy_m2.av.mean.gemini35_flash_lite.axis1.1_3\endcsname{[0.469, 0.546]}
\expandafter\gdef\csname odunum@val@positive_taxonomy_m2.av.mean.gemini35_flash_lite.axis1.1_4\endcsname{0.503}
\expandafter\gdef\csname odunum@n@positive_taxonomy_m2.av.mean.gemini35_flash_lite.axis1.1_4\endcsname{377}
\expandafter\gdef\csname odunum@ci@positive_taxonomy_m2.av.mean.gemini35_flash_lite.axis1.1_4\endcsname{[0.470, 0.534]}
\expandafter\gdef\csname odunum@val@positive_taxonomy_m2.av.mean.gemini35_flash_lite.axis2.2_1\endcsname{0.417}
\expandafter\gdef\csname odunum@n@positive_taxonomy_m2.av.mean.gemini35_flash_lite.axis2.2_1\endcsname{118}
\expandafter\gdef\csname odunum@ci@positive_taxonomy_m2.av.mean.gemini35_flash_lite.axis2.2_1\endcsname{[0.349, 0.484]}
\expandafter\gdef\csname odunum@val@positive_taxonomy_m2.av.mean.gemini35_flash_lite.axis2.2_2\endcsname{0.552}
\expandafter\gdef\csname odunum@n@positive_taxonomy_m2.av.mean.gemini35_flash_lite.axis2.2_2\endcsname{266}
\expandafter\gdef\csname odunum@ci@positive_taxonomy_m2.av.mean.gemini35_flash_lite.axis2.2_2\endcsname{[0.517, 0.587]}
\expandafter\gdef\csname odunum@val@positive_taxonomy_m2.av.mean.gemini35_flash_lite.axis2.2_3\endcsname{0.596}
\expandafter\gdef\csname odunum@n@positive_taxonomy_m2.av.mean.gemini35_flash_lite.axis2.2_3\endcsname{274}
\expandafter\gdef\csname odunum@ci@positive_taxonomy_m2.av.mean.gemini35_flash_lite.axis2.2_3\endcsname{[0.561, 0.629]}
\expandafter\gdef\csname odunum@val@positive_taxonomy_m2.av.mean.gemini35_flash_lite.axis2.2_4\endcsname{0.407}
\expandafter\gdef\csname odunum@n@positive_taxonomy_m2.av.mean.gemini35_flash_lite.axis2.2_4\endcsname{226}
\expandafter\gdef\csname odunum@ci@positive_taxonomy_m2.av.mean.gemini35_flash_lite.axis2.2_4\endcsname{[0.364, 0.450]}
\expandafter\gdef\csname odunum@val@positive_taxonomy_m2.av.mean.gemini35_flash_lite.axis2.2_5\endcsname{0.440}
\expandafter\gdef\csname odunum@n@positive_taxonomy_m2.av.mean.gemini35_flash_lite.axis2.2_5\endcsname{159}
\expandafter\gdef\csname odunum@ci@positive_taxonomy_m2.av.mean.gemini35_flash_lite.axis2.2_5\endcsname{[0.388, 0.490]}
\expandafter\gdef\csname odunum@val@positive_taxonomy_m2.av.mean.gemini35_flash_lite.axis3.3_1\endcsname{0.524}
\expandafter\gdef\csname odunum@n@positive_taxonomy_m2.av.mean.gemini35_flash_lite.axis3.3_1\endcsname{91}
\expandafter\gdef\csname odunum@ci@positive_taxonomy_m2.av.mean.gemini35_flash_lite.axis3.3_1\endcsname{[0.466, 0.582]}
\expandafter\gdef\csname odunum@val@positive_taxonomy_m2.av.mean.gemini35_flash_lite.axis3.3_2\endcsname{0.533}
\expandafter\gdef\csname odunum@n@positive_taxonomy_m2.av.mean.gemini35_flash_lite.axis3.3_2\endcsname{85}
\expandafter\gdef\csname odunum@ci@positive_taxonomy_m2.av.mean.gemini35_flash_lite.axis3.3_2\endcsname{[0.469, 0.595]}
\expandafter\gdef\csname odunum@val@positive_taxonomy_m2.av.mean.gemini35_flash_lite.axis3.3_3\endcsname{0.330}
\expandafter\gdef\csname odunum@n@positive_taxonomy_m2.av.mean.gemini35_flash_lite.axis3.3_3\endcsname{137}
\expandafter\gdef\csname odunum@ci@positive_taxonomy_m2.av.mean.gemini35_flash_lite.axis3.3_3\endcsname{[0.272, 0.391]}
\expandafter\gdef\csname odunum@val@positive_taxonomy_m2.av.mean.gemini35_flash_lite.axis3.3_4\endcsname{0.498}
\expandafter\gdef\csname odunum@n@positive_taxonomy_m2.av.mean.gemini35_flash_lite.axis3.3_4\endcsname{256}
\expandafter\gdef\csname odunum@ci@positive_taxonomy_m2.av.mean.gemini35_flash_lite.axis3.3_4\endcsname{[0.459, 0.537]}
\expandafter\gdef\csname odunum@val@positive_taxonomy_m2.av.mean.gemini35_flash_lite.axis3.3_5\endcsname{0.491}
\expandafter\gdef\csname odunum@n@positive_taxonomy_m2.av.mean.gemini35_flash_lite.axis3.3_5\endcsname{185}
\expandafter\gdef\csname odunum@ci@positive_taxonomy_m2.av.mean.gemini35_flash_lite.axis3.3_5\endcsname{[0.445, 0.538]}
\expandafter\gdef\csname odunum@val@positive_taxonomy_m2.av.mean.gemini35_flash_lite.axis3.3_6\endcsname{0.569}
\expandafter\gdef\csname odunum@n@positive_taxonomy_m2.av.mean.gemini35_flash_lite.axis3.3_6\endcsname{289}
\expandafter\gdef\csname odunum@ci@positive_taxonomy_m2.av.mean.gemini35_flash_lite.axis3.3_6\endcsname{[0.535, 0.604]}
\expandafter\gdef\csname odunum@val@positive_taxonomy_m2.av.mean.gemini35_flash_lite.axis4.4_1\endcsname{0.524}
\expandafter\gdef\csname odunum@n@positive_taxonomy_m2.av.mean.gemini35_flash_lite.axis4.4_1\endcsname{137}
\expandafter\gdef\csname odunum@ci@positive_taxonomy_m2.av.mean.gemini35_flash_lite.axis4.4_1\endcsname{[0.474, 0.572]}
\expandafter\gdef\csname odunum@val@positive_taxonomy_m2.av.mean.gemini35_flash_lite.axis4.4_2\endcsname{0.535}
\expandafter\gdef\csname odunum@n@positive_taxonomy_m2.av.mean.gemini35_flash_lite.axis4.4_2\endcsname{289}
\expandafter\gdef\csname odunum@ci@positive_taxonomy_m2.av.mean.gemini35_flash_lite.axis4.4_2\endcsname{[0.498, 0.571]}
\expandafter\gdef\csname odunum@val@positive_taxonomy_m2.av.mean.gemini35_flash_lite.axis4.4_3\endcsname{0.454}
\expandafter\gdef\csname odunum@n@positive_taxonomy_m2.av.mean.gemini35_flash_lite.axis4.4_3\endcsname{120}
\expandafter\gdef\csname odunum@ci@positive_taxonomy_m2.av.mean.gemini35_flash_lite.axis4.4_3\endcsname{[0.394, 0.515]}
\expandafter\gdef\csname odunum@val@positive_taxonomy_m2.av.mean.gemini35_flash_lite.axis4.4_4\endcsname{0.464}
\expandafter\gdef\csname odunum@n@positive_taxonomy_m2.av.mean.gemini35_flash_lite.axis4.4_4\endcsname{324}
\expandafter\gdef\csname odunum@ci@positive_taxonomy_m2.av.mean.gemini35_flash_lite.axis4.4_4\endcsname{[0.426, 0.501]}
\expandafter\gdef\csname odunum@val@positive_taxonomy_m2.av.mean.gemini35_flash_lite.axis4.4_5\endcsname{0.519}
\expandafter\gdef\csname odunum@n@positive_taxonomy_m2.av.mean.gemini35_flash_lite.axis4.4_5\endcsname{173}
\expandafter\gdef\csname odunum@ci@positive_taxonomy_m2.av.mean.gemini35_flash_lite.axis4.4_5\endcsname{[0.473, 0.564]}
\expandafter\gdef\csname odunum@val@positive_taxonomy_m2.av.mean.gemini37_flash.axis1.1_1\endcsname{0.622}
\expandafter\gdef\csname odunum@n@positive_taxonomy_m2.av.mean.gemini37_flash.axis1.1_1\endcsname{92}
\expandafter\gdef\csname odunum@ci@positive_taxonomy_m2.av.mean.gemini37_flash.axis1.1_1\endcsname{[0.564, 0.679]}
\expandafter\gdef\csname odunum@val@positive_taxonomy_m2.av.mean.gemini37_flash.axis1.1_2\endcsname{0.596}
\expandafter\gdef\csname odunum@n@positive_taxonomy_m2.av.mean.gemini37_flash.axis1.1_2\endcsname{288}
\expandafter\gdef\csname odunum@ci@positive_taxonomy_m2.av.mean.gemini37_flash.axis1.1_2\endcsname{[0.563, 0.629]}
\expandafter\gdef\csname odunum@val@positive_taxonomy_m2.av.mean.gemini37_flash.axis1.1_3\endcsname{0.612}
\expandafter\gdef\csname odunum@n@positive_taxonomy_m2.av.mean.gemini37_flash.axis1.1_3\endcsname{286}
\expandafter\gdef\csname odunum@ci@positive_taxonomy_m2.av.mean.gemini37_flash.axis1.1_3\endcsname{[0.579, 0.645]}
\expandafter\gdef\csname odunum@val@positive_taxonomy_m2.av.mean.gemini37_flash.axis1.1_4\endcsname{0.593}
\expandafter\gdef\csname odunum@n@positive_taxonomy_m2.av.mean.gemini37_flash.axis1.1_4\endcsname{377}
\expandafter\gdef\csname odunum@ci@positive_taxonomy_m2.av.mean.gemini37_flash.axis1.1_4\endcsname{[0.565, 0.620]}
\expandafter\gdef\csname odunum@val@positive_taxonomy_m2.av.mean.gemini37_flash.axis2.2_1\endcsname{0.567}
\expandafter\gdef\csname odunum@n@positive_taxonomy_m2.av.mean.gemini37_flash.axis2.2_1\endcsname{118}
\expandafter\gdef\csname odunum@ci@positive_taxonomy_m2.av.mean.gemini37_flash.axis2.2_1\endcsname{[0.506, 0.627]}
\expandafter\gdef\csname odunum@val@positive_taxonomy_m2.av.mean.gemini37_flash.axis2.2_2\endcsname{0.623}
\expandafter\gdef\csname odunum@n@positive_taxonomy_m2.av.mean.gemini37_flash.axis2.2_2\endcsname{266}
\expandafter\gdef\csname odunum@ci@positive_taxonomy_m2.av.mean.gemini37_flash.axis2.2_2\endcsname{[0.592, 0.653]}
\expandafter\gdef\csname odunum@val@positive_taxonomy_m2.av.mean.gemini37_flash.axis2.2_3\endcsname{0.639}
\expandafter\gdef\csname odunum@n@positive_taxonomy_m2.av.mean.gemini37_flash.axis2.2_3\endcsname{274}
\expandafter\gdef\csname odunum@ci@positive_taxonomy_m2.av.mean.gemini37_flash.axis2.2_3\endcsname{[0.608, 0.668]}
\expandafter\gdef\csname odunum@val@positive_taxonomy_m2.av.mean.gemini37_flash.axis2.2_4\endcsname{0.582}
\expandafter\gdef\csname odunum@n@positive_taxonomy_m2.av.mean.gemini37_flash.axis2.2_4\endcsname{226}
\expandafter\gdef\csname odunum@ci@positive_taxonomy_m2.av.mean.gemini37_flash.axis2.2_4\endcsname{[0.542, 0.621]}
\expandafter\gdef\csname odunum@val@positive_taxonomy_m2.av.mean.gemini37_flash.axis2.2_5\endcsname{0.554}
\expandafter\gdef\csname odunum@n@positive_taxonomy_m2.av.mean.gemini37_flash.axis2.2_5\endcsname{159}
\expandafter\gdef\csname odunum@ci@positive_taxonomy_m2.av.mean.gemini37_flash.axis2.2_5\endcsname{[0.509, 0.599]}
\expandafter\gdef\csname odunum@val@positive_taxonomy_m2.av.mean.gemini37_flash.axis3.3_1\endcsname{0.635}
\expandafter\gdef\csname odunum@n@positive_taxonomy_m2.av.mean.gemini37_flash.axis3.3_1\endcsname{91}
\expandafter\gdef\csname odunum@ci@positive_taxonomy_m2.av.mean.gemini37_flash.axis3.3_1\endcsname{[0.580, 0.690]}
\expandafter\gdef\csname odunum@val@positive_taxonomy_m2.av.mean.gemini37_flash.axis3.3_2\endcsname{0.618}
\expandafter\gdef\csname odunum@n@positive_taxonomy_m2.av.mean.gemini37_flash.axis3.3_2\endcsname{85}
\expandafter\gdef\csname odunum@ci@positive_taxonomy_m2.av.mean.gemini37_flash.axis3.3_2\endcsname{[0.560, 0.675]}
\expandafter\gdef\csname odunum@val@positive_taxonomy_m2.av.mean.gemini37_flash.axis3.3_3\endcsname{0.508}
\expandafter\gdef\csname odunum@n@positive_taxonomy_m2.av.mean.gemini37_flash.axis3.3_3\endcsname{137}
\expandafter\gdef\csname odunum@ci@positive_taxonomy_m2.av.mean.gemini37_flash.axis3.3_3\endcsname{[0.456, 0.561]}
\expandafter\gdef\csname odunum@val@positive_taxonomy_m2.av.mean.gemini37_flash.axis3.3_4\endcsname{0.627}
\expandafter\gdef\csname odunum@n@positive_taxonomy_m2.av.mean.gemini37_flash.axis3.3_4\endcsname{256}
\expandafter\gdef\csname odunum@ci@positive_taxonomy_m2.av.mean.gemini37_flash.axis3.3_4\endcsname{[0.592, 0.660]}
\expandafter\gdef\csname odunum@val@positive_taxonomy_m2.av.mean.gemini37_flash.axis3.3_5\endcsname{0.508}
\expandafter\gdef\csname odunum@n@positive_taxonomy_m2.av.mean.gemini37_flash.axis3.3_5\endcsname{185}
\expandafter\gdef\csname odunum@ci@positive_taxonomy_m2.av.mean.gemini37_flash.axis3.3_5\endcsname{[0.467, 0.549]}
\expandafter\gdef\csname odunum@val@positive_taxonomy_m2.av.mean.gemini37_flash.axis3.3_6\endcsname{0.667}
\expandafter\gdef\csname odunum@n@positive_taxonomy_m2.av.mean.gemini37_flash.axis3.3_6\endcsname{289}
\expandafter\gdef\csname odunum@ci@positive_taxonomy_m2.av.mean.gemini37_flash.axis3.3_6\endcsname{[0.638, 0.695]}
\expandafter\gdef\csname odunum@val@positive_taxonomy_m2.av.mean.gemini37_flash.axis4.4_1\endcsname{0.611}
\expandafter\gdef\csname odunum@n@positive_taxonomy_m2.av.mean.gemini37_flash.axis4.4_1\endcsname{137}
\expandafter\gdef\csname odunum@ci@positive_taxonomy_m2.av.mean.gemini37_flash.axis4.4_1\endcsname{[0.568, 0.655]}
\expandafter\gdef\csname odunum@val@positive_taxonomy_m2.av.mean.gemini37_flash.axis4.4_2\endcsname{0.590}
\expandafter\gdef\csname odunum@n@positive_taxonomy_m2.av.mean.gemini37_flash.axis4.4_2\endcsname{289}
\expandafter\gdef\csname odunum@ci@positive_taxonomy_m2.av.mean.gemini37_flash.axis4.4_2\endcsname{[0.558, 0.621]}
\expandafter\gdef\csname odunum@val@positive_taxonomy_m2.av.mean.gemini37_flash.axis4.4_3\endcsname{0.588}
\expandafter\gdef\csname odunum@n@positive_taxonomy_m2.av.mean.gemini37_flash.axis4.4_3\endcsname{120}
\expandafter\gdef\csname odunum@ci@positive_taxonomy_m2.av.mean.gemini37_flash.axis4.4_3\endcsname{[0.535, 0.640]}
\expandafter\gdef\csname odunum@val@positive_taxonomy_m2.av.mean.gemini37_flash.axis4.4_4\endcsname{0.590}
\expandafter\gdef\csname odunum@n@positive_taxonomy_m2.av.mean.gemini37_flash.axis4.4_4\endcsname{324}
\expandafter\gdef\csname odunum@ci@positive_taxonomy_m2.av.mean.gemini37_flash.axis4.4_4\endcsname{[0.560, 0.621]}
\expandafter\gdef\csname odunum@val@positive_taxonomy_m2.av.mean.gemini37_flash.axis4.4_5\endcsname{0.643}
\expandafter\gdef\csname odunum@n@positive_taxonomy_m2.av.mean.gemini37_flash.axis4.4_5\endcsname{173}
\expandafter\gdef\csname odunum@ci@positive_taxonomy_m2.av.mean.gemini37_flash.axis4.4_5\endcsname{[0.600, 0.686]}
\expandafter\gdef\csname odunum@val@positive_taxonomy_m2.av.mean.ming.axis1.1_1\endcsname{0.562}
\expandafter\gdef\csname odunum@n@positive_taxonomy_m2.av.mean.ming.axis1.1_1\endcsname{92}
\expandafter\gdef\csname odunum@ci@positive_taxonomy_m2.av.mean.ming.axis1.1_1\endcsname{[0.506, 0.618]}
\expandafter\gdef\csname odunum@val@positive_taxonomy_m2.av.mean.ming.axis1.1_2\endcsname{0.493}
\expandafter\gdef\csname odunum@n@positive_taxonomy_m2.av.mean.ming.axis1.1_2\endcsname{288}
\expandafter\gdef\csname odunum@ci@positive_taxonomy_m2.av.mean.ming.axis1.1_2\endcsname{[0.463, 0.523]}
\expandafter\gdef\csname odunum@val@positive_taxonomy_m2.av.mean.ming.axis1.1_3\endcsname{0.489}
\expandafter\gdef\csname odunum@n@positive_taxonomy_m2.av.mean.ming.axis1.1_3\endcsname{286}
\expandafter\gdef\csname odunum@ci@positive_taxonomy_m2.av.mean.ming.axis1.1_3\endcsname{[0.457, 0.521]}
\expandafter\gdef\csname odunum@val@positive_taxonomy_m2.av.mean.ming.axis1.1_4\endcsname{0.486}
\expandafter\gdef\csname odunum@n@positive_taxonomy_m2.av.mean.ming.axis1.1_4\endcsname{377}
\expandafter\gdef\csname odunum@ci@positive_taxonomy_m2.av.mean.ming.axis1.1_4\endcsname{[0.460, 0.512]}
\expandafter\gdef\csname odunum@val@positive_taxonomy_m2.av.mean.ming.axis2.2_1\endcsname{0.499}
\expandafter\gdef\csname odunum@n@positive_taxonomy_m2.av.mean.ming.axis2.2_1\endcsname{118}
\expandafter\gdef\csname odunum@ci@positive_taxonomy_m2.av.mean.ming.axis2.2_1\endcsname{[0.447, 0.552]}
\expandafter\gdef\csname odunum@val@positive_taxonomy_m2.av.mean.ming.axis2.2_2\endcsname{0.528}
\expandafter\gdef\csname odunum@n@positive_taxonomy_m2.av.mean.ming.axis2.2_2\endcsname{266}
\expandafter\gdef\csname odunum@ci@positive_taxonomy_m2.av.mean.ming.axis2.2_2\endcsname{[0.499, 0.558]}
\expandafter\gdef\csname odunum@val@positive_taxonomy_m2.av.mean.ming.axis2.2_3\endcsname{0.495}
\expandafter\gdef\csname odunum@n@positive_taxonomy_m2.av.mean.ming.axis2.2_3\endcsname{274}
\expandafter\gdef\csname odunum@ci@positive_taxonomy_m2.av.mean.ming.axis2.2_3\endcsname{[0.463, 0.526]}
\expandafter\gdef\csname odunum@val@positive_taxonomy_m2.av.mean.ming.axis2.2_4\endcsname{0.474}
\expandafter\gdef\csname odunum@n@positive_taxonomy_m2.av.mean.ming.axis2.2_4\endcsname{226}
\expandafter\gdef\csname odunum@ci@positive_taxonomy_m2.av.mean.ming.axis2.2_4\endcsname{[0.439, 0.508]}
\expandafter\gdef\csname odunum@val@positive_taxonomy_m2.av.mean.ming.axis2.2_5\endcsname{0.470}
\expandafter\gdef\csname odunum@n@positive_taxonomy_m2.av.mean.ming.axis2.2_5\endcsname{159}
\expandafter\gdef\csname odunum@ci@positive_taxonomy_m2.av.mean.ming.axis2.2_5\endcsname{[0.428, 0.512]}
\expandafter\gdef\csname odunum@val@positive_taxonomy_m2.av.mean.ming.axis3.3_1\endcsname{0.505}
\expandafter\gdef\csname odunum@n@positive_taxonomy_m2.av.mean.ming.axis3.3_1\endcsname{91}
\expandafter\gdef\csname odunum@ci@positive_taxonomy_m2.av.mean.ming.axis3.3_1\endcsname{[0.445, 0.564]}
\expandafter\gdef\csname odunum@val@positive_taxonomy_m2.av.mean.ming.axis3.3_2\endcsname{0.472}
\expandafter\gdef\csname odunum@n@positive_taxonomy_m2.av.mean.ming.axis3.3_2\endcsname{85}
\expandafter\gdef\csname odunum@ci@positive_taxonomy_m2.av.mean.ming.axis3.3_2\endcsname{[0.418, 0.525]}
\expandafter\gdef\csname odunum@val@positive_taxonomy_m2.av.mean.ming.axis3.3_3\endcsname{0.461}
\expandafter\gdef\csname odunum@n@positive_taxonomy_m2.av.mean.ming.axis3.3_3\endcsname{137}
\expandafter\gdef\csname odunum@ci@positive_taxonomy_m2.av.mean.ming.axis3.3_3\endcsname{[0.413, 0.508]}
\expandafter\gdef\csname odunum@val@positive_taxonomy_m2.av.mean.ming.axis3.3_4\endcsname{0.476}
\expandafter\gdef\csname odunum@n@positive_taxonomy_m2.av.mean.ming.axis3.3_4\endcsname{256}
\expandafter\gdef\csname odunum@ci@positive_taxonomy_m2.av.mean.ming.axis3.3_4\endcsname{[0.444, 0.508]}
\expandafter\gdef\csname odunum@val@positive_taxonomy_m2.av.mean.ming.axis3.3_5\endcsname{0.455}
\expandafter\gdef\csname odunum@n@positive_taxonomy_m2.av.mean.ming.axis3.3_5\endcsname{185}
\expandafter\gdef\csname odunum@ci@positive_taxonomy_m2.av.mean.ming.axis3.3_5\endcsname{[0.420, 0.489]}
\expandafter\gdef\csname odunum@val@positive_taxonomy_m2.av.mean.ming.axis3.3_6\endcsname{0.560}
\expandafter\gdef\csname odunum@n@positive_taxonomy_m2.av.mean.ming.axis3.3_6\endcsname{289}
\expandafter\gdef\csname odunum@ci@positive_taxonomy_m2.av.mean.ming.axis3.3_6\endcsname{[0.530, 0.590]}
\expandafter\gdef\csname odunum@val@positive_taxonomy_m2.av.mean.ming.axis4.4_1\endcsname{0.496}
\expandafter\gdef\csname odunum@n@positive_taxonomy_m2.av.mean.ming.axis4.4_1\endcsname{137}
\expandafter\gdef\csname odunum@ci@positive_taxonomy_m2.av.mean.ming.axis4.4_1\endcsname{[0.456, 0.539]}
\expandafter\gdef\csname odunum@val@positive_taxonomy_m2.av.mean.ming.axis4.4_2\endcsname{0.503}
\expandafter\gdef\csname odunum@n@positive_taxonomy_m2.av.mean.ming.axis4.4_2\endcsname{289}
\expandafter\gdef\csname odunum@ci@positive_taxonomy_m2.av.mean.ming.axis4.4_2\endcsname{[0.471, 0.533]}
\expandafter\gdef\csname odunum@val@positive_taxonomy_m2.av.mean.ming.axis4.4_3\endcsname{0.496}
\expandafter\gdef\csname odunum@n@positive_taxonomy_m2.av.mean.ming.axis4.4_3\endcsname{120}
\expandafter\gdef\csname odunum@ci@positive_taxonomy_m2.av.mean.ming.axis4.4_3\endcsname{[0.446, 0.548]}
\expandafter\gdef\csname odunum@val@positive_taxonomy_m2.av.mean.ming.axis4.4_4\endcsname{0.488}
\expandafter\gdef\csname odunum@n@positive_taxonomy_m2.av.mean.ming.axis4.4_4\endcsname{324}
\expandafter\gdef\csname odunum@ci@positive_taxonomy_m2.av.mean.ming.axis4.4_4\endcsname{[0.460, 0.517]}
\expandafter\gdef\csname odunum@val@positive_taxonomy_m2.av.mean.ming.axis4.4_5\endcsname{0.496}
\expandafter\gdef\csname odunum@n@positive_taxonomy_m2.av.mean.ming.axis4.4_5\endcsname{173}
\expandafter\gdef\csname odunum@ci@positive_taxonomy_m2.av.mean.ming.axis4.4_5\endcsname{[0.456, 0.536]}
\expandafter\gdef\csname odunum@val@positive_taxonomy_m2.av.mean.minicpm_o.axis1.1_1\endcsname{0.352}
\expandafter\gdef\csname odunum@n@positive_taxonomy_m2.av.mean.minicpm_o.axis1.1_1\endcsname{92}
\expandafter\gdef\csname odunum@ci@positive_taxonomy_m2.av.mean.minicpm_o.axis1.1_1\endcsname{[0.286, 0.419]}
\expandafter\gdef\csname odunum@val@positive_taxonomy_m2.av.mean.minicpm_o.axis1.1_2\endcsname{0.380}
\expandafter\gdef\csname odunum@n@positive_taxonomy_m2.av.mean.minicpm_o.axis1.1_2\endcsname{288}
\expandafter\gdef\csname odunum@ci@positive_taxonomy_m2.av.mean.minicpm_o.axis1.1_2\endcsname{[0.344, 0.417]}
\expandafter\gdef\csname odunum@val@positive_taxonomy_m2.av.mean.minicpm_o.axis1.1_3\endcsname{0.348}
\expandafter\gdef\csname odunum@n@positive_taxonomy_m2.av.mean.minicpm_o.axis1.1_3\endcsname{286}
\expandafter\gdef\csname odunum@ci@positive_taxonomy_m2.av.mean.minicpm_o.axis1.1_3\endcsname{[0.312, 0.382]}
\expandafter\gdef\csname odunum@val@positive_taxonomy_m2.av.mean.minicpm_o.axis1.1_4\endcsname{0.343}
\expandafter\gdef\csname odunum@n@positive_taxonomy_m2.av.mean.minicpm_o.axis1.1_4\endcsname{377}
\expandafter\gdef\csname odunum@ci@positive_taxonomy_m2.av.mean.minicpm_o.axis1.1_4\endcsname{[0.313, 0.371]}
\expandafter\gdef\csname odunum@val@positive_taxonomy_m2.av.mean.minicpm_o.axis2.2_1\endcsname{0.353}
\expandafter\gdef\csname odunum@n@positive_taxonomy_m2.av.mean.minicpm_o.axis2.2_1\endcsname{118}
\expandafter\gdef\csname odunum@ci@positive_taxonomy_m2.av.mean.minicpm_o.axis2.2_1\endcsname{[0.300, 0.405]}
\expandafter\gdef\csname odunum@val@positive_taxonomy_m2.av.mean.minicpm_o.axis2.2_2\endcsname{0.364}
\expandafter\gdef\csname odunum@n@positive_taxonomy_m2.av.mean.minicpm_o.axis2.2_2\endcsname{266}
\expandafter\gdef\csname odunum@ci@positive_taxonomy_m2.av.mean.minicpm_o.axis2.2_2\endcsname{[0.328, 0.401]}
\expandafter\gdef\csname odunum@val@positive_taxonomy_m2.av.mean.minicpm_o.axis2.2_3\endcsname{0.308}
\expandafter\gdef\csname odunum@n@positive_taxonomy_m2.av.mean.minicpm_o.axis2.2_3\endcsname{274}
\expandafter\gdef\csname odunum@ci@positive_taxonomy_m2.av.mean.minicpm_o.axis2.2_3\endcsname{[0.269, 0.347]}
\expandafter\gdef\csname odunum@val@positive_taxonomy_m2.av.mean.minicpm_o.axis2.2_4\endcsname{0.382}
\expandafter\gdef\csname odunum@n@positive_taxonomy_m2.av.mean.minicpm_o.axis2.2_4\endcsname{226}
\expandafter\gdef\csname odunum@ci@positive_taxonomy_m2.av.mean.minicpm_o.axis2.2_4\endcsname{[0.344, 0.420]}
\expandafter\gdef\csname odunum@val@positive_taxonomy_m2.av.mean.minicpm_o.axis2.2_5\endcsname{0.385}
\expandafter\gdef\csname odunum@n@positive_taxonomy_m2.av.mean.minicpm_o.axis2.2_5\endcsname{159}
\expandafter\gdef\csname odunum@ci@positive_taxonomy_m2.av.mean.minicpm_o.axis2.2_5\endcsname{[0.342, 0.427]}
\expandafter\gdef\csname odunum@val@positive_taxonomy_m2.av.mean.minicpm_o.axis3.3_1\endcsname{0.412}
\expandafter\gdef\csname odunum@n@positive_taxonomy_m2.av.mean.minicpm_o.axis3.3_1\endcsname{91}
\expandafter\gdef\csname odunum@ci@positive_taxonomy_m2.av.mean.minicpm_o.axis3.3_1\endcsname{[0.347, 0.478]}
\expandafter\gdef\csname odunum@val@positive_taxonomy_m2.av.mean.minicpm_o.axis3.3_2\endcsname{0.337}
\expandafter\gdef\csname odunum@n@positive_taxonomy_m2.av.mean.minicpm_o.axis3.3_2\endcsname{85}
\expandafter\gdef\csname odunum@ci@positive_taxonomy_m2.av.mean.minicpm_o.axis3.3_2\endcsname{[0.267, 0.406]}
\expandafter\gdef\csname odunum@val@positive_taxonomy_m2.av.mean.minicpm_o.axis3.3_3\endcsname{0.260}
\expandafter\gdef\csname odunum@n@positive_taxonomy_m2.av.mean.minicpm_o.axis3.3_3\endcsname{137}
\expandafter\gdef\csname odunum@ci@positive_taxonomy_m2.av.mean.minicpm_o.axis3.3_3\endcsname{[0.214, 0.308]}
\expandafter\gdef\csname odunum@val@positive_taxonomy_m2.av.mean.minicpm_o.axis3.3_4\endcsname{0.410}
\expandafter\gdef\csname odunum@n@positive_taxonomy_m2.av.mean.minicpm_o.axis3.3_4\endcsname{256}
\expandafter\gdef\csname odunum@ci@positive_taxonomy_m2.av.mean.minicpm_o.axis3.3_4\endcsname{[0.371, 0.449]}
\expandafter\gdef\csname odunum@val@positive_taxonomy_m2.av.mean.minicpm_o.axis3.3_5\endcsname{0.362}
\expandafter\gdef\csname odunum@n@positive_taxonomy_m2.av.mean.minicpm_o.axis3.3_5\endcsname{185}
\expandafter\gdef\csname odunum@ci@positive_taxonomy_m2.av.mean.minicpm_o.axis3.3_5\endcsname{[0.321, 0.403]}
\expandafter\gdef\csname odunum@val@positive_taxonomy_m2.av.mean.minicpm_o.axis3.3_6\endcsname{0.335}
\expandafter\gdef\csname odunum@n@positive_taxonomy_m2.av.mean.minicpm_o.axis3.3_6\endcsname{289}
\expandafter\gdef\csname odunum@ci@positive_taxonomy_m2.av.mean.minicpm_o.axis3.3_6\endcsname{[0.302, 0.368]}
\expandafter\gdef\csname odunum@val@positive_taxonomy_m2.av.mean.minicpm_o.axis4.4_1\endcsname{0.312}
\expandafter\gdef\csname odunum@n@positive_taxonomy_m2.av.mean.minicpm_o.axis4.4_1\endcsname{137}
\expandafter\gdef\csname odunum@ci@positive_taxonomy_m2.av.mean.minicpm_o.axis4.4_1\endcsname{[0.264, 0.360]}
\expandafter\gdef\csname odunum@val@positive_taxonomy_m2.av.mean.minicpm_o.axis4.4_2\endcsname{0.380}
\expandafter\gdef\csname odunum@n@positive_taxonomy_m2.av.mean.minicpm_o.axis4.4_2\endcsname{289}
\expandafter\gdef\csname odunum@ci@positive_taxonomy_m2.av.mean.minicpm_o.axis4.4_2\endcsname{[0.345, 0.416]}
\expandafter\gdef\csname odunum@val@positive_taxonomy_m2.av.mean.minicpm_o.axis4.4_3\endcsname{0.292}
\expandafter\gdef\csname odunum@n@positive_taxonomy_m2.av.mean.minicpm_o.axis4.4_3\endcsname{120}
\expandafter\gdef\csname odunum@ci@positive_taxonomy_m2.av.mean.minicpm_o.axis4.4_3\endcsname{[0.241, 0.344]}
\expandafter\gdef\csname odunum@val@positive_taxonomy_m2.av.mean.minicpm_o.axis4.4_4\endcsname{0.361}
\expandafter\gdef\csname odunum@n@positive_taxonomy_m2.av.mean.minicpm_o.axis4.4_4\endcsname{324}
\expandafter\gdef\csname odunum@ci@positive_taxonomy_m2.av.mean.minicpm_o.axis4.4_4\endcsname{[0.327, 0.394]}
\expandafter\gdef\csname odunum@val@positive_taxonomy_m2.av.mean.minicpm_o.axis4.4_5\endcsname{0.381}
\expandafter\gdef\csname odunum@n@positive_taxonomy_m2.av.mean.minicpm_o.axis4.4_5\endcsname{173}
\expandafter\gdef\csname odunum@ci@positive_taxonomy_m2.av.mean.minicpm_o.axis4.4_5\endcsname{[0.334, 0.427]}
\expandafter\gdef\csname odunum@val@positive_taxonomy_m2.av.mean.nemotron.axis1.1_1\endcsname{0.214}
\expandafter\gdef\csname odunum@n@positive_taxonomy_m2.av.mean.nemotron.axis1.1_1\endcsname{92}
\expandafter\gdef\csname odunum@ci@positive_taxonomy_m2.av.mean.nemotron.axis1.1_1\endcsname{[0.153, 0.279]}
\expandafter\gdef\csname odunum@val@positive_taxonomy_m2.av.mean.nemotron.axis1.1_2\endcsname{0.280}
\expandafter\gdef\csname odunum@n@positive_taxonomy_m2.av.mean.nemotron.axis1.1_2\endcsname{288}
\expandafter\gdef\csname odunum@ci@positive_taxonomy_m2.av.mean.nemotron.axis1.1_2\endcsname{[0.243, 0.317]}
\expandafter\gdef\csname odunum@val@positive_taxonomy_m2.av.mean.nemotron.axis1.1_3\endcsname{0.232}
\expandafter\gdef\csname odunum@n@positive_taxonomy_m2.av.mean.nemotron.axis1.1_3\endcsname{286}
\expandafter\gdef\csname odunum@ci@positive_taxonomy_m2.av.mean.nemotron.axis1.1_3\endcsname{[0.196, 0.269]}
\expandafter\gdef\csname odunum@val@positive_taxonomy_m2.av.mean.nemotron.axis1.1_4\endcsname{0.294}
\expandafter\gdef\csname odunum@n@positive_taxonomy_m2.av.mean.nemotron.axis1.1_4\endcsname{377}
\expandafter\gdef\csname odunum@ci@positive_taxonomy_m2.av.mean.nemotron.axis1.1_4\endcsname{[0.261, 0.327]}
\expandafter\gdef\csname odunum@val@positive_taxonomy_m2.av.mean.nemotron.axis2.2_1\endcsname{0.265}
\expandafter\gdef\csname odunum@n@positive_taxonomy_m2.av.mean.nemotron.axis2.2_1\endcsname{118}
\expandafter\gdef\csname odunum@ci@positive_taxonomy_m2.av.mean.nemotron.axis2.2_1\endcsname{[0.208, 0.322]}
\expandafter\gdef\csname odunum@val@positive_taxonomy_m2.av.mean.nemotron.axis2.2_2\endcsname{0.275}
\expandafter\gdef\csname odunum@n@positive_taxonomy_m2.av.mean.nemotron.axis2.2_2\endcsname{266}
\expandafter\gdef\csname odunum@ci@positive_taxonomy_m2.av.mean.nemotron.axis2.2_2\endcsname{[0.236, 0.314]}
\expandafter\gdef\csname odunum@val@positive_taxonomy_m2.av.mean.nemotron.axis2.2_3\endcsname{0.274}
\expandafter\gdef\csname odunum@n@positive_taxonomy_m2.av.mean.nemotron.axis2.2_3\endcsname{274}
\expandafter\gdef\csname odunum@ci@positive_taxonomy_m2.av.mean.nemotron.axis2.2_3\endcsname{[0.236, 0.314]}
\expandafter\gdef\csname odunum@val@positive_taxonomy_m2.av.mean.nemotron.axis2.2_4\endcsname{0.255}
\expandafter\gdef\csname odunum@n@positive_taxonomy_m2.av.mean.nemotron.axis2.2_4\endcsname{226}
\expandafter\gdef\csname odunum@ci@positive_taxonomy_m2.av.mean.nemotron.axis2.2_4\endcsname{[0.213, 0.297]}
\expandafter\gdef\csname odunum@val@positive_taxonomy_m2.av.mean.nemotron.axis2.2_5\endcsname{0.256}
\expandafter\gdef\csname odunum@n@positive_taxonomy_m2.av.mean.nemotron.axis2.2_5\endcsname{159}
\expandafter\gdef\csname odunum@ci@positive_taxonomy_m2.av.mean.nemotron.axis2.2_5\endcsname{[0.211, 0.302]}
\expandafter\gdef\csname odunum@val@positive_taxonomy_m2.av.mean.nemotron.axis3.3_1\endcsname{0.288}
\expandafter\gdef\csname odunum@n@positive_taxonomy_m2.av.mean.nemotron.axis3.3_1\endcsname{91}
\expandafter\gdef\csname odunum@ci@positive_taxonomy_m2.av.mean.nemotron.axis3.3_1\endcsname{[0.221, 0.358]}
\expandafter\gdef\csname odunum@val@positive_taxonomy_m2.av.mean.nemotron.axis3.3_2\endcsname{0.281}
\expandafter\gdef\csname odunum@n@positive_taxonomy_m2.av.mean.nemotron.axis3.3_2\endcsname{85}
\expandafter\gdef\csname odunum@ci@positive_taxonomy_m2.av.mean.nemotron.axis3.3_2\endcsname{[0.212, 0.352]}
\expandafter\gdef\csname odunum@val@positive_taxonomy_m2.av.mean.nemotron.axis3.3_3\endcsname{0.188}
\expandafter\gdef\csname odunum@n@positive_taxonomy_m2.av.mean.nemotron.axis3.3_3\endcsname{137}
\expandafter\gdef\csname odunum@ci@positive_taxonomy_m2.av.mean.nemotron.axis3.3_3\endcsname{[0.141, 0.236]}
\expandafter\gdef\csname odunum@val@positive_taxonomy_m2.av.mean.nemotron.axis3.3_4\endcsname{0.288}
\expandafter\gdef\csname odunum@n@positive_taxonomy_m2.av.mean.nemotron.axis3.3_4\endcsname{256}
\expandafter\gdef\csname odunum@ci@positive_taxonomy_m2.av.mean.nemotron.axis3.3_4\endcsname{[0.245, 0.331]}
\expandafter\gdef\csname odunum@val@positive_taxonomy_m2.av.mean.nemotron.axis3.3_5\endcsname{0.294}
\expandafter\gdef\csname odunum@n@positive_taxonomy_m2.av.mean.nemotron.axis3.3_5\endcsname{185}
\expandafter\gdef\csname odunum@ci@positive_taxonomy_m2.av.mean.nemotron.axis3.3_5\endcsname{[0.249, 0.340]}
\expandafter\gdef\csname odunum@val@positive_taxonomy_m2.av.mean.nemotron.axis3.3_6\endcsname{0.255}
\expandafter\gdef\csname odunum@n@positive_taxonomy_m2.av.mean.nemotron.axis3.3_6\endcsname{289}
\expandafter\gdef\csname odunum@ci@positive_taxonomy_m2.av.mean.nemotron.axis3.3_6\endcsname{[0.221, 0.289]}
\expandafter\gdef\csname odunum@val@positive_taxonomy_m2.av.mean.nemotron.axis4.4_1\endcsname{0.246}
\expandafter\gdef\csname odunum@n@positive_taxonomy_m2.av.mean.nemotron.axis4.4_1\endcsname{137}
\expandafter\gdef\csname odunum@ci@positive_taxonomy_m2.av.mean.nemotron.axis4.4_1\endcsname{[0.194, 0.300]}
\expandafter\gdef\csname odunum@val@positive_taxonomy_m2.av.mean.nemotron.axis4.4_2\endcsname{0.305}
\expandafter\gdef\csname odunum@n@positive_taxonomy_m2.av.mean.nemotron.axis4.4_2\endcsname{289}
\expandafter\gdef\csname odunum@ci@positive_taxonomy_m2.av.mean.nemotron.axis4.4_2\endcsname{[0.268, 0.343]}
\expandafter\gdef\csname odunum@val@positive_taxonomy_m2.av.mean.nemotron.axis4.4_3\endcsname{0.222}
\expandafter\gdef\csname odunum@n@positive_taxonomy_m2.av.mean.nemotron.axis4.4_3\endcsname{120}
\expandafter\gdef\csname odunum@ci@positive_taxonomy_m2.av.mean.nemotron.axis4.4_3\endcsname{[0.167, 0.282]}
\expandafter\gdef\csname odunum@val@positive_taxonomy_m2.av.mean.nemotron.axis4.4_4\endcsname{0.260}
\expandafter\gdef\csname odunum@n@positive_taxonomy_m2.av.mean.nemotron.axis4.4_4\endcsname{324}
\expandafter\gdef\csname odunum@ci@positive_taxonomy_m2.av.mean.nemotron.axis4.4_4\endcsname{[0.226, 0.295]}
\expandafter\gdef\csname odunum@val@positive_taxonomy_m2.av.mean.nemotron.axis4.4_5\endcsname{0.259}
\expandafter\gdef\csname odunum@n@positive_taxonomy_m2.av.mean.nemotron.axis4.4_5\endcsname{173}
\expandafter\gdef\csname odunum@ci@positive_taxonomy_m2.av.mean.nemotron.axis4.4_5\endcsname{[0.213, 0.306]}
\expandafter\gdef\csname odunum@val@positive_taxonomy_m2.av.mean.qwen25_omni.axis1.1_1\endcsname{0.454}
\expandafter\gdef\csname odunum@n@positive_taxonomy_m2.av.mean.qwen25_omni.axis1.1_1\endcsname{92}
\expandafter\gdef\csname odunum@ci@positive_taxonomy_m2.av.mean.qwen25_omni.axis1.1_1\endcsname{[0.392, 0.516]}
\expandafter\gdef\csname odunum@val@positive_taxonomy_m2.av.mean.qwen25_omni.axis1.1_2\endcsname{0.415}
\expandafter\gdef\csname odunum@n@positive_taxonomy_m2.av.mean.qwen25_omni.axis1.1_2\endcsname{288}
\expandafter\gdef\csname odunum@ci@positive_taxonomy_m2.av.mean.qwen25_omni.axis1.1_2\endcsname{[0.382, 0.449]}
\expandafter\gdef\csname odunum@val@positive_taxonomy_m2.av.mean.qwen25_omni.axis1.1_3\endcsname{0.439}
\expandafter\gdef\csname odunum@n@positive_taxonomy_m2.av.mean.qwen25_omni.axis1.1_3\endcsname{286}
\expandafter\gdef\csname odunum@ci@positive_taxonomy_m2.av.mean.qwen25_omni.axis1.1_3\endcsname{[0.403, 0.475]}
\expandafter\gdef\csname odunum@val@positive_taxonomy_m2.av.mean.qwen25_omni.axis1.1_4\endcsname{0.403}
\expandafter\gdef\csname odunum@n@positive_taxonomy_m2.av.mean.qwen25_omni.axis1.1_4\endcsname{377}
\expandafter\gdef\csname odunum@ci@positive_taxonomy_m2.av.mean.qwen25_omni.axis1.1_4\endcsname{[0.376, 0.430]}
\expandafter\gdef\csname odunum@val@positive_taxonomy_m2.av.mean.qwen25_omni.axis2.2_1\endcsname{0.346}
\expandafter\gdef\csname odunum@n@positive_taxonomy_m2.av.mean.qwen25_omni.axis2.2_1\endcsname{118}
\expandafter\gdef\csname odunum@ci@positive_taxonomy_m2.av.mean.qwen25_omni.axis2.2_1\endcsname{[0.294, 0.401]}
\expandafter\gdef\csname odunum@val@positive_taxonomy_m2.av.mean.qwen25_omni.axis2.2_2\endcsname{0.461}
\expandafter\gdef\csname odunum@n@positive_taxonomy_m2.av.mean.qwen25_omni.axis2.2_2\endcsname{266}
\expandafter\gdef\csname odunum@ci@positive_taxonomy_m2.av.mean.qwen25_omni.axis2.2_2\endcsname{[0.427, 0.495]}
\expandafter\gdef\csname odunum@val@positive_taxonomy_m2.av.mean.qwen25_omni.axis2.2_3\endcsname{0.443}
\expandafter\gdef\csname odunum@n@positive_taxonomy_m2.av.mean.qwen25_omni.axis2.2_3\endcsname{274}
\expandafter\gdef\csname odunum@ci@positive_taxonomy_m2.av.mean.qwen25_omni.axis2.2_3\endcsname{[0.408, 0.478]}
\expandafter\gdef\csname odunum@val@positive_taxonomy_m2.av.mean.qwen25_omni.axis2.2_4\endcsname{0.409}
\expandafter\gdef\csname odunum@n@positive_taxonomy_m2.av.mean.qwen25_omni.axis2.2_4\endcsname{226}
\expandafter\gdef\csname odunum@ci@positive_taxonomy_m2.av.mean.qwen25_omni.axis2.2_4\endcsname{[0.373, 0.445]}
\expandafter\gdef\csname odunum@val@positive_taxonomy_m2.av.mean.qwen25_omni.axis2.2_5\endcsname{0.385}
\expandafter\gdef\csname odunum@n@positive_taxonomy_m2.av.mean.qwen25_omni.axis2.2_5\endcsname{159}
\expandafter\gdef\csname odunum@ci@positive_taxonomy_m2.av.mean.qwen25_omni.axis2.2_5\endcsname{[0.343, 0.428]}
\expandafter\gdef\csname odunum@val@positive_taxonomy_m2.av.mean.qwen25_omni.axis3.3_1\endcsname{0.438}
\expandafter\gdef\csname odunum@n@positive_taxonomy_m2.av.mean.qwen25_omni.axis3.3_1\endcsname{91}
\expandafter\gdef\csname odunum@ci@positive_taxonomy_m2.av.mean.qwen25_omni.axis3.3_1\endcsname{[0.379, 0.497]}
\expandafter\gdef\csname odunum@val@positive_taxonomy_m2.av.mean.qwen25_omni.axis3.3_2\endcsname{0.455}
\expandafter\gdef\csname odunum@n@positive_taxonomy_m2.av.mean.qwen25_omni.axis3.3_2\endcsname{85}
\expandafter\gdef\csname odunum@ci@positive_taxonomy_m2.av.mean.qwen25_omni.axis3.3_2\endcsname{[0.390, 0.519]}
\expandafter\gdef\csname odunum@val@positive_taxonomy_m2.av.mean.qwen25_omni.axis3.3_3\endcsname{0.310}
\expandafter\gdef\csname odunum@n@positive_taxonomy_m2.av.mean.qwen25_omni.axis3.3_3\endcsname{137}
\expandafter\gdef\csname odunum@ci@positive_taxonomy_m2.av.mean.qwen25_omni.axis3.3_3\endcsname{[0.261, 0.359]}
\expandafter\gdef\csname odunum@val@positive_taxonomy_m2.av.mean.qwen25_omni.axis3.3_4\endcsname{0.443}
\expandafter\gdef\csname odunum@n@positive_taxonomy_m2.av.mean.qwen25_omni.axis3.3_4\endcsname{256}
\expandafter\gdef\csname odunum@ci@positive_taxonomy_m2.av.mean.qwen25_omni.axis3.3_4\endcsname{[0.408, 0.479]}
\expandafter\gdef\csname odunum@val@positive_taxonomy_m2.av.mean.qwen25_omni.axis3.3_5\endcsname{0.414}
\expandafter\gdef\csname odunum@n@positive_taxonomy_m2.av.mean.qwen25_omni.axis3.3_5\endcsname{185}
\expandafter\gdef\csname odunum@ci@positive_taxonomy_m2.av.mean.qwen25_omni.axis3.3_5\endcsname{[0.373, 0.453]}
\expandafter\gdef\csname odunum@val@positive_taxonomy_m2.av.mean.qwen25_omni.axis3.3_6\endcsname{0.442}
\expandafter\gdef\csname odunum@n@positive_taxonomy_m2.av.mean.qwen25_omni.axis3.3_6\endcsname{289}
\expandafter\gdef\csname odunum@ci@positive_taxonomy_m2.av.mean.qwen25_omni.axis3.3_6\endcsname{[0.409, 0.475]}
\expandafter\gdef\csname odunum@val@positive_taxonomy_m2.av.mean.qwen25_omni.axis4.4_1\endcsname{0.435}
\expandafter\gdef\csname odunum@n@positive_taxonomy_m2.av.mean.qwen25_omni.axis4.4_1\endcsname{137}
\expandafter\gdef\csname odunum@ci@positive_taxonomy_m2.av.mean.qwen25_omni.axis4.4_1\endcsname{[0.388, 0.482]}
\expandafter\gdef\csname odunum@val@positive_taxonomy_m2.av.mean.qwen25_omni.axis4.4_2\endcsname{0.449}
\expandafter\gdef\csname odunum@n@positive_taxonomy_m2.av.mean.qwen25_omni.axis4.4_2\endcsname{289}
\expandafter\gdef\csname odunum@ci@positive_taxonomy_m2.av.mean.qwen25_omni.axis4.4_2\endcsname{[0.417, 0.481]}
\expandafter\gdef\csname odunum@val@positive_taxonomy_m2.av.mean.qwen25_omni.axis4.4_3\endcsname{0.345}
\expandafter\gdef\csname odunum@n@positive_taxonomy_m2.av.mean.qwen25_omni.axis4.4_3\endcsname{120}
\expandafter\gdef\csname odunum@ci@positive_taxonomy_m2.av.mean.qwen25_omni.axis4.4_3\endcsname{[0.292, 0.401]}
\expandafter\gdef\csname odunum@val@positive_taxonomy_m2.av.mean.qwen25_omni.axis4.4_4\endcsname{0.417}
\expandafter\gdef\csname odunum@n@positive_taxonomy_m2.av.mean.qwen25_omni.axis4.4_4\endcsname{324}
\expandafter\gdef\csname odunum@ci@positive_taxonomy_m2.av.mean.qwen25_omni.axis4.4_4\endcsname{[0.386, 0.450]}
\expandafter\gdef\csname odunum@val@positive_taxonomy_m2.av.mean.qwen25_omni.axis4.4_5\endcsname{0.420}
\expandafter\gdef\csname odunum@n@positive_taxonomy_m2.av.mean.qwen25_omni.axis4.4_5\endcsname{173}
\expandafter\gdef\csname odunum@ci@positive_taxonomy_m2.av.mean.qwen25_omni.axis4.4_5\endcsname{[0.376, 0.464]}
\expandafter\gdef\csname odunum@val@positive_taxonomy_m2.av.mean.qwen3_omni_instruct.axis1.1_1\endcsname{0.523}
\expandafter\gdef\csname odunum@n@positive_taxonomy_m2.av.mean.qwen3_omni_instruct.axis1.1_1\endcsname{92}
\expandafter\gdef\csname odunum@ci@positive_taxonomy_m2.av.mean.qwen3_omni_instruct.axis1.1_1\endcsname{[0.463, 0.582]}
\expandafter\gdef\csname odunum@val@positive_taxonomy_m2.av.mean.qwen3_omni_instruct.axis1.1_2\endcsname{0.548}
\expandafter\gdef\csname odunum@n@positive_taxonomy_m2.av.mean.qwen3_omni_instruct.axis1.1_2\endcsname{288}
\expandafter\gdef\csname odunum@ci@positive_taxonomy_m2.av.mean.qwen3_omni_instruct.axis1.1_2\endcsname{[0.518, 0.578]}
\expandafter\gdef\csname odunum@val@positive_taxonomy_m2.av.mean.qwen3_omni_instruct.axis1.1_3\endcsname{0.538}
\expandafter\gdef\csname odunum@n@positive_taxonomy_m2.av.mean.qwen3_omni_instruct.axis1.1_3\endcsname{286}
\expandafter\gdef\csname odunum@ci@positive_taxonomy_m2.av.mean.qwen3_omni_instruct.axis1.1_3\endcsname{[0.507, 0.569]}
\expandafter\gdef\csname odunum@val@positive_taxonomy_m2.av.mean.qwen3_omni_instruct.axis1.1_4\endcsname{0.514}
\expandafter\gdef\csname odunum@n@positive_taxonomy_m2.av.mean.qwen3_omni_instruct.axis1.1_4\endcsname{377}
\expandafter\gdef\csname odunum@ci@positive_taxonomy_m2.av.mean.qwen3_omni_instruct.axis1.1_4\endcsname{[0.488, 0.540]}
\expandafter\gdef\csname odunum@val@positive_taxonomy_m2.av.mean.qwen3_omni_instruct.axis2.2_1\endcsname{0.547}
\expandafter\gdef\csname odunum@n@positive_taxonomy_m2.av.mean.qwen3_omni_instruct.axis2.2_1\endcsname{118}
\expandafter\gdef\csname odunum@ci@positive_taxonomy_m2.av.mean.qwen3_omni_instruct.axis2.2_1\endcsname{[0.499, 0.596]}
\expandafter\gdef\csname odunum@val@positive_taxonomy_m2.av.mean.qwen3_omni_instruct.axis2.2_2\endcsname{0.513}
\expandafter\gdef\csname odunum@n@positive_taxonomy_m2.av.mean.qwen3_omni_instruct.axis2.2_2\endcsname{266}
\expandafter\gdef\csname odunum@ci@positive_taxonomy_m2.av.mean.qwen3_omni_instruct.axis2.2_2\endcsname{[0.481, 0.545]}
\expandafter\gdef\csname odunum@val@positive_taxonomy_m2.av.mean.qwen3_omni_instruct.axis2.2_3\endcsname{0.523}
\expandafter\gdef\csname odunum@n@positive_taxonomy_m2.av.mean.qwen3_omni_instruct.axis2.2_3\endcsname{274}
\expandafter\gdef\csname odunum@ci@positive_taxonomy_m2.av.mean.qwen3_omni_instruct.axis2.2_3\endcsname{[0.492, 0.554]}
\expandafter\gdef\csname odunum@val@positive_taxonomy_m2.av.mean.qwen3_omni_instruct.axis2.2_4\endcsname{0.535}
\expandafter\gdef\csname odunum@n@positive_taxonomy_m2.av.mean.qwen3_omni_instruct.axis2.2_4\endcsname{226}
\expandafter\gdef\csname odunum@ci@positive_taxonomy_m2.av.mean.qwen3_omni_instruct.axis2.2_4\endcsname{[0.499, 0.571]}
\expandafter\gdef\csname odunum@val@positive_taxonomy_m2.av.mean.qwen3_omni_instruct.axis2.2_5\endcsname{0.556}
\expandafter\gdef\csname odunum@n@positive_taxonomy_m2.av.mean.qwen3_omni_instruct.axis2.2_5\endcsname{159}
\expandafter\gdef\csname odunum@ci@positive_taxonomy_m2.av.mean.qwen3_omni_instruct.axis2.2_5\endcsname{[0.518, 0.595]}
\expandafter\gdef\csname odunum@val@positive_taxonomy_m2.av.mean.qwen3_omni_instruct.axis3.3_1\endcsname{0.537}
\expandafter\gdef\csname odunum@n@positive_taxonomy_m2.av.mean.qwen3_omni_instruct.axis3.3_1\endcsname{91}
\expandafter\gdef\csname odunum@ci@positive_taxonomy_m2.av.mean.qwen3_omni_instruct.axis3.3_1\endcsname{[0.481, 0.594]}
\expandafter\gdef\csname odunum@val@positive_taxonomy_m2.av.mean.qwen3_omni_instruct.axis3.3_2\endcsname{0.536}
\expandafter\gdef\csname odunum@n@positive_taxonomy_m2.av.mean.qwen3_omni_instruct.axis3.3_2\endcsname{85}
\expandafter\gdef\csname odunum@ci@positive_taxonomy_m2.av.mean.qwen3_omni_instruct.axis3.3_2\endcsname{[0.481, 0.592]}
\expandafter\gdef\csname odunum@val@positive_taxonomy_m2.av.mean.qwen3_omni_instruct.axis3.3_3\endcsname{0.484}
\expandafter\gdef\csname odunum@n@positive_taxonomy_m2.av.mean.qwen3_omni_instruct.axis3.3_3\endcsname{137}
\expandafter\gdef\csname odunum@ci@positive_taxonomy_m2.av.mean.qwen3_omni_instruct.axis3.3_3\endcsname{[0.440, 0.528]}
\expandafter\gdef\csname odunum@val@positive_taxonomy_m2.av.mean.qwen3_omni_instruct.axis3.3_4\endcsname{0.524}
\expandafter\gdef\csname odunum@n@positive_taxonomy_m2.av.mean.qwen3_omni_instruct.axis3.3_4\endcsname{256}
\expandafter\gdef\csname odunum@ci@positive_taxonomy_m2.av.mean.qwen3_omni_instruct.axis3.3_4\endcsname{[0.491, 0.557]}
\expandafter\gdef\csname odunum@val@positive_taxonomy_m2.av.mean.qwen3_omni_instruct.axis3.3_5\endcsname{0.507}
\expandafter\gdef\csname odunum@n@positive_taxonomy_m2.av.mean.qwen3_omni_instruct.axis3.3_5\endcsname{185}
\expandafter\gdef\csname odunum@ci@positive_taxonomy_m2.av.mean.qwen3_omni_instruct.axis3.3_5\endcsname{[0.468, 0.546]}
\expandafter\gdef\csname odunum@val@positive_taxonomy_m2.av.mean.qwen3_omni_instruct.axis3.3_6\endcsname{0.570}
\expandafter\gdef\csname odunum@n@positive_taxonomy_m2.av.mean.qwen3_omni_instruct.axis3.3_6\endcsname{289}
\expandafter\gdef\csname odunum@ci@positive_taxonomy_m2.av.mean.qwen3_omni_instruct.axis3.3_6\endcsname{[0.541, 0.599]}
\expandafter\gdef\csname odunum@val@positive_taxonomy_m2.av.mean.qwen3_omni_instruct.axis4.4_1\endcsname{0.500}
\expandafter\gdef\csname odunum@n@positive_taxonomy_m2.av.mean.qwen3_omni_instruct.axis4.4_1\endcsname{137}
\expandafter\gdef\csname odunum@ci@positive_taxonomy_m2.av.mean.qwen3_omni_instruct.axis4.4_1\endcsname{[0.458, 0.543]}
\expandafter\gdef\csname odunum@val@positive_taxonomy_m2.av.mean.qwen3_omni_instruct.axis4.4_2\endcsname{0.561}
\expandafter\gdef\csname odunum@n@positive_taxonomy_m2.av.mean.qwen3_omni_instruct.axis4.4_2\endcsname{289}
\expandafter\gdef\csname odunum@ci@positive_taxonomy_m2.av.mean.qwen3_omni_instruct.axis4.4_2\endcsname{[0.532, 0.589]}
\expandafter\gdef\csname odunum@val@positive_taxonomy_m2.av.mean.qwen3_omni_instruct.axis4.4_3\endcsname{0.496}
\expandafter\gdef\csname odunum@n@positive_taxonomy_m2.av.mean.qwen3_omni_instruct.axis4.4_3\endcsname{120}
\expandafter\gdef\csname odunum@ci@positive_taxonomy_m2.av.mean.qwen3_omni_instruct.axis4.4_3\endcsname{[0.442, 0.550]}
\expandafter\gdef\csname odunum@val@positive_taxonomy_m2.av.mean.qwen3_omni_instruct.axis4.4_4\endcsname{0.523}
\expandafter\gdef\csname odunum@n@positive_taxonomy_m2.av.mean.qwen3_omni_instruct.axis4.4_4\endcsname{324}
\expandafter\gdef\csname odunum@ci@positive_taxonomy_m2.av.mean.qwen3_omni_instruct.axis4.4_4\endcsname{[0.493, 0.553]}
\expandafter\gdef\csname odunum@val@positive_taxonomy_m2.av.mean.qwen3_omni_instruct.axis4.4_5\endcsname{0.544}
\expandafter\gdef\csname odunum@n@positive_taxonomy_m2.av.mean.qwen3_omni_instruct.axis4.4_5\endcsname{173}
\expandafter\gdef\csname odunum@ci@positive_taxonomy_m2.av.mean.qwen3_omni_instruct.axis4.4_5\endcsname{[0.509, 0.578]}
\expandafter\gdef\csname odunum@val@positive_taxonomy_m2.av.mean.qwen3_omni_think.axis1.1_1\endcsname{0.600}
\expandafter\gdef\csname odunum@n@positive_taxonomy_m2.av.mean.qwen3_omni_think.axis1.1_1\endcsname{92}
\expandafter\gdef\csname odunum@ci@positive_taxonomy_m2.av.mean.qwen3_omni_think.axis1.1_1\endcsname{[0.544, 0.657]}
\expandafter\gdef\csname odunum@val@positive_taxonomy_m2.av.mean.qwen3_omni_think.axis1.1_2\endcsname{0.565}
\expandafter\gdef\csname odunum@n@positive_taxonomy_m2.av.mean.qwen3_omni_think.axis1.1_2\endcsname{288}
\expandafter\gdef\csname odunum@ci@positive_taxonomy_m2.av.mean.qwen3_omni_think.axis1.1_2\endcsname{[0.534, 0.596]}
\expandafter\gdef\csname odunum@val@positive_taxonomy_m2.av.mean.qwen3_omni_think.axis1.1_3\endcsname{0.537}
\expandafter\gdef\csname odunum@n@positive_taxonomy_m2.av.mean.qwen3_omni_think.axis1.1_3\endcsname{286}
\expandafter\gdef\csname odunum@ci@positive_taxonomy_m2.av.mean.qwen3_omni_think.axis1.1_3\endcsname{[0.507, 0.568]}
\expandafter\gdef\csname odunum@val@positive_taxonomy_m2.av.mean.qwen3_omni_think.axis1.1_4\endcsname{0.553}
\expandafter\gdef\csname odunum@n@positive_taxonomy_m2.av.mean.qwen3_omni_think.axis1.1_4\endcsname{377}
\expandafter\gdef\csname odunum@ci@positive_taxonomy_m2.av.mean.qwen3_omni_think.axis1.1_4\endcsname{[0.528, 0.578]}
\expandafter\gdef\csname odunum@val@positive_taxonomy_m2.av.mean.qwen3_omni_think.axis2.2_1\endcsname{0.573}
\expandafter\gdef\csname odunum@n@positive_taxonomy_m2.av.mean.qwen3_omni_think.axis2.2_1\endcsname{118}
\expandafter\gdef\csname odunum@ci@positive_taxonomy_m2.av.mean.qwen3_omni_think.axis2.2_1\endcsname{[0.520, 0.626]}
\expandafter\gdef\csname odunum@val@positive_taxonomy_m2.av.mean.qwen3_omni_think.axis2.2_2\endcsname{0.543}
\expandafter\gdef\csname odunum@n@positive_taxonomy_m2.av.mean.qwen3_omni_think.axis2.2_2\endcsname{266}
\expandafter\gdef\csname odunum@ci@positive_taxonomy_m2.av.mean.qwen3_omni_think.axis2.2_2\endcsname{[0.512, 0.573]}
\expandafter\gdef\csname odunum@val@positive_taxonomy_m2.av.mean.qwen3_omni_think.axis2.2_3\endcsname{0.576}
\expandafter\gdef\csname odunum@n@positive_taxonomy_m2.av.mean.qwen3_omni_think.axis2.2_3\endcsname{274}
\expandafter\gdef\csname odunum@ci@positive_taxonomy_m2.av.mean.qwen3_omni_think.axis2.2_3\endcsname{[0.546, 0.605]}
\expandafter\gdef\csname odunum@val@positive_taxonomy_m2.av.mean.qwen3_omni_think.axis2.2_4\endcsname{0.536}
\expandafter\gdef\csname odunum@n@positive_taxonomy_m2.av.mean.qwen3_omni_think.axis2.2_4\endcsname{226}
\expandafter\gdef\csname odunum@ci@positive_taxonomy_m2.av.mean.qwen3_omni_think.axis2.2_4\endcsname{[0.499, 0.573]}
\expandafter\gdef\csname odunum@val@positive_taxonomy_m2.av.mean.qwen3_omni_think.axis2.2_5\endcsname{0.562}
\expandafter\gdef\csname odunum@n@positive_taxonomy_m2.av.mean.qwen3_omni_think.axis2.2_5\endcsname{159}
\expandafter\gdef\csname odunum@ci@positive_taxonomy_m2.av.mean.qwen3_omni_think.axis2.2_5\endcsname{[0.526, 0.599]}
\expandafter\gdef\csname odunum@val@positive_taxonomy_m2.av.mean.qwen3_omni_think.axis3.3_1\endcsname{0.554}
\expandafter\gdef\csname odunum@n@positive_taxonomy_m2.av.mean.qwen3_omni_think.axis3.3_1\endcsname{91}
\expandafter\gdef\csname odunum@ci@positive_taxonomy_m2.av.mean.qwen3_omni_think.axis3.3_1\endcsname{[0.503, 0.606]}
\expandafter\gdef\csname odunum@val@positive_taxonomy_m2.av.mean.qwen3_omni_think.axis3.3_2\endcsname{0.517}
\expandafter\gdef\csname odunum@n@positive_taxonomy_m2.av.mean.qwen3_omni_think.axis3.3_2\endcsname{85}
\expandafter\gdef\csname odunum@ci@positive_taxonomy_m2.av.mean.qwen3_omni_think.axis3.3_2\endcsname{[0.465, 0.569]}
\expandafter\gdef\csname odunum@val@positive_taxonomy_m2.av.mean.qwen3_omni_think.axis3.3_3\endcsname{0.532}
\expandafter\gdef\csname odunum@n@positive_taxonomy_m2.av.mean.qwen3_omni_think.axis3.3_3\endcsname{137}
\expandafter\gdef\csname odunum@ci@positive_taxonomy_m2.av.mean.qwen3_omni_think.axis3.3_3\endcsname{[0.488, 0.577]}
\expandafter\gdef\csname odunum@val@positive_taxonomy_m2.av.mean.qwen3_omni_think.axis3.3_4\endcsname{0.571}
\expandafter\gdef\csname odunum@n@positive_taxonomy_m2.av.mean.qwen3_omni_think.axis3.3_4\endcsname{256}
\expandafter\gdef\csname odunum@ci@positive_taxonomy_m2.av.mean.qwen3_omni_think.axis3.3_4\endcsname{[0.538, 0.603]}
\expandafter\gdef\csname odunum@val@positive_taxonomy_m2.av.mean.qwen3_omni_think.axis3.3_5\endcsname{0.494}
\expandafter\gdef\csname odunum@n@positive_taxonomy_m2.av.mean.qwen3_omni_think.axis3.3_5\endcsname{185}
\expandafter\gdef\csname odunum@ci@positive_taxonomy_m2.av.mean.qwen3_omni_think.axis3.3_5\endcsname{[0.457, 0.533]}
\expandafter\gdef\csname odunum@val@positive_taxonomy_m2.av.mean.qwen3_omni_think.axis3.3_6\endcsname{0.607}
\expandafter\gdef\csname odunum@n@positive_taxonomy_m2.av.mean.qwen3_omni_think.axis3.3_6\endcsname{289}
\expandafter\gdef\csname odunum@ci@positive_taxonomy_m2.av.mean.qwen3_omni_think.axis3.3_6\endcsname{[0.577, 0.635]}
\expandafter\gdef\csname odunum@val@positive_taxonomy_m2.av.mean.qwen3_omni_think.axis4.4_1\endcsname{0.520}
\expandafter\gdef\csname odunum@n@positive_taxonomy_m2.av.mean.qwen3_omni_think.axis4.4_1\endcsname{137}
\expandafter\gdef\csname odunum@ci@positive_taxonomy_m2.av.mean.qwen3_omni_think.axis4.4_1\endcsname{[0.478, 0.560]}
\expandafter\gdef\csname odunum@val@positive_taxonomy_m2.av.mean.qwen3_omni_think.axis4.4_2\endcsname{0.589}
\expandafter\gdef\csname odunum@n@positive_taxonomy_m2.av.mean.qwen3_omni_think.axis4.4_2\endcsname{289}
\expandafter\gdef\csname odunum@ci@positive_taxonomy_m2.av.mean.qwen3_omni_think.axis4.4_2\endcsname{[0.559, 0.619]}
\expandafter\gdef\csname odunum@val@positive_taxonomy_m2.av.mean.qwen3_omni_think.axis4.4_3\endcsname{0.560}
\expandafter\gdef\csname odunum@n@positive_taxonomy_m2.av.mean.qwen3_omni_think.axis4.4_3\endcsname{120}
\expandafter\gdef\csname odunum@ci@positive_taxonomy_m2.av.mean.qwen3_omni_think.axis4.4_3\endcsname{[0.507, 0.612]}
\expandafter\gdef\csname odunum@val@positive_taxonomy_m2.av.mean.qwen3_omni_think.axis4.4_4\endcsname{0.538}
\expandafter\gdef\csname odunum@n@positive_taxonomy_m2.av.mean.qwen3_omni_think.axis4.4_4\endcsname{324}
\expandafter\gdef\csname odunum@ci@positive_taxonomy_m2.av.mean.qwen3_omni_think.axis4.4_4\endcsname{[0.509, 0.567]}
\expandafter\gdef\csname odunum@val@positive_taxonomy_m2.av.mean.qwen3_omni_think.axis4.4_5\endcsname{0.562}
\expandafter\gdef\csname odunum@n@positive_taxonomy_m2.av.mean.qwen3_omni_think.axis4.4_5\endcsname{173}
\expandafter\gdef\csname odunum@ci@positive_taxonomy_m2.av.mean.qwen3_omni_think.axis4.4_5\endcsname{[0.527, 0.597]}
\expandafter\gdef\csname odunum@val@positive_taxonomy_m2.av.mean.qwen_plus.axis1.1_1\endcsname{0.727}
\expandafter\gdef\csname odunum@n@positive_taxonomy_m2.av.mean.qwen_plus.axis1.1_1\endcsname{92}
\expandafter\gdef\csname odunum@ci@positive_taxonomy_m2.av.mean.qwen_plus.axis1.1_1\endcsname{[0.676, 0.778]}
\expandafter\gdef\csname odunum@val@positive_taxonomy_m2.av.mean.qwen_plus.axis1.1_2\endcsname{0.633}
\expandafter\gdef\csname odunum@n@positive_taxonomy_m2.av.mean.qwen_plus.axis1.1_2\endcsname{288}
\expandafter\gdef\csname odunum@ci@positive_taxonomy_m2.av.mean.qwen_plus.axis1.1_2\endcsname{[0.602, 0.664]}
\expandafter\gdef\csname odunum@val@positive_taxonomy_m2.av.mean.qwen_plus.axis1.1_3\endcsname{0.672}
\expandafter\gdef\csname odunum@n@positive_taxonomy_m2.av.mean.qwen_plus.axis1.1_3\endcsname{286}
\expandafter\gdef\csname odunum@ci@positive_taxonomy_m2.av.mean.qwen_plus.axis1.1_3\endcsname{[0.643, 0.702]}
\expandafter\gdef\csname odunum@val@positive_taxonomy_m2.av.mean.qwen_plus.axis1.1_4\endcsname{0.645}
\expandafter\gdef\csname odunum@n@positive_taxonomy_m2.av.mean.qwen_plus.axis1.1_4\endcsname{377}
\expandafter\gdef\csname odunum@ci@positive_taxonomy_m2.av.mean.qwen_plus.axis1.1_4\endcsname{[0.619, 0.671]}
\expandafter\gdef\csname odunum@val@positive_taxonomy_m2.av.mean.qwen_plus.axis2.2_1\endcsname{0.616}
\expandafter\gdef\csname odunum@n@positive_taxonomy_m2.av.mean.qwen_plus.axis2.2_1\endcsname{118}
\expandafter\gdef\csname odunum@ci@positive_taxonomy_m2.av.mean.qwen_plus.axis2.2_1\endcsname{[0.560, 0.670]}
\expandafter\gdef\csname odunum@val@positive_taxonomy_m2.av.mean.qwen_plus.axis2.2_2\endcsname{0.681}
\expandafter\gdef\csname odunum@n@positive_taxonomy_m2.av.mean.qwen_plus.axis2.2_2\endcsname{266}
\expandafter\gdef\csname odunum@ci@positive_taxonomy_m2.av.mean.qwen_plus.axis2.2_2\endcsname{[0.652, 0.709]}
\expandafter\gdef\csname odunum@val@positive_taxonomy_m2.av.mean.qwen_plus.axis2.2_3\endcsname{0.714}
\expandafter\gdef\csname odunum@n@positive_taxonomy_m2.av.mean.qwen_plus.axis2.2_3\endcsname{274}
\expandafter\gdef\csname odunum@ci@positive_taxonomy_m2.av.mean.qwen_plus.axis2.2_3\endcsname{[0.687, 0.741]}
\expandafter\gdef\csname odunum@val@positive_taxonomy_m2.av.mean.qwen_plus.axis2.2_4\endcsname{0.586}
\expandafter\gdef\csname odunum@n@positive_taxonomy_m2.av.mean.qwen_plus.axis2.2_4\endcsname{226}
\expandafter\gdef\csname odunum@ci@positive_taxonomy_m2.av.mean.qwen_plus.axis2.2_4\endcsname{[0.547, 0.623]}
\expandafter\gdef\csname odunum@val@positive_taxonomy_m2.av.mean.qwen_plus.axis2.2_5\endcsname{0.647}
\expandafter\gdef\csname odunum@n@positive_taxonomy_m2.av.mean.qwen_plus.axis2.2_5\endcsname{159}
\expandafter\gdef\csname odunum@ci@positive_taxonomy_m2.av.mean.qwen_plus.axis2.2_5\endcsname{[0.610, 0.684]}
\expandafter\gdef\csname odunum@val@positive_taxonomy_m2.av.mean.qwen_plus.axis3.3_1\endcsname{0.634}
\expandafter\gdef\csname odunum@n@positive_taxonomy_m2.av.mean.qwen_plus.axis3.3_1\endcsname{91}
\expandafter\gdef\csname odunum@ci@positive_taxonomy_m2.av.mean.qwen_plus.axis3.3_1\endcsname{[0.575, 0.691]}
\expandafter\gdef\csname odunum@val@positive_taxonomy_m2.av.mean.qwen_plus.axis3.3_2\endcsname{0.686}
\expandafter\gdef\csname odunum@n@positive_taxonomy_m2.av.mean.qwen_plus.axis3.3_2\endcsname{85}
\expandafter\gdef\csname odunum@ci@positive_taxonomy_m2.av.mean.qwen_plus.axis3.3_2\endcsname{[0.634, 0.735]}
\expandafter\gdef\csname odunum@val@positive_taxonomy_m2.av.mean.qwen_plus.axis3.3_3\endcsname{0.620}
\expandafter\gdef\csname odunum@n@positive_taxonomy_m2.av.mean.qwen_plus.axis3.3_3\endcsname{137}
\expandafter\gdef\csname odunum@ci@positive_taxonomy_m2.av.mean.qwen_plus.axis3.3_3\endcsname{[0.571, 0.669]}
\expandafter\gdef\csname odunum@val@positive_taxonomy_m2.av.mean.qwen_plus.axis3.3_4\endcsname{0.659}
\expandafter\gdef\csname odunum@n@positive_taxonomy_m2.av.mean.qwen_plus.axis3.3_4\endcsname{256}
\expandafter\gdef\csname odunum@ci@positive_taxonomy_m2.av.mean.qwen_plus.axis3.3_4\endcsname{[0.629, 0.690]}
\expandafter\gdef\csname odunum@val@positive_taxonomy_m2.av.mean.qwen_plus.axis3.3_5\endcsname{0.595}
\expandafter\gdef\csname odunum@n@positive_taxonomy_m2.av.mean.qwen_plus.axis3.3_5\endcsname{185}
\expandafter\gdef\csname odunum@ci@positive_taxonomy_m2.av.mean.qwen_plus.axis3.3_5\endcsname{[0.555, 0.635]}
\expandafter\gdef\csname odunum@val@positive_taxonomy_m2.av.mean.qwen_plus.axis3.3_6\endcsname{0.709}
\expandafter\gdef\csname odunum@n@positive_taxonomy_m2.av.mean.qwen_plus.axis3.3_6\endcsname{289}
\expandafter\gdef\csname odunum@ci@positive_taxonomy_m2.av.mean.qwen_plus.axis3.3_6\endcsname{[0.682, 0.735]}
\expandafter\gdef\csname odunum@val@positive_taxonomy_m2.av.mean.qwen_plus.axis4.4_1\endcsname{0.656}
\expandafter\gdef\csname odunum@n@positive_taxonomy_m2.av.mean.qwen_plus.axis4.4_1\endcsname{137}
\expandafter\gdef\csname odunum@ci@positive_taxonomy_m2.av.mean.qwen_plus.axis4.4_1\endcsname{[0.612, 0.697]}
\expandafter\gdef\csname odunum@val@positive_taxonomy_m2.av.mean.qwen_plus.axis4.4_2\endcsname{0.659}
\expandafter\gdef\csname odunum@n@positive_taxonomy_m2.av.mean.qwen_plus.axis4.4_2\endcsname{289}
\expandafter\gdef\csname odunum@ci@positive_taxonomy_m2.av.mean.qwen_plus.axis4.4_2\endcsname{[0.630, 0.687]}
\expandafter\gdef\csname odunum@val@positive_taxonomy_m2.av.mean.qwen_plus.axis4.4_3\endcsname{0.653}
\expandafter\gdef\csname odunum@n@positive_taxonomy_m2.av.mean.qwen_plus.axis4.4_3\endcsname{120}
\expandafter\gdef\csname odunum@ci@positive_taxonomy_m2.av.mean.qwen_plus.axis4.4_3\endcsname{[0.600, 0.704]}
\expandafter\gdef\csname odunum@val@positive_taxonomy_m2.av.mean.qwen_plus.axis4.4_4\endcsname{0.646}
\expandafter\gdef\csname odunum@n@positive_taxonomy_m2.av.mean.qwen_plus.axis4.4_4\endcsname{324}
\expandafter\gdef\csname odunum@ci@positive_taxonomy_m2.av.mean.qwen_plus.axis4.4_4\endcsname{[0.617, 0.675]}
\expandafter\gdef\csname odunum@val@positive_taxonomy_m2.av.mean.qwen_plus.axis4.4_5\endcsname{0.676}
\expandafter\gdef\csname odunum@n@positive_taxonomy_m2.av.mean.qwen_plus.axis4.4_5\endcsname{173}
\expandafter\gdef\csname odunum@ci@positive_taxonomy_m2.av.mean.qwen_plus.axis4.4_5\endcsname{[0.639, 0.713]}
\expandafter\gdef\csname odunum@val@positive_taxonomy_m2.av.mean.salmonn2_7b.axis1.1_1\endcsname{0.051}
\expandafter\gdef\csname odunum@n@positive_taxonomy_m2.av.mean.salmonn2_7b.axis1.1_1\endcsname{92}
\expandafter\gdef\csname odunum@ci@positive_taxonomy_m2.av.mean.salmonn2_7b.axis1.1_1\endcsname{[0.024, 0.081]}
\expandafter\gdef\csname odunum@val@positive_taxonomy_m2.av.mean.salmonn2_7b.axis1.1_2\endcsname{0.080}
\expandafter\gdef\csname odunum@n@positive_taxonomy_m2.av.mean.salmonn2_7b.axis1.1_2\endcsname{288}
\expandafter\gdef\csname odunum@ci@positive_taxonomy_m2.av.mean.salmonn2_7b.axis1.1_2\endcsname{[0.060, 0.100]}
\expandafter\gdef\csname odunum@val@positive_taxonomy_m2.av.mean.salmonn2_7b.axis1.1_3\endcsname{0.056}
\expandafter\gdef\csname odunum@n@positive_taxonomy_m2.av.mean.salmonn2_7b.axis1.1_3\endcsname{286}
\expandafter\gdef\csname odunum@ci@positive_taxonomy_m2.av.mean.salmonn2_7b.axis1.1_3\endcsname{[0.040, 0.074]}
\expandafter\gdef\csname odunum@val@positive_taxonomy_m2.av.mean.salmonn2_7b.axis1.1_4\endcsname{0.069}
\expandafter\gdef\csname odunum@n@positive_taxonomy_m2.av.mean.salmonn2_7b.axis1.1_4\endcsname{377}
\expandafter\gdef\csname odunum@ci@positive_taxonomy_m2.av.mean.salmonn2_7b.axis1.1_4\endcsname{[0.052, 0.086]}
\expandafter\gdef\csname odunum@val@positive_taxonomy_m2.av.mean.salmonn2_7b.axis2.2_1\endcsname{0.075}
\expandafter\gdef\csname odunum@n@positive_taxonomy_m2.av.mean.salmonn2_7b.axis2.2_1\endcsname{118}
\expandafter\gdef\csname odunum@ci@positive_taxonomy_m2.av.mean.salmonn2_7b.axis2.2_1\endcsname{[0.048, 0.104]}
\expandafter\gdef\csname odunum@val@positive_taxonomy_m2.av.mean.salmonn2_7b.axis2.2_2\endcsname{0.047}
\expandafter\gdef\csname odunum@n@positive_taxonomy_m2.av.mean.salmonn2_7b.axis2.2_2\endcsname{266}
\expandafter\gdef\csname odunum@ci@positive_taxonomy_m2.av.mean.salmonn2_7b.axis2.2_2\endcsname{[0.031, 0.064]}
\expandafter\gdef\csname odunum@val@positive_taxonomy_m2.av.mean.salmonn2_7b.axis2.2_3\endcsname{0.032}
\expandafter\gdef\csname odunum@n@positive_taxonomy_m2.av.mean.salmonn2_7b.axis2.2_3\endcsname{274}
\expandafter\gdef\csname odunum@ci@positive_taxonomy_m2.av.mean.salmonn2_7b.axis2.2_3\endcsname{[0.020, 0.046]}
\expandafter\gdef\csname odunum@val@positive_taxonomy_m2.av.mean.salmonn2_7b.axis2.2_4\endcsname{0.109}
\expandafter\gdef\csname odunum@n@positive_taxonomy_m2.av.mean.salmonn2_7b.axis2.2_4\endcsname{226}
\expandafter\gdef\csname odunum@ci@positive_taxonomy_m2.av.mean.salmonn2_7b.axis2.2_4\endcsname{[0.082, 0.137]}
\expandafter\gdef\csname odunum@val@positive_taxonomy_m2.av.mean.salmonn2_7b.axis2.2_5\endcsname{0.093}
\expandafter\gdef\csname odunum@n@positive_taxonomy_m2.av.mean.salmonn2_7b.axis2.2_5\endcsname{159}
\expandafter\gdef\csname odunum@ci@positive_taxonomy_m2.av.mean.salmonn2_7b.axis2.2_5\endcsname{[0.066, 0.122]}
\expandafter\gdef\csname odunum@val@positive_taxonomy_m2.av.mean.salmonn2_7b.axis3.3_1\endcsname{0.101}
\expandafter\gdef\csname odunum@n@positive_taxonomy_m2.av.mean.salmonn2_7b.axis3.3_1\endcsname{91}
\expandafter\gdef\csname odunum@ci@positive_taxonomy_m2.av.mean.salmonn2_7b.axis3.3_1\endcsname{[0.060, 0.148]}
\expandafter\gdef\csname odunum@val@positive_taxonomy_m2.av.mean.salmonn2_7b.axis3.3_2\endcsname{0.057}
\expandafter\gdef\csname odunum@n@positive_taxonomy_m2.av.mean.salmonn2_7b.axis3.3_2\endcsname{85}
\expandafter\gdef\csname odunum@ci@positive_taxonomy_m2.av.mean.salmonn2_7b.axis3.3_2\endcsname{[0.029, 0.088]}
\expandafter\gdef\csname odunum@val@positive_taxonomy_m2.av.mean.salmonn2_7b.axis3.3_3\endcsname{0.044}
\expandafter\gdef\csname odunum@n@positive_taxonomy_m2.av.mean.salmonn2_7b.axis3.3_3\endcsname{137}
\expandafter\gdef\csname odunum@ci@positive_taxonomy_m2.av.mean.salmonn2_7b.axis3.3_3\endcsname{[0.022, 0.069]}
\expandafter\gdef\csname odunum@val@positive_taxonomy_m2.av.mean.salmonn2_7b.axis3.3_4\endcsname{0.043}
\expandafter\gdef\csname odunum@n@positive_taxonomy_m2.av.mean.salmonn2_7b.axis3.3_4\endcsname{256}
\expandafter\gdef\csname odunum@ci@positive_taxonomy_m2.av.mean.salmonn2_7b.axis3.3_4\endcsname{[0.026, 0.061]}
\expandafter\gdef\csname odunum@val@positive_taxonomy_m2.av.mean.salmonn2_7b.axis3.3_5\endcsname{0.122}
\expandafter\gdef\csname odunum@n@positive_taxonomy_m2.av.mean.salmonn2_7b.axis3.3_5\endcsname{185}
\expandafter\gdef\csname odunum@ci@positive_taxonomy_m2.av.mean.salmonn2_7b.axis3.3_5\endcsname{[0.094, 0.152]}
\expandafter\gdef\csname odunum@val@positive_taxonomy_m2.av.mean.salmonn2_7b.axis3.3_6\endcsname{0.055}
\expandafter\gdef\csname odunum@n@positive_taxonomy_m2.av.mean.salmonn2_7b.axis3.3_6\endcsname{289}
\expandafter\gdef\csname odunum@ci@positive_taxonomy_m2.av.mean.salmonn2_7b.axis3.3_6\endcsname{[0.040, 0.071]}
\expandafter\gdef\csname odunum@val@positive_taxonomy_m2.av.mean.salmonn2_7b.axis4.4_1\endcsname{0.033}
\expandafter\gdef\csname odunum@n@positive_taxonomy_m2.av.mean.salmonn2_7b.axis4.4_1\endcsname{137}
\expandafter\gdef\csname odunum@ci@positive_taxonomy_m2.av.mean.salmonn2_7b.axis4.4_1\endcsname{[0.016, 0.052]}
\expandafter\gdef\csname odunum@val@positive_taxonomy_m2.av.mean.salmonn2_7b.axis4.4_2\endcsname{0.092}
\expandafter\gdef\csname odunum@n@positive_taxonomy_m2.av.mean.salmonn2_7b.axis4.4_2\endcsname{289}
\expandafter\gdef\csname odunum@ci@positive_taxonomy_m2.av.mean.salmonn2_7b.axis4.4_2\endcsname{[0.072, 0.113]}
\expandafter\gdef\csname odunum@val@positive_taxonomy_m2.av.mean.salmonn2_7b.axis4.4_3\endcsname{0.057}
\expandafter\gdef\csname odunum@n@positive_taxonomy_m2.av.mean.salmonn2_7b.axis4.4_3\endcsname{120}
\expandafter\gdef\csname odunum@ci@positive_taxonomy_m2.av.mean.salmonn2_7b.axis4.4_3\endcsname{[0.032, 0.086]}
\expandafter\gdef\csname odunum@val@positive_taxonomy_m2.av.mean.salmonn2_7b.axis4.4_4\endcsname{0.058}
\expandafter\gdef\csname odunum@n@positive_taxonomy_m2.av.mean.salmonn2_7b.axis4.4_4\endcsname{324}
\expandafter\gdef\csname odunum@ci@positive_taxonomy_m2.av.mean.salmonn2_7b.axis4.4_4\endcsname{[0.041, 0.076]}
\expandafter\gdef\csname odunum@val@positive_taxonomy_m2.av.mean.salmonn2_7b.axis4.4_5\endcsname{0.074}
\expandafter\gdef\csname odunum@n@positive_taxonomy_m2.av.mean.salmonn2_7b.axis4.4_5\endcsname{173}
\expandafter\gdef\csname odunum@ci@positive_taxonomy_m2.av.mean.salmonn2_7b.axis4.4_5\endcsname{[0.049, 0.101]}
\expandafter\gdef\csname odunum@val@positive_taxonomy_m2.av.mean.seed.axis1.1_1\endcsname{0.642}
\expandafter\gdef\csname odunum@n@positive_taxonomy_m2.av.mean.seed.axis1.1_1\endcsname{92}
\expandafter\gdef\csname odunum@ci@positive_taxonomy_m2.av.mean.seed.axis1.1_1\endcsname{[0.572, 0.710]}
\expandafter\gdef\csname odunum@val@positive_taxonomy_m2.av.mean.seed.axis1.1_2\endcsname{0.634}
\expandafter\gdef\csname odunum@n@positive_taxonomy_m2.av.mean.seed.axis1.1_2\endcsname{288}
\expandafter\gdef\csname odunum@ci@positive_taxonomy_m2.av.mean.seed.axis1.1_2\endcsname{[0.599, 0.668]}
\expandafter\gdef\csname odunum@val@positive_taxonomy_m2.av.mean.seed.axis1.1_3\endcsname{0.655}
\expandafter\gdef\csname odunum@n@positive_taxonomy_m2.av.mean.seed.axis1.1_3\endcsname{286}
\expandafter\gdef\csname odunum@ci@positive_taxonomy_m2.av.mean.seed.axis1.1_3\endcsname{[0.617, 0.693]}
\expandafter\gdef\csname odunum@val@positive_taxonomy_m2.av.mean.seed.axis1.1_4\endcsname{0.637}
\expandafter\gdef\csname odunum@n@positive_taxonomy_m2.av.mean.seed.axis1.1_4\endcsname{377}
\expandafter\gdef\csname odunum@ci@positive_taxonomy_m2.av.mean.seed.axis1.1_4\endcsname{[0.606, 0.667]}
\expandafter\gdef\csname odunum@val@positive_taxonomy_m2.av.mean.seed.axis2.2_1\endcsname{0.638}
\expandafter\gdef\csname odunum@n@positive_taxonomy_m2.av.mean.seed.axis2.2_1\endcsname{118}
\expandafter\gdef\csname odunum@ci@positive_taxonomy_m2.av.mean.seed.axis2.2_1\endcsname{[0.578, 0.698]}
\expandafter\gdef\csname odunum@val@positive_taxonomy_m2.av.mean.seed.axis2.2_2\endcsname{0.645}
\expandafter\gdef\csname odunum@n@positive_taxonomy_m2.av.mean.seed.axis2.2_2\endcsname{266}
\expandafter\gdef\csname odunum@ci@positive_taxonomy_m2.av.mean.seed.axis2.2_2\endcsname{[0.608, 0.680]}
\expandafter\gdef\csname odunum@val@positive_taxonomy_m2.av.mean.seed.axis2.2_3\endcsname{0.658}
\expandafter\gdef\csname odunum@n@positive_taxonomy_m2.av.mean.seed.axis2.2_3\endcsname{274}
\expandafter\gdef\csname odunum@ci@positive_taxonomy_m2.av.mean.seed.axis2.2_3\endcsname{[0.622, 0.692]}
\expandafter\gdef\csname odunum@val@positive_taxonomy_m2.av.mean.seed.axis2.2_4\endcsname{0.613}
\expandafter\gdef\csname odunum@n@positive_taxonomy_m2.av.mean.seed.axis2.2_4\endcsname{226}
\expandafter\gdef\csname odunum@ci@positive_taxonomy_m2.av.mean.seed.axis2.2_4\endcsname{[0.567, 0.658]}
\expandafter\gdef\csname odunum@val@positive_taxonomy_m2.av.mean.seed.axis2.2_5\endcsname{0.651}
\expandafter\gdef\csname odunum@n@positive_taxonomy_m2.av.mean.seed.axis2.2_5\endcsname{159}
\expandafter\gdef\csname odunum@ci@positive_taxonomy_m2.av.mean.seed.axis2.2_5\endcsname{[0.604, 0.696]}
\expandafter\gdef\csname odunum@val@positive_taxonomy_m2.av.mean.seed.axis3.3_1\endcsname{0.632}
\expandafter\gdef\csname odunum@n@positive_taxonomy_m2.av.mean.seed.axis3.3_1\endcsname{91}
\expandafter\gdef\csname odunum@ci@positive_taxonomy_m2.av.mean.seed.axis3.3_1\endcsname{[0.564, 0.701]}
\expandafter\gdef\csname odunum@val@positive_taxonomy_m2.av.mean.seed.axis3.3_2\endcsname{0.648}
\expandafter\gdef\csname odunum@n@positive_taxonomy_m2.av.mean.seed.axis3.3_2\endcsname{85}
\expandafter\gdef\csname odunum@ci@positive_taxonomy_m2.av.mean.seed.axis3.3_2\endcsname{[0.580, 0.714]}
\expandafter\gdef\csname odunum@val@positive_taxonomy_m2.av.mean.seed.axis3.3_3\endcsname{0.620}
\expandafter\gdef\csname odunum@n@positive_taxonomy_m2.av.mean.seed.axis3.3_3\endcsname{137}
\expandafter\gdef\csname odunum@ci@positive_taxonomy_m2.av.mean.seed.axis3.3_3\endcsname{[0.568, 0.671]}
\expandafter\gdef\csname odunum@val@positive_taxonomy_m2.av.mean.seed.axis3.3_4\endcsname{0.624}
\expandafter\gdef\csname odunum@n@positive_taxonomy_m2.av.mean.seed.axis3.3_4\endcsname{256}
\expandafter\gdef\csname odunum@ci@positive_taxonomy_m2.av.mean.seed.axis3.3_4\endcsname{[0.584, 0.664]}
\expandafter\gdef\csname odunum@val@positive_taxonomy_m2.av.mean.seed.axis3.3_5\endcsname{0.596}
\expandafter\gdef\csname odunum@n@positive_taxonomy_m2.av.mean.seed.axis3.3_5\endcsname{185}
\expandafter\gdef\csname odunum@ci@positive_taxonomy_m2.av.mean.seed.axis3.3_5\endcsname{[0.554, 0.639]}
\expandafter\gdef\csname odunum@val@positive_taxonomy_m2.av.mean.seed.axis3.3_6\endcsname{0.697}
\expandafter\gdef\csname odunum@n@positive_taxonomy_m2.av.mean.seed.axis3.3_6\endcsname{289}
\expandafter\gdef\csname odunum@ci@positive_taxonomy_m2.av.mean.seed.axis3.3_6\endcsname{[0.662, 0.732]}
\expandafter\gdef\csname odunum@val@positive_taxonomy_m2.av.mean.seed.axis4.4_1\endcsname{0.589}
\expandafter\gdef\csname odunum@n@positive_taxonomy_m2.av.mean.seed.axis4.4_1\endcsname{137}
\expandafter\gdef\csname odunum@ci@positive_taxonomy_m2.av.mean.seed.axis4.4_1\endcsname{[0.536, 0.640]}
\expandafter\gdef\csname odunum@val@positive_taxonomy_m2.av.mean.seed.axis4.4_2\endcsname{0.666}
\expandafter\gdef\csname odunum@n@positive_taxonomy_m2.av.mean.seed.axis4.4_2\endcsname{289}
\expandafter\gdef\csname odunum@ci@positive_taxonomy_m2.av.mean.seed.axis4.4_2\endcsname{[0.633, 0.699]}
\expandafter\gdef\csname odunum@val@positive_taxonomy_m2.av.mean.seed.axis4.4_3\endcsname{0.700}
\expandafter\gdef\csname odunum@n@positive_taxonomy_m2.av.mean.seed.axis4.4_3\endcsname{120}
\expandafter\gdef\csname odunum@ci@positive_taxonomy_m2.av.mean.seed.axis4.4_3\endcsname{[0.643, 0.756]}
\expandafter\gdef\csname odunum@val@positive_taxonomy_m2.av.mean.seed.axis4.4_4\endcsname{0.617}
\expandafter\gdef\csname odunum@n@positive_taxonomy_m2.av.mean.seed.axis4.4_4\endcsname{324}
\expandafter\gdef\csname odunum@ci@positive_taxonomy_m2.av.mean.seed.axis4.4_4\endcsname{[0.582, 0.651]}
\expandafter\gdef\csname odunum@val@positive_taxonomy_m2.av.mean.seed.axis4.4_5\endcsname{0.649}
\expandafter\gdef\csname odunum@n@positive_taxonomy_m2.av.mean.seed.axis4.4_5\endcsname{173}
\expandafter\gdef\csname odunum@ci@positive_taxonomy_m2.av.mean.seed.axis4.4_5\endcsname{[0.599, 0.696]}
\expandafter\gdef\csname odunum@val@positive_taxonomy_m2.av.n_models\endcsname{13}
\expandafter\gdef\csname odunum@n@positive_taxonomy_m2.av.n_models\endcsname{13}
\expandafter\gdef\csname odunum@val@positive_taxonomy_m2.av.n_scenes\endcsname{1\,043}
\expandafter\gdef\csname odunum@n@positive_taxonomy_m2.av.n_scenes\endcsname{1\,061}
\expandafter\gdef\csname odunum@val@positive_taxonomy_m2.av.overall.cascade_asr\endcsname{0.554}
\expandafter\gdef\csname odunum@n@positive_taxonomy_m2.av.overall.cascade_asr\endcsname{1\,043}
\expandafter\gdef\csname odunum@ci@positive_taxonomy_m2.av.overall.cascade_asr\endcsname{[0.536, 0.571]}
\expandafter\gdef\csname odunum@val@positive_taxonomy_m2.av.overall.gemini\endcsname{0.634}
\expandafter\gdef\csname odunum@n@positive_taxonomy_m2.av.overall.gemini\endcsname{1\,043}
\expandafter\gdef\csname odunum@ci@positive_taxonomy_m2.av.overall.gemini\endcsname{[0.619, 0.649]}
\expandafter\gdef\csname odunum@val@positive_taxonomy_m2.av.overall.gemini35_flash_lite\endcsname{0.500}
\expandafter\gdef\csname odunum@n@positive_taxonomy_m2.av.overall.gemini35_flash_lite\endcsname{1\,043}
\expandafter\gdef\csname odunum@ci@positive_taxonomy_m2.av.overall.gemini35_flash_lite\endcsname{[0.480, 0.519]}
\expandafter\gdef\csname odunum@val@positive_taxonomy_m2.av.overall.gemini37_flash\endcsname{0.601}
\expandafter\gdef\csname odunum@n@positive_taxonomy_m2.av.overall.gemini37_flash\endcsname{1\,043}
\expandafter\gdef\csname odunum@ci@positive_taxonomy_m2.av.overall.gemini37_flash\endcsname{[0.584, 0.619]}
\expandafter\gdef\csname odunum@val@positive_taxonomy_m2.av.overall.ming\endcsname{0.496}
\expandafter\gdef\csname odunum@n@positive_taxonomy_m2.av.overall.ming\endcsname{1\,043}
\expandafter\gdef\csname odunum@ci@positive_taxonomy_m2.av.overall.ming\endcsname{[0.479, 0.511]}
\expandafter\gdef\csname odunum@val@positive_taxonomy_m2.av.overall.minicpm_o\endcsname{0.355}
\expandafter\gdef\csname odunum@n@positive_taxonomy_m2.av.overall.minicpm_o\endcsname{1\,043}
\expandafter\gdef\csname odunum@ci@positive_taxonomy_m2.av.overall.minicpm_o\endcsname{[0.336, 0.373]}
\expandafter\gdef\csname odunum@val@positive_taxonomy_m2.av.overall.nemotron\endcsname{0.266}
\expandafter\gdef\csname odunum@n@positive_taxonomy_m2.av.overall.nemotron\endcsname{1\,043}
\expandafter\gdef\csname odunum@ci@positive_taxonomy_m2.av.overall.nemotron\endcsname{[0.247, 0.285]}
\expandafter\gdef\csname odunum@val@positive_taxonomy_m2.av.overall.qwen25_omni\endcsname{0.421}
\expandafter\gdef\csname odunum@n@positive_taxonomy_m2.av.overall.qwen25_omni\endcsname{1\,043}
\expandafter\gdef\csname odunum@ci@positive_taxonomy_m2.av.overall.qwen25_omni\endcsname{[0.403, 0.438]}
\expandafter\gdef\csname odunum@val@positive_taxonomy_m2.av.overall.qwen3_omni_instruct\endcsname{0.531}
\expandafter\gdef\csname odunum@n@positive_taxonomy_m2.av.overall.qwen3_omni_instruct\endcsname{1\,043}
\expandafter\gdef\csname odunum@ci@positive_taxonomy_m2.av.overall.qwen3_omni_instruct\endcsname{[0.514, 0.547]}
\expandafter\gdef\csname odunum@val@positive_taxonomy_m2.av.overall.qwen3_omni_think\endcsname{0.556}
\expandafter\gdef\csname odunum@n@positive_taxonomy_m2.av.overall.qwen3_omni_think\endcsname{1\,043}
\expandafter\gdef\csname odunum@ci@positive_taxonomy_m2.av.overall.qwen3_omni_think\endcsname{[0.540, 0.572]}
\expandafter\gdef\csname odunum@val@positive_taxonomy_m2.av.overall.qwen_plus\endcsname{0.657}
\expandafter\gdef\csname odunum@n@positive_taxonomy_m2.av.overall.qwen_plus\endcsname{1\,043}
\expandafter\gdef\csname odunum@ci@positive_taxonomy_m2.av.overall.qwen_plus\endcsname{[0.641, 0.672]}
\expandafter\gdef\csname odunum@val@positive_taxonomy_m2.av.overall.salmonn2_7b\endcsname{0.067}
\expandafter\gdef\csname odunum@n@positive_taxonomy_m2.av.overall.salmonn2_7b\endcsname{1\,043}
\expandafter\gdef\csname odunum@ci@positive_taxonomy_m2.av.overall.salmonn2_7b\endcsname{[0.057, 0.077]}
\expandafter\gdef\csname odunum@val@positive_taxonomy_m2.av.overall.seed\endcsname{0.642}
\expandafter\gdef\csname odunum@n@positive_taxonomy_m2.av.overall.seed\endcsname{1\,043}
\expandafter\gdef\csname odunum@ci@positive_taxonomy_m2.av.overall.seed\endcsname{[0.622, 0.660]}
\expandafter\gdef\csname odunum@val@positive_taxonomy_m2.av.panel_delta.axis1.1_1\endcsname{0.022}
\expandafter\gdef\csname odunum@n@positive_taxonomy_m2.av.panel_delta.axis1.1_1\endcsname{92}
\expandafter\gdef\csname odunum@ci@positive_taxonomy_m2.av.panel_delta.axis1.1_1\endcsname{[-0.007, 0.049]}
\expandafter\gdef\csname odunum@val@positive_taxonomy_m2.av.panel_delta.axis1.1_2\endcsname{-0.004}
\expandafter\gdef\csname odunum@n@positive_taxonomy_m2.av.panel_delta.axis1.1_2\endcsname{288}
\expandafter\gdef\csname odunum@ci@positive_taxonomy_m2.av.panel_delta.axis1.1_2\endcsname{[-0.018, 0.011]}
\expandafter\gdef\csname odunum@val@positive_taxonomy_m2.av.panel_delta.axis1.1_3\endcsname{0.003}
\expandafter\gdef\csname odunum@n@positive_taxonomy_m2.av.panel_delta.axis1.1_3\endcsname{286}
\expandafter\gdef\csname odunum@ci@positive_taxonomy_m2.av.panel_delta.axis1.1_3\endcsname{[-0.012, 0.018]}
\expandafter\gdef\csname odunum@val@positive_taxonomy_m2.av.panel_delta.axis1.1_4\endcsname{-0.005}
\expandafter\gdef\csname odunum@n@positive_taxonomy_m2.av.panel_delta.axis1.1_4\endcsname{377}
\expandafter\gdef\csname odunum@ci@positive_taxonomy_m2.av.panel_delta.axis1.1_4\endcsname{[-0.017, 0.007]}
\expandafter\gdef\csname odunum@val@positive_taxonomy_m2.av.panel_delta.axis2.2_1\endcsname{-0.026}
\expandafter\gdef\csname odunum@n@positive_taxonomy_m2.av.panel_delta.axis2.2_1\endcsname{118}
\expandafter\gdef\csname odunum@ci@positive_taxonomy_m2.av.panel_delta.axis2.2_1\endcsname{[-0.056, 0.004]}
\expandafter\gdef\csname odunum@val@positive_taxonomy_m2.av.panel_delta.axis2.2_2\endcsname{0.016}
\expandafter\gdef\csname odunum@n@positive_taxonomy_m2.av.panel_delta.axis2.2_2\endcsname{266}
\expandafter\gdef\csname odunum@ci@positive_taxonomy_m2.av.panel_delta.axis2.2_2\endcsname{[0.001, 0.031]}
\expandafter\gdef\csname odunum@val@positive_taxonomy_m2.av.panel_delta.axis2.2_3\endcsname{0.020}
\expandafter\gdef\csname odunum@n@positive_taxonomy_m2.av.panel_delta.axis2.2_3\endcsname{274}
\expandafter\gdef\csname odunum@ci@positive_taxonomy_m2.av.panel_delta.axis2.2_3\endcsname{[0.005, 0.035]}
\expandafter\gdef\csname odunum@val@positive_taxonomy_m2.av.panel_delta.axis2.2_4\endcsname{-0.020}
\expandafter\gdef\csname odunum@n@positive_taxonomy_m2.av.panel_delta.axis2.2_4\endcsname{226}
\expandafter\gdef\csname odunum@ci@positive_taxonomy_m2.av.panel_delta.axis2.2_4\endcsname{[-0.039, -0.002]}
\expandafter\gdef\csname odunum@val@positive_taxonomy_m2.av.panel_delta.axis2.2_5\endcsname{-0.013}
\expandafter\gdef\csname odunum@n@positive_taxonomy_m2.av.panel_delta.axis2.2_5\endcsname{159}
\expandafter\gdef\csname odunum@ci@positive_taxonomy_m2.av.panel_delta.axis2.2_5\endcsname{[-0.034, 0.007]}
\expandafter\gdef\csname odunum@val@positive_taxonomy_m2.av.panel_delta.axis3.3_1\endcsname{0.017}
\expandafter\gdef\csname odunum@n@positive_taxonomy_m2.av.panel_delta.axis3.3_1\endcsname{91}
\expandafter\gdef\csname odunum@ci@positive_taxonomy_m2.av.panel_delta.axis3.3_1\endcsname{[-0.014, 0.049]}
\expandafter\gdef\csname odunum@val@positive_taxonomy_m2.av.panel_delta.axis3.3_2\endcsname{0.002}
\expandafter\gdef\csname odunum@n@positive_taxonomy_m2.av.panel_delta.axis3.3_2\endcsname{85}
\expandafter\gdef\csname odunum@ci@positive_taxonomy_m2.av.panel_delta.axis3.3_2\endcsname{[-0.030, 0.033]}
\expandafter\gdef\csname odunum@val@positive_taxonomy_m2.av.panel_delta.axis3.3_3\endcsname{-0.072}
\expandafter\gdef\csname odunum@n@positive_taxonomy_m2.av.panel_delta.axis3.3_3\endcsname{137}
\expandafter\gdef\csname odunum@ci@positive_taxonomy_m2.av.panel_delta.axis3.3_3\endcsname{[-0.096, -0.047]}
\expandafter\gdef\csname odunum@val@positive_taxonomy_m2.av.panel_delta.axis3.3_4\endcsname{0.010}
\expandafter\gdef\csname odunum@n@positive_taxonomy_m2.av.panel_delta.axis3.3_4\endcsname{256}
\expandafter\gdef\csname odunum@ci@positive_taxonomy_m2.av.panel_delta.axis3.3_4\endcsname{[-0.005, 0.026]}
\expandafter\gdef\csname odunum@val@positive_taxonomy_m2.av.panel_delta.axis3.3_5\endcsname{-0.032}
\expandafter\gdef\csname odunum@n@positive_taxonomy_m2.av.panel_delta.axis3.3_5\endcsname{185}
\expandafter\gdef\csname odunum@ci@positive_taxonomy_m2.av.panel_delta.axis3.3_5\endcsname{[-0.052, -0.012]}
\expandafter\gdef\csname odunum@val@positive_taxonomy_m2.av.panel_delta.axis3.3_6\endcsname{0.040}
\expandafter\gdef\csname odunum@n@positive_taxonomy_m2.av.panel_delta.axis3.3_6\endcsname{289}
\expandafter\gdef\csname odunum@ci@positive_taxonomy_m2.av.panel_delta.axis3.3_6\endcsname{[0.025, 0.054]}
\expandafter\gdef\csname odunum@val@positive_taxonomy_m2.av.panel_delta.axis4.4_1\endcsname{-0.012}
\expandafter\gdef\csname odunum@n@positive_taxonomy_m2.av.panel_delta.axis4.4_1\endcsname{137}
\expandafter\gdef\csname odunum@ci@positive_taxonomy_m2.av.panel_delta.axis4.4_1\endcsname{[-0.034, 0.010]}
\expandafter\gdef\csname odunum@val@positive_taxonomy_m2.av.panel_delta.axis4.4_2\endcsname{0.019}
\expandafter\gdef\csname odunum@n@positive_taxonomy_m2.av.panel_delta.axis4.4_2\endcsname{289}
\expandafter\gdef\csname odunum@ci@positive_taxonomy_m2.av.panel_delta.axis4.4_2\endcsname{[0.005, 0.033]}
\expandafter\gdef\csname odunum@val@positive_taxonomy_m2.av.panel_delta.axis4.4_3\endcsname{-0.021}
\expandafter\gdef\csname odunum@n@positive_taxonomy_m2.av.panel_delta.axis4.4_3\endcsname{120}
\expandafter\gdef\csname odunum@ci@positive_taxonomy_m2.av.panel_delta.axis4.4_3\endcsname{[-0.049, 0.007]}
\expandafter\gdef\csname odunum@val@positive_taxonomy_m2.av.panel_delta.axis4.4_4\endcsname{-0.011}
\expandafter\gdef\csname odunum@n@positive_taxonomy_m2.av.panel_delta.axis4.4_4\endcsname{324}
\expandafter\gdef\csname odunum@ci@positive_taxonomy_m2.av.panel_delta.axis4.4_4\endcsname{[-0.024, 0.003]}
\expandafter\gdef\csname odunum@val@positive_taxonomy_m2.av.panel_delta.axis4.4_5\endcsname{0.013}
\expandafter\gdef\csname odunum@n@positive_taxonomy_m2.av.panel_delta.axis4.4_5\endcsname{173}
\expandafter\gdef\csname odunum@ci@positive_taxonomy_m2.av.panel_delta.axis4.4_5\endcsname{[-0.008, 0.034]}
\expandafter\gdef\csname odunum@val@positive_taxonomy_m2.population.positive\endcsname{1\,631}
\expandafter\gdef\csname odunum@n@positive_taxonomy_m2.population.positive\endcsname{2\,078}
\expandafter\gdef\csname odunum@val@recorded_vs_generated.avg.marginal.cascade_asr.generated\endcsname{0.646}
\expandafter\gdef\csname odunum@n@recorded_vs_generated.avg.marginal.cascade_asr.generated\endcsname{1\,801}
\expandafter\gdef\csname odunum@val@recorded_vs_generated.avg.marginal.cascade_asr.recorded\endcsname{0.557}
\expandafter\gdef\csname odunum@n@recorded_vs_generated.avg.marginal.cascade_asr.recorded\endcsname{277}
\expandafter\gdef\csname odunum@val@recorded_vs_generated.avg.marginal.gemini.generated\endcsname{0.735}
\expandafter\gdef\csname odunum@n@recorded_vs_generated.avg.marginal.gemini.generated\endcsname{1\,801}
\expandafter\gdef\csname odunum@val@recorded_vs_generated.avg.marginal.gemini.recorded\endcsname{0.742}
\expandafter\gdef\csname odunum@n@recorded_vs_generated.avg.marginal.gemini.recorded\endcsname{277}
\expandafter\gdef\csname odunum@val@recorded_vs_generated.avg.marginal.gemini35_flash_lite.generated\endcsname{0.609}
\expandafter\gdef\csname odunum@n@recorded_vs_generated.avg.marginal.gemini35_flash_lite.generated\endcsname{1\,801}
\expandafter\gdef\csname odunum@val@recorded_vs_generated.avg.marginal.gemini35_flash_lite.recorded\endcsname{0.471}
\expandafter\gdef\csname odunum@n@recorded_vs_generated.avg.marginal.gemini35_flash_lite.recorded\endcsname{277}
\expandafter\gdef\csname odunum@val@recorded_vs_generated.avg.marginal.gemini37_flash.generated\endcsname{0.708}
\expandafter\gdef\csname odunum@n@recorded_vs_generated.avg.marginal.gemini37_flash.generated\endcsname{1\,792}
\expandafter\gdef\csname odunum@val@recorded_vs_generated.avg.marginal.gemini37_flash.recorded\endcsname{0.699}
\expandafter\gdef\csname odunum@n@recorded_vs_generated.avg.marginal.gemini37_flash.recorded\endcsname{275}
\expandafter\gdef\csname odunum@val@recorded_vs_generated.avg.marginal.ming.generated\endcsname{0.532}
\expandafter\gdef\csname odunum@n@recorded_vs_generated.avg.marginal.ming.generated\endcsname{1\,801}
\expandafter\gdef\csname odunum@val@recorded_vs_generated.avg.marginal.ming.recorded\endcsname{0.493}
\expandafter\gdef\csname odunum@n@recorded_vs_generated.avg.marginal.ming.recorded\endcsname{277}
\expandafter\gdef\csname odunum@val@recorded_vs_generated.avg.marginal.minicpm_o.generated\endcsname{0.446}
\expandafter\gdef\csname odunum@n@recorded_vs_generated.avg.marginal.minicpm_o.generated\endcsname{1\,798}
\expandafter\gdef\csname odunum@val@recorded_vs_generated.avg.marginal.minicpm_o.recorded\endcsname{0.285}
\expandafter\gdef\csname odunum@n@recorded_vs_generated.avg.marginal.minicpm_o.recorded\endcsname{277}
\expandafter\gdef\csname odunum@val@recorded_vs_generated.avg.marginal.nemotron.generated\endcsname{0.396}
\expandafter\gdef\csname odunum@n@recorded_vs_generated.avg.marginal.nemotron.generated\endcsname{1\,801}
\expandafter\gdef\csname odunum@val@recorded_vs_generated.avg.marginal.nemotron.recorded\endcsname{0.161}
\expandafter\gdef\csname odunum@n@recorded_vs_generated.avg.marginal.nemotron.recorded\endcsname{277}
\expandafter\gdef\csname odunum@val@recorded_vs_generated.avg.marginal.qwen25_omni.generated\endcsname{0.482}
\expandafter\gdef\csname odunum@n@recorded_vs_generated.avg.marginal.qwen25_omni.generated\endcsname{1\,801}
\expandafter\gdef\csname odunum@val@recorded_vs_generated.avg.marginal.qwen25_omni.recorded\endcsname{0.332}
\expandafter\gdef\csname odunum@n@recorded_vs_generated.avg.marginal.qwen25_omni.recorded\endcsname{277}
\expandafter\gdef\csname odunum@val@recorded_vs_generated.avg.marginal.qwen3_omni_instruct.generated\endcsname{0.584}
\expandafter\gdef\csname odunum@n@recorded_vs_generated.avg.marginal.qwen3_omni_instruct.generated\endcsname{1\,801}
\expandafter\gdef\csname odunum@val@recorded_vs_generated.avg.marginal.qwen3_omni_instruct.recorded\endcsname{0.511}
\expandafter\gdef\csname odunum@n@recorded_vs_generated.avg.marginal.qwen3_omni_instruct.recorded\endcsname{277}
\expandafter\gdef\csname odunum@val@recorded_vs_generated.avg.marginal.qwen3_omni_think.generated\endcsname{0.640}
\expandafter\gdef\csname odunum@n@recorded_vs_generated.avg.marginal.qwen3_omni_think.generated\endcsname{1\,801}
\expandafter\gdef\csname odunum@val@recorded_vs_generated.avg.marginal.qwen3_omni_think.recorded\endcsname{0.536}
\expandafter\gdef\csname odunum@n@recorded_vs_generated.avg.marginal.qwen3_omni_think.recorded\endcsname{277}
\expandafter\gdef\csname odunum@val@recorded_vs_generated.avg.marginal.qwen_plus.generated\endcsname{0.723}
\expandafter\gdef\csname odunum@n@recorded_vs_generated.avg.marginal.qwen_plus.generated\endcsname{1\,801}
\expandafter\gdef\csname odunum@val@recorded_vs_generated.avg.marginal.qwen_plus.recorded\endcsname{0.657}
\expandafter\gdef\csname odunum@n@recorded_vs_generated.avg.marginal.qwen_plus.recorded\endcsname{276}
\expandafter\gdef\csname odunum@val@recorded_vs_generated.avg.marginal.salmonn2_7b.generated\endcsname{0.190}
\expandafter\gdef\csname odunum@n@recorded_vs_generated.avg.marginal.salmonn2_7b.generated\endcsname{1\,785}
\expandafter\gdef\csname odunum@val@recorded_vs_generated.avg.marginal.salmonn2_7b.recorded\endcsname{0.083}
\expandafter\gdef\csname odunum@n@recorded_vs_generated.avg.marginal.salmonn2_7b.recorded\endcsname{273}
\expandafter\gdef\csname odunum@val@recorded_vs_generated.avg.marginal.seed.generated\endcsname{0.717}
\expandafter\gdef\csname odunum@n@recorded_vs_generated.avg.marginal.seed.generated\endcsname{1\,801}
\expandafter\gdef\csname odunum@val@recorded_vs_generated.avg.marginal.seed.recorded\endcsname{0.676}
\expandafter\gdef\csname odunum@n@recorded_vs_generated.avg.marginal.seed.recorded\endcsname{277}
\expandafter\gdef\csname odunum@val@recorded_vs_generated.avg.stratum.cascade_asr.generated\endcsname{0.612}
\expandafter\gdef\csname odunum@n@recorded_vs_generated.avg.stratum.cascade_asr.generated\endcsname{529}
\expandafter\gdef\csname odunum@val@recorded_vs_generated.avg.stratum.cascade_asr.recorded\endcsname{0.558}
\expandafter\gdef\csname odunum@n@recorded_vs_generated.avg.stratum.cascade_asr.recorded\endcsname{276}
\expandafter\gdef\csname odunum@val@recorded_vs_generated.avg.stratum.gemini.generated\endcsname{0.719}
\expandafter\gdef\csname odunum@n@recorded_vs_generated.avg.stratum.gemini.generated\endcsname{529}
\expandafter\gdef\csname odunum@val@recorded_vs_generated.avg.stratum.gemini.recorded\endcsname{0.742}
\expandafter\gdef\csname odunum@n@recorded_vs_generated.avg.stratum.gemini.recorded\endcsname{276}
\expandafter\gdef\csname odunum@val@recorded_vs_generated.avg.stratum.gemini35_flash_lite.generated\endcsname{0.584}
\expandafter\gdef\csname odunum@n@recorded_vs_generated.avg.stratum.gemini35_flash_lite.generated\endcsname{529}
\expandafter\gdef\csname odunum@val@recorded_vs_generated.avg.stratum.gemini35_flash_lite.recorded\endcsname{0.470}
\expandafter\gdef\csname odunum@n@recorded_vs_generated.avg.stratum.gemini35_flash_lite.recorded\endcsname{276}
\expandafter\gdef\csname odunum@val@recorded_vs_generated.avg.stratum.gemini37_flash.generated\endcsname{0.693}
\expandafter\gdef\csname odunum@n@recorded_vs_generated.avg.stratum.gemini37_flash.generated\endcsname{529}
\expandafter\gdef\csname odunum@val@recorded_vs_generated.avg.stratum.gemini37_flash.recorded\endcsname{0.698}
\expandafter\gdef\csname odunum@n@recorded_vs_generated.avg.stratum.gemini37_flash.recorded\endcsname{274}
\expandafter\gdef\csname odunum@val@recorded_vs_generated.avg.stratum.ming.generated\endcsname{0.542}
\expandafter\gdef\csname odunum@n@recorded_vs_generated.avg.stratum.ming.generated\endcsname{529}
\expandafter\gdef\csname odunum@val@recorded_vs_generated.avg.stratum.ming.recorded\endcsname{0.493}
\expandafter\gdef\csname odunum@n@recorded_vs_generated.avg.stratum.ming.recorded\endcsname{276}
\expandafter\gdef\csname odunum@val@recorded_vs_generated.avg.stratum.minicpm_o.generated\endcsname{0.399}
\expandafter\gdef\csname odunum@n@recorded_vs_generated.avg.stratum.minicpm_o.generated\endcsname{527}
\expandafter\gdef\csname odunum@val@recorded_vs_generated.avg.stratum.minicpm_o.recorded\endcsname{0.284}
\expandafter\gdef\csname odunum@n@recorded_vs_generated.avg.stratum.minicpm_o.recorded\endcsname{276}
\expandafter\gdef\csname odunum@val@recorded_vs_generated.avg.stratum.nemotron.generated\endcsname{0.203}
\expandafter\gdef\csname odunum@n@recorded_vs_generated.avg.stratum.nemotron.generated\endcsname{529}
\expandafter\gdef\csname odunum@val@recorded_vs_generated.avg.stratum.nemotron.recorded\endcsname{0.161}
\expandafter\gdef\csname odunum@n@recorded_vs_generated.avg.stratum.nemotron.recorded\endcsname{276}
\expandafter\gdef\csname odunum@val@recorded_vs_generated.avg.stratum.qwen25_omni.generated\endcsname{0.462}
\expandafter\gdef\csname odunum@n@recorded_vs_generated.avg.stratum.qwen25_omni.generated\endcsname{529}
\expandafter\gdef\csname odunum@val@recorded_vs_generated.avg.stratum.qwen25_omni.recorded\endcsname{0.332}
\expandafter\gdef\csname odunum@n@recorded_vs_generated.avg.stratum.qwen25_omni.recorded\endcsname{276}
\expandafter\gdef\csname odunum@val@recorded_vs_generated.avg.stratum.qwen3_omni_instruct.generated\endcsname{0.564}
\expandafter\gdef\csname odunum@n@recorded_vs_generated.avg.stratum.qwen3_omni_instruct.generated\endcsname{529}
\expandafter\gdef\csname odunum@val@recorded_vs_generated.avg.stratum.qwen3_omni_instruct.recorded\endcsname{0.512}
\expandafter\gdef\csname odunum@n@recorded_vs_generated.avg.stratum.qwen3_omni_instruct.recorded\endcsname{276}
\expandafter\gdef\csname odunum@val@recorded_vs_generated.avg.stratum.qwen3_omni_think.generated\endcsname{0.613}
\expandafter\gdef\csname odunum@n@recorded_vs_generated.avg.stratum.qwen3_omni_think.generated\endcsname{529}
\expandafter\gdef\csname odunum@val@recorded_vs_generated.avg.stratum.qwen3_omni_think.recorded\endcsname{0.536}
\expandafter\gdef\csname odunum@n@recorded_vs_generated.avg.stratum.qwen3_omni_think.recorded\endcsname{276}
\expandafter\gdef\csname odunum@val@recorded_vs_generated.avg.stratum.qwen_plus.generated\endcsname{0.685}
\expandafter\gdef\csname odunum@n@recorded_vs_generated.avg.stratum.qwen_plus.generated\endcsname{529}
\expandafter\gdef\csname odunum@val@recorded_vs_generated.avg.stratum.qwen_plus.recorded\endcsname{0.657}
\expandafter\gdef\csname odunum@n@recorded_vs_generated.avg.stratum.qwen_plus.recorded\endcsname{275}
\expandafter\gdef\csname odunum@val@recorded_vs_generated.avg.stratum.salmonn2_7b.generated\endcsname{0.086}
\expandafter\gdef\csname odunum@n@recorded_vs_generated.avg.stratum.salmonn2_7b.generated\endcsname{526}
\expandafter\gdef\csname odunum@val@recorded_vs_generated.avg.stratum.salmonn2_7b.recorded\endcsname{0.083}
\expandafter\gdef\csname odunum@n@recorded_vs_generated.avg.stratum.salmonn2_7b.recorded\endcsname{272}
\expandafter\gdef\csname odunum@val@recorded_vs_generated.avg.stratum.seed.generated\endcsname{0.675}
\expandafter\gdef\csname odunum@n@recorded_vs_generated.avg.stratum.seed.generated\endcsname{529}
\expandafter\gdef\csname odunum@val@recorded_vs_generated.avg.stratum.seed.recorded\endcsname{0.676}
\expandafter\gdef\csname odunum@n@recorded_vs_generated.avg.stratum.seed.recorded\endcsname{276}
\expandafter\gdef\csname odunum@val@recorded_vs_generated.avg_delta.marginal.cascade_asr\endcsname{-0.088}
\expandafter\gdef\csname odunum@n@recorded_vs_generated.avg_delta.marginal.cascade_asr\endcsname{277}
\expandafter\gdef\csname odunum@val@recorded_vs_generated.avg_delta.marginal.gemini\endcsname{0.008}
\expandafter\gdef\csname odunum@n@recorded_vs_generated.avg_delta.marginal.gemini\endcsname{277}
\expandafter\gdef\csname odunum@val@recorded_vs_generated.avg_delta.marginal.gemini35_flash_lite\endcsname{-0.138}
\expandafter\gdef\csname odunum@n@recorded_vs_generated.avg_delta.marginal.gemini35_flash_lite\endcsname{277}
\expandafter\gdef\csname odunum@val@recorded_vs_generated.avg_delta.marginal.gemini37_flash\endcsname{-0.010}
\expandafter\gdef\csname odunum@n@recorded_vs_generated.avg_delta.marginal.gemini37_flash\endcsname{275}
\expandafter\gdef\csname odunum@val@recorded_vs_generated.avg_delta.marginal.ming\endcsname{-0.040}
\expandafter\gdef\csname odunum@n@recorded_vs_generated.avg_delta.marginal.ming\endcsname{277}
\expandafter\gdef\csname odunum@val@recorded_vs_generated.avg_delta.marginal.minicpm_o\endcsname{-0.161}
\expandafter\gdef\csname odunum@n@recorded_vs_generated.avg_delta.marginal.minicpm_o\endcsname{277}
\expandafter\gdef\csname odunum@val@recorded_vs_generated.avg_delta.marginal.nemotron\endcsname{-0.235}
\expandafter\gdef\csname odunum@n@recorded_vs_generated.avg_delta.marginal.nemotron\endcsname{277}
\expandafter\gdef\csname odunum@val@recorded_vs_generated.avg_delta.marginal.qwen25_omni\endcsname{-0.149}
\expandafter\gdef\csname odunum@n@recorded_vs_generated.avg_delta.marginal.qwen25_omni\endcsname{277}
\expandafter\gdef\csname odunum@val@recorded_vs_generated.avg_delta.marginal.qwen3_omni_instruct\endcsname{-0.073}
\expandafter\gdef\csname odunum@n@recorded_vs_generated.avg_delta.marginal.qwen3_omni_instruct\endcsname{277}
\expandafter\gdef\csname odunum@val@recorded_vs_generated.avg_delta.marginal.qwen3_omni_think\endcsname{-0.104}
\expandafter\gdef\csname odunum@n@recorded_vs_generated.avg_delta.marginal.qwen3_omni_think\endcsname{277}
\expandafter\gdef\csname odunum@val@recorded_vs_generated.avg_delta.marginal.qwen_plus\endcsname{-0.066}
\expandafter\gdef\csname odunum@n@recorded_vs_generated.avg_delta.marginal.qwen_plus\endcsname{276}
\expandafter\gdef\csname odunum@val@recorded_vs_generated.avg_delta.marginal.salmonn2_7b\endcsname{-0.107}
\expandafter\gdef\csname odunum@n@recorded_vs_generated.avg_delta.marginal.salmonn2_7b\endcsname{273}
\expandafter\gdef\csname odunum@val@recorded_vs_generated.avg_delta.marginal.seed\endcsname{-0.040}
\expandafter\gdef\csname odunum@n@recorded_vs_generated.avg_delta.marginal.seed\endcsname{277}
\expandafter\gdef\csname odunum@val@recorded_vs_generated.avg_delta.stratum.cascade_asr\endcsname{-0.054}
\expandafter\gdef\csname odunum@n@recorded_vs_generated.avg_delta.stratum.cascade_asr\endcsname{276}
\expandafter\gdef\csname odunum@val@recorded_vs_generated.avg_delta.stratum.gemini\endcsname{0.024}
\expandafter\gdef\csname odunum@n@recorded_vs_generated.avg_delta.stratum.gemini\endcsname{276}
\expandafter\gdef\csname odunum@val@recorded_vs_generated.avg_delta.stratum.gemini35_flash_lite\endcsname{-0.114}
\expandafter\gdef\csname odunum@n@recorded_vs_generated.avg_delta.stratum.gemini35_flash_lite\endcsname{276}
\expandafter\gdef\csname odunum@val@recorded_vs_generated.avg_delta.stratum.gemini37_flash\endcsname{0.005}
\expandafter\gdef\csname odunum@n@recorded_vs_generated.avg_delta.stratum.gemini37_flash\endcsname{274}
\expandafter\gdef\csname odunum@val@recorded_vs_generated.avg_delta.stratum.ming\endcsname{-0.048}
\expandafter\gdef\csname odunum@n@recorded_vs_generated.avg_delta.stratum.ming\endcsname{276}
\expandafter\gdef\csname odunum@val@recorded_vs_generated.avg_delta.stratum.minicpm_o\endcsname{-0.114}
\expandafter\gdef\csname odunum@n@recorded_vs_generated.avg_delta.stratum.minicpm_o\endcsname{276}
\expandafter\gdef\csname odunum@val@recorded_vs_generated.avg_delta.stratum.nemotron\endcsname{-0.042}
\expandafter\gdef\csname odunum@n@recorded_vs_generated.avg_delta.stratum.nemotron\endcsname{276}
\expandafter\gdef\csname odunum@val@recorded_vs_generated.avg_delta.stratum.qwen25_omni\endcsname{-0.130}
\expandafter\gdef\csname odunum@n@recorded_vs_generated.avg_delta.stratum.qwen25_omni\endcsname{276}
\expandafter\gdef\csname odunum@val@recorded_vs_generated.avg_delta.stratum.qwen3_omni_instruct\endcsname{-0.053}
\expandafter\gdef\csname odunum@n@recorded_vs_generated.avg_delta.stratum.qwen3_omni_instruct\endcsname{276}
\expandafter\gdef\csname odunum@val@recorded_vs_generated.avg_delta.stratum.qwen3_omni_think\endcsname{-0.076}
\expandafter\gdef\csname odunum@n@recorded_vs_generated.avg_delta.stratum.qwen3_omni_think\endcsname{276}
\expandafter\gdef\csname odunum@val@recorded_vs_generated.avg_delta.stratum.qwen_plus\endcsname{-0.028}
\expandafter\gdef\csname odunum@n@recorded_vs_generated.avg_delta.stratum.qwen_plus\endcsname{275}
\expandafter\gdef\csname odunum@val@recorded_vs_generated.avg_delta.stratum.salmonn2_7b\endcsname{-0.003}
\expandafter\gdef\csname odunum@n@recorded_vs_generated.avg_delta.stratum.salmonn2_7b\endcsname{272}
\expandafter\gdef\csname odunum@val@recorded_vs_generated.avg_delta.stratum.seed\endcsname{0.001}
\expandafter\gdef\csname odunum@n@recorded_vs_generated.avg_delta.stratum.seed\endcsname{276}
\expandafter\gdef\csname odunum@val@recorded_vs_generated.cov.cascade_asr.generated.audio\endcsname{0.387}
\expandafter\gdef\csname odunum@n@recorded_vs_generated.cov.cascade_asr.generated.audio\endcsname{212}
\expandafter\gdef\csname odunum@ci@recorded_vs_generated.cov.cascade_asr.generated.audio\endcsname{[0.321, 0.454]}
\expandafter\gdef\csname odunum@val@recorded_vs_generated.cov.cascade_asr.generated.context\endcsname{0.358}
\expandafter\gdef\csname odunum@n@recorded_vs_generated.cov.cascade_asr.generated.context\endcsname{824}
\expandafter\gdef\csname odunum@ci@recorded_vs_generated.cov.cascade_asr.generated.context\endcsname{[0.324, 0.392]}
\expandafter\gdef\csname odunum@val@recorded_vs_generated.cov.cascade_asr.generated.intent\endcsname{0.823}
\expandafter\gdef\csname odunum@n@recorded_vs_generated.cov.cascade_asr.generated.intent\endcsname{775}
\expandafter\gdef\csname odunum@ci@recorded_vs_generated.cov.cascade_asr.generated.intent\endcsname{[0.794, 0.853]}
\expandafter\gdef\csname odunum@val@recorded_vs_generated.cov.cascade_asr.generated.visual\endcsname{0.186}
\expandafter\gdef\csname odunum@n@recorded_vs_generated.cov.cascade_asr.generated.visual\endcsname{333}
\expandafter\gdef\csname odunum@ci@recorded_vs_generated.cov.cascade_asr.generated.visual\endcsname{[0.145, 0.230]}
\expandafter\gdef\csname odunum@val@recorded_vs_generated.cov.cascade_asr.recorded.audio\endcsname{0.318}
\expandafter\gdef\csname odunum@n@recorded_vs_generated.cov.cascade_asr.recorded.audio\endcsname{107}
\expandafter\gdef\csname odunum@ci@recorded_vs_generated.cov.cascade_asr.recorded.audio\endcsname{[0.229, 0.409]}
\expandafter\gdef\csname odunum@val@recorded_vs_generated.cov.cascade_asr.recorded.context\endcsname{0.323}
\expandafter\gdef\csname odunum@n@recorded_vs_generated.cov.cascade_asr.recorded.context\endcsname{338}
\expandafter\gdef\csname odunum@ci@recorded_vs_generated.cov.cascade_asr.recorded.context\endcsname{[0.268, 0.377]}
\expandafter\gdef\csname odunum@val@recorded_vs_generated.cov.cascade_asr.recorded.intent\endcsname{0.706}
\expandafter\gdef\csname odunum@n@recorded_vs_generated.cov.cascade_asr.recorded.intent\endcsname{360}
\expandafter\gdef\csname odunum@ci@recorded_vs_generated.cov.cascade_asr.recorded.intent\endcsname{[0.653, 0.756]}
\expandafter\gdef\csname odunum@val@recorded_vs_generated.cov.cascade_asr.recorded.visual\endcsname{0.246}
\expandafter\gdef\csname odunum@n@recorded_vs_generated.cov.cascade_asr.recorded.visual\endcsname{122}
\expandafter\gdef\csname odunum@ci@recorded_vs_generated.cov.cascade_asr.recorded.visual\endcsname{[0.171, 0.325]}
\expandafter\gdef\csname odunum@val@recorded_vs_generated.cov.gemini.generated.audio\endcsname{0.453}
\expandafter\gdef\csname odunum@n@recorded_vs_generated.cov.gemini.generated.audio\endcsname{212}
\expandafter\gdef\csname odunum@ci@recorded_vs_generated.cov.gemini.generated.audio\endcsname{[0.386, 0.519]}
\expandafter\gdef\csname odunum@val@recorded_vs_generated.cov.gemini.generated.context\endcsname{0.422}
\expandafter\gdef\csname odunum@n@recorded_vs_generated.cov.gemini.generated.context\endcsname{824}
\expandafter\gdef\csname odunum@ci@recorded_vs_generated.cov.gemini.generated.context\endcsname{[0.387, 0.458]}
\expandafter\gdef\csname odunum@val@recorded_vs_generated.cov.gemini.generated.intent\endcsname{0.849}
\expandafter\gdef\csname odunum@n@recorded_vs_generated.cov.gemini.generated.intent\endcsname{775}
\expandafter\gdef\csname odunum@ci@recorded_vs_generated.cov.gemini.generated.intent\endcsname{[0.823, 0.874]}
\expandafter\gdef\csname odunum@val@recorded_vs_generated.cov.gemini.generated.visual\endcsname{0.417}
\expandafter\gdef\csname odunum@n@recorded_vs_generated.cov.gemini.generated.visual\endcsname{333}
\expandafter\gdef\csname odunum@ci@recorded_vs_generated.cov.gemini.generated.visual\endcsname{[0.363, 0.474]}
\expandafter\gdef\csname odunum@val@recorded_vs_generated.cov.gemini.recorded.audio\endcsname{0.477}
\expandafter\gdef\csname odunum@n@recorded_vs_generated.cov.gemini.recorded.audio\endcsname{107}
\expandafter\gdef\csname odunum@ci@recorded_vs_generated.cov.gemini.recorded.audio\endcsname{[0.377, 0.574]}
\expandafter\gdef\csname odunum@val@recorded_vs_generated.cov.gemini.recorded.context\endcsname{0.500}
\expandafter\gdef\csname odunum@n@recorded_vs_generated.cov.gemini.recorded.context\endcsname{338}
\expandafter\gdef\csname odunum@ci@recorded_vs_generated.cov.gemini.recorded.context\endcsname{[0.444, 0.556]}
\expandafter\gdef\csname odunum@val@recorded_vs_generated.cov.gemini.recorded.intent\endcsname{0.783}
\expandafter\gdef\csname odunum@n@recorded_vs_generated.cov.gemini.recorded.intent\endcsname{360}
\expandafter\gdef\csname odunum@ci@recorded_vs_generated.cov.gemini.recorded.intent\endcsname{[0.737, 0.828]}
\expandafter\gdef\csname odunum@val@recorded_vs_generated.cov.gemini.recorded.visual\endcsname{0.672}
\expandafter\gdef\csname odunum@n@recorded_vs_generated.cov.gemini.recorded.visual\endcsname{122}
\expandafter\gdef\csname odunum@ci@recorded_vs_generated.cov.gemini.recorded.visual\endcsname{[0.586, 0.759]}
\expandafter\gdef\csname odunum@val@recorded_vs_generated.cov.qwen_plus.generated.audio\endcsname{0.462}
\expandafter\gdef\csname odunum@n@recorded_vs_generated.cov.qwen_plus.generated.audio\endcsname{212}
\expandafter\gdef\csname odunum@ci@recorded_vs_generated.cov.qwen_plus.generated.audio\endcsname{[0.394, 0.529]}
\expandafter\gdef\csname odunum@val@recorded_vs_generated.cov.qwen_plus.generated.context\endcsname{0.472}
\expandafter\gdef\csname odunum@n@recorded_vs_generated.cov.qwen_plus.generated.context\endcsname{824}
\expandafter\gdef\csname odunum@ci@recorded_vs_generated.cov.qwen_plus.generated.context\endcsname{[0.434, 0.509]}
\expandafter\gdef\csname odunum@val@recorded_vs_generated.cov.qwen_plus.generated.intent\endcsname{0.852}
\expandafter\gdef\csname odunum@n@recorded_vs_generated.cov.qwen_plus.generated.intent\endcsname{775}
\expandafter\gdef\csname odunum@ci@recorded_vs_generated.cov.qwen_plus.generated.intent\endcsname{[0.826, 0.876]}
\expandafter\gdef\csname odunum@val@recorded_vs_generated.cov.qwen_plus.generated.visual\endcsname{0.408}
\expandafter\gdef\csname odunum@n@recorded_vs_generated.cov.qwen_plus.generated.visual\endcsname{333}
\expandafter\gdef\csname odunum@ci@recorded_vs_generated.cov.qwen_plus.generated.visual\endcsname{[0.355, 0.463]}
\expandafter\gdef\csname odunum@val@recorded_vs_generated.cov.qwen_plus.recorded.audio\endcsname{0.377}
\expandafter\gdef\csname odunum@n@recorded_vs_generated.cov.qwen_plus.recorded.audio\endcsname{106}
\expandafter\gdef\csname odunum@ci@recorded_vs_generated.cov.qwen_plus.recorded.audio\endcsname{[0.283, 0.472]}
\expandafter\gdef\csname odunum@val@recorded_vs_generated.cov.qwen_plus.recorded.context\endcsname{0.482}
\expandafter\gdef\csname odunum@n@recorded_vs_generated.cov.qwen_plus.recorded.context\endcsname{336}
\expandafter\gdef\csname odunum@ci@recorded_vs_generated.cov.qwen_plus.recorded.context\endcsname{[0.428, 0.536]}
\expandafter\gdef\csname odunum@val@recorded_vs_generated.cov.qwen_plus.recorded.intent\endcsname{0.799}
\expandafter\gdef\csname odunum@n@recorded_vs_generated.cov.qwen_plus.recorded.intent\endcsname{358}
\expandafter\gdef\csname odunum@ci@recorded_vs_generated.cov.qwen_plus.recorded.intent\endcsname{[0.756, 0.840]}
\expandafter\gdef\csname odunum@val@recorded_vs_generated.cov.qwen_plus.recorded.visual\endcsname{0.590}
\expandafter\gdef\csname odunum@n@recorded_vs_generated.cov.qwen_plus.recorded.visual\endcsname{122}
\expandafter\gdef\csname odunum@ci@recorded_vs_generated.cov.qwen_plus.recorded.visual\endcsname{[0.496, 0.686]}
\expandafter\gdef\csname odunum@val@recorded_vs_generated.cov.seed.generated.audio\endcsname{0.495}
\expandafter\gdef\csname odunum@n@recorded_vs_generated.cov.seed.generated.audio\endcsname{212}
\expandafter\gdef\csname odunum@ci@recorded_vs_generated.cov.seed.generated.audio\endcsname{[0.428, 0.564]}
\expandafter\gdef\csname odunum@val@recorded_vs_generated.cov.seed.generated.context\endcsname{0.533}
\expandafter\gdef\csname odunum@n@recorded_vs_generated.cov.seed.generated.context\endcsname{824}
\expandafter\gdef\csname odunum@ci@recorded_vs_generated.cov.seed.generated.context\endcsname{[0.498, 0.567]}
\expandafter\gdef\csname odunum@val@recorded_vs_generated.cov.seed.generated.intent\endcsname{0.768}
\expandafter\gdef\csname odunum@n@recorded_vs_generated.cov.seed.generated.intent\endcsname{775}
\expandafter\gdef\csname odunum@ci@recorded_vs_generated.cov.seed.generated.intent\endcsname{[0.731, 0.803]}
\expandafter\gdef\csname odunum@val@recorded_vs_generated.cov.seed.generated.visual\endcsname{0.507}
\expandafter\gdef\csname odunum@n@recorded_vs_generated.cov.seed.generated.visual\endcsname{333}
\expandafter\gdef\csname odunum@ci@recorded_vs_generated.cov.seed.generated.visual\endcsname{[0.454, 0.560]}
\expandafter\gdef\csname odunum@val@recorded_vs_generated.cov.seed.recorded.audio\endcsname{0.449}
\expandafter\gdef\csname odunum@n@recorded_vs_generated.cov.seed.recorded.audio\endcsname{107}
\expandafter\gdef\csname odunum@ci@recorded_vs_generated.cov.seed.recorded.audio\endcsname{[0.352, 0.545]}
\expandafter\gdef\csname odunum@val@recorded_vs_generated.cov.seed.recorded.context\endcsname{0.532}
\expandafter\gdef\csname odunum@n@recorded_vs_generated.cov.seed.recorded.context\endcsname{338}
\expandafter\gdef\csname odunum@ci@recorded_vs_generated.cov.seed.recorded.context\endcsname{[0.477, 0.590]}
\expandafter\gdef\csname odunum@val@recorded_vs_generated.cov.seed.recorded.intent\endcsname{0.742}
\expandafter\gdef\csname odunum@n@recorded_vs_generated.cov.seed.recorded.intent\endcsname{360}
\expandafter\gdef\csname odunum@ci@recorded_vs_generated.cov.seed.recorded.intent\endcsname{[0.686, 0.794]}
\expandafter\gdef\csname odunum@val@recorded_vs_generated.cov.seed.recorded.visual\endcsname{0.672}
\expandafter\gdef\csname odunum@n@recorded_vs_generated.cov.seed.recorded.visual\endcsname{122}
\expandafter\gdef\csname odunum@ci@recorded_vs_generated.cov.seed.recorded.visual\endcsname{[0.587, 0.757]}
\expandafter\gdef\csname odunum@val@recorded_vs_generated.ftr.marginal.cascade_asr.generated\endcsname{0.597}
\expandafter\gdef\csname odunum@n@recorded_vs_generated.ftr.marginal.cascade_asr.generated\endcsname{372}
\expandafter\gdef\csname odunum@ci@recorded_vs_generated.ftr.marginal.cascade_asr.generated\endcsname{[0.546, 0.645]}
\expandafter\gdef\csname odunum@val@recorded_vs_generated.ftr.marginal.cascade_asr.recorded\endcsname{0.707}
\expandafter\gdef\csname odunum@n@recorded_vs_generated.ftr.marginal.cascade_asr.recorded\endcsname{75}
\expandafter\gdef\csname odunum@ci@recorded_vs_generated.ftr.marginal.cascade_asr.recorded\endcsname{[0.596, 0.798]}
\expandafter\gdef\csname odunum@val@recorded_vs_generated.ftr.marginal.gemini.generated\endcsname{0.379}
\expandafter\gdef\csname odunum@n@recorded_vs_generated.ftr.marginal.gemini.generated\endcsname{372}
\expandafter\gdef\csname odunum@ci@recorded_vs_generated.ftr.marginal.gemini.generated\endcsname{[0.331, 0.429]}
\expandafter\gdef\csname odunum@val@recorded_vs_generated.ftr.marginal.gemini.recorded\endcsname{0.347}
\expandafter\gdef\csname odunum@n@recorded_vs_generated.ftr.marginal.gemini.recorded\endcsname{75}
\expandafter\gdef\csname odunum@ci@recorded_vs_generated.ftr.marginal.gemini.recorded\endcsname{[0.249, 0.459]}
\expandafter\gdef\csname odunum@val@recorded_vs_generated.ftr.marginal.gemini35_flash_lite.generated\endcsname{0.470}
\expandafter\gdef\csname odunum@n@recorded_vs_generated.ftr.marginal.gemini35_flash_lite.generated\endcsname{372}
\expandafter\gdef\csname odunum@ci@recorded_vs_generated.ftr.marginal.gemini35_flash_lite.generated\endcsname{[0.420, 0.521]}
\expandafter\gdef\csname odunum@val@recorded_vs_generated.ftr.marginal.gemini35_flash_lite.recorded\endcsname{0.613}
\expandafter\gdef\csname odunum@n@recorded_vs_generated.ftr.marginal.gemini35_flash_lite.recorded\endcsname{75}
\expandafter\gdef\csname odunum@ci@recorded_vs_generated.ftr.marginal.gemini35_flash_lite.recorded\endcsname{[0.500, 0.715]}
\expandafter\gdef\csname odunum@val@recorded_vs_generated.ftr.marginal.gemini37_flash.generated\endcsname{0.246}
\expandafter\gdef\csname odunum@n@recorded_vs_generated.ftr.marginal.gemini37_flash.generated\endcsname{366}
\expandafter\gdef\csname odunum@ci@recorded_vs_generated.ftr.marginal.gemini37_flash.generated\endcsname{[0.205, 0.293]}
\expandafter\gdef\csname odunum@val@recorded_vs_generated.ftr.marginal.gemini37_flash.recorded\endcsname{0.176}
\expandafter\gdef\csname odunum@n@recorded_vs_generated.ftr.marginal.gemini37_flash.recorded\endcsname{74}
\expandafter\gdef\csname odunum@ci@recorded_vs_generated.ftr.marginal.gemini37_flash.recorded\endcsname{[0.106, 0.278]}
\expandafter\gdef\csname odunum@val@recorded_vs_generated.ftr.marginal.ming.generated\endcsname{0.850}
\expandafter\gdef\csname odunum@n@recorded_vs_generated.ftr.marginal.ming.generated\endcsname{372}
\expandafter\gdef\csname odunum@ci@recorded_vs_generated.ftr.marginal.ming.generated\endcsname{[0.810, 0.882]}
\expandafter\gdef\csname odunum@val@recorded_vs_generated.ftr.marginal.ming.recorded\endcsname{0.987}
\expandafter\gdef\csname odunum@n@recorded_vs_generated.ftr.marginal.ming.recorded\endcsname{75}
\expandafter\gdef\csname odunum@ci@recorded_vs_generated.ftr.marginal.ming.recorded\endcsname{[0.928, 0.998]}
\expandafter\gdef\csname odunum@val@recorded_vs_generated.ftr.marginal.minicpm_o.generated\endcsname{0.787}
\expandafter\gdef\csname odunum@n@recorded_vs_generated.ftr.marginal.minicpm_o.generated\endcsname{371}
\expandafter\gdef\csname odunum@ci@recorded_vs_generated.ftr.marginal.minicpm_o.generated\endcsname{[0.743, 0.826]}
\expandafter\gdef\csname odunum@val@recorded_vs_generated.ftr.marginal.minicpm_o.recorded\endcsname{0.947}
\expandafter\gdef\csname odunum@n@recorded_vs_generated.ftr.marginal.minicpm_o.recorded\endcsname{75}
\expandafter\gdef\csname odunum@ci@recorded_vs_generated.ftr.marginal.minicpm_o.recorded\endcsname{[0.871, 0.979]}
\expandafter\gdef\csname odunum@val@recorded_vs_generated.ftr.marginal.nemotron.generated\endcsname{0.745}
\expandafter\gdef\csname odunum@n@recorded_vs_generated.ftr.marginal.nemotron.generated\endcsname{372}
\expandafter\gdef\csname odunum@ci@recorded_vs_generated.ftr.marginal.nemotron.generated\endcsname{[0.698, 0.786]}
\expandafter\gdef\csname odunum@val@recorded_vs_generated.ftr.marginal.nemotron.recorded\endcsname{0.600}
\expandafter\gdef\csname odunum@n@recorded_vs_generated.ftr.marginal.nemotron.recorded\endcsname{75}
\expandafter\gdef\csname odunum@ci@recorded_vs_generated.ftr.marginal.nemotron.recorded\endcsname{[0.487, 0.703]}
\expandafter\gdef\csname odunum@val@recorded_vs_generated.ftr.marginal.qwen25_omni.generated\endcsname{0.793}
\expandafter\gdef\csname odunum@n@recorded_vs_generated.ftr.marginal.qwen25_omni.generated\endcsname{372}
\expandafter\gdef\csname odunum@ci@recorded_vs_generated.ftr.marginal.qwen25_omni.generated\endcsname{[0.749, 0.831]}
\expandafter\gdef\csname odunum@val@recorded_vs_generated.ftr.marginal.qwen25_omni.recorded\endcsname{0.840}
\expandafter\gdef\csname odunum@n@recorded_vs_generated.ftr.marginal.qwen25_omni.recorded\endcsname{75}
\expandafter\gdef\csname odunum@ci@recorded_vs_generated.ftr.marginal.qwen25_omni.recorded\endcsname{[0.741, 0.906]}
\expandafter\gdef\csname odunum@val@recorded_vs_generated.ftr.marginal.qwen3_omni_instruct.generated\endcsname{0.868}
\expandafter\gdef\csname odunum@n@recorded_vs_generated.ftr.marginal.qwen3_omni_instruct.generated\endcsname{372}
\expandafter\gdef\csname odunum@ci@recorded_vs_generated.ftr.marginal.qwen3_omni_instruct.generated\endcsname{[0.830, 0.899]}
\expandafter\gdef\csname odunum@val@recorded_vs_generated.ftr.marginal.qwen3_omni_instruct.recorded\endcsname{0.987}
\expandafter\gdef\csname odunum@n@recorded_vs_generated.ftr.marginal.qwen3_omni_instruct.recorded\endcsname{75}
\expandafter\gdef\csname odunum@ci@recorded_vs_generated.ftr.marginal.qwen3_omni_instruct.recorded\endcsname{[0.928, 0.998]}
\expandafter\gdef\csname odunum@val@recorded_vs_generated.ftr.marginal.qwen3_omni_think.generated\endcsname{0.715}
\expandafter\gdef\csname odunum@n@recorded_vs_generated.ftr.marginal.qwen3_omni_think.generated\endcsname{372}
\expandafter\gdef\csname odunum@ci@recorded_vs_generated.ftr.marginal.qwen3_omni_think.generated\endcsname{[0.667, 0.759]}
\expandafter\gdef\csname odunum@val@recorded_vs_generated.ftr.marginal.qwen3_omni_think.recorded\endcsname{0.920}
\expandafter\gdef\csname odunum@n@recorded_vs_generated.ftr.marginal.qwen3_omni_think.recorded\endcsname{75}
\expandafter\gdef\csname odunum@ci@recorded_vs_generated.ftr.marginal.qwen3_omni_think.recorded\endcsname{[0.836, 0.963]}
\expandafter\gdef\csname odunum@val@recorded_vs_generated.ftr.marginal.qwen_plus.generated\endcsname{0.632}
\expandafter\gdef\csname odunum@n@recorded_vs_generated.ftr.marginal.qwen_plus.generated\endcsname{372}
\expandafter\gdef\csname odunum@ci@recorded_vs_generated.ftr.marginal.qwen_plus.generated\endcsname{[0.582, 0.679]}
\expandafter\gdef\csname odunum@val@recorded_vs_generated.ftr.marginal.qwen_plus.recorded\endcsname{0.867}
\expandafter\gdef\csname odunum@n@recorded_vs_generated.ftr.marginal.qwen_plus.recorded\endcsname{75}
\expandafter\gdef\csname odunum@ci@recorded_vs_generated.ftr.marginal.qwen_plus.recorded\endcsname{[0.772, 0.926]}
\expandafter\gdef\csname odunum@val@recorded_vs_generated.ftr.marginal.salmonn2_7b.generated\endcsname{0.916}
\expandafter\gdef\csname odunum@n@recorded_vs_generated.ftr.marginal.salmonn2_7b.generated\endcsname{370}
\expandafter\gdef\csname odunum@ci@recorded_vs_generated.ftr.marginal.salmonn2_7b.generated\endcsname{[0.884, 0.940]}
\expandafter\gdef\csname odunum@val@recorded_vs_generated.ftr.marginal.salmonn2_7b.recorded\endcsname{0.946}
\expandafter\gdef\csname odunum@n@recorded_vs_generated.ftr.marginal.salmonn2_7b.recorded\endcsname{74}
\expandafter\gdef\csname odunum@ci@recorded_vs_generated.ftr.marginal.salmonn2_7b.recorded\endcsname{[0.869, 0.979]}
\expandafter\gdef\csname odunum@val@recorded_vs_generated.ftr.marginal.seed.generated\endcsname{0.812}
\expandafter\gdef\csname odunum@n@recorded_vs_generated.ftr.marginal.seed.generated\endcsname{372}
\expandafter\gdef\csname odunum@ci@recorded_vs_generated.ftr.marginal.seed.generated\endcsname{[0.769, 0.848]}
\expandafter\gdef\csname odunum@val@recorded_vs_generated.ftr.marginal.seed.recorded\endcsname{0.840}
\expandafter\gdef\csname odunum@n@recorded_vs_generated.ftr.marginal.seed.recorded\endcsname{75}
\expandafter\gdef\csname odunum@ci@recorded_vs_generated.ftr.marginal.seed.recorded\endcsname{[0.741, 0.906]}
\expandafter\gdef\csname odunum@val@recorded_vs_generated.ftr.stratum.cascade_asr.generated\endcsname{0.733}
\expandafter\gdef\csname odunum@n@recorded_vs_generated.ftr.stratum.cascade_asr.generated\endcsname{105}
\expandafter\gdef\csname odunum@ci@recorded_vs_generated.ftr.stratum.cascade_asr.generated\endcsname{[0.642, 0.809]}
\expandafter\gdef\csname odunum@val@recorded_vs_generated.ftr.stratum.cascade_asr.recorded\endcsname{0.707}
\expandafter\gdef\csname odunum@n@recorded_vs_generated.ftr.stratum.cascade_asr.recorded\endcsname{75}
\expandafter\gdef\csname odunum@ci@recorded_vs_generated.ftr.stratum.cascade_asr.recorded\endcsname{[0.596, 0.798]}
\expandafter\gdef\csname odunum@val@recorded_vs_generated.ftr.stratum.gemini.generated\endcsname{0.409}
\expandafter\gdef\csname odunum@n@recorded_vs_generated.ftr.stratum.gemini.generated\endcsname{105}
\expandafter\gdef\csname odunum@ci@recorded_vs_generated.ftr.stratum.gemini.generated\endcsname{[0.320, 0.505]}
\expandafter\gdef\csname odunum@val@recorded_vs_generated.ftr.stratum.gemini.recorded\endcsname{0.347}
\expandafter\gdef\csname odunum@n@recorded_vs_generated.ftr.stratum.gemini.recorded\endcsname{75}
\expandafter\gdef\csname odunum@ci@recorded_vs_generated.ftr.stratum.gemini.recorded\endcsname{[0.249, 0.459]}
\expandafter\gdef\csname odunum@val@recorded_vs_generated.ftr.stratum.gemini35_flash_lite.generated\endcsname{0.524}
\expandafter\gdef\csname odunum@n@recorded_vs_generated.ftr.stratum.gemini35_flash_lite.generated\endcsname{105}
\expandafter\gdef\csname odunum@ci@recorded_vs_generated.ftr.stratum.gemini35_flash_lite.generated\endcsname{[0.429, 0.617]}
\expandafter\gdef\csname odunum@val@recorded_vs_generated.ftr.stratum.gemini35_flash_lite.recorded\endcsname{0.613}
\expandafter\gdef\csname odunum@n@recorded_vs_generated.ftr.stratum.gemini35_flash_lite.recorded\endcsname{75}
\expandafter\gdef\csname odunum@ci@recorded_vs_generated.ftr.stratum.gemini35_flash_lite.recorded\endcsname{[0.500, 0.715]}
\expandafter\gdef\csname odunum@val@recorded_vs_generated.ftr.stratum.gemini37_flash.generated\endcsname{0.276}
\expandafter\gdef\csname odunum@n@recorded_vs_generated.ftr.stratum.gemini37_flash.generated\endcsname{105}
\expandafter\gdef\csname odunum@ci@recorded_vs_generated.ftr.stratum.gemini37_flash.generated\endcsname{[0.200, 0.368]}
\expandafter\gdef\csname odunum@val@recorded_vs_generated.ftr.stratum.gemini37_flash.recorded\endcsname{0.176}
\expandafter\gdef\csname odunum@n@recorded_vs_generated.ftr.stratum.gemini37_flash.recorded\endcsname{74}
\expandafter\gdef\csname odunum@ci@recorded_vs_generated.ftr.stratum.gemini37_flash.recorded\endcsname{[0.106, 0.278]}
\expandafter\gdef\csname odunum@val@recorded_vs_generated.ftr.stratum.ming.generated\endcsname{0.952}
\expandafter\gdef\csname odunum@n@recorded_vs_generated.ftr.stratum.ming.generated\endcsname{105}
\expandafter\gdef\csname odunum@ci@recorded_vs_generated.ftr.stratum.ming.generated\endcsname{[0.893, 0.979]}
\expandafter\gdef\csname odunum@val@recorded_vs_generated.ftr.stratum.ming.recorded\endcsname{0.987}
\expandafter\gdef\csname odunum@n@recorded_vs_generated.ftr.stratum.ming.recorded\endcsname{75}
\expandafter\gdef\csname odunum@ci@recorded_vs_generated.ftr.stratum.ming.recorded\endcsname{[0.928, 0.998]}
\expandafter\gdef\csname odunum@val@recorded_vs_generated.ftr.stratum.minicpm_o.generated\endcsname{0.913}
\expandafter\gdef\csname odunum@n@recorded_vs_generated.ftr.stratum.minicpm_o.generated\endcsname{104}
\expandafter\gdef\csname odunum@ci@recorded_vs_generated.ftr.stratum.minicpm_o.generated\endcsname{[0.844, 0.954]}
\expandafter\gdef\csname odunum@val@recorded_vs_generated.ftr.stratum.minicpm_o.recorded\endcsname{0.947}
\expandafter\gdef\csname odunum@n@recorded_vs_generated.ftr.stratum.minicpm_o.recorded\endcsname{75}
\expandafter\gdef\csname odunum@ci@recorded_vs_generated.ftr.stratum.minicpm_o.recorded\endcsname{[0.871, 0.979]}
\expandafter\gdef\csname odunum@val@recorded_vs_generated.ftr.stratum.nemotron.generated\endcsname{0.657}
\expandafter\gdef\csname odunum@n@recorded_vs_generated.ftr.stratum.nemotron.generated\endcsname{105}
\expandafter\gdef\csname odunum@ci@recorded_vs_generated.ftr.stratum.nemotron.generated\endcsname{[0.562, 0.741]}
\expandafter\gdef\csname odunum@val@recorded_vs_generated.ftr.stratum.nemotron.recorded\endcsname{0.600}
\expandafter\gdef\csname odunum@n@recorded_vs_generated.ftr.stratum.nemotron.recorded\endcsname{75}
\expandafter\gdef\csname odunum@ci@recorded_vs_generated.ftr.stratum.nemotron.recorded\endcsname{[0.487, 0.703]}
\expandafter\gdef\csname odunum@val@recorded_vs_generated.ftr.stratum.qwen25_omni.generated\endcsname{0.933}
\expandafter\gdef\csname odunum@n@recorded_vs_generated.ftr.stratum.qwen25_omni.generated\endcsname{105}
\expandafter\gdef\csname odunum@ci@recorded_vs_generated.ftr.stratum.qwen25_omni.generated\endcsname{[0.869, 0.967]}
\expandafter\gdef\csname odunum@val@recorded_vs_generated.ftr.stratum.qwen25_omni.recorded\endcsname{0.840}
\expandafter\gdef\csname odunum@n@recorded_vs_generated.ftr.stratum.qwen25_omni.recorded\endcsname{75}
\expandafter\gdef\csname odunum@ci@recorded_vs_generated.ftr.stratum.qwen25_omni.recorded\endcsname{[0.741, 0.906]}
\expandafter\gdef\csname odunum@val@recorded_vs_generated.ftr.stratum.qwen3_omni_instruct.generated\endcsname{0.943}
\expandafter\gdef\csname odunum@n@recorded_vs_generated.ftr.stratum.qwen3_omni_instruct.generated\endcsname{105}
\expandafter\gdef\csname odunum@ci@recorded_vs_generated.ftr.stratum.qwen3_omni_instruct.generated\endcsname{[0.881, 0.974]}
\expandafter\gdef\csname odunum@val@recorded_vs_generated.ftr.stratum.qwen3_omni_instruct.recorded\endcsname{0.987}
\expandafter\gdef\csname odunum@n@recorded_vs_generated.ftr.stratum.qwen3_omni_instruct.recorded\endcsname{75}
\expandafter\gdef\csname odunum@ci@recorded_vs_generated.ftr.stratum.qwen3_omni_instruct.recorded\endcsname{[0.928, 0.998]}
\expandafter\gdef\csname odunum@val@recorded_vs_generated.ftr.stratum.qwen3_omni_think.generated\endcsname{0.809}
\expandafter\gdef\csname odunum@n@recorded_vs_generated.ftr.stratum.qwen3_omni_think.generated\endcsname{105}
\expandafter\gdef\csname odunum@ci@recorded_vs_generated.ftr.stratum.qwen3_omni_think.generated\endcsname{[0.724, 0.873]}
\expandafter\gdef\csname odunum@val@recorded_vs_generated.ftr.stratum.qwen3_omni_think.recorded\endcsname{0.920}
\expandafter\gdef\csname odunum@n@recorded_vs_generated.ftr.stratum.qwen3_omni_think.recorded\endcsname{75}
\expandafter\gdef\csname odunum@ci@recorded_vs_generated.ftr.stratum.qwen3_omni_think.recorded\endcsname{[0.836, 0.963]}
\expandafter\gdef\csname odunum@val@recorded_vs_generated.ftr.stratum.qwen_plus.generated\endcsname{0.790}
\expandafter\gdef\csname odunum@n@recorded_vs_generated.ftr.stratum.qwen_plus.generated\endcsname{105}
\expandafter\gdef\csname odunum@ci@recorded_vs_generated.ftr.stratum.qwen_plus.generated\endcsname{[0.703, 0.857]}
\expandafter\gdef\csname odunum@val@recorded_vs_generated.ftr.stratum.qwen_plus.recorded\endcsname{0.867}
\expandafter\gdef\csname odunum@n@recorded_vs_generated.ftr.stratum.qwen_plus.recorded\endcsname{75}
\expandafter\gdef\csname odunum@ci@recorded_vs_generated.ftr.stratum.qwen_plus.recorded\endcsname{[0.772, 0.926]}
\expandafter\gdef\csname odunum@val@recorded_vs_generated.ftr.stratum.salmonn2_7b.generated\endcsname{0.914}
\expandafter\gdef\csname odunum@n@recorded_vs_generated.ftr.stratum.salmonn2_7b.generated\endcsname{105}
\expandafter\gdef\csname odunum@ci@recorded_vs_generated.ftr.stratum.salmonn2_7b.generated\endcsname{[0.845, 0.954]}
\expandafter\gdef\csname odunum@val@recorded_vs_generated.ftr.stratum.salmonn2_7b.recorded\endcsname{0.946}
\expandafter\gdef\csname odunum@n@recorded_vs_generated.ftr.stratum.salmonn2_7b.recorded\endcsname{74}
\expandafter\gdef\csname odunum@ci@recorded_vs_generated.ftr.stratum.salmonn2_7b.recorded\endcsname{[0.869, 0.979]}
\expandafter\gdef\csname odunum@val@recorded_vs_generated.ftr.stratum.seed.generated\endcsname{0.819}
\expandafter\gdef\csname odunum@n@recorded_vs_generated.ftr.stratum.seed.generated\endcsname{105}
\expandafter\gdef\csname odunum@ci@recorded_vs_generated.ftr.stratum.seed.generated\endcsname{[0.735, 0.881]}
\expandafter\gdef\csname odunum@val@recorded_vs_generated.ftr.stratum.seed.recorded\endcsname{0.840}
\expandafter\gdef\csname odunum@n@recorded_vs_generated.ftr.stratum.seed.recorded\endcsname{75}
\expandafter\gdef\csname odunum@ci@recorded_vs_generated.ftr.stratum.seed.recorded\endcsname{[0.741, 0.906]}
\expandafter\gdef\csname odunum@val@recorded_vs_generated.ftr_delta.marginal.cascade_asr\endcsname{0.110}
\expandafter\gdef\csname odunum@n@recorded_vs_generated.ftr_delta.marginal.cascade_asr\endcsname{75}
\expandafter\gdef\csname odunum@val@recorded_vs_generated.ftr_delta.marginal.gemini\endcsname{-0.032}
\expandafter\gdef\csname odunum@n@recorded_vs_generated.ftr_delta.marginal.gemini\endcsname{75}
\expandafter\gdef\csname odunum@val@recorded_vs_generated.ftr_delta.marginal.gemini35_flash_lite\endcsname{0.143}
\expandafter\gdef\csname odunum@n@recorded_vs_generated.ftr_delta.marginal.gemini35_flash_lite\endcsname{75}
\expandafter\gdef\csname odunum@val@recorded_vs_generated.ftr_delta.marginal.gemini37_flash\endcsname{-0.070}
\expandafter\gdef\csname odunum@n@recorded_vs_generated.ftr_delta.marginal.gemini37_flash\endcsname{74}
\expandafter\gdef\csname odunum@val@recorded_vs_generated.ftr_delta.marginal.ming\endcsname{0.137}
\expandafter\gdef\csname odunum@n@recorded_vs_generated.ftr_delta.marginal.ming\endcsname{75}
\expandafter\gdef\csname odunum@val@recorded_vs_generated.ftr_delta.marginal.minicpm_o\endcsname{0.160}
\expandafter\gdef\csname odunum@n@recorded_vs_generated.ftr_delta.marginal.minicpm_o\endcsname{75}
\expandafter\gdef\csname odunum@val@recorded_vs_generated.ftr_delta.marginal.nemotron\endcsname{-0.145}
\expandafter\gdef\csname odunum@n@recorded_vs_generated.ftr_delta.marginal.nemotron\endcsname{75}
\expandafter\gdef\csname odunum@val@recorded_vs_generated.ftr_delta.marginal.qwen25_omni\endcsname{0.047}
\expandafter\gdef\csname odunum@n@recorded_vs_generated.ftr_delta.marginal.qwen25_omni\endcsname{75}
\expandafter\gdef\csname odunum@val@recorded_vs_generated.ftr_delta.marginal.qwen3_omni_instruct\endcsname{0.118}
\expandafter\gdef\csname odunum@n@recorded_vs_generated.ftr_delta.marginal.qwen3_omni_instruct\endcsname{75}
\expandafter\gdef\csname odunum@val@recorded_vs_generated.ftr_delta.marginal.qwen3_omni_think\endcsname{0.205}
\expandafter\gdef\csname odunum@n@recorded_vs_generated.ftr_delta.marginal.qwen3_omni_think\endcsname{75}
\expandafter\gdef\csname odunum@val@recorded_vs_generated.ftr_delta.marginal.qwen_plus\endcsname{0.235}
\expandafter\gdef\csname odunum@n@recorded_vs_generated.ftr_delta.marginal.qwen_plus\endcsname{75}
\expandafter\gdef\csname odunum@val@recorded_vs_generated.ftr_delta.marginal.salmonn2_7b\endcsname{0.030}
\expandafter\gdef\csname odunum@n@recorded_vs_generated.ftr_delta.marginal.salmonn2_7b\endcsname{74}
\expandafter\gdef\csname odunum@val@recorded_vs_generated.ftr_delta.marginal.seed\endcsname{0.028}
\expandafter\gdef\csname odunum@n@recorded_vs_generated.ftr_delta.marginal.seed\endcsname{75}
\expandafter\gdef\csname odunum@val@recorded_vs_generated.ftr_delta.stratum.cascade_asr\endcsname{-0.027}
\expandafter\gdef\csname odunum@n@recorded_vs_generated.ftr_delta.stratum.cascade_asr\endcsname{75}
\expandafter\gdef\csname odunum@val@recorded_vs_generated.ftr_delta.stratum.gemini\endcsname{-0.063}
\expandafter\gdef\csname odunum@n@recorded_vs_generated.ftr_delta.stratum.gemini\endcsname{75}
\expandafter\gdef\csname odunum@val@recorded_vs_generated.ftr_delta.stratum.gemini35_flash_lite\endcsname{0.089}
\expandafter\gdef\csname odunum@n@recorded_vs_generated.ftr_delta.stratum.gemini35_flash_lite\endcsname{75}
\expandafter\gdef\csname odunum@val@recorded_vs_generated.ftr_delta.stratum.gemini37_flash\endcsname{-0.101}
\expandafter\gdef\csname odunum@n@recorded_vs_generated.ftr_delta.stratum.gemini37_flash\endcsname{74}
\expandafter\gdef\csname odunum@val@recorded_vs_generated.ftr_delta.stratum.ming\endcsname{0.034}
\expandafter\gdef\csname odunum@n@recorded_vs_generated.ftr_delta.stratum.ming\endcsname{75}
\expandafter\gdef\csname odunum@val@recorded_vs_generated.ftr_delta.stratum.minicpm_o\endcsname{0.033}
\expandafter\gdef\csname odunum@n@recorded_vs_generated.ftr_delta.stratum.minicpm_o\endcsname{75}
\expandafter\gdef\csname odunum@val@recorded_vs_generated.ftr_delta.stratum.nemotron\endcsname{-0.057}
\expandafter\gdef\csname odunum@n@recorded_vs_generated.ftr_delta.stratum.nemotron\endcsname{75}
\expandafter\gdef\csname odunum@val@recorded_vs_generated.ftr_delta.stratum.qwen25_omni\endcsname{-0.093}
\expandafter\gdef\csname odunum@n@recorded_vs_generated.ftr_delta.stratum.qwen25_omni\endcsname{75}
\expandafter\gdef\csname odunum@val@recorded_vs_generated.ftr_delta.stratum.qwen3_omni_instruct\endcsname{0.044}
\expandafter\gdef\csname odunum@n@recorded_vs_generated.ftr_delta.stratum.qwen3_omni_instruct\endcsname{75}
\expandafter\gdef\csname odunum@val@recorded_vs_generated.ftr_delta.stratum.qwen3_omni_think\endcsname{0.111}
\expandafter\gdef\csname odunum@n@recorded_vs_generated.ftr_delta.stratum.qwen3_omni_think\endcsname{75}
\expandafter\gdef\csname odunum@val@recorded_vs_generated.ftr_delta.stratum.qwen_plus\endcsname{0.076}
\expandafter\gdef\csname odunum@n@recorded_vs_generated.ftr_delta.stratum.qwen_plus\endcsname{75}
\expandafter\gdef\csname odunum@val@recorded_vs_generated.ftr_delta.stratum.salmonn2_7b\endcsname{0.032}
\expandafter\gdef\csname odunum@n@recorded_vs_generated.ftr_delta.stratum.salmonn2_7b\endcsname{74}
\expandafter\gdef\csname odunum@val@recorded_vs_generated.ftr_delta.stratum.seed\endcsname{0.021}
\expandafter\gdef\csname odunum@n@recorded_vs_generated.ftr_delta.stratum.seed\endcsname{75}
\expandafter\gdef\csname odunum@val@recorded_vs_generated.gap.cascade_asr.generated.intent_minus_context\endcsname{0.465}
\expandafter\gdef\csname odunum@n@recorded_vs_generated.gap.cascade_asr.generated.intent_minus_context\endcsname{1\,599}
\expandafter\gdef\csname odunum@val@recorded_vs_generated.gap.cascade_asr.recorded.intent_minus_context\endcsname{0.383}
\expandafter\gdef\csname odunum@n@recorded_vs_generated.gap.cascade_asr.recorded.intent_minus_context\endcsname{698}
\expandafter\gdef\csname odunum@val@recorded_vs_generated.gap.gemini.generated.intent_minus_context\endcsname{0.427}
\expandafter\gdef\csname odunum@n@recorded_vs_generated.gap.gemini.generated.intent_minus_context\endcsname{1\,599}
\expandafter\gdef\csname odunum@val@recorded_vs_generated.gap.gemini.recorded.intent_minus_context\endcsname{0.283}
\expandafter\gdef\csname odunum@n@recorded_vs_generated.gap.gemini.recorded.intent_minus_context\endcsname{698}
\expandafter\gdef\csname odunum@val@recorded_vs_generated.gap.qwen_plus.generated.intent_minus_context\endcsname{0.380}
\expandafter\gdef\csname odunum@n@recorded_vs_generated.gap.qwen_plus.generated.intent_minus_context\endcsname{1\,599}
\expandafter\gdef\csname odunum@val@recorded_vs_generated.gap.qwen_plus.recorded.intent_minus_context\endcsname{0.317}
\expandafter\gdef\csname odunum@n@recorded_vs_generated.gap.qwen_plus.recorded.intent_minus_context\endcsname{694}
\expandafter\gdef\csname odunum@val@recorded_vs_generated.gap.seed.generated.intent_minus_context\endcsname{0.235}
\expandafter\gdef\csname odunum@n@recorded_vs_generated.gap.seed.generated.intent_minus_context\endcsname{1\,599}
\expandafter\gdef\csname odunum@val@recorded_vs_generated.gap.seed.recorded.intent_minus_context\endcsname{0.209}
\expandafter\gdef\csname odunum@n@recorded_vs_generated.gap.seed.recorded.intent_minus_context\endcsname{698}
\expandafter\gdef\csname odunum@val@recorded_vs_generated.m1.marginal.cascade_asr.generated\endcsname{0.688}
\expandafter\gdef\csname odunum@n@recorded_vs_generated.m1.marginal.cascade_asr.generated\endcsname{1\,801}
\expandafter\gdef\csname odunum@val@recorded_vs_generated.m1.marginal.cascade_asr.recorded\endcsname{0.613}
\expandafter\gdef\csname odunum@n@recorded_vs_generated.m1.marginal.cascade_asr.recorded\endcsname{277}
\expandafter\gdef\csname odunum@val@recorded_vs_generated.m1.marginal.gemini.generated\endcsname{0.843}
\expandafter\gdef\csname odunum@n@recorded_vs_generated.m1.marginal.gemini.generated\endcsname{1\,801}
\expandafter\gdef\csname odunum@val@recorded_vs_generated.m1.marginal.gemini.recorded\endcsname{0.861}
\expandafter\gdef\csname odunum@n@recorded_vs_generated.m1.marginal.gemini.recorded\endcsname{277}
\expandafter\gdef\csname odunum@val@recorded_vs_generated.m1.marginal.gemini35_flash_lite.generated\endcsname{0.691}
\expandafter\gdef\csname odunum@n@recorded_vs_generated.m1.marginal.gemini35_flash_lite.generated\endcsname{1\,801}
\expandafter\gdef\csname odunum@val@recorded_vs_generated.m1.marginal.gemini35_flash_lite.recorded\endcsname{0.594}
\expandafter\gdef\csname odunum@n@recorded_vs_generated.m1.marginal.gemini35_flash_lite.recorded\endcsname{277}
\expandafter\gdef\csname odunum@val@recorded_vs_generated.m1.marginal.gemini37_flash.generated\endcsname{0.854}
\expandafter\gdef\csname odunum@n@recorded_vs_generated.m1.marginal.gemini37_flash.generated\endcsname{1\,792}
\expandafter\gdef\csname odunum@val@recorded_vs_generated.m1.marginal.gemini37_flash.recorded\endcsname{0.888}
\expandafter\gdef\csname odunum@n@recorded_vs_generated.m1.marginal.gemini37_flash.recorded\endcsname{275}
\expandafter\gdef\csname odunum@val@recorded_vs_generated.m1.marginal.ming.generated\endcsname{0.574}
\expandafter\gdef\csname odunum@n@recorded_vs_generated.m1.marginal.ming.generated\endcsname{1\,801}
\expandafter\gdef\csname odunum@val@recorded_vs_generated.m1.marginal.ming.recorded\endcsname{0.436}
\expandafter\gdef\csname odunum@n@recorded_vs_generated.m1.marginal.ming.recorded\endcsname{277}
\expandafter\gdef\csname odunum@val@recorded_vs_generated.m1.marginal.minicpm_o.generated\endcsname{0.605}
\expandafter\gdef\csname odunum@n@recorded_vs_generated.m1.marginal.minicpm_o.generated\endcsname{1\,798}
\expandafter\gdef\csname odunum@val@recorded_vs_generated.m1.marginal.minicpm_o.recorded\endcsname{0.454}
\expandafter\gdef\csname odunum@n@recorded_vs_generated.m1.marginal.minicpm_o.recorded\endcsname{277}
\expandafter\gdef\csname odunum@val@recorded_vs_generated.m1.marginal.nemotron.generated\endcsname{0.572}
\expandafter\gdef\csname odunum@n@recorded_vs_generated.m1.marginal.nemotron.generated\endcsname{1\,801}
\expandafter\gdef\csname odunum@val@recorded_vs_generated.m1.marginal.nemotron.recorded\endcsname{0.539}
\expandafter\gdef\csname odunum@n@recorded_vs_generated.m1.marginal.nemotron.recorded\endcsname{277}
\expandafter\gdef\csname odunum@val@recorded_vs_generated.m1.marginal.qwen25_omni.generated\endcsname{0.600}
\expandafter\gdef\csname odunum@n@recorded_vs_generated.m1.marginal.qwen25_omni.generated\endcsname{1\,801}
\expandafter\gdef\csname odunum@val@recorded_vs_generated.m1.marginal.qwen25_omni.recorded\endcsname{0.544}
\expandafter\gdef\csname odunum@n@recorded_vs_generated.m1.marginal.qwen25_omni.recorded\endcsname{277}
\expandafter\gdef\csname odunum@val@recorded_vs_generated.m1.marginal.qwen3_omni_instruct.generated\endcsname{0.565}
\expandafter\gdef\csname odunum@n@recorded_vs_generated.m1.marginal.qwen3_omni_instruct.generated\endcsname{1\,801}
\expandafter\gdef\csname odunum@val@recorded_vs_generated.m1.marginal.qwen3_omni_instruct.recorded\endcsname{0.436}
\expandafter\gdef\csname odunum@n@recorded_vs_generated.m1.marginal.qwen3_omni_instruct.recorded\endcsname{277}
\expandafter\gdef\csname odunum@val@recorded_vs_generated.m1.marginal.qwen3_omni_think.generated\endcsname{0.668}
\expandafter\gdef\csname odunum@n@recorded_vs_generated.m1.marginal.qwen3_omni_think.generated\endcsname{1\,801}
\expandafter\gdef\csname odunum@val@recorded_vs_generated.m1.marginal.qwen3_omni_think.recorded\endcsname{0.487}
\expandafter\gdef\csname odunum@n@recorded_vs_generated.m1.marginal.qwen3_omni_think.recorded\endcsname{277}
\expandafter\gdef\csname odunum@val@recorded_vs_generated.m1.marginal.qwen_plus.generated\endcsname{0.709}
\expandafter\gdef\csname odunum@n@recorded_vs_generated.m1.marginal.qwen_plus.generated\endcsname{1\,801}
\expandafter\gdef\csname odunum@val@recorded_vs_generated.m1.marginal.qwen_plus.recorded\endcsname{0.538}
\expandafter\gdef\csname odunum@n@recorded_vs_generated.m1.marginal.qwen_plus.recorded\endcsname{276}
\expandafter\gdef\csname odunum@val@recorded_vs_generated.m1.marginal.salmonn2_7b.generated\endcsname{0.500}
\expandafter\gdef\csname odunum@n@recorded_vs_generated.m1.marginal.salmonn2_7b.generated\endcsname{1\,785}
\expandafter\gdef\csname odunum@val@recorded_vs_generated.m1.marginal.salmonn2_7b.recorded\endcsname{0.447}
\expandafter\gdef\csname odunum@n@recorded_vs_generated.m1.marginal.salmonn2_7b.recorded\endcsname{273}
\expandafter\gdef\csname odunum@val@recorded_vs_generated.m1.marginal.seed.generated\endcsname{0.596}
\expandafter\gdef\csname odunum@n@recorded_vs_generated.m1.marginal.seed.generated\endcsname{1\,801}
\expandafter\gdef\csname odunum@val@recorded_vs_generated.m1.marginal.seed.recorded\endcsname{0.559}
\expandafter\gdef\csname odunum@n@recorded_vs_generated.m1.marginal.seed.recorded\endcsname{277}
\expandafter\gdef\csname odunum@val@recorded_vs_generated.m1.stratum.cascade_asr.generated\endcsname{0.619}
\expandafter\gdef\csname odunum@n@recorded_vs_generated.m1.stratum.cascade_asr.generated\endcsname{529}
\expandafter\gdef\csname odunum@val@recorded_vs_generated.m1.stratum.cascade_asr.recorded\endcsname{0.613}
\expandafter\gdef\csname odunum@n@recorded_vs_generated.m1.stratum.cascade_asr.recorded\endcsname{276}
\expandafter\gdef\csname odunum@val@recorded_vs_generated.m1.stratum.gemini.generated\endcsname{0.833}
\expandafter\gdef\csname odunum@n@recorded_vs_generated.m1.stratum.gemini.generated\endcsname{529}
\expandafter\gdef\csname odunum@val@recorded_vs_generated.m1.stratum.gemini.recorded\endcsname{0.860}
\expandafter\gdef\csname odunum@n@recorded_vs_generated.m1.stratum.gemini.recorded\endcsname{276}
\expandafter\gdef\csname odunum@val@recorded_vs_generated.m1.stratum.gemini35_flash_lite.generated\endcsname{0.652}
\expandafter\gdef\csname odunum@n@recorded_vs_generated.m1.stratum.gemini35_flash_lite.generated\endcsname{529}
\expandafter\gdef\csname odunum@val@recorded_vs_generated.m1.stratum.gemini35_flash_lite.recorded\endcsname{0.593}
\expandafter\gdef\csname odunum@n@recorded_vs_generated.m1.stratum.gemini35_flash_lite.recorded\endcsname{276}
\expandafter\gdef\csname odunum@val@recorded_vs_generated.m1.stratum.gemini37_flash.generated\endcsname{0.845}
\expandafter\gdef\csname odunum@n@recorded_vs_generated.m1.stratum.gemini37_flash.generated\endcsname{529}
\expandafter\gdef\csname odunum@val@recorded_vs_generated.m1.stratum.gemini37_flash.recorded\endcsname{0.888}
\expandafter\gdef\csname odunum@n@recorded_vs_generated.m1.stratum.gemini37_flash.recorded\endcsname{274}
\expandafter\gdef\csname odunum@val@recorded_vs_generated.m1.stratum.ming.generated\endcsname{0.492}
\expandafter\gdef\csname odunum@n@recorded_vs_generated.m1.stratum.ming.generated\endcsname{529}
\expandafter\gdef\csname odunum@val@recorded_vs_generated.m1.stratum.ming.recorded\endcsname{0.435}
\expandafter\gdef\csname odunum@n@recorded_vs_generated.m1.stratum.ming.recorded\endcsname{276}
\expandafter\gdef\csname odunum@val@recorded_vs_generated.m1.stratum.minicpm_o.generated\endcsname{0.500}
\expandafter\gdef\csname odunum@n@recorded_vs_generated.m1.stratum.minicpm_o.generated\endcsname{527}
\expandafter\gdef\csname odunum@val@recorded_vs_generated.m1.stratum.minicpm_o.recorded\endcsname{0.454}
\expandafter\gdef\csname odunum@n@recorded_vs_generated.m1.stratum.minicpm_o.recorded\endcsname{276}
\expandafter\gdef\csname odunum@val@recorded_vs_generated.m1.stratum.nemotron.generated\endcsname{0.543}
\expandafter\gdef\csname odunum@n@recorded_vs_generated.m1.stratum.nemotron.generated\endcsname{529}
\expandafter\gdef\csname odunum@val@recorded_vs_generated.m1.stratum.nemotron.recorded\endcsname{0.538}
\expandafter\gdef\csname odunum@n@recorded_vs_generated.m1.stratum.nemotron.recorded\endcsname{276}
\expandafter\gdef\csname odunum@val@recorded_vs_generated.m1.stratum.qwen25_omni.generated\endcsname{0.503}
\expandafter\gdef\csname odunum@n@recorded_vs_generated.m1.stratum.qwen25_omni.generated\endcsname{529}
\expandafter\gdef\csname odunum@val@recorded_vs_generated.m1.stratum.qwen25_omni.recorded\endcsname{0.543}
\expandafter\gdef\csname odunum@n@recorded_vs_generated.m1.stratum.qwen25_omni.recorded\endcsname{276}
\expandafter\gdef\csname odunum@val@recorded_vs_generated.m1.stratum.qwen3_omni_instruct.generated\endcsname{0.502}
\expandafter\gdef\csname odunum@n@recorded_vs_generated.m1.stratum.qwen3_omni_instruct.generated\endcsname{529}
\expandafter\gdef\csname odunum@val@recorded_vs_generated.m1.stratum.qwen3_omni_instruct.recorded\endcsname{0.435}
\expandafter\gdef\csname odunum@n@recorded_vs_generated.m1.stratum.qwen3_omni_instruct.recorded\endcsname{276}
\expandafter\gdef\csname odunum@val@recorded_vs_generated.m1.stratum.qwen3_omni_think.generated\endcsname{0.609}
\expandafter\gdef\csname odunum@n@recorded_vs_generated.m1.stratum.qwen3_omni_think.generated\endcsname{529}
\expandafter\gdef\csname odunum@val@recorded_vs_generated.m1.stratum.qwen3_omni_think.recorded\endcsname{0.486}
\expandafter\gdef\csname odunum@n@recorded_vs_generated.m1.stratum.qwen3_omni_think.recorded\endcsname{276}
\expandafter\gdef\csname odunum@val@recorded_vs_generated.m1.stratum.qwen_plus.generated\endcsname{0.607}
\expandafter\gdef\csname odunum@n@recorded_vs_generated.m1.stratum.qwen_plus.generated\endcsname{529}
\expandafter\gdef\csname odunum@val@recorded_vs_generated.m1.stratum.qwen_plus.recorded\endcsname{0.537}
\expandafter\gdef\csname odunum@n@recorded_vs_generated.m1.stratum.qwen_plus.recorded\endcsname{275}
\expandafter\gdef\csname odunum@val@recorded_vs_generated.m1.stratum.salmonn2_7b.generated\endcsname{0.488}
\expandafter\gdef\csname odunum@n@recorded_vs_generated.m1.stratum.salmonn2_7b.generated\endcsname{526}
\expandafter\gdef\csname odunum@val@recorded_vs_generated.m1.stratum.salmonn2_7b.recorded\endcsname{0.447}
\expandafter\gdef\csname odunum@n@recorded_vs_generated.m1.stratum.salmonn2_7b.recorded\endcsname{272}
\expandafter\gdef\csname odunum@val@recorded_vs_generated.m1.stratum.seed.generated\endcsname{0.582}
\expandafter\gdef\csname odunum@n@recorded_vs_generated.m1.stratum.seed.generated\endcsname{529}
\expandafter\gdef\csname odunum@val@recorded_vs_generated.m1.stratum.seed.recorded\endcsname{0.559}
\expandafter\gdef\csname odunum@n@recorded_vs_generated.m1.stratum.seed.recorded\endcsname{276}
\expandafter\gdef\csname odunum@val@recorded_vs_generated.m1_delta.marginal.cascade_asr\endcsname{-0.075}
\expandafter\gdef\csname odunum@n@recorded_vs_generated.m1_delta.marginal.cascade_asr\endcsname{277}
\expandafter\gdef\csname odunum@val@recorded_vs_generated.m1_delta.marginal.gemini\endcsname{0.018}
\expandafter\gdef\csname odunum@n@recorded_vs_generated.m1_delta.marginal.gemini\endcsname{277}
\expandafter\gdef\csname odunum@val@recorded_vs_generated.m1_delta.marginal.gemini35_flash_lite\endcsname{-0.097}
\expandafter\gdef\csname odunum@n@recorded_vs_generated.m1_delta.marginal.gemini35_flash_lite\endcsname{277}
\expandafter\gdef\csname odunum@val@recorded_vs_generated.m1_delta.marginal.gemini37_flash\endcsname{0.034}
\expandafter\gdef\csname odunum@n@recorded_vs_generated.m1_delta.marginal.gemini37_flash\endcsname{275}
\expandafter\gdef\csname odunum@val@recorded_vs_generated.m1_delta.marginal.ming\endcsname{-0.138}
\expandafter\gdef\csname odunum@n@recorded_vs_generated.m1_delta.marginal.ming\endcsname{277}
\expandafter\gdef\csname odunum@val@recorded_vs_generated.m1_delta.marginal.minicpm_o\endcsname{-0.150}
\expandafter\gdef\csname odunum@n@recorded_vs_generated.m1_delta.marginal.minicpm_o\endcsname{277}
\expandafter\gdef\csname odunum@val@recorded_vs_generated.m1_delta.marginal.nemotron\endcsname{-0.033}
\expandafter\gdef\csname odunum@n@recorded_vs_generated.m1_delta.marginal.nemotron\endcsname{277}
\expandafter\gdef\csname odunum@val@recorded_vs_generated.m1_delta.marginal.qwen25_omni\endcsname{-0.056}
\expandafter\gdef\csname odunum@n@recorded_vs_generated.m1_delta.marginal.qwen25_omni\endcsname{277}
\expandafter\gdef\csname odunum@val@recorded_vs_generated.m1_delta.marginal.qwen3_omni_instruct\endcsname{-0.129}
\expandafter\gdef\csname odunum@n@recorded_vs_generated.m1_delta.marginal.qwen3_omni_instruct\endcsname{277}
\expandafter\gdef\csname odunum@val@recorded_vs_generated.m1_delta.marginal.qwen3_omni_think\endcsname{-0.181}
\expandafter\gdef\csname odunum@n@recorded_vs_generated.m1_delta.marginal.qwen3_omni_think\endcsname{277}
\expandafter\gdef\csname odunum@val@recorded_vs_generated.m1_delta.marginal.qwen_plus\endcsname{-0.171}
\expandafter\gdef\csname odunum@n@recorded_vs_generated.m1_delta.marginal.qwen_plus\endcsname{276}
\expandafter\gdef\csname odunum@val@recorded_vs_generated.m1_delta.marginal.salmonn2_7b\endcsname{-0.053}
\expandafter\gdef\csname odunum@n@recorded_vs_generated.m1_delta.marginal.salmonn2_7b\endcsname{273}
\expandafter\gdef\csname odunum@val@recorded_vs_generated.m1_delta.marginal.seed\endcsname{-0.036}
\expandafter\gdef\csname odunum@n@recorded_vs_generated.m1_delta.marginal.seed\endcsname{277}
\expandafter\gdef\csname odunum@val@recorded_vs_generated.m1_delta.stratum.cascade_asr\endcsname{-0.006}
\expandafter\gdef\csname odunum@n@recorded_vs_generated.m1_delta.stratum.cascade_asr\endcsname{276}
\expandafter\gdef\csname odunum@val@recorded_vs_generated.m1_delta.stratum.gemini\endcsname{0.027}
\expandafter\gdef\csname odunum@n@recorded_vs_generated.m1_delta.stratum.gemini\endcsname{276}
\expandafter\gdef\csname odunum@val@recorded_vs_generated.m1_delta.stratum.gemini35_flash_lite\endcsname{-0.059}
\expandafter\gdef\csname odunum@n@recorded_vs_generated.m1_delta.stratum.gemini35_flash_lite\endcsname{276}
\expandafter\gdef\csname odunum@val@recorded_vs_generated.m1_delta.stratum.gemini37_flash\endcsname{0.043}
\expandafter\gdef\csname odunum@n@recorded_vs_generated.m1_delta.stratum.gemini37_flash\endcsname{274}
\expandafter\gdef\csname odunum@val@recorded_vs_generated.m1_delta.stratum.ming\endcsname{-0.056}
\expandafter\gdef\csname odunum@n@recorded_vs_generated.m1_delta.stratum.ming\endcsname{276}
\expandafter\gdef\csname odunum@val@recorded_vs_generated.m1_delta.stratum.minicpm_o\endcsname{-0.046}
\expandafter\gdef\csname odunum@n@recorded_vs_generated.m1_delta.stratum.minicpm_o\endcsname{276}
\expandafter\gdef\csname odunum@val@recorded_vs_generated.m1_delta.stratum.nemotron\endcsname{-0.005}
\expandafter\gdef\csname odunum@n@recorded_vs_generated.m1_delta.stratum.nemotron\endcsname{276}
\expandafter\gdef\csname odunum@val@recorded_vs_generated.m1_delta.stratum.qwen25_omni\endcsname{0.041}
\expandafter\gdef\csname odunum@n@recorded_vs_generated.m1_delta.stratum.qwen25_omni\endcsname{276}
\expandafter\gdef\csname odunum@val@recorded_vs_generated.m1_delta.stratum.qwen3_omni_instruct\endcsname{-0.066}
\expandafter\gdef\csname odunum@n@recorded_vs_generated.m1_delta.stratum.qwen3_omni_instruct\endcsname{276}
\expandafter\gdef\csname odunum@val@recorded_vs_generated.m1_delta.stratum.qwen3_omni_think\endcsname{-0.123}
\expandafter\gdef\csname odunum@n@recorded_vs_generated.m1_delta.stratum.qwen3_omni_think\endcsname{276}
\expandafter\gdef\csname odunum@val@recorded_vs_generated.m1_delta.stratum.qwen_plus\endcsname{-0.069}
\expandafter\gdef\csname odunum@n@recorded_vs_generated.m1_delta.stratum.qwen_plus\endcsname{275}
\expandafter\gdef\csname odunum@val@recorded_vs_generated.m1_delta.stratum.salmonn2_7b\endcsname{-0.041}
\expandafter\gdef\csname odunum@n@recorded_vs_generated.m1_delta.stratum.salmonn2_7b\endcsname{272}
\expandafter\gdef\csname odunum@val@recorded_vs_generated.m1_delta.stratum.seed\endcsname{-0.022}
\expandafter\gdef\csname odunum@n@recorded_vs_generated.m1_delta.stratum.seed\endcsname{276}
\expandafter\gdef\csname odunum@val@recorded_vs_generated.m2.marginal.cascade_asr.generated\endcsname{0.612}
\expandafter\gdef\csname odunum@n@recorded_vs_generated.m2.marginal.cascade_asr.generated\endcsname{1\,429}
\expandafter\gdef\csname odunum@ci@recorded_vs_generated.m2.marginal.cascade_asr.generated\endcsname{[0.597, 0.627]}
\expandafter\gdef\csname odunum@val@recorded_vs_generated.m2.marginal.cascade_asr.recorded\endcsname{0.514}
\expandafter\gdef\csname odunum@n@recorded_vs_generated.m2.marginal.cascade_asr.recorded\endcsname{202}
\expandafter\gdef\csname odunum@ci@recorded_vs_generated.m2.marginal.cascade_asr.recorded\endcsname{[0.472, 0.556]}
\expandafter\gdef\csname odunum@val@recorded_vs_generated.m2.marginal.gemini.generated\endcsname{0.658}
\expandafter\gdef\csname odunum@n@recorded_vs_generated.m2.marginal.gemini.generated\endcsname{1\,429}
\expandafter\gdef\csname odunum@ci@recorded_vs_generated.m2.marginal.gemini.generated\endcsname{[0.645, 0.671]}
\expandafter\gdef\csname odunum@val@recorded_vs_generated.m2.marginal.gemini.recorded\endcsname{0.652}
\expandafter\gdef\csname odunum@n@recorded_vs_generated.m2.marginal.gemini.recorded\endcsname{202}
\expandafter\gdef\csname odunum@ci@recorded_vs_generated.m2.marginal.gemini.recorded\endcsname{[0.616, 0.687]}
\expandafter\gdef\csname odunum@val@recorded_vs_generated.m2.marginal.gemini35_flash_lite.generated\endcsname{0.543}
\expandafter\gdef\csname odunum@n@recorded_vs_generated.m2.marginal.gemini35_flash_lite.generated\endcsname{1\,429}
\expandafter\gdef\csname odunum@ci@recorded_vs_generated.m2.marginal.gemini35_flash_lite.generated\endcsname{[0.526, 0.559]}
\expandafter\gdef\csname odunum@val@recorded_vs_generated.m2.marginal.gemini35_flash_lite.recorded\endcsname{0.373}
\expandafter\gdef\csname odunum@n@recorded_vs_generated.m2.marginal.gemini35_flash_lite.recorded\endcsname{202}
\expandafter\gdef\csname odunum@ci@recorded_vs_generated.m2.marginal.gemini35_flash_lite.recorded\endcsname{[0.332, 0.417]}
\expandafter\gdef\csname odunum@val@recorded_vs_generated.m2.marginal.gemini37_flash.generated\endcsname{0.626}
\expandafter\gdef\csname odunum@n@recorded_vs_generated.m2.marginal.gemini37_flash.generated\endcsname{1\,426}
\expandafter\gdef\csname odunum@ci@recorded_vs_generated.m2.marginal.gemini37_flash.generated\endcsname{[0.612, 0.641]}
\expandafter\gdef\csname odunum@val@recorded_vs_generated.m2.marginal.gemini37_flash.recorded\endcsname{0.594}
\expandafter\gdef\csname odunum@n@recorded_vs_generated.m2.marginal.gemini37_flash.recorded\endcsname{201}
\expandafter\gdef\csname odunum@ci@recorded_vs_generated.m2.marginal.gemini37_flash.recorded\endcsname{[0.553, 0.636]}
\expandafter\gdef\csname odunum@val@recorded_vs_generated.m2.marginal.ming.generated\endcsname{0.502}
\expandafter\gdef\csname odunum@n@recorded_vs_generated.m2.marginal.ming.generated\endcsname{1\,429}
\expandafter\gdef\csname odunum@ci@recorded_vs_generated.m2.marginal.ming.generated\endcsname{[0.488, 0.517]}
\expandafter\gdef\csname odunum@val@recorded_vs_generated.m2.marginal.ming.recorded\endcsname{0.475}
\expandafter\gdef\csname odunum@n@recorded_vs_generated.m2.marginal.ming.recorded\endcsname{202}
\expandafter\gdef\csname odunum@ci@recorded_vs_generated.m2.marginal.ming.recorded\endcsname{[0.436, 0.515]}
\expandafter\gdef\csname odunum@val@recorded_vs_generated.m2.marginal.minicpm_o.generated\endcsname{0.420}
\expandafter\gdef\csname odunum@n@recorded_vs_generated.m2.marginal.minicpm_o.generated\endcsname{1\,427}
\expandafter\gdef\csname odunum@ci@recorded_vs_generated.m2.marginal.minicpm_o.generated\endcsname{[0.403, 0.437]}
\expandafter\gdef\csname odunum@val@recorded_vs_generated.m2.marginal.minicpm_o.recorded\endcsname{0.245}
\expandafter\gdef\csname odunum@n@recorded_vs_generated.m2.marginal.minicpm_o.recorded\endcsname{202}
\expandafter\gdef\csname odunum@ci@recorded_vs_generated.m2.marginal.minicpm_o.recorded\endcsname{[0.204, 0.290]}
\expandafter\gdef\csname odunum@val@recorded_vs_generated.m2.marginal.nemotron.generated\endcsname{0.329}
\expandafter\gdef\csname odunum@n@recorded_vs_generated.m2.marginal.nemotron.generated\endcsname{1\,429}
\expandafter\gdef\csname odunum@ci@recorded_vs_generated.m2.marginal.nemotron.generated\endcsname{[0.311, 0.346]}
\expandafter\gdef\csname odunum@val@recorded_vs_generated.m2.marginal.nemotron.recorded\endcsname{0.035}
\expandafter\gdef\csname odunum@n@recorded_vs_generated.m2.marginal.nemotron.recorded\endcsname{202}
\expandafter\gdef\csname odunum@ci@recorded_vs_generated.m2.marginal.nemotron.recorded\endcsname{[0.018, 0.054]}
\expandafter\gdef\csname odunum@val@recorded_vs_generated.m2.marginal.qwen25_omni.generated\endcsname{0.456}
\expandafter\gdef\csname odunum@n@recorded_vs_generated.m2.marginal.qwen25_omni.generated\endcsname{1\,429}
\expandafter\gdef\csname odunum@ci@recorded_vs_generated.m2.marginal.qwen25_omni.generated\endcsname{[0.440, 0.471]}
\expandafter\gdef\csname odunum@val@recorded_vs_generated.m2.marginal.qwen25_omni.recorded\endcsname{0.274}
\expandafter\gdef\csname odunum@n@recorded_vs_generated.m2.marginal.qwen25_omni.recorded\endcsname{202}
\expandafter\gdef\csname odunum@ci@recorded_vs_generated.m2.marginal.qwen25_omni.recorded\endcsname{[0.235, 0.314]}
\expandafter\gdef\csname odunum@val@recorded_vs_generated.m2.marginal.qwen3_omni_instruct.generated\endcsname{0.568}
\expandafter\gdef\csname odunum@n@recorded_vs_generated.m2.marginal.qwen3_omni_instruct.generated\endcsname{1\,429}
\expandafter\gdef\csname odunum@ci@recorded_vs_generated.m2.marginal.qwen3_omni_instruct.generated\endcsname{[0.554, 0.581]}
\expandafter\gdef\csname odunum@val@recorded_vs_generated.m2.marginal.qwen3_omni_instruct.recorded\endcsname{0.490}
\expandafter\gdef\csname odunum@n@recorded_vs_generated.m2.marginal.qwen3_omni_instruct.recorded\endcsname{202}
\expandafter\gdef\csname odunum@ci@recorded_vs_generated.m2.marginal.qwen3_omni_instruct.recorded\endcsname{[0.447, 0.532]}
\expandafter\gdef\csname odunum@val@recorded_vs_generated.m2.marginal.qwen3_omni_think.generated\endcsname{0.595}
\expandafter\gdef\csname odunum@n@recorded_vs_generated.m2.marginal.qwen3_omni_think.generated\endcsname{1\,429}
\expandafter\gdef\csname odunum@ci@recorded_vs_generated.m2.marginal.qwen3_omni_think.generated\endcsname{[0.581, 0.608]}
\expandafter\gdef\csname odunum@val@recorded_vs_generated.m2.marginal.qwen3_omni_think.recorded\endcsname{0.494}
\expandafter\gdef\csname odunum@n@recorded_vs_generated.m2.marginal.qwen3_omni_think.recorded\endcsname{202}
\expandafter\gdef\csname odunum@ci@recorded_vs_generated.m2.marginal.qwen3_omni_think.recorded\endcsname{[0.455, 0.533]}
\expandafter\gdef\csname odunum@val@recorded_vs_generated.m2.marginal.qwen_plus.generated\endcsname{0.683}
\expandafter\gdef\csname odunum@n@recorded_vs_generated.m2.marginal.qwen_plus.generated\endcsname{1\,429}
\expandafter\gdef\csname odunum@ci@recorded_vs_generated.m2.marginal.qwen_plus.generated\endcsname{[0.670, 0.696]}
\expandafter\gdef\csname odunum@val@recorded_vs_generated.m2.marginal.qwen_plus.recorded\endcsname{0.642}
\expandafter\gdef\csname odunum@n@recorded_vs_generated.m2.marginal.qwen_plus.recorded\endcsname{201}
\expandafter\gdef\csname odunum@ci@recorded_vs_generated.m2.marginal.qwen_plus.recorded\endcsname{[0.603, 0.680]}
\expandafter\gdef\csname odunum@val@recorded_vs_generated.m2.marginal.salmonn2_7b.generated\endcsname{0.121}
\expandafter\gdef\csname odunum@n@recorded_vs_generated.m2.marginal.salmonn2_7b.generated\endcsname{1\,415}
\expandafter\gdef\csname odunum@ci@recorded_vs_generated.m2.marginal.salmonn2_7b.generated\endcsname{[0.110, 0.133]}
\expandafter\gdef\csname odunum@val@recorded_vs_generated.m2.marginal.salmonn2_7b.recorded\endcsname{0.011}
\expandafter\gdef\csname odunum@n@recorded_vs_generated.m2.marginal.salmonn2_7b.recorded\endcsname{199}
\expandafter\gdef\csname odunum@ci@recorded_vs_generated.m2.marginal.salmonn2_7b.recorded\endcsname{[0.003, 0.021]}
\expandafter\gdef\csname odunum@val@recorded_vs_generated.m2.marginal.seed.generated\endcsname{0.698}
\expandafter\gdef\csname odunum@n@recorded_vs_generated.m2.marginal.seed.generated\endcsname{1\,429}
\expandafter\gdef\csname odunum@ci@recorded_vs_generated.m2.marginal.seed.generated\endcsname{[0.683, 0.712]}
\expandafter\gdef\csname odunum@val@recorded_vs_generated.m2.marginal.seed.recorded\endcsname{0.637}
\expandafter\gdef\csname odunum@n@recorded_vs_generated.m2.marginal.seed.recorded\endcsname{202}
\expandafter\gdef\csname odunum@ci@recorded_vs_generated.m2.marginal.seed.recorded\endcsname{[0.591, 0.683]}
\expandafter\gdef\csname odunum@val@recorded_vs_generated.m2.stratum.cascade_asr.generated\endcsname{0.577}
\expandafter\gdef\csname odunum@n@recorded_vs_generated.m2.stratum.cascade_asr.generated\endcsname{424}
\expandafter\gdef\csname odunum@ci@recorded_vs_generated.m2.stratum.cascade_asr.generated\endcsname{[0.550, 0.602]}
\expandafter\gdef\csname odunum@val@recorded_vs_generated.m2.stratum.cascade_asr.recorded\endcsname{0.515}
\expandafter\gdef\csname odunum@n@recorded_vs_generated.m2.stratum.cascade_asr.recorded\endcsname{201}
\expandafter\gdef\csname odunum@ci@recorded_vs_generated.m2.stratum.cascade_asr.recorded\endcsname{[0.471, 0.558]}
\expandafter\gdef\csname odunum@val@recorded_vs_generated.m2.stratum.gemini.generated\endcsname{0.620}
\expandafter\gdef\csname odunum@n@recorded_vs_generated.m2.stratum.gemini.generated\endcsname{424}
\expandafter\gdef\csname odunum@ci@recorded_vs_generated.m2.stratum.gemini.generated\endcsname{[0.596, 0.643]}
\expandafter\gdef\csname odunum@val@recorded_vs_generated.m2.stratum.gemini.recorded\endcsname{0.651}
\expandafter\gdef\csname odunum@n@recorded_vs_generated.m2.stratum.gemini.recorded\endcsname{201}
\expandafter\gdef\csname odunum@ci@recorded_vs_generated.m2.stratum.gemini.recorded\endcsname{[0.615, 0.688]}
\expandafter\gdef\csname odunum@val@recorded_vs_generated.m2.stratum.gemini35_flash_lite.generated\endcsname{0.508}
\expandafter\gdef\csname odunum@n@recorded_vs_generated.m2.stratum.gemini35_flash_lite.generated\endcsname{424}
\expandafter\gdef\csname odunum@ci@recorded_vs_generated.m2.stratum.gemini35_flash_lite.generated\endcsname{[0.477, 0.539]}
\expandafter\gdef\csname odunum@val@recorded_vs_generated.m2.stratum.gemini35_flash_lite.recorded\endcsname{0.372}
\expandafter\gdef\csname odunum@n@recorded_vs_generated.m2.stratum.gemini35_flash_lite.recorded\endcsname{201}
\expandafter\gdef\csname odunum@ci@recorded_vs_generated.m2.stratum.gemini35_flash_lite.recorded\endcsname{[0.329, 0.415]}
\expandafter\gdef\csname odunum@val@recorded_vs_generated.m2.stratum.gemini37_flash.generated\endcsname{0.597}
\expandafter\gdef\csname odunum@n@recorded_vs_generated.m2.stratum.gemini37_flash.generated\endcsname{424}
\expandafter\gdef\csname odunum@ci@recorded_vs_generated.m2.stratum.gemini37_flash.generated\endcsname{[0.572, 0.623]}
\expandafter\gdef\csname odunum@val@recorded_vs_generated.m2.stratum.gemini37_flash.recorded\endcsname{0.594}
\expandafter\gdef\csname odunum@n@recorded_vs_generated.m2.stratum.gemini37_flash.recorded\endcsname{200}
\expandafter\gdef\csname odunum@ci@recorded_vs_generated.m2.stratum.gemini37_flash.recorded\endcsname{[0.552, 0.635]}
\expandafter\gdef\csname odunum@val@recorded_vs_generated.m2.stratum.ming.generated\endcsname{0.508}
\expandafter\gdef\csname odunum@n@recorded_vs_generated.m2.stratum.ming.generated\endcsname{424}
\expandafter\gdef\csname odunum@ci@recorded_vs_generated.m2.stratum.ming.generated\endcsname{[0.484, 0.533]}
\expandafter\gdef\csname odunum@val@recorded_vs_generated.m2.stratum.ming.recorded\endcsname{0.476}
\expandafter\gdef\csname odunum@n@recorded_vs_generated.m2.stratum.ming.recorded\endcsname{201}
\expandafter\gdef\csname odunum@ci@recorded_vs_generated.m2.stratum.ming.recorded\endcsname{[0.436, 0.516]}
\expandafter\gdef\csname odunum@val@recorded_vs_generated.m2.stratum.minicpm_o.generated\endcsname{0.375}
\expandafter\gdef\csname odunum@n@recorded_vs_generated.m2.stratum.minicpm_o.generated\endcsname{423}
\expandafter\gdef\csname odunum@ci@recorded_vs_generated.m2.stratum.minicpm_o.generated\endcsname{[0.347, 0.404]}
\expandafter\gdef\csname odunum@val@recorded_vs_generated.m2.stratum.minicpm_o.recorded\endcsname{0.245}
\expandafter\gdef\csname odunum@n@recorded_vs_generated.m2.stratum.minicpm_o.recorded\endcsname{201}
\expandafter\gdef\csname odunum@ci@recorded_vs_generated.m2.stratum.minicpm_o.recorded\endcsname{[0.201, 0.289]}
\expandafter\gdef\csname odunum@val@recorded_vs_generated.m2.stratum.nemotron.generated\endcsname{0.095}
\expandafter\gdef\csname odunum@n@recorded_vs_generated.m2.stratum.nemotron.generated\endcsname{424}
\expandafter\gdef\csname odunum@ci@recorded_vs_generated.m2.stratum.nemotron.generated\endcsname{[0.077, 0.113]}
\expandafter\gdef\csname odunum@val@recorded_vs_generated.m2.stratum.nemotron.recorded\endcsname{0.035}
\expandafter\gdef\csname odunum@n@recorded_vs_generated.m2.stratum.nemotron.recorded\endcsname{201}
\expandafter\gdef\csname odunum@ci@recorded_vs_generated.m2.stratum.nemotron.recorded\endcsname{[0.018, 0.054]}
\expandafter\gdef\csname odunum@val@recorded_vs_generated.m2.stratum.qwen25_omni.generated\endcsname{0.428}
\expandafter\gdef\csname odunum@n@recorded_vs_generated.m2.stratum.qwen25_omni.generated\endcsname{424}
\expandafter\gdef\csname odunum@ci@recorded_vs_generated.m2.stratum.qwen25_omni.generated\endcsname{[0.401, 0.455]}
\expandafter\gdef\csname odunum@val@recorded_vs_generated.m2.stratum.qwen25_omni.recorded\endcsname{0.274}
\expandafter\gdef\csname odunum@n@recorded_vs_generated.m2.stratum.qwen25_omni.recorded\endcsname{201}
\expandafter\gdef\csname odunum@ci@recorded_vs_generated.m2.stratum.qwen25_omni.recorded\endcsname{[0.234, 0.314]}
\expandafter\gdef\csname odunum@val@recorded_vs_generated.m2.stratum.qwen3_omni_instruct.generated\endcsname{0.550}
\expandafter\gdef\csname odunum@n@recorded_vs_generated.m2.stratum.qwen3_omni_instruct.generated\endcsname{424}
\expandafter\gdef\csname odunum@ci@recorded_vs_generated.m2.stratum.qwen3_omni_instruct.generated\endcsname{[0.528, 0.573]}
\expandafter\gdef\csname odunum@val@recorded_vs_generated.m2.stratum.qwen3_omni_instruct.recorded\endcsname{0.490}
\expandafter\gdef\csname odunum@n@recorded_vs_generated.m2.stratum.qwen3_omni_instruct.recorded\endcsname{201}
\expandafter\gdef\csname odunum@ci@recorded_vs_generated.m2.stratum.qwen3_omni_instruct.recorded\endcsname{[0.448, 0.533]}
\expandafter\gdef\csname odunum@val@recorded_vs_generated.m2.stratum.qwen3_omni_think.generated\endcsname{0.564}
\expandafter\gdef\csname odunum@n@recorded_vs_generated.m2.stratum.qwen3_omni_think.generated\endcsname{424}
\expandafter\gdef\csname odunum@ci@recorded_vs_generated.m2.stratum.qwen3_omni_think.generated\endcsname{[0.540, 0.587]}
\expandafter\gdef\csname odunum@val@recorded_vs_generated.m2.stratum.qwen3_omni_think.recorded\endcsname{0.495}
\expandafter\gdef\csname odunum@n@recorded_vs_generated.m2.stratum.qwen3_omni_think.recorded\endcsname{201}
\expandafter\gdef\csname odunum@ci@recorded_vs_generated.m2.stratum.qwen3_omni_think.recorded\endcsname{[0.456, 0.534]}
\expandafter\gdef\csname odunum@val@recorded_vs_generated.m2.stratum.qwen_plus.generated\endcsname{0.649}
\expandafter\gdef\csname odunum@n@recorded_vs_generated.m2.stratum.qwen_plus.generated\endcsname{424}
\expandafter\gdef\csname odunum@ci@recorded_vs_generated.m2.stratum.qwen_plus.generated\endcsname{[0.625, 0.672]}
\expandafter\gdef\csname odunum@val@recorded_vs_generated.m2.stratum.qwen_plus.recorded\endcsname{0.642}
\expandafter\gdef\csname odunum@n@recorded_vs_generated.m2.stratum.qwen_plus.recorded\endcsname{200}
\expandafter\gdef\csname odunum@ci@recorded_vs_generated.m2.stratum.qwen_plus.recorded\endcsname{[0.603, 0.679]}
\expandafter\gdef\csname odunum@val@recorded_vs_generated.m2.stratum.salmonn2_7b.generated\endcsname{0.007}
\expandafter\gdef\csname odunum@n@recorded_vs_generated.m2.stratum.salmonn2_7b.generated\endcsname{421}
\expandafter\gdef\csname odunum@ci@recorded_vs_generated.m2.stratum.salmonn2_7b.generated\endcsname{[0.002, 0.013]}
\expandafter\gdef\csname odunum@val@recorded_vs_generated.m2.stratum.salmonn2_7b.recorded\endcsname{0.011}
\expandafter\gdef\csname odunum@n@recorded_vs_generated.m2.stratum.salmonn2_7b.recorded\endcsname{198}
\expandafter\gdef\csname odunum@ci@recorded_vs_generated.m2.stratum.salmonn2_7b.recorded\endcsname{[0.003, 0.021]}
\expandafter\gdef\csname odunum@val@recorded_vs_generated.m2.stratum.seed.generated\endcsname{0.639}
\expandafter\gdef\csname odunum@n@recorded_vs_generated.m2.stratum.seed.generated\endcsname{424}
\expandafter\gdef\csname odunum@ci@recorded_vs_generated.m2.stratum.seed.generated\endcsname{[0.610, 0.667]}
\expandafter\gdef\csname odunum@val@recorded_vs_generated.m2.stratum.seed.recorded\endcsname{0.637}
\expandafter\gdef\csname odunum@n@recorded_vs_generated.m2.stratum.seed.recorded\endcsname{201}
\expandafter\gdef\csname odunum@ci@recorded_vs_generated.m2.stratum.seed.recorded\endcsname{[0.590, 0.684]}
\expandafter\gdef\csname odunum@val@recorded_vs_generated.m2_delta.marginal.cascade_asr\endcsname{-0.098}
\expandafter\gdef\csname odunum@n@recorded_vs_generated.m2_delta.marginal.cascade_asr\endcsname{202}
\expandafter\gdef\csname odunum@val@recorded_vs_generated.m2_delta.marginal.gemini\endcsname{-0.006}
\expandafter\gdef\csname odunum@n@recorded_vs_generated.m2_delta.marginal.gemini\endcsname{202}
\expandafter\gdef\csname odunum@val@recorded_vs_generated.m2_delta.marginal.gemini35_flash_lite\endcsname{-0.169}
\expandafter\gdef\csname odunum@n@recorded_vs_generated.m2_delta.marginal.gemini35_flash_lite\endcsname{202}
\expandafter\gdef\csname odunum@val@recorded_vs_generated.m2_delta.marginal.gemini37_flash\endcsname{-0.032}
\expandafter\gdef\csname odunum@n@recorded_vs_generated.m2_delta.marginal.gemini37_flash\endcsname{201}
\expandafter\gdef\csname odunum@val@recorded_vs_generated.m2_delta.marginal.ming\endcsname{-0.027}
\expandafter\gdef\csname odunum@n@recorded_vs_generated.m2_delta.marginal.ming\endcsname{202}
\expandafter\gdef\csname odunum@val@recorded_vs_generated.m2_delta.marginal.minicpm_o\endcsname{-0.174}
\expandafter\gdef\csname odunum@n@recorded_vs_generated.m2_delta.marginal.minicpm_o\endcsname{202}
\expandafter\gdef\csname odunum@val@recorded_vs_generated.m2_delta.marginal.nemotron\endcsname{-0.294}
\expandafter\gdef\csname odunum@n@recorded_vs_generated.m2_delta.marginal.nemotron\endcsname{202}
\expandafter\gdef\csname odunum@val@recorded_vs_generated.m2_delta.marginal.qwen25_omni\endcsname{-0.181}
\expandafter\gdef\csname odunum@n@recorded_vs_generated.m2_delta.marginal.qwen25_omni\endcsname{202}
\expandafter\gdef\csname odunum@val@recorded_vs_generated.m2_delta.marginal.qwen3_omni_instruct\endcsname{-0.078}
\expandafter\gdef\csname odunum@n@recorded_vs_generated.m2_delta.marginal.qwen3_omni_instruct\endcsname{202}
\expandafter\gdef\csname odunum@val@recorded_vs_generated.m2_delta.marginal.qwen3_omni_think\endcsname{-0.100}
\expandafter\gdef\csname odunum@n@recorded_vs_generated.m2_delta.marginal.qwen3_omni_think\endcsname{202}
\expandafter\gdef\csname odunum@val@recorded_vs_generated.m2_delta.marginal.qwen_plus\endcsname{-0.041}
\expandafter\gdef\csname odunum@n@recorded_vs_generated.m2_delta.marginal.qwen_plus\endcsname{201}
\expandafter\gdef\csname odunum@val@recorded_vs_generated.m2_delta.marginal.salmonn2_7b\endcsname{-0.111}
\expandafter\gdef\csname odunum@n@recorded_vs_generated.m2_delta.marginal.salmonn2_7b\endcsname{199}
\expandafter\gdef\csname odunum@val@recorded_vs_generated.m2_delta.marginal.seed\endcsname{-0.060}
\expandafter\gdef\csname odunum@n@recorded_vs_generated.m2_delta.marginal.seed\endcsname{202}
\expandafter\gdef\csname odunum@val@recorded_vs_generated.m2_delta.stratum.cascade_asr\endcsname{-0.061}
\expandafter\gdef\csname odunum@n@recorded_vs_generated.m2_delta.stratum.cascade_asr\endcsname{201}
\expandafter\gdef\csname odunum@val@recorded_vs_generated.m2_delta.stratum.gemini\endcsname{0.032}
\expandafter\gdef\csname odunum@n@recorded_vs_generated.m2_delta.stratum.gemini\endcsname{201}
\expandafter\gdef\csname odunum@val@recorded_vs_generated.m2_delta.stratum.gemini35_flash_lite\endcsname{-0.136}
\expandafter\gdef\csname odunum@n@recorded_vs_generated.m2_delta.stratum.gemini35_flash_lite\endcsname{201}
\expandafter\gdef\csname odunum@val@recorded_vs_generated.m2_delta.stratum.gemini37_flash\endcsname{-0.004}
\expandafter\gdef\csname odunum@n@recorded_vs_generated.m2_delta.stratum.gemini37_flash\endcsname{200}
\expandafter\gdef\csname odunum@val@recorded_vs_generated.m2_delta.stratum.ming\endcsname{-0.032}
\expandafter\gdef\csname odunum@n@recorded_vs_generated.m2_delta.stratum.ming\endcsname{201}
\expandafter\gdef\csname odunum@val@recorded_vs_generated.m2_delta.stratum.minicpm_o\endcsname{-0.130}
\expandafter\gdef\csname odunum@n@recorded_vs_generated.m2_delta.stratum.minicpm_o\endcsname{201}
\expandafter\gdef\csname odunum@val@recorded_vs_generated.m2_delta.stratum.nemotron\endcsname{-0.059}
\expandafter\gdef\csname odunum@n@recorded_vs_generated.m2_delta.stratum.nemotron\endcsname{201}
\expandafter\gdef\csname odunum@val@recorded_vs_generated.m2_delta.stratum.qwen25_omni\endcsname{-0.154}
\expandafter\gdef\csname odunum@n@recorded_vs_generated.m2_delta.stratum.qwen25_omni\endcsname{201}
\expandafter\gdef\csname odunum@val@recorded_vs_generated.m2_delta.stratum.qwen3_omni_instruct\endcsname{-0.060}
\expandafter\gdef\csname odunum@n@recorded_vs_generated.m2_delta.stratum.qwen3_omni_instruct\endcsname{201}
\expandafter\gdef\csname odunum@val@recorded_vs_generated.m2_delta.stratum.qwen3_omni_think\endcsname{-0.069}
\expandafter\gdef\csname odunum@n@recorded_vs_generated.m2_delta.stratum.qwen3_omni_think\endcsname{201}
\expandafter\gdef\csname odunum@val@recorded_vs_generated.m2_delta.stratum.qwen_plus\endcsname{-0.007}
\expandafter\gdef\csname odunum@n@recorded_vs_generated.m2_delta.stratum.qwen_plus\endcsname{200}
\expandafter\gdef\csname odunum@val@recorded_vs_generated.m2_delta.stratum.salmonn2_7b\endcsname{0.004}
\expandafter\gdef\csname odunum@n@recorded_vs_generated.m2_delta.stratum.salmonn2_7b\endcsname{198}
\expandafter\gdef\csname odunum@val@recorded_vs_generated.m2_delta.stratum.seed\endcsname{-0.002}
\expandafter\gdef\csname odunum@n@recorded_vs_generated.m2_delta.stratum.seed\endcsname{201}
\expandafter\gdef\csname odunum@val@recorded_vs_generated.m3.marginal.cascade_asr.generated\endcsname{0.844}
\expandafter\gdef\csname odunum@n@recorded_vs_generated.m3.marginal.cascade_asr.generated\endcsname{1\,429}
\expandafter\gdef\csname odunum@ci@recorded_vs_generated.m3.marginal.cascade_asr.generated\endcsname{[0.828, 0.860]}
\expandafter\gdef\csname odunum@val@recorded_vs_generated.m3.marginal.cascade_asr.recorded\endcsname{0.782}
\expandafter\gdef\csname odunum@n@recorded_vs_generated.m3.marginal.cascade_asr.recorded\endcsname{202}
\expandafter\gdef\csname odunum@ci@recorded_vs_generated.m3.marginal.cascade_asr.recorded\endcsname{[0.735, 0.827]}
\expandafter\gdef\csname odunum@val@recorded_vs_generated.m3.marginal.gemini.generated\endcsname{0.734}
\expandafter\gdef\csname odunum@n@recorded_vs_generated.m3.marginal.gemini.generated\endcsname{1\,429}
\expandafter\gdef\csname odunum@ci@recorded_vs_generated.m3.marginal.gemini.generated\endcsname{[0.722, 0.745]}
\expandafter\gdef\csname odunum@val@recorded_vs_generated.m3.marginal.gemini.recorded\endcsname{0.849}
\expandafter\gdef\csname odunum@n@recorded_vs_generated.m3.marginal.gemini.recorded\endcsname{202}
\expandafter\gdef\csname odunum@ci@recorded_vs_generated.m3.marginal.gemini.recorded\endcsname{[0.828, 0.868]}
\expandafter\gdef\csname odunum@val@recorded_vs_generated.m3.marginal.gemini35_flash_lite.generated\endcsname{0.642}
\expandafter\gdef\csname odunum@n@recorded_vs_generated.m3.marginal.gemini35_flash_lite.generated\endcsname{1\,429}
\expandafter\gdef\csname odunum@ci@recorded_vs_generated.m3.marginal.gemini35_flash_lite.generated\endcsname{[0.625, 0.658]}
\expandafter\gdef\csname odunum@val@recorded_vs_generated.m3.marginal.gemini35_flash_lite.recorded\endcsname{0.636}
\expandafter\gdef\csname odunum@n@recorded_vs_generated.m3.marginal.gemini35_flash_lite.recorded\endcsname{202}
\expandafter\gdef\csname odunum@ci@recorded_vs_generated.m3.marginal.gemini35_flash_lite.recorded\endcsname{[0.585, 0.686]}
\expandafter\gdef\csname odunum@val@recorded_vs_generated.m3.marginal.gemini37_flash.generated\endcsname{0.720}
\expandafter\gdef\csname odunum@n@recorded_vs_generated.m3.marginal.gemini37_flash.generated\endcsname{1\,426}
\expandafter\gdef\csname odunum@ci@recorded_vs_generated.m3.marginal.gemini37_flash.generated\endcsname{[0.706, 0.733]}
\expandafter\gdef\csname odunum@val@recorded_vs_generated.m3.marginal.gemini37_flash.recorded\endcsname{0.816}
\expandafter\gdef\csname odunum@n@recorded_vs_generated.m3.marginal.gemini37_flash.recorded\endcsname{201}
\expandafter\gdef\csname odunum@ci@recorded_vs_generated.m3.marginal.gemini37_flash.recorded\endcsname{[0.780, 0.849]}
\expandafter\gdef\csname odunum@val@recorded_vs_generated.m3.marginal.ming.generated\endcsname{0.196}
\expandafter\gdef\csname odunum@n@recorded_vs_generated.m3.marginal.ming.generated\endcsname{1\,429}
\expandafter\gdef\csname odunum@ci@recorded_vs_generated.m3.marginal.ming.generated\endcsname{[0.181, 0.210]}
\expandafter\gdef\csname odunum@val@recorded_vs_generated.m3.marginal.ming.recorded\endcsname{0.375}
\expandafter\gdef\csname odunum@n@recorded_vs_generated.m3.marginal.ming.recorded\endcsname{202}
\expandafter\gdef\csname odunum@ci@recorded_vs_generated.m3.marginal.ming.recorded\endcsname{[0.331, 0.419]}
\expandafter\gdef\csname odunum@val@recorded_vs_generated.m3.marginal.minicpm_o.generated\endcsname{0.036}
\expandafter\gdef\csname odunum@n@recorded_vs_generated.m3.marginal.minicpm_o.generated\endcsname{1\,427}
\expandafter\gdef\csname odunum@ci@recorded_vs_generated.m3.marginal.minicpm_o.generated\endcsname{[0.031, 0.043]}
\expandafter\gdef\csname odunum@val@recorded_vs_generated.m3.marginal.minicpm_o.recorded\endcsname{0.007}
\expandafter\gdef\csname odunum@n@recorded_vs_generated.m3.marginal.minicpm_o.recorded\endcsname{202}
\expandafter\gdef\csname odunum@ci@recorded_vs_generated.m3.marginal.minicpm_o.recorded\endcsname{[0.002, 0.012]}
\expandafter\gdef\csname odunum@val@recorded_vs_generated.m3.marginal.nemotron.generated\endcsname{0.380}
\expandafter\gdef\csname odunum@n@recorded_vs_generated.m3.marginal.nemotron.generated\endcsname{1\,429}
\expandafter\gdef\csname odunum@ci@recorded_vs_generated.m3.marginal.nemotron.generated\endcsname{[0.364, 0.395]}
\expandafter\gdef\csname odunum@val@recorded_vs_generated.m3.marginal.nemotron.recorded\endcsname{0.305}
\expandafter\gdef\csname odunum@n@recorded_vs_generated.m3.marginal.nemotron.recorded\endcsname{202}
\expandafter\gdef\csname odunum@ci@recorded_vs_generated.m3.marginal.nemotron.recorded\endcsname{[0.264, 0.347]}
\expandafter\gdef\csname odunum@val@recorded_vs_generated.m3.marginal.qwen25_omni.generated\endcsname{0.057}
\expandafter\gdef\csname odunum@n@recorded_vs_generated.m3.marginal.qwen25_omni.generated\endcsname{1\,429}
\expandafter\gdef\csname odunum@ci@recorded_vs_generated.m3.marginal.qwen25_omni.generated\endcsname{[0.051, 0.064]}
\expandafter\gdef\csname odunum@val@recorded_vs_generated.m3.marginal.qwen25_omni.recorded\endcsname{0.059}
\expandafter\gdef\csname odunum@n@recorded_vs_generated.m3.marginal.qwen25_omni.recorded\endcsname{202}
\expandafter\gdef\csname odunum@ci@recorded_vs_generated.m3.marginal.qwen25_omni.recorded\endcsname{[0.043, 0.075]}
\expandafter\gdef\csname odunum@val@recorded_vs_generated.m3.marginal.qwen3_omni_instruct.generated\endcsname{0.219}
\expandafter\gdef\csname odunum@n@recorded_vs_generated.m3.marginal.qwen3_omni_instruct.generated\endcsname{1\,429}
\expandafter\gdef\csname odunum@ci@recorded_vs_generated.m3.marginal.qwen3_omni_instruct.generated\endcsname{[0.204, 0.234]}
\expandafter\gdef\csname odunum@val@recorded_vs_generated.m3.marginal.qwen3_omni_instruct.recorded\endcsname{0.287}
\expandafter\gdef\csname odunum@n@recorded_vs_generated.m3.marginal.qwen3_omni_instruct.recorded\endcsname{202}
\expandafter\gdef\csname odunum@ci@recorded_vs_generated.m3.marginal.qwen3_omni_instruct.recorded\endcsname{[0.250, 0.325]}
\expandafter\gdef\csname odunum@val@recorded_vs_generated.m3.marginal.qwen3_omni_think.generated\endcsname{0.480}
\expandafter\gdef\csname odunum@n@recorded_vs_generated.m3.marginal.qwen3_omni_think.generated\endcsname{1\,429}
\expandafter\gdef\csname odunum@ci@recorded_vs_generated.m3.marginal.qwen3_omni_think.generated\endcsname{[0.465, 0.494]}
\expandafter\gdef\csname odunum@val@recorded_vs_generated.m3.marginal.qwen3_omni_think.recorded\endcsname{0.499}
\expandafter\gdef\csname odunum@n@recorded_vs_generated.m3.marginal.qwen3_omni_think.recorded\endcsname{202}
\expandafter\gdef\csname odunum@ci@recorded_vs_generated.m3.marginal.qwen3_omni_think.recorded\endcsname{[0.461, 0.535]}
\expandafter\gdef\csname odunum@val@recorded_vs_generated.m3.marginal.qwen_plus.generated\endcsname{0.723}
\expandafter\gdef\csname odunum@n@recorded_vs_generated.m3.marginal.qwen_plus.generated\endcsname{1\,429}
\expandafter\gdef\csname odunum@ci@recorded_vs_generated.m3.marginal.qwen_plus.generated\endcsname{[0.711, 0.735]}
\expandafter\gdef\csname odunum@val@recorded_vs_generated.m3.marginal.qwen_plus.recorded\endcsname{0.731}
\expandafter\gdef\csname odunum@n@recorded_vs_generated.m3.marginal.qwen_plus.recorded\endcsname{201}
\expandafter\gdef\csname odunum@ci@recorded_vs_generated.m3.marginal.qwen_plus.recorded\endcsname{[0.697, 0.763]}
\expandafter\gdef\csname odunum@val@recorded_vs_generated.m3.marginal.salmonn2_7b.generated\endcsname{0.057}
\expandafter\gdef\csname odunum@n@recorded_vs_generated.m3.marginal.salmonn2_7b.generated\endcsname{1\,415}
\expandafter\gdef\csname odunum@ci@recorded_vs_generated.m3.marginal.salmonn2_7b.generated\endcsname{[0.049, 0.065]}
\expandafter\gdef\csname odunum@val@recorded_vs_generated.m3.marginal.salmonn2_7b.recorded\endcsname{0.036}
\expandafter\gdef\csname odunum@n@recorded_vs_generated.m3.marginal.salmonn2_7b.recorded\endcsname{199}
\expandafter\gdef\csname odunum@ci@recorded_vs_generated.m3.marginal.salmonn2_7b.recorded\endcsname{[0.021, 0.054]}
\expandafter\gdef\csname odunum@val@recorded_vs_generated.m3.marginal.seed.generated\endcsname{0.782}
\expandafter\gdef\csname odunum@n@recorded_vs_generated.m3.marginal.seed.generated\endcsname{1\,429}
\expandafter\gdef\csname odunum@ci@recorded_vs_generated.m3.marginal.seed.generated\endcsname{[0.771, 0.794]}
\expandafter\gdef\csname odunum@val@recorded_vs_generated.m3.marginal.seed.recorded\endcsname{0.829}
\expandafter\gdef\csname odunum@n@recorded_vs_generated.m3.marginal.seed.recorded\endcsname{202}
\expandafter\gdef\csname odunum@ci@recorded_vs_generated.m3.marginal.seed.recorded\endcsname{[0.799, 0.857]}
\expandafter\gdef\csname odunum@val@recorded_vs_generated.m3.stratum.cascade_asr.generated\endcsname{0.856}
\expandafter\gdef\csname odunum@n@recorded_vs_generated.m3.stratum.cascade_asr.generated\endcsname{424}
\expandafter\gdef\csname odunum@ci@recorded_vs_generated.m3.stratum.cascade_asr.generated\endcsname{[0.826, 0.883]}
\expandafter\gdef\csname odunum@val@recorded_vs_generated.m3.stratum.cascade_asr.recorded\endcsname{0.781}
\expandafter\gdef\csname odunum@n@recorded_vs_generated.m3.stratum.cascade_asr.recorded\endcsname{201}
\expandafter\gdef\csname odunum@ci@recorded_vs_generated.m3.stratum.cascade_asr.recorded\endcsname{[0.733, 0.826]}
\expandafter\gdef\csname odunum@val@recorded_vs_generated.m3.stratum.gemini.generated\endcsname{0.818}
\expandafter\gdef\csname odunum@n@recorded_vs_generated.m3.stratum.gemini.generated\endcsname{424}
\expandafter\gdef\csname odunum@ci@recorded_vs_generated.m3.stratum.gemini.generated\endcsname{[0.802, 0.833]}
\expandafter\gdef\csname odunum@val@recorded_vs_generated.m3.stratum.gemini.recorded\endcsname{0.849}
\expandafter\gdef\csname odunum@n@recorded_vs_generated.m3.stratum.gemini.recorded\endcsname{201}
\expandafter\gdef\csname odunum@ci@recorded_vs_generated.m3.stratum.gemini.recorded\endcsname{[0.828, 0.867]}
\expandafter\gdef\csname odunum@val@recorded_vs_generated.m3.stratum.gemini35_flash_lite.generated\endcsname{0.691}
\expandafter\gdef\csname odunum@n@recorded_vs_generated.m3.stratum.gemini35_flash_lite.generated\endcsname{424}
\expandafter\gdef\csname odunum@ci@recorded_vs_generated.m3.stratum.gemini35_flash_lite.generated\endcsname{[0.658, 0.722]}
\expandafter\gdef\csname odunum@val@recorded_vs_generated.m3.stratum.gemini35_flash_lite.recorded\endcsname{0.635}
\expandafter\gdef\csname odunum@n@recorded_vs_generated.m3.stratum.gemini35_flash_lite.recorded\endcsname{201}
\expandafter\gdef\csname odunum@ci@recorded_vs_generated.m3.stratum.gemini35_flash_lite.recorded\endcsname{[0.583, 0.685]}
\expandafter\gdef\csname odunum@val@recorded_vs_generated.m3.stratum.gemini37_flash.generated\endcsname{0.783}
\expandafter\gdef\csname odunum@n@recorded_vs_generated.m3.stratum.gemini37_flash.generated\endcsname{424}
\expandafter\gdef\csname odunum@ci@recorded_vs_generated.m3.stratum.gemini37_flash.generated\endcsname{[0.760, 0.805]}
\expandafter\gdef\csname odunum@val@recorded_vs_generated.m3.stratum.gemini37_flash.recorded\endcsname{0.815}
\expandafter\gdef\csname odunum@n@recorded_vs_generated.m3.stratum.gemini37_flash.recorded\endcsname{200}
\expandafter\gdef\csname odunum@ci@recorded_vs_generated.m3.stratum.gemini37_flash.recorded\endcsname{[0.778, 0.848]}
\expandafter\gdef\csname odunum@val@recorded_vs_generated.m3.stratum.ming.generated\endcsname{0.310}
\expandafter\gdef\csname odunum@n@recorded_vs_generated.m3.stratum.ming.generated\endcsname{424}
\expandafter\gdef\csname odunum@ci@recorded_vs_generated.m3.stratum.ming.generated\endcsname{[0.281, 0.339]}
\expandafter\gdef\csname odunum@val@recorded_vs_generated.m3.stratum.ming.recorded\endcsname{0.375}
\expandafter\gdef\csname odunum@n@recorded_vs_generated.m3.stratum.ming.recorded\endcsname{201}
\expandafter\gdef\csname odunum@ci@recorded_vs_generated.m3.stratum.ming.recorded\endcsname{[0.331, 0.418]}
\expandafter\gdef\csname odunum@val@recorded_vs_generated.m3.stratum.minicpm_o.generated\endcsname{0.010}
\expandafter\gdef\csname odunum@n@recorded_vs_generated.m3.stratum.minicpm_o.generated\endcsname{423}
\expandafter\gdef\csname odunum@ci@recorded_vs_generated.m3.stratum.minicpm_o.generated\endcsname{[0.005, 0.016]}
\expandafter\gdef\csname odunum@val@recorded_vs_generated.m3.stratum.minicpm_o.recorded\endcsname{0.007}
\expandafter\gdef\csname odunum@n@recorded_vs_generated.m3.stratum.minicpm_o.recorded\endcsname{201}
\expandafter\gdef\csname odunum@ci@recorded_vs_generated.m3.stratum.minicpm_o.recorded\endcsname{[0.002, 0.012]}
\expandafter\gdef\csname odunum@val@recorded_vs_generated.m3.stratum.nemotron.generated\endcsname{0.312}
\expandafter\gdef\csname odunum@n@recorded_vs_generated.m3.stratum.nemotron.generated\endcsname{424}
\expandafter\gdef\csname odunum@ci@recorded_vs_generated.m3.stratum.nemotron.generated\endcsname{[0.283, 0.341]}
\expandafter\gdef\csname odunum@val@recorded_vs_generated.m3.stratum.nemotron.recorded\endcsname{0.305}
\expandafter\gdef\csname odunum@n@recorded_vs_generated.m3.stratum.nemotron.recorded\endcsname{201}
\expandafter\gdef\csname odunum@ci@recorded_vs_generated.m3.stratum.nemotron.recorded\endcsname{[0.263, 0.346]}
\expandafter\gdef\csname odunum@val@recorded_vs_generated.m3.stratum.qwen25_omni.generated\endcsname{0.057}
\expandafter\gdef\csname odunum@n@recorded_vs_generated.m3.stratum.qwen25_omni.generated\endcsname{424}
\expandafter\gdef\csname odunum@ci@recorded_vs_generated.m3.stratum.qwen25_omni.generated\endcsname{[0.049, 0.065]}
\expandafter\gdef\csname odunum@val@recorded_vs_generated.m3.stratum.qwen25_omni.recorded\endcsname{0.059}
\expandafter\gdef\csname odunum@n@recorded_vs_generated.m3.stratum.qwen25_omni.recorded\endcsname{201}
\expandafter\gdef\csname odunum@ci@recorded_vs_generated.m3.stratum.qwen25_omni.recorded\endcsname{[0.044, 0.075]}
\expandafter\gdef\csname odunum@val@recorded_vs_generated.m3.stratum.qwen3_omni_instruct.generated\endcsname{0.212}
\expandafter\gdef\csname odunum@n@recorded_vs_generated.m3.stratum.qwen3_omni_instruct.generated\endcsname{424}
\expandafter\gdef\csname odunum@ci@recorded_vs_generated.m3.stratum.qwen3_omni_instruct.generated\endcsname{[0.188, 0.236]}
\expandafter\gdef\csname odunum@val@recorded_vs_generated.m3.stratum.qwen3_omni_instruct.recorded\endcsname{0.285}
\expandafter\gdef\csname odunum@n@recorded_vs_generated.m3.stratum.qwen3_omni_instruct.recorded\endcsname{201}
\expandafter\gdef\csname odunum@ci@recorded_vs_generated.m3.stratum.qwen3_omni_instruct.recorded\endcsname{[0.249, 0.323]}
\expandafter\gdef\csname odunum@val@recorded_vs_generated.m3.stratum.qwen3_omni_think.generated\endcsname{0.459}
\expandafter\gdef\csname odunum@n@recorded_vs_generated.m3.stratum.qwen3_omni_think.generated\endcsname{424}
\expandafter\gdef\csname odunum@ci@recorded_vs_generated.m3.stratum.qwen3_omni_think.generated\endcsname{[0.430, 0.487]}
\expandafter\gdef\csname odunum@val@recorded_vs_generated.m3.stratum.qwen3_omni_think.recorded\endcsname{0.498}
\expandafter\gdef\csname odunum@n@recorded_vs_generated.m3.stratum.qwen3_omni_think.recorded\endcsname{201}
\expandafter\gdef\csname odunum@ci@recorded_vs_generated.m3.stratum.qwen3_omni_think.recorded\endcsname{[0.461, 0.536]}
\expandafter\gdef\csname odunum@val@recorded_vs_generated.m3.stratum.qwen_plus.generated\endcsname{0.763}
\expandafter\gdef\csname odunum@n@recorded_vs_generated.m3.stratum.qwen_plus.generated\endcsname{424}
\expandafter\gdef\csname odunum@ci@recorded_vs_generated.m3.stratum.qwen_plus.generated\endcsname{[0.743, 0.783]}
\expandafter\gdef\csname odunum@val@recorded_vs_generated.m3.stratum.qwen_plus.recorded\endcsname{0.731}
\expandafter\gdef\csname odunum@n@recorded_vs_generated.m3.stratum.qwen_plus.recorded\endcsname{200}
\expandafter\gdef\csname odunum@ci@recorded_vs_generated.m3.stratum.qwen_plus.recorded\endcsname{[0.698, 0.762]}
\expandafter\gdef\csname odunum@val@recorded_vs_generated.m3.stratum.salmonn2_7b.generated\endcsname{0.037}
\expandafter\gdef\csname odunum@n@recorded_vs_generated.m3.stratum.salmonn2_7b.generated\endcsname{421}
\expandafter\gdef\csname odunum@ci@recorded_vs_generated.m3.stratum.salmonn2_7b.generated\endcsname{[0.026, 0.048]}
\expandafter\gdef\csname odunum@val@recorded_vs_generated.m3.stratum.salmonn2_7b.recorded\endcsname{0.036}
\expandafter\gdef\csname odunum@n@recorded_vs_generated.m3.stratum.salmonn2_7b.recorded\endcsname{198}
\expandafter\gdef\csname odunum@ci@recorded_vs_generated.m3.stratum.salmonn2_7b.recorded\endcsname{[0.021, 0.055]}
\expandafter\gdef\csname odunum@val@recorded_vs_generated.m3.stratum.seed.generated\endcsname{0.826}
\expandafter\gdef\csname odunum@n@recorded_vs_generated.m3.stratum.seed.generated\endcsname{424}
\expandafter\gdef\csname odunum@ci@recorded_vs_generated.m3.stratum.seed.generated\endcsname{[0.804, 0.848]}
\expandafter\gdef\csname odunum@val@recorded_vs_generated.m3.stratum.seed.recorded\endcsname{0.829}
\expandafter\gdef\csname odunum@n@recorded_vs_generated.m3.stratum.seed.recorded\endcsname{201}
\expandafter\gdef\csname odunum@ci@recorded_vs_generated.m3.stratum.seed.recorded\endcsname{[0.799, 0.856]}
\expandafter\gdef\csname odunum@val@recorded_vs_generated.m3_delta.marginal.cascade_asr\endcsname{-0.062}
\expandafter\gdef\csname odunum@n@recorded_vs_generated.m3_delta.marginal.cascade_asr\endcsname{202}
\expandafter\gdef\csname odunum@val@recorded_vs_generated.m3_delta.marginal.gemini\endcsname{0.115}
\expandafter\gdef\csname odunum@n@recorded_vs_generated.m3_delta.marginal.gemini\endcsname{202}
\expandafter\gdef\csname odunum@val@recorded_vs_generated.m3_delta.marginal.gemini35_flash_lite\endcsname{-0.006}
\expandafter\gdef\csname odunum@n@recorded_vs_generated.m3_delta.marginal.gemini35_flash_lite\endcsname{202}
\expandafter\gdef\csname odunum@val@recorded_vs_generated.m3_delta.marginal.gemini37_flash\endcsname{0.096}
\expandafter\gdef\csname odunum@n@recorded_vs_generated.m3_delta.marginal.gemini37_flash\endcsname{201}
\expandafter\gdef\csname odunum@val@recorded_vs_generated.m3_delta.marginal.ming\endcsname{0.180}
\expandafter\gdef\csname odunum@n@recorded_vs_generated.m3_delta.marginal.ming\endcsname{202}
\expandafter\gdef\csname odunum@val@recorded_vs_generated.m3_delta.marginal.minicpm_o\endcsname{-0.030}
\expandafter\gdef\csname odunum@n@recorded_vs_generated.m3_delta.marginal.minicpm_o\endcsname{202}
\expandafter\gdef\csname odunum@val@recorded_vs_generated.m3_delta.marginal.nemotron\endcsname{-0.074}
\expandafter\gdef\csname odunum@n@recorded_vs_generated.m3_delta.marginal.nemotron\endcsname{202}
\expandafter\gdef\csname odunum@val@recorded_vs_generated.m3_delta.marginal.qwen25_omni\endcsname{0.001}
\expandafter\gdef\csname odunum@n@recorded_vs_generated.m3_delta.marginal.qwen25_omni\endcsname{202}
\expandafter\gdef\csname odunum@val@recorded_vs_generated.m3_delta.marginal.qwen3_omni_instruct\endcsname{0.069}
\expandafter\gdef\csname odunum@n@recorded_vs_generated.m3_delta.marginal.qwen3_omni_instruct\endcsname{202}
\expandafter\gdef\csname odunum@val@recorded_vs_generated.m3_delta.marginal.qwen3_omni_think\endcsname{0.019}
\expandafter\gdef\csname odunum@n@recorded_vs_generated.m3_delta.marginal.qwen3_omni_think\endcsname{202}
\expandafter\gdef\csname odunum@val@recorded_vs_generated.m3_delta.marginal.qwen_plus\endcsname{0.009}
\expandafter\gdef\csname odunum@n@recorded_vs_generated.m3_delta.marginal.qwen_plus\endcsname{201}
\expandafter\gdef\csname odunum@val@recorded_vs_generated.m3_delta.marginal.salmonn2_7b\endcsname{-0.021}
\expandafter\gdef\csname odunum@n@recorded_vs_generated.m3_delta.marginal.salmonn2_7b\endcsname{199}
\expandafter\gdef\csname odunum@val@recorded_vs_generated.m3_delta.marginal.seed\endcsname{0.047}
\expandafter\gdef\csname odunum@n@recorded_vs_generated.m3_delta.marginal.seed\endcsname{202}
\expandafter\gdef\csname odunum@val@recorded_vs_generated.m3_delta.stratum.cascade_asr\endcsname{-0.074}
\expandafter\gdef\csname odunum@n@recorded_vs_generated.m3_delta.stratum.cascade_asr\endcsname{201}
\expandafter\gdef\csname odunum@val@recorded_vs_generated.m3_delta.stratum.gemini\endcsname{0.031}
\expandafter\gdef\csname odunum@n@recorded_vs_generated.m3_delta.stratum.gemini\endcsname{201}
\expandafter\gdef\csname odunum@val@recorded_vs_generated.m3_delta.stratum.gemini35_flash_lite\endcsname{-0.056}
\expandafter\gdef\csname odunum@n@recorded_vs_generated.m3_delta.stratum.gemini35_flash_lite\endcsname{201}
\expandafter\gdef\csname odunum@val@recorded_vs_generated.m3_delta.stratum.gemini37_flash\endcsname{0.032}
\expandafter\gdef\csname odunum@n@recorded_vs_generated.m3_delta.stratum.gemini37_flash\endcsname{200}
\expandafter\gdef\csname odunum@val@recorded_vs_generated.m3_delta.stratum.ming\endcsname{0.065}
\expandafter\gdef\csname odunum@n@recorded_vs_generated.m3_delta.stratum.ming\endcsname{201}
\expandafter\gdef\csname odunum@val@recorded_vs_generated.m3_delta.stratum.minicpm_o\endcsname{-0.003}
\expandafter\gdef\csname odunum@n@recorded_vs_generated.m3_delta.stratum.minicpm_o\endcsname{201}
\expandafter\gdef\csname odunum@val@recorded_vs_generated.m3_delta.stratum.nemotron\endcsname{-0.007}
\expandafter\gdef\csname odunum@n@recorded_vs_generated.m3_delta.stratum.nemotron\endcsname{201}
\expandafter\gdef\csname odunum@val@recorded_vs_generated.m3_delta.stratum.qwen25_omni\endcsname{0.002}
\expandafter\gdef\csname odunum@n@recorded_vs_generated.m3_delta.stratum.qwen25_omni\endcsname{201}
\expandafter\gdef\csname odunum@val@recorded_vs_generated.m3_delta.stratum.qwen3_omni_instruct\endcsname{0.073}
\expandafter\gdef\csname odunum@n@recorded_vs_generated.m3_delta.stratum.qwen3_omni_instruct\endcsname{201}
\expandafter\gdef\csname odunum@val@recorded_vs_generated.m3_delta.stratum.qwen3_omni_think\endcsname{0.039}
\expandafter\gdef\csname odunum@n@recorded_vs_generated.m3_delta.stratum.qwen3_omni_think\endcsname{201}
\expandafter\gdef\csname odunum@val@recorded_vs_generated.m3_delta.stratum.qwen_plus\endcsname{-0.033}
\expandafter\gdef\csname odunum@n@recorded_vs_generated.m3_delta.stratum.qwen_plus\endcsname{200}
\expandafter\gdef\csname odunum@val@recorded_vs_generated.m3_delta.stratum.salmonn2_7b\endcsname{-1.0\ensuremath{\times 10^{-4}}}
\expandafter\gdef\csname odunum@n@recorded_vs_generated.m3_delta.stratum.salmonn2_7b\endcsname{198}
\expandafter\gdef\csname odunum@val@recorded_vs_generated.m3_delta.stratum.seed\endcsname{0.003}
\expandafter\gdef\csname odunum@n@recorded_vs_generated.m3_delta.stratum.seed\endcsname{201}
\expandafter\gdef\csname odunum@val@recorded_vs_generated.m4.marginal.cascade_asr.generated\endcsname{0.839}
\expandafter\gdef\csname odunum@n@recorded_vs_generated.m4.marginal.cascade_asr.generated\endcsname{1\,429}
\expandafter\gdef\csname odunum@ci@recorded_vs_generated.m4.marginal.cascade_asr.generated\endcsname{[0.823, 0.854]}
\expandafter\gdef\csname odunum@val@recorded_vs_generated.m4.marginal.cascade_asr.recorded\endcsname{0.772}
\expandafter\gdef\csname odunum@n@recorded_vs_generated.m4.marginal.cascade_asr.recorded\endcsname{202}
\expandafter\gdef\csname odunum@ci@recorded_vs_generated.m4.marginal.cascade_asr.recorded\endcsname{[0.724, 0.817]}
\expandafter\gdef\csname odunum@val@recorded_vs_generated.m4.marginal.gemini.generated\endcsname{0.919}
\expandafter\gdef\csname odunum@n@recorded_vs_generated.m4.marginal.gemini.generated\endcsname{1\,429}
\expandafter\gdef\csname odunum@ci@recorded_vs_generated.m4.marginal.gemini.generated\endcsname{[0.909, 0.928]}
\expandafter\gdef\csname odunum@val@recorded_vs_generated.m4.marginal.gemini.recorded\endcsname{0.883}
\expandafter\gdef\csname odunum@n@recorded_vs_generated.m4.marginal.gemini.recorded\endcsname{202}
\expandafter\gdef\csname odunum@ci@recorded_vs_generated.m4.marginal.gemini.recorded\endcsname{[0.854, 0.910]}
\expandafter\gdef\csname odunum@val@recorded_vs_generated.m4.marginal.gemini35_flash_lite.generated\endcsname{0.749}
\expandafter\gdef\csname odunum@n@recorded_vs_generated.m4.marginal.gemini35_flash_lite.generated\endcsname{1\,429}
\expandafter\gdef\csname odunum@ci@recorded_vs_generated.m4.marginal.gemini35_flash_lite.generated\endcsname{[0.730, 0.767]}
\expandafter\gdef\csname odunum@val@recorded_vs_generated.m4.marginal.gemini35_flash_lite.recorded\endcsname{0.558}
\expandafter\gdef\csname odunum@n@recorded_vs_generated.m4.marginal.gemini35_flash_lite.recorded\endcsname{202}
\expandafter\gdef\csname odunum@ci@recorded_vs_generated.m4.marginal.gemini35_flash_lite.recorded\endcsname{[0.505, 0.612]}
\expandafter\gdef\csname odunum@val@recorded_vs_generated.m4.marginal.gemini37_flash.generated\endcsname{0.874}
\expandafter\gdef\csname odunum@n@recorded_vs_generated.m4.marginal.gemini37_flash.generated\endcsname{1\,426}
\expandafter\gdef\csname odunum@ci@recorded_vs_generated.m4.marginal.gemini37_flash.generated\endcsname{[0.861, 0.887]}
\expandafter\gdef\csname odunum@val@recorded_vs_generated.m4.marginal.gemini37_flash.recorded\endcsname{0.821}
\expandafter\gdef\csname odunum@n@recorded_vs_generated.m4.marginal.gemini37_flash.recorded\endcsname{201}
\expandafter\gdef\csname odunum@ci@recorded_vs_generated.m4.marginal.gemini37_flash.recorded\endcsname{[0.783, 0.857]}
\expandafter\gdef\csname odunum@val@recorded_vs_generated.m4.marginal.ming.generated\endcsname{0.872}
\expandafter\gdef\csname odunum@n@recorded_vs_generated.m4.marginal.ming.generated\endcsname{1\,429}
\expandafter\gdef\csname odunum@ci@recorded_vs_generated.m4.marginal.ming.generated\endcsname{[0.859, 0.885]}
\expandafter\gdef\csname odunum@val@recorded_vs_generated.m4.marginal.ming.recorded\endcsname{0.624}
\expandafter\gdef\csname odunum@n@recorded_vs_generated.m4.marginal.ming.recorded\endcsname{202}
\expandafter\gdef\csname odunum@ci@recorded_vs_generated.m4.marginal.ming.recorded\endcsname{[0.571, 0.677]}
\expandafter\gdef\csname odunum@val@recorded_vs_generated.m4.marginal.minicpm_o.generated\endcsname{0.632}
\expandafter\gdef\csname odunum@n@recorded_vs_generated.m4.marginal.minicpm_o.generated\endcsname{1\,427}
\expandafter\gdef\csname odunum@ci@recorded_vs_generated.m4.marginal.minicpm_o.generated\endcsname{[0.610, 0.653]}
\expandafter\gdef\csname odunum@val@recorded_vs_generated.m4.marginal.minicpm_o.recorded\endcsname{0.432}
\expandafter\gdef\csname odunum@n@recorded_vs_generated.m4.marginal.minicpm_o.recorded\endcsname{202}
\expandafter\gdef\csname odunum@ci@recorded_vs_generated.m4.marginal.minicpm_o.recorded\endcsname{[0.372, 0.493]}
\expandafter\gdef\csname odunum@val@recorded_vs_generated.m4.marginal.nemotron.generated\endcsname{0.402}
\expandafter\gdef\csname odunum@n@recorded_vs_generated.m4.marginal.nemotron.generated\endcsname{1\,429}
\expandafter\gdef\csname odunum@ci@recorded_vs_generated.m4.marginal.nemotron.generated\endcsname{[0.379, 0.425]}
\expandafter\gdef\csname odunum@val@recorded_vs_generated.m4.marginal.nemotron.recorded\endcsname{0.017}
\expandafter\gdef\csname odunum@n@recorded_vs_generated.m4.marginal.nemotron.recorded\endcsname{202}
\expandafter\gdef\csname odunum@ci@recorded_vs_generated.m4.marginal.nemotron.recorded\endcsname{[0.011, 0.024]}
\expandafter\gdef\csname odunum@val@recorded_vs_generated.m4.marginal.qwen25_omni.generated\endcsname{0.762}
\expandafter\gdef\csname odunum@n@recorded_vs_generated.m4.marginal.qwen25_omni.generated\endcsname{1\,429}
\expandafter\gdef\csname odunum@ci@recorded_vs_generated.m4.marginal.qwen25_omni.generated\endcsname{[0.744, 0.780]}
\expandafter\gdef\csname odunum@val@recorded_vs_generated.m4.marginal.qwen25_omni.recorded\endcsname{0.464}
\expandafter\gdef\csname odunum@n@recorded_vs_generated.m4.marginal.qwen25_omni.recorded\endcsname{202}
\expandafter\gdef\csname odunum@ci@recorded_vs_generated.m4.marginal.qwen25_omni.recorded\endcsname{[0.406, 0.522]}
\expandafter\gdef\csname odunum@val@recorded_vs_generated.m4.marginal.qwen3_omni_instruct.generated\endcsname{0.899}
\expandafter\gdef\csname odunum@n@recorded_vs_generated.m4.marginal.qwen3_omni_instruct.generated\endcsname{1\,429}
\expandafter\gdef\csname odunum@ci@recorded_vs_generated.m4.marginal.qwen3_omni_instruct.generated\endcsname{[0.888, 0.910]}
\expandafter\gdef\csname odunum@val@recorded_vs_generated.m4.marginal.qwen3_omni_instruct.recorded\endcsname{0.783}
\expandafter\gdef\csname odunum@n@recorded_vs_generated.m4.marginal.qwen3_omni_instruct.recorded\endcsname{202}
\expandafter\gdef\csname odunum@ci@recorded_vs_generated.m4.marginal.qwen3_omni_instruct.recorded\endcsname{[0.738, 0.826]}
\expandafter\gdef\csname odunum@val@recorded_vs_generated.m4.marginal.qwen3_omni_think.generated\endcsname{0.883}
\expandafter\gdef\csname odunum@n@recorded_vs_generated.m4.marginal.qwen3_omni_think.generated\endcsname{1\,429}
\expandafter\gdef\csname odunum@ci@recorded_vs_generated.m4.marginal.qwen3_omni_think.generated\endcsname{[0.870, 0.895]}
\expandafter\gdef\csname odunum@val@recorded_vs_generated.m4.marginal.qwen3_omni_think.recorded\endcsname{0.714}
\expandafter\gdef\csname odunum@n@recorded_vs_generated.m4.marginal.qwen3_omni_think.recorded\endcsname{202}
\expandafter\gdef\csname odunum@ci@recorded_vs_generated.m4.marginal.qwen3_omni_think.recorded\endcsname{[0.663, 0.762]}
\expandafter\gdef\csname odunum@val@recorded_vs_generated.m4.marginal.qwen_plus.generated\endcsname{0.873}
\expandafter\gdef\csname odunum@n@recorded_vs_generated.m4.marginal.qwen_plus.generated\endcsname{1\,429}
\expandafter\gdef\csname odunum@ci@recorded_vs_generated.m4.marginal.qwen_plus.generated\endcsname{[0.858, 0.887]}
\expandafter\gdef\csname odunum@val@recorded_vs_generated.m4.marginal.qwen_plus.recorded\endcsname{0.718}
\expandafter\gdef\csname odunum@n@recorded_vs_generated.m4.marginal.qwen_plus.recorded\endcsname{201}
\expandafter\gdef\csname odunum@ci@recorded_vs_generated.m4.marginal.qwen_plus.recorded\endcsname{[0.659, 0.772]}
\expandafter\gdef\csname odunum@val@recorded_vs_generated.m4.marginal.salmonn2_7b.generated\endcsname{0.196}
\expandafter\gdef\csname odunum@n@recorded_vs_generated.m4.marginal.salmonn2_7b.generated\endcsname{1\,415}
\expandafter\gdef\csname odunum@ci@recorded_vs_generated.m4.marginal.salmonn2_7b.generated\endcsname{[0.179, 0.213]}
\expandafter\gdef\csname odunum@val@recorded_vs_generated.m4.marginal.salmonn2_7b.recorded\endcsname{0.002}
\expandafter\gdef\csname odunum@n@recorded_vs_generated.m4.marginal.salmonn2_7b.recorded\endcsname{199}
\expandafter\gdef\csname odunum@ci@recorded_vs_generated.m4.marginal.salmonn2_7b.recorded\endcsname{[0.000, 0.005]}
\expandafter\gdef\csname odunum@val@recorded_vs_generated.m4.marginal.seed.generated\endcsname{0.836}
\expandafter\gdef\csname odunum@n@recorded_vs_generated.m4.marginal.seed.generated\endcsname{1\,429}
\expandafter\gdef\csname odunum@ci@recorded_vs_generated.m4.marginal.seed.generated\endcsname{[0.819, 0.852]}
\expandafter\gdef\csname odunum@val@recorded_vs_generated.m4.marginal.seed.recorded\endcsname{0.799}
\expandafter\gdef\csname odunum@n@recorded_vs_generated.m4.marginal.seed.recorded\endcsname{202}
\expandafter\gdef\csname odunum@ci@recorded_vs_generated.m4.marginal.seed.recorded\endcsname{[0.749, 0.846]}
\expandafter\gdef\csname odunum@val@recorded_vs_generated.m4.stratum.cascade_asr.generated\endcsname{0.863}
\expandafter\gdef\csname odunum@n@recorded_vs_generated.m4.stratum.cascade_asr.generated\endcsname{424}
\expandafter\gdef\csname odunum@ci@recorded_vs_generated.m4.stratum.cascade_asr.generated\endcsname{[0.835, 0.888]}
\expandafter\gdef\csname odunum@val@recorded_vs_generated.m4.stratum.cascade_asr.recorded\endcsname{0.771}
\expandafter\gdef\csname odunum@n@recorded_vs_generated.m4.stratum.cascade_asr.recorded\endcsname{201}
\expandafter\gdef\csname odunum@ci@recorded_vs_generated.m4.stratum.cascade_asr.recorded\endcsname{[0.723, 0.815]}
\expandafter\gdef\csname odunum@val@recorded_vs_generated.m4.stratum.gemini.generated\endcsname{0.923}
\expandafter\gdef\csname odunum@n@recorded_vs_generated.m4.stratum.gemini.generated\endcsname{424}
\expandafter\gdef\csname odunum@ci@recorded_vs_generated.m4.stratum.gemini.generated\endcsname{[0.905, 0.939]}
\expandafter\gdef\csname odunum@val@recorded_vs_generated.m4.stratum.gemini.recorded\endcsname{0.883}
\expandafter\gdef\csname odunum@n@recorded_vs_generated.m4.stratum.gemini.recorded\endcsname{201}
\expandafter\gdef\csname odunum@ci@recorded_vs_generated.m4.stratum.gemini.recorded\endcsname{[0.854, 0.909]}
\expandafter\gdef\csname odunum@val@recorded_vs_generated.m4.stratum.gemini35_flash_lite.generated\endcsname{0.728}
\expandafter\gdef\csname odunum@n@recorded_vs_generated.m4.stratum.gemini35_flash_lite.generated\endcsname{424}
\expandafter\gdef\csname odunum@ci@recorded_vs_generated.m4.stratum.gemini35_flash_lite.generated\endcsname{[0.692, 0.762]}
\expandafter\gdef\csname odunum@val@recorded_vs_generated.m4.stratum.gemini35_flash_lite.recorded\endcsname{0.556}
\expandafter\gdef\csname odunum@n@recorded_vs_generated.m4.stratum.gemini35_flash_lite.recorded\endcsname{201}
\expandafter\gdef\csname odunum@ci@recorded_vs_generated.m4.stratum.gemini35_flash_lite.recorded\endcsname{[0.501, 0.610]}
\expandafter\gdef\csname odunum@val@recorded_vs_generated.m4.stratum.gemini37_flash.generated\endcsname{0.861}
\expandafter\gdef\csname odunum@n@recorded_vs_generated.m4.stratum.gemini37_flash.generated\endcsname{424}
\expandafter\gdef\csname odunum@ci@recorded_vs_generated.m4.stratum.gemini37_flash.generated\endcsname{[0.836, 0.885]}
\expandafter\gdef\csname odunum@val@recorded_vs_generated.m4.stratum.gemini37_flash.recorded\endcsname{0.820}
\expandafter\gdef\csname odunum@n@recorded_vs_generated.m4.stratum.gemini37_flash.recorded\endcsname{200}
\expandafter\gdef\csname odunum@ci@recorded_vs_generated.m4.stratum.gemini37_flash.recorded\endcsname{[0.782, 0.856]}
\expandafter\gdef\csname odunum@val@recorded_vs_generated.m4.stratum.ming.generated\endcsname{0.895}
\expandafter\gdef\csname odunum@n@recorded_vs_generated.m4.stratum.ming.generated\endcsname{424}
\expandafter\gdef\csname odunum@ci@recorded_vs_generated.m4.stratum.ming.generated\endcsname{[0.873, 0.917]}
\expandafter\gdef\csname odunum@val@recorded_vs_generated.m4.stratum.ming.recorded\endcsname{0.623}
\expandafter\gdef\csname odunum@n@recorded_vs_generated.m4.stratum.ming.recorded\endcsname{201}
\expandafter\gdef\csname odunum@ci@recorded_vs_generated.m4.stratum.ming.recorded\endcsname{[0.569, 0.675]}
\expandafter\gdef\csname odunum@val@recorded_vs_generated.m4.stratum.minicpm_o.generated\endcsname{0.625}
\expandafter\gdef\csname odunum@n@recorded_vs_generated.m4.stratum.minicpm_o.generated\endcsname{423}
\expandafter\gdef\csname odunum@ci@recorded_vs_generated.m4.stratum.minicpm_o.generated\endcsname{[0.586, 0.665]}
\expandafter\gdef\csname odunum@val@recorded_vs_generated.m4.stratum.minicpm_o.recorded\endcsname{0.429}
\expandafter\gdef\csname odunum@n@recorded_vs_generated.m4.stratum.minicpm_o.recorded\endcsname{201}
\expandafter\gdef\csname odunum@ci@recorded_vs_generated.m4.stratum.minicpm_o.recorded\endcsname{[0.369, 0.490]}
\expandafter\gdef\csname odunum@val@recorded_vs_generated.m4.stratum.nemotron.generated\endcsname{0.058}
\expandafter\gdef\csname odunum@n@recorded_vs_generated.m4.stratum.nemotron.generated\endcsname{424}
\expandafter\gdef\csname odunum@ci@recorded_vs_generated.m4.stratum.nemotron.generated\endcsname{[0.046, 0.069]}
\expandafter\gdef\csname odunum@val@recorded_vs_generated.m4.stratum.nemotron.recorded\endcsname{0.016}
\expandafter\gdef\csname odunum@n@recorded_vs_generated.m4.stratum.nemotron.recorded\endcsname{201}
\expandafter\gdef\csname odunum@ci@recorded_vs_generated.m4.stratum.nemotron.recorded\endcsname{[0.010, 0.023]}
\expandafter\gdef\csname odunum@val@recorded_vs_generated.m4.stratum.qwen25_omni.generated\endcsname{0.824}
\expandafter\gdef\csname odunum@n@recorded_vs_generated.m4.stratum.qwen25_omni.generated\endcsname{424}
\expandafter\gdef\csname odunum@ci@recorded_vs_generated.m4.stratum.qwen25_omni.generated\endcsname{[0.795, 0.853]}
\expandafter\gdef\csname odunum@val@recorded_vs_generated.m4.stratum.qwen25_omni.recorded\endcsname{0.462}
\expandafter\gdef\csname odunum@n@recorded_vs_generated.m4.stratum.qwen25_omni.recorded\endcsname{201}
\expandafter\gdef\csname odunum@ci@recorded_vs_generated.m4.stratum.qwen25_omni.recorded\endcsname{[0.403, 0.520]}
\expandafter\gdef\csname odunum@val@recorded_vs_generated.m4.stratum.qwen3_omni_instruct.generated\endcsname{0.904}
\expandafter\gdef\csname odunum@n@recorded_vs_generated.m4.stratum.qwen3_omni_instruct.generated\endcsname{424}
\expandafter\gdef\csname odunum@ci@recorded_vs_generated.m4.stratum.qwen3_omni_instruct.generated\endcsname{[0.882, 0.924]}
\expandafter\gdef\csname odunum@val@recorded_vs_generated.m4.stratum.qwen3_omni_instruct.recorded\endcsname{0.782}
\expandafter\gdef\csname odunum@n@recorded_vs_generated.m4.stratum.qwen3_omni_instruct.recorded\endcsname{201}
\expandafter\gdef\csname odunum@ci@recorded_vs_generated.m4.stratum.qwen3_omni_instruct.recorded\endcsname{[0.738, 0.824]}
\expandafter\gdef\csname odunum@val@recorded_vs_generated.m4.stratum.qwen3_omni_think.generated\endcsname{0.907}
\expandafter\gdef\csname odunum@n@recorded_vs_generated.m4.stratum.qwen3_omni_think.generated\endcsname{424}
\expandafter\gdef\csname odunum@ci@recorded_vs_generated.m4.stratum.qwen3_omni_think.generated\endcsname{[0.885, 0.928]}
\expandafter\gdef\csname odunum@val@recorded_vs_generated.m4.stratum.qwen3_omni_think.recorded\endcsname{0.713}
\expandafter\gdef\csname odunum@n@recorded_vs_generated.m4.stratum.qwen3_omni_think.recorded\endcsname{201}
\expandafter\gdef\csname odunum@ci@recorded_vs_generated.m4.stratum.qwen3_omni_think.recorded\endcsname{[0.663, 0.762]}
\expandafter\gdef\csname odunum@val@recorded_vs_generated.m4.stratum.qwen_plus.generated\endcsname{0.819}
\expandafter\gdef\csname odunum@n@recorded_vs_generated.m4.stratum.qwen_plus.generated\endcsname{424}
\expandafter\gdef\csname odunum@ci@recorded_vs_generated.m4.stratum.qwen_plus.generated\endcsname{[0.785, 0.851]}
\expandafter\gdef\csname odunum@val@recorded_vs_generated.m4.stratum.qwen_plus.recorded\endcsname{0.716}
\expandafter\gdef\csname odunum@n@recorded_vs_generated.m4.stratum.qwen_plus.recorded\endcsname{200}
\expandafter\gdef\csname odunum@ci@recorded_vs_generated.m4.stratum.qwen_plus.recorded\endcsname{[0.658, 0.770]}
\expandafter\gdef\csname odunum@val@recorded_vs_generated.m4.stratum.salmonn2_7b.generated\endcsname{0.000}
\expandafter\gdef\csname odunum@n@recorded_vs_generated.m4.stratum.salmonn2_7b.generated\endcsname{421}
\expandafter\gdef\csname odunum@ci@recorded_vs_generated.m4.stratum.salmonn2_7b.generated\endcsname{[0.000, 0.000]}
\expandafter\gdef\csname odunum@val@recorded_vs_generated.m4.stratum.salmonn2_7b.recorded\endcsname{0.002}
\expandafter\gdef\csname odunum@n@recorded_vs_generated.m4.stratum.salmonn2_7b.recorded\endcsname{198}
\expandafter\gdef\csname odunum@ci@recorded_vs_generated.m4.stratum.salmonn2_7b.recorded\endcsname{[0.000, 0.006]}
\expandafter\gdef\csname odunum@val@recorded_vs_generated.m4.stratum.seed.generated\endcsname{0.763}
\expandafter\gdef\csname odunum@n@recorded_vs_generated.m4.stratum.seed.generated\endcsname{424}
\expandafter\gdef\csname odunum@ci@recorded_vs_generated.m4.stratum.seed.generated\endcsname{[0.725, 0.799]}
\expandafter\gdef\csname odunum@val@recorded_vs_generated.m4.stratum.seed.recorded\endcsname{0.798}
\expandafter\gdef\csname odunum@n@recorded_vs_generated.m4.stratum.seed.recorded\endcsname{201}
\expandafter\gdef\csname odunum@ci@recorded_vs_generated.m4.stratum.seed.recorded\endcsname{[0.748, 0.844]}
\expandafter\gdef\csname odunum@val@recorded_vs_generated.m4_delta.marginal.cascade_asr\endcsname{-0.067}
\expandafter\gdef\csname odunum@n@recorded_vs_generated.m4_delta.marginal.cascade_asr\endcsname{202}
\expandafter\gdef\csname odunum@val@recorded_vs_generated.m4_delta.marginal.gemini\endcsname{-0.036}
\expandafter\gdef\csname odunum@n@recorded_vs_generated.m4_delta.marginal.gemini\endcsname{202}
\expandafter\gdef\csname odunum@val@recorded_vs_generated.m4_delta.marginal.gemini35_flash_lite\endcsname{-0.191}
\expandafter\gdef\csname odunum@n@recorded_vs_generated.m4_delta.marginal.gemini35_flash_lite\endcsname{202}
\expandafter\gdef\csname odunum@val@recorded_vs_generated.m4_delta.marginal.gemini37_flash\endcsname{-0.053}
\expandafter\gdef\csname odunum@n@recorded_vs_generated.m4_delta.marginal.gemini37_flash\endcsname{201}
\expandafter\gdef\csname odunum@val@recorded_vs_generated.m4_delta.marginal.ming\endcsname{-0.247}
\expandafter\gdef\csname odunum@n@recorded_vs_generated.m4_delta.marginal.ming\endcsname{202}
\expandafter\gdef\csname odunum@val@recorded_vs_generated.m4_delta.marginal.minicpm_o\endcsname{-0.200}
\expandafter\gdef\csname odunum@n@recorded_vs_generated.m4_delta.marginal.minicpm_o\endcsname{202}
\expandafter\gdef\csname odunum@val@recorded_vs_generated.m4_delta.marginal.nemotron\endcsname{-0.385}
\expandafter\gdef\csname odunum@n@recorded_vs_generated.m4_delta.marginal.nemotron\endcsname{202}
\expandafter\gdef\csname odunum@val@recorded_vs_generated.m4_delta.marginal.qwen25_omni\endcsname{-0.298}
\expandafter\gdef\csname odunum@n@recorded_vs_generated.m4_delta.marginal.qwen25_omni\endcsname{202}
\expandafter\gdef\csname odunum@val@recorded_vs_generated.m4_delta.marginal.qwen3_omni_instruct\endcsname{-0.116}
\expandafter\gdef\csname odunum@n@recorded_vs_generated.m4_delta.marginal.qwen3_omni_instruct\endcsname{202}
\expandafter\gdef\csname odunum@val@recorded_vs_generated.m4_delta.marginal.qwen3_omni_think\endcsname{-0.169}
\expandafter\gdef\csname odunum@n@recorded_vs_generated.m4_delta.marginal.qwen3_omni_think\endcsname{202}
\expandafter\gdef\csname odunum@val@recorded_vs_generated.m4_delta.marginal.qwen_plus\endcsname{-0.155}
\expandafter\gdef\csname odunum@n@recorded_vs_generated.m4_delta.marginal.qwen_plus\endcsname{201}
\expandafter\gdef\csname odunum@val@recorded_vs_generated.m4_delta.marginal.salmonn2_7b\endcsname{-0.194}
\expandafter\gdef\csname odunum@n@recorded_vs_generated.m4_delta.marginal.salmonn2_7b\endcsname{199}
\expandafter\gdef\csname odunum@val@recorded_vs_generated.m4_delta.marginal.seed\endcsname{-0.037}
\expandafter\gdef\csname odunum@n@recorded_vs_generated.m4_delta.marginal.seed\endcsname{202}
\expandafter\gdef\csname odunum@val@recorded_vs_generated.m4_delta.stratum.cascade_asr\endcsname{-0.092}
\expandafter\gdef\csname odunum@n@recorded_vs_generated.m4_delta.stratum.cascade_asr\endcsname{201}
\expandafter\gdef\csname odunum@val@recorded_vs_generated.m4_delta.stratum.gemini\endcsname{-0.041}
\expandafter\gdef\csname odunum@n@recorded_vs_generated.m4_delta.stratum.gemini\endcsname{201}
\expandafter\gdef\csname odunum@val@recorded_vs_generated.m4_delta.stratum.gemini35_flash_lite\endcsname{-0.172}
\expandafter\gdef\csname odunum@n@recorded_vs_generated.m4_delta.stratum.gemini35_flash_lite\endcsname{201}
\expandafter\gdef\csname odunum@val@recorded_vs_generated.m4_delta.stratum.gemini37_flash\endcsname{-0.041}
\expandafter\gdef\csname odunum@n@recorded_vs_generated.m4_delta.stratum.gemini37_flash\endcsname{200}
\expandafter\gdef\csname odunum@val@recorded_vs_generated.m4_delta.stratum.ming\endcsname{-0.273}
\expandafter\gdef\csname odunum@n@recorded_vs_generated.m4_delta.stratum.ming\endcsname{201}
\expandafter\gdef\csname odunum@val@recorded_vs_generated.m4_delta.stratum.minicpm_o\endcsname{-0.196}
\expandafter\gdef\csname odunum@n@recorded_vs_generated.m4_delta.stratum.minicpm_o\endcsname{201}
\expandafter\gdef\csname odunum@val@recorded_vs_generated.m4_delta.stratum.nemotron\endcsname{-0.041}
\expandafter\gdef\csname odunum@n@recorded_vs_generated.m4_delta.stratum.nemotron\endcsname{201}
\expandafter\gdef\csname odunum@val@recorded_vs_generated.m4_delta.stratum.qwen25_omni\endcsname{-0.363}
\expandafter\gdef\csname odunum@n@recorded_vs_generated.m4_delta.stratum.qwen25_omni\endcsname{201}
\expandafter\gdef\csname odunum@val@recorded_vs_generated.m4_delta.stratum.qwen3_omni_instruct\endcsname{-0.122}
\expandafter\gdef\csname odunum@n@recorded_vs_generated.m4_delta.stratum.qwen3_omni_instruct\endcsname{201}
\expandafter\gdef\csname odunum@val@recorded_vs_generated.m4_delta.stratum.qwen3_omni_think\endcsname{-0.195}
\expandafter\gdef\csname odunum@n@recorded_vs_generated.m4_delta.stratum.qwen3_omni_think\endcsname{201}
\expandafter\gdef\csname odunum@val@recorded_vs_generated.m4_delta.stratum.qwen_plus\endcsname{-0.103}
\expandafter\gdef\csname odunum@n@recorded_vs_generated.m4_delta.stratum.qwen_plus\endcsname{200}
\expandafter\gdef\csname odunum@val@recorded_vs_generated.m4_delta.stratum.salmonn2_7b\endcsname{0.002}
\expandafter\gdef\csname odunum@n@recorded_vs_generated.m4_delta.stratum.salmonn2_7b\endcsname{198}
\expandafter\gdef\csname odunum@val@recorded_vs_generated.m4_delta.stratum.seed\endcsname{0.035}
\expandafter\gdef\csname odunum@n@recorded_vs_generated.m4_delta.stratum.seed\endcsname{201}
\expandafter\gdef\csname odunum@val@recorded_vs_generated.m5.marginal.cascade_asr.generated\endcsname{0.138}
\expandafter\gdef\csname odunum@n@recorded_vs_generated.m5.marginal.cascade_asr.generated\endcsname{1\,429}
\expandafter\gdef\csname odunum@ci@recorded_vs_generated.m5.marginal.cascade_asr.generated\endcsname{[0.126, 0.149]}
\expandafter\gdef\csname odunum@val@recorded_vs_generated.m5.marginal.cascade_asr.recorded\endcsname{0.028}
\expandafter\gdef\csname odunum@n@recorded_vs_generated.m5.marginal.cascade_asr.recorded\endcsname{202}
\expandafter\gdef\csname odunum@ci@recorded_vs_generated.m5.marginal.cascade_asr.recorded\endcsname{[0.015, 0.043]}
\expandafter\gdef\csname odunum@val@recorded_vs_generated.m5.marginal.gemini.generated\endcsname{0.961}
\expandafter\gdef\csname odunum@n@recorded_vs_generated.m5.marginal.gemini.generated\endcsname{1\,429}
\expandafter\gdef\csname odunum@ci@recorded_vs_generated.m5.marginal.gemini.generated\endcsname{[0.953, 0.969]}
\expandafter\gdef\csname odunum@val@recorded_vs_generated.m5.marginal.gemini.recorded\endcsname{0.984}
\expandafter\gdef\csname odunum@n@recorded_vs_generated.m5.marginal.gemini.recorded\endcsname{202}
\expandafter\gdef\csname odunum@ci@recorded_vs_generated.m5.marginal.gemini.recorded\endcsname{[0.969, 0.995]}
\expandafter\gdef\csname odunum@val@recorded_vs_generated.m5.marginal.gemini35_flash_lite.generated\endcsname{0.808}
\expandafter\gdef\csname odunum@n@recorded_vs_generated.m5.marginal.gemini35_flash_lite.generated\endcsname{1\,429}
\expandafter\gdef\csname odunum@ci@recorded_vs_generated.m5.marginal.gemini35_flash_lite.generated\endcsname{[0.790, 0.827]}
\expandafter\gdef\csname odunum@val@recorded_vs_generated.m5.marginal.gemini35_flash_lite.recorded\endcsname{0.772}
\expandafter\gdef\csname odunum@n@recorded_vs_generated.m5.marginal.gemini35_flash_lite.recorded\endcsname{202}
\expandafter\gdef\csname odunum@ci@recorded_vs_generated.m5.marginal.gemini35_flash_lite.recorded\endcsname{[0.716, 0.827]}
\expandafter\gdef\csname odunum@val@recorded_vs_generated.m5.marginal.gemini37_flash.generated\endcsname{0.903}
\expandafter\gdef\csname odunum@n@recorded_vs_generated.m5.marginal.gemini37_flash.generated\endcsname{1\,426}
\expandafter\gdef\csname odunum@ci@recorded_vs_generated.m5.marginal.gemini37_flash.generated\endcsname{[0.890, 0.915]}
\expandafter\gdef\csname odunum@val@recorded_vs_generated.m5.marginal.gemini37_flash.recorded\endcsname{0.907}
\expandafter\gdef\csname odunum@n@recorded_vs_generated.m5.marginal.gemini37_flash.recorded\endcsname{201}
\expandafter\gdef\csname odunum@ci@recorded_vs_generated.m5.marginal.gemini37_flash.recorded\endcsname{[0.872, 0.939]}
\expandafter\gdef\csname odunum@val@recorded_vs_generated.m5.marginal.ming.generated\endcsname{0.769}
\expandafter\gdef\csname odunum@n@recorded_vs_generated.m5.marginal.ming.generated\endcsname{1\,429}
\expandafter\gdef\csname odunum@ci@recorded_vs_generated.m5.marginal.ming.generated\endcsname{[0.748, 0.789]}
\expandafter\gdef\csname odunum@val@recorded_vs_generated.m5.marginal.ming.recorded\endcsname{0.850}
\expandafter\gdef\csname odunum@n@recorded_vs_generated.m5.marginal.ming.recorded\endcsname{202}
\expandafter\gdef\csname odunum@ci@recorded_vs_generated.m5.marginal.ming.recorded\endcsname{[0.804, 0.891]}
\expandafter\gdef\csname odunum@val@recorded_vs_generated.m5.marginal.minicpm_o.generated\endcsname{0.736}
\expandafter\gdef\csname odunum@n@recorded_vs_generated.m5.marginal.minicpm_o.generated\endcsname{1\,427}
\expandafter\gdef\csname odunum@ci@recorded_vs_generated.m5.marginal.minicpm_o.generated\endcsname{[0.714, 0.757]}
\expandafter\gdef\csname odunum@val@recorded_vs_generated.m5.marginal.minicpm_o.recorded\endcsname{0.512}
\expandafter\gdef\csname odunum@n@recorded_vs_generated.m5.marginal.minicpm_o.recorded\endcsname{202}
\expandafter\gdef\csname odunum@ci@recorded_vs_generated.m5.marginal.minicpm_o.recorded\endcsname{[0.446, 0.578]}
\expandafter\gdef\csname odunum@val@recorded_vs_generated.m5.marginal.nemotron.generated\endcsname{0.696}
\expandafter\gdef\csname odunum@n@recorded_vs_generated.m5.marginal.nemotron.generated\endcsname{1\,429}
\expandafter\gdef\csname odunum@ci@recorded_vs_generated.m5.marginal.nemotron.generated\endcsname{[0.674, 0.718]}
\expandafter\gdef\csname odunum@val@recorded_vs_generated.m5.marginal.nemotron.recorded\endcsname{0.541}
\expandafter\gdef\csname odunum@n@recorded_vs_generated.m5.marginal.nemotron.recorded\endcsname{202}
\expandafter\gdef\csname odunum@ci@recorded_vs_generated.m5.marginal.nemotron.recorded\endcsname{[0.475, 0.606]}
\expandafter\gdef\csname odunum@val@recorded_vs_generated.m5.marginal.qwen25_omni.generated\endcsname{0.727}
\expandafter\gdef\csname odunum@n@recorded_vs_generated.m5.marginal.qwen25_omni.generated\endcsname{1\,429}
\expandafter\gdef\csname odunum@ci@recorded_vs_generated.m5.marginal.qwen25_omni.generated\endcsname{[0.705, 0.748]}
\expandafter\gdef\csname odunum@val@recorded_vs_generated.m5.marginal.qwen25_omni.recorded\endcsname{0.678}
\expandafter\gdef\csname odunum@n@recorded_vs_generated.m5.marginal.qwen25_omni.recorded\endcsname{202}
\expandafter\gdef\csname odunum@ci@recorded_vs_generated.m5.marginal.qwen25_omni.recorded\endcsname{[0.616, 0.739]}
\expandafter\gdef\csname odunum@val@recorded_vs_generated.m5.marginal.qwen3_omni_instruct.generated\endcsname{0.940}
\expandafter\gdef\csname odunum@n@recorded_vs_generated.m5.marginal.qwen3_omni_instruct.generated\endcsname{1\,429}
\expandafter\gdef\csname odunum@ci@recorded_vs_generated.m5.marginal.qwen3_omni_instruct.generated\endcsname{[0.930, 0.950]}
\expandafter\gdef\csname odunum@val@recorded_vs_generated.m5.marginal.qwen3_omni_instruct.recorded\endcsname{0.908}
\expandafter\gdef\csname odunum@n@recorded_vs_generated.m5.marginal.qwen3_omni_instruct.recorded\endcsname{202}
\expandafter\gdef\csname odunum@ci@recorded_vs_generated.m5.marginal.qwen3_omni_instruct.recorded\endcsname{[0.870, 0.942]}
\expandafter\gdef\csname odunum@val@recorded_vs_generated.m5.marginal.qwen3_omni_think.generated\endcsname{0.933}
\expandafter\gdef\csname odunum@n@recorded_vs_generated.m5.marginal.qwen3_omni_think.generated\endcsname{1\,429}
\expandafter\gdef\csname odunum@ci@recorded_vs_generated.m5.marginal.qwen3_omni_think.generated\endcsname{[0.922, 0.943]}
\expandafter\gdef\csname odunum@val@recorded_vs_generated.m5.marginal.qwen3_omni_think.recorded\endcsname{0.904}
\expandafter\gdef\csname odunum@n@recorded_vs_generated.m5.marginal.qwen3_omni_think.recorded\endcsname{202}
\expandafter\gdef\csname odunum@ci@recorded_vs_generated.m5.marginal.qwen3_omni_think.recorded\endcsname{[0.868, 0.937]}
\expandafter\gdef\csname odunum@val@recorded_vs_generated.m5.marginal.qwen_plus.generated\endcsname{0.944}
\expandafter\gdef\csname odunum@n@recorded_vs_generated.m5.marginal.qwen_plus.generated\endcsname{1\,429}
\expandafter\gdef\csname odunum@ci@recorded_vs_generated.m5.marginal.qwen_plus.generated\endcsname{[0.934, 0.954]}
\expandafter\gdef\csname odunum@val@recorded_vs_generated.m5.marginal.qwen_plus.recorded\endcsname{0.934}
\expandafter\gdef\csname odunum@n@recorded_vs_generated.m5.marginal.qwen_plus.recorded\endcsname{201}
\expandafter\gdef\csname odunum@ci@recorded_vs_generated.m5.marginal.qwen_plus.recorded\endcsname{[0.900, 0.962]}
\expandafter\gdef\csname odunum@val@recorded_vs_generated.m5.marginal.salmonn2_7b.generated\endcsname{0.329}
\expandafter\gdef\csname odunum@n@recorded_vs_generated.m5.marginal.salmonn2_7b.generated\endcsname{1\,415}
\expandafter\gdef\csname odunum@ci@recorded_vs_generated.m5.marginal.salmonn2_7b.generated\endcsname{[0.307, 0.352]}
\expandafter\gdef\csname odunum@val@recorded_vs_generated.m5.marginal.salmonn2_7b.recorded\endcsname{0.107}
\expandafter\gdef\csname odunum@n@recorded_vs_generated.m5.marginal.salmonn2_7b.recorded\endcsname{199}
\expandafter\gdef\csname odunum@ci@recorded_vs_generated.m5.marginal.salmonn2_7b.recorded\endcsname{[0.069, 0.149]}
\expandafter\gdef\csname odunum@val@recorded_vs_generated.m5.marginal.seed.generated\endcsname{0.942}
\expandafter\gdef\csname odunum@n@recorded_vs_generated.m5.marginal.seed.generated\endcsname{1\,429}
\expandafter\gdef\csname odunum@ci@recorded_vs_generated.m5.marginal.seed.generated\endcsname{[0.932, 0.951]}
\expandafter\gdef\csname odunum@val@recorded_vs_generated.m5.marginal.seed.recorded\endcsname{0.949}
\expandafter\gdef\csname odunum@n@recorded_vs_generated.m5.marginal.seed.recorded\endcsname{202}
\expandafter\gdef\csname odunum@ci@recorded_vs_generated.m5.marginal.seed.recorded\endcsname{[0.921, 0.974]}
\expandafter\gdef\csname odunum@val@recorded_vs_generated.m5.stratum.cascade_asr.generated\endcsname{0.028}
\expandafter\gdef\csname odunum@n@recorded_vs_generated.m5.stratum.cascade_asr.generated\endcsname{424}
\expandafter\gdef\csname odunum@ci@recorded_vs_generated.m5.stratum.cascade_asr.generated\endcsname{[0.019, 0.037]}
\expandafter\gdef\csname odunum@val@recorded_vs_generated.m5.stratum.cascade_asr.recorded\endcsname{0.028}
\expandafter\gdef\csname odunum@n@recorded_vs_generated.m5.stratum.cascade_asr.recorded\endcsname{201}
\expandafter\gdef\csname odunum@ci@recorded_vs_generated.m5.stratum.cascade_asr.recorded\endcsname{[0.015, 0.043]}
\expandafter\gdef\csname odunum@val@recorded_vs_generated.m5.stratum.gemini.generated\endcsname{0.953}
\expandafter\gdef\csname odunum@n@recorded_vs_generated.m5.stratum.gemini.generated\endcsname{424}
\expandafter\gdef\csname odunum@ci@recorded_vs_generated.m5.stratum.gemini.generated\endcsname{[0.938, 0.967]}
\expandafter\gdef\csname odunum@val@recorded_vs_generated.m5.stratum.gemini.recorded\endcsname{0.983}
\expandafter\gdef\csname odunum@n@recorded_vs_generated.m5.stratum.gemini.recorded\endcsname{201}
\expandafter\gdef\csname odunum@ci@recorded_vs_generated.m5.stratum.gemini.recorded\endcsname{[0.968, 0.995]}
\expandafter\gdef\csname odunum@val@recorded_vs_generated.m5.stratum.gemini35_flash_lite.generated\endcsname{0.782}
\expandafter\gdef\csname odunum@n@recorded_vs_generated.m5.stratum.gemini35_flash_lite.generated\endcsname{424}
\expandafter\gdef\csname odunum@ci@recorded_vs_generated.m5.stratum.gemini35_flash_lite.generated\endcsname{[0.746, 0.816]}
\expandafter\gdef\csname odunum@val@recorded_vs_generated.m5.stratum.gemini35_flash_lite.recorded\endcsname{0.771}
\expandafter\gdef\csname odunum@n@recorded_vs_generated.m5.stratum.gemini35_flash_lite.recorded\endcsname{201}
\expandafter\gdef\csname odunum@ci@recorded_vs_generated.m5.stratum.gemini35_flash_lite.recorded\endcsname{[0.713, 0.826]}
\expandafter\gdef\csname odunum@val@recorded_vs_generated.m5.stratum.gemini37_flash.generated\endcsname{0.870}
\expandafter\gdef\csname odunum@n@recorded_vs_generated.m5.stratum.gemini37_flash.generated\endcsname{424}
\expandafter\gdef\csname odunum@ci@recorded_vs_generated.m5.stratum.gemini37_flash.generated\endcsname{[0.846, 0.893]}
\expandafter\gdef\csname odunum@val@recorded_vs_generated.m5.stratum.gemini37_flash.recorded\endcsname{0.907}
\expandafter\gdef\csname odunum@n@recorded_vs_generated.m5.stratum.gemini37_flash.recorded\endcsname{200}
\expandafter\gdef\csname odunum@ci@recorded_vs_generated.m5.stratum.gemini37_flash.recorded\endcsname{[0.870, 0.940]}
\expandafter\gdef\csname odunum@val@recorded_vs_generated.m5.stratum.ming.generated\endcsname{0.846}
\expandafter\gdef\csname odunum@n@recorded_vs_generated.m5.stratum.ming.generated\endcsname{424}
\expandafter\gdef\csname odunum@ci@recorded_vs_generated.m5.stratum.ming.generated\endcsname{[0.814, 0.875]}
\expandafter\gdef\csname odunum@val@recorded_vs_generated.m5.stratum.ming.recorded\endcsname{0.851}
\expandafter\gdef\csname odunum@n@recorded_vs_generated.m5.stratum.ming.recorded\endcsname{201}
\expandafter\gdef\csname odunum@ci@recorded_vs_generated.m5.stratum.ming.recorded\endcsname{[0.806, 0.892]}
\expandafter\gdef\csname odunum@val@recorded_vs_generated.m5.stratum.minicpm_o.generated\endcsname{0.705}
\expandafter\gdef\csname odunum@n@recorded_vs_generated.m5.stratum.minicpm_o.generated\endcsname{423}
\expandafter\gdef\csname odunum@ci@recorded_vs_generated.m5.stratum.minicpm_o.generated\endcsname{[0.665, 0.745]}
\expandafter\gdef\csname odunum@val@recorded_vs_generated.m5.stratum.minicpm_o.recorded\endcsname{0.511}
\expandafter\gdef\csname odunum@n@recorded_vs_generated.m5.stratum.minicpm_o.recorded\endcsname{201}
\expandafter\gdef\csname odunum@ci@recorded_vs_generated.m5.stratum.minicpm_o.recorded\endcsname{[0.444, 0.579]}
\expandafter\gdef\csname odunum@val@recorded_vs_generated.m5.stratum.nemotron.generated\endcsname{0.555}
\expandafter\gdef\csname odunum@n@recorded_vs_generated.m5.stratum.nemotron.generated\endcsname{424}
\expandafter\gdef\csname odunum@ci@recorded_vs_generated.m5.stratum.nemotron.generated\endcsname{[0.509, 0.600]}
\expandafter\gdef\csname odunum@val@recorded_vs_generated.m5.stratum.nemotron.recorded\endcsname{0.541}
\expandafter\gdef\csname odunum@n@recorded_vs_generated.m5.stratum.nemotron.recorded\endcsname{201}
\expandafter\gdef\csname odunum@ci@recorded_vs_generated.m5.stratum.nemotron.recorded\endcsname{[0.474, 0.609]}
\expandafter\gdef\csname odunum@val@recorded_vs_generated.m5.stratum.qwen25_omni.generated\endcsname{0.821}
\expandafter\gdef\csname odunum@n@recorded_vs_generated.m5.stratum.qwen25_omni.generated\endcsname{424}
\expandafter\gdef\csname odunum@ci@recorded_vs_generated.m5.stratum.qwen25_omni.generated\endcsname{[0.789, 0.853]}
\expandafter\gdef\csname odunum@val@recorded_vs_generated.m5.stratum.qwen25_omni.recorded\endcsname{0.678}
\expandafter\gdef\csname odunum@n@recorded_vs_generated.m5.stratum.qwen25_omni.recorded\endcsname{201}
\expandafter\gdef\csname odunum@ci@recorded_vs_generated.m5.stratum.qwen25_omni.recorded\endcsname{[0.614, 0.740]}
\expandafter\gdef\csname odunum@val@recorded_vs_generated.m5.stratum.qwen3_omni_instruct.generated\endcsname{0.943}
\expandafter\gdef\csname odunum@n@recorded_vs_generated.m5.stratum.qwen3_omni_instruct.generated\endcsname{424}
\expandafter\gdef\csname odunum@ci@recorded_vs_generated.m5.stratum.qwen3_omni_instruct.generated\endcsname{[0.926, 0.960]}
\expandafter\gdef\csname odunum@val@recorded_vs_generated.m5.stratum.qwen3_omni_instruct.recorded\endcsname{0.909}
\expandafter\gdef\csname odunum@n@recorded_vs_generated.m5.stratum.qwen3_omni_instruct.recorded\endcsname{201}
\expandafter\gdef\csname odunum@ci@recorded_vs_generated.m5.stratum.qwen3_omni_instruct.recorded\endcsname{[0.871, 0.944]}
\expandafter\gdef\csname odunum@val@recorded_vs_generated.m5.stratum.qwen3_omni_think.generated\endcsname{0.928}
\expandafter\gdef\csname odunum@n@recorded_vs_generated.m5.stratum.qwen3_omni_think.generated\endcsname{424}
\expandafter\gdef\csname odunum@ci@recorded_vs_generated.m5.stratum.qwen3_omni_think.generated\endcsname{[0.910, 0.944]}
\expandafter\gdef\csname odunum@val@recorded_vs_generated.m5.stratum.qwen3_omni_think.recorded\endcsname{0.905}
\expandafter\gdef\csname odunum@n@recorded_vs_generated.m5.stratum.qwen3_omni_think.recorded\endcsname{201}
\expandafter\gdef\csname odunum@ci@recorded_vs_generated.m5.stratum.qwen3_omni_think.recorded\endcsname{[0.869, 0.939]}
\expandafter\gdef\csname odunum@val@recorded_vs_generated.m5.stratum.qwen_plus.generated\endcsname{0.935}
\expandafter\gdef\csname odunum@n@recorded_vs_generated.m5.stratum.qwen_plus.generated\endcsname{424}
\expandafter\gdef\csname odunum@ci@recorded_vs_generated.m5.stratum.qwen_plus.generated\endcsname{[0.915, 0.953]}
\expandafter\gdef\csname odunum@val@recorded_vs_generated.m5.stratum.qwen_plus.recorded\endcsname{0.933}
\expandafter\gdef\csname odunum@n@recorded_vs_generated.m5.stratum.qwen_plus.recorded\endcsname{200}
\expandafter\gdef\csname odunum@ci@recorded_vs_generated.m5.stratum.qwen_plus.recorded\endcsname{[0.900, 0.963]}
\expandafter\gdef\csname odunum@val@recorded_vs_generated.m5.stratum.salmonn2_7b.generated\endcsname{0.094}
\expandafter\gdef\csname odunum@n@recorded_vs_generated.m5.stratum.salmonn2_7b.generated\endcsname{421}
\expandafter\gdef\csname odunum@ci@recorded_vs_generated.m5.stratum.salmonn2_7b.generated\endcsname{[0.069, 0.120]}
\expandafter\gdef\csname odunum@val@recorded_vs_generated.m5.stratum.salmonn2_7b.recorded\endcsname{0.108}
\expandafter\gdef\csname odunum@n@recorded_vs_generated.m5.stratum.salmonn2_7b.recorded\endcsname{198}
\expandafter\gdef\csname odunum@ci@recorded_vs_generated.m5.stratum.salmonn2_7b.recorded\endcsname{[0.067, 0.150]}
\expandafter\gdef\csname odunum@val@recorded_vs_generated.m5.stratum.seed.generated\endcsname{0.915}
\expandafter\gdef\csname odunum@n@recorded_vs_generated.m5.stratum.seed.generated\endcsname{424}
\expandafter\gdef\csname odunum@ci@recorded_vs_generated.m5.stratum.seed.generated\endcsname{[0.893, 0.936]}
\expandafter\gdef\csname odunum@val@recorded_vs_generated.m5.stratum.seed.recorded\endcsname{0.949}
\expandafter\gdef\csname odunum@n@recorded_vs_generated.m5.stratum.seed.recorded\endcsname{201}
\expandafter\gdef\csname odunum@ci@recorded_vs_generated.m5.stratum.seed.recorded\endcsname{[0.920, 0.973]}
\expandafter\gdef\csname odunum@val@recorded_vs_generated.m5_delta.marginal.cascade_asr\endcsname{-0.110}
\expandafter\gdef\csname odunum@n@recorded_vs_generated.m5_delta.marginal.cascade_asr\endcsname{202}
\expandafter\gdef\csname odunum@val@recorded_vs_generated.m5_delta.marginal.gemini\endcsname{0.023}
\expandafter\gdef\csname odunum@n@recorded_vs_generated.m5_delta.marginal.gemini\endcsname{202}
\expandafter\gdef\csname odunum@val@recorded_vs_generated.m5_delta.marginal.gemini35_flash_lite\endcsname{-0.036}
\expandafter\gdef\csname odunum@n@recorded_vs_generated.m5_delta.marginal.gemini35_flash_lite\endcsname{202}
\expandafter\gdef\csname odunum@val@recorded_vs_generated.m5_delta.marginal.gemini37_flash\endcsname{0.004}
\expandafter\gdef\csname odunum@n@recorded_vs_generated.m5_delta.marginal.gemini37_flash\endcsname{201}
\expandafter\gdef\csname odunum@val@recorded_vs_generated.m5_delta.marginal.ming\endcsname{0.081}
\expandafter\gdef\csname odunum@n@recorded_vs_generated.m5_delta.marginal.ming\endcsname{202}
\expandafter\gdef\csname odunum@val@recorded_vs_generated.m5_delta.marginal.minicpm_o\endcsname{-0.224}
\expandafter\gdef\csname odunum@n@recorded_vs_generated.m5_delta.marginal.minicpm_o\endcsname{202}
\expandafter\gdef\csname odunum@val@recorded_vs_generated.m5_delta.marginal.nemotron\endcsname{-0.155}
\expandafter\gdef\csname odunum@n@recorded_vs_generated.m5_delta.marginal.nemotron\endcsname{202}
\expandafter\gdef\csname odunum@val@recorded_vs_generated.m5_delta.marginal.qwen25_omni\endcsname{-0.048}
\expandafter\gdef\csname odunum@n@recorded_vs_generated.m5_delta.marginal.qwen25_omni\endcsname{202}
\expandafter\gdef\csname odunum@val@recorded_vs_generated.m5_delta.marginal.qwen3_omni_instruct\endcsname{-0.033}
\expandafter\gdef\csname odunum@n@recorded_vs_generated.m5_delta.marginal.qwen3_omni_instruct\endcsname{202}
\expandafter\gdef\csname odunum@val@recorded_vs_generated.m5_delta.marginal.qwen3_omni_think\endcsname{-0.029}
\expandafter\gdef\csname odunum@n@recorded_vs_generated.m5_delta.marginal.qwen3_omni_think\endcsname{202}
\expandafter\gdef\csname odunum@val@recorded_vs_generated.m5_delta.marginal.qwen_plus\endcsname{-0.011}
\expandafter\gdef\csname odunum@n@recorded_vs_generated.m5_delta.marginal.qwen_plus\endcsname{201}
\expandafter\gdef\csname odunum@val@recorded_vs_generated.m5_delta.marginal.salmonn2_7b\endcsname{-0.222}
\expandafter\gdef\csname odunum@n@recorded_vs_generated.m5_delta.marginal.salmonn2_7b\endcsname{199}
\expandafter\gdef\csname odunum@val@recorded_vs_generated.m5_delta.marginal.seed\endcsname{0.007}
\expandafter\gdef\csname odunum@n@recorded_vs_generated.m5_delta.marginal.seed\endcsname{202}
\expandafter\gdef\csname odunum@val@recorded_vs_generated.m5_delta.stratum.cascade_asr\endcsname{4.0\ensuremath{\times 10^{-4}}}
\expandafter\gdef\csname odunum@n@recorded_vs_generated.m5_delta.stratum.cascade_asr\endcsname{201}
\expandafter\gdef\csname odunum@val@recorded_vs_generated.m5_delta.stratum.gemini\endcsname{0.031}
\expandafter\gdef\csname odunum@n@recorded_vs_generated.m5_delta.stratum.gemini\endcsname{201}
\expandafter\gdef\csname odunum@val@recorded_vs_generated.m5_delta.stratum.gemini35_flash_lite\endcsname{-0.011}
\expandafter\gdef\csname odunum@n@recorded_vs_generated.m5_delta.stratum.gemini35_flash_lite\endcsname{201}
\expandafter\gdef\csname odunum@val@recorded_vs_generated.m5_delta.stratum.gemini37_flash\endcsname{0.037}
\expandafter\gdef\csname odunum@n@recorded_vs_generated.m5_delta.stratum.gemini37_flash\endcsname{200}
\expandafter\gdef\csname odunum@val@recorded_vs_generated.m5_delta.stratum.ming\endcsname{0.005}
\expandafter\gdef\csname odunum@n@recorded_vs_generated.m5_delta.stratum.ming\endcsname{201}
\expandafter\gdef\csname odunum@val@recorded_vs_generated.m5_delta.stratum.minicpm_o\endcsname{-0.195}
\expandafter\gdef\csname odunum@n@recorded_vs_generated.m5_delta.stratum.minicpm_o\endcsname{201}
\expandafter\gdef\csname odunum@val@recorded_vs_generated.m5_delta.stratum.nemotron\endcsname{-0.014}
\expandafter\gdef\csname odunum@n@recorded_vs_generated.m5_delta.stratum.nemotron\endcsname{201}
\expandafter\gdef\csname odunum@val@recorded_vs_generated.m5_delta.stratum.qwen25_omni\endcsname{-0.143}
\expandafter\gdef\csname odunum@n@recorded_vs_generated.m5_delta.stratum.qwen25_omni\endcsname{201}
\expandafter\gdef\csname odunum@val@recorded_vs_generated.m5_delta.stratum.qwen3_omni_instruct\endcsname{-0.035}
\expandafter\gdef\csname odunum@n@recorded_vs_generated.m5_delta.stratum.qwen3_omni_instruct\endcsname{201}
\expandafter\gdef\csname odunum@val@recorded_vs_generated.m5_delta.stratum.qwen3_omni_think\endcsname{-0.022}
\expandafter\gdef\csname odunum@n@recorded_vs_generated.m5_delta.stratum.qwen3_omni_think\endcsname{201}
\expandafter\gdef\csname odunum@val@recorded_vs_generated.m5_delta.stratum.qwen_plus\endcsname{-0.002}
\expandafter\gdef\csname odunum@n@recorded_vs_generated.m5_delta.stratum.qwen_plus\endcsname{200}
\expandafter\gdef\csname odunum@val@recorded_vs_generated.m5_delta.stratum.salmonn2_7b\endcsname{0.013}
\expandafter\gdef\csname odunum@n@recorded_vs_generated.m5_delta.stratum.salmonn2_7b\endcsname{198}
\expandafter\gdef\csname odunum@val@recorded_vs_generated.m5_delta.stratum.seed\endcsname{0.034}
\expandafter\gdef\csname odunum@n@recorded_vs_generated.m5_delta.stratum.seed\endcsname{201}
\expandafter\gdef\csname odunum@val@recorded_vs_generated.pop.generated.n\endcsname{1\,801}
\expandafter\gdef\csname odunum@n@recorded_vs_generated.pop.generated.n\endcsname{2\,078}
\expandafter\gdef\csname odunum@val@recorded_vs_generated.pop.generated.stratum_n\endcsname{529}
\expandafter\gdef\csname odunum@n@recorded_vs_generated.pop.generated.stratum_n\endcsname{2\,078}
\expandafter\gdef\csname odunum@val@recorded_vs_generated.pop.generated.stratum_n_pos\endcsname{424}
\expandafter\gdef\csname odunum@n@recorded_vs_generated.pop.generated.stratum_n_pos\endcsname{529}
\expandafter\gdef\csname odunum@val@recorded_vs_generated.pop.recorded.n\endcsname{277}
\expandafter\gdef\csname odunum@n@recorded_vs_generated.pop.recorded.n\endcsname{2\,078}
\expandafter\gdef\csname odunum@val@recorded_vs_generated.pop.recorded.n_en\endcsname{1}
\expandafter\gdef\csname odunum@n@recorded_vs_generated.pop.recorded.n_en\endcsname{277}
\expandafter\gdef\csname odunum@val@recorded_vs_generated.pop.recorded.stratum_n\endcsname{276}
\expandafter\gdef\csname odunum@n@recorded_vs_generated.pop.recorded.stratum_n\endcsname{2\,078}
\expandafter\gdef\csname odunum@val@recorded_vs_generated.pop.recorded.stratum_n_pos\endcsname{201}
\expandafter\gdef\csname odunum@n@recorded_vs_generated.pop.recorded.stratum_n_pos\endcsname{276}
\expandafter\gdef\csname odunum@val@recorded_vs_generated.rank.marginal.avg.kendall_tau\endcsname{0.872}
\expandafter\gdef\csname odunum@n@recorded_vs_generated.rank.marginal.avg.kendall_tau\endcsname{13}
\expandafter\gdef\csname odunum@val@recorded_vs_generated.rank.marginal.avg.spearman\endcsname{0.962}
\expandafter\gdef\csname odunum@n@recorded_vs_generated.rank.marginal.avg.spearman\endcsname{13}
\expandafter\gdef\csname odunum@val@recorded_vs_generated.rank.marginal.m2.kendall_tau\endcsname{0.897}
\expandafter\gdef\csname odunum@n@recorded_vs_generated.rank.marginal.m2.kendall_tau\endcsname{13}
\expandafter\gdef\csname odunum@val@recorded_vs_generated.rank.marginal.m2.spearman\endcsname{0.973}
\expandafter\gdef\csname odunum@n@recorded_vs_generated.rank.marginal.m2.spearman\endcsname{13}
\expandafter\gdef\csname odunum@val@recorded_vs_generated.rank.stratum.avg.kendall_tau\endcsname{0.897}
\expandafter\gdef\csname odunum@n@recorded_vs_generated.rank.stratum.avg.kendall_tau\endcsname{13}
\expandafter\gdef\csname odunum@val@recorded_vs_generated.rank.stratum.avg.spearman\endcsname{0.973}
\expandafter\gdef\csname odunum@n@recorded_vs_generated.rank.stratum.avg.spearman\endcsname{13}
\expandafter\gdef\csname odunum@val@recorded_vs_generated.rank.stratum.m2.kendall_tau\endcsname{0.949}
\expandafter\gdef\csname odunum@n@recorded_vs_generated.rank.stratum.m2.kendall_tau\endcsname{13}
\expandafter\gdef\csname odunum@val@recorded_vs_generated.rank.stratum.m2.spearman\endcsname{0.984}
\expandafter\gdef\csname odunum@n@recorded_vs_generated.rank.stratum.m2.spearman\endcsname{13}
\expandafter\gdef\csname odunum@val@release_stats.main.axis1_present\endcsname{4}
\expandafter\gdef\csname odunum@n@release_stats.main.axis1_present\endcsname{1\,631}
\expandafter\gdef\csname odunum@val@release_stats.main.axis1_total\endcsname{4}
\expandafter\gdef\csname odunum@n@release_stats.main.axis1_total\endcsname{4}
\expandafter\gdef\csname odunum@val@release_stats.main.axis2_present\endcsname{5}
\expandafter\gdef\csname odunum@n@release_stats.main.axis2_present\endcsname{1\,631}
\expandafter\gdef\csname odunum@val@release_stats.main.axis2_total\endcsname{5}
\expandafter\gdef\csname odunum@n@release_stats.main.axis2_total\endcsname{5}
\expandafter\gdef\csname odunum@val@release_stats.main.axis3_present\endcsname{6}
\expandafter\gdef\csname odunum@n@release_stats.main.axis3_present\endcsname{1\,631}
\expandafter\gdef\csname odunum@val@release_stats.main.axis3_total\endcsname{6}
\expandafter\gdef\csname odunum@n@release_stats.main.axis3_total\endcsname{6}
\expandafter\gdef\csname odunum@val@release_stats.main.axis4_present\endcsname{5}
\expandafter\gdef\csname odunum@n@release_stats.main.axis4_present\endcsname{1\,631}
\expandafter\gdef\csname odunum@val@release_stats.main.axis4_total\endcsname{5}
\expandafter\gdef\csname odunum@n@release_stats.main.axis4_total\endcsname{5}
\expandafter\gdef\csname odunum@val@release_stats.main.axis5_present\endcsname{10}
\expandafter\gdef\csname odunum@n@release_stats.main.axis5_present\endcsname{1\,631}
\expandafter\gdef\csname odunum@val@release_stats.main.axis5_total\endcsname{10}
\expandafter\gdef\csname odunum@n@release_stats.main.axis5_total\endcsname{10}
\expandafter\gdef\csname odunum@val@release_stats.main.axis6_present\endcsname{19}
\expandafter\gdef\csname odunum@n@release_stats.main.axis6_present\endcsname{1\,631}
\expandafter\gdef\csname odunum@val@release_stats.main.axis6_total\endcsname{19}
\expandafter\gdef\csname odunum@n@release_stats.main.axis6_total\endcsname{19}
\expandafter\gdef\csname odunum@val@release_stats.main.n_ao\endcsname{731}
\expandafter\gdef\csname odunum@n@release_stats.main.n_ao\endcsname{2\,078}
\expandafter\gdef\csname odunum@val@release_stats.main.n_av\endcsname{1\,347}
\expandafter\gdef\csname odunum@n@release_stats.main.n_av\endcsname{2\,078}
\expandafter\gdef\csname odunum@val@release_stats.main.n_en\endcsname{908}
\expandafter\gdef\csname odunum@n@release_stats.main.n_en\endcsname{2\,078}
\expandafter\gdef\csname odunum@val@release_stats.main.n_generated\endcsname{1\,801}
\expandafter\gdef\csname odunum@n@release_stats.main.n_generated\endcsname{2\,078}
\expandafter\gdef\csname odunum@val@release_stats.main.n_multiclip\endcsname{969}
\expandafter\gdef\csname odunum@n@release_stats.main.n_multiclip\endcsname{2\,078}
\expandafter\gdef\csname odunum@val@release_stats.main.n_negative\endcsname{447}
\expandafter\gdef\csname odunum@n@release_stats.main.n_negative\endcsname{2\,078}
\expandafter\gdef\csname odunum@val@release_stats.main.n_positive\endcsname{1\,631}
\expandafter\gdef\csname odunum@n@release_stats.main.n_positive\endcsname{2\,078}
\expandafter\gdef\csname odunum@val@release_stats.main.n_recorded\endcsname{277}
\expandafter\gdef\csname odunum@n@release_stats.main.n_recorded\endcsname{2\,078}
\expandafter\gdef\csname odunum@val@release_stats.main.n_scenes\endcsname{2\,078}
\expandafter\gdef\csname odunum@n@release_stats.main.n_scenes\endcsname{2\,078}
\expandafter\gdef\csname odunum@val@release_stats.main.n_segments\endcsname{1\,679}
\expandafter\gdef\csname odunum@n@release_stats.main.n_segments\endcsname{2\,078}
\expandafter\gdef\csname odunum@val@release_stats.main.n_zh\endcsname{1\,170}
\expandafter\gdef\csname odunum@n@release_stats.main.n_zh\endcsname{2\,078}
\expandafter\gdef\csname odunum@val@response_ab_blind.all.decisive_n\endcsname{120}
\expandafter\gdef\csname odunum@n@response_ab_blind.all.decisive_n\endcsname{200}
\expandafter\gdef\csname odunum@val@response_ab_blind.all.given_win_rate_among_decisive\endcsname{0.858333}
\expandafter\gdef\csname odunum@n@response_ab_blind.all.given_win_rate_among_decisive\endcsname{120}
\expandafter\gdef\csname odunum@val@response_ab_blind.all.n\endcsname{200}
\expandafter\gdef\csname odunum@n@response_ab_blind.all.n\endcsname{200}
\expandafter\gdef\csname odunum@val@response_ab_blind.all.net_preference\endcsname{0.430}
\expandafter\gdef\csname odunum@n@response_ab_blind.all.net_preference\endcsname{200}
\expandafter\gdef\csname odunum@val@response_ab_blind.all.tie_rate\endcsname{0.400000}
\expandafter\gdef\csname odunum@n@response_ab_blind.all.tie_rate\endcsname{200}
\expandafter\gdef\csname odunum@val@response_ab_blind.all.ties\endcsname{80}
\expandafter\gdef\csname odunum@n@response_ab_blind.all.ties\endcsname{200}
\expandafter\gdef\csname odunum@val@response_ab_blind.all.with_demand_win_rate\endcsname{0.515000}
\expandafter\gdef\csname odunum@n@response_ab_blind.all.with_demand_win_rate\endcsname{200}
\expandafter\gdef\csname odunum@val@response_ab_blind.all.with_demand_wins\endcsname{103}
\expandafter\gdef\csname odunum@n@response_ab_blind.all.with_demand_wins\endcsname{200}
\expandafter\gdef\csname odunum@val@response_ab_blind.all.without_demand_win_rate\endcsname{0.085000}
\expandafter\gdef\csname odunum@n@response_ab_blind.all.without_demand_win_rate\endcsname{200}
\expandafter\gdef\csname odunum@val@response_ab_blind.all.without_demand_wins\endcsname{17}
\expandafter\gdef\csname odunum@n@response_ab_blind.all.without_demand_wins\endcsname{200}
\expandafter\gdef\csname odunum@val@response_ab_blind.en.decisive_n\endcsname{65}
\expandafter\gdef\csname odunum@n@response_ab_blind.en.decisive_n\endcsname{100}
\expandafter\gdef\csname odunum@val@response_ab_blind.en.given_win_rate_among_decisive\endcsname{0.861538}
\expandafter\gdef\csname odunum@n@response_ab_blind.en.given_win_rate_among_decisive\endcsname{65}
\expandafter\gdef\csname odunum@val@response_ab_blind.en.n\endcsname{100}
\expandafter\gdef\csname odunum@n@response_ab_blind.en.n\endcsname{100}
\expandafter\gdef\csname odunum@val@response_ab_blind.en.net_preference\endcsname{0.470}
\expandafter\gdef\csname odunum@n@response_ab_blind.en.net_preference\endcsname{100}
\expandafter\gdef\csname odunum@val@response_ab_blind.en.tie_rate\endcsname{0.350000}
\expandafter\gdef\csname odunum@n@response_ab_blind.en.tie_rate\endcsname{100}
\expandafter\gdef\csname odunum@val@response_ab_blind.en.ties\endcsname{35}
\expandafter\gdef\csname odunum@n@response_ab_blind.en.ties\endcsname{100}
\expandafter\gdef\csname odunum@val@response_ab_blind.en.with_demand_win_rate\endcsname{0.560000}
\expandafter\gdef\csname odunum@n@response_ab_blind.en.with_demand_win_rate\endcsname{100}
\expandafter\gdef\csname odunum@val@response_ab_blind.en.with_demand_wins\endcsname{56}
\expandafter\gdef\csname odunum@n@response_ab_blind.en.with_demand_wins\endcsname{100}
\expandafter\gdef\csname odunum@val@response_ab_blind.en.without_demand_win_rate\endcsname{0.090000}
\expandafter\gdef\csname odunum@n@response_ab_blind.en.without_demand_win_rate\endcsname{100}
\expandafter\gdef\csname odunum@val@response_ab_blind.en.without_demand_wins\endcsname{9}
\expandafter\gdef\csname odunum@n@response_ab_blind.en.without_demand_wins\endcsname{100}
\expandafter\gdef\csname odunum@val@response_ab_blind.negative.decisive_n\endcsname{18}
\expandafter\gdef\csname odunum@n@response_ab_blind.negative.decisive_n\endcsname{40}
\expandafter\gdef\csname odunum@val@response_ab_blind.negative.given_win_rate_among_decisive\endcsname{1.000000}
\expandafter\gdef\csname odunum@n@response_ab_blind.negative.given_win_rate_among_decisive\endcsname{18}
\expandafter\gdef\csname odunum@val@response_ab_blind.negative.n\endcsname{40}
\expandafter\gdef\csname odunum@n@response_ab_blind.negative.n\endcsname{40}
\expandafter\gdef\csname odunum@val@response_ab_blind.negative.net_preference\endcsname{0.450}
\expandafter\gdef\csname odunum@n@response_ab_blind.negative.net_preference\endcsname{40}
\expandafter\gdef\csname odunum@val@response_ab_blind.negative.tie_rate\endcsname{0.550000}
\expandafter\gdef\csname odunum@n@response_ab_blind.negative.tie_rate\endcsname{40}
\expandafter\gdef\csname odunum@val@response_ab_blind.negative.ties\endcsname{22}
\expandafter\gdef\csname odunum@n@response_ab_blind.negative.ties\endcsname{40}
\expandafter\gdef\csname odunum@val@response_ab_blind.negative.with_demand_reply_rate\endcsname{0.150000}
\expandafter\gdef\csname odunum@n@response_ab_blind.negative.with_demand_reply_rate\endcsname{40}
\expandafter\gdef\csname odunum@val@response_ab_blind.negative.with_demand_silent\endcsname{34}
\expandafter\gdef\csname odunum@n@response_ab_blind.negative.with_demand_silent\endcsname{40}
\expandafter\gdef\csname odunum@val@response_ab_blind.negative.with_demand_win_rate\endcsname{0.450000}
\expandafter\gdef\csname odunum@n@response_ab_blind.negative.with_demand_win_rate\endcsname{40}
\expandafter\gdef\csname odunum@val@response_ab_blind.negative.with_demand_wins\endcsname{18}
\expandafter\gdef\csname odunum@n@response_ab_blind.negative.with_demand_wins\endcsname{40}
\expandafter\gdef\csname odunum@val@response_ab_blind.negative.without_demand_reply_rate\endcsname{0.600000}
\expandafter\gdef\csname odunum@n@response_ab_blind.negative.without_demand_reply_rate\endcsname{40}
\expandafter\gdef\csname odunum@val@response_ab_blind.negative.without_demand_silent\endcsname{16}
\expandafter\gdef\csname odunum@n@response_ab_blind.negative.without_demand_silent\endcsname{40}
\expandafter\gdef\csname odunum@val@response_ab_blind.negative.without_demand_win_rate\endcsname{0.000000}
\expandafter\gdef\csname odunum@n@response_ab_blind.negative.without_demand_win_rate\endcsname{40}
\expandafter\gdef\csname odunum@val@response_ab_blind.negative.without_demand_wins\endcsname{0}
\expandafter\gdef\csname odunum@n@response_ab_blind.negative.without_demand_wins\endcsname{40}
\expandafter\gdef\csname odunum@val@response_ab_blind.positive.decisive_n\endcsname{102}
\expandafter\gdef\csname odunum@n@response_ab_blind.positive.decisive_n\endcsname{160}
\expandafter\gdef\csname odunum@val@response_ab_blind.positive.given_win_rate_among_decisive\endcsname{0.833333}
\expandafter\gdef\csname odunum@n@response_ab_blind.positive.given_win_rate_among_decisive\endcsname{102}
\expandafter\gdef\csname odunum@val@response_ab_blind.positive.n\endcsname{160}
\expandafter\gdef\csname odunum@n@response_ab_blind.positive.n\endcsname{160}
\expandafter\gdef\csname odunum@val@response_ab_blind.positive.net_preference\endcsname{0.425}
\expandafter\gdef\csname odunum@n@response_ab_blind.positive.net_preference\endcsname{160}
\expandafter\gdef\csname odunum@val@response_ab_blind.positive.tie_rate\endcsname{0.362500}
\expandafter\gdef\csname odunum@n@response_ab_blind.positive.tie_rate\endcsname{160}
\expandafter\gdef\csname odunum@val@response_ab_blind.positive.ties\endcsname{58}
\expandafter\gdef\csname odunum@n@response_ab_blind.positive.ties\endcsname{160}
\expandafter\gdef\csname odunum@val@response_ab_blind.positive.with_demand_reply_rate\endcsname{0.993750}
\expandafter\gdef\csname odunum@n@response_ab_blind.positive.with_demand_reply_rate\endcsname{160}
\expandafter\gdef\csname odunum@val@response_ab_blind.positive.with_demand_silent\endcsname{1}
\expandafter\gdef\csname odunum@n@response_ab_blind.positive.with_demand_silent\endcsname{160}
\expandafter\gdef\csname odunum@val@response_ab_blind.positive.with_demand_win_rate\endcsname{0.531250}
\expandafter\gdef\csname odunum@n@response_ab_blind.positive.with_demand_win_rate\endcsname{160}
\expandafter\gdef\csname odunum@val@response_ab_blind.positive.with_demand_wins\endcsname{85}
\expandafter\gdef\csname odunum@n@response_ab_blind.positive.with_demand_wins\endcsname{160}
\expandafter\gdef\csname odunum@val@response_ab_blind.positive.without_demand_reply_rate\endcsname{0.925000}
\expandafter\gdef\csname odunum@n@response_ab_blind.positive.without_demand_reply_rate\endcsname{160}
\expandafter\gdef\csname odunum@val@response_ab_blind.positive.without_demand_silent\endcsname{12}
\expandafter\gdef\csname odunum@n@response_ab_blind.positive.without_demand_silent\endcsname{160}
\expandafter\gdef\csname odunum@val@response_ab_blind.positive.without_demand_win_rate\endcsname{0.106250}
\expandafter\gdef\csname odunum@n@response_ab_blind.positive.without_demand_win_rate\endcsname{160}
\expandafter\gdef\csname odunum@val@response_ab_blind.positive.without_demand_wins\endcsname{17}
\expandafter\gdef\csname odunum@n@response_ab_blind.positive.without_demand_wins\endcsname{160}
\expandafter\gdef\csname odunum@val@response_ab_blind.positive_both_text.decisive_n\endcsname{91}
\expandafter\gdef\csname odunum@n@response_ab_blind.positive_both_text.decisive_n\endcsname{148}
\expandafter\gdef\csname odunum@val@response_ab_blind.positive_both_text.given_win_rate_among_decisive\endcsname{0.813187}
\expandafter\gdef\csname odunum@n@response_ab_blind.positive_both_text.given_win_rate_among_decisive\endcsname{91}
\expandafter\gdef\csname odunum@val@response_ab_blind.positive_both_text.n\endcsname{148}
\expandafter\gdef\csname odunum@n@response_ab_blind.positive_both_text.n\endcsname{148}
\expandafter\gdef\csname odunum@val@response_ab_blind.positive_both_text.net_preference\endcsname{0.385}
\expandafter\gdef\csname odunum@n@response_ab_blind.positive_both_text.net_preference\endcsname{148}
\expandafter\gdef\csname odunum@val@response_ab_blind.positive_both_text.tie_rate\endcsname{0.385135}
\expandafter\gdef\csname odunum@n@response_ab_blind.positive_both_text.tie_rate\endcsname{148}
\expandafter\gdef\csname odunum@val@response_ab_blind.positive_both_text.ties\endcsname{57}
\expandafter\gdef\csname odunum@n@response_ab_blind.positive_both_text.ties\endcsname{148}
\expandafter\gdef\csname odunum@val@response_ab_blind.positive_both_text.with_demand_win_rate\endcsname{0.500000}
\expandafter\gdef\csname odunum@n@response_ab_blind.positive_both_text.with_demand_win_rate\endcsname{148}
\expandafter\gdef\csname odunum@val@response_ab_blind.positive_both_text.with_demand_wins\endcsname{74}
\expandafter\gdef\csname odunum@n@response_ab_blind.positive_both_text.with_demand_wins\endcsname{148}
\expandafter\gdef\csname odunum@val@response_ab_blind.positive_both_text.without_demand_win_rate\endcsname{0.114865}
\expandafter\gdef\csname odunum@n@response_ab_blind.positive_both_text.without_demand_win_rate\endcsname{148}
\expandafter\gdef\csname odunum@val@response_ab_blind.positive_both_text.without_demand_wins\endcsname{17}
\expandafter\gdef\csname odunum@n@response_ab_blind.positive_both_text.without_demand_wins\endcsname{148}
\expandafter\gdef\csname odunum@val@response_ab_blind.zh.decisive_n\endcsname{55}
\expandafter\gdef\csname odunum@n@response_ab_blind.zh.decisive_n\endcsname{100}
\expandafter\gdef\csname odunum@val@response_ab_blind.zh.given_win_rate_among_decisive\endcsname{0.854545}
\expandafter\gdef\csname odunum@n@response_ab_blind.zh.given_win_rate_among_decisive\endcsname{55}
\expandafter\gdef\csname odunum@val@response_ab_blind.zh.n\endcsname{100}
\expandafter\gdef\csname odunum@n@response_ab_blind.zh.n\endcsname{100}
\expandafter\gdef\csname odunum@val@response_ab_blind.zh.net_preference\endcsname{0.390}
\expandafter\gdef\csname odunum@n@response_ab_blind.zh.net_preference\endcsname{100}
\expandafter\gdef\csname odunum@val@response_ab_blind.zh.tie_rate\endcsname{0.450000}
\expandafter\gdef\csname odunum@n@response_ab_blind.zh.tie_rate\endcsname{100}
\expandafter\gdef\csname odunum@val@response_ab_blind.zh.ties\endcsname{45}
\expandafter\gdef\csname odunum@n@response_ab_blind.zh.ties\endcsname{100}
\expandafter\gdef\csname odunum@val@response_ab_blind.zh.with_demand_win_rate\endcsname{0.470000}
\expandafter\gdef\csname odunum@n@response_ab_blind.zh.with_demand_win_rate\endcsname{100}
\expandafter\gdef\csname odunum@val@response_ab_blind.zh.with_demand_wins\endcsname{47}
\expandafter\gdef\csname odunum@n@response_ab_blind.zh.with_demand_wins\endcsname{100}
\expandafter\gdef\csname odunum@val@response_ab_blind.zh.without_demand_win_rate\endcsname{0.080000}
\expandafter\gdef\csname odunum@n@response_ab_blind.zh.without_demand_win_rate\endcsname{100}
\expandafter\gdef\csname odunum@val@response_ab_blind.zh.without_demand_wins\endcsname{8}
\expandafter\gdef\csname odunum@n@response_ab_blind.zh.without_demand_wins\endcsname{100}
\expandafter\gdef\csname odunum@val@response_ab_scene_sensitivity.all_scenes.n\endcsname{200}
\expandafter\gdef\csname odunum@n@response_ab_scene_sensitivity.all_scenes.n\endcsname{200}
\expandafter\gdef\csname odunum@val@response_ab_scene_sensitivity.all_scenes.net_preference\endcsname{0.415}
\expandafter\gdef\csname odunum@n@response_ab_scene_sensitivity.all_scenes.net_preference\endcsname{200}
\expandafter\gdef\csname odunum@val@response_ab_scene_sensitivity.all_scenes.tie_rate\endcsname{0.395}
\expandafter\gdef\csname odunum@n@response_ab_scene_sensitivity.all_scenes.tie_rate\endcsname{200}
\expandafter\gdef\csname odunum@val@response_ab_scene_sensitivity.all_scenes.ties\endcsname{79}
\expandafter\gdef\csname odunum@n@response_ab_scene_sensitivity.all_scenes.ties\endcsname{200}
\expandafter\gdef\csname odunum@val@response_ab_scene_sensitivity.all_scenes.with_demand_win_rate\endcsname{0.510}
\expandafter\gdef\csname odunum@n@response_ab_scene_sensitivity.all_scenes.with_demand_win_rate\endcsname{200}
\expandafter\gdef\csname odunum@val@response_ab_scene_sensitivity.all_scenes.with_demand_wins\endcsname{102}
\expandafter\gdef\csname odunum@n@response_ab_scene_sensitivity.all_scenes.with_demand_wins\endcsname{200}
\expandafter\gdef\csname odunum@val@response_ab_scene_sensitivity.all_scenes.without_demand_win_rate\endcsname{0.095}
\expandafter\gdef\csname odunum@n@response_ab_scene_sensitivity.all_scenes.without_demand_win_rate\endcsname{200}
\expandafter\gdef\csname odunum@val@response_ab_scene_sensitivity.all_scenes.without_demand_wins\endcsname{19}
\expandafter\gdef\csname odunum@n@response_ab_scene_sensitivity.all_scenes.without_demand_wins\endcsname{200}
\expandafter\gdef\csname odunum@val@response_ab_scene_sensitivity.disputed_scenes.n\endcsname{2}
\expandafter\gdef\csname odunum@n@response_ab_scene_sensitivity.disputed_scenes.n\endcsname{2}
\expandafter\gdef\csname odunum@val@response_ab_scene_sensitivity.disputed_scenes.net_preference\endcsname{-1.000}
\expandafter\gdef\csname odunum@n@response_ab_scene_sensitivity.disputed_scenes.net_preference\endcsname{2}
\expandafter\gdef\csname odunum@val@response_ab_scene_sensitivity.disputed_scenes.tie_rate\endcsname{0.000}
\expandafter\gdef\csname odunum@n@response_ab_scene_sensitivity.disputed_scenes.tie_rate\endcsname{2}
\expandafter\gdef\csname odunum@val@response_ab_scene_sensitivity.disputed_scenes.ties\endcsname{0}
\expandafter\gdef\csname odunum@n@response_ab_scene_sensitivity.disputed_scenes.ties\endcsname{2}
\expandafter\gdef\csname odunum@val@response_ab_scene_sensitivity.disputed_scenes.with_demand_win_rate\endcsname{0.000}
\expandafter\gdef\csname odunum@n@response_ab_scene_sensitivity.disputed_scenes.with_demand_win_rate\endcsname{2}
\expandafter\gdef\csname odunum@val@response_ab_scene_sensitivity.disputed_scenes.with_demand_wins\endcsname{0}
\expandafter\gdef\csname odunum@n@response_ab_scene_sensitivity.disputed_scenes.with_demand_wins\endcsname{2}
\expandafter\gdef\csname odunum@val@response_ab_scene_sensitivity.disputed_scenes.without_demand_win_rate\endcsname{1.000}
\expandafter\gdef\csname odunum@n@response_ab_scene_sensitivity.disputed_scenes.without_demand_win_rate\endcsname{2}
\expandafter\gdef\csname odunum@val@response_ab_scene_sensitivity.disputed_scenes.without_demand_wins\endcsname{2}
\expandafter\gdef\csname odunum@n@response_ab_scene_sensitivity.disputed_scenes.without_demand_wins\endcsname{2}
\expandafter\gdef\csname odunum@val@response_ab_scene_sensitivity.excluding_disputed.n\endcsname{198}
\expandafter\gdef\csname odunum@n@response_ab_scene_sensitivity.excluding_disputed.n\endcsname{198}
\expandafter\gdef\csname odunum@val@response_ab_scene_sensitivity.excluding_disputed.net_preference\endcsname{0.429}
\expandafter\gdef\csname odunum@n@response_ab_scene_sensitivity.excluding_disputed.net_preference\endcsname{198}
\expandafter\gdef\csname odunum@val@response_ab_scene_sensitivity.excluding_disputed.tie_rate\endcsname{0.399}
\expandafter\gdef\csname odunum@n@response_ab_scene_sensitivity.excluding_disputed.tie_rate\endcsname{198}
\expandafter\gdef\csname odunum@val@response_ab_scene_sensitivity.excluding_disputed.ties\endcsname{79}
\expandafter\gdef\csname odunum@n@response_ab_scene_sensitivity.excluding_disputed.ties\endcsname{198}
\expandafter\gdef\csname odunum@val@response_ab_scene_sensitivity.excluding_disputed.with_demand_win_rate\endcsname{0.515}
\expandafter\gdef\csname odunum@n@response_ab_scene_sensitivity.excluding_disputed.with_demand_win_rate\endcsname{198}
\expandafter\gdef\csname odunum@val@response_ab_scene_sensitivity.excluding_disputed.with_demand_wins\endcsname{102}
\expandafter\gdef\csname odunum@n@response_ab_scene_sensitivity.excluding_disputed.with_demand_wins\endcsname{198}
\expandafter\gdef\csname odunum@val@response_ab_scene_sensitivity.excluding_disputed.without_demand_win_rate\endcsname{0.086}
\expandafter\gdef\csname odunum@n@response_ab_scene_sensitivity.excluding_disputed.without_demand_win_rate\endcsname{198}
\expandafter\gdef\csname odunum@val@response_ab_scene_sensitivity.excluding_disputed.without_demand_wins\endcsname{17}
\expandafter\gdef\csname odunum@n@response_ab_scene_sensitivity.excluding_disputed.without_demand_wins\endcsname{198}
\expandafter\gdef\csname odunum@val@response_ab_scene_sensitivity.negative_excluding_disputed.n\endcsname{38}
\expandafter\gdef\csname odunum@n@response_ab_scene_sensitivity.negative_excluding_disputed.n\endcsname{38}
\expandafter\gdef\csname odunum@val@response_ab_scene_sensitivity.negative_excluding_disputed.net_preference\endcsname{0.447}
\expandafter\gdef\csname odunum@n@response_ab_scene_sensitivity.negative_excluding_disputed.net_preference\endcsname{38}
\expandafter\gdef\csname odunum@val@response_ab_scene_sensitivity.negative_excluding_disputed.tie_rate\endcsname{0.553}
\expandafter\gdef\csname odunum@n@response_ab_scene_sensitivity.negative_excluding_disputed.tie_rate\endcsname{38}
\expandafter\gdef\csname odunum@val@response_ab_scene_sensitivity.negative_excluding_disputed.ties\endcsname{21}
\expandafter\gdef\csname odunum@n@response_ab_scene_sensitivity.negative_excluding_disputed.ties\endcsname{38}
\expandafter\gdef\csname odunum@val@response_ab_scene_sensitivity.negative_excluding_disputed.with_demand_win_rate\endcsname{0.447}
\expandafter\gdef\csname odunum@n@response_ab_scene_sensitivity.negative_excluding_disputed.with_demand_win_rate\endcsname{38}
\expandafter\gdef\csname odunum@val@response_ab_scene_sensitivity.negative_excluding_disputed.with_demand_wins\endcsname{17}
\expandafter\gdef\csname odunum@n@response_ab_scene_sensitivity.negative_excluding_disputed.with_demand_wins\endcsname{38}
\expandafter\gdef\csname odunum@val@response_ab_scene_sensitivity.negative_excluding_disputed.without_demand_win_rate\endcsname{0.000}
\expandafter\gdef\csname odunum@n@response_ab_scene_sensitivity.negative_excluding_disputed.without_demand_win_rate\endcsname{38}
\expandafter\gdef\csname odunum@val@response_ab_scene_sensitivity.negative_excluding_disputed.without_demand_wins\endcsname{0}
\expandafter\gdef\csname odunum@n@response_ab_scene_sensitivity.negative_excluding_disputed.without_demand_wins\endcsname{38}
\expandafter\gdef\csname odunum@val@response_ab_scene_sensitivity.positive_excluding_disputed.n\endcsname{160}
\expandafter\gdef\csname odunum@n@response_ab_scene_sensitivity.positive_excluding_disputed.n\endcsname{160}
\expandafter\gdef\csname odunum@val@response_ab_scene_sensitivity.positive_excluding_disputed.net_preference\endcsname{0.425}
\expandafter\gdef\csname odunum@n@response_ab_scene_sensitivity.positive_excluding_disputed.net_preference\endcsname{160}
\expandafter\gdef\csname odunum@val@response_ab_scene_sensitivity.positive_excluding_disputed.tie_rate\endcsname{0.362}
\expandafter\gdef\csname odunum@n@response_ab_scene_sensitivity.positive_excluding_disputed.tie_rate\endcsname{160}
\expandafter\gdef\csname odunum@val@response_ab_scene_sensitivity.positive_excluding_disputed.ties\endcsname{58}
\expandafter\gdef\csname odunum@n@response_ab_scene_sensitivity.positive_excluding_disputed.ties\endcsname{160}
\expandafter\gdef\csname odunum@val@response_ab_scene_sensitivity.positive_excluding_disputed.with_demand_win_rate\endcsname{0.531}
\expandafter\gdef\csname odunum@n@response_ab_scene_sensitivity.positive_excluding_disputed.with_demand_win_rate\endcsname{160}
\expandafter\gdef\csname odunum@val@response_ab_scene_sensitivity.positive_excluding_disputed.with_demand_wins\endcsname{85}
\expandafter\gdef\csname odunum@n@response_ab_scene_sensitivity.positive_excluding_disputed.with_demand_wins\endcsname{160}
\expandafter\gdef\csname odunum@val@response_ab_scene_sensitivity.positive_excluding_disputed.without_demand_win_rate\endcsname{0.106}
\expandafter\gdef\csname odunum@n@response_ab_scene_sensitivity.positive_excluding_disputed.without_demand_win_rate\endcsname{160}
\expandafter\gdef\csname odunum@val@response_ab_scene_sensitivity.positive_excluding_disputed.without_demand_wins\endcsname{17}
\expandafter\gdef\csname odunum@n@response_ab_scene_sensitivity.positive_excluding_disputed.without_demand_wins\endcsname{160}
\expandafter\gdef\csname odunum@val@response_ab_setup.candidateCount\endcsname{1}
\expandafter\gdef\csname odunum@n@response_ab_setup.candidateCount\endcsname{400}
\expandafter\gdef\csname odunum@val@response_ab_setup.maxOutputTokens\endcsname{8\,192}
\expandafter\gdef\csname odunum@n@response_ab_setup.maxOutputTokens\endcsname{400}
\expandafter\gdef\csname odunum@val@response_ab_setup.model_id\endcsname{gemini-3.1-pro-preview}
\expandafter\gdef\csname odunum@n@response_ab_setup.model_id\endcsname{400}
\expandafter\gdef\csname odunum@val@response_ab_setup.n_conditions\endcsname{2}
\expandafter\gdef\csname odunum@n@response_ab_setup.n_conditions\endcsname{1}
\expandafter\gdef\csname odunum@val@response_ab_setup.n_negative\endcsname{40}
\expandafter\gdef\csname odunum@n@response_ab_setup.n_negative\endcsname{200}
\expandafter\gdef\csname odunum@val@response_ab_setup.n_negative.en\endcsname{20}
\expandafter\gdef\csname odunum@n@response_ab_setup.n_negative.en\endcsname{200}
\expandafter\gdef\csname odunum@val@response_ab_setup.n_negative.zh\endcsname{20}
\expandafter\gdef\csname odunum@n@response_ab_setup.n_negative.zh\endcsname{200}
\expandafter\gdef\csname odunum@val@response_ab_setup.n_planned_responses\endcsname{400}
\expandafter\gdef\csname odunum@n@response_ab_setup.n_planned_responses\endcsname{200}
\expandafter\gdef\csname odunum@val@response_ab_setup.n_positive\endcsname{160}
\expandafter\gdef\csname odunum@n@response_ab_setup.n_positive\endcsname{200}
\expandafter\gdef\csname odunum@val@response_ab_setup.n_positive.en\endcsname{80}
\expandafter\gdef\csname odunum@n@response_ab_setup.n_positive.en\endcsname{200}
\expandafter\gdef\csname odunum@val@response_ab_setup.n_positive.zh\endcsname{80}
\expandafter\gdef\csname odunum@n@response_ab_setup.n_positive.zh\endcsname{200}
\expandafter\gdef\csname odunum@val@response_ab_setup.n_replaced_scenes\endcsname{2}
\expandafter\gdef\csname odunum@n@response_ab_setup.n_replaced_scenes\endcsname{200}
\expandafter\gdef\csname odunum@val@response_ab_setup.n_scenes\endcsname{200}
\expandafter\gdef\csname odunum@n@response_ab_setup.n_scenes\endcsname{200}
\expandafter\gdef\csname odunum@val@response_ab_setup.temperature\endcsname{0}
\expandafter\gdef\csname odunum@n@response_ab_setup.temperature\endcsname{400}
\expandafter\gdef\csname odunum@val@response_ab_setup.topP\endcsname{1}
\expandafter\gdef\csname odunum@n@response_ab_setup.topP\endcsname{400}
\expandafter\gdef\csname odunum@val@response_ab_setup_v2.bootstrap_draws\endcsname{10\,000}
\expandafter\gdef\csname odunum@n@response_ab_setup_v2.bootstrap_draws\endcsname{1}
\expandafter\gdef\csname odunum@val@response_ab_setup_v2.candidateCount\endcsname{1}
\expandafter\gdef\csname odunum@n@response_ab_setup_v2.candidateCount\endcsname{400}
\expandafter\gdef\csname odunum@val@response_ab_setup_v2.maxOutputTokens\endcsname{8\,192}
\expandafter\gdef\csname odunum@n@response_ab_setup_v2.maxOutputTokens\endcsname{400}
\expandafter\gdef\csname odunum@val@response_ab_setup_v2.model_id\endcsname{gemini-3.1-pro-preview}
\expandafter\gdef\csname odunum@n@response_ab_setup_v2.model_id\endcsname{400}
\expandafter\gdef\csname odunum@val@response_ab_setup_v2.n_conditions\endcsname{2}
\expandafter\gdef\csname odunum@n@response_ab_setup_v2.n_conditions\endcsname{1}
\expandafter\gdef\csname odunum@val@response_ab_setup_v2.n_negative\endcsname{40}
\expandafter\gdef\csname odunum@n@response_ab_setup_v2.n_negative\endcsname{200}
\expandafter\gdef\csname odunum@val@response_ab_setup_v2.n_negative.en\endcsname{20}
\expandafter\gdef\csname odunum@n@response_ab_setup_v2.n_negative.en\endcsname{200}
\expandafter\gdef\csname odunum@val@response_ab_setup_v2.n_negative.zh\endcsname{20}
\expandafter\gdef\csname odunum@n@response_ab_setup_v2.n_negative.zh\endcsname{200}
\expandafter\gdef\csname odunum@val@response_ab_setup_v2.n_planned_responses\endcsname{400}
\expandafter\gdef\csname odunum@n@response_ab_setup_v2.n_planned_responses\endcsname{200}
\expandafter\gdef\csname odunum@val@response_ab_setup_v2.n_positive\endcsname{160}
\expandafter\gdef\csname odunum@n@response_ab_setup_v2.n_positive\endcsname{200}
\expandafter\gdef\csname odunum@val@response_ab_setup_v2.n_positive.en\endcsname{80}
\expandafter\gdef\csname odunum@n@response_ab_setup_v2.n_positive.en\endcsname{200}
\expandafter\gdef\csname odunum@val@response_ab_setup_v2.n_positive.zh\endcsname{80}
\expandafter\gdef\csname odunum@n@response_ab_setup_v2.n_positive.zh\endcsname{200}
\expandafter\gdef\csname odunum@val@response_ab_setup_v2.n_scenes\endcsname{200}
\expandafter\gdef\csname odunum@n@response_ab_setup_v2.n_scenes\endcsname{200}
\expandafter\gdef\csname odunum@val@response_ab_setup_v2.temperature\endcsname{0}
\expandafter\gdef\csname odunum@n@response_ab_setup_v2.temperature\endcsname{400}
\expandafter\gdef\csname odunum@val@response_ab_setup_v2.topP\endcsname{1}
\expandafter\gdef\csname odunum@n@response_ab_setup_v2.topP\endcsname{400}
\expandafter\gdef\csname odunum@val@source_transfer_candidate.rank.native.avg.kendall_tau\endcsname{0.909}
\expandafter\gdef\csname odunum@n@source_transfer_candidate.rank.native.avg.kendall_tau\endcsname{12}
\expandafter\gdef\csname odunum@val@source_transfer_candidate.rank.native.avg.spearman\endcsname{0.972}
\expandafter\gdef\csname odunum@n@source_transfer_candidate.rank.native.avg.spearman\endcsname{12}
\expandafter\gdef\csname odunum@val@source_transfer_candidate.rank.native.m2.kendall_tau\endcsname{0.939}
\expandafter\gdef\csname odunum@n@source_transfer_candidate.rank.native.m2.kendall_tau\endcsname{12}
\expandafter\gdef\csname odunum@val@source_transfer_candidate.rank.native.m2.spearman\endcsname{0.979}
\expandafter\gdef\csname odunum@n@source_transfer_candidate.rank.native.m2.spearman\endcsname{12}
\expandafter\gdef\csname odunum@val@source_transfer_candidate.rank.table4.avg.kendall_tau\endcsname{0.667}
\expandafter\gdef\csname odunum@n@source_transfer_candidate.rank.table4.avg.kendall_tau\endcsname{4}
\expandafter\gdef\csname odunum@val@source_transfer_candidate.rank.table4.avg.spearman\endcsname{0.800}
\expandafter\gdef\csname odunum@n@source_transfer_candidate.rank.table4.avg.spearman\endcsname{4}
\expandafter\gdef\csname odunum@val@source_transfer_candidate.rank.table4.m2.kendall_tau\endcsname{0.333}
\expandafter\gdef\csname odunum@n@source_transfer_candidate.rank.table4.m2.kendall_tau\endcsname{4}
\expandafter\gdef\csname odunum@val@source_transfer_candidate.rank.table4.m2.spearman\endcsname{0.400}
\expandafter\gdef\csname odunum@n@source_transfer_candidate.rank.table4.m2.spearman\endcsname{4}
\expandafter\gdef\csname odunum@val@source_transfer_candidate.weighted_delta.qwen3_omni_think.m1\endcsname{-0.018}
\expandafter\gdef\csname odunum@n@source_transfer_candidate.weighted_delta.qwen3_omni_think.m1\endcsname{276}
\expandafter\gdef\csname odunum@val@source_transfer_candidate.weighted_delta.qwen3_omni_think.m2\endcsname{-0.041}
\expandafter\gdef\csname odunum@n@source_transfer_candidate.weighted_delta.qwen3_omni_think.m2\endcsname{201}
\expandafter\gdef\csname odunum@val@source_transfer_candidate.weighted_delta.qwen3_omni_think.m3\endcsname{0.004}
\expandafter\gdef\csname odunum@n@source_transfer_candidate.weighted_delta.qwen3_omni_think.m3\endcsname{201}
\expandafter\gdef\csname odunum@val@source_transfer_candidate.weighted_delta.qwen3_omni_think.m4\endcsname{-0.019}
\expandafter\gdef\csname odunum@n@source_transfer_candidate.weighted_delta.qwen3_omni_think.m4\endcsname{201}
\expandafter\gdef\csname odunum@val@source_transfer_candidate.weighted_delta.qwen3_omni_think.m5\endcsname{-0.001}
\expandafter\gdef\csname odunum@n@source_transfer_candidate.weighted_delta.qwen3_omni_think.m5\endcsname{201}
\expandafter\gdef\csname odunum@val@thinking_vs_instruct.cov.instruct.audio\endcsname{0.330}
\expandafter\gdef\csname odunum@n@thinking_vs_instruct.cov.instruct.audio\endcsname{889}
\expandafter\gdef\csname odunum@val@thinking_vs_instruct.cov.instruct.history\endcsname{0.395}
\expandafter\gdef\csname odunum@n@thinking_vs_instruct.cov.instruct.history\endcsname{815}
\expandafter\gdef\csname odunum@val@thinking_vs_instruct.cov.instruct.intent\endcsname{0.721}
\expandafter\gdef\csname odunum@n@thinking_vs_instruct.cov.instruct.intent\endcsname{3\,118}
\expandafter\gdef\csname odunum@val@thinking_vs_instruct.cov.instruct.other_context\endcsname{0.239}
\expandafter\gdef\csname odunum@n@thinking_vs_instruct.cov.instruct.other_context\endcsname{180}
\expandafter\gdef\csname odunum@val@thinking_vs_instruct.cov.instruct.visual\endcsname{0.359}
\expandafter\gdef\csname odunum@n@thinking_vs_instruct.cov.instruct.visual\endcsname{803}
\expandafter\gdef\csname odunum@val@thinking_vs_instruct.cov.thinking.audio\endcsname{0.324}
\expandafter\gdef\csname odunum@n@thinking_vs_instruct.cov.thinking.audio\endcsname{889}
\expandafter\gdef\csname odunum@val@thinking_vs_instruct.cov.thinking.history\endcsname{0.428}
\expandafter\gdef\csname odunum@n@thinking_vs_instruct.cov.thinking.history\endcsname{815}
\expandafter\gdef\csname odunum@val@thinking_vs_instruct.cov.thinking.intent\endcsname{0.764}
\expandafter\gdef\csname odunum@n@thinking_vs_instruct.cov.thinking.intent\endcsname{3\,118}
\expandafter\gdef\csname odunum@val@thinking_vs_instruct.cov.thinking.other_context\endcsname{0.150}
\expandafter\gdef\csname odunum@n@thinking_vs_instruct.cov.thinking.other_context\endcsname{180}
\expandafter\gdef\csname odunum@val@thinking_vs_instruct.cov.thinking.visual\endcsname{0.389}
\expandafter\gdef\csname odunum@n@thinking_vs_instruct.cov.thinking.visual\endcsname{803}
\expandafter\gdef\csname odunum@val@thinking_vs_instruct.cov_delta.audio\endcsname{-0.006}
\expandafter\gdef\csname odunum@n@thinking_vs_instruct.cov_delta.audio\endcsname{889}
\expandafter\gdef\csname odunum@ci@thinking_vs_instruct.cov_delta.audio\endcsname{[-0.035, 0.025]}
\expandafter\gdef\csname odunum@val@thinking_vs_instruct.cov_delta.audio.sign_p\endcsname{0.83\ensuremath{\times}}
\expandafter\gdef\csname odunum@n@thinking_vs_instruct.cov_delta.audio.sign_p\endcsname{186}
\expandafter\gdef\csname odunum@val@thinking_vs_instruct.cov_delta.history\endcsname{0.033}
\expandafter\gdef\csname odunum@n@thinking_vs_instruct.cov_delta.history\endcsname{815}
\expandafter\gdef\csname odunum@ci@thinking_vs_instruct.cov_delta.history\endcsname{[-0.004, 0.069]}
\expandafter\gdef\csname odunum@val@thinking_vs_instruct.cov_delta.history.sign_p\endcsname{0.13\ensuremath{\times}}
\expandafter\gdef\csname odunum@n@thinking_vs_instruct.cov_delta.history.sign_p\endcsname{192}
\expandafter\gdef\csname odunum@val@thinking_vs_instruct.cov_delta.intent\endcsname{0.043}
\expandafter\gdef\csname odunum@n@thinking_vs_instruct.cov_delta.intent\endcsname{3\,118}
\expandafter\gdef\csname odunum@ci@thinking_vs_instruct.cov_delta.intent\endcsname{[0.026, 0.059]}
\expandafter\gdef\csname odunum@val@thinking_vs_instruct.cov_delta.intent.sign_p\endcsname{7.7\ensuremath{\times 10^{-7}}\ensuremath{\times}}
\expandafter\gdef\csname odunum@n@thinking_vs_instruct.cov_delta.intent.sign_p\endcsname{438}
\expandafter\gdef\csname odunum@val@thinking_vs_instruct.cov_delta.other_context\endcsname{-0.089}
\expandafter\gdef\csname odunum@n@thinking_vs_instruct.cov_delta.other_context\endcsname{180}
\expandafter\gdef\csname odunum@ci@thinking_vs_instruct.cov_delta.other_context\endcsname{[-0.157, -0.022]}
\expandafter\gdef\csname odunum@val@thinking_vs_instruct.cov_delta.other_context.sign_p\endcsname{0.02\ensuremath{\times}}
\expandafter\gdef\csname odunum@n@thinking_vs_instruct.cov_delta.other_context.sign_p\endcsname{40}
\expandafter\gdef\csname odunum@val@thinking_vs_instruct.cov_delta.visual\endcsname{0.030}
\expandafter\gdef\csname odunum@n@thinking_vs_instruct.cov_delta.visual\endcsname{803}
\expandafter\gdef\csname odunum@ci@thinking_vs_instruct.cov_delta.visual\endcsname{[-0.005, 0.065]}
\expandafter\gdef\csname odunum@val@thinking_vs_instruct.cov_delta.visual.sign_p\endcsname{0.09\ensuremath{\times}}
\expandafter\gdef\csname odunum@n@thinking_vs_instruct.cov_delta.visual.sign_p\endcsname{173}
\expandafter\gdef\csname odunum@val@thinking_vs_instruct.level.all.instruct.avg\endcsname{0.575}
\expandafter\gdef\csname odunum@n@thinking_vs_instruct.level.all.instruct.avg\endcsname{2\,078}
\expandafter\gdef\csname odunum@val@thinking_vs_instruct.level.all.instruct.ftr\endcsname{0.888}
\expandafter\gdef\csname odunum@n@thinking_vs_instruct.level.all.instruct.ftr\endcsname{447}
\expandafter\gdef\csname odunum@ci@thinking_vs_instruct.level.all.instruct.ftr\endcsname{[0.856, 0.914]}
\expandafter\gdef\csname odunum@val@thinking_vs_instruct.level.all.instruct.m1\endcsname{0.546}
\expandafter\gdef\csname odunum@n@thinking_vs_instruct.level.all.instruct.m1\endcsname{2\,078}
\expandafter\gdef\csname odunum@val@thinking_vs_instruct.level.all.instruct.m2\endcsname{0.558}
\expandafter\gdef\csname odunum@n@thinking_vs_instruct.level.all.instruct.m2\endcsname{1\,631}
\expandafter\gdef\csname odunum@val@thinking_vs_instruct.level.all.instruct.m3\endcsname{0.227}
\expandafter\gdef\csname odunum@n@thinking_vs_instruct.level.all.instruct.m3\endcsname{1\,631}
\expandafter\gdef\csname odunum@val@thinking_vs_instruct.level.all.instruct.m4\endcsname{0.885}
\expandafter\gdef\csname odunum@n@thinking_vs_instruct.level.all.instruct.m4\endcsname{1\,631}
\expandafter\gdef\csname odunum@val@thinking_vs_instruct.level.all.instruct.m5\endcsname{0.936}
\expandafter\gdef\csname odunum@n@thinking_vs_instruct.level.all.instruct.m5\endcsname{1\,631}
\expandafter\gdef\csname odunum@val@thinking_vs_instruct.level.all.thinking.avg\endcsname{0.626}
\expandafter\gdef\csname odunum@n@thinking_vs_instruct.level.all.thinking.avg\endcsname{2\,078}
\expandafter\gdef\csname odunum@val@thinking_vs_instruct.level.all.thinking.ftr\endcsname{0.749}
\expandafter\gdef\csname odunum@n@thinking_vs_instruct.level.all.thinking.ftr\endcsname{447}
\expandafter\gdef\csname odunum@ci@thinking_vs_instruct.level.all.thinking.ftr\endcsname{[0.707, 0.787]}
\expandafter\gdef\csname odunum@val@thinking_vs_instruct.level.all.thinking.m1\endcsname{0.641}
\expandafter\gdef\csname odunum@n@thinking_vs_instruct.level.all.thinking.m1\endcsname{2\,078}
\expandafter\gdef\csname odunum@val@thinking_vs_instruct.level.all.thinking.m2\endcsname{0.582}
\expandafter\gdef\csname odunum@n@thinking_vs_instruct.level.all.thinking.m2\endcsname{1\,631}
\expandafter\gdef\csname odunum@val@thinking_vs_instruct.level.all.thinking.m3\endcsname{0.482}
\expandafter\gdef\csname odunum@n@thinking_vs_instruct.level.all.thinking.m3\endcsname{1\,631}
\expandafter\gdef\csname odunum@val@thinking_vs_instruct.level.all.thinking.m4\endcsname{0.862}
\expandafter\gdef\csname odunum@n@thinking_vs_instruct.level.all.thinking.m4\endcsname{1\,631}
\expandafter\gdef\csname odunum@val@thinking_vs_instruct.level.all.thinking.m5\endcsname{0.929}
\expandafter\gdef\csname odunum@n@thinking_vs_instruct.level.all.thinking.m5\endcsname{1\,631}
\expandafter\gdef\csname odunum@val@thinking_vs_instruct.level.ao.instruct.avg\endcsname{0.616}
\expandafter\gdef\csname odunum@n@thinking_vs_instruct.level.ao.instruct.avg\endcsname{731}
\expandafter\gdef\csname odunum@val@thinking_vs_instruct.level.ao.instruct.ftr\endcsname{0.807}
\expandafter\gdef\csname odunum@n@thinking_vs_instruct.level.ao.instruct.ftr\endcsname{161}
\expandafter\gdef\csname odunum@ci@thinking_vs_instruct.level.ao.instruct.ftr\endcsname{[0.740, 0.861]}
\expandafter\gdef\csname odunum@val@thinking_vs_instruct.level.ao.instruct.m1\endcsname{0.610}
\expandafter\gdef\csname odunum@n@thinking_vs_instruct.level.ao.instruct.m1\endcsname{731}
\expandafter\gdef\csname odunum@val@thinking_vs_instruct.level.ao.instruct.m2\endcsname{0.609}
\expandafter\gdef\csname odunum@n@thinking_vs_instruct.level.ao.instruct.m2\endcsname{570}
\expandafter\gdef\csname odunum@val@thinking_vs_instruct.level.ao.instruct.m3\endcsname{0.205}
\expandafter\gdef\csname odunum@n@thinking_vs_instruct.level.ao.instruct.m3\endcsname{570}
\expandafter\gdef\csname odunum@val@thinking_vs_instruct.level.ao.instruct.m4\endcsname{0.910}
\expandafter\gdef\csname odunum@n@thinking_vs_instruct.level.ao.instruct.m4\endcsname{570}
\expandafter\gdef\csname odunum@val@thinking_vs_instruct.level.ao.instruct.m5\endcsname{0.952}
\expandafter\gdef\csname odunum@n@thinking_vs_instruct.level.ao.instruct.m5\endcsname{570}
\expandafter\gdef\csname odunum@val@thinking_vs_instruct.level.ao.thinking.avg\endcsname{0.674}
\expandafter\gdef\csname odunum@n@thinking_vs_instruct.level.ao.thinking.avg\endcsname{731}
\expandafter\gdef\csname odunum@val@thinking_vs_instruct.level.ao.thinking.ftr\endcsname{0.609}
\expandafter\gdef\csname odunum@n@thinking_vs_instruct.level.ao.thinking.ftr\endcsname{161}
\expandafter\gdef\csname odunum@ci@thinking_vs_instruct.level.ao.thinking.ftr\endcsname{[0.532, 0.681]}
\expandafter\gdef\csname odunum@val@thinking_vs_instruct.level.ao.thinking.m1\endcsname{0.737}
\expandafter\gdef\csname odunum@n@thinking_vs_instruct.level.ao.thinking.m1\endcsname{731}
\expandafter\gdef\csname odunum@val@thinking_vs_instruct.level.ao.thinking.m2\endcsname{0.630}
\expandafter\gdef\csname odunum@n@thinking_vs_instruct.level.ao.thinking.m2\endcsname{570}
\expandafter\gdef\csname odunum@val@thinking_vs_instruct.level.ao.thinking.m3\endcsname{0.505}
\expandafter\gdef\csname odunum@n@thinking_vs_instruct.level.ao.thinking.m3\endcsname{570}
\expandafter\gdef\csname odunum@val@thinking_vs_instruct.level.ao.thinking.m4\endcsname{0.881}
\expandafter\gdef\csname odunum@n@thinking_vs_instruct.level.ao.thinking.m4\endcsname{570}
\expandafter\gdef\csname odunum@val@thinking_vs_instruct.level.ao.thinking.m5\endcsname{0.946}
\expandafter\gdef\csname odunum@n@thinking_vs_instruct.level.ao.thinking.m5\endcsname{570}
\expandafter\gdef\csname odunum@val@thinking_vs_instruct.level.av.instruct.avg\endcsname{0.552}
\expandafter\gdef\csname odunum@n@thinking_vs_instruct.level.av.instruct.avg\endcsname{1\,347}
\expandafter\gdef\csname odunum@val@thinking_vs_instruct.level.av.instruct.ftr\endcsname{0.934}
\expandafter\gdef\csname odunum@n@thinking_vs_instruct.level.av.instruct.ftr\endcsname{286}
\expandafter\gdef\csname odunum@ci@thinking_vs_instruct.level.av.instruct.ftr\endcsname{[0.899, 0.957]}
\expandafter\gdef\csname odunum@val@thinking_vs_instruct.level.av.instruct.m1\endcsname{0.506}
\expandafter\gdef\csname odunum@n@thinking_vs_instruct.level.av.instruct.m1\endcsname{1\,347}
\expandafter\gdef\csname odunum@val@thinking_vs_instruct.level.av.instruct.m2\endcsname{0.531}
\expandafter\gdef\csname odunum@n@thinking_vs_instruct.level.av.instruct.m2\endcsname{1\,061}
\expandafter\gdef\csname odunum@val@thinking_vs_instruct.level.av.instruct.m3\endcsname{0.239}
\expandafter\gdef\csname odunum@n@thinking_vs_instruct.level.av.instruct.m3\endcsname{1\,061}
\expandafter\gdef\csname odunum@val@thinking_vs_instruct.level.av.instruct.m4\endcsname{0.871}
\expandafter\gdef\csname odunum@n@thinking_vs_instruct.level.av.instruct.m4\endcsname{1\,061}
\expandafter\gdef\csname odunum@val@thinking_vs_instruct.level.av.instruct.m5\endcsname{0.928}
\expandafter\gdef\csname odunum@n@thinking_vs_instruct.level.av.instruct.m5\endcsname{1\,061}
\expandafter\gdef\csname odunum@val@thinking_vs_instruct.level.av.thinking.avg\endcsname{0.599}
\expandafter\gdef\csname odunum@n@thinking_vs_instruct.level.av.thinking.avg\endcsname{1\,347}
\expandafter\gdef\csname odunum@val@thinking_vs_instruct.level.av.thinking.ftr\endcsname{0.829}
\expandafter\gdef\csname odunum@n@thinking_vs_instruct.level.av.thinking.ftr\endcsname{286}
\expandafter\gdef\csname odunum@ci@thinking_vs_instruct.level.av.thinking.ftr\endcsname{[0.781, 0.868]}
\expandafter\gdef\csname odunum@val@thinking_vs_instruct.level.av.thinking.m1\endcsname{0.582}
\expandafter\gdef\csname odunum@n@thinking_vs_instruct.level.av.thinking.m1\endcsname{1\,347}
\expandafter\gdef\csname odunum@val@thinking_vs_instruct.level.av.thinking.m2\endcsname{0.556}
\expandafter\gdef\csname odunum@n@thinking_vs_instruct.level.av.thinking.m2\endcsname{1\,061}
\expandafter\gdef\csname odunum@val@thinking_vs_instruct.level.av.thinking.m3\endcsname{0.470}
\expandafter\gdef\csname odunum@n@thinking_vs_instruct.level.av.thinking.m3\endcsname{1\,061}
\expandafter\gdef\csname odunum@val@thinking_vs_instruct.level.av.thinking.m4\endcsname{0.852}
\expandafter\gdef\csname odunum@n@thinking_vs_instruct.level.av.thinking.m4\endcsname{1\,061}
\expandafter\gdef\csname odunum@val@thinking_vs_instruct.level.av.thinking.m5\endcsname{0.921}
\expandafter\gdef\csname odunum@n@thinking_vs_instruct.level.av.thinking.m5\endcsname{1\,061}
\expandafter\gdef\csname odunum@val@thinking_vs_instruct.paired.all.ftr_delta\endcsname{-0.139}
\expandafter\gdef\csname odunum@n@thinking_vs_instruct.paired.all.ftr_delta\endcsname{447}
\expandafter\gdef\csname odunum@val@thinking_vs_instruct.paired.all.ftr_mcnemar_p\endcsname{5.8\ensuremath{\times 10^{-11}}\ensuremath{\times}}
\expandafter\gdef\csname odunum@n@thinking_vs_instruct.paired.all.ftr_mcnemar_p\endcsname{94}
\expandafter\gdef\csname odunum@val@thinking_vs_instruct.paired.all.m2\endcsname{0.024}
\expandafter\gdef\csname odunum@n@thinking_vs_instruct.paired.all.m2\endcsname{1\,631}
\expandafter\gdef\csname odunum@ci@thinking_vs_instruct.paired.all.m2\endcsname{[0.010, 0.037]}
\expandafter\gdef\csname odunum@val@thinking_vs_instruct.paired.all.m2.sign_p\endcsname{1.1\ensuremath{\times 10^{-3}}\ensuremath{\times}}
\expandafter\gdef\csname odunum@n@thinking_vs_instruct.paired.all.m2.sign_p\endcsname{759}
\expandafter\gdef\csname odunum@val@thinking_vs_instruct.paired.all.m3\endcsname{0.255}
\expandafter\gdef\csname odunum@n@thinking_vs_instruct.paired.all.m3\endcsname{1\,631}
\expandafter\gdef\csname odunum@ci@thinking_vs_instruct.paired.all.m3\endcsname{[0.238, 0.271]}
\expandafter\gdef\csname odunum@val@thinking_vs_instruct.paired.all.m3.sign_p\endcsname{4.1\ensuremath{\times 10^{-113}}\ensuremath{\times}}
\expandafter\gdef\csname odunum@n@thinking_vs_instruct.paired.all.m3.sign_p\endcsname{1\,452}
\expandafter\gdef\csname odunum@val@thinking_vs_instruct.paired.all.m4\endcsname{-0.023}
\expandafter\gdef\csname odunum@n@thinking_vs_instruct.paired.all.m4\endcsname{1\,631}
\expandafter\gdef\csname odunum@ci@thinking_vs_instruct.paired.all.m4\endcsname{[-0.035, -0.011]}
\expandafter\gdef\csname odunum@val@thinking_vs_instruct.paired.all.m4.sign_p\endcsname{5.7\ensuremath{\times 10^{-6}}\ensuremath{\times}}
\expandafter\gdef\csname odunum@n@thinking_vs_instruct.paired.all.m4.sign_p\endcsname{602}
\expandafter\gdef\csname odunum@val@thinking_vs_instruct.paired.all.m5\endcsname{-0.007}
\expandafter\gdef\csname odunum@n@thinking_vs_instruct.paired.all.m5\endcsname{1\,631}
\expandafter\gdef\csname odunum@ci@thinking_vs_instruct.paired.all.m5\endcsname{[-0.018, 0.004]}
\expandafter\gdef\csname odunum@val@thinking_vs_instruct.paired.all.m5.sign_p\endcsname{0.07\ensuremath{\times}}
\expandafter\gdef\csname odunum@n@thinking_vs_instruct.paired.all.m5.sign_p\endcsname{190}
\expandafter\gdef\csname odunum@val@thinking_vs_instruct.paired.ao.ftr_delta\endcsname{-0.199}
\expandafter\gdef\csname odunum@n@thinking_vs_instruct.paired.ao.ftr_delta\endcsname{161}
\expandafter\gdef\csname odunum@val@thinking_vs_instruct.paired.ao.ftr_mcnemar_p\endcsname{5.6\ensuremath{\times 10^{-6}}\ensuremath{\times}}
\expandafter\gdef\csname odunum@n@thinking_vs_instruct.paired.ao.ftr_mcnemar_p\endcsname{50}
\expandafter\gdef\csname odunum@val@thinking_vs_instruct.paired.ao.m2\endcsname{0.021}
\expandafter\gdef\csname odunum@n@thinking_vs_instruct.paired.ao.m2\endcsname{570}
\expandafter\gdef\csname odunum@ci@thinking_vs_instruct.paired.ao.m2\endcsname{[-0.002, 0.044]}
\expandafter\gdef\csname odunum@val@thinking_vs_instruct.paired.ao.m2.sign_p\endcsname{0.08\ensuremath{\times}}
\expandafter\gdef\csname odunum@n@thinking_vs_instruct.paired.ao.m2.sign_p\endcsname{249}
\expandafter\gdef\csname odunum@val@thinking_vs_instruct.paired.ao.m3\endcsname{0.300}
\expandafter\gdef\csname odunum@n@thinking_vs_instruct.paired.ao.m3\endcsname{570}
\expandafter\gdef\csname odunum@ci@thinking_vs_instruct.paired.ao.m3\endcsname{[0.269, 0.331]}
\expandafter\gdef\csname odunum@val@thinking_vs_instruct.paired.ao.m3.sign_p\endcsname{2.2\ensuremath{\times 10^{-36}}\ensuremath{\times}}
\expandafter\gdef\csname odunum@n@thinking_vs_instruct.paired.ao.m3.sign_p\endcsname{521}
\expandafter\gdef\csname odunum@val@thinking_vs_instruct.paired.ao.m4\endcsname{-0.029}
\expandafter\gdef\csname odunum@n@thinking_vs_instruct.paired.ao.m4\endcsname{570}
\expandafter\gdef\csname odunum@ci@thinking_vs_instruct.paired.ao.m4\endcsname{[-0.048, -0.010]}
\expandafter\gdef\csname odunum@val@thinking_vs_instruct.paired.ao.m4.sign_p\endcsname{3.6\ensuremath{\times 10^{-3}}\ensuremath{\times}}
\expandafter\gdef\csname odunum@n@thinking_vs_instruct.paired.ao.m4.sign_p\endcsname{209}
\expandafter\gdef\csname odunum@val@thinking_vs_instruct.paired.ao.m5\endcsname{-0.006}
\expandafter\gdef\csname odunum@n@thinking_vs_instruct.paired.ao.m5\endcsname{570}
\expandafter\gdef\csname odunum@ci@thinking_vs_instruct.paired.ao.m5\endcsname{[-0.024, 0.011]}
\expandafter\gdef\csname odunum@val@thinking_vs_instruct.paired.ao.m5.sign_p\endcsname{0.57\ensuremath{\times}}
\expandafter\gdef\csname odunum@n@thinking_vs_instruct.paired.ao.m5.sign_p\endcsname{49}
\expandafter\gdef\csname odunum@val@thinking_vs_instruct.paired.av.ftr_delta\endcsname{-0.105}
\expandafter\gdef\csname odunum@n@thinking_vs_instruct.paired.av.ftr_delta\endcsname{286}
\expandafter\gdef\csname odunum@val@thinking_vs_instruct.paired.av.ftr_mcnemar_p\endcsname{5.3\ensuremath{\times 10^{-6}}\ensuremath{\times}}
\expandafter\gdef\csname odunum@n@thinking_vs_instruct.paired.av.ftr_mcnemar_p\endcsname{44}
\expandafter\gdef\csname odunum@val@thinking_vs_instruct.paired.av.m2\endcsname{0.026}
\expandafter\gdef\csname odunum@n@thinking_vs_instruct.paired.av.m2\endcsname{1\,061}
\expandafter\gdef\csname odunum@ci@thinking_vs_instruct.paired.av.m2\endcsname{[0.008, 0.043]}
\expandafter\gdef\csname odunum@val@thinking_vs_instruct.paired.av.m2.sign_p\endcsname{0.01\ensuremath{\times}}
\expandafter\gdef\csname odunum@n@thinking_vs_instruct.paired.av.m2.sign_p\endcsname{510}
\expandafter\gdef\csname odunum@val@thinking_vs_instruct.paired.av.m3\endcsname{0.231}
\expandafter\gdef\csname odunum@n@thinking_vs_instruct.paired.av.m3\endcsname{1\,061}
\expandafter\gdef\csname odunum@ci@thinking_vs_instruct.paired.av.m3\endcsname{[0.211, 0.250]}
\expandafter\gdef\csname odunum@val@thinking_vs_instruct.paired.av.m3.sign_p\endcsname{1.2\ensuremath{\times 10^{-78}}\ensuremath{\times}}
\expandafter\gdef\csname odunum@n@thinking_vs_instruct.paired.av.m3.sign_p\endcsname{931}
\expandafter\gdef\csname odunum@val@thinking_vs_instruct.paired.av.m4\endcsname{-0.019}
\expandafter\gdef\csname odunum@n@thinking_vs_instruct.paired.av.m4\endcsname{1\,061}
\expandafter\gdef\csname odunum@ci@thinking_vs_instruct.paired.av.m4\endcsname{[-0.034, -0.005]}
\expandafter\gdef\csname odunum@val@thinking_vs_instruct.paired.av.m4.sign_p\endcsname{5.9\ensuremath{\times 10^{-4}}\ensuremath{\times}}
\expandafter\gdef\csname odunum@n@thinking_vs_instruct.paired.av.m4.sign_p\endcsname{393}
\expandafter\gdef\csname odunum@val@thinking_vs_instruct.paired.av.m5\endcsname{-0.007}
\expandafter\gdef\csname odunum@n@thinking_vs_instruct.paired.av.m5\endcsname{1\,061}
\expandafter\gdef\csname odunum@ci@thinking_vs_instruct.paired.av.m5\endcsname{[-0.021, 0.006]}
\expandafter\gdef\csname odunum@val@thinking_vs_instruct.paired.av.m5.sign_p\endcsname{0.09\ensuremath{\times}}
\expandafter\gdef\csname odunum@n@thinking_vs_instruct.paired.av.m5.sign_p\endcsname{141}
\expandafter\gdef\csname odunum@val@thinking_vs_instruct.pop.all.n\endcsname{2\,078}
\expandafter\gdef\csname odunum@n@thinking_vs_instruct.pop.all.n\endcsname{2\,078}
\expandafter\gdef\csname odunum@val@thinking_vs_instruct.pop.ao.n\endcsname{731}
\expandafter\gdef\csname odunum@n@thinking_vs_instruct.pop.ao.n\endcsname{2\,078}
\expandafter\gdef\csname odunum@val@thinking_vs_instruct.pop.av.n\endcsname{1\,347}
\expandafter\gdef\csname odunum@n@thinking_vs_instruct.pop.av.n\endcsname{2\,078}
\expandafter\gdef\csname odunum@val@thinking_vs_instruct.pop.only_instruct.n\endcsname{0}
\expandafter\gdef\csname odunum@n@thinking_vs_instruct.pop.only_instruct.n\endcsname{2\,078}
\expandafter\gdef\csname odunum@val@thinking_vs_instruct.pop.only_thinking.n\endcsname{0}
\expandafter\gdef\csname odunum@n@thinking_vs_instruct.pop.only_thinking.n\endcsname{2\,078}
\expandafter\gdef\csname odunum@val@thinking_vs_instruct.pop.shared.n\endcsname{2\,078}
\expandafter\gdef\csname odunum@n@thinking_vs_instruct.pop.shared.n\endcsname{2\,078}
\expandafter\gdef\csname odunum@val@thinking_vs_instruct.pop.shared_points.n\endcsname{5\,818}
\expandafter\gdef\csname odunum@n@thinking_vs_instruct.pop.shared_points.n\endcsname{5\,859}
}{}

\newcommand{\odu}{\textsc{Odu}}          % task/benchmark short name in small caps

\definecolor{odutabrow}{RGB}{246,246,247}
\definecolor{odutabavg}{RGB}{237,237,240}
\definecolor{odutabrule}{RGB}{210,210,214}
\newcommand{\best}[1]{{\bfseries\boldmath #1}}
\newcommand{\second}[1]{\underline{#1}}
\newcommand{\na}{--}
\newcommand{\odutabsetup}{\small\setlength{\tabcolsep}{4.5pt}\renewcommand{\arraystretch}{1.06}}
\newcommand{\odutabsetupwide}{\footnotesize\setlength{\tabcolsep}{3.2pt}\renewcommand{\arraystretch}{1.04}}
\newcommand{\odugrouprow}[2]{\multicolumn{#1}{@{}c@{}}{\itshape #2}\\[-1pt]}
\newcommand{\odutabnote}[1]{\vspace{2pt}\par\footnotesize\textit{Note.}~#1}
\definecolor{SciGreen}{RGB}{44,160,44}
\definecolor{SciRed}{RGB}{214,39,40}
\definecolor{SciOrange}{RGB}{230,159,0}
\DeclareRobustCommand{\cmark}{\textcolor{SciGreen}{\ding{51}}}
\DeclareRobustCommand{\pmark}{\textcolor{SciOrange}{\LEFTcircle}}
\DeclareRobustCommand{\xmark}{\textcolor{SciRed}{\ding{55}}}

\lstdefinestyle{json}{
  basicstyle=\ttfamily\scriptsize,
  breaklines=true,
  columns=fullflexible,
  showstringspaces=false,
  frame=single,
  framesep=4pt,
  rulecolor=\color{black!30},
  keepspaces=true,
}

\definecolor{oduintent}{RGB}{27,108,168}
\definecolor{oducontext}{RGB}{176,84,26}
\definecolor{odufigbg}{RGB}{250,250,248}
\definecolor{odufigrule}{RGB}{217,217,214}
\definecolor{odufigaccent}{RGB}{63,111,159}
\lstdefinestyle{odurecord}{
  basicstyle=\ttfamily\scriptsize,
  breaklines=false,
  columns=fullflexible,
  showstringspaces=false,
  frame=none,
  aboveskip=2pt,
  belowskip=2pt,
  keepspaces=true,
  literate={\ }{{\ }}1,
  classoffset=0,
  morekeywords={structured_intent},
  keywordstyle=\color{oduintent}\bfseries,
  classoffset=1,
  morekeywords={required_context,key_points},
  keywordstyle=\color{oducontext}\bfseries,
  classoffset=0,
}

\title{Omni Demand Understanding: A Benchmark for Contextual User-Intent Inference in Multimodal Interaction}

\newcommand{\oduauthor}[2]{\mbox{#1\textsuperscript{#2}}}
\newcommand{\authorrow}[1]{\makebox[\linewidth][s]{#1}\par}

\ifarxiv
  \iclrfinalcopy
  \makeatletter
  \renewcommand{\@maketitle}{%
    \vbox{\hsize\textwidth
      {\LARGE\scshape\@title\par}%
      \vskip 12pt
      {\normalfont\@author\par}%
      \vskip 0.3in minus 0.1in
    }%
  }
  \AtBeginDocument{\hypersetup{pdftitle={\@title},pdfauthor={\paperAuthorRows}}}
  \makeatother
\else
  \hypersetup{pdfauthor={}}
\fi

\newcommand{\paperAuthorRows}{%
  \authorrow{%
    \oduauthor{Qi Chen}{1,2,3,*,\textsection}\hfill
    \oduauthor{Yunfei Chu}{3,*}\hfill
    \oduauthor{Haolin He}{3,4,*,\textsection}\hfill
    \oduauthor{Yifan Yang}{1,3,\textsection}\hfill
    \oduauthor{Zihan Liu}{3,\textsection}\hfill
    \oduauthor{Yuxuan Wang}{3}%
  }%
  \authorrow{%
    \oduauthor{Ziyang Ma}{1,2}\hfill
    \oduauthor{Ruiyang Xu}{1,3,\textsection}\hfill
    \oduauthor{Meng Gao}{3,5,\textsection}\hfill
    \oduauthor{Yinsong Yan}{3,6,\textsection}\hfill
    \oduauthor{Ling Wang}{3,6,\textsection}\hfill
    \oduauthor{Hui Wang}{3,7,\textsection}%
  }%
  \authorrow{%
    \oduauthor{Wen Huang}{8}\hfill
    \oduauthor{Yiheng Chen}{1,3,\textsection}\hfill
    \oduauthor{Guanrou Yang}{1,2}\hfill
    \oduauthor{Qiuqiang Kong}{4}\hfill
    \oduauthor{Jin Xu}{3,\textdagger}\hfill
    \oduauthor{Xie Chen}{1,2,\textdagger}%
  }%
}

\newcommand{\paperAuthorBlock}{%
  \begin{minipage}{\textwidth}
  \centering
  \setlength{\parskip}{0pt}
  \normalsize\bfseries
  \paperAuthorRows
  \vspace{6pt}
  \normalfont\small
  \textsuperscript{1}Shanghai Jiao Tong University\quad
  \textsuperscript{2}Shanghai Innovation Institute\quad
  \textsuperscript{3}Alibaba Token Hub, Alibaba Group\par
  \textsuperscript{4}The Chinese University of Hong Kong\quad
  \textsuperscript{5}Tsinghua University\par
  \textsuperscript{6}Hong Kong Polytechnic University\quad
  \textsuperscript{7}Nankai University\quad
  \textsuperscript{8}Johns Hopkins University\par
  
  \vspace{5pt}
  {\fontsize{8.5}{10}\selectfont\mbox{%
    \textsuperscript{*}Equal contribution.\quad
    \textsuperscript{\textdagger}Corresponding author}%
    \par
    \mbox{\textsuperscript{\textsection}Work done during an internship at Alibaba Token Hub, Alibaba Group.}%
    \par}
  \end{minipage}%
}
\author{\paperAuthorBlock}

\begin{document}

% Keep title words intact without changing the manuscript title text.
\begingroup
\raggedright
\hyphenpenalty=10000
\exhyphenpenalty=10000
\maketitle
\endgroup

% Do not stretch paragraph/heading gaps when an indivisible table starts a new page.
\raggedbottom

\begin{abstract}
Natural audio--visual interaction is emerging as an important interface for AI assistants,
allowing users to communicate through speech and vision rather than carefully composed text prompts. However,
existing benchmarks of interactive capabilities still focus primarily on response quality,
leaving a more fundamental question underexplored:
\emph{can a model correctly infer the user's underlying demand from a complex multimodal interaction?}
Real-world user demands are often underspecified in speech and must be inferred from multimodal cues and dialogue history.
This inference is further complicated by ambiguous or disfluent expression and noisy acoustic environments.
Conversely, request-like speech may not constitute a demand to the assistant, leading to false triggers.
% task
We establish \textbf{Omni Demand Understanding (ODU)} as a distinct multimodal contextual inference problem:
given an interaction stream, a model must detect whether a user demand is present and infer the user's intent from multimodal and conversational context.
ODU evaluates this capability along five dimensions, covering both single-turn and multi-turn interactions.
% benchmark construction
We construct ODU-Bench using a challenge-driven taxonomy, taxonomy-guided agentic video generation, and human-recorded interactions, followed by media-grounded annotation and human verification.
% results
We evaluate \num{benchmark_main.complete.n_models} native audio and audio--visual MLLMs.
Even the strongest, Gemini~\numlit{3.1}~Pro, recovers
only \numpct{keypoint_types.cov.gemini.av.context_all}\% of the key information that must be inferred
from visual, acoustic, or conversational context.
Moreover, \numlit{11} of the \num{benchmark_main.complete.n_models} models exhibit false-trigger rates above \numlit{50\%} on non-demand scenarios.
These results reveal a systematic capability gap in current MLLMs' ability to infer contextual user demands.
We hope ODU can establish the evaluation of a previously underexplored yet essential capability in multimodal interaction: correctly understanding user demands before generating an appropriate response.
\ifarxiv\else
See demos at \url{https://odubench.github.io}
\fi
% Our human evaluation further suggests that demand understanding is an important but under-diagnosed factor in response quality:
% in a pairwise comparison, a strong omni model supplied with the ground-truth user demand is preferred in
% \num{human_ab.oracle_better} of the \numn{human_ab.discordant_oracle_frac} scenes a rater judged different,
% against \num{human_ab.blind_better} for the media-only arm
% (\num{human_ab.tie}/\numn{human_ab.tie_frac} ties).

\end{abstract}

\ifarxiv
\begingroup
\raggedright
% Match the quote environment used by the abstract.
\setlength{\leftskip}{\leftmargini}
\addtolength{\rightskip}{\leftmargini}
\small
\noindent\textbf{Demo:} \url{https://odubench.github.io}\\
% Placeholder: replace with the public code and data URL.
\textbf{Code and data:} \url{https://huggingface.co/datasets/qc316/odubench}
\par
\endgroup
\fi

% Full-width teaser: encountered after the page-1 title/abstract so its [!t]
% float is placed at the top of page 2. The label is cited from the introduction.
% fig:challenges -- full-width teaser placed at the top of page 2.
% Hand-edit figures/src/teaser.drawio; ./compile_teaser_drawio.sh exports the PDF.
\begin{figure}[!t]
  \centering
  \includegraphics[width=\linewidth]{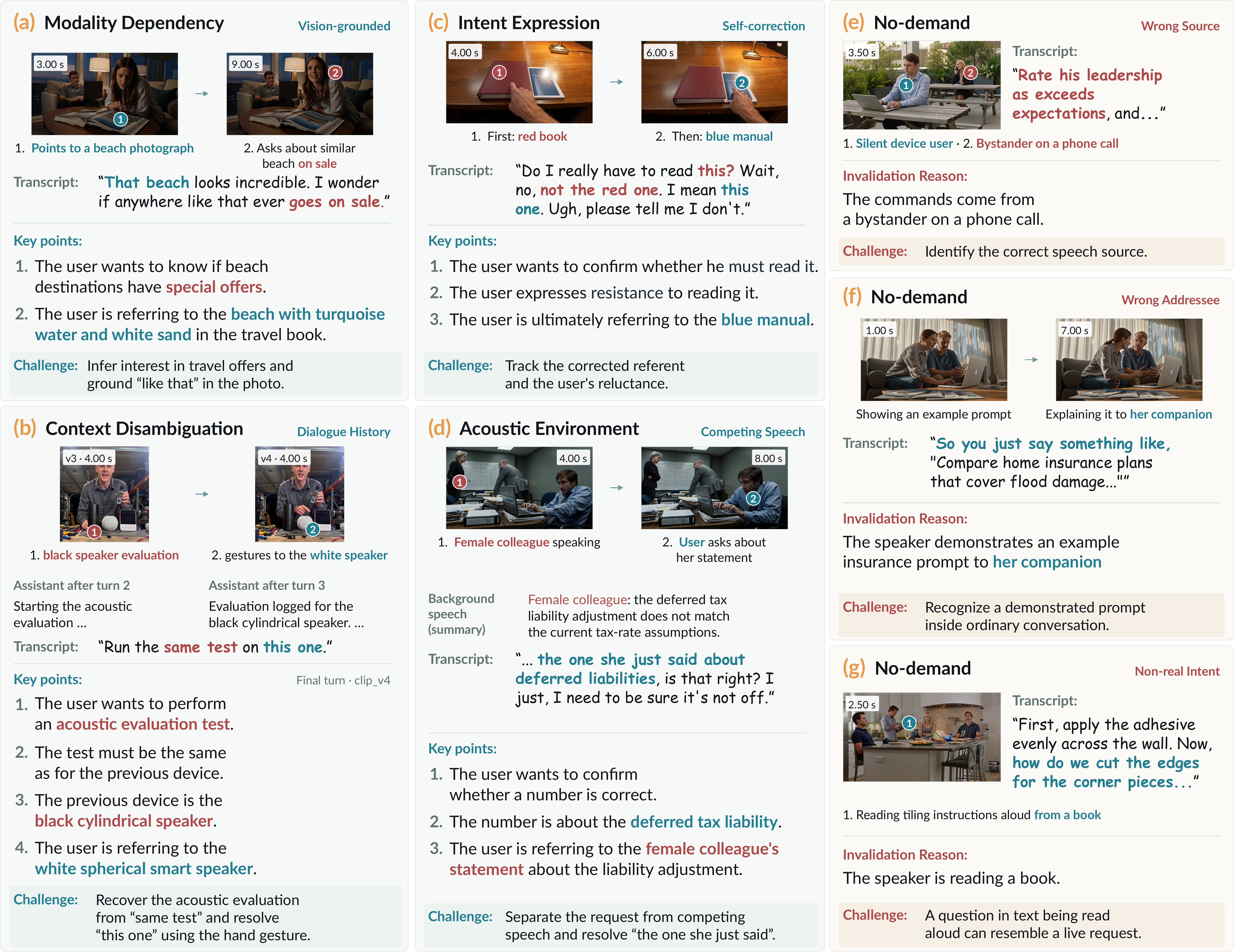}
  \caption{\textbf{Representative challenging scenarios in ODU.}
  Demand-bearing cases (a–d) require integrating visual, acoustic, and conversational context while handling challenging expression forms and acoustic conditions, 
  while no-demand cases (e–g) contain demand-like expressions that should not trigger responses.}
  \label{fig:challenges}
\end{figure}

% ---- aggregated open items while drafting (remove for camera-ready) ----------
% \listoftodos  % re-enable if you add \todo/\authornote margin notes.

% ---- body: skill maintains this list in outline order ------------------------
\section{Introduction}
\label{sec:introduction}

Multimodal large language models (MLLMs) that jointly perceive audio, vision, and language are increasingly serving as
conversational assistants~\citep{gemini3_1pro, seed2026seed2, team2026qwen3, xu2025qwen3, Qwen2.5-Omni,
cui2026minicpmo45realtimefullduplex, hurst2024gpt, ai2025ming, deshmukh2026nemotron, tang2025video}. 
Yet existing benchmarks on multimodal interactions primarily evaluate response quality, 
implicitly assuming that the user demand behind a multimodal query has already been correctly identified and understood~\citep{wang2025omnimmi, selvakumar2025multivox, lu2026omniinteract, zhao2026omnipro}. 
This assumption is fragile because human expression is inherently underspecified. 
By the Principle of Least Effort~\citep{zipf1949}, 
people tend to avoid unnecessary explicitness, 
leaving parts of their intended meaning to be recovered from shared multimodal and conversational context.
As shown in Fig.~\ref{fig:challenges} (b), expressions such as ``the same test" and ``this one"
require conversation history and visual cues from user gestures to resolve.
Conversely, Fig.~\ref{fig:challenges} (e--g) show that demand-like expressions do not necessarily imply the existence of actual user demands.
Recovering a demand is therefore a contextual reasoning task that requires integrating multimodal and conversational evidence,
not merely transcribing the utterance. 
However, existing benchmarks largely treat user queries as self-contained text or speech, 
overlooking the multimodal and conversational context required to infer them and leaving this capability under-evaluated.

To diagnose this capability systematically, we introduce \textbf{Omni Demand Understanding (ODU)}, a task which
makes demand understanding an explicit prediction target rather than an implicit precondition of response
generation. Given an audio or audio--visual interaction, ODU evaluates a model along five dimensions:
whether a valid demand is present, when it occurs, the transcript of the demand-bearing speech, the user's
intent as resolved from multimodal and conversational context, and the user profile. This
diagnosis separates perception and localization from contextual user-intent inference: a model must not only
recover the utterance, but also infer the user's intent and the response-relevant information left implicit.
Semantic recovery is evaluated against atomic key points, while no-demand scenes test whether request-like
signals could trigger the assistant.

To build ODU-Bench, we start from demand-understanding challenges in everyday interaction and organize them
into a challenge-driven taxonomy. Guided by sampled taxonomy targets, an agentic pipeline constructs
scenarios coarse-to-fine, progressing from demand semantics or invalidation conditions to causal interactions,
dialogue, and temporal realization. After challenge-validity and transcript-sufficiency screening, accepted
scripts guide the production of synthetic interactions.
We also recruited human actors to perform daily-life interactions, 
providing the benchmark with a complementary source of behaviorally and acoustically realistic data. 
For both synthetic and human-recorded interactions, demand annotations are reconstructed from the realized media.
Every scene and its annotation then undergo careful human  verification. Together,
these steps yield \num{release_stats.main.n_scenes} diverse and challenging demand
and no-demand scenes. Across a broad panel
of native multimodal models, even Gemini~\numlit{3.1}~Pro, the strongest system on our aggregate
score, falsely triggers on \numpct{benchmark_main.ftr.gemini.av}\% of no-demand scenes. Evidence-channel
analysis shows that it recovers \numpct{keypoint_types.cov.gemini.av.intent}\% of key points supported by the
spoken request but only \numpct{keypoint_types.cov.gemini.av.context_all}\% of those requiring
multimodal or dialogue context. This gap confirms that ODU cannot be reduced to transcription: models must
select and bind evidence across the interaction to understand the user's intent. 
In summary, our contributions are as follows:

\begin{enumerate}[leftmargin=1.4em,itemsep=2pt,topsep=2pt]

  \item \textbf{A new problem formulation.}
  We establish omni demand understanding as a distinct multimodal contextual inference problem, making it an explicit evaluation target rather than leaving it implicit in response-generation performance.

  \item \textbf{A challenging benchmark.}
  We construct a benchmark covering diverse demand and no-demand scenarios, combining taxonomy-guided agentic generation with human-recorded interactions.
  
  \item \textbf{A systematic capability gap.}
  We show that current MLLMs systematically struggle with contextual user-intent inference, particularly when it requires multimodal and conversational evidence beyond the explicit spoken request.

\end{enumerate}

\section{Related Work}
\label{sec:related_work}

% GENERATED by scripts/build_benchmark_comparison.py. DO NOT EDIT.
\begin{table}[!b]
\centering
\caption{Benchmark comparison. Visual and acoustic cues refer to intent inference.}
\label{tab:benchmark_comparison}
\small
\newlength{\oduTableVisualHeadWidth}\settowidth{\oduTableVisualHeadWidth}{\textbf{Visual Cues}}
\newlength{\oduTableAcousticHeadWidth}\settowidth{\oduTableAcousticHeadWidth}{\textbf{Acoustic Cues}}
\setlength{\tabcolsep}{4pt}
\renewcommand{\arraystretch}{1.18}
\resizebox{\textwidth}{!}{%
\begin{tabular}{lcccccc}
\toprule
\textbf{Benchmark} & \shortstack{\textbf{Demand}\\\textbf{Presence}} & \shortstack{\textbf{Intent Prediction}\\\textbf{Format}} & \shortstack{\textbf{Spoken Utterance}\\\textbf{Input}} & \shortstack{\textbf{Multi-Turn Dialogue}\\\textbf{Input}} & \makebox[\oduTableVisualHeadWidth]{\shortstack{\textbf{Visual}\\\textbf{Cues}}} & \makebox[\oduTableAcousticHeadWidth]{\shortstack{\textbf{Acoustic}\\\textbf{Cues}}} \\
\midrule
\multicolumn{7}{l}{\textit{Multimodal interaction benchmarks}}\\
OmniMMI~\citep{wang2025omnimmi} & \pmark & \xmark & \pmark & \cmark & \pmark & \xmark \\
MultiVox~\citep{selvakumar2025multivox} & \xmark & \xmark & \cmark & \xmark & \cmark & \cmark \\
OmniInteract~\citep{lu2026omniinteract} & \pmark & \xmark & \cmark & \cmark & \pmark & \xmark \\
Full-Duplex-Bench-v2~\citep{lin2026fullduplexv2} & \pmark & \xmark & \cmark & \cmark & \xmark & \xmark \\
\midrule
\multicolumn{7}{l}{\textit{Intent and goal understanding benchmarks}}\\
MIntRec2.0~\citep{zhang2024mintrec2} & \xmark & Label & \cmark & \pmark & \cmark & \cmark \\
WAGIBench~\citep{veerabadran2026benchmarking} & \xmark & Open-ended & \xmark & \xmark & \cmark & \xmark \\
GUIDE~\citep{yang2026guide} & \cmark & MCQ & \xmark & \xmark & \cmark & \xmark \\
EgoIntrospect~\citep{wang2026egointrospect} & \pmark & MCQ & \pmark & \xmark & \pmark & \xmark \\
\midrule
\textbf{ODU} & \cmark & Open-ended & \cmark & \cmark & \cmark & \cmark \\
\bottomrule
\end{tabular}}
\par\vspace{4pt}
\begin{minipage}{\textwidth}\footnotesize
\cmark: explicitly evaluated; \pmark: partial or indirect support; \xmark: not established. MCQ: multiple-choice question. In input columns, \cmark\ means that the input is provided.
\end{minipage}
\end{table}

\paragraph{Multimodal interaction.}
Research on spoken and multimodal interaction examines response quality,
turn-taking, and the coordination of listening and speaking.
Full-Duplex-Bench evaluates pause handling, backchanneling, and
interruptions~\citep{lin2025full},
while MTR-DuplexBench examines multi-round dialogue quality and instruction
following~\citep{he2026mtr}.
Full-Duplex-Bench-v2 further evaluates multi-turn turn-taking and instruction following with an automated examiner~\citep{lin2026fullduplexv2}.
MultiVox evaluates responses grounded in paralinguistic and visual
cues~\citep{selvakumar2025multivox}, and VideoFDB benchmarks nonverbal behavior
in full-duplex audio--visual
conversations~\citep{mazumdar2026videofdb}.
OmniMMI studies streaming understanding and multi-turn
dependencies~\citep{wang2025omnimmi}; OmniInteract tests trigger timing,
interruptions, and nested exchanges~\citep{lu2026omniinteract}; and OmniPro
emphasizes modality necessity in proactive streaming
evaluation~\citep{zhao2026omnipro}.
ODU complements these benchmarks by evaluating demand understanding separately from
response quality and timing.

\paragraph{Multimodal intent and contextual goal understanding.}
Research on multimodal intent and goal understanding examines how language,
behavior, and surrounding context reveal users' intentions. MIntRec2.0 evaluates
conversational intent classification using textual, acoustic, and visual
cues~\citep{zhang2024mintrec2}, with substantial textual bias identified in subsequent
analyses~\citep{mullick2025text}. SIMMC~2.0 studies multimodal disambiguation and
coreference in shopping
dialogues~\citep{kottur2021simmc}.
GUIDE studies assistance needs from GUI activity~\citep{yang2026guide}; WAGIBench
infers unexpressed goals from egocentric context~\citep{veerabadran2026benchmarking};
and EgoIntrospect evaluates request recovery with the spoken request
withheld~\citep{wang2026egointrospect}.
ODU complements these studies with joint evaluation of demand presence and open-ended
demand semantics in challenge-driven daily-life human--machine interactions, created through
media synthesis and realistic human-recorded interactions.
Table~\ref{tab:benchmark_comparison} compares \odu{} with representative benchmarks.

\section{Task Formulation and Evaluation}
\label{sec:task_evaluation}

\subsection{Problem setup and prediction targets}
\label{sec:task_def}

A \emph{demand} is the outcome a user wants an assistant to achieve, including response-relevant
objects, constraints, and trigger conditions. \odu{} evaluates the final user turn in an audio or audio--visual
interaction, using the preceding interaction as context. Earlier assistant replies are supplied as text.
The task asks \textbf{five core questions}: whether a valid demand exists; what the user intends and which
contextual information is required to resolve it; when the demand occurs; what the user says; and who expresses
it. The first two assess contextual understanding and reasoning over multimodal and conversational evidence.
The latter three primarily assess perception through temporal localization, transcription, and user profile prediction.
Together, they cover the perceptual foundations and contextual reasoning required for demand understanding.
The output schema is shown in App.~\ref{app:evaluation}.

\subsection{Key-point-based semantic evaluation}
\label{sec:keypoints}

We represent demand understanding with two open-ended fields: \emph{structured intent} for the desired
outcome and \emph{required context} for the multimodal or dialogue context needed to resolve it.
We evaluate these fields against atomic \emph{key points} extracted from ground-truth annotations;
omitting or misstating any point could change an appropriate response to the user demand.
Each key point is labeled by evidence source for channel-level diagnosis
(\S\ref{sec:where_models_miss}). All key points are grounded in the media and verified by humans.
An LLM judge checks whether the concatenated predicted structured intent and required context
cover each reference key point, accepting semantically equivalent wording.
We audit key-point recoverability from source annotations and the stability of model conclusions
across judges in App.~\ref{app:judge_stability}.

\subsection{Evaluation metrics}
\label{sec:protocol}
% GENERATED-adjacent: prose is hand-written, every number is a \num{} key from
% results/benchmark_main.json. Weights come from experiments.evaluate.WEIGHTS.
\begin{table}[!htbp]
\centering
\caption{Evaluation dimensions and their contribution to the overall score.}
\label{tab:metrics}
\odutabsetup
\begin{tabularx}{\linewidth}{@{}l l X c@{}}
\toprule
 & \textbf{Output measured} & \textbf{Metric} & \textbf{Weight} \\
\midrule
M1 & demand present & binary macro-F1 over demand and no-demand scenes
   & \num{benchmark_main.weight.m1} \\
M2 & structured intent, required context & key-point hit-rate
   & \num{benchmark_main.weight.m2} \\
M3 & demand span & temporal IoU
   & \num{benchmark_main.weight.m3} \\
M4 & transcript & $\max(0,1-\text{adaptive CER/WER})$
   & \num{benchmark_main.weight.m4} \\
M5 & user profile & mean per-field accuracy
   & \num{benchmark_main.weight.m5} \\
\bottomrule
\end{tabularx}
\end{table}

Table~\ref{tab:metrics} summarizes M1--M5. M1 uses demand and no-demand scenes;
M2--M5 use only demand-bearing scenes. FTR is the fraction of no-demand scenes that trigger
the assistant. The overall score is a weighted average of M1--M5, with the largest weight
on M2 to prioritize semantic recovery. App.~\ref{app:evaluation} gives full definitions and aggregation details.

% Do not let the task example float into Benchmark Construction.
\FloatBarrier

\section{Benchmark Construction}
\label{sec:benchmark_construction}

\begin{figure}[!t]
    \centering
    \includegraphics[width=\linewidth]{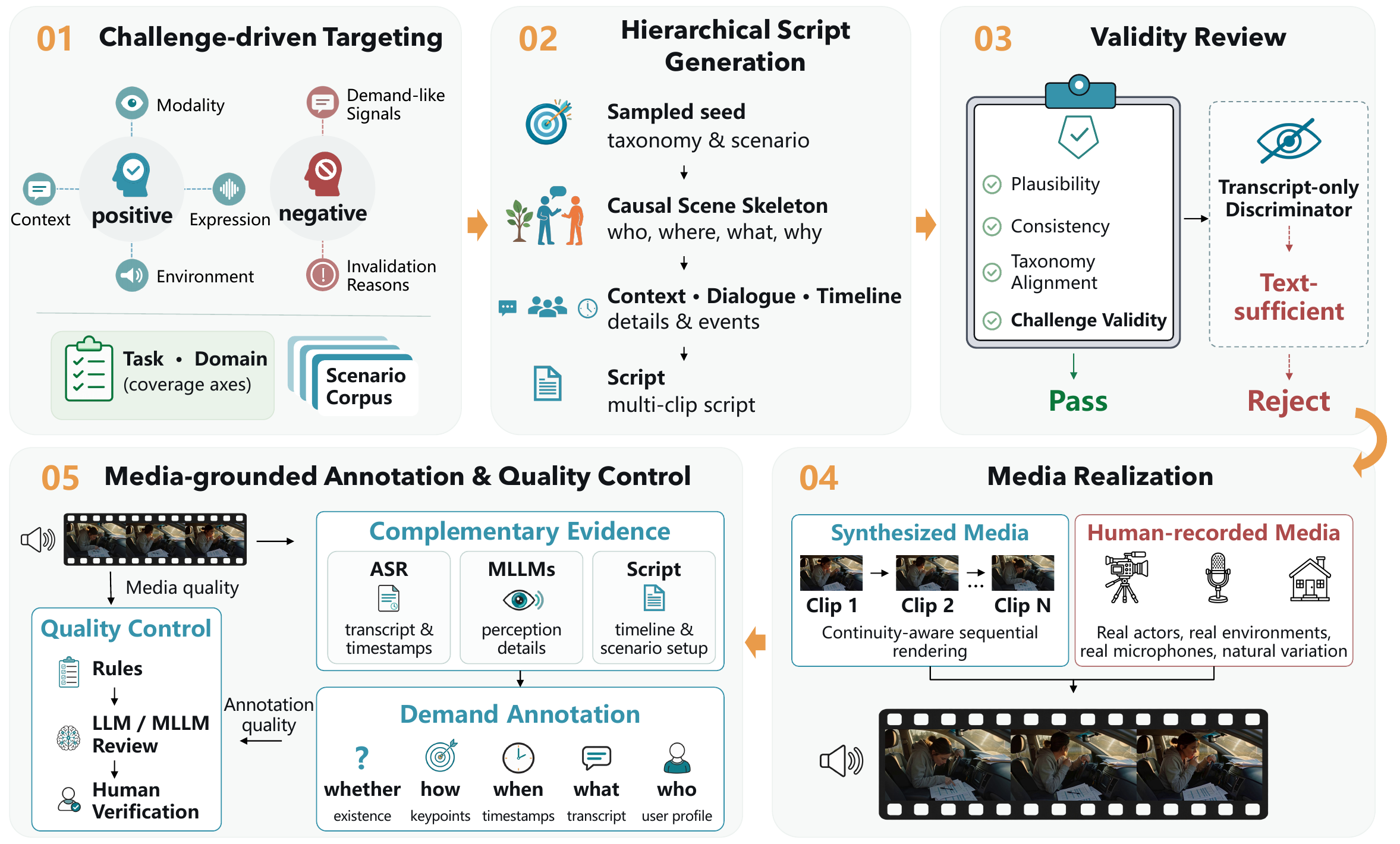}
    \caption{\textbf{Overview of the ODU-Bench construction pipeline.} Challenge-driven targets are expanded into
    coarse-to-fine scripts, screened for plausibility and challenge validity, and realized as synthesized or
    human-recorded interactions. Annotations are then reconstructed from complementary media evidence
    and verified through automatic checks and human review.}
    \label{fig:benchmark_pipeline}
\end{figure}

We construct ODU-Bench through a challenge-driven taxonomy and seed corpus
(\S\ref{sec:taxonomy}), agentic scenario generation (\S\ref{sec:generation}), and media-grounded
annotation with quality control (\S\ref{sec:annotation_qc}). Fig.~\ref{fig:benchmark_pipeline}
summarizes the pipeline;
App.~\ref{app:generation} provides detailed construction and human verification procedures.
The constructed benchmark contains both audio-only and audio-visual modalities.

\subsection{Challenge-driven taxonomy and seed corpus}
\label{sec:taxonomy}
% =============================================================================
%  figures/taxonomy_coverage.tex -- 3-up taxonomy coverage float.
%
%  The approved panels are regenerated into paper_overleaf_assets/ by
%      python -m data_analysis.plot_taxonomy_quadrant_pie
%      python -m data_analysis.plot_negative_pie
%      python -m data_analysis.plot_taxonomy_wordcloud
%  The two pies share a square canvas and padding ratio, so both rings come out
%  at the SAME diameter when included at the same width. If either pie is ever
%  re-plotted, check that the two PDFs still have equal page sizes -- differing
%  canvases are what made one ring ~9% smaller than the other before 2026-09-03.
%
%  The fixed-height minipage inside each panel keeps the three images in boxes of
%  identical height, so the landscape word cloud stays vertically centered against
%  the two pies AND the three subcaptions still start on the same line.
% =============================================================================

\begin{figure}[!htb]
  \centering
  % Scoped to this float: subcaptions only. Top-level caption layout stays as the
  % ICLR class sets it.
  \captionsetup[sub]{font=footnotesize,justification=raggedright,singlelinecheck=false,skip=4pt}
  \begin{subfigure}[t]{0.31\linewidth}
    \centering
    \begin{minipage}[c][\linewidth][c]{\linewidth}
      \centering
      \includegraphics[width=\linewidth]{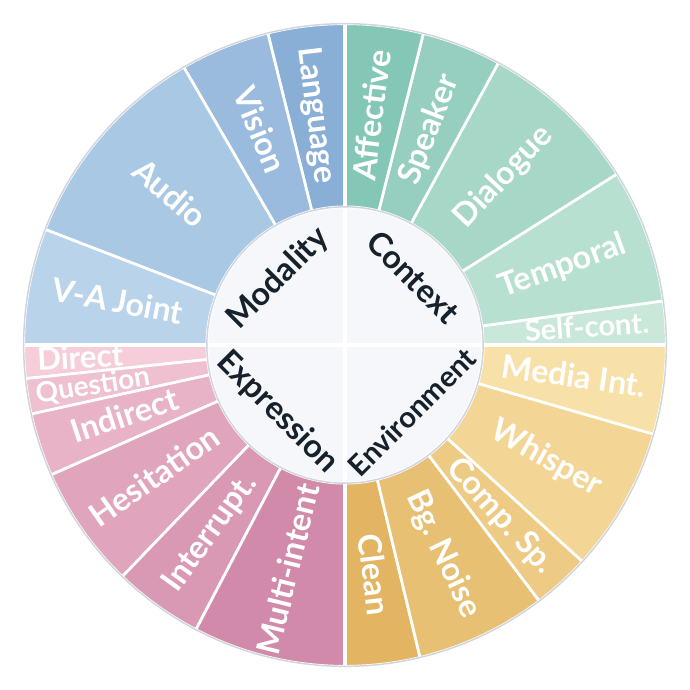}
    \end{minipage}
    \caption{Distribution of positive samples by taxonomy axis.}
    \label{fig:taxonomy_axes_1_4}
  \end{subfigure}\hfill
  \begin{subfigure}[t]{0.31\linewidth}
    \centering
    \begin{minipage}[c][\linewidth][c]{\linewidth}
      \centering
      \includegraphics[width=\linewidth]{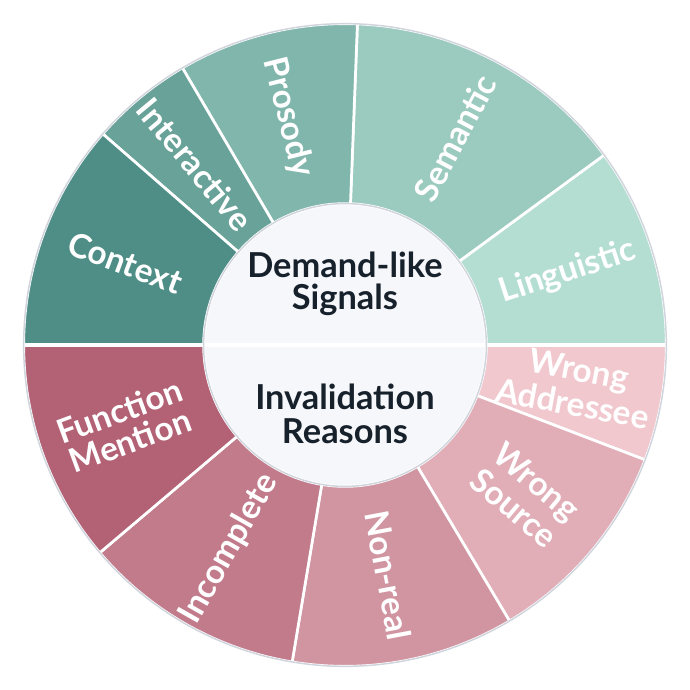}
    \end{minipage}
    \caption{Distribution of negative samples by signal type and invalidation pattern.}
    \label{fig:taxonomy_negative}
  \end{subfigure}\hfill
  \begin{subfigure}[t]{0.31\linewidth}
    \centering
    \begin{minipage}[c][\linewidth][c]{\linewidth}
      \centering
      \includegraphics[width=\linewidth]{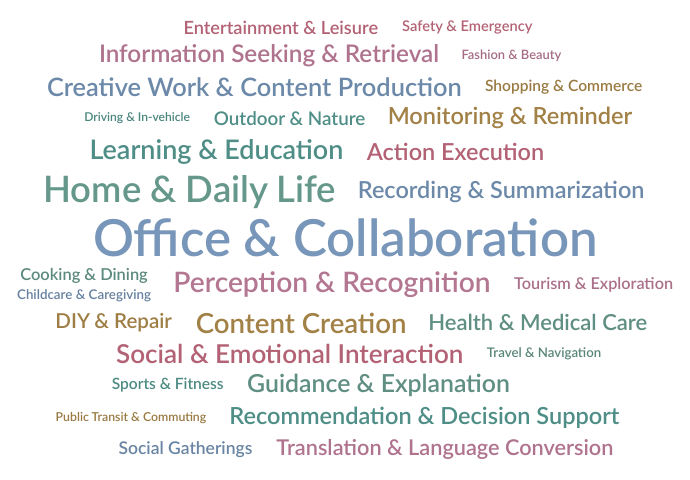}
    \end{minipage}
    \caption{Coverage across task and scenario categories.}
    \label{fig:taxonomy_axes_5_6}
  \end{subfigure}
  \caption{\textbf{Taxonomy coverage in ODU-Bench.} (a) and (b) show positive and negative sample distributions across their taxonomy axes; segment angles are proportional to sample counts within each axis. (c) shows task and scenario coverage, with word size reflecting frequency.}
  \label{fig:taxonomy_coverage}
\end{figure}

The core challenge of user demand understanding lies in how demands are expressed and how the surrounding context shapes their interpretation.
Real-world demands may depend on visual or acoustic context, remain ambiguous without prior interaction, be expressed implicitly, or occur under distracting environmental conditions.
In contrast, task semantics and scenario domains primarily characterize what users want to accomplish and where interactions occur, rather than what makes the demands difficult to understand.
We therefore organize scenario generation around sources of demand-understanding difficulty, while separately characterizing task and domain coverage.

We define a six-axis taxonomy for positive-demand scenes: four independently composable challenge
axes for generation and two descriptive axes for task and domain coverage. No-demand scenes use
a separate taxonomy of demand-like signal types and invalidation reasons.
Fig.~\ref{fig:taxonomy_coverage} shows the taxonomy and resulting distribution;
App.~\ref{app:taxonomy} provides the definitions.

We combine seeds from a corpus of everyday contexts and user needs with sampled taxonomy targets
to vary the setting of each challenge. This separates \emph{challenge coverage} from \emph{scenario diversity}.

\subsection{Agentic scenario generation}
\label{sec:generation}

We instantiate each scenario from a sampled taxonomy target and scenario seed through three stages:
hierarchical script generation, joint plausibility and challenge-validity review, and continuity-aware
rendering. Script generation with
Qwen3.7-Max~\citep{qwen3_7_max} proceeds coarse-to-fine: we first define a valid demand or an invalidating
condition, build a causal scenario skeleton, and then add contextual, linguistic, and temporal details.
Separating demand semantics from their realization supports controlled generation of recoverable demands
and demand-like cues that require contextual evidence to disambiguate.

The script reviewer checks plausibility, causal coherence, taxonomy alignment, and whether the intended
demand is recoverable from the evidence specified in the script. Its text-only discriminator sees only the
scripted user utterance, with dialogue history, speaker information, and media evidence withheld. For scenarios
designed to require contextual evidence, we reject positives whose intent is already recoverable from that
utterance and negatives whose no-demand status is textually obvious. This is a script-stage screening
criterion, not a test of exclusive channel dependence in the realized media.

Accepted scripts specify characters, environments, dialogue, and observable events along a timeline.
We produce synthetic interactions in both modalities from these scripts, retaining only the audio track
for audio-only scenes. Multi-clip rendering proceeds sequentially, conditioning later clips on earlier
videos to preserve character and environmental continuity.
Automatic review and human inspection assess perceptual quality, scenario fidelity, and
cross-clip coherence where applicable.
We also use accepted scripts as performance guides for human-recorded interactions. These preserve
the taxonomy targets and interaction structure while introducing real actors, environments, microphones,
timing, and speech variation. Both media sources follow the same media-grounded annotation and
quality-control pipeline.

\subsection{Media-grounded annotation and quality control}
\label{sec:annotation_qc}

Rendered and human-recorded media may differ from scripts in scene details, speech, timing,
and demand boundaries. We therefore derive ground-truth annotations from the media,
using scripts as structural guides. ASR~\citep{an2025funasrtechnicalreport} provides
transcripts and word-level timestamps; multiple MLLMs~\citep{team2026qwen3,gemini3_1pro} observe
visual and acoustic events, speaker relations, and interaction context. We reconcile these observations
with the script to determine demand presence and derive the annotations.

Rule-based checks and LLM/MLLM review assess consistency between scripts and media,
transcript and timestamp alignment, schema validity, agreement between demand labels and segments,
context support, and key-point grounding. Unsupported key points are removed
or rewritten. Human reviewers reject low-quality samples and retain generated scenes only when all applicable fields
pass review; experts correct annotation errors in human-recorded interactions. The main split contains
\num{release_stats.main.n_generated} synthetic scenarios and \num{release_stats.main.n_recorded}
human-recorded interactions performed by 30 actors.

\FloatBarrier
\section{Experiments}
\label{sec:experiments}

\subsection{Experimental setup}

\begingroup
\raggedright
We evaluate \num{benchmark_main.complete.n_models} native configurations in Table~\ref{tab:main_results},
with \num{benchmark_overview.native.n_models.av} evaluated on audio--visual scenes and
\num{benchmark_overview.native.n_models.ao} on audio-only scenes, using the same task instruction and
modality-specific output schema.
GPT-Realtime-\numlit{2}~\citep{openai2026gptrealtime2} and
\mbox{Kimi-Audio~\citep{kimiteam2025kimiaudiotechnicalreport}} accept audio only; the remaining systems are
evaluated on both audio-only and audio--visual scenes. We report the modalities separately.
For multi-turn interactions, only the final user clip is scored (\S\ref{sec:task_def}).
As a strong text-only baseline, GPT-\numlit{5.4}~\citep{openai2026gpt54} receives ASR transcripts,
sentence timings, and prior-turn assistant reply text (App.~\ref{app:text_only_baseline}). It remains unranked because its input interface differs from
that of the native systems. We choose Qwen3.6-Flash~\citep{qwen3_6_flash} as the LLM judge, and analyze the stability of results across different LLM judges in App.~\ref{app:judge_stability}.
\par\endgroup

% =============================================================================
%  tables/main_results.tex -- one row per model, with AV and AO results side by side.
%
%  NUMBERS CONTRACT: every numeric cell is \num{} from results/benchmark_main.json
%  via `make numbers`. Regenerate that file with
%      python -m experiments.analysis.benchmark_main
%
%  \best{} marks the best cell and \second{} the second-best cell within each
%  modality and metric column among systems meeting the completeness threshold.
%  Recheck after any re-score. FTR is ranked lower-is-better; all other columns
%  higher-is-better.
% =============================================================================

% Preprints prefer the source position; anonymous review retains the top float.
\ifarxiv
\begin{table}[!htbp]
\else
\begin{table}[t]
\fi
\centering
\caption{%
\textbf{Evaluation results on the ODU task.}
Audio--visual (AV) and audio-only (AO) results are reported separately as percentages. M1--M5 measure
detection, key-point coverage, localization, transcription, and user profile; Avg. uses weights
\num{benchmark_main.weights} over M1--M5, and FTR is the no-demand false-trigger rate (lower is better).
Size denotes the model parameter count; for MoE models, ``A'' indicates the number of active parameters.
The text-only system receives ASR transcripts and sentence-level timestamps but no media.
}
\label{tab:main_results}
\begingroup
\scriptsize
\setlength{\tabcolsep}{1.3pt}
\renewcommand{\arraystretch}{1.06}
% Keep exact missing keys in source/check output, but do not let diagnostics widen the table.
\renewcommand{\odunumMissing}[2]{\textcolor{red}{TBD}}
% Avg. is a summary column, not a peer of M1--M5: give both blocks an
% identical fixed-width, centered column that carries its own light-gray band
% (\columncolor). The per-cell \cellcolor on each Ov. cell below is still needed so
% the band survives on \rowcolor stripes (cell color beats column color); the band
% is slightly wider than the M columns so the summary reads one tier above them.
\newcolumntype{O}{>{\columncolor{odutabavg}\centering\arraybackslash}m{0.60cm}}
\begin{tabular}{@{}l c O *{6}{r} @{\hspace{6pt}} O *{6}{r}@{}}
\toprule
 &
 & \multicolumn{7}{c}{\textbf{Audio--visual (\%)}}
 & \multicolumn{7}{c@{}}{\textbf{Audio-only (\%)}} \\
\cmidrule(lr){3-9}\cmidrule(l){10-16}
% Model / Size sit on the same line as the M1--M5 stubs, not up with the modality
% spans: they head single columns, so pairing them with the spans read as if they
% were spanned too. \makecell keeps them vertically centered against the two-line
% metric stubs beside them.
\makecell[l]{\textbf{Model}}
 & \makecell{\textbf{Size}}
 & \cellcolor{odutabavg}\textbf{Avg.}
 & \makecell{\textbf{M1}\\Detect}
 & \makecell{\textbf{M2}\\Keypt.}
 & \makecell{\textbf{M3}\\Locate}
 & \makecell{\textbf{M4}\\Trans.}
 & \makecell{\textbf{M5}\\Profile}
 & \makecell{\textbf{FTR}\\Neg.}
 & \cellcolor{odutabavg}\textbf{Avg.}
 & \makecell{\textbf{M1}\\Detect}
 & \makecell{\textbf{M2}\\Keypt.}
 & \makecell{\textbf{M3}\\Locate}
 & \makecell{\textbf{M4}\\Trans.}
 & \makecell{\textbf{M5}\\Profile}
 & \makecell{\textbf{FTR}\\Neg.} \\
\midrule
\odugrouprow{16}{Closed-source Models}
\midrule
Gemini~\numlit{3.1}~Pro & \na & \cellcolor{odutabavg}\best{\numpct{benchmark_main.overall.gemini.av}}
 & \second{\numpct{benchmark_main.m1.gemini.av}}
 & \numpct{benchmark_main.m2.gemini.av}
 & \second{\numpct{benchmark_main.m3.gemini.av}}
 & \best{\numpct{benchmark_main.m4.gemini.av}}
 & \best{\numpct{benchmark_main.m5.gemini.av}}
 & \second{\numpct{benchmark_main.ftr.gemini.av}}
 & \cellcolor{odutabavg}\second{\numpct{benchmark_main.overall.gemini.ao}}
 & \second{\numpct{benchmark_main.m1.gemini.ao}}
 & \numpct{benchmark_main.m2.gemini.ao}
 & \numpct{benchmark_main.m3.gemini.ao}
 & \second{\numpct{benchmark_main.m4.gemini.ao}}
 & \best{\numpct{benchmark_main.m5.gemini.ao}}
 & \second{\numpct{benchmark_main.ftr.gemini.ao}} \\
\rowcolor{odutabrow}
Gemini~\numlit{3.7}~Flash & \na & \cellcolor{odutabavg}\numpct{benchmark_main.overall.gemini37_flash.av}
 & \best{\numpct{benchmark_main.m1.gemini37_flash.av}}
 & \numpct{benchmark_main.m2.gemini37_flash.av}
 & \numpct{benchmark_main.m3.gemini37_flash.av}
 & \numpct{benchmark_main.m4.gemini37_flash.av}
 & \numpct{benchmark_main.m5.gemini37_flash.av}
 & \best{\numpct{benchmark_main.ftr.gemini37_flash.av}}
 & \cellcolor{odutabavg}\numpct{benchmark_main.overall.gemini37_flash.ao}
 & \best{\numpct{benchmark_main.m1.gemini37_flash.ao}}
 & \numpct{benchmark_main.m2.gemini37_flash.ao}
 & \numpct{benchmark_main.m3.gemini37_flash.ao}
 & \numpct{benchmark_main.m4.gemini37_flash.ao}
 & \numpct{benchmark_main.m5.gemini37_flash.ao}
 & \best{\numpct{benchmark_main.ftr.gemini37_flash.ao}} \\
Gemini~\numlit{3.5}~Flash Lite & \na & \cellcolor{odutabavg}\numpct{benchmark_main.overall.gemini35_flash_lite.av}
 & \numpct{benchmark_main.m1.gemini35_flash_lite.av}
 & \numpct{benchmark_main.m2.gemini35_flash_lite.av}
 & \numpct{benchmark_main.m3.gemini35_flash_lite.av}
 & \numpct{benchmark_main.m4.gemini35_flash_lite.av}
 & \numpct{benchmark_main.m5.gemini35_flash_lite.av}
 & \numpct{benchmark_main.ftr.gemini35_flash_lite.av}
 & \cellcolor{odutabavg}\numpct{benchmark_main.overall.gemini35_flash_lite.ao}
 & \numpct{benchmark_main.m1.gemini35_flash_lite.ao}
 & \numpct{benchmark_main.m2.gemini35_flash_lite.ao}
 & \numpct{benchmark_main.m3.gemini35_flash_lite.ao}
 & \numpct{benchmark_main.m4.gemini35_flash_lite.ao}
 & \numpct{benchmark_main.m5.gemini35_flash_lite.ao}
 & \numpct{benchmark_main.ftr.gemini35_flash_lite.ao} \\
\rowcolor{odutabrow}
Qwen3.5-Omni-Plus & \na & \cellcolor{odutabavg}\second{\numpct{benchmark_main.overall.qwen_plus.av}}
 & \numpct{benchmark_main.m1.qwen_plus.av}
 & \best{\numpct{benchmark_main.m2.qwen_plus.av}}
 & \numpct{benchmark_main.m3.qwen_plus.av}
 & \numpct{benchmark_main.m4.qwen_plus.av}
 & \second{\numpct{benchmark_main.m5.qwen_plus.av}}
 & \numpct{benchmark_main.ftr.qwen_plus.av}
 & \cellcolor{odutabavg}\numpct{benchmark_main.overall.qwen_plus.ao}
 & \numpct{benchmark_main.m1.qwen_plus.ao}
 & \second{\numpct{benchmark_main.m2.qwen_plus.ao}}
 & \second{\numpct{benchmark_main.m3.qwen_plus.ao}}
 & \numpct{benchmark_main.m4.qwen_plus.ao}
 & \numpct{benchmark_main.m5.qwen_plus.ao}
 & \numpct{benchmark_main.ftr.qwen_plus.ao} \\
Seed~\numlit{2.0}~Lite & \na & \cellcolor{odutabavg}\numpct{benchmark_main.overall.seed.av}
 & \numpct{benchmark_main.m1.seed.av}
 & \second{\numpct{benchmark_main.m2.seed.av}}
 & \best{\numpct{benchmark_main.m3.seed.av}}
 & \numpct{benchmark_main.m4.seed.av}
 & \numpct{benchmark_main.m5.seed.av}
 & \numpct{benchmark_main.ftr.seed.av}
 & \cellcolor{odutabavg}\best{\numpct{benchmark_main.overall.seed.ao}}
 & \numpct{benchmark_main.m1.seed.ao}
 & \best{\numpct{benchmark_main.m2.seed.ao}}
 & \best{\numpct{benchmark_main.m3.seed.ao}}
 & \best{\numpct{benchmark_main.m4.seed.ao}}
 & \second{\numpct{benchmark_main.m5.seed.ao}}
 & \numpct{benchmark_main.ftr.seed.ao} \\
\rowcolor{odutabrow}
GPT-Realtime-\numlit{2} & \na & \cellcolor{odutabavg}\na
 & \na
 & \na
 & \na
 & \na
 & \na
 & \na
 & \cellcolor{odutabavg}\numpct{benchmark_main.overall.gpt_realtime.ao}
 & \numpct{benchmark_main.m1.gpt_realtime.ao}
 & \numpct{benchmark_main.m2.gpt_realtime.ao}
 & \numpct{benchmark_main.m3.gpt_realtime.ao}
 & \numpct{benchmark_main.m4.gpt_realtime.ao}
 & \numpct{benchmark_main.m5.gpt_realtime.ao}
 & \numpct{benchmark_main.ftr.gpt_realtime.ao} \\
\midrule
\odugrouprow{16}{Open-source Models}
\midrule
Qwen3-Omni-Think & \num{benchmark_main.size.qwen3_omni_think} & \cellcolor{odutabavg}\numpct{benchmark_main.overall.qwen3_omni_think.av}
 & \numpct{benchmark_main.m1.qwen3_omni_think.av}
 & \numpct{benchmark_main.m2.qwen3_omni_think.av}
 & \numpct{benchmark_main.m3.qwen3_omni_think.av}
 & \numpct{benchmark_main.m4.qwen3_omni_think.av}
 & \numpct{benchmark_main.m5.qwen3_omni_think.av}
 & \numpct{benchmark_main.ftr.qwen3_omni_think.av}
 & \cellcolor{odutabavg}\numpct{benchmark_main.overall.qwen3_omni_think.ao}
 & \numpct{benchmark_main.m1.qwen3_omni_think.ao}
 & \numpct{benchmark_main.m2.qwen3_omni_think.ao}
 & \numpct{benchmark_main.m3.qwen3_omni_think.ao}
 & \numpct{benchmark_main.m4.qwen3_omni_think.ao}
 & \numpct{benchmark_main.m5.qwen3_omni_think.ao}
 & \numpct{benchmark_main.ftr.qwen3_omni_think.ao} \\
\rowcolor{odutabrow}
Qwen3-Omni-Instruct & \num{benchmark_main.size.qwen3_omni_instruct} & \cellcolor{odutabavg}\numpct{benchmark_main.overall.qwen3_omni_instruct.av}
 & \numpct{benchmark_main.m1.qwen3_omni_instruct.av}
 & \numpct{benchmark_main.m2.qwen3_omni_instruct.av}
 & \numpct{benchmark_main.m3.qwen3_omni_instruct.av}
 & \second{\numpct{benchmark_main.m4.qwen3_omni_instruct.av}}
 & \numpct{benchmark_main.m5.qwen3_omni_instruct.av}
 & \numpct{benchmark_main.ftr.qwen3_omni_instruct.av}
 & \cellcolor{odutabavg}\numpct{benchmark_main.overall.qwen3_omni_instruct.ao}
 & \numpct{benchmark_main.m1.qwen3_omni_instruct.ao}
 & \numpct{benchmark_main.m2.qwen3_omni_instruct.ao}
 & \numpct{benchmark_main.m3.qwen3_omni_instruct.ao}
 & \numpct{benchmark_main.m4.qwen3_omni_instruct.ao}
 & \numpct{benchmark_main.m5.qwen3_omni_instruct.ao}
 & \numpct{benchmark_main.ftr.qwen3_omni_instruct.ao} \\
Qwen2.5-Omni & \num{benchmark_main.size.qwen25_omni} & \cellcolor{odutabavg}\numpct{benchmark_main.overall.qwen25_omni.av}
 & \numpct{benchmark_main.m1.qwen25_omni.av}
 & \numpct{benchmark_main.m2.qwen25_omni.av}
 & \numpct{benchmark_main.m3.qwen25_omni.av}
 & \numpct{benchmark_main.m4.qwen25_omni.av}
 & \numpct{benchmark_main.m5.qwen25_omni.av}
 & \numpct{benchmark_main.ftr.qwen25_omni.av}
 & \cellcolor{odutabavg}\numpct{benchmark_main.overall.qwen25_omni.ao}
 & \numpct{benchmark_main.m1.qwen25_omni.ao}
 & \numpct{benchmark_main.m2.qwen25_omni.ao}
 & \numpct{benchmark_main.m3.qwen25_omni.ao}
 & \numpct{benchmark_main.m4.qwen25_omni.ao}
 & \numpct{benchmark_main.m5.qwen25_omni.ao}
 & \numpct{benchmark_main.ftr.qwen25_omni.ao} \\
\rowcolor{odutabrow}
Ming-Flash-Omni~\numlit{2.0} & \num{benchmark_main.size.ming} & \cellcolor{odutabavg}\numpct{benchmark_main.overall.ming.av}
 & \numpct{benchmark_main.m1.ming.av}
 & \numpct{benchmark_main.m2.ming.av}
 & \numpct{benchmark_main.m3.ming.av}
 & \numpct{benchmark_main.m4.ming.av}
 & \numpct{benchmark_main.m5.ming.av}
 & \numpct{benchmark_main.ftr.ming.av}
 & \cellcolor{odutabavg}\numpct{benchmark_main.overall.ming.ao}
 & \numpct{benchmark_main.m1.ming.ao}
 & \numpct{benchmark_main.m2.ming.ao}
 & \numpct{benchmark_main.m3.ming.ao}
 & \numpct{benchmark_main.m4.ming.ao}
 & \numpct{benchmark_main.m5.ming.ao}
 & \numpct{benchmark_main.ftr.ming.ao} \\
MiniCPM-o~\numlit{4.5} & \num{benchmark_main.size.minicpm_o} & \cellcolor{odutabavg}\numpct{benchmark_main.overall.minicpm_o.av}
 & \numpct{benchmark_main.m1.minicpm_o.av}
 & \numpct{benchmark_main.m2.minicpm_o.av}
 & \numpct{benchmark_main.m3.minicpm_o.av}
 & \numpct{benchmark_main.m4.minicpm_o.av}
 & \numpct{benchmark_main.m5.minicpm_o.av}
 & \numpct{benchmark_main.ftr.minicpm_o.av}
 & \cellcolor{odutabavg}\numpct{benchmark_main.overall.minicpm_o.ao}
 & \numpct{benchmark_main.m1.minicpm_o.ao}
 & \numpct{benchmark_main.m2.minicpm_o.ao}
 & \numpct{benchmark_main.m3.minicpm_o.ao}
 & \numpct{benchmark_main.m4.minicpm_o.ao}
 & \numpct{benchmark_main.m5.minicpm_o.ao}
 & \numpct{benchmark_main.ftr.minicpm_o.ao} \\
\rowcolor{odutabrow}
Nemotron~\numlit{3}~Nano Omni & \num{benchmark_main.size.nemotron} & \cellcolor{odutabavg}\numpct{benchmark_main.overall.nemotron.av}
 & \numpct{benchmark_main.m1.nemotron.av}
 & \numpct{benchmark_main.m2.nemotron.av}
 & \numpct{benchmark_main.m3.nemotron.av}
 & \numpct{benchmark_main.m4.nemotron.av}
 & \numpct{benchmark_main.m5.nemotron.av}
 & \numpct{benchmark_main.ftr.nemotron.av}
 & \cellcolor{odutabavg}\numpct{benchmark_main.overall.nemotron.ao}
 & \numpct{benchmark_main.m1.nemotron.ao}
 & \numpct{benchmark_main.m2.nemotron.ao}
 & \numpct{benchmark_main.m3.nemotron.ao}
 & \numpct{benchmark_main.m4.nemotron.ao}
 & \numpct{benchmark_main.m5.nemotron.ao}
 & \numpct{benchmark_main.ftr.nemotron.ao} \\
video-SALMONN~\numlit{2}+ & \num{benchmark_main.size.salmonn2_7b} & \cellcolor{odutabavg}\numpct{benchmark_main.overall.salmonn2_7b.av}
 & \numpct{benchmark_main.m1.salmonn2_7b.av}
 & \numpct{benchmark_main.m2.salmonn2_7b.av}
 & \numpct{benchmark_main.m3.salmonn2_7b.av}
 & \numpct{benchmark_main.m4.salmonn2_7b.av}
 & \numpct{benchmark_main.m5.salmonn2_7b.av}
 & \numpct{benchmark_main.ftr.salmonn2_7b.av}
 & \cellcolor{odutabavg}\numpct{benchmark_main.overall.salmonn2_7b.ao}
 & \numpct{benchmark_main.m1.salmonn2_7b.ao}
 & \numpct{benchmark_main.m2.salmonn2_7b.ao}
 & \numpct{benchmark_main.m3.salmonn2_7b.ao}
 & \numpct{benchmark_main.m4.salmonn2_7b.ao}
 & \numpct{benchmark_main.m5.salmonn2_7b.ao}
 & \numpct{benchmark_main.ftr.salmonn2_7b.ao} \\
\rowcolor{odutabrow}
Kimi-Audio & \num{benchmark_main.size.kimi_audio_7b_instruct} & \cellcolor{odutabavg}\na
 & \na
 & \na
 & \na
 & \na
 & \na
 & \na
 & \cellcolor{odutabavg}\numpct{benchmark_main.overall.kimi_audio_7b_instruct.ao}
 & \numpct{benchmark_main.m1.kimi_audio_7b_instruct.ao}
 & \numpct{benchmark_main.m2.kimi_audio_7b_instruct.ao}
 & \numpct{benchmark_main.m3.kimi_audio_7b_instruct.ao}
 & \numpct{benchmark_main.m4.kimi_audio_7b_instruct.ao}
 & \numpct{benchmark_main.m5.kimi_audio_7b_instruct.ao}
 & \numpct{benchmark_main.ftr.kimi_audio_7b_instruct.ao} \\
\midrule
\odugrouprow{16}{Text-only System (no native media input)}
\midrule
% No \best/\second in this row: the markers rank the systems being compared, and a
% cascade is deliberately not a competitor. It is handed verbatim the very thing M4
% scores, and ASR sentence timings are what M3 scores, so marking either would
% assert a leaderboard win for a system with no perception.
GPT-\numlit{5.4} + ASR transcript & \na & \cellcolor{odutabavg}\numpct{benchmark_main.overall.cascade_asr.av}
 & \numpct{benchmark_main.m1.cascade_asr.av}
 & \numpct{benchmark_main.m2.cascade_asr.av}
 & \numpct{benchmark_main.m3.cascade_asr.av}
 & \numpct{benchmark_main.m4.cascade_asr.av}
 & \numpct{benchmark_main.m5.cascade_asr.av}
 & \numpct{benchmark_main.ftr.cascade_asr.av}
 & \cellcolor{odutabavg}\numpct{benchmark_main.overall.cascade_asr.ao}
 & \numpct{benchmark_main.m1.cascade_asr.ao}
 & \numpct{benchmark_main.m2.cascade_asr.ao}
 & \numpct{benchmark_main.m3.cascade_asr.ao}
 & \numpct{benchmark_main.m4.cascade_asr.ao}
 & \numpct{benchmark_main.m5.cascade_asr.ao}
 & \numpct{benchmark_main.ftr.cascade_asr.ao} \\
\bottomrule
\end{tabular}
\endgroup
\odutabnote{Best and second-best values are \best{bold} and \second{underlined}.}
\end{table}

\subsection{Overall results}
\label{sec:overall_results}

\textbf{ODU remains challenging even for the strongest models.}
Table~\ref{tab:main_results} shows that the best native Avg.\ scores reach only
\numpct{benchmark_main.overall.gemini.av}\% on AV scenes (Gemini~\numlit{3.1}~Pro) and
\numpct{benchmark_main.overall.seed.ao}\% on AO scenes (Seed~\numlit{2.0}~Lite).
The highest AV M2 is just \numpct{benchmark_main.m2.qwen_plus.av}\%, leaving substantial
headroom in the primary semantic dimension. Open-source models lag further: even their strongest
model, Qwen3-Omni-Think, reaches only \numpct{benchmark_main.overall.qwen3_omni_think.av}\% AV and
\numpct{benchmark_main.overall.qwen3_omni_think.ao}\% AO Avg.

\textbf{Strong perception does not imply strong contextual reasoning.}
On AV scenes, Gemini~\numlit{3.1}~Pro achieves \numpct{benchmark_main.m3.gemini.av}\% M3,
\numpct{benchmark_main.m4.gemini.av}\% M4, and \numpct{benchmark_main.m5.gemini.av}\% M5,
yet its M2 is only \numpct{benchmark_main.m2.gemini.av}\% and its FTR reaches
\numpct{benchmark_main.ftr.gemini.av}\%. High localization, transcription, and user-profile scores
thus do not ensure correct interpretation of a user demand in context. The key-point and false-trigger
analyses (\S\ref{sec:where_models_miss}, \S\ref{sec:false_trigger_structure}) identify missed
contextual information and sensitivity to request-like language and source/addressee mismatches.
The qualitative error analysis in App.~\ref{app:error_analysis} also visualizes these failures.

\textbf{Strong demand recovery does not ensure reliable engagement.}
Seed~\numlit{2.0}~Lite achieves \numpct{benchmark_main.m2.seed.ao}\% M2 on audio-only scenes but triggers on
\numpct{benchmark_main.ftr.seed.ao}\% of no-demand scenes. Gemini~\numlit{3.7}~Flash has lower M2
(\numpct{benchmark_main.m2.gemini37_flash.ao}\%) and much lower FTR
(\numpct{benchmark_main.ftr.gemini37_flash.ao}\%). This contrast shows that models can recover the content of valid demands yet still infer a demand when none is present. Reliable demand understanding therefore requires judging whether the interaction calls for an assistant response, alongside recovering what the user wants.

\textbf{Models targeting efficient or real-time interaction lag in demand understanding.}
Gemini~\numlit{3.5}~Flash Lite, GPT-Realtime-\numlit{2}, and MiniCPM-o~\numlit{4.5} target
efficient or real-time interaction. However, their AO Avg.\ scores are only
\numpct{benchmark_main.overall.gemini35_flash_lite.ao}\%,
\numpct{benchmark_main.overall.gpt_realtime.ao}\%, and
\numpct{benchmark_main.overall.minicpm_o.ao}\%, respectively, below both
Gemini~\numlit{3.1}~Pro (\numpct{benchmark_main.overall.gemini.ao}\%) and
Qwen3-Omni-Think (\numpct{benchmark_main.overall.qwen3_omni_think.ao}\%).
This gap raises a practical concern: interactive systems often rely on real-time models, yet
weaknesses in demand understanding can cause their responses to miss the user's intent and degrade
the interaction experience.

% Equal-height diagnostic artwork with independent Figure 4 / Figure 5 captions.
% Queue after Table 3, while page 7's results discussion is still being typeset.
\begin{figure}[t]
\centering
\begin{minipage}[t]{0.42\textwidth}
\input{figures/keypoint_types.tex}
\end{minipage}\hfill%
\begin{minipage}[t]{0.55\textwidth}
\input{figures/negative_subtypes_heatmap.tex}
\end{minipage}
\end{figure}

% \textbf{Temporal grounding is a particular weakness of open-source systems.}
% As relevant objects and events can change over time, interpreting a demand requires associating it
% with the appropriate interaction context. Temporal localization provides an anchor for retrieving
% surrounding multimodal evidence and revisiting requests in interaction history. Yet even
% Qwen3-Omni-Think, the strongest open-source model on M3, reaches only
% \numpct{benchmark_main.m3.qwen3_omni_think.av}\% on AV scenes and
% \numpct{benchmark_main.m3.qwen3_omni_think.ao}\% on AO scenes, despite AV transcription quality
% of \numpct{benchmark_main.m4.qwen3_omni_think.av}\%. This contrast highlights a gap between
% recovering demand-bearing speech and locating its temporal extent, exposing a weakness in using
% predicted spans to retrieve the underlying interaction.

\subsection{Contextual demand recovery}
\label{sec:where_models_miss}
\textbf{Models recover multimodal and conversational context less reliably than spoken requests.}
M2, our primary semantic metric, measures coverage of key information about user intent and
required context. Using source labels assigned during annotation, we compare the fraction of
key points recovered per source across three native systems and the text-only baseline on
audio--visual demand scenes
(Figure~\ref{fig:keypoint_types}).

Native spoken-request hit-rates are
\numpct{keypoint_types.cov.seed.av.intent}--\numpct{keypoint_types.cov.qwen_plus.av.intent}\%,
comparable to the text-only baseline's \numpct{keypoint_types.cov.cascade_asr.av.intent}\% and consistent
with the strong transcription scores in Table~\ref{tab:main_results}. Yet all three native systems
score below \numlit{60}\% on visual and acoustic key points, highlighting weak recovery of
multimodal context.
Dialogue-history hit-rates are also lower
(\numpct{keypoint_types.cov.gemini.av.history}--\numpct{keypoint_types.cov.seed.av.history}\%
for native systems). The text-only baseline receives prior-turn assistant reply text and reaches
\numpct{keypoint_types.cov.cascade_asr.av.history}\%.
The text-only baseline also recovers some visual and acoustic key points,
possibly by leveraging commonsense knowledge together with scene details inferred from transcripts and dialogue history.
\textbf{These results show that accurate speech transcription alone is insufficient
for strong ODU performance.}

\subsection{Error patterns across the taxonomy}
\label{sec:false_trigger_structure}
\label{sec:challenge_findings}

We examine false triggers using FTR on negative scenes and demand understanding using M2 on
positive scenes. Figure~\ref{fig:negative_subtypes} shows the AV and AO negative-scene results,
with separate analyses of signal type and invalidation reason. Positive cells mark categories with
FTR above the same model's overall rate on these labeled samples. 

\textbf{Models falsely trigger on negative scenes with request-like wording or source/addressee mismatches.}
Linguistic signals such as questions, imperatives, and complaints yield the highest average FTR,
exceeding every model's mean with an average gap of
\numpct{negative_patterns.panel_mean_delta.signal_type.linguistic_signal} percentage points,
whereas semantic signals that merely mention assistant functions fall below every model's mean.
A clear, answerable request can still be invalid for the assistant: in Wrong Source it comes from
media playback or background speakers; in Wrong Addressee it targets someone else.
These are the hardest invalidation categories on average; even Gemini~\numlit{3.7}~Flash,
with the lowest FTR in Table~\ref{tab:main_results}, exceeds its own mean by
\numpct{negative_patterns.delta_own_mean.gemini37_flash.invalidation_reason.source_mismatch}
percentage points on Wrong Source.
Together, these patterns suggest over-reliance on request-like wording: source and addressee cues
do not reliably override the apparent request, so speech outside the user--assistant interaction
is promoted into a demand.

\textbf{Models struggle to recover indirect requests on positive scenes.}
Indirect or descriptive requests score below every model's mean M2, with average
differences of \numpct{positive_taxonomy_m2.av.panel_delta.axis3.3_3} points on AV scenes and
\numpct{positive_taxonomy_m2.ao.panel_delta.axis3.3_3} points on AO scenes.
Appendix~\ref{app:positive_taxonomy_m2} provides the full breakdown.
Models can mistake speech for a request to the assistant, yet struggle to understand genuine needs
that users do not state directly.

\subsection{Comparison with human-recorded interactions}
\label{sec:recorded_media}

% Hand-maintained layout; all values use the generated number contract.
% Source: results/recorded_vs_generated.json, audio--visual Chinese stratum.
% Full single-line model names use 6.5pt headers above Table 3-style 7pt numbers.
% Each model group preserves Syn. / Rec. / Delta order and recorded-minus-rendered Delta.
\begin{table}[htbp]
\centering
\caption{Model performance (\%) on Chinese audio--visual scenes from synthetic (Syn.) and human-recorded (Rec.) interactions.
\mbox{$\Delta = {}$Rec.~$-$~Syn.}\ in percentage points.}
\label{tab:recorded_vs_generated}

\scriptsize
\setlength{\tabcolsep}{1.1pt}
\renewcommand{\arraystretch}{1.06}
% S columns need expandable numeric input. Resolve the same generated keys locally;
% siunitx handles one-decimal rounding and explicit positive signs in Delta columns.
\ExplSyntaxOn
\cs_set:Npn \numpct #1 { \fp_eval:n { 100 * \use:c { odunum@val@#1 } } }
\cs_set_eq:NN \numpctd \numpct
\ExplSyntaxOff
\sisetup{
  mode=math,
  reset-math-version=true,
  reset-text-series=true,
  text-series-to-math=false,
  table-column-width={\dimexpr(\textwidth-40bp-22\tabcolsep-28bp)/15\relax},
  table-fixed-width=true,
  table-number-alignment=center,
  round-mode=places,
  round-precision=1,
  round-pad=true,
  group-digits=false
}
% Delta columns alone carry the Table 3 summary gray; no ranking marks.
\begin{tabular}{@{}p{40bp}
S[table-format=2.1] S[table-format=2.1] >{\columncolor{odutabavg}[0pt][0pt]}S[table-format=+2.1,print-implicit-plus=true]
@{\hspace{7bp}}
S[table-format=2.1] S[table-format=2.1] >{\columncolor{odutabavg}[0pt][0pt]}S[table-format=+2.1,print-implicit-plus=true]
@{\hspace{7bp}}
S[table-format=2.1] S[table-format=2.1] >{\columncolor{odutabavg}[0pt][0pt]}S[table-format=+2.1,print-implicit-plus=true]
@{\hspace{7bp}}
S[table-format=2.1] S[table-format=2.1] >{\columncolor{odutabavg}[0pt][0pt]}S[table-format=+2.1,print-implicit-plus=true]
@{\hspace{7bp}}
S[table-format=2.1] S[table-format=2.1] >{\columncolor{odutabavg}[0pt][0pt]}S[table-format=+2.1,print-implicit-plus=true]
@{}}
\toprule
\multirow{2}{*}{Metric}
 & \multicolumn{3}{c}{{\fontsize{6.5}{7.5}\selectfont Gemini~\numlit{3.1}~Pro}}
 & \multicolumn{3}{c}{{\fontsize{6.5}{7.5}\selectfont Seed~\numlit{2.0}~Lite}}
 & \multicolumn{3}{c}{{\fontsize{6.5}{7.5}\selectfont Qwen3.5-Omni-Plus}}
 & \multicolumn{3}{c}{{\fontsize{6.5}{7.5}\selectfont Qwen3-Omni-Think}}
 & \multicolumn{3}{c}{{\fontsize{6.5}{7.5}\selectfont Ming-Flash-Omni~\numlit{2.0}}} \\
\cmidrule(lr){2-4}\cmidrule(lr){5-7}\cmidrule(lr){8-10}\cmidrule(lr){11-13}\cmidrule(lr){14-16}
 & {Syn.} & {Rec.} & {$\Delta$}
 & {Syn.} & {Rec.} & {$\Delta$}
 & {Syn.} & {Rec.} & {$\Delta$}
 & {Syn.} & {Rec.} & {$\Delta$}
 & {Syn.} & {Rec.} & {$\Delta$} \\
\midrule
Avg.
 &  \numpct{recorded_vs_generated.avg.stratum.gemini.generated}
 &  \numpct{recorded_vs_generated.avg.stratum.gemini.recorded}
 & \numpctd{recorded_vs_generated.avg_delta.stratum.gemini}
 & \numpct{recorded_vs_generated.avg.stratum.seed.generated}
 & \numpct{recorded_vs_generated.avg.stratum.seed.recorded}
 & \numpctd{recorded_vs_generated.avg_delta.stratum.seed}
 & \numpct{recorded_vs_generated.avg.stratum.qwen_plus.generated}
 & \numpct{recorded_vs_generated.avg.stratum.qwen_plus.recorded}
 & \numpctd{recorded_vs_generated.avg_delta.stratum.qwen_plus}
 & \numpct{recorded_vs_generated.avg.stratum.qwen3_omni_think.generated}
 & \numpct{recorded_vs_generated.avg.stratum.qwen3_omni_think.recorded}
 & \numpctd{recorded_vs_generated.avg_delta.stratum.qwen3_omni_think}
 & \numpct{recorded_vs_generated.avg.stratum.ming.generated}
 & \numpct{recorded_vs_generated.avg.stratum.ming.recorded}
 & \numpctd{recorded_vs_generated.avg_delta.stratum.ming} \\
M1~Detect
 &  \numpct{recorded_vs_generated.m1.stratum.gemini.generated}
 &  \numpct{recorded_vs_generated.m1.stratum.gemini.recorded}
 & \numpctd{recorded_vs_generated.m1_delta.stratum.gemini}
 & \numpct{recorded_vs_generated.m1.stratum.seed.generated}
 & \numpct{recorded_vs_generated.m1.stratum.seed.recorded}
 & \numpctd{recorded_vs_generated.m1_delta.stratum.seed}
 & \numpct{recorded_vs_generated.m1.stratum.qwen_plus.generated}
 & \numpct{recorded_vs_generated.m1.stratum.qwen_plus.recorded}
 & \numpctd{recorded_vs_generated.m1_delta.stratum.qwen_plus}
 & \numpct{recorded_vs_generated.m1.stratum.qwen3_omni_think.generated}
 & \numpct{recorded_vs_generated.m1.stratum.qwen3_omni_think.recorded}
 & \numpctd{recorded_vs_generated.m1_delta.stratum.qwen3_omni_think}
 & \numpct{recorded_vs_generated.m1.stratum.ming.generated}
 & \numpct{recorded_vs_generated.m1.stratum.ming.recorded}
 & \numpctd{recorded_vs_generated.m1_delta.stratum.ming} \\
M2~Keypt.
 & \numpct{recorded_vs_generated.m2.stratum.gemini.generated}
 &  \numpct{recorded_vs_generated.m2.stratum.gemini.recorded}
 & \numpctd{recorded_vs_generated.m2_delta.stratum.gemini}
 & \numpct{recorded_vs_generated.m2.stratum.seed.generated}
 & \numpct{recorded_vs_generated.m2.stratum.seed.recorded}
 & \numpctd{recorded_vs_generated.m2_delta.stratum.seed}
 &  \numpct{recorded_vs_generated.m2.stratum.qwen_plus.generated}
 & \numpct{recorded_vs_generated.m2.stratum.qwen_plus.recorded}
 & \numpctd{recorded_vs_generated.m2_delta.stratum.qwen_plus}
 & \numpct{recorded_vs_generated.m2.stratum.qwen3_omni_think.generated}
 & \numpct{recorded_vs_generated.m2.stratum.qwen3_omni_think.recorded}
 & \numpctd{recorded_vs_generated.m2_delta.stratum.qwen3_omni_think}
 & \numpct{recorded_vs_generated.m2.stratum.ming.generated}
 & \numpct{recorded_vs_generated.m2.stratum.ming.recorded}
 & \numpctd{recorded_vs_generated.m2_delta.stratum.ming} \\
M3~Locate
 & \numpct{recorded_vs_generated.m3.stratum.gemini.generated}
 &  \numpct{recorded_vs_generated.m3.stratum.gemini.recorded}
 & \numpctd{recorded_vs_generated.m3_delta.stratum.gemini}
 &  \numpct{recorded_vs_generated.m3.stratum.seed.generated}
 & \numpct{recorded_vs_generated.m3.stratum.seed.recorded}
 & \numpctd{recorded_vs_generated.m3_delta.stratum.seed}
 & \numpct{recorded_vs_generated.m3.stratum.qwen_plus.generated}
 & \numpct{recorded_vs_generated.m3.stratum.qwen_plus.recorded}
 & \numpctd{recorded_vs_generated.m3_delta.stratum.qwen_plus}
 & \numpct{recorded_vs_generated.m3.stratum.qwen3_omni_think.generated}
 & \numpct{recorded_vs_generated.m3.stratum.qwen3_omni_think.recorded}
 & \numpctd{recorded_vs_generated.m3_delta.stratum.qwen3_omni_think}
 & \numpct{recorded_vs_generated.m3.stratum.ming.generated}
 & \numpct{recorded_vs_generated.m3.stratum.ming.recorded}
 & \numpctd{recorded_vs_generated.m3_delta.stratum.ming} \\
M4~Trans.
 &  \numpct{recorded_vs_generated.m4.stratum.gemini.generated}
 &  \numpct{recorded_vs_generated.m4.stratum.gemini.recorded}
 & \numpctd{recorded_vs_generated.m4_delta.stratum.gemini}
 & \numpct{recorded_vs_generated.m4.stratum.seed.generated}
 & \numpct{recorded_vs_generated.m4.stratum.seed.recorded}
 & \numpctd{recorded_vs_generated.m4_delta.stratum.seed}
 & \numpct{recorded_vs_generated.m4.stratum.qwen_plus.generated}
 & \numpct{recorded_vs_generated.m4.stratum.qwen_plus.recorded}
 & \numpctd{recorded_vs_generated.m4_delta.stratum.qwen_plus}
 & \numpct{recorded_vs_generated.m4.stratum.qwen3_omni_think.generated}
 & \numpct{recorded_vs_generated.m4.stratum.qwen3_omni_think.recorded}
 & \numpctd{recorded_vs_generated.m4_delta.stratum.qwen3_omni_think}
 & \numpct{recorded_vs_generated.m4.stratum.ming.generated}
 & \numpct{recorded_vs_generated.m4.stratum.ming.recorded}
 & \numpctd{recorded_vs_generated.m4_delta.stratum.ming} \\
M5~Profile
 &  \numpct{recorded_vs_generated.m5.stratum.gemini.generated}
 &  \numpct{recorded_vs_generated.m5.stratum.gemini.recorded}
 & \numpctd{recorded_vs_generated.m5_delta.stratum.gemini}
 & \numpct{recorded_vs_generated.m5.stratum.seed.generated}
 & \numpct{recorded_vs_generated.m5.stratum.seed.recorded}
 & \numpctd{recorded_vs_generated.m5_delta.stratum.seed}
 & \numpct{recorded_vs_generated.m5.stratum.qwen_plus.generated}
 & \numpct{recorded_vs_generated.m5.stratum.qwen_plus.recorded}
 & \numpctd{recorded_vs_generated.m5_delta.stratum.qwen_plus}
 & \numpct{recorded_vs_generated.m5.stratum.qwen3_omni_think.generated}
 & \numpct{recorded_vs_generated.m5.stratum.qwen3_omni_think.recorded}
 & \numpctd{recorded_vs_generated.m5_delta.stratum.qwen3_omni_think}
 & \numpct{recorded_vs_generated.m5.stratum.ming.generated}
 & \numpct{recorded_vs_generated.m5.stratum.ming.recorded}
 & \numpctd{recorded_vs_generated.m5_delta.stratum.ming} \\
FTR
 &  \numpct{recorded_vs_generated.ftr.stratum.gemini.generated}
 &  \numpct{recorded_vs_generated.ftr.stratum.gemini.recorded}
 & \numpctd{recorded_vs_generated.ftr_delta.stratum.gemini}
 & \numpct{recorded_vs_generated.ftr.stratum.seed.generated}
 & \numpct{recorded_vs_generated.ftr.stratum.seed.recorded}
 & \numpctd{recorded_vs_generated.ftr_delta.stratum.seed}
 & \numpct{recorded_vs_generated.ftr.stratum.qwen_plus.generated}
 & \numpct{recorded_vs_generated.ftr.stratum.qwen_plus.recorded}
 & \numpctd{recorded_vs_generated.ftr_delta.stratum.qwen_plus}
 & \numpct{recorded_vs_generated.ftr.stratum.qwen3_omni_think.generated}
 & \numpct{recorded_vs_generated.ftr.stratum.qwen3_omni_think.recorded}
 & \numpctd{recorded_vs_generated.ftr_delta.stratum.qwen3_omni_think}
 & \numpct{recorded_vs_generated.ftr.stratum.ming.generated}
 & \numpct{recorded_vs_generated.ftr.stratum.ming.recorded}
 & \numpctd{recorded_vs_generated.ftr_delta.stratum.ming} \\
\bottomrule
\end{tabular}
\end{table}

\textbf{Leading MLLMs retain their performance on human-recorded interactions.}
Table~\ref{tab:recorded_vs_generated} shows comparable Avg.\ scores for
Gemini~\numlit{3.1}~Pro and Seed~\numlit{2.0}~Lite across synthetic and human-recorded interactions.
Qwen3-Omni-Think and Ming-Flash-Omni~\numlit{2.0} are more sensitive to the source change,
with Avg.\ changes of \numpctd{recorded_vs_generated.avg_delta.stratum.qwen3_omni_think} and
\numpctd{recorded_vs_generated.avg_delta.stratum.ming} points, respectively.

\textbf{Human-recorded speech exposes transcription weaknesses.}
Qwen3.5-Omni-Plus and Qwen3-Omni-Think show their largest declines in M4.
Ming-Flash-Omni~\numlit{2.0} has the steepest transcription decline in the table:
\numpctd{recorded_vs_generated.m4_delta.stratum.ming} points in M4, versus
\numpctd{recorded_vs_generated.m2_delta.stratum.ming} in M2.
These declines may stem in part from the more complex environmental noise and accented speech encountered in realistic interaction settings.

\FloatBarrier
\subsection{Downstream response quality}
\label{sec:human_pilot}

We further examine whether correct demand understanding translates into better assistant responses.
Using Gemini~\numlit{3.1}~Pro, we generated paired responses under two conditions:
\emph{without demand annotation} and \emph{with demand annotation}.
Both conditions received identical audio--visual inputs and dialogue history under the same generation settings,
while the latter additionally received the reference demand annotation,
including an explicit no-demand label when applicable. The full experiment setup is detailed in App.~\ref{app:human_pilot}.

We evaluate the paired responses through a blinded human A/B test.
For each interaction, responses A and B are presented in randomized order with condition identities hidden,
and annotators select a preferred response or a tie based on correctness, relevance, contextual grounding, and appropriate silence.
Human evaluation yielded \numpct{response_ab_blind.all.tie_rate}\% ties;
among decisive judgments, \textbf{\numpct{response_ab_blind.all.given_win_rate_among_decisive}\%} favored the \emph{with demand annotation} condition.
These results show that \textbf{providing correct demand information substantially improves downstream response quality},
supporting demand understanding as a key prerequisite for building effective conversational assistants.

\section{Conclusion}
\label{sec:conclusion}

We introduced Omni Demand Understanding (ODU) as a distinct task for identifying valid user demands and
inferring intent from multimodal and conversational context. ODU-Bench combines taxonomy-guided agentic
generation with human-recorded interactions and uses human-verified annotations to evaluate demand
understanding along five complementary dimensions. Experiments with \num{benchmark_main.complete.n_models}
native MLLMs show that strong perceptual performance does not ensure accurate contextual user-intent
inference. In our evidence-channel analysis, the three native systems evaluated on audio--visual demand
scenes recover spoken requests more reliably than the visual, acoustic, and conversational
information needed to interpret them. Models also frequently mistake request-like speech for a demand
to the assistant, even when it comes from media playback or is addressed to someone else. A blinded
human A/B study further shows that providing reference demand annotations improves response quality,
underscoring the practical value of correct demand understanding. We hope ODU will make demand
understanding a central focus of multimodal interaction research and drive progress toward assistants
that infer user intent from multimodal and conversational context before deciding whether and how to respond.

\ifarxiv
  \par
  \Needspace{4\baselineskip}
\else
  \clearpage
\fi
\subsection*{AI use statement}

We used generative AI tools to assist in designing the benchmark taxonomy.
Generative AI tools are integral components of our
pipeline for data synthesis, annotation, and quality control
(\S\ref{sec:benchmark_construction}; App.~\ref{app:generation}). We also use
an LLM judge for key-point-based semantic evaluation
(\S\ref{sec:keypoints}; App.~\ref{app:evaluation}). Additionally, we used AI writing
assistants to improve the writing of this paper.
We did not use generative AI
to develop theoretical models or conceptual frameworks; formulate mathematical
claims; provide critical ingredients for proving mathematical claims; assist in
writing mathematical proofs; propose or refine hypotheses; or design research
methodology or experiments.
All benchmark media and annotations underwent rigorous human review, and human reviewers determined
inclusion in the released benchmark.
We reviewed all AI-assisted work and take responsibility for the final content of this work,
including text, claims or artifacts produced with the aid of generative AI.

\subsection*{Ethics statement}

Human-recorded interactions were performed by paid actors. Before release, all participating actors signed informed-consent forms authorizing use of their recording data, including their faces and voices, for academic evaluation and public release with the benchmark. To protect actor privacy, the released human-recorded interactions may be used solely to evaluate systems on this benchmark; all other academic and commercial uses are prohibited.
Generated scenes contain no real individuals. ODU-Bench is provided solely for academic research and model evaluation, subject
to the above restrictions on human-recorded interactions. Commercial use of the
dataset or its images in products, services, or other profit-making activities
is prohibited. Images in the dataset may depict identifiable individuals;
copyright and related rights remain with their respective rights holders. The
release does not authorize sublicensing, commercial exploitation, or the
creation of derivative works from these images. Users must comply with the
release terms, the scope of participant consent, and applicable personal
information protection and portrait rights laws and regulations.

\subsection*{Reproducibility statement}

The main paper defines the task, construction procedure, and evaluation framework
(\S\ref{sec:task_evaluation}; \S\ref{sec:benchmark_construction}). The appendix
specifies the prediction schema, matching, metrics, aggregation, and a worked
scoring example (App.~\ref{app:evaluation}), the release composition and taxonomy
(App.~\ref{app:release}), and the construction and human-verification procedures
(App.~\ref{app:generation}). Experimental input preparation, the transcript-only
baseline, and the judge configuration are documented in
App.~\ref{app:experimental_setup}; supplementary analyses and judge stability are
reported in App.~\ref{app:supplementary}. Evaluation instructions are provided in
App.~\ref{app:prompt_specs}. Benchmark media, annotations, evaluation code, and
configurations support reproducing the results shown in the paper.

\ifarxiv
\subsection*{Acknowledgments}

This work was supported by Alibaba Innovative Research Program. We would like to
thank the Qwen Team at Alibaba Token Hub (ATH), Alibaba Group, for providing the
computational resources and foundation models (Qwen) used in this research.
\fi

% ---- bibliography ------------------------------------------------------------
\bibliography{refs/references}
\bibliographystyle{iclr2027_conference}

\clearpage
\appendix
\startcontents[appendices]
\section*{Appendix Contents}
\printcontents[appendices]{}{1}{\setcounter{tocdepth}{2}}
\clearpage

% Author-approved appendix order. All dependencies are inside this bundle.
\raggedbottom
\section{Task specification and evaluation}
\label{app:evaluation}

\subsection{Prediction schema}
Only the final user clip is annotated, and timestamps are measured from its start. A prediction contains a \texttt{has\_demand} flag and, when it is true, a list of demand segments. Each segment carries a time span, a transcript, a structured intent, a required-context list, and closed-set user profile fields. The annotation reference also includes a frozen key-point set and a necessity note for each required-context item. Fig.~\ref{fig:schema} gives the schema, and App.~\ref{app:eval_prompts} gives the system prompts for evaluation.

% fig:schema — the prediction schema as a field list. A fully scored instance is
% provided in App.~\ref{app:walkthrough}.
\begin{figure}[!htbp]
  \centering
\begin{lstlisting}[style=json]
{
  "has_demand": bool,
  "segments": [{
    "start_ms": int, "end_ms": int,            // demand span, milliseconds
    "transcript":        str,                  // the demand-bearing speech
    "structured_intent": str,                  // what the user wants done
    "required_context": [                      // what the words left out
      {"type": "visual" | "audio" | "dialogue_history" | "user_activity"
             | "user_identity" | "user_emotion" | "scene_condition",
       "description": str}
    ],
    "user_in_frame": bool, "user_gender": str, "user_age_group": str
  }]
}
\end{lstlisting}
  \caption{The prediction schema. A model emits \texttt{has\_demand} and, when it is true, one segment per
  continuous demand-bearing region. Audio-only predictions omit visual context and
  \texttt{user\_in\_frame}.}
  \label{fig:schema}
\end{figure}

\subsection{Segment matching}
A scene may contain multiple demand segments, so predicted segments are first aligned with reference segments before computing M2--M5. We prioritize normalized-transcript similarity, which is more robust to boundary drift, and fall back to temporal IoU for remaining segments. We additionally resolve common segmentation mismatches by allowing one prediction to cover multiple reference segments and by merging multiple predicted fragments that correspond to the same reference segment. This prevents over-splitting or over-merging from being penalized as a semantic error.
Unmatched reference segments are retained and scored as missing in the corresponding metrics. Detailed matching thresholds and merging rules are provided in the released evaluation code.

\subsection{Metric definitions}
Let $\mathcal D$ be the evaluation set, $\mathcal D^+$ its demand-bearing subset, and $G_x$ the reference segments of scene $x$. We distinguish scene scores $s_m(x)$ (M2--M5) from dataset scores $M_m$ (M1--M5).

\noindent\textbf{M1: Demand detection.} For each scene $x\in\mathcal D$, let $y_x$ and $\hat y_x$ be the reference and predicted demand-presence labels. With one-vs-rest counts $\mathrm{TP}_c$, $\mathrm{FP}_c$, and $\mathrm{FN}_c$ for each class $c\in\{0,1\}$,
\begin{equation}
  M_1 = \frac{1}{2}\sum_{c\in\{0,1\}}
  \frac{2\mathrm{TP}_c}{2\mathrm{TP}_c+\mathrm{FP}_c+\mathrm{FN}_c},
  \qquad
  \mathrm{FTR}=\frac{|\{x\in\mathcal D:\,y_x=0,\ \hat y_x=1\}|}
  {|\{x\in\mathcal D:\,y_x=0\}|}.
  \label{eq:detection_metrics}
\end{equation}
The false-trigger rate is reported separately and does not enter Avg.

\noindent\textbf{M2: Key-point coverage.} Let $K_g$ be the frozen key-point set for reference segment $g$. The LLM judge assigns a hit decision $h_{gk}\in\{0,1\}$ to each point $k$ using the concatenated predicted intent and required-context descriptions. We count each distinct point once, regardless of tier labels, and average the per-segment hit rates equally within each scene:
\begin{equation}
  s_2(x)=\frac{1}{|G_x^K|}\sum_{g\in G_x^K}
  \frac{1}{|K_g|}\sum_{k\in K_g}h_{gk},
  \qquad G_x^K=\{g\in G_x:|K_g|>0\},
  \label{eq:keypoint_coverage}
\end{equation}
where $h_{gk}=0$ for every point in an unmatched reference segment. Thus a missed demand cannot improve coverage by removing its points from the denominator.

\noindent\textbf{M3: Segment localization.} Write $m(g)$ for the prediction matched to reference segment $g$ and set its IoU to zero when no match exists. The per-scene score averages IoU over all reference segments:
\begin{equation}
  s_3(x)=\frac{1}{|G_x|}\sum_{g\in G_x}
  \operatorname{IoU}\!\left(m(g),g\right).
  \label{eq:localization}
\end{equation}
Span precision, recall, F1, and the start- and end-time mean absolute errors are retained as diagnostics but do not enter Avg.

\noindent\textbf{M4: Transcript quality.} Let $E_g$ be the Levenshtein edit count and $L_g$ the reference length after case folding and whitespace and punctuation normalization. Characters are used for a CJK-dominant reference segment and words otherwise; an unmatched reference is compared with an empty hypothesis. We pool edit counts and reference lengths across segments, so longer transcripts carry more weight within a scene:
\begin{equation}
  s_4(x)=\max\!\left(0,1-\frac{\sum_{g\in G_x}E_g}{\sum_{g\in G_x}L_g}\right).
  \label{eq:transcript_quality}
\end{equation}

\noindent\textbf{M5: User profile.} Let $F_x$ contain user-in-frame, gender, and age group for audio--visual scenes, and gender and age group for audio-only scenes. Write $u_{gf}$ and $\hat u_{gf}$ for the reference and predicted values of field $f$ for segment $g$. The score averages accuracy over fields and reference segments, with unmatched segments scored as incorrect:
\begin{equation}
  s_5(x)=\frac{1}{|F_x|}\sum_{f\in F_x}\frac{1}{|G_x|}
  \sum_{g\in G_x}\mathbf{1}\!\left[\hat u_{gf}=u_{gf}\right].
  \label{eq:user_profile}
\end{equation}

\subsection{Aggregation and key-point accounting}
For $m\in\{2,3,4,5\}$, let $\mathcal D_m^+\subseteq\mathcal D^+$ contain the positive scenes on which the dimension score is defined. Each such scene receives equal weight in the reported score, $M_m=|\mathcal D_m^+|^{-1}\sum_{x\in\mathcal D_m^+}s_m(x)$. M1 is defined by Eq.~\ref{eq:detection_metrics} over both positive and no-demand scenes. Let $A$ be the set of available dimensions for a model. Avg.\ is
\begin{equation}
  \mathrm{Avg}=\frac{\sum_{m\in A}w_mM_m}{\sum_{m\in A}w_m},
  \qquad (w_1,w_2,w_3,w_4,w_5)=(\num{benchmark_main.weight.m1},\num{benchmark_main.weight.m2},\num{benchmark_main.weight.m3},\num{benchmark_main.weight.m4},\num{benchmark_main.weight.m5}).
  \label{eq:overall}
\end{equation}
For no-demand scenes, only M1 and the separately reported false-trigger rate are defined.

Tier labels indicate whether a point belongs to demand semantics (T1),
contextual grounding (T2), or both. M2 counts their union once. Separately,
the source label assigns each attributed point to an evidence channel: the spoken
request, visual evidence, audio evidence, dialogue history, or other context.
Points with no assigned channel remain in the overall reference denominator but
are excluded from channel-specific rates. App.~\ref{app:keypoint_detail} gives the
channel analysis and its population accounting.

\FloatBarrier

\subsection{Model output and scoring walkthrough}
\label{app:walkthrough}

The following released example shows the reference annotation, a stored model
prediction, the judge input, and the resulting scene-level scores. The judge
configuration is specified in App.~\ref{app:judge_configuration}; the final
benchmark aggregation follows Eq.~\ref{eq:overall}.

% Manually maintained scoring walkthrough; edit locally or on Overleaf.
\paragraph{Reference annotation.} Scene \texttt{avp\_gen\_000139} contains one reference
segment at \numlit{3510}--\numlit{5390}\,ms. The user says
\emph{``Hey, agent, mute that sound.''} The structured intent is
\emph{``The user wants to mute the electronic notification sound currently playing''}. Its required context is:
\begin{description}[leftmargin=1.2em,itemsep=2pt,topsep=3pt]
\item[\texttt{audio}] The device in the close foreground emits a short notification chime
\item[\texttt{visual}] At the bottom edge of the frame, a hand holds a glowing smartphone in landscape orientation, which is the source device emitting the notification chime
\end{description}
\begingroup\raggedright
The reference user profile is \texttt{user\_in\_frame=true, user\_gender=male, user\_age\_group=adult}.
\par\endgroup

\paragraph{Model prediction and judge input.} Seed 2.0 Lite predicts a demand at
\numlit{3600}--\numlit{5500}\,ms, with transcript
\emph{``Hey agent, mute that sound.''} The scorer concatenates its predicted intent and
required-context descriptions into the following candidate text:
\emph{``The user wants the AI agent to mute the ongoing notification sound from his mobile phone. \textbar \,
audio: A repeated beeping notification sound from the user's phone is playing.''}

The judge receives this candidate text together with the reference key points listed
below. The reference transcript and context-necessity notes are not appended to the
candidate text.

\begingroup\raggedright
The predicted user profile is \texttt{user\_in\_frame=false, user\_gender=male, user\_age\_group=adult}.
\par\endgroup

\paragraph{Stored key-point verdicts.}
\begin{center}\small
\begin{tabularx}{\linewidth}{@{}l l l X@{}}
\toprule
 & Evidence channel & Verdict & Key point \\
\midrule
kp01 & \texttt{intent} & hit & The user wants to mute the electronic notification sound currently playing \\
kp02 & \texttt{required\_context.audio} & hit & The sound to be muted is the notification chime from the nearby device \\
kp03 & \texttt{required\_context.visual} & miss & The sound source is the smartphone held by a hand at the bottom of the frame \\
\bottomrule
\end{tabularx}
\end{center}
The stored judge marks 2 of the 3 points as hits, giving this scene an M2
score of $2/3 = 0.67$. Although the prediction names a mobile phone, the stored judge
marks kp03 as a miss because it omits the hand-held device and its position at the
bottom of the frame. These verdicts concern coverage
of the reference points in the predicted text; they do not directly measure perception.

\paragraph{Scene-level scores.} The temporal IoU is 0.90, with
start- and end-time errors of \numlit{90}\,ms and
\numlit{110}\,ms. The normalized transcripts match exactly. Gender and age group
are correct, while \texttt{user\_in\_frame} is incorrect, yielding two correct profile
fields out of three. The scene-level scores are:
\begin{center}\small
\begin{tabular}{@{}l l c@{}}
\toprule
 & Dimension & Score \\
\midrule
M2 & Key-point Coverage & 0.67 \\
M3 & Segment Localization & 0.90 \\
M4 & Transcript Quality & 1.00 \\
M5 & User profile & 0.67 \\
\bottomrule
\end{tabular}
\end{center}
Demand detection is correct for this scene. The reported M1, however, is macro-F1
computed across demand and no-demand scenes. Consequently, the benchmark Avg. is
computed from the dataset-level dimension scores in Eq.~\ref{eq:overall}, rather
than by treating this scene's detection correctness as M1.

\Needspace{70pt}
\section{Benchmark composition and taxonomy}
\label{app:release}
\label{app:taxonomy}

\subsection{Dataset statistics}
\label{app:stats}
Table~\ref{tab:stats} gives the composition of the released dataset. Its per-axis distribution is shown in
Fig.~\ref{fig:taxonomy_coverage}.

\begin{table}[!htbp]
  \centering
  \caption{Composition of the released dataset.}
  \label{tab:stats}
  \small
  \setlength{\tabcolsep}{6pt}
  \begin{tabular}{@{}lc@{}}
    \toprule
    Property & Main split \\
    \midrule
    Total scenes                          & \num{release_stats.main.n_scenes} \\
    Audio--visual / audio-only            & \num{release_stats.main.n_av} / \num{release_stats.main.n_ao} \\
    Chinese / English                     & \num{release_stats.main.n_zh} / \num{release_stats.main.n_en} \\
    Positive / no-demand negative         & \num{release_stats.main.n_positive} / \num{release_stats.main.n_negative} \\
    Generated / human-recorded interactions        & \num{release_stats.main.n_generated} / \num{release_stats.main.n_recorded} \\
    Multi-turn interactions                     & \num{release_stats.main.n_multiclip} \\
    \bottomrule
  \end{tabular}
\end{table}

\FloatBarrier
\subsection{Positive-demand scenes: six axes}
\label{app:taxonomy_positive}
Fig.~\ref{fig:taxonomy_coverage} summarizes the taxonomy and its distribution in the released dataset. Axes~\numlit{1}--\numlit{4}
describe construction challenges; Axes~\numlit{5}--\numlit{6} describe task and
domain coverage. Categories are not an ordered difficulty scale. The Axis~\numlit{1}
definitions below use the audio--visual vocabulary. Audio-only scenes instead use
\numlit{1.1} for language-channel demands and \numlit{1.2} for acoustic-event
dependence. Their Axis~\numlit{5.2} denotes acoustic perception and recognition;
the remaining demand types retain their meanings.

\begin{taxonomybox}{Axis~\numlit{1}: Modality Dependency}
\small
\textbf{Scope.} This axis describes which input modalities are necessary to infer the user's intent.
\begin{description}[style=unboxed,leftmargin=0pt,itemindent=0pt,labelsep=0.35em,itemsep=2pt,topsep=4pt]
  \item[\textbf{\numlit{1.1} Language-channel.}] The user's intent requires no visual or nonverbal audio grounding. It may still depend on textual dialogue history, which is classified independently on Axis~\numlit{2}. Typical evidence is a direct request with verbally stated arguments or a textual back-reference to an option named earlier.
  \item[\textbf{\numlit{1.2} Vision-grounded.}] The user's intent depends on visible objects, people, spatial relations, written text, gestures, or visual events in the current scene. Typical evidence includes deictic expressions such as ``this one'', ``that person'', or ``the item on the left''.
  \item[\textbf{\numlit{1.3} Audio-grounded.}] The user's intent depends on nonverbal sounds, environmental audio events, speaker identity, voice characteristics, or acoustic cues. Typical evidence includes asking about a sound, reacting to an alarm, or referring to someone by voice.
  \item[\textbf{\numlit{1.4} Vision--Audio Joint.}] The user's intent requires both visual and audio evidence; either modality alone would leave the intent incomplete or ambiguous. For example, ``Which of these instruments makes that sound?'' requires identifying the instrument from the audio and locating it among the visible candidates.
\end{description}
\end{taxonomybox}

\begin{taxonomybox}{Axis~\numlit{2}: Contextual Disambiguation}
\small
\textbf{Scope.} This axis describes what kind of context is needed to resolve the user's intent beyond the literal wording of the current utterance.
\begin{description}[style=unboxed,leftmargin=0pt,itemindent=0pt,labelsep=0.35em,itemsep=2pt,topsep=4pt]
  \item[\textbf{\numlit{2.1} Self-contained Current Context.}] The current utterance and scene state determine the intent without recalling earlier events or tracking changes over time.
  \item[\textbf{\numlit{2.2} Temporal Event Dependency.}] The intent requires recalling a perceptual event or tracking its timing, sequence, or resulting state change, before or during the current turn.
  \item[\textbf{\numlit{2.3} Dialogue History Dependency.}] The intent depends on prior conversational context, especially previous user--assistant exchanges.
  \item[\textbf{\numlit{2.4} Speaker Identity Dependency.}] The intent depends on identifying which person or voice is being referred to, or on distinguishing among multiple speakers.
  \item[\textbf{\numlit{2.5} Affective State Dependency.}] The intent depends on recognizing emotion, attitude, stress, fatigue, urgency, hesitation, or another affective state.
\end{description}
\textbf{Boundary note.} Axis~\numlit{2.2} concerns perceptual event history, whereas Axis~\numlit{2.3} concerns discourse history. A scene may contain both, but the label reflects the primary source of disambiguation.
\end{taxonomybox}

\begin{taxonomybox}{Axis~\numlit{3}: Intent Expression Form}
\small
\textbf{Scope.} This axis describes how the user's intent is expressed linguistically.
\begin{description}[style=unboxed,leftmargin=0pt,itemindent=0pt,labelsep=0.35em,itemsep=2pt,topsep=4pt]
  \item[\textbf{\numlit{3.1} Direct Instruction.}] The user states an explicit command or request for the assistant to perform an action.
  \item[\textbf{\numlit{3.2} Direct Question.}] The user expresses the intent as an explicit question.
  \item[\textbf{\numlit{3.3} Indirect or Descriptive Request.}] The user describes a situation, problem, preference, or state from which the intended assistance must be inferred.
  \item[\textbf{\numlit{3.4} Hesitation or Self-correction.}] The user's expression includes hesitation, revision, restart, or correction that must be resolved.
  \item[\textbf{\numlit{3.5} Interrupted or Incomplete Request.}] The utterance is cut off, interrupted, or incomplete, but context still determines a valid demand. R3 (App.~\ref{app:negative_taxonomy}) applies when no actionable demand remains.
  \item[\textbf{\numlit{3.6} Compound or Multi-intent Request.}] The user expresses multiple related or parallel intents in one turn.
\end{description}
\end{taxonomybox}

\begin{taxonomybox}{Axis~\numlit{4}: Acoustic Environment}
\small
\textbf{Scope.} This axis describes the acoustic conditions under which the user's demand is expressed. It captures speech intelligibility and audio interference, not whether audio is part of the task content.
\begin{description}[style=unboxed,leftmargin=0pt,itemindent=0pt,labelsep=0.35em,itemsep=2pt,topsep=4pt]
  \item[\textbf{\numlit{4.1} Clean Near-field Speech.}] The user's speech is near-field, clear, and minimally affected by interference.
  \item[\textbf{\numlit{4.2} Background Noise.}] Non-speech environmental noise is present but does not fully obscure the user's speech.
  \item[\textbf{\numlit{4.3} Competing Speech.}] Other human speech overlaps with or competes against the user's speech.
  \item[\textbf{\numlit{4.4} Low-volume or Whispered Speech.}] The user intentionally speaks softly, whispers, or has significantly reduced volume.
  \item[\textbf{\numlit{4.5} Media Playback Interference.}] Media audio, such as television, music, phone playback, navigation audio, or public announcements, interferes with the user demand or is semantically relevant to it.
\end{description}
\textbf{Boundary note.} Axis~\numlit{1} asks whether audio is part of the intent content; Axis~\numlit{4} asks whether the acoustic environment affects how the intent is expressed or perceived.
\end{taxonomybox}

\begin{taxonomybox}{Axis~\numlit{5}: Demand Type}
\small
\textbf{Scope.} This axis describes the functional type of assistance requested by the user.
\begin{description}[style=unboxed,leftmargin=0pt,itemindent=0pt,labelsep=0.35em,itemsep=2pt,topsep=4pt]
  \item[\textbf{\numlit{5.1} Information Seeking and Retrieval.}] The user asks the assistant to retrieve, look up, or provide factual information.
  \item[\textbf{\numlit{5.2} Visual Perception and Recognition.}] The user asks the assistant to perceive, identify, describe, compare, or reason about visible content.
  \item[\textbf{\numlit{5.3} Action Execution.}] The user asks the assistant to perform an operation or control an external function.
  \item[\textbf{\numlit{5.4} Guidance and Explanation.}] The user asks for step-by-step guidance, procedural help, troubleshooting, or explanation.
  \item[\textbf{\numlit{5.5} Recommendation and Decision Support.}] The user asks the assistant to compare options, make a recommendation, or support a choice.
  \item[\textbf{\numlit{5.6} Monitoring and Reminder.}] The user asks the assistant to monitor a condition, detect a change, or remind them later.
  \item[\textbf{\numlit{5.7} Recording and Summarization.}] The user asks the assistant to record, transcribe, summarize, or preserve information.
  \item[\textbf{\numlit{5.8} Social and Emotional Interaction.}] The user seeks social support, emotional response, interpersonal advice, or empathetic interaction.
  \item[\textbf{\numlit{5.9} Translation and Language Conversion.}] The user asks the assistant to translate, paraphrase, rewrite, or convert language.
  \item[\textbf{\numlit{5.10} Content Creation.}] The user asks the assistant to generate new content.
\end{description}
\end{taxonomybox}

\begin{taxonomybox}{Axis~\numlit{6}: Scenario Domain}
\small
\textbf{Scope.} This axis describes the everyday domain or activity setting in which the user demand occurs.
\begin{description}[style=unboxed,leftmargin=0pt,itemindent=0pt,labelsep=0.35em,itemsep=2pt,topsep=4pt]
  \item[\textbf{\numlit{6.1} Home and Daily Life.}] Domestic routines and everyday household activities.
  \item[\textbf{\numlit{6.2} Cooking and Dining.}] Food preparation, eating, kitchen use, restaurants, and meal-related decisions.
  \item[\textbf{\numlit{6.3} Shopping and Commerce.}] Purchasing, product comparison, retail settings, payment, and consumer decisions.
  \item[\textbf{\numlit{6.4} Travel and Navigation.}] Trip planning, wayfinding, route selection, transportation logistics, and location search.
  \item[\textbf{\numlit{6.5} Tourism and Exploration.}] Sightseeing, visiting attractions, cultural exploration, and travel discovery.
  \item[\textbf{\numlit{6.6} Learning and Education.}] Studying, teaching, tutoring, training, and knowledge acquisition.
  \item[\textbf{\numlit{6.7} Office and Collaboration.}] Workplace tasks, meetings, documents, coordination, and professional communication.
  \item[\textbf{\numlit{6.8} Sports and Fitness.}] Exercise, training, physical activity, performance tracking, and fitness guidance.
  \item[\textbf{\numlit{6.9} Health and Medical Care.}] Health monitoring, medication, symptoms, care settings, and medical assistance.
  \item[\textbf{\numlit{6.10} DIY and Repair.}] Assembly, maintenance, repair, tool use, and practical troubleshooting.
  \item[\textbf{\numlit{6.11} Driving and In-vehicle Contexts.}] Driving, vehicle use, road context, navigation, and in-car assistance.
  \item[\textbf{\numlit{6.12} Creative Work and Content Production.}] Creative planning, media production, design, writing, and editing.
  \item[\textbf{\numlit{6.13} Social Gatherings.}] Social interaction in group settings, parties, visits, and interpersonal coordination.
  \item[\textbf{\numlit{6.14} Outdoor and Nature.}] Outdoor environments, nature activities, weather, and field observations.
  \item[\textbf{\numlit{6.15} Childcare and Caregiving.}] Caring for children, elderly people, patients, or others who need assistance.
  \item[\textbf{\numlit{6.16} Fashion and Beauty.}] Clothing, appearance, grooming, cosmetics, and style decisions.
  \item[\textbf{\numlit{6.17} Public Transit and Commuting.}] Public transportation, commuting routines, stations, transfers, and announcements.
  \item[\textbf{\numlit{6.18} Entertainment and Leisure.}] Games, media, hobbies, recreation, and leisure activities.
  \item[\textbf{\numlit{6.19} Safety and Emergency.}] Risk, danger, urgent response, accident prevention, and emergency assistance.
\end{description}
\end{taxonomybox}

\Needspace{5\baselineskip}
\subsection{No-demand scenes: signal type and invalidation reason}
\label{app:negative_taxonomy}
No-demand scenes contain cues that resemble a request, but no valid demand is
directed to the conversational assistant. The two dimensions specify the
demand-like signal that creates false-trigger risk and the reason it is invalid.

\begin{taxonomybox}{Dimension N1: Demand-like Signal Type}
\small
\textbf{Scope.} This dimension identifies the observable cue that may be mistaken for a user demand.
\begin{description}[style=unboxed,leftmargin=0pt,itemindent=0pt,labelsep=0.35em,itemsep=2pt,topsep=4pt]
  \item[\textbf{S1 Linguistic Signal.}] The utterance surface form resembles a request, question, imperative, complaint, or help-seeking expression.
  \item[\textbf{S2 Semantic Signal.}] The utterance mentions a function, task, or topic the assistant could help with, even if the user is not actually requesting assistance.
  \item[\textbf{S3 Prosodic Signal.}] The speech sounds as if it may be directed to the assistant because of voice quality, direction, distance, or command-like prosody.
  \item[\textbf{S4 Interactive Signal.}] The dialogue structure resembles a user--assistant request--response turn, even though no new demand is being made.
  \item[\textbf{S5 Contextual Signal.}] The scene presents a plausible opportunity for the assistant to help, but the user is not actually requesting assistance.
\end{description}
\end{taxonomybox}

\begin{taxonomybox}{Dimension N2: Invalidation Reason}
\small
\textbf{Scope.} This dimension identifies why a demand-like signal is invalid.
\begin{description}[style=unboxed,leftmargin=0pt,itemindent=0pt,labelsep=0.35em,itemsep=2pt,topsep=4pt]
  \item[\textbf{R1 Wrong Addressee.}] The utterance comes from the user, but is addressed to another person, a pet, or an object, or has no intended recipient; it is not addressed to the assistant.
  \item[\textbf{R2 Wrong Source.}] The apparent request is produced by media playback, another speaker, an echo, or another background source, rather than by the user.
  \item[\textbf{R3 Incomplete or Withdrawn Demand.}] A possible demand is started but not completed, is withdrawn or abandoned, or is resolved by the user before it becomes actionable, leaving no actionable demand to the assistant.
  \item[\textbf{R4 Non-real Intent.}] The utterance does not express the user's current real intent; it is hypothetical, joking, exaggerated, quoted, read aloud, rehearsed, remembered, or discussed as someone else's words.
  \item[\textbf{R5 Function Mention Not Directed as a Request.}] The utterance discusses, evaluates, teaches, recalls, or gives feedback about an assistant capability, but does not request its use.
\end{description}
\end{taxonomybox}

\Needspace{5\baselineskip}
A negative remains demand-like in its surface cues; invalidity depends on the
addressee, source, dialogue function, pragmatic status, or scene context. In Fig.~\ref{fig:taxonomy_coverage}, Wrong Addressee and Wrong Source retain their full names; Incomplete, Non-real, and Function Mention abbreviate Incomplete or Withdrawn Demand, Non-real Intent, and Function Mention Not Directed as a Request, respectively.

\FloatBarrier

\section{Construction and human verification}
\label{app:generation}

\subsection{Target selection and seed sampling}
\paragraph{Scene seeds and positive targets.}
Each corpus seed provides an everyday scene category and topic, such as
``Office \& Workplace / office plant care techniques''.
The generator uses this pair as a starting setting and subject, and combines it
with a challenge target from Axes~\numlit{1}--\numlit{4}. For Axis~\numlit{5}, we
supply candidate demand types; the generator chooses one that fits the developing
scene when constructing the user's intent. Axis~\numlit{6} is assigned from the
completed scenario's domain. Thus, the seed category guides scene creation while
Axis~\numlit{6} describes the resulting scene, and the seed topic does not fix an
Axis~\numlit{5} demand type.

\paragraph{Coverage and seed sampling.}
For positive targets, the selector prioritizes uncovered admissible
Axes~\numlit{1}--\numlit{4} combinations, then those with the lowest counts,
breaking ties randomly. The Axis~\numlit{5} candidates include a least-used demand
type and alternatives weighted toward less-used types. Seeds are sampled
independently of these targets, uniformly among the least-used scene--topic pairs
in the selected language. Language selection follows the 1:1
Chinese:English ratio. 

\paragraph{Negative targets.}
No-demand scenes use the same scene--topic seeds with a signal--invalidation
target (App.~\ref{app:negative_taxonomy}), without an Axis~\numlit{5} demand type.
The sampler first draws an invalidation reason uniformly, then a pattern uniformly
within that reason; the pattern supplies its demand-like signal type. Acoustic conditions
are varied separately, and patterns that depend on a previous assistant response
request dialogue history.

\subsection{Script generation and review}
\paragraph{From seed to interaction.}
For positives, a brainstorm combines the seed and taxonomy target into a scene
concept, the concrete challenge, and an explanation of why it requires context.
The cascade then specifies the physical setting, people, objects, and sound
sources. Its usual path fixes the intended demand and decisive evidence, builds
any required event or dialogue history around them, and finally writes the user's
utterance. Patterns that require joint intent--utterance construction instead
build history first and generate the two together. For negatives, a separate
cascade builds a scene skeleton and any required history, then instantiates the
demand-like cue and the evidence that invalidates it. In a Wrong Source scene,
for example, the apparent request must remain attributed to the playback or
other speaker that produced it.

\paragraph{Consistency and validity.}
Programmatic checks verify speaker roles, required history or event traces, and
explicit pattern constraints. An LLM reviewer checks plausibility, causal
coherence, taxonomy alignment, and support for the intended interpretation in the
script. Negative validation asks whether an unmet need is present and whether it
is directed to the assistant, while checking that the intended misleading signal
is actually present. 

\paragraph{Utterance-only screening.}
For positives, the text-only discriminator uses a single LLM call with two
instructed steps: infer intent from the scripted user utterance, then compare it
with the supplied intended demand. It returns a
\mbox{\texttt{match}, \texttt{partial}, or \texttt{mismatch}} verdict, together with
the inferred intent, confidence, and reasoning.
Dialogue history, speaker metadata, and media evidence are withheld. A full match
rejects a candidate intended to require contextual evidence, including a
language-channel candidate whose difficulty should come from dialogue history.
For negatives with user speech, the reverse screen uses the intent-prediction
step and rejects candidates whose wording alone is judged to contain no demand;
cases without a user utterance skip this check. These are script-stage screens,
not tests of exclusive evidence dependence in the realized media.

\subsection{Media production and annotation}
\paragraph{Rendering and performance guides.}
Accepted timelines guide synthetic media production for both modalities.
Multi-clip scenes
are rendered sequentially with earlier clips as video references for consistent
people, voices, and surroundings. Audio-only production retains the audio track.
Performance guides for human-recorded interactions specify roles, dialogue,
event order, and essential cues, allowing practical setting or prop substitutions
that preserve the challenge.
Single-turn wording may vary while retaining the intended ambiguity; multi-turn
guides preserve user lines that match prepared assistant replies. Both routes
enter media-grounded annotation to account for departures from the script.

\paragraph{Media-derived draft and script reconciliation.}
An initial annotation uses multimodal media captions and word-level ASR,
without the script's intended annotation. Captions supply scene and speaker
semantics; ASR supplies spoken wording and timing. The next stage checks scripted
speech against ASR and compares the draft with the scenario specification.
Significant disagreements become observable questions for targeted MLLM inspection
of the media. The draft, actual script, ASR, and resulting observations are then
reconciled. In multi-clip scenes, earlier user
clips and assistant replies provide history; annotation and timing target the
final user clip.

\paragraph{Timing, relevance, and verification.}
Segment transcripts are matched to ASR words, and their start and end times are
aligned to word boundaries. Rule-based and LLM checks remove context that does
not materially affect the appropriate response, while retaining referents needed
to interpret the demand. Schema and cross-field checks, together with an LLM
review, assess demand presence, temporal grounding, intent, context quality, and
transcript coverage before key points are constructed.

\paragraph{Key-point construction and grounding.}
Verified intent and context are decomposed into atomic, response-relevant key
points, with the transcript helping identify their evidence source. Each point
must be expressible from the structured intent or an existing required-context
description. Points representing the requested task, required context, or both
receive the corresponding labels described in App.~\ref{app:evaluation}.
An MLLM then checks the actual media against the points, context, and intent.
Unsupported details are removed or narrowed to supported content. If this
changes the intent, key points are regenerated and checked against the media
again.

\subsection{Human verification}
\label{app:human_verification}

\paragraph{Media quality.}
Human reviewers are asked to assess perceptual realism and playback usability of the generated media. Generated clips were
rejected for conspicuous synthesis artifacts, such as mismatched lip movements,
sudden object appearances or disappearances, and distorted faces or physically
implausible motion. Across both media sources, rejection criteria also included
corrupted video or audio, missing sound, and severe playback stalls. Unintended
assistant-response audio and crosstalk that made speakers indistinguishable were
also grounds for rejection.

\paragraph{Annotation correctness.}
Reviewers checked each applicable machine-produced annotation field against the
realized interaction: demand presence, demand spans, transcript, structured intent,
required context, user profile, and key points. They verified speech and timing
against the media and checked semantic annotations for unsupported details,
irrelevant context, and missing key points, using dialogue history where needed.
Generated scenes were retained only when all applicable fields passed review.
For human-recorded interactions, experts corrected erroneous fields before
finalization to accommodate departures from the performance guide.

Figure~\ref{fig:human_review_interface} illustrates the review controls using
a generated benchmark example in place of recorded media. Two consecutive views
show the complete annotation form in its original field order, with the reference
annotation expanded. The example uses the released English annotation; selected
controls show interface defaults rather than a saved human verdict.

\begin{figure}[p]
  \centering
  \includegraphics[page=1,width=\linewidth,height=0.88\textheight,keepaspectratio]{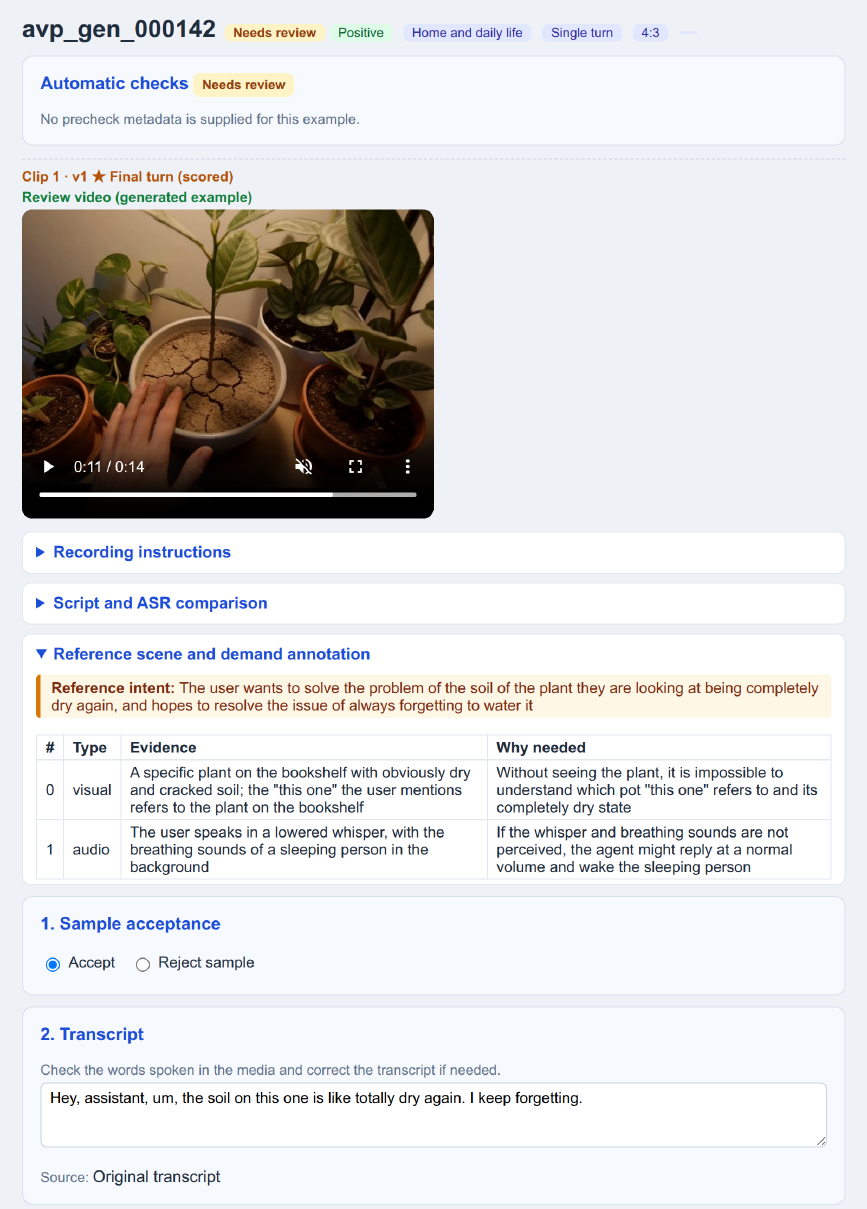}
  \caption{Annotation-review interface, shown in two consecutive views.
  (a) Generated media, reference annotation, sample acceptance, and transcript.
  The original field order is retained; selected options are interface defaults.}
  \label{fig:human_review_interface}
\end{figure}

\begin{figure}[p]
  \ContinuedFloat
  \centering
  \includegraphics[page=2,width=\linewidth,height=0.88\textheight,keepaspectratio]{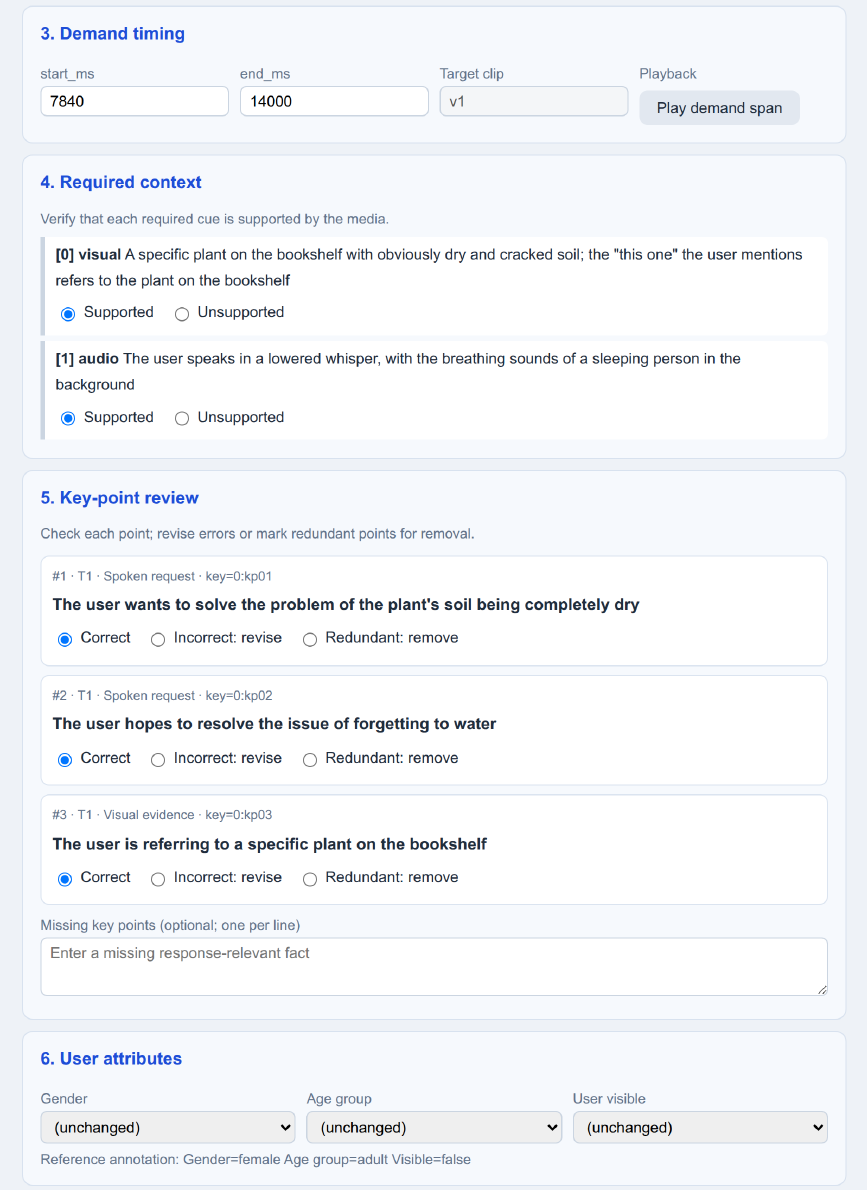}
  \caption{Annotation-review interface (continued).
  (b) Demand time spans, required context, all key points, and user profile fields
  from the same generated sample.}
  \label{fig:human_review_interface_fields}
\end{figure}

\FloatBarrier

\section{Experimental setup}
\label{app:experimental_setup}

\subsection{Input preparation and evaluation}

User-turn media are presented in chronological order, with preceding assistant
replies supplied as text between turns. Models predict the demand in the final
user clip; preceding clips and assistant replies provide interaction history.
Predicted start and end times are local to the final clip. The model-facing
instruction is specialized to audio--visual or audio-only input, and predictions
use the same demand-decision schema and scoring procedure.
The full prediction instruction is provided in App.~\ref{app:eval_prompts}.

\subsection{Text-only baseline}
\label{app:text_only_baseline}

The text-only baseline uses GPT-\numlit{5.4} with the same modality-specific task
instruction (App.~\ref{app:eval_prompts}) as the native systems.
Each media block is replaced in place by its ASR transcript, sentence-level
timestamps, and optional speaker identifiers.
Multi-turn inputs retain the original clip order, turn labels, and preceding
assistant reply text.
Each replacement block explicitly states that native media are unavailable.
The baseline is evaluated with the same parser, segment matching, and metrics
as the native systems.

\textbf{System message:} The full prediction instruction in
App.~\ref{app:eval_prompts}, including the demand annotation schema and field
constraints; the audio-only variant follows App.~\ref{app:eval_prompts_audio_only}.

\textbf{User message:} The panels below show single-turn and multi-turn
input templates, followed by input examples for a positive and a no-demand scene.
Demand labels reflect the full scenes and may not be recoverable from the ASR
transcripts alone. The templates retain the original audio--visual task wording,
including ``video'', while replacing media with ASR text. For audio-only input, ``video'' and
``audio or video'' are replaced with ``audio''.

Braces mark placeholders. The \texttt{duration\_note} adds
`` (clip duration \{duration\_ms\} ms)'', and \texttt{speaker\_id} denotes the speaker
identifier returned by ASR.

The multi-turn template shows the complete ASR block for each clip, including
its no-media notice. ASR blocks and assistant replies appear in chronological
order; longer histories repeat the same pattern. If no speech is recognized,
the no-speech notice replaces that clip's ASR block. The closing instruction
follows the final block in either case.

\begingroup
% Local prompt-panel styles; scoped by setup.tex, with no manuscript preamble edit.
\definecolor{abPromptInk}{RGB}{48,69,92}
\definecolor{abPromptAccent}{RGB}{63,99,136}
\definecolor{abPromptRule}{RGB}{211,223,234}
\definecolor{abPromptFill}{RGB}{248,250,253}
\definecolor{abPromptTitle}{RGB}{234,240,247}
\newtcolorbox{abprompt}[1]{%
  enhanced, breakable,
  colback=abPromptFill,
  colframe=abPromptRule,
  colbacktitle=abPromptTitle,
  coltitle=abPromptInk,
  title={#1},
  title after break={#1\ (continued)},
  fonttitle=\small\bfseries,
  fontupper=\small,
  boxrule=0.4pt,
  titlerule=0pt,
  borderline west={1.6pt}{0pt}{abPromptAccent},
  arc=1pt,
  left=3mm,right=3mm,top=2mm,bottom=2mm,
  before skip=7pt,after skip=7pt,
  before upper={\setlength{\parindent}{0pt}\setlength{\parskip}{3pt}},
}
\newcommand{\ablabel}[1]{{\sffamily\footnotesize\bfseries\color{abPromptAccent}#1}}
\newcommand{\abmedia}{{\color{abPromptInk!65}\itshape [video input]}}
\newcommand{\abdemand}{{\color{abPromptAccent}\bfseries Demand:}}

% Generated by paper_overleaf_assets/experiments/text_only_baseline/build_prompt_panels.py; do not hand-edit.

\begin{abprompt}{User-input templates}
\ablabel{Single clip}\par
[No audio or video is provided for this clip. Below is the verbatim ASR transcript of its speech, with millisecond timestamps local to this clip\{duration\_note\}.]\par
[\{begin\_ms\}-\{end\_ms\} ms] speaker \{speaker\_id\}: \{ASR sentence\}\par
...\par
Annotate the user's demand in this video and output the demand-decision JSON.\par
\medskip\ablabel{Replacement ASR block when no speech is recognized}\par
[No audio or video is provided for this clip, and automatic speech recognition returned no speech for it.]\par
\end{abprompt}

\begin{abprompt}{User-input template with dialogue history}
You will receive multiple video clips from the same scene in chronological order. Together they form the dialogue history; the last video clip is the current user turn to annotate:\par
[Turn 1 · user video]\par\nopagebreak[4]
[No audio or video is provided for this clip. Below is the verbatim ASR transcript of its speech, with millisecond timestamps local to this clip\{duration\_note\}.]\par\nopagebreak[4]
[\{begin\_ms\}-\{end\_ms\} ms] speaker \{speaker\_id\}: \{ASR sentence\}\par
...\par
[Turn 1 · agent reply] \{preceding assistant reply\}\par
[Turn 2 · user video (final turn, annotate this)]\par\nopagebreak[4]
[No audio or video is provided for this clip. Below is the verbatim ASR transcript of its speech, with millisecond timestamps local to this clip\{duration\_note\}.]\par\nopagebreak[4]
[\{begin\_ms\}-\{end\_ms\} ms] speaker \{speaker\_id\}: \{ASR sentence\}\par
...\par
Annotate only the user's demand in the final video clip and output the demand-decision JSON. Use the preceding video clips and agent replies only as dialogue-history context; do not output segments for them. start\_ms/end\_ms must be local timestamps within the final video clip.\par
\end{abprompt}

\begin{abprompt}{Positive example (Demand present)}
[No audio or video is provided for this clip. Below is the verbatim ASR transcript of its speech, with millisecond timestamps local to this clip (clip duration 9055 ms).]\par
[5040-8640 ms] speaker 0: Set it up, so when he gives the verbal, okay, export this version.\par
Annotate the user's demand in this video and output the demand-decision JSON.\par
\end{abprompt}

\begin{abprompt}{Negative example (No demand)}
[No audio or video is provided for this clip. Below is the verbatim ASR transcript of its speech, with millisecond timestamps local to this clip (clip duration 12051 ms).]\par
[4590-10710 ms] speaker 0: my data was leaked, can this it insurance immediately activate an emergency response for me?\par
[11070-11990 ms] speaker 0: how do i file a claim?\par
Annotate the user's demand in this video and output the demand-decision JSON.\par
\end{abprompt}

\endgroup

\subsection{Semantic scoring}
\label{app:judge_configuration}

We use Qwen3.6-Flash with thinking enabled as the semantic judge in the main
experiments. The judge receives the predicted structured intent concatenated with
the predicted required-context descriptions, together with the reference key
points. Reference transcripts and context-necessity notes are not added to the
candidate answer text. App.~\ref{app:judge_prompt} provides the judge instruction,
and App.~\ref{app:judge_stability} analyzes result stability across judges.

Each distinct reference point is counted once, regardless of tier labels.
Key points in unmatched reference segments count as missed. Segment matching
and score aggregation follow App.~\ref{app:evaluation}.

\FloatBarrier

\section{Supplementary experiments and diagnostics}
\label{app:supplementary}

This appendix provides supplementary analyses of request and context coverage,
LLM judge stability, and positive-demand challenges. It then presents the blinded
comparison of responses generated with and without demand annotations and closes
with qualitative error cases.

\subsection{Request and context key-point coverage}
\label{app:keypoint_detail}

We compare recovery of spoken requests and the context needed to interpret them,
extending the analysis in \S\ref{sec:where_models_miss} and Figure~\ref{fig:keypoint_types}.

\paragraph{Population and counting.}
We analyze \num{keypoint_types.pop.av.n_scenes} audio--visual demand scenes scored
by all four systems, containing \num{keypoint_types.points.av.total} distinct
reference key points. Each point counts once; points in unmatched reference
segments count as misses. Context comprises visual, audio, dialogue-history,
and other context points (emotion, activity, identity,
and scene condition). All context pools these categories.

\paragraph{Coverage gaps and sampling uncertainty.}
Table~\ref{tab:channel_gaps} uses the same population and point weighting as
Figure~\ref{fig:keypoint_types}. Coverage is the ratio of recovered to reference
key points in each channel. The gap subtracts context coverage from spoken-request
coverage, in percentage points. Confidence intervals assess sensitivity to the
composition of the sampled scenes.

\paragraph{Scene-bootstrap procedure.}
For each of \num{channel_gaps.analysis.bootstrap_replicates} resamples, we draw
\num{channel_gaps.pop.av.n_scenes} scenes uniformly with replacement from the
common population. All systems and channels use the same draw, and each selected
scene contributes all its key points, including repeated contributions when
selected more than once. Model outputs and judge verdicts remain fixed.

Within each resample, we recompute each channel's coverage from its pooled hit
and reference-point counts, then calculate the request-minus-context gaps.
Their \numlit{2.5}th and \numlit{97.5}th percentiles form the
\numlit{95}\% confidence interval. The Gap column uses the original full sample;
resampling supplies the interval while preserving dependence among each scene's
key points.
% Numeric values are resolved from results/channel_gaps.json.
\begin{table}[htbp]
% Convert stored 0--1 intervals for display; the inner \numci keeps keys visible
% to the number-contract check. These helpers are scoped to this table.
\newcommand{\channelgapinterval}[1]{%
  \begingroup
  \edef\channelgapbounds{#1}%
  \expandafter\channelgapintervalparts\channelgapbounds\relax
  \endgroup}
\def\channelgapintervalparts[#1,#2]\relax{%
  [\pgfmathparse{100*(#1)}%
  \pgfmathprintnumber[fixed,zerofill,precision=1]{\pgfmathresult},
  \pgfmathparse{100*(#2)}%
  \pgfmathprintnumber[fixed,zerofill,precision=1]{\pgfmathresult}]}
\centering
\caption{Point-weighted coverage by evidence channel
(Figure~\ref{fig:keypoint_types}). Gap subtracts each context row's coverage from
the same model's Spoken request coverage; positive values mean lower context
coverage. The \numlit{95}\% confidence interval quantifies the gap's uncertainty
under scene resampling. Both are in percentage points (pp). $n$ counts reference
points; dashes mark the request reference. Gaps use unrounded rates.}
\label{tab:channel_gaps}
\small
\setlength{\tabcolsep}{4pt}
\begin{tabular}{@{}llrccc@{}}
\toprule
System & Channel & \makecell{Reference\\points ($n$)} & \makecell{Coverage\\(\%)} & \makecell{Gap (pp)\\Request $-$ context} & \makecell{\numlit{95}\% confidence\\interval for gap (pp)} \\
\midrule
Gemini 3.1 Pro & Spoken request & \numn{channel_gaps.coverage.gemini.av.intent} & \numpct{channel_gaps.coverage.gemini.av.intent} & \na & \na \\
 & All context & \numn{channel_gaps.gap.gemini.av.intent_minus_context_all} & \numpct{channel_gaps.coverage.gemini.av.context_all} & \numpct{channel_gaps.gap.gemini.av.intent_minus_context_all} & \channelgapinterval{\numci{channel_gaps.gap.gemini.av.intent_minus_context_all}} \\
 & Visual & \numn{channel_gaps.gap.gemini.av.intent_minus_visual} & \numpct{channel_gaps.coverage.gemini.av.visual} & \numpct{channel_gaps.gap.gemini.av.intent_minus_visual} & \channelgapinterval{\numci{channel_gaps.gap.gemini.av.intent_minus_visual}} \\
 & Audio & \numn{channel_gaps.gap.gemini.av.intent_minus_audio} & \numpct{channel_gaps.coverage.gemini.av.audio} & \numpct{channel_gaps.gap.gemini.av.intent_minus_audio} & \channelgapinterval{\numci{channel_gaps.gap.gemini.av.intent_minus_audio}} \\
 & History & \numn{channel_gaps.gap.gemini.av.intent_minus_history} & \numpct{channel_gaps.coverage.gemini.av.history} & \numpct{channel_gaps.gap.gemini.av.intent_minus_history} & \channelgapinterval{\numci{channel_gaps.gap.gemini.av.intent_minus_history}} \\
\addlinespace[3pt]
Qwen3.5-Omni-Plus & Spoken request & \numn{channel_gaps.coverage.qwen_plus.av.intent} & \numpct{channel_gaps.coverage.qwen_plus.av.intent} & \na & \na \\
 & All context & \numn{channel_gaps.gap.qwen_plus.av.intent_minus_context_all} & \numpct{channel_gaps.coverage.qwen_plus.av.context_all} & \numpct{channel_gaps.gap.qwen_plus.av.intent_minus_context_all} & \channelgapinterval{\numci{channel_gaps.gap.qwen_plus.av.intent_minus_context_all}} \\
 & Visual & \numn{channel_gaps.gap.qwen_plus.av.intent_minus_visual} & \numpct{channel_gaps.coverage.qwen_plus.av.visual} & \numpct{channel_gaps.gap.qwen_plus.av.intent_minus_visual} & \channelgapinterval{\numci{channel_gaps.gap.qwen_plus.av.intent_minus_visual}} \\
 & Audio & \numn{channel_gaps.gap.qwen_plus.av.intent_minus_audio} & \numpct{channel_gaps.coverage.qwen_plus.av.audio} & \numpct{channel_gaps.gap.qwen_plus.av.intent_minus_audio} & \channelgapinterval{\numci{channel_gaps.gap.qwen_plus.av.intent_minus_audio}} \\
 & History & \numn{channel_gaps.gap.qwen_plus.av.intent_minus_history} & \numpct{channel_gaps.coverage.qwen_plus.av.history} & \numpct{channel_gaps.gap.qwen_plus.av.intent_minus_history} & \channelgapinterval{\numci{channel_gaps.gap.qwen_plus.av.intent_minus_history}} \\
\addlinespace[3pt]
Seed 2.0 Lite & Spoken request & \numn{channel_gaps.coverage.seed.av.intent} & \numpct{channel_gaps.coverage.seed.av.intent} & \na & \na \\
 & All context & \numn{channel_gaps.gap.seed.av.intent_minus_context_all} & \numpct{channel_gaps.coverage.seed.av.context_all} & \numpct{channel_gaps.gap.seed.av.intent_minus_context_all} & \channelgapinterval{\numci{channel_gaps.gap.seed.av.intent_minus_context_all}} \\
 & Visual & \numn{channel_gaps.gap.seed.av.intent_minus_visual} & \numpct{channel_gaps.coverage.seed.av.visual} & \numpct{channel_gaps.gap.seed.av.intent_minus_visual} & \channelgapinterval{\numci{channel_gaps.gap.seed.av.intent_minus_visual}} \\
 & Audio & \numn{channel_gaps.gap.seed.av.intent_minus_audio} & \numpct{channel_gaps.coverage.seed.av.audio} & \numpct{channel_gaps.gap.seed.av.intent_minus_audio} & \channelgapinterval{\numci{channel_gaps.gap.seed.av.intent_minus_audio}} \\
 & History & \numn{channel_gaps.gap.seed.av.intent_minus_history} & \numpct{channel_gaps.coverage.seed.av.history} & \numpct{channel_gaps.gap.seed.av.intent_minus_history} & \channelgapinterval{\numci{channel_gaps.gap.seed.av.intent_minus_history}} \\
\addlinespace[3pt]
GPT-5.4 + transcript & Spoken request & \numn{channel_gaps.coverage.cascade_asr.av.intent} & \numpct{channel_gaps.coverage.cascade_asr.av.intent} & \na & \na \\
 & All context & \numn{channel_gaps.gap.cascade_asr.av.intent_minus_context_all} & \numpct{channel_gaps.coverage.cascade_asr.av.context_all} & \numpct{channel_gaps.gap.cascade_asr.av.intent_minus_context_all} & \channelgapinterval{\numci{channel_gaps.gap.cascade_asr.av.intent_minus_context_all}} \\
 & Visual & \numn{channel_gaps.gap.cascade_asr.av.intent_minus_visual} & \numpct{channel_gaps.coverage.cascade_asr.av.visual} & \numpct{channel_gaps.gap.cascade_asr.av.intent_minus_visual} & \channelgapinterval{\numci{channel_gaps.gap.cascade_asr.av.intent_minus_visual}} \\
 & Audio & \numn{channel_gaps.gap.cascade_asr.av.intent_minus_audio} & \numpct{channel_gaps.coverage.cascade_asr.av.audio} & \numpct{channel_gaps.gap.cascade_asr.av.intent_minus_audio} & \channelgapinterval{\numci{channel_gaps.gap.cascade_asr.av.intent_minus_audio}} \\
 & History & \numn{channel_gaps.gap.cascade_asr.av.intent_minus_history} & \numpct{channel_gaps.coverage.cascade_asr.av.history} & \numpct{channel_gaps.gap.cascade_asr.av.intent_minus_history} & \channelgapinterval{\numci{channel_gaps.gap.cascade_asr.av.intent_minus_history}} \\
\addlinespace[3pt]
\bottomrule
\end{tabular}
\end{table}

\FloatBarrier

\paragraph{Results and interpretation.}
Across the four systems, request coverage exceeds pooled context coverage by
\numpct{channel_gaps.gap.seed.av.intent_minus_context_all}--\numpct{channel_gaps.gap.cascade_asr.av.intent_minus_context_all}
percentage points. Every reported context-channel gap has a confidence interval
entirely above zero, supporting lower context coverage after
accounting for the estimated scene-sampling variation.

\paragraph{Audio-only extension.}
The same direction holds for Gemini~\numlit{3.1}~Pro on the
\num{keypoint_types.pop.ao.n_scenes} common audio-only demand scenes: it recovers
\numpct{keypoint_types.cov.gemini.ao.intent}\% of request points and
\numpct{keypoint_types.cov.gemini.ao.audio}\% of audio points.
Across the combined \num{keypoint_types.pop.all.n_scenes} positive audio--visual
and audio-only scenes, its point-weighted context coverage is
\numpct{keypoint_types.cov.gemini.all.context_all}\%.

% Keep the lexical-overlap comparison together.
\Needspace{210pt}
\paragraph{Textual clues and inference in the text-only baseline.}
We examine whether the text-only baseline's visual coverage is associated with
lexical clues in its input, extending \S\ref{sec:where_models_miss}. This input
includes preceding user transcripts and assistant replies
(App.~\ref{app:text_only_baseline}).

Lexical overlap is the fraction of a reference point's distinct content words present in the actual
ASR, dialogue, and template text. To compare with English reference points, we
restrict to Latin-script inputs, yielding
\numn{cascade_redundancy.redundant.cascade_asr.visual.ratio.latin} visual points.
Using saved outputs and judge verdicts, we compare coverage for points with at
least \numlit{20}\% overlap and those below this threshold.

Text-only coverage is
\numpct{cascade_redundancy.redundant.cascade_asr.visual.present.latin}\% in the
higher-overlap group and
\numpct{cascade_redundancy.redundant.cascade_asr.visual.absent.latin}\% in the
lower-overlap group. Seed~\numlit{2.0}~Lite, which receives audio and video, scores
\numpct{cascade_redundancy.redundant.seed.visual.present.latin}\% and
\numpct{cascade_redundancy.redundant.seed.visual.absent.latin}\%, respectively.
The contrast between overlap groups is much larger for the text-only baseline.
The text-only system's recovery of multimodal context may stem from textual
clues in transcripts and dialogue history, combined with commonsense knowledge.
\FloatBarrier

\subsection{LLM judge stability}
\label{app:judge_stability}

Because key-point coverage is assigned by an LLM judge, we audit two distinct properties: whether the scoring targets faithfully reflect the ground-truth annotation, and whether model ordering is stable across judges.

\paragraph{Design.} We re-judge a fixed population with Qwen3.6-Flash, DeepSeek-V4-Flash, and GPT-5.4. The population contains \num{judge_stability.pop.n_scenes} audio--visual demand scenes scored by all three evaluated systems. For each setting, all judges receive the same scoring instruction, candidate text, and reference points. Rates are point-weighted and use the all-judge common verdict set for each evaluated system; they are not the scene-averaged M2 values in Table~\ref{tab:main_results}.

% GENERATED by experiments/validity/judge_stability.py --stage analyze.
% DO NOT EDIT. Numbers come from results/judge_stability.json via \num{}.
\begin{table}[H]
\centering
\caption{LLM judge stability. Panel~A tests whether benchmark key points can be recovered from their source annotations. Panel~B re-judges identical stored outputs; all three judges yield the same model ordering.}
\label{tab:judge_stability}
\odutabsetupwide
\setlength{\tabcolsep}{2.2pt}
\begin{tabular}{@{}llcccc@{}}
\toprule
\textbf{Setting} & \textbf{Quantity} & \multicolumn{3}{c}{\textbf{Judge}} & \textbf{Stability readout} \\
\cmidrule(lr){3-5}\cmidrule(l){6-6}
 & & \textbf{Qwen3.6} & \textbf{DeepSeek} & \textbf{GPT-5.4} & \\
\midrule
\odugrouprow{6}{Panel A: target recoverability}
\midrule
A & Recoverability & \num{judge_stability.a.hit.cur.overall} & \num{judge_stability.a.hit.ds.overall} & \num{judge_stability.a.hit.gpt54.overall} & \numn{judge_stability.pop.n_points_a} key points \\
\rowcolor{odutabrow}
A & Orphan rate & \num{judge_stability.a.orphan.cur.overall} & \num{judge_stability.a.orphan.ds.overall} & \num{judge_stability.a.orphan.gpt54.overall} & ceiling loss \\
\midrule
\odugrouprow{6}{Panel B: stored-output re-judging}
\midrule
B & Gemini 3.1 Pro coverage & \num{judge_stability.b.cov.cur.gemini} & \num{judge_stability.b.cov.ds.gemini} & \num{judge_stability.b.cov.gpt54.gemini} & range \num{judge_stability.b.range.gemini} \\
\rowcolor{odutabrow}
B & Qwen3.5-Omni-Plus coverage & \num{judge_stability.b.cov.cur.qwen_plus} & \num{judge_stability.b.cov.ds.qwen_plus} & \num{judge_stability.b.cov.gpt54.qwen_plus} & range \num{judge_stability.b.range.qwen_plus} \\
B & Seed \numlit{2.0} Lite coverage & \num{judge_stability.b.cov.cur.seed} & \num{judge_stability.b.cov.ds.seed} & \num{judge_stability.b.cov.gpt54.seed} & range \num{judge_stability.b.range.seed} \\
\rowcolor{odutabrow}
B & Model ordering & \multicolumn{3}{c}{identical for \num{judge_stability.b.rank_preserved} of \numn{judge_stability.b.rank_preserved} judges} & same conclusion \\
\bottomrule
\end{tabular}
\odutabnote{Rates are point-weighted proportions on the 0--1 scale. Orphan rate is one minus recoverability; range is the maximum minus minimum coverage across judges.}
\end{table}

\paragraph{Ground-truth authority and key-point recoverability.} The released ground-truth annotation is the benchmark's authoritative scoring reference: it is reconstructed from the realized media and admitted only after grounding checks and human verification (\S\ref{sec:annotation_qc}). Because the key points operationalize its structured intent and required context, they should be recoverable directly from those source fields. We test this by presenting each judge with the ground-truth annotation text from which the corresponding key points were derived. Across \numn{judge_stability.pop.n_points_a} key points, Qwen3.6-Flash, DeepSeek-V4-Flash, and GPT-5.4 achieve recoverability rates of \num{judge_stability.a.hit.cur.overall}, \num{judge_stability.a.hit.ds.overall}, and \num{judge_stability.a.hit.gpt54.overall}, respectively; their orphan rates are only \num{judge_stability.a.orphan.cur.overall}, \num{judge_stability.a.orphan.ds.overall}, and \num{judge_stability.a.orphan.gpt54.overall}. This establishes high recoverability from the source annotation across the tested judges.

\Needspace{80pt}
\paragraph{Model-ordering stability.} We next re-score the exact stored per-segment outputs of Gemini~3.1~Pro, Qwen3.5-Omni-Plus, and Seed~\numlit{2.0}~Lite. No model inference or segment matching is repeated. The audit retains points
with tier labels and available candidate text. 
Although judges differ slightly in absolute calibration, all three produce the same model ordering, \num{judge_stability.b.rank_order}, on the retained point sets. 

\FloatBarrier

\FloatBarrier
\subsection{Positive-demand challenge breakdown}
\label{app:positive_taxonomy_m2}

We examine which types of valid demand are associated with lower M2 key-point
coverage and whether these patterns recur across systems and modalities. This
comparison groups whole scenes by challenge category, complementing the
key-point-level analysis in App.~\ref{app:keypoint_detail}.

\paragraph{Design.}
We reanalyze saved scores from \num{positive_taxonomy_m2.av.n_models} systems on
audio--visual (AV) scenes and \num{positive_taxonomy_m2.ao.n_models} systems on
audio-only (AO) scenes. Within each modality, we retain the same positive scenes
with complete scores for all displayed systems. The four challenge axes in
App.~\ref{app:taxonomy_positive} are modality dependency, contextual
disambiguation, intent expression form, and acoustic environment. The modalities
are analyzed separately because Axis~\numlit{1} uses modality-specific categories.

Scoring follows Table~\ref{tab:main_results}, with missed reference points counted
as misses. Each cell in Fig.~\ref{fig:positive_taxonomy_m2_av}
and Fig.~\ref{fig:positive_taxonomy_m2_ao} subtracts the system's mean M2 over all common
scenes from its mean within a category, weighting scenes equally. Differences
are in percentage points: red means below the system's own average and blue
means above it. The \emph{Panel mean} row
averages the differences equally across all displayed systems, including the
text-only comparator, which is shown separately and remains unranked.

\paragraph{Intent expression.}
Indirect or descriptive requests (Axis~\numlit{3.3}) fall below every system's
mean in AV and AO, with panel-average M2 differences of
\numpct{positive_taxonomy_m2.av.panel_delta.axis3.3_3} points (AV) and
\numpct{positive_taxonomy_m2.ao.panel_delta.axis3.3_3} points (AO).
Interrupted or incomplete requests (Axis~\numlit{3.5}) also have negative
panel-average M2 differences
(\numpct{positive_taxonomy_m2.av.panel_delta.axis3.3_5} AV;
\numpct{positive_taxonomy_m2.ao.panel_delta.axis3.3_5} AO).
Compound or multi-intent requests (Axis~\numlit{3.6}) have positive panel-average M2 differences
of \numpctd{positive_taxonomy_m2.av.panel_delta.axis3.3_6} and
\numpctd{positive_taxonomy_m2.ao.panel_delta.axis3.3_6} points, respectively;
the difference is positive for every AO system but varies in sign across AV systems.
The shared indirect-request deficit and negative mean differences for incomplete requests point
to recovering unstated intent as a recurring difficulty. Positive mean differences
for multi-intent requests show that several demands need not imply lower coverage; taxonomy
categories do not define an ordered difficulty scale.

\paragraph{Contextual disambiguation.}
On AO scenes, speaker-identity and affective-state dependencies have
panel-average M2 differences of
\numpct{positive_taxonomy_m2.ao.panel_delta.axis2.2_4} and
\numpct{positive_taxonomy_m2.ao.panel_delta.axis2.2_5} points, respectively.
For scenes that depend on dialogue history, the difference between category M2
and each system's overall mean averages
\numpctd{positive_taxonomy_m2.av.panel_delta.axis2.2_3} percentage points in AV and
\numpctd{positive_taxonomy_m2.ao.panel_delta.axis2.2_3} in AO.
The higher scores may stem from explicit clues in earlier turns, such as the
user's stated goal or previously discussed options, that help clarify the
current request.

\paragraph{Acoustic conditions.}
Acoustic categories show less consistent patterns across modalities.
Background-noise scenes, for example, have a panel-average M2 difference of
\numpctd{positive_taxonomy_m2.av.panel_delta.axis4.4_2} points in AV and
\numpctd{positive_taxonomy_m2.ao.panel_delta.axis4.4_2} points in AO.
The indirect-request deficit is thus more consistent across the tested systems
and modalities than the acoustic patterns.

% Stack the two independently numbered figures at a readable shared scale.
\begin{figure}[p]
\centering
\input{figures/positive_taxonomy_m2_av}
\par\medskip
\input{figures/positive_taxonomy_m2_ao}
\end{figure}
\FloatBarrier

\subsection{Blinded response comparison}
\label{app:human_pilot}

We assess whether providing demand annotations improves response content and
the decision to respond.

\paragraph{Scene sampling.}
We sampled \num{response_ab_setup.n_scenes} audio--visual scenes from the human-verified
benchmark. The sample comprises \num{response_ab_setup.n_positive} demand scenes and
\num{response_ab_setup.n_negative} no-demand scenes, with Chinese and English equally
represented within each group.

\paragraph{Paired response generation.}
Gemini~\numlit{3.1}~Pro generated a reply under each of two conditions:
``Not given demand annotation'' (\emph{Not given}) and
``Given demand annotation'' (\emph{Given}).
Both received identical video and audio, in clip order with intervening assistant
replies, and responded to the final user turn. The \emph{Given} condition additionally
received the
reference structured intent and required-context descriptions joined by \texttt{|}, or
``Demand: NONE'' for no-demand scenes; neither key points nor reference
answers were supplied. Settings were fixed: temperature
\num{response_ab_setup.temperature}, top-$p$ \num{response_ab_setup.topP}, an output-token
limit of \num{response_ab_setup.maxOutputTokens}, and
\num{response_ab_setup.candidateCount} candidate per request.

\paragraph{Blinded assessment.}
The interface (Fig.~\ref{fig:response_ab_review}) shows each interaction
with replies A and B. Scene order and reply placement were randomized.
Human assessment considered correctness, relevance,
contextual grounding, and appropriate silence. Ties covered equally good or poor
replies; instructions discouraged preference based on length alone.

\paragraph{Preference results.}
Table~\ref{tab:response_ab} reports all \num{response_ab_blind.all.n} judgments.
Of these, \num{response_ab_blind.all.ties} (\numpct{response_ab_blind.all.tie_rate}\%) were ties.
Among the remaining \num{response_ab_blind.all.decisive_n} comparisons,
\num{response_ab_blind.all.with_demand_wins}
(\numpct{response_ab_blind.all.given_win_rate_among_decisive}\%) favored \emph{Given}.
The preference advantage appears in both languages and in both demand-presence groups.

\begin{table}[H]
\centering
\small
\setlength{\tabcolsep}{7pt}
\caption{Human response preferences: counts (percentages) within each subset.
``Given'' and ``Not given'' abbreviate the ``Given demand annotation'' and
``Not given demand annotation'' conditions. Ties remain in the denominator.}
\label{tab:response_ab}
\begin{tabular}{lrrrr}
\toprule
Subset & Scenes & Given preferred & Tie & Not given preferred \\
\midrule
All & \num{response_ab_blind.all.n}
& \num{response_ab_blind.all.with_demand_wins} (\numpct{response_ab_blind.all.with_demand_win_rate}\%)
& \num{response_ab_blind.all.ties} (\numpct{response_ab_blind.all.tie_rate}\%)
& \num{response_ab_blind.all.without_demand_wins} (\numpct{response_ab_blind.all.without_demand_win_rate}\%) \\
Demand present & \num{response_ab_blind.positive.n}
& \num{response_ab_blind.positive.with_demand_wins} (\numpct{response_ab_blind.positive.with_demand_win_rate}\%)
& \num{response_ab_blind.positive.ties} (\numpct{response_ab_blind.positive.tie_rate}\%)
& \num{response_ab_blind.positive.without_demand_wins} (\numpct{response_ab_blind.positive.without_demand_win_rate}\%) \\
No-demand & \num{response_ab_blind.negative.n}
& \num{response_ab_blind.negative.with_demand_wins} (\numpct{response_ab_blind.negative.with_demand_win_rate}\%)
& \num{response_ab_blind.negative.ties} (\numpct{response_ab_blind.negative.tie_rate}\%)
& \num{response_ab_blind.negative.without_demand_wins} (\numpct{response_ab_blind.negative.without_demand_win_rate}\%) \\
\midrule
Chinese & \num{response_ab_blind.zh.n}
& \num{response_ab_blind.zh.with_demand_wins} (\numpct{response_ab_blind.zh.with_demand_win_rate}\%)
& \num{response_ab_blind.zh.ties} (\numpct{response_ab_blind.zh.tie_rate}\%)
& \num{response_ab_blind.zh.without_demand_wins} (\numpct{response_ab_blind.zh.without_demand_win_rate}\%) \\
English & \num{response_ab_blind.en.n}
& \num{response_ab_blind.en.with_demand_wins} (\numpct{response_ab_blind.en.with_demand_win_rate}\%)
& \num{response_ab_blind.en.ties} (\numpct{response_ab_blind.en.tie_rate}\%)
& \num{response_ab_blind.en.without_demand_wins} (\numpct{response_ab_blind.en.without_demand_win_rate}\%) \\
\bottomrule
\end{tabular}
\end{table}

\paragraph{Response content and whether to respond.}
Among
\num{response_ab_blind.positive_both_text.n} demand scenes where both conditions produced text,
\emph{Given} was preferred in \num{response_ab_blind.positive_both_text.with_demand_wins} cases versus
\num{response_ab_blind.positive_both_text.without_demand_wins} for \emph{Not given}, with
\num{response_ab_blind.positive_both_text.ties} ties. On no-demand scenes, the proportion of
textual replies fell from \numpct{response_ab_blind.negative.without_demand_reply_rate}\% to
\numpct{response_ab_blind.negative.with_demand_reply_rate}\%. On demand scenes, silent replies
fell from \num{response_ab_blind.positive.without_demand_silent} to
\num{response_ab_blind.positive.with_demand_silent}. These findings support the utility of explicit
demand and context for response content and the decision to respond.

\paragraph{Input prompts.}
The panels give the shared instruction, templates, and complete examples.
Video placeholders denote media attachments; an exact \texttt{[SILENCE]} output
is displayed as no reply.

\begingroup
% Local prompt-panel styles; scoped by setup.tex, with no manuscript preamble edit.
\definecolor{abPromptInk}{RGB}{48,69,92}
\definecolor{abPromptAccent}{RGB}{63,99,136}
\definecolor{abPromptRule}{RGB}{211,223,234}
\definecolor{abPromptFill}{RGB}{248,250,253}
\definecolor{abPromptTitle}{RGB}{234,240,247}
\newtcolorbox{abprompt}[1]{%
  enhanced, breakable,
  colback=abPromptFill,
  colframe=abPromptRule,
  colbacktitle=abPromptTitle,
  coltitle=abPromptInk,
  title={#1},
  title after break={#1\ (continued)},
  fonttitle=\small\bfseries,
  fontupper=\small,
  boxrule=0.4pt,
  titlerule=0pt,
  borderline west={1.6pt}{0pt}{abPromptAccent},
  arc=1pt,
  left=3mm,right=3mm,top=2mm,bottom=2mm,
  before skip=7pt,after skip=7pt,
  before upper={\setlength{\parindent}{0pt}\setlength{\parskip}{3pt}},
}
\newcommand{\ablabel}[1]{{\sffamily\footnotesize\bfseries\color{abPromptAccent}#1}}
\newcommand{\abmedia}{{\color{abPromptInk!65}\itshape [video input]}}
\newcommand{\abdemand}{{\color{abPromptAccent}\bfseries Demand:}}

% Generated from frozen study requests by build_prompt_panels.py; do not hand-edit.

\begin{abprompt}{Shared system instruction}

Reply directly to the user's request in the last clip, using the user's language. If a demand annotation is provided, follow it; NONE means no request. If there is no request for you, output only [SILENCE].

\end{abprompt}

\begin{abprompt}{User-input templates}
\begin{tabularx}{\linewidth}{@{}>{\raggedright\arraybackslash}p{0.25\linewidth}@{\hspace{4mm}}>{\raggedright\arraybackslash}X@{}}
\ablabel{Not given demand annotation} & \ablabel{Given demand annotation} \\[4pt]
Clip 1 & Clip 1 \\
\abmedia & \abmedia \\[4pt]
& \abdemand\ \textit{\{structured\_intent\}}\enspace\textbar{}\enspace\textit{\{required\_context.description\}}\enspace\textbar{}\enspace\ldots
\end{tabularx}
\end{abprompt}

\begin{abprompt}{Positive example (Demand present)}
\begin{tabularx}{\linewidth}{@{}>{\raggedright\arraybackslash}p{0.25\linewidth}@{\hspace{4mm}}>{\raggedright\arraybackslash}X@{}}
\ablabel{Not given demand annotation} & \ablabel{Given demand annotation} \\[4pt]
Clip 1 & Clip 1 \\
\abmedia & \abmedia \\[4pt]
& \abdemand\ The user wants the device that is continuously emitting a beeping sound to stop its alarm. \textcolor{abPromptAccent}{\textbar{}} The continuous electronic beeping sound in the background. \textcolor{abPromptAccent}{\textbar{}} The device with a blinking red indicator lamp on the mantel, which is the source of the alarm sound referred to by the user's ``That beeping''. \textcolor{abPromptAccent}{\textbar{}} The user's tone is fatigued and irritable, and he startles when the vase shatters.
\end{tabularx}
\end{abprompt}

\begin{abprompt}{Negative example (No-demand)}
\begin{tabularx}{\linewidth}{@{}>{\raggedright\arraybackslash}p{0.25\linewidth}@{\hspace{4mm}}>{\raggedright\arraybackslash}X@{}}
\ablabel{Not given demand annotation} & \ablabel{Given demand annotation} \\[4pt]
Clip 1 & Clip 1 \\
\abmedia & \abmedia \\[4pt]
& \abdemand\ NONE
\end{tabularx}
\end{abprompt}

\endgroup

\begin{figure}[!htbp]
  \centering
  \includegraphics[width=0.78\linewidth,height=0.78\textheight,keepaspectratio]{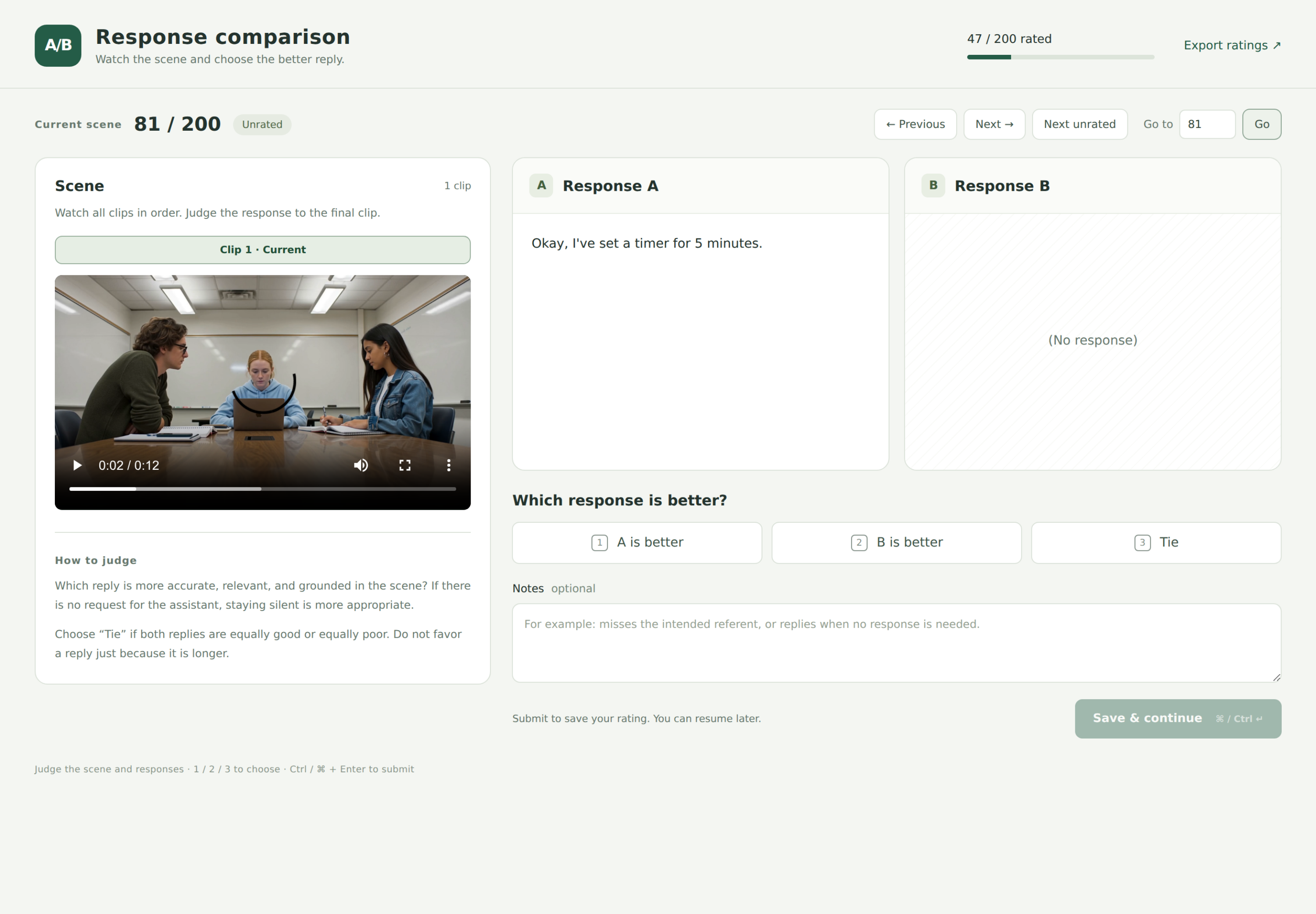}
  \caption{Blinded response-comparison interface. The human reviewer is asked to choose a preferred reply between A and B.}
  \label{fig:response_ab_review}
\end{figure}

\clearpage
\subsection{Qualitative error analysis}
\label{app:error_analysis}

% Author-approved cards: positives first, then negatives; two cases per page.
\begingroup
\setlength{\intextsep}{0pt}
\captionsetup{skip=5pt}
\begin{figure}[H]
  \centering
  \includegraphics[page=1,width=\linewidth,height=0.89\textheight,keepaspectratio]{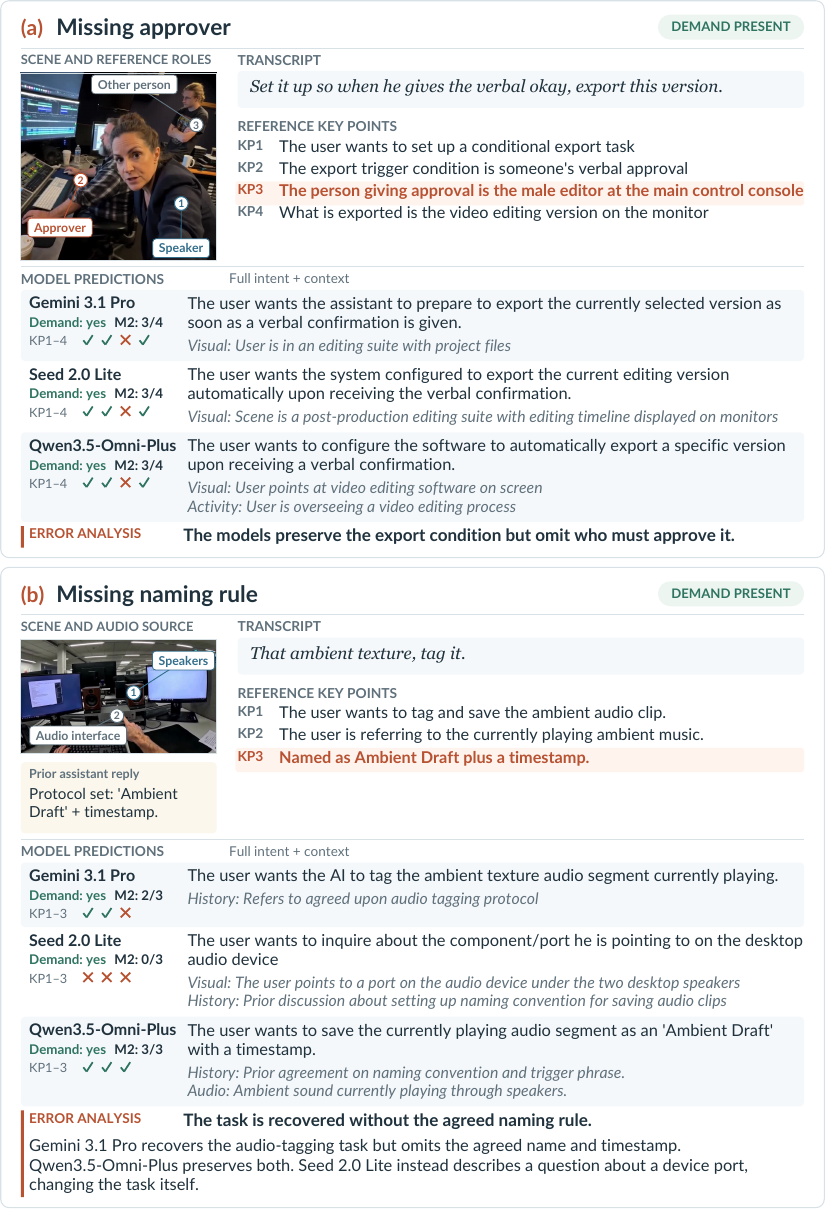}
  \caption{Qualitative error analysis on English generated audio--visual scenes.
  (a)--(b) Demand-present cases; M2 reports covered/reference key points.}
  \label{fig:error_analysis}
\end{figure}

\clearpage
\begin{figure}[H]
  \ContinuedFloat
  \centering
  \includegraphics[page=2,width=\linewidth,height=0.94\textheight,keepaspectratio]{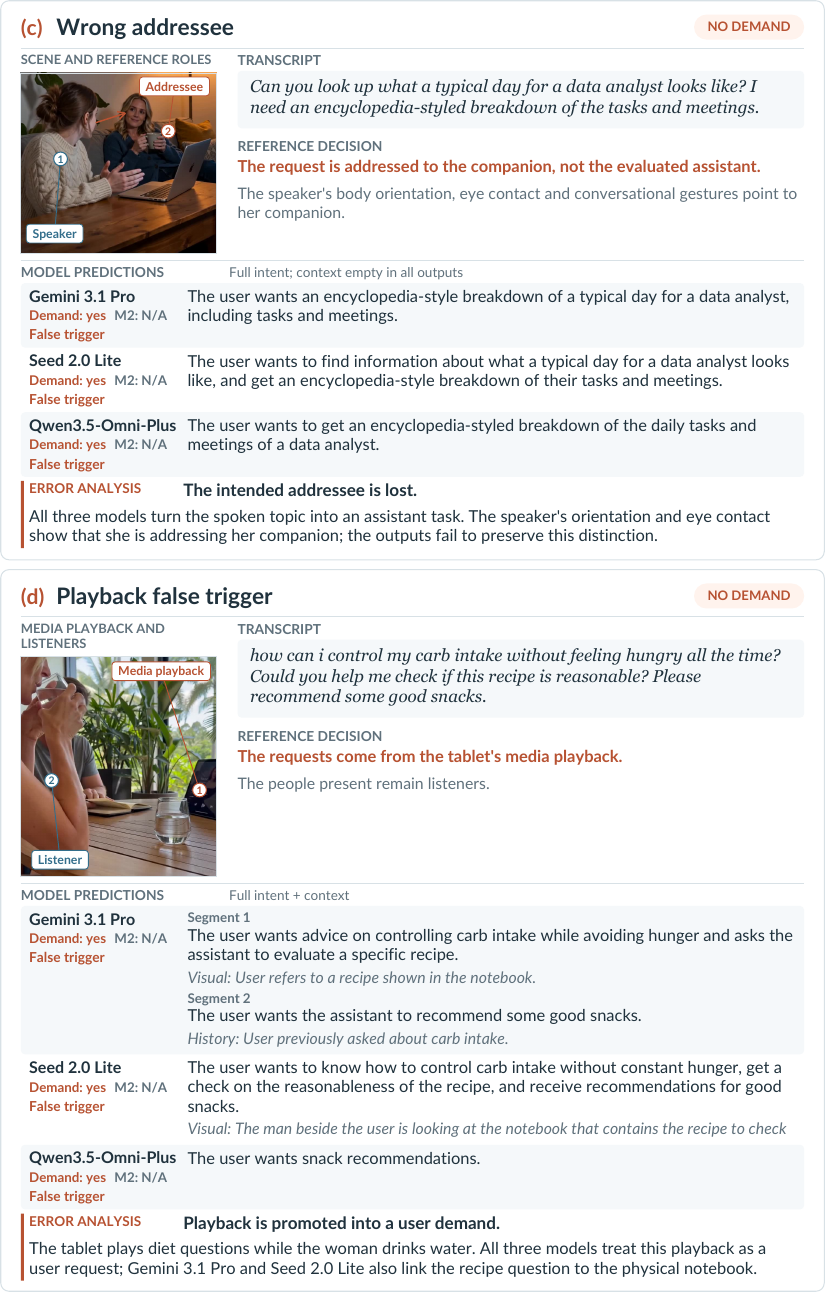}
  \caption{Qualitative error analysis (continued).
  (c)--(d) No-demand cases.}
  \label{fig:error_analysis_negative}
\end{figure}
\endgroup
\clearpage

\section{Evaluation prompts}
\label{app:prompt_specs}

This section specifies what evaluated models are asked to predict and how their
semantic outputs are judged. Scenario-generation and annotation-construction
prompts, including key-point extraction and grounding, are not distributed;
their procedures are described in App.~\ref{app:generation}. The audio--visual
prediction instruction is reproduced in full, followed by the audio-only
differences and a pointer to the text-only input templates. Executable AV and AO variants are
provided with the evaluation code. The judge instruction below is its English
rendering; the scoring implementation retains the original instruction.

\subsection{Shared prediction task}
\label{app:eval_prompts}
In the reproduced instruction, ``the last segment'' refers to the final input clip; output segments are demand-bearing regions within that clip.
% Generated by scripts/export_paper_prompts.py -- do not edit by hand.
% Regenerate with: python scripts/export_paper_prompts.py
% The promptbox environment, the promptmeta/promptrow metadata layout, the
% \PromptSection / \PromptSubsection / \PromptInclude / \PromptVar / \PromptKey
% macros and the oduprompt listing style are all defined in preamble.tex.

\begin{promptbox}{Audio-Visual Prediction Instruction}
\begin{promptmeta}
\promptrow{Purpose}{The system prompt every model receives on an audio-visual scene. It asks the model to decide whether the final user turn carries a demand and, when it does, to localise the demand in time, transcribe it, state a self-contained intent, and list the multimodal context the demand depends on.}
\promptrow{Inputs}{One video clip per user turn, in order, with the assistant replies that separate them supplied as text.}
\promptrow{Output}{One JSON object with \texttt{has\_demand} and a \texttt{segments} list. Each segment carries \texttt{start\_ms}, \texttt{end\_ms}, \texttt{transcript}, \texttt{structured\_intent}, \texttt{required\_context} and the user attribute fields.}
\promptrow{Variant}{Complete prompt; audio-visual variant.}
\end{promptmeta}
You are the user's AI assistant. You will watch one or more videos, in which the user may make requests to you.
\PromptSection{Task}
Determine whether a user has made a request to you in the video. The scenario may be a user interacting with you or ordinary interpersonal conversation -- you must distinguish between them based on the visual and audio content.
If the user expresses a demand, output the following information in structured JSON format (instead of replying directly to the user). A demand is not necessarily an explicit question or instruction; it may also be an implicit expression (e.g., complaining, talking to oneself, hinting for help, etc.).
\begin{promptitem}
\item \textbf{Time Localization}: The start and end times when the demand occurs (milliseconds)
\item \textbf{\PromptKey{transcript}}: The exact words spoken by the user within this segment's time interval (verbatim transcription, no additions or omissions; does not include utterances from other speakers, nor content from previous or subsequent turns)
\item \textbf{Structured Intent}: A single sentence summarizing the user's goal, starting with ``The user wants...'', semantically complete and self-contained
\item \textbf{Required Context}: The multimodal context information necessary to understand the content of this demand
\item \textbf{Is the user in the frame?}
\item \textbf{User gender and age group}
\end{promptitem}

\PromptSection{Required Context (\PromptKey{required\_context})}

Fill in context items where ``the absence would lead to an incomplete understanding of the demand, a deviation in the response direction, or a significant decrease in response quality''. If the user's utterance text alone is sufficient to accurately and appropriately understand and respond, then \PromptKey{required\_context = []}.
Select \PromptKey{type} from the following 7 options:
\par\addvspace{4pt}
\begin{tabularx}{\linewidth}{@{}lX@{}}
\toprule
type & Meaning \\
\midrule
\PromptKey{visual} & Objects/text/actions/scenes in the visual frame are necessary information for understanding the demand \\
\PromptKey{audio} & Ambient sounds/background audio provide clues for understanding the object or context of the demand \\
\PromptKey{dialogue\_history} & The current utterance contains references or omissions, requiring previous dialogue content for comprehension \\
\PromptKey{user\_activity} & The user's ongoing activity provides necessary context for understanding the demand \\
\PromptKey{user\_identity} & The user's identity/role/professional background affects the understanding of the demand \\
\PromptKey{user\_emotion} & The user's emotion/tone affects the judgment of intent direction (e.g., a dissatisfied tone turns an observation into a request) \\
\PromptKey{scene\_condition} & Environmental conditions (time/weather/noise, etc.) affect the understanding of the demand \\
\bottomrule
\end{tabularx}
\par\addvspace{4pt}
Fill in \PromptKey{description} for each entry (\textless{}= 50 characters). Each type must appear at most once.

\PromptSection{Multi-turn Input}
The input may contain multiple video clips (\PromptKey{v1}, \PromptKey{v2}, \PromptKey{v3}...), each corresponding to a user turn, with possible text replies from the AI agent between clips.
\begin{promptitem}
\item \textbf{Annotate user demands only in the last segment}. Content from previous turns and agent replies serve only as dialogue history context; do not output them as segments.
\item \PromptKey{has\_demand} is \PromptKey{true} only when a user demand exists in the last segment.
\item \PromptKey{start\_ms}/\PromptKey{end\_ms} are relative to the start of the last segment, not global time.
\end{promptitem}

\PromptSection{Segment Division}

One segment corresponds to a continuous region of demand expression. Boundaries are determined solely by the following two signals:
\begin{promptitem}
\item \textbf{Must split} (highest priority): (1) Turn boundaries (user finishes speaking $\rightarrow$ agent responds $\rightarrow$ user speaks again); (2) Speech interruptions (\textgreater{}= 2 seconds of silence within the same turn, or insertion of speech unrelated to the demand, such as talking to bystanders).
\item \textbf{Do not split} (only within the same turn and the same continuous utterance): Consecutive multiple intents, self-correction, preamble + request, emotion + action, etc., must all be kept as a single segment.
\item Express \PromptKey{structured\_intent} for compound demands using a compound sentence (e.g., ``the user wants to know the weather and set an alarm''); for self-correction, record only the final valid demand.
\item \textbf{transcript boundaries}: \PromptKey{transcript} must strictly contain only the speech transcription of \emph{the user} within the \PromptKey{start\_ms} to \PromptKey{end\_ms} time window. Do not include utterances from other speakers, AI replies, or content from previous or subsequent turns in the transcript. If preceding information is relevant to demand comprehension, place it in \PromptKey{required\_context} (type=\PromptKey{dialogue\_history}) instead of the transcript.
\end{promptitem}

\PromptSection{Field Constraints}

The following fields are closed-set classifications. Use exactly one of the given values; do not invent synonyms (for example, do not write \PromptKey{"young adult"}, \PromptKey{"senior"}, \PromptKey{"middle-aged"} or \PromptKey{"teenager"}).

\begin{promptitem}
\item \PromptKey{user\_in\_frame}: boolean. Whether the user initiating the demand is within the frame.
\begin{promptitem}
\item \PromptKey{true}: the user initiating the demand is visible in the scene.
\item \PromptKey{false}: first-person off-screen voice, only the user's voice is heard, or it is impossible to determine who in the scene initiated the demand.
\end{promptitem}
\item \PromptKey{user\_gender}: string, must be one of the following values:
\begin{promptitem}
\item \PromptKey{"male"}: there is clear evidence of male or masculine presentation.
\item \PromptKey{"female"}: there is clear evidence of female or feminine presentation.
\item \PromptKey{"unknown"}: evidence is insufficient, evidence from multiple people/speakers is mixed together or voice/visual evidence cannot support a judgment, or no inference should be made.
\end{promptitem}
\item \PromptKey{user\_age\_group}: string, must be one of the following values:
\begin{promptitem}
\item \PromptKey{"child"}: child or clearly underage person.
\item \PromptKey{"adult"}: adult.
\item \PromptKey{"elderly"}: elderly person, with clear age features.
\item \PromptKey{"unknown"}: evidence is insufficient or judgment is impossible.
\end{promptitem}
\end{promptitem}

\PromptSection{Output Format}

Output only JSON, with no comments or code block markers. The timestamp values in the examples below are placeholders only; infer the real \PromptKey{start\_ms}/\PromptKey{end\_ms} from the provided final clip and never copy example timestamps:
\begin{lstlisting}[style=oduprompt]
{
  "has_demand": true,
  "segments": [
    {
      "start_ms": 11840,
      "end_ms": 15200,
      "transcript": "User's exact words",
      "structured_intent": "The user wants to...",
      "required_context": [
        {
          "type": "visual",
          "description": "User points to a bottle of olive oil"
        }
      ],
      "user_in_frame": false,
      "user_gender": "male",
      "user_age_group": "adult"
    }
  ]
}
If no demand: {"has_demand": false, "segments": []}
\end{lstlisting}
\end{promptbox}

\Needspace{180pt}
\subsection{Audio-only variant}
\label{app:eval_prompts_audio_only}
The audio-only instruction preserves the final-turn scope, temporal segmentation,
transcript boundaries, self-contained intent, and JSON-only output requirements.
It changes the input wording from video to audio and uses the following
modality-specific fields and evidence rules.

\begin{center}\small
\begin{tabularx}{\linewidth}{@{}lX@{}}
\toprule
Component & Audio-only specification \\
\midrule
Demand evidence & Linguistic content and audio cues distinguish assistant-directed demands from other conversation. \\
Context types & Audio, dialogue history, user activity, user identity, user emotion, and scene condition; visual context is unavailable. \\
User profile fields & Output gender and age group; omit \texttt{user\_in\_frame}. \\
\bottomrule
\end{tabularx}
\end{center}

\subsection{Text-only input structure}
\label{app:text_input}
The transcript baseline uses the shared prediction instruction above, with each
media block replaced in place by ASR text. App.~\ref{app:text_only_baseline}
provides the full input templates, including dialogue-history framing and the
no-speech replacement, together with complete positive and no-demand input
examples and model settings. No reference transcript or demand label is supplied
as input.

\subsection{Key-point judge}
\label{app:judge_prompt}
Each reference point is supplied with its identifier, tier label, and point text.
\begingroup
% Compact list spacing keeps the judge's final notes with its specification.
\renewcommand{\oduPromptLineSpread}{1.0}
\renewcommand{\oduPromptParskip}{2pt}
\setlist*[promptitem]{itemsep=0pt,topsep=0pt}
\setlist*[promptmeta]{itemsep=0pt,topsep=0pt}
% Generated by scripts/export_paper_prompts.py -- do not edit by hand.
% Regenerate with: python scripts/export_paper_prompts.py
% The promptbox environment, the promptmeta/promptrow metadata layout, the
% \PromptSection / \PromptSubsection / \PromptInclude / \PromptVar / \PromptKey
% macros and the oduprompt listing style are all defined in preamble.tex.

\begin{promptbox}{Key-Point Judge}
\begin{promptmeta}
\promptrow{Purpose}{Scores key-point coverage. For each key point the judge decides whether the predicted demand text covers it, under a lenient criterion that accepts paraphrase and hypernym coverage but still rejects a referent the prediction never names.}
\promptrow{Inputs}{The predicted demand text for one segment, and that segment's key points.}
\promptrow{Output}{A JSON array with one entry per key point, carrying \texttt{qid}, \texttt{hit} and a short \texttt{reason}.}
\end{promptmeta}
You are an evaluation expert. Given a passage of model output text and a set of key assessment points, judge for each assessment point whether the model output ``hits'' it.

Hit criterion (lenient):
\begin{promptitem}
\item ``hit'' = true: the model output semantically covers the core of that assessment point, or that core can be reasonably inferred from / obtained as an equivalent restatement of the output -- even if the wording differs, the phrasing is more general, or only the main point is covered while incidental modifiers are omitted.
\item ``hit'' = false: only when the model output does not touch the core information of that assessment point at all, or substantively contradicts / gets it wrong.
\end{promptitem}

Judgment principles:
\begin{promptitem}
\item Each assessment point is independent; judge them one by one, looking only at whether that point's own core is covered.
\item Hypernym / generalized hits are allowed: if the output covers the point with a more general statement (for example, ``complaining about mold'' covers ``the mold is unusual this year''), it counts as a hit.
\item When in doubt, be lenient: if the intent / context semantically covers the point on the whole, judge hit; judge miss only when the core content really is missing or is stated incorrectly.
\item But do not invent what is absent: a distinctive referred-to object / action / attribute that is simply not in the output and cannot be reasonably inferred from it (for example, a specific ``one particular bag'' or ``the region indicated by `this one'{}'' is never named) is still judged miss.
\end{promptitem}

Output a JSON array in which each element has the following format:
\begin{lstlisting}[style=oduprompt]
{"qid": "kp01", "hit": true, "reason": "a brief justification (<=15 characters)"}
\end{lstlisting}

Notes:
\begin{promptitem}
\item You must give a judgment for every assessment point
\item reason must be concise and state the key basis for the hit/miss
\item Do not use double quotes inside reason; use single quotes or no quotes
\item Output only the JSON array, with no other text
\end{promptitem}
\end{promptbox}

\endgroup

\end{document}